\documentclass[examplefnt,biber]{nowfnt-arxiv} 

\usepackage{etoolbox}

\AfterEndPreamble{\input{fnfix}}

\PassOptionsToPackage{dvipsnames}{xcolor} 

\usepackage{tikz}
\usetikzlibrary{arrows.meta,calc,positioning,shapes.geometric}
\usepackage{subcaption}
\usepackage{mathrsfs}
\usepackage{tcolorbox}
\usepackage{booktabs} 
\usetikzlibrary{positioning,calc,arrows.meta,shapes.geometric}

\usepackage[labelfont={bf,sf}]{caption}
\usetikzlibrary{shapes, arrows, positioning, backgrounds, shadows} 
\usepackage{stmaryrd}
\usepackage{placeins} 
\newcommand{\graymidrule}{\arrayrulecolor{black!25}\midrule\arrayrulecolor{black}}
\usepackage{cancel}
\usepackage{tabularx}
\usepackage{array}

\usepackage{mathtools}
\usepackage{fancybox}
\usepackage[dvipsnames,table]{xcolor}
\usepackage{amsmath, amssymb, amsfonts}
\usepackage{graphicx}
\usepackage{setspace}
\usepackage{geometry}
\usepackage{float}
\usepackage[utf8]{inputenc}
\usepackage[english]{babel}
\usepackage{environ}
\usepackage{xr}
\usepackage{mwe}
\usepackage{adjustbox} 
\usepackage{multirow}       
\usepackage{makecell}
\usepackage{tikz}
\usetikzlibrary{arrows.meta, positioning}

\usepackage{annotate_eqs}
\usetikzlibrary{positioning, arrows.meta, fit, shapes.multipart}

\definecolor{StoryBrown}{HTML}{9A5B32}
\definecolor{StoryBrownDark}{HTML}{704023}
\definecolor{StoryBrownLight}{HTML}{FBF6F0}

\newtcolorbox{storychapterguidebox}{
  enhanced,
  breakable,
  colback=StoryBrownLight,
  colframe=StoryBrownLight,
  boxrule=0pt,
  borderline west={2.2pt}{0pt}{StoryBrown},
  arc=1.5mm,
  left=3.5mm,
  right=3mm,
  top=2.5mm,
  bottom=2.5mm,
}

\NewDocumentCommand{\chapterlearning}{+m +m}{%
  \begin{storychapterguidebox}
  \noindent\textcolor{StoryBrownDark}{%
    \textbfs{After this chapter, you will be able to...}%
  }
  \begin{itemize}[leftmargin=1.5em,itemsep=2pt,topsep=4pt,parsep=0pt]
  #1
  \end{itemize}
  \vspace{-1mm}
  \noindent\textcolor{StoryBrownDark}{\textbfs{Reader route.}} #2
  \end{storychapterguidebox}
}

\newtheorem*{lemma*}{Lemma}

\usepackage{epigraph} 
\usepackage{thmbox}  
\newtheorem[M]{question}{\sffamily\bfseries Question}[section]
\tcbuselibrary{theorems,skins,breakable}

\newcommand{\cc}[1]{\cellcolor{gray!#1}}
\newcommand{\textbfs}[1]{{\sffamily\textbf{#1}}}
\usepackage{multicol}

\usepackage{algorithm}
\usepackage{algpseudocode}
\floatname{algorithm}{\textbfs{Algorithm}}
\DeclareCaptionFormat{bfsformat}{\textbfs{#1#2}#3}
\usepackage{cleveref}
\crefname{algorithm}{Algorithm}{Algorithms}

\usepackage{subcaption}
\usepackage{graphicx}

\usepackage[framemethod=TikZ]{mdframed}
\mdfsetup{%
   middlelinecolor=black,
   middlelinewidth=1.1pt,
   backgroundcolor=gray!10,
   roundcorner=6pt}

\usepackage{amsmath,amsfonts,bm} 
\usepackage{mathrsfs}
\usepackage{enumitem}
\usepackage{stmaryrd}

\def\eqref#1{equation~\ref{#1}}

\def\1{\bm{1}}

\def\btheta{{\bm{\theta}}}
\def\bphi{{\bm{\phi}}}
\def\beps{{\bm{\epsilon}}}

\def\rva{{\mathbf{a}}}
\def\rvb{{\mathbf{b}}}
\def\rvc{{\mathbf{c}}}

\def\rve{{\mathbf{e}}}
\def\rvf{{\mathbf{f}}}
\def\rvg{{\mathbf{g}}}
\def\rvh{{\mathbf{h}}}
\def\rvu{{\mathbf{i}}}

\def\rvm{{\mathbf{m}}}

\def\rvp{{\mathbf{p}}}

\def\rvs{{\mathbf{s}}}

\def\rvu{{\mathbf{u}}}
\def\rvv{{\mathbf{v}}}
\def\rvw{{\mathbf{w}}}
\def\rvx{{\mathbf{x}}}
\def\rvy{{\mathbf{y}}}
\def\rvz{{\mathbf{z}}}

\def\rmA{{\mathbf{A}}}

\def\rmD{{\mathbf{D}}}
\def\rmE{{\mathbf{E}}}
\def\rmF{{\mathbf{F}}}
\def\rmG{{\mathbf{G}}}
\def\rmH{{\mathbf{H}}}
\def\rmI{{\mathbf{I}}}

\def\rmN{{\mathbf{N}}}

\def\rmP{{\mathbf{P}}}
\def\rmQ{{\mathbf{Q}}}

\def\rmS{{\mathbf{S}}}
\def\rmT{{\mathbf{T}}}

\def\rmY{{\mathbf{Y}}}
\def\rmZ{{\mathbf{Z}}}

\def\vmu{{\bm{\mu}}}

\def\mSigma{{\bm{\Sigma}}}

\DeclareMathAlphabet{\mathsfit}{\encodingdefault}{\sfdefault}{m}{sl}
\SetMathAlphabet{\mathsfit}{bold}{\encodingdefault}{\sfdefault}{bx}{n}

\newcommand{\bfI}{\mathbf{I}}
\newcommand{\E}{\mathbb{E}}

\newcommand{\R}{\mathbb{R}}

\newcommand{\Var}{\mathrm{Var}}

\newcommand{\Cov}{\mathrm{Cov}}
\newcommand*\diff{\mathop{}\!\mathrm{d}}

\newcommand{\norm}[1]{\left\lVert#1\right\rVert}
\newcommand{\abs}[1]{\left|#1\right|}

\usepackage{tikz}

\DeclareMathOperator*{\argmin}{arg\,min}

\DeclareMathOperator{\Tr}{Tr}

\def\btheta{{\bm{\theta}}}

\def\bphi{{\bm{\phi}}}
\def\bPsi{{\bm{\Psi}}}
\def\bpsi{{\bm{\psi}}}
\def\bPhi{{\bm{\Phi}}}

\def\beps{{\bm{\epsilon}}}

\newcommand{\methodl}[2]{\mathrm{#1}\text{-}\mathrm{#2}}

\usepackage{xcolor}

\definecolor{brightmaroon}{rgb}{0.76, 0.13, 0.28}
\definecolor{brown(web)}{rgb}{0.65, 0.16, 0.16}
\definecolor{customblue}{HTML}{5D94D1}

\definecolor{scoreblue}{RGB}{31,78,121}
\definecolor{fmorange}{RGB}{176,88,10}
\newcommand{\proofparagraph}[1]{%
    \vspace{0.3cm}\noindent{\normalfont\sffamily\bfseries\boldmath #1\hspace{0.2cm}}%
}

\definecolor{Pink}{rgb}{1.0, 0.44, 0.37}

\definecolor{PersB}{HTML}{FF7F0E} 
\definecolor{PersA}{HTML}{1F77B4} 
\definecolor{PersC}{HTML}{2CA02C} 

\newcommand{\sqbullet}{\leavevmode\hbox{\rule[0.35ex]{0.65ex}{0.65ex}}}

\usepackage{xcolor}

\definecolor{defscol}{HTML}{ecd8d7} 
\definecolor{asumscol}{HTML}{ecd8d7} 

\definecolor{rmkscol}{HTML}{313160} 
\definecolor{exmscol}{HTML}{e04b52} 

\definecolor{lemscol}{HTML}{2c3943} 
\definecolor{thmscol}{RGB}{230,120,20}
\definecolor{prpscol}{HTML}{8A9A5B}
\definecolor{corscol}{HTML}{dfd9fd} 

\definecolor{clmscol}{HTML}{165c58} 
\definecolor{facscol}{HTML}{28a8a1} 

\newtcbtheorem[number within=section]{mydefinition}{Definition}
{
    enhanced,
    frame hidden,
    titlerule=0mm,
    toptitle=1mm,
    bottomtitle=1mm,
    fonttitle=\sffamily\bfseries,
    coltitle=black,
    colbacktitle=defscol!50!white,
    colback=defscol!20!white,
}{defn}

\newtcbtheorem[number within=section]{myproblem}{Problem Setup}
{
    enhanced,
    frame hidden,
    titlerule=0mm,
    toptitle=1mm,
    bottomtitle=1mm,
    fonttitle=\sffamily\bfseries,
    coltitle=black,
    colbacktitle=defscol!50!white,
    colback=defscol!20!white,
}{defn}

\newcounter{msgcounter}[section]
\renewcommand{\themsgcounter}{\thesection.\arabic{msgcounter}}

\newtcolorbox{custommessagebox}[2][]{
    enhanced,
    coltitle=black,
    colbacktitle=defscol!50!white,
    colback=defscol!20!white,
    fonttitle=\sffamily\bfseries,
    title=#1,        
    toptitle=1mm,
    bottomtitle=1mm,
    titlerule=0mm,
    frame hidden,
    #2
}

\NewDocumentCommand{\msg}{mm+m}{
    \refstepcounter{msgcounter} 
    \begin{custommessagebox}[#1~\themsgcounter:~\textbfs{#2}]{}
        #3
    \end{custommessagebox}
}

\newtcbtheorem{myresearchproblem}{Research Problem}
{
    enhanced,
    frame hidden,
    titlerule=0mm,
    toptitle=1mm,
    bottomtitle=1mm,
    fonttitle=\sffamily\bfseries,
    coltitle=black,
    colbacktitle=defscol!40!white,
    colback=defscol!20!white,
}{defn}

\newtcbtheorem[use counter from=mydefinition]{myassumption}{Assumption}
{
    enhanced,
    frame hidden,
    titlerule=0mm,
    toptitle=1mm,
    bottomtitle=1mm,
    fonttitle=\sffamily\bfseries,
    coltitle=black,
    colbacktitle=asumscol!40!white,
    colback=asumscol!20!white,
}{asum}

\numberwithin{equation}{section}

\newcounter{mytheorem}[section]

\newtcbtheorem[use counter=mytheorem, number within=section]{mytheorem}{Theorem}{
    enhanced,
    frame hidden,
    titlerule=0.0mm,
    fonttitle=\sffamily\bfseries,
    coltitle=black,
    colbacktitle=thmscol!20!white,
    colback=thmscol!8!white,
}{thm}

\makeatletter
\newenvironment{thmpf}{%
  \par\noindent{\it \textbfs{Proof for Theorem.}}\par
  \tcolorbox[
    blanker,
    breakable,
    enhanced jigsaw,
    parbox=false,
    colback=white,
    left=5mm,
    before upper={\parindent15pt\relax\allowdisplaybreaks},
    after skip=10pt,
    borderline west={1mm}{0pt}{thmscol!20!white}
  ]%
}{%
  \textcolor{thmscol!18!white}{\hbox{}\nobreak\hfill$\blacksquare$}%
  \endtcolorbox
}
\makeatother

\newcommand{\thmp}[4]{%
    \begin{mytheorem}{#1}{#2}%
        #3%
    \end{mytheorem}%
     \par
    \begin{thmpf}%
        #4%
    \end{thmpf}%
    
}

\newtcbtheorem[
  use counter=mytheorem,
  number within=section
]{mylemma}{Lemma}{
  enhanced,
  frame hidden,
  titlerule=0mm,
  toptitle=1mm,
  bottomtitle=1mm,
  fonttitle=\sffamily\bfseries,
  coltitle=black,
  colbacktitle=corscol!50!white,
  colback=corscol!20!white,
}{lem}

\NewDocumentCommand{\lem}{m m +m}{%
  \begin{mylemma}[label=#2]{#1}{}%
    #3%
  \end{mylemma}%
}

\newenvironment{lempf}{%
  \par\noindent{\it \textbfs{Proof for Lemma.}}\par
  \tcolorbox[
    blanker,breakable,left=5mm,parbox=false,
    before upper={\parindent15pt},
    after skip=10pt,
    colback=white,
    borderline west={1mm}{0pt}{corscol!50!white}
  ]%
}{%
  \textcolor{corscol!50!white}{\hbox{}\nobreak\hfill$\blacksquare$}%
  \endtcolorbox
}

\NewDocumentCommand{\lemp}{m m +m +m}{%
  \begin{mylemma}[label=#2]{#1}{}%
    #3%
  \end{mylemma}%
  \par
  \begin{lempf}%
    #4%
  \end{lempf}%
}

\newtcbtheorem[use counter from=mydefinition]{mycorollary}{Corollary}
{
    enhanced,
    frame hidden,
    titlerule=0mm,
    toptitle=1mm,
    bottomtitle=1mm,
    fonttitle=\sffamily\bfseries,
    coltitle=black,
    colbacktitle=corscol!40!white,
    colback=corscol!20!white,
}{cor}

\makeatletter
\newenvironment{corpf}{%
  \par\noindent{\it \textbfs{Proof for Corollary.}}\par
  \tcolorbox[
    blanker,
    breakable,          
    enhanced jigsaw,    
    colback=white,
    parbox=false,
    left=5mm,
    before upper={\parindent15pt\allowdisplaybreaks}, 
    after skip=10pt,
    borderline west={1mm}{0pt}{corscol!40!white}
  ]%
}{%
  \textcolor{corscol!40!white}{\hbox{}\nobreak\hfill$\blacksquare$}%
  \endtcolorbox
}
\makeatother

\NewDocumentCommand{\corp}{m+m+m}{
    \begin{mycorollary}{{{#1}}}{}
        #2
    \end{mycorollary}

    \begin{corpf}
     {   #3}
    \end{corpf}
}

\NewDocumentCommand{\cornp}{m+m}{ 
    \begin{mycorollary}{{{#1}}}{}
        #2
    \end{mycorollary}
}

\newtcbtheorem[use counter=mytheorem, number within=section]{myproposition}{Proposition}
{ 
  enhanced,
  frame hidden,
  titlerule=0mm,
  toptitle=1mm,
  bottomtitle=1mm,
  fonttitle=\sffamily\bfseries,
  coltitle=black,
  colbacktitle=prpscol!18!white,
  colback=prpscol!8!white,
}
{prop}

\NewDocumentCommand{\proppnp}{m m m}{%
  \begin{myproposition}[label=#2]{#1}{}%
    #3%
  \end{myproposition}%
}

\newenvironment{proppf}
{ 
  {\noindent{\it \textbfs{Proof for Proposition.}}} 
  \tcolorbox[blanker,breakable,left=5mm,parbox=false, colback=white,
    before upper={\parindent15pt}, after skip=10pt, 
    borderline west={1mm}{0pt}{prpscol!18!white}] 
}
{ 
  \textcolor{prpscol!18!white}{\hbox{}\nobreak\hfill$\blacksquare$} 
  \endtcolorbox 
}

\NewDocumentCommand{\proppp}{m m m m}{%
  \begin{myproposition}[label=#2]{#1}{}%
    #3%
  \end{myproposition}%
  \begin{proppf}%
    #4%
  \end{proppf}%
}

\newtcbtheorem[use counter from=mydefinition]{myclaim}{Claim}
{
    enhanced,
    frame hidden,
    titlerule=0mm,
    toptitle=1mm,
    bottomtitle=1mm,
    fonttitle=\sffamily\bfseries,
    coltitle=black,
    colbacktitle=clmscol!40!white,
    colback=clmscol!20!white,
}{clm}

\newenvironment{clmpf}{
	{\noindent{\it \textbfs{Proof for Claim.}}}
	\tcolorbox[blanker,breakable,left=5mm,parbox=false,
    before upper={\parindent15pt},
    after skip=10pt,
	borderline west={1mm}{0pt}{clmscol!40!white}]
}{
    \textcolor{clmscol!40!white}{\hbox{}\nobreak\hfill$\blacksquare$} 
    \endtcolorbox
}

\newtcbtheorem[use counter from=mydefinition]{myfact}{Fact}
{
    enhanced,
    frame hidden,
    titlerule=0mm,
    toptitle=1mm,
    bottomtitle=1mm,
    fonttitle=\sffamily\bfseries,
    coltitle=black,
    colbacktitle=facscol!40!white,
    colback=facscol!20!white,
}{fact}

\newenvironment{myexample}{
    \tcolorbox[blanker,breakable,left=5mm,parbox=false,
    colback=white,
    before upper={\parindent15pt},
    after skip=10pt,
	borderline west={1mm}{0pt}{clmscol!25!white}]
}{
    \textcolor{clmscol!25!white}{\hbox{}\nobreak\hfill$\blacksquare$} 
    \endtcolorbox
}

\NewDocumentCommand{\exm}{m+m}{\vspace{0.5cm}
{\noindent{\textbfs{Example: #1 }}}
    \begin{myexample}
	
        #2
    \end{myexample}
}

\newenvironment{myremark}{
    \tcolorbox[blanker,breakable,left=5mm,parbox=false,
    colback=white,
    before upper={\parindent15pt},
    after skip=10pt,
	borderline west={1mm}{0pt}{rmkscol!20!white}]
}{
    \textcolor{rmkscol!20!white}{\hbox{}} 
    \endtcolorbox
}

\NewDocumentCommand{\rmkb}{+m}{\vspace{0.5cm}{\par\noindent{\textbfs{Remark.}}}
    \begin{myremark}
        #1
    \end{myremark}
}

\tcbuselibrary{theorems,skins,hooks}
\newtcbtheorem[number within=section]{Theorem}{Theorem}
{%
	enhanced,
	breakable,
	colback = mytheorembg,
	frame hidden,
	boxrule = 0sp,
	borderline west = {2pt}{0pt}{mytheoremfr},
	sharp corners,
	detach title,
	before upper = \tcbtitle\par\smallskip,
	coltitle = mytheoremfr,
	fonttitle = \sffamily\bfseries\sffamily,
	description font = \mdseries,
	separator sign none,
	segmentation style={solid, mytheoremfr},
}
{th}

\NewDocumentCommand{\msgu}{m m +m}{
    \begin{custommessagebox}[
        \IfBlankTF{#2}
            {#1}
            {#1:~\textbfs{#2}}
    ]{}
        #3
    \end{custommessagebox}
}

\definecolor{StoryNavy}{HTML}{23415C}       
\definecolor{StoryBlue}{HTML}{3E789B}       
\definecolor{StoryBluePale}{HTML}{EEF5F8}
\definecolor{StoryGreen}{HTML}{4F7C63}      
\definecolor{StoryGreenPale}{HTML}{EEF6F0}
\definecolor{StoryPurple}{HTML}{6D5B8C}     
\definecolor{StoryPurplePale}{HTML}{F3F0F7}
\definecolor{StoryWarm}{HTML}{9A6A3A}       
\definecolor{StoryWarmPale}{HTML}{FAF4EC}
\definecolor{StoryGray}{HTML}{5C6670}       
\definecolor{StoryGrayPale}{HTML}{F5F6F7}

\newtcolorbox{storytakeawaybox}{
  enhanced,
  breakable,
  frame hidden,
  colback=StoryGreenPale,
  borderline west={1.2mm}{0pt}{StoryGreen},
  left=3.5mm,right=3.5mm,top=2mm,bottom=2mm,
  before skip=8pt,after skip=10pt
}

\NewDocumentCommand{\proofidea}{o +m}{%
  \par\smallskip
  \noindent\textcolor{StoryPurple}{\textbfs{Proof idea.}} #2%
  \IfNoValueF{#1}{\space\textit{Full details: #1.}}%
  \par\smallskip
}

\newtcolorbox{runningexample}[1]{
  enhanced,
  breakable,
  frame hidden,
  colback=StoryGrayPale,
  borderline west={1.2mm}{0pt}{StoryGray},
  title={\textbfs{Running Example: #1}},
  fonttitle=\sffamily\bfseries,
  coltitle=StoryNavy,
  colbacktitle=StoryGrayPale,
  titlerule=0mm,
  left=3.5mm,right=3.5mm,top=2mm,bottom=2mm,
  before skip=8pt,after skip=10pt
}

\newtcolorbox{storymethodbox}[1]{
  enhanced,
  breakable,
  frame hidden,
  colback=StoryWarmPale,
  borderline west={1.2mm}{0pt}{StoryWarm},
  title={\textbfs{Method Lens: #1}},
  fonttitle=\sffamily\bfseries,
  coltitle=StoryWarm,
  colbacktitle=StoryWarmPale,
  titlerule=0mm,
  left=3.5mm,right=3.5mm,top=2mm,bottom=2mm,
  before skip=8pt,after skip=10pt
}

\newtcolorbox{storychapterclosebox}{
  enhanced,
  breakable,
  frame hidden,
  colback=StoryPurplePale,
  borderline west={1.4mm}{0pt}{StoryPurple},
  left=4mm,right=4mm,top=2.5mm,bottom=2.5mm,
  before skip=10pt,after skip=8pt
}

\renewcommand\thepart{\Alph{part}}

\newlist{todolist}{itemize}{2}
\setlist[todolist]{label=$\square$}
\usepackage{pifont}
\usepackage{pstricks}
\usepackage{pstricks-add}

\usepackage{kotex}

\DeclareRobustCommand{\figcredit}[2][Source]{%
  \par\smallskip
  \makebox[\linewidth][r]{\footnotesize\itshape #1: #2}%
}

\title{The Principles of Diffusion Models}

\maintitleauthorlist{
Chieh-Hsin Lai \\
Sony AI \\
\and
Yang Song\\
OpenAI \\
\and
Dongjun Kim \\
Stanford University \\
\and
Yuki Mitsufuji \\
Sony Group Corporation, Sony AI \\
\and
Stefano Ermon \\
Stanford University
}

\issuesetup
{%
 copyrightowner={Chieh-Hsin Lai, Yang Song, Dongjun Kim, Yuki Mitsufuji, and Stefano Ermon},
 volume        = xx,
 issue         = xx,
 pubyear       = 2024,
 isbn          = xxx-x-xxxxx-xxx-x,
 eisbn         = xxx-x-xxxxx-xxx-x,
 doi           = 10.1561/XXXXXXXXX,
 firstpage     = 1, 
 lastpage      = 18
 }

\author[1]{Chieh-Hsin Lai}
\author[2]{Yang Song}
\author[3]{Dongjun Kim}
\author[4]{Yuki Mitsufuji}
\author[5]{Stefano Ermon}

\affil[1]{Sony AI; \url{chieh-hsin.lai@sony.com} / \url{chiehhsinlai@gmail.com}}
\affil[2]{OpenAI\footnote{Affiliation reflects the institution at the time of the work.}; \url{thusongyang@gmail.com}}
\affil[3]{Stanford University; \url{dongjun@stanford.edu}}
\affil[4]{Sony Group Corporation, Sony AI; \url{yuhki.mitsufuji@sony.com}}
\affil[5]{Stanford University; \url{ermon@cs.stanford.edu}}

\articledatabox{\nowfntstandardcitation}

\usepackage{eso-pic}

\definecolor{MITTagGold}{HTML}{B8872E}
\definecolor{MITTagBrown}{HTML}{5A3518}
\definecolor{MITTagCream}{HTML}{FFF4D8}
\definecolor{MITTagOrange}{HTML}{D9982F}

\newcommand{\MITPressTagXShift}{-13mm}
\newcommand{\MITPressTagYShift}{-13mm}

\newcommand{\CoverTitleDrop}{-5mm}

\makeatletter
\newif\ifMITPressTagFirstCover
\MITPressTagFirstCovertrue

\newcommand{\MITPressBadgeContent}{%
  \begin{minipage}{6.2cm}
    \centering
    {\sffamily\bfseries\fontsize{7.2}{8.2}\selectfont
     \color{MITTagGold!75!black} FORTHCOMING FROM}%
    \hspace{0.32em}%
    {\sffamily\bfseries\fontsize{11.4}{12.4}\selectfont
     \color{MITTagBrown} MIT Press}%
    {\sffamily\bfseries\fontsize{7.2}{8.2}\selectfont
     \color{MITTagGold!75!black}, 2027}%
  \end{minipage}%
}

\newcommand{\MITPressForthcomingTag}{%
  \AddToShipoutPictureFG*{%
    \begin{tikzpicture}[remember picture,overlay]

      \node[
        anchor=north east,
        xshift=\MITPressTagXShift,
        yshift=\MITPressTagYShift,
        rounded corners=7pt,
        inner xsep=11pt,
        inner ysep=7pt,
        fill=black,
        fill opacity=0.16,
        draw=none,
        text opacity=0
      ] at ([xshift=1.15pt,yshift=-1.15pt]current page.north east) {%
        \MITPressBadgeContent
      };

      \node[
        anchor=north east,
        xshift=\MITPressTagXShift,
        yshift=\MITPressTagYShift,
        rounded corners=7pt,
        inner xsep=11pt,
        inner ysep=7pt,
        draw=none,
        top color=MITTagCream,
        bottom color=MITTagOrange!18
      ] at (current page.north east) {%
        \MITPressBadgeContent
      };

    \end{tikzpicture}%
  }%
}

\patchcmd{\nowfnt@bookauthortitlepage}
  {\thispagestyle{\nowfnt@titlepagestyle}}
  {\thispagestyle{\nowfnt@titlepagestyle}%
   \ifMITPressTagFirstCover
     \MITPressForthcomingTag
   \fi}
  {}
  {\PackageWarning{MITPressTag}{Could not attach the MIT Press tag to the cover page}}

\patchcmd{\nowfnt@bookauthortitlepage}
  {\raggedleft}
  {\raggedleft
   \ifMITPressTagFirstCover
     \vspace*{\CoverTitleDrop}%
     \global\MITPressTagFirstCoverfalse
   \fi}
  {}
  {\PackageWarning{MITPressTag}{Could not move the first-cover text down}}

\makeatother

\begin{document}

\makeabstracttitle

\begin{abstract}
This book focuses on the principles that have shaped the development of diffusion models, tracing their origins and showing how different formulations arise from common mathematical ideas.

Diffusion modeling begins by specifying a \emph{forward corruption process} that gradually turns data into noise. This forward process links the data distribution to a simple noise distribution by defining a continuous family of intermediate distributions. The core objective of a diffusion model is to construct another process that runs in the opposite direction, transforming noise into data while recovering the same intermediate distributions defined by the forward corruption process.

We describe three complementary ways to formalize this idea. The \emph{variational view}, inspired by variational autoencoders, sees diffusion as learning to remove noise step by step, solving small denoising objectives that together teach the model to turn noise back into data. The \emph{score-based view}, rooted in energy-based modeling, learns the gradient of the evolving data distribution, which indicates how to nudge samples toward more likely regions. The \emph{flow-based view}, related to normalizing flows, treats generation as following a smooth path that moves samples from noise to data under a learned velocity field.

These perspectives share a common backbone: a learned time-dependent velocity field whose flow transports a simple prior to the data. With this in hand, sampling amounts to solving a differential equation that evolves noise into data along a continuous generative trajectory. On this foundation, the book discusses \emph{guidance} for controllable generation, \emph{advanced numerical solvers} for efficient sampling, and diffusion-motivated \emph{flow-map models} that learn direct mappings between arbitrary times along this trajectory.

This book is intended for readers with a basic background in deep learning who seek a clear, conceptual, and mathematically grounded understanding of diffusion models. It develops the core principles underlying the subject, explains the ideas that unify its many formulations, and provides a solid foundation for further study in this rapidly evolving area. As such, it serves both as a principled reference for researchers and as an accessible entry point for students and newcomers. Supplementary materials for the book are available at the book website: 

{\centering
\url{https://the-principles-of-diffusion-models.github.io/}\par
}
\end{abstract}

\begin{acknowledgements}

We would like to express our sincere gratitude to the many members of the community who have offered valuable feedback and helped us identify errata. In particular, we thank the following individuals, listed in alphabetical order: \emph{Francis Bach}, \emph{Ilia Badanin}, \emph{Rwiddhi Chakraborty}, \emph{Mauricio Delbracio}, \emph{Jacob Lessing}, \emph{Kaiming He}, \emph{Yutong (Kelly) He}, \emph{Durk Kingma}, \emph{Shucheng Li}, \emph{Ramtin Moslemi}, \emph{Rukmangadh Sai Myana}, \emph{Stefano Peluchetti}, \emph{Yuri Plotkin}, \emph{François Rozet}, \emph{Yair Shenfeld}, \emph{Molei Tao}, \emph{Mohamad Ternanni}, and \emph{Baojian Zhou}, as well as anonymous reviewers and other readers who kindly shared feedback.

The authors would also like to express their deep gratitude to Professor \emph{Dohyun Kwon} from the University of Seoul and KIAS for his generous time and effort in engaging with this work. He carefully reviewed parts of \Cref{ch:ot-eot}, helping to ensure the correctness of statements and proofs, and contributed to several valuable discussions that clarified the presentation. Beyond his technical suggestions, his thoughtful feedback and willingness to share his perspectives have been a source of encouragement throughout the writing of this book. We sincerely appreciate his support and collegial spirit, which have enriched the final version.

\end{acknowledgements}

\chapter*{Preface and Roadmap}
\markboth{\sffamily\slshape Preface}
{\sffamily\slshape Preface}

Diffusion models have rapidly become a central paradigm in generative modeling, with a vast body of work spanning machine learning, computer vision, natural language processing, and beyond. This literature is dispersed across communities and highlights different dimensions of progress, including theoretical foundations that concern modeling principles, training objectives, sampler design, and the mathematical ideas behind them; implementation advances that cover engineering practices and architectural choices; practical applications that adapt the models to specific domains or tasks; and system level optimizations that improve efficiency in computation, memory, and deployment.

This book sets out to provide a \emph{principled foundation} of diffusion models, focusing on the following central themes:
\begin{itemize}
    \item We present the essential concepts and formulations that anchor diffusion model research, giving readers the core understanding needed to navigate the broader literature. We do not survey all variants or domain specific applications; instead we establish a stable conceptual foundation from which such developments can be understood.
    \item Unlike classical generative models that learn a direct mapping from noise to data, diffusion models view generation as a gradual transformation over time, refining coarse structures into fine details. This central idea has been developed through three main perspectives, i.e., \emph{variational}, \emph{score-based}, and \emph{flow-based} methods, which offer complementary ways to understand and implement diffusion modeling. We focus on the core principles and foundations of these formulations, aiming to trace the origins of their key ideas, clarify the relations among different formulations, and develop a coherent understanding that connects intuitive insight with rigorous mathematical formulation.
    \item Building on these foundations, we examine how diffusion models can be further developed to generate samples more efficiently, provide greater control over the generative process, and inspire standalone forms of generative modeling grounded in the principles of diffusion.
\end{itemize}

This book is intended for researchers, graduate students, and practitioners who have a basic understanding of deep learning (for example, what a neural network is and how training works), or more specifically, deep generative modeling, and who wish to deepen their grasp of diffusion models beyond surface-level familiarity. By the end, readers will have a principled understanding of the foundations of diffusion modeling, the ability to interpret different formulations within a coherent framework, and the background needed to both apply existing models with confidence and pursue new research directions.



\section*{\centering \Large Roadmap of This Book}

This book systematically introduces the foundations of diffusion models, tracing them back to their core underlying principles.

\paragraph{Suggested Reading Path.}  
We recommend reading this book in the presented order to build a comprehensive understanding. Sections marked as \emph{Optional} can be skipped by readers already familiar with the fundamentals. For instance, those comfortable with deep generative models (DGM) may bypass the overview in \Cref{ch:overview-dgm}. Similarly, prior knowledge of Variational Autoencoders (\Cref{sec:vae}), Energy-Based Models (\Cref{sec:ebm}), or Normalizing Flows (\Cref{sec:flow-based-method}) allows skipping these introductory sections. Other optional parts provide deeper insights into advanced or specialized topics and can be consulted as needed.

The book is organized into four main parts.
\paragraph{Parts A \& B: Foundations of Diffusion Models.}
This section traces the origins of diffusion models by reviewing three foundational perspectives that have shaped the field\footnote{We remark that our organization of the three perspectives on diffusion
models is conceptual rather than chronological. For instance,
\emph{energy-based approaches} draw on a much longer history.
Statistical mechanics connected probability with energy, while studies
of Brownian motion and subsequent work described how probability
distributions evolve over time
~\citep{gibbs1902elementary,langevin1908brownian,fokker1914mittlere,planck1917satz,kolmogorov1931}. Modern diffusion probabilistic
models later brought related ideas into generative modeling and
explicitly drew inspiration from nonequilibrium statistical physics
~\citep{sohl2015deep}. The three perspectives presented here
should therefore be understood as modern ways of explaining diffusion
models, rather than as a strict historical ordering of where their
underlying ideas first arose.}. \Cref{fig:part-b-roadmap} provides an overview of this part.

\subparagraph{Part A: Introduction to Deep Generative Modeling (DGM).}
We begin in \Cref{ch:overview-dgm} with a review of the fundamental goals of deep generative modeling. Starting from a collection of data examples, the aim is to build a model that can produce new examples that appear to come from the same underlying, and generally unknown, data distribution. Many approaches achieve this by learning how the data are distributed, either explicitly through a probability model or implicitly through a learned transformation. We then explain how such models represent the data distribution with neural networks, how they learn from examples, and how they generate new samples. The chapter concludes with a taxonomy of major generative frameworks, highlighting their central ideas and key distinctions.


\subparagraph{Part B: Core Perspectives on Diffusion Models.}

\begin{figure}[th!]
\vspace{-0.5cm}
    \centering
    \includegraphics[width=\linewidth]{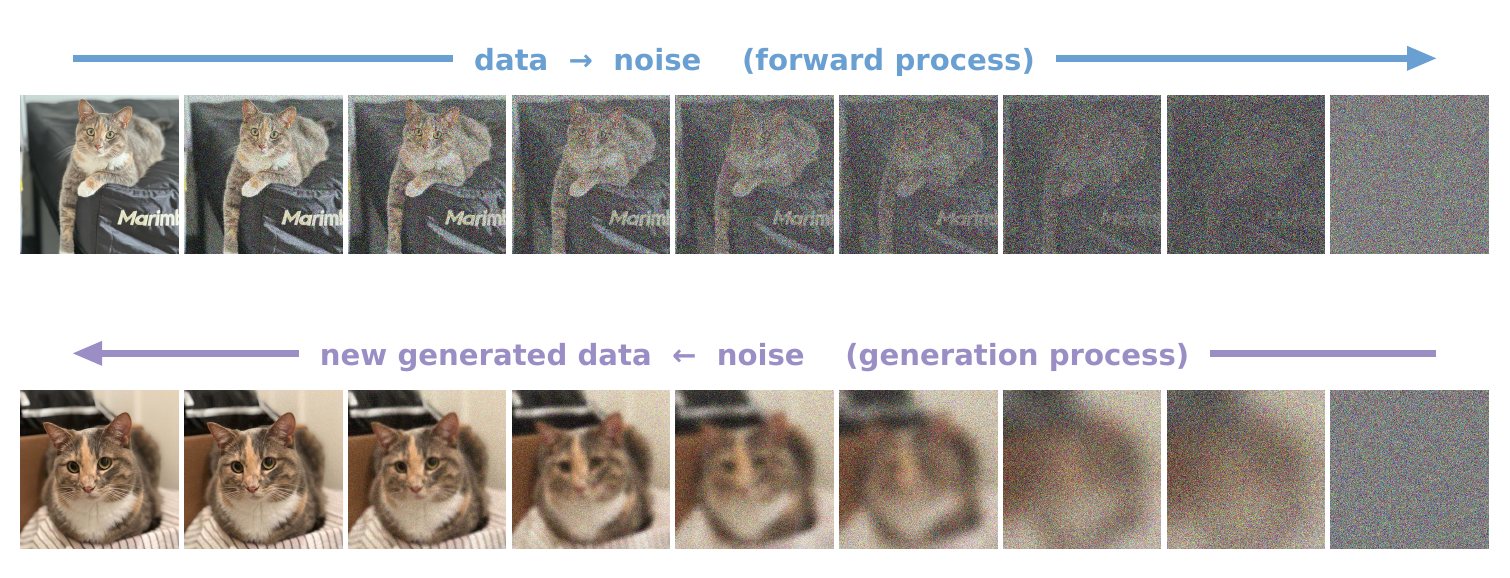}
    \caption{\textbfs{Forward and reverse diffusion trajectories.} In the forward process (top), a real data sample is gradually corrupted by noise, passing through a sequence of intermediate noisy stages until it becomes indistinguishable from pure noise. In the reverse process (bottom), generation starts from a fresh noise sample and gradually removes noise to produce a new image. The reverse process is not about reconstructing a specific example seen during training. Instead, it begins from a fresh noise and gradually shapes it into a realistic new sample. \figcredit{Created by the authors.}}
    \label{fig:ill-forward-reverse}
\end{figure}

Having outlined the general goals and mechanisms of deep generative modeling, we now turn to diffusion models, a class of methods that realize generation as a gradual transformation from noise to data (see \Cref{fig:ill-forward-reverse}). The core idea is simple: a data sample can be progressively corrupted by adding noise until it becomes indistinguishable from pure randomness. This \emph{forward process} moves different data samples into the same simple noise space, which is easy to sample from and serves as the starting point for generation. Along the way, it also creates a smooth sequence of intermediate noisy stages connecting data to noise. To generate, we draw a fresh noise sample from this space and gradually remove noise through a \emph{reverse-time (generation) process}, eventually producing a new data sample. This reverse process does not recover one particular original example; instead, it starts from fresh noise and gradually turns it into realistic data. We examine three interconnected frameworks, each built on this same principle: a forward process that gradually adds noise, and a reverse-time process approximated by a sequence of models that gradually denoise:
\begin{itemize}
    \item \textbfs{Variational View} (\Cref{ch:variational}): Originating from Variational Autoencoders (VAEs)~\citep{kingma2013auto}, it frames diffusion as learning a denoising process through a variational objective, giving rise to Denoising Diffusion Probabilistic Models (DDPMs)~\citep{sohl2015deep,ho2020denoising}.
    \item \textbfs{Score-Based View} (\Cref{ch:score-based}): Rooted in Energy-Based Models (EBMs)\\~\citep{ackley1985learning} and developed into Noise Conditional Score Networks (NCSN)~\citep{song2019generative}. It learns the score function, the gradient of the log data density, which guides how to gradually remove noise from samples. In continuous time, \Cref{ch:score-sde} introduces the \emph{Score SDE framework}, which describes this denoising process as a Stochastic Differential Equation (SDE) and its deterministic counterpart as an Ordinary Differential Equation (ODE). This view connects diffusion modeling with classical differential equation theory, providing a clear mathematical basis for analysis and algorithm design.
    \item \textbfs{Flow-Based View} (\Cref{ch:flow-based}): Building on Normalizing Flows~\citep{rezende2015variational} and generalized by Flow Matching~\citep{lipman2022flow}, this view models generation as a continuous transformation that transports samples from a simple prior toward the data distribution. The evolution is governed by a velocity field through an ODE, which explicitly defines how probability mass moves over time. This flow-based formulation naturally extends beyond prior-to-data generation to more general \emph{distribution-to-distribution translation} problems, where one seeks to learn a flow connecting any pair of source and target distributions.
\end{itemize}

\begin{figure}[th!]
  \centering
  \resizebox{0.95\linewidth}{!}{%
    \begin{tikzpicture}[x=2.6cm,y=1cm,line cap=butt]
      \tikzset{>={Stealth[length=12pt,width=16pt]}}
      \def\DotR{4pt}
      \def\LabelAngle{45}
      \def\DateAboveA{32pt}
      \def\DateAboveB{32pt}
      \def\DateXShiftA{0pt}
      \def\DateXShiftB{0pt}
      \def\TextBelow{20pt}
      \def\dx{0.9}      
      \def\XStart{0.4}  
      \newcommand{\nl}{\\}

      \definecolor{PersB}{HTML}{FF7F0E} 
      \definecolor{PersA}{HTML}{1F77B4} 
      \definecolor{PersC}{HTML}{2CA02C} 

      \def\Events{
        {1985/01}/{\color{PersB}{EBM}},
        {2013/12}/{\color{PersA}{VAE}},
        {2014/12}/{\color{PersC}{NF}},
        {2015/05}/{\color{PersA}{DPM}},
        {2018/06}/{\color{PersC}{NODE}},
        {2019/07}/{\color{PersB}{NCSN}},
        {2020/06}/{\color{PersA}{DDPM}},
        {2020/11}/{\color{PersB}{Score SDE}},
        {2022/10}/{\color{PersC}{FM}}
      }

      \pgfmathsetmacro{\N}{9}
      \draw[ultra thick,-Stealth] (0,0) -- ({\XStart + (\N-1)*\dx + 0.8},0);

      \foreach [count=\i] \date/\desc in \Events {
        \pgfmathsetmacro{\xx}{\XStart + (\i-1)*\dx}
        \fill[black] (\xx,0) circle[radius=\DotR];

        \ifodd\i
          \node[above=\DateAboveA,align=center,rotate=90,inner sep=1pt,
                xshift=\DateXShiftA] at (\xx,0) {\date};
        \else
          \node[above=\DateAboveB,align=center,rotate=90,inner sep=1pt,
                xshift=\DateXShiftB] at (\xx,0) {\date};
        \fi

        \node[below=\TextBelow,anchor=north,rotate=\LabelAngle,inner sep=1pt] at (\xx,0)
          {\shortstack[c]{\strut \desc}};
      }
    \end{tikzpicture}%
  }
  \caption{\textbfs{Timeline of modern machine-learning perspectives on diffusion models.} Each group shares the same color. 
  \\
\sqbullet\ In \Cref{ch:variational}, {Variational Autoencoder (VAE)}~\citep{kingma2013auto} $\to$ {Diffusion Probabilistic Models (DPM)}~\citep{sohl2015deep} $\to$ 
{DDPM}~\citep{ho2020denoising}. 
 \\
\sqbullet\ In \Cref{ch:score-based,ch:score-sde}, {Energy-Based Model (EBM)}~\citep{ackley1985learning} $\to$ {Noise Conditional Score Network (NCSN)}~\citep{song2019generative} $\to$ {Score SDE}~\citep{song2020score}.  \\
\sqbullet\
In \Cref{ch:flow-based}, {Normalizing Flow (NF)}~\citep{rezende2015variational} $\to$ {Neural ODE (NODE)}~\citep{chen2018neural} $\to$ {Flow Matching (FM)}~\citep{lipman2022flow}.
\figcredit{Created by the authors.}}
  \label{fig:timeline-diffusion-perspectives-equal}
\end{figure}

Although these perspectives may seem different at first, \Cref{ch:all-equivalent} shows that they are deeply connected. 
Each uses a \emph{conditioning strategy} that turns the learning objective into a tractable regression problem. 
At a deeper level, they all describe the same temporal evolution of probability distributions, from the prior toward the data. 
This evolution is governed by the \emph{Fokker–Planck equation}, which can be viewed as the continuous-time change of variables for densities, ensuring consistency between the stochastic and deterministic formulations.

Since diffusion models can be viewed as approaches for transporting one distribution to another, \Cref{ch:ot-eot} develops their connections to classical optimal transport and the Schrödinger bridge, interpreted as optimal transport with entropy regularization. We review both the static and dynamic formulations and explain their relations to the continuity equation and the Fokker--Planck perspective. This chapter is optional for readers focused on practical aspects, but it provides rigorous mathematical background and pointers to the classical literature for those who wish to study these links in depth. 


\begin{figure}[ht!]
\begin{center}
\resizebox{\textwidth}{!}{%
\begin{tikzpicture}[
  box/.style = {draw, rounded corners, minimum width=3.6cm, minimum height=1.5cm, align=center},
  model/.style = {draw, fill=gray!20, minimum width=3.5cm, minimum height=1cm, align=center},
  arrow/.style = {ultra thick, -{Stealth[length=4mm, width=4mm]}}
  ]

\node[box] (vae) {Variational \\Autoencoder};
\node[box, right=1.2cm of vae] (ebm) {Energy-Based Model};
\node[box, right=1.2cm of ebm] (flow) {Normalizing Flows};

\node[draw, rounded corners, minimum width=13.6cm, minimum height=1.0cm, thick, above=3.5cm of ebm, label=above:{\parbox{5cm}{ \centering {Chapter}~\ref{ch:overview-dgm} 
 }}] (dgm) {Overview of Deep Generative Modeling};
 
\node[model, above=1.5cm of vae] (variational-based) {\large\textbfs{Variational View}};
\node[model, above=1.5cm of ebm] (score-based) {\large\textbfs{Score-Based View}};
\node[model, above=1.5cm of flow] (flow-based) {\large\textbfs{Flow-Based View}};

\node[box, below=2.2cm of vae] (ddpm) {Denoising Diffusion\\Probabilistic Model\\ (DDPM)};
\node[box, below=2.2cm of ebm] (score) {Noise Conditional\\Score Network \\(NCSN)};
\node[box, below=2.2cm of flow] (gfm) {Gaussian\\Flow Matching};

\draw[arrow] (vae.south) -- node[midway, anchor=center, yshift=3.5pt, xshift=-8pt] {\rotatebox{90}{\small {Chapter~\ref{ch:variational}}}} (ddpm.north);

\draw[arrow] (ebm.south) -- node[midway, anchor=center, yshift=3.5pt, xshift=-8pt] {\rotatebox{90}{\small {Chapter}~\ref{ch:score-based}}} (score.north);

\draw[arrow] (flow.south) -- node[midway, anchor=center, yshift=3.5pt, xshift=-8pt] {\rotatebox{90}{\small {Chapter}~\ref{ch:flow-based}}} (gfm.north);

\node[box, below=2cm of score] (scoresde) { Continuous-Time Formulation \\ (e.g., Score SDE) \\ \small {Chapter}~\ref{ch:score-sde}};

\node[draw, rounded corners, thick, fit={(ddpm) (score) (gfm) (scoresde)}, inner sep=0.3cm, label=below:{\parbox{5cm}{\centering\textbfs{\\ \large Unifying Principles}  \\ \vspace{0.2cm}\small {Chapter}~\ref{ch:all-equivalent} 
\begin{itemize}
    \item Conditional Strategy 
    \item Fokker-Planck Equation
\end{itemize} }}] (box) {};

\draw[ultra thick, -{Stealth[length=4mm, width=4mm]}, rounded corners=4pt] (ddpm.south) -- ++(0,-0.4) -- ([xshift=-1.0cm]scoresde.north);
\draw[ultra thick, -{Stealth[length=4mm, width=4mm]}, rounded corners=4pt] (score.south) -- ++(0,-0.4) -- (scoresde.north);
\draw[ultra thick, -{Stealth[length=4mm, width=4mm]}, rounded corners=4pt] (gfm.south) -- ++(0,-0.4) -- ([xshift=1.0cm]scoresde.north);

\node[align=center, left=0.5cm of box, yshift=8.3cm] {\rotatebox{90}{\textbfs{\large{Perspective}}}};
\node[align=center, left=0.5cm of box, yshift=5.5cm] {\rotatebox{90}{\textbfs{\large{Origin}}}};
\node[align=center, left=0.5cm of box] {\rotatebox{90}{\textbfs{\large{Diffusion Model}}}};
\end{tikzpicture}
    }
\end{center}
\caption{\textbfs{Part B. Unifying and Principled Perspectives on Diffusion Models.} This diagram visually connects classical generative modeling approaches—Variational Autoencoders, Energy-Based Models, and Normalizing Flows—with their corresponding diffusion model formulations. Each vertical path illustrates a conceptual lineage, culminating in the continuous-time framework. The three views (Variational, Score-Based, and Flow-Based) offer distinct yet mathematically equivalent interpretations. \figcredit{Created by the authors.}}
\label{fig:part-b-roadmap}
\end{figure}



\paragraph{Part C \& D: Controlling and Accelerating the Diffusion Sampling.}

With the foundational principles unified, we now turn to practical aspects of utilizing diffusion models for efficient generation. Sampling from a diffusion model corresponds to solving a differential equation. However, this procedure is typically computationally expensive. Parts C and D focus on improving generation quality, controllability, and efficiency through enhanced sampling and learned acceleration techniques.

\subparagraph{Part C: Sampling from Diffusion Models.} The generation process of diffusion models exhibits a distinctive coarse-to-fine refinement: noise is removed step by step, yielding samples with increasingly coherent structure and detail. This property comes with trade-offs. On the positive side, it affords fine-grained control; by adding a guidance term to the learned, time-dependent velocity field, we can steer the ODE flow to reflect user intent and make sampling controllable. On the negative side, the required iterative integration makes sampling slow compared with single-shot generators.
This part focuses on improving the generative process at inference time, without retraining.
\begin{itemize}
    \item \textbfs{Steering Generation} (\Cref{ch:guidance})\textbfs{:} Techniques such as classifier guidance and classifier-free guidance make it possible to condition the generation process on user-defined objectives or attributes. Building on this, we next discuss how the use of a preference dataset can further align diffusion models with such preferences.
    \item \textbfs{Fast Generation with Numerical Solvers} (\Cref{ch:solvers})\textbfs{:} Sampling can be significantly accelerated using advanced numerical solvers that approximate the reverse process in fewer steps, reducing cost while preserving quality.
\end{itemize}
\subparagraph{Part D: Learning Fast Generative Models.} Beyond improving existing sampling algorithms, we investigate how to directly learn fast generators that approximate the diffusion process.
\begin{itemize}
    \item \textbfs{Distillation-Based Methods} (\Cref{ch:distillation}):  This approach focuses on training a student model to imitate the behavior of a pre-trained, slow diffusion model (the teacher). Instead of reducing the teacher’s size, the goal is to reproduce its sampling trajectory or output distribution with far fewer integration steps, often only a few or even one.
    \item \textbfs{Learning from Scratch} (\Cref{ch:fast-scratch})\textbfs{:} Since sampling in diffusion models can be seen as solving an ODE, this approach learns the solution map (i.e., the flow map) directly from scratch, without relying on a teacher model. The learned map can take noise directly to data, or more generally perform anytime-to-anytime jumps along the solution trajectory.
\end{itemize}


\paragraph{Epilogue: Beyond Diffusion on Continuous State Spaces.}
We close in \Cref{ch:epilogue} by making explicit the principle that has guided the entire book. At its heart, diffusion modeling is not tied to one particular noise type, model class, or even to continuous state spaces. Rather, it is the study of how probability mass evolves under a prescribed time-dependent forward corruption process, and how that evolution can be reversed, approximated, or exploited for generation. In the continuous setting, this principle appears through the change-of-variable formula, the continuity equation, and the Fokker--Planck equation. In the discrete setting, the same structural role is played by transition kernels, continuous-time Markov chains, and the master equation.

This final chapter also shows that the three perspectives developed throughout the book, namely the variational, score-based, and flow-based viewpoints, extend beyond continuous data modeling. Even for discrete data such as text, protein sequences, and other token-based objects, one can formulate reverse modeling through variational objectives, score-like ratio or reverse-rate characterizations, and flow-like transport viewpoints. The mathematical objects change, but the underlying principles do not. In this way, the epilogue serves both as a conceptual culmination of the book and as an outlook toward a broader landscape of generative modeling.

\paragraph{Appendices.}
To ensure our journey is accessible to all, the appendices provide background for foundational concepts. \Cref{app:de} offers a crash course on the differential equations that have become the language of diffusion models. 

The core insight behind diffusion models, despite their varied perspectives and origins, lies in the \emph{change-of-variables formula}. This foundation naturally extends to deeper concepts such as the \emph{Fokker--Planck equation} and the \emph{continuity equation}, which describe how probability densities transform and evolve under mappings defined by functions (discrete time) or differential equations (continuous time). 
\Cref{app:continuity} offers a gentle introduction that bridges these foundational ideas to more advanced concepts. 
In \Cref{app:Ito}, we present two powerful but often overlooked tools underlying diffusion models: \emph{Itô's formula} and \emph{Girsanov's theorem}, which provide rigorous support for the Fokker--Planck equation and the reverse-time sampling process. 
Finally, \Cref{app:proof} gathers proofs of selected propositions and theorems discussed in the main chapters.



\paragraph{What This Book Covers and What It Does Not.}
We aim for durability. From a top-down viewpoint, this book begins with a single principle: construct continuous-time dynamics that transport a simple prior to the data distribution while ensuring that the marginal distribution at each time matches the marginal induced by a prescribed forward process from data to noise. From this principle, we develop the stochastic and deterministic flows that enable sampling, show how to steer the trajectory (guidance), and explain how to accelerate it (numerical solvers). We then study diffusion-motivated fast generators, including distillation methods and flow-map models. With these tools, readers can place new papers within a common template, understand why methods work, and design improved models.

We do not attempt to provide an exhaustive survey of the diffusion model literature, nor do we catalog architectures, training practices, hyperparameters, compare empirical results across methods, cover datasets and leaderboards, describe domain- or modality-specific applications, address system-level deployment, provide recipes for large-scale training, or discuss hardware engineering. These topics evolve rapidly and are better covered by focused surveys, open repositories, and implementation guides.

\chapter*{Notations}
\markboth{\sffamily\slshape Notations}{\sffamily\slshape Notations}

\newcommand{\notationsection}[1]{%
  \vspace{1.2em}%
  \noindent{\textbfs{#1}}%
  \vspace{0.4em}%
  \par\nobreak
}

\newcommand{\notationtable}[1]{%
  \bgroup
  \def\arraystretch{1.25}%
  \noindent\begin{tabular}{@{}p{0.3\linewidth}@{\hspace{1.5em}}p{0.64\linewidth}@{}}
  #1
  \end{tabular}
  \egroup
}

\notationsection{Numbers and Arrays}
\notationtable{
$a$                                       & A scalar.\\
$\rva$                                    & A column vector (e.g., $\rva\in\mathbb{R}^D$).\\
$\rmA$                                    & A matrix (e.g., $\rmA\in\mathbb{R}^{m\times n}$).\\
$\rmA^\top$                               & Transpose of $\rmA$.\\
$\Tr(\rmA)$                               & Trace of $\rmA$.\\
$\rmI_D$                                  & Identity matrix of size $D\times D$.\\
$\rmI$                                    & Identity matrix; dimension implied by context.\\
$\mathrm{diag}(\rva)$                     & Diagonal matrix with diagonal entries given by $\rva$.\\
$\bm{\phi},\,\bm{\theta}$                & Learnable neural network parameters.\\
$\bm{\phi}^{\times},\,\bm{\theta}^{\times}$ & Parameters after training (fixed during inference).\\
$\bm{\phi}^{*},\,\bm{\theta}^{*}$        & Optimal parameters of an optimization problem.\\
}

\notationsection{Calculus}
\notationtable{
$\dfrac{\partial \rvy}{\partial \rvx}$
    & Partial derivatives of $\rvy$ w.r.t.\ $\rvx$ (componentwise).\\[6pt]
$\dfrac{\diff \rvy}{\diff \rvx}$ \text{or} $\mathrm{D}\rvy(\rvx)$
    & Total (Fréchet) derivative of $\rvy$ w.r.t.\ $\rvx$.\\[6pt]
$\nabla_\rvx y$
    & Gradient of scalar $y:\mathbb{R}^D \to\mathbb{R}$; a column in $\mathbb{R}^D$.\\
$\dfrac{\partial \rmF}{\partial \rvx}$ \text{or} $\nabla_\rvx \rmF$
    & Jacobian of $\rmF:\mathbb{R}^n \to\mathbb{R}^m$; shape $m\times n$.\\[6pt]
$\nabla\cdot\rvy$
    & Divergence of a vector field $\rvy:\mathbb{R}^D \to\mathbb{R}^D$; a scalar.\\
$\nabla^2_\rvx f(\rvx)$ \text{or} $\rmH(f)(\rvx)$
    & Hessian of $f:\mathbb{R}^D \to\mathbb{R}$; shape $D\times D$.\\
$\displaystyle\int f(\rvx)\diff\rvx$
    & Integral of $f$ over the domain of $\rvx$.\\
}

\notationsection{Probability and Information Theory}
\notationtable{
$p(\rvx)$                   & Density/distribution over a continuous vector $\rvx$.\\
$\Pr(\rvx = \rva)$, or informally $\Pr(\rvx)$
    & Point probability in the discrete setting.\\
$\Pr(\rvx = \rva |\rvy = \rvb)$, or informally $\Pr(\rvx |\rvy)$
    & Conditional point probability in the discrete setting.\\
$p_{\mathrm{data}}$        & Data distribution.\\
$p_{\mathrm{prior}}$       & Prior distribution (e.g., standard normal).\\
$p_{\mathrm{src}}$         & Source distribution.\\
$p_{\mathrm{tgt}}$         & Target distribution.\\
$\rva \sim p$              & Random vector $\rva$ is distributed as $p$.\\
$\E_{\rvx\sim p}\big[\rvf(\rvx)\big]$
    & Expectation of $\rvf(\rvx)$ under $p(\rvx)$.\\[3pt]
\parbox[t]{0.22\linewidth}{\raggedright
  $\E\big[\rvf(\rvx)\,|\,\rvz\big]$,\\[2pt]
  or $\E_{\rvx\sim p(\cdot|\rvz)}\big[\rvf(\rvx)\big]$}
    & Conditional expectation of $\rvf(\rvx)$ given $\rvz$, with $\rvx$ distributed as $p(\cdot|\rvz)$.\\[3pt]
$\Var\big(\rvf(\rvx)\big)$
    & Variance under $p(\rvx)$.\\
$\Cov\big(\rvf(\rvx),\rvg(\rvx)\big)$
    & Covariance under $p(\rvx)$.\\
$\mathcal{D}_{\mathrm{KL}}\left(p\,\Vert\, q\right)$
    & Kullback--Leibler divergence from $q$ to $p$.\\
$\bm{\epsilon}\sim\mathcal{N}(\bm{0},\rmI)$
    & Standard normal sample.\\
$\mathcal{N}(\rvx;\vmu,\mSigma)$
    & Gaussian over $\rvx$ with mean $\vmu$ and covariance $\mSigma$.\\
}

\vspace{1.2em}

\paragraph{Notational Conventions.}
We collect several conventions used throughout this book.

\subparagraph{I. Random Variables and Their Realizations.}
We use the same symbol for a random vector and its realized value.
This convention, common in deep learning and generative modeling,
keeps notation compact. The intended meaning is determined by context.

For densities, $p(\rvx)$ denotes the functional form of the
distribution rather than evaluation at a particular sample. When
evaluation at a specific point is intended, we state so explicitly.
Conditional densities are read by context: in $p(\rvx|\rvy)$, fixing
$\rvy$ gives a density in $\rvx$, while fixing $\rvx$ gives a
function of $\rvy$.

\subparagraph{II. Conditional Expectations.}
The expression $\E[\rvf(\rvx)|\rvz]$ denotes a function of $\rvz$,
giving the expected value of $\rvf(\rvx)$ conditional on $\rvz$.
When conditioning on a specific realized value, we write
$\E[\rvf(\rvx)|\rmZ = \rvz]$. Equivalently, this can be expressed as
an integral with respect to the conditional distribution:
\[
\E_{\rvx \sim p(\cdot|\rvz)}[\rvf(\rvx)]
= \int \rvf(\rvx)\, p(\rvx|\rvz)\, \diff\rvx.
\]
This distinction clarifies whether $\rvz$ is treated as a variable
defining a function $\rvz \mapsto \E[\rvf(\rvx)|\rvz]$, or as a
fixed value at which that function is evaluated.

\subparagraph{III. Implicit Time Indices.}
Since many objects in this book are indexed by time, writing every
time variable explicitly can make formulas heavy. When the relevant
time indices are clear from context, we leave them implicit. For
example,
\[
\E[\rvx_s | \rvx_t]
\quad\text{is shorthand for}\quad
\E[\rvx_s | \rvx_t,\, s,\, t],
\]
meaning the average of $\rvx_s$ at time $s$ given $\rvx_t$ at time
$t$, with $s$ and $t$ understood from the surrounding discussion.
Similarly, $p(\rvx_s | \rvx_t)$ denotes the conditional distribution
of $\rvx_s$ given $\rvx_t$, with both time indices left implicit.

\subparagraph{IV. Gaussian Perturbation and Equality in Distribution.}
For a fixed index $t$, we frequently use the Gaussian perturbation
kernel
\[
p_t(\rvx_t|\rvx_0)
= \mathcal{N}\big(\rvx_t;\, \alpha_t \rvx_0,\, \sigma_t^2 \rmI\big),
\]
where $\rvx_0 \sim p_{\mathrm{data}}$, $\alpha_t$ and $\sigma_t$ are
deterministic scalars. We equivalently write this as the identity
\emph{in distribution}
\[
\rvx_t \stackrel{d}{=} \alpha_t \rvx_0 + \sigma_t \beps,
\qquad
\beps \sim \mathcal{N}(\bm{0}, \rmI),
\]
where $\beps$ is independent of $\rvx_0$. The symbol
``$\stackrel{d}{=}$'' means the two sides have the same probability
law, so that 
\[
    \E[\phi(\rvx_t)] = \E[\phi(\alpha_t \rvx_0 + \sigma_t \beps)]
\]
for any test function $\phi$. For brevity, we will simply
write $\rvx_t = \alpha_t \rvx_0 + \sigma_t \beps$, understood either
as an equality in distribution or as a sample realization depending
on context; this shorthand is used throughout.
\part{Introduction to Deep Generative Modeling}
\chapter{Deep Generative Modeling}\label{ch:overview-dgm}

\epigraph{
    \textit{What I cannot create, I do not understand.}}{Richard P. Feynman}




Deep generative models (DGMs) are neural networks that learn a probability distribution over high-dimensional data (e.g., images, text, audio) so they can generate new examples that resemble the dataset. We denote the model distribution by $p_{\bphi}$ and the data distribution by $p_{\mathrm{data}}$. Given a finite dataset, we fit $\bphi$ by minimizing a loss that measures how far $p_{\bphi}$ is from $p_{\mathrm{data}}$. After training, generation amounts to running the model’s sampling procedure to draw $\rvx \sim p_{\bphi}$ (the density $p_{\bphi}(\rvx)$ may or may not be directly computable, depending on the model class). Model quality is judged by how well generated samples and their summary statistics match those of $p_{\mathrm{data}}$, together with task-specific or perceptual metrics.

This chapter builds the mathematical and conceptual foundations behind these ideas. We formalize the problem in \Cref{sec:formulation-dgm}, present representative model classes in \Cref{sec:examples-dgm}, and summarize a practical taxonomy in \Cref{sec:taxonomy}.

\chapterlearning{
  \item explain, in plain language, what a generative model is trying to learn from a dataset and what it means to model the data distribution;
  \item understand the core idea behind autoregressive models, VAEs, energy-based models, normalizing flows, GANs, and diffusion models, and see how they differ in the way they learn and generate new samples;
}{
If you are already familiar with the basic ideas behind modern generative models, you may skim this chapter and use it later as a quick reference.
}




\newpage
\section{What is Deep Generative Modeling?}\label{sec:formulation-dgm}


DGMs take as input a large collection of real-world examples 
(e.g., images, text) drawn from an unknown and complex distribution $p_{\mathrm{data}}$ 
and output a trained neural network that parameterizes an approximate distribution $p_{\bphi}$. Their goals are twofold: 
\begin{enumerate}
\item \textbfs{Realistic Generation:} To generate novel, realistic samples indistinguishable from real data.
\item \textbfs{Controllable Generation:} To enable fine-grained and interpretable control over the generative process.
\end{enumerate}



This section presents the fundamental concepts and motivations behind DGMs, preparing for a detailed exploration of their mathematical framework and practical applications.

\subsection{Mathematical Setup}\label{subsec:intro-math-setup}
Suppose we are given a finite collection of examples
$\{\rvx^{(i)}\}_{i=1}^N$, such as images, audio, or text.
We imagine that these examples were drawn from some underlying data distribution
$p_{\mathrm{data}}$, but we do not know this distribution explicitly.
In particular, we only observe the available training examples; we do not have a
formula for $p_{\mathrm{data}}$ or a direct procedure for drawing arbitrary new
samples from it\footnote{This is a common assumption in machine learning. For simplicity, we use the symbol $ p $ to represent either a probability distribution or its probability density/mass function, depending on the context.}.

\paragraph{Goal of DGM.}
The goal of deep generative modeling is therefore to learn a model that can
produce \emph{new} samples whose statistical characteristics resemble those of
the observed data.
We denote the distribution represented by a model with parameters
$\bm{\phi}$ by $p_{\bm{\phi}}$.
After training, we denote the learned parameters by $\bm{\phi}^{\times}$ and
hope that, in the aspects relevant to generation,
\[
    p_{\bm{\phi}^{\times}}
    \approx
    p_{\mathrm{data}}.
\]
This expression should be understood conceptually: with finite data, finite
model capacity, and imperfect optimization, we do not expect the two
distributions to be exactly identical.
Rather, the learned model should capture enough of the structure of the data
distribution to generate realistic and diverse new examples.

Once the model is trained, generation simply means drawing a new sample
\[
    \rvx \sim p_{\bm{\phi}^{\times}}.
\]
Different families of generative models provide different mechanisms for doing
this.
Some models additionally allow us to evaluate how much density they assign to
an observed sample, while others are primarily defined through their sampling
procedure.
This distinction will become useful later when we discuss how different
generative models are trained; for now, the central objective is simply to
learn a distribution from data from which we can generate new samples.

\begin{figure}
\vspace{-0.5cm}
    \centering
    \includegraphics[width=0.9\linewidth]{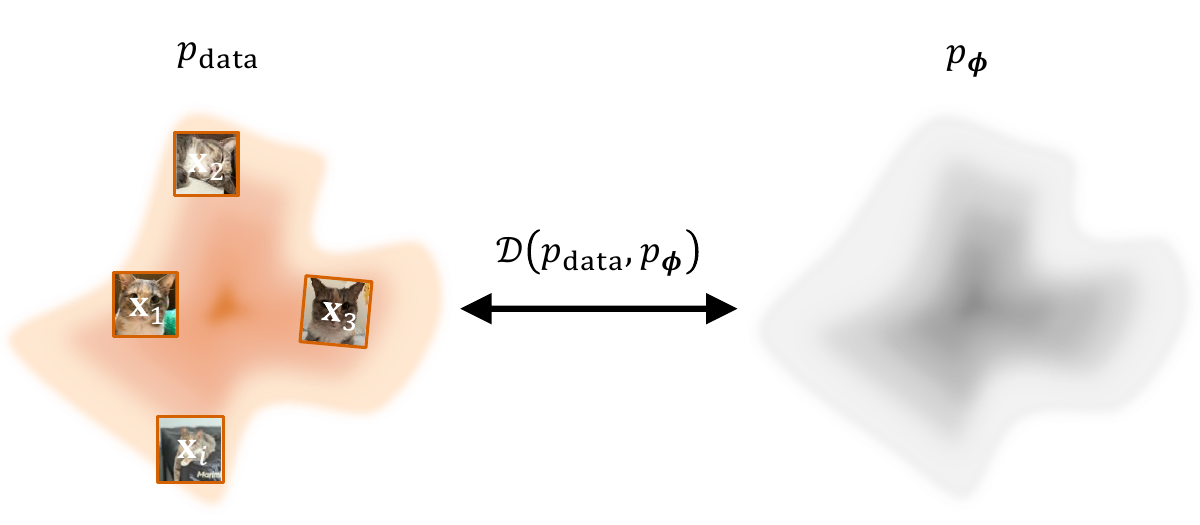}
    \vspace{-0.2cm}
    \caption{\textbfs{Illustration of the target in DGM.} Training a DGM is essentially minimizing the discrepancy between the model distribution $p_{\bm{\phi}}$ and the unknown data distribution $p_{\mathrm{data}}$. Since $p_{\mathrm{data}}$ is not directly accessible, this discrepancy must be estimated efficiently using a finite set of independent and identically distributed (i.i.d.) samples, $\rvx_i$, drawn from it. \figcredit{Created by the authors.}}
    \label{fig:dgm-training}
\end{figure}

\paragraph{Training of DGM.}
We learn parameters $\bm{\phi}$ of a model family $\{p_{\bm{\phi}}\}$ by minimizing a discrepancy $\mathcal{D}(p_{\mathrm{data}},p_{\bm{\phi}})$:
\begin{align}\label{eq:dgm-optimization}
  \bm{\phi}^\ast \in \arg\min_{\bm{\phi}}\; \mathcal{D}(p_{\mathrm{data}},p_{\bm{\phi}}).
\end{align}
Because $p_{\mathrm{data}}$ is unknown, a practical choice of $\mathcal{D}$ must admit efficient estimation from i.i.d.\ samples
from $ p_{\mathrm{data}}$. With sufficient capacity, $p_{\bm{\phi}^\ast}$ can closely approximate $p_{\mathrm{data}}$.

\subparagraph{Forward KL and Maximum Likelihood Estimation (MLE).}
How should we choose the model parameters so that
$p_{\bm{\phi}}$ resembles $p_{\mathrm{data}}$?
When the model density $p_{\bm{\phi}}(\rvx)$ can be evaluated, a natural idea is
to make the observed data receive high density under the model.
This leads to \emph{maximum likelihood estimation} (MLE): we choose
$\bm{\phi}$ so that the training examples are collectively assigned high model
density\footnote{For a fixed observed sample $\rvx$, the quantity
$p_{\bm{\phi}}(\rvx)$, viewed as a function of the model parameters
$\bm{\phi}$, is called the \emph{likelihood}. For continuous data,
$p_{\bm{\phi}}(\rvx)$ is a probability density rather than the probability of
observing exactly $\rvx$.}.

A standard way to formalize this idea is through the (forward)
Kullback--Leibler divergence\footnote{All integrals are in the Lebesgue sense
and reduce to sums under counting measures.}
\begin{align*}
\mathcal{D}_{\mathrm{KL}}
\big(p_{\mathrm{data}}\|p_{\bm{\phi}}\big)
:=\;&
\int
p_{\mathrm{data}}(\rvx)
\log
\frac{p_{\mathrm{data}}(\rvx)}
     {p_{\bm{\phi}}(\rvx)}
\,\diff \rvx
\\
=\;&
\mathbb{E}_{\rvx\sim p_{\mathrm{data}}}
\big[
\log p_{\mathrm{data}}(\rvx)
-
\log p_{\bm{\phi}}(\rvx)
\big].
\end{align*}
The KL divergence is asymmetric, i.e.,
\[
\mathcal{D}_{\mathrm{KL}}
\big(p_{\mathrm{data}}\|p_{\bm{\phi}}\big)
\neq
\mathcal{D}_{\mathrm{KL}}
\big(p_{\bm{\phi}}\|p_{\mathrm{data}}\big).
\]

At first, the forward KL may appear unusable because the data density
$p_{\mathrm{data}}(\rvx)$ is unknown.
However, it can be decomposed as
\begin{align*}
\mathcal{D}_{\mathrm{KL}}
\big(p_{\mathrm{data}}\|p_{\bm{\phi}}\big)
=
\mathbb{E}_{\rvx\sim p_{\mathrm{data}}}
\left[
\log
\frac{p_{\mathrm{data}}(\rvx)}
     {p_{\bm{\phi}}(\rvx)}
\right]
=
-\mathbb{E}_{\rvx\sim p_{\mathrm{data}}}
\big[
\log p_{\bm{\phi}}(\rvx)
\big]
-
\mathcal{H}\big(p_{\mathrm{data}}\big),
\end{align*}
where
\[
\mathcal{H}\big(p_{\mathrm{data}}\big)
:=
-\mathbb{E}_{\rvx\sim p_{\mathrm{data}}}
\big[
\log p_{\mathrm{data}}(\rvx)
\big]
\]
is the entropy of the data distribution.
Crucially, this term does not depend on $\bm{\phi}$.
Therefore, although we cannot evaluate $p_{\mathrm{data}}(\rvx)$ itself,
we do not need to know it in order to optimize the model parameters. This gives the following equivalence:
\lem{Minimizing KL $\Leftrightarrow$ MLE}{mle-kl}{
\begin{align}\label{eq:MLE}
\min_{\bm{\phi}}\,
\mathcal{D}_{\mathrm{KL}}
\big(
p_{\mathrm{data}}
\,\|\,
p_{\bm{\phi}}
\big)
\;\Longleftrightarrow\;
\max_{\bm{\phi}}\,
\mathbb{E}_{\rvx\sim p_{\mathrm{data}}}
\big[
\log p_{\bm{\phi}}(\rvx)
\big].
\end{align}
}

In words, minimizing the forward KL divergence is equivalent to making real
data samples receive high log-likelihood under the model.
This is exactly maximum likelihood estimation.

In practice, the expectation over $p_{\mathrm{data}}$ is replaced by the
available training samples
$\{\rvx^{(i)}\}_{i=1}^{N}$, assumed to be drawn i.i.d.\ from
$p_{\mathrm{data}}$.
This gives the empirical negative log-likelihood objective
\begin{align*}
\hat{\mathcal{L}}_{\mathrm{MLE}}(\bm{\phi})
:=
-\frac{1}{N}
\sum_{i=1}^{N}
\log p_{\bm{\phi}}\big(\rvx^{(i)}\big),
\end{align*}
which can be optimized using stochastic gradients over minibatches.
Thus, MLE requires only samples from $p_{\mathrm{data}}$; it does not require
evaluating the unknown data density itself.

The direction of the KL divergence matters because it determines
\emph{where the discrepancy is averaged}.
For the forward KL,
\[
\mathcal{D}_{\mathrm{KL}}
\big(p_{\mathrm{data}}\|p_{\bm{\phi}}\big)
=
\mathbb{E}_{\rvx\sim p_{\mathrm{data}}}
\left[
\log
\frac{p_{\mathrm{data}}(\rvx)}
     {p_{\bm{\phi}}(\rvx)}
\right],
\]
the expectation is taken over the data distribution.
Thus, regions that occur under $p_{\mathrm{data}}$ must be accounted for by
the model.
By contrast, the reverse KL,
$\mathcal{D}_{\mathrm{KL}}
(p_{\bm{\phi}}\|p_{\mathrm{data}})$,
averages instead over samples from the model distribution.

This difference has an important consequence.
Suppose there exists a region $A$ with positive data probability,
$p_{\mathrm{data}}(A)>0$, but the model assigns zero density there,
$p_{\bm{\phi}}(\rvx)=0$ for $\rvx\in A$.
Then the forward KL becomes infinite, since
\[
\log
\frac{p_{\mathrm{data}}(\rvx)}
     {p_{\bm{\phi}}(\rvx)}
=
+\infty
\]
on that region.
Therefore, minimizing the forward KL strongly penalizes a model for completely
missing regions where the data distribution places probability mass.
This behavior is commonly referred to as \emph{mode covering}.

\subparagraph{Fisher Divergence.} The Fisher divergence is another important concept for (score-based) diffusion modeling (see \Cref{ch:score-based}). For two distributions $p$ and $q$, it is defined as
\begin{align}\label{eq:fisher}
    \mathcal D_{\mathrm F}(p \|  q)
:=
\mathbb{E}_{\mathbf{x}\sim p} \left[
\left\|\nabla_{\mathbf x}\log p(\mathbf x)-\nabla_{\mathbf x}\log q(\mathbf x)\right\|_2^{2}
\right].
\end{align}
It measures the discrepancy between the \emph{score functions} 
$\nabla_{\mathbf x}\log p(\mathbf x)$ and $\nabla_{\mathbf x}\log q(\mathbf x)$,
which are vector fields pointing toward regions of higher probability. 
In short, $\mathcal D_{\mathrm F}(p \|  q)\ge 0$ with equality if and only if $p=q$ almost everywhere. 
It is invariant to normalization constants, since scores depend only on gradients of log-densities, 
and it forms the basis of \emph{score matching} (\Cref{eq:ebm-sm,eq:sm}): a method that learns the gradient of the log-density for generation (score-based models). 
In this setting, the data distribution $p=p_{\mathrm{data}}$ serves as the target, 
while the model $q=p_{\bm{\phi}}$ is trained to align its score field with that of the data.

\subparagraph{Beyond KL.} Although the KL divergence is the most widely used measure of difference between probability distributions, it is not the only one. Different divergences capture different geometric or statistical notions of discrepancy, which in turn affect the optimization dynamics of learning algorithms.  A broad family is the \emph{$f$-divergences}~\citep{csiszar1963informationstheoretische}:
\begin{align}\label{eq:f-div}
    \mathcal{D}_f(p\|q)
=\int q(\rvx) f \left(\frac{p(\rvx)}{q(\rvx)}\right)\diff \rvx,
\qquad f(1)=0,
\end{align}
where $f:\mathbb{R}_+ \to\mathbb{R}$ is a convex function.  
By changing $f$, we obtain many well-known divergences:
\[
\begin{aligned}
f(u)&=u\log u &&\Rightarrow&& \mathcal{D}_f=\mathcal{D}_{\mathrm{KL}}(p\|q)\quad\text{(forward KL)},\\
f(u)&=\tfrac12 \left[u\log u-(u+1)\log \tfrac{1+u}{2}\right] &&\Rightarrow&& \mathcal{D}_f=\mathcal{D}_{\mathrm{JS}}(p\|q)\quad\text{(Jensen--Shannon)},\\
f(u)&=\tfrac12|u-1| &&\Rightarrow&& \mathcal{D}_f=\mathcal{D}_{\mathrm{TV}}(p,q)\quad\text{(total variation)}.
\end{aligned}
\]
For clarity, the explicit forms are
\[
\mathcal{D}_{\mathrm{JS}}(p\|q)=\tfrac12 \mathcal{D}_{\mathrm{KL}}\big(p\| \tfrac12(p+q)\big)+\tfrac12 \mathcal{D}_{\mathrm{KL}}\big(q\| \tfrac12(p+q)\big),
\]
and
\[
\mathcal{D}_{\mathrm{TV}}(p,q)=\tfrac12 \int_{\mathbb{R}^D
} |p-q|\diff \rvx
=\sup_{A\subset \mathbb{R}^D } |p(A)-q(A)|.
\]
Intuitively, the JS divergence provides a smooth and symmetric measure that balances both distributions and avoids the unbounded penalties of KL (we will later see that it helps interpret the Generative Adversarial Network (GAN) framework), while the total variation distance captures the largest possible probability difference between the two.

A different viewpoint comes from \emph{optimal transport} (see \Cref{ch:ot-eot}), whose representative is the Wasserstein distance (see \Cref{eq: Monge's Formulation}). It measures the minimal cost of moving probability mass from one distribution to another. Unlike $f$-divergences, which compare density ratios, Wasserstein distances depend on the geometry of the sample space and remain meaningful even when the supports of $p$ and $q$ do not overlap.

Each divergence embodies a different notion of closeness between distributions and thus induces distinct learning behavior. We will revisit these divergences when they arise naturally in the context of generative modeling throughout this monograph.

\subsection{Challenges in Modeling Distributions}



To model a complex data distribution, we can parameterize the probability density function $p_{\mathrm{data}}$ using a neural network with parameters $\bm{\phi}$, creating a model we denote as $p_{\bm{\phi}}$. For $p_{\bm{\phi}}$ to be a valid probability density function, it must satisfy two fundamental properties:
\begin{enumerate}
    \item[(i)] \textbfs{Non-Negativity:} $p_{\bm{\phi}}(\rvx) \ge 0$ for all $\rvx$ in the domain.
    \item[(ii)] \textbfs{Normalization:} The integral over the entire domain must equal one, i.e., $\int p_{\bm{\phi}}(\rvx) \diff \rvx = 1$.
\end{enumerate}

A network can naturally produce a real scalar $E_{\bphi}(\rvx) \in \R$ for input $\rvx$. To interpret this output as a valid density, it must be transformed to satisfy conditions (i) and (ii).  
A practical alternative is to view $E_{\bphi}\colon\mathbb{R}^D\to\mathbb{R}$ as defining an \emph{unnormalized} density and then enforce these properties explicitly.

\paragraph{Step 1: Ensuring Non-Negativity.}
We can guarantee that our model's output is always non-negative by applying a positive function to the raw output of the neural network $E_{\bm{\phi}}(\rvx)$, such as $\abs{E_{\bm{\phi}}(\rvx)}$, $E^2_{\bm{\phi}}(\rvx)$. A standard and convenient choice is the exponential function. This gives us an unnormalized density, $\tilde{p}_{\bm{\phi}}(\rvx)$, that is guaranteed to be positive:
\[
\tilde{p}_{\bm{\phi}}(\rvx) = \exp(E_{\bm{\phi}}(\rvx)).
\]

\paragraph{Step 2: Enforcing Normalization.}
The function $\tilde{p}_{\bm{\phi}}(\rvx)$ is positive but does not integrate to one. To create a valid probability density, we must divide it by its integral over the entire space. This leads to the final form of our model:
\[
p_{\bm{\phi}}(\rvx) = \frac{\tilde{p}_{\bm{\phi}}(\rvx)}{\int \tilde{p}_{\bm{\phi}}(\rvx') \diff \rvx'} = \frac{\exp(E_{\bm{\phi}}(\rvx))}{\int \exp(E_{\bm{\phi}}(\rvx')) \diff \rvx'}.
\]
The denominator in this expression is known as the \emph{normalizing constant} or \emph{partition function}, denoted by $Z(\bm{\phi})$:
\[
Z(\bm{\phi}) := \int \exp(E_{\bm{\phi}}(\rvx')) \diff \rvx'.
\]
While this procedure provides a valid construction for $p_{\bm{\phi}}(\rvx)$, it introduces a major computational challenge. For most high-dimensional problems, the integral required to compute the normalizing constant $Z(\bm{\phi})$ is intractable. This intractability is a central problem that motivates the development of many different families of deep generative models.

In the following sections, we introduce several prominent approaches of DGM. Each is designed to circumvent or reduce the computational cost of evaluating this normalizing constant.
\newpage

\begin{figure}[th!]
\centering
\resizebox{\linewidth}{!}{%
\begin{tikzpicture}[
  ->, >=Stealth, thick, font=\normalsize,
  x=2.6cm, y=2.0cm,
  every node/.style={outer sep=0pt}
]
\tikzset{
  state/.style={
    rectangle, rounded corners=3pt,
    draw=black, fill=gray!10,
    minimum height=10mm, minimum width=14mm, 
    align=center, inner sep=2pt,
    font=\Large 
  },
  stategap/.style={ 
    rectangle, rounded corners=3pt,
    draw=none, fill=none,
    minimum height=10mm, minimum width=14mm,
    align=center, inner sep=2pt
  },
  processor/.style={
    trapezium, trapezium stretches=true,
    trapezium left angle=80, trapezium right angle=80,
    draw=black, fill=gray!20,
    minimum height=12mm, minimum width=28mm, align=center, inner sep=2pt
  },
  ovalval/.style={
    draw, ellipse, fill=gray!10,
    minimum height=7mm, minimum width=20mm, align=center
  },
  flowblock/.style={ 
    rectangle, rounded corners=3pt,
    draw=black, fill=gray!20,
    minimum height=12mm, minimum width=28mm, align=center, inner sep=3pt
  }
}

\coordinate (L) at (0,0);    
\coordinate (M) at (2.0,0);  
\coordinate (R) at (4.8,0);  
\coordinate (C) at ($ (L)!0.5!(R) $); 

\def\rowA{0}       
\def\rowB{-1.3}    
\def\rowC{-3.0}    
\def\rowD{-4.7}    
\def\rowE{-7.1}    
\def\rowF{-8.5}    

\node[anchor=east] at ($(L)+(-0.5,\rowA)$) {\textbfs{EBM}};
\node[anchor=east] at ($(L)+(-0.5,\rowB)$) {\textbfs{AR}};
\node[anchor=east] at ($(L)+(-0.5,\rowC)$) {\textbfs{VAE}};
\node[anchor=east] at ($(L)+(-0.5,\rowD)$) {\textbfs{NF}};
\node[anchor=east] at ($(L)+(-0.5,\rowE)$) {\textbfs{GAN}};
\node[anchor=east] at ($(L)+(-0.5,\rowF)$) {\textbfs{DM}};

\node[state]   (ebm-x)   at ($(L)+(0,\rowA)$) {$\rvx$};
\node[ovalval] (ebm-val) at ($(C)+(0,\rowA)$) {value};
\coordinate (midE) at ($ (ebm-x.east)!0.5!(ebm-val.west) $);
\node[processor, shape border rotate=270] (ebm-energy) at (midE)
  {\footnotesize\textbfs{Energy}\\[-2pt]\footnotesize $E_{\bm\phi}(\rvx)$};
\draw (ebm-x.east) -- (ebm-energy.west);
\draw (ebm-energy.east) -- (ebm-val.west);

\coordinate (Lb) at ($(L)+(0,\rowB)$);
\coordinate (Rb) at ($(R)+(0,\rowB)$);
\node[state]   (ar-x)    at ($ (Lb)!0/6!(Rb) $) {$\rvx_0$};
\node[state]   (ar-x0)   at ($ (Lb)!1/6!(Rb) $) {$\rvx_{1}$};
\node[state]   (ar-x1)   at ($ (Lb)!2/6!(Rb) $) {$\rvx_{2}$};
\node[state]   (ar-x2)   at ($ (Lb)!3/6!(Rb) $) {$\rvx_{3}$}; 
\node[stategap](ar-gap)  at ($ (Lb)!4/6!(Rb) $) {};
\node[state]   (ar-xLm1) at ($ (Lb)!5/6!(Rb) $) {$\rvx_{L-1}$};
\node[state]   (ar-xL)   at ($ (Lb)!6/6!(Rb) $) {$\rvx_{L}$};
\node at (ar-gap) {$\boldsymbol{\cdots}$};
\draw (ar-x.east)    -- (ar-x0.west);
\draw (ar-x0.east)   -- (ar-x1.west);
\draw (ar-x1.east)   -- (ar-x2.west);
\draw (ar-x2.east)   -- (ar-gap.west);
\draw (ar-gap.east)  -- (ar-xLm1.west);
\draw (ar-xLm1.east) -- (ar-xL.west);
\draw (ar-x.south)    to[out=-35, in=-145] (ar-x2.south);
\draw (ar-x0.south)   to[out=-40, in=-140] (ar-x2.south);
\draw (ar-gap.south)  to[out=-40, in=-140] (ar-xL.south);
\draw (ar-xLm1.south) to[out=-55, in=-125] (ar-xL.south);

\node[state] (vae-x)   at ($(L)+(0,\rowC)$) {$\rvx$};
\node[state] (vae-z)   at ($(C)+(0,\rowC)$) {$\rvz$};
\node[state] (vae-xp)  at ($(R)+(0,\rowC)$) {$\rvx'$};
\coordinate (midEnc) at ($ (vae-x.east)!0.5!(vae-z.west) $);
\node[processor, shape border rotate=270] (vae-enc) at (midEnc)
  {\footnotesize\textbfs{Encoder}\\[-2pt]\footnotesize $q_{\bm\theta}(\rvz|\rvx)$};
\coordinate (midDec) at ($ (vae-z.east)!0.5!(vae-xp.west) $);
\node[processor, shape border rotate=90] (vae-dec) at (midDec)
  {\footnotesize\textbfs{Decoder}\\[-2pt]\footnotesize $p_{\bm\phi}(\rvx|\rvz)$};
\draw (vae-x.east)   -- (vae-enc.west);
\draw (vae-enc.east) -- (vae-z.west);
\draw (vae-z.east)   -- (vae-dec.west);
\draw (vae-dec.east) -- (vae-xp.west);

\node[state] (nf-x)  at ($(L)+(0,\rowD)$) {$\rvx$};
\node[state] (nf-z)  at ($(C)+(0,\rowD)$) {$\rvz$};
\node[state] (nf-xp) at ($(R)+(0,\rowD)$) {$\rvx'$};
\coordinate (midFwd) at ($ (nf-x.east)!0.5!(nf-z.west) $);
\coordinate (midInv) at ($ (nf-z.east)!0.5!(nf-xp.west) $);
\node[flowblock] (nf-forward) at (midFwd)
  {\footnotesize\textbfs{Forward}\\[-1pt]\footnotesize $\rvf_{\bm\phi}(\rvx)$};
\node[flowblock] (nf-inverse) at (midInv)
  {\footnotesize\textbfs{Inverse}\\[-1pt]\footnotesize $\rvf_{\bm\phi}^{-1}(\rvz)$};
\draw (nf-x.east)       -- (nf-forward.west);
\draw (nf-forward.east) -- (nf-z.west);
\draw (nf-z.east)       -- (nf-inverse.west);
\draw (nf-inverse.east) -- (nf-xp.west);

\def\gansep{0.9} 
\node[state] (gan-z)  at ($(L)+(0,\rowE)$) {$\rvz$};
\node[state] (gan-xp) at ($(C)+(0,\rowE)$) {$\rvx'$};
\coordinate (midGen) at ($ (gan-z.east)!0.5!(gan-xp.west) $);
\node[processor, shape border rotate=90] (gan-gen) at (midGen)
  {\footnotesize\textbfs{Generator}\\[-2pt]\footnotesize $\rmG_{\bm\phi}(\rvz)$};
\node[state]   (gan-x)   at ($(C)+(0,\rowE+\gansep)$) {$\rvx$};
\node[ovalval] (gan-val) at ($(R)+(0,\rowE+\gansep)$) { 0/1};
\coordinate (midDisc) at ($ (gan-x.east)!0.5!(gan-val.west) $);
\node[processor, shape border rotate=270] (gan-disc) at (midDisc)
  {\footnotesize\textbfs{Discriminator}\\[-2pt]\footnotesize $D_{\bm\zeta}$};
\draw (gan-z.east)   -- (gan-gen.west);
\draw (gan-gen.east) -- (gan-xp.west);
\draw (gan-x.east)   -- (gan-disc.west);
\draw (gan-xp.east)  -- (gan-disc);
\draw (gan-disc.east) -- (gan-val.west);

\coordinate (Lf) at ($(L)+(0,\rowF)$);
\coordinate (Rf) at ($(R)+(0,\rowF)$);

\node[state]    (dm-x)     at ($ (Lf)!0/6!(Rf) $) {$\rvx_0$};
\node[state]    (dm-x0)    at ($ (Lf)!1/6!(Rf) $) {$\rvx_{1}$};
\node[state]    (dm-x1)    at ($ (Lf)!2/6!(Rf) $) {$\rvx_{2}$};
\node[state]    (dm-x2)    at ($ (Lf)!3/6!(Rf) $) {$\rvx_{3}$}; 
\node[stategap] (dm-gap)   at ($ (Lf)!4/6!(Rf) $) {};
\node[state]    (dm-xLm1)  at ($ (Lf)!5/6!(Rf) $) {$\rvx_{L-1}$};
\node[state]    (dm-xL)    at ($ (Lf)!6/6!(Rf) $) {$\rvx_{L}$};

\node at (dm-gap) {$\boldsymbol{\cdots}$};

\def\dmshift{4.5pt}

\draw[dashed] ([yshift=\dmshift]dm-x.east)     -- ([yshift=\dmshift]dm-x0.west);
\draw[dashed] ([yshift=\dmshift]dm-x0.east)    -- ([yshift=\dmshift]dm-x1.west);
\draw[dashed] ([yshift=\dmshift]dm-x1.east)    -- ([yshift=\dmshift]dm-x2.west);
\draw[dashed] ([yshift=\dmshift]dm-x2.east)    -- ([yshift=\dmshift]dm-gap.west);
\draw[dashed] ([yshift=\dmshift]dm-gap.east)   -- ([yshift=\dmshift]dm-xLm1.west);
\draw[dashed] ([yshift=\dmshift]dm-xLm1.east)  -- ([yshift=\dmshift]dm-xL.west);

\draw ([yshift=-\dmshift]dm-x0.west)  -- ([yshift=-\dmshift]dm-x.east);
\draw ([yshift=-\dmshift]dm-x1.west)  -- ([yshift=-\dmshift]dm-x0.east);
\draw ([yshift=-\dmshift]dm-x2.west)  -- ([yshift=-\dmshift]dm-x1.east);
\draw ([yshift=-\dmshift]dm-gap.west) -- ([yshift=-\dmshift]dm-x2.east);
\draw ([yshift=-\dmshift]dm-xLm1.west)-- ([yshift=-\dmshift]dm-gap.east);
\draw ([yshift=-\dmshift]dm-xL.west)  -- ([yshift=-\dmshift]dm-xLm1.east);

\end{tikzpicture}}%
\vspace{1.0cm}
\caption{\textbfs{Computation graphs of prominent deep generative models.} Top to bottom: \textbfs{EBM} maps an input $\rvx$ to a scalar energy; \textbfs{AR} generates a sequence $\{\rvx_\ell\}_{\ell=0}^{L}$ left to right with causal dependencies; \textbfs{VAE} encodes $\rvx$ to a latent $\rvz$ and decodes to a reconstruction $\rvx'$; \textbfs{NF} applies an invertible map $\rvf_\bphi$ between $\rvx$ and $\rvz$ and uses $\rvf_\bphi^{-1}$ to produce $\rvx'$; \textbfs{GAN} transforms noise $\rvz$ to a sample $\rvx'$ that is judged against real $\rvx$ by a discriminator $D_{\bm\zeta}$; \textbfs{DM} iteratively refines a noisy sample through a multi-step denoising chain $\{\rvx_\ell\}_{\ell=0}^{L}$. Boxes denote variables, trapezoids are learnable networks, ovals are scalars; arrows indicate computation flow. The trapezoid was intended to indicate a dimension-changing transformation, while rectangles denote variables or mappings that preserve dimensionality. \figcredit{Created by the authors.}}
\label{fig:ebm-ar-vae-nf-gan-dm-unified}
\end{figure}
\newpage

\section{Prominent Deep Generative Models}\label{sec:examples-dgm}

A central challenge in generative modeling is to learn expressive probabilistic models that can capture the rich and complex structure of high-dimensional data. Over the years, various modeling strategies have been developed, each making different trade-offs between tractability, expressiveness, and training efficiency. In this section, we explore some of the most influential strategies that have
shaped the field, accompanied by a comparison of their computation graphs in
\Cref{fig:ebm-ar-vae-nf-gan-dm-unified}.




\paragraph{Energy-Based Models (EBMs).}
EBMs~\citep{ackley1985learning,lecun2006tutorial} define a probability distribution through an energy function $ E_{\bm{\phi}}(\rvx) $ that assigns lower energy to more probable data points. The probability of a data point is defined as:
\begin{equation*}
    p_{\bm{\phi}}(\rvx) := \frac{1}{Z({\bm{\phi}})} \exp(-E_{\bm{\phi}}(\rvx)),
\end{equation*}
where 
\[Z({\bm{\phi}}) = \int \exp(-E_{\bm{\phi}}(\rvx)) \diff\rvx\] 
is the partition function. Training EBMs typically involves maximizing the log-likelihood of the data. However, this requires techniques to address the computational challenges arising from the intractability of the partition function. In the following chapter, we will explore how Diffusion Models offer an alternative by generating data from \emph{the gradient of the log density}, which does not depend on the normalizing constant, thereby circumventing the need for partition function computation.

\paragraph{Autoregressive Models.}
Deep autoregressive (AR) models~\citep{frey1995does,larochelle2011neural,uria2016neural}
build on a simple but powerful observation: any joint distribution can be decomposed into a sequence of conditional distributions using the \emph{chain rule of probability}. 
For a chosen ordering of the variables $\mathbf{x}=(x_1,\ldots,x_D)$, the data distribution satisfies
\begin{equation*}
    p_{\mathrm{data}}(\mathbf{x})
    =
    \prod_{i=1}^{D}
    p_{\mathrm{data}}(x_i | \mathbf{x}_{<i}),
    \qquad
    \mathbf{x}_{<i}=(x_1,\ldots,x_{i-1}).
\end{equation*}
Importantly, this factorization is an exact identity rather than a modeling assumption: any joint distribution admits such a decomposition for any chosen ordering of its variables.

The challenge is that the true conditional distributions
$p_{\mathrm{data}}(x_i|\mathbf{x}_{<i})$ are unknown.
AR models therefore parameterize a model with the same factorized form,
\begin{equation*}
    p_{\bm{\phi}}(\mathbf{x})
    =
    \prod_{i=1}^{D}
    p_{\bm{\phi}}(x_i|\mathbf{x}_{<i}),
\end{equation*}
and use a neural network, such as a Transformer, to approximate these conditional distributions from data.
For example, for discrete data the network may output a categorical distribution through a softmax, whereas for continuous data it may output the parameters of a tractable conditional density.
Because every conditional distribution is normalized, their product automatically defines a normalized joint distribution.

This construction also makes likelihood-based training particularly simple.
For a data sample $\mathbf{x}$, the log-likelihood decomposes as
\begin{equation*}
    \log p_{\bm{\phi}}(\mathbf{x})
    =
    \sum_{i=1}^{D}
    \log p_{\bm{\phi}}(x_i|\mathbf{x}_{<i}),
\end{equation*}
so the model can be trained directly by maximum likelihood, or equivalently by minimizing the negative log-likelihood.
During training, the observed preceding variables $\mathbf{x}_{<i}$ are available, so many of these conditional predictions can typically be evaluated in parallel.

Generation, however, remains inherently sequential:
\begin{equation*}
    x_1 \sim p_{\bm{\phi}}(x_1),
    \qquad
    x_2 \sim p_{\bm{\phi}}(x_2| x_1),
    \qquad
    \ldots,
    \qquad
    x_D \sim p_{\bm{\phi}}(x_D|\mathbf{x}_{<D}).
\end{equation*}
Thus, AR models provide tractable exact likelihoods and flexible density modeling, but sampling requires generating variables one after another and can therefore become computationally expensive for long sequences. AR models remain a foundational class of likelihood-based generative models and underlie many modern sequence and language models.

\paragraph{Variational Autoencoders (VAEs).}
VAEs~\citep{kingma2013auto} extend classical autoencoders by introducing latent variables $\rvz$ that capture hidden structure in the data $\rvx$. 
Instead of directly learning a mapping between $\rvx$ and $\rvz$, VAEs adopt a probabilistic view: they learn both an \emph{encoder}, $q_{\btheta}(\rvz|\rvx)$, which approximates the unknown distribution of latent variables given the data, and a \emph{decoder}, $p_{\bphi}(\rvx|\rvz)$, which reconstructs data from these latent variables. 
To make training feasible, VAEs maximize a tractable surrogate to the true log-likelihood, called the Evidence Lower Bound (ELBO):
\begin{equation*}
    \mathcal{L}_{\text{ELBO}}(\btheta, \bphi; \rvx) 
    = \mathbb{E}_{q_{\btheta}(\rvz | \rvx)} \left[ \log p_{\bphi}(\rvx | \rvz) \right] 
    - \mathcal D_{\mathrm{KL}} \left( q_{\btheta}(\rvz | \rvx) \,\|\, p_{\mathrm{prior}}(\rvz) \right).
\end{equation*}
Here, the first term encourages accurate reconstruction of the data, while the second regularizes the latent variables by keeping them close to a simple prior distribution $p_{\mathrm{prior}}(\rvz)$ (often Gaussian).  

VAEs provide a principled way to combine neural networks with latent-variable models and remain one of the most widely used likelihood-based approaches. 
However, they also face practical challenges, such as limited sample sharpness and training pathologies (e.g., the tendency of the decoder to ignore latent variables). 
Despite these limitations, VAEs laid important foundations for later advances, including diffusion models.


\paragraph{Normalizing Flows.}
Classic flow-based generative models include Normalizing Flows (NFs)~\citep{rezende2015variational}
and their continuous-time counterparts based on Neural Ordinary Differential Equations
(NODEs)~\citep{chen2018neural}, aim to learn a bijective mapping $\rvf_{\bm{\phi}}$ between a simple latent distribution $ \rvz $ and a complex data distribution $ \rvx $ via an invertible operator. This is achieved either through a sequence of bijective transformations (in NFs) or by modeling the transformation as an Ordinary Differential Equation (in NODEs). These models leverage the ``change-of-variable formula for densities'', enabling MLE training:
\begin{equation*}
    \log p_{\bm{\phi}}(\rvx) = \log p(\rvz) + \log \left| \det \frac{\partial \rvf_{\bm{\phi}}^{-1}(\rvx)}{\partial \rvx} \right|,
\end{equation*}
where $ \rvf_{\bm{\phi}} $ represents the invertible transformation mapping $ \rvz $ to $ \rvx $. NFs explicitly model normalized densities using invertible transformations with tractable Jacobian determinants. The normalization constant is absorbed analytically via the change-of-variables formula, making likelihood computation exact and tractable.

Despite their conceptual elegance, classic flow-based models often face practical limitations. For instance, NFs typically impose restrictive architectural constraints to ensure bijectivity, while NODEs may encounter training inefficiencies due to the computational overhead of solving ODEs. Both approaches face challenges when scaling to high-dimensional data. In later chapters, we will explore how Diffusion Models relate to and build upon these classic flow-based methods.










\paragraph{Generative Adversarial Networks (GANs).}
GANs~\citep{goodfellow2014generative} consist of two neural networks, a generator $ \rmG_{\bm{\phi}} $ and a discriminator $ D_{\bm{\zeta}} $, that compete against each other. The generator aims to create realistic samples $ \rmG_{\bm{\phi}}(\rvz) $ from random noise $ \rvz \sim p_{\mathrm{prior}}$, while the discriminator attempts to distinguish between real samples $ \rvx $ and generated samples $ \rmG_{\bm{\phi}}(\rvz) $. The objective function for GANs can be formulated as:
\begin{equation*}
    \min_{\rmG_{\bm{\phi}}} \max_{D_{\bm{\zeta}}} \underbrace{\mathbb{E}_{\rvx \sim p_{\mathrm{data}}(\rvx)}[\log D_{\bm{\zeta}}(\rvx)]}_{\text{real}} + \underbrace{\mathbb{E}_{\rvz \sim p_{\mathrm{prior}}(\rvz)}\left[\log(1 - D_{\bm{\zeta}}\left(\rmG_{\bm{\phi}}(\rvz))\right)\right]}_{\text{fake}}.
\end{equation*}
GANs do not define an explicit density function and therefore bypass likelihood estimation entirely. 
Instead of computing a normalization constant, they focus on generating samples that closely mimic the data distribution. 

From a divergence perspective, the discriminator implicitly measures the discrepancy between the true data distribution $p_{\mathrm{data}}$ 
and the generator distribution $p_{\rmG_{\bphi}}$, where $p_{\rmG_{\bphi}}$ denotes the distribution of generated samples $\rmG_{\bphi}(\rvz)$ obtained from noise $\rvz \sim p_{\mathrm{prior}}$. 
With an optimal discriminator for a fixed generator $\rmG_{\bphi}$ computed as
\begin{equation*}
     \frac{p_{\mathrm{data}}(\rvx)}{p_{\mathrm{data}}(\rvx) + p_{\rmG_{\bphi}}(\rvx)},
\end{equation*}
the generator’s minimization reduces to
\begin{equation*}
    \min_{\rmG_{\bphi}} \; 2\,\mathcal{D}_\mathrm{JS} \left(p_{\mathrm{data}} \,\|\, p_{\rmG_\bphi}\right) - \log 4.
\end{equation*}
Here, $\mathcal{D}_\mathrm{JS}$ denotes the Jensen–Shannon divergence, defined as
\begin{equation*}
    \mathcal{D}_\mathrm{JS}(p \,\|\, q) 
    := \tfrac{1}{2} \mathcal{D}_\mathrm{KL} \left(p \,\middle\|\, \tfrac{p+q}{2}\right) 
    + \tfrac{1}{2} \mathcal{D}_\mathrm{KL} \left(q \,\middle\|\, \tfrac{p+q}{2}\right).
\end{equation*}
This shows that GANs implicitly minimize $\mathcal{D}_\mathrm{JS}(p_{\mathrm{data}} \,\|\, p_{\rmG_\bphi})$. 
More broadly, extensions such as $f$-GANs~\citep{nowozin2016f} generalize this view by demonstrating that adversarial training can minimize a family of $f$-divergences, 
placing GANs within the same divergence-minimization framework as other generative models.

Although GANs are capable of generating high-quality data, their min-max training process is notoriously unstable, often requiring carefully designed architectures and engineering techniques to achieve satisfactory performance. However, GANs have since been revived as an auxiliary component to enhance other generative models, particularly Diffusion Models.

\newpage
\section{Taxonomy of Modeling}\label{sec:taxonomy}


As we have seen, DGMs span a wide spectrum of modeling strategies. A fundamental distinction lies in how these models \emph{parameterize} the underlying data distribution, that is, whether they specify $p_\bphi(\mathbf{x})$ \emph{explicitly} or only \emph{implicitly}, irrespective of the training objective.

\begin{itemize}
  \item \textbfs{Explicit Models:}  
  These models directly parameterize a probability distribution $p_\bphi(\mathbf{x})$ via a tractable or approximately tractable density or mass function. Examples include ARs, NFs, VAEs, and DMs, all of which define $p_\bphi(\mathbf{x})$ either exactly or through a tractable bound.

  \item \textbfs{Implicit Models:}  
  These models specify a distribution only through a sampling procedure, typically of the form $\mathbf{x} = \rmG_\bphi(\mathbf{z})$ for some noise variable $\mathbf{z} \sim p_{\mathrm{prior}}$. In this case, $p_\bphi(\mathbf{x})$ is not available in closed form and may not be defined at all.
\end{itemize}

The table in \Cref{tb:comparison-explicit-implicit} offers a concise summary of these contrasting approaches.

\begin{table}[th]
  \caption{Comparison of Explicit and Implicit Generative Models}
  \small
  \centering
  \resizebox{\textwidth}{!}{
  \begin{tabular}{cccc}
    \toprule
    & \multicolumn{2}{c}{\textbfs{Explicit}} & \textbfs{Implicit} \\
    \cmidrule(lr){2-3}
    & \textbfs{Exact Likelihood} & \textbfs{Approx.\ Likelihood} & \\
    \midrule
    \textbfs{Likelihood}
      & Tractable
      & Bound/Approx.
      & \makecell{Not Directly Modeled/\\Intractable} \\

    \textbfs{Objective}
      & MLE
      & ELBO
      & Adversarial \\

    \textbfs{Examples}
      & NFs, ARs
      & VAEs, DMs
      & GANs \\
    \bottomrule
  \end{tabular}
  }
  \label{tb:comparison-explicit-implicit}
\end{table}

\paragraph{Connection to Diffusion Models.}
Taken together, these classical families of DGMs illustrate complementary strategies for modeling complex distributions.  
Beyond their standalone importance, they also provide guiding principles for understanding diffusion models.  
Diffusion methods inherit ideas from several of these perspectives: they connect to VAEs through variational training objectives, to EBMs through score-matching approaches that learn gradients of the log-density (closely tied to energy functions), and to NFs through continuous-time transformations.  

To lay the groundwork for the diffusion methods discussed in later chapters, we will focus on three central paradigms: VAEs (\Cref{sec:vae}), EBMs (\Cref{sec:ebm}), and NFs (\Cref{sec:flow-based-method}).  
This exploration provides a foundation for the core principles that underlie modern diffusion-based generative modeling, which will be developed further in the chapters that follow.



\newpage
\section{Closing Remarks}\label{sec:ch1_cr}

This chapter has established the foundational concepts of deep generative modeling. We begin by defining the primary objective: to learn a tractable model distribution $p_{\text{model}}$ (parametrized by $\bm{\phi}$) that approximates an unknown, complex data distribution $p_{\text{data}}$. A central challenge is the computational intractability of the normalizing constant, or partition function $Z(\bm{\phi})$, which is required to define a valid probability density.

To circumvent this problem, various families of deep generative models have been developed, each employing a distinct strategy. We surveyed several prominent approaches, including Energy-Based Models (EBMs), Autoregressive Models (ARs), Variational Autoencoders (VAEs), Normalizing Flows (NFs), and Generative Adversarial Networks (GANs). These models can be broadly categorized into explicit models, which define a tractable density, and implicit models, which define a distribution only through a sampling procedure.

While each of these classical frameworks is significant, three in particular serve as the conceptual origins for the diffusion models that are the focus of this monograph: VAEs, EBMs, and NFs. In the chapters that follow, we will trace the evolution of diffusion models from these three foundational paradigms:
\begin{enumerate}
    \item Part B will begin by exploring the variational perspective (\Cref{ch:variational}), showing how (the hierarchical latent variable structure of) VAEs leads naturally to the formulation of Denoising Diffusion Probabilistic Models (DDPMs).
    \item Next, we will examine the score-based perspective (\Cref{ch:score-based}), which originates from EBMs and score matching, and develops into Noise Conditional Score Networks (NCSN) and the more general Score SDE framework (\Cref{ch:score-sde}).
    \item Finally, we will investigate the flow-based perspective (\Cref{ch:flow-based}), which builds upon the principles of Normalizing Flows to frame generation as a continuous transformation, generalized by the concept of Flow Matching.
\end{enumerate}

By understanding these origins, we will build a coherent framework for interpreting the diverse formulations of diffusion models and uncovering the deep principles that unify them.

\part{Foundations and Perspectives of Diffusion Models}
\vspace*{\fill}  

\begin{figure}[ht!]
\begin{center}
\resizebox{\textwidth}{!}{%
\begin{tikzpicture}[
  box/.style = {draw, rounded corners, minimum width=3.6cm, minimum height=1.5cm, align=center},
  model/.style = {draw, fill=gray!10, minimum width=3.5cm, minimum height=1cm, align=center},
  arrow/.style = {ultra thick, -{Stealth[length=4mm, width=4mm]}}
  ]

\node[box] (vae) {Variational \\Autoencoder};
\node[box, right=1.2cm of vae] (ebm) {Energy-Based Model};
\node[box, right=1.2cm of ebm] (flow) {Normalizing Flows};

\node[draw, rounded corners, minimum width=13.6cm, minimum height=1.0cm, thick, above=3.5cm of ebm, label=above:{\parbox{5cm}{ \centering {Chapter}~\ref{ch:overview-dgm} 
 }}] (dgm) {Overview of Deep Generative Modeling};
 
\node[model, above=1.5cm of vae] (variational-based) {\large\textbfs{Variational View}};
\node[model, above=1.5cm of ebm] (score-based) {\large\textbfs{Score-Based View}};
\node[model, above=1.5cm of flow] (flow-based) {\large\textbfs{Flow-Based View}};

\node[box, below=2.2cm of vae] (ddpm) {Denoising Diffusion\\Probabilistic Model\\ (DDPM)};
\node[box, below=2.2cm of ebm] (score) {Noise Conditional\\Score Network \\(NCSN)};
\node[box, below=2.2cm of flow] (gfm) {Gaussian\\Flow Matching};

\draw[arrow] (vae.south) -- node[midway, anchor=center, yshift=3.5pt, xshift=-8pt] {\rotatebox{90}{\small {Chapter~\ref{ch:variational}}}} (ddpm.north);

\draw[arrow] (ebm.south) -- node[midway, anchor=center, yshift=3.5pt, xshift=-8pt] {\rotatebox{90}{\small {Chapter}~\ref{ch:score-based}}} (score.north);

\draw[arrow] (flow.south) -- node[midway, anchor=center, yshift=3.5pt, xshift=-8pt] {\rotatebox{90}{\small {Chapter}~\ref{ch:flow-based}}} (gfm.north);

\node[box, below=2cm of score] (scoresde) { Continuous-Time Formulation \\ (e.g., Score SDE) \\ \small {Chapter}~\ref{ch:score-sde}};

\node[draw, rounded corners, thick, fit={(ddpm) (score) (gfm) (scoresde)}, inner sep=0.3cm, label=below:{\parbox{5cm}{\centering\textbfs{\\ \large Unifying Principles}  \\ \vspace{0.2cm}\small {Chapter}~\ref{ch:all-equivalent} 
\begin{itemize}
    \item Conditional Strategy
    \item Fokker-Planck Equation
\end{itemize} }}] (box) {};

\draw[ultra thick, -{Stealth[length=4mm, width=4mm]}, rounded corners=4pt] (ddpm.south) -- ++(0,-0.4) -- ([xshift=-1.0cm]scoresde.north);
\draw[ultra thick, -{Stealth[length=4mm, width=4mm]}, rounded corners=4pt] (score.south) -- ++(0,-0.4) -- (scoresde.north);
\draw[ultra thick, -{Stealth[length=4mm, width=4mm]}, rounded corners=4pt] (gfm.south) -- ++(0,-0.4) -- ([xshift=1.0cm]scoresde.north);

\node[align=center, left=0.5cm of box, yshift=8.3cm] {\rotatebox{90}{\textbfs{\large{Perspective}}}};
\node[align=center, left=0.5cm of box, yshift=5.5cm] {\rotatebox{90}{\textbfs{\large{Origin}}}};
\node[align=center, left=0.5cm of box] {\rotatebox{90}{\textbfs{\large{Diffusion Model}}}};
\end{tikzpicture}
    }
\end{center}
\end{figure}

\newpage
\section*{A First Operational View of Diffusion Model Training and Sampling}

Before diving into the mathematics of diffusion models, let us first develop
a simple operational view of how they are trained and how they generate
samples.

The basic construction has two parts. First, we define a known corruption
rule that can turn clean data into noisy observations at any chosen noise
level. Second, we train a neural network to infer the clean signal from such
a noisy observation. Once this prediction rule has been learned across many noise levels, it can
also be used for generation. We start from a fresh noise sample and repeatedly
apply the network while moving from high noise levels to lower ones. The
result is a new sample that gradually acquires the structure of the data
distribution.

\msgu{Message}{}{
Create noisy data at many noise levels, learn to predict the clean signal,
and generate by repeatedly applying this prediction from high noise to low.
}
\paragraph{Creating Noisy Data.}

Let us make precise what a noisy version of the data actually means. Take
a a clean data sample
\[
\rvx_0\sim p_{\mathrm{data}},
\]
such as an image or any other kind of data we care about. We corrupt it at
noise level $t$ by mixing in independent Gaussian noise\footnote{
The formula
$\rvx_t=\alpha_t\rvx_0+\sigma_t\beps$
gives the accumulated corruption at a chosen noise level $t$ directly.
In DDPMs and Score SDEs, the same marginal distribution at that level can
also be obtained by adding noise through many smaller sequential steps; see
\Cref{sec:ddpm} and \Cref{ch:score-sde}. In the discrete DDPM notation introduced later,
this accumulated corruption is written as
\[
\rvx_i
=
\bar{\alpha}_i\rvx_0
+
\sqrt{1-\bar{\alpha}_i^2}\,\beps.
\]
}:
\[
\rvx_t
=
\alpha_t\rvx_0+\sigma_t\beps,
\qquad
\beps\sim\mathcal{N}(\bm{0},\rmI).
\]
Think of $\alpha_t$ and $\sigma_t$ as two knobs, fixed in advance, that
set how much clean signal and how much noise enter the mix at noise level
$t$.

The simplest possible choice, $\alpha_t=1$ and $\sigma_t=t$, gives
\[
\rvx_t=\rvx_0+t\beps.
\]
Now $t$ is literally a noise dial: turn it toward zero and $\rvx_t$ is
nearly the clean sample; turn it up and the noise term $t\beps$
increasingly takes over.

\begin{figure}[!ht]
\centering
\includegraphics[width=0.6\linewidth]{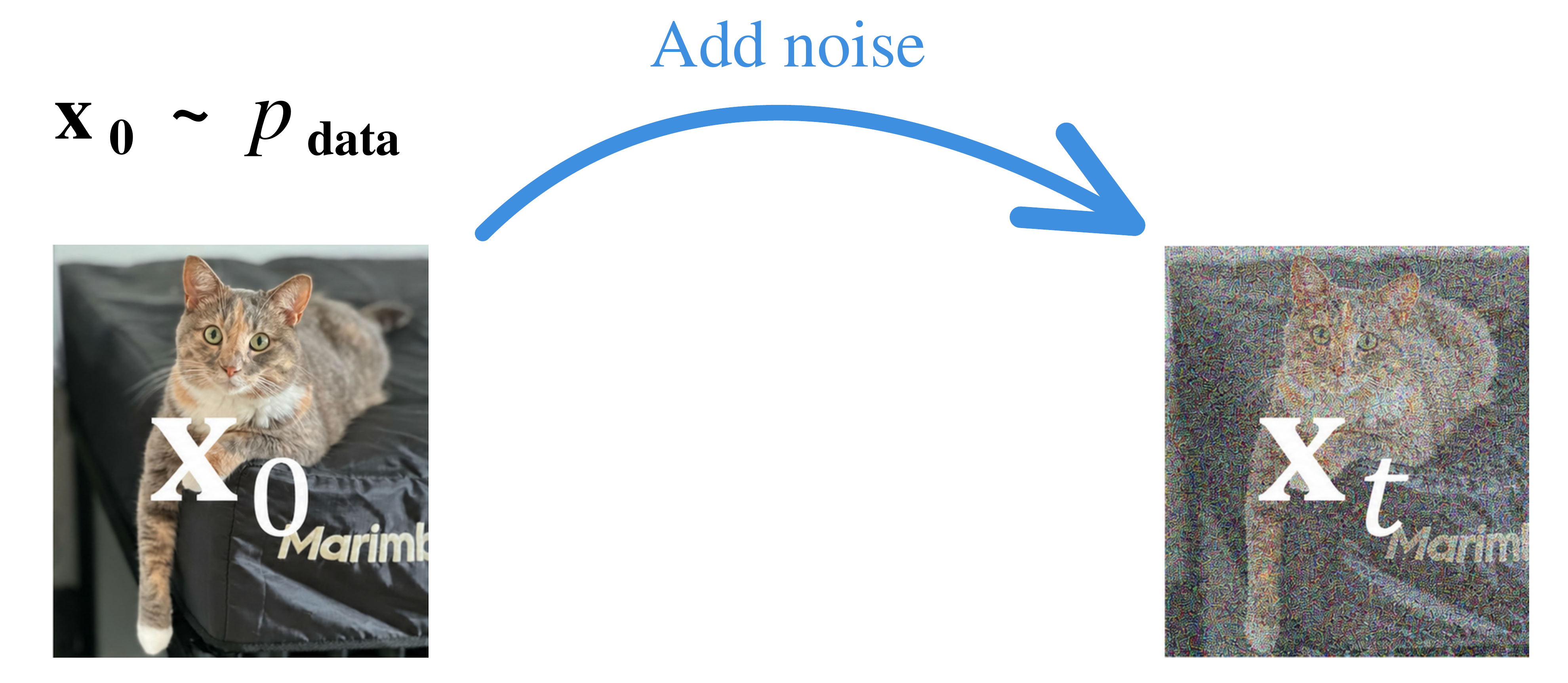}
\caption{\textbfs{Forward diffusion.} A clean sample $\mathbf{x}_0$ is
corrupted to produce a noisy observation $\mathbf{x}_t$ at noise level $t$.
\figcredit{Created by the authors.}}
\label{fig:ill-forward}
\end{figure}

More generally, every noise level $t$ defines a distribution $p_t$ over
noisy samples. It is useful to think of $p_t$ as a \emph{snapshot}: if we
generated many noisy samples using the same value of $t$, their distribution
would be $p_t$.

The schedule $(\alpha_t,\sigma_t)$ is chosen so that these snapshots move
from clean data toward a simple noise distribution:
\[
p_0=p_{\mathrm{data}}
\qquad\longrightarrow\qquad
p_t
\qquad\longrightarrow\qquad
p_T\approx p_{\mathrm{prior}},
\]
where $p_{\mathrm{prior}}$ is typically a Gaussian distribution that is easy
to sample from. Thus, the family
\[
\{p_t\}_{t\in[0,T]}
\]
describes how the distribution gradually changes as the noise level
increases.

\paragraph{Training: Learn to Predict the Clean Signal.}

With a way to corrupt data now in hand, training becomes almost
mechanical, and it hinges on one key trick: because we are the ones
corrupting the data, we always know the right answer. To build a single
training example, we sample a clean datum $\rvx_0$, choose a noise level
$t$, draw noise $\beps$, and form
\[
\rvx_t
=
\alpha_t\rvx_0+\sigma_t\beps.
\]
We manufactured $\rvx_t$ ourselves, so we already know exactly which
clean $\rvx_0$ produced it.

The network, however, never sees $\rvx_0$ or $\beps$ directly. It sees
only the pair $(\rvx_t,t)$, the corrupted observation and how corrupted
it is. Its task is to guess the clean data behind it:
\[
\rvx_0^{\mathrm{pred}}
:=
\rvx_{\bphi}(\rvx_t,t).
\]
Because we know the true target, we can score this guess and improve it
using an ordinary squared-error loss:
\[
\left\|
\rvx_0^{\mathrm{pred}}-\rvx_0
\right\|_2^2.
\]

\msgu{Message}{}{
In a nutshell: create a noisy observation with a known clean target, hide
that target from the network, and train it to predict the target back.
}

Repeat this many times, over different data samples, noise draws, and
noise levels, and a single network learns to make clean-data predictions
across the entire range of corruption, from barely touched to nearly
pure noise.

It is worth pausing on what this loss teaches the network to predict.
Under squared error, the best possible prediction is
\[
\rvx_0^*(\rvx_t,t)
=
\E[\rvx_0|\rvx_t,t].
\]
A noisy observation generally does not correspond to one unique clean sample.
Several different clean samples could have produced a similar $\rvx_t$.
This ambiguity exists at any nonzero noise level, but it becomes much stronger
when the noise level is high. The network therefore learns to average over
the clean samples that are plausible given the current noisy observation
\footnote{When the noise level is clear from context, we will often shorten
this to $\E[\rvx_0|\rvx_t]$.}.

\paragraph{Sampling: Start from Fresh Noise.}

Training leaves us with a network that is good at guessing clean data
from noisy data. The next question is how to put that skill to use for
generating something new.

Once training is finished, we freeze the network's parameters and call
them $\bphi^\times$. Generation now runs in the direction opposite to the
one we have been describing: rather than starting at data and adding
noise, it starts from noise alone,
\[
\tilde{\rvx}_T\sim p_{\mathrm{prior}}.
\]
We use a tilde to mark these generated states, $\tilde{\rvx}_t$, and keep
them visually distinct from the training variables $\rvx_t$ produced by
forward corruption.

This distinction matters more than it might first appear. The state
$\tilde{\rvx}_T$ was not produced by secretly corrupting some real,
hidden sample: there is no true $\rvx_0$ behind it waiting to be
recovered. Generation is therefore not ``rewinding a recording'' of some
specific corruption, because no such recording exists. Instead, at every
noise level we simply ask the trained network what it currently believes
the clean signal looks like, and use that belief to take one step back
toward data.

Concretely, suppose we currently hold the state $\tilde{\rvx}_t$. Feeding
it to the network gives a clean-data guess,
\[
\rvx_0^{\mathrm{pred}}
:=
\rvx_{\bphi^\times}(\tilde{\rvx}_t,t).
\]
For $t>0$, we can also compute an implied noise residual, defined as
whatever is left over once the predicted signal is removed from the
current state:
\[
\beps_t^{\mathrm{pred}}
:=
\frac{
\tilde{\rvx}_t-\alpha_t\rvx_0^{\mathrm{pred}}
}{
\sigma_t
}.
\]
These two quantities are not independent guesses: by construction, they
exactly reassemble the current state,
\[
\tilde{\rvx}_t
=
\alpha_t\rvx_0^{\mathrm{pred}}
+
\sigma_t\beps_t^{\mathrm{pred}}.
\]

Here is the idea behind a single sampling step: keep the predicted
signal and predicted noise as they are, but re-mix them using the
coefficients for a slightly lower noise level $t-\Delta t$ in place of
$t$:
\[
\tilde{\rvx}_{t-\Delta t}
=
\alpha_{t-\Delta t}\rvx_0^{\mathrm{pred}}
+
\sigma_{t-\Delta t}\beps_t^{\mathrm{pred}}.
\]
Equivalently, this is a small update on top of the current state,
\begin{equation}\label{eq:ddpm-first-try}
\tilde{\rvx}_{t-\Delta t}
=
\tilde{\rvx}_t
+
(\alpha_{t-\Delta t}-\alpha_t)\rvx_0^{\mathrm{pred}}
+
(\sigma_{t-\Delta t}-\sigma_t)\beps_t^{\mathrm{pred}}.
\end{equation}

That is one step. We now feed the new state $\tilde{\rvx}_{t-\Delta t}$
back into the network, obtain an updated, usually different, clean-data
prediction, and take another step down in noise. Sampling is exactly
this loop, repeated from $t=T$ down to $t=0$.

\msgu{Message}{}{
In a nutshell: start from fresh noise, predict the clean signal, step to
a lower noise level, and predict again.
}

\begin{figure}[th!]
\vspace{-0.5cm}
\centering
\includegraphics[width=\linewidth]{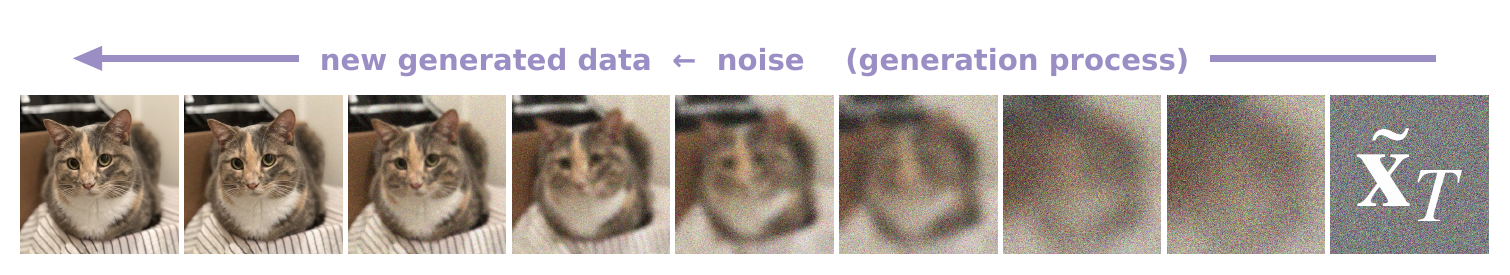}
\caption{\textbfs{Diffusion generation.} Starting from fresh noise
$\tilde{\mathbf{x}}_T\sim p_{\mathrm{prior}}$, repeated model predictions
construct a new sample as the noise level decreases.
\figcredit{Created by the authors.}}
\label{fig:ill-reverse}
\end{figure}

A natural question at this point: why not simply output that very first
prediction $\rvx_0^{\mathrm{pred}}$ and stop? Because at high noise
levels the current state carries very little information about clean
structure, so the prediction is little more than a rough average over
many plausible samples. Moving to a slightly lower noise level gives the
network a slightly less ambiguous state to work from, so its next
prediction can sharpen. Repeating this many times is what lets a
genuinely detailed sample gradually emerge\footnote{We will later connect
this deterministic update to DDIM sampling and its Euler-type
interpretation~\citep{song2020denoising} in \Cref{sec:revisit-ddim}; the
stochastic DDPM sampler is introduced in \Cref{sec:ddpm}.}.

\paragraph{A Simple Example.}

Let us make this tangible using our earlier running example where $\alpha_t=1$ and $\sigma_t=t$. Substituting into the update above gives
\[
\tilde{\rvx}_{t-\Delta t}
=
\rvx_0^{\mathrm{pred}}
+
(t-\Delta t)
\frac{
\tilde{\rvx}_t-\rvx_0^{\mathrm{pred}}
}{
t
},
\]
which rearranges into a simple weighted average,
\[
\tilde{\rvx}_{t-\Delta t}
=
\frac{t-\Delta t}{t}\tilde{\rvx}_t
+
\frac{\Delta t}{t}\rvx_0^{\mathrm{pred}}.
\]
In words, for small $\Delta t$ the next state is simply a blend of where
we currently stand and where the network thinks we should end up.

So each step nudges the sample toward whatever the network currently
predicts to be clean. The word \emph{currently} is doing real work here:
once we move, the network sees a new, less noisy state and typically
changes its prediction. The final sample is therefore not the output of
a single clean-data guess. It is built from an entire sequence of
guesses, each one revised in light of the last.

\paragraph{How this Connects to the Next Chapters?}

The clean-data prediction template developed here is only a first operational
view. The basic pattern will remain the same throughout the book: during
training, we create noisy observations from clean data and learn how to
predict useful information from them; during generation, we start from fresh
noise and use these learned predictions to construct a new sample over
decreasing noise levels.

The following chapters explain this same process from different mathematical
perspectives. \Cref{ch:variational} connects DDPMs to latent-variable models,
where the fixed noising process plays the role of an encoder and the learned
reverse transitions play the role of a decoder.
\Cref{ch:score-based} describes denoising through the score, which provides a
direction for updating noisy samples.
\Cref{ch:score-sde} moves from discrete noise levels to continuous time and
describes generation through reverse-time stochastic dynamics.
\Cref{ch:flow-based} instead views generation as transporting samples along a
continuous path from a simple prior toward the data distribution.
Finally, \Cref{ch:all-equivalent} explains how these viewpoints are connected.

The main takeaway is simple: diffusion models learn from noisy versions of
data at many noise levels, and generation uses these learned predictions
repeatedly to turn fresh noise into a new sample.
\chapter{Variational Perspective: From VAEs to DDPMs}\label{ch:variational}

In this chapter we view diffusion models through a variational lens. We begin with the Variational Autoencoders (VAEs), which represent data with latent variables and are trained by maximizing a tractable lower bound on the log likelihood. In this setting a learned encoder maps observations to latents, and a learned decoder maps latents back to observations, closing the modeling loop. 

Building on this pattern, hierarchical variants (Hierarchical VAEs) stack several latent layers to capture structure at multiple scales. With this setup, Denoising Diffusion Probabilistic Models (DDPM) conceptually follow the similar template: instead of jointly training both the encoder and decoder, the encoder is fixed as a forward noising process that gradually maps data to noise, and training learns a decoder that reverses this path in successive denoising steps. In this view, VAEs, hierarchical VAEs, and diffusion models all optimize a likelihood surrogate defined by a variational bound, providing a common foundation for the methods introduced here.

\chapterlearning{
  \item understand why VAEs optimize the ELBO when exact maximum-likelihood training is difficult;
  \item see how DDPM extends the latent-variable view of VAEs into many noisy layers, with a fixed noising process and a learned reverse process;
  \item understand the conditioning trick that turns the difficult reverse-denoising problem into a tractable training target;
  \item compare mean, clean-data, and noise prediction, and see how they represent the same underlying denoising information;
  \item connect the practical DDPM denoising loss back to its variational foundation through the ELBO.
}{
Readers already familiar with VAEs and ELBO training may skim \Cref{sec:vae} and begin with \Cref{sec:ddpm}.
}

\newpage
\section{Variational Autoencoder}\label{sec:vae}
How can a neural network learn to generate realistic data?  
A natural starting point is the \emph{autoencoder}, which consists of two networks:  
a deterministic \emph{encoder} that compresses an input to a low-dimensional latent code, and a deterministic \emph{decoder} that reconstructs the input from this code.  
Training minimizes the reconstruction error between the original input and its reconstruction.  
While this setup enables accurate reconstruction, the latent space is unstructured: randomly sampling latent codes usually produces meaningless outputs, limiting the model’s use for generation.


The \emph{Variational Autoencoder (VAE)} \citep{kingma2013auto}  solves this by imposing a probabilistic structure on the latent space. This transforms the model from a simple reconstruction tool into a true generative model, capable of producing novel and realistic data.

\subsection{Probabilistic Encoder and Decoder}

\begin{figure}[tbh!]
\centering
\resizebox{0.85\linewidth}{!}{%
\begin{tikzpicture}[>=Latex, thick, font=\sffamily, node distance=0.8cm]

\tikzset{
  state/.style={
    rectangle, rounded corners=3pt,
    draw=black, fill=gray!10,
    minimum height=45pt, minimum width=25pt, align=center,
    outer sep=0pt 
  },
  latent/.style={
    rectangle, rounded corners=3pt,
    draw=black, fill=gray!10,
    minimum height=28pt, minimum width=28pt, align=center,
    outer sep=0pt
  },
  encoderblock/.style={
    trapezium, trapezium stretches=true,
    trapezium left angle=80,  
    trapezium right angle=80, 
    shape border rotate=270,   
    draw=black, fill=gray!20,
    minimum height=8pt, minimum width=15pt, inner sep=4pt, align=center
  },
  decoderblock/.style={
    trapezium, trapezium stretches=true,
    trapezium left angle=80,
    trapezium right angle=80,
    shape border rotate=90,  
    draw=black, fill=gray!20,
    minimum height=8pt, minimum width=15pt, inner sep=4pt, align=center
  }
}

\node[state] (x) {$\mathbf{x}$};
\node[encoderblock, right=of x] (enc) {\footnotesize\textbf{Encoder}\\[-2pt]
 \footnotesize $q_{\bm{\theta}}(\mathbf{z}|\mathbf{x})$}; 
\node[latent, right=of enc] (z) {$\mathbf{z}$};
\node[decoderblock, right=of z] (dec) {\footnotesize\textbf{Decoder}\\[-2pt]
  \footnotesize$p_{\bm{\phi}}(\mathbf{x}|\mathbf{z})$}; 
\node[state, right=of dec] (xprime) {$\mathbf{x}'$};

\draw[->] (x) -- (enc); 
\draw[->] (enc) -- (z);
\draw[->] (z) -- (dec);
\draw[->] (dec) -- (xprime);

\end{tikzpicture}}
\caption{\textbfs{Illustration of a VAE.} It consists of a stochastic encoder $q_{\bm{\theta}}(\rvz|\rvx)$ that maps data $\rvx$ to a latent variable $\rvz$, and a decoder $p_{\bm{\phi}}(\rvx|\rvz)$ that reconstructs data from the latent. \figcredit{Created by the authors.}}
\label{fig:vae-graph}
\end{figure}

\paragraph{Construction of Decoder (Generator).}



In VAEs, we distinguish between two types of variables: 
\emph{observed variables} $\rvx$, which correspond to the data we see (e.g., an image), 
and \emph{latent variables} $\rvz$, which capture the hidden factors of variation 
(e.g., object shape, color, or style). 
The model assumes that each observation $\rvx$ is generated from a latent variable 
sampled from a simple \emph{prior distribution}, typically a standard Gaussian, 
$\rvz \sim p_{\text{prior}} := \mathcal{N}(\mathbf{0}, \mathbf{I})$.

To map $\rvz$ back to data space, we define a \emph{decoder (generator)} distribution 
$p_{\bm{\phi}}(\rvx|\rvz)$. In practice, this decoder is kept simple, often a 
factorized Gaussian (see \Cref{subsec:optimization-vae}) or similar distribution, so that learning focuses on extracting 
useful latent features rather than memorizing data. 
Intuitively, directly generating pixels one by one is extremely hard; instead, the 
latent variable provides a compact representation, 
from which decoding the exact pixel arrangement becomes much easier. 
New samples are drawn by first sampling $\rvz \sim p_{\text{prior}}$ and then 
decoding via $\rvx \sim p_{\bm{\phi}}(\rvx|\rvz)$.

The VAE thereby defines a latent-variable generative model through the marginal likelihood:
\[
p_{\bm{\phi}}(\rvx) = \int p_{\bm{\phi}}(\rvx|\rvz) p(\rvz) \diff \rvz.
\]
Ideally, the decoder parameters $\bm{\phi}$ are learned by maximizing this marginal 
likelihood, as in maximum likelihood estimation (see \Cref{eq:MLE}). 
However, because the integral over $\rvz$ is intractable for expressive, non-linear 
decoders, direct MLE is computationally infeasible, motivating the variational 
approach used in VAEs.


\paragraph{Construction of Encoder (Inference Network).}


To connect our intractable generator to real data, consider the reverse question:  
given an observation $\rvx$, what latent codes $\rvz$ could have produced it?  
By Bayes’ rule, the posterior distribution is
\[
p_{\bm{\phi}}(\rvz|\rvx) \;=\; \frac{p_{\bm{\phi}}(\rvx|\rvz) p(\rvz)}{p_{\bm{\phi}}(\rvx)}.
\]
The difficulty is that the denominator involves the marginal likelihood $p_{\bm{\phi}}(\rvx)$,  
which requires integrating over all latent variables and is intractable for nonlinear decoders.  
Thus, exact inference of $\rvz$ from $\rvx$ is computationally prohibitive.

The ``variational'' step in VAEs addresses this by replacing the intractable posterior with a tractable approximation.  
We introduce an \emph{encoder} (or inference network) $q_{\bm{\theta}}(\rvz|\rvx)$, parameterized by a neural network,  
whose role is to serve as a learnable proxy:
\[
q_{\bm{\theta}}(\rvz|\rvx) \;\approx\; p_{\bm{\phi}}(\rvz|\rvx).
\]
In practice, the encoder maps each observed data point $\rvx$ to a distribution over latent codes,  
providing a feasible and trainable pathway from $\rvx$ back to $\rvz$ that enables learning.

\subsection{Training via the Evidence Lower Bound (ELBO)}

We now define a computable training objective. While we cannot directly optimize $\log p_{\bm{\phi}}(\mathbf{x})$, we can maximize a lower bound on it—the \emph{Evidence Lower Bound (ELBO)}:

\thmp{Evidence Lower Bound (ELBO)}{elbo}{
For any data point $\mathbf{x}$, the log-likelihood satisfies:
\[
\log p_{\bm{\phi}}(\mathbf{x}) \geq \mathcal{L}_{\text{ELBO}}(\bm{\theta}, \bm{\phi}; \mathbf{x}),
\]
where the ELBO is given by: 
\begin{align}\label{eq:vae-elbo}
    \mathcal{L}_{\text{ELBO}} = \underbrace{\mathbb{E}_{\mathbf{z} \sim q_{\bm{\theta}}(\mathbf{z}|\mathbf{x})} \left[ \log p_{\bm{\phi}}(\mathbf{x}|\mathbf{z}) \right]}_{\text{Reconstruction Term}} - \underbrace{\mathcal{D}_{\mathrm{KL}}\left( q_{\bm{\theta}}(\mathbf{z}|\mathbf{x}) \| p(\mathbf{z}) \right)}_{\text{Latent Regularization}}.
\end{align}
}{
The ELBO arises from Jensen’s inequality:
\begin{align*}
\log p_{\bm{\phi}}(\mathbf{x}) &= \log \int p_{\bm{\phi}}(\mathbf{x}, \mathbf{z})   \mathrm{d}\mathbf{z} = \log \int q_{\bm{\theta}}(\mathbf{z}|\mathbf{x}) \frac{p_{\bm{\phi}}(\mathbf{x}, \mathbf{z})}{q_{\bm{\theta}}(\mathbf{z}|\mathbf{x})}   \mathrm{d}\mathbf{z} \\
&= \log \mathbb{E}_{\mathbf{z} \sim q_{\bm{\theta}}(\mathbf{z}|\mathbf{x})} \left[ \frac{p_{\bm{\phi}}(\mathbf{x}, \mathbf{z})}{q_{\bm{\theta}}(\mathbf{z}|\mathbf{x})} \right] \geq \mathbb{E}_{\mathbf{z} \sim q_{\bm{\theta}}(\mathbf{z}|\mathbf{x})} \left[ \log \frac{p_{\bm{\phi}}(\mathbf{x}, \mathbf{z})}{q_{\bm{\theta}}(\mathbf{z}|\mathbf{x})} \right].
\end{align*}
}
The ELBO objective naturally decomposes into two parts:
\begin{itemize}
    \item \textbfs{Reconstruction:} Encourages accurate recovery of $\mathbf{x}$ from its latent code $\mathbf{z}$. With Gaussian encoder and decoder assumptions, this term reduces exactly to the familiar reconstruction loss of an autoencoder (cf.\ \Cref{subsec:optimization-vae}). However, as in autoencoders, optimizing this term alone risks memorizing the training data, motivating an additional regularization.
    \item \textbfs{Latent KL:} Encourages the encoder distribution $q_{\bm{\theta}}(\mathbf{z}|\mathbf{x})$ to stay close to a simple Gaussian prior $p_{\mathrm{prior}}(\mathbf{z})$. This regularization shapes the latent space into a smooth and continuous structure, enabling meaningful generation by ensuring that samples drawn from the prior can be reliably decoded.
\end{itemize}
This trade-off ensures both faithful reconstructions and coherent sampling.

\paragraph{Information-Theoretic View: ELBO as a Divergence Bound.}

The ELBO objective has a natural information-theoretic interpretation. 
Recall that maximum likelihood training amounts to minimizing the KL divergence
\[
\mathcal{D}_{\mathrm{KL}}(p_{\text{data}}(\rvx) \| p_{\bm{\phi}}(\rvx)),
\]
which measures how well the model distribution approximates the data distribution. 
Since this term is intractable in general, the variational framework introduces a joint comparison.

Specifically, consider two joint distributions:
\begin{itemize}
    \item The \textbfs{generative joint}, $p_{\bm{\phi}}(\rvx, \rvz) = p(\rvz) p_{\bm{\phi}}(\rvx|\rvz)$, which describes how the model generates data;
    \item The \textbfs{inference joint}, $q_{\bm{\theta}}(\rvx, \rvz) = p_{\text{data}}(\rvx) q_{\bm{\theta}}(\rvz|\rvx)$, which couples real data with its inferred latent.
\end{itemize}
Comparing these distributions yields the inequality
\begin{align}\label{eq:joint-bound}
    \mathcal{D}_{\mathrm{KL}}(p_{\text{data}}(\rvx) \| p_{\bm{\phi}}(\rvx)) 
    \leq \mathcal{D}_{\mathrm{KL}}(q_{\bm{\theta}}(\rvx,\rvz) \| p_{\bm{\phi}}(\rvx,\rvz)),
\end{align}
sometimes referred to as the chain rule for KL divergence. 
Intuitively, comparing only marginals ($\rvx$) can hide mismatches that are revealed when the full latent–data joint is considered.

Formally, one can expand the joint KL as
\begin{align*}
    &\underbrace{\mathcal{D}_{\mathrm{KL}}(q_{\bm{\theta}}(\mathbf{x}, \mathbf{z}) \| p_{\bm{\phi}}(\mathbf{x}, \mathbf{z}))}_{\text{Total Error Bound}} 
    \\=& \mathbb{E}_{q_{\bm{\theta}}(\mathbf{x}, \mathbf{z})} \left[ \log \frac{p_{\text{data}}(\mathbf{x}) q_{\bm{\theta}}(\mathbf{z}|\mathbf{x})}{p_{\bm{\phi}}(\mathbf{x}) p_{\bm{\phi}}(\mathbf{z}|\mathbf{x})} \right] \\=&\mathbb{E}_{p_{\text{data}}(\mathbf{x})} \left[ \log \frac{p_{\text{data}}(\mathbf{x})}{p_{\bm{\phi}}(\mathbf{x})} + \mathcal{D}_{\mathrm{KL}}\left( q_{\bm{\theta}}(\mathbf{z}|\mathbf{x}) \| p_{\bm{\phi}}(\mathbf{z}|\mathbf{x}) \right) \right] \\
    \\=&  \underbrace{\mathcal{D}_{\mathrm{KL}}(p_{\text{data}} \| p_{\bm{\phi}})}_{\text{True Modeling Error}} + \underbrace{\mathbb{E}_{p_{\text{data}}(\mathbf{x})} \big[ \mathcal{D}_{\mathrm{KL}}( q_{\bm{\theta}}(\mathbf{z}|\mathbf{x}) \| p_{\bm{\phi}}(\mathbf{z}|\mathbf{x}) ) \big]}_{\text{Inference Error}},
\end{align*}
where the first term is the true modeling error and the second is the inference error, i.e., the gap between the approximate and true posteriors. The latter is always non-negative, which explains \Cref{eq:joint-bound}.

Finally, note that
\[
\log p_{\bm{\phi}}(\rvx) - \mathcal{L}_{\text{ELBO}}(\bm{\theta},\bm{\phi};\rvx) 
= \mathcal{D}_{\mathrm{KL}} \left(q_{\bm{\theta}}(\rvz|\rvx)\|p_{\bm{\phi}}(\rvz|\rvx)\right).
\]
Thus the inference error is exactly the gap between the log-likelihood and the ELBO. 
Maximizing the ELBO therefore corresponds to directly reducing inference error, ensuring that training minimizes a meaningful part of the overall bound.

\subsection{Gaussian VAE}\label{subsec:optimization-vae}
A standard formulation of the VAE employs Gaussian distributions for both the encoder and decoder.
\paragraph{Encoder Part.}
The encoder $ q_{\bm{\theta}}(\mathbf{z} | \mathbf{x}) $ is typically modeled as a Gaussian distribution as:
\[
q_{\bm{\theta}}(\mathbf{z} | \mathbf{x}) := \mathcal{N}\left(\mathbf{z}; \bm{\mu}_{\bm{\theta}}(\mathbf{x}), \operatorname{diag}(\bm{\sigma}_{\bm{\theta}}^2(\mathbf{x}))\right),
\]
where $\bm{\mu}_{\bm{\theta}} : \mathbb{R}^D \to \mathbb{R}^d$ and $\bm{\sigma}_{\bm{\theta}} : \mathbb{R}^D \to \mathbb{R}_+^d$ are deterministic outputs of the encoder network.
\paragraph{Decoder Part.} The decoder is typically modeled as a Gaussian distribution with fixed variance:
\[
p_{\bm{\phi}}(\mathbf{x}|\mathbf{z}) := \mathcal{N}\bigl(\mathbf{x}; \bm{\mu}_{\bm{\phi}}(\mathbf{z}), \sigma^2 \rmI\bigr),
\]
where $\bm{\mu}_{\bm{\phi}} : \mathbb{R}^d \to \mathbb{R}^D$ is a neural network, and $\sigma > 0$ is a small constant controlling the variance.
 
Under this assumption, the reconstruction term in the ELBO simplifies as
\[
\mathbb{E}_{q_{\bm{\theta}}(\mathbf{z} | \mathbf{x})} \left[\log p_{\bm{\phi}}(\mathbf{x} | \mathbf{z})\right]
= - \frac{1}{2\sigma^2} \mathbb{E}_{q_{\bm{\theta}}(\mathbf{z} | \mathbf{x})} \left[\|\mathbf{x} - \bm{\mu}_{\bm{\phi}}(\mathbf{z})\|^2\right] + C,
\]
where $C$ is a constant independent of $\bm{\theta}$ and $\bm{\phi}$. The ELBO objective thus reduces to:
\[
\min_{\bm{\theta}, \bm{\phi}}  \mathbb{E}_{q_{\bm{\theta}}(\mathbf{z} | \mathbf{x})} \left[ \frac{1}{2\sigma^2} \|\mathbf{x} - \bm{\mu}_{\bm{\phi}}(\mathbf{z})\|^2 \right]
+ \mathcal{D}_{\mathrm{KL}}\big(q_{\bm{\theta}}(\mathbf{z} | \mathbf{x}) \| p_{\mathrm{prior}}(\mathbf{z})\big),
\]
where the KL term admits a closed-form solution due to the Gaussian assumption. Training the VAE therefore reduces to minimizing a regularized reconstruction loss.

\subsection{Drawbacks of Standard VAE}\label{sec:drawback_standard_VAE}
Despite the theoretical appeal of the VAE framework, it suffers from a critical drawback: it often produces blurry outputs\footnote{The limitation we study here is specific to the common VAE setup introduced
above: a simple Gaussian decoder trained through a squared reconstruction
error. In this setting, the population-optimal decoder returns a conditional
mean. If several visually different inputs correspond to overlapping latent
regions, this averaging can appear as blur. This should not be read as a
universal statement about every VAE architecture or observation model.}.

\paragraph{Blurry Generations in VAEs.}
To understand this phenomenon, consider a fixed Gaussian encoder $ q_{\text{enc}}(\mathbf{z}|\mathbf{x}) $, and a decoder of the form
\[
p_{\text{dec}}(\mathbf{x}|\mathbf{z}) = \mathcal{N}(\mathbf{x}; \bm{\mu}(\rvz), \sigma^2 \rmI),
\]
where $ \bm{\mu}(\rvz) $ denotes the decoder network. With an arbitrary encoder, optimizing the ELBO reduces (up to an additive constant) to minimizing the expected reconstruction error:
\[
\argmin_{\bm{\mu}} \mathbb{E}_{p_{\text{data}}(\mathbf{x})q_{\text{enc}}(\mathbf{z}|\mathbf{x})} \left[ \| \mathbf{x} - \bm{\mu}(\rvz) \|^2 \right].
\]
This is a standard least squares problem in $ \bm{\mu}(\rvz) $, and its solution is given in closed form by the conditional mean:
\[
\bm{\mu}^*(\mathbf{z}) = \mathbb{E}_{q_{\text{enc}}(\mathbf{x}|\mathbf{z})}[\mathbf{x}],
\]
where $ q_{\text{enc}}(\mathbf{x}|\mathbf{z}) $ is the encoder-induced posterior on inputs given latents, defined via Bayes’ rule:
\[
q_{\text{enc}}(\mathbf{x}|\mathbf{z})
=
\frac{q_{\text{enc}}(\mathbf{z}|\mathbf{x})\, p_{\text{data}}(\mathbf{x})}{q_{\text{enc}}(\mathbf{z})},
\qquad
q_{\text{enc}}(\mathbf{z})
:=
\int q_{\text{enc}}(\mathbf{z}|\mathbf{x})\, p_{\text{data}}(\mathbf{x})\, \diff \mathbf{x}.
\]

An equivalent form of the optimal generator via Bayes' rule is:
\[
\bm{\mu}^*(\mathbf{z})
=
\frac{\mathbb{E}_{p_{\text{data}}(\mathbf{x})}\!\left[q_{\text{enc}}(\mathbf{z}|\mathbf{x}) \cdot \mathbf{x}\right]}
{q_{\text{enc}}(\mathbf{z})}
=
\frac{\mathbb{E}_{p_{\text{data}}(\mathbf{x})}\!\left[q_{\text{enc}}(\mathbf{z}|\mathbf{x}) \cdot \mathbf{x}\right]}
{\mathbb{E}_{p_{\text{data}}(\mathbf{x})}\!\left[q_{\text{enc}}(\mathbf{z}|\mathbf{x})\right]}.
\]
Now suppose that two distinct inputs $ \mathbf{x} \neq \mathbf{x}' $ are mapped to overlapping regions in latent space, i.e., the supports of $ q_{\text{enc}}(\cdot|\mathbf{x}) $ and $ q_{\text{enc}}(\cdot|\mathbf{x}') $ intersect. Then $\bm{\mu}^*(\mathbf{z})$ averages over multiple, potentially unrelated inputs, which leads to blurry, non-distinct outputs. This averaging effect over conflicting modes is a fundamental reason for the characteristic blurriness in VAE-generated samples.

\newpage
\subsection{(Optional) From Standard VAE to Hierarchical VAEs}
To model complex data, Hierarchical Variational Autoencoders (HVAEs)
\citep{vahdat2020nvae} enhance VAEs by introducing a hierarchy of latent variables. This deep, layered structure allows the model to capture data features at multiple levels of abstraction, significantly boosting expressive power and mirroring the compositional nature of real-world data.
\vspace{0.3cm}
\begin{figure}[tbh!]
    \centering
\begin{tikzpicture}[
    ->, 
    >=Stealth, 
    node_dist/.style={right=1.7cm of #1}, 
    font=\small,
    thick
  ]

  \tikzset{state/.style={
      rectangle,
      rounded corners=3pt,
      draw=black,
      fill=gray!10,
      minimum height=35pt,
      minimum width=25pt
    }
  }

  \node[state] (x) {$\rvx$};
  \node[state] (z1) [node_dist=x] {$\rvz_1$};
  \node[state] (z2) [node_dist=z1] {$\rvz_2$};
  \node (dots) [right=0.7cm of z2] {$\boldsymbol{\cdots}$};
  \node[state] (zT) [node_dist=dots] {$\rvz_L$};

  
  \path ([yshift=4.5pt]x.east)  edge node[above] {$q_{\bm{\theta}}(\rvz_1|\rvx)$} ([yshift=4.5pt]z1.west);
  \path ([yshift=4.5pt]z1.east) edge node[above] {$q_{\bm{\theta}}(\rvz_2|\rvz_1)$} ([yshift=4.5pt]z2.west);
  \path ([yshift=4.5pt]z2.east) edge ([yshift=4.5pt]dots.west);
  \path ([yshift=4.5pt]dots.east) edge node[above, pos=0.4] {$q_{\bm{\theta}}(\rvz_L|\rvz_{L-1})$} ([yshift=4.5pt]zT.west);
  
  \path ([yshift=-4.5pt]z1.west) edge node[below] {$p_{\bm{\phi}}(\rvx|\rvz_1)$} ([yshift=-4.5pt]x.east);
  \path ([yshift=-4.5pt]z2.west) edge node[below] {$p_{\bm{\phi}}(\rvz_1|\rvz_2)$} ([yshift=-4.5pt]z1.east);
  \path ([yshift=-4.5pt]dots.west) edge ([yshift=-4.5pt]z2.east);
  \path ([yshift=-4.5pt]zT.west) edge node[below, pos=0.6] {$p_{\bm{\phi}}(\rvz_{L-1}|\rvz_L)$} ([yshift=-4.5pt]dots.east);

\end{tikzpicture}

    \caption{\textbfs{Computation graph of the HVAE.} It has a hierarchical structure with stacked, trainable encoders and decoders across multiple latent layers. \figcredit{Created by the authors.}
}
    \label{fig:hvae-graph}
\end{figure}

\paragraph{HVAE's Modeling.}
Unlike standard VAEs that use a single latent code $\mathbf{z}$, hierarchical VAEs (HVAEs) introduce multiple layers of latent variables arranged in a top-down hierarchy.  
Each latent layer conditions the one below it, forming a chain of conditional priors that captures structure at progressively finer levels of abstraction.  
This leads to the following top-down factorization of the joint distribution:

\[
p_{\bm{\phi}}(\mathbf{x}, \mathbf{z}_{1:L}) = p_{\bm{\phi}}(\mathbf{x} | \mathbf{z}_1) \prod_{i=2}^{L} p_{\bm{\phi}}(\mathbf{z}_{i-1}|\mathbf{z}_i)   p(\mathbf{z}_L).
\]
This structure defines the marginal data distribution,
\[
p_{\text{HVAE}}(\mathbf{x}) := \int p_{\bm{\phi}}(\mathbf{x}, \mathbf{z}_{1:L}) \diff \mathbf{z}_{1:L}.
\]
 Generation proceeds progressively: starting from the top latent variable $ \mathbf{z}_L $, each latent is decoded sequentially down to $ \mathbf{z}_1 $, followed by generating the final observation $ \mathbf{x} $.

For encoding part, HVAEs utilize a structured, learnable variational encoder $ q_{\bm{\theta}}(\mathbf{z}_{1:L}|\mathbf{x}) $ that mirrors the generative hierarchy. A common choice is a bottom-up Markov factorization:
\[
q_{\bm{\theta}}(\mathbf{z}_{1:L} | \mathbf{x}) = q_{\bm{\theta}}(\mathbf{z}_1 | \mathbf{x}) \prod_{i=2}^{L} q_{\bm{\theta}}(\mathbf{z}_i | \mathbf{z}_{i-1}).
\]

\paragraph{HVAE's ELBO.}
Similar to \Cref{eq:vae-elbo}, ELBO is derived via Jensen's inequality:
\begin{align}\label{eq:hvae-elbo}
\begin{aligned}
    \log p_{\text{HVAE}}(\mathbf{x}) 
&= \log \int p_{\bm{\phi}}(\mathbf{x}, \mathbf{z}_{1:L}) \diff \mathbf{z}_{1:L} \\
&= \log \int \frac{p_{\bm{\phi}}(\mathbf{x}, \mathbf{z}_{1:L})}{q_{\bm{\theta}}(\mathbf{z}_{1:L}|\mathbf{x})}   q_{\bm{\theta}}(\mathbf{z}_{1:L}|\mathbf{x}) \diff \mathbf{z}_{1:L} \\
&= \log \mathbb{E}_{q_{\bm{\theta}}(\mathbf{z}_{1:L}|\mathbf{x})} \left[ \frac{p_{\bm{\phi}}(\mathbf{x}, \mathbf{z}_{1:L})}{q_{\bm{\theta}}(\mathbf{z}_{1:L}|\mathbf{x})} \right] \\
&\geq \mathbb{E}_{q_{\bm{\theta}}(\mathbf{z}_{1:L}|\mathbf{x})} \left[ \log \frac{p_{\bm{\phi}}(\mathbf{x}, \mathbf{z}_{1:L})}{q_{\bm{\theta}}(\mathbf{z}_{1:L}|\mathbf{x})} \right]\\
&=:\mathcal{L}_{\text{ELBO}}(\bm{\phi}).
\end{aligned}
\end{align}
Substituting the factorized forms yields:
\begin{align*}
\mathcal{L}_{\text{ELBO}} 
= \mathbb{E}_{q_{\bm{\theta}}(\mathbf{z}_{1:L}|\mathbf{x})} \left[ 
\log \frac{
p(\mathbf{z}_L) \prod_{i=2}^{L} p_{\bm{\phi}}(\mathbf{z}_{i-1} | \mathbf{z}_i)   p_{\bm{\phi}}(\mathbf{x} | \mathbf{z}_1)
}{
q_{\bm{\theta}}(\mathbf{z}_1|\mathbf{x}) \prod_{i=2}^{L} q_{\bm{\theta}}(\mathbf{z}_i|\mathbf{z}_{i-1})
}
\right].
\end{align*}

This hierarchical ELBO decomposes into interpretable terms, including a
reconstruction term, adjacent-layer matching terms, and a top-level KL
regularizer to the prior.

The leap from shallow to deep networks revolutionized machine learning, and a similar idea transformed generative models. HVAEs showed the power of using deep, stacked layers to build data. This concept of a layered hierarchy is a cornerstone of modern generative modeling, appearing again in score-based methods (\Cref{sec:SMLD}) and normalizing flows (\Cref{sec:flow-based-method}). The core insight is simple yet powerful: 
\msg{Observation}{}{Stacking layers allows the model to generate data progressively, starting with coarse details and adding finer ones at each step. This process makes it far easier to capture the complex structure of high-dimensional data.}


\paragraph{Why Deeper Networks in a Flat VAE are Not Enough.} There are two fundamental limitations of a standard flat VAE that are not resolved by simply making the encoder and decoder deeper.

The first limitation is the variational family. In a standard VAE,
\[
q_\btheta(\rvz| \rvx)=\mathcal N \big(\rvz;\,\bm{\mu}_{\btheta}(\rvx), \operatorname{diag}(\sigma_\theta^2(\rvx))\big),
\]
so for each fixed $\rvx$ the encoder posterior is a single Gaussian with diagonal
covariance. Greater network depth improves the accuracy of $\bm{\mu}_{\btheta}$ and
$\sigma_\btheta$ but does not expand the family; even a full covariance remains
one unimodal ellipsoid. When $p_\bphi(\rvz| \rvx)$ is multi-peaked, this family cannot match it, which loosens the ELBO and
weakens inference. Addressing this requires a richer posterior class, not merely deeper networks.

Second, if the decoder is too expressive, the model may suffer from posterior collapse. To see why, let us recall that the objective of the VAE is
\begin{align*}
&\mathbb{E}_{p_{\text{data}}(\mathbf{x})}[\mathcal L_{\mathrm{ELBO}}(\rvx)]\\
&\quad\quad=\E_{p_{\text{data}}(\mathbf{x})q_\btheta(\rvz|\rvx)}[\log p_\bphi(\rvx|\rvz)]
-\mathbb{E}_{p_{\text{data}}(\mathbf{x})}\big[\mathcal{D}_\mathrm{KL} \big(q_\btheta(\rvz|\rvx) \| p(\rvz)\big)\big]\\
&\quad\quad=\E_{p_{\text{data}}(\mathbf{x})q_\btheta(\rvz|\rvx)}[\log p_\bphi(\rvx|\rvz)]-\mathcal I_q(\rvx;\rvz) - \mathcal{D}_\mathrm{KL}(q_{\bm{\theta}}(\rvz) \| p(\rvz)),
\end{align*}
where $\mathcal I_q(\rvx;\rvz)$ is the mutual information defined by
\[
\mathcal I_q(\rvx;\rvz)
=\E_{q(\rvx,\rvz)} \Big[\log \tfrac{q_\btheta(\rvz|\rvx)}{q(\rvz)}\Big]
=\E_{p_{\mathrm{data}}(\rvx)} \big[\mathcal{D}_\mathrm{KL}(q_\btheta(\rvz|\rvx) \| q(\rvz))\big],
\]
and the aggregated posterior is $q_{\bm{\theta}}(\rvz)= \int p_{\mathrm{data}}(\rvx) q_\btheta(\rvz|\rvx) \diff\rvx$.

If the decoder class can model the data well without using $\rvz$ (i.e., it contains some $p_\bphi(\rvx|\rvz)=r(\rvx)$ close to $p_{\mathrm{data}}$), then a maximizer of the ELBO sets $q_\btheta(\rvz|\rvx)=p(\rvz)$, so $\mathcal I_q(\rvx;\rvz)=0$ and $q_\btheta(\rvz)=p(\rvz)$. This ``ignore $\rvz$'' solution does not disappear by making the networks deeper: (1) the learned code becomes independent of $\rvx$ (so it carries no data-dependent structure useful for downstream tasks), and (2) conditioning or moving in $\rvz$ has no effect on generated samples, so controllable generation fails.

\paragraph{What Hierarchy Changes?}
An HVAE introduces multiple latent levels,
\[
p_\bphi(\rvx,\rvz_{1:L})=p_\bphi(\rvx|\rvz_1)\prod_{i=2}^L p_\bphi(\rvz_{i-1}|\rvz_i) p(\rvz_L),
\]
with ELBO
\begin{align*}
\mathcal L_{\mathrm{ELBO}}(\rvx)
&= \E_{q_\btheta(\rvz_1|\rvx)}[\log p_\bphi(\rvx|\rvz_1)]
\\
&\quad - \E_{q_\btheta(\rvz_{1:2}|\rvx)}
\big[\log q_\btheta(\rvz_1|\rvx)-\log p_\bphi(\rvz_1|\rvz_2)\big]
\\
&\quad - \sum_{i=2}^{L-1}
\E_{q_\btheta(\rvz_{1:i+1}|\rvx)}
\big[\log q_\btheta(\rvz_i|\rvz_{i-1})-\log p_\bphi(\rvz_i|\rvz_{i+1})\big]
\\
&\quad - \E_{q_\btheta(\rvz_{L-1}|\rvx)}
\Big[\mathcal{D}_{\mathrm{KL}}(q_\btheta(\rvz_L|\rvz_{L-1}) \| p(\rvz_L))\Big].
\end{align*}
Here, we denote $\E_q := \E_{p_{\mathrm{data}}(\rvx) q_{\btheta}(\rvz_{1:L}|\rvx)}$.
Each latent level interacts only with its neighboring levels in the
hierarchy: the encoder passes information upward through
$q_\btheta(\rvz_i|\rvz_{i-1})$, while the decoder passes information
downward through $p_\bphi(\rvz_{i-1}|\rvz_i)$. Accordingly, the ELBO
decomposes into a reconstruction term, adjacent-layer matching terms,
and a top-level KL regularizer to the prior. These properties stem from
the hierarchical latent graph, not from simply deepening networks in a
flat VAE.

\paragraph{What Will be Ahead?}
While HVAEs extend the VAE framework with multiple latent layers for expressiveness, their training poses unique challenges. Because the encoder and decoder must be optimized jointly, learning becomes unstable: lower layers and the decoder can already reconstruct $\rvx$, leaving higher-level latents with little effective signal. Moreover, gradient information reaching deeper variables is often indirect and weak, making it difficult for them to contribute meaningfully. An additional difficulty lies in balancing model capacity, since overly expressive conditionals  can dominate the reconstruction task and suppress the utility of higher latents.

Interestingly, the core idea of a deep, layered hierarchy finds a more powerful incarnation in variational diffusion models, a topic we explore in \Cref{sec:ddpm}. Diffusion models inherit the progressive structure of HVAEs but elegantly sidestep their central weakness. By fixing the encoding process and focusing solely on learning the generative reversal, they unlock newfound stability and modeling flexibility, leading to a significant leap in the quality of generated outputs.

For notational simplicity, we deviate from the common VAE convention that uses $q$ for the encoder and $p$ for the generator. To avoid ambiguity, we denote distributions as 
$p$ and will always specify their roles through appropriate subscripts or superscripts, clarifying them in context.

\clearpage
\newpage

\section{Variational Perspective: DDPM}\label{sec:ddpm}

Denoising Diffusion Probabilistic Models (DDPMs) \citep{sohl2015deep,ho2020denoising} represent a cornerstone of diffusion modeling. Conceptually, they operate within a variational framework, much like VAEs and HVAEs. However, DDPMs introduce a clever twist that tackles some of the challenges faced by their predecessors.

At their core, DDPMs involve two distinct stochastic processes:
\begin{itemize}
    \item \textbfs{The Forward Pass (Fixed Encoder):} This process gradually corrupts data by injecting Gaussian noise over multiple steps via a transition kernel $p(\mathbf{x}_{i}|\mathbf{x}_{i-1})$. The data evolves into an isotropic Gaussian distribution, effectively becoming pure noise. This means the encoder is fixed and not learned.
    \item \textbfs{The Reverse Denoising Process (Learnable Decoder):} Here, a neural network learns to reverse the noise corruption through a parameterized distribution $p_{\bm{\phi}}(\mathbf{x}_{i-1}|\mathbf{x}_{i})$. Starting from pure noise, this process iteratively denoises to generate realistic samples. Crucially, each individual denoising step is a more manageable task than generating a complete sample from scratch, as VAEs often attempt to do.
\end{itemize}
By fixing the encoder and concentrating learning on the gradual generative trajectory, DDPMs achieve remarkable stability and expressive power. 

\vspace{0.3cm}
\begin{figure}[tbh!]
    \centering
\begin{tikzpicture}[
    ->, 
    >=Stealth, 
    node_dist/.style={right=1.7cm of #1}, 
    font=\small,
    thick
  ]

  \tikzset{state/.style={
      rectangle,
      rounded corners=3pt,
      draw=black,
      fill=gray!10, 
      minimum height=35pt,
      minimum width=25pt
    }
  }

  \node[state] (x) {$\rvx_0$};
  \node[state] (z1) [node_dist=x] {$\rvx_1$};
  \node[state] (z2) [node_dist=z1] {$\rvx_2$};
  \node (dots) [right=0.7cm of z2] {$\boldsymbol{\cdots}$};
  \node[state] (zT) [node_dist=dots] {$\rvx_L$};

  
  \path [gray] ([yshift=4.5pt]x.east)  edge node[above] {$p(\rvx_1|\rvx_0)$} ([yshift=4.5pt]z1.west);
  \path [gray] ([yshift=4.5pt]z1.east) edge node[above] {$p(\rvx_2|\rvx_1)$} ([yshift=4.5pt]z2.west);
  \path [gray] ([yshift=4.5pt]z2.east) edge ([yshift=4.5pt]dots.west);
  \path [gray] ([yshift=4.5pt]dots.east) edge node[above, pos=0.4] {$p(\rvx_L|\rvx_{L-1})$} ([yshift=4.5pt]zT.west);
  
  \path ([yshift=-4.5pt]z1.west) edge node[below] {$p_{\bm{\phi}}(\rvx_0|\rvx_1)$} ([yshift=-4.5pt]x.east);
  \path ([yshift=-4.5pt]z2.west) edge node[below] {$p_{\bm{\phi}}(\rvx_1|\rvx_2)$} ([yshift=-4.5pt]z1.east);
  \path ([yshift=-4.5pt]dots.west) edge ([yshift=-4.5pt]z2.east);
  \path ([yshift=-4.5pt]zT.west) edge node[below, pos=0.6] {$p_{\bm{\phi}}(\rvx_{L-1}|\rvx_L)$} ([yshift=-4.5pt]dots.east);

\end{tikzpicture}
    \caption{\textbfs{Illustration of DDPM.} It consists of a fixed forward process (in gray) that gradually adds Gaussian noise to the data, and a learned reverse process that denoises step-by-step to generate new samples. \figcredit{Created by the authors.}}
     \label{fig:vdm-grap}
\end{figure}


In this section, we focus on DDPMs, postponing the broader discussion to \Cref{sec:variational-perspective}, where we present a more general and flexible framework.

\subsection{Forward Process (Fixed Encoder)}
\begin{figure}[tbh!]
    \centering
    \includegraphics[width=\linewidth]{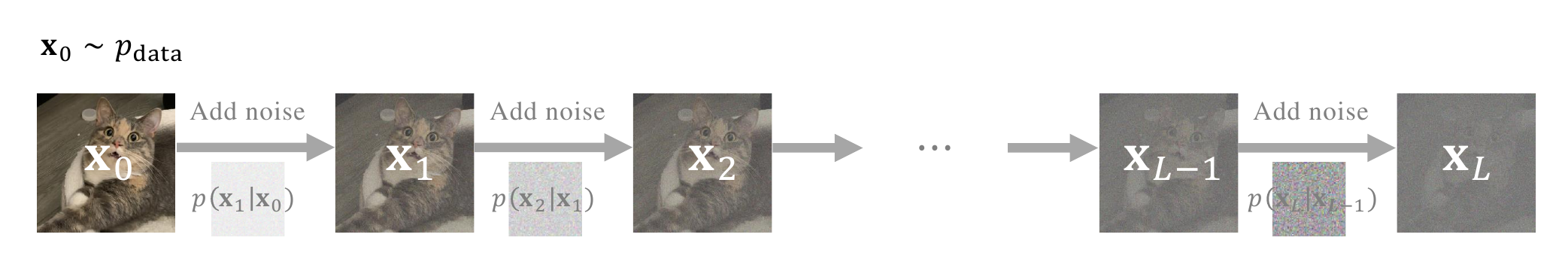}
    \caption{Illustration of the DDPM forward process, wherein Gaussian noise is incrementally added to corrupt a data sample into pure noise. \figcredit{Created by the authors.}}
    \label{fig:vdm-forward}
\end{figure}

In DDPMs, the forward process is a fixed, non-trainable operation that serves as an encoder. It progressively corrupts the original data by adding noise over multiple steps, eventually transforming it into a simple prior distribution $p_{\text{prior}} := \mathcal{N}(\mathbf{0}, \mathbf{I})$. This transformation is depicted as the forward chain in \Cref{fig:vdm-grap} or as illustrated in \Cref{fig:vdm-forward}.

Let us formalize this step-by-step degradation:
\paragraph{Fixed Gaussian Transitions.}
Each step in the forward process is governed by a fixed Gaussian transition kernel\footnote{This formulation, while potentially appearing different, is mathematically equivalent to the original DDPM transition kernel.}:
\[
p(\rvx_i|\rvx_{i-1}) := \mathcal{N}(\rvx_i; \sqrt{1 - \beta_i^2}   \rvx_{i-1}, \beta_i^2 \rmI).
\]
Here, the process begins with $\rvx_{0} $, representing a sample drawn from the real data distribution $p_{\text{data}}$. The sequence $\{\beta_i\}_{i=1}^L$ denotes a pre-determined, monotonically increasing noise schedule, where each 
$\beta_i \in (0, 1)$ controls the variance of the Gaussian noise injected at step $i$.  For convenience, we define $\alpha_i := \sqrt{1 - \beta_i^2}$. This mathematical definition is precisely equivalent to the following intuitive iterative update:
\[
\rvx_i = \alpha_i \rvx_{i-1} + \beta_i \bm{\epsilon}_i,  
\]
where $\bm{\epsilon}_i \sim \mathcal{N}(\bm{0}, \rmI)$ are independently and identically distributed. This means at each step $i$, we scale down the previous state $\rvx_{i-1}$ by $\alpha_i$ and add a controlled amount of Gaussian noise scaled by $\beta_i$.

\paragraph{Perturbation Kernel and Prior Distribution.}
By recursively applying the transition kernels, we obtain a closed-form expression for the distribution of noisy samples at step $i$ given the original data $\rvx_0$:
\[
p_i(\rvx_i|\rvx_0) = \mathcal{N}\left(\rvx_i; \bar{\alpha}_i \rvx_0, (1 - \bar{\alpha}_i^2) \rmI \right),
\]
where 
\[
\bar{\alpha}_i := \prod_{k=1}^i \sqrt{1-\beta_k^2} = \prod_{k=1}^i \alpha_k.
\]
This means we can sample $\rvx_i$ directly from $\rvx$ as
\begin{align}\label{eq:ddpm-forward}
    \rvx_i = \bar{\alpha}_i \rvx_0 + \sqrt{1-\bar{\alpha}_i^2}   \bm{\epsilon}, \quad \bm{\epsilon} \sim \mathcal{N}(\bm{0}, \rmI).
\end{align}
Let the noise schedule $ \{\beta_i\}_{i=1}^L $ be an increasing sequence, then the marginal distribution of the forward process converges as
\[
p_L(\mathbf{x}_L | \rvx_0) \longrightarrow \mathcal{N}(\mathbf{0}, \mathbf{I}) \quad \text{as } L \to \infty,
\]
which motivates the choice of the prior distribution as
\[
p_{\text{prior}} := \mathcal{N}(\mathbf{0}, \mathbf{I})
\]
with no reliance on data $\rvx_0$.


\subsection{Reverse Denoising Process (Learnable Decoder)}

\begin{figure}[tbh!]
    \centering
    \includegraphics[width=\linewidth]{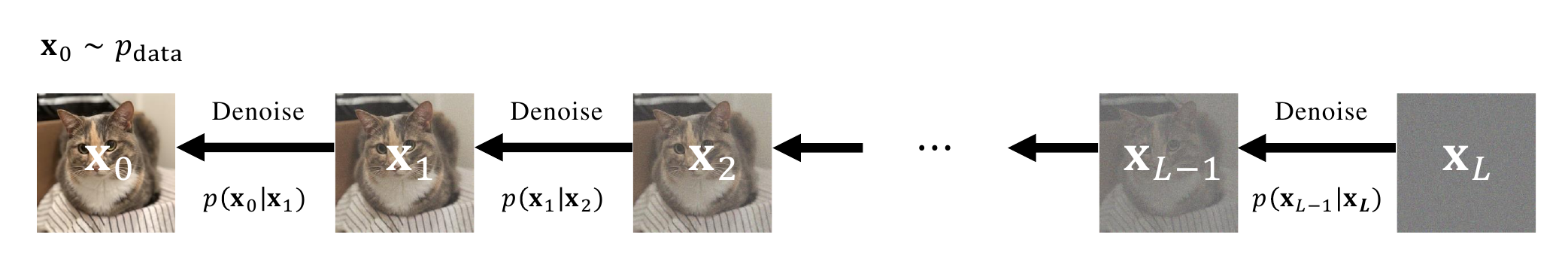}
\caption{\textbfs{Illustration of DDPM reverse (denoising) process.} Starting from noise $\mathbf{x}_L \sim  p_{\mathrm{prior}}$, the model sequentially samples $\mathbf{x}_{i-1} \sim  p(\mathbf{x}_{i-1}|\mathbf{x}_i)$ for $i=L,\ldots,1$ to obtain a newly generated data $\mathbf{x}_0$. The oracle transition $p(\mathbf{x}_{i-1}|\mathbf{x}_i)$ is unknown; thus, we aim to approximate it.
\figcredit{Created by the authors.}
}

    \label{fig:ddpm-sampling-formal-rev}
\end{figure}

At its core, the essence of DDPMs lies in their ability to \emph{reverse} the controlled degradation imposed by the forward diffusion process. Starting from pure, unstructured noise, $\mathbf{x}_L \sim p_{\text{prior}}$, the objective is to progressively \emph{denoise} this randomness, step by step, until a coherent and meaningful data sample emerges. This reverse generation proceeds through a Markov chain, illustrated by \Cref{fig:ddpm-sampling-formal-rev}.

The fundamental challenge, and the central question guiding DDPM development, then becomes:
\begin{question}
Can we precisely compute, or at least effectively approximate, these reverse transition kernels $ p(\mathbf{x}_{i-1}|\mathbf{x}_i) $, especially when considering the complex distribution of $\mathbf{x}_i \sim p_i(\mathbf{x}_i)$?
\end{question}

Rather than diving immediately into the mathematically intricate derivation of the Evidence Lower Bound (ELBO), as the original DDPM paper does (for which a detailed discussion awaits in \Cref{subsec:ddpm-elbo}), we will instead approach the training objective from a more intuitive perspective: by leveraging conditional probabilities to achieve a tractable formulation.

\paragraph{Overview: Modeling and Training Objective.} 
To enable the generative process, our goal is to approximate the unknown true reverse transition kernel, $p(\mathbf{x}_{i-1}|\mathbf{x}_i)$. We achieve this by introducing a learnable parametric model, $p_{\bm{\phi}}(\mathbf{x}_{i-1}|\mathbf{x}_i)$, and training it to minimize the expected KL divergence:
\begin{align}\label{eq:marginal-kl-matching}
    \mathbb{E}_{p_i(\mathbf{x}_i)} \left[
    \mathcal{D}_{\mathrm{KL}}\big(p(\mathbf{x}_{i-1}|\mathbf{x}_i) \| p_{\bm{\phi}}(\mathbf{x}_{i-1}|\mathbf{x}_i)\big)
    \right].
\end{align}

However, a direct computation of the target distribution $p(\mathbf{x}_{i-1}|\mathbf{x}_i)$ is challenging. By Bayes' theorem, we would need to evaluate:
\[
p(\mathbf{x}_{i-1}|\mathbf{x}_i) = p(\mathbf{x}_i|\mathbf{x}_{i-1}) \underbrace{\frac{p_{i-1}(\mathbf{x}_{i-1})}{p_i(\mathbf{x}_i)}}_{\text{intractable}}.
\]
The marginals $p_i(\rvx_i)$ and $p_{i-1}(\rvx_{i-1})$ are expectations over the unknown data distribution $p_{\mathrm{data}}$, given by:
\[
p_i(\rvx_i) = \int p_i(\rvx_i | \rvx_0) p_{\mathrm{data}}(\rvx_0) \mathrm{d}\rvx_0,
\]
and analogously for $p_{i-1}(\rvx_{i-1})$. Since $p_{\mathrm{data}}$ is unknown, these integrals have no closed-form evaluation; at best they can be approximated from samples, so the exact densities are not available in practice.

\begin{figure}[th!]
    \centering
    \includegraphics[width=\linewidth]{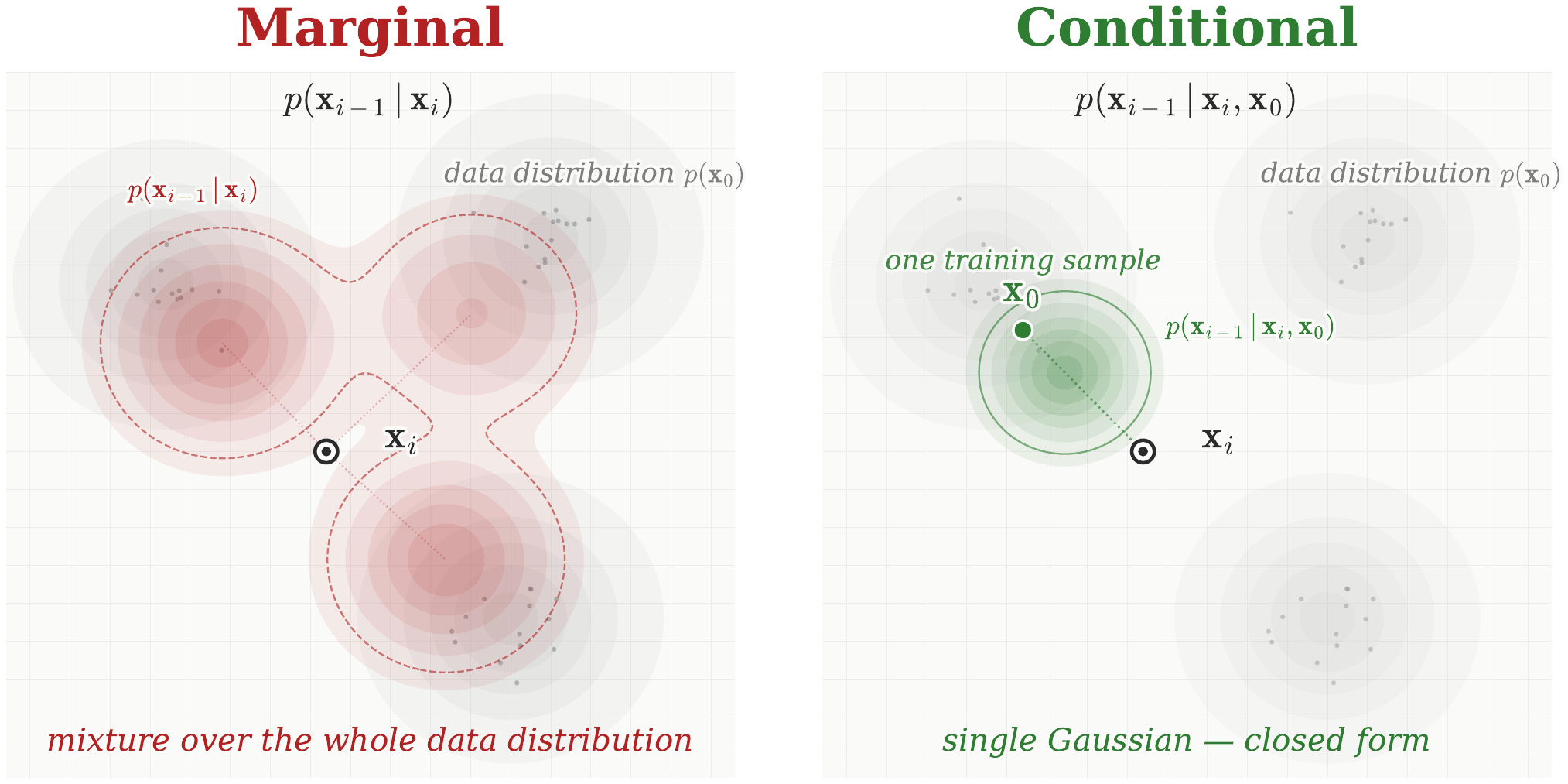}
    \caption{\textbfs{The conditioning trick in DDPM.}
\textbfs{(Left)}
For a fixed noisy sample $\rvx_i$, the reverse distribution
$p(\rvx_{i-1}|\rvx_i)$ averages over all possible clean samples
$\rvx_0$ that could have produced $\rvx_i$. It can therefore be a
complicated mixture.
\textbfs{(Right)}
During training, however, the clean sample $\rvx_0$ used to generate
$\rvx_i$ is known. Conditioning on this sample gives
$p(\rvx_{i-1}|\rvx_i,\rvx_0)$, a Gaussian with closed-form mean and
variance. DDPM can therefore use this tractable conditional distribution
as a simple regression target. Although each training example conditions
on only one clean sample, averaging the conditional training objective
over all samples gives the same optimal model as directly learning the
marginal reverse distribution; see Theorem~\ref{thm:equiv-marginal-kl}.
Thus, DDPM can learn the same optimal reverse prediction without directly
regressing onto the complicated, generally intractable target induced by
the marginal mixture.    \figcredit{Created by the authors with AI-assisted coding.}}
    \label{fig:conditional-trick}
\end{figure}

\paragraph{Overcoming Intractability with Conditioning.}
A central insight in DDPMs resolves this intractability: we condition the reverse transition on a clean data sample $\mathbf{x}_0$. This subtle yet powerful step transforms the intractable kernel into one that is mathematically tractable:
\[
p(\mathbf{x}_{i-1}|\mathbf{x}_i, \mathbf{x}_0) = p(\mathbf{x}_i|\mathbf{x}_{i-1}) \frac{p(\mathbf{x}_{i-1}|\mathbf{x}_0)}{p_i(\mathbf{x}_i|\mathbf{x}_0)}.
\]
This tractability arises from two key properties of the forward process: its Markov property, meaning $p(\mathbf{x}_i|\mathbf{x}_{i-1}, \mathbf{x}_0) = p(\mathbf{x}_i|\mathbf{x}_{i-1})$, and the Gaussian nature of all involved distributions. As a result, $p(\mathbf{x}_{i-1}|\mathbf{x}_i, \mathbf{x}_0)$ itself is Gaussian and admits a closed-form expression (which we will see in \Cref{eq:ddpm-reverse-mean}). We visualize it in \Cref{fig:conditional-trick}. Crucially, this elegant conditioning strategy allows us to derive a tractable objective that is functionally equivalent to the seemingly intractable marginal KL divergence in \Cref{eq:marginal-kl-matching}.

\thmp{Equivalence Between Marginal and Conditional KL Minimization}{equiv-marginal-kl}{
The following equality holds:
\begin{align}\label{eq:kl-matching}
\begin{aligned}
    &\mathbb{E}_{p_i(\rvx_i)}\left[
        \mathcal{D}_{\mathrm{KL}}\big(p(\rvx_{i-1}|\rvx_i) \| p_{\bm{\phi}}(\rvx_{i-1}|\rvx_i)\big)
    \right] \\
    ={}&
    \mathbb{E}_{p_{\mathrm{data}}(\rvx_0)}
    \mathbb{E}_{p_i(\rvx_i|\rvx_0)} \left[
        \mathcal{D}_{\mathrm{KL}}\big(p(\rvx_{i-1}|\rvx_i, \rvx_0) \| p_{\bm{\phi}}(\rvx_{i-1}|\rvx_i)\big)
    \right] + C.
\end{aligned}
\end{align}
where $C$ is a constant independent of $\bm{\phi}$. Moreover, the minimizer of \Cref{eq:kl-matching} satisfies
\[
p^*(\mathbf{x}_{i-1}|\mathbf{x}_i) = \mathbb{E}_{p(\mathbf{x}_0|\mathbf{x}_i)} \big[p(\mathbf{x}_{i-1}|\mathbf{x}_i, \mathbf{x}_0)\big] = p(\mathbf{x}_{i-1}|\mathbf{x}_i), \quad \mathbf{x}_i \sim p_i.
\]
}{The proof rewrites a KL-divergence expectation by expanding definitions, applying the chain rule of probability, and using a logarithmic identity to decompose it into the sum of an expected conditional KL divergence and a marginal KL divergence. A complete derivation is in \Cref{subsec:proof-equiv-marginal-kl}.}

This alternative viewpoint: conditioning to obtain a tractable objective, forms the foundation of DDPMs and reveals a profound commonality with other influential diffusion models, as we will explore in \Cref{ch:score-based} and \Cref{ch:flow-based}.

It reveals a powerful equivalence: minimizing the KL divergence between marginal distributions is mathematically identical to minimizing the KL divergence between specific conditional distributions. This latter formulation is exceptionally useful because the crucial conditional distribution, $p(\mathbf{x}_{i-1}|\mathbf{x}_i, \mathbf{x})$, possesses a convenient closed-form expression:
\lem{Reverse Conditional Transition Kernel}{ddpm-reverse-kernel}{$p(\mathbf{x}_{i-1} | \mathbf{x}_i, \mathbf{x})$ is Gaussian with the closed-form expression:
\begin{align*}
    p(\rvx_{i-1} | \rvx_i, \rvx) =\mathcal{N}\left(\rvx_{i-1}; \bm{\mu}\left(\rvx_i, \rvx, i\right), \sigma^2(i)\rmI \right),
\end{align*}
where
\begin{align}\label{eq:ddpm-reverse-mean}
\begin{aligned}
            \bm{\mu}\left(\rvx_i, \rvx, i\right):= \frac{\bar{\alpha}_{i-1}\beta_i^2}{1-\bar{\alpha}_i^2}\rvx + \frac{(1-\bar{\alpha}_{i-1}^2)\alpha_i }{1-\bar{\alpha}_i^2}\rvx_i,
\quad  \sigma^2(i) := \frac{1-\bar{\alpha}_{i-1}^2}{1-\bar{\alpha}_i^2}\beta_i^2.
\end{aligned}
\end{align}
}
Later in Lemma~\ref{reverse-kernel}, we present a more general formula that extends beyond the DDPM noising process described in \Cref{eq:ddpm-forward}.

\subsection{Modeling of Reverse Transition Kernel $p_{\bm{\phi}}(\mathbf{x}_{i-1}|\mathbf{x}_i)$} Leveraging the gradient-level equivalence as in Theorem~\ref{thm:equiv-marginal-kl} and the Gaussian form of the reverse conditional $p(\mathbf{x}_{i-1}|\mathbf{x}_i, \mathbf{x})$ as in Lemma~\ref{ddpm-reverse-kernel}, DDPM assumes that each reverse transition $p_{\bm{\phi}}(\rvx_{i-1}| \rvx_i)$ is Gaussian, parameterized as
\begin{align}\label{eq:def-reverse}
    p_{\bm{\phi}}(\mathbf{x}_{i-1}|\mathbf{x}_i) := \mathcal{N}\big(\mathbf{x}_{i-1}; \bm{\mu}_{\bm{\phi}}(\mathbf{x}_i, i), \sigma^2(i) \mathbf{I}\big),
\end{align}
where $\bm{\mu}_{\bm{\phi}}(\cdot, i) \colon \mathbb{R}^D \to \mathbb{R}^D$ is a learnable mean function, and $\sigma^2(i) > 0$ is fixed as defined in \Cref{eq:ddpm-reverse-mean}.

We denote the KL divergence, averaged over time steps $i$ and conditioned on data $\rvx_0 \sim p_{\text{data}}$, to match all layers of distributions as:
\begin{align}\label{eq:ddpm-diffusion}
    \mathcal{L}_{\text{diffusion}}(\rvx_0;\bm{\phi}) := \sum_{i=2}^L \mathbb{E}_{p_i(\rvx_i|\rvx_0)}\left[\mathcal{D}_{\text{KL}}\big(p(\rvx_{i-1}|\rvx_i, \rvx_0)  \Vert  p_{\bm{\phi}}(\rvx_{i-1}|\rvx_i)\big)\right].
\end{align}
Thanks to the Gaussian forms of both distributions and the parameterization defined in \Cref{eq:def-reverse}, the objective admits a closed-form expression and can be simplified as:
\begin{align}\label{eq:ddpm-diffusion-simp}
    \mathcal{L}_{\text{diffusion}}(\rvx_0;\bm{\phi}) = \sum_{i=2}^L \frac{1}{2\sigma^2(i)}  \mathbb{E}_{p_i(\rvx_i|\rvx_0)}\left\| \bm{\mu}_{\bm{\phi}}(\rvx_i, i) - \bm{\mu}(\rvx_i, \rvx_0, i) \right\|_2^2 + C,
\end{align}
where $C$ is a constant independent of $\bm{\phi}$. Averaging over the data distribution and omitting the constant $C$ (which does not affect the optimization), the final DDPM training objective is
\begin{align}
\begin{aligned}
    \label{eq:ddpm-mean-loss}
    \mathcal{L}_{\text{DDPM}}(\bm{\phi}) := \sum_{i=2}^L \frac{1}{2\sigma^2(i)}  \mathbb{E}_{\rvx_0} \mathbb{E}_{p_i(\rvx_i|\rvx_0)} \left[ \left\| \bm{\mu}_{\bm{\phi}}(\mathbf{x}_i, i) - \bm{\mu}(\mathbf{x}_i, \rvx_0, i) \right\|_2^2 \right],
\end{aligned}
\end{align}
where $\rvx_0\sim p_{\mathrm{data}}$.

We remark that the forward perturbation kernel $p_i(\rvx_i | \rvx_0)
$
plays an important computational role in the training objective. It gives the
distribution of the noisy state $\rvx_i$ obtained from a clean sample $\rvx_0$
after $i$ forward steps. Because each forward transition is linear and
Gaussian, their composition also has a closed-form Gaussian distribution; see
the discussion around \Cref{eq:ddpm-forward}. Consequently, after selecting a
time step $i$, we can sample $\rvx_i \sim p_i(\rvx_i | \rvx_0)
$
directly from $\rvx_0$, without first generating the intermediate states
$\rvx_1,\ldots,\rvx_{i-1}$.  Thus, each
loss term can be estimated using only a clean sample, a selected time step, and
a directly corrupted sample, making stochastic training computationally
efficient.

\subsection{Practical Choices of Predictions and Loss}\label{subsec:ddpm-prediction}

\paragraph{$\beps$-Prediction.}
In typical DDPM implementations, training is not conducted directly using the original loss based on the \emph{mean prediction} parameterization from \Cref{eq:ddpm-mean-loss}. Instead, an equivalent reparameterization, known as the \emph{$\beps$-prediction} (noise prediction) formulation, is commonly adopted.  

Recall that in the DDPM forward process, a noisy sample $\rvx_i \sim p(\rvx_i |\rvx_0)$ at noise level $i$ is generated by
\begin{equation}\label{eq:ddpm-perturbation}
    \rvx_i = \bar{\alpha}_i \rvx_0 + \sqrt{1 - \bar{\alpha}_i^2}   \boldsymbol{\epsilon}, \quad \rvx_0 \sim p_{\text{data}}, \quad \boldsymbol{\epsilon} \sim \mathcal{N}(\mathbf{0}, \mathbf{I}).
\end{equation}
Using this expression, the reverse mean $\bm{\mu}(\rvx_i, \rvx_0, i)$ from \Cref{eq:ddpm-reverse-mean} can be rewritten as:
\[
\bm{\mu}(\rvx_i, \rvx_0, i) = \frac{1}{\alpha_i} \left( \rvx_i - \frac{1 - \alpha_i^2}{\sqrt{1 - \bar{\alpha}_i^2}} \boldsymbol{\epsilon} \right).
\]

This motivates a parameterization of the model mean $\bm{\mu}_{\bm{\phi}}$ using a neural network $\boldsymbol{\epsilon}_{\bm{\phi}}(\rvx_i, i)$ that directly predicts the noise:
\[
\bm{\mu}_{\bm{\phi}}(\rvx_i, i) = \frac{1}{\alpha_i} \left( \rvx_i - \frac{1 - \alpha_i^2}{\sqrt{1 - \bar{\alpha}_i^2}} \underbrace{\boldsymbol{\epsilon}_{\bm{\phi}}(\rvx_i, i)}_{\beps\text{-prediction}} \right).
\]
Substituting this into the original loss leads to a squared $\ell_2$ error between predicted and true noise:
\[
\left\| \bm{\mu}_{\bm{\phi}}(\rvx_i, i) - \bm{\mu}(\rvx_i, \rvx_0, i) \right\|_2^2 \propto \left\| \boldsymbol{\epsilon}_{\bm{\phi}}(\rvx_i, i) - \boldsymbol{\epsilon} \right\|_2^2,
\]
up to a weighting factor depending on $i$. Intuitively, the model acts as a ``noise detective'', estimating the random noise added at each step of the forward process. Subtracting this estimate from the corrupted sample moves it closer to the clean original, and repeating this step-by-step reconstructs the data from pure noise.

\paragraph{Simplified Loss with $\beps$-Prediction.}
In practice, this expression is further simplified by omitting the weighting term, yielding the widely used DDPM training loss:
\begin{equation}\label{eq:ddpm-simple-loss}
    \mathcal{L}_{\text{simple}}(\bm{\phi}) := \mathbb{E}_{i} \mathbb{E}_{\rvx \sim p_{\text{data}}(\rvx)} \mathbb{E}_{\boldsymbol{\epsilon} \sim \mathcal{N}(\mathbf{0}, \mathbf{I})} \left[ \left\| \boldsymbol{\epsilon}_{\bm{\phi}}(\rvx_i, i) - \boldsymbol{\epsilon} \right\|_2^2 \right],
\end{equation}
where $\rvx_i = \bar{\alpha}_i \rvx_0 + \sqrt{1 - \bar{\alpha}_i^2} \boldsymbol{\epsilon}$ with $\rvx_0 \sim p_{\text{data}}$. Since the target noise has unit variance at every timestep $t$, the $\ell_2$ loss in \Cref{eq:ddpm-simple-loss} maintains a consistent scale across all $t$. This prevents vanishing or exploding targets and eliminates the need for explicit loss weighting.

Importantly, both $\mathcal{L}_{\text{DDPM}}$ and $\mathcal{L}_{\text{simple}}$ share the same optimal solution $\boldsymbol{\epsilon}^*$, this is because \Cref{eq:ddpm-simple-loss} essentially reduces to a least-squares problem (as shown similarly in Proposition~\ref{dsm-minimizer} or Proposition~\ref{equi-para})\footnote{We remark that removing the ELBO-derived weights in $\mathcal{L}_{\text{DDPM}}$ changes the training objective,
but not its population-optimal prediction: $\beps^*(\rvx_i,i)
=
\E[\beps| \rvx_i]$.
Thus, the weighted and simplified DDPM losses have the same ideal predictor, but
they emphasize different noise levels during training. In practice, this weighting
can still affect optimization and performance.}:
\[
\boldsymbol{\epsilon}^*(\rvx_i, i) = \mathbb{E}\left[ \boldsymbol{\epsilon}|\rvx_i \right], \quad \rvx_i \sim p_i.
\]
Intuitively, the $\beps$-prediction network $\boldsymbol{\epsilon}_{\bm{\phi}}(\rvx_i, i)$ estimates the noise added by the forward process to produce $\rvx_i$. At optimality, this estimate coincides with the conditional expectation of the true noise, even though $\rvx_i$ does not uniquely determine the original clean sample.

\paragraph{Another Equivalent Parametrization: $\rvx$-Prediction.} 
\Cref{eq:ddpm-reverse-mean} motivates an alternative yet equivalent parameterization, known as \emph{$\rvx$-prediction} (clean prediction), in which a neural network $\rvx_{\bm{\phi}}(\rvx_i, i)$ is trained to predict a clean (denoised) sample from a given noisy input $\rvx_i \sim p_i(\rvx_i)$ at noise level $i$. Replacing the ground-truth clean sample $\rvx_0$ in the reverse mean expression with $\rvx_{\bm{\phi}}(\rvx_i, i)$ leads to the following model:
\begin{align*}
\bm{\mu}_{\bm{\phi}}(\rvx_i, i) = \frac{\bar{\alpha}_{i-1} \beta_i^2}{1 - \bar{\alpha}_i^2} \rvx_{\bm{\phi}}(\rvx_i, i) + \frac{(1 - \bar{\alpha}_{i-1}^2)\alpha_i}{1 - \bar{\alpha}_i^2} \rvx_i.
\end{align*}
Analogous to the $\beps$-prediction formulation, the training objective can be expressed as
\[
\left\| \bm{\mu}_{\bm{\phi}}(\rvx_i, i) - \bm{\mu}(\rvx_i, \rvx_0, i) \right\|_2^2 \propto \left\| \rvx_{\bm{\phi}}(\rvx_i, i) - \rvx_0 \right\|_2^2, \quad \rvx_0 \sim p_{\text{data}},
\]
where the model is trained to predict the original data sample $\rvx_0$ from its noisy version $\rvx_i$. This equivalence reduces the mean-matching loss in \Cref{eq:ddpm-mean-loss} to
\[
\mathbb{E}_{i} \mathbb{E}_{\rvx_0 \sim p_{\text{data}}} \mathbb{E}_{\boldsymbol{\epsilon} \sim \mathcal{N}(\mathbf{0}, \mathbf{I})} \left[\omega_i \left\| \rvx_{\bm{\phi}}(\rvx_i, i) - \rvx_0 \right\|_2^2 \right],
\]
for some weighting function $\omega_i$. Since this is a least-squares problem, the optimal solution is given by (see Proposition~\ref{dsm-minimizer} or Proposition~\ref{equi-para}) 
\begin{align}\label{eq:ddpm-clean-minimizer}
    \rvx^*(\rvx_i, i) = \mathbb{E}\left[ \rvx_0|\rvx_i \right], \quad \rvx_i \sim p_i,
\end{align}
that is, the model should predict the expected clean data given a noisy observation $\rvx_i$ at timestep $i$.

The $\rvx$-prediction and $\beps$-prediction parameterizations are mathematically equivalent and connected via the forward process:
\begin{equation}\label{eq:ddpm-clean-noise}
    \rvx_i = \bar{\alpha}_i \rvx_{\bm{\phi}}(\rvx_i, i) + \sqrt{1 - \bar{\alpha}_i^2}   \boldsymbol{\epsilon}_{\bm{\phi}}(\rvx_i, i).
\end{equation}
That is, one may either predict the clean sample $\rvx_{\bm{\phi}}(\rvx_i, i)$ or the noise $\boldsymbol{\epsilon}_{\bm{\phi}}(\rvx_i, i)$, such that their combination reproduces $\rvx_i$ under the forward noising process.

\subsection{DDPM's ELBO}\label{subsec:ddpm-elbo} With the reverse transitions defined as in \Cref{eq:def-reverse}, this leads to the definition of the joint generative distribution in DDPM as:
\[
p_{\bm{\phi}}(\rvx_0, \rvx_{1:L}) := p_{\bm{\phi}}(\rvx_0|\rvx_1)  p_{\bm{\phi}}(\rvx_1|\rvx_2)\cdots p_{\bm{\phi}}(\rvx_{L-1}|\rvx_L)   p_{\text{prior}}(\rvx_L),
\]
and the marginal generative model for data is given by:
\begin{align*}
    p_{\bm{\phi}}(\rvx_0) := \int p_{\bm{\phi}}(\rvx_0, \rvx_{1:L}) \diff \rvx_{1:L}.
\end{align*}

Indeed, DDPM training via \Cref{eq:ddpm-diffusion} can be rigorously grounded in maximum likelihood estimation (\Cref{eq:MLE}). Specifically, its objective forms an ELBO, similar to those in VAEs and HVAEs from \Cref{sec:vae}, which serves as a lower bound on the log-density:
\thmp{DDPM's ELBO}{ddpm-elbo}{    
\begin{align}\label{eq:ddpm-elbo}
        \begin{aligned}
        -\log p_{\bm{\phi}}(\rvx_0)  &\leq-\mathcal{L}_{\text{ELBO}}(\rvx_0; \bm{\phi}) 
        \\&:= \mathcal{L}_{\text{prior}}(\rvx_0) + \mathcal{L}_{\text{recon.}}(\rvx_0;\bm{\phi})  + \mathcal{L}_{\text{diffusion}}(\rvx_0;\bm{\phi})
        \end{aligned}
\end{align}
Here, each component of losses are defined as:
\begin{align*}
   \mathcal{L}_{\text{prior}}(\rvx_0) &:= \mathcal{D}_{\text{KL}}\Big( p(\rvx_L|\rvx_0) \Vert p_{\text{prior}}(\rvx_L)\Big)\\
   \mathcal{L}_{\text{recon.}}(\rvx_0;\bm{\phi}) &:= \mathbb{E}_{p(\rvx_1|\rvx_0)}\left[-\log p_{\bm{\phi}}(\rvx_0|\rvx_1)\right]\\
   \mathcal{L}_{\text{diffusion}}(\rvx_0;\bm{\phi}) &= \sum_{i=2}^L  \mathbb{E}_{p_i(\rvx_i|\rvx_0)}\left[\mathcal{D}_{\text{KL}}\Big(p(\rvx_{i-1}|\rvx_i, \rvx_0) \Vert p_{\bm{\phi}}(\rvx_{i-1}|\rvx_i) \Big)\right]. 
\end{align*}}{The derivation applies Jensen’s inequality, as in the HVAE/VAE ELBO (\Cref{eq:hvae-elbo}), with further simplifications. The detailed proof is deferred to \Cref{app:vlb-proof}.
}
The ELBO $\mathcal{L}_{\text{ELBO}}$ consists of three terms:
\begin{itemize}
    \item $\mathcal{L}_{\text{prior}}$ can be made negligible by choosing the noise schedule $\{\beta_i\}$ such that $p(\cdot|\rvx_0) \approx p_{\text{prior}}(\cdot)$.
    \item For $\mathcal{L}_{\text{recon.}}$, this can be approximated and optimized using a Monte Carlo estimate; see \citep{ho2020denoising, kingma2021variational} for practical implementations. 
    \item $\mathcal{L}_{\text{diffusion}}$ (cf. \Cref{eq:ddpm-diffusion}) matches the reverse conditionals $p_{\bm{\phi}}(\mathbf{x}_{i-1}|\mathbf{x}_i)$ to $p(\mathbf{x}_{i-1}|\mathbf{x}_i)$ at all steps $i$.
\end{itemize}

The ELBO objective $\mathcal{L}_{\text{ELBO}}$ can also be interpreted through the lens of the Data Processing Inequality with latents $\rvz = \rvx_{1:L}$, as illustrated in \Cref{eq:joint-bound}:
\begin{align*}
    \mathcal{D}_{\mathrm{KL}}(p_{\mathrm{data}}(\rvx_0) \| p_{\bm{\phi}}(\rvx_0)) \leq \mathcal{D}_{\mathrm{KL}}\left(p(\rvx_0, \rvx_{1:L}) \| p_{\bm{\phi}}(\rvx_0, \rvx_{1:L})\right),
\end{align*}
where $p(\rvx_0, \rvx_{1:L}) := p_{\mathrm{data}}(\rvx_0)  p(\rvx_1|\rvx_0)  p(\rvx_2|\rvx_1) \cdots p(\rvx_L|\rvx_{L-1})$ denotes the joint distribution along the forward process.

\rmkb{Diffusion's variational view fits the hierarchical latent-variable
picture in a particularly convenient way: the ``encoder'' is the fixed forward
noising chain, so there is no learned inference network. The ELBO nevertheless
does decompose into per-step KL terms, together with the terminal-prior
and reconstruction terms in \Cref{eq:ddpm-elbo}. What makes DDPM training
especially tractable is that, under the linear--Gaussian forward process, these
per-step KL terms can be rewritten as simple denoising regression problems with
closed-form conditional targets. Thus the key simplification is not the absence
of per-level KL terms, but the fact that the forward process is fixed and turns
them into supervised denoising objectives that can be sampled efficiently.}

\subsection{Sampling} 
After training the $\beps$-prediction model, $\bm{\epsilon}_{\bm{\phi}^\times}(\rvx_i, i)$\footnote{We use the symbol ``$\times$'' to indicate that the model has been trained and is now frozen.}, sampling is performed sequentially as illustrated in \Cref{fig:ddpm-sampling-formal-rev}, using the parametrized transition $p_{\bm{\phi}^\times}(\rvx_{i-1} | \rvx_i)$ instead.

More specifically, starting from a random seed $\rvx_L \sim p_{\text{prior}} = \mathcal{N}(\bm{0},\rmI)$, we recursively sample from $p_{\bm{\phi}^\times}(\rvx_{i-1} | \rvx_i)$ following the update rule below for $i = L, L-1, \dots, 1$:
\begin{align}\label{eq:ddpm-sampling}
    \rvx_{i-1} \leftarrow \underbrace{\frac{1}{\alpha_i}\left( \rvx_i  - \frac{1-\alpha_i^2}{\sqrt{1-\bar{\alpha}_i^2}} \bm{\epsilon}_{\bm{\phi}^\times}(\rvx_i, i)\right)}_{\bm{\mu}_{\bm{\phi}^\times}(\rvx_i, i)} +\sigma(i) \bm{\epsilon}_i,\quad \bm{\epsilon}_i\sim  \mathcal{N}(\bm{0},\rmI).
\end{align}
This ``denoising'' process continues until $\rvx_0$ is obtained as the final clean generated sample.

\paragraph{Another Interpretation of DDPM Sampling.}

A useful way to interpret DDPM sampling is to rewrite the update rule
so that the noise-level structure becomes explicit.
Recall from \Cref{eq:ddpm-clean-noise} that the predicted noise can be
expressed in terms of the current sample $\rvx_i$ and the predicted
clean sample $\rvx_{\bm{\phi}^\times}(\rvx_i,i)$:
\[
\boldsymbol{\epsilon}_{\bm{\phi}^\times}(\rvx_i,i)
=
\frac{\rvx_i - \bar{\alpha}_i\,\rvx_{\bm{\phi}^\times}(\rvx_i,i)}
     {\sqrt{1-\bar{\alpha}_i^2}}.
\]
Substituting this into the DDPM sampling rule
(\Cref{eq:ddpm-sampling}) and rearranging, we obtain
\begin{equation}\label{eq:ddpm-renoise}
\rvx_{i-1}
\;\leftarrow\;
\bar{\alpha}_{i-1}\,\rvx_{\bm{\phi}^\times}(\rvx_i,i)
\;+\;
\underbrace{
  \sqrt{1-\bar{\alpha}_{i-1}^2 - \tilde{\beta}_i^2}\;
  \boldsymbol{\epsilon}_{\bm{\phi}^\times}(\rvx_i,i)
  \;+\;
  \tilde{\beta}_i\,\boldsymbol{\epsilon}_i
}_{\text{combined std.\ }=\;\sqrt{1-\bar{\alpha}_{i-1}^2}},
\end{equation}
where $\tilde{\beta}_i^2
= \frac{(1-\bar{\alpha}_{i-1}^2)(1-\alpha_i^2)}{1-\bar{\alpha}_i^2}$
is the DDPM posterior variance\footnote{%
The coefficient
$\sqrt{1-\bar{\alpha}_{i-1}^2 - \tilde{\beta}_i^2}$
simplifies to
$\frac{\alpha_i(1-\bar{\alpha}_{i-1}^2)}{\sqrt{1-\bar{\alpha}_i^2}}$,
which is always non-negative, so the square root is well-defined.}.
Moreover, the residual coefficients satisfy
\[
\bigl(1-\bar{\alpha}_{i-1}^2-\tilde{\beta}_i^2\bigr)+\tilde{\beta}_i^2
=
1-\bar{\alpha}_{i-1}^2,
\]
so the update has the same noise-level structure as the forward marginal
at level $i-1$. This returns to the first operational sampling template introduced in
\Cref{eq:ddpm-first-try}: predict the clean part and rewrite the sample
at a lower noise level. DDPM carries out this rewrite stochastically,
combining the noise direction estimated from $\rvx_i$ with a fresh
Gaussian noise sample. A deterministic sampler follows the same basic
idea, but removes the fresh randomness and adjusts the remaining noise
coefficient accordingly. We will revisit this construction when studying
diffusion samplers more systematically in \Cref{sec:revisit-ddim}.

Compare \Cref{eq:ddpm-renoise} with the forward marginal
$\rvx_{i-1} = \bar{\alpha}_{i-1}\,\rvx_0
+ \sqrt{1-\bar{\alpha}_{i-1}^2}\,\bar{\boldsymbol{\epsilon}}$.
The structure is identical: signal coefficient $\bar{\alpha}_{i-1}$
and noise standard deviation $\sqrt{1-\bar{\alpha}_{i-1}^2}$,
except that the unknown $\rvx_0$ is replaced by the network
prediction $\rvx_{\bm{\phi}^\times}(\rvx_i,i)$.
In other words, each DDPM step \emph{predicts the clean data and
then re-noises it to exactly noise level $i{-}1$}.
One can therefore view DDPM sampling as iterating two operations:
\begin{enumerate}
    \item \textbfs{Denoise.}
          Estimate the clean data
          $\rvx_{\bm{\phi}^\times}(\rvx_i, i)$
          from the current noisy input $\rvx_i$.
    \item \textbfs{Re-noise.}
          Add back noise at the reduced level $i{-}1$
          via \Cref{eq:ddpm-renoise},
          producing a sample $\rvx_{i-1}$ whose
          noise level matches the forward process at step $i{-}1$.
\end{enumerate}

However, even if $\rvx_{\bm{\phi}^\times}$ is trained as the optimal
denoiser (i.e., the conditional expectation minimizer;
see \Cref{eq:ddpm-clean-minimizer}), it can only predict the
\emph{average} clean sample given $\rvx_i$.
This limitation leads to blurry predictions, particularly at high
noise levels, where recovering detailed structure from severely
corrupted inputs becomes difficult.
From this viewpoint, diffusion sampling moves from high to low noise
and progressively refines an estimate of the clean signal.
Early steps set the global structure, later steps add fine detail,
and the sample becomes more realistic as the noise is removed.

\begin{figure}[th!]
    \centering
    \includegraphics[width=\linewidth]{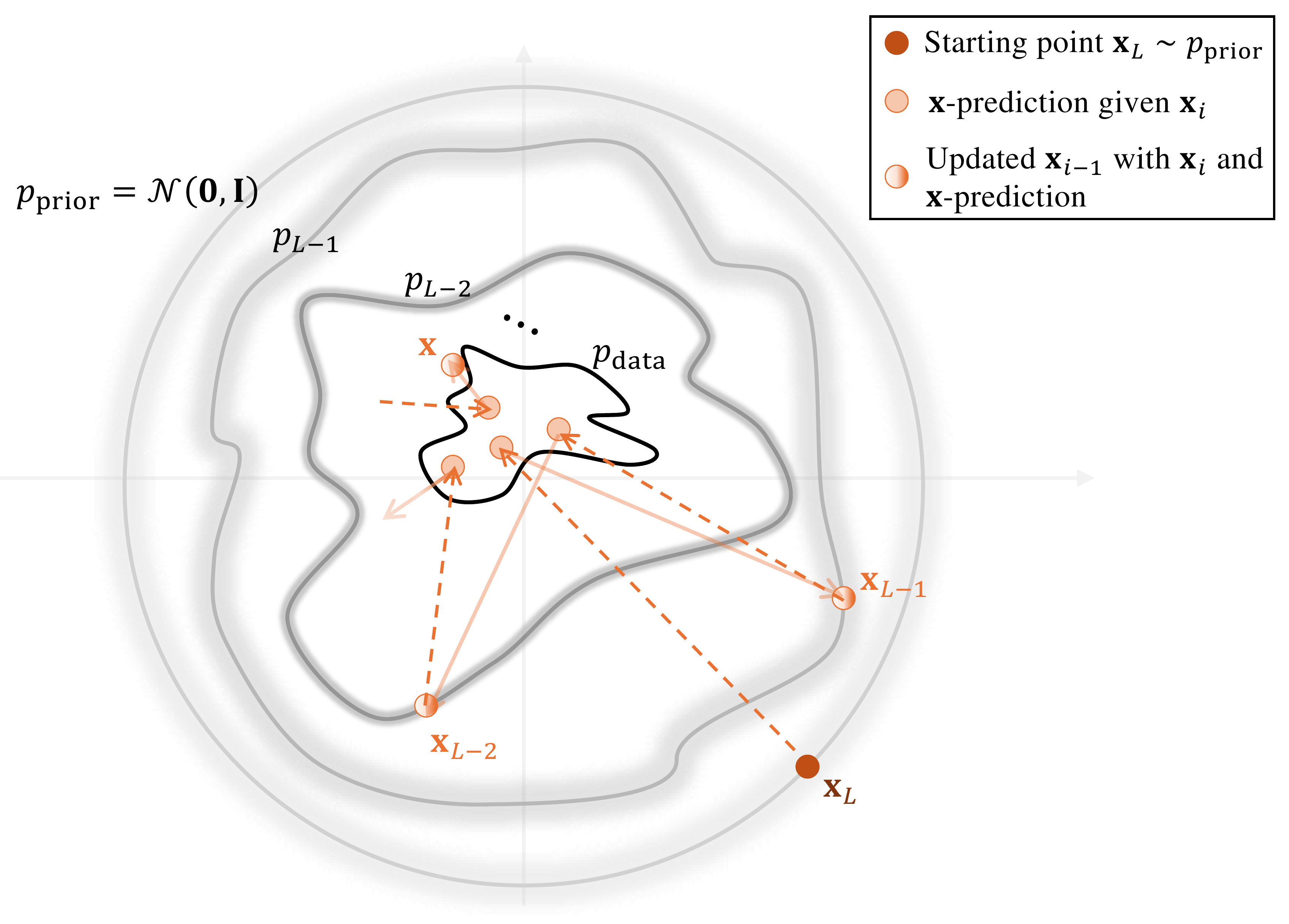}
\caption{\textbfs{Illustration of the denoise-then-re-noise view of
    DDPM sampling.} From the current noisy sample $\mathbf{x}_i$, the model first estimates the underlying clean signal $\mathbf{x}_{\bm{\phi}^\times}(\mathbf{x}_i,i)$, and then samples a less noisy point $\mathbf{x}_{i-1}$ by re-noising this estimate to the $(i-1)$-th noise level. \figcredit{Created by the authors.}}
    \label{fig:vdm-clean}
\end{figure}

\paragraph{Slow Sampling Speed of DDPM.} Sampling in DDPMs (i.e., diffusion models) is inherently slow. In the original formulation and implementation, generating a sample typically requires around $1{,}000$ denoising steps. This is because sampling proceeds through a long sequence of small refinements: at each step, the model updates the current noisy sample slightly toward a cleaner one, and the next update must start from the result of the previous step. As a result, standard DDPM sampling is time-stepping and strongly sequential, and good sample quality usually requires many such steps to closely track the reverse diffusion path.

Theorem~\ref{thm:equiv-marginal-kl} shows that an expressive $ p_{\bm{\phi}}(\mathbf{x}_{i-1} | \mathbf{x}_i) $ can theoretically match the true reverse distribution $ p(\mathbf{x}_{i-1} | \mathbf{x}_i) $. However, in practice, $ p_{\bm{\phi}}(\mathbf{x}_{i-1} | \mathbf{x}_i) $ is typically modeled as a Gaussian to approximate $ p(\mathbf{x}_{i-1} | \mathbf{x}_i) $, limiting its expressiveness.

For small forward noise scales $\beta_i$, the true reverse distribution is approximately Gaussian, enabling accurate approximation. Conversely, large $\beta_i$ induce multimodality or strong non-Gaussianity that a single Gaussian cannot capture. To maintain accuracy, DDPM employs many small $\beta_i$ steps, forming a sequential chain where each step depends on the previous and requires a neural network evaluation $\boldsymbol{\epsilon}_{\bm{\phi}^\times}(\mathbf{x}_i, i)$. This results in $\mathcal{O}(L)$ sequential passes, preventing parallelization and slowing generation.

Later in \Cref{ch:score-sde} we show a more principled interpretation of this inherent sampling bottleneck as a differential-equation problem, which motivates continuous-time numerical strategies for accelerating generation.

\newpage
\section{Closing Remarks}\label{sec:ch2_cr}

In this chapter, we have traced the origins of diffusion models through the variational lens. We began with the Variational Autoencoder (VAE), a foundational generative model that learns a probabilistic mapping between data and a structured latent space via the Evidence Lower Bound (ELBO). We saw how Hierarchical VAEs (HVAEs) extended this idea by stacking latent layers, introducing the powerful concept of progressive, coarse-to-fine generation. However, these models face challenges with training stability and sample quality.

We then framed Denoising Diffusion Probabilistic Models (DDPMs) as a pivotal evolution within this variational framework. By fixing the encoder to a gradual noising process and learning only the reverse denoising steps, DDPMs elegantly sidestep the training instabilities of HVAEs. Crucially, we demonstrated that DDPMs are also trained by maximizing a variational bound on the log-likelihood, with a training objective that decomposes into a series of simple denoising tasks. This tractability is enabled by a powerful conditioning strategy that transforms an intractable marginal objective into a tractable conditional one, a recurring theme in diffusion models.

While this variational framework provides a complete and powerful foundation for DDPMs, it is not the only way to understand them. An alternative and equally fundamental perspective emerges from the principles of energy-based modeling. In the next chapter, we will explore this score-based perspective:
\begin{enumerate}
    \item We will shift our focus from learning the denoising transition probabilities $p_{\bm{\phi}}(\mathbf{x}_{i-1}\vert\mathbf{x}_{i})$ to directly learning the gradient of the data's log-density, i.e., the score function.
    \item We will see how this approach, originating from EBMs, gives rise to Noise Conditional Score Networks (NCSN) and reveals a deep, mathematical equivalence between the noise prediction ($\bm{\epsilon}$-prediction) learned in DDPMs and the score function itself.
\end{enumerate}

This alternative viewpoint will not only offer new insights but also serve as another cornerstone for the unified, continuous-time framework of diffusion models to be developed later.

\chapter{Score-Based Perspective: From EBMs to NCSN}\label{ch:score-based}


In the previous chapters we traced diffusion models to their variational roots and showed how they arise within the framework of VAEs. We now turn to a second, equally fundamental viewpoint: \emph{Energy-Based Models} (EBMs)~\citep{ackley1985learning,lecun2006tutorial}. An EBM represents a distribution by an energy landscape that is low on data and high elsewhere. Sampling typically relies on Langevin dynamics, which moves samples toward high density regions by following the gradient of this landscape. This gradient field, known as the \emph{score}, points toward directions of higher probability.

The central observation is that knowing the score is enough for generation: it moves samples toward likely regions without computing the intractable normalization constant. \emph{Score-based} diffusion models build directly on this idea. Instead of focusing only on the clean data distribution, they consider a sequence of Gaussian noise–perturbed distributions whose scores are easier to approximate. Learning these scores yields a family of vector fields that guide noisy samples step by step back to data, turning generation into progressive denoising.

\chapterlearning{
  \item understand the core idea of score-based modeling: what a score function is, and how it can be used with Langevin dynamics for generation;
  \item understand how denoising score matching learns the score at a single noise level through conditioning trick, and how Tweedie's formula connects score prediction to denoising;
  \item understand how Noise Conditional Score Networks (NCSNs) extend denoising score matching to multiple noise levels, why multiple noise levels are useful, and how NCSN score prediction relates to DDPM noise prediction.
}{
Readers already familiar with energy-based models may skim \Cref{sec:ebm} and begin with \Cref{subsec:training-sm}.
}


\newpage

\section{Energy-Based Models}\label{sec:ebm}

For readers already familiar with EBMs, this section is meant as a concise refresher and a bridge to the score-based view of diffusion.
\subsection{Modeling Probability Distributions Using Energy Functions}

Let $\mathbf{x} \in \mathbb{R}^D$ denote a data point. EBMs define a probability density via an energy function $E_{\bm{\phi}}(\mathbf{x})$, parameterized by ${\bm{\phi}}$, which assigns lower energy to more likely configurations. The resulting distribution is given by
\begin{align*}
    p_{{\bm{\phi}}}(\mathbf{x}) := \frac{\exp(-E_{\bm{\phi}}(\mathbf{x}))}{Z_\bphi}, \quad 
    Z_\bphi := \int_{\mathbb{R}^D} \exp(-E_{\bm{\phi}}(\mathbf{x})) \diff\mathbf{x},
\end{align*}
where $Z_\bphi$ is called the \emph{partition function} ensuring normalization:
\[
\int_{\mathbb{R}^D} p_{{\bm{\phi}}}(\mathbf{x})  \diff\mathbf{x} = 1.
\]
\begin{figure}[th!]
    \centering
    \includegraphics[width=\linewidth]{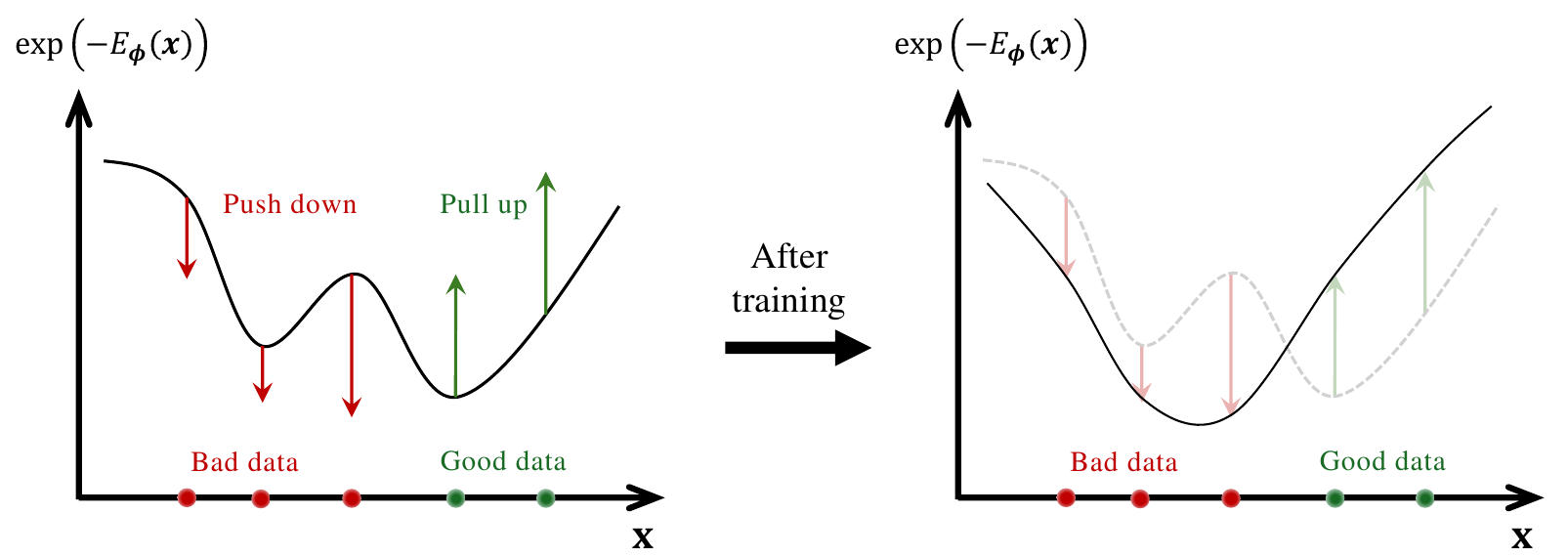}
    \caption{\textbfs{Illustration of EBM training.} The model lowers density (raises energy) at ``bad'' data points (red arrows), and raises density (lowers energy) at ``good'' data points (green arrows). \figcredit{Created by the authors.}}
    \label{fig:ebm-graph}
\end{figure}

In this view, points with lower energy correspond to higher probability, much like a ball rolling down into a valley. The partition function $Z_{\bphi}$ ensures that all probabilities add up to one, and as a result only the \emph{relative} values of energy matter. For instance, adding a constant to all energies multiplies both numerator and denominator by the same factor, leaving the distribution unchanged. 

Moreover, because the partition function $Z_{\bphi}$ enforces that probabilities sum to one, it follows mathematically that decreasing the energy within a region increases its probability, while the probability of its complement decreases accordingly. Thus, EBMs obey a strict global trade-off: making one valley deeper inevitably makes others shallower, and probability mass is redistributed across the entire space rather than assigned independently to each region.

\paragraph{Challenges of Maximum Likelihood Training in EBMs.}
In principle, EBMs can be trained by maximum likelihood, which naturally balances fitting the data with global regularization (see \Cref{eq:MLE}):
\begin{align}
\label{eq:mle-ebm}
\mathcal{L}_{\text{MLE}}(\bphi) 
&= \mathbb{E}_{p_{\text{data}}(\rvx)}  \left[ \log \frac{\exp(-E_{\bphi}(\rvx))}{Z_\bphi} \right] \\
&= -  \underbrace{\mathbb{E}_{p_{\text{data}}}[E_{\bphi}(\rvx)]}_{\text{lowers energy of data}}
   -  \underbrace{\log \int \exp(-E_{\bphi}(\rvx))\diff\rvx}_{\text{global regularization}}, \nonumber
\end{align}
with $Z_\bphi = \int \exp(-E_\bphi(\rvx))\diff\rvx$. 
The first term lowers the energy of real data, while the second enforces normalization via the partition function.

However, in high dimensions computing $\log Z_{\bphi}$ and its gradient is intractable, as it requires expectations under the model distribution. 
This motivates alternative objectives that either approximate the term, such as contrastive divergence \citep{hinton2002training}, or avoid it altogether through \emph{score matching}. 

In what follows, we first introduce the notion of the score function in \Cref{subsec:motivation-score} and present score matching as a tractable training objective that bypasses the partition function in \Cref{subsec:training-ebm}, and then discuss Langevin dynamics as a practical sampling method with score functions in \Cref{subsec:sampling-ebm}.

\subsection{Motivation: What Is the Score?}\label{subsec:motivation-score}

\begin{figure}[th]
    \centering
    \includegraphics[width=0.95\linewidth]{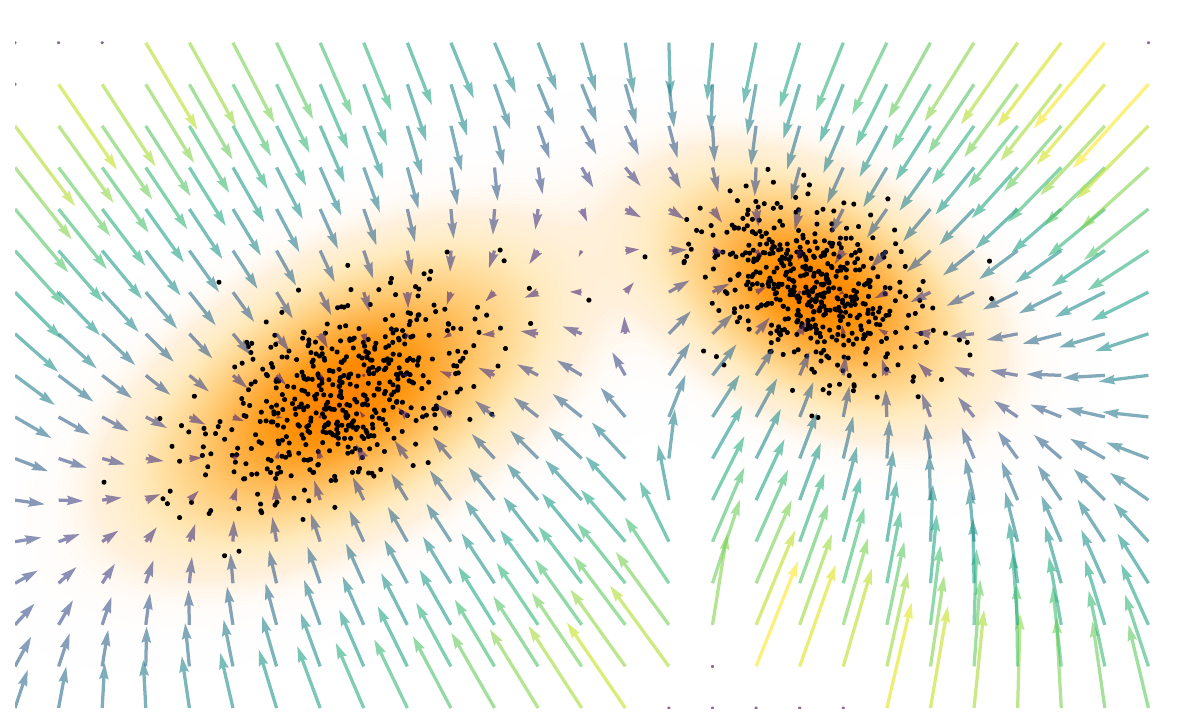}
    \caption{\textbfs{Illustration of score vector fields.} Score vector fields $\nabla_{\rvx} \log p(\rvx)$ indicate directions of increasing density. \figcredit{Created by the authors.}}
    \label{fig:score-function}
\end{figure}

For a density $p(\rvx)$ on $\mathbb{R}^D$, the \emph{score function} is the gradient of the log-density:
\[
\rvs(\rvx) := \nabla_{\rvx} \log p(\rvx), \qquad \rvs \colon \mathbb{R}^D \to \mathbb{R}^D.
\]
Intuitively, the score forms a vector field that points toward regions of higher probability, providing a local guide to where the data is most likely to occur (see \Cref{fig:score-function}).

\paragraph{Why Model Scores Instead of Densities?}
Modeling the score offers both theoretical and practical benefits:

\subparagraph{1. Freedom from Normalization Constants.}  
    Many distributions are defined only up to an unnormalized density $\tilde{p}(\rvx)$, e.g., $\exp(-E_{\bm{\phi}}(\rvx))$ in EBMs:
    \[
    p(\rvx) = \frac{\tilde{p}(\rvx)}{Z}, 
    \qquad 
    Z = \int \tilde{p}(\rvx)\diff\rvx.
    \]
    While computing $Z$ is intractable, the score depends only on $\tilde{p}$:
\begin{align}\label{eq:score-compute}
\nabla_{\rvx}\log p(\rvx) = \nabla_{\rvx}\log \tilde{p}(\rvx) - \underbrace{\nabla_{\rvx}\log Z}_{=0} = \nabla_{\rvx}\log \tilde{p}(\rvx), \end{align} 
since $Z$ is constant in $\rvx$. This bypasses the partition function entirely.

\subparagraph{2. A Complete Representation.}  
The score function fully characterizes the underlying distribution.  
Since it is the gradient of the log-density, the density can be recovered (up to a constant) via
\[
\log p(\rvx) 
= \log p(\rvx_0) + \int_0^1 \rvs(\rvx_0 + t(\rvx - \rvx_0))^\top (\rvx - \rvx_0)  \diff t,
\]
where $\rvx_0$ is a reference point and $\log p(\rvx_0)$ is fixed by normalization.  
Thus, modeling the score is as expressive as modeling $p(\rvx)$ itself, while often more tractable for generative modeling.

\subsection{Training EBMs via Score Matching}\label{subsec:training-ebm}
In EBMs, the density is defined as $p_{{\bm{\phi}}}(\mathbf{x}) = \tfrac{\exp(-E_{\bm{\phi}}(\mathbf{x}))}{Z_\bphi}$. Maximum likelihood training requires computing $Z_{\bm{\phi}}$, which is generally intractable.  
A key observation is that the model score of $p_{{\bm{\phi}}}$ simplifies to: $-\nabla_{\rvx}E_{\bm{\phi}}(\rvx)$, independent of $Z_{\bm{\phi}}$ (see \Cref{eq:score-compute}).  

\emph{Score matching}~\citep{hyvarinen2005estimation} leverages the fact that scores depend only on the energy function.  
Instead of fitting normalized probabilities, it trains EBMs by aligning the model score with the (unknown) data score:
\begin{align}\label{eq:ebm-sm}
    \mathcal{L}_{\mathrm{SM}}(\bm{\phi}) 
= \tfrac{1}{2}  \mathbb{E}_{p_{\mathrm{data}}(\mathbf{x})} 
\big\| \nabla_{\mathbf{x}} \log p_{\bm{\phi}}(\mathbf{x}) 
- \nabla_{\mathbf{x}} \log p_{\mathrm{data}}(\mathbf{x}) \big\|_2^2.
\end{align}

Although the data score is inaccessible, integration by parts yields an equivalent expression involving only the energy and its derivatives (see Proposition~\ref{sm-trace} for more details):
\[
\mathcal{L}_{\mathrm{SM}}(\bm{\phi}) =
\mathbb{E}_{p_{\mathrm{data}}(\mathbf{x})} \left[
-\Tr \big(\nabla_{\mathbf{x}}^2 E_{\bm{\phi}}(\rvx)\big)
+ \tfrac{1}{2}  \|\nabla_{\mathbf{x}} E_{\bm{\phi}}(\rvx)\|_2^2
\right] + C,
\]
where $\nabla_{\mathbf{x}}^2 E_{\bm{\phi}}(\rvx)$ is the Hessian of $E_{\bm{\phi}}$ and $C$ is a constant independent of $\bm{\phi}$.  

This formulation is attractive because it eliminates the partition function and avoids sampling from the model during training.  
Its main drawback is the need for second-order derivatives: although one only needs the trace of the Hessian ($\Tr \big(\nabla_{\mathbf{x}}^2 E_{\bm{\phi}}\big)$), rather than the full Hessian matrix, exact computation of this term generally requires aggregating second-derivative information across input coordinates, so its cost typically increases with the input dimension $D$.  
We will revisit approaches to addressing this limitation later in the chapter.

\subsection{Langevin Sampling with Score Functions}\label{subsec:sampling-ebm}
Sampling from EBMs, defined by the energy function $E_{\bm{\phi}}(\mathbf{x})$, can be performed using \emph{Langevin dynamics}. We first present the discrete-time Langevin update and then its continuous-time limit as a stochastic differential equation (SDE). Finally, we discuss the physical intuition behind how Langevin dynamics enables efficient exploration of complex energy landscapes.

\begin{figure}
    \centering
    \includegraphics[width=\linewidth]{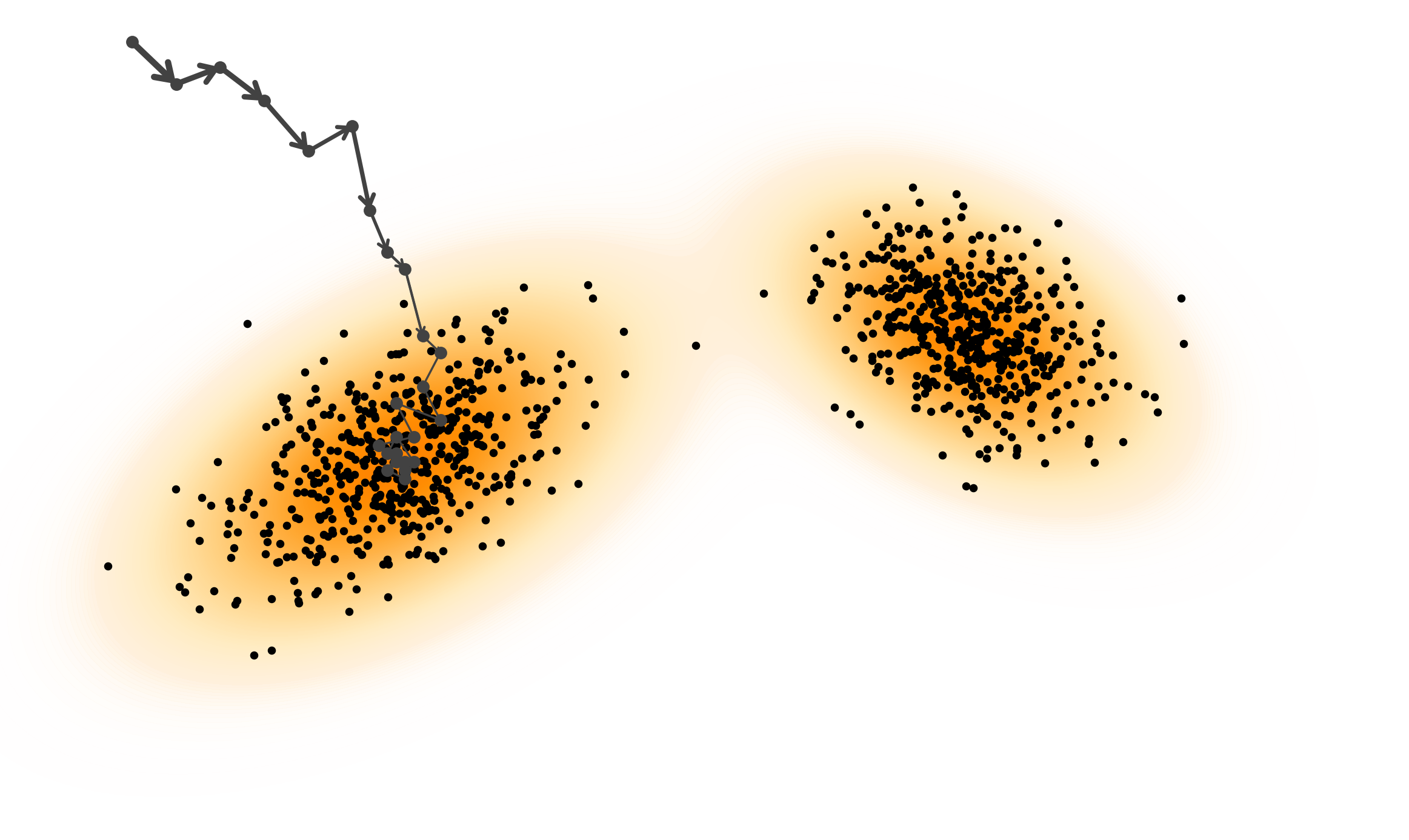}
    \vspace{-1.5cm}
    \caption{\textbfs{Illustration of Langevin sampling.} Langevin sampling using the score function $\nabla_{\mathbf{x}} \log p_{\bm{\phi}}(\mathbf{x})$ to guide trajectories toward high-density regions via the update in \Cref{eq:discrete-langevin-score} (indicating by arrows). \figcredit{Created by the authors.}}
    \label{fig:langevin}
\end{figure}

\paragraph{Discrete-Time Langevin Dynamics.}
The discrete-time Langevin update is
\begin{align}\label{eq:discrete-langevin}
    \mathbf{x}_{n+1} = \mathbf{x}_n - \eta \nabla_{\mathbf{x}} E_{\bm{\phi}}(\mathbf{x}_n) +  \sqrt{2 \eta} \bm{\epsilon}_n, \quad n=0,1,2,\ldots,
\end{align}
where $\mathbf{x}_0$ is initialized from some distribution (often Gaussian), $\eta > 0$ is the step size, and $\bm{\epsilon}_n \sim \mathcal{N}(\mathbf{0}, \mathbf{I})$ is Gaussian noise. The noise enables exploration beyond local minima by adding stochasticity. 

Since the score function can be computed as 
\[
\nabla_{\mathbf{x}} \log p_{\bm{\phi}}(\mathbf{x}) = -\nabla_{\mathbf{x}} E_{\bm{\phi}}(\rvx).
\]
the update can equivalently be written as
\begin{align}\label{eq:discrete-langevin-score}
    \mathbf{x}_{n+1} = \mathbf{x}_n + \eta \nabla_{\mathbf{x}} \log p_{\bm{\phi}}(\mathbf{x}_n) + \sqrt{2 \eta}\bm{\epsilon}_n,
\end{align}
where the score function guides the samples toward high-density regions. This formulation is central to diffusion models, as will be detailed later.

\paragraph{Continuous-Time Langevin Dynamics.}
As the step size $\eta$ approaches zero, the discrete Langevin updates naturally converge to a continuous-time process described by the \emph{Langevin Stochastic Differential Equation (SDE)}\footnote{
With the factor $\sqrt{2}$, the Langevin dynamics leave $p_{\bm\phi}$ unchanged in time. Namely, $p_{\bm\phi}$ is stationary: if $\rvx(0)\sim p_{\bm\phi}$ then $\rvx(t)\sim p_{\bm\phi}$ for all $t\ge 0$. Equivalently, $p_{\bm\phi}$ is the stationary solution of the Fokker–Planck equation (see \Cref{app:continuity}):
\[
\partial_t \rho = -\nabla \cdot(\rho  \nabla\log p_{\bm\phi}) + \tfrac{\sigma^2}{2}\Delta \rho.
\]
Setting $\rho=p_{\bm\phi}$ gives $(\tfrac{\sigma^2}{2}-1)\Delta p_{\bm\phi}=0$, which holds only if $\sigma=\sqrt{2}$.
}
:
\begin{align}\label{eq:continuous-langevin}
 \diff \mathbf{x}(t) = \nabla_{\mathbf{x}} \log p_{\bm{\phi}}(\mathbf{x}(t)) \diff t + \sqrt{2} \diff \mathbf{w}(t),
\end{align}
where $\mathbf{w}(t)$ denotes a standard Brownian motion (also known as a Wiener process\footnote{Brownian increments satisfy
$\mathbf{w}(t+\eta)-\mathbf{w}(t)\sim \mathcal{N}(\mathbf{0},  \eta  \mathbf{I})$.
Euler–Maruyama therefore uses a step noise $\sqrt{2}  [\mathbf{w}(t+\eta)-\mathbf{w}(t)]
= \sqrt{2\eta}  \bm{\epsilon}_n$ with $\bm{\epsilon}_n \sim \mathcal{N}(\mathbf{0},\mathbf{I})$,
which explains the $\sqrt{\eta}$ factor.; this is the source of the square-root scaling. For a detailed introduction to Brownian motion and SDEs, please refer to \Cref{app:de}.}). It is important to understand that the discrete update rule in \Cref{eq:discrete-langevin} serves as the Euler–Maruyama discretization of this continuous SDE.

Under standard regularity assumptions (e.g., $p_{\bm{\phi}}\propto e^{-E_{\bm{\phi}}}$ with a confining, sufficiently smooth $E_{\bm{\phi}}$), the distribution of $\mathbf{x}(t)$ converges (exponentially fast) to $p_{\bm{\phi}}$ as $t\to\infty$; thus we can sample by simulating (solving) the SDE \Cref{eq:continuous-langevin}.

\paragraph{Why Langevin Sampling?}
A natural way to understand Langevin sampling is through the lens of physics, where the energy function $ E_{\bm{\phi}}(\mathbf{x}) $ defines a potential landscape that shapes the behavior of particles. According to Newtonian dynamics, the motion of a particle under the force field derived from this energy is described by the ordinary differential equation (ODE)
\[
\diff \mathbf{x}(t) = -\nabla_{\mathbf{x}} E_{\bm{\phi}}\big(\mathbf{x}(t)\big)  \diff t,
\]
which deterministically drives the particle downhill toward a local minimum of the energy function. However, such deterministic dynamics can become trapped in local minima, preventing exploration of the full data distribution.

To overcome this limitation, Langevin dynamics introduces stochastic perturbations, resulting in the SDE
\[
\diff \mathbf{x}(t) = -\nabla_{\mathbf{x}} E_{\bm{\phi}}\big(\mathbf{x}(t)\big)  \diff t + \underbrace{\sqrt{2 } \diff \rvw(t)}_{\text{injected noise}},
\]
where $\rvw(t)$ is a standard Brownian motion. 
The noise term allows the particle to escape local minima by crossing energy barriers, making the trajectory a stochastic process whose stationary distribution converges to the Boltzmann distribution
\[
p_{\bm{\phi}}(\mathbf{x}) \propto e^{-E_{\bm{\phi}}(\mathbf{x})}.
\]

From this perspective, EBMs can be viewed as learning a force field that pushes samples toward regions of high probability. Langevin sampling is particularly useful for EBMs because it provides a practical method to generate samples from the model distribution $p_{\bm{\phi}}(\mathbf{x})$ without explicitly computing the partition function. By iteratively applying the Langevin update, one obtains samples that approximate the target distribution.

\paragraph{Inherent Challenges of Langevin Sampling.}
Langevin dynamics, a widely used MCMC-based sampler, faces serious limitations in high-dimensional spaces. Its efficiency is highly sensitive to the choice of step size $\eta$, noise scale, and the number of iterations required to approximate the target distribution accurately. 

At the heart of this inefficiency lies the issue of poor ``mixing time'': In complex data distributions with many isolated modes, Langevin sampling often requires an extremely long time to transition between regions of high probability. This problem becomes significantly worse as dimensionality increases, leading to prohibitively slow convergence.

One can think of sampling as exploring a vast and rugged landscape with many distant valleys, each corresponding to a different data mode. Langevin dynamics, relying on local stochastic updates, struggles to traverse between these valleys efficiently. As a result, it often fails to capture the full diversity of the distribution.

This inefficiency hints the need for more structured and guided sampling methods that can navigate complex data manifolds more effectively than purely random exploration.


\clearpage
\newpage

\section{\texorpdfstring{From Energy-Based to Score-Based Generative Models}{From Energy-Based to Score-Based Generative Models}}
EBMs show that generation depends only on the score, which points toward regions of higher probability, rather than on the full normalized density. 
While score matching avoids the partition function, training through the energy still requires expensive second derivatives. 
The key idea is that since sampling with Langevin dynamics needs only the score, we can learn it directly with a neural network. 
This shift, from modeling energies to modeling scores, forms the foundation of score-based generative models.

\begin{figure}[ht!]
    \centering
    \includegraphics[width=0.9\linewidth]{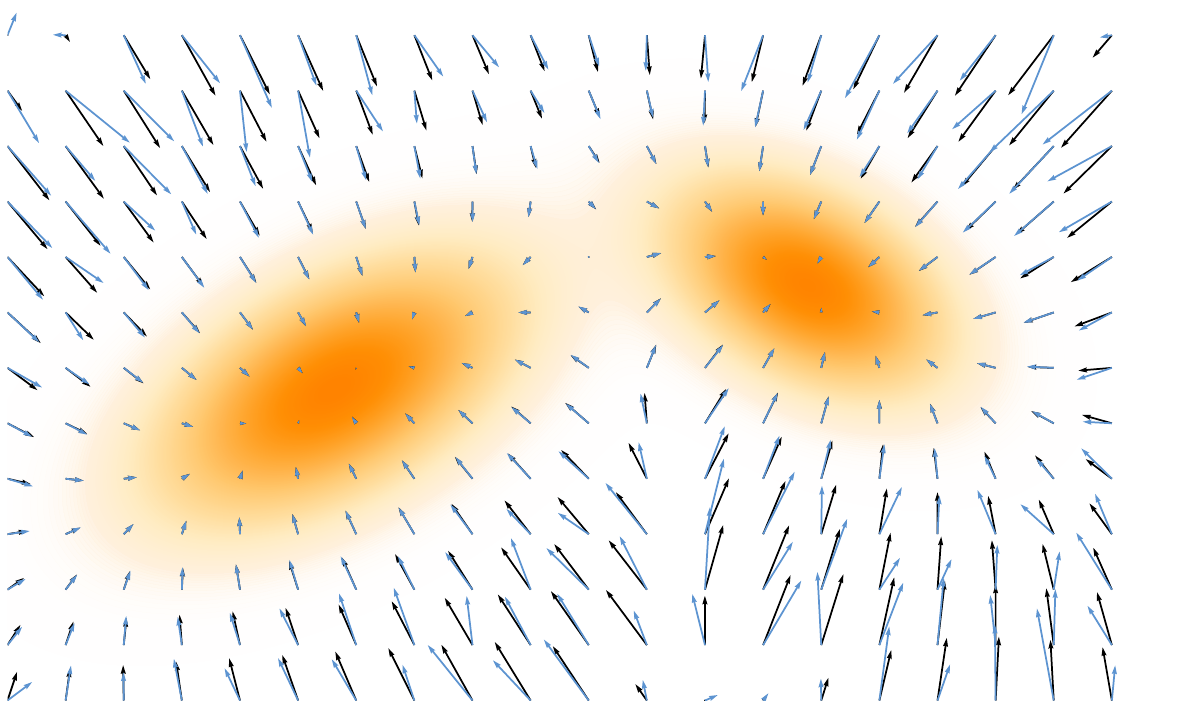}
    \caption{\textbfs{Illustration of Score Matching.} The neural network score ${\color{customblue}\rvs_{\bm{\phi}}(\rvx)}$ is trained to match the ground truth score $\rvs(\rvx)$ using a MSE loss. Both are represented as vector fields. \figcredit{Created by the authors.}}
    \label{fig:sm-illustration}
\end{figure}

\subsection{Training with Score Matching}\label{subsec:training-sm}
\paragraph{Score Matching.}
To approximate the score function $\rvs(\rvx) = \nabla_{\rvx} \log p_{\mathrm{data}}(\rvx)$ from samples of $p_{\mathrm{data}}$,  
we approximate it directly as a vector field parameterized by a neural network $\rvs_{\bm{\phi}}(\rvx)$ (see \Cref{fig:sm-illustration}):
\[
\rvs_{\bm{\phi}}(\rvx) \approx \rvs(\rvx).
\]
\emph{Score matching} fits this vector field by minimizing the mean squared error (MSE) between the true and estimated scores:
\begin{mdframed}
\begin{align}\label{eq:sm}
    \mathcal{L}_{\mathrm{SM}}(\bm{\phi}) 
    := \frac{1}{2}  \mathbb{E}_{\rvx \sim p_{\mathrm{data}}}
    \Big[\| \rvs_{\bm{\phi}}(\rvx) - \rvs(\rvx)\|_2^2\Big].
\end{align}
\end{mdframed}

\paragraph{Tractable Score Matching.}
At first glance, this objective seems infeasible because the true score $\rvs(\rvx)$, which serves as the regression target, is unknown.  
Fortunately, \citet{hyvarinen2005estimation} showed that integration by parts yields an equivalent objective that depends only on the model $\rvs_{\bm{\phi}}$ and the data samples, without requiring access to the true score.  
We state this key result in the following proposition: 
\proppp{Hyvärinen’s Tractable Form of SM}{sm-trace}
        {
        We can express the following equation as:
        \begin{align*}
            \mathcal{L}_{\text{SM}}(\bm{\phi}) = \widetilde{\mathcal{L}}_{\text{SM}}(\bm{\phi}) + C.
        \end{align*}
        where
        \begin{align}\label{eq:sm-alternative}
            \widetilde{\mathcal{L}}_{\text{SM}}(\bm{\phi})
            := \mathbb{E}_{\rvx \sim p_{\text{data}}(\rvx)} \left[
            \Tr\left(\nabla_{\rvx}\rvs_{\bm{\phi}}(\rvx)\right)
            + \frac{1}{2} \norm{\rvs_{\bm{\phi}}(\rvx)}_2^2
            \right].
        \end{align}
        and $C$ is a constant that does not depend on $\bm{\phi}$. The minimizer $\rvs^*$ is obtained as: $\rvs^*(\cdot) = \nabla_{\rvx} \log p(\cdot)$.
        }
        {The result follows by expanding the MSE in $ \mathcal{L}_{\text{SM}} $ and applying integration by parts. The proof is given in \Cref{app-sec:sm-trace}. 
        }

Using the equivalent objective in \Cref{eq:sm-alternative}, we train the score model $\rvs_{\bm{\phi}}(\rvx)$ solely from observed samples of $p_{\mathrm{data}}$, eliminating the need for the true score function.

\paragraph{Intuition of \Cref{eq:sm-alternative}.}
The alternative score matching objective $\widetilde{\mathcal{L}}_{\text{SM}}(\bm\phi)$
can be understood from its two terms.
The norm term $\tfrac12\|\rvs_{\bm\phi}(\rvx)\|^2$ penalizes large score values,
especially in regions where $p_{\mathrm{data}}$ is large.
The divergence term $\Tr(\nabla_{\rvx} \rvs_{\bm\phi}(\rvx))$ appears with a positive sign in the loss, so minimizing the objective also favors locally contracting behavior.
Together, these two terms yield the population-optimal score field
$\rvs^*(\rvx)=\nabla_{\rvx}\log p_{\mathrm{data}}(\rvx)$.
We explain the geometric intuition in more detail below.

\subparagraph{Effect of the Magnitude Term.}
Since the expectation in $\widetilde{\mathcal{L}}_{\mathrm{SM}}(\bm\phi)$ is taken under
$p_{\mathrm{data}}$, regions where $p_{\mathrm{data}}(\rvx)$ is large (high data density)
contribute more strongly to the loss. The magnitude term
$\tfrac12\|\rvs_{\bm\phi}(\rvx)\|^2$ therefore places a stronger penalty on large
score values in these regions. By itself, however, this term does not imply that
$\rvs_{\bm\phi}(\rvx)=\mathbf{0}$ at every high-density point; at the population
optimum,
\[
\rvs^*(\rvx)=\nabla_{\rvx}\log p_{\mathrm{data}}(\rvx),
\]
which vanishes only at stationary points of the log-density.

\subparagraph{Concavity When the Field is (Approximately) a Gradient.}
Because the divergence term $\Tr(\nabla_{\rvx} \rvs_{\bm\phi}(\rvx))$ enters the loss with a positive sign, minimizing the objective favors
negative divergence in regions emphasized by the data distribution.
Negative divergence indicates net local contraction, although by itself it does
not guarantee that a stationary point is a sink.
To make this precise, assume
$\rvs_{\bm\phi}=\nabla_{\rvx}u$ for a scalar function $u:\R^D\to\R$, as is natural when matching a
log density. Then $\nabla_{\rvx}\rvs_{\bm\phi}=\nabla_{\rvx}^2 u$ (the Hessian) and
$\nabla \cdot \rvs_{\bm\phi}(\rvx)=\Tr(\nabla_{\rvx}^2 u(\rvx))$ (the divergence).

At a stationary point $\rvx_\star$, where $\rvs_{\bm\phi}(\rvx_\star)=\nabla_{\rvx}u(\rvx_\star)=\mathbf{0}$,
a second order Taylor expansion gives
\[
u(\rvx)
= u(\rvx_\star)
+ \tfrac12(\rvx-\rvx_\star)^\top \nabla_{\rvx}^2 u(\rvx_\star)(\rvx-\rvx_\star)
+ o(\|\rvx-\rvx_\star\|^2).
\]
If the Hessian $\nabla_{\rvx}^2 u(\rvx_\star)$ is negative definite, then $u$ is locally concave at
$\rvx_\star$ and the log density attains a strict local maximum\footnote{We remark that strict concavity (and thus a strict local maximum of the log density) requires the entire Hessian
$\nabla_{\rvx}^2 u$ to be negative definite, not merely to have negative trace.
A negative trace guarantees that the sum of eigenvalues is negative, but some eigenvalues could still
be positive, leading to a saddle point rather than a maximum.
} there.
Because all eigenvalues of the Hessian are negative, the trace is also negative:
$\Tr(\nabla_{\rvx}^2 u(\rvx_\star))<0$.
Thus, in this case, the stationary point is locally attracting under the score field:
small perturbations are pulled back toward $\rvx_\star$.

\subsection{Sampling with Langevin Dynamics}
Once trained by minimizing \Cref{eq:sm-alternative}, the score model $\rvs_{\bm{\phi}^\times}(\rvx)$ can replace the oracle score in Langevin dynamics for sampling:
\begin{align}\label{eq:score-langevin-emp}
    \mathbf{x}_{n+1} = \mathbf{x}_n + \eta   \rvs_{\bm{\phi}^\times}(\rvx_n) + \sqrt{2\eta}    \bm{\epsilon}_n, \quad \bm{\epsilon}_n \sim \mathcal{N}(\bm{0}, \bm{I}),
\end{align}
for $n=0,1,2,\dots$, initialized at $\rvx_0$. As in the EBM case \Cref{eq:continuous-langevin}, this recursion is precisely the Euler–Maruyama discretization of the continuous-time Langevin SDE:
\begin{align*}
    \diff \rvx(t) = \rvs_{\bm{\phi}^\times}(\rvx(t)) \diff t + \sqrt{2} \diff \mathbf{w}(t),
\end{align*}
with initialization $\rvx(0)$. Hence, in the limit of small step size, the discrete and continuous formulations coincide.  In practice, one can either run the discrete sampler or directly simulate the SDE.

\subsection{Prologue: Score-Based Generative Models}

In the remainder of this chapter, we examine the foundational role of the score function in modern diffusion models. Initially introduced to enable efficient training of EBMs, the score function has evolved into a central component of a new generation of generative models. Building on this foundation, we explore how the score function informs the theoretical formulation and practical implementation of \emph{score-based diffusion models}, offering a principled framework for data generation via stochastic processes.

\clearpage
\newpage

\section{\texorpdfstring{Denoising Score Matching}{Denoising Score Matching}}\label{sec:dsm}

\subsection{Motivation}
Although the alternative objective in \Cref{eq:sm-alternative}
\begin{align*}
    \widetilde{\mathcal{L}}_{\text{SM}}(\bm{\phi}) = \mathbb{E}_{\rvx \sim p_{\mathrm{data}}} \left[\mathrm{Tr}\big(\nabla_{\rvx}\rvs_{\bm{\phi}}(\rvx)\big) + \frac{1}{2} \|\rvs_{\bm{\phi}}(\rvx)\|_2^2 \right]
\end{align*}
is more tractable, it still requires computing the trace of the Jacobian, $\mathrm{Tr}(\nabla_{\rvx}\rvs_{\bm{\phi}}(\rvx))$. Although this does not require forming the full Jacobian matrix, exact evaluation generally becomes more expensive as the input dimension $D$ increases, since it requires aggregating derivative information across input coordinates. This can therefore be computationally expensive in high dimensions.

To address this, sliced score matching~\citep{song2020sliced} replaces the trace term with a stochastic estimate based on random projections. We briefly outline the idea below.
\paragraph{Sliced Score Matching and Hutchinson's Estimator.} Sliced score matching  replaces the trace in score matching by averaging directional derivatives along random ``slices''. Let $\rvu\in\R^D$ be an \emph{isotropic} random vector (e.g., Rademacher or standard Gaussian) with
$\E[\rvu]=0$ and $\E[\rvu\rvu^\top]=\mathbf I$. By Hutchinson’s identity 
\[
\Tr(\rmA)=\E_{\rvu}[\rvu^\top \rmA \rvu],\quad\text{and}\quad\E_{\rvu}[(\rvu^\top\rvs_{\bphi}(\rvx))^2]=\|\rvs_{\bphi}(\rvx)\|_2^2,
\]
 we obtain the exact form
\[
\widetilde{\mathcal{L}}_{\mathrm{SM}}(\bphi)
=\E_{\rvx,\rvu} \Big[\rvu^\top\big(\nabla_{\rvx}\rvs_{\bphi}(\rvx)\big)\rvu+\tfrac12(\rvu^\top\rvs_{\bphi}(\rvx))^2\Big].
\]
This objective can be evaluated efficiently with automatic differentiation, using Jacobian- and vector-Jacobian-product operations (JVP/VJP) instead of explicitly computing large Jacobian or Hessian matrices.
Averaging over $K$ random probes yields an unbiased estimator with variance $\mathcal{O}(1/K)$, 
and the directional term $\rvu^\top(\nabla_{\rvx}\rvs_{\bphi})\rvu$ can be computed 
efficiently using JVP/VJP routines without explicit Jacobians.  Intuitively, this means we only check the model’s behavior along random directions: 
the projected score is nudged to align with regions of higher data density, 
so data points become stationary in expectation. 

\paragraph{From Sliced to Denoising Score Matching.} Sliced score matching sidesteps Jacobians but still relies on the raw data distribution. 
This makes it fragile: for image data lying on low-dimensional manifolds, 
the score $\nabla_{\rvx}\log p_{\mathrm{data}}(\rvx)$ may be undefined or unstable, 
and the method only constrains the vector field at observed points, 
providing weak control in their neighborhoods. 
It further suffers from probe-induced variance and repeated JVP/VJP costs. 

A more robust alternative, which we focus on here, is \emph{Denoising Score Matching} (DSM)~\citep{vincent2011connection}, 
which offers a principled and scalable solution.

\subsection{Training}
Let us revisit the SM loss in \Cref{eq:sm}:
\begin{align*}
     \mathcal{L}_{\text{SM}}(\bm{\phi}) =  \frac{1}{2} \mathbb{E}_{\rvx \sim p_{\text{data}}(\rvx)} \left[ \norm{\rvs_{\bm{\phi}}(\rvx) - \nabla_{\rvx} \log p_{\text{data}}(\rvx)}_2^2 \right],
\end{align*}
where the issue arises from the intractable term $\nabla_{\rvx} \log p_{\text{data}}(\rvx)$.

\paragraph{\citet{vincent2011connection}'s Solution by Conditioning.}
To overcome the intractability of $\nabla_{\rvx} \log p_{\text{data}}(\rvx)$, \citet{vincent2011connection} proposed injecting noise into the data $\rvx \sim p_{\text{data}}$ via a known conditional distribution $p_{\sigma}(\tilde{\rvx}|\rvx)$ with scale $\sigma$. The neural network $\rvs_{\bm{\phi}}(\tilde{\rvx}, \sigma)$ is trained to approximate the score of the marginal perturbed distribution
\[
p_\sigma(\tilde{\rvx}) = \int p_{\sigma}(\tilde{\rvx}|\rvx) p_{\text{data}}(\rvx) \diff \rvx
\]
by minimizing the loss
\begin{align}\label{eq:sm-noise-loss}
\mathcal{L}_{\text{SM}}(\bm{\phi}; \sigma) := \frac{1}{2} \mathbb{E}_{\tilde{\rvx} \sim p_\sigma} \left[ \norm{\rvs_{\bm{\phi}}(\tilde{\rvx}, \sigma) - \nabla_{\tilde{\rvx}} \log p_\sigma(\tilde{\rvx})}_2^2 \right].
\end{align}

Even though $\nabla_{\tilde{\rvx}} \log p_\sigma(\tilde{\rvx})$ is generally intractable, \citet{vincent2011connection} showed that \emph{conditioning} on $\rvx \sim p_{\text{data}}$ yields an equivalent, tractable objective—the \emph{Denoising Score Matching} (DSM) loss:
\begin{mdframed}
\begin{align}\label{eq:dsm-loss}
\mathcal{L}_{\text{DSM}}(\bm{\phi}; \sigma) := \frac{1}{2} \mathbb{E}_{\rvx \sim p_{\text{data}}, \tilde{\rvx} \sim p_{\sigma}(\cdot|\rvx)} \left[ \norm{\rvs_{\bm{\phi}}(\tilde{\rvx}, \sigma) - \nabla_{\tilde{\rvx}} \log p_{\sigma}(\tilde{\rvx}|\rvx)}_2^2 \right].
\end{align}
\end{mdframed}

The optimal minimizer $\rvs^*$ of \Cref{eq:dsm-loss} satisfies
\[
\rvs^*(\tilde{\rvx}, \sigma) = \nabla_{\tilde{\rvx}} \log p_\sigma(\tilde{\rvx}),
\]
which is also optimal for \Cref{eq:sm-noise-loss}.

For example, when $p_{\sigma}(\tilde{\rvx}|\rvx)$ is Gaussian noise with variance $\sigma^2$,
\[
p_{\sigma}(\tilde{\rvx}|\rvx) = \mathcal{N}(\tilde{\rvx}; \rvx, \sigma^2 \mathbf{I}),
\]
the gradient $\nabla_{\tilde{\rvx}} \log p_{\sigma}(\tilde{\rvx}|\rvx)$ has a closed form (see \Cref{eq:gaussian-dsm}), making the regression target explicit and computationally tractable.

Moreover, as $\sigma \approx 0$, $p_\sigma(\tilde{\rvx}) \approx p_{\text{data}}(\rvx)$ and
\[
\rvs^*(\tilde{\rvx}, \sigma) = \nabla_{\tilde{\rvx}} \log p_\sigma(\tilde{\rvx}) \approx \nabla_{\rvx} \log p_{\text{data}}(\rvx),
\]
indicating the learned score approximates the original data score, enabling its use in generation.

We formalize this discussion on the gradient equivalence between $\mathcal{L}_{\text{SM}}$ and $\mathcal{L}_{\text{DSM}}$ in the following theorem:
\thmp{Equivalence of $\mathcal{L}_{\text{SM}}$ and $\mathcal{L}_{\text{DSM}}$}{sm-dsm}{
For any fixed noise scale $\sigma > 0$, the following holds:
\begin{align}\label{eq:score-matching}
    \mathcal{L}_{\text{SM}}(\bm{\phi}; \sigma) = \mathcal{L}_{\text{DSM}}(\bm{\phi}; \sigma) + C,
\end{align}
where $C$ is a constant independent of the parameter $\bm{\phi}$. Furthermore, the minimizer $\rvs^*(\cdot, \sigma)$ of both losses satisfies
\[
    \rvs^*(\tilde{\rvx}, \sigma) = \nabla_{\tilde{\rvx}} \log p_\sigma(\tilde{\rvx}), \quad \text{for almost every } \tilde{\rvx}.
\]
}{The equivalence follows from a direct computation: by expanding the MSE in $\mathcal{L}_{\text{SM}}$ and $\mathcal{L}_{\text{DSM}}$, all $\bm{\phi}$-dependent terms cancel, leaving only a constant difference independent of $\bm{\phi}$.\\
 The derivation of the minimizer follows the same argument as in Proposition~\ref{dsm-minimizer}. We defer the detailed derivation to \Cref{app-sec:sm-dsm}.
}

This theorem, like Theorem~\ref{thm:equiv-marginal-kl} in DDPM, illustrates a key shared principle:
\msg{Insight}{Conditioning Technique}{The conditioning technique also appears in the variational view of diffusion models in DDPM (see Theorem~\ref{thm:equiv-marginal-kl}), where conditioning on a data point $\rvx$ turns an intractable loss into a tractable one for Monte Carlo estimation. A similar idea arises in the flow-based perspective (e.g., Flow Matching~\citep{lipman2022flow}), as we will see in \Cref{sec:flow-matching-framework}.}

\begin{figure}
    \centering
    \includegraphics[width=0.8\linewidth]{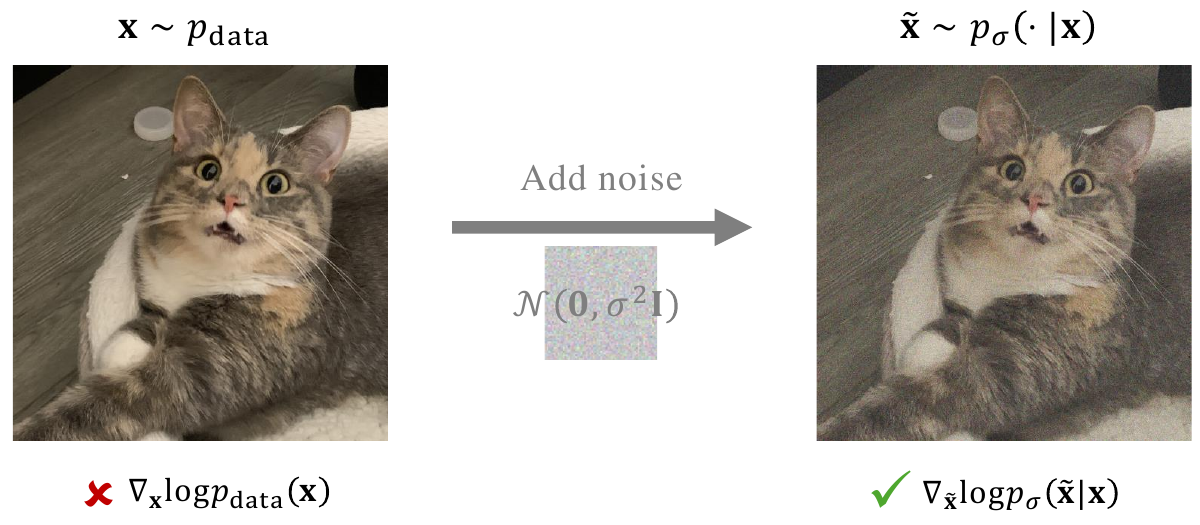}
    \caption{\textbfs{Illustration of DSM via the conditioning technique.} By perturbing the data distribution $p_{\mathrm{data}}$ with small additive Gaussian noise $\mathcal{N}(\mathbf{0}, \sigma^2 \rmI)$, the resulting conditional distribution $p_{\sigma}(\tilde{\rvx}|\rvx) = \mathcal{N}(\tilde{\rvx}; \rvx, \sigma^2 \rmI)$ admits a closed-form score function.
\figcredit{Created by the authors.}}
    \label{fig:dsm-graph}
\end{figure}

\paragraph{Special Case: Additive Gaussian Noise.} We now consider the common case where Gaussian noise $\mathcal{N}(\mathbf{0}, \sigma^2 \rmI)$ with variance $\sigma^2$ is added to each data point $\rvx \sim p_{\text{data}}$:
\[
\tilde{\rvx} = \rvx + \sigma \bm{\epsilon}, \quad \bm{\epsilon} \sim \mathcal{N}(\mathbf{0}, \rmI),
\]
so that the corrupted data $\tilde{\rvx}$ follows
\[
p_{\sigma}(\tilde{\rvx}|\rvx) = \mathcal{N}(\tilde{\rvx}; \rvx, \sigma^2 \rmI).
\]
In this setting, the conditional score is analytically given by
\[
\nabla_{\tilde{\rvx}} \log p_{\sigma}(\tilde{\rvx}|\rvx) = \frac{\rvx - \tilde{\rvx}}{\sigma^2}.
\]
Hence, the DSM loss simplifies to:
\begin{mdframed}
\begin{align}\label{eq:gaussian-dsm}
\begin{aligned}
    \mathcal{L}_{\text{DSM}}(\bm{\phi}; \sigma) &= \frac{1}{2} \mathbb{E}_{\rvx, \tilde{\rvx}|\rvx} \left[ \norm{\rvs_{\bm{\phi}}(\tilde{\rvx}, \sigma) - \frac{\rvx - \tilde{\rvx}}{\sigma^2}}_2^2 \right] \\
    &= \frac{1}{2} \mathbb{E}_{\rvx, \bm{\epsilon}} \left[ \norm{\rvs_{\bm{\phi}}(\rvx + \sigma \bm{\epsilon}, \sigma) + \frac{\bm{\epsilon}}{\sigma}}_2^2 \right],
\end{aligned}
\end{align}
\end{mdframed}
where $\bm{\epsilon} \sim \mathcal{N}(\mathbf{0}, \rmI)$. This objective forms the core of the (score-based) Diffusion Model.

At a fixed noise level $\sigma>0$, DSM learns the score of the
Gaussian-smoothed distribution
\[
p_\sigma
=
p_{\mathrm{data}} * \mathcal{N}(\bm{0},\sigma^2\rmI),
\]
rather than the score of $p_{\mathrm{data}}$ itself.
When $\sigma$ is small, the corruption is mild, so $p_\sigma$ still closely
preserves the local structure of the data distribution. Its score therefore
provides a useful direction for moving noisy samples back toward likely data
regions. As $\sigma$ becomes larger, the Gaussian smoothing progressively removes fine
details and merges nearby modes. The corresponding score then captures mainly
coarser, more global structure. This behavior is useful rather than undesirable:
the multi-noise construction in \Cref{sec:SMLD} deliberately combines many noise
levels, using large noise for coarse exploration and small noise for fine
refinement.

To better see why the objective naturally corresponds to a ``denoising'' process,
we expand on the discussion in \Cref{subsec:why-dsm-denoising-tweedie,subsec:why-dsm-denoising-sure}.


\subsection{Sampling} Once we have a trained score model $\rvs_{\bm{\phi}^\times}(\tilde{\rvx}, \sigma)$ at noise level $\sigma$, we generate samples using Langevin dynamics by replacing the true score with the learned model. The update rule is:
\begin{align}\label{eq:score-langevin-emp-dsm}
   \tilde{\rvx}_{n+1} = \tilde{\rvx}_n + \eta \underbrace{\rvs_{\bm{\phi}^\times}(\tilde{\rvx}_n, \sigma)}_{\approx \nabla_{\tilde{\rvx}} \log p_\sigma(\tilde{\rvx}_n)} + \sqrt{2\eta}    \bm{\epsilon}_n, \quad \bm{\epsilon}_n \sim \mathcal{N}(\mathbf{0}, \rmI),
\end{align}
for $n = 0, 1, 2, \dots$, starting from an initial value $\tilde{\rvx}_0$. If $\sigma$ is sufficiently small, then after enough iterations, $\tilde{\rvx}_n$ approximates samples from $p_{\text{data}}$.

\paragraph{Advantages of Noise Injection.}
We additionally remark that, compared to vanilla score matching in \Cref{eq:sm}, injecting Gaussian noise to form $p_\sigma$ (e.g., \Cref{eq:gaussian-dsm}) provides two key advantages~\citep{song2019generative}:
\begin{itemize}
  \item \textbfs{Well-Defined Gradients.} The noise perturbs data away from its lower-dimensional manifold, resulting in a distribution $p_\sigma$ with full support in $\mathbb{R}^D$. Consequently, the score function $\nabla_{\tilde{\rvx}} \log p_\sigma(\tilde{\rvx})$ is well-defined everywhere.
  \item \textbfs{Improved Coverage.} The noise smooths out sparse regions between modes, enhancing training signal quality and facilitating Langevin dynamics to traverse low-density regions more effectively.
\end{itemize}

\subsection{Why DSM is Denoising: Tweedie's Formula}\label{subsec:why-dsm-denoising-tweedie}

We begin with \emph{Tweedie's formula}~\citep{efron2011tweedie}, which provides a principled basis for denoising from noisy observations alone.
Concretely, it states that: given a single Gaussian–corrupted observation $\tilde{\rvx}\sim\mathcal N(\,\cdot\,;\alpha\rvx,\sigma^2\mathrm{I})$  from an unknown $\rvx \sim p_{\mathrm{data}}$, a denoised estimate (the
average over all plausible clean signals given $\tilde{\rvx}$) is obtained by nudging
$\tilde{\rvx}$ a step of size $\sigma^2$ in the direction of the score
$\nabla_{\tilde{\rvx}}\log p_\sigma(\tilde{\rvx})$ of its noisy marginal defined as:
\[
p_\sigma(\tilde{\rvx}):= \int \mathcal{N}(\tilde{\rvx};\alpha \rvx,\sigma^2\mathrm{I})p_{\mathrm{data}}(\rvx)\diff\rvx.
\]
We present the proposition formally below.
\lemp{Tweedie's Formula}{tweedie}{Assume $\rvx\sim p_{\mathrm{data}}$ and, conditionally on $\rvx$,
$\tilde{\rvx}\sim\mathcal N(\,\cdot\,;\alpha\rvx,\sigma^2\mathrm{I})$ with $\alpha\neq0$. Then Tweedie's formula states
 \begin{align}\label{eq:tweedie} 
 \alpha   \mathbb{E}_{\rvx \sim p(\rvx |\tilde\rvx)}\big[\rvx  \big|  \tilde\rvx\big] = \tilde\rvx + \sigma^2 \nabla_{\tilde\rvx} \log p_\sigma(\tilde\rvx), 
 \end{align} where the expectation is taken over the posterior distribution $p(\rvx |\tilde\rvx)$ of $\rvx$ given  $\tilde\rvx$.
}{
The proof proceeds by computing the score of the marginal
\[
p(\tilde\rvx)=\int p(\tilde\rvx|\rvx)p_{\mathrm{data}}(\rvx)\diff\rvx.
\]
Differentiating under the integral and using the Gaussian form of the
conditional density leads directly to an expression that rearranges into
the desired identity linking the score with the posterior mean. See \Cref{app-sec:tweedie} for details.
}
Tweedie’s formula plays a central role in diffusion models, where multiple layers of noise are introduced as in DDPM. 
It enables the estimation of clean samples from noisy observations via the score function, thereby establishing a fundamental link between score prediction and denoiser:
\begin{equation*}
\underbrace{\mathbb{E} \left[\rvx|\tilde\rvx\right]}_{\substack{\text{denoiser}\\\text{estimated from }\tilde\rvx}}
= \frac{1}{\alpha}\left(\tilde\rvx + \sigma^2 \nabla_{\tilde\rvx} \log p_\sigma(\tilde\rvx)\right).
\end{equation*}

Especially,  a single gradient-ascent step on the noisy log-likelihood with the particular step size $\sigma^2$
is the denoised estimate (the conditional average clean signal). This makes
DSM training and denoising tightly related: if $\rvs_\bphi(\tilde{\rvx})\approx
\nabla_{\tilde{\rvx}}\log p_\sigma(\tilde{\rvx})$ trained from DSM, then
\[
\frac{1}{\alpha}\left(\tilde{\rvx}+\sigma^2 \rvs_\bphi(\tilde{\rvx})\right)
\]
is the denoiser.

\paragraph{(Optional) Higher Order Tweedie's Formula.} The classical Tweedie's formula expresses the posterior mean $\E[\rvx_0|\tilde{\rvx}]$ through the gradient $\nabla_{\tilde{\rvx}}\log p(\tilde{\rvx})$.
Higher order extensions~\citep{meng2021estimating} express the posterior covariance and higher cumulants through higher derivatives of $\log p(\tilde{\rvx})$.

\subparagraph{Exponential Family Setup with the Log-Normalizer $\lambda(\tilde{\rvx})$.}
Assume the conditional law of $\tilde{\rvx}$ given a latent natural parameter $\boldsymbol\eta\in\R^D$ belongs to a natural exponential family written as
\[
q_\sigma(\tilde{\rvx}|\boldsymbol\eta)
=\exp\!\big(\boldsymbol\eta^\top \tilde{\rvx}-\psi(\boldsymbol\eta)\big)\,q_0(\tilde{\rvx}).
\]
Here $q_0(\tilde{\rvx})$ is the \emph{base measure}, namely the part that does not depend on $\boldsymbol\eta$; for additive Gaussian noise with variance $\sigma^2\rmI$ it equals $(2\pi\sigma^2)^{-D/2}\exp(-\|\tilde{\rvx}\|^2/2\sigma^2)$.
Let $p(\boldsymbol\eta)$ be the pre-defined distribution of the latent natural parameter, which can be viewed as the reparameterized clean-data distribution (for Gaussian location, $\boldsymbol\eta=\rvx/\sigma^2$).
The observed noisy marginal is
\[
p_\sigma(\tilde{\rvx})=\int q_\sigma(\tilde{\rvx}|\boldsymbol\eta)\,p(\boldsymbol\eta)\diff\boldsymbol\eta.
\]
Define the \emph{log-partition} (log-normalizer) in $\tilde{\rvx}$ by
\[
\lambda(\tilde{\rvx}) := \log p_\sigma(\tilde{\rvx})-\log q_0(\tilde{\rvx}).
\]
Then the posterior of $\boldsymbol\eta$ given $\tilde{\rvx}$ is
\[
p(\boldsymbol\eta|\tilde{\rvx})
\ \propto\
\exp\!\big(\boldsymbol\eta^\top \tilde{\rvx}-\psi(\boldsymbol\eta)-\lambda(\tilde{\rvx})\big)\,p(\boldsymbol\eta),
\]
which shows that, as a function of $\tilde{\rvx}$, the posterior has exponential-family form with natural parameter $\tilde{\rvx}$, sufficient statistic $\boldsymbol\eta$, and log-partition  $\lambda(\tilde{\rvx})$.

\subparagraph{Derivatives of $\lambda$ Produce Posterior Cumulants.}
Two simple rules are at play.
First, normalization: for every $\tilde{\rvx}$,
\[
\int \exp\!\big(\boldsymbol\eta^\top \tilde{\rvx}-\psi(\boldsymbol\eta)-\lambda(\tilde{\rvx})\big)\,p(\boldsymbol\eta)\diff\boldsymbol\eta
=1.
\]
Differentiating this identity with respect to $\tilde{\rvx}$ brings down powers of $\boldsymbol\eta$ from the exponential and derivatives of $\lambda(\tilde{\rvx})$; setting the result to zero yields equalities between derivatives of $\lambda$ and posterior moments of $\boldsymbol\eta$.
Second, a standard property of exponential families: the log-partition is the cumulant generating function of the sufficient statistic.
Therefore
\[
\nabla_{\tilde{\rvx}}\lambda(\tilde{\rvx})=\E[\boldsymbol\eta|\tilde{\rvx}],\quad
\nabla_{\tilde{\rvx}}^{2}\lambda(\tilde{\rvx})=\Cov[\boldsymbol\eta|\tilde{\rvx}],\quad
\nabla_{\tilde{\rvx}}^{(k)}\lambda(\tilde{\rvx})=\kappa_k(\boldsymbol\eta|\tilde{\rvx})\quad(k\ge 3),
\]
where $\kappa_k$ are the \emph{conditional cumulants} of order $k$ of the random vector $\boldsymbol\eta$ given $\tilde{\rvx}$, obtained via the standard moment–cumulant relations.

These are the higher order Tweedie's formulas.
Specializing to the Gaussian location model with $\boldsymbol\eta=\rvx/\sigma^2$ yields the familiar forms in terms of derivatives of $\log p_\sigma(\tilde{\rvx})$:
\[
\E[\rvx|\tilde{\rvx}]
=\tilde{\rvx}+\sigma^2\nabla_{\tilde{\rvx}}\log p_\sigma(\tilde{\rvx}),
\qquad
\Cov[\rvx|\tilde{\rvx}]
=\sigma^2\mathrm I+\sigma^4\nabla_{\tilde{\rvx}}^{2}\log p_\sigma(\tilde{\rvx}),
\]
and higher cumulants scale with higher derivatives of $\log p_\sigma(\tilde{\rvx})$.

Several studies have explored training neural networks to estimate higher order scores~\citep{meng2021estimating,lu2022maximum,lai2023fp}. In contrast, our aim is to clarify their relationship with statistical quantities, and we refer the reader to these works for methodological details.

\subsection{(Optional) Why DSM is Denoising: SURE}\label{subsec:why-dsm-denoising-sure}

\paragraph{SURE (Stein’s Unbiased Risk Estimator).}
At a high level, Stein’s Unbiased Risk Estimator (SURE) is a technique that allows one to 
estimate the mean squared error (MSE) of a denoiser $\rmD$ \emph{without knowing the clean signal}. 
In other words, SURE provides a way to select or train denoisers when only noisy data are available.

For clarity, consider the additive Gaussian noise setting:
\[
\tilde{\rvx} = \rvx + \sigma \bm{\epsilon}, 
\qquad \bm{\epsilon} \sim \mathcal{N}(\mathbf{0},\rmI),
\]
where $\rvx \in \R^D$ is the unknown clean signal and $\tilde{\rvx}$ is the observed noisy version. 
A denoiser is any (weakly differentiable) mapping $\rmD:\R^D \to \R^D$ that produces an estimate $\rmD(\tilde{\rvx})$ of $\rvx$.

The natural quality measure is the conditional MSE
\[
R(\rmD;\rvx) := \E_{\tilde{\rvx}|\rvx}\left[\|\rmD(\tilde{\rvx})-\rvx\|_2^2 \,\big|\, \rvx\right].
\]
This quantity depends on the unknown ground truth $\rvx$, and therefore cannot be computed directly. 
Stein’s identity (see \Cref{app-sec:stein-identity}), however, yields the following \emph{observable} surrogate:
\begin{align}\label{eq:sure-obj}
    \mathrm{SURE}(\rmD;\tilde{\rvx})
= \|\rmD(\tilde{\rvx})-\tilde{\rvx}\|_2^2
+ 2\sigma^2\,\nabla_{\tilde{\rvx}}\cdot \rmD(\tilde{\rvx})
- D\sigma^2,
\end{align}
where $\nabla_{\tilde{\rvx}}\cdot \rmD(\tilde{\rvx})$ denotes the divergence of $\rmD$. 
We emphasize that $\mathrm{SURE}(\rmD;\tilde{\rvx})$ requires only the noisy observation $\tilde\rvx$, not the clean $\rvx$.

Intuitively, $\mathrm{SURE}$ consists of two parts that complement each other. 
The term $\|\rmD(\tilde{\rvx})-\tilde{\rvx}\|^2$ measures how far the denoiser’s output is from the noisy input; 
by itself this underestimates the true error since $\tilde{\rvx}$ is already corrupted. 
The divergence term acts as a correction: it captures how sensitive the denoiser is to small perturbations in its input, 
effectively accounting for the variance introduced by the noise.

Importantly, for any fixed but unknown $\rvx$,
\[
\E_{\tilde{\rvx}|\rvx}\left[\mathrm{SURE}(\rmD;\rvx+\sigma\beps) \,\big|\, \rvx\right]
= R(\rmD;\rvx),
\]
where the expectation is over the Gaussian noise $\beps\sim\mathcal{N}(\bm{0},\rmI)$. Thus, minimizing SURE (in expectation or empirically) is equivalent to minimizing the true MSE, 
while relying only on noisy data. 
In practice, averaging $\mathrm{SURE}$ over both $\rvx\sim p_{\mathrm{data}}$ and the corruption noise $\beps$ 
yields an unbiased estimate of the global MSE risk.

\paragraph{Link to Tweedie’s Formula and Bayes Optimality.}
Let $p_\sigma(\tilde{\rvx}) = \big(p_{\mathrm{data}} * \mathcal{N}(\mathbf{0}, \sigma^2 \rmI)\big)(\tilde{\rvx})$ denote the noisy marginal considered in this section.

SURE is an unbiased estimator of the mean squared error with respect to the noise, conditional on $\rvx$:
\[
\mathbb{E}_{\tilde{\rvx}| \rvx}\!\big[\mathrm{SURE}(\rmD; \tilde{\rvx})\big]
=
\mathbb{E}_{\tilde{\rvx}| \rvx}\!\big[\|\rmD(\tilde{\rvx})-\rvx\|^2\big].
\]
Hence minimizing the expected SURE equals minimizing the Bayes risk
$\mathbb{E}_{(\rvx, \tilde{\rvx})}\big[\|\rmD(\tilde{\rvx})-\rvx\|^2\big]
=\mathbb{E}_{\tilde{\rvx}}\!\big[\mathbb{E}_{\rvx| \tilde{\rvx}}\!\big[\|\rmD(\tilde{\rvx})-\rvx\|^2\big]\big]$ by the law of total expectation (tower property). This decomposition yields a pointwise optimization: for almost every $\tilde{\rvx}$,
\[
\rmD^*(\tilde{\rvx})
=\arg\min_{\mathbf z}\ \mathbb{E}_{\rvx| \tilde{\rvx}}\!\big[\|\mathbf z-\rvx\|^2\big]
=\mathbb{E}[\rvx| \tilde{\rvx}].
\]
Therefore the SURE-optimal denoiser coincides with the Bayes estimator in \Cref{subsec:why-dsm-denoising-tweedie}, and by Tweedie’s identity:
\begin{align}\label{eq:sure-tweedie-denoiser}
    \rmD^*(\tilde{\rvx})=\mathbb{E}[\rvx| \tilde{\rvx}]=\tilde{\rvx}+\sigma^2\nabla_{\tilde{\rvx}}\log p_\sigma(\tilde{\rvx}).
\end{align}

\paragraph{Relationship of SURE and Score Matching.}
The identity in \Cref{eq:sure-tweedie-denoiser} motivates parameterizing the denoiser $\rmD$ via a score field:
\[
\rmD(\tilde{\rvx})=\tilde{\rvx}+\sigma^2\rvs_{\bm\phi}(\tilde{\rvx}, \sigma),
\]
with $\rvs_{\bm\phi}(\cdot, \sigma)$ meant to approximate the noisy score
$\nabla_{\tilde{\rvx}}\log p_\sigma(\cdot)$.
Plugging  $\rmD(\tilde{\rvx})=\tilde{\rvx}+\sigma^2 \rvs_{\bm\phi}(\tilde{\rvx}, \sigma)$ in \Cref{eq:sure-obj} gives
\[
\frac{1}{2\sigma^4}\,\mathrm{SURE}(\rmD;\tilde{\rvx})
=
\Tr\!\big(\nabla_{\tilde{\rvx}} s_{\bm\phi}(\tilde{\rvx}, \sigma)\big)
+\tfrac12\|\rvs_{\bm\phi}(\tilde{\rvx}, \sigma)\|_2^2
+\text{const}(\sigma).
\]
Therefore, taking expectation with respect to $\tilde{\rvx}\sim p_\sigma$, minimizing SURE is
equivalent (up to an additive constant) to minimizing Hyv\"arinen’s alternative
score matching objective at noise level $\sigma$, with the expectation taken under $p_\sigma$ (see \Cref{eq:sm-alternative}).
Consequently, both objectives share the same minimizer, namely the denoiser in \Cref{eq:sure-tweedie-denoiser}.

\subsection{(Optional) Generalized Score Matching}

\paragraph{Motivation.}
Classical score matching, denoising score matching, and higher order variants all target
\[
\frac{\mathcal L p(\rvx)}{p(\rvx)},\quad\text{for some density } p
\]
with a linear operator $\mathcal L$ acting on the density. 
In the classical case $\mathcal L=\nabla_{\rvx}$, this gives 
\[
\nabla_{\rvx}\log p(\rvx) = \frac{\nabla_{\rvx} p(\rvx)}{p(\rvx)}
\]
The $\frac{\mathcal L p}{p}$ structure allows integration by parts to remove normalizing constants, yielding a tractable objective that depends only on samples from $p$ and the learned field $\rvs_\bphi$.
This viewpoint motivates the generalized score matching framework.

\paragraph{Generalized Fisher Divergence.}
Let $p$ be the data distribution and $q$ any model distribution.
For a linear operator $\mathcal L$ on scalar functions of $\rvx$, define the generalized Fisher divergence
\[
\mathcal D_{\mathcal L}(p \,\|\, q)
:= \int p(\rvx) 
\left\|
\frac{\mathcal L p(\rvx)}{p(\rvx)}
-
\frac{\mathcal L q(\rvx)}{q(\rvx)}
\right\|_2^2 \diff \rvx.
\]
If $\mathcal L$ is \emph{complete}, i.e., 
\[
\frac{\mathcal L p_1}{p_1}=\frac{\mathcal L p_2}{p_2} \text{ a.e.}\quad \text{implies}\quad p_1=p_2 \text{ a.e.,}
\]
then $\mathcal D_{\mathcal L}(p \,\|\, q)=0$ identifies $q=p$.
For $\mathcal L=\nabla_{\tilde\rvx}$ this recovers the classical Fisher divergence (see \Cref{eq:fisher}).

\paragraph{Score Parameterization.}
In practice we do not model a normalized density $q$.
Instead, we directly parameterize a vector field $\rvs_\bphi(\rvx)$ to approximate the generalized score $\frac{\mathcal L p(\rvx)}{p(\rvx)}$.
Consider
\[
\mathcal D_{\mathcal L}\!\left(p \,\|\, \rvs_\bphi\right)
:= \mathbb E_{\rvx\sim p}\!\left[\left\|\,\rvs_\bphi(\rvx) - \frac{\mathcal L p(\rvx)}{p(\rvx)}\right\|_2^2\right].
\]
Although $\frac{\mathcal L p(\rvx)}{p(\rvx)}$ is unknown, ``integration by parts'' makes the loss depend only on $\rvs_\bphi$.
Let $\mathcal L^\dagger$ be the adjoint of $\mathcal L$, defined by
\[
\int \big(\mathcal L f\big)^\top g
= \int f\,(\mathcal L^\dagger g)
\quad\text{for all test functions } f,g,
\]
which formally “moves” $\mathcal L$ across the integral when boundary terms vanish.
Expanding the square and applying this identity yields the tractable objective
\[
\mathcal L_{\text{GSM}}(\bphi)
= \mathbb E_{\rvx\sim p}\!\left[
\frac{1}{2}\,\|\rvs_\bphi(\rvx)\|_2^2
-
\big(\mathcal L^\dagger \rvs_\bphi\big)(\rvx)
\right] + \text{const},
\]
where the constant does not depend on $\bphi$. We use $p$ only through expectations, so the generalized score matching loss admits an empirical estimator from training data, exactly as in classical score matching.

For $\mathcal L=\nabla$ we have $\mathcal L^\dagger=-\nabla\!\,\cdot$, which recovers Hyv\"arinen’s score matching objective $\mathbb E_p\!\big[\tfrac{1}{2}\|\rvs_\bphi\|_2^2 + \nabla\!\cdot \rvs_\bphi\big]$ in \Cref{eq:sm-alternative}.

\paragraph{Examples of Operators.}
\begin{itemize}
\item \textbfs{Classical Score Matching.} Consider $\mathcal L = \nabla_{\rvx}$. 
Then the generalized score reduces to the classical score function
\[
\frac{\mathcal L p(\rvx)}{p(\rvx)}
= \nabla_{\rvx}\log p(\rvx).
\]
\item \textbfs{Denoising Score Matching.}
For additive Gaussian noise, define the operator
\[
(\mathcal L f)(\tilde{\rvx})
= \tilde{\rvx}\,f(\tilde{\rvx}) + \sigma^2\nabla_{\tilde{\rvx}} f(\tilde{\rvx}).
\]
Then
\[
\frac{\mathcal L p_\sigma(\tilde{\rvx})}{p_\sigma(\tilde{\rvx})}
= \tilde{\rvx} + \sigma^2\nabla_{\tilde{\rvx}}\log p_\sigma(\tilde{\rvx})
= \mathbb E[\rvx_0|\tilde{\rvx}],
\]
with
\[
p_\sigma(\tilde\rvx)
:=
\int
\mathcal{N}(\tilde\rvx;\rvx_0,\sigma^2\rmI)
\,p_{\mathrm{data}}(\rvx_0)\,\diff\rvx_0,
\qquad
\tilde\rvx=\rvx_0+\sigma\beps.
\]
This is exactly Tweedie's identity for additive Gaussian noise. Minimizing
$\mathcal{L}_{\mathrm{GSM}}$ with this operator therefore trains
$\rvs_{\bphi}$ to approximate the posterior-mean denoiser. If one instead uses
a scaled Gaussian location model
$\tilde\rvx=\alpha\rvx_0+\sigma\beps$, the corresponding identity contains the
factor $\alpha$ as in \Cref{eq:tweedie}.

\item \textbfs{Higher Order Targets.}
Stacking derivatives inside $\mathcal L$ exposes $\nabla^2\log p$ and higher derivatives, which align with posterior covariance and higher order cumulants.
\end{itemize}

\paragraph{Extensions and Use Cases.}
Generalized score matching extends beyond continuous variables to discrete settings, including language modeling \citep{meng2022concrete,lou2024discrete}. It also motivates score inspired training that yields denoising style objectives. This operator view unifies a range of objectives, admits empirical estimation from data, and offers a general principle for designing loss functions through suitable choices of $\mathcal L$.

\clearpage
\newpage

\section{\texorpdfstring{Multi-Noise Levels of Denoising Score Matching (NCSN)}{Multi-Noise Levels of Denoising Score Matching (NCSN)}}\label{sec:SMLD}

\subsection{Motivation}
Adding Gaussian noise with a single fixed variance to the data distribution smooths it to a certain extent, but training a score-based model at only one noise level introduces key limitations. At low levels of injected noise, Langevin dynamics struggles to traverse modes in multi-modal distributions due to vanishing gradients in low-density regions. In contrast, at high noise levels, sampling becomes easier, but the model captures only coarse structures, resulting in blurry samples that lack fine detail. Furthermore, Langevin dynamics can be slow to converge or even fail in high-dimensional spaces. Since it depends on the gradient of the log-density for guidance, poor initialization, particularly in plateau regions or near saddle points, can impede exploration or cause the sampler to get trapped in a single mode.

\begin{figure}[ht!]
    \centering
    \includegraphics[width=0.85\linewidth]{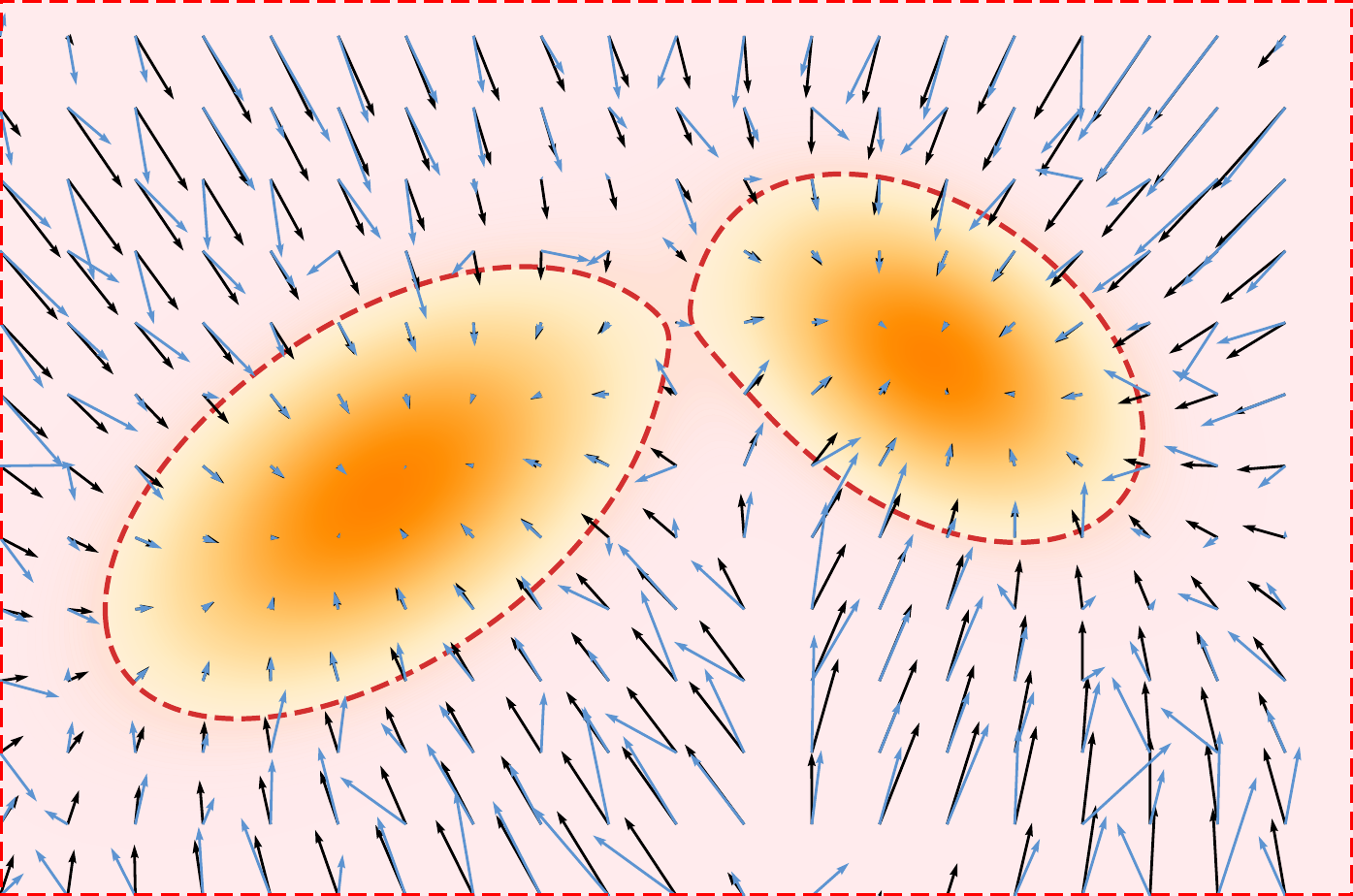}
    \caption{\textbfs{Illustration of SM inaccuracy (revisiting \Cref{fig:sm-illustration}).} the red region indicates low-density areas with potentially inaccurate score estimates due to limited sample coverage, while high-density regions tend to yield more accurate estimates. \figcredit{Created by the authors.}}
    \label{fig:sm-acc-illustration}
\end{figure}

To address these challenges, \citet{song2019generative} propose injecting Gaussian noise at multiple levels into the data distribution and jointly training a noise-conditional score network (NCSN) to estimate score functions across a range of noise scales. During generation, Langevin dynamics is applied in a noise-annealed fashion: beginning with high-noise levels to enable coarse exploration, and gradually refining toward low-noise levels to recover fine details.

\begin{figure}[th]
    \centering
    \includegraphics[width=0.95\linewidth]{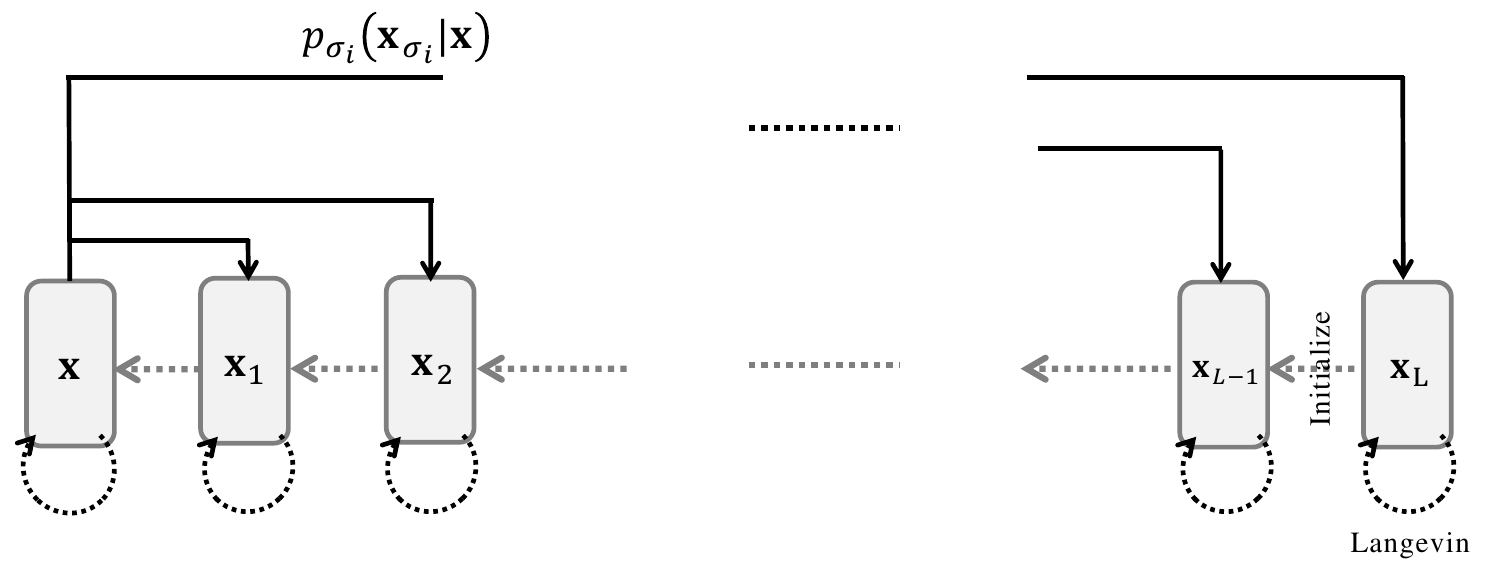}
    \caption{\textbfs{Illustration of NCSN.} The forward process perturbs the data with multiple levels of additive Gaussian noise $p_{\sigma}(\rvx_{\sigma} | \rvx)$. Generation proceeds (in dash lines) via Langevin sampling at each noise level, using the result from the current level to initialize sampling at the next lower variance. \figcredit{Created by the authors.}}
    \label{fig:smld-ddpm-graphical-model}
\end{figure}

\subsection{Training}
To overcome the limitations of score-based models trained at a single noise level, \citet{song2019generative} propose adding Gaussian noise at multiple levels to the data distribution. Specifically, a sequence of $L$ noise levels $\{\sigma_i\}_{i=1}^L$ is chosen such that
\[
0 < \sigma_1 < \sigma_2 < \cdots < \sigma_L,
\]
where $\sigma_1$ is small enough to preserve most of the data's fine details, and $\sigma_L$ is large enough to sufficiently smooth the distribution, facilitating easier training.

Each noisy sample is constructed by perturbing a clean data point $\rvx \sim p_{\text{data}}$ as $\rvx_{\sigma} = \rvx + \sigma \bm{\epsilon}$ with $ \bm{\epsilon} \sim \mathcal{N}(\bm{0}, \rmI)$. This defines the  
\vspace{-0.4cm}
\subparagraph{Perturbation Kernel:}
\[
p_{\sigma}(\rvx_{\sigma} | \rvx) := \mathcal{N}(\rvx_{\sigma}; \rvx, \sigma^2 \rmI),
\]
which induces the  
\vspace{-0.4cm}
\subparagraph{Marginal Distribution:}
\[
p_\sigma(\rvx_{\sigma}) = \int p_{\sigma} (\rvx_{\sigma} | \rvx)    p_{\text{data}}(\rvx) \diff \rvx,
\]
at each noise level $\sigma$. It presents the Gaussian smoothed data distribution.

\paragraph{Training Objective of NCSN.}

The goal is to train a noise-conditional score network $\rvs_{\bm{\phi}}(\rvx, \sigma)$ to estimate the score function $\nabla_{\rvx} \log p_\sigma(\rvx)$ for all $\sigma \in \{\sigma_i\}_{i=1}^L$. This is achieved by minimizing the DSM objective across all noise levels:
\begin{align}\label{eq:nscn}
\mathcal{L}_{\text{NCSN}}(\bm{\phi}) := \sum_{i=1}^{L} \lambda(\sigma_i)   \mathcal{L}_{\text{DSM}}(\bm{\phi}; \sigma_i),
\end{align}
where
\begin{align*}
\mathcal{L}_{\text{DSM}}(\bm{\phi}; \sigma) 
= \frac{1}{2} \mathbb{E}_{\rvx \sim p_{\text{data}}(\rvx),   \tilde{\rvx} \sim p_{\sigma}(\tilde{\rvx} | \rvx)} 
\left[ \norm{\rvs_{\bm{\phi}}(\tilde{\rvx}, \sigma) - \left( \frac{\rvx - \tilde{\rvx}}{\sigma^2} \right)}_2^2 \right],
\end{align*}
and $\lambda(\sigma_i)>0$ is a weighting function for each scale.

Minimizing this objective yields the score model $\rvs^*(\rvx, \sigma)$ that recovers the true score at each noise level:
\[
\rvs^*(\cdot, \sigma) = \nabla_{\rvx} \log p_\sigma(\cdot),  \quad \text{for all } \sigma \in \{\sigma_i\}_{i=1}^L,
\]
as it is essentially DSM minimization (see Theorem~\ref{thm:sm-dsm}).

\paragraph{Relationship with DDPM Loss.}
Let $\rvx_\sigma=\rvx+\sigma\boldsymbol{\epsilon}$ with $\boldsymbol{\epsilon}\sim\mathcal{N}(\mathbf{0},\mathbf{I})$ and let $p_\sigma$ denote the  marginal distribution. By Tweedie’s formula,
\[
\nabla_{\rvx_\sigma}\log p_\sigma(\rvx_\sigma)
= -\frac{1}{\sigma} \E \left[\boldsymbol{\epsilon} \middle|  \rvx_\sigma\right].
\]
Thus the NCSN optimum is the true score $\rvs^*(\rvx_\sigma,\sigma)=\nabla_{\rvx_\sigma}\log p_\sigma(\rvx_\sigma)$, while the Bayes optimal noise predictor under the DDPM loss \Cref{eq:ddpm-simple-loss} is $\boldsymbol{\epsilon}^*(\rvx_\sigma,\sigma)=\E[\boldsymbol{\epsilon}|\rvx_\sigma]$. They are exactly equivalent via
\[
\rvs^*(\rvx_\sigma,\sigma) = - \frac{1}{\sigma} \boldsymbol{\epsilon}^*(\rvx_\sigma,\sigma),
\qquad
\boldsymbol{\epsilon}^*(\rvx_\sigma,\sigma) = - \sigma \rvs^*(\rvx_\sigma,\sigma).
\]
In the DDPM setting, the noisy sample at step $i$ can be written as
\[
\sigma_i:=\sqrt{1-\bar{\alpha}_i^2},
\qquad
\rvx_i
=
\bar{\alpha}_i\rvx_0+\sigma_i\beps,
\qquad
\beps\sim\mathcal{N}(\bm{0},\rmI).
\]
Therefore, the score and the optimal noise prediction contain the same
information:
\[
\rvs^*(\rvx_i,i)
=
-\frac{1}{\sigma_i}
\E\left[\beps|\rvx_i\right].
\]
In other words, predicting the noise in DDPM is equivalent, up to the known
scale factor $1/\sigma_i$, to predicting the score at noise level $i$.

We will systematically compare and summarize this equivalence of parameterizations in \Cref{ch:all-equivalent}.

\subsection{Sampling}

\begin{algorithm}[th]
\caption{\label{alg:smld} Annealed Langevin Dynamics}
\begin{algorithmic}[1]
\Require Trained score $\rvs_{\bm{\phi}^\times}(\cdot,\sigma_\ell)$,
step sizes $\eta_\ell$, and Langevin iteration budgets $N_\ell$
for each noise level $\ell=L,\ldots,1$; initialization distribution
$p_{\mathrm{init}}$
\State $\rvx \sim p_{\mathrm{init}}$
\Statex\Comment{Initialize from unstructured noise at the largest noise level}
\For{$\ell=L,L-1,\ldots,1$}
    \State $\tilde{\rvx}_0 \gets \rvx$
    \Statex\Comment{Initialize this noise level from the previous level's output}
    \For{$n=0$ \textbf{to} $N_\ell-1$}
        \State $\bm{\epsilon}_n \sim \mathcal{N}(\bm{0},\rmI)$
        \State $\tilde{\rvx}_{n+1}
        \gets
        \tilde{\rvx}_n
        +\eta_\ell
        \rvs_{\bm{\phi}^\times}(\tilde{\rvx}_n,\sigma_\ell)
        +\sqrt{2\eta_\ell}\,\bm{\epsilon}_n$
    \EndFor
    \State $\rvx \gets \tilde{\rvx}_{N_\ell}$
    \Statex\Comment{Pass the final state to the next smaller noise level, if any}
\EndFor
\Ensure $\rvx$
\end{algorithmic}
\end{algorithm}

With trained score networks available at multiple noise levels
\[
\rvs_{\bm{\phi}^\times}(\cdot, \sigma_1), \quad\rvs_{\bm{\phi}^\times}(\cdot, \sigma_2), \quad \cdots, \quad \rvs_{\bm{\phi}^\times}(\cdot, \sigma_{L-1}), \quad    \rvs_{\bm{\phi}^\times}(\cdot, \sigma_L),
\]
the sampling procedure known as \emph{annealed Langevin dynamics}~\citep{song2019generative} generates data by progressively denoising from a high noise level $\sigma_L$ down to a low noise level $\sigma_1 \approx 0$.

With trained scores at noise levels
$0<\sigma_1<\cdots<\sigma_L$, annealed Langevin dynamics generates samples by
moving gradually from high noise to low noise.
Starting from a simple initialization
$\rvx\sim p_{\mathrm{init}}$, it first applies Langevin dynamics at the largest
noise level $\sigma_L$, whose score
$\rvs_{\bm{\phi}^\times}(\cdot,\sigma_L)$ guides the samples toward the
corresponding smoothed distribution $p_{\sigma_L}$.
The resulting samples are then used to initialize Langevin dynamics at the next
smaller noise level. Repeating this procedure progressively refines the samples
until the smallest noise level $\sigma_1$ is reached.

At each noise level $\sigma_\ell$, Langevin dynamics performs the update
\[
\tilde{\rvx}_{n+1}
=
\tilde{\rvx}_n
+
\eta_\ell
\rvs_{\bm{\phi}^\times}(\tilde{\rvx}_n,\sigma_\ell)
+
\sqrt{2\eta_\ell}\,\bm{\epsilon}_n,
\qquad
\bm{\epsilon}_n\sim\mathcal{N}(\bm{0},\rmI),
\]
where the output from the previous noise level provides the initialization
$\tilde{\rvx}_0$.
A common choice is to scale the step size with the noise level as
\[
\eta_\ell
=
\delta\frac{\sigma_\ell^2}{\sigma_1^2},
\qquad
\delta>0.
\]
The algorithm performs Langevin updates at every noise level, including
$\sigma_1$, so the final sample is obtained after refinement at the smallest
noise level. \Cref{alg:smld} summarizes the procedure.

The initial distribution $p_{\mathrm{init}}$ simply provides a starting point
for the Markov chain. At the largest noise level, Langevin dynamics moves this
initial distribution toward $p_{\sigma_L}$, after which the successive noise
levels gradually guide the samples toward increasingly less smoothed versions
of the data distribution.

\paragraph{Slow Sampling Speed of NCSN.}
NCSN generates samples using annealed MCMC (commonly Langevin dynamics) across noise scales $\{\sigma_i\}_{i=1}^L$.  
For each scale $\sigma_i$, it performs $K$ iterative updates of the form ``update $\tilde{\mathbf{x}}_n$ using the score $\rvs_{\bm{\phi}^\times}(\tilde{\mathbf{x}}_n,\sigma_i)$ plus a small random perturbation'', each requiring a forward pass through the score network. Two factors necessitate large $L \times K$:  
\begin{enumerate}
    \item[(i)] \textbfs{Local Accuracy and Stability:} the learned score is reliable only for small perturbations, requiring small step sizes and many iterations per noise level to avoid bias or instability;
    \item[(ii)] \textbfs{Slow Mixing in High Dimensions:} local MCMC moves explore multimodal, high-dimensional targets inefficiently, demanding many iterations to reach typical data regions.
\end{enumerate}
Because updates are strictly sequential (each iteration depends on the previous one) and each requires an expensive network evaluation, the overall cost is $\mathcal{O}(L  K)$ sequential network passes, making sampling computationally slow.



\clearpage
\newpage

\section{Summary: A Comparative View of NCSN and DDPM}

We begin by comparing the graphical models of NCSN and DDPM in \Cref{fig:smld-ddpm-graphical-model}, with key differences and similarities summarized in \Cref{tb:comparison-smld-ddpm}.

\begin{table}[th]
  \caption{Comparisons of NCSN and DDPM}
  \small
  \centering
  \resizebox{\textwidth}{!}{
  \begin{tabular}{cccc}
     \toprule
    
         &   \textbfs{NCSN}   & \textbfs{DDPM}  \\
        \midrule
    $\rvx_{i+1}|\rvx_{i}$ & Derive as $\rvx_{i+1} =  \rvx_i + \sqrt{\sigma^2_{i+1}-\sigma^2_{i}}\bm{\epsilon}$   &  \cc{15}Define as $\rvx_{i+1} = \sqrt{1-\beta_{i}} \rvx_{i} + \sqrt{\beta_{i}}\bm{\epsilon}$  \\
      $\rvx_{i}|\rvx$   & \cc{15}Define as $\rvx_{i} =  \rvx + \sigma_{i}\bm{\epsilon}$
       &  Derive as $\rvx_i = \bar{\alpha}_i\rvx + \sqrt{1 - \bar{\alpha}_i^2}\bm{\epsilon}$ \\
   $p_{\text{prior}}$   & $ \mathcal{N}(\bm{0}, \sigma_L^2\rmI)$    &  $\mathcal{N}(\bm{0}, \rmI)$  \\
   \midrule
        Loss 
          &  
        $\mathbb{E}_{i} \mathbb{E}_{p_{\text{data}}(\rvx)} 
        \mathbb{E}_{\bm{\epsilon} \sim \mathcal{N}\left(\mathbf{0}, \rmI\right)}   
    \left[ \norm{\rvs_{\bm{\phi}}(\rvx_i, \sigma_i) +   \frac{\bm{\epsilon}}{\sigma_i}}_2^2 \right] $ 
            &  $      \mathbb{E}_{i}\mathbb{E}_{p_{\text{data}}(\rvx)} \mathbb{E}_{\bm{\epsilon}\sim \mathcal{N}(\bm{0},\rmI)}\left[ \norm{\bm{\epsilon}_{\bm{\phi}}(\rvx_i, i) - \bm{\epsilon}}_2^2 \right]$ \\
    \midrule
     Sampling   & \begin{tabular}{@{}c@{}} Apply Langevin per layer;\\ use output to initialize the next \end{tabular}  &  \begin{tabular}{@{}c@{}}Traversing the Markovian chain with \\ $p_{\bm{\phi}^\times}(\rvx_{i-1} | \rvx_i)$   \end{tabular} \\
 \bottomrule
  \end{tabular}
  }
  \label{tb:comparison-smld-ddpm}
\end{table}

\paragraph{A Shared Bottleneck.}
Despite their different formulations, both NCSN and DDPM rely on dense time discretization. This leads to a critical limitation: sampling often requires hundreds or even thousands of iterations, making generation slow and computationally intensive.

\begin{question}\label{ques:dm-sampling-slow}
    How can we accelerate sampling in diffusion models?
\end{question}

We return to this challenge in \Cref{ch:solvers}, which develops more efficient samplers; \Cref{ch:distillation}, which compresses pretrained diffusion models requiring hundreds of sampling steps into fewer-step student generators; and \Cref{ch:fast-scratch}, which studies how to learn fast few-step generators directly from scratch under diffusion-inspired principles.




\newpage

\section{Closing Remarks}\label{sec:ch3_cr}

This chapter has charted a second major path to diffusion models, beginning from the score-based perspective rooted in Energy-Based Models (EBMs). We started by identifying the core challenge of EBMs—the intractable partition function—and introduced the score function, $\nabla_{\mathbf{x}}\log{p(\mathbf{x})}$, as a powerful tool that circumvents this issue entirely.

Our journey led us from classic score matching to its more scalable and robust variant, Denoising Score Matching (DSM). Through DSM, we saw how perturbing data with noise enables a tractable training objective, once again leveraging a conditioning strategy to create a simple regression target. Furthermore, we established a profound connection between score estimation and the act of denoising via Tweedie's formula, which showed that the score provides the precise direction needed to estimate a clean signal from its noisy observation.

This principle was then extended from a single noise level to a \emph{ladder}
of noise levels with Noise Conditional Score Networks (NCSN), which learn a
single score model conditioned on multiple noise scales and generate samples
via annealed Langevin dynamics. The next chapter takes the additional step from
this discrete ladder to a continuum of noise levels through the Score SDE
framework.

This convergence is no coincidence; it hints at a deeper, unified mathematical structure. The limitations of these discrete-time models motivate the need for a more general framework. In the next chapter, we will take this crucial step:
\begin{enumerate}
    \item We will move into a continuous-time perspective, showing that both DDPMs and NCSNs can be elegantly unified as different discretizations of a single, powerful process described by a Stochastic Differential Equation (SDE).
    \item This Score SDE framework will formally connect the variational and score-based views, recasting the problem of generation as one of solving a differential equation.
\end{enumerate}

This unifying lens will not only provide profound theoretical clarity but also unlock a new class of advanced numerical methods designed to tackle the fundamental challenge of slow sampling. 
\chapter{Diffusion Models Today: Score SDE Framework}\label{ch:score-sde}

\epigraph{
    \textit{There is only one precise way of presenting the laws, and that is by means of differential equations. They have the advantage of being fundamental and, so far as we know, precise.
}}{Richard P. Feynman}

So far, we have studied diffusion models from two perspectives: the variational
view and the score-based view, the latter naturally emerging from the EBM
formulation. We now take the next step and move to the \emph{continuous-time
framework}. At its core lies the \emph{Score SDE}~\citep{song2020score}, which
provides a continuous-time view that brings DDPM and NCSN under the same
framework.

Instead of working with only a finite sequence of noise levels, we imagine
inserting more and more intermediate levels between data and noise. In the
continuous-time limit, the noise level varies smoothly with time, giving the
basic picture
\[
    p_{\mathrm{data}}
    \;\xrightarrow{\text{forward SDE}}\;
    p_T \approx p_{\mathrm{prior}},
    \qquad
    p_{\mathrm{prior}}
    \;\xrightarrow{\text{reverse dynamics}}\;
    p_{\mathrm{data}}.
\]
The forward SDE describes how data is gradually corrupted into noise, playing
the continuous-time role of the noise schedule in DDPM or the sequence of noise
levels in NCSN. Once this forward process is specified, the main challenge is
generation in the opposite direction: starting from a simple noise distribution,
how can we recover the data distribution?

The key quantity is the time-dependent score
\[
    \nabla_{\rvx}\log p_t(\rvx),
\]
which describes the local geometry of the noisy distribution at every time
$t$. Once this score is known, or approximated by a neural network, it determines the reverse-time dynamics used for generation, transforming noise into data. This leads to two complementary
descriptions: a stochastic \emph{reverse-time SDE} and a deterministic
\emph{probability flow ODE} (PF-ODE). Although their sample trajectories are
different, they can follow the same evolution of marginal distributions.

This continuous-time view turns generation into the problem of solving
differential equations, opening the door to a broad set of tools from stochastic
processes, differential equations, and numerical analysis. In particular, once
the generative dynamics are written as an SDE or ODE, standard numerical
methods can be used to simulate them, while more advanced solvers can improve
sampling efficiency and accuracy. This perspective therefore provides not only
a unified mathematical description of diffusion models, but also a practical
foundation for understanding and designing faster sampling methods developed
later in this book.

\chapterlearning{
  \item understand how the discrete noise levels of DDPM and NCSN lead to a continuous-time forward SDE that gradually transforms data into noise;
  \item understand how generation from noise back to data can be described either by a stochastic reverse-time SDE or by a deterministic probability flow ODE (PF-ODE);
  \item explain how the reverse-time SDE and PF-ODE can follow the same sequence of marginal distributions even though their individual sample trajectories are different, and understand how the Fokker--Planck equation describes the continuous-time evolution of probability densities;
  \item understand how the time-dependent score is learned by denoising score matching and then used for sampling through the reverse-time SDE or PF-ODE;
  \item understand how the PF-ODE can also be run forward for inversion and track changes in probability density through the continuous change-of-variables formula, enabling likelihood evaluation;
  \item recognize VE and VP SDEs as two important instances of the same continuous-time diffusion framework, connecting back to NCSN and DDPM.
}{
Readers new to stochastic differential equations may consult the differential-equation appendices in \Cref{app:de} as needed. For a first pass, focus on the roles of the forward SDE, reverse-time SDE, and PF-ODE, together with their training and sampling procedures. The Gaussian example and optional derivations can be revisited after the main picture is clear.
}

\newpage

\section{Score SDE: Its Principles}\label{sec:dm-today}

\begin{figure}[th!]
    \centering
    \includegraphics[width=\linewidth]{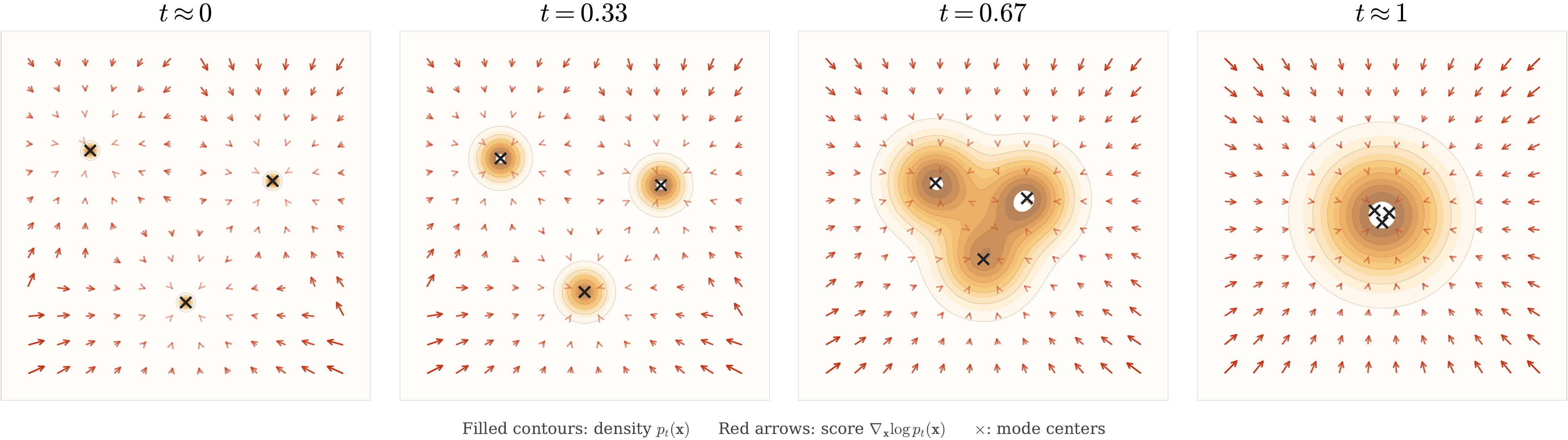}
    \caption{\textbfs{Illustration of the time-dependent score landscape.}
As the noise level increases, the perturbed density $p_t(\rvx)$ evolves from a multimodal distribution (data) concentrated near the data modes to a broad nearly unimodal distribution (prior).
Accordingly, the score field $\nabla_{\rvx}\log p_t(\rvx)$ changes from a sharply local vector field that pulls samples toward nearby modes to a smooth global field that points toward the overall center of mass.
This time-dependent score field is the key object learned in score-based diffusion models (i.e., Score SDE). \figcredit{Created by the authors with AI-assisted coding.}}
    \label{fig:time-dependent-score}
\end{figure}


The use of multiple noise scales has been a crucial ingredient in the success of NCSN and DDPM frameworks. In this section, we introduce the foundation of the \emph{Score SDE}~\citep{song2020score}, which elevates this idea by considering a continuum of noise levels. A continuous-time limit of forward and reverse diffusion processes had already been noted by \citet{sohl2015deep}, but \citet{song2020score} make this perspective central by formulating the data evolution as a stochastic/ordinary differential equation, where the noise level increases smoothly over time. Central to this formulation is the \emph{time-dependent score function} $\nabla_{\rvx}\log p_t(\rvx)$: at each noise level~$t$, it points toward regions of higher density under the current noisy distribution~$p_t$, and it is precisely this quantity that determines how to reverse the diffusion. As \Cref{fig:time-dependent-score} illustrates, the score field evolves from sharply converging toward individual data modes at small~$t$ to pointing gently toward the global center at large~$t$. This continuous-time formulation not only unifies prior discrete-time models but also provides a principled and flexible foundation for generative modeling: learning this family of score functions across all noise levels reduces generation to the problem of solving differential equations.



\subsection{Motivation: From Discrete to Continuous-Time Processes}\label{subsec:scoresde-motivation}

We revisit the forward noise injection schemes of NCSN and DDPM. NCSN uses a sequence of increasing noise levels $\{\sigma_i\}_{i=1}^L$. 
Each clean sample $\rvx_0 \sim p_{\mathrm{data}}$ is perturbed as
\[
\rvx_{\sigma_i} = \rvx_0 + \sigma_i \bm{\epsilon}_i, 
\qquad \bm{\epsilon}_i \sim \mathcal{N}(\bm{0}, \rmI).
\]
DDPM instead injects noise incrementally with a variance schedule $\{\beta_i\}_{i=1}^L$\footnote{In \Cref{sec:ddpm}, the DDPM forward step is written as $\rvx_i = \sqrt{1-\beta_i^2}\,\rvx_{i-1} + \beta_i\,\bm{\epsilon}_i$, where $\beta_i$ denotes the standard deviation of the injected noise and $\alpha_i := \sqrt{1-\beta_i^2}$. Here and throughout this chapter, we follow the convention of \citet{song2020score}, where $\beta_i$ instead denotes the variance of the injected noise, so that $\rvx_i = \sqrt{1-\beta_i}\,\rvx_{i-1} + \sqrt{\beta_i}\,\bm{\epsilon}_i$. The two conventions are related by $\beta_i^{\textup{(here)}} = \bigl(\beta_i^{\textup{(Ch.\ \ref{ch:variational})}}\bigr)^2$ and yield identical dynamics.}:
\[
\rvx_i = \sqrt{1 - \beta_i}\, \rvx_{i-1} + \sqrt{\beta_i}\, \bm{\epsilon}_i, 
\qquad \bm{\epsilon}_i \sim \mathcal{N}(\bm{0}, \rmI).
\]

We view them together on a discrete time grid, where the sequential update from $\rvx_t$ to $\rvx_{t+\Delta t}$ takes the form\footnote{For convenience, we use $\rvx(t)$ and $\rvx_t$ interchangeably (and similarly for other time-dependent variables) to denote samples at time $t$.}:
\begin{align*}
    \text{\textbfs{NCSN:}} \,\,
    & \rvx_{t+\Delta t} = \rvx_t + \sqrt{\sigma_{t+\Delta t}^2 - \sigma_t^2} \bm{\epsilon}_t 
    &&\approx \rvx_t + \sqrt{\frac{\diff \sigma_t^2}{\diff t} \Delta t}  \bm{\epsilon}_t \\[0.5em]
    \text{\textbfs{DDPM:}} \,\,
    & \rvx_{t+\Delta t} = \sqrt{1 - \beta(t)\Delta t} \rvx_t + \sqrt{\beta(t)\Delta t} \bm{\epsilon}_t 
    &&\approx \rvx_t - \frac{1}{2} \beta(t) \rvx_t  \Delta t + \sqrt{\beta(t)\Delta t}  \bm{\epsilon}_t,
\end{align*}
where $\bm{\epsilon}_t \sim \mathcal{N}(\bm{0}, \rmI)$. Interestingly, both noise injection processes follow a common structural pattern:
\begin{align}\label{eq:euler-discrete-sde}
    \rvx_{t+\Delta t} \approx \rvx_t + \rvf(\rvx_t, t) \Delta t + g(t) \sqrt{\Delta t}\, \bm{\epsilon}_t,
\end{align}
with $\rvf: \mathbb{R}^D \times \mathbb{R} \to \mathbb{R}^D$ and $g: \mathbb{R} \to \mathbb{R}$ given by:
\begin{align*}
    \text{\textbfs{NCSN:}} \quad 
    & \rvf(\rvx, t) = 0, \qquad g(t) = \sqrt{\frac{\diff \sigma^2(t)}{\diff t}} \\[0.5em]
    \text{\textbfs{DDPM:}} \quad 
    & \rvf(\rvx, t) = -\frac{1}{2} \beta(t)\, \rvx, \qquad g(t) = \sqrt{\beta(t)}.
\end{align*}

\begin{figure}
    \centering
    \includegraphics[width=0.9\linewidth]{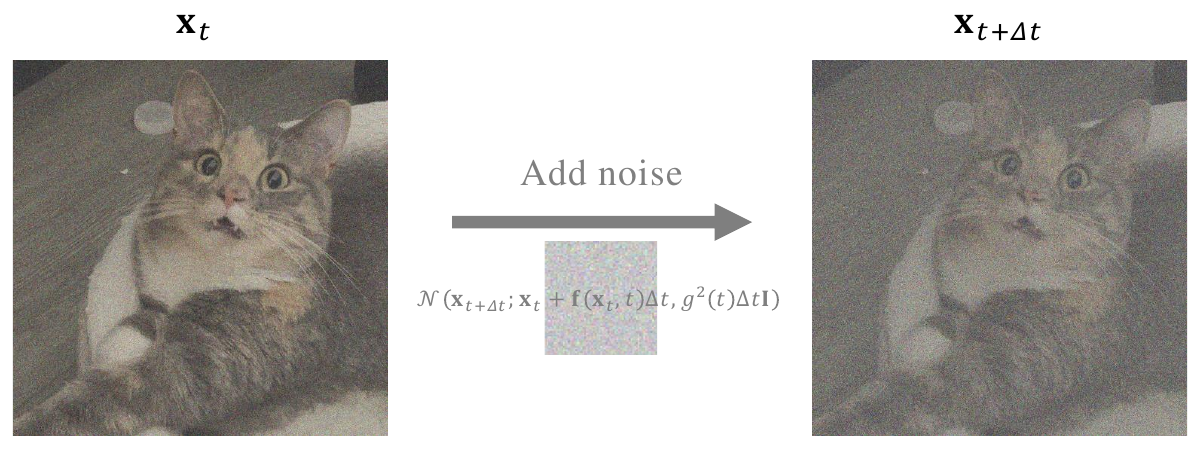}
    \caption{\textbfs{Illustration of the discrete-time noise-adding step.} It adds noise from $t$ to $t + \Delta t$ with mean drift $\rvf(\rvx_t, t)$ and diffusion coefficient $g(t)$. \figcredit{Created by the authors.}}
    \label{fig:sde-forward-noise}
\end{figure}

This formulation corresponds to the following Gaussian transition:
\begin{align}\label{eq:transition-forward-f-g}
    p(\rvx_{t+\Delta t} |\rvx_t) := \mathcal{N}\!\left(\rvx_{t+\Delta t}; \rvx_t + \rvf(\rvx_t, t)\Delta t,\, g^2(t)\Delta t\, \rmI\right),
\end{align}
where, by a slight abuse of notation, we treat $\rvx_t$ as a fixed sample and $\rvx_{t+\Delta t}$ as a random variable.

As $\Delta t \to 0$ (which can be conceptually understood as preparing \emph{infinitely many} layers of noises), the discrete-time process converges to a continuous time SDE evolving forward in time\footnote{The forward kernel in \Cref{eq:transition-forward-f-g} converges, as $\Delta t\to0$, to the solution of the corresponding Itô SDE. A fully rigorous proof relies on advanced results which we defer to the literature.}:
\begin{align*}
    \diff \rvx(t) = \rvf(\rvx(t), t) \diff t + g(t) \diff \rvw(t),
\end{align*}
where $\rvw(t)$ is a standard Wiener process (or Brownian motion).
\rmkb{While a full formal definition is not necessary here, a \emph{Wiener process} is a continuous-time stochastic process $ \rvw(t) $ that starts at zero, has independent increments, and satisfies that for any $ s < t $, the increment $ \rvw(t) - \rvw(s) $ is normally distributed with mean zero and variance $ t - s $. It represents the accumulation of independent Gaussian fluctuations over time, and although it is almost surely continuous, it is nowhere differentiable.

Over an infinitesimal time interval $[t, t + \diff t]$, the increment of a Wiener process is defined as
\[
\diff \rvw(t) := \rvw(t + \diff t) - \rvw(t),
\]
which is modeled as a Gaussian random variable with zero mean and variance $\diff t$:
\[
\diff \rvw(t) \sim \mathcal{N}(\bm{0}, \diff t \rmI).
\]}

A brief introduction to the foundations of SDEs is provided in \Cref{app:sde-intro}, with a more advanced discussion in \Cref{app:Ito}. However, we can conceptually understand the connection between the discrete and continuous formulations as follows:
\begin{itemize}
    \item $\rvx(t+\Delta t) - \rvx(t) \approx \diff \rvx(t)$,
    \item $\Delta t \approx \diff t$,
    \item $\sqrt{\Delta t} \bm{\epsilon}_t \sim \mathcal{N}(\bm{0}, \Delta t \rmI) \approx \diff \rvw(t)$.
\end{itemize}
Once the drift $\rvf(\rvx,t)$ and diffusion $g(t)$ are specified, the forward time SDE automatically induces a reverse time SDE that transports the terminal noise distribution back to the data distribution. The reverse dynamics involve only a single unknown term, surprisingly the \emph{score function at each continuous-time level}. This identifies score matching as the training objective; once the score is learned, sampling amounts to numerically integrating the reverse time SDE with the learned score.

While \Cref{sec:scoresde-training-sampling} presents practical implementations, we first examine the theoretical foundations of the forward and reverse processes in \Cref{subsec:forward-sde} and \Cref{subsec:reverse-sde}.

\subsection{Forward-Time SDEs: From Data to Noise}\label{subsec:forward-sde}
\begin{figure}[th!]
    \centering
    \includegraphics[width=\textwidth]{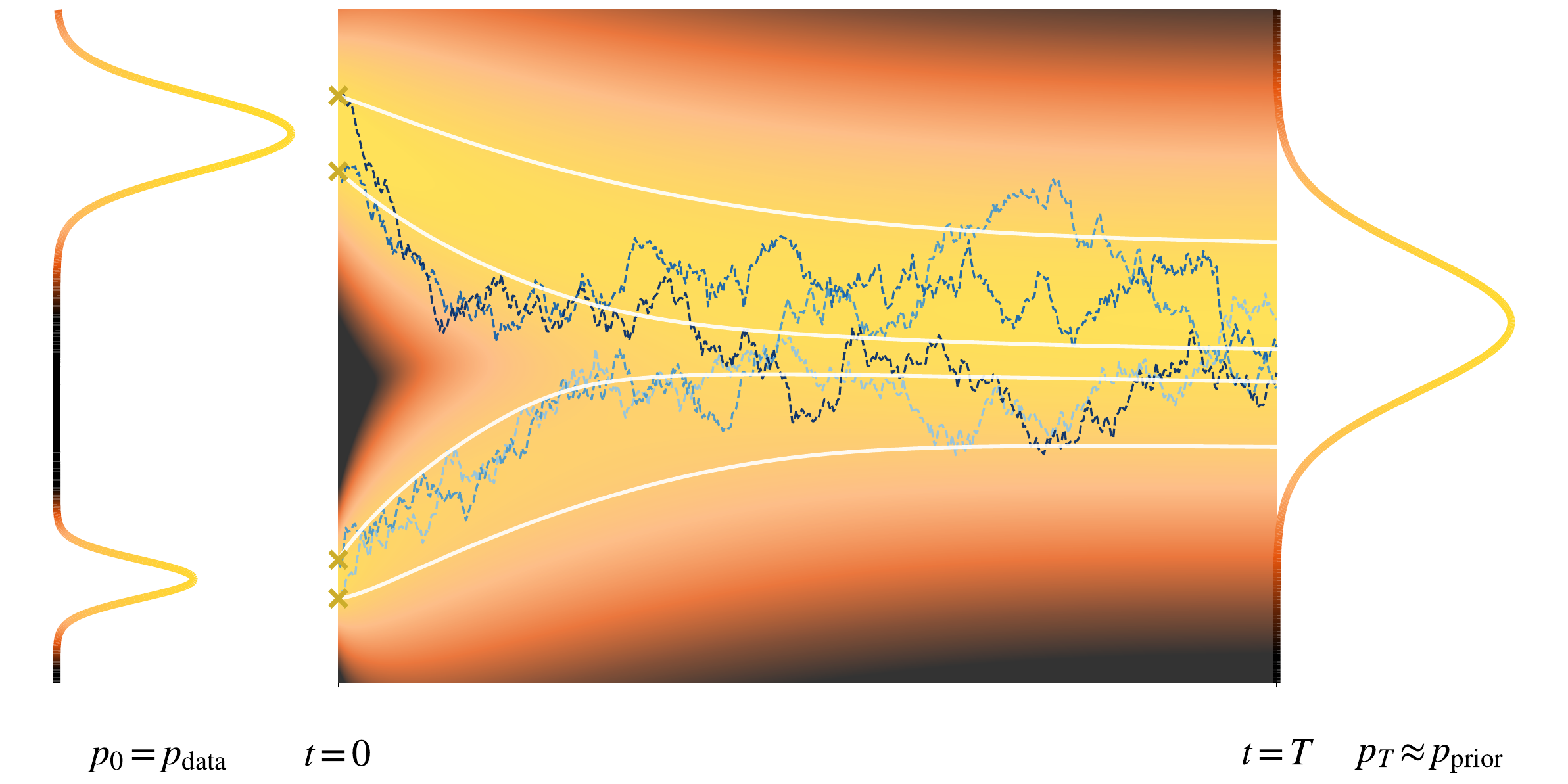}
        \caption{\textbfs{(1D) Visualization of the forward process in a diffusion model.} The process starts from initial points sampled (denoted as ``$\times$'') from a  complex bimodal data distribution ($p_0 = p_{\mathrm{data}}$) and evolves toward a simple, unimodal Gaussian prior ($p_T \approx p_{\mathrm{prior}}$). The background heatmap illustrates the evolving marginal probability density, $p_t$, which smooths over time. Sample trajectories are shown evolving from $t=0$ to $t=T$, comparing the stochastic forward SDE process (blue paths) with its deterministic counterpart, the PF-ODE (white paths). Note that the PF-ODE is a deterministic transport map for densities, not generally the mean of sample paths started from a single point. \figcredit{Created by the authors.}}

    \label{fig:sde-density-traj}
\end{figure}

With this formulation, earlier methods based on discrete time, such as NCSN~\citep{song2019generative} and DDPM~\citep{sohl2015deep,ho2020denoising}, can be unified under the \emph{continuous-time} framework through a stochastic process $\rvx(t)$ governed by a forward SDE defined on the interval $[0, T]$:
\begin{mdframed}
\begin{equation}\label{eq:sde_forward}
    \diff\rvx(t) = \mathbf{f}(\rvx(t), t) \diff t + g(t) \diff\rvw(t), \quad \rvx(0) \sim p_{\mathrm{data}}.
\end{equation}
\end{mdframed}
Here, $\mathbf{f}(\cdot, t)\colon\mathbb{R}^D\to\mathbb{R}^D$ is the drift, $g(t) \in \mathbb{R}$ is the scalar diffusion coefficient, and $\rvw(t)$ denotes a standard Wiener process. We refer to this as the \emph{forward SDE}, which describes how clean data is gradually perturbed into noise over time.

Once the drift $\mathbf{f}$ and diffusion coefficient $g$ are specified, the forward process is fully determined, describing how the data variable is progressively corrupted through the injection of Gaussian noise. In particular, two families of time-dependent densities are induced:
\paragraph{Perturbation Kernels.} 
The conditional law
\[
p_t(\rvx_t |\rvx_0)
\]
describes how a clean data sample $ \rvx_0 \sim p_{\mathrm{data}} $ evolves into its noisy counterpart $\rvx_t$ at time $t$. In general, the drift term $\rvf(\rvx,t)$ in \Cref{eq:sde_forward} can be an arbitrary function of $\rvx$, but a common and analytically convenient choice is to assume it is affine:
\begin{equation}\label{eq:affine-drift}
    \rvf(\rvx,t) = f(t) \rvx, 
\end{equation}
where $f(t)$ is a scalar function of $t$, typically taken to be non-positive. 
Under this structure, the conditional process remains Gaussian given
$\rvx_0$, and the perturbation kernel admits a closed-form solution obtained
by solving the associated mean--variance ODEs~\citep{sarkka2019applied}
(see also \Cref{subsec:scoresde-perturbation-kernel}). {In particular,
\[
p_t(\rvx_t |\rvx_0) = \mathcal{N} \big(\rvx_t; \rvm(t), P(t)\mathbf I_D\big),
\]
with
\[
\rvm(t) = \exp \Big(\int_0^t f(u) \diff u\Big) \rvx_0,
\quad
P(t) = \int_0^t \exp \Big(2 \int_s^t f(u) \diff u\Big) g^2(s) \diff s,
\]
and initial conditions $\rvm(0)=\rvx_0$, $P(0)=0$.}

This explicit form allows one to sample $\rvx_t$ given $\rvx_0$ directly, without numerically integrating the SDE, hence the term \emph{simulation-free}. Both NCSN and DDPM fall into this affine-drift setting. 

In the remainder, we develop the general theory for arbitrary drifts $\rvf(\rvx,t)$, but will return to the affine drift when closed-form analysis is useful.

\paragraph{Marginal Densities.} 
The time-marginal density $ p_t(\rvx_t) $ is obtained by integrating over the perturbation kernel:
\begin{equation}\label{eq:scoresde-oracle-marginal}
    p_t(\rvx_t) := \int p_t(\rvx_t|\rvx_0) p_{\mathrm{data}}(\rvx_0) \diff \rvx_0, \quad \text{with } p_0 = p_{\mathrm{data}}.
\end{equation}

By choosing the coefficients $f(t)$ and $g(t)$ appropriately, the forward process gradually adds noise until the influence of the initial state is effectively forgotten. As $T$ becomes large, the conditional distribution $p_T(\mathbf{x}_T |\mathbf{x}_0)$ no longer depends on $\mathbf{x}_0$, because its mean evolves as
\[
\rvm(T) = \exp\!\Big(\int_0^T f(u)\diff u\Big) \mathbf{x}_0  \longrightarrow  \mathbf{0}, \quad \text{as } T\to\infty,
\]
provided $f(u)$ is non-positive so that the exponential factor decays. At the same time, the variance grows and stabilizes to match a chosen prior distribution. Consequently, the marginal
\[
p_T(\mathbf{x}_T) = \int p_T(\mathbf{x}_T |\mathbf{x}_0) p_{\mathrm{data}}(\mathbf{x}_0)\diff\mathbf{x}_0,
\]
which initially represents a complicated mixture over data samples, converges to a simple prior $p_{\mathrm{prior}}$, typically Gaussian. In this limit,
\[
p_T(\mathbf{x}_T) \approx p_{\mathrm{prior}}(\mathbf{x}_T) 
\quad \text{and} \quad 
p_T(\mathbf{x}_T |\mathbf{x}_0) \approx p_{\mathrm{prior}}(\mathbf{x}_T),
\]
so the forward process maps any data distribution into a tractable prior, providing a convenient starting point for reversal and generation.

\subsection{Reverse-Time Stochastic Process for Generation}\label{subsec:reverse-sde}

\begin{figure}[th!]
    \centering
    \includegraphics[width=\textwidth]{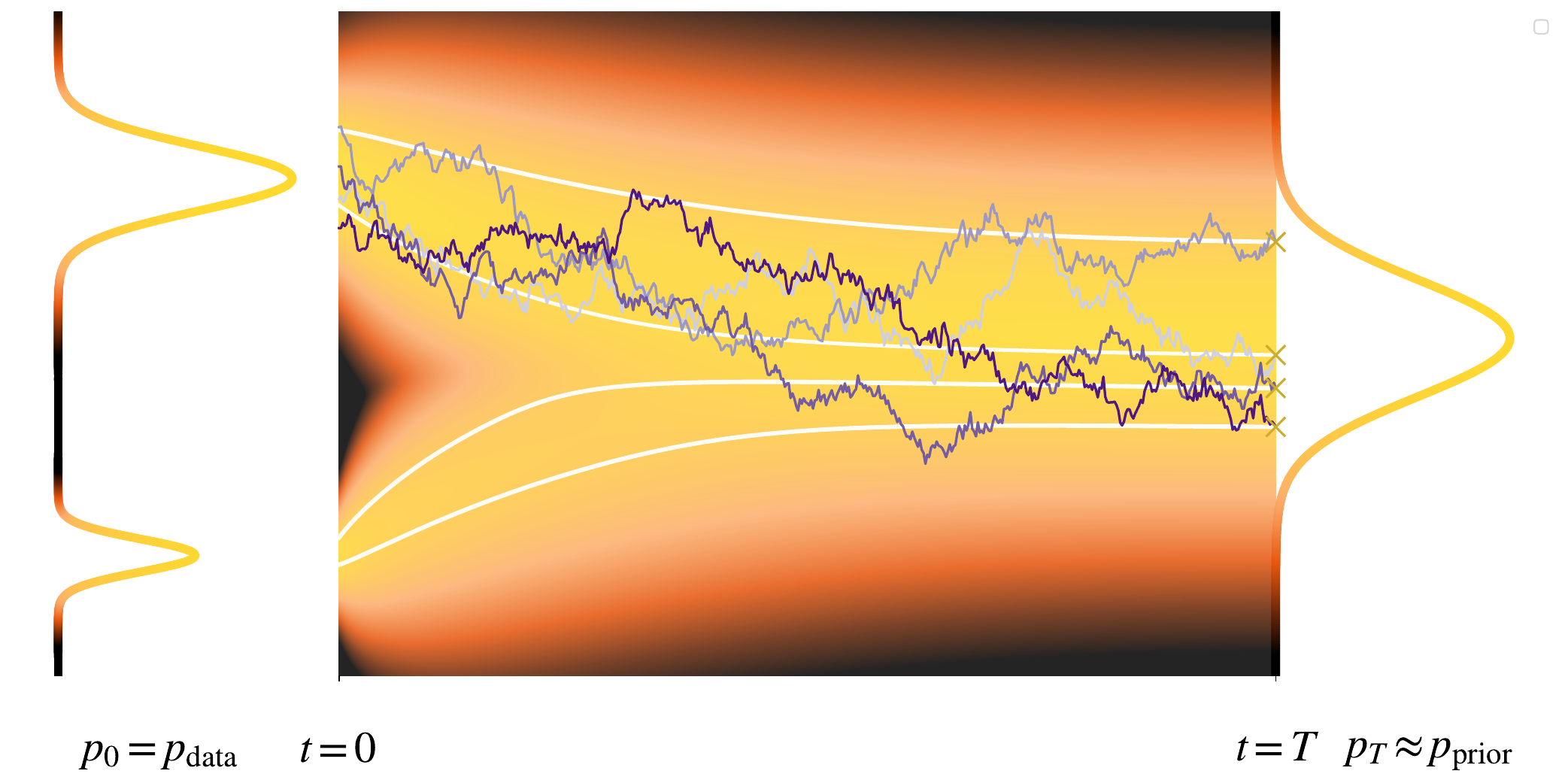}
    \caption{\textbfs{Visualization of the reverse-time stochastic process for data generation.} 
It begins from samples drawn from a simple prior distribution 
($p_{\mathrm{prior}}$) at $t=T$ (denoted as ``$\times$''), which are evolved backward 
in time using a reverse-SDE. The resulting trajectories terminate at $t=0$ 
and collectively form the target bimodal data distribution 
($p_0 = p_{\mathrm{data}}$). The background heatmap illustrates how the 
probability density is gradually transformed from a simple Gaussian into 
the complex target distribution.
\figcredit{Created by the authors.}}
    \label{fig:sde-density-back-traj}
\end{figure}
Intuitively, data generation from noise can be achieved by ``reversing'' the forward process: starting from a random point sampled from the prior distribution and evolving it backward in time to obtain a generated sample. For deterministic systems (that is, ODEs), this idea works naturally. Since no randomness is involved, reversing time simply means tracing the trajectory of a point in the opposite direction along the same path as in the forward process\footnote{Technically, this corresponds to solving the ODE with a time-flipping substitution $t \leftrightarrow T - t$.}. In contrast, SDEs incorporate stochasticity at every time step, meaning that a single point can evolve along many plausible random trajectories. As a result, reversing such processes is more subtle\footnote{Naively flipping time does not yield the correct reverse process.}. 

While individual stochastic trajectories are not reversible, the remarkable insight is that the distribution over these trajectories can be reversed. This is formalized by a foundational result from \citet{anderson1982reverse}, which shows that the time-reversed process $\{ \bar{\rvx}(t) \}_{t \in [0, T]}$\footnote{We use the ``bar'' notation to distinguish the reverse process $\{ \bar{\rvx}(t) \}_{t \in [0, T]}$ from the forward process $\{ \rvx(t) \}_{t \in [0, T]}$, defined by the forward-time SDE.} of the forward process in \Cref{eq:sde_forward} is itself governed by a well-defined SDE. This reverse-time process evolves from $T$ to $0$, and its dynamics are given by:
\begin{mdframed}
\begin{align}\label{eq:sde_backward}
\begin{aligned}[t]
    \diff\bar{\rvx}(t) = \left[\mathbf{f}(\bar{\rvx}(t), t) {\color{orange} - g^2(t) \nabla_{\rvx} \log p_t(\bar{\rvx}(t))} \right] \diff t + g(t) \diff \bar{\rvw}(t), \\
    \hfill\bar{\rvx}(T) \sim p_{\mathrm{prior}} \approx p_T.
\end{aligned}
\end{align}
\end{mdframed}
Here, $\bar{\rvw}(t)$ denotes a standard Wiener process in reverse time, defined as $\bar{\rvw}(t) := \rvw(T - t) - \rvw(T)$.

\rmkb{
Throughout this chapter, $t$ denotes the original diffusion clock:
$t=0$ corresponds to data and $t=T$ to noise.
Accordingly, the reverse-time SDE in \Cref{eq:sde_backward} is integrated
from $t=T$ down to $t=0$, so $\diff t<0$ along the reverse process.
Equivalently, one may introduce a forward-running reverse clock
$s:=T-t$; unless stated otherwise, we keep the original clock $t$.
}

To build intuition for \Cref{eq:sde_backward}, we present a concrete example in \Cref{subsec:example-gaussian} with a Gaussian data distribution and linear–Gaussian dynamics. This setting is analytically tractable: one can derive the time-reversal formula directly using basic calculus and linear algebra, without invoking the full general theory of \citet{anderson1982reverse}.

Note that the presence of stochasticity ($g\neq 0$) introduces an additional correction term, $ -g^2(t) \nabla_{\rvx} \log p_t(\bar{\rvx}(t))$, which accounts for the effect of diffusion and ensures that the reversed dynamics correctly reproduce the evolution of marginal distributions induced by the forward SDE (see \Cref{subsec:fpe-sde-ode}).

\paragraph{Conceptually, Why Does the Reverse Process Work?} \Cref{subsec:reverse-sde-reason} presents an intuitive derivation of the reverse-time SDE by connecting it to the DDPM variational framework (optional but insightful). Here, we provide complementary intuition for how the reverse-time dynamics recover structured data from noise.

At first glance, the presence of Brownian noise in the reverse time process may seem paradoxical. If the forward diffusion spreads data into increasingly noisy configurations, it is unclear how reversing this process, particularly one that introduces additional randomness through $\bar{\rvw}(t)$, can produce clean, structured samples concentrated near the data manifold. The key point is that the reverse time SDE does not inject arbitrary randomness. The diffusion term $g(t)\diff\bar{\rvw}(t)$ is always coupled with the score–driven drift $- g^2(t)\nabla_{\rvx}\log p_t(\bar{\rvx}(t))$. Together, these terms balance one another: the score guides trajectories toward regions of higher density, while the noise introduces controlled stochasticity that allows exploration without overwhelming the dynamics.

To see this more clearly, return to the Langevin intuition in \Cref{eq:continuous-langevin}. When $f(t)\equiv 0$, \Cref{eq:sde_backward} reads
\[
\diff\bar{\rvx}(t)
= - g^2(t) \nabla_{\rvx}\log p_t\!\big(\bar{\rvx}(t)\big) \diff t
+ g(t)  \diff\bar{\rvw}(t).
\]
Reparameterize time forward via $s:=T-t$ (so $\diff t=-\diff s$), and rename the Brownian motion in law so that $\diff\bar{\rvw}(t)=-\diff\rvw_s$. Writing $\bar{\rvx}_s:=\bar{\rvx}(T-s)$ and $\pi_s:=p_{T-s}$ then gives
\begin{align*}
    \diff\bar{\rvx}_s
&= g^2(T-s) \nabla\log \pi_s \big(\bar{\rvx}_s\big) \diff s
+ g(T-s) \diff\rvw_s
\\&=
2\tau(s) \nabla\log \pi_s\!\big(\bar{\rvx}_s\big) \diff s
+ \sqrt{2\tau(s)} \diff\rvw_s,
\quad
\tau(s):=\tfrac{1}{2}g^2(T-s).
\end{align*}
This has the Langevin form with a time-varying temperature $\tau(s)$, targeting the evolving density $\pi_s$. By Tweedie's formula (\Cref{eq:tweedie}), the score direction $\nabla\log \pi_s$ points toward the conditional clean signal at each time slice, so the drift continually ``pulls back'' denoised structure.

Crucially, $g(t)$ is \emph{annealing} along the reverse trajectory. Early on ($s\approx 0$, i.e., $t\approx T$), $g(T-s)$ is typically larger, so the injected noise is stronger and the process explores broadly. As $s$ increases, $g(T-s)$ decreases, the stochastic term weakens, and the score term dominates, pulling samples into high-density regions of $\pi_s$; by $s=T$ (i.e., $t=0$), trajectories concentrate near the data manifold.


\paragraph{Overview of Reverse-Time SDE Capabilities.}

It is fascinating how the time-dependent score function
\[
\rvs(\rvx, t) := \nabla_{\rvx} \log p_t(\rvx)
\]
naturally appears in \Cref{eq:sde_backward}. Once the forward coefficients $f(t)$ and $g(t)$ are specified, the score is the only remaining unknown in the reverse dynamics. This highlights its central role: with the score in hand, the reverse process is determined, and sampling amounts to numerically integrating \Cref{eq:sde_backward} with the learned score.

Since the oracle score generally lacks a closed-form expression, we adopt the approach of \Cref{ch:score-based} and train a neural network $\mathbf{s}_{\bm{\phi}}(\rvx, t)$ to approximate it via score matching; see \Cref{subsec:sde-train} for details. Substituting $\rvs(\rvx, t)$ with $\mathbf{s}_{\bm{\phi}}(\rvx, t)$ in \Cref{eq:sde_backward} then specifies the reverse dynamics completely.

Generation corresponds to solving the reverse-time SDE reversely from $t = T$, starting with $\rvx_T \sim p_{\mathrm{prior}}$, to $t = 0$. Importantly, \citet{anderson1982reverse} proves that the marginal densities of the forward and reverse processes coincide, ensuring that samples at $t=0$ approximately follow $p_{\mathrm{data}}$ when $p_{\mathrm{prior}} \approx p_T$. We will explore this further in \Cref{subsec:sde-sample}.

\subsection{Deterministic Process (Probability Flow ODE)  for Generation}\label{subsec:pf-ode}
 Although the SDE in \Cref{eq:sde_backward} introduces stochasticity and potentially increases the diversity of generated samples, a question arises:
\begin{question}
    Is it necessary to sample using the SDE in \Cref{eq:sde_backward}?
\end{question}

Inspired by \citet{maoutsa2020interacting}, \citet{song2020score} also introduced a deterministic process, an ODE, that evolves samples with the same marginal distributions as the forward SDE. This process $\{\tilde\rvx(t)\}_{t\in[0,T]}$\footnote{We use a tilde to distinguish processes associated with the forward and reverse-time SDEs. Going forward, we omit this notational distinction for simplicity.}, called the \emph{Probability Flow ODE (PF-ODE)}, is given by:
\begin{mdframed}
    \begin{equation}\label{eq:prob_ode_gt}
    \frac{\diff}{\diff t}\tilde\rvx(t) = \mathbf{f}(\tilde\rvx(t), t) - \frac{1}{2} g^2(t) \nabla_{\rvx} \log p_t(\tilde\rvx(t)).
    \end{equation}
\end{mdframed}

Analogous to the SDE case, one can replace the true score with a learned approximation and integrate the reverse-time ODE from $t = T$ to $t = 0$ to generate samples. Concretely, the generated sample (solution of PF-ODE at time $t=0$) takes the form
\[
\tilde\rvx(T) + \int_{T}^{0} \Big[ \mathbf{f}(\tilde\rvx(\tau), \tau) - \frac{1}{2} g^2(\tau)  \nabla_{\rvx} \log p_\tau(\tilde\rvx(\tau)) \Big]   \diff \tau,
\]
where the initial condition $\tilde\rvx(T) \sim p_{\mathrm{prior}}$. Since this integral is intractable in closed form, practical generation relies on numerical solvers (e.g., Euler method, see \Cref{eq:euler-maruyama}).

Compared to the reverse-time SDE, the PF-ODE offers two key advantages:
\begin{itemize}
    \item The ODE can be integrated in either direction, from $t=0$ to $t=T$ or from $t=T$ to $t=0$, using the same formulation of the equation, provided the corresponding initial condition is specified at the chosen endpoint. This bidirectionality contrasts with SDEs, which generally admit only forward time integration.
    \item It benefits from a wide range of well-established, off-the-shelf numerical solvers developed for ODEs.
\end{itemize}

We emphasize that the PF-ODE is not obtained by simply removing the diffusion term in \Cref{eq:sde_backward}; notably, the factor of $\frac{1}{2}$ in its drift term has a principled origin. At a high level, \Cref{eq:prob_ode_gt} arises by choosing the drift of an ODE such that its evolution preserves the same marginal densities as the forward SDE in \Cref{eq:sde_forward}. The underlying principle (i.e., Fokker-Planck Equation~\citep{oksendal2003stochastic}) ensuring this alignment of marginals will be detailed in the next section.

\clearpage
\newpage

\section{Matching Marginals in Forward/Reverse-Time SDEs and PF-ODE}\label{subsec:fpe-sde-ode}
\begin{figure}[th!]
    \centering
    \includegraphics[width=\linewidth]{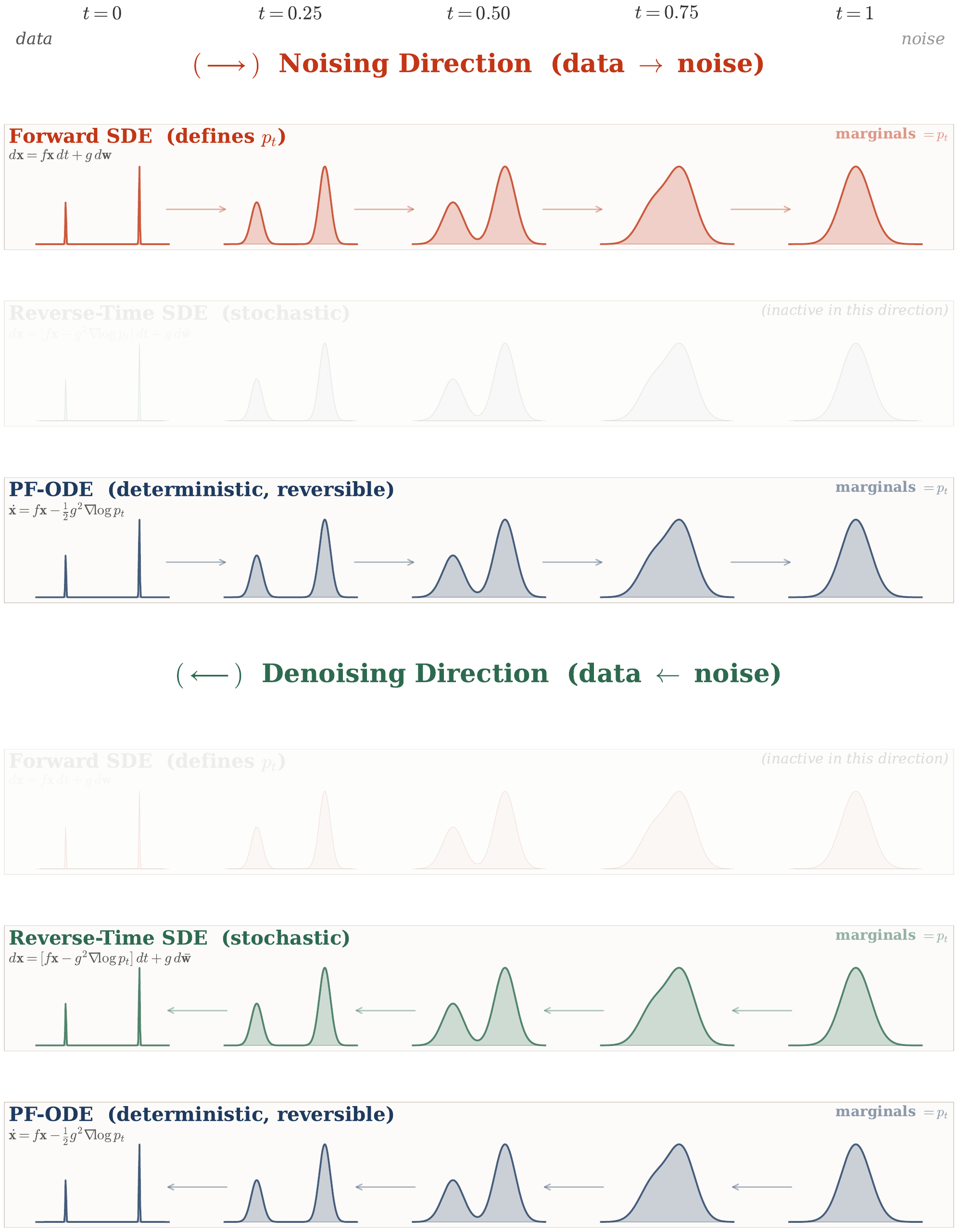}
\caption{\textbfs{Three dynamics sharing the same marginals $p_t$.}
\textbfs{Upper panel: noising direction.}
The forward SDE, which serves as an anchor reference, defines the marginals
$p_t$ by diffusing data toward a Gaussian prior. The PF-ODE follows the same
marginal path deterministically when \emph{run forward in time}. \textbfs{Lower panel: denoising direction.}
The reverse-time SDE and the PF-ODE \emph{run backward in time} yield the same
marginals in reverse, as guaranteed by the Fokker--Planck equation. The reverse-time SDE gives a stochastic
denoising process, whereas the PF-ODE gives a deterministic one. Thus, the
repeated density snapshots are intentional: the processes agree at the level of
one-time marginals, but differ in their sample-level dynamics. The PF-ODE
appears in both panels because it is a deterministic flow that can be integrated
in either time direction.
\figcredit{Created by the authors with AI-assisted coding.}}
    \label{fig:score-sde-three-dynamics}
    \vspace{-2cm}
\end{figure}
In this section, we present the \emph{Fokker-Planck equation} as the central principle governing diffusion models. It describes how probability densities evolve under noisy dynamics and serves as the natural analogue of a change-of-variables formula when randomness is present (see \Cref{app:continuity}). Starting from the forward diffusion process, which defines the marginals $p_t$, we explain how both the reverse-time SDE and the PF-ODE are designed to reproduce the same marginal family. We then make this principle concrete through a Gaussian example, where the reverse-time SDE and the PF-ODE can be computed directly and explicitly.

\subsection{Fokker-Planck Equation to Ensure Alignment of Marginal Densities}
A central concept in diffusion models is that \emph{different processes can lead to the same sequence of marginal distributions} (as we will illustrate later in this subsection).
 The objective is to construct a process that transforms $p_{\mathrm{prior}}$ into $p_{\mathrm{data}}$ by aligning the marginals across time, and in particular at $t=0$. The exact form of the process is secondary, provided it is tractable and admits efficient sampling. This naturally leads to a fundamental question:

\begin{question}
How can we ensure that different processes yield identical marginal distributions?
\end{question}

Returning to our setup, once the forward SDE is specified, it defines the evolution of marginal densities from $p_{\mathrm{data}}$ to $p_{\mathrm{prior}}$. The reverse-time SDE and PF-ODE are then constructed so that their trajectories yield marginal distributions that exactly match those of the forward process. The key to this correspondence lies in the Fokker–Planck equation, which governs how marginal densities evolve under diffusion processes. An illustration is provided in \Cref{fig:score-sde-three-dynamics}. The following theorem~\citep{anderson1982reverse,song2020score} establishes the foundation for this connection:
\thmp{Marginal Alignment via the Fokker--Planck Equation}{fpe}{
Consider the forward SDE
\[
\diff \rvx(t)
=
\rvf(\rvx(t),t)\diff t
+
g(t)\diff \rvw(t),
\qquad
\rvx(0)\sim p_0=p_{\mathrm{data}}.
\]
Under standard regularity conditions, its marginal density $p_t$ evolves
according to the Fokker--Planck equation
\begin{align}\label{eq:fp_general}
\begin{aligned}
\partial_t p_t(\rvx)
&=
-\nabla_{\rvx}\cdot
\bigl[\rvf(\rvx,t)p_t(\rvx)\bigr]
+
\tfrac12 g^2(t)\Delta_{\rvx}p_t(\rvx)
\\
&=
-\nabla_{\rvx}\cdot
\bigl[\rvv(\rvx,t)p_t(\rvx)\bigr],
\end{aligned}
\end{align}
where $\Delta_{\rvx}$ denotes the Laplacian operator and
\[
\rvv(\rvx,t)
:=
\rvf(\rvx,t)
-
\tfrac12 g^2(t)
\nabla_{\rvx}\log p_t(\rvx).
\]

Importantly, the same family of marginal distributions
$\{p_t\}_{t\in[0,T]}$ can be realized by two different dynamics:

\begin{enumerate}
\item[\textup{(i)}]
The \emph{PF-ODE}
\[
\frac{\diff\tilde\rvx(t)}{\diff t}
=
\rvv(\tilde\rvx(t),t).
\]
Starting from $\tilde\rvx(0)\sim p_0$ and solving forward from $0$ to $T$
gives $\tilde\rvx(t)\sim p_t$.
Equivalently, starting from $\tilde\rvx(T)\sim p_T$ and solving the same ODE
backward from $T$ to $0$ follows the same marginals in reverse.

\item[\textup{(ii)}]
The \emph{reverse-time SDE}
\[
\diff\bar\rvx(t)
=
\Bigl[
\rvf(\bar\rvx(t),t)
-
g^2(t)\nabla_{\rvx}\log p_t(\bar\rvx(t))
\Bigr]\diff t
+
g(t)\diff\bar\rvw(t).
\]
Starting from $\bar\rvx(T)\sim p_T$, this SDE is solved backward in time from
$t=T$ to $t=0$. Along this reverse evolution, its marginal distribution at
time $t$ is
\[
\bar\rvx(t)\sim p_t.
\]
\end{enumerate}

Thus, the reverse-time SDE and PF-ODE can have the same marginal distribution
at every noise level even though one is stochastic and the other is
deterministic.
}{
The proof is provided in \Cref{app-sec:scoresde-fpe-proof}, while \Cref{subsec:fpe-reason} offers further intuition behind the Fokker–Planck equation using the marginalization technique of probability. 
}



\paragraph{Multiple Conditional Distributions for a Fixed Marginal.}
To understand how the PF-ODE transports $p_{\mathrm{data}}$ forward in time (or equivalently $p_{\mathrm{prior}}$ in reverse), consider the \emph{flow map} $\bPsi_{s \to t}: \mathbb{R}^D \to \mathbb{R}^D$, where $\bPsi_{s \to t}(\rvx_s)$ denotes the PF-ODE solution at time $t$ initialized from $\rvx_s$ at time $s$, for any time $s, t \in[0,T]$. In other words, this map takes an initial state $\rvx_s$ and directly jumps to its state at $t$:
\begin{align}\label{eq:flow-map-def}
\bPsi_{s \to t}(\rvx_s) \coloneqq \rvx_s + \int_s^t \rvv(\rvx_\tau, \tau)\mathrm{d}\tau,
\end{align}
with velocity field
\begin{align*}
\rvv(\rvx, \tau) := \rvf(\rvx, \tau) - \frac{1}{2} g^2(\tau)\nabla_{\rvx}\log p_\tau(\rvx).
\end{align*}
Here, the integral captures the net displacement accumulated along the
PF-ODE trajectory. The term \emph{flow map} is indeed classical in
differential equations, dynamical systems, and continuum mechanics: it
denotes the map that carries a state from its position at time $s$ to its
position at time $t$ under the given dynamics. The velocity field
$\rvv(\rvx,t)$ describes the instantaneous motion, whereas the flow map
$\bPsi_{s\to t}$ records the accumulated motion over a finite time
interval~\citep{arnold1988geometrical,chorin1993mathematical}\footnote{The physical idea underlying this map is much
older. Eighteenth-century fluid mechanics already studied motion through
velocity fields and by following the evolution of material particles,
notably in the foundational works of Euler and Lagrange
~\citep{euler1757fluid,lagrange1783fluid}.}. We use the term in this standard mathematical and physical sense
throughout the book, particularly when discussing fast diffusion-based
generators in \Cref{ch:distillation,ch:fast-scratch}. Under mild smoothness assumptions on $\rvv$, the flow map
$\bPsi_{s \to t}: \mathbb{R}^D \to \mathbb{R}^D$ is a smooth
bijection\footnote{%
Spoiler: the PF-ODE flow map $\bPsi_{s\to t}$ is exactly the Normalizing
Flow (NF) bijection carrying $p_s$ to $p_t$ (to be detailed in
\Cref{sec:flow-based-method}). The difference is that, once the forward SDE and its marginal densities
are fixed, the PF-ODE formula uniquely specifies its associated velocity
field, $\rvv(\rvx,t)
=
\rvf(\rvx,t)
-\frac{1}{2}g^2(t)\nabla_{\rvx}\log p_t(\rvx)$,
whereas a normalizing flow, or continuous-time normalizing flow,
parameterizes and learns a velocity field directly. Both rely on the same
continuous change-of-variables principle.
}.

For any $t\in[0,T]$, the \emph{pushforward density} is defined as
\[
p_t^{\mathrm{fwd}}(\rvx_t) := \int \delta\bigl(\rvx_t - \bPsi_{0 \to t}(\rvx_0)\bigr)\, p_{\mathrm{data}}(\rvx_0) \,\diff \rvx_0,
\]
denoted $\bm{\Psi}_{0\to t} \sharp p_{\mathrm{data}}$, representing the distribution at time $t$ under $\bm{\Psi}_{0\to t}$. Theorem~\ref{thm:fpe} ensures $p_t^{\mathrm{fwd}} = p_t$, where $p_t$ is the marginal density of the forward SDE, equating the deterministic PF-ODE and stochastic kernel:
\[
p_t(\rvx_t) = \int p_t(\rvx_t |\rvx_0)\, p_{\mathrm{data}}(\rvx_0)\,\diff \rvx_0
= \int \delta\bigl(\rvx_t - \bm{\Psi}_{0\to t}(\rvx_0)\bigr)\, p_{\mathrm{data}}(\rvx_0)\,\diff \rvx_0.
\]

This implies infinitely many conditionals $Q_t(\rvx_t |\rvx_0)$ yield the same $p_t(\rvx_t)$, for instance:
\begin{itemize}
  \item \textbfs{Stochastic (Simulation-Free):} $Q_t(\rvx_t |\rvx_0) = p_t(\rvx_t |\rvx_0)$,
  \item \textbfs{Deterministic (Requires ODE Solving):} $Q_t(\rvx_t |\rvx_0) = \delta\bigl(\rvx_t - \bm{\Psi}_{0\to t}(\rvx_0)\bigr)$,
  \item \textbfs{Mixture:} $Q_t(\rvx_t |\rvx_0) = \lambda p_t(\rvx_t |\rvx_0) + (1-\lambda)\delta\bigl(\rvx_t - \bm{\Psi}_{0\to t}(\rvx_0)\bigr)$, $\lambda\in[0,1]$.
\end{itemize}
This nonuniqueness of $Q_t(\rvx_t |\rvx_0)$ arises from the fact that the marginal constraint does not uniquely determine the conditional distribution. This concept reappears in \Cref{subsec:fm-framework} and \Cref{subsec:ddim}. In particular, there exists an entire family of reverse-time SDEs that are consistent with the same marginal $p_t$.

\msg{Observation}{Matching Prescribed Marginal Densities}{
Multiple processes can give rise to the same sequence of marginal densities; what truly matters is satisfying the Fokker–Planck equation. This fundamental insight affords us remarkable flexibility in designing generative processes that transition from $ p_{\mathrm{prior}} $ to $ p_{\mathrm{data}} $, or vice versa.
}
The Fokker–Planck equation lies at the heart of diffusion models and is rooted in the fundamental \emph{change-of-variable formula} for probability densities (see \Cref{app:continuity} for a systematic treatment). Far from being a minor technical detail, this principle recurs throughout our development, most notably in \Cref{sec:flow-matching-framework}.

\subsection{A Computable Example: Evolutions of Gaussian Dynamics}\label{subsec:example-gaussian}
When $p_{\mathrm{data}}$ is a normal distribution (or a mixture of Gaussians), the score function admits a closed-form expression. 
This makes it an ideal setting for building intuition about diffusion processes: we can explicitly derive the reverse-time SDE and the PF-ODE using only basic calculus, without resorting to advanced mathematical tools. 
In this subsection, we illustrate how these equations behave in such a tractable case.

\paragraph{Exact Computation of the Reverse-Time SDE with a Gaussian.}
 When $p_{\mathrm{data}}$ is Gaussian, the formula in \Cref{eq:sde_backward} can be derived directly, without relying on the general theory and proofs of \citet{anderson1982reverse}. To illustrate the core idea, we consider the one-dimensional case; the extension to higher dimensions follows in the same way.

Start from the forward SDE
\[
\diff x(t)  =  f(t)  x(t)  \diff t  +  g(t)  \diff w_t,
\]
and take one small Euler step of size $\Delta t>0$:
\[
x_{t+\Delta t}  =  a  x_t  +  r  \epsilon,
\]
where $a:=1+f(t)\Delta t$, $r:=g(t)\sqrt{\Delta t}$, and $ \epsilon\sim\mathcal N(0,1)$. Equivalently, the forward one-step transition kernel is Gaussian:
\[
x_{t+\Delta t}   |   x_t  \sim  \mathcal N\!\big(\cdot;a  x_t,   r^2\big).
\]
Since $p_{\mathrm{data}}$ is assumed to be Gaussian, the current marginal at time $t$ is also Gaussian, which takes the following form:
\[
x_t \sim \mathcal N(\cdot;m_t,   s_t^2),
\]
for some scalar $m_t$ and $s_t$. So conditioning will amount to multiplying two Gaussians and renormalizing. This keeps the algebra elementary.

By Bayes’ rule the conditional density is, up to a constant, the product of the prior and the transition kernel:
\begin{align*}
    p(x_t   |   x_{t+\Delta t})  &\propto p(x_{t+\Delta t} | x_t   ) p_t(\rvx_t) \\&\propto
\exp\Big(-\frac{(x_t-m_t)^2}{2s_t^2}\Big)  
\exp\Big(-\frac{(x_{t+\Delta t}-a  x_t)^2}{2r^2}\Big).
\end{align*}
The exponent is a quadratic in $x_t$. Expanding both squares and grouping terms shows exactly which coefficients matter:
\[
-2\log p(x_t   |   x_{t+\Delta t})
 = 
A  x_t^2  -  2 B  x_t  +  \text{const},
\]
with
\[
A:=\frac{1}{s_t^2}+\frac{a^2}{r^2},\quad
B:=\frac{m_t}{s_t^2}+\frac{a  x_{t+\Delta t}}{r^2}.
\]
Here $A$ is the sum of precisions (prior precision plus the transition-kernel precision transported through $a$), while $B$ is the corresponding precision-weighted sum of targets. With these in hand, completing the square gives the posterior in one line:
\[
A x_t^2 - 2 B x_t
  =  
A\Big(x_t-\frac{B}{A}\Big)^2  -  \frac{B^2}{A},
\]
so the conditional distribution is Gaussian with variance $1/A$ and mean $B/A$:
\[
\Var(x_t   |   x_{t+\Delta t})   =   \frac{1}{\frac{1}{s_t^2}+\frac{a^2}{r^2}},
\qquad
\mathbb E[x_t   |   x_{t+\Delta t}]   =  
\frac{\frac{m_t}{s_t^2}+\frac{a  x_{t+\Delta t}}{r^2}}{\frac{1}{s_t^2}+\frac{a^2}{r^2}}.
\]
These closed forms already describe the reverse transition for any small $\Delta t$. To read off a reverse-time SDE, we now expand them for small $\Delta t$.

Use $a=1+f(t)\Delta t$ and $r^2=g^2(t)\Delta t$. As $\Delta t\to 0$, the contribution $\frac{a^2}{r^2}\sim \frac{1}{g^2(t)\Delta t}$ dominates the precision, so the variance becomes
\[
\Var(x_t   |   x_{t+\Delta t})
= \left(\frac{1}{s_t^2}+\frac{a^2}{r^2}\right)^{-1}
= g^2(t)  \Delta t  +  \mathcal{O}(\Delta t^2),
\]
which tells us the reverse step has the same diffusion scale $g(t)$ as the forward step. For the mean, expand the ratio $B/A$ to first order:
\[
\mathbb E[x_t   |   x_{t+\Delta t}]
  =  x_{t+\Delta t}
 +  \Delta t\left[
-\left(f(t)+\frac{g^2(t)}{s_t^2}\right)  x_{t+\Delta t}
 +  \frac{g^2(t)}{s_t^2}  m_t
\right]  +  \mathcal{O}(\Delta t^2).
\]

Putting the mean and variance together yields the one-step reverse transition kernel
\[
x_t    |    x_{t+\Delta t}
 \sim 
\mathcal N\!\left(
x_{t+\Delta t} + \Delta t\!\left[-\left(f+\frac{g^2}{s_t^2}\right)  x_{t+\Delta t} + \frac{g^2}{s_t^2}  m_t\right],
 g^2  \Delta t\right)
 +  \mathcal{O}(\Delta t^2).
\]
This is recognized as the Euler--Maruyama update, run backward from
$t+\Delta t$ to $t$:
\[
x_t - x_{t+\Delta t}
= \Delta t\!\left[-\left(f+\frac{g^2}{s_t^2}\right) x_{t+\Delta t}
+ \frac{g^2}{s_t^2} m_t\right]
+ g\sqrt{\Delta t}\,\epsilon.
\]
This backward drift can be written with the score because for a Gaussian
marginal $p_t=\mathcal N(m_t,s_t^2)$,
\[
\partial_x \log p_t(x) = -\frac{x-m_t}{s_t^2}
 \Longrightarrow
-\left(f+\frac{g^2}{s_t^2}\right)x
+\frac{g^2}{s_t^2}m_t
=
-\left[f x-g^2\partial_x\log p_t(x)\right].
\]
Therefore, letting $\Delta t\to0$ gives the reverse-time SDE on the original
diffusion clock,
\[
\diff \bar x(t)
=
\left[
f(t)\bar x(t)
-
g^2(t)\partial_x\log p_t(\bar x(t))
\right]\diff t
+
g(t)\diff\bar w(t),
\qquad t:T\to0.
\]
This exactly matches the general reverse-time SDE in
\Cref{eq:sde_backward}. In vector form, the scalar derivative
$\partial_x\log p_t$ is replaced by
$\nabla_{\rvx}\log p_t$.

\paragraph{Exact Computation of PF–ODE with a Gaussian.}
When the data distribution is assumed to be Gaussian, we can also directly derive the PF-ODE formula, avoiding heavy machinery such as the Fokker–Planck equation. In the end, we will see that the marginal densities of the PF-ODE coincide with those of both the forward SDE and the reverse-time SDE, providing a constructive verification of the Fokker–Planck theory to be discussed in \Cref{subsec:fpe-sde-ode}.

Assume $x_t\sim\mathcal N(\cdot;m_t,s_t^2)$ at time $t$. A small deterministic step of size $\Delta t$ can be written as a smooth map
\[
x_{t+\Delta t}=\Phi_{t,\Delta t}(x_t)
= x_t + \Delta t v_t(x_t) + \mathcal{O}(\Delta t^2),
\]
which is simply the first–order Taylor expansion in $\Delta t$.
Our goal is to construct a velocity field whose deterministic flow reproduces
the same Gaussian marginal evolution. A particularly simple choice is an affine velocity field
\[
v_t(x)=a_t x+b_t.
\]
Since affine dynamics preserve Gaussianity, we can determine $a_t$ and $b_t$
by matching the evolution of the mean and variance to those of the forward
SDE. Plugging this back into the step gives
\[
x_{t+\Delta t}=(1+\alpha_t \Delta t) x_t+\beta_t \Delta t+\mathcal{O}(\Delta t^2),
\qquad
\alpha_t:=a_t,\ \ \beta_t:=b_t.
\]


We now push $x_t\sim\mathcal N(\cdot;m_t,s_t^2)$ through this map and track mean and variance to first order:
\begin{align*}
\mathbb E[x_{t+\Delta t}]
&= m_t+\Delta t(\alpha_t m_t+\beta_t)+\mathcal{O}(\Delta t^2),
\\
\Var(x_{t+\Delta t})
&= s_t^2+\Delta t (2\alpha_t s_t^2)+\mathcal{O}(\Delta t^2).
\end{align*}
On the other hand, the forward SDE $\diff x=f(t) x \diff t+g(t) \diff w_t$ has the elementary moment formulas (see \Cref{eq:mean-var-evolution}):
\[
m_t' = f(t) m_t,
\qquad
(s_t^2)' = 2 f(t) s_t^2 + g^2(t).
\]
Matching the coefficients of $\Delta t$ gives
\[
\alpha_t = f(t)+\frac{g^2(t)}{2 s_t^2},
\qquad
\beta_t = - \frac{g^2(t)}{2 s_t^2} m_t.
\]

With these choices, the step becomes
\[
x_{t+\Delta t}
= x_t + \Delta t\!\left[\Big(f(t)+\frac{g^2(t)}{2s_t^2}\Big)x_t
-\frac{g^2(t)}{2s_t^2} m_t\right]+\mathcal{O}(\Delta t^2).
\]
Since for a Gaussian $p_t=\mathcal N(m_t,s_t^2)$ we have $\partial_x\log p_t(x)=-(x-m_t)/s_t^2$, we can rewrite the bracket as
$f(t) x_t - \tfrac12 g^2(t) \partial_x\log p_t(x_t)$.
Therefore,
\[
x_{t+\Delta t}
= x_t + \Delta t\Big[f(t) x_t - \tfrac12 g^2(t) \partial_x\log p_t(x_t)\Big] + \mathcal{O}(\Delta t^2).
\]
Finally, dividing by $\Delta t$ and letting $\Delta t\to 0$ yields the PF-ODE
\[
x'(t)  =  f(t) x(t)  -  \tfrac12 g^2(t) \partial_x\log p_t\!\big(x(t)\big).
\]

To see why this ODE has the same marginals as the forward SDE (and the reverse–time SDE), observe that the drift above is linear plus a shift. Thus $x(t)$ depends affinely on $x(0)$, and affine maps send Gaussians to Gaussians. Moreover, the mean $m_t$ and variance $s_t^2$ along this ODE satisfy exactly the same two scalar ODEs as the forward SDE (by our matching), with the same initial values. Hence $p_t=\mathcal N(m_t,s_t^2)$ is identical for both evolutions at every time $t$.

\clearpage
\newpage

\section{Score SDE: Its Training and Sampling}\label{sec:scoresde-training-sampling}

\subsection{Training}\label{subsec:sde-train}
Building on the philosophy as in \Cref{ch:score-based}, we approximate the oracle score $\nabla_{\rvx} \log p_t(\rvx)$ using a time-conditioned neural network 
\[
\mathbf{s}_{\bm{\phi}} = \mathbf{s}_{\bm{\phi}}(\rvx, t)
\]
across all $t\in[0,T]$, by minimizing a score-matching objective as in \Cref{eq:sm}:
\begin{equation*}
\mathcal{L}_{\text{SM}}(\bm{\phi}; \omega(\cdot)) := \frac{1}{2} \mathbb{E}_{t\sim p_{\text{time}}}  \mathbb{E}_{\rvx_t \sim p_t}\left[\omega(t) \left\|\mathbf{s}_{\bm{\phi}}(\rvx_t, t) - \nabla_{\rvx} \log p_t(\rvx_t)\right\|_2^2\right],
\end{equation*}
where $p_{\text{time}}$ is some time distribution (e.g., uniform on $[0,T]$), $\omega(\cdot)$ is a time-weighting function.

\begin{figure}[th!]
    \centering
    \includegraphics[width=\linewidth]{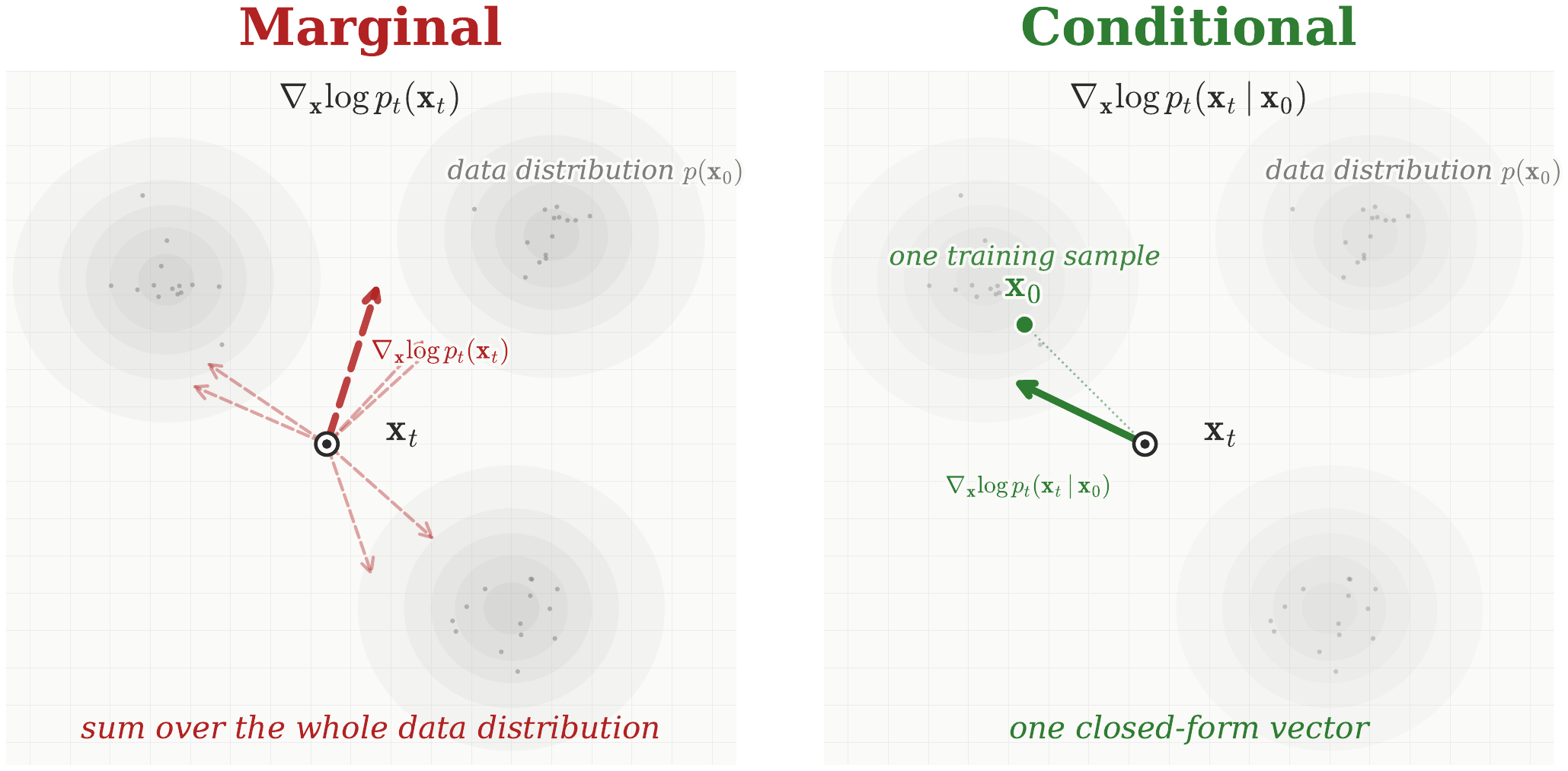}
    \caption{\textbfs{The denoising score matching trick.}
    \textbfs{(Left)}
    For a fixed noisy sample $\rvx_t$, the marginal score
    $\nabla_{\rvx_t}\log p_t(\rvx_t)$ points toward regions where noisy
    samples are more likely. Computing it directly requires combining the
    contributions from all possible clean samples $\rvx_0$ that could have
    produced $\rvx_t$, which is generally intractable.
    \textbfs{(Right)}
    During training, however, the clean sample $\rvx_0$ used to generate
    $\rvx_t$ is known. For Gaussian corruption, the corresponding
    conditional score
    $\nabla_{\rvx_t}\log p_t(\rvx_t|\rvx_0)$ is available in closed form,
    providing one simple vector as the regression target. Across training
    examples, these conditional score vectors average to the marginal
    score. Thus, denoising score matching learns the desired marginal score without
directly regressing on the generally intractable score
$\nabla_{\rvx_t}\log p_t(\rvx_t)$, whose evaluation requires averaging
over all possible samples from the data distribution.
    \figcredit{Created by the authors with AI-assisted coding.}}
    \label{fig:score-and-denoising-score}
\end{figure}

To avoid relying on the intractable oracle score $\nabla_{\rvx} \log p_t(\rvx)$, the DSM loss in \Cref{eq:dsm-loss} is employed. Conditioned on a data point $\rvx_0$, this approach allows the use of the analytically tractable score $\nabla_{\rvx_t} \log p_t(\rvx_t|\rvx_0)$ via \Cref{eq:score-gaussian-formula}, with concrete examples given in \Cref{sec:instances}. Specifically, we exploit the following loss function:
\begin{mdframed}
    \begin{equation}\label{eq:conti-dsm_loss-real}
\begin{aligned}
     &\mathcal{L}_{\text{DSM}}(\bm{\phi}; \omega(\cdot)) := \\
     &\frac{1}{2} \mathbb{E}_{t}  \mathbb{E}_{\rvx_0} \mathbb{E}_{ p_t(\rvx_t|\rvx_0)} \Big[\omega(t)\norm{\mathbf{s}_{\bm{\phi}}(\rvx_t, t) - \nabla_{\rvx_t} \log p_t(\rvx_t|\rvx_0)}_2^2 \Big],
\end{aligned}
\end{equation}
\end{mdframed}
where $\rvx_0 \sim p_{\mathrm{data}}$. \Cref{eq:conti-dsm_loss-real} can be interpreted as the continuous-time counterpart of \Cref{eq:nscn}, with the summation in the discrete case replaced by integration.

Similar to the result in Theorem~\ref{thm:sm-dsm}, the minimizer of \Cref{eq:conti-dsm_loss-real} is uniquely determined as follows:

\proppp{Minimizer of DSM}{dsm-minimizer}{
The minimizer $\mathbf{s}^*$ satisfies
\begin{align}\label{eq:optimal-score}
\begin{aligned}
    \mathbf{s}^*(\rvx_t, t) 
    = \mathbb{E}_{\rvx_0 \sim p(\rvx_0 | \rvx_t)} \big[ \nabla_{\rvx_t} \log p_t(\rvx_t | \rvx_0) \big]= \nabla_{\rvx_t} \log p_t(\rvx_t),
\end{aligned}
\end{align}
for almost every $\rvx_t \sim p_t$ and $t \in [0,T]$. }{ 
DSM objective can be understood as a least-squares error problem. Specifically, at each time $t$, the optimal score function is given by the conditional expectation of the gradient of the log conditional density, which, under Bayes' rule, is equivalent to the gradient of the log marginal density. For a detailed proof, see Appendix~\ref{app-sec:min-dsm}.
}

\subsection{Sampling and Inference}\label{subsec:sde-sample}
\begin{figure}[th!]
    \centering
    \vspace{-0.8cm}
    \includegraphics[width=\linewidth]{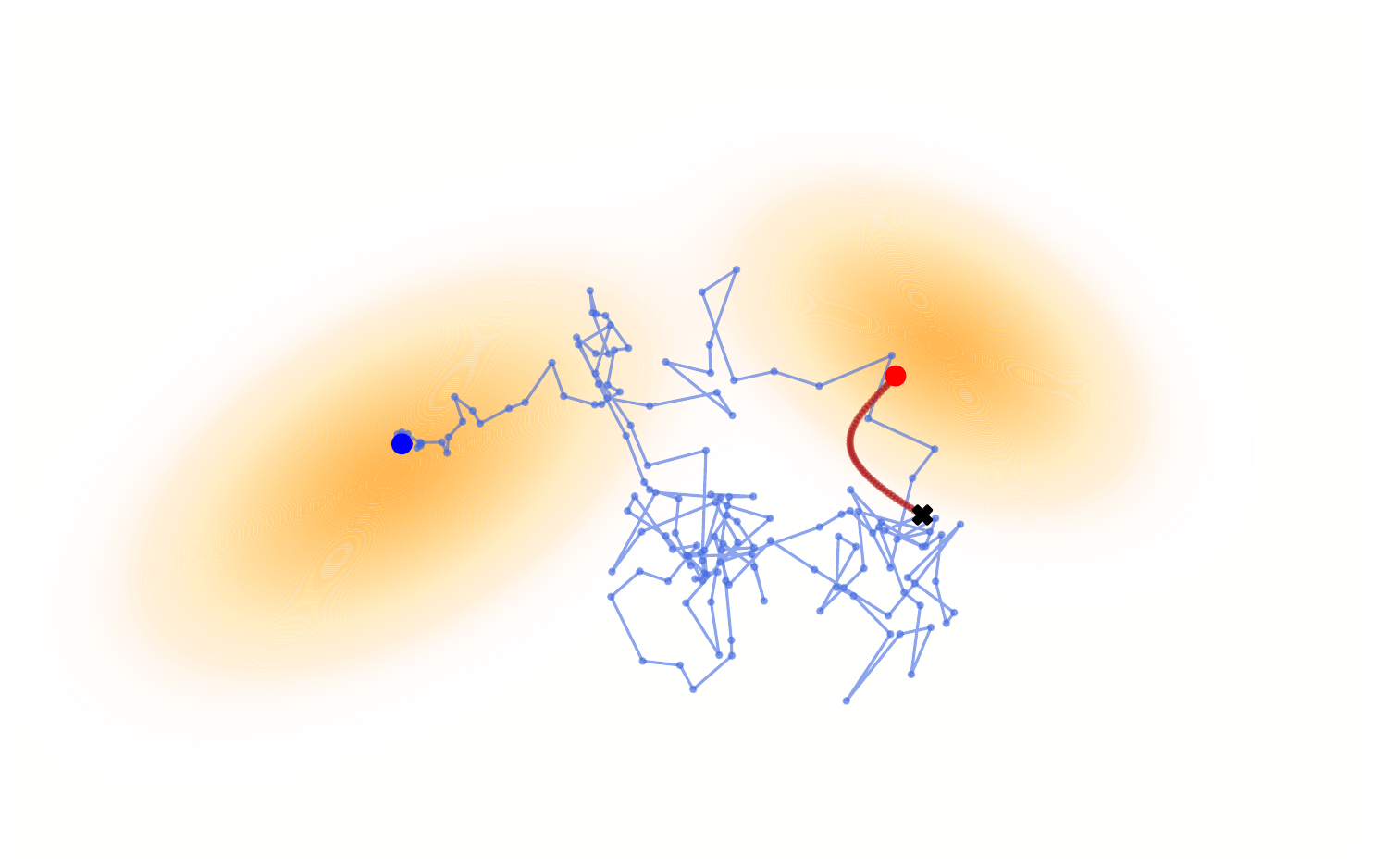}
    \vspace{-0.6cm}
    \caption{\textbfs{(2D) Illustration of sampling from the Score SDE.} The forward SDE is designed as $\diff\rvx(t)=\sqrt{2t}\diff\rvw(t)$, with drift $\mathbf{f}\equiv\mathbf{0}$ and diffusion coefficient $g(t)=\sqrt{2t}$ on $[0,T]$, evolves a two-mode Gaussian mixture $p_0=p_{\mathrm{data}}$ into $p_T\approx p_{\mathrm{prior}}:=\mathcal{N}(\mathbf{0},T^2\mathbf{I})$. Shown are sampling trajectories of the reverse-time SDE (blue; solved by \Cref{eq:euler-maruyama}) and the PF-ODE (red; solved by \Cref{eq:euler-ode}). Starting from a random point $\rvx_T\sim p_{\mathrm{prior}}$ (dark ``$\times$''), both trajectories move toward the support of $p_{\mathrm{data}}$ as $t\to 0$.
    \figcredit{Created by the authors.}}
    \label{fig:sde_ode_trajectories}
\end{figure}

After learning 
\[
\mathbf{s}_{\bm{\phi}^\times} := \mathbf{s}_{\bm{\phi}^\times}(\rvx, t) \approx \nabla_{\rvx} \log p_t(\rvx),
\]
we replace the intractable oracle score $\nabla_{\rvx} \log p_t(\rvx)$ in the reverse-time SDE (\Cref{eq:sde_backward}) and PF-ODE (\Cref{eq:prob_ode_gt}) with the learned proxy $\mathbf{s}_{\bm{\phi}^\times}(\rvx, t)$. This substitution enables tractable inference via either the SDE or the ODE. For clarity, we distinguish the resulting processes as $\rvx^{\mathrm{SDE}}_{\bm{\phi}^\times}(t)$ and $\rvx^{\mathrm{ODE}}_{\bm{\phi}^\times}(t)$, respectively, but will omit this distinction in later sections.\footnote{This is to simplify notation after this subsection.}

\paragraph{Empirical Reverse-Time SDE.}  
By substituting the trained score model $\mathbf{s}_{\bm{\phi}^\times}$ for the true score in \Cref{eq:sde_backward}, we obtain the parameterized reverse-time SDE used for generation:
\begin{equation}\label{eq:sde_backward_sub}
    \diff \rvx^{\mathrm{SDE}}_{\bm{\phi}^\times}(t) = \left[\mathbf{f}\left(\rvx^{\mathrm{SDE}}_{\bm{\phi}^\times}(t), t\right) - g^2(t) \mathbf{s}_{\bm{\phi}^\times}\left(\rvx^{\mathrm{SDE}}_{\bm{\phi}^\times}(t), t\right) \right] \diff t + g(t) \diff \bar{\rvw}(t).
\end{equation}
To generate a sample, we first draw an initial value $\rvx_T$ from the prior distribution $p_{\mathrm{prior}}$ and then numerically solve \Cref{eq:sde_backward_sub} backward in time from $t=T$ to $t=0$. A standard numerical solver for this is the Euler–Maruyama method, which provides the discrete update rule:
\begin{equation}\label{eq:euler-maruyama}
\rvx_{t - \Delta t} \leftarrow \rvx_t - \left[ \mathbf{f}(\rvx_t, t) - g^2(t)  \mathbf{s}_{\bm{\phi}^\times}(\rvx_t, t) \right] \Delta t + g(t) \sqrt{\Delta t} \cdot \boldsymbol{\epsilon},
\end{equation}
where $\boldsymbol{\epsilon} \sim \mathcal{N}(\bm{0}, \mathbf{I})$ and $\Delta t > 0$ is the step size.

Iterating this update rule yields a final sample $\rvx^{\mathrm{SDE}}_{\bm{\phi}^\times}(0)$. If the score model is accurate, the distribution of these generated samples, denoted $p_{\bm{\phi}^\times}^{\mathrm{SDE}}(\cdot; 0)$, provides a close approximation to the true data distribution\footnote{Theoretically, estimation accuracy depends on the discrepancy between $p_T$ and $p_{\mathrm{prior}}$ (typically negligible), model training error, and numerical discretization error~\citep{de2022convergence,wang2024evaluating}. We do not pursue formal bounds here.}:
\[
p_{\bm{\phi}^\times}^{\mathrm{SDE}}(\cdot; 0) \approx p_{\mathrm{data}}(\cdot).
\]
Indeed, for the VP construction, the DDPM sampling scheme in
\Cref{eq:ddpm-sampling} has the same continuous-time reverse-SDE limit as the
Euler--Maruyama discretization above; see \Cref{sec:instances}.

\paragraph{Empirical PF-ODE.}  
The PF-ODE defines a continuous flow connecting $p_{\mathrm{prior}}$ and $p_{\mathrm{data}}$, enabling sampling, encoding, and exact likelihood evaluation. The following section provides further details on each of these operations.
\subparagraph{I. Sampling with PF-ODE.} Replacing the oracle score in \Cref{eq:prob_ode_gt} with $\mathbf{s}_{\bm{\phi}^\times}$ yields the empirical PF-ODE:
\begin{equation}\label{eq:prob_ode_sub}
    \frac{\diff}{\diff t} \rvx^{\mathrm{ODE}}_{\bm{\phi}^\times}(t) = \mathbf{f}\left(\rvx^{\mathrm{ODE}}_{\bm{\phi}^\times}(t), t\right) - \frac{1}{2} g^2(t) \mathbf{s}_{\bm{\phi}^\times}\left(\rvx^{\mathrm{ODE}}_{\bm{\phi}^\times}(t), t\right).
\end{equation}
To generate samples, we begin by drawing an initial sample $\rvx_T$ from the prior distribution, $p_{\mathrm{prior}}$. We then numerically solve the PF-ODE from \Cref{eq:prob_ode_sub} backward in time from $t=T$ to $t=0$. This process is equivalent to approximating the integral:
\[
\rvx^{\mathrm{ODE}}_{\bm{\phi}^\times}(0) = \rvx_T + \int_T^0 \left[ \mathbf{f}\left(\rvx^{\mathrm{ODE}}_{\bm{\phi}^\times}(\tau), \tau\right) - \frac{1}{2} g^2(\tau) \mathbf{s}_{\bm{\phi}^\times}\left(\rvx^{\mathrm{ODE}}_{\bm{\phi}^\times}(\tau), \tau\right) \right] \mathrm{d}\tau.
\]
Solving this integral yields a final sample, $\rvx^{\mathrm{ODE}}_{\bm{\phi}^\times}(0)$. The distribution of samples generated via this deterministic process, denoted $p_{\bm{\phi}^\times}^{\mathrm{ODE}}(\cdot; 0)$, provides an approximation to the data distribution, such that $p_{\bm{\phi}^\times}^{\mathrm{ODE}}(\cdot; 0) \approx p_{\mathrm{data}}$. 

Let $\Delta t > 0$ denote a discretization step size. A standard numerical integration approach is the Euler method, which estimates
\[
\mathbf{f}(\rvx_\tau, \tau) - \frac{1}{2} g^2(\tau)  
\mathbf{s}_{\bm{\phi}^\times}(\rvx_\tau, \tau) \approx \mathbf{f}(\rvx_t, t) - \frac{1}{2} g^2(t)  
\mathbf{s}_{\bm{\phi}^\times}(\rvx_t, t), \quad \tau \in [t-\Delta t, t],
\]
leading to the following update rule:
\begin{equation}\label{eq:euler-ode}
\rvx_{t - \Delta t} \leftarrow \rvx_t -
\left[ \mathbf{f}(\rvx_t, t) - \frac{1}{2} g^2(t)  
\mathbf{s}_{\bm{\phi}^\times}(\rvx_t, t) \right] \Delta t.
\end{equation}

This connection allows us to reframe the process of generation with the following core insight:
\msg{Insight}{Generation $\Leftrightarrow$ ODE/SDE Solving}{Sampling from diffusion models is fundamentally equivalent to solving a corresponding probability flow ODE or a reverse-time SDE.}

This equivalence provides a clear explanation for the slow sampling speeds of diffusion models, as raised in Question~\ref{ques:dm-sampling-slow}. Generation is computationally intensive because numerical solvers for these differential equations are inherently iterative, often requiring many steps to accurately approximate a solution trajectory\footnote{For example, DDPM and Score SDE typically use $1,000$ function evaluations for generation.}. However, the PF-ODE formulation is also advantageous, as it allows us to leverage the extensive literature on accelerated numerical solvers. Exploring these techniques to speed up diffusion model sampling is the primary focus of \Cref{ch:solvers}.

\subparagraph{II. Inversion with PF-ODE.} As discussed, unlike in the case of SDEs, we can solve the same \Cref{eq:prob_ode_sub} both forward (from $0$ to $T$) and reverse (from $T$ to $0$) in time. When solving it forward, the ODE flow maps data to its (noisy) latent representations across all $t \in [0,T]$, which plays a role of an encoder. This concept enables powerful applications for controllable generation, such as image translation and editing and beyond~\citep{mokady2023null,su2022dual}.

\subparagraph{III. Log-Likelihood Evaluation via PF-ODE.}
The PF-ODE also allows us to evaluate the likelihood of a data sample.
Recall from \Cref{eq:instant-change-of-var} that, for an ODE, the change in
log-density along a trajectory is determined by the divergence of its velocity
field.

For the learned PF-ODE, define
\[
\rvv_{\bm{\phi}^\times}(\rvx,t)
:=
\rvf(\rvx,t)
-\frac{1}{2}g^2(t)\rvs_{\bm{\phi}^\times}(\rvx,t).
\]
If $\rvx(t)$ evolves according to
\[
\frac{\diff}{\diff t}\rvx(t)
=
\rvv_{\bm{\phi}^\times}(\rvx(t),t),
\]
then its log-density satisfies
\[
\frac{\diff}{\diff t}
\log p^{\textup{ODE}}_{\bm{\phi}^\times}(\rvx(t);t)
=
-\nabla_{\rvx}\cdot
\rvv_{\bm{\phi}^\times}(\rvx(t),t).
\]

To evaluate the likelihood of an observed sample $\rvx_0$, we run the PF-ODE
forward from $t=0$ to $t=T$, mapping $\rvx_0$ to a terminal point where the
model density is the tractable prior. We can simultaneously track the
log-density change by defining
\[
\delta(t)
:=
\log p^{\textup{ODE}}_{\bm{\phi}^\times}(\rvx(t);t)
-
\log p^{\textup{ODE}}_{\bm{\phi}^\times}(\rvx_0;0).
\]
Thus, $\delta(0)=0$, and the state and log-density change can be computed
together through
\begin{align}\label{eq:score-sde-likelihood}
\frac{\diff}{\diff t}
\begin{bmatrix}
\rvx(t)\\
\delta(t)
\end{bmatrix}
=
\begin{bmatrix}
\rvv_{\bm{\phi}^\times}(\rvx(t),t)\\
-\nabla_{\rvx}\cdot
\rvv_{\bm{\phi}^\times}(\rvx(t),t)
\end{bmatrix},
\qquad
\begin{bmatrix}
\rvx(0)\\
\delta(0)
\end{bmatrix}
=
\begin{bmatrix}
\rvx_0\\
0
\end{bmatrix}.
\end{align}

After integrating to $t=T$, the terminal density is
$p_{\mathrm{prior}}(\rvx(T))$. Therefore,
\[
\log
p^{\textup{ODE}}_{\bm{\phi}^\times}(\rvx_0;0)
=
\log p_{\mathrm{prior}}(\rvx(T))
-
\delta(T).
\]
Equivalently,
\[
\log
p^{\textup{ODE}}_{\bm{\phi}^\times}(\rvx_0;0)
=
\log p_{\mathrm{prior}}(\rvx(T))
+
\int_0^T
\nabla_{\rvx}\cdot
\rvv_{\bm{\phi}^\times}(\rvx(t),t)
\,\diff t.
\]
Thus, the same PF-ODE used for generation and inversion also provides a
continuous change-of-variables formula for likelihood evaluation.

\newpage

\section{Instantiations of SDEs}\label{sec:instances}

\citet{song2020score} categorize the drift term $\rvf(\rvx, t)$ and the diffusion term $g(t)$ in the forward SDE into three types based on the behavior of the variance during evolution. Here, we focus on two commonly used types: the \emph{\underline{V}ariance \underline{E}xplosion} (VE) SDE and \emph{\underline{V}ariance \underline{P}reserving} (VP) SDE. While it is possible to design custom noise schedulers, their design can substantially influence empirical performance. 
Table~\ref{tb:sde_instances} summarizes these two SDE instantiations.
\begin{table}[th]
  \caption{Summary of the forward SDEs}
  \small
  \centering
  \resizebox{\textwidth}{!}{
  \begin{tabular}{cccc}
     \toprule
    
         &   \textbfs{VE SDE}   & \textbfs{VP SDE}  \\
       \midrule
    $\mathbf{f}(\rvx, t)$ & $\mathbf{0} $   &  $-\frac{1}{2}\beta(t)\rvx$  \\
      $g(t)$   & $\sqrt{\frac{\diff \sigma^2(t)}{\diff t}}$ 
       &  $\sqrt{\beta(t)}$  \\

    SDE   &   $\diff\rvx(t) =  g(t) \diff\rvw(t)$    &   $\diff\rvx(t) = -\frac{1}{2}\beta(t) \rvx(t) \diff t + \sqrt{\beta(t)} \diff\rvw(t)$ \\
   \midrule
     $p_t(\rvx_t|\rvx_0)$   & $        \mathcal{N}\left(\rvx_t ; \rvx_0, \left(\sigma^2(t) - \sigma^2(0)\right)\rmI\right)$     &  $ \mathcal{N}\left(\rvx_t ; \rvx_0 e^{-\frac{1}{2}\int_{0}^{t} \beta(\tau)\diff \tau}, \rmI - \rmI e^{-\int_{0}^{t} \beta(\tau)\diff \tau}\right)$ \\
   
     $p_{\mathrm{prior}}$   & $ \mathcal{N}(\bm{0}, \sigma^2(T)\rmI)$    &  $\mathcal{N}(\bm{0}, \rmI)$  \\
 \bottomrule
  \end{tabular}
  }
  \label{tb:sde_instances}
\end{table}

\subsection{VE SDE}
VE SDE has the following components:
\begin{itemize}
    \item \textbfs{Drift Term:} A zero drift term $\mathbf{f}=\bm{0}$.
    \item \textbfs{Diffusion Term:} $g(t)=\sqrt{\frac{\diff \sigma^2(t)}{\diff t}}$ for some function $\sigma(t)$.
\end{itemize}
The forward SDE then takes the following form:
\begin{equation}\label{eq:vesde}
    \diff\rvx(t) =  \sqrt{\frac{\diff \sigma^2(t)}{\diff t}} \diff\rvw(t).
\end{equation} 
Similarly, the results from \Cref{subsec:scoresde-perturbation-kernel} imply the perturbation kernel for the VE SDE and suggest selecting an appropriate prior distribution:
\begin{itemize}
    \item \textbfs{Perturbation Kernel:}
\begin{align*}
        p_t(\rvx_t | \rvx_0) &= \mathcal{N}\left(\rvx_t ; \rvx_0, \left(\sigma^2(t) - \sigma^2(0)\right)\rmI\right) 
\end{align*}
    \item \textbfs{Prior Distribution:} Assume that $\sigma(t)$ is an increasing function for $t \in [0,T]$ and that $\sigma^2(T) \gg \sigma^2(0)$. The prior distribution is given by:
    \[
    p_{\mathrm{prior}} := \mathcal{N}(\bm{0}, \sigma^2(T)\rmI).
    \]
\end{itemize}

A typical instance of a VE SDE is NCSN with the following design:
\[
\sigma(t):=\sigma_{\textup{min}}\left(\frac{\sigma_{\textup{max}}}{\sigma_{\textup{min}}} \right)^t,\quad \text{for } t\in(0,1],
\]
where $\sigma_{\textup{min}}$ and $\sigma_{\textup{max}}$ are pre-specified constants. Namely, the sequence of variances is designed as a geometric sequence. With this, NCSN is viewed as a discretized version of VE SDE, as discussed in \Cref{subsec:scoresde-motivation}.


\subsection{VP SDE}
Let $\beta\colon [0,T] \rightarrow \mathbb{R}_{\geq 0 }$ be a non-negative function of $t$. A VP SDE is defined with the following components:
\begin{itemize}
    \item \textbfs{Drift Term:} A linear drift given by $\mathbf{f}(\rvx, t) = -\frac{1}{2}\beta(t)\rvx$.
    \item \textbfs{Diffusion Term:} $g(t) = \sqrt{\beta(t)}$.
\end{itemize}
Thus, the forward SDE is expressed as:
\begin{align}\label{eq:forward-linear-sde}
\diff\rvx(t) = -\frac{1}{2}\beta(t) \rvx(t) \diff t + \sqrt{\beta(t)} \diff\rvw(t).
\end{align}

Using the results from \Cref{subsec:scoresde-perturbation-kernel}, we can derive the perturbation kernel for the VP SDE and select an appropriate prior distribution:
\begin{itemize}
    \item \textbfs{Perturbation Kernel:}
    \[
    p_t(\rvx_t|\rvx_0) = \mathcal{N}\left(\rvx_t ; \rvx_0 e^{-\frac{1}{2}\int_{0}^{t} \beta(\tau)\diff \tau}, \rmI - \rmI e^{-\int_{0}^{t} \beta(\tau)\diff \tau}\right).
    \]
    \item \textbfs{Prior Distribution:} $p_{\mathrm{prior}} := \mathcal{N}(\bm{0}, \rmI)$.
\end{itemize}
We remark that since the perturbation kernel is a Gaussian with a known mean and covariance, we can apply \Cref{eq:gaussian-score} to compute its score function.

A classic example of a VP SDE is the DDPM, where the noise schedule $\beta(t)$ is defined as:
\[
\beta(t) := \beta_{\textup{min}} + t(\beta_{\textup{max}} - \beta_{\textup{min}}), \quad \text{for all } t \in [0,1].
\]
Here, $\beta_{\textup{min}}$ and $\beta_{\textup{max}}$ are pre-defined constants. With this, DDPM can be interpreted as a discretization of the VP SDE, as discussed in \Cref{subsec:scoresde-motivation}.

\subsection{(Optional) How Is the Perturbation Kernel 
$ p_t(\rvx_t|\rvx_0) $ Derived?}\label{subsec:scoresde-perturbation-kernel}
If the drift term in the forward SDE \Cref{eq:sde_forward} is linear in $\rvx$, taking the form
\begin{align*}
    \rvf(\rvx, t) = f(t) \rvx,
\end{align*}
for some scalar-valued, time-dependent function $f(t) \in \mathbb{R}$, then \Cref{eq:sde_forward} becomes a linear SDE:
\[
\diff\rvx(t) = f(t)\rvx(t) \diff t + g(t) \diff\rvw(t).
\]

Even if the initial distribution $p_{\mathrm{data}}$ is non-Gaussian, the linearity of the drift ensures that the conditional process remains Gaussian. In particular, for $t>0$, the transition kernel admits the form:
\[
p_t( \rvx_t | \rvx_0 ) = \mathcal{N} \left( \rvx_t;  \rvm(t),  P(t) \mathbf{I}_D \right),
\]
where $\rvx_0 \sim p_{\mathrm{data}}$, and $\rvm(t) \in \mathbb{R}^D$, $P(t)\in\mathbb{R}_{\ge 0}$ denote the conditional mean and (scalar) variance given $\rvx_0$, defined as:
\[
\rvm(t) = \mathbb{E} \left[\rvx_t | \rvx(0) = \rvx_0\right], \qquad
P(t) \mathbf{I}_D = \operatorname{Cov} \left[\rvx_t | \rvx(0) = \rvx_0\right].
\]

These first and second moments evolve according to the following ODEs~\citep{sarkka2019applied}:
\begin{align}\label{eq:mean-var-evolution}
\begin{aligned}
    \frac{\diff \rvm(t)}{\diff t} &= f(t) \rvm(t), \\
    \frac{\diff P(t)}{\diff t} &= 2f(t) P(t) + g^2(t),
\end{aligned}
\end{align}
provided that the initial mean $\rvm(0)$ and variance $P(0)$ are finite.

Since both ODEs are linear, they admit closed-form solutions via the integrating factor method. 
Given the initial condition $\rvx_0$, the mean and variance evolve as
\begin{align}\label{eq:mean-var-solution}
    \rvm(t) = \mathcal{E}(0  \to  t) \rvx_0, 
\qquad 
P(t) = \int_{0}^{t} \mathcal{E}^{2}(s  \to  t)  g(s)^{2} \diff s,
\end{align}
with $\rvm(0)=\rvx_0$ and $P(0)=0$. 
Here $\mathcal{E}(s  \to  t)$ denotes the exponential integrating factor
\[
\mathcal{E}(s  \to  t) := \exp \left(\int_{s}^{t} f(u) \diff u\right),
\]
which captures the accumulated effect of the drift from time $s$ to $t$. Consequently, the transition kernel $p_t(\rvx_t | \rvx_0)$ also admits a closed-form expression. 

We defer the justification that the conditional covariance of $p_t(\rvx_t |\rvx_0)$ is isotropic, that is 
$\operatorname{Cov}[\rvx_t | \rvx_0] = P(t) \mathbf I_D$ under a $D$-dimensional Wiener process with independent coordinates and diffusion $g(t)\mathbf I_D$, as well as the derivation of \Cref{eq:mean-var-evolution}, to \Cref{subsec:rigorous-proof-linear-sde}, which relies on Itô calculus.

\exm{VE SDE's Transition Kernel}{In the special case of VE SDE: $\rvf \equiv 0$ and $g(t) = \sqrt{\frac{\diff \sigma^2(t)}{\diff t}}$, the mean and covariance of the solution to the SDE evolve as follows.

\proofparagraph{Mean.}
\[
\frac{\diff \rvm(t)}{\diff t} = 0, 
\quad \text{with} \quad \rvm(0) = \rvx_0
 \Longrightarrow 
\rvm(t) = \rvx_0.
\]

\proofparagraph{Variance.}
\[
\frac{\diff P(t)}{\diff t} = \frac{\diff \sigma^2(t)}{\diff t},
\quad \text{with} \quad P(0) = 0
 \Longrightarrow 
P(t) = \sigma^2(t) - \sigma^2(0).
\]
Therefore
\[
p_t(\rvx_t | \rvx_0)=\mathcal{N} \left(\rvx_t;  \rvx_0,  \big(\sigma^2(t)-\sigma^2(0)\big)\mathbf{I}_D\right).
\]
}

\exm{VP SDE's Transition Kernel}{In the VP SDE case with drift $\rvf(\rvx, t) = -\tfrac{1}{2}\beta(t) \rvx$ and diffusion $g(t) = \sqrt{\beta(t)}$:

\proofparagraph{Mean $\rvm(t)$.}
\[
\frac{\diff \rvm}{\diff t}
= -\frac{1}{2}\beta(t) \rvm(t),
\quad
B(t) \coloneqq \int_{0}^{t} \beta(s) \diff s,
\quad
\rvm(t) = e^{-\frac{1}{2}B(t)} \rvx_0.
\]

\proofparagraph{Variance $P(t)$.} The variance satisfies
\[
\frac{\diff P}{\diff t} = -\beta(t) P(t) + \beta(t).
\]
Applying the integrating factor $e^{B(t)}$ with $B(t) = \int_0^t \beta(s) \diff s$, we obtain
\[
\frac{\diff}{\diff t} \left[P(t)e^{B(t)}\right] = \beta(t) e^{B(t)}.
\]
Integrating both sides gives
\[
P(t) = 1 - e^{-B(t)}.
\]
Hence the covariance is isotropic with
\[
\rmP(t) = P(t) \mathbf I_D = \bigl(1 - e^{-B(t)}\bigr)\mathbf I_D.
\]

\proofparagraph{Final Closed-Form Transition Kernel.}
\[
p_t(\rvx_t|\rvx_0)
= \mathcal{N} \left(
   \rvx_t; 
   \underbrace{e^{-\frac{1}{2} B(t)} \rvx_0}_{\rvm(t)}, 
   \underbrace{\big(1 - e^{-B(t)}\big)\mathbf{I}_D}_{P(t)\mathbf{I}_D}
\right),
\quad
B(t) = \int_0^t \beta(s) \diff s.
\]
}

\clearpage
\newpage

\section{\texorpdfstring{A Direct View Through Accumulated Perturbation Kernels}{A Direct View Through Accumulated Perturbation Kernels}}
\label{sec:variational-perspective}

DDPM and Score SDE are commonly introduced by specifying how noise is added
locally in time: through discrete forward transitions
$p(\rvx_t|\rvx_{t-\Delta t})$ in DDPM, or through a continuous-time SDE in
Score SDE. For training, however, a particularly useful object is the
\emph{accumulated perturbation kernel}
\[
p_t(\rvx_t|\rvx_0),
\]
which describes the distribution at time $t$ directly conditioned on the
original data sample. This is precisely the form that appears naturally in
the DDPM and continuous-time denoising objectives
\Cref{eq:ddpm-mean-loss,eq:conti-dsm_loss-real}.

This suggests a complementary viewpoint: rather than beginning with the local
noise injection and then deriving $p_t(\rvx_t|\rvx_0)$, we can specify the
accumulated perturbation kernel directly. For the linear-Gaussian processes
considered below, an admissible accumulated kernel determines the corresponding
linear SDE, and vice versa. This viewpoint makes the amount of signal and noise
at each time explicit, connects directly to practical training objectives, and
will also provide a useful bridge to the probability-path viewpoint of Flow
Matching in \Cref{ch:flow-based}.

\begin{figure}[th!]
    \centering
    \includegraphics[width=\linewidth]{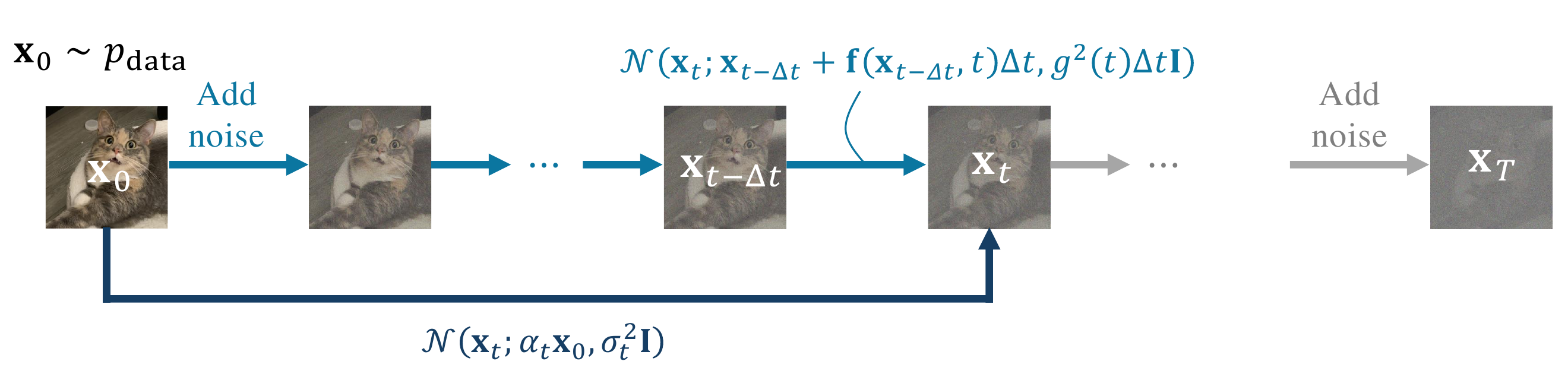}
    \caption{\textbfs{Two views of the same affine forward process.}
    For an admissible affine diffusion schedule, incremental noise injection
    through a continuous-time linear SDE and direct specification of the
    accumulated perturbation kernel in \Cref{eq:forward_kernel} describe the
    same conditional distributions $p_t(\rvx_t|\rvx_0)$.
    \figcredit{Created by the authors.}}
    \label{fig:vdm-sde}
\end{figure}

\paragraph{A General Affine Forward Process $p_t(\rvx_t|\rvx_0)$.}
We begin by defining the accumulated forward perturbation kernel
\begin{equation}\label{eq:forward_kernel}
p_t(\rvx_t|\rvx_0)
:=
\mathcal{N}\bigl(
\rvx_t;
\alpha_t\rvx_0,
\sigma_t^2\mathbf{I}
\bigr),
\end{equation}
where $\rvx_0\sim p_{\mathrm{data}}$, and $\alpha_t$ and $\sigma_t$ are
nonnegative scalar functions of $t\in[0,T]$. In the standard forward
diffusion setup, we take
\[
\alpha_0=1,
\qquad
\sigma_0=0,
\]
so that the process starts exactly from the data, and assume
$\alpha_t>0$ and $\sigma_t>0$ for $t\in(0,T]$ whenever the corresponding
ratios are used.

Equivalently, a sample from \Cref{eq:forward_kernel} can be written as
\[
\rvx_t
=
\alpha_t\rvx_0
+
\sigma_t\bm{\epsilon},
\qquad
\bm{\epsilon}\sim\mathcal{N}(\mathbf{0},\mathbf{I}).
\]

This simple form includes several familiar probability paths:
\begin{itemize}
    \item \textbfs{VE (NCSN):}
    $\alpha_t\equiv1$, with the noise scale $\sigma_t$ increasing over time;
    \item \textbfs{VP (DDPM):}
    $\alpha_t^2+\sigma_t^2=1$;
    \item \textbfs{Linear Gaussian interpolation:}
    $\alpha_t=1-t$ and $\sigma_t=t$ on $t\in[0,1]$, corresponding to the
    familiar linear path used in Flow Matching and Rectified Flow when written
    in the data-to-noise time convention.
\end{itemize}

\subsection{Connection to Score SDE}
For Score SDE, it is often more intuitive to specify the accumulated
perturbation kernel $p_t(\rvx_t|\rvx_0)$ first and then recover the
corresponding SDE coefficients, rather than starting from the drift and
diffusion terms and solving ODEs for the moments
(see \Cref{subsec:scoresde-perturbation-kernel}).

For an admissible affine schedule, the corresponding forward SDE takes the
linear-in-$\rvx$ form in \Cref{eq:forward-linear-sde}:
\begin{align*}
    \diff \rvx(t)
    =
    \underbrace{f(t)\rvx(t)}_{\rvf(\rvx(t),t)}
    \diff t
    +
    g(t)\diff\rvw(t),
\end{align*}
where $f,g\colon[0,T]\rightarrow\mathbb{R}$ are real-valued functions of time.

The coefficients $f(t)$ and $g(t)$ can be recovered analytically from
$\alpha_t$ and $\sigma_t$. Importantly, not every choice of
$\alpha_t$ and $\sigma_t$ defines a valid diffusion: the implied diffusion
variance $g^2(t)$ must remain nonnegative. Equivalently, in this
linear-Gaussian setting, the signal-to-noise ratio
$\alpha_t^2/\sigma_t^2$ cannot increase with time. The following lemma makes
this relationship precise.

\lemp{Forward Perturbation Kernel $\Leftrightarrow$ Linear SDE}{forward-sde}{
Assume $\alpha_0=1$, $\sigma_0=0$, and that $\alpha_t>0$ and
$\sigma_t>0$ are differentiable for $t\in(0,T]$. Define
$\lambda_t:=\log\tfrac{\alpha_t}{\sigma_t}$.

Given the forward perturbation kernel
\[
\rvx_t=\alpha_t\rvx_0+\sigma_t\beps,
\qquad
\beps\sim\mathcal N(\mathbf 0,\rmI),
\]
there exists a corresponding linear SDE
\[
\diff \rvx(t)=f(t)\rvx(t)\diff t+g(t)\diff\rvw(t)
\]
with the same conditional kernel
$p_t(\rvx_t|\rvx_0)
=\mathcal N(\rvx_t;\alpha_t\rvx_0,\sigma_t^2\rmI)$
if and only if the coefficients
\begin{align}\label{eq:kernel-sde-equiv}
\begin{aligned}
f(t)
&=\frac{\diff}{\diff t}\log\alpha_t,\\
g^2(t)
&=\frac{\diff\sigma_t^2}{\diff t}
-2\frac{\diff}{\diff t}\log\alpha_t\,\sigma_t^2
=-2\sigma_t^2\frac{\diff}{\diff t}\lambda_t
\end{aligned}
\end{align}
satisfy $g^2(t)\geq0$ for $t\in(0,T]$.
Equivalently, $\lambda_t$ must be non-increasing.

Conversely, if a linear SDE has conditional kernels
$\mathcal N(\rvx_t;\alpha_t\rvx_0,\sigma_t^2\rmI)$ with
$\alpha_t>0$ and $\sigma_t>0$, then its coefficients necessarily satisfy
\Cref{eq:kernel-sde-equiv}.
}{
From \Cref{subsec:scoresde-perturbation-kernel}, the result follows by
matching the mean and covariance ODEs
$\rvm'(t)=f(t)\rvm(t)$ and
$\rmP'(t)=2f(t)\rmP(t)+g^2(t)\rmI$
with $\rvm(t)=\alpha_t\rvx_0$ and
$\rmP(t)=\sigma_t^2\rmI$.
}

\rmkb{To make the terminal conditional distribution independent of $\rvx_0$ and
equal to a prescribed Gaussian prior, we need
$\alpha_T=0$ and $\sigma_T^2$ equal to the prior variance. Since
\[
\alpha_t
=
\exp\Big(\int_0^t f(u)\diff u\Big),
\]
achieving $\alpha_T=0$ at a finite terminal time requires
\[
\int_0^T f(u)\diff u=-\infty.
\]
Thus, the contraction induced by the drift must become singular near the
terminal time; in particular, $f$ cannot remain bounded on $[0,T]$.

The diffusion coefficient is related to the prescribed variance through
\[
g^2(t)
=
(\sigma_t^2)'
-
2\frac{\alpha_t'}{\alpha_t}\sigma_t^2.
\]
For the usual regular schedules in which
$\sigma_t^2$ approaches a finite nonzero terminal variance,
$(\sigma_t^2)'$ remains bounded, and
$\alpha_t'/\alpha_t\to-\infty$, this also gives
\[
g^2(t)\to\infty
\qquad\text{as }t\to T.
\]
More generally, the precise behavior of $g(t)$ depends on the chosen
schedule.

If $f$ and $g$ remain bounded on a finite interval $[0,T]$, then
$\alpha_T>0$, so exact forgetting cannot occur at finite $T$.
In practice, one therefore chooses a terminal time for which the residual
dependence on $\rvx_0$ is negligible and $p_T\approx p_{\mathrm{prior}}$, or
considers an infinite-time formulation in which $\alpha_t\to0$
asymptotically.
}

From the above lemma, specifying the incremental noise injection via a linear SDE with coefficients $f(t)$ and $g(t)$ is mathematically equivalent to defining the perturbation kernel with parameters $\alpha_t$ and $\sigma_t$. In the diffusion model literature, these two viewpoints are used interchangeably. Therefore, we conclude:
\msg{Observation}{}{
For a valid affine diffusion schedule, the two descriptions are equivalent:
we may either specify the accumulated perturbation kernel
$p_t(\rvx_t|\rvx_0)$ through $\alpha_t$ and $\sigma_t$, or specify the
corresponding linear SDE through $f(t)$ and $g(t)$.
}

\subsection{Connection to Variational-Based Diffusion Model}\label{subsec:connection-vdm}
We revisit a core identity from DDPM, derived via Bayes’ rule:
\begin{align}\label{eq:ddpm-reverse-condition-kernel}
    p(\rvx_{t-\Delta t}|\rvx_t, \rvx) = p(\rvx_t|\rvx_{t-\Delta t}) \cdot \frac{p_{t-\Delta t}(\rvx_{t-\Delta t}|\rvx)}{p_t(\rvx_t|\rvx)},
\end{align}
for any $\rvx$ (usually $\rvx\sim p_{\mathrm{data}}$). This reverse conditional $ p(\rvx_{t-\Delta t} | \rvx_t, \rvx) $ is central to modeling, enabling both tractable training targets and efficient sampling.

Although DDPM typically defines the incremental kernel $p(\rvx_t|\rvx_{t-\Delta t})$ first, the accumulated transition $p_t(\rvx_t|\rvx_0)$ often provides a more interpretable and practical formulation, especially for the prior and loss design.

\paragraph{Deriving Transition Kernels.}
We now extend this to the continuous-time setting. Let $0 \leq t < s \leq T$ be two (continuous) time points. Given the perturbation kernel $p_t(\rvx_t|\rvx_0)$, we can compute the reverse conditional $p(\rvx_t|\rvx_s, \rvx)$ for any $\rvx$ by applying \Cref{eq:ddpm-reverse-condition-kernel}\footnote{This identity extends naturally to continuous time by treating $s$ as a general earlier time.}, using the forward kernel $p(\rvx_s|\rvx_t)$ as an intermediate. The following lemma summarizes this derivation, extending Lemma~\ref{ddpm-reverse-kernel} without assuming $\alpha_t^2 + \sigma_t^2 = 1$.

\lemp{Reverse Conditional Transition Kernels}{reverse-kernel}{
Let $0 \leq t < s \leq T$. The reverse \emph{conditional} transition kernel is:
\begin{equation*}
    p(\rvx_t|\rvx_s, \rvx) = \mathcal{N}\left(\rvx_t; \bm{\mu}(\rvx_s, \rvx; s, t), \sigma^2(s, t) \bfI\right),
\end{equation*}
where
\begin{align}\label{eq:p_backward_mean_var}
\begin{aligned}
    \bm{\mu}(\rvx_s, \rvx; s, t) := \frac{\alpha_{s|t} \sigma_t^2}{\sigma_s^2} \rvx_s + \frac{\alpha_t \sigma^2_{s|t}}{\sigma_s^2} \rvx, \quad
    \sigma^2(s, t) := \sigma^2_{s|t} \frac{\sigma_t^2}{\sigma_s^2}.
\end{aligned}
\end{align}
Here, $\alpha_{s|t}$ and $\sigma_{s|t}$ are defined as:
\begin{equation*}
    \alpha_{s|t} := \frac{\alpha_s}{\alpha_t}, \qquad \sigma^2_{s|t} := \sigma_s^2 - \alpha_{s|t}^2 \sigma_t^2.
\end{equation*}
}{
We first compute the forward transition kernel:
\begin{equation}\label{eq:zt_given_zs}
    p(\rvx_s|\rvx_t) = \mathcal{N}\left(\rvx_s; \alpha_{s|t} \rvx_t, \sigma^2_{s|t} \bfI\right).
\end{equation}
The reverse kernel then follows from Bayes’ rule, and since all involved distributions are Gaussian, the result can be derived by direct computation. For further details, see Appendix A of \citep{kingma2021variational}.
}

Although $p(\rvx_{t + \Delta t}|\rvx_t)$ and $p_t(\rvx_t|\rvx_0)$ are theoretically equivalent, $p_t(\rvx_t|\rvx_0)$ often takes a more central role. The step-wise transition in \Cref{eq:zt_given_zs} mainly serves to obtain a closed-form reverse kernel. Recent works~\citep{kingma2021variational} thus favor directly specifying $p_t(\rvx_t|\rvx_0)$ for its clarity and interpretability.

\paragraph{Reverse Process Modeling, Training, and Sampling.}  
The training objective (ELBO in \Cref{eq:ddpm-elbo}) and the modeling framework introduced in \Cref{sec:ddpm} remain applicable under our generalized setting. For clarity, we adopt the $\rvx$-prediction formulation, denoted by $\rvx_{\bm{\phi}}(\rvx_s, s)$, following~\citet{kingma2021variational}. However, the equivalent $\beps$-prediction perspective, represented by $\bm{\epsilon}_{\bm{\phi}}(\rvx_s, s)$, is also valid due to the relationship (as in \Cref{eq:ddpm-clean-noise})
\[
\rvx_s = \alpha_s \rvx_{\bm{\phi}}(\rvx_s, s) + \sigma_s \bm{\epsilon}_{\bm{\phi}}(\rvx_s, s),\quad \text{for any given } \rvx_s \sim q_s.
\]

\subparagraph{Modeling and Diffusion Loss $\mathcal{L}_{\text{diffusion}}$.}  
Similar to DDPM, the conditional distribution $p(\rvx_t|\rvx_s, \rvx)$ in \Cref{eq:p_backward_mean_var} motivates replacing the clean signal $\rvx$ with a learnable predictor $\rvx_{\bm{\phi}}(\rvx_s, s)$, yielding a parameterized reverse model of the form:
\begin{align}\label{eq:general-ddpm-sampling}
    p_{\bm{\phi}}(\rvx_t | \rvx_s) := \mathcal{N}\left(\rvx_t; \bm{\mu}_{\bm{\phi}}(\rvx_s, s, t), \sigma^2(s,t) \mathbf{I}\right),
\end{align}
with the mean parametrized as:
\[
    \bm{\mu}_{\bm{\phi}}(\rvx_s, s, t) = \frac{\alpha_{s|t} \sigma_t^2}{\sigma_s^2} \rvx_s + \frac{\alpha_t \sigma_{s|t}^2}{\sigma_s^2} \rvx_{\bm{\phi}}(\rvx_s, s).
\]
Given the forward kernel in \Cref{eq:forward_kernel}, the KL divergence in $\mathcal{L}_{\text{diffusion}}(\rvx; \bm{\phi})$ reduces to a weighted regression loss:
\begin{align}
\begin{aligned}
    \label{eq:snr-kl}
    \mathcal{D}_{\text{KL}}\big(p(\rvx_t|\rvx_s, \rvx_0)  \| p_{\bm{\phi}}(\rvx_t|\rvx_s)\big)  
    &= \frac{1}{2\sigma^2(s,t)} \left\| \bm{\mu}(\rvx_s, \rvx_0; s,t) - \bm{\mu}_{\bm{\phi}}(\rvx_s, s, t) \right\|_2^2 \\
    &= \frac{1}{2} \big(\text{SNR}(t) - \text{SNR}(s)\big) \left\| \rvx_0 - \rvx_{\bm{\phi}}(\rvx_s, s) \right\|_2^2,
\end{aligned}
\end{align}
where $\rvx_s = \alpha_s\rvx_0 + \sigma_s\bm{\epsilon}$, with $\rvx_0 \sim p_{\mathrm{data}}$, $\bm{\epsilon} \sim \mathcal{N}(\bm{0}, \rmI)$, and $\text{SNR}(s) := \alpha_s^2 / \sigma_s^2$ denotes the signal-to-noise ratio at time $s$.

\rmkb{
In \citep{kingma2021variational}, the authors study the continuous-time limit of \Cref{eq:snr-kl} as $t \to s$, yielding:
\begin{align*}
\mathcal{L}_{\text{VDM}}^\infty(\rvx_0) 
= -\frac{1}{2}  \mathbb{E}_{s, \bm{\epsilon} \sim \mathcal{N}(\bm{0}, \bfI)}  \text{SNR}'(s)  \big\| \rvx_0 - \rvx_{\bm{\phi}}(\rvx_s, s) \big\|_2^2.
\end{align*}
This setup also introduces a learnable noise schedule, and while it generalizes beyond continuous data, such extensions fall outside the scope of our current discussion.
}

\subparagraph{Sampling.}  
Sampling proceeds similarly to DDPM using the parameterized kernel from \Cref{eq:general-ddpm-sampling}:
\begin{align}\label{eq:general-ddpm-sampling-discrete}
    \rvx_t = \underbrace{\frac{\alpha_{s|t} \sigma_t^2}{\sigma_s^2} \rvx_s + \frac{\alpha_t \sigma_{s|t}^2}{\sigma_s^2} \rvx_{\bm{\phi}^\times}(\rvx_s, s)}_{\bm{\mu}_{\bm{\phi}^\times}(\rvx_s, t, s)} + \sigma_{s|t} \frac{\sigma_t}{\sigma_s} \bm{\epsilon}_s,\quad  \bm{\epsilon}_s \sim \mathcal{N}(\bm{0}, \rmI).
\end{align}

\clearpage
\newpage

\section{(Optional) Fokker–Planck Equation and Reverse-Time SDEs \\via Marginalization and Bayes’ Rule}\label{eq:heuristic-fpe-sde}
In this section, we offer a probabilistic perspective on the structure of the Fokker–Planck equation and the reverse-time SDE. By leveraging fundamental tools such as the marginalization trick and Bayes' rule, we illuminate the connection between the statistical formulation of stochastic processes and their corresponding differential equations.

We emphasize that the ``derivation'' presented here is not  mathematically rigorous proofs, but rather heuristic arguments intended to convey the underlying connections.

\subsection{Fokker-Planck Equation from the Marginalization of Transition Kernels}\label{subsec:fpe-reason}
Given the forward transition probability as in \Cref{eq:transition-forward-f-g}
\begin{align*}
    p(\mathbf{x}_{t+\Delta t} | \mathbf{x}_t) = \mathcal{N}\left(\mathbf{x}_{t+\Delta t}; \mathbf{x}_t + \mathbf{f}(\mathbf{x}_t, t)\Delta t, g^2(t)\Delta t \mathbf{I}\right),
\end{align*}
and the marginal distributions
\begin{align*}
    p_t(\mathbf{x}_t), \quad p_{t+\Delta t}(\mathbf{x}_{t+\Delta t}),
\end{align*}
we aim to derive the Fokker-Planck equation that governs the time evolution of the marginal distribution $p_t$.

\paragraph{Change of Variables.}
By the Markov property, the marginal distribution at time $t + \Delta t$ can be expressed as an integral over the previous state $\mathbf{x}_t$ (i.e., Chapman-Kolmogorov equation):
\begin{align*}
p_{t+\Delta t}(\rvx)
= \int \mathcal N \big(\rvx; \rvy+\rvf(\rvy,t)\Delta t, g^2(t)\Delta t \rmI\big) p_t(\rvy) \diff\rvy.
\end{align*}
 We introduce a new variable
\[
\rvu  :=  \rvy+\rvf(\rvy,t)\Delta t,
\]
so the Gaussian is centered at $\rvu$. For small $\Delta t$, this map is invertible with
\[
\rvy  =  \rvu-\rvf(\rvu,t)\Delta t+\mathcal O(\Delta t^2),
\quad
\Bigg|\det\frac{\partial \rvy}{\partial \rvu}\Bigg|
 = 1-(\nabla_\rvu \cdot\rvf)(\rvu,t) \Delta t+\mathcal O(\Delta t^2).
\]
Hence, change-of-variable formula leads us to:
\begin{align*}
    p_{t+\Delta t}(\rvx)
&= \int \mathcal N\left(\rvx;\rvu,g^2(t)\Delta t\rmI\right) \cdot
\\\quad\Big[p_t(\rvu)&-\Delta t \rvf(\rvu,t) \cdot \nabla_\rvu p_t(\rvu)
-\Delta t (\nabla_\rvu \cdot\rvf)(\rvu,t) p_t(\rvu)\Big] \diff\rvu
+\mathcal O(\Delta t^2),
\end{align*}

\paragraph{Taylor Expansion.}
For any smooth function $\phi:\R^D\to\R$ and scale $\sigma>0$,  
if $\rvz\sim\mathcal N(\bm{0},\rmI)$, the following approximation holds  
(known as the \emph{Taylor–Gaussian smoothing formula}):

\[
\int \mathcal N(\rvx;\rvu,\sigma^2\rmI) \phi(\rvu) \diff\rvu
=\E \left[\phi(\rvx+\sigma\rvz)\right]
=\phi(\rvx)+\frac{\sigma^2}{2} \Delta_\rvx\phi(\rvx)+\mathcal O(\sigma^4).
\]
This is because Taylor expansion for: 
\[
\phi(\rvx+\sigma\rvz)=\phi(\rvx)+\sigma\nabla_\rvx\phi(\rvx) \cdot \rvz+\frac{\sigma^2}{2}\rvz^\top\nabla_\rvx^2\phi(\rvx)\rvz+\mathcal O(\sigma^3)
\]
and $\E[\rvz]=\bm{0}$, $\E[\rvz\rvz^\top]=\rmI$.

Apply this with $\phi=p_t$, $\phi=\rvf \cdot \nabla_\rvu p_t$, and $\phi=(\nabla_\rvu \cdot\rvf) p_t$, and use $\sigma^2=g^2(t)\Delta t$, we can obtain
\[
\begin{aligned}
&p_{t+\Delta t}(\rvx) - p_t(\rvx)
\\=& -\Delta t \rvf(\rvx,t) \cdot \nabla_\rvx p_t(\rvx)
-\Delta t (\nabla_\rvx \cdot\rvf)(\rvx,t) p_t(\rvx)
+\frac{g^2(t)}{2}\Delta t \Delta_\rvx p_t(\rvx)
+\mathcal O(\Delta t^2) 
\\=&  - \Delta t \nabla_\rvx \cdot \big(\rvf(\rvx,t) p_t(\rvx)\big)
 + \frac{g^2(t)}{2} \Delta t \Delta_\rvx p_t(\rvx)
 + \mathcal O(\Delta t^2).
\end{aligned}
\]
Divide by $\Delta t$ and let $\Delta t\to0$ to obtain the Fokker–Planck equation.

In \Cref{subsec:rigorous-proof-fpe}, we present the It\^o--based derivation to complement the discrete--time view above.

\subsection{Why Does Reverse-Time SDE Take The Form?}\label{subsec:reverse-sde-reason}

The rigorous derivation of the reverse-time SDE is technical and requires delving into the properties of the Fokker-Planck equation. However, the form of the reverse-time SDE can be understood intuitively through Bayes' theorem. Here, we present a heuristic derivation to provide insight into why \Cref{eq:sde_backward} takes its form, with the appearance of score functions\footnote{This derivation is inspired by the approach in \href{https://alexxthiery.github.io/notes/reverse_and_tweedie/reverse_and_tweedie.html}{this post}.}.

\paragraph{Using Bayes' Rule for Inversion.} 
Our goal is to determine the reverse-time transition kernel by first considering the discrete-time case:
\[
p(\rvx_t | \rvx_{t+\Delta t}),
\]
and then taking $\Delta t \rightarrow 0$ to obtain the continuous-time formulation.
Using Bayes' theorem, we express:
\begin{align}\label{eq:reverse-sde-bayes-discrete}
    p(\rvx_t | \rvx_{t+\Delta t}) 
    &= p(\rvx_{t+\Delta t} | \rvx_t) \frac{p_t(\rvx_t)}{p_{t+\Delta t}(\rvx_{t+\Delta t})} \nonumber \\
    &= p(\rvx_{t+\Delta t} | \rvx_t) \exp\left(\log p_t(\rvx_t) - \log p_{t+\Delta t}(\rvx_{t+\Delta t})\right).
\end{align}

The forward transition kernel is assumed to be as in \Cref{eq:transition-forward-f-g}:
\begin{equation*}
p(\rvx_{t+\Delta t} | \rvx_t) = \mathcal{N}\left(\rvx_{t+\Delta t}; \rvx_t + \rvf(\rvx_t, t)\Delta t, g^2(t)\Delta t \mathbf{I}\right)
\end{equation*}

\paragraph{Taylor Expansion.} 
To handle the exponential term, we apply a first-order Taylor expansion. The key insight is to expand around the point $(\rvx_t, t)$ in both space and time:
\begin{align*}
    \log p_{t+\Delta t}(\rvx_{t+\Delta t}) &= \log p_t(\rvx_t) + \nabla_{\rvx} \log p_t(\rvx_t) \cdot (\rvx_{t+\Delta t} - \rvx_t) \nonumber \\
    &\quad + \frac{\partial \log p_t(\rvx_t)}{\partial t} \Delta t + \mathcal{O}(\|\rvh\|_2^2) 
\end{align*}
where $\rvh := (\rvx_{t+\Delta t} - \rvx_t, \Delta t)$. Therefore:
\begin{align}
    \log p_t(\rvx_t) - \log p_{t+\Delta t}(\rvx_{t+\Delta t}) &= -\nabla_{\rvx} \log p_t(\rvx_t) \cdot (\rvx_{t+\Delta t} - \rvx_t) \nonumber \\
    &\quad - \frac{\partial \log p_t(\rvx_t)}{\partial t} \Delta t + \mathcal{O}(\|\rvh\|_2^2) \label{eq:log-difference}
\end{align}

For a diffusion process,
$\rvx_{t+\Delta t}-\rvx_t=\mathcal O_p(\sqrt{\Delta t})$, so the
second-order spatial Taylor terms are generally of order
$\mathcal O_p(\Delta t)$. Accordingly, the following calculation should be
understood as a leading-order heuristic.

\paragraph{Substituting into the Reverse Transition.}
Substituting equations \Cref{eq:transition-forward-f-g} and \Cref{eq:log-difference} into \Cref{eq:reverse-sde-bayes-discrete}:
\begin{align*}
    &p(\rvx_t | \rvx_{t+\Delta t}) \nonumber \\
    = &\frac{1}{(2\pi g^2(t) \Delta t)^{D/2}} \exp\left(-\frac{\|\rvx_{t+\Delta t} - \rvx_t - \rvf(\rvx_t, t) \Delta t\|_2^2}{2 g^2(t) \Delta t}\right) \nonumber \\
    &\qquad \cdot \exp\left(-\nabla_{\rvx} \log p_t(\rvx_t) \cdot (\rvx_{t+\Delta t} - \rvx_t) - \frac{\partial \log p_t(\rvx_t)}{\partial t} \Delta t +\mathcal{O}(\Delta t)\right). 
\end{align*}
\paragraph{Algebraic Manipulation.}
The key step is to complete the square in the exponent. We have:
\begin{align*}
    &-\frac{\|\rvx_{t+\Delta t} - \rvx_t - \rvf(\rvx_t, t) \Delta t\|_2^2}{2 g^2(t) \Delta t} - \nabla_{\rvx} \log p_t(\rvx_t) \cdot (\rvx_{t+\Delta t} - \rvx_t) \nonumber \\
    = &-\frac{\left[\|\rvx_{t+\Delta t} - \rvx_t - \rvf(\rvx_t, t) \Delta t\|_2^2 + 2 g^2(t) \Delta t \nabla_{\rvx} \log p_t(\rvx_t) \cdot (\rvx_{t+\Delta t} - \rvx_t)\right]}{2 g^2(t) \Delta t} 
\end{align*}
Let $\boldsymbol{\delta} := \rvx_{t+\Delta t} - \rvx_t$ and $\boldsymbol{\mu} := \rvf(\rvx_t, t) \Delta t$. Then:
\begin{align*}
    &\|\boldsymbol{\delta} - \boldsymbol{\mu}\|_2^2 + 2 g^2(t) \Delta t \nabla_{\rvx} \log p_t(\rvx_t) \cdot \boldsymbol{\delta} \nonumber \\
    = &\|\boldsymbol{\delta}\|_2^2 - 2\boldsymbol{\delta} \cdot \boldsymbol{\mu} + \|\boldsymbol{\mu}\|_2^2 + 2 g^2(t) \Delta t \nabla_{\rvx} \log p_t(\rvx_t) \cdot \boldsymbol{\delta} \nonumber \\
    = &\|\boldsymbol{\delta}\|_2^2 - 2\boldsymbol{\delta} \cdot [\boldsymbol{\mu} - g^2(t) \Delta t \nabla_{\rvx} \log p_t(\rvx_t)] + \|\boldsymbol{\mu}\|_2^2 \nonumber \\
    = &\|\boldsymbol{\delta} - [\boldsymbol{\mu} - g^2(t) \Delta t \nabla_{\rvx} \log p_t(\rvx_t)]\|_2^2 - \|g^2(t) \Delta t \nabla_{\rvx} \log p_t(\rvx_t)\|_2^2 
\end{align*}
Substituting back:
\begin{align*}
    &\|\boldsymbol{\delta} - [\rvf(\rvx_t, t) \Delta t - g^2(t) \Delta t \nabla_{\rvx} \log p_t(\rvx_t)]\|_2^2 \nonumber \\
    = &\|\rvx_{t+\Delta t} - \rvx_t - [\rvf(\rvx_t, t) - g^2(t) \nabla_{\rvx} \log p_t(\rvx_t)] \Delta t\|_2^2. 
\end{align*}
Therefore,
\begin{align*}
    p(\rvx_t | \rvx_{t+\Delta t}) 
    =~&\frac{1}{(2\pi g^2(t) \Delta t)^{D/2}} \nonumber \\
    \quad &\cdot \exp\left(-\frac{\|\rvx_{t+\Delta t} - \rvx_t - [\rvf(\rvx_t, t) - g^2(t) \nabla_{\rvx} \log p_t(\rvx_t)] \Delta t\|_2^2}{2 g^2(t) \Delta t}\right) \nonumber \\
    \quad &\cdot \exp(\mathcal{O}(\Delta t)) \nonumber \\
    =~&\mathcal{N}\left(\rvx_t; \rvx_{t+\Delta t} - [\rvf(\rvx_t, t) - g^2(t) \nabla_{\rvx} \log p_t(\rvx_t)] \Delta t, g^2(t) \Delta t \mathbf{I}\right) \nonumber \\
    \quad &\cdot (1 + \mathcal{O}(\Delta t)).
\end{align*}
The additional term $\|g^2(t) \Delta t \nabla_{\rvx} \log p_t(\rvx_t)\|_2^2$ from completing the square is $\mathcal{O}((\Delta t)^2)$ and can be absorbed into the error term. Similarly, the time derivative term $\frac{\partial \log p_t(\rvx_t)}{\partial t} \Delta t$ is $\mathcal{O}(\Delta t)$ and will vanish in the continuous limit.

\paragraph{Taking $\Delta t \rightarrow 0$ Limit.}
As $\Delta t \approx 0$, under smoothness assumptions, the following approximations hold:
\begin{align*}
    \rvf(\rvx_t, t) &\approx \rvf(\rvx_{t+\Delta t}, t+\Delta t), \\
    g(t) &\approx g(t+\Delta t), \\
    \nabla_{\rvx} \log p_t(\rvx_t) &\approx \nabla_{\rvx} \log p_{t+\Delta t}(\rvx_{t+\Delta t})  = \rvs(\rvx_{t+\Delta t}, t+\Delta t).
\end{align*}
Using these approximations and some rearrangements, we obtain:
\begin{align*}
    &p(\rvx_t | \rvx_{t+\Delta t}) \\ \approx &\frac{1}{(2\pi g^2(t) \Delta t)^{D/2}}  \exp \Bigg(
            \\ &-\frac{\Big\|\rvx_{t} - \Big(\rvx_{t+\Delta t} - \big[\rvf(\rvx_{t+\Delta t}, t+\Delta t) - g^2({t+\Delta t}) \rvs(\rvx_{t+\Delta t}, t+\Delta t)\big] \Delta t\Big) \Big\|_2^2 }{2 g^2(t+\Delta t) \Delta t} \\
            &\Bigg).
\end{align*}
This implies that $p(\rvx_t | \rvx_{t+\Delta t})$ is roughly a normal distribution with:
\begin{align*}
    \text{\textbfs{Mean:}} &  \quad\rvx_{t+\Delta t} - \big[\rvf(\rvx_{t+\Delta t}, t+\Delta t) - g^2(t+\Delta t) \rvs (\rvx_{t+\Delta t}, t+\Delta t)\big] \Delta t, \\
    \text{\textbfs{Covariance:}} & \quad g^2(t+\Delta t) \Delta t \mathbf{I}.
\end{align*}
Taking the limit as $\Delta t \rightarrow 0$, we ``derive'' the reverse-time continuous SDE given in \Cref{eq:sde_backward}.

\clearpage
\newpage

\section{Closing Remarks}

This chapter marked a pivotal moment in our journey, unifying the discrete-time diffusion processes from the variational and score-based perspectives into a single, elegant continuous-time framework. We demonstrated that both DDPM and NCSN can be understood as discretizations of Stochastic Differential Equations (SDEs) with different drift/volatility coefficients.

The cornerstone of this framework is the existence of a corresponding reverse-time SDE, which formally defines a generative process that reverses the noise corruption. Crucially, the drift of this reverse process depends on a single unknown quantity: the score function, $\nabla_{\mathbf{x}}\log{p_{t}(\mathbf{x})}$, of the marginal data distributions at every point in time. This insight solidifies the score function's central role in generative modeling.

Furthermore, we introduced a purely deterministic counterpart, the Probability Flow Ordinary Differential Equation (PF-ODE), whose solution trajectories evolve along the same marginal densities $\{p_{t}\}$ as the SDEs. This remarkable consistency is guaranteed by the underlying Fokker-Planck equation. The profound implication is that the complex task of generation is fundamentally equivalent to solving a differential equation. Training reduces to learning the score function that defines the equation's vector field, while sampling becomes a problem of numerical integration.

The introduction of the PF-ODE, a purely deterministic flow, provides a powerful bridge to the third and final perspective on diffusion models. This concept of learning a deterministic transformation governed by a velocity field is the central principle of recent major family of generative models. In the next chapter, we will:
\begin{enumerate}
    \item Explore this flow-based perspective, starting from its origins in Normalizing Flows and Neural ODEs.
    \item Show how this viewpoint leads to the modern framework of Flow Matching, which directly learns a velocity field to transport samples between distributions.
\end{enumerate}

Ultimately, we will see how the deterministic PF-ODE, which we derived from stochastic principles, can be constructed and generalized from this entirely different, flow-based origin, completing our unified picture of diffusion modeling.

\chapter{Flow-Based Perspective: From NFs to Flow Matching}\label{ch:flow-based}

\epigraph{
    \textit{Everything flows.
}}{Heraclitus}

The \emph{change-of-variables formula}, a cornerstone of probability
theory~\citep{tabak2010density,tabak2013family}, takes on new life in modern
generative modeling. In \Cref{ch:score-sde}, we saw that the Fokker--Planck
equation describes how probability densities evolve under a diffusion process,
and that the same marginal density evolution can also be realized
deterministically through the associated PF-ODE. This naturally suggests
studying generative modeling directly from the viewpoint of deterministic
probability transport.

\paragraph{Change-of-Variables Formula of Densities.}
Given an invertible transformation $\rvf$, the density of
$\rvx=\rvf(\rvz)$, where $\rvz\sim p_{\mathrm{prior}}$, is
\begin{mdframed}
\begin{align}\label{eq:nf-change-of-var}
p(\rvx)
=
p_{\mathrm{prior}}(\rvz)
\left|
\det
\frac{\partial \rvf^{-1}(\rvx)}{\partial \rvx}
\right|,
\qquad
\text{where }
\rvz=\rvf^{-1}(\rvx).
\end{align}
\end{mdframed}
This deceptively simple formula enables bidirectional transport of samples and
densities when $\rvf$ is tractable, forming the foundation of Normalizing
Flows, which we introduce in \Cref{sec:flow-based-method}. But what happens
when this transformation is viewed continuously in time?

In this chapter, we follow this idea from Normalizing Flows to Neural ODEs and
then to Flow Matching. The key connection to \Cref{ch:score-sde} is especially
important. In Score SDEs, the learned score is used to construct the velocity field of
the PF-ODE. Flow Matching instead learns a velocity field directly. For the
same diffusion probability path, these provide two views of the same
deterministic probability flow.

A useful mental picture for this chapter is therefore, where both rely on the same conditioning principle:

\begin{figure}[th!]
\centering
\newcommand{\subnote}[1]{{\scriptsize\itshape\textcolor{black!62}{#1}}}
\resizebox{\ifdim\width>\linewidth \linewidth\else \width\fi}{!}{%
\begin{tikzpicture}[
  font=\small,
  >={Latex[length=2.4mm,width=1.7mm]},
  block/.style={rectangle, rounded corners=2pt, line width=.6pt, align=center,
                text width=3.6cm, inner sep=6pt, minimum height=1.25cm},
  sblock/.style={block, draw=scoreblue!75, fill=scoreblue!7},
  fblock/.style={block, draw=fmorange!75,  fill=fmorange!7},
  bridge/.style={block, draw=black!70, fill=white, line width=.8pt},
  shared/.style={block, draw=black!75, fill=black!5, line width=.9pt},
  panel/.style={rounded corners=5pt, line width=.7pt},
  head/.style={font=\footnotesize, inner sep=4pt, fill=white},
  arr/.style={->, semithick, black!60},
  lbl/.style={draw=black!25, rounded corners=2pt, fill=white, inner sep=3pt,
              font=\footnotesize, align=center}
]
 
\node[sblock] (cs) at (-3.0,2.55)
  {\textbfs{Conditional Score}\\[1pt]
   $\nabla_{\rvx}\log p_t(\rvx\,|\,\rvx_0)$\\[1pt]
   \subnote{closed form}};
 
\node[fblock] (cv) at (3.0,2.55)
  {\textbfs{Conditional Velocity}\\[1pt]
   $\rvv_t(\rvx\,|\,\rvz)$\\[1pt]
   \subnote{closed form}};
 
\node[sblock] (ms) at (-3.0,0)
  {\textbfs{Marginal Score}\\[1pt]
   $\nabla_{\rvx}\log p_t(\rvx)$\\[1pt]
   \subnote{intractable}};
 
\node[fblock] (mv) at (3.0,0)
  {\textbfs{Marginal Velocity}\\[1pt]
   $\rvv_t(\rvx)$\\[1pt]
   \subnote{intractable}};
 
\begin{pgfonlayer}{background}
  \node[panel, draw=scoreblue!40, fill=scoreblue!4,
        fit=(cs)(ms), inner xsep=8pt, inner ysep=12pt] (panelS) {};
  \node[panel, draw=fmorange!40, fill=fmorange!4,
        fit=(cv)(mv), inner xsep=8pt, inner ysep=12pt] (panelF) {};
\end{pgfonlayer}
 
\node[head, text=scoreblue] at (panelS.north) {\textbfs{Score SDE}};
\node[head, text=fmorange]  at (panelF.north) {\textbfs{Flow Matching}};
 
\draw[dashed, black!22, line width=.5pt]
  ($(panelS.north east)!0.5!(panelF.north west)$) --
  ($(panelS.south east)!0.5!(panelF.south west)$);
\node[fill=white, inner sep=2pt, text=black!45, font=\large]
  at ($(panelS.east)!0.5!(panelF.west)$) {$\Longleftrightarrow$};
 
\draw[arr] (cs) -- (ms);
\draw[arr] (cv) -- (mv);
\node[lbl] at ($(cs)!0.5!(ms)$) {$\E_{\rvx_0\,|\,\rvx}\big[\,\cdot\,\big]$};
\node[lbl] at ($(cv)!0.5!(mv)$) {$\E_{\rvz\,|\,\rvx}\big[\,\cdot\,\big]$};
 
\node[bridge, text width=5.8cm] (conv) at (0,-3.45)
  {$\displaystyle \rvv_t(\rvx)=f(t)\rvx-\tfrac{1}{2}g^{2}(t)\,\nabla_{\rvx}\log p_t(\rvx)$};
 
\coordinate (bus) at ($(panelS.south)+(0,-0.75)$);
\coordinate (busC) at (bus -| conv.north);
\draw[semithick, black!60, rounded corners=4pt] (panelS.south) |- (busC);
\draw[semithick, black!60, rounded corners=4pt] (panelF.south) |- (busC);
\draw[arr] (busC) -- (conv.north);
 
\node[shared, text width=7.8cm] (ode) at (0,-5.75)
  {\textbfs{The same probability flow}\\[4pt]
   $\displaystyle \frac{\diff \rvx(t)}{\diff t}=\rvv_t\big(\rvx(t)\big),$ \\[4pt] transporting between $p_{\mathrm{data}}$ and $p_{\mathrm{prior}}$.};
 
\draw[arr] (conv.south) -- (ode.north);
 
\end{tikzpicture}}
\caption{\textbfs{Score SDEs and Flow Matching.}
Both frameworks train on the closed-form \emph{conditional} quantity in the top
row. Averaging it over the conditioning variable (vertical arrows) gives the
intractable \emph{marginal} quantity below, which is what the trained model
recovers. The columns differ only in which quantity that is. On a diffusion
path the score determines the velocity through the conversion in the middle, so
both routes end at the same ODE; Flow Matching learns that velocity directly and
needs neither endpoint to be Gaussian.
\figcredit{Created by the authors.}}
\label{fig:score-fm-mental-picture}
\end{figure}

Flow Matching carries this principle beyond the usual Gaussian diffusion
setting. It can construct continuous transport between more general source and
target distributions from which samples are available, provided that the
chosen conditional probability paths can be sampled efficiently and their
conditional velocities can be computed tractably.

To support a solid understanding of this chapter, we provide in
\Cref{app:continuity} an intuitive, self-contained overview of the different
forms of the change-of-variables principle, progressing from the ordinary
change-of-variables formula to the continuity equation and, with stochastic
dynamics, to the Fokker--Planck equation.

\chapterlearning{
  \item connect Normalizing Flows, Neural ODEs, and Flow Matching through the common idea of transporting probability distributions, from discrete invertible transformations to continuous-time velocity fields;
  \item recognize the key analogy with the Score SDE framework in \Cref{ch:score-sde}: for the same diffusion probability path, Score SDEs learn a score that determines the PF-ODE velocity, whereas Flow Matching can learn that velocity field directly; in both cases, an intractable marginal target can be learned through tractable conditional targets;
  \item understand the continuity equation as the density-level rule that connects an ODE velocity field to the evolution of its probability distribution;
  \item understand how Flow Matching extends this conditional strategy beyond Gaussian diffusion paths to transport between general source and target distributions from which samples are available, while retaining simulation-free training through tractable conditional paths and velocities;
  \item distinguish the prescribed probability path $\{p_t\}_{t\in[0,1]}$ from the particular velocity field, coupling, and particle trajectories used to realize it, and understand why different flows can share the same marginal distributions;
  \item understand the two complementary ways of constructing tractable conditional flows: starting from a conditional probability path and deriving its velocity, or starting from a conditional flow map and deriving its velocity along trajectories.
}{
Readers already comfortable with Normalizing Flows and Neural ODEs may skim
\Cref{sec:flow-based-method} and begin with \Cref{subsec:lesson-score}.
For a first pass, focus on the connection between the Score SDE and Flow
Matching views, the continuity equation, and the conditional-to-marginal
velocity construction. The detailed path constructions and optional theoretical
properties can be revisited afterward.
}

\newpage

\section{Flow-Based Models: Normalizing Flows and Neural ODEs}\label{sec:flow-based-method}
In this section, we will introduce Flow-Based Models, including Normalizing Flows (NFs)~\citep{rezende2015variational} and Neural Ordinary Differential Equations (NODEs)~\citep{chen2018neural}.

NFs enable flexible and tractable probability density estimation by applying a series of invertible transformations to a simple base distribution. NODEs extend this framework to continuous time, where the transformation is governed by an ODE. By treating the transformations as continuous-time dynamics, NODEs provide a smooth, scalable extension to the NF paradigm.

\begin{figure}[th!]
    \centering
    \includegraphics[width=0.76\linewidth]{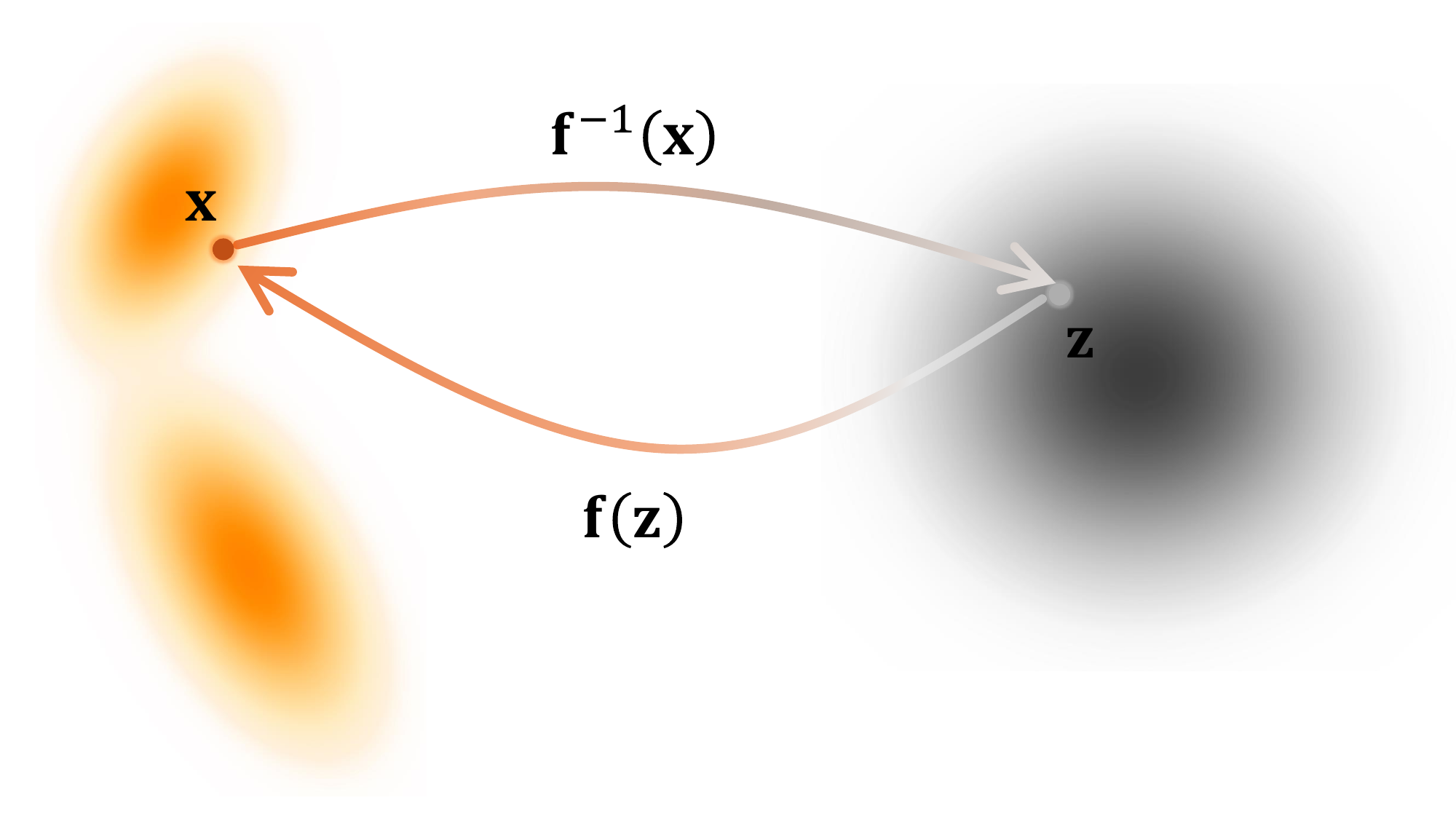}
    \caption{\textbfs{Illustration of sample movement of NF under an invertible map.} It consists of a sequence of invertible functions $\rvf: \rvz \mapsto \rvx$ that transform latent variable $\rvz$ into a data $\rvx$, together with the inverse mapping $\rvf^{-1}: \rvx \mapsto \rvz$ that reconstructs the data. An NF resembles an encoder--decoder structure, but with the encoder realized as a smooth invertible map and the decoder given exactly by its inverse.    The corresponding change in density can be computed via the change-of-variables formula, as given in \Cref{eq:nf-change-of-var}. \figcredit{Created by the authors.}}
    \label{fig:sample-and-change-of-variable}
\end{figure}

\subsection{Normalizing Flows}

NFs~\citep{rezende2015variational} model a complex data distribution $p_{\text{data}}(\rvx)$ by transforming a simple prior $p_{\text{prior}}(\rvz)$ (e.g., standard Gaussian $\mathcal{N}(\bm{0}, \rmI)$) via an invertible mapping
\[
\rvf_{\bm{\phi}}: \mathbb{R}^D \to \mathbb{R}^D,
\]
with $\rvx = \rvf_{\bm{\phi}}(\rvz)$ and $\rvz \sim p_{\text{prior}}$. Here, $\rvx$ and $\rvz$ share the same dimension.
Using the change-of-variables formula in \Cref{eq:nf-change-of-var}, the model likelihood is\footnote{If the map is further constrained to be the gradient of a convex potential,
$\rvf_\bphi = \nabla \psi_\bphi$ with $\psi_\bphi$ convex, then
\Cref{eq:nf-change-of-var-log} reduces to the Monge--Ampère
relation in \Cref{eq:monge-ampere}.
This PDE characterizes the optimal transformation of one distribution into another under the quadratic cost. See \Cref{ch:ot-eot} and \citep{huangconvex} for further details.
}
\begin{align}\label{eq:nf-change-of-var-log}
\log p_{\bm{\phi}}(\rvx) = \log p_{\text{prior}}(\rvz) + \log \left| \det \frac{\partial \rvf_{\bm{\phi}}^{-1}(\rvx)}{\partial \rvx} \right|.
\end{align}

\paragraph{Training Objective.}
Parameters $\bm{\phi}$ are learned by maximizing the likelihood over data:
\begin{align}\label{eq:nf-training}
\mathcal{L}_{\text{NF}}(\bm{\phi}) = \mathbb{E}_{\rvx \sim p_{\text{data}}} \left[ \log p_{\bm{\phi}}(\rvx) \right].
\end{align}
Computing the Jacobian determinant in \Cref{eq:nf-change-of-var-log} can be costly, scaling as $\mathcal{O}(D^3)$ in general.

\paragraph{Constructing Invertible Transformations.}

A single complex invertible network can be expensive due to its Jacobian determinant. Conversely, simple transforms (e.g., linear) are efficient but lack expressivity.

To balance this, NFs employ a sequence of $L$ trainable invertible mappings $\{\rvf_k\}_{k=0}^{L-1}$, each with efficiently computable Jacobians:
\[
\rvf_{\bm{\phi}} = \rvf_{L-1} \circ \rvf_{L-2} \circ \cdots \circ \rvf_0.
\]
Each $\rvf_k$ is parameterized by a neural network, though we omit the explicit dependence on ${\bm{\phi}}$ for notational simplicity.

Samples transform via
\begin{align}\label{eq:nf-seq}
\rvx_{k+1} = \rvf_k(\rvx_k), \quad k = 0, \ldots, L-1,
\end{align}
with $\rvz=\rvx_0 \sim p_{\text{prior}}$ and $\rvx = \rvx_L$, corresponding to data. The resulting (log-)density is derived as 
\begin{align}\label{eq:multi-nf-density}
\begin{aligned}
    p_{\bm{\phi}}(\rvx) &= p_{\text{prior}}(\rvx_0) \prod_{k=0}^{L-1} \left| \det \frac{\partial \rvf_k}{\partial \rvx_k} \right|^{-1}, \text{ or equivalently,}\\
\log p_{\bm{\phi}}(\rvx) &= \log p_{\text{prior}}(\rvx_0) + \sum_{k=0}^{L-1} \log \left| \det \frac{\partial \rvf_k}{\partial \rvx_k} \right|^{-1}.
\end{aligned}
\end{align}

\vspace{0.3cm}
\begin{figure}[tbh!]
    \centering
\begin{tikzpicture}[
    ->, 
    >=Stealth, 
    node_dist/.style={left=1.7cm of #1}, 
    font=\small,
    thick
  ]

  \tikzset{state/.style={
      rectangle,
      rounded corners=3pt,
      draw=black,
      fill=gray!10,
      minimum height=35pt,
      minimum width=25pt
    }
  }

  \node[state] (x0) {$\rvx_0 $};
  \node[state] (x1) [node_dist=x0] {$\rvx_1$};
  \node[state] (x2) [node_dist=x1] {$\rvx_2$};
  \node (dots) [left=0.7cm of x2] {$\boldsymbol{\cdots}$};
  \node[state] (xL) [node_dist=dots] {$\rvx_L$};

  
  \path [black] ([yshift=4.5pt]x0.west)  edge node[above] {$\rvf_0$} ([yshift=4.5pt]x1.east);
  \path [black] ([yshift=4.5pt]x1.west) edge node[above] {$\rvf_1$} ([yshift=4.5pt]x2.east);
  \path [black] ([yshift=4.5pt]x2.west) edge ([yshift=4.5pt]dots.east);
  \path [black] ([yshift=4.5pt]dots.west) edge node[above, pos=0.4] {$\rvf_{L-1}$} ([yshift=4.5pt]xL.east);
  
  \path ([yshift=-4.5pt]x1.east) edge node[below] {$\rvf_0^{-1}$} ([yshift=-4.5pt]x0.west);
  \path ([yshift=-4.5pt]x2.east) edge node[below] {$\rvf_1^{-1}$} ([yshift=-4.5pt]x1.west);
  \path ([yshift=-4.5pt]dots.east) edge ([yshift=-4.5pt]x2.west);
  \path ([yshift=-4.5pt]xL.east) edge node[below, pos=0.6] {$\rvf_{L-1}^{-1}$} ([yshift=-4.5pt]dots.west);

\end{tikzpicture}
\caption{\textbfs{Illustration of a NF.} An NF consists of a stack of invertible maps
$\rvf_{\bm{\phi}} = \rvf_{L-1} \circ \rvf_{L-2} \circ \cdots \circ \rvf_0$. 
The transformation maps latent samples $\rvx_0 \sim p_{\mathrm{prior}}$ to data samples $\rvx_L \sim p_{\mathrm{data}}$. \figcredit{Created by the authors.}}
    \label{fig:nf-stack}
\end{figure}

\paragraph{Examples of Invertible Flows.} Extensive literature has focused on designing single-layer flow constructions that enable efficient computation of the Jacobian. Below, we introduce two representative types: Planar Flows~\citep{rezende2015variational} and Residual Flows~\citep{chen2019residual,behrmann2019invertible}, with the latter motivating the developments in \Cref{subsec:NODE}.
\subparagraph{Planar Flows:}
Planar flows~\citep{rezende2015variational} apply a simple transformation
\[
\rvf(\rvz)
=
\rvz
+
\rvu h(\rvw^\top\rvz+b),
\]
where $\rvu,\rvw\in\mathbb{R}^D$, $b\in\mathbb{R}$, and $h(\cdot)$ is an
activation function. The parameters are appropriately constrained or
reparameterized to ensure that the transformation is invertible. Its Jacobian
determinant has the simple form
\[
\left|
\det\frac{\partial\rvf(\rvz)}{\partial\rvz}
\right|
=
\left|
1+h'(\rvw^\top\rvz+b)\,\rvw^\top\rvu
\right|,
\]
making the change-of-variables computation efficient.

\subparagraph{Residual Flows:}
Define the transform $\rvf$ as
\begin{align}\label{eq:residual-flow}
\rvf(\rvz) = \rvz + \rvv(\rvz),
\end{align}
with $\rvv$ contractive (Lipschitz constant $< 1$). This ensures invertibility via the Banach fixed-point theorem.

The log-determinant of the Jacobian reduces to a trace expansion:
\begin{align}\label{eq:log-det-tr}
    \log \Bigg|\det \frac{\partial \rvf(\rvz)}{\partial \rvz} \Bigg| &= \log \left( \det \frac{\partial \rvf(\rvz)}{\partial \rvz} \right) \nonumber\\
    &= \Tr\left(\log\left(\frac{\partial \rvf(\rvz)}{\partial \rvz}\right)\right) \nonumber\\
    &= \Tr\left(\log\left(\rmI + \frac{\partial \rvv(\rvz)}{\partial \rvz}\right)\right) \nonumber\\
    &= \sum_{k=1}^{\infty} \frac{(-1)^{k+1}}{k} \Tr\left(\left(\frac{\partial \rvv(\rvz)}{\partial \rvz}\right)^k\right),
\end{align}
making evaluation efficient via trace estimators~\citep{hutchinson1989stochastic}.

\paragraph{Sampling and Inference.}
Sampling from NFs is straightforward: draw $\rvx_0 \sim p_{\text{prior}}$ and compute $\rvx = \rvf_{\bm{\phi}}(\rvx_0)$. Exact likelihoods are obtained from \Cref{eq:multi-nf-density}.

\subsection{Neural ODEs}\label{subsec:NODE}
\begin{figure}[th!]
    \centering
    \includegraphics[width=\linewidth]{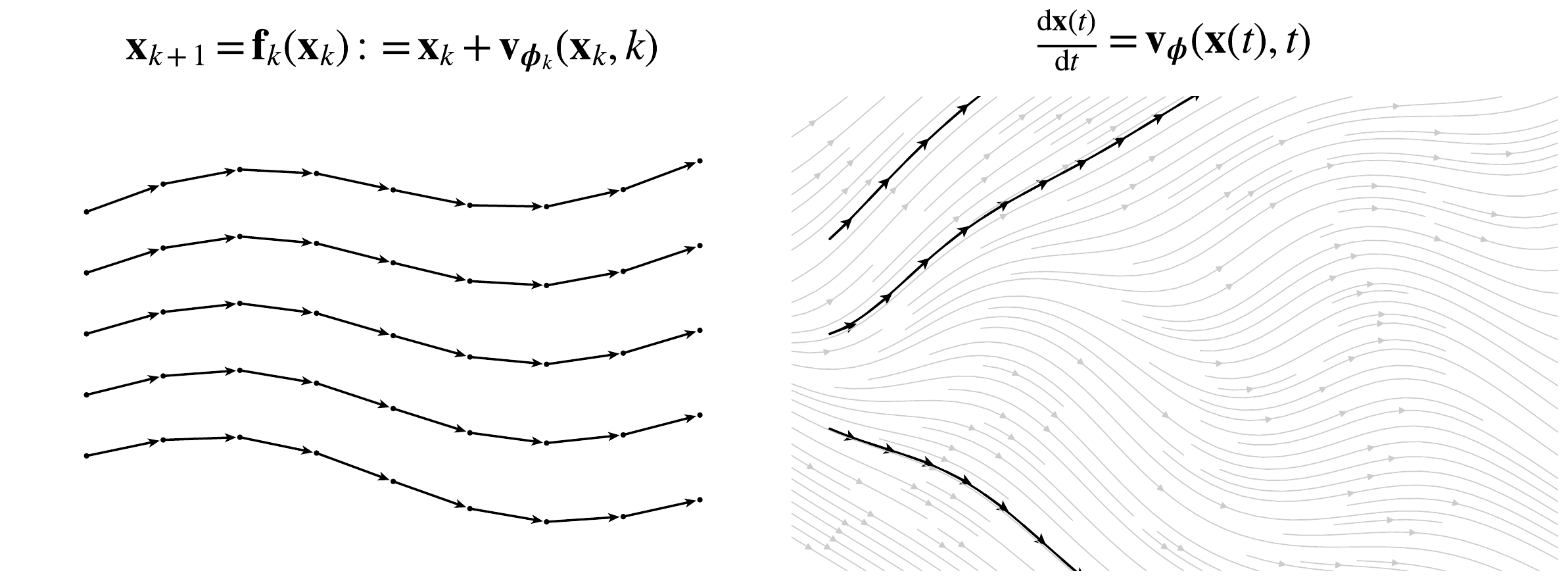}
    \caption{\textbfs{Discrete- vs. continuous-time normalizing flows.} (Left) A discrete NF transports samples by a finite sequence of invertible maps  $\mathbf{x}_{k+1} = \mathbf{f}_k(\mathbf{x}_k)$,
yielding stepwise, non-crossing trajectories (dots with arrows). (Right) A continuous NF (Neural ODE) evolves states along integral curves of $\frac{\diff\mathbf{x}(t)}{\diff t} = \mathbf{v}_{\boldsymbol{\phi}}(\mathbf{x}(t),t)$,
where black paths with tangent arrows are shown over the gray vector field. \figcredit{Created by the authors.}}
    \label{fig:nf-cnf}
\end{figure}



\paragraph{From Discrete-Time NFs to Continuous-Time NFs (Neural ODEs).}
A discrete NF transports samples through a composition of $L$ invertible
maps. To obtain a meaningful continuous-depth limit, we write each layer
as a small residual update:
\[
\rvx_{k+1}
=
\rvx_k+\Delta t\,\rvv_{\bm{\phi}}(\rvx_k,t_k),
\qquad
t_k=k\Delta t,
\qquad
\Delta t=\frac{T}{L}.
\]
Here, $\rvv_{\bm{\phi}}$ specifies the direction in which each point is
moved, while $\Delta t$ controls the size of one layer.

\begin{figure}[th!]
    \centering
    \includegraphics[width=\linewidth]{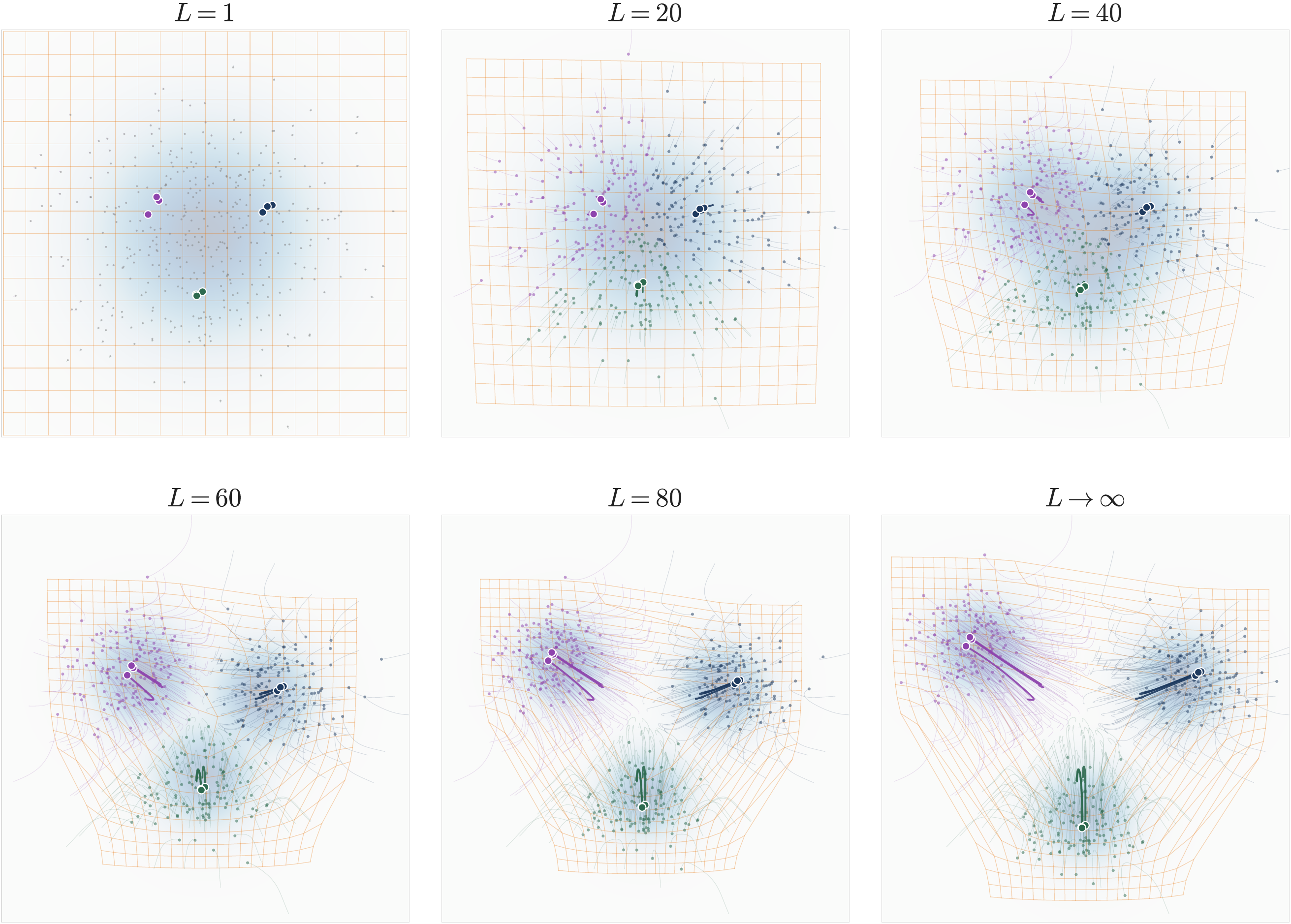}
    \caption{\textbfs{From discrete bijections to a continuous transport flow.}
    Each panel shows a transport built by composing $L$ invertible layers.
    Starting from a source distribution (Gaussian), the layers gradually move
    the colored samples toward a target distribution with three modes. The
    orange reference grid shows how the surrounding space is deformed during
    this transport. With only a few layers, each layer makes a relatively
    large and discrete change. As $L$ increases, the transport is divided
    into more, smaller updates, so the sample trajectories and grid
    deformation become smoother. If the total transport time $T$ is fixed
    and each layer has step size $\Delta t=T/L$, then, as
    $L\rightarrow\infty$, the stacked bijections approach a continuous
    time-dependent flow. The transported density evolves according to the
    continuity equation in \Cref{eq:continuity_eq}.
    \figcredit{Created by the authors with AI-assisted coding.}}
    \label{fig:bijection-continuity}
\end{figure}

The geometric meaning of this limit is illustrated in
\Cref{fig:bijection-continuity}. As $L$ increases, the same
source-to-target transport is divided into more near-identity bijections.
The sample trajectories are resolved more finely, and the deformation of
the surrounding reference grid becomes progressively smoother.

As $L\rightarrow\infty$ and $\Delta t\rightarrow 0$, the discrete layer
index becomes continuous time, yielding the ODE
\[
\frac{\diff \rvx(t)}{\diff t}
=
\rvv_{\bm{\phi}}(\rvx(t),t).
\]
This continuous-depth model is known as a Neural ODE or Continuous
Normalizing Flow.

\paragraph{Formal Setup of Neural ODEs.}
A Neural ODE defines a continuous transformation through:
\begin{align}\label{eq:neural-ode}
    \frac{\diff \rvx(t)}{\diff t} = \rvv_{\bm{\phi}}(\rvx(t), t),\quad t\in [0,T]
\end{align}
where:
\begin{itemize}
    \item $ \rvx(t) \in \mathbb{R}^D $ is the state at time $t$; we sometimes write $\rvx_t$ for brevity;
    \item $\rvv_{\bm{\phi}}(\rvx(t), t)$ is a neural network parameterized by $\bm{\phi}$.
\end{itemize}

\subparagraph{Goal of NODE.}
Starting from the initial condition $\rvx(0) \sim p_{\text{prior}}$, the ODE evolves the state continuously over time, inducing a family of marginal distributions $p_{\bm{\phi}}(\rvx_t, t)$ (similar to PF-ODEs!)\footnote{We adopt a flipped time convention, with $t = 0$ denoting the prior (source) and $t = 1$ the data (target) distribution. The prior is interchangeably written as $p_{\bm{\phi}}(\rvx(0), 0)$, $p_{\text{prior}}(\rvx(0), 0)$, or simply $p_{\text{prior}}(\rvz)$.
}. 

The goal is to learn the neural vector field $\rvv_{\bm{\phi}}$, which intuitively represents a velocity that transports points along continuous trajectories in data space. By learning this velocity, the terminal distribution at $t=T$ matches the target distribution $p_{\mathrm{data}}(\cdot)$. This continuous transformation unifies discrete normalizing flows and neural ODEs within a single framework.

\paragraph{Continuous-Time Change-of-Variables Formula.}
Analogous to \Cref{eq:nf-change-of-var} or \Cref{eq:multi-nf-density},
\citet{chen2018neural} derived a continuous-time analogue of the
change-of-variables formula. For the time-dependent density
$p_{\bm{\phi}}(\rvx(t),t)$ of a process $\rvx(t)$ evolving under
\Cref{eq:neural-ode}, the so-called \emph{Instantaneous
Change-of-Variables Formula} is
\[
\frac{\diff}{\diff t}
\log p_{\bm{\phi}}(\rvx(t),t)
=
-\nabla_{\rvx}\cdot
\rvv_{\bm{\phi}}(\rvx(t),t).
\]
Here, the derivative is taken along the ODE trajectory $\rvx(t)$.

Therefore, if $\rvx(0)\sim p_{\mathrm{prior}}$ and the ODE transports the
prior from $t=0$ to $t=T$, integrating this identity gives
\begin{align}\label{eq:NODE-likelihood}
\log p_{\bm{\phi}}(\rvx(T),T)
=
\log p_{\mathrm{prior}}(\rvx(0))
-
\int_0^T
\nabla_{\rvx}\cdot
\rvv_{\bm{\phi}}(\rvx(t),t)
\,\diff t.
\end{align}
This expression provides likelihood evaluation through numerical integration,
which in turn enables maximum likelihood training of the model.

Although this formula may appear unfamiliar at first, it is closely connected
to the Fokker--Planck equation. In particular, it can be understood as the
along-trajectory form of the \emph{continuity equation}, which is the
deterministic, zero-diffusion case of the Fokker--Planck equation
(see \Cref{app:continuity}). It can also be interpreted as the continuous-time
limit of the discrete change-of-variables formula in
\Cref{eq:multi-nf-density}. We summarize the result below.

\lemp{Instantaneous Change of Variables}{instant-change-of-var}{
Let $\rvz(t)$ be a continuous random process with time-dependent density
$p(\rvz(t),t)$, and suppose it evolves according to the ODE
\begin{align*}
\frac{\diff\rvz(t)}{\diff t}
=
\mathbf{F}(\rvz(t),t).
\end{align*}
Assuming sufficient regularity so that the ODE flow and the evolving density
are well defined, the log-density along the trajectory satisfies
\begin{align}\label{eq:instant-change-of-var}
\frac{\diff}{\diff t}
\log p(\rvz(t),t)
=
-\nabla_{\rvz}\cdot
\mathbf{F}(\rvz(t),t).
\end{align}
}{
This follows from the continuity equation together with the chain rule. We present detailed alternative derivations in \Cref{app-sec:flow-proof}.}

\subparagraph{Connection to Discrete-Time Formula.}  
The NODE likelihood in \Cref{eq:NODE-likelihood},
\[
\log p_{\bm{\phi}}(\rvx(T), T) = \log p_{\text{prior}}(\rvx(0), 0) - \int_0^T \nabla_\rvx \cdot \rvv_{\bm{\phi}}(\rvx(t), t) \diff t,
\]
can be seen as the continuous-time analogue of the discrete normalizing flow formulation in \Cref{eq:multi-nf-density}:
\[
\log p_{\bm{\phi}}(\rvx_L) = \log p_{\text{prior}}(\rvx_0) - \sum_{k=0}^{L-1} \log \left| \det \frac{\partial \rvf_k}{\partial \rvx_k} \right|.
\]
The integral mirrors the summation, and the trace operator replaces the log-determinant, as discussed in \Cref{eq:log-det-tr}. These parallels are further explored in the proof of the lemma.

\paragraph{Training NODEs.}  
Based on \Cref{eq:NODE-likelihood}, NODEs learn a parameterized velocity field $\rvv_{\bm{\phi}}$ such that the terminal distribution
\[
p_{\bm{\phi}}(\cdot, T) \approx p_{\text{data}},
\]
where trajectories evolve from latent variables $\rvx(0) \sim p_{\text{prior}}$ via the ODE flow. Training follows the MLE framework from \Cref{eq:MLE}:
\[
\mathcal{L}_{\text{NODE}}(\bm{\phi}) := \mathbb{E}_{\rvx \sim p_{\text{data}}} \big[\log p_{\bm{\phi}}(\rvx, T)\big].
\]

\subparagraph{Exact Log-Likelihood Computation.} 
To compute $\log p_{\bm{\phi}}(\mathbf{x}, T)$ for data point $\mathbf{x}$, we integrate the change-of-variables formula \eqref{eq:NODE-likelihood}:
\begin{align}
\log p_{\bm{\phi}}(\mathbf{x}, T) = \log p_{\text{prior}}(\mathbf{z}(0)) - \int_0^T \nabla_{\mathbf{z}} \cdot \mathbf{v}_{\bm{\phi}}(\mathbf{z}(t), t) \diff t.
\end{align}
Here, $\mathbf{z}(t)$ solves the ODE reversely from $t = T$ to $t = 0$:
\[
\frac{\diff\mathbf{z}(t)}{\diff t} = \mathbf{v}_{\bm{\phi}}(\mathbf{z}(t), t)
\]
with $\mathbf{z}(T) = \mathbf{x}$. The prior term $\log p_{\text{prior}}(\mathbf{z}(0))$ is tractable for standard distributions. This enables exact likelihood-based training and evaluation in neural ODEs.

\subparagraph{Gradient-Based Optimization.}  
Maximizing $\mathcal{L}_{\text{NODE}}$ requires backpropagation through the ODE solver. The adjoint sensitivity method~\citep{pontryagin2018mathematical, chen2018neural} computes gradients via an auxiliary ODE with $\mathcal{O}(1)$ memory complexity, but NODE training remains expensive due to numerical integration at each step.

\paragraph{Inference with NODEs.}  
Sampling with a trained model $\rvv_{\bm{\phi}^\times}$ proceeds by drawing $\rvx(0) \sim p_{\text{prior}}$ and integrating forward (by numerical solvers):
\[
\rvx(T) = \rvx(0) + \int_0^T \rvv_{\bm{\phi}^\times}(\rvx(t), t)   \diff t.
\]
The terminal state $\rvx(T)$ approximates a sample from $p_{\text{data}}$.

Moreover, we note that for any vector field $\rmF$, the following identity holds:
\[
    \Tr\left(\frac{\partial \rmF}{\partial \rvz(t)} \right) = \nabla_{\rvz} \cdot \rmF.
\]
Hence, the divergence can be estimated efficiently using stochastic trace
estimators, such as Hutchinson's estimator~\citep{hutchinson1989stochastic},
without explicitly constructing the full Jacobian. This makes likelihood
evaluation much more practical in high-dimensional settings. However, because
the trace is then estimated from random samples, the resulting numerical
log-likelihood estimate is stochastic rather than an exact evaluation of the
divergence.

Having seen in \Cref{fig:bijection-continuity} how many small bijections
approach a continuous transport flow, we now ask a practical question:
\begin{question}
    How can we learn its velocity field without repeatedly solving an ODE
during training?
\end{question}
 This question leads naturally to Flow Matching, which
learns the velocity field directly through a simulation-free regression
objective and provides the flow-based perspective on diffusion models.

\newpage
\clearpage
\section{\texorpdfstring{Flow Matching Framework}{Flow Matching Framework}}\label{sec:flow-matching-framework}

Both Score SDEs~(\Cref{ch:score-sde}) and
NODEs~(\Cref{sec:flow-based-method}) describe generative modeling as a
continuous transport process. Score SDEs formulate this transport through
stochastic dynamics together with an associated PF-ODE,
whereas NODEs directly parameterize a deterministic ODE. In the standard
generative setting, both aim to transport a simple Gaussian prior
$p_{\mathrm{prior}}$ into the data distribution $p_{\mathrm{data}}$.

A practical limitation of NODEs is that likelihood-based training requires
repeatedly solving an ODE. The \emph{Flow Matching} (FM)
framework~\citep{lipman2022flow,lipman2024flow,tongimproving} avoids this
cost by training a neural network to predict the velocity field directly
through a regression objective. ODE integration is therefore needed for
generation, but not inside the training loop, making FM training
simulation-free.

More generally, FM is not restricted to Gaussian-to-data transport. It
considers a source distribution $p_{\mathrm{src}}$ and a target
distribution $p_{\mathrm{tgt}}$, both assumed to be easy to sample from.
Standard generative modeling is recovered as the special case
\[
p_{\mathrm{src}}=p_{\mathrm{prior}},
\qquad
p_{\mathrm{tgt}}=p_{\mathrm{data}},
\]
where $p_{\mathrm{prior}}$ is typically Gaussian.

In this section, we adopt the FM viewpoint\footnote{Several related
approaches share the core idea of transporting between endpoint
distributions using a continuous-time flow, although their formulations
differ slightly. These include Flow Matching
(FM)~\citep{lipman2022flow,neklyudov2023action}, Rectified Flow
(RF)~\citep{liu2022rectified,heitz2023iterative}, and Stochastic
Interpolants~\citep{albergo2023stochastic,albergobuilding,ma2024sit}.
Here, we use FM as a unifying terminology.}
Its goal is to learn a time-dependent velocity field
$\mathbf{v}_t(\rvx_t)$ whose associated ODE transports samples along a
prescribed probability path $\{p_t\}_{t\in[0,1]}$ satisfying
\[
p_0=p_{\mathrm{src}},
\qquad
p_1=p_{\mathrm{tgt}}.
\]
When $p_{\mathrm{src}}$ is Gaussian, we refer to this setting as
\emph{Gaussian Flow Matching}. Thus, FM provides both a simulation-free
training method for generative modeling and a general framework for
learning continuous transport between arbitrary endpoint distributions.

\subsection{Lesson from Score-Based Methods}\label{subsec:lesson-score}

We revisit the Score SDE framework (\Cref{ch:score-sde}) using a slightly different but equivalent formulation to extract key insights that motivate the FM approach. This analysis reveals how diffusion models implicitly learn probability flows and motivates a more direct formulation.

\paragraph{Step 1: Defining a Conditional Path and Its Marginal Densities.}

\begin{figure}[th!]
    \centering
    \includegraphics[width=\linewidth]{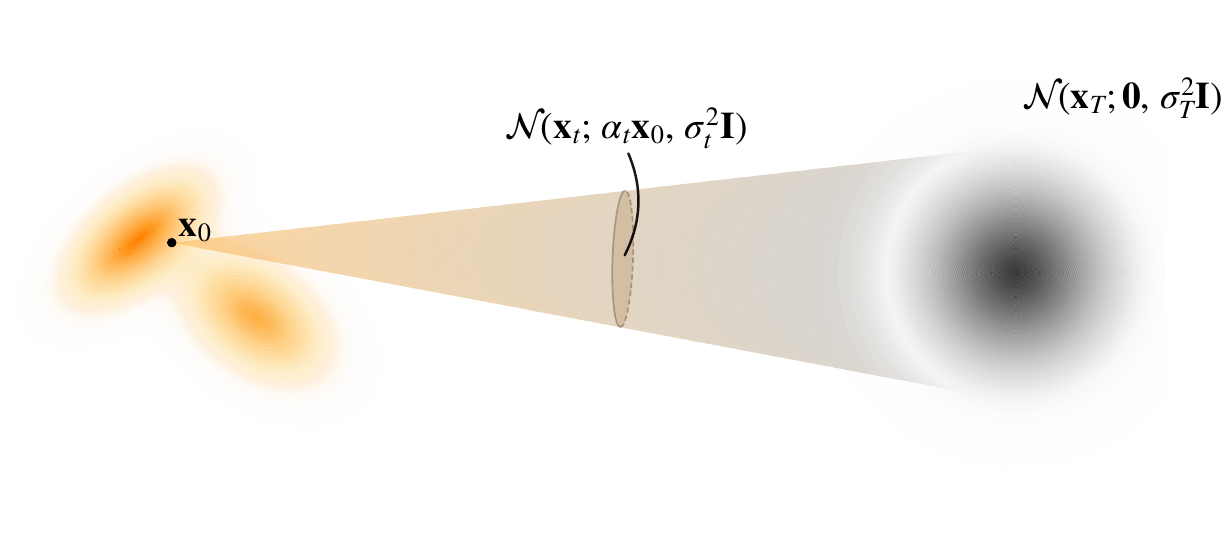}
    \caption{\textbfs{Illustration of the conditional transition distribution.} 
$p_t(\rvx_t | \rvx_0) = \mathcal{N}(\rvx_t;\alpha_t \rvx_0,\sigma_t^2 \mathbf{I})$, 
defines a (Gaussian) conditional probability path from a data sample 
$\rvx_0 \sim p_{\text{data}}$ (left) towards the Gaussian prior 
$p_{\text{prior}}$ (right). \figcredit{Created by the authors.}}
    \label{fig:data-prior-conditioning}
\end{figure}

A diffusion model specifies a continuous-time family of densities $\{p_t\}_{t \in [0,1]}$ that transports a simple prior $p_{\text{prior}}$ (e.g., Gaussian) at $t=1$, used as the source, to a target data distribution $p_{\text{data}}$ at $t=0$:
\[
p_1(\rvx_1) = p_{\text{prior}}(\rvx_1), \quad p_0(\rvx_0) = p_{\text{data}}(\rvx_0).
\]
This path is implicitly defined via the forward conditional distribution
\begin{align}\label{eq:lesson-dm-conditional-path}
    p_t(\rvx_t|\rvx_0) = \mathcal{N}(\rvx_t; \alpha_t \rvx_0, \sigma_t^2 \mathbf{I}), \quad \rvx_0\sim p_{\text{data}}
\end{align}
which induces the marginal density
\[
p_t(\rvx_t) := \int p_t(\rvx_t |\rvx_0)  p_{\text{data}}(\rvx_0) \mathrm{d}\rvx_0.
\]
The increasing variance $\sigma_t^2$ of the conditional Gaussian drives the evolution of $p_t$ toward the Gaussian prior.

\paragraph{Step 2: Velocity Field.}
The time evolution of the marginal density $p_t$ is governed by a velocity field $\rvv_t : \mathbb{R}^D \to \mathbb{R}^D$, derived from the Fokker-Planck equation:
\begin{align}\label{eq:dm-velocity}
    \rvv_t(\rvx) := f(t)\rvx - \frac{1}{2}g^2(t)\nabla_{\rvx} \log p_t(\rvx),
\end{align}
which defines a deterministic particle flow through the PF-ODE:
\[
\frac{\diff \rvx(t)}{\diff t} = \underbrace{f(t)\rvx(t) - \frac{1}{2}g^2(t)\nabla_{\rvx} \log p_t\big(\rvx(t)\big)}_{\rvv_t(\rvx(t))}.
\]
This ODE transports an initial random variable $\rvx(0) \sim p_{\text{data}}$ forward in time or $\rvx(1) \sim p_{\text{prior}}$ backward in time, such that the evolving marginal density of $\rvx(t)$ matches $p_t$ at every $t \in [0,1]$  (see ``Underlying Rule'' below).

The scalar functions $f(t)$ and $g(t)$ are determined by the coefficients of the associated forward SDE, or equivalently the Gaussian kernel parameters $\alpha_t$ and $\sigma_t$ defined in the conditional path (see Lemma~\ref{forward-sde}).

\paragraph{Step 3: Learning via the Conditional Strategy.}
The PF-ODE velocity in \Cref{eq:dm-velocity} depends on the marginal score
$\nabla_{\rvx_t}\log p_t(\rvx_t)$, which is generally unknown.
Score-based methods therefore learn this score using a neural network
$\rvs_{\bm{\phi}}(\rvx_t,t)$ through the ideal score-matching objective
\[
\mathcal{L}_{\text{SM}}(\bm{\phi})
=
\mathbb{E}_{t\sim\mathcal{U}[0,1],\,\rvx_t\sim p_t}
\left[
\left\|
\rvs_{\bm{\phi}}(\rvx_t,t)
-
\nabla_{\rvx_t}\log p_t(\rvx_t)
\right\|^2
\right].
\]
Once the score is learned, the PF-ODE velocity is obtained from it as
\[
\rvv_t(\rvx_t)
=
f(t)\rvx_t
-
\frac{1}{2}g^2(t)
\nabla_{\rvx_t}\log p_t(\rvx_t),
\]
with $\rvs_{\bm{\phi}^\times}(\rvx,t) \approx \nabla_{\rvx_t}\log p_t(\rvx)$

The difficulty is that the marginal score itself is inaccessible when computing the regression target.
As in \Cref{ch:score-based}, we overcome this using the conditioning trick.
The conditional distribution $p_t(\rvx_t|\rvx_0)$ is tractable, and its
conditional score satisfies
\begin{align}\label{eq:marginal-oracle-score}
\nabla_{\rvx_t}\log p_t(\rvx_t)
=
\mathbb{E}_{\rvx_0\sim p(\cdot|\rvx_t)}
\left[
\nabla_{\rvx_t}\log p_t(\rvx_t|\rvx_0)
\right].
\end{align}
This motivates the denoising score matching objective
\[
\mathcal{L}_{\text{DSM}}(\bm{\phi})
:=
\mathbb{E}_{t,\,\rvx_0\sim p_{\mathrm{data}},\,
\rvx_t\sim p_t(\cdot|\rvx_0)}
\left[
\left\|
\rvs_{\bm{\phi}}(\rvx_t,t)
-
\nabla_{\rvx_t}\log p_t(\rvx_t|\rvx_0)
\right\|^2
\right].
\]
As shown in Theorem~\ref{thm:sm-dsm}, $\mathcal{L}_{\text{SM}}$ and
$\mathcal{L}_{\text{DSM}}$ differ only by a constant independent of
$\bm{\phi}$ and therefore have the same population minimizer:
\[
\rvs^*(\rvx_t,t)
=
\mathbb{E}_{\rvx_0\sim p(\cdot|\rvx_t)}
\left[
\nabla_{\rvx_t}\log p_t(\rvx_t|\rvx_0)
\right]
=
\nabla_{\rvx_t}\log p_t(\rvx_t).
\]

Importantly, the same conditioning principle can also be expressed directly
at the level of the PF-ODE velocity. Define the conditional velocity
\[
\rvv_t(\rvx_t|\rvx_0)
:=
f(t)\rvx_t
-
\frac{1}{2}g^2(t)
\nabla_{\rvx_t}\log p_t(\rvx_t|\rvx_0).
\]
Using \Cref{eq:marginal-oracle-score}, we immediately obtain
\[
\rvv_t(\rvx_t)
=
\mathbb{E}_{\rvx_0\sim p(\cdot|\rvx_t)}
\left[
\rvv_t(\rvx_t|\rvx_0)
\right].
\]
Thus, score-based diffusion learns the score through tractable conditional
targets and then uses it to construct the PF-ODE velocity. Flow Matching will
apply the same conditional-to-marginal principle directly to the velocity
field.
\begin{figure}
    \centering
    \includegraphics[width=\linewidth]{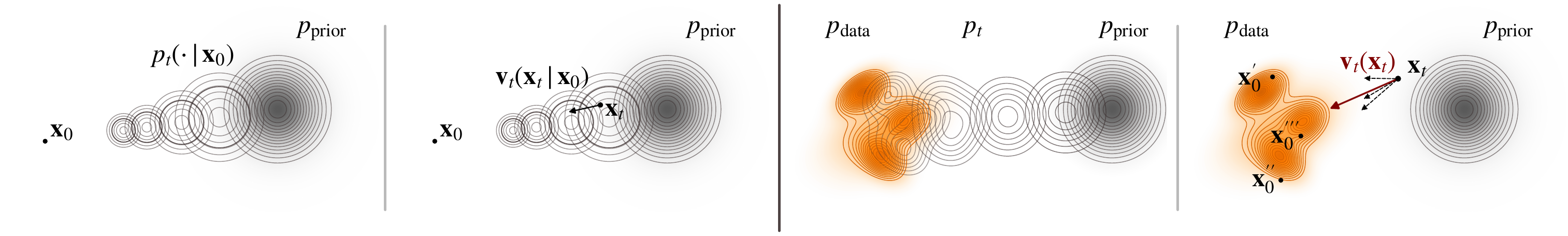}
    \caption{\textbfs{Illustration of conditional versus marginal perspectives in diffusion.} (This figure is motivated by \citet{lipman2024flow}.) 
(1) Conditional Gaussian path $p_t(\cdot|\mathbf{x}_0)$, showing expanding densities from a fixed $\mathbf{x}_0$ toward the prior. 
(2) Conditional velocity $\mathbf{v}_t(\mathbf{x}_t|\mathbf{x}_0)=f(t)\rvx_t - \tfrac{1}{2}g^2(t) \nabla_{\rvx_t} \log p_t(\rvx_t|\rvx_0)$. 
(3) Marginal density $p_t$, transporting the data distribution $p_{\mathrm{data}}$ (orange) into the prior $p_{\mathrm{prior}}$ (gray). 
(4) Marginal velocity $\mathbf{v}_t(\mathbf{x}_t)=f(t)\rvx_t - \tfrac{1}{2}g^2(t) \nabla_{\rvx_t} \log p_t(\rvx_t)$, obtained by averaging conditional directions from $\mathbf{x}_t$ to multiple plausible origins (dashed), yielding the red arrow. In the FM framework with one-sided conditioning $\rvz=\rvx_0$, 
the same illustration applies to $\rvv_t(\rvx_t|\rvx_0)$ and $\rvv_t(\rvx_t)$, 
without requiring them to be written explicitly in terms of the scores 
$\nabla_{\rvx_t}\log p_t(\rvx_t|\rvx_0)$ or $\nabla_{\rvx_t}\log p_t(\rvx_t)$.
\figcredit{Created by the authors.}}
    \label{fig:conditional-marginal}
\end{figure}

\paragraph{Underlying Rule: The Fokker–Planck Equation.}
The marginal density $p_t$ evolves according to the Fokker–Planck equation:
\begin{mdframed}
\begin{align*}
    \frac{\partial p_t(\rvx)}{\partial t} + \nabla \cdot \Bigg( \underbrace{\left(f(t)\rvx - \frac{1}{2}g^2(t) \nabla_\rvx \log p_t(\rvx)\right)}_{\rvv_t(\rvx)}  p_t(\rvx) \Bigg) = 0.
\end{align*}
\end{mdframed}
This PDE ensures that the density given by the PF-ODE matches the marginal distribution of the forward SDE.  
To see this, recall the \emph{flow map} $\bPsi_{s \to t}(\rvx_s)$ of the PF-ODE as defined in \Cref{eq:flow-map-def}, which carries an initial state $\rvx_s$ at time $s$ directly to its state at $t$.  
Running the PF-ODE backward from $t=1$ to $t=0$, starting with $\rvx_1 \sim p_{\mathrm{prior}}$, we obtain time-dependent densities through the pushforward formula:
\begin{align}\label{eq:pushforward-ode}
    p_t^{\mathrm{rev}}(\rvx)
    = \int \delta \left(\rvx - \bPsi_{1 \to t}(\rvx_1)\right) 
      p_{\mathrm{prior}}(\rvx_1) \mathrm{d}\rvx_1.
\end{align}

The Fokker–Planck equation ensures that the induced density path coincides with the same evolving density:
\begin{align}\label{eq:p_t-fwd-rev}
p_t^{\mathrm{rev}} = p_t.
\end{align}
In particular, this implies $p_0^{\mathrm{rev}} = p_0 = p_{\text{data}}$, thereby recovering the data distribution at time $t=0$. Since the ODE solution map is bidirectional, we can similarly consider initializing at 
$\rvx_0 \sim p_{\text{data}}$ and solving the ODE forward in time, enabling a parallel analysis.

\subsection{Flow Matching Framework}\label{subsec:fm-framework}

The analysis in \Cref{subsec:lesson-score} reveals that diffusion models succeed by learning a velocity field, specifically, the score, that transports between distributions while satisfying boundary conditions. The design of the Gaussian conditional path in \Cref{eq:lesson-dm-conditional-path}, with increasing variance $\sigma_t^2$, implicitly anchors one endpoint to a Gaussian prior while allowing the conditional density to be defined over the entire space, enabling score-based gradient computation.

In this subsection, we introduce the FM framework, which builds on this insight 
(the same illustration in \Cref{fig:conditional-marginal} also applies to the FM framework) 
and extends it to learning continuous flows that transport samples between two arbitrary 
distributions, $p_{\text{src}}$ and $p_{\text{tgt}}$.

\paragraph{Step 1: Defining a Conditional Path and Its Marginal Densities.}
Consider arbitrary source and target probability distributions $ p_{\mathrm{src}} $ and $ p_{\mathrm{tgt}} $ on $\mathbb{R}^D$. We set\footnote{To align with the standard notation in FM literature, we reverse the time axis compared to earlier sections: $t=0$ corresponds to the source distribution and $t=1$ to the target.}
\begin{align}\label{eq:marginal-endpoint}
    p_0(\mathbf{x}) = p_{\mathrm{src}}(\mathbf{x}), \quad p_1(\mathbf{x}) = p_{\mathrm{tgt}}(\mathbf{x}).
\end{align}

FM implicitly defines a continuous family of intermediate densities $\{ p_t \}_{t \in [0,1]}$ interpolating between these endpoints. Each marginal $ p_t $ is expressed via a latent variable $\mathbf{z}$ drawn from a known distribution $\pi(\mathbf{z})$ and a conditional distribution $p_t(\mathbf{x}_t | \mathbf{z})$:
\begin{equation}\label{eq:p_t}
    p_t(\mathbf{x}_t) = \int p_t(\mathbf{x}_t | \mathbf{z}) \pi(\mathbf{z}) \diff \mathbf{z},
\end{equation}
with $(\pi(\mathbf{z}), \{p_t(\cdot|\mathbf{z})\})$ chosen to satisfy the boundary conditions in \Cref{eq:marginal-endpoint}.

We remark that, in general, the marginal densities $p_t$ are not tractable, since they require integrating over $\pi(\mathbf{z})$, and both $\pi(\mathbf{z})$ and the conditional distributions $p_t(\mathbf{x}_t|\mathbf{z})$ can be complex.
Nonetheless, conditioning on the latent $\mathbf{z}$ grants FM the flexibility to model a broad class of interpolation paths beyond those discussed in \Cref{subsec:lesson-score}. Common choices for $\mathbf{z}$ include:
\begin{itemize}
    \item \textbfs{Two-sided conditioning:} $\mathbf{z} = (\mathbf{x}_0, \mathbf{x}_1) \sim p_{\text{src}}(\mathbf{x}_0) p_{\text{tgt}}(\mathbf{x}_1)$, where $\pi$ couples source and target distributions. This allows FM to define transport between arbitrary distributions.
    \item \textbfs{One-sided conditioning:} $\mathbf{z} = \mathbf{x}_0$ or $\mathbf{z} = \mathbf{x}_1$. It especially recovers diffusion-like setups when the source distribution is chosen to be Gaussian.  
\end{itemize}

\begin{figure}[th!]
    \centering
    \includegraphics[width=\linewidth]{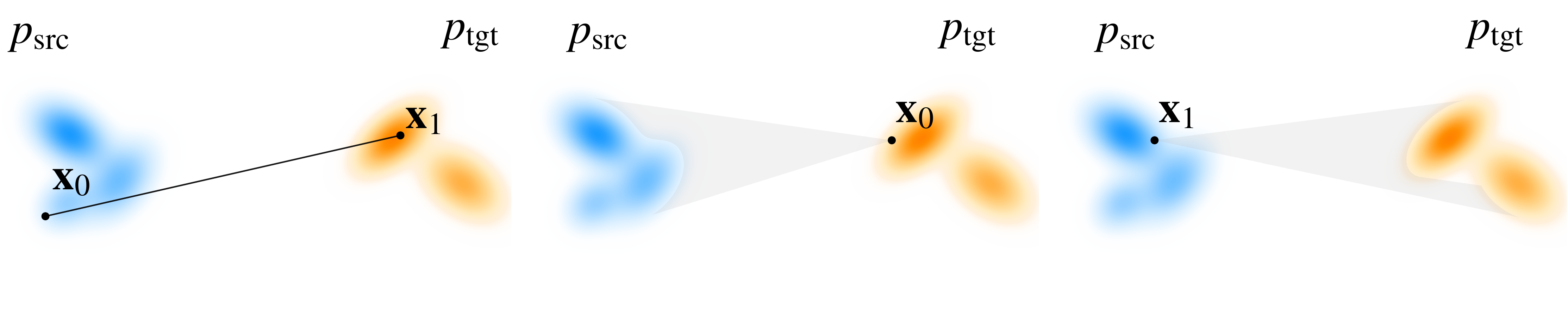}
    \caption{\textbfs{Illustrations of two common types of conditioning probability paths}. It includes:
(i) \emph{two-sided}, conditioned on $\rvx_0 \sim p_{\text{tgt}}$ and $\rvx_1 \sim p_{\text{src}}$ with general endpoint distributions; 
(ii) \emph{one-sided}, conditioned at either $\rvx_0 \sim p_{\text{tgt}}$ or $\rvx_1 \sim p_{\text{src}}$.
\figcredit{Created by the authors.}}
    \label{fig:conditional-path}
\end{figure}

In all cases, the conditional path $p_t(\mathbf{x}_t|\mathbf{z})$ should be
efficiently sampleable, and the corresponding conditional velocity should be
tractable. In the probability-path-first constructions below, we further use
conditional distributions with tractable closed-form expressions. We present
specific constructions in \Cref{subsec:fm-instantiations}, with illustrations
in \Cref{fig:conditional-path}.

\paragraph{Step 2: Velocity Field.}
A probability path $\{p_t\}_{t\in[0,1]}$ specifies how the distribution
should evolve over time, but it does not generally specify how individual
samples should move. In other words, the same sequence of marginal densities
can be realized by different velocity fields\footnote{
The key distinction is that a probability path specifies the evolution of
the \emph{distribution}, but generally not the particle-level transport that
realizes it. A particular velocity becomes determined only after additional
transport structure is specified. For example, a sufficiently regular and
invertible flow map $\bPsi_{0\to t}$ determines its Eulerian velocity through
$\mathbf{v}_t=\partial_t\bPsi_{0\to t}\circ\bPsi_{0\to t}^{-1}$.
Likewise, once a forward diffusion SDE is specified, its associated PF-ODE
selects the particular velocity
$\mathbf{v}_t(\mathbf{x})
=
f(t)\mathbf{x}
-\tfrac12g^2(t)\nabla_{\mathbf{x}}\log p_t(\mathbf{x})$.
Thus, both cases add particle-level transport structure beyond the marginal
path $\{p_t\}$ itself.
}.

The goal of FM is therefore to find a velocity field
$\mathbf{v}_t(\mathbf{x})$ whose induced ODE
\[
\frac{\diff \mathbf{x}(t)}{\diff t}
=
\mathbf{v}_t(\mathbf{x}(t)),
\qquad
t\in[0,1],
\]
produces marginal distributions that match the prescribed $p_t$ at every
time $t$. When this holds, the ODE can transport samples forward from
$\mathbf{x}(0)\sim p_{\text{src}}$ to $p_{\text{tgt}}$, or backward from
$\mathbf{x}(1)\sim p_{\text{tgt}}$ to $p_{\text{src}}$
(see \Cref{subsec:fm-theory} for a more formal discussion).

This requirement is captured by the \emph{continuity equation}\footnote{
The deterministic analogue of the Fokker--Planck equation, without the
diffusion term.}:
\begin{mdframed}
\begin{equation}\label{eq:continuity_eq}
    \frac{\partial p_t(\mathbf{x})}{\partial t}
    +
    \nabla\cdot
    \bigl(
    \mathbf{v}_t(\mathbf{x})p_t(\mathbf{x})
    \bigr)
    =
    0.
\end{equation}
\end{mdframed}

Any velocity field $\mathbf{v}_t$ satisfying \Cref{eq:continuity_eq},
together with the initial condition $p_0=p_{\mathrm{src}}$, generates an ODE
whose evolving marginals follow the prescribed path $\{p_t\}$, under suitable
regularity conditions. Thus, solving the ODE provides a sample-wise transport
between $p_{\mathrm{src}}$ and $p_{\mathrm{tgt}}$ while matching the
intermediate distributions.

Importantly, the continuity equation constrains the evolution of probability
density, but generally does not uniquely determine the velocity field.
Different sample trajectories can therefore produce exactly the same marginal
density evolution. For example, on regions where $p_t>0$, if
$\mathbf{v}_t$ satisfies \Cref{eq:continuity_eq}, then
\[
\mathbf{v}_t
+
\frac{1}{p_t}\tilde{\mathbf{v}}_t
\]
also satisfies it for any sufficiently regular divergence-free field
$\tilde{\mathbf{v}}_t$ with
$\nabla\cdot\tilde{\mathbf{v}}_t=0$.

FM therefore seeks \emph{a} velocity field that realizes the prescribed
probability path. For general source and target distributions, neither the
marginal density $p_t$ nor such a velocity field is usually available in
closed form. This motivates the conditional construction introduced next,
where tractable conditional paths and velocities are used to learn a valid
marginal velocity field.

As a concrete illustration, in
\Cref{subsec:special-fm-instantiations} we consider a Gaussian-to-Gaussian
path together with a particular affine flow map, for which the corresponding
velocity field can be computed explicitly.

\paragraph{Step 3: Learning via the Conditional Strategy.}
Once a probability path and a corresponding marginal velocity field
$\rvv_t$ have been chosen, FM learns this velocity field directly using a
neural network $\rvv_{\bm{\phi}}$. The ideal FM objective is
\[
\mathcal{L}_{\mathrm{FM}}(\bm{\phi})
=
\mathbb{E}_{t,\,\rvx_t\sim p_t}
\left[
\left\|
\rvv_{\bm{\phi}}(\rvx_t,t)
-
\rvv_t(\rvx_t)
\right\|^2
\right].
\]
We refer to this neural network parameterization as
\emph{$\rvv$-prediction} (velocity prediction), since the network directly
predicts the drift of the ODE. This differs from the score-based construction
in \Cref{subsec:lesson-score}, where the score is learned first and then used
to construct the PF-ODE velocity.

The difficulty is that the marginal velocity $\rvv_t(\rvx_t)$ is generally
intractable. As in denoising score matching, we overcome this using a
conditioning strategy. We introduce a latent variable
$\rvz\sim\pi(\rvz)$ and construct a tractable conditional probability path
$p_t(\rvx_t|\rvz)$ together with a corresponding conditional velocity field
$\rvv_t(\rvx_t|\rvz)$.

If each conditional velocity field generates its corresponding conditional
probability path, then marginalizing over $\rvz$ gives a valid marginal
velocity field:
\begin{align}\label{eq:marginal-oracle-velocity}
\rvv_t(\rvx_t)
:=
\mathbb{E}_{\rvz\sim p(\cdot|\rvx_t)}
\left[
\rvv_t(\rvx_t|\rvz)
\right].
\end{align}
Thus, the generally intractable marginal velocity is the conditional average
of tractable conditional velocities. This is directly analogous to the
conditional-score identity in \Cref{eq:marginal-oracle-score}.

This observation leads to the \emph{conditional flow matching} (CFM)
objective:
\begin{mdframed}
\begin{align}\label{eq:fm-cfm}
\mathcal{L}_{\mathrm{CFM}}(\bm{\phi})
:=
\mathbb{E}_{t,\,\rvz\sim\pi(\rvz),\,
\rvx_t\sim{\color{orange}p_t(\cdot|\rvz)}}
\left[
\left\|
\rvv_{\bm{\phi}}(\rvx_t,t)
-
{\color{orange}\rvv_t(\rvx_t|\rvz)}
\right\|^2
\right].
\end{align}
\end{mdframed}
Here, $\rvz$ is used to construct the training sample $\rvx_t$ and
the conditional regression target $\rvv_t(\rvx_t|\rvz)$; it is not an
additional input to the neural network $\rvv_{\bm{\phi}}(\rvx_t,t)$. Importantly, the two objectives differ only by a constant $C\leq0$ independent of $\bm{\phi}$:
\[
\mathcal{L}_{\mathrm{FM}}(\bm{\phi})
=
\mathcal{L}_{\mathrm{CFM}}(\bm{\phi})
+
C.
\]
They therefore have the same population minimizer,
\begin{align}\label{eq:minimum_velocity}
\rvv^*(\rvx_t,t)
=
\rvv_t(\rvx_t).
\end{align}
In other words, although each training example uses a conditional velocity
$\rvv_t(\rvx_t|\rvz)$ as its target, averaging over these examples recovers
the desired marginal velocity field.

For CFM to provide tractable, simulation-free training, two practical
conditions are needed:
\begin{itemize}
    \item[(i)] we can easily sample
    $\rvx_t\sim p_t(\cdot|\rvz)$ without solving an ODE;
    \item[(ii)] the conditional velocity
    $\rvv_t(\rvx_t|\rvz)$ has a tractable expression that can be used as the
    regression target.
\end{itemize}
We provide concrete constructions satisfying these conditions in
\Cref{subsec:fm-instantiations}. Thus, just as denoising score matching learns
an intractable marginal score through tractable conditional scores, Flow
Matching learns an intractable marginal velocity through tractable conditional-velocity targets.

\begin{figure}[th!]
    \centering
    \includegraphics[width=\linewidth]{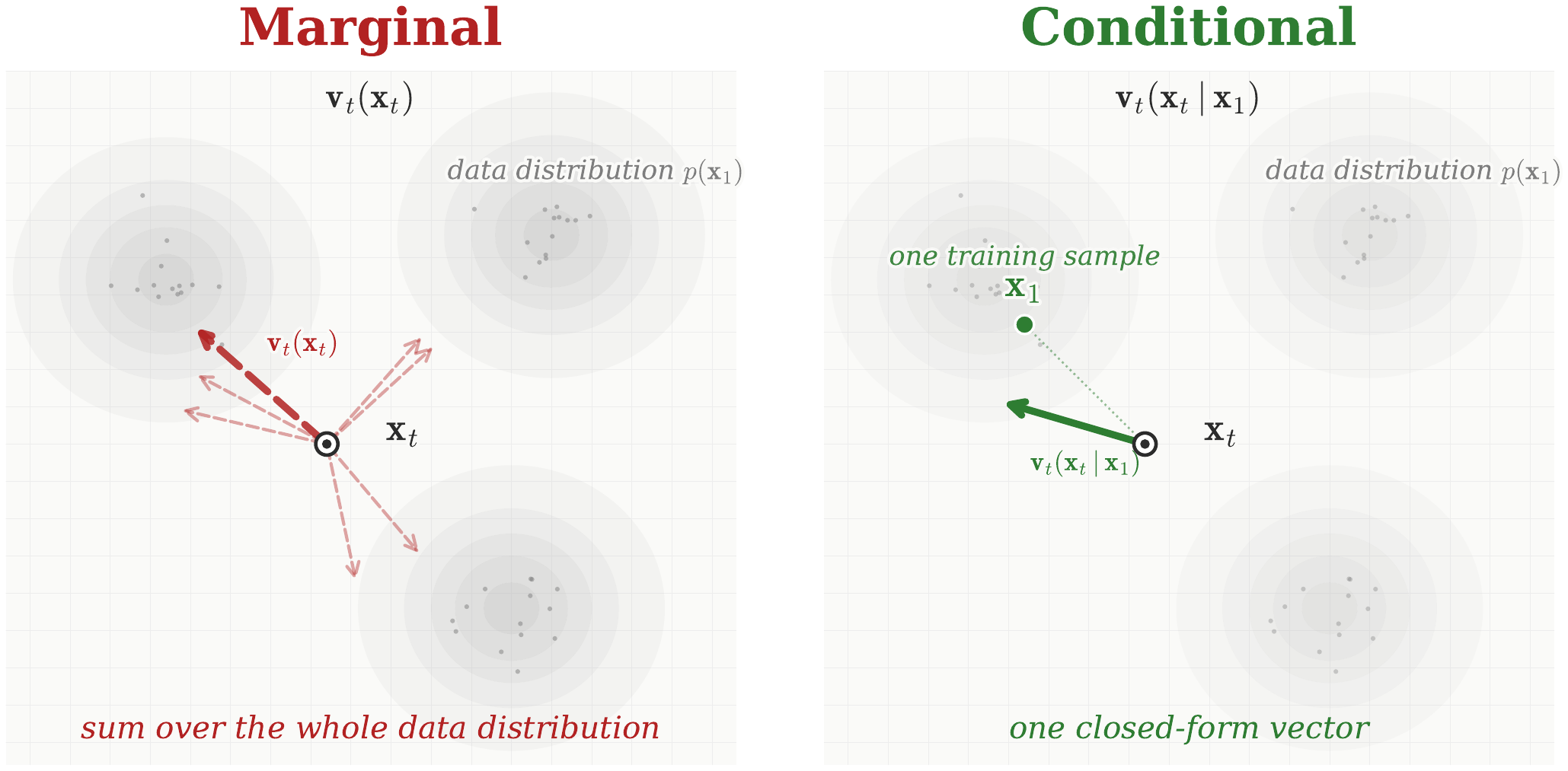}
    \caption{\textbfs{The conditional trick in flow matching.}
    Here, $\rvx_1\sim p_{\mathrm{tgt}}$ denotes a target data sample.
    \textbfs{(Left)}
    For a fixed intermediate sample $\rvx_t$, the marginal velocity
    $\rvv_t(\rvx_t)$ averages the conditional velocities associated with
    all possible target samples $\rvx_1$ that could have produced
    $\rvx_t$. The contribution of each $\rvx_1$ is weighted by how
    compatible it is with the observed $\rvx_t$. Computing this average
    directly requires integrating over the whole target distribution and
    is generally intractable.
    \textbfs{(Right)}
    During training, however, the particular target sample $\rvx_1$ used
    to construct $\rvx_t$ is known. Conditioning on this sample gives the
    closed-form velocity $\rvv_t(\rvx_t|\rvx_1)
        =
        b_t'\rvx_1
        +
        \frac{a_t'}{a_t}
        \bigl(\rvx_t-b_t\rvx_1\bigr)$,
    which provides one simple vector as the regression target
    (Proposition~\ref{closed-form-vf}). Although each training example
    conditions on only one target sample, averaging the conditional
    training objective over the data gives the same optimal marginal
    velocity field; see Theorem~\ref{thm:fm-cfm}. Thus, conditional flow
    matching learns $\rvv_t(\rvx_t)$ without directly regressing on the
    generally intractable marginal velocity. This is the same conditioning
    principle used in the variational perspective
    (Theorem~\ref{thm:equiv-marginal-kl}) and in denoising score matching
    (Theorem~\ref{thm:sm-dsm}).
    \figcredit{Created by the authors with AI-assisted coding.}}
    \label{fig:cfm-illustration}
\end{figure}

We summarize the above discussion as follows:
\thmp{Equivalence of $\mathcal{L}_{\mathrm{FM}}$ and $\mathcal{L}_{\mathrm{CFM}}$}{fm-cfm}{
The following holds:
\begin{align*}
    \mathcal{L}_{\mathrm{FM}}(\bm{\phi}) = \mathcal{L}_{\mathrm{CFM}}(\bm{\phi}) + C,
\end{align*}
where $C$ is a constant independent of the parameter $\bm{\phi}$. Furthermore, the minimizer $\rvv^*$ of both losses satisfies
\[
    \rvv^*(\rvx_t, t) = \rvv_t(\rvx_t), \quad \text{for almost every } \rvx_t \sim p_t,
\]
where $\rvv_t(\rvx_t)$ is defined in \Cref{eq:marginal-oracle-velocity}.
}{The argument and derivation of the minimizer follows exactly the same reasoning as in the score matching case of Proposition~\ref{dsm-minimizer}.
}

This marks the third instance where the conditioning trick yields a tractable training objective, as illustrated in \Cref{fig:cfm-illustration}. Notably, the variational, score based, and flow based approaches all reflect the same underlying principle.

\rmkb{
Taking $\pi = p_{\mathrm{data}}$, we can apply Bayes' rule:
\[
p(\rvx_0|\rvx_t) = \frac{p_t(\rvx_t|\rvx_0) p_{\mathrm{data}}(\rvx_0)}{p_t(\rvx_t)},
\]
a similar decomposition of \Cref{eq:marginal-oracle-velocity} appears in score-based models:
\begin{align*}
    \nabla_{\rvx_t} \log p_t(\rvx_t) 
    &= \mathbb{E}_{\rvx_0 \sim p(\cdot|\rvx_t)} \left[ \nabla_{\rvx_t} \log p_t(\rvx_t|\rvx_0) \right] \\
    &= \mathbb{E}_{\rvx_0 \sim p_{\text{data}}} \left[ \nabla_{\rvx_t} \log p_t(\rvx_t|\rvx_0) \cdot \frac{p_t(\rvx_t|\rvx_0)}{p_t(\rvx_t)} \right],
\end{align*}
which mirrors the marginalization strategy in \Cref{eq:marginal-oracle-velocity}.
}

As in \Cref{subsec:lesson-score}, where the conditional density $p_t(\rvx_t|\rvz)$ and conditional velocity field $\rvv_t(\rvx_t|\rvz)$ must be explicitly specified, with $p_t(\rvx_t|\rvx_0) = \mathcal{N}(\rvx_t; \alpha_t \rvx_0, \sigma_t^2 \rmI)$ and $\rvv_t(\rvx_t|\rvx_0) = f(t)\rvx_t - \frac{1}{2}g^2(t)\nabla_{\rvx_t} \log p_t(\rvx_t|\rvx_0)$, the general conditional flow matching framework also requires these two components. However, we have not yet construct the conditional density $p_t(\rvx_t|\rvz)$ or the conditional velocity field $\rvv_t(\rvx_t|\rvz)$ in this general case. In \Cref{sec:instance-fm}, we introduce several common instantiations of these components.

\subsection{Comparison of Diffusion Models, General Flow Matching, and NODEs}\label{subsec:compare-dm-general-fm}

\paragraph{Comparison of Diffusion Models and General Flow Matching.}
The insight from \Cref{subsec:lesson-score} leads to an extended FM framework that retains the same underlying principles. To highlight their similarities, we summarize them in \Cref{tab:framework_comparison}.

\begin{table}[th]
\caption{Comparison between diffusion models (or Gaussian FM) and the general FM framework.  
Here, the general FM framework refers to the setting with two-sided conditioning, 
where $\mathbf{x}_0 \sim p_{\text{src}}$ and $\mathbf{x}_1 \sim p_{\text{tgt}}$ 
are sampled independently.}
\label{tab:framework_comparison}
\centering
\resizebox{\textwidth}{!}{
\begin{tabular}{lcc}
\toprule
\textbfs{Aspect} & \textbfs{(Score-Based) Diffusion Model} & \textbfs{General FM} \\
\midrule
Source dist. $p_{\mathrm{src}}$ & Gaussian prior & Any \\
Target dist. $p_{\mathrm{tgt}}$ & Data distribution & Any \\
Latent  dist. $\pi(\rvz)$ & $p_{\text{data}}$ & See \Cref{subsec:fm-instantiations} \\ 
Cond. dist. $p_t(\rvx_t|\rvz)$ & $\mathcal{N}(\rvx_t;\alpha_t\rvx_0,\sigma_t^2\rmI)$ & See \Cref{subsec:fm-instantiations} \\
Marginal dist. $p_t(\rvx_t)$ & $\int p_t(\rvx_t|\rvx_0) p_{\mathrm{data}}(\rvx_0) \diff\rvx_0$ & $ \int p_t(\rvx_t|\rvz) \pi(\rvz) \diff\rvz$ \\
Cond. velocity $\rvv_t(\rvx|\rvz)$  & $f(t) \rvx - \frac{1}{2} g^2(t) \nabla \log p_t(\rvx|\rvx_0)$ & See \Cref{subsec:fm-instantiations} \\
Marginal velocity $\rvv_t(\rvx)$  & $f(t) \rvx - \frac{1}{2} g^2(t) \nabla \log p_t(\rvx)$ & See \Cref{eq:marginal-oracle-velocity} \\
Learning objective & $\mathcal{L}_{\text{SM}}=\mathcal{L}_{\text{DSM}}+C$ & $\mathcal{L}_{\mathrm{FM}}=\mathcal{L}_{\mathrm{CFM}}+C$ \\
Underlying Rule & \multicolumn{2}{c}{Fokker-Planck / Continuity Equation} \\
\bottomrule
\end{tabular}
}
\end{table}
We remark that, assuming the Gaussian schedule satisfies the diffusion-SDE
compatibility condition in Lemma~\ref{forward-sde}, Gaussian FM is essentially
equivalent to the corresponding diffusion formulation (see more in
\Cref{ch:all-equivalent}); we therefore do not differentiate between them
unless explicitly stated.

\paragraph{Connection to NODEs.}
FM can be viewed as a simulation-free alternative to NODEs, introduced in \Cref{subsec:NODE}. While CNFs require solving ODEs during maximum likelihood training, which is computationally intensive, FM bypasses this by directly regressing a prescribed velocity field through a simple regression loss. The key insight is that when the marginal density path connecting the source and target distributions is fixed, exact simulation during training becomes unnecessary.

\subsection{(Optional) Underlying Rules}\label{subsec:fm-theory}

\paragraph{Continuity Equation: Mass Conservation Criterion.}  
Similar to the PF-ODE and Fokker–Planck analysis in \Cref{subsec:lesson-score}, we now present a criterion for verifying whether the density path induced by an ODE flow aligns with a prescribed path $\{p_t\}_{t \in [0,1]}$.

Consider the ODE describing the flow of particles under a time-dependent velocity field $\rvv_t$:
\[
\frac{\diff \rvx(t)}{\diff t} = \rvv_t\left(\rvx(t)\right).
\]
As in \Cref{eq:pushforward-ode}, this ODE defines a flow map 
$\bm{\Psi}_{s\to t}(\rvx_0)$ for any $s,t \in [0,1]$, which in particular 
transports an initial point $\rvx_0 \sim p_{\mathrm{src}}$ at time $0$ to its 
state at time $t$.  
The induced distribution at time $t$ is given by the pushforward
\begin{align}\label{eq:ode-pushforward}
    p_t^{\mathrm{fwd}}(\rvx) 
    = \int \delta \left(\rvx - \bm{\Psi}_{0\to t}(\rvx_0)\right) 
      p_{\mathrm{src}}(\rvx_0) \mathrm{d}\rvx_0=: \bm{\Psi}_{0\to t}\# p_{\mathrm{src}},
\end{align}
so that $\bm{\Psi}_{0\to t}(\rvx_0) \sim p_t^{\mathrm{fwd}}$ whenever 
$\rvx_0 \sim p_{\mathrm{src}}$.  
Similarly, one can transport backward from $\rvx_1 \sim p_{\mathrm{tgt}}$ 
to $p_{\mathrm{src}}$ via $\bm{\Psi}_{1\to 0}(\rvx_1)$.

Suppose we are given a prescribed density path $\{p_t\}_{t \in [0,1]}$, and we construct a velocity field $\{\rvv_t\}_{t \in [0,1]}$ to define a particle flow. This naturally raises the question:

\begin{question}
    Under what conditions does the flow-induced density $p_t^{\mathrm{fwd}}$  exactly match the target density $p_t$ for all $t \in [0,1]$?
\end{question}
Once the two density evolutions align, we can leverage the ODE flow to flexibly transport samples between $p_{\mathrm{src}}$ and $p_{\mathrm{tgt}}$ by solving the ODE.

As in \Cref{eq:p_t-fwd-rev}, a principled way to verify this alignment is via the \emph{continuity equation}, which captures the conservation of mass in time-evolving densities:

\thmp{Mass Conservation Criterion}{continuity-mass}{
The flow-induced density $p_t^{\mathrm{fwd}}$ equals the prescribed path $p_t$ for all $t \in [0,1]$; i.e., 
\[
p_t^{\mathrm{fwd}} = p_t, \quad \text{for all } t \in [0,1],
\]
if and only if the pair $(p_t, \rvv_t)$  satisfies the continuity equation:
\[
\partial_t p_t(\rvx) + \nabla_\rvx \cdot (p_t(\rvx) \rvv_t(\rvx)) = 0,
\]
for all $t\in[0,1]$ and $\rvx$.
}{A conceptual derivation is provided in \Cref{app-sec:mass-conservation}, while a more rigorous treatment can be found in \citep{villani2008optimal} (see ``Mass Conservation Formula'').}

\paragraph{From Conditional to Marginal Paths.} As seen in \Cref{subsec:fm-framework}, we begin by defining a conditional probability path $p_t(\cdot| \rvz)$ and a corresponding conditional velocity field $\rvv_t(\cdot| \rvz)$. We then construct the marginal velocity field via:
\[
    \rvv_t(\rvx) 
    = \int \rvv_t(\rvx| \rvz) \frac{p_t(\rvx| \rvz) \pi(\rvz)}{p_t(\rvx)}   \diff \rvz,
\]
as in \Cref{eq:marginal-oracle-velocity}. However, we still need to ensure that the resulting marginal velocity $\rvv_t$ induces an ODE flow whose density path aligns with the prescribed $p_t$. Fortunately, this verification can be done entirely at the conditional level: if each conditional velocity field $\rvv_t(\cdot| \rvz)$ induces the conditional density path $p_t(\cdot| \rvz)$, then the resulting marginal velocity $\rvv_t$ also induces the correct marginal path. Formally, we state the result below and provide the complete derivation in one place to show explicitly how the marginal velocity field takes this form:

\proppp{Marginal VF Generates Given Marginal Density}{marginal-vf}{
If the conditional velocity fields $\rvv_t(\cdot|\rvz)$ induce conditional density paths that match $p_t(\cdot|\rvz)$ (starting from $p_0(\cdot|\rvz)$), then the marginal velocity field $\rvv_t(\cdot)$ defined in \Cref{eq:marginal-oracle-velocity} induces a marginal density path that aligns with $p_t(\cdot)$, starting from $p_0(\cdot)$.
}{
This result follows by verifying that the pair $(p_t, \rvv_t)$ satisfies the Continuity Equation. We present the argument in a converse manner to provide intuition for why the marginalized velocity field takes the form in \Cref{eq:marginal-oracle-velocity}.
Since the conditional velocity fields $\rvv_t(\cdot|\rvz)$ induce density paths matching the conditional densities $p_t(\cdot|\rvz)$ for $\rvz \sim \pi$, the continuity equation holds for each conditional pair:
\begin{align}\label{eq:continuity_conditional_pt}
    \frac{\diff}{\diff t} p_t(\rvx|\rvz) =- \nabla_\rvx \cdot \big(\rvv_t(\rvx|\rvz) p_t(\rvx|\rvz)\big).
\end{align}
We aim to find a velocity field ${\color{orange}\rvv_t(\cdot)}$ whose induced densities align with the marginal density $p_t$, i.e., satisfy
\begin{align}\label{eq:p_t_continuity}
    \frac{\diff}{\diff t} p_t(\rvx) =- \nabla_\rvx \cdot \big({\color{orange}\rvv_t(\rvx)} p_t(\rvx)\big).
\end{align}
Starting from the definition of $p_t$ in \Cref{eq:p_t}, 
\begin{align*}
  \frac{\diff}{\diff t} p_t(\rvx) &=\int \frac{\diff}{\diff t} p_t(\rvx_t|\rvz)\pi(\rvz)\diff \rvz \nonumber\\
  &=  - \int \nabla_\rvx \cdot\Big(  \rvv_t(\rvx|\rvz) p_t(\rvx | \rvz)  \Big)\pi(\rvz) \diff \rvz \quad \nonumber \\
  &=  -\nabla_\rvx \cdot \Big( \int  \rvv_t(\rvx|\rvz) p_t(\rvx | \rvz) 
 \pi(\rvz) \diff \rvz\Big), \nonumber  
\end{align*}
where the second equality follows by applying \Cref{eq:continuity_conditional_pt}. Comparing this with the right-hand side of \Cref{eq:p_t_continuity} shows that, up to a divergence-free term,
\[
    {\color{orange}\rvv_t(\rvx)} p_t(\rvx) = \int \rvv_t(\rvx|\rvz) p_t(\rvx|\rvz) \pi(\rvz) \diff \rvz.
\]
Therefore, we can define
\[
    {\color{orange}\rvv_t(\rvx)} := \int \rvv_t(\rvx|\rvz) \frac{p_t(\rvx|\rvz)}{p_t(\rvx)} \pi(\rvz) \diff \rvz,
\]
which is precisely the form in \Cref{eq:marginal-oracle-velocity}.
The proof of this theorem essentially follows the reverse of this argument.
}

This connection allows us to reduce the construction of the potentially intractable marginal velocity field to defining simpler conditional fields $\rvv_t(\cdot| \rvz)$, which are easier to work with by construction.

\newpage

\section{Constructing Probability Paths and Velocities Between Distributions}\label{sec:instance-fm}

The essence of flow matching lies in the gradual transformation of a source distribution into a target. To direct this transformation, two key elements are needed: the \emph{probability path} $p_{t}$, which provides a snapshot of the evolving distribution at each time $t$, and the \emph{velocity field} $\rvv_{t}$, which describes how individual particles move along the path. These two objects are not independent; they are linked through the continuity equation, which ensures that particle dynamics are consistent with the evolution of the distribution. Thus, the learning task reduces to finding a velocity field $\rvv_t$ that faithfully drives the process. The difficulty, however, is that for general and complex distributions, the true marginal velocity $\rvv_t$ is unknown, leaving us with an intractable target that cannot be accessed directly. 

The core idea of Conditional Flow Matching is to address the intractability of the true marginal velocity by constructing an artificial but tractable process. To do this, we introduce a conditioning variable $\rvz$ and design either a conditional velocity $\rvv_t(\rvx_t|\rvz)$ and/or a conditional path $p_t(\rvx_t|\rvz)$, which are deliberately chosen to be simple.

 Because these conditional objects are known in closed form, they serve as surrogate targets that the model can regress against. This leads to a valid training loss $\mathcal{L}_{\mathrm{CFM}}$, provided two practical requirements are met: (i) we can sample efficiently from $p_t(\cdot|\rvz)$, and (ii) the corresponding velocity $\rvv_t(\cdot|\rvz)$ admits a closed-form expression.

How should we design a well behaved conditional process? For inspiration, we turn to the one case that is fully understood: the Gaussian to Gaussian bridge (\Cref{subsec:special-fm-instantiations}). This example highlights two natural design strategies: adopt a Gaussian probability path at each time $t$, or prescribe an affine velocity field, both of which are analytically tractable. 


Guided by this insight, we extend to general endpoint distributions using
two complementary construction views (see also
\Cref{subsec:cov-continuity-eq}). These correspond to the classical
Eulerian and Lagrangian descriptions of transport:

\begin{itemize}
    \item \textbfs{Conditional Probability-Path First (Eulerian View).}
    We begin with the conditional density path
    $p_t(\cdot|\rvz)$ and seek a conditional velocity field
    $\rvv_t(\cdot|\rvz)$ satisfying the continuity equation. Since a
    density path alone generally does not determine a unique velocity,
    this construction also requires choosing a particular compatible
    velocity field.

    \item \textbfs{Conditional Flow First (Lagrangian View).}
    We instead specify how individual particles move through a conditional
    flow map $\bPsi_{0\to t}(\cdot|\rvz)$. The induced density is obtained
    by pushforward, and, when the flow map is invertible, the corresponding
    Eulerian velocity field is
    \[
    \rvv_t
    =
    \partial_t\bPsi_{0\to t}
    \circ
    \bPsi_{0\to t}^{-1}.
    \]
\end{itemize}

In \Cref{subsec:fm-instantiations}, we detail the first approach, which shows its close analogy to diffusion model construction discussed in \Cref{subsec:lesson-score}, while in \Cref{subsec:prob-path-flow} we present the second. Together, these perspectives provide a practical framework for defining $p_t(\rvx_t| \rvz)$ and $\rvv_t(\rvx_t| \rvz)$, enabling simulation-free training and the construction of flows between arbitrary source and target distributions.

\subsection{A Key Special Case: Marginal $p_t(\rvx_t)$ and Velocity $\rvv_t(\rvx_t)$ in the Gaussian-to-Gaussian Bridge}\label{subsec:special-fm-instantiations}

We begin with the Gaussian–endpoint case, where we can compute the marginal density $p_t(\rvx_t)$ and velocity field $\rvv_t(\rvx_t)$ analytically. This serves as a template for the general construction of the conditional density $p_t(\rvx_t |\rvz_t)$ and velocity field $\rvv_t(\rvx_t |\rvz_t)$.

When the source and target distributions, $p_{\text{src}}$ and $p_{\text{tgt}}$, are both Gaussian, the velocity field $\rvv_t(\cdot)$ admits a closed-form expression. We consider the interpolated marginal density path:
\begin{align}\label{eq:gaussian-density-path}
    p_t(\rvx_t) = \mathcal{N}\left(\rvx_t; \bm{\mu}(t), \sigma^2(t) \rmI\right),
\end{align}
with time-varying mean $\bm{\mu}(t)$ and variance $\sigma^2(t)>0$. The two endpoints are given by
\[
p_{\text{src}} = p_0 =  \mathcal{N}\left(\rvx; \bm{\mu}(0), \sigma^2(0) \rmI\right), \quad
p_{\text{tgt}} = p_1 =  \mathcal{N}\left(\rvx; \bm{\mu}(1), \sigma^2(1) \rmI\right),
\]
so that the path $\{p_t\}_{t \in [0,1]}$ connects these distributions.

With the given path $\{p_t\}_{t \in [0,1]}$, there are indeed many velocity fields that induce an ODE flow $\bm{\Psi}_{0\to t}(\rvx)$ such that $\rvx \sim p_0$ implies $\bm{\Psi}_{0\to t}(\rvx) \sim p_t$. For this Gaussian path, a particularly simple realization is given by\footnote{In \citep{lipman2022flow}, the authors consider $\bm{\Psi}_{0\to t}(\rvx) = \bm{\mu}(t) + \sigma(t)\rvx$, which requires constraints on $\bm{\mu}(t)$ and $\sigma(t)$ to ensure boundary conditions. We adopt an equivalent normalized formulation that avoids such constraints.}:
\begin{align}\label{eq:gaussian-sol-map}
    \bm{\Psi}_{0\to t}(\rvx) := \bm{\mu}(t) + \sigma(t) \left( \frac{\rvx - \bm{\mu}(0)}{\sigma(0)} \right).
\end{align}

For the defined Gaussian path $p_t$ (Gaussian for all $t$), the velocity field $\rvv_t(\cdot)$ inducing the ODE flow in \Cref{eq:gaussian-sol-map} is uniquely and analytically characterized as follows~\citep{lipman2022flow}:

\proppp{Closed-Form Velocity Field for Gaussian Density Path}{closed-form-vf}{
Let $p_t$ be the Gaussian path in \Cref{eq:gaussian-density-path}. Then the velocity field $\rvv_t(\cdot)$ that generates the ODE flow \Cref{eq:gaussian-sol-map} is unique for the defined $\bPsi_{0\to t}$ and has the closed-form expression:
\begin{equation*}
    \rvv_t(\rvx) = \frac{\sigma'(t)}{\sigma(t)} \left( \rvx - \bm{\mu}(t) \right) + \bm{\mu}'(t).
\end{equation*}
}{Consider the ODE with initial condition $\rvy$:
\[
    \frac{\diff}{\diff t} \bm{\Psi}_{0\to t}(\rvy) = \rvv_t(\bm{\Psi}_{0\to t}(\rvy)).
\]
Since $\bm{\Psi}_{0\to t}$ is invertible (as $\sigma(t)>0$), we may set $\rvx = \bm{\Psi}_{0\to t}(\rvy)$ and $\rvy = \bm{\Psi}_{0\to t}^{-1}(\rvx) = \bm{\Psi}_{t\to 0}(\rvx)$ to obtain
\[
    \bm{\Psi}_{0\to t}'\big(\bm{\Psi}_{0\to t}^{-1}(\rvx)\big) = \rvv_t(\rvx).
\]
Differentiating \Cref{eq:gaussian-sol-map} with respect to $t$ gives
\begin{equation*}
    \bm{\Psi}_{0\to t}'(\rvx) = \bm{\mu}'(t) + \sigma'(t) \left( \frac{\rvx - \bm{\mu}(0)}{\sigma(0)} \right).
\end{equation*}
Solving for $\rvy = \bm{\Psi}_{0\to t}^{-1}(\rvx)$ yields
\[
    \rvy = \bm{\mu}(0) + \sigma(0) \left( \frac{\rvx - \bm{\mu}(t)}{\sigma(t)} \right).
\]
Substituting this into $\bm{\Psi}_{0\to t}'(\rvx)$ gives
\[
    \rvv_t(\rvx) = \frac{\sigma'(t)}{\sigma(t)} \left( \rvx - \bm{\mu}(t) \right) + \bm{\mu}'(t),
\]
as claimed.
}
We note that for a fixed flow map $\bPsi_{0\to t}$ (flow-first view), the velocity is uniquely determined by
\[
\rvv_t  =  \partial_t \bPsi_{0\to t} \circ \bPsi_{0\to t}^{-1}.
\]
Under this construction, the pair $(p_t,\rvv_t)$ automatically satisfies the continuity equation.
By contrast, for a given density path $t\mapsto p_t$ without fixing $\bPsi_{0\to t}$ (probability-path-first view),
the velocity field is not unique.

This distinction precisely characterizes the difference between the flow-first and probability-path-first perspectives.

This closed-form characterization remains valid when conditioning on a latent variable $\rvz$.
In the following, we extend this insight to construct a conditional Gaussian path $p_t(\cdot|\rvz)$
and derive the corresponding conditional velocity field $\rvv_t(\cdot|\rvz)$ for the general marginal setting.

\subsection{Conditional Probability-Path-First Construction of $\rvv_t(\cdot|\rvz)$ and $p_t(\cdot|\rvz)$}\label{subsec:fm-instantiations}

We aim to construct a conditional density path $p_t(\cdot|\rvz)$ first and then derive its corresponding conditional velocity field $\rvv_t(\cdot|\rvz)$ (via Proposition~\ref{closed-form-vf}), under conditioning with respect to $\pi(\rvz)$. 
Depending on how $\rvz$ is chosen, there are two natural scenarios: 
(i) two-sided conditioning with $\rvz=(\rvx_0,\rvx_1)$, or 
(ii) one-sided conditioning with $\rvz=\rvx_0$ or $\rvx_1$. 
In either case, the construction must match the boundary distributions:
\begin{align*}
   p_{\mathrm{src}}(\rvx_0) = \int p_0(\rvx_0|\rvz) \pi(\rvz)\diff\rvz, \quad
   p_{\mathrm{tgt}}(\rvx_1) = \int p_1(\rvx_1|\rvz) \pi(\rvz) \diff\rvz.
\end{align*}
Since verifying these constraints is straightforward once a concrete construction is specified, we do not emphasize the verification step here.

\paragraph{I. Two-Sided $\rvz = (\rvx_0, \rvx_1)$ — ``Beam-Like'' Path.}
    \subparagraph{Choice of $\pi(\rvz)$.} Consider general distributions $p_{\mathrm{src}}$ and $p_{\mathrm{tgt}}$ over $\mathbb{R}^D$. Let $\rvz = (\rvx_0, \rvx_1)$ with $\rvx_0 \sim p_{\mathrm{src}}$ and $\rvx_1 \sim p_{\mathrm{tgt}}$ independently, i.e.,
    \[
    \pi(\rvz) = p_{\mathrm{src}}(\rvx_0)  p_{\mathrm{tgt}}(\rvx_1).
    \]
    \subparagraph{Choice of Conditional Path $p_t(\cdot|\rvz)$.} Define the conditional path by linear interpolation with a fixed and small variance $\sigma > 0$\footnote{
One may instead add a fixed conditional variance $\sigma^2\rmI$.
The conditional velocity remains $a_t'\rvx_0+b_t'\rvx_1$ because the
variance is constant in time, but the endpoint marginals then become Gaussian
smoothings of $p_{\mathrm{src}}$ and $p_{\mathrm{tgt}}$, approaching the
desired endpoints as $\sigma\to0$.
}:
    \[
    p_t(\rvx_t | \rvz = (\rvx_0, \rvx_1)) = \mathcal{N}(\rvx_t; a_t\rvx_0 + b_t\rvx_1,  \sigma^2 \rmI),
    \]
    where $a_t$ and $b_t$ are time-dependent functions satisfying  $a_0 = 1,  b_0 = 0$ and $a_1 = 0,  b_1 = 1$. A choice suggested by \citep{lipman2022flow,liu2022rectified} is $a_t = 1-t$, $b_t = t$.  In the deterministic case $\sigma = 0$, we obtain  
\[
p_t(\rvx_t |\rvz) = \delta \big(\rvx_t - [a_t\rvx_0 + b_t\rvx_1]\big),
\]
which describes a deterministic interpolating path from $\rvx_0$ to $\rvx_1$.
    \subparagraph{Derived Conditional Velocity  $\rvv_t(\cdot|\rvz)$.} By Proposition~\ref{closed-form-vf}, the conditional velocity is
    \[
    \rvv_t(\rvx | \rvz) = a'_t\rvx_0 + b'_t\rvx_1.
    \]
    \subparagraph{CFM Loss.} When $\sigma=0$ so that $\rvx_t = a_t\rvx_0 + b_t\rvx_1$, the CFM loss reduces to
    \[
    \mathcal{L}_{\mathrm{CFM}} = \mathbb{E}_{t, \rvx_0\sim p_{\text{src}}, \rvx_1\sim p_{\text{tgt}}} \left\| \rvv_{\bm{\phi}}(\rvx_t, t) - \left(a'_t\rvx_0 + b'_t\rvx_1\right) \right\|^2.
    \]
From \Cref{eq:marginal-oracle-velocity,eq:minimum_velocity}, the optimal velocity field is
\[
\rvv^*(\rvx_t, t) = \mathbb{E} \left[ \rvx_t'|\rvx_t \right] = \mathbb{E} \left[ a'_t\rvx_0 + b'_t\rvx_1|\rvx_t \right].
\]
Here, the expectation is taken over $p(\mathbf{x}_0, \mathbf{x}_1|\mathbf{x}_t)$, the conditional distribution over source-target pairs $(\mathbf{x}_0, \mathbf{x}_1)$ that could have produced the observed interpolation $\mathbf{x}_t = a_t\mathbf{x}_0 + b_t\mathbf{x}_1$ at time $t$.

\paragraph{II. One-Sided $\rvz = \rvx_0$ or $\rvx_1$ — ``Spotlight-Like'' Path.}
We illustrate the conditional probability–path–first construction in the one-sided setting, considering the standard generative setup with $p_{\mathrm{src}}=\mathcal N(\bm 0,\rmI)$ and $p_{\mathrm{tgt}}=p_{\mathrm{data}}$.
Crucially, this Gaussian source is not an additional assumption but a direct consequence of the conditional path defined below.
A more general treatment of arbitrary endpoints will be given in \Cref{subsec:prob-path-flow}.

\subparagraph{Choice of $\pi(\rvz)$.}
We take $\rvz=\rvx_1$ with $\pi(\rvz)=p_{\mathrm{data}}(\rvx_1)$ (the case $\rvz=\rvx_0\sim p_{\mathrm{prior}}$ follows analogously).

\subparagraph{Choice of Conditional Path $p_t(\cdot|\rvz)$.}
For fixed $\rvx_1\sim p_{\mathrm{data}}$, define
\[
p_t(\rvx_t|\rvz=\rvx_1)=\mathcal N \big(\rvx_t; b_t \rvx_1, a_t^2 \rmI\big),
\]
with $ a_0=1, b_0=0, a_1=0, b_1=1$ (usually interpreted as the limit).
At the boundaries,
\[
p_0(\cdot|\rvz=\rvx_1)=\mathcal N(\cdot; \bm 0,\rmI),\qquad
p_1(\cdot|\rvz=\rvx_1)=\delta(\cdot - \rvx_1).
\]
Marginalizing over $\rvx_1$ yields $\{p_t\}_{t\in[0,1]}$ with
$p_0=\mathcal N(\bm 0,\rmI)$ (independent of $p_{\mathrm{data}}$) and $p_1=p_{\mathrm{data}}$.

\subparagraph{Derived Conditional Velocity  $\rvv_t(\cdot|\rvz)$.}
For $t\in(0,1)$ with $a_t>0$, applying Proposition~\ref{closed-form-vf} to the conditional Gaussian path gives
\[
\mathbf v_t(\rvx|\rvx_1)
= b_t' \rvx_1 + \frac{a_t'}{a_t} \big(\rvx - b_t \rvx_1\big).
\]

\subparagraph{One-Sided CFM Objective.}
With $t\sim\mathcal U(0,1)$ (or any fixed sampling distribution) and $\rvx_1\sim p_{\mathrm{data}}$, the CFM loss becomes
\begin{align}\label{eq:one-sided-cfm}
    \mathcal L_{\mathrm{CFM}}
=\E_{t,\rvx_1} \E_{\rvx_t\sim p_t(\cdot|\rvx_1)}
\Big\|\mathbf v_\phi(\rvx_t,t)-
\Big[b_t'\rvx_1+\tfrac{a_t'}{a_t}\big(\rvx_t-b_t\rvx_1\big)\Big]\Big\|_2^2.
\end{align}
By MSE optimality, the unique minimizer is the marginal velocity field
\[
\mathbf v^*(\rvx,t)
= \E \left[\mathbf v_t(\rvx|\rvx_1) |dle|  \rvx_t=\rvx\right]
= \E \left[a_t'\rvx_0+b_t'\rvx_1 |dle|  \rvx_t=\rvx\right].
\]

\subparagraph{Equivalence to Two-Sided Target.}
For paired samples $(\rvx_0,\rvx_1)$ with $\rvx_t=a_t\rvx_0+b_t\rvx_1$,
\[
\mathbf v_t(\rvx_t|\rvx_1)
= b_t'\rvx_1 + \tfrac{a_t'}{a_t}\big(\rvx_t-b_t\rvx_1\big)
= a_t'\rvx_0 + b_t'\rvx_1.
\]
Thus the one-sided loss regresses to the \emph{conditional expectation} of the two-sided target given $\rvx_t$:
\[
\mathbf v^*(\rvx,t)=\E \big[a_t'\rvx_0 + b_t'\rvx_1 | \rvx_t=\rvx\big],
\]
so the one-sided and two-sided CFM objectives share the same minimizer.

\paragraph{Gaussian FM = Diffusion Model.}
We use the FM convention where $t=0$ denotes the source/prior and $t=1$ denotes the target/data:
\[
p_{\mathrm{src}}=p_{\mathrm{prior}},\qquad p_{\mathrm{tgt}}=p_{\mathrm{data}}.
\]
By contrast, diffusion models typically index time from data to noise (i.e., $t=0$ is data and $t=1$ is prior). 
Here, we adopt the FM convention indexing in the main discussion. For comparison, however, the VE and VP entries in \Cref{tb:interpolant_instances} are summarized in their usual diffusion-time indexing, where $t=0$ corresponds to data and $t=1$ to prior/noise.
If further $p_{\mathrm{src}}=\mathcal N(\bm 0,\rmI)$, then for fixed condition $\rvx_1\sim p_{\mathrm{data}}$, 
the conditional path $p_t(\cdot |\rvx_1)$ is naturally chosen to be Gaussian, 
while the target distribution $p_{\mathrm{tgt}}$ itself need not be Gaussian. We refer to this Gaussian conditional-path setting as \emph{Gaussian FM}.
Diffusion probability paths form an important subclass of such Gaussian
paths. A generic Gaussian schedule, however, need not correspond to a valid
forward diffusion SDE; the schedule must additionally satisfy the SDE
compatibility condition in Lemma~\ref{forward-sde}.

Choosing $a_t = 1-t$ and $b_t = t$ (equivalently, $\alpha_t = t$ and $\sigma_t = 1-t$ under the relabeling $a_t:=\sigma_t$, $b_t:=\alpha_t$ in diffusion model) recovers the familiar FM/RF schedule~\citep{lipman2022flow,liu2022rectified}.

\begin{table}[th]
\caption{Summary of different interpolants written as
$\rvx_t=a_t\rvx_0+b_t\rvx_1$, where
$\rvx_0\sim p_{\mathrm{src}}=p_{\mathrm{prior}}$ and
$\rvx_1\sim p_{\mathrm{tgt}}=p_{\mathrm{data}}$.
The FM/RF and trigonometric columns use the FM convention
($t=0$ source, $t=1$ target), whereas the VE/VP columns retain the usual
diffusion-time convention ($t=0$ data, $t=1$ prior/noise), with
$a_t:=\sigma_t$ and $b_t:=\alpha_t$.
For VE at a finite terminal noise level, the exact terminal marginal is
$p_{\mathrm{data}}*\mathcal{N}(\bm{0},a_1^2\rmI)$; the Gaussian prior shown
in the table is the usual approximation when the terminal noise dominates
the scale of the data.}
  \small
  \centering
  \resizebox{\textwidth}{!}{
  \begin{tabular}{lcccc}
     \toprule
         & \textbfs{VE} & \textbfs{VP} & \textbfs{FM/RF} & \textbfs{Trig.}~\citep{albergo2023stochastic} \\
    \midrule
     $a_t$ (prior coeff.) 
       & $a_t$ 
       & $\sqrt{1-b_t^2}$ 
       & $1-t$ 
       & $\cos \bigl(\tfrac{\pi}{2}t\bigr)$ \\
     $b_t$ (data coeff.) 
       & $1$ 
       & $b_t$ 
       & $t$ 
       & $\sin \bigl(\tfrac{\pi}{2}t\bigr)$ \\
    \midrule
     $a_0$ & $0$ & $0$ & $1$ & $1$ \\
     $b_0$ & $1$ & $1$ & $0$ & $0$ \\
     $a_1$ & $a_1$ & $1$ & $0$ & $0$ \\
     $b_1$ & $1$ & $0$ & $1$ & $1$ \\
    \midrule
     $p_{\text{prior}}$ 
       & $p_{\mathrm{data}} * \mathcal{N}(\bm{0},a_1^2\rmI)
   \approx \mathcal{N}(\bm{0},a_1^2\rmI)$ 
       & $\mathcal{N}(\bm{0},\rmI)$ 
       & $\mathcal{N}(\bm{0},\rmI)$ 
       & $\mathcal{N}(\bm{0},\rmI)$ \\
     \bottomrule
  \end{tabular}}
  \label{tb:interpolant_instances}
\end{table}

In the Gaussian FM setting, both the beam-like and spotlight-like conditional paths lead to training objectives that are  similar to the standard diffusion losses. As we will elaborate in \Cref{ch:all-equivalent}, Gaussian FM can in fact be equivalently interpreted as a diffusion model trained to predict the \emph{velocity}, under the linear schedule $a_t = 1-t$ and $b_t = t$. This perspective highlights that flow matching and diffusion are not fundamentally different, but rather two equivalent formulations that can be transformed into one another. The Gaussian FM objective is particularly appealing in practice: its loss function ($\E_{t,\rvx_t} \left[\norm{\rvv_\bphi(\rvx_t, t) - (\rvx_1 - \rvx_0)}_2^2\right]$) is simple, and it has been shown to achieve competitive performance at scale~\citep{esser2024scaling}.

\subsection{Conditional Flow-First Construction of $\rvv_t(\cdot|\rvz)$ and $p_t(\cdot|\rvz)$}\label{subsec:prob-path-flow}

We treat the general case where the endpoints $p_{\mathrm{src}}$ (at $t=0$) and $p_{\mathrm{tgt}}$ (at $t=1$) are arbitrary. 
Our goal is to design, directly in trajectory space, a conditional flow that transports samples from $p_{\mathrm{src}}$ to $p_{\mathrm{tgt}}$ and yields a closed-form $\rvv_t(\rvx_t|\rvz)$ usable as a regression target.

\paragraph{Motivation.}
Instead of first designing conditional density path, we may directly specify a conditional flow map $\bPsi_{0\to t}(\cdot;\rvz)$ that moves samples along trajectories. This has two practical advantages: (i) it immediately yields a regression target for training via a time derivative along trajectories; (ii) on geometry-structured spaces (Riemannian manifolds, Lie groups, or constrained submanifolds), it is often natural to construct the conditional flow map 
$\bPsi_{0\to t}$ directly from the geometry (e.g., geodesics, exponential maps, or premetrics)~\citep{lipman2024flow} which yields analytic, simulation-free target velocities for training.

\paragraph{Conditional Affine Flow (Link to Proposition~\ref{closed-form-vf}).}
We fix a conditioning variable $\rvz\sim\pi$ (e.g., $\rvz=\rvx_1\sim p_{\mathrm{tgt}}$ in one-sided “spotlight’’ training) and push forward $\rvx_0\sim p_{\mathrm{src}}$ through the time-varying \emph{conditional affine flow} 
\[
\bPsi_{0\to t}(\rvx_0;\rvz) := \bm\mu_t(\rvz) + \rmA_t(\rvz) \rvx_0,\qquad t\in[0,1],
\]
where $\bm\mu_t(\rvz)\in\R^D$ and $\rmA_t(\rvz)\in\R^{D\times D}$ is invertible for $t\in(0,1)$. The boundary 
$\rmA_0(\rvz)=\rmI,\ \bm\mu_0(\rvz)=\mathbf 0$ recovers $p_{\mathrm{src}}$ at $t=0$. It is standard to interpret boundary when $t\to1$ as a limit (the terminal map may concentrate mass on a lower-dimensional set or a point)\footnote{Allowing $\rmA_1(\rvz)$ to be singular (e.g., $\mathbf 0$) is compatible with invertibility on $(0,1)$ and causes the path to contract onto the prescribed endpoint at $t=1$.}.

\subparagraph{Induced Conditional Path $p_t(\cdot|\rvz)$.}
The construction defines
\[
p_t(\cdot|\rvz) = \bigl(\bPsi_{0\to t}( \cdot ;\rvz)\bigr)_{\#} p_{\mathrm{src}},
\qquad
p_t(\cdot)  =  \int p_t(\cdot|\rvz)\pi(\rvz)\diff\rvz.
\]
What ultimately matters in $\mathcal{L}_{\mathrm{CFM}}$ is how to sample from it:  
first draw $\rvz \sim \pi$, then draw $\rvx_0 \sim p_{\mathrm{src}}$, and finally set
\[
\rvx_t = \bm\mu_t(\rvz) + \rmA_t(\rvz)\rvx_0.
\]

We remark that when $\bPsi_{0\to t}$ is affine in $\rvx_0$, then $p_t( \cdot | \rvz)$ is Gaussian 
if and only if $p_{\mathrm{src}}$ is Gaussian. 
In particular, for arbitrary (non-Gaussian) $p_{\mathrm{src}}$, an affine flow yields a
generally non-Gaussian $p_t(\cdot|\rvz)$.

\subparagraph{Derived Conditional Velocity $\rvv_t(\cdot|\rvz)$.} The conditional velocity $\rvv_t(\cdot|\rvz)$ is obtained by $t$-differentiating the conditional flow map $\bPsi_{0\to t}$. Following the derivation in Proposition~\ref{closed-form-vf}, consider the \emph{conditional} ODE defined by the flow map $\bPsi_{0\to t}(\rvy;\rvz)$ with initial condition $\rvy$, where the goal is to identify the corresponding conditional velocity field $\rvv_t(\cdot|\rvz)$:
\[
\frac{\diff}{\diff t} \bPsi_{0\to t}(\rvy;\rvz)
= \rvv_t \Big(\underbrace{\bPsi_{0\to t}(\rvy;\rvz)}_{\rvx} \Big| \rvz\Big).
\]
Since $\bPsi_{0\to t}(\cdot;\rvz)$ is invertible for $t\in(0,1)$, we may
express the initial point $\rvy$ corresponding to the current state
$\rvx=\bPsi_{0\to t}(\rvy;\rvz)$ as
\[
\rvy
=
\bPsi_{0\to t}^{-1}(\rvx;\rvz)
=
\bPsi_{t\to0}(\rvx;\rvz).
\]
The corresponding Eulerian velocity field is therefore
\[
\rvv_t(\rvx|\rvz)
:=
\left.
\frac{\partial}{\partial t}
\bPsi_{0\to t}(\rvy;\rvz)
\right|_{\rvy=\bPsi_{0\to t}^{-1}(\rvx;\rvz)}.
\]
That is, we first differentiate the forward flow map with respect to $t$
while holding its initial point $\rvy$ fixed, and then substitute the initial
point whose trajectory reaches $\rvx$ at time $t$.

Since $\rvx_t=\bm\mu_t(\rvz)+\rmA_t(\rvz)\rvx_0$ and $\rmA_t(\rvz)$ is invertible on $(0,1)$, we have $\rvx_0=\rmA_t(\rvz)^{-1}\bigl(\rvx-\bm\mu_t(\rvz)\bigr)$, giving
\[
\rvv_t(\rvx|\rvz)
 = 
\bm\mu_t'(\rvz) + \rmA_t'(\rvz) \rmA_t(\rvz)^{-1} \bigl(\rvx-\bm\mu_t(\rvz)\bigr).
\]

\subparagraph{One-Sided Conditioning ($\rvz=\rvx_1$).}
Choosing $\bm\mu_t(\rvz)=b_t \rvz$ and $\rmA_t(\rvz)=a_t \rmI$ with $a_0=1,a_1=0$ and $b_0=0,b_1=1$ (with $a_t>0$ for $t\in(0,1)$) yields
\[
\rvx_t = a_t \rvx_0 + b_t \rvx_1,
\qquad
\rvv_t(\rvx|\rvx_1) = b_t' \rvx_1 + \frac{a_t'}{a_t} \bigl(\rvx-b_t\rvx_1\bigr).
\]
On paired samples $(\rvx_0,\rvx_1)$ (with $\rvx_t=a_t\rvx_0+b_t\rvx_1$), this simplifies to the usual CFM target:
\[
\rvv_t(\rvx_t|\rvx_1)=a_t' \rvx_0 + b_t' \rvx_1.
\]

\subparagraph{Two-Sided Conditioning ($\rvz=(\rvx_0,\rvx_1)$).}
The same template with $\bm\mu_t(\rvx_0,\rvx_1)=b_t \rvx_1$ and $\rmA_t(\rvx_0,\rvx_1)=a_t \rmI$ makes the conditional path \emph{deterministic}:
\[
\rvx_t=a_t \rvx_0+b_t \rvx_1,
\qquad
p_t(\cdot|\rvx_0,\rvx_1)=\delta\left(\cdot - (a_t\rvx_0+b_t\rvx_1)\right),
\]
and the conditional velocity is
\[
\rvv_t(\rvx_t|\rvx_0,\rvx_1)=a_t' \rvx_0+b_t' \rvx_1,
\]
i.e., the standard two-sided CFM target. 

\subparagraph{Unconditional Gaussian Path as a Special Case.}
If $\bm{\mu}_t$ is independent of $\rvz$ (denoted $\bm{\mu}(t)$) and 
$\rmA_t = \tfrac{\sigma(t)}{\sigma(0)} \rmI$, then
\[
\bPsi_{0\to t}(\rvx_0)=\bm\mu(t)+\sigma(t) \frac{\rvx_0-\bm\mu(0)}{\sigma(0)},
\quad
\rvv_t(\rvx)=\bm\mu'(t)+\frac{\sigma'(t)}{\sigma(t)}\bigl(\rvx-\bm\mu(t)\bigr),
\]
which recovers the Gaussian density path and the closed-form velocity in Proposition~\ref{closed-form-vf}.

\subsection{Probability-Path-First vs. Flow-First Construction}

The two constructions correspond to complementary Eulerian and Lagrangian
descriptions of transport. The \emph{probability-path-first} (Eulerian)
view begins by specifying a conditional density path
$p_t(\cdot|\rvz)$ and then selects a compatible velocity field through the
continuity equation. Because a density path alone generally does not
determine the velocity uniquely, additional transport structure may be
needed to select one.

The \emph{flow-first} (Lagrangian) view instead specifies a conditional
flow map $\bPsi_{0\to t}(\cdot|\rvz)$, which directly describes particle
trajectories. The corresponding density path follows by pushforward, and
when the map is invertible its Eulerian velocity is uniquely recovered as
\[
\rvv_t
=
\partial_t\bPsi_{0\to t}
\circ
\bPsi_{0\to t}^{-1}.
\]
Both constructions therefore describe the same kinds of transport objects
from different starting points, but a prescribed density path need not
select the same velocity as a separately prescribed flow map.

The following table summarizes these contrasts. The takeaway: path-first is natural when conditional paths admit closed-form velocities; flow-first is natural when you have strong structural priors on trajectories.

\begin{table}[th!]
\caption{Comparison of conditional probability-path-first and conditional flow-first constructions.}
\label{tb:conditional-path-vs-flow}
\centering
\begingroup
\scriptsize
\setlength{\tabcolsep}{4pt}
\renewcommand{\arraystretch}{1.05}
\begin{tabular}{p{0.13\linewidth} p{0.41\linewidth} p{0.41\linewidth}}
\toprule
\textbfs{Axis}
&
\textbfs{Conditional Probability-Path-First}
&
\textbfs{Conditional Flow-First}
\\
\midrule

\textbfs{Given}
&
Conditional density path $p_t(\cdot|\rvz)$.
&
Conditional flow map $\bPsi_{0\to t}(\cdot|\rvz)$
(trajectories, for each fixed $\rvz$).
\\

\graymidrule

\textbfs{Get Velocity}
&
For each $\rvz$, find $\rvv_t(\cdot|\rvz)$ s.t.\
$$
\partial_t p_t(\cdot|\rvz)
+
\nabla \cdot
\left(
p_t(\cdot|\rvz)\rvv_t(\cdot|\rvz)
\right)
=
0;
$$
Non-unique: if
$\nabla\!\cdot~\left(p_t \mathbf w_t\right)=0$
then $\rvv_t+\mathbf w_t$ yields the same $p_t$.
&
Along paths (for each $\rvz$):
$$
\rvv_t
\left(
\bPsi_{0\to t}(\cdot|\rvz)
\mid
\rvz
\right)
=
\tfrac{\diff}{\diff t}
\bPsi_{0\to t}(\cdot|\rvz).
$$
When $\bPsi_{0\to t}$ is invertible, one can solve
$$
\rvv_t(\rvx|\rvz)
=
\left.
\partial_t
\bPsi_{0\to t}(\rvy|\rvz)
\right|_{\rvy=\bPsi_{0\to t}^{-1}(\rvx|\rvz)}.
$$
\\

\graymidrule

\textbfs{Closed Form \newline of $\rvv_t(\cdot|\rvz)$}
&
Convenient when $p_t(\cdot|\rvz)$ is Gaussian / exponential-family;
otherwise obtaining $\rvv_t(\cdot|\rvz)$ is nontrivial.
&
Convenient when $\bPsi_{0\to t}(\cdot|\rvz)$ has structure
(affine/low-rank); avoids density evaluation.
\\

\graymidrule

\textbfs{Uniqueness \newline of $\rvv_t(\cdot|\rvz)$}
&
For each $\rvz$, $\rvv_t(\cdot|\rvz)$ is underdetermined unless a
selection rule (e.g., potential flow / min.\ kinetic energy) is imposed.
&
Given $\bPsi_{0\to t}(\cdot|\rvz)$, both
$p_t(\cdot|\rvz)=(\bPsi_{0\to t}(\cdot|\rvz))_\#p_0$
and the particle trajectories are determined. If the flow map is invertible,
the corresponding Eulerian velocity field is uniquely determined; otherwise,
a unique global Eulerian field need not exist.
\\

\graymidrule

\textbfs{Realizability}
&
Must verify the constructed $\rvv_t(\cdot|\rvz)$ solving the continuity
equation on the intended support.
&
Holds by construction:
$$
p_t(\cdot|\rvz)
=
\left(
\bPsi_{0\to t}(\cdot|\rvz)
\right)_\#
p_0(\cdot|\rvz).
$$
\\

\graymidrule

\textbfs{Match \newline $(p_{\mathrm{src}},p_{\mathrm{tgt}})$}
&
Mix conditionals:
\newline
\hspace{3cm}
$p_{\mathrm{src}}
=
\int p_0(\cdot|\rvz)\pi(\rvz)\diff\rvz,$
\newline
$p_{\mathrm{tgt}}
=
\int p_1(\cdot|\rvz)\pi(\rvz)\diff\rvz.$
\newline
Under Gaussian--affine conditional paths with $\rvz$-independent coefficients,
$p_{\mathrm{src}}$ can be forced to be Gaussian.
For a general fixed endpoint $p_{\mathrm{src}}$ (possibly non-Gaussian),
the choice of $p_t(\cdot|\rvz)$ does not generally pin $p_{\mathrm{src}}$.
&
Set $\bPsi_{0\to 0}=\mathrm{Id}$ and choose boundary condition to hit any
$p_{\mathrm{tgt}}$.
\\

\graymidrule

\textbfs{Preferred \newline Scenarios}
&
Diffusion-style constructions; analytic targets via conditional Gaussians
$p_t(\cdot|\rvz)$.
&
Strong structural priors via maps $\bPsi_{0\to t}(\cdot|\rvz)$;
easy boundary control; accommodates singular/low-dimensional endpoints;
natural for map-based regularization/transport costs.
\\

\bottomrule
\end{tabular}
\endgroup
\end{table}

\subsection{Conditional Straightness Does Not Imply Marginal-Flow
Straightness}

The canonical affine interpolation
\[
\rvx_t
=
(1-t)\rvx_0+t\rvx_1
\]
has an appealing geometric property: conditioned on a particular
endpoint pair $(\rvx_0,\rvx_1)$, the interpolation is a straight line
with constant velocity $\rvx_1-\rvx_0$.
This conditional notion of straightness is used extensively in Flow
Matching and Rectified Flow.

It is important, however, to distinguish the straightness of these
\emph{conditional interpolation paths} from the straightness of the
\emph{marginal ODE trajectories} used at generation time.
Under a coupling $\pi(\rvx_0,\rvx_1)$, the population-optimal marginal
velocity is
\[
\rvv^\star(\rvx,t)
=
\mathbb{E}
\left[
\rvx_1-\rvx_0
\,\middle|\,
\rvx_t=\rvx
\right].
\]
The posterior over endpoint pairs generally changes with $(\rvx,t)$.
Consequently, the velocity encountered along a marginal ODE trajectory
need not remain constant, even though every conditional interpolation
used for training is a straight line.

Thus, choosing the canonical affine schedule alone does not imply that
the learned marginal ODE is exactly straight or that one Euler step is
exact. Rectified Flow addresses precisely this distinction: its
rectification or \texttt{Reflow} procedure changes the endpoint coupling
so that the induced causal flow can become progressively closer to a
straight flow~\citep{liu2022flow}.

The following example makes the distinction explicit.

\begin{figure}[th]
        \label{fig:straightness-counterexample}
    \centering
    \includegraphics[width=\linewidth]{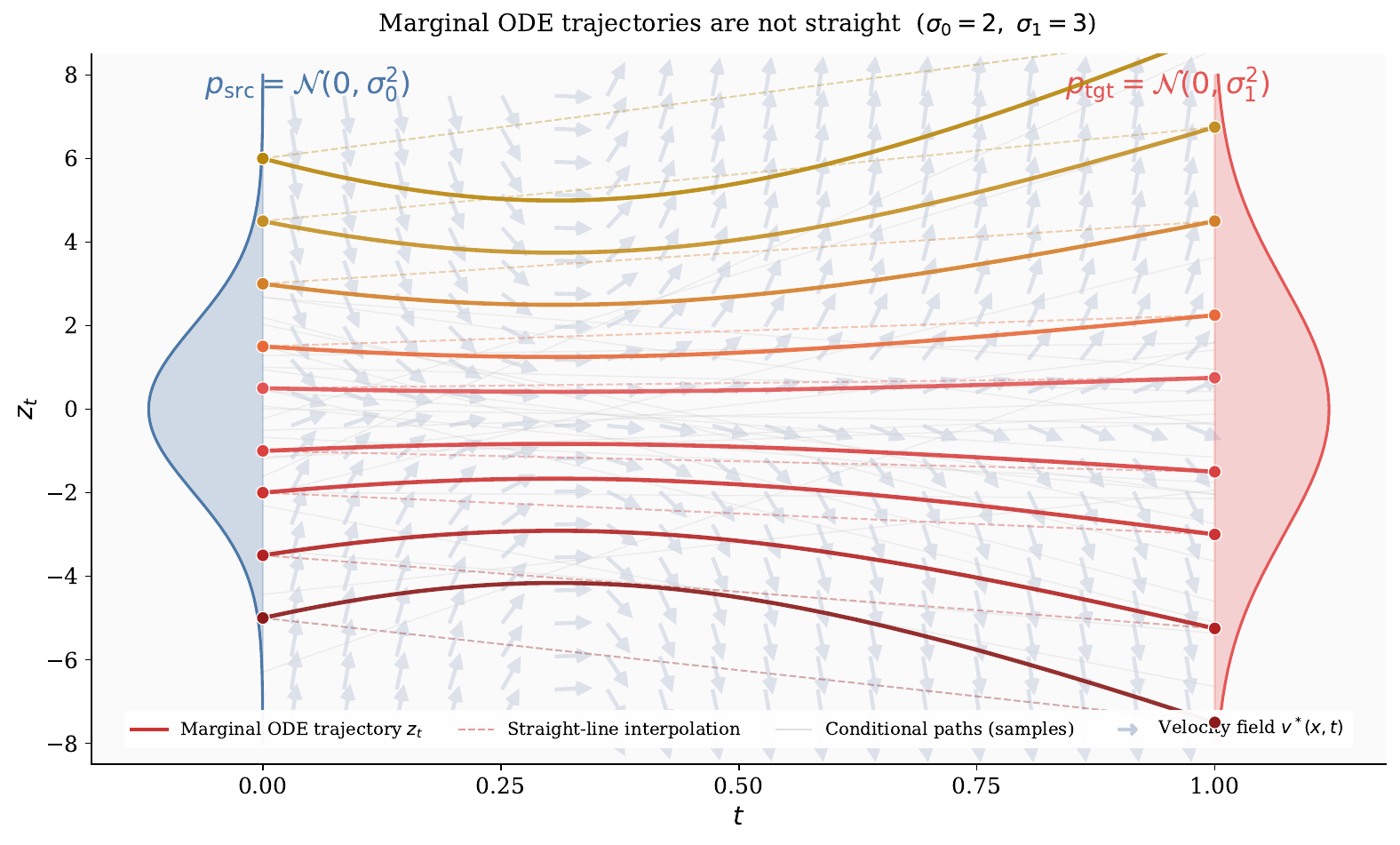}
    \caption{\textbfs{Marginal ODE trajectories are not straight under the canonical affine interpolation.}
    We take $x_0 \sim \mathcal{N}(0, \sigma_0^2)$ and $x_1 \sim \mathcal{N}(0, \sigma_1^2)$ with $\sigma_0=2$, $\sigma_1=3$, and independent coupling.
    Gray lines show sampled conditional paths $(1-t)x_0 + tx_1$, which are straight by construction.
    Solid curves show the exact marginal ODE trajectories $z_t = z_0\sqrt{(1-t)^2 + (\sigma_1/\sigma_0)^2 t^2}$, and dashed lines show the corresponding straight-line interpolations from $z_0$ to $z_1$.
    The visible gap between solid and dashed curves confirms that the marginal ODE trajectories are not affine in $t$, even though every conditional path is a straight line.
    The marginal densities $p_{\mathrm{src}}$ and $p_{\mathrm{tgt}}$ are shown on the left and right margins, respectively.
    \figcredit{Created by the authors with AI-assisted coding.}}
\end{figure}

\exm{A Simple 1D Counterexample}{
Consider the one-dimensional setting with
\[
x_0 \sim \mathcal{N}(0, \sigma_0^2),
\qquad
x_1 \sim \mathcal{N}(0, \sigma_1^2),
\]
independent coupling, and the canonical affine interpolation
$x_t = (1-t)x_0 + tx_1$.
Each conditional path $(1-t)x_0 + tx_1$ is a straight line in
time. We now compute the induced marginal velocity field.

Since $(x_1 - x_0,\, x_t)$ is jointly Gaussian, the
conditional expectation is linear in $x$, so
\[
v^*(x,t)
= \E[x_1 - x_0| x_t = x]
= \frac{\operatorname{Cov}(x_1 - x_0,\, x_t)}
       {\operatorname{Var}(x_t)}\, x.
\]
A direct calculation gives
\[
\operatorname{Var}(x_t)
= (1-t)^2\sigma_0^2 + t^2\sigma_1^2
=: V(t),
\]
and
\[
\operatorname{Cov}(x_1 - x_0,\, x_t)
= t\sigma_1^2 - (1-t)\sigma_0^2
= \tfrac{1}{2} V'(t).
\]
Hence
\[
v^*(x,t) = \frac{V'(t)}{2V(t)}\, x.
\]

The marginal ODE
\[
\frac{\diff z_t}{\diff t} = v^*(z_t, t)
\]
therefore has the exact solution
\[
z_t
= z_0 \sqrt{\frac{V(t)}{V(0)}}
= z_0 \sqrt{(1-t)^2 + \frac{\sigma_1^2}{\sigma_0^2}\, t^2}.
\]

Now, a straight-line interpolation in time would require $z_t$ to be
affine in $t$, namely $z_t = (1-t)z_0 + t\,z_1$ for some endpoint
$z_1$. But
\[
\sqrt{(1-t)^2 + \frac{\sigma_1^2}{\sigma_0^2}\, t^2}
\]
is not affine in $t$ for any positive $\sigma_0, \sigma_1$. Thus the ODE
trajectory is generally not a straight-line interpolation in time. We illustrate this phenomenon in \Cref{fig:straightness-counterexample}
with the example $\sigma_0=2$ and $\sigma_1=3$.

For example, when $\sigma_0 = \sigma_1$, we have $z_1 = z_0$, so a
straight-line interpolation would simply be the constant path
$z_t = z_0$. In contrast, the actual ODE trajectory satisfies
\[
z_{1/2} = \frac{z_0}{\sqrt{2}} \neq z_0.
\]
Therefore, even though every conditional interpolation path is straight,
the induced marginal ODE trajectory need not be.
}

\clearpage
\newpage

\section{\texorpdfstring{(Optional) Properties of the Canonical Affine Flow}{(Optional) Properties of the Canonical Affine Flow}}\label{sec:optimal-canonical}

Given two endpoint distributions $p_0=p_{\mathrm{src}}$ and $p_1=p_{\mathrm{tgt}}$, 
a natural and widely used choice for defining the conditional path in flow matching (FM)~\citep{lipman2022flow} 
and rectified flow (RF)~\citep{liu2022flow} is the linear interpolation
\[
a_t = 1 - t, \qquad b_t = t,
\]
which yields the interpolant
\[
\mathbf{x}_t = (1 - t)\mathbf{x}_0 + t \mathbf{x}_1, 
\qquad \mathbf{x}_0 \sim p_{\mathrm{src}},\ \mathbf{x}_1 \sim p_{\mathrm{tgt}}.
\]
Under this choice, the training objective simplifies to
\[
\mathbb{E}_{t \sim \mathcal{U}[0,1]}   
\mathbb{E}_{\mathbf{x}_0, \mathbf{x}_1} 
\left[ \bigl\|\mathbf{v}_{\bm{\phi}}(\mathbf{x}_t, t) - (\mathbf{x}_1 - \mathbf{x}_0)\bigr\|_2^2 \right].
\]

This linear flow enjoys several appealing properties. 
In particular, it admits an iterative refinement scheme, known as \emph{\texttt{Reflow}}, 
which progressively straightens the path between distributions while preserving the marginals.

\subsection{Rectifying Flows: From Noisy Guesses to Structured Pairings}\label{subsec:rectify}

\paragraph{Why Change the Coupling?}
Under the independent coupling
\[
\pi(\rvx_0,\rvx_1)
=
p_{\mathrm{src}}(\rvx_0)
p_{\mathrm{tgt}}(\rvx_1),
\]
each conditional interpolation
$(1-t)\rvx_0+t\rvx_1$ is a straight line. As emphasized in
\Cref{sec:optimal-canonical}, however, the corresponding marginal velocity
averages directions from many unrelated endpoint pairs, so the induced
marginal ODE trajectories need not be straight. Such curved marginal dynamics
can be harder to integrate accurately with only a few numerical steps.

Rectification addresses this issue by changing the endpoint coupling. A
learned ODE transports each source sample to a model-generated endpoint,
producing a dependent coupling that is subsequently used to define a new
conditional interpolation.

\paragraph{Rectify the Flow via Dependent Coupling.}
Rather than relying on arbitrary pairings, we use a pre-trained diffusion model $\rvv_{\bphi^\times}(\cdot,t)$ as the drift in a PF-ODE to \emph{deterministically} transport each source point. Starting from $\rvz(0)=\rvz_0 \sim p_{\text{src}}$, we integrate
\[
\frac{\diff \rvz(t)}{\diff t} = \rvv_{\bphi^\times}(\rvz(t),t), \qquad t \in [0,1],
\]
to obtain $\hat\rvz_1 := \rvz(1)$ positioned near the data space learned from the pre-trained model. The resulting pair $(\rvz_0,\hat\rvz_1)$ forms a \emph{dependent coupling}: it follows a structured, model-guided path rather than an arbitrary interpolation. This idea extends naturally to affine reference paths of the form $\rvx_t = a_t \rvx_0 + b_t \rvx_1$, where $\rvx_0 \sim p_{\text{src}}$ and $\rvx_1 \sim p_{\text{tgt}}$.

\begin{algorithm}[H]
\caption{\texttt{Rectify} Operation\label{alg:Rectify}}
\begin{algorithmic}[1]
\Require Reference path $\{\rvx_t\}_{t\in[0,1]}$ (e.g.\ $\rvx_t=a_t\rvx_0+b_t\rvx_1$)
\State \textbfs{Pre-Train Diffusion.} Fit $\rvv_{\bphi^\times}$ on the chosen path by minimizing
\[
\bphi^\times \in \arg\min_{\bphi} 
\mathbb{E}_{t,\rvx_0,\rvx_1} \left[\Big\|\rvv_{\bphi}(\rvx_t,t)-\frac{\diff \rvx_t}{\diff t}\Big\|_2^2\right].
\]
\State \textbfs{Rectify.} Sample $\rvz_0 \sim p_{\text{src}}$ and integrate
\[
\frac{\diff \rvz(t)}{\diff t}=\rvv_{\bphi^\times}(\rvz(t),t),\qquad \rvz(0)=\rvz_0,\quad t\in[0,1],
\]
to obtain $\hat\rvz_1=\rvz(1)$ and the trajectory $\{\rvz(t)\}_{t\in[0,1]}$.
\Ensure Dependent (coherent) pair $(\rvz_0,\hat\rvz_1)$ or the full trajectory.
\end{algorithmic}
\end{algorithm}

\paragraph{Why It Works: Marginal-Preserving Structure.}
Let $\bPhi_{0\to t}$ denote the flow map generated by the above ODE defined by the pre-trained diffusion $\rvv_{\bphi^\times}$; then $\rvz(t)=\bPhi_{0\to t}(\rvz_0)$ and $\hat\rvz_1=\bPhi_{0\to 1}(\rvz_0)$. The \texttt{Rectify} procedure pairs each source point with its flow endpoint, giving the deterministic joint
\[
\pi_{\text{Rectify}}(\rvz_0,\rvz_1)
= p_{\text{src}}(\rvz_0) \delta \big(\rvz_1-\bPhi_{0\to 1}(\rvz_0)\big).
\]
We have two immediate consequences:
\begin{itemize}
    \item \textbfs{Source Marginal is Preserved:} 
    $\displaystyle \int \pi_{\text{Rectify}}(\rvz_0,\rvz_1) \diff \rvz_1
    = p_{\text{src}}(\rvz_0)$.
    \item \textbfs{Pushforward Along the Flow:} 
    $\displaystyle (\bPhi_{0\to t})_{\#}p_{\text{src}}=\mathrm{Law}(\rvz(t))$, i.e., the time–$t$ distribution is the pushforward of $p_{\text{src}}$ by $\bPhi_{0\to t}$.
\end{itemize}
If $\rvv_{\bphi^\times}$ matches the oracle drift of a given reference path $\rvx_t$, then all intermediate marginals coincide:
\[
\mathrm{Law}(\rvz(t))=\mathrm{Law}(\rvx_t),\quad \text{for all }t\in[0,1],
\quad\text{and}\quad
(\bPhi_{0\to 1})_{\#}p_{\mathrm{src}}=p_{\mathrm{tgt}}.
\]

\paragraph{Summary.}
Rectification replaces the independent endpoint coupling with a coupling
induced by the learned flow. Under repeated \texttt{Reflow}, this can produce
marginal transport that is progressively easier to integrate with few
numerical steps.

For the canonical path, repeatedly applying \texttt{Rectify} (``\emph{\texttt{Reflow}}'') further straightens trajectories without increasing transport cost, making training still easier.

\subsection{\texttt{Reflow}: Iteratively Straightening Flows}\label{subsec:reflow-def}

\paragraph{Why \texttt{Reflow}?}
Independent pairings often induce irregular and meandering ODE trajectories between $p_{\text{src}}$ and $p_{\text{tgt}}$, which increase discretization error and variance during simulation. This raises a natural question:

\begin{question}\label{ques:straight}
Can we learn couplings that induce transport paths that are closer to straight lines between the two distributions, while still preserving the correct marginals?
\end{question}

This motivates \emph{\texttt{Reflow}}: repeatedly apply \texttt{Rectify} to update the coupling so that successive flows become easier to integrate.

\paragraph{Core Idea: Recursive Straightening via \texttt{Rectify}.}

Start from the canonical interpolation on the product coupling
$\pi^{(0)}:=p_{\text{src}}(\rvx_0)p_{\text{tgt}}(\rvx_1)$,
\[
\rvx_t=(1-t)\rvx_0+t\rvx_1.
\]
Applying \texttt{Rectify} replaces the independent pairing with a dependent
one $(\rvz_0,\hat\rvz_1)$ induced by the learned ODE. Repeating this procedure,
known as \emph{\texttt{Reflow}}, tends to produce straighter trajectories that
are easier to integrate accurately with a small number of numerical steps.
The precise straightening guarantee is discussed later in
\Cref{subsec:reflow-properties}.

\paragraph{The \texttt{Reflow} Procedure.}
Each iteration performs two steps:
\begin{itemize} 
    \item \textbfs{Re-Fit Flow:} Train a new velocity field from samples of the current coupling: \begin{align} \bm{\phi}_{k+1} &= \arg\min_{\bm{\phi}} \mathcal{L}\left(\bm{\phi}\Big\vert \pi^{(k)}\right), \quad \text{where} \notag \\ \mathcal{L}\left(\bm{\phi}\Big\vert \pi^{(k)}\right) &:= \mathbb{E}_{t, (\rvz_0^{(k)}, \hat\rvz_1^{(k)})\sim \pi^{(k)}} \left[\left\|\rvv_{\bm{\phi}}(\rvz_t, t) - (\hat\rvz_1^{(k)} - \rvz_0^{(k)})\right\|^2\right] \label{eq:reflow-loss} \end{align} with $\rvz_t = (1-t)\rvz_0^{(k)} + t\hat\rvz_1^{(k)}$. 
    \item \textbfs{Generate New Coupling:} Solve the learned ODE starting from new source samples $\rvz_0^{(k+1)} \sim p_{\text{src}}$: \[ \hat\rvz_1^{(k+1)} \gets \rvz_0^{(k+1)} + \int_0^1 \rvv_{\bm{\phi}_{k+1}}(\rvz(t), t) \mathrm{d}t, \] and define the updated coupling: \[ \pi^{(k+1)}(\rvz_0, \rvz_1) := p_{\text{src}}(\rvz_0) \delta\left(\rvz_1 - \hat\rvz_1^{(k+1)}\right). \] \end{itemize}
In other words, \texttt{Reflow} can be viewed as repeatedly applying the \texttt{Rectify} operator, producing a sequence of progressively refined couplings:
\begin{align}\label{eq:rectify-op}
    \pi^{(k+1)} = \texttt{Rectify} \left(\pi^{(k)}\right)
\end{align}
so that both the flow and the coupling evolve together, yielding progressively more stable transport paths.

\subsection{Properties of \texttt{Reflow}}\label{subsec:reflow-properties}

Two key theoretical properties drive the usefulness of \texttt{Reflow}: it reduces transport cost and it straightens the trajectories.

\paragraph{I. \texttt{Reflow} Never Increases Transport Cost.}
Let $c(\rvy)$ be a convex cost function (e.g., $\|\rvy\|_2^p$ with $p \geq 1$). Each \texttt{Rectify} step forms a new coupling $(\rvz_0, \hat\rvz_1)$ whose cost is no worse than the original:
\proppp{\texttt{Rectify} May Reduce Transport Costs}{rec-transport-cost}{
Assuming an ideal velocity field $\rvv^* = \rvv_{\bm{\phi}^\times}$, we have:
\[
\mathbb{E}\left[c\left(\hat\rvz_1 - \rvz_0\right)\right] \leq \mathbb{E}\left[c\left(\rvx_1 - \rvx_0\right)\right].
\]
}{Follows from Jensen's inequality. See \citet{liu2022flow} for a full derivation.}
Applying this result recursively shows that the \texttt{Reflow} process does not increase the transport cost.

\paragraph{II. \texttt{Reflow} Straightens the Path.}
Recursive application of \texttt{Reflow} promotes increasingly straight
trajectories, although the straightness measure is not guaranteed to improve
monotonically at every iteration. To measure this, define the \emph{straightness functional} of a path $\mathbf{Y} = \{\rvy_t\}_{t\in[0,1]}$ as
\[
\mathcal{S}(\mathbf{Y}) := \int_0^1 \mathbb{E}\left[\left\|\rvy_1 - \rvy_0 - \frac{\mathrm{d}\rvy_t}{\mathrm{d}t} \right\|_2^2 \right]  \mathrm{d}t.
\]
If $\mathcal{S}(\mathbf{Y}) = 0$, then $\mathbf{Y}$ is exactly a straight line.

\proppp{\texttt{Reflow} Straightens the Stochastic Path}{reflow-straight}{
For rectified paths $\rmZ^{(k)}$, we have:
\[
\min_{k \in \{0,\dots,K\}} \mathcal{S}(\rmZ^{(k)}) \leq \frac{\mathbb{E}\left[\|\rvx_1 - \rvx_0\|^2\right]}{K}.
\]
}{See Theorem 3.7 of \citet{liu2022flow}.}

\begin{figure}[t]
    \centering
    \includegraphics[width=\textwidth]{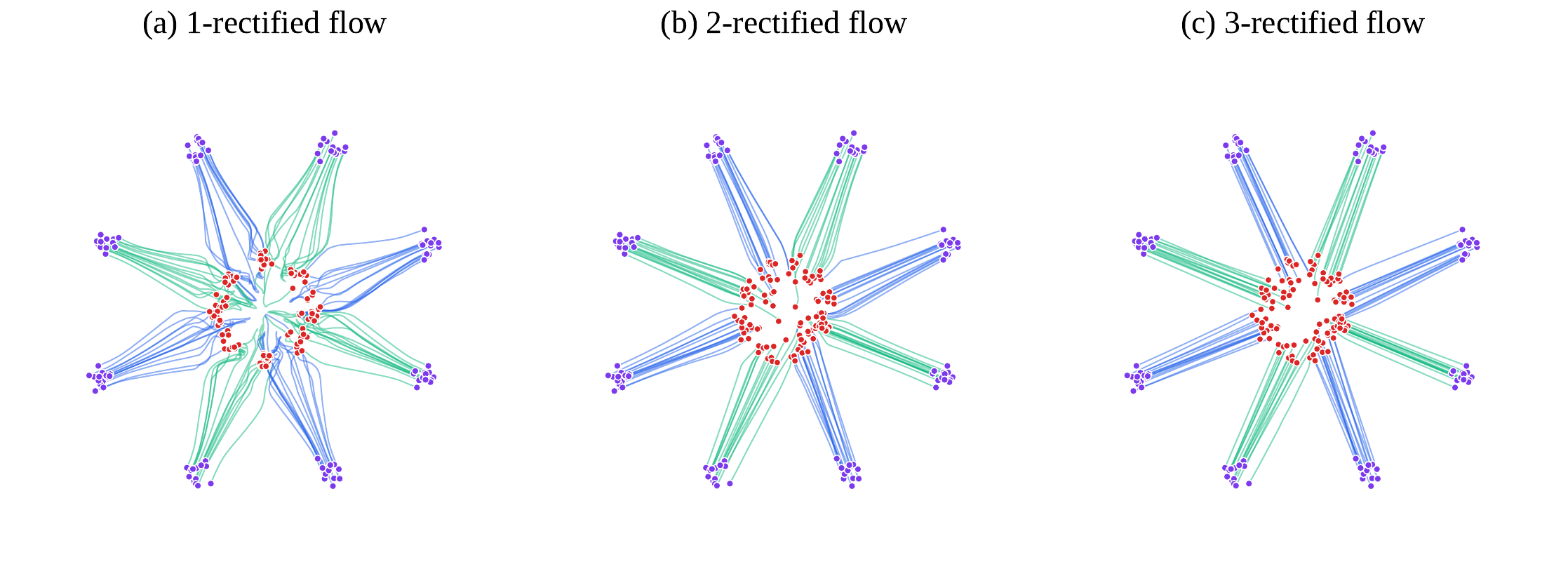}
    \caption{\textbfs{Illustration of \texttt{Reflow}.} Paths become progressively straighter with \texttt{Rectify} procedure.  \figcredit{Adapted from \citet{liu2022flow}.}}
    \label{fig:ill_Reflow}
\end{figure}

The FM or RF formulation with linear interpolation kernels, together with the \texttt{Reflow} procedure, provides a simpler training objective and a practical method for refining stochastic couplings. For theoretical details, we refer readers to \citep{liu2022flow, liu2022rectified}.

\paragraph{III. Connection to Optimal Transport.}
Lastly, we note that straight-line couplings are not necessarily optimal in the sense of optimal transport (OT). This involves some terminology that will be introduced in \Cref{sec:ot-intro}; we therefore refer readers who are not familiar with OT to that section.

A hallmark of quadratic-cost optimal transport is that particles travel along straight lines: a particle at $\rvx_0$ moves to $\rmT(\rvx_0)$ via $\rvx_t = (1 - t)\rvx_0 + t \rmT(\rvx_0)$, where $\rmT$ is the optimal transport map. However, not every map $\rmS$ generating such straight-line paths, i.e., $\rvx_t = (1 - t)\rvx_0 + t \rmS(\rvx_0)$, is optimal. The map $\rmS$ yields the optimal flow only if it minimizes the Monge cost $\mathbb{E}[\|\rvx_0 - \rmS(\rvx_0)\|^2]$. Thus, while straight-line paths are necessary, they are not sufficient; optimality also depends on the correct endpoint map $\rmT$.

\exm{Straight Couplings Need Not Be Optimal}{
Let $p_{\text{src}} = p_{\text{tgt}} = \mathcal{N}(\bm{0}, \rmI)$. For the cost $c(\rvx, \rvy) = \|\rvx - \rvy\|^p$ with $p > 0$, the $c$-optimal coupling is the identity coupling $\pi^{*}$, where $\pi^{*}$ is the law of $(\rvx, \rvx)$ with $\rvx \sim p_{\text{src}}$.

Now consider the coupling $\pi_{\rmA}$ defined as the law of $(\rvx, \rmA \rvx)$, where $\rvx \sim p_{\text{src}}$ and $\rmA$ is a rotation matrix satisfying $\rmA^\top \rmA = \rmI$, $\det(\rmA) = 1$, $\rmA \neq \rmI$, and $-1$ is not an eigenvalue. Then $\pi_{\rmA}$ is a valid coupling of $p_{\text{src}}$ and $p_{\text{tgt}}$, and corresponds to straight-line paths between $\rvx$ and $\rmA \rvx$, but it is not $c$-optimal for any twice-differentiable strictly convex cost $c$ with invertible Hessian. The suboptimality arises from the rotational transformation. As discussed in \Cref{eq:reflow-loss-gradient}, even removing the rotation may not lead to an optimal coupling.
}

We will continue exploring the connection to OT in \Cref{subsec:ot-linear-reflow}.

\clearpage
\newpage

\newpage
\section{Closing Remarks}\label{sec:ch5_cr}
     
This chapter has illuminated the third and final foundational perspective on diffusion models, one rooted in the principles of deterministic flows. Our exploration began with Normalizing Flows (NFs), which leverage the change-of-variables formula to learn an exact, invertible mapping between a simple prior and the data distribution. We then saw this concept evolve into a continuous-time process with Neural ODEs, where a learned velocity field governs the transformation. However, this approach comes with the significant drawback of requiring costly ODE simulations within the training loop.

The modern framework of Flow Matching (FM) was presented as an elegant and efficient solution to this challenge. By pre-defining a probability path $\{p_{t}\}_t$ and a corresponding velocity field that satisfies the continuity equation, FM establishes a clear target for the ODE flow. Crucially, just as we saw in the variational and score-based views, FM employs a powerful conditioning trick. This transforms the intractable problem of matching the marginal velocity field into a simple and tractable regression against a known conditional velocity, making training entirely simulation-free. This perspective recasts diffusion models themselves as a special case of learning a deterministic flow to transport a Gaussian prior to the data distribution.

With the introduction of the flow-based view, our survey of the three conceptual pillars of diffusion modeling is now complete. Throughout this journey, a remarkable pattern has emerged: each framework, despite its unique origins in VAEs, EBMs, or NFs, has converged on a continuous-time generative process and has relied on a conditioning strategy to enable tractable learning.

In the next chapter, we will finally synthesize these parallel threads into a single, unified framework. We will:
\begin{enumerate}
    \item Formally demonstrate that the variational, score-based, and flow-based perspectives are not merely analogous but are mathematically equivalent at a fundamental level.
    \item Show how the Fokker-Planck equation serves as the universal law governing density evolution across all three views, revealing that they are simply different lenses for describing the same core generative principle.
\end{enumerate}

This unified lens will provide a complete and systematic understanding of the modern diffusion paradigm.

\chapter{A Unified and Systematic Lens on Diffusion Models}\label{ch:all-equivalent} 

\epigraph{
    \textit{Mathematics is the art of giving the same name to different things.
}}{Henri Poincaré}

This chapter presents a systematic viewpoint that connects the variational, score-based, and flow-based perspectives within a coherent picture. Although motivated by different intuitions, these approaches converge on the same core mechanism underlying modern diffusion models: define a forward corruption process that traces a path of marginal distributions, then learn the local
quantities needed to construct generative dynamics that follow this path in reverse. Building on \Cref{ch:variational,ch:score-based,ch:score-sde,ch:flow-based}, we observe a common recipe: define a forward corruption process that traces a path of marginals, then learn a time varying vector field that transports a simple prior to the data distribution along this path.

A key ingredient across all perspectives is the conditioning trick introduced in \Cref{sec:conditional-trick}, which transforms an intractable marginal objective into a tractable conditional one, leading to stable and efficient training.

In \Cref{sec:elucidating} we analyze the training objective in a systematic way, identifying its essential components and clarifying how loss functions are formulated in the variational, score-based, and flow-based viewpoints.

\Cref{sec:equivalent-parametrizations} shows that any affine forward noise injection of the form $\rvx_t = \alpha_t \rvx_0 + \sigma_t \beps$ can be equivalently transformed into the standard linear schedule $\rvx_t = (1 - t)\rvx_0 + t\beps$. Moreover, common parameterizations such as noise prediction, clean data prediction, score prediction, and velocity prediction are interchangeable at the level of gradients. Thus, the choices of noise schedulers and parameterizations both adhere to the same  modeling principle.

Finally, \Cref{sec:comparison-connection} brings the discussion together and identifies the governing rule: the Fokker–Planck equation. Whether viewed as a variational scheme (discrete time denoising), a score-based method (SDE formulation), or a flow-based method (ODE formulation), each constructs a generator whose marginals follow the same density evolution. The Fokker–Planck equation thus serves as the universal constraint respected by all three viewpoints, with differences arising only in parameterization and training objectives.

\chapterlearning{
  \item recognize the common conditioning principle behind variational, score-based, and flow-based training: an intractable marginal target can be learned through tractable conditional targets;
  \item understand the four common prediction types---clean data, noise, score, and velocity---and convert between their oracle predictions when the forward schedule is nondegenerate;
  \item disentangle the main design choices in diffusion training: the forward noise schedule, prediction target, time-sampling distribution, and time weighting, and understand which changes are mathematical reparameterizations and which can still affect practical optimization;
  \item distinguish different notions of ``equivalence'': representing the same oracle information, having equivalent population objectives up to weighting or additive constants, tracing the same marginal density path, and behaving identically as a practical training or sampling algorithm;
  \item understand how different affine probability paths are related through time reparameterization and spatial rescaling, while keeping track of what these transformations do and do not preserve;
  \item understand how the continuity and Fokker--Planck equations provide the density-level principle connecting variational, deterministic ODE, and stochastic SDE descriptions of diffusion models.
}{
This chapter is the conceptual keystone of the book. For a first pass, focus
on the conditioning principle, the relationships among the four prediction
types, and the continuity/Fokker--Planck view that connects the three
perspectives. The central lesson is not to memorize every conversion formula,
but to ask what notion of equivalence is being used and what that equivalence
actually preserves. The detailed analysis of weighting, schedules, and
parameterization choices can then be revisited to understand why mathematically
equivalent formulations may still behave differently in practice.
}

\newpage

\section{Conditional Tricks: The Secret Sauce of Diffusion Models}\label{sec:conditional-trick}
Until now, we have explored diffusion models from three seemingly distinct origins: variational, score-based, and flow based perspectives. Each was originally motivated by different goals and led to its own training objectives (with a fixed $t$):
\begin{itemize}
    \item \textbfs{Variational View:} Learn a parametrized density $p_{\bm{\phi}}(\mathbf{x}_{t-\Delta t}|\mathbf{x}_t)$ to approximate the oracle reverse transition $p(\mathbf{x}_{t-\Delta t}|\mathbf{x}_t)$ by minimizing:
    \[
    \mathcal{J}_{\text{KL}}(\bm{\phi}) := \mathbb{E}_{p_t(\mathbf{x}_t)}\left[
        \mathcal{D}_{\mathrm{KL}}\big(p(\mathbf{x}_{t-\Delta t}|\mathbf{x}_t) \| p_{\bm{\phi}}(\mathbf{x}_{t-\Delta t}|\mathbf{x}_t)\big)
    \right];
    \]
    \item \textbfs{Score-Based View:} Learn a score model $\rvs_{\bm{\phi}}(\rvx_t, t)$ to approximate the marginal score $\nabla_{\rvx} \log p_t(\rvx_t)$ via:
    \[
    \mathcal{J}_{\text{SM}}(\bm{\phi}) := \mathbb{E}_{p_t(\rvx_t)}\left[
       \left\| \rvs_{\bm{\phi}}(\rvx_t, t) - \nabla_{\rvx} \log p_t(\rvx_t) \right\|_2^2
    \right];
    \]
    \item \textbfs{Flow-Based View:} Learn a velocity model $\rvv_{\bm{\phi}}(\rvx_t, t)$ to match the oracle velocity $\rvv_t(\rvx_t)$ (e.g., defined by \Cref{eq:marginal-oracle-velocity}) by minimizing:
    \[
    \mathcal{J}_{\text{FM}}(\bm{\phi}) := \mathbb{E}_{p_t(\rvx_t)} \left[
        \left\| \rvv_{\bm{\phi}}(\rvx_t, t) - \rvv_t(\rvx_t) \right\|_2^2
    \right].
    \]
\end{itemize}

At first glance, these objectives seem hopelessly intractable, since they all require access to oracle quantities that are fundamentally unknowable in general. But here comes the exciting twist: each method independently arrives at the same elegant solution to this problem: \emph{conditioning on the data $\rvx_0$}. This technique transforms each intractable training target into a tractable one.

This elegant ``conditioning technique'' rewrites the objectives as expectations
over the known Gaussian conditionals $p_t(\rvx_t | \rvx_0)$, yielding
tractable conditional targets and training objectives that are
gradient-equivalent to their marginal counterparts:
\begin{itemize}
    \item \textbfs{Variational View} (\Cref{eq:kl-matching})\textbfs{:}
    \[
    \mathcal{J}_{\text{KL}}(\bm{\phi}) = \underbrace{\mathbb{E}_{\rvx_0} \mathbb{E}_{p_t(\rvx_t|\rvx_0)} \left[
        \mathcal{D}_{\mathrm{KL}}\big(p(\rvx_{t-\Delta t}|\rvx_t, \rvx_0) \| p_{\bm{\phi}}(\rvx_{t-\Delta t}|\rvx_t)\big)
    \right]}_{\mathcal{J}_{\text{CKL}}(\bm{\phi})} + C;
    \]    
    \item \textbfs{Score-Based View} (\Cref{eq:score-matching})\textbfs{:}
    \[
    \mathcal{J}_{\text{SM}}(\bm{\phi}) = \underbrace{\mathbb{E}_{\rvx_0} \mathbb{E}_{p_t(\rvx_t|\rvx_0)} \left[
        \left\| \rvs_{\bm{\phi}}(\rvx_t, t) - \nabla_{\rvx_t} \log p_t(\rvx_t | \rvx_0) \right\|_2^2
    \right]}_{\mathcal{J}_{\text{DSM}}(\bm{\phi})} + C;
    \]
    \item \textbfs{Flow-Based View} (\Cref{eq:fm-cfm})\textbfs{:}
    \[
    \mathcal{J}_{\text{FM}}(\bm{\phi}) = \underbrace{\mathbb{E}_{\rvx_0} \mathbb{E}_{p_t(\rvx_t|\rvx_0)} \left[
        \left\| \rvv_{\bm{\phi}}(\rvx_t, t) - \rvv_t(\rvx_t | \rvx_0) \right\|^2
    \right]}_{\mathcal{J}_{\text{CFM}}(\bm{\phi})} + C.
    \]
\end{itemize}
To build a unified view, we next revisit the conditional KL, score, and velocity objectives in a systematic manner. Crucially, these objectives are not only tractable but also equivalent to their original forms up to a constant vertical shift. 
The conditional versions ($\mathcal{J}_{\text{CKL}}$,
$\mathcal{J}_{\text{DSM}}$, $\mathcal{J}_{\text{CFM}}$) differ from the
originals ($\mathcal{J}_{\text{KL}}$, $\mathcal{J}_{\text{SM}}$,
$\mathcal{J}_{\text{FM}}$) only by a constant that is independent of the
model parameters. Consequently, each conditional objective has the same
minimizers as its original counterpart within the same model class.

To see what these minimizers ideally recover, suppose the model is expressive
enough to represent the exact oracle. Despite their different objectives, all
three perspectives then share the same conceptual pattern:
\[
\underbrace{\text{marginal oracle}}_{\text{generally intractable}}
=
\mathbb{E}_{\rvx_0\sim p(\cdot|\rvx_t)}
\left[
\underbrace{\text{conditional oracle}}_{\text{tractable given }\rvx_0}
\right].
\]
That is, conditioning on the clean sample $\rvx_0$ gives a tractable target,
and averaging these conditional targets over the possible clean samples
consistent with $\rvx_t$ recovers the desired marginal oracle. Concretely,

\begin{mdframed}
    \begin{align}\label{eq:three-minimizer-oracle}
\begin{aligned}
        p^*(\mathbf{x}_{t-\Delta t}|\mathbf{x}_t) &= \mathbb{E}_{\mathbf{x}_0 \sim p(\cdot|\mathbf{x}_t)} \big[p(\mathbf{x}_{t-\Delta t}|\mathbf{x}_t, \mathbf{x}_0)\big] &&= p(\mathbf{x}_{t-\Delta t}|\mathbf{x}_t), \\
    \mathbf{s}^*(\mathbf{x}_t, t) &= \mathbb{E}_{\mathbf{x}_0 \sim p(\cdot|\mathbf{x}_t)} \big[\nabla_{\mathbf{x}_t} \log p_t(\mathbf{x}_t|\mathbf{x}_0)\big] &&= \nabla_{\mathbf{x}_t} \log p_t(\mathbf{x}_t), \\
    \mathbf{v}^*(\mathbf{x}_t, t) &= \mathbb{E}_{\mathbf{x}_0 \sim p(\cdot|\mathbf{x}_t)} \big[\mathbf{v}_t(\mathbf{x}_t|\mathbf{x}_0)\big] &&= \mathbf{v}_t(\mathbf{x}_t).
\end{aligned}
\end{align}
\end{mdframed}

For the variational view, the first identity follows from marginalizing the
conditional reverse transition over $\rvx_0$, and the KL objective is minimized
by the resulting marginal reverse transition. For score and velocity matching,
the squared-error objectives are least-squares regressions whose population
minimizers are the corresponding conditional expectations. These identities
describe the unrestricted population oracles; a restricted neural model need
not represent them exactly. Below is a concrete example of a scenario with a finite number of data points.

\exm{Closed-Form Oracle Targets on a Finite Dataset}{
To make \Cref{eq:three-minimizer-oracle} more concrete, suppose the data distribution is the empirical measure
\[
\hat p_0=\frac{1}{N}\sum_{i=1}^N \delta_{\rvx_0^{(i)}},
\]
where $\{\rvx_0^{(i)}\}_{i=1}^N$ denotes the finite training set. 

In \Cref{eq:three-minimizer-oracle}, the notation
\[
\rvx_0 \sim p(\cdot|\rvx_t)
\]
means that we average over the posterior distribution of the clean sample given the noisy observation $\rvx_t$. Equivalently, for any quantity $\rvh(\rvx_0)$,
\[
\mathbb{E}_{\rvx_0 \sim p(\cdot|\rvx_t)}[\rvh(\rvx_0)]
=
\int \rvh(\rvx_0)\, p(\rvx_0|\rvx_t)\diff\rvx_0.
\]

By Bayes' rule,
\[
p(\rvx_0|\rvx_t)
=
\frac{p_t(\rvx_t|\rvx_0)\,p_{\text{data}}(\rvx_0)}
{\int p_t(\rvx_t|\tilde{\rvx}_0)\,p_{\text{data}}(\tilde{\rvx}_0)\diff\tilde{\rvx}_0}.
\]
Under the standard Gaussian diffusion forward process,
\[
p_t(\rvx_t|\rvx_0^{(i)})
=
\mathcal{N}\left(\rvx_t;\alpha_t \rvx_0^{(i)}, \sigma_t^2 \rmI\right),
\]
so the conditional likelihood is explicit. However, at the population level, the posterior $p(\rvx_0|\rvx_t)$ still requires averaging over the full data distribution $p_{\text{data}}$.

If we now replace $p_{\text{data}}$ by the empirical measure
\[
\hat p_0=\frac{1}{N}\sum_{i=1}^N \delta_{\rvx_0^{(i)}},
\]
then the posterior $p(\rvx_0|\rvx_t)$ is supported only on the training examples, and its mass at $\rvx_0^{(i)}$ is
\[
p(\rvx_0=\rvx_0^{(i)}|\rvx_t)
=
\frac{p_t(\rvx_t|\rvx_0^{(i)})}
{\sum_{j=1}^N p_t(\rvx_t|\rvx_0^{(j)})}
=: w_i(\rvx_t,t),
\qquad
\sum_{i=1}^N w_i(\rvx_t,t)=1.
\]
Hence the posterior expectation above reduces to the finite sum
\[
\mathbb{E}_{\rvx_0 \sim p(\cdot|\rvx_t)}[\rvh(\rvx_0)]
=
\sum_{i=1}^N w_i(\rvx_t,t)\rvh(\rvx_0^{(i)}).
\]
Thus, in the finite-dataset case, the abstract posterior averaging in \Cref{eq:three-minimizer-oracle} becomes a concrete weighted average over the training set, with weights $w_i(\rvx_t,t)$. We now make these three cases explicit.

\paragraph{Variational View.}
The oracle reverse transition becomes
\[
p^*(\rvx_{t-\Delta t}|\rvx_t)
=
\sum_{i=1}^N
w_i(\rvx_t,t) 
p(\rvx_{t-\Delta t}|\rvx_t,\rvx_0^{(i)}).
\]
Here each conditional reverse kernel
\[
p(\rvx_{t-\Delta t}|\rvx_t,\rvx_0^{(i)})
\]
is itself available in closed form under Gaussian diffusion; in fact, it is Gaussian as shown in Lemma~\ref{ddpm-reverse-kernel}. So the true reverse kernel is simply a posterior-weighted mixture of explicit Gaussian bridges. 

\paragraph{Score-Based View.}
Likewise, the oracle score becomes
\[
\mathbf{s}^*(\rvx_t,t)
=
\sum_{i=1}^N
w_i(\rvx_t,t) 
\nabla_{\rvx_t}\log p_t(\rvx_t|\rvx_0^{(i)}).
\]
The conditional score is explicit:
\[
\nabla_{\rvx_t}\log p_t(\rvx_t|\rvx_0^{(i)})
=
-\frac{\rvx_t-\alpha_t\rvx_0^{(i)}}{\sigma_t^2}.
\]
Therefore,
\[
\mathbf{s}^*(\rvx_t,t)
=
-\frac{1}{\sigma_t^2}
\left(
\rvx_t-\alpha_t\sum_{i=1}^N w_i(\rvx_t,t) \rvx_0^{(i)}
\right).
\]
So the marginal score points from the noisy sample toward a posterior-weighted average of training examples.

\paragraph{Flow-Based View.}
The same phenomenon appears for flow matching:
\[
\mathbf{v}^*(\rvx_t,t)
=
\sum_{i=1}^N
w_i(\rvx_t,t) 
\mathbf{v}_t(\rvx_t|\rvx_0^{(i)}).
\]
Again, the building blocks are explicit. The corresponding conditional velocity field has closed form:
\[
\mathbf{v}_t(\rvx_t|\rvx_0^{(i)})
=
{\alpha}'_t \rvx_0^{(i)}
+
\frac{{\sigma}'_t}{\sigma_t}\bigl(\rvx_t-\alpha_t\rvx_0^{(i)}\bigr).
\]
Hence the oracle velocity is also an explicit posterior-weighted average over the dataset.

Taken together, these formulas show that, on a finite dataset, all three oracle targets reduce to explicit posterior-weighted averages of closed-form conditional quantities.

\paragraph{Remark on Memorization.} The formulas above characterize the exact oracle targets induced by the empirical distribution $\hat p_0$, rather than by the unknown population distribution $p_{\mathrm{data}}$. If the model learned these empirical targets perfectly at every noise level, and if the reverse process were computed without error, then generation would exactly reproduce the empirical distribution $\hat p_0$. Since $\hat p_0$ assigns probability only to the training examples, the generated samples would be drawn from the training set itself. In this idealized sense, perfect fitting of the empirical oracle amounts to memorization.

However, a practically trained diffusion model, need not realize this oracle exactly: it is learned with finite model capacity and finite optimization, and its generated distribution is further affected by model approximation and numerical sampling errors. Therefore, the finite-dataset calculation establishes memorization as an exact empirical solution, but does not imply that every practically trained model necessarily memorizes its training set.

}

The common conditional reformulation is no coincidence: by making training tractable, it reveals a profound unification. Variational diffusion, score-based SDEs, and flow matching are simply different facets of the same principle. Three perspectives, one insight, elegantly connected. We will continue to explore their equivalence throughout the rest of this chapter.

\newpage
\clearpage

\section{A Roadmap for Elucidating Training Losses in Diffusion Models
}\label{sec:elucidating}

This section builds a systematic view of training losses in diffusion models.  
In \Cref{subsec:four-predictions}, we extend the standard three objectives to a broader set of four parameterizations, showing how they arise from different modeling perspectives.  
In \Cref{subsec:summary-disentagle}, we then distill these results into a general framework that disentangles the structure of diffusion objectives, laying the groundwork for the equivalence results in \Cref{sec:equivalent-parametrizations}.

\subsection{Four Common Parameterizations in Diffusion Models}\label{subsec:four-predictions}
Throughout this section, we consider the forward perturbation kernel
\[
p_t(\mathbf{x}_t|\mathbf{x}_0) = \mathcal{N}\left(\mathbf{x}_t; \alpha_t \mathbf{x}_0, \sigma_t^2 \mathbf{I} \right),
\]
where $\mathbf{x}_0 \sim p_{\mathrm{data}}$, as defined in \Cref{eq:forward_kernel}, unless stated otherwise.

Let $\omega: [0, T] \to \mathbb{R}_{>0}$ denote a positive time-weighting function.  
The four standard parameterizations (noise $\bm{\epsilon}_{\bm{\phi}}$, clean $\rvx_{\bm{\phi}}$, score $\rvs_{\bm{\phi}}$, and velocity $\rvv_{\bm{\phi}}$), together with their respective minimizers $\bm{\epsilon}^*$, $\rvx^*$, $\rvs^*$, and $\rvv^*$, are summarized below for clarity and to facilitate further discussion.

\paragraph{Variational View.} Based on the KL divergence in DDPMs (see \Cref{subsec:ddpm-prediction,subsec:connection-vdm}), this approach reduces to predicting either the expected noise that produces $\mathbf{x}_t$ or the expected clean signal that $\mathbf{x}_t$ was perturbed from.
\begin{enumerate}
    \item \textbfs{$\beps$-Prediction (Noise Prediction)} \citep{ho2020denoising}:
    \begin{equation}\label{eq:noise-def}
        \bm{\epsilon}_{\bm{\phi}}(\mathbf{x}_t,t) \approx \mathbb{E}[\bm{\epsilon}|\mathbf{x}_t]=\bm{\epsilon}^*(\rvx_t, t)
    \end{equation}
    with training objective
    \[
        \mathcal{L}_{\text{noise}}(\bm{\phi}) := \mathbb{E}_t \left[\omega(t) \mathbb{E}_{\mathbf{x}_0, \bm{\epsilon}} \left\| \bm{\epsilon}_{\bm{\phi}}(\mathbf{x}_t,t) - \bm{\epsilon} \right\|_2^2 \right].
    \]
    Here, $\beps^*$ means the average noise that was injected to obtain the given $\rvx_t$.

    \item \textbfs{$\rvx$-Prediction (Clean Prediction)} \citep{kingma2021variational,karras2022elucidating,song2023consistency}:
    \begin{equation}\label{eq:clean-def}
        \mathbf{x}_{\bm{\phi}}(\mathbf{x}_t,t) \approx \mathbb{E}[\mathbf{x}_0|\mathbf{x}_t]=\rvx^*(\rvx_t, t)
    \end{equation}
    with training objective
    \[
        \mathcal{L}_{\text{clean}}(\bm{\phi}) := \mathbb{E}_t \left[\omega(t) \mathbb{E}_{\mathbf{x}_0, \bm{\epsilon}} \left\| \mathbf{x}_{\bm{\phi}}(\mathbf{x}_t,t) - \mathbf{x}_0 \right\|_2^2 \right].
    \]
    Here, $\rvx^*$ means the average of all plausible clean guesses, given the noisy observation $\rvx_t$.
\end{enumerate}

\paragraph{Score-Based View.} Predicts the score function at noise level $t$, which points in the average direction to denoise $\mathbf{x}_t$ back toward all possible clean samples that could have generated it:
\begin{enumerate}
    \item[3.]  \textbfs{Score Prediction} \citep{song2019generative,song2020score}:
    \begin{equation}\label{eq:score-def}
        \mathbf{s}_{\bm{\phi}}(\mathbf{x}_t,t) \approx \nabla_{\mathbf{x}_t} \log p_t(\mathbf{x}_t) = \mathbb{E}\left[\nabla_{\mathbf{x}_t} \log p_t(\mathbf{x}_t|\mathbf{x}_0) |\mathbf{x}_t\right] = \rvs^*(\rvx_t, t)
    \end{equation}
    with training objective
    \[
        \mathcal{L}_{\text{score}}(\bm{\phi}) := \mathbb{E}_t \left[\omega(t) \mathbb{E}_{\mathbf{x}_0, \bm{\epsilon}} \left\| \mathbf{s}_{\bm{\phi}}(\mathbf{x}_t,t) - \nabla_{\mathbf{x}_t} \log p_t(\mathbf{x}_t|\mathbf{x}_0) \right\|_2^2 \right],
    \]
    where the conditional score satisfies $\nabla_{\mathbf{x}_t} \log p_t(\mathbf{x}_t|\mathbf{x}_0) = -\frac{1}{\sigma_t} \bm{\epsilon}$.
\end{enumerate}

\paragraph{Flow-Based View.} Predicts the instantaneous average velocity of the data as it evolves through $\mathbf{x}_t$:
\begin{enumerate}
    \item[4.] \textbfs{$\rvv$-Prediction (Velocity Prediction)} \citep{lipman2022flow,liu2022rectified,salimans2021progressive,albergo2023stochastic}:
    \begin{equation}\label{eq:velocity-def}
        \mathbf{v}_{\bm{\phi}}(\mathbf{x}_t, t) \approx \mathbb{E}\left[\left. \frac{\diff  \mathbf{x}_t}{\diff  t} \right| {\mathbf{x}_t}\right] =\rvv^*(\rvx_t, t)
    \end{equation}
    with training objective
    \[
        \mathcal{L}_{\text{velocity}}(\bm{\phi}) := \mathbb{E}_t \left[\omega(t) \mathbb{E}_{\mathbf{x}_0, \bm{\epsilon}} \left\| \mathbf{v}_{\bm{\phi}}(\mathbf{x}_t,t) - \rvv_t(\rvx_t | \rvx_0,\bm{\epsilon}) \right\|_2^2 \right],
    \]
    where the conditional velocity is $\rvv_t(\rvx_t|\rvx_0,\bm{\epsilon}) = \alpha_t' \mathbf{x}_0 + \sigma_t' \bm{\epsilon}$.

Here, $\rvv^*$ indicates the average velocity vector passing through the observation point $\rvx_t$.
\end{enumerate}

Building on the insight from \Cref{eq:three-minimizer-oracle}, all four prediction types ultimately aim to approximate a conditional expectation in the form of the average noise, clean data, score, or velocity given an observed $\mathbf{x}_t$.

\subsection{Disentangling the Training Objective of Diffusion Models}\label{subsec:summary-disentagle}
As shown in \Cref{subsec:four-predictions}, the objective functions for the four prediction types commonly share the following template form for diffusion model training:

\begin{align}    \label{eq:general-dm-loss-D}
\begin{aligned}
   \mathcal{L}(\bm{\phi}):=\mathbb{E}_{\rvx_0, \bm{\epsilon}} \underbrace{\eqnmarkbox[NavyBlue]{tsample}{\mathbb{E}_{p_{\text{time}}(t)}}}_{\substack{\text{time} \\ \vspace{0.5cm} \text{distribution}}}\Big[ \underbrace{\eqnmarkbox[OliveGreen]{wgt}{\omega(t)}}_{\substack{\text{time} \\ \vspace{0.5cm}\text{weighting}}}\underbrace{\norm{ \eqnmarkbox[Plum]{target}{\mathrm{NN}_{\bm{\phi}}}(\eqnmarkbox[Pink]{para}{\rvx_t}, t) - \eqnmarkbox[Plum]{target}{(A_t \rvx_0 + B_t \bm{\epsilon}) }}_2^2}_{\text{MSE part}}\Big].
\end{aligned}
\end{align}

Here, to enhance training efficiency and optimize the diffusion model learning pipeline, several key design choices are crucial~\citep{karras2022elucidating,lu2024simplifying}:
\begin{enumerate}
    \item[(A)] Noise schedule in the forward process of $\eqnmarkbox[Pink]{para}{\rvx_t}$ via $\alpha_t$ and $\sigma_t$;
    \item[(B)] Prediction types of $\eqnmarkbox[Plum]{target}{\mathrm{NN}_{\bm{\phi}}}$ and their associated regression targets $\eqnmarkbox[Plum]{target}{(A_t \rvx_0 + B_t \bm{\epsilon}) }$;
    \item[(C)] Time-weighting function $\eqnmarkbox[OliveGreen]{wgt}{\omega(\cdot)} \colon [0,T] \to \mathbb{R}_{\geq 0}$;
    \item[(D)] Time distribution $\eqnmarkbox[NavyBlue]{tsample}{p_{\text{time}}}$.
\end{enumerate}

We elaborate on these four components here to serve as a roadmap for the discussions in the following sections.

\paragraph{(A) Noise Schedule $\alpha_t$ and $\sigma_t$.}
Users have the flexibility to choose schedules tailored to their applications, with common examples summarized in \Cref{tb:interpolant_instances}. Importantly, as we will demonstrate in \Cref{eq:fm-general-affine-formula,eq:any-to-trig}, all affine flows of the form $\mathbf{x}_t = \alpha_t \mathbf{x}_0 + \sigma_t \bm{\epsilon}$ are mathematically equivalent. Specifically, any such interpolation can be converted to the canonical linear schedule ($\alpha_t = 1 - t$, $\sigma_t = t$) or to a trigonometric schedule ($\alpha_t = \cos t$, $\sigma_t = \sin t$) by appropriate time reparametrization and spatial rescaling.

\paragraph{(B) Parameterization $\mathrm{NN}_{\bm{\phi}}$ and Training Target $A_t \rvx_0 + B_t \bm{\epsilon}$.}
\begin{table}[th!]
  \caption{Summary of the Relationships Between Different Parameterizations.
All four parameterizations are mathematically equivalent and can be converted into one another through straightforward algebraic transformations.
}
  \small
  \centering
  \resizebox{0.43\textwidth}{!}{
  \begin{tabular}{ccc}
     \toprule
      Regression Target =    &  \multicolumn{2}{c}{$A_t\rvx_0 + B_t \bm{\epsilon}$} \\
              &   $A_t$   & $B_t$ \\
       \midrule
    Clean & $1 $   & $0$ \\
      Noise & $0$   &  $1$\\
    Conditional Score & $0 $   &  -$\frac{1}{\sigma_t}$ \\
     Conditional  Velocity & $\alpha_t' $   &  $\sigma_t'$ \\
 \bottomrule
  \end{tabular}
  }
  \label{tb:A_t-B_t-parametrizations}
\end{table}
There are two choices that are often coupled in diffusion training:
\emph{what quantity the network directly predicts} and \emph{which quantity
is compared in the training loss}. Clean data, noise, score, and velocity can
be used for either purpose.

\subparagraph{Matched Prediction and Loss.}
The common choice, used in \Cref{eq:general-dm-loss-D}, is to train the
network directly in the same prediction type that it outputs. Each prediction
type then has a corresponding conditional training target of the form
\[
\text{Regression Target}
=
A_t\rvx_0+B_t\beps,
\]
where $A_t$ and $B_t$ depend on the prediction type and the forward schedule
$(\alpha_t,\sigma_t)$. These targets are summarized in
\Cref{tb:A_t-B_t-parametrizations}.

For example, an $\rvx$-prediction network is compared with the clean target
$\rvx_0$, whereas a $\rvv$-prediction network is compared with the conditional
velocity
\[
\rvv_t
=
\alpha_t'\rvx_0+\sigma_t'\beps.
\]
As shown later in \Cref{eq:predictions-equivalence}, the four prediction types
encode the same oracle information and can be converted into one another.

\subparagraph{Predicting One Quantity and Evaluating the Loss in Another.}
The network output and the quantity used in the loss can also be chosen
separately. The network prediction is first converted using the relations in
\Cref{eq:predictions-equivalence}, and the loss is then computed against the
corresponding target in the converted prediction type.

For example, under the canonical schedule
\[
\rvx_t=(1-t)\rvx_0+t\beps,
\]
suppose the network directly predicts clean data
$\rvx_{\bm{\phi}}(\rvx_t,t)$. Converting this prediction to velocity gives
\[
\rvv_{\bm{\phi}}(\rvx_t,t)
=
\frac{\rvx_t-\rvx_{\bm{\phi}}(\rvx_t,t)}{t},
\]
while the conditional velocity target satisfies
\[
\rvv_t
=
\frac{\rvx_t-\rvx_0}{t}.
\]
Therefore,
\[
\left\|
\rvv_{\bm{\phi}}(\rvx_t,t)-\rvv_t
\right\|_2^2
=
\frac{1}{t^2}
\left\|
\rvx_{\bm{\phi}}(\rvx_t,t)-\rvx_0
\right\|_2^2.
\]
Thus, with the network output fixed as clean prediction, evaluating a velocity
loss is exactly equivalent to applying an additional time-dependent weight to
the clean-prediction loss. The corresponding relations for general affine
schedules and other prediction types are summarized later in
\Cref{eq:s-e-x-v-equi}.

Changing the quantity predicted directly by the network has an additional
effect. Although the oracle predictions are algebraically convertible, a
finite neural network is then trained to represent a different function, with
different output scales and parameter gradients. The choice of network
prediction can therefore interact with model architecture and optimization,
while the choice of loss also controls the relative weighting of prediction
errors across noise levels.

For clarity, \Cref{eq:general-dm-loss-D} uses the matched prediction-and-loss
case unless stated otherwise.

\paragraph{(C) Time Distribution $p_{\text{time}}(t)$.}
We next consider how time $t$, and hence the noise level, is sampled during
training. For convenience, define the fixed-time regression loss
\[
\mathcal{L}_t(\bm{\phi})
:=
\mathbb{E}_{\rvx_0,\beps}
\left[
\left\|
\mathrm{NN}_{\bm{\phi}}(\rvx_t,t)
-
\bigl(A_t\rvx_0+B_t\beps\bigr)
\right\|_2^2
\right],
\qquad
\rvx_t=\alpha_t\rvx_0+\sigma_t\beps.
\]
Then \Cref{eq:general-dm-loss-D} can be written compactly as
\[
\mathcal{L}(\bm{\phi})
=
\mathbb{E}_{t\sim p_{\text{time}}}
\left[
\omega(t)\mathcal{L}_t(\bm{\phi})
\right].
\]
Thus, if no correction is applied, sampling some times more frequently also
gives them more weight in the objective. At the level of the exact expected
loss, only the product $p_{\text{time}}(t)\omega(t)$ matters.

A different use of nonuniform sampling is \emph{importance sampling}.
Suppose the objective above is fixed, but we instead draw times from a proposal
$q_{\text{time}}(t)$. Then
\[
\mathcal{L}(\bm{\phi})
=
\mathbb{E}_{t\sim q_{\text{time}}}
\left[
\frac{p_{\text{time}}(t)}{q_{\text{time}}(t)}
\omega(t)\mathcal{L}_t(\bm{\phi})
\right],
\]
provided $q_{\text{time}}(t)>0$ wherever
$p_{\text{time}}(t)\omega(t)>0$.
With this correction, changing $q_{\text{time}}$ leaves the objective
unchanged and changes only its Monte Carlo estimator, for example its variance.

The key distinction is therefore:
\begin{center}
\small
\renewcommand{\arraystretch}{1.3}
\begin{tabular}{@{}r@{\;\;$\Longrightarrow$\;\;}l@{}}
change $p_{\text{time}}$ in the objective
&
different weighting, different objective
\\
change $q_{\text{time}}$ $+$ importance correction
&
same objective, different estimator
\end{tabular}
\end{center}

We therefore keep time sampling and loss weighting conceptually separate.
Uniform time sampling makes this distinction less visible because its density
contributes only a constant factor, whereas with nonuniform sampling one must
specify whether the sampling distribution defines the objective or merely
serves as an importance-sampling proposal.

Both uses appear in diffusion-model training. ScoreFlow explicitly uses
importance sampling to reduce the variance of a fixed likelihood-weighted
objective~\citep{song2021maximum}. VDM and subsequent analyses similarly show
that, when the objective is fixed in noise-level coordinates, the noise
schedule can act as an importance-sampling distribution
~\citep{kingma2021variational,kingma2023understanding}.
EDM~\citep{karras2022elucidating}, by contrast, specifies both a nonuniform
training distribution over noise levels and a separate loss weighting, whose
combination determines the effective training weight.

\paragraph{(D) Time-Weighting Function $\omega(t)$.}
For a fixed $p_{\text{time}}$, the weighting function $\omega(t)$ determines
the relative contribution of different times to the training objective.
Together with the previous paragraph, the effective weighting over time is
therefore $p_{\text{time}}(t)\omega(t)$.

A simple baseline is uniform time sampling with $\omega(t)\equiv1$, as in the
standard simplified $\beps$-prediction and FM/RF objectives
~\citep{ho2020denoising,lipman2022flow,liu2022rectified}.
Other weightings emphasize different noise regimes and can also carry
likelihood or variational interpretations. For example, for the score-matching
objective, the likelihood weighting $\omega(t)=g^2(t)$ yields an upper bound
on the negative log-likelihood under suitable regularity conditions
~\citep{song2021maximum}. More generally, SNR-dependent weightings provide a
common way to redistribute training emphasis across noise levels
~\citep{kingma2021variational,kingma2023understanding}.

\subparagraph{Monotonic Weighting and the ELBO-with-Augmentation View.}
A useful way to describe the corruption level is through the
\emph{log signal-to-noise ratio} (log-SNR)
\[
\Lambda_t
:=
\log\frac{\alpha_t^2}{\sigma_t^2}.
\]
We use $\Lambda$ for the full log-SNR, distinguishing it from the
half-log-SNR coordinate
$\lambda_t=\log(\alpha_t/\sigma_t)=\tfrac12\Lambda_t$
used later in \Cref{ch:solvers}.
Since $\Lambda_t$ decreases as the forward process becomes noisier, after
expressing the loss in a common prediction parameterization, let
$w(\Lambda)$ denote the effective weight assigned to each log-SNR level.

A useful result of \citet{kingma2023understanding} is that, if
$w(\Lambda)$ is non-increasing in $\Lambda$ (equivalently, its weight does
not decrease as the forward process becomes noisier), then the weighted
diffusion objective can be rewritten, up to terms independent of the model
parameters, as an average negative ELBO over Gaussian-perturbed data%
\footnote{
To see why monotonicity matters, define the remaining variational cost from
noise level $t$ onward as
\[
L(t;\rvx_0)
:=
\mathcal{D}_{\mathrm{KL}}
\left(
p(\rvx_{[t,T]}|\rvx_0)
\,\Vert\,
p_{\bm{\phi}}(\rvx_{[t,T]})
\right),
\]
where $\rvx_{[t,T]}:=\{\rvx_s\}_{s\in[t,T]}$ denotes the remaining
continuous diffusion path. This is the continuous-time counterpart of the
negative ELBO in \Cref{eq:ddpm-elbo}, starting from corruption level $t$;
more precisely, it equals the expected negative ELBO of the perturbed sample
$\rvx_t$, up to terms independent of $\bm{\phi}$.

For $\beps$-prediction, \citet{kingma2023understanding} show that
\[
\frac{\diff}{\diff t}L(t;\rvx_0)
=
\frac{1}{2}\frac{\diff\Lambda_t}{\diff t}
\E_{\beps}
\left[
\left\|
\beps_{\bm{\phi}}(\rvx_t,t)-\beps
\right\|_2^2
\right].
\]
Hence the weighted diffusion loss is
\[
-\int_0^T
w(\Lambda_t)
\frac{\diff}{\diff t}L(t;\rvx_0)\diff t
=
w(\Lambda_0)L(0;\rvx_0)
+
\int_0^T
\frac{\diff}{\diff t}w(\Lambda_t)
L(t;\rvx_0)\diff t
+
C,
\]
where $C$ is independent of $\bm{\phi}$.
If $w(\Lambda)$ is non-increasing in $\Lambda$, then
$\frac{\diff}{\diff t}w(\Lambda_t)\geq0$ because $\Lambda_t$ decreases with
$t$. Thus, after normalization, the terms above form a nonnegative weighting
over ELBOs at different Gaussian corruption levels, which gives the
ELBO-with-Gaussian-augmentation interpretation.
}.

Thus, monotonic weighting has a concrete probabilistic interpretation:
rather than merely emphasizing some noise levels more than others, it can be
viewed as averaging variational training over Gaussian-corrupted versions of
the data at different corruption strengths.

\paragraph{Conclusion.}
For our purposes, the main lesson is that many apparent design choices
in diffusion training can be understood through how they reshape the
effective weighting over noise levels in the objective. This weighting
in turn influences both the optimization landscape in practice and, in
special cases such as monotonic weighting, the variational
interpretation of the loss.

\clearpage
\newpage

\section{Equivalence in Diffusion Models}\label{sec:equivalent-parametrizations}
Many formulations of diffusion models look different because they use
different prediction types, training objectives, noise schedules, or
continuous-time dynamics. In this section, we clarify in what precise sense
these formulations can be equivalent at the exact theoretical level, and how
their behavior can still differ for learned models in practice.

At the exact or oracle level, several notions of equivalence will appear:
\begin{itemize}
    \item \textbfs{Predictions.}
    Given the noisy observation $\rvx_t$ and the forward schedule, clean-data,
    noise, score, and velocity oracles contain the same information and can be
    converted into one another algebraically.
    
    \item \textbfs{Training Objectives.}
    Squared-error objectives written in different prediction types can differ by
    positive time-dependent factors while sharing the same pointwise oracle
    minimizer. These factors determine how strongly different noise levels are
    weighted during training.
    
    \item \textbfs{Affine Forward Process.}
    At the particle level, affine forward processes
    \[
    \rvx_t=\alpha_t\rvx_0+\sigma_t\beps
    \]
    can be related to canonical affine forms through state rescaling and a
    corresponding canonical coordinate. When this coordinate evolves
    monotonically with time, the relation also defines a time reparameterization.
    
    \item \textbfs{Marginal Density Paths.}
    At the density level, different deterministic and stochastic dynamics can
    produce the same marginal path $\{p_t\}_{t\in[0,T]}$. Their sample
    trajectories may differ, while the evolution of $p_t$ is governed by the
    corresponding continuity or Fokker-Planck equation.
\end{itemize}
We develop these connections from prediction to sampling.
\Cref{subsec:four-equiv-para} first relates the four common oracle prediction
types and their training losses.
\Cref{subsec:pf-ode-different-para} then shows how these predictions lead to
different expressions of the same oracle PF--ODE.
Next, \Cref{subsec:all-flows-are-equiv} relates affine probability paths to
canonical linear and trigonometric forms.
Finally, \Cref{subsec:concept-why-v-affine} develops additional intuition for
the canonical linear schedule and velocity prediction, including their
implications for training and numerical integration.

These exact relationships provide a common mathematical foundation, while
the practical choices remain meaningful. Finite neural networks may
approximate equivalent oracle quantities differently, and loss weighting,
optimization, model architecture, and numerical integration can lead to
different training and sampling behavior.

\subsection{Four Prediction Types Are Equivalent}\label{subsec:four-equiv-para}
We begin by analyzing the design choices for component (B) in \Cref{eq:general-dm-loss-D}. 

We have seen that the four prediction types are not independent choices but different views of the same underlying quantity.  
For example, noise and clean predictions are directly related (\Cref{subsec:ddpm-prediction}), as are score and noise predictions (\Cref{sec:SMLD}).  
This recurring pattern points to a deeper principle: all four parameterizations are algebraically equivalent and can be converted into one another through simple transformations.  
To make this connection precise, we state the following proposition, illustrated in \Cref{fig:equiv-parametrizations}, following~\citep{kingma2021variational}.
\proppp{Equivalence of Parametrizations}{equi-para}{
Let the optimal predictions minimizing their respective objectives be
\[
\bm{\epsilon}^*(\rvx_t,t) , \quad \rvx^*(\rvx_t,t), \quad \rvs^*(\rvx_t,t) , \quad \rvv^*(\rvx_t,t),
\]
corresponding to noise, clean, score, and velocity parameterizations. These satisfy the following equivalences:
\begin{equation}\label{eq:predictions-equivalence}
\begin{aligned}
\bm{\epsilon}^*(\rvx_t,t) &= -\sigma_t \rvs^*(\rvx_t,t), \\
\rvx^*(\rvx_t,t) &= \frac{1}{\alpha_t} \rvx_t
+ \frac{\sigma_t^2}{\alpha_t} \rvs^*(\rvx_t,t), \\
\rvv^*(\rvx_t,t) &= \alpha_t' \rvx^* + \sigma_t' \bm{\epsilon}^* 
= f(t)\rvx_t - \frac{1}{2} g^2(t)\rvs^*(\rvx_t,t).
\end{aligned}
\end{equation}
Here, $f(t)$ and $g(t)$ are related to $\alpha_t$ and $\sigma_t$ via Lemma~\ref{forward-sde}.
Moreover, these minimizers satisfy the identities given in \Cref{eq:noise-def,eq:clean-def,eq:score-def,eq:velocity-def}. 
}{The proof is similar to that of Theorem~\ref{dsm-minimizer}, which analyzes the global optimum of various matching losses under the DSM objective. See \Cref{app-sec:equiv-para} for details.
}

\begin{figure}[th]
    \centering
\begin{tikzpicture}[
    node distance=2cm and 3cm,
    box/.style={rectangle,draw,fill=gray!10,minimum width=2.5cm,minimum height=1cm,align=center},
    arr/.style={-{Latex[length=3mm,width=2mm]},thick,font=\footnotesize},
    lbl/.style={draw,rounded corners,fill=white,fill opacity=1,inner sep=2pt}
  ]
  \node[box]            (Score)    {\textbfs{Score}};
  \node[box,right=of Score]   (Noise)    {\textbfs{Noise}};
  \node[box,below=of Noise]   (Velocity) {\textbfs{Velocity}};
  \node[box,below=of Score]   (Clean)    {\textbfs{Clean}};

  \begin{pgfonlayer}{background}
    \draw[arr] (Score)    -- (Noise);
    \draw[arr] (Score)    -- (Clean);
    \draw[arr] (Score)    -- (Velocity);
    \draw[arr] (Noise)    -- (Velocity);
    \draw[arr] (Noise)    -- (Score);
    \draw[arr] (Clean)    -- (Score);
    \draw[arr] (Clean)    -- (Velocity);
    \draw[arr] (Velocity) -- (Clean);
    \draw[arr] (Velocity) -- (Score);
    \draw[arr] (Velocity) -- (Noise);
  \end{pgfonlayer}

  \node[lbl] at ($(Score)!0.5!(Noise)+(0,0.5)$)
    (eqlabel) {$\displaystyle \bm{\epsilon}^* = -\sigma_t\rvs^*$};
  \node[lbl] at ($(Score)!0.5!(Clean)+(-1.2,0)$)
    (xlabel) {$\displaystyle \rvx^* =\tfrac1{\alpha_t}\rvx_t + \tfrac{\sigma_t^2}{\alpha_t}\rvs^*$};
  \node[lbl] at ($(Velocity)!0.5!(Score)-(0,0.0)$)
    (vlabel) {$\displaystyle \rvv^* = f(t)\rvx_t - \tfrac12 g^2(t)\rvs^*$};
  \node[lbl] at ($(Clean)!0.5!(Noise)+(4.2,0)$)
    (vvlabel) {$\displaystyle  \rvv^*=\alpha_t'\rvx^*+\sigma_t'\bm{\epsilon}^*(\bm{\star})$};
  \node[lbl] at ($(Clean)!0.5!(Velocity)+(0,-0.4)$)
    (vvlabel) {$\displaystyle (\bm{\star})$};

\end{tikzpicture}
\caption{\textbfs{Equivalent relations among the four oracle
parameterizations.}
Each oracle prediction can be converted
into the others using the forward schedule and the observed state $\rvx_t$.
For example,
$\rvv^*=\alpha_t'\rvx^*+\sigma_t'\bm{\epsilon}^*$, while $\rvx$- and
$\beps$-predictions satisfy
$\rvx_t=\alpha_t\rvx^*+\sigma_t\bm{\epsilon}^*$.
\figcredit{Created by the authors.}}
    \label{fig:equiv-parametrizations}
\end{figure}
The proposition is most naturally understood as an equivalence of
\emph{oracle information}. At any noise level, knowing one oracle prediction,
together with the noisy observation $\rvx_t$ and the forward schedule,
determines the others through the identities in
\Cref{eq:predictions-equivalence}. For example, the score determines the
posterior-mean clean prediction, and either of these determines the PF-ODE
velocity. Thus, at the oracle level, these parameterizations contain the same
information about the underlying diffusion process.

For a learned model, however, one must distinguish this oracle equivalence
from the quantity predicted directly by the neural network. One may choose
any of
\[
\bm{\epsilon}_{\bm{\phi}}(\rvx_t,t),
\quad
\rvx_{\bm{\phi}}(\rvx_t,t),
\quad
\rvs_{\bm{\phi}}(\rvx_t,t),
\quad
\rvv_{\bm{\phi}}(\rvx_t,t)
\]
as the network output. Once one prediction is taken as primary, the others
can be defined through the same algebraic conversions as in
\Cref{eq:predictions-equivalence}. In particular, for a trained model with
parameters $\bm{\phi}^\times$, these converted predictions are exactly
consistent by construction; for example,
\[
\rvx
=
\alpha_t
\rvx_{\bm{\phi}^\times}(\rvx,t)
+
\sigma_t
\beps_{\bm{\phi}^\times}(\rvx,t).
\]
These conversions are useful both during and after training. For instance,
a noise prediction can be converted into a score for reverse-SDE sampling
or into a velocity for PF-ODE sampling.

The network prediction parameterization should also be distinguished from
the \emph{training loss}. As discussed in component (B) of
\Cref{eq:general-dm-loss-D}, the quantity output by the network need not be
the quantity on which the loss is evaluated: the prediction may first be
converted into another parameterization before computing the objective.
Such a conversion generally changes a squared-error objective by a
time-dependent weighting, as shown later in \Cref{eq:s-e-x-v-equi}.

These distinctions matter for finite neural networks. Although the oracle
targets are information-equivalent, directly predicting different quantities
leads the network to represent and optimize different functions, and the
corresponding objectives may weight noise levels differently. Consequently,
models trained independently under different prediction parameterizations
need not produce identical empirical predictions or identical PF-ODEs.
The equivalence in this proposition therefore concerns the information
carried by the ideal oracle predictions; the network output parameterization
and the training objective remain separate and meaningful design choices.

\subsection{PF-ODE in Different Parameterizations}\label{subsec:pf-ode-different-para} 
The PF-ODE admits several equivalent parameterizations (score, noise, denoised, and velocity). 
Although interchangeable in principle, the choice has practical consequences: it changes the stiffness of the vector field, the behavior of discretization error, and the ease of optimization.
For fast sampling with advanced ODE solvers (see \Cref{ch:solvers}), practitioners often work with $\beps$ or $\rvx$ prediction because they align well with solver inputs and reduce error accumulation.
For training generators that use only a few function evaluations (see \Cref{ch:fast-scratch}), $\rvx$ or $\rvv$ prediction often yields smoother objectives and better step to step consistency.

We write the PF-ODE under each parameterization and make
the conversions explicit using \Cref{eq:predictions-equivalence}. The
results are collected in the following proposition.

\proppnp{PF-ODE in Different Parameterizations}{pf-ode-para}{
Let $\alpha_t$ and $\sigma_t$ be the forward perturbation schedules, and denote time derivatives by
$\alpha_t' := \tfrac{\diff \alpha_t}{\diff t}$ and $\sigma_t' := \tfrac{\diff \sigma_t}{\diff t}$.
Then the empirical PF-ODE admits the equivalent forms
\begin{align}\label{eq:clean_pf_ode}
\begin{aligned}
    \frac{\diff \rvx(t)}{\diff t}
&= \frac{\alpha_t'}{\alpha_t} \rvx(t)
 - \sigma_t \left(\frac{\alpha_t'}{\alpha_t}-\frac{\sigma_t'}{\sigma_t}\right)
   \bm{\epsilon}^*(\rvx(t), t) \\    
&= \frac{\sigma_t'}{\sigma_t} \rvx(t)
 + \alpha_t \left(\frac{\alpha_t'}{\alpha_t}-\frac{\sigma_t'}{\sigma_t}\right)
   \rvx^*(\rvx(t), t) \\
&= \frac{\alpha_t'}{\alpha_t} \rvx(t)
 + \sigma_t^{2} \left(\frac{\alpha_t'}{\alpha_t}-\frac{\sigma_t'}{\sigma_t}\right)
   \rvs^*(\rvx(t), t) \\ 
   &= \alpha_t'\rvx^*(\rvx(t), t) + \sigma_t' \beps^*(\rvx(t), t)\\
&= \rvv^*(\rvx(t), t).
\end{aligned}
\end{align}
}
To see the Score SDE notation, we recall Lemma~\ref{forward-sde}.  If we set
\[
f(t) = \frac{\alpha_t'}{\alpha_t}, 
\quad
g^2(t) = \frac{\diff}{\diff t} \big(\sigma_t^2\big) 
         - 2 \frac{\alpha_t'}{\alpha_t} \sigma_t^2
       = 2\sigma_t\sigma_t' - 2 \frac{\alpha_t'}{\alpha_t} \sigma_t^2,
\]
then the PF-ODE can be written in the familiar Score SDE form:
\[
\frac{\diff \rvx(t)}{\diff t}
= f(t) \rvx(t)  -  \frac{1}{2} g^2(t) \rvs^*(\rvx(t),t).
\]

To give a concrete sense of how the PF-ODE is discretized
for sampling, we will present in \Cref{sec:revisit-ddim} the update rule of a
widely used diffusion-based ODE sampler, the DDIM scheme. This example will show how an Euler discretization naturally connects with the PF-ODE.

\subsection{All Affine Flows Are Equivalent}\label{subsec:all-flows-are-equiv}
We next analyze the design choices for component (A) in \Cref{eq:general-dm-loss-D}.

\paragraph{State-Level Equivalence.}
A convenient canonical interpolation used in FM~\citep{lipman2022flow} and RF~\citep{liu2022rectified} is
\[
\rvx_t^{\mathrm{FM}}=(1-t) \rvx_0+t \beps = \rvx_{0}+t(\beps-\rvx_0),
\]
whose velocity is the constant vector $\beps-\rvx_0$. The key point of this subsection is that the apparent simplicity of this choice is not essential: any affine interpolation
\[
\rvx_t=\alpha_t \rvx_0+\sigma_t \beps
\]
can be written as a time–reparameterized and rescaled version of the canonical path. Define
\[
c(t):=\alpha_t+\sigma_t,
\qquad
\tau(t):=\frac{\sigma_t}{\alpha_t+\sigma_t}
\quad\big(c(t)\neq 0\big).
\]
A direct algebraic rewrite yields
\begin{align*}
\rvx_t
&=\alpha_t \rvx_0+\sigma_t \beps
\\
&=\big(\alpha_t+\sigma_t\big) \left(\frac{\alpha_t}{\alpha_t+\sigma_t}\rvx_0+\frac{\sigma_t}{\alpha_t+\sigma_t}\beps\right) \\
&=c(t) \left(\big(1-\tau(t)\big)\rvx_0+\tau(t)\beps\right)
=c(t) \rvx_{\tau(t)}^{\mathrm{FM}}.
\end{align*}
Hence, whenever $c(t)\neq0$, every affine path can be represented pointwise
as a scalar rescaling of the canonical path. If, in addition, $\tau(t)$ is
continuous and strictly monotone over the time interval of interest, then
$t\mapsto\tau(t)$ is also a genuine change of time variable, and the two paths
are related by time reparameterization together with the rescaling $c(t)$.
For the usual forward diffusion schedules, monotonic decay of the
signal-to-noise ratio implies monotonic evolution of $\tau(t)$.

For the associated velocities, apply the chain rule to $\rvx_t=c(t)\rvx_{\tau(t)}^{\mathrm{FM}}$:
\begin{align*}
\rvv(\rvx_t,t)
&:= \frac{\diff}{\diff t}\left(\alpha_t\rvx_0 +\sigma_t \beps\right)
\\&=\frac{\diff}{\diff t}\big(c(t)\rvx_{\tau(t)}^{\mathrm{FM}}\big)
\\
&=c'(t) \rvx_{\tau(t)}^{\mathrm{FM}}+c(t) \tau'(t) \frac{\diff}{\diff s}\rvx^{\mathrm{FM}}_{s}\bigg|_{s=\tau(t)} \\
&=c'(t) \rvx_{\tau(t)}^{\mathrm{FM}}+c(t) \tau'(t) \rvv^{\mathrm{FM}} \left(\rvx_{\tau(t)}^{\mathrm{FM}},\tau(t)\right),
\end{align*}
since $\rvv^{\mathrm{FM}}(\rvx^{\mathrm{FM}}_{\tau},\tau)=-\rvx_0+\beps$ along the canonical path.

We summarize the above derivation as a formal statement in the following proposition.

\proppnp{Equivalence of Affine Flows}{equiv-interpolation}{
Let
$\rvx_t^{\mathrm{FM}}=(1-t)\rvx_0+t\beps$
and
$\rvx_t=\alpha_t\rvx_0+\sigma_t\beps$,
with
$c(t):=\alpha_t+\sigma_t\neq0$
and
$\tau(t):=\sigma_t/(\alpha_t+\sigma_t)$.
Then
\begin{align}\label{eq:fm-general-affine-formula}
\begin{aligned}
\rvx_t
&=
c(t)\rvx_{\tau(t)}^{\mathrm{FM}},
\\
\rvv(\rvx_t,t)
&=
c'(t)\rvx_{\tau(t)}^{\mathrm{FM}}
+
c(t)\tau'(t)
\rvv^{\mathrm{FM}}
\left(
\rvx_{\tau(t)}^{\mathrm{FM}},\tau(t)
\right).
\end{aligned}
\end{align}
Thus, every affine interpolation with $c(t)\neq0$ admits this canonical
state-level representation. When $\tau(t)$ is additionally continuous and
strictly monotone, it can also be interpreted as a time reparameterization
together with the scalar rescaling $c(t)$.
}

\subparagraph{Equivalence with Trigonometric Flow.}
Another widely used affine flow is the trigonometric interpolation~\citep{salimans2021progressive,albergo2023stochastic,lu2024simplifying}. 
As a concrete example, we also show that \emph{any} affine flow can be expressed in this form. 
The trigonometric path is defined by
\begin{align}\label{eq:trig-para}
\rvx^{\mathrm{Trig}}_u := \cos(u) \rvx_0 + \sin(u)\beps .
\end{align}
Let $R_t := \sqrt{\alpha_t^2 + \sigma_t^2}$ and assume $R_t>0$. Choose an angle $\tau_t$ so that
\[
\cos \tau_t = \frac{\alpha_t}{R_t},
\qquad
\sin \tau_t = \frac{\sigma_t}{R_t}.
\]
Then every affine interpolation $\rvx_t=\alpha_t\rvx_0+\sigma_t\beps$ admits a pointwise rescaled and re-timed trigonometric path:
\begin{align}\label{eq:any-to-trig}
\rvx_t
= \alpha_t\rvx_0+\sigma_t\beps
= R_t \left(\frac{\alpha_t}{R_t}\rvx_0+\frac{\sigma_t}{R_t}\beps\right)
= R_t \rvx^{\mathrm{Trig}}_{\tau_t}.
\end{align}
The pair $(\alpha_t,\sigma_t)$ is a point in the plane. Normalizing by $R_t$ places it on the unit circle, which fixes the angle $\tau_t$ and hence the state $\rvx^{\mathrm{Trig}}_{\tau_t}$; the radius $R_t$ gives the overall scale. If $\tau_t$ can additionally be chosen as a continuous and strictly monotone
function of $t$ on the interval of interest, then this state-level
representation also defines a genuine time reparameterization.

Differentiating $\rvx^{\mathrm{Trig}}_u$ with respect to $u$ gives its velocity,  
\[
\rvv^{\mathrm{Trig}}_u = -\sin(u) \rvx_0 + \cos(u) \beps.
\]
Through the same change of variables as in \Cref{eq:any-to-trig}, this relation provides closed-form conversions for the velocity (and analogously for other parameterizations).

Summarizing the above discussion, we arrive at the following conclusion:
\msg{Conclusion}{}{Regardless of the schedule $(\alpha_t,\sigma_t)$, including VE, VP (such as trigonometric), FM, or RF, affine interpolations are mutually convertible by a suitable change of time variable and a scalar rescaling.}

\paragraph{Training Objectives of Four Parameterizations.}
Let $\rvx_t=\alpha_t \rvx_0+\sigma_t \beps$ with $\sigma_t>0$ and differentiable $(\alpha_t,\sigma_t)$ such that $\alpha_t'\sigma_t-\alpha_t\sigma_t'\neq 0$. Consider the oracle targets
\[
\beps^*(\rvx_t,t)=\E[\beps|\rvx_t],\quad 
\rvx_0^*(\rvx_t,t)=\E[\rvx_0|\rvx_t],\quad
\rvv^*(\rvx_t,t)=\E[\alpha_t'\rvx_0+\sigma_t'\beps|\rvx_t].
\]
From Proposition~\ref{equi-para}, they satisfy
\[
\nabla_{\rvx_t}\log p_t(\rvx_t)
= -\frac{1}{\sigma_t} \beps^*(\rvx_t,t)
= \frac{\alpha_t}{\sigma_t^2} \left(\rvx_0^*(\rvx_t,t)-\frac{\rvx_t}{\alpha_t}\right),
\quad
\rvv^*=\alpha_t' \rvx_0^*+\sigma_t' \beps^*.
\]
Under the head conversions
\[
\rvs_{\bphi} \equiv -\frac{1}{\sigma_t} \beps_{\bphi}
\equiv \frac{\alpha_t}{\sigma_t^2} \left(\rvx_{\bphi}-\frac{\rvx_t}{\alpha_t}\right),
\]
and the velocity-to-score conversion is
\[
\rvs_{\bphi}
= \frac{\alpha_t}{\sigma_t(\alpha_t'\sigma_t-\alpha_t\sigma_t')} \rvv_{\bphi}
- \frac{\alpha_t'}{\sigma_t(\alpha_t'\sigma_t-\alpha_t\sigma_t')} \rvx_t,
\]
the per–sample squared losses match up to time–dependent weights:
\begin{align}\label{eq:s-e-x-v-equi}
\begin{aligned}
\big\|\rvs_{\bphi}-\nabla_{\rvx_t}\log p_t(\rvx_t)\big\|_2^2
&=\frac{1}{\sigma_t^2} \big\|\beps_{\bphi}-\beps^*\big\|_2^2 \\
&=\frac{\alpha_t^2}{\sigma_t^4} \big\|\rvx_{\bphi}-\rvx_0^*\big\|_2^2 \\
&=\left(\frac{\alpha_t}{\sigma_t(\alpha_t'\sigma_t-\alpha_t\sigma_t')}\right)^{ 2}
\big\|\rvv_{\bphi}-\rvv^*\big\|_2^2 .
\end{aligned}
\end{align}

By Proposition~\ref{equiv-interpolation}, any affine flow $\rvx_t=\alpha_t\rvx_0+\sigma_t\beps$
is transferable to the canonical FM path via $\rvx_t=c(t)\rvx^{\mathrm{FM}}_{\tau(t)}$ with
$c(t)=\alpha_t+\sigma_t$ and $\tau(t)=\sigma_t/(\alpha_t+\sigma_t)$. Differentiating gives
\[
\rvv_{\bphi}(\rvx_t,t)=c'(t)\rvx^{\mathrm{FM}}_{\tau(t)}
+c(t)\tau'(t)\rvv^{\mathrm{FM}}_{\bphi} \left(\rvx_{\tau(t)}^{\mathrm{FM}},\tau(t)\right),
\qquad
\rvx_{\tau(t)}^{\mathrm{FM}}=\frac{\rvx_t}{c(t)},
\]
and the same relation holds for $\rvv^*$. Hence the velocity loss transforms by
\begin{align*}
    \big\|\rvv_{\bphi}(\rvx_t,t)-\rvv^*(\rvx_t,t)\big\|_2^2
= \big(c(t) \tau'(t)\big)^2
\Big\|\rvv^{\mathrm{FM}}_{\bphi} \left(\frac{\rvx_t}{c(t)},\tau(t)\right)
-\big(\rvv^{\mathrm{FM}}\big)^* \left(\frac{\rvx_t}{c(t)},\tau(t)\right)\Big\|_2^2.
\end{align*}

With the above observation, we arrive at the following conclusion:
\msg{Conclusion}{}{Score, noise, clean, and velocity training objectives are theoretically equivalent up to time–dependent weights (and, for velocity, an affine head conversion involving $\rvx_t$) determined by $(\alpha_t,\sigma_t)$.}

\subsection{(Optional) Analysis of Parametrizations and the Canonical Flow}\label{subsec:concept-why-v-affine}
Even though we have shown in the previous sections that all four parameterizations are mathematically equivalent and can be transformed into one another, and that the forward affine noise-injection flow is equivalent to the canonical form
\[
\rvx_t^{\mathrm{FM}} = (1-t) \rvx_0 + t \beps,
\]
in this subsection we provide further intuition and analyze the potential advantages of using the $\rvv$-prediction parameterization together with this canonical affine flow.

This subsection asks a simple question: \emph{how do different parameterizations and schedules shape what the model learns and how we sample?} We proceed in two conceptual perspectives:
\begin{itemize}
    \item \textbfs{Regression Targets and Schedules.}
    We focus on why combining $\rvv$-prediction with the canonical linear
    schedule $(\alpha_t,\sigma_t)=(1-t,t)$ is natural: it keeps the conditional
    velocity target scale constant in time and removes the explicit
    schedule-curvature terms from the velocity decomposition.
  \item \textbfs{Solver Implications.}
We examine how this parameterization interacts with numerical integration,
while deferring the detailed solver analysis to \Cref{ch:solvers}.
\end{itemize}

Before proceeding, we distinguish between two types of velocity fields to avoid ambiguity. The \emph{conditional velocity}, which serves as a tractable training target, is defined as
\[
\rvv_t(\rvx_t |\rvz) = \rvx_t' = \alpha_t' \rvx_0 + \sigma_t' \beps, 
\quad \text{where } \rvz = (\rvx_0, \beps),
\]
while the \emph{oracle (marginalized) velocity}, used to move samples during inference of PF-ODE solving, is given by
\[
\rvv^*(\rvx, t) = \E\big[\rvv_t(\cdot |\rvz)  \big|  \rvx_t = \rvx\big].
\]

\paragraph{Perspective 1: Why $(\alpha_t,\sigma_t)=(1-t,t)$ is a Natural Schedule.}
Writing $\sigma_t:=\zeta(t)$ and $\alpha_t:=1-\zeta(t)$ for a time-varying $\zeta(t)$, the conditional velocity becomes
\[
\rvv_t(\rvx_t |\rvz)
=\zeta'(t)(\beps-\rvx_0),
\quad\text{where } \rvz=(\rvx_0,\beps).
\]

\subparagraph{Unit-Scale Regression Targets.}
For the canonical schedule $\zeta(t)=t$, the conditional velocity $\rvv_t(\cdot |\rvz)$ satisfies
\begin{align}\label{eq:unit-velocity}
\mathbb{E}\left[\|\rvv_t(\cdot |\rvz)\|_2^2\right]
=\E_{\beps}\norm{\beps}_2^2+\E_{\rvx_0}\norm{\rvx_0}^2_2 = D + \underbrace{\Tr\Cov[\rvx_0]}_{\text{total variance}}+\underbrace{\norm{\E\rvx_0}^2_2}_{\text{mean}}.
\end{align}
Thus the expected target magnitude is constant in $t$. After standardizing the data to zero mean and identity covariance (i.e., $\Cov[\rvx_0]=\rmI$), the two components
$\alpha_t'\rvx_0$ and $\sigma_t'\beps$ contribute comparably for all $t$, avoiding gradient explosion/vanishing near the endpoints.
To see this, we consider the diffusion's training objective:
\[
\mathcal{L}_{\mathrm{velocity}}(\bphi)
= \E_{t}  \E_{\rvx_0,\beps}\left[\big\|\rvv_{\bphi}(\rvx_t,t) - \rvv_t(\rvx_t|\rvz)\big\|_2^2
\right].
\]
By applying the chain rule, the gradient of this loss with respect to the model parameters $\bphi$ is
\[
\nabla_{\bphi}\mathcal{L}_{\mathrm{velocity}}(\bphi)
= 2  \E_{t}  \E_{\rvx_0,\beps} \left[
  \partial_{\bphi}\rvv_{\bphi}(\rvx_t,t)^{\top} \left(\rvv_{\bphi}(\rvx_t,t) - \rvv_t(\rvx_t|\rvz)\right)
\right].
\]
Thus the scale of the target $\|\rvv_t(\rvx_t|\rvz)\|_2$ influences gradient stability:
if it collapses to $0$ (or blows up) at some $t$, gradients tend to vanish (or explode), 
all else equal. With the canonical choice $\zeta(t)=t$, 
\Cref{eq:unit-velocity} gives a $t$-independent target magnitude, so there is no 
endpoint ($t=0$ or $t=1$) collapse or blow-up arising from the regression signal
(assuming $\E\|\partial_{\bphi}\rvv_{\bphi}(\rvx_t,t)\|^2$ 
and any time-weights are controlled).

\subparagraph{Interplay of the Canonical Schedule and $\rvv$-Prediction.}
Under the affine path $\rvx_t=\alpha_t \rvx_0+\sigma_t\beps$, the oracle velocity decomposes as
\[
\rvv^*(\rvx, t)=\alpha_t' \rvx^*(\rvx,t)+\sigma_t' \beps^*(\rvx,t),
\]
with $\rvx^*=\E[\rvx_0|\rvx_t=\rvx]$ and $ \beps^*=\E[\beps|\rvx_t=\rvx]$.
Differentiating at fixed $\rvx$ gives
\[
\partial_t \rvv^*
=\underbrace{\alpha_t'' \rvx^*+\sigma_t'' \beps^*}_{\text{schedule curvature}}
+ \alpha_t' \partial_t \rvx^*+\sigma_t' \partial_t \beps^*.
\]
With the linear schedule $\alpha_t=1-t,\ \sigma_t=t$, the curvature terms vanish
($\alpha_t''=\sigma_t''=0$), so the time-variation of $\rvv^*$ primarily reflects the
posterior evolution $(\partial_t \rvx^*,\partial_t \beps^*)$ rather than the schedule.
This effect is especially clean for $\rvv$-prediction: the coefficients $\alpha_t',\sigma_t'$
are constants ($-1$ and $+1$), avoiding extra $t$-dependent rescaling in the drift.
By contrast, score-, $\rvx_0$, or $\beps$-parameterizations often introduce ratios such as
$\sigma_t'/\sigma_t$ or $\alpha_t'/\alpha_t$ that can vary sharply near the endpoints,
even under a linear schedule. Hence, while not exclusive in principle, the linear
$(1-t,t)$ schedule combined with $v$-prediction offers a particularly stable and
transparent time dependence for the oracle velocity.

\subparagraph{Minimizing the Conditional Energy.} We next adopt a more theoretical perspective of optimal transport (see \Cref{ch:ot-eot}). 
Here, the \emph{conditional kinetic energy} quantifies the total expected motion 
of the conditional velocity along the forward path, that is,
the amount of instantaneous movement (or kinetic effort) required to traverse from 
$\rvx_0$ to $\beps$:
\[
\mathcal K[\zeta]
:=\int_0^1 \E_{\rvx_0,\beps}\!\big[\|\rvv_t(\cdot |\rvz)\|_2^2\big]\diff t
=\Big(D+\Tr\Cov[\rvx_0]+\|\E\rvx_0\|_2^2\Big)
\int_0^1 \big(\zeta'(t)\big)^2\diff t.
\]
Minimizing $\mathcal K[\zeta]$ therefore corresponds to finding the smoothest, least-energy
path in expectation within the affine interpolation family
\[
\rvx_t=(1-\zeta(t))\rvx_0+\zeta(t)\beps.
\]
With the boundary conditions $\zeta(0)=0$ and $\zeta(1)=1$, the
Euler--Lagrange equation $\zeta''(t)=0$ gives the minimizer $\zeta(t)=t$, corresponding to a
straight conditional path. This means that, among all smooth schedules $\zeta$ in this family,
the canonical flow $\zeta(t)=t$ is the most energy-efficient way to move from $\rvx_0$ to
$\beps$. We will revisit this point in Proposition~\ref{ot-conditional-flow}
for a more detailed treatment.

\subparagraph{Remark on the Energy of the Oracle Velocity.}
The linear schedule $\zeta(t)=t$ always minimizes the \emph{conditional}
velocity energy derived above, but it need not minimize the energy of the
marginalized oracle velocity $\rvv^*$. The reason is that the magnitude of
$\rvv^*$ can vary along the probability path, so the energy-minimizing schedule
may traverse different parts of the path at different speeds.

To make this dependence explicit, let
\[
\rvy_u=(1-u)\rvx_0+u\beps,
\qquad u\in[0,1],
\]
where $u$ denotes position along the affine probability path, and define
\[
\mathbf{b}_u(\rvy)
:=
\E\!\left[\beps-\rvx_0|\rvy_u=\rvy\right],
\qquad
\kappa(u)
:=
\E_{\rvy_u}
\left[
\|\mathbf{b}_u(\rvy_u)\|_2^2
\right].
\]
Here, $\kappa(u)$ measures the expected squared magnitude of the oracle
velocity per unit movement along the path.

A schedule $u=\zeta(t)$ specifies how quickly this path is traversed. Since
$\rvx_t=\rvy_{\zeta(t)}$, the oracle velocity becomes
\[
\rvv^*(\rvx,t)
=
\zeta'(t)\mathbf{b}_{\zeta(t)}(\rvx),
\]
with kinetic energy
\[
\mathcal{K}_{\mathrm{marg}}[\zeta]
=
\int_0^1
\kappa(\zeta(t))
\bigl(\zeta'(t)\bigr)^2
\diff t.
\]
Assuming $\kappa(u)>0$ and sufficient regularity, minimizing this energy gives
\[
\kappa(\zeta(t))
\bigl(\zeta'(t)\bigr)^2
=
\text{constant},
\qquad\text{or equivalently}\qquad
\zeta'(t)
\propto
\frac{1}{\sqrt{\kappa(\zeta(t))}}.
\]
Thus, the energy-minimizing schedule moves more slowly where the oracle
velocity has larger expected magnitude and faster where it has smaller
magnitude.

Consequently, $\zeta(t)=t$ minimizes the marginal-velocity energy only when
$\kappa(u)$ is constant along the path. This highlights the distinction from
the conditional-velocity result above, where the linear schedule is always the
constant-speed minimum-energy choice within the affine family.

\paragraph{Perspective 2: Why Velocity Prediction Can Be Considered Natural for Sampling.}

Sampling with the PF-ODE ultimately requires its velocity field.
This makes $\rvv$-prediction particularly direct: the network predicts the
quantity that appears on the right-hand side of the ODE itself. By comparison,
$\rvx$-, $\beps$-, and $\rvs$-predictions encode the same oracle velocity
through a combination of the current state and the network prediction.

At the oracle level, these are exactly convertible forms of the same
$\rvv^*(\rvx,t)$ by \Cref{eq:clean_pf_ode}; hence, the underlying oracle
PF-ODE is unchanged by the choice of parameterization. Here, we focus on how
the learned PF-ODE is expressed during sampling; the loss used to train the
corresponding prediction can be chosen separately.

\subparagraph{PF-ODE under $\rvx$-, $\beps$-, and $\rvs$-Predictions.}
The first three identities in \Cref{eq:clean_pf_ode} share the form
\[
\frac{\diff\rvx(t)}{\diff t}
=
\underbrace{L(t)\rvx(t)}_{\text{known linear term}}
+
\underbrace{\rmN_{\bm{\phi}^{\times}}(\rvx(t),t)}_{\text{network-dependent term}},
\]
where $L(t)$ denotes the corresponding known scalar coefficient determined by
the noise schedule, and $\rmN_{\bm{\phi}^{\times}}$ contains the appropriate
time-dependent coefficient multiplying the clean, noise, or score prediction.
For example, under $\beps$-prediction,
\[
L(t)
=
\frac{\alpha_t'}{\alpha_t},
\qquad
\rmN_{\bm{\phi}^{\times}}(\rvx,t)
=
-\sigma_t
\left(
\frac{\alpha_t'}{\alpha_t}
-
\frac{\sigma_t'}{\sigma_t}
\right)
\beps_{\bm{\phi}^{\times}}(\rvx,t).
\]

This form is useful for numerical integration because the known linear term can
be handled analytically while the network-dependent term is approximated
numerically. Exponential-integrator-based diffusion solvers such as
DPM-Solver~\citep{lu2022dpm} exploit precisely this structure. We develop this
solver viewpoint in detail in \Cref{ch:solvers}.

\subparagraph{PF-ODE under $\rvv$-Prediction.}
With $\rvv$-prediction, the PF-ODE takes the direct and simple form
\[
\frac{\diff\rvx(t)}{\diff t}
=
\rvv_{\bm{\phi}^{\times}}(\rvx(t),t)
\approx
\rvv^*(\rvx(t),t).
\]
Thus, a single network evaluation directly provides the quantity required by
a standard ODE solver, without first combining the prediction with a separate
linear term.

The canonical linear schedule makes this connection especially transparent.
For
\[
\rvx_t=(1-t)\rvx_0+t\beps,
\]
a velocity prediction determines the corresponding endpoint predictions
\[
\widehat{\rvx}_0
=
\rvx_t-t\rvv_{\bm{\phi}^{\times}}(\rvx_t,t),
\qquad
\widehat{\beps}
=
\rvx_t+(1-t)\rvv_{\bm{\phi}^{\times}}(\rvx_t,t).
\]
Using this single prediction to move from $t$ to $s$ gives
\[
(1-s)\widehat{\rvx}_0+s\widehat{\beps}
=
\rvx_t
+
(s-t)\rvv_{\bm{\phi}^{\times}}(\rvx_t,t),
\]
which is exactly the plain Euler update for the learned velocity field under
the canonical linear schedule. Thus, the combination of $\rvv$-prediction and
the linear schedule gives a particularly direct connection between the model
prediction and numerical integration of the PF-ODE. We defer a detailed
discussion of numerical solvers and more general schedules to
\Cref{ch:solvers}.

Taken together, $\rvv$-prediction directly provides the velocity required by
the PF-ODE, whereas the other prediction types represent the same oracle
dynamics through schedule-dependent transformations. This direct
correspondence is the main sense in which velocity prediction can be
considered natural for sampling.

\paragraph{Conclusion.}
While $\rvv$-prediction combined with the canonical linear schedule offers certain theoretical advantages, such as a constant target magnitude and the absence of schedule curvature, these properties do not necessarily make it universally superior. In practice, model performance depends on a range of interacting factors, including network architecture, normalization schemes, loss weighting over time, the choice of sampler and discretization steps, guidance strength, regularization strategy, data scaling, and overall training budget. Different datasets and objectives may favor alternative parameterizations or schedules, and the optimal configuration is ultimately an empirical question that should be resolved through validation and ablation studies.

\clearpage
\newpage

\section{Beneath It All: The Fokker–Planck Equation}\label{sec:comparison-connection}

\begin{figure}[htbp]
\begin{center}
\scalebox{0.9}{ 
\centering
\begin{tikzpicture}[
    node distance=1.5cm,
    box/.style={draw, rounded corners=8pt, minimum width=4.5cm, minimum height=1.2cm, align=center, font=\footnotesize},
    section/.style={draw, thick, minimum width=12cm, minimum height=3cm, align=center},
    central/.style={draw, thick, fill=gray!10, minimum width=8cm, minimum height=1.5cm, align=center},
    arrow/.style={->, ultra thick, >=Stealth},
    bidir/.style={<->, thick, >=Stealth},
    label/.style={font=\small, align=center}
]

\node[section] (forward_section) at (0, 4) {\\ \\ \\ \\ \\ \\ \small Forward from $p_{\mathrm{data}}$ at $t = 0$};
\node[font=\Large\sffamily\bfseries] at (0, 5.1) {Forward Process};

\node[box] (var_forward) at (-3.5, 4) {
    \textbfs{Variational Approach:}\\
    Defining $p(\mathbf{x}_t|\mathbf{x}_{t-\Delta t})$, or\\
    $p_t(\mathbf{x}_t|\mathbf{x}_0) \Leftrightarrow \mathbf{x}_t = \alpha_t \mathbf{x}_0 + \sigma_t \boldsymbol{\epsilon}$
};

\node[box] (sde_forward) at (3.5, 4) {
    \textbfs{Forward SDE:}\\
    $\diff\mathbf{x}(t) = f(t)\mathbf{x}(t)\diff t + g(t)\diff \rvw(t)$
};

\node[central] (central) at (0, 1) {
    \textbfs{Continuity Equation/Fokker-Planck Equation}\\
    for $p_t(\mathbf{x})$
};

\node[section] (reverse_section) at (0, -3) [minimum height=5.2cm] {\\ \\ \\ \\  \\ \\ \\ \\ \\ \\ \small Reverse from $p_{\mathrm{prior}}$ at $t = T$};
\node[font=\Large\sffamily\bfseries] at (0, -0.8) {Reverse Process}; 

\node[box, minimum width=3.6cm] (var_reverse) at (-4.0, -2.0) {
    \textbfs{Variational Approach:}\\
    $p(\mathbf{x}_{t-\Delta t}|\mathbf{x}_t)$\\ from Bayes' rule
};

\node[box] (ode) at (3, -4.3) {
    \textbfs{PF-ODE:}\\
    $\diff\mathbf{x} (t) = \mathbf{v}^*(\mathbf{x} (t), t)\diff t$
};

\node[box] (sde_reverse) at (3, -2.0) {
    \textbfs{Reverse SDE:}\\
    $\diff\bar{\mathbf{x}}  = \left[\mathbf{v}^* - \frac{1}{2}\gamma^2(t)\mathbf{s}^*\right]\diff t + \gamma(t)\diff\bar{ \rvw}(t)$
};

\draw[arrow] ($(forward_section.south)+(0,-0.)$) -- ($(central.north)+(0,0.)$);
\draw[arrow] ($(reverse_section.north)+(0,0.)$) -- ($(central.south)+(0,-0.)$);

\draw[bidir] (var_forward) -- (sde_forward) node[midway, above, label] {Lemma~\ref{forward-sde}};


\draw[->, thick, >=Stealth]
    ($(var_reverse.east)+(0,0.1)$) -- ($(sde_reverse.west)+(-0,0.1)$)
    node[midway, above, label] { $\Delta t \to 0$ };

\draw[->, thick, >=Stealth]
    ($(sde_reverse.west)+(0,-0.2)$) -- ($(var_reverse.east)+(0,-0.2)$)
    node[midway, below, label] {\scriptsize{discretization}};

\draw[->, thick, >=Stealth]
    ($(sde_reverse.south)+(0, 0)$) -- ($(ode.north)+(0, 0)$)
    node[midway, right] {$\gamma \equiv 0$};

\draw[->, thick, >=Stealth]
    ($(ode.north)+(-0.4, 0)$) -- ($(sde_reverse.south)+(-0.4, 0)$)
    node[midway, left] {$\gamma \not\equiv 0$};

\end{tikzpicture}
}
\end{center}
\caption{\textbfs{Unified perspective connecting variational, SDE, and ODE formulations through the continuity equation, where all $p_t(\rvx)$ evolve under a shared dynamic.} The velocity field $\rvv^*(\rvx, t) = f(t)\rvx - \frac{1}{2}g^2(t)\rvs^*(\rvx, t)$ is governed by the score function $\rvs^*(\rvx, t) := \nabla_\rvx \log p_t(\rvx)$. Coefficients $f(t)$, $g(t)$, $\sigma_t$, and $\alpha_t$ are pre-defined time-dependent functions, and $\gamma(t)$ is a tunable time-varying hyperparameter.
\figcredit{Created by the authors.}}
\label{fig:unified_framework}
\end{figure}

In this section, we show that the three main perspectives of diffusion models—variational, score-based, and flow-based—are not separate constructions but arise from a single unifying principle: the continuity (Fokker–Planck) equation that governs density evolution under a chosen forward process.

First, we recall that the analysis in \Cref{eq:heuristic-fpe-sde} unifies the variational perspective, based on discrete kernels and Bayes' rule, with the score-based SDE perspective of continuous dynamics. We establish this connection by showing that variational models act as consistent discretizations of the underlying forward and reverse SDEs. Specifically, the marginal densities calculated step-by-step via the discrete kernels evolve in a manner consistent with the Fokker–Planck equation that governs the continuous-time dynamics. This confirms that the two perspectives are fundamentally equivalent.

 We then connect the flow-based and score-based views. In \Cref{subsec:flow-score}, we show that an ODE flow determines a density path whose marginals can always be realized by a family of stochastic processes. This places deterministic flows and stochastic SDEs within the same family.  

Together, these results unify the three perspectives under one framework (see \Cref{fig:unified_framework}). At last, we conclude this chapter  in \Cref{subsec:takeaway-dm-gaussianfm}.

\subsection{Connection of Flow-Based Approach and Score SDE}\label{subsec:flow-score}

A remarkable aspect of diffusion model lies in how different dynamic systems, deterministic or stochastic, can trace out the same evolution of probability distributions. In this section, we reveal a natural and elegant connection between ODE-based flows of \Cref{sec:flow-matching-framework} and Score SDEs. Specifically, we show that the velocity field defining a generative ODE can be transformed into a stochastic counterpart that follows the same Fokker-Planck dynamics, providing a principled bridge between deterministic interpolation and stochastic sampling. This offers us a continuous family of models, ranging from ODEs to SDEs, that generate the same data distribution path.

We consider the continuous-time setup where the perturbation kernel is given by
\[
p_t(\rvx_t|\rvx_0) = \mathcal{N}(\rvx_t; \alpha_t \rvx_0, \sigma_t^2 \rmI),
\]
where $\rvx_0 \sim p_{\mathrm{data}}$. This conditional distribution induces a marginal density path $p_t(\rvx_t) = \mathbb{E}_{\rvx_0 \sim p_{\mathrm{data}}}[p(\rvx_t|\rvx_0)]$ as usual, with $p_T\approx p_{\mathrm{prior}}$.

To match this density path, consider the ODE
\begin{align}\label{eq:ode-velocity}
    \frac{\diff\rvx(t)}{\diff t} = \mathbf{v}_t(\rvx(t)), \quad t \in [0,T],
\end{align}
where $\mathbf{v}_t(\rvx) =\E\left[\alpha_t'\rvx_0+\sigma_t'\beps|\rvx\right]$ is the oracle velocity  as shown in \Cref{eq:marginal-oracle-velocity} (noting that time is flipped to follow the diffusion model convention). Integrating \eqref{eq:ode-velocity} backward from $\rvx(T)\sim p_{\mathrm{prior}}$ yields samples from $p_0$.

Although this ODE suffices for generating high-quality samples, incorporating stochasticity may improve sample diversity. This motivates the following question:
\begin{question}
Is there an SDE whose dynamics, starting from $p_{\mathrm{prior}}$, yield the same marginal densities as the ODE in \Cref{eq:ode-velocity}?
\end{question}
This statement affirms that there exists a family of reverse-time SDEs that induce the same marginal density path as the corresponding PF-ODE.  
 The densities induced by these SDEs satisfy the same Fokker–Planck equation for this path, and therefore their one time marginals coincide with the prescribed interpolation $\{p_t\}_{t\in[0,T]}$\footnote{For completeness, a forward SDE representation (not needed here) is
\[
\diff \rvx(t)=f(t) \rvx(t) \diff t+g(t) \diff \rvw(t),
\]
with $f(t)$ and $g(t)$ related to $(\alpha_t,\sigma_t)$ via \Cref{eq:kernel-sde-equiv}.}.


\proppp{Reverse-Time SDEs Generate the Same Marginals as Interpolations}{rev-sde-interpolant}{
Let $\gamma(t) \geq 0$ be an arbitrary time-dependent coefficient. Consider
the reverse-time SDE, written in the original time variable $t$ and integrated
backward from $T$ to $0$,
\begin{align}\label{eq:general-reverse-sde}
    \diff \bar\rvx(t)
    =
    \Big[
        \rvv^*(\bar\rvx(t), t)
        - \tfrac{1}{2}\gamma^2(t)\rvs^*(\bar\rvx(t), t)
    \Big]\diff t
    + \gamma(t)\diff\bar{\rvw}(t),
    \qquad t:T\to 0 .
\end{align}
with terminal condition $\bar\rvx(T)\sim p_T$. Then this process
$\{\bar\rvx(t)\}_{t\in[0,T]}$ matches the prescribed marginals
$\{p_t\}_{t\in[0,T]}$ induced by the ODE's density path. Here,
$\rvs^*(\rvx,t):=\nabla_{\rvx}\log p_t(\rvx)$ is the score function, and it is
related to the velocity field $\rvv^*(\rvx,t)$ by
\begin{align}\label{eq:score-velocity}
    \rvv^*(\rvx, t)
    =
    f(t)\rvx
    -
    \frac{1}{2}g^2(t)\rvs^*(\rvx,t),
    \quad
    \rvs^*(\rvx,t)
    =
    \frac{1}{\sigma_t}
    \frac{\alpha_t\rvv^*(\rvx,t)-\alpha_t'\rvx}
    {\alpha_t'\sigma_t-\alpha_t\sigma_t'} .
\end{align}
}{
The reverse-time Fokker--Planck equation, written with respect to the original
time variable $t$ while the process is integrated backward, is
\[
\partial_t p_t
=
-\nabla\cdot
\Big(
\big[
\rvv^*
-
\tfrac{1}{2}\gamma^2\rvs^*
\big]p_t
\Big)
-
\tfrac{1}{2}\gamma^2\Delta p_t .
\]
Using the identity $\nabla\cdot(\rvs^*p_t)=\Delta p_t$ since
$\rvs^*=\nabla\log p_t$, the second-order terms cancel, yielding
\[
\partial_t p_t
=
-\nabla\cdot(\rvv^*p_t),
\]
i.e., the first-order continuity equation associated with the PF-ODE. Hence
the reverse-time SDE and the ODE induce the same marginal density path
$\{p_t\}$. See Appendix~A.2--A.3 of \citet{ma2024sit}.
}

At the oracle level of Proposition~\ref{rev-sde-interpolant},
$\gamma(t)$ can be chosen arbitrarily without changing the prescribed
marginal density path: the $\gamma$-dependent drift and diffusion terms cancel
exactly in the Fokker--Planck equation. Some representative choices are:
\begin{itemize}
    \item Setting $\gamma(t) = 0$ recovers the ODE in \Cref{eq:ode-velocity}.
    \item When $\gamma(t) = g(t)$, \Cref{eq:general-reverse-sde} becomes the reverse-time SDE in \Cref{eq:sde_backward}, since the oracle velocity $\rvv(\rvx, t)$ satisfies (see Proposition~\ref{equi-para}):
    \[
    \rvv^*(\rvx, t) = f(t)\rvx - \tfrac{1}{2}g^2(t)\rvs^*(\rvx, t).
    \]
    \item Other choices for $\gamma(t)$ have been explored; e.g., \citet{ma2024sit} select $\gamma(t)$ to minimize the KL gap between $p_{\mathrm{data}}$ and the $t=0$ density obtained by solving \Cref{eq:general-reverse-sde} from $t=T$.
\end{itemize}
Following Score SDE, the trained velocity field $\rvv_{\bm{\phi}^\times}(\rvx, t)$ can be converted into a parameterized score function $\rvs_{\bm{\phi}^\times}(\rvx, t)$ via \Cref{eq:score-velocity}. Plugging this into \Cref{eq:general-reverse-sde} defines an \emph{empirical reverse-time SDE}, which can be sampled by numerically integrating from $t=T$ with $\bar\rvx(T) \sim p_{\mathrm{prior}}$.

This proposition highlights a remarkable flexibility of diffusion models:
once a \emph{marginal density path} $\{p_t\}_{t\in[0,T]}$ is induced by the
forward process, it can be realized by an entire family of reverse-time
dynamics, including the PF-ODE and the reverse-time SDEs
\[
\diff\bar\rvx(t)
=
\big[
\rvv^*(\bar\rvx,t)
-\tfrac12\gamma^2(t)\rvs^*(\bar\rvx,t)
\big]\diff t
+
\gamma(t)\diff\bar\rvw(t),
\quad
\gamma(t)\geq0,
\]
from $t=T$ to $t=0$. Although $\gamma(t)$ changes the stochastic dynamics of individual
trajectories, the corresponding Fokker--Planck equations all reduce, for the
prescribed density path, to the same governing density equation:
\[
\partial_t p_t
=
-\nabla_{\rvx}\cdot
\bigl(\rvv^*(\rvx,t)p_t(\rvx)\bigr).
\]
Indeed, using
$\rvs^*(\rvx,t)=\nabla_{\rvx}\log p_t(\rvx)$, the additional
$\gamma$-dependent drift and diffusion terms cancel at the density level.
Thus, $\gamma(t)$ controls the stochasticity of sample trajectories while
preserving the same oracle one-time marginals, connecting the deterministic
PF-ODE with a family of stochastic SDE counterparts, as illustrated in
\Cref{fig:unified_framework}.

\newpage

\section{Closing Remarks}\label{subsec:takeaway-dm-gaussianfm}




This chapter has served as the keystone of our theoretical exploration, synthesizing the variational, score-based, and flow-based perspectives into a single, cohesive framework. We have shown that these three seemingly distinct approaches are not merely parallel but are deeply and fundamentally interconnected.

Our unification rests on two core insights. First, we identified the secret sauce common to all frameworks: a conditioning trick that transforms an intractable marginal training objective into a tractable conditional one, enabling stable and efficient learning. Second, we established that the Fokker-Planck equation is the universal law governing the evolution of probability densities. All three perspectives, in their own way, construct a generative process that respects this fundamental dynamic.

Furthermore, we showed that the oracle noise, clean-data, score, and velocity
predictions contain the same information and can be converted into one another
algebraically. This oracle equivalence does not make the prediction
parameterization merely cosmetic: finite-capacity networks may approximate
these targets differently, their training objectives can weight errors across
noise levels differently, and finite-step numerical solvers can behave
differently depending on the representation in which the dynamics are
approximated. The important lesson is therefore to distinguish equivalence of
the underlying continuous theory from equivalence of a practical training or
sampling algorithm. At the density level, however, the common principle remains: modern diffusion
methods construct dynamics whose marginals transport a simple prior toward the
data distribution along a prescribed probability path.

With this unified and principled foundation firmly established, we are now equipped to move from the foundational theory to the practical application and acceleration of diffusion models. The central insight that generation is equivalent to solving a differential equation provides a powerful platform for control and optimization. The subsequent parts of this monograph will leverage this unified understanding to address key practical challenges:
\begin{enumerate}
    \item Part C will focus on improving the sampling process at inference time. We will explore how to steer the generative trajectory for controllable generation (\Cref{ch:guidance}) and investigate advanced numerical solvers to dramatically accelerate the slow, iterative sampling process (\Cref{ch:solvers}).
    \item Part D will then look beyond iterative solvers to learn fast generators directly. We will examine methods that can produce high-quality samples in just one or a few steps, either through distillation from a teacher model (\Cref{ch:distillation}) or by training from scratch (\Cref{ch:fast-scratch}).
\end{enumerate}

Having unified the \textit{what} and \textit{why} of diffusion models, we now turn our attention to the exciting and practical frontiers of \textit{how}.

\chapter{(Optional) Diffusion Models and Optimal Transport}\label{ch:ot-eot}

Mapping one distribution to another (with generation as a special case) 
is a central challenge. Flow matching addresses this by learning a 
time-dependent velocity field that transports mass from source to target. 
This naturally connects to transport theory: classical optimal transport 
seeks the minimal cost path between distributions, while its entropy 
regularized form, the Schr\"odinger bridge, selects the most likely 
controlled diffusion relative to a reference such as Brownian motion. 

In this chapter we review the foundations of optimal transport, entropic 
optimal transport, and Schr\"odinger bridges as formulations of the 
distribution–to–distribution problem. This leads to a central question: 
to what extent do diffusion models realize such optimal transports? 
They admit two views: as \emph{stochastic processes}, defined through 
forward and reverse SDEs, and as \emph{deterministic processes}, given 
by PF-ODEs. The stochastic view aligns directly with 
entropic optimal transport, while the PF-ODE does not 
generally correspond to any known transport objective. This gap leaves 
an open question: under what conditions can diffusion models be regarded 
as solving an optimal transport problem?

\chapterlearning{

  \item distinguish a transport coupling, a deterministic transport map, a dynamic transport flow, and a stochastic Schr\"odinger bridge, and connect these notions conceptually to deep generative modeling;

  \item understand what classical OT, entropic OT, and the Schr\"odinger bridge problem optimize, and how these formulations are related;

  \item explain why a standard diffusion forward/reverse SDE does not, in general, solve a prescribed two-endpoint Schr\"odinger bridge problem;

  \item explain why a diffusion PF-ODE defines a deterministic transport map but does not, in general, coincide with the quadratic-cost optimal transport map.

}{This chapter is optional for the main diffusion-model story. Read it for a precise answer to the question ``in what sense, if any, is diffusion optimal transport?''}

\newpage
\section{Prologue of Distribution-to-Distribution Translation} 
Diffusion models fix the terminal distribution to a standard Gaussian, $p_{\mathrm{prior}}$.  
However, many applications require \emph{distribution-to-distribution} translation: transforming a source distribution $p_{\mathrm{src}}$ into a different target $p_{\mathrm{tgt}}$.  
Examples include converting sketches to photorealistic images or translating between artistic styles.

Modern diffusion methods provide practical ways to achieve this.  
\emph{One-endpoint} methods such as SDEdit~\citep{meng2021sdedit} begin with a source image at $t=0$, diffuse it to an intermediate step $t$, and then use a pre-trained diffusion model for the target domain to reverse the process.  
This produces an output that matches the style and content of the target distribution.  

\emph{Two-endpoint} methods, like Dual Diffusion Bridge~\citep{su2022dual}, instead connect the two domains through a shared latent distribution, typically a Gaussian at $t=1$.  
A forward probability–flow ODE transports samples from $p_{\mathrm{src}}$ into this latent space, while a reverse ODE trained on the target domain maps them back to $p_{\mathrm{tgt}}$.  
Beyond such sampling-time approaches, the Flow Matching framework described in \Cref{sec:flow-matching-framework} offers a training-based alternative: it directly learns an ODE flow that continuously moves mass from $p_{\mathrm{src}}$ to $p_{\mathrm{tgt}}$.  

Crucially, transforming between distributions requires more than two separately trained models. 
It demands a principled mapping that aligns the dynamics at \emph{both} endpoints and does so in the ``cheapest'' (cost-efficient) way. 

In this section, rather than surveying the many diffusion-based translation applications, we shift our focus to the mathematical foundations of this classic distribution-to-distribution problem.  
In particular, we highlight optimal transport (OT) and its entropic variant, the Schrödinger bridge (SB), which have long been studied in the theoretical community as canonical formulations of cost-efficient (in mathematical sense) distributional transformation.

At its core,  the fundamental question is: 
\begin{question}
    Given two probability distributions, what is the most efficient way to transform one into the other while minimizing the total \emph{cost}?
\end{question}
Here, the cost, $ c(\mathbf{x}, \mathbf{y}) $, is a non-negative function that assigns a penalty for moving a unit of mass from a point $ \mathbf{x}  $ to a point $ \mathbf{y}$. A common choice is the squared distance, $c(\mathbf{x}, \mathbf{y}) = \|\mathbf{x} - \mathbf{y}\|^2$.

This section provides a brief overview to clarify how diffusion-based approaches, including flow matching, connect to classical and regularized optimal transport. The central question we aim to explore is:
\begin{question}
    Is a diffusion model a form of optimal transport connecting $p_{\text{data}}$ and $p_{\text{prior}}$, and in what sense?
\end{question}

To address this question, we first clarify what ``optimality'' means in
\Cref{sec:ot-intro}. We review classical \emph{optimal transport (OT)} in the static
Monge--Kantorovich form (\Cref{eq:ot}) and its dynamic Benamou--Brenier formulation (\Cref{eq:bb-ot}; minimizing
kinetic energy subject to the continuity equation), as well as the entropy
regularized variant (\emph{entropic OT}) in \Cref{eq:eot-epsilon}, which is equivalent to the \emph{Schr\"odinger
Bridge Problem} (\Cref{eq:sb-kinetic}). In the dynamic view, OT induces a deterministic flow
satisfying the continuity equation, whereas SB induces a controlled diffusion
whose marginals evolve by the Fokker-Planck equation. We provide a high level
map between these formulations in \Cref{sec:relation-ot}.

We then split the discussion into two parts. First, in \Cref{sec:dm-and-sb}, we
explain that the fixed forward noising SDE used in standard diffusion models is
not, by itself, a Schr\"odinger bridge between arbitrary $p_{\mathrm{src}}$ and
$p_{\mathrm{tgt}}$: the forward process is a chosen reference diffusion and both forward-in-time or reverse-time SDE do not, in general, enforce exact endpoint matching to a prescribed target. Hence
it is not entropic OT optimal unless  one explicitly solves the SB problem with those
endpoints; while it is an optimal solution to the half-bridge problem as it is anchored with one starting point.

Second, in \Cref{sec:pfode-ot}, we return to the generative setting with
$p_{\mathrm{src}}=p_{\mathrm{prior}}$ (Gaussian) and
$p_{\mathrm{tgt}}=p_{\mathrm{data}}$. The PF-ODE  defines a deterministic map that transports
$p_{\mathrm{prior}}$ to $p_{\mathrm{data}}$ by construction. However, this flow is generally not an OT map for a prescribed transport cost (e.g., quadratic $W_2$): it realizes one admissible deterministic coupling among many and does not minimize the Benamou–Brenier action. What follows is we discuss if the ``rectify flow'' procedure  (\Cref{subsec:rectify}) can lead to a OT map; however, in general, there is no such theoretical guarantee. Therefore, the exact characterization between diffusion model's PF-ODE map and OT remains a challenging and unsolved problem.

\newpage
\section{Taxonomy of the Problem Setups}\label{sec:ot-intro} 
In this section, we introduce notions of what constitutes the most ``efficient'' or ``optimal'' way to transport mass from $p_{\mathrm{src}}$ to $p_{\mathrm{tgt}}$. These include classical optimal transport (OT) and its entropy-regularized variant, which admits an equivalent formulation known as the Schrödinger Bridge. This taxonomy provides background that will later clarify connections with diffusion models.

\subsection{Optimal Transport (OT)}  

\begin{figure}[th!]
    \centering
    \includegraphics[width=\linewidth]{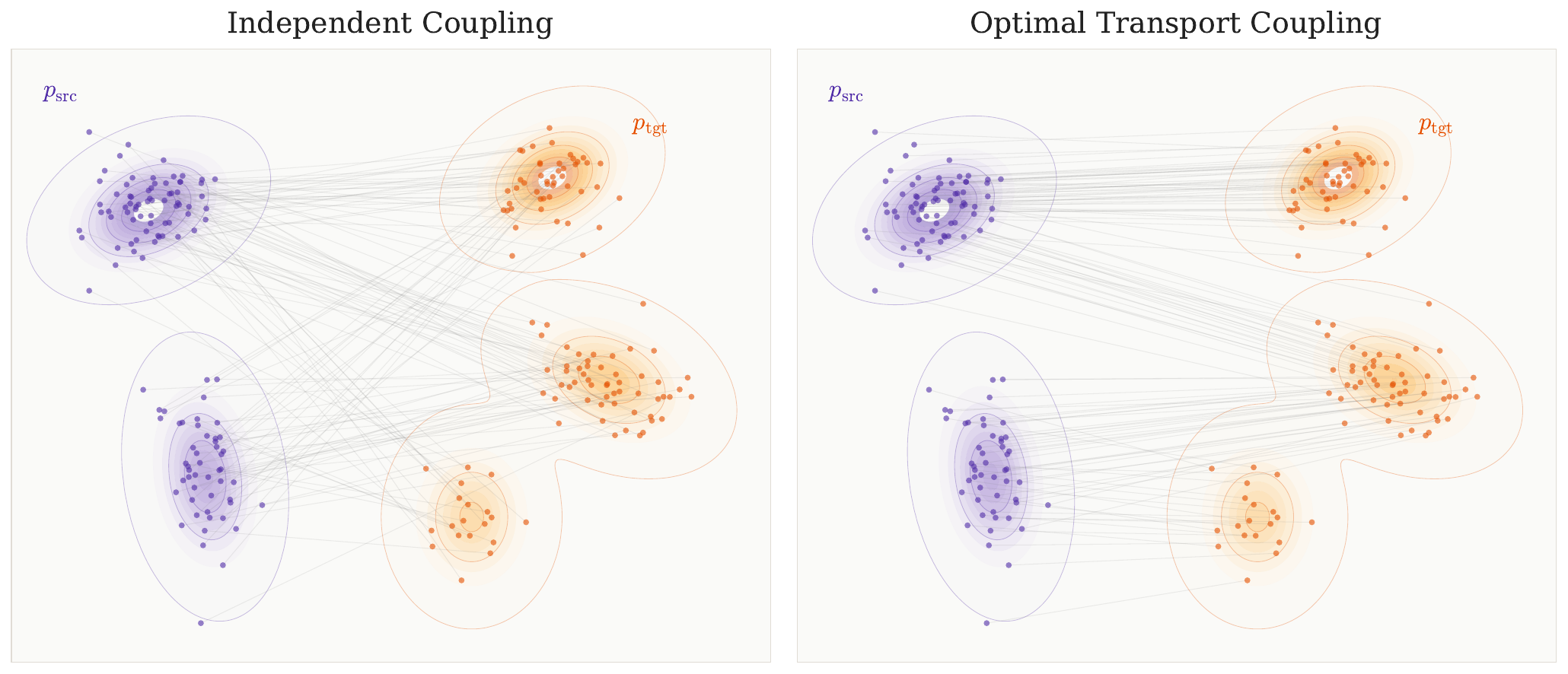}
    \caption{\textbfs{Two couplings between the same $p_{\mathrm{src}}$ and $p_{\mathrm{tgt}}$ in 2D.}
    Left: independent (random) pairing produces crossing paths with high cost.
    Right: the optimal transport coupling pairs nearby mass, avoiding crossings.
    \figcredit{Created by the authors with AI-assisted coding.}}
    \label{fig:ot-2d-coupling}
\end{figure}

\paragraph{Monge--Kantorovich (Static) Formulation of OT Problem.}
We fix a cost function $c:\R^D\times\R^D\to\R$ that specifies the expense of sending probability mass
from $\rvx$ to $\rvy$. The goal is to transform the source distribution $p_{\mathrm{src}}$ into the target
distribution $p_{\mathrm{tgt}}$ as cheaply as possible.

To even define a cost, we must know which pairs $(\rvx,\rvy)$ are matched. This role is played by a
\emph{coupling}: a joint distribution $\gamma$ on $\R^D\times\R^D$ whose marginals are
$p_{\mathrm{src}}$ and $p_{\mathrm{tgt}}$. In other words, sampling $(\rvx,\rvy)\sim\gamma$ means we match
$\rvx$ from the source with $\rvy$ from the target. If $\gamma$ admits a density $\gamma(\rvx,\rvy)$ with
respect to Lebesgue measure, the marginal constraints read
\[
\int_{\R^D} \gamma(\rvx,\rvy)\,\diff \rvy = p_{\mathrm{src}}(\rvx), 
\qquad 
\int_{\R^D} \gamma(\rvx,\rvy)\,\diff \rvx = p_{\mathrm{tgt}}(\rvy).
\]
That is, integrating out $\rvy$ recovers the source density in $\rvx$, while integrating out $\rvx$
recovers the target density in $\rvy$.

We give two standard examples for illustration:
\begin{enumerate}
    \item \textbfs{Discrete Supports.}  
    If $p_{\mathrm{src}}$ and $p_{\mathrm{tgt}}$ are supported on finitely many points, a coupling is
    represented by a nonnegative matrix $(\gamma_{ij})$ whose row sums equal $p_{\mathrm{src}}(i)$ and
    column sums equal $p_{\mathrm{tgt}}(j)$. Each entry $\gamma_{ij}$ is the amount of mass sent from
    $i$ to $j$.
    \item \textbfs{Deterministic Map.}  
    If there exists a measurable map $\rmT$ with $\rmT_\#p_{\mathrm{src}}=p_{\mathrm{tgt}}$, then
    $\gamma=(\rmI,\rmT)_\#p_{\mathrm{src}}$ is a deterministic coupling that moves each point
    $\rvx$ directly to $\rmT(\rvx)$.
\end{enumerate}

Once a coupling $\gamma$ is fixed, the transport cost is simply the average unit cost under this plan:
\[
\int c(\rvx,\rvy)\,\diff\gamma(\rvx,\rvy)
=\E_{(\rvx,\rvy)\sim\gamma}\!\big[c(\rvx,\rvy)\big].
\]
In the discrete case, this reduces to $\sum_{i,j} c_{ij}\gamma_{ij}$, whereas in
the continuous setting it becomes a double integral. In what follows, we will
focus only on the continuous case.

The optimal transport problem is then to choose, among all admissible couplings,
the one that minimizes this expected cost.
\begin{mdframed}
\begin{equation}\label{eq:ot}
\mathrm{OT}\big(p_{\mathrm{src}},p_{\mathrm{tgt}}\big)
:= \inf_{\gamma\in\Gamma(p_{\mathrm{src}},p_{\mathrm{tgt}})}
\int c(\rvx,\rvy)\,\diff\gamma(\rvx,\rvy),
\end{equation}
\end{mdframed}
where the feasible set simply enforces the marginal, or mass-conservation, constraints:
\begin{align*}
    \Gamma(p_{\mathrm{src}},p_{\mathrm{tgt}})
= \Big\{\gamma \in &\mathcal{P}(\R^D\times\R^D): \\
&\int \gamma(\rvx,\rvy) \diff \rvy = p_{\mathrm{src}}(\rvx),\,\,
\int \gamma(\rvx,\rvy)\diff \rvx = p_{\mathrm{tgt}}(\rvy)\Big\},
\end{align*}
where $\mathcal{P}(\R^D\times\R^D)$ denotes the set of all probability measures on
$\R^D\times\R^D$.

\subparagraph{A Special Case: Wasserstein-2 Distance.}

The Wasserstein-2 distance  is a special case of the Monge--Kantorovich problem with the quadratic cost $c(\mathbf{x}, \mathbf{y}) = \|\mathbf{x} - \mathbf{y}\|^2$. It measures the distance between two probability distributions as follows:
\begin{align*}
 \mathcal{W}_2^2(p_{\mathrm{src}}, p_{\mathrm{tgt}}) := \inf_{\gamma \in \Gamma(p_{\mathrm{src}}, p_{\mathrm{tgt}})} \mathbb{E}_{(\mathbf{x}, \mathbf{y}) \sim \gamma} \left[ \|\mathbf{x} - \mathbf{y}\|^2 \right].
\end{align*}

Under suitable assumptions on $p_{\mathrm{src}}$ and $p_{\mathrm{tgt}}$, Brenier's theorem (see \Cref{thm:brenier})\footnote{Brenier's theorem is about the existence and structure of the optimal transport map for quadratic cost. In particular, if $p_{\mathrm{src}}$ does not give mass to sets of dimension at most $D-1$, then an optimal transport map $\rmT^*$ uniquely exists.} guarantees that the optimal coupling $\gamma$ for the quadratic cost is concentrated on the graph of a deterministic map $\rmT:\mathbb{R}^D \rightarrow \mathbb{R}^D$. Consequently, the Wasserstein-2 distance can be equivalently expressed as\footnote{There are three commonly used formulations of the $\mathcal{W}_2$ distance: the Monge formulation (based on an optimal transport map), the Kantorovich formulation (based on couplings), and the Benamou--Brenier dynamic formulation (see \Cref{eq:bb-ot}). These are equivalent under appropriate regularity conditions.}:
\begin{equation}\label{eq: Monge's Formulation}    \mathcal{W}_2^2(p_{\mathrm{src}}, p_{\mathrm{tgt}}) = \inf_{\substack{\rmT:\mathbb{R}^D \rightarrow \mathbb{R}^D,\\ \text{ s.t. }\rmT\#p_{\mathrm{src}} = p_{\mathrm{tgt}}}} \mathbb{E}_{\rvx \sim p_{\mathrm{src}}} \left[ \| \rmT(\rvx) - \rvx \|^2 \right].
\end{equation}
Here, $\rmT\#p_{\mathrm{src}} = p_{\mathrm{tgt}}$ means that $ \rmT $ pushes $ p_{\mathrm{src}} $ forward to $ p_{\mathrm{tgt}} $, i.e., $ \rmT(\rvx) \sim p_{\mathrm{tgt}} $ for $ \rvx \sim p_{\mathrm{src}}$.

Thus, the Wasserstein-2 distance represents the minimal expected squared transport cost among all couplings or transport maps that match the given marginals. The optimal transport map denoted by $\rmT^*(\mathbf{x})$, known as the \emph{Monge map}, yields the most efficient way to transform $p_{\mathrm{src}}$ into $p_{\mathrm{tgt}}$.

\paragraph{Benamou--Brenier (Dynamic) Formulation of OT.} 
\begin{figure}[th!]
    \centering
    \includegraphics[width=0.8\linewidth]{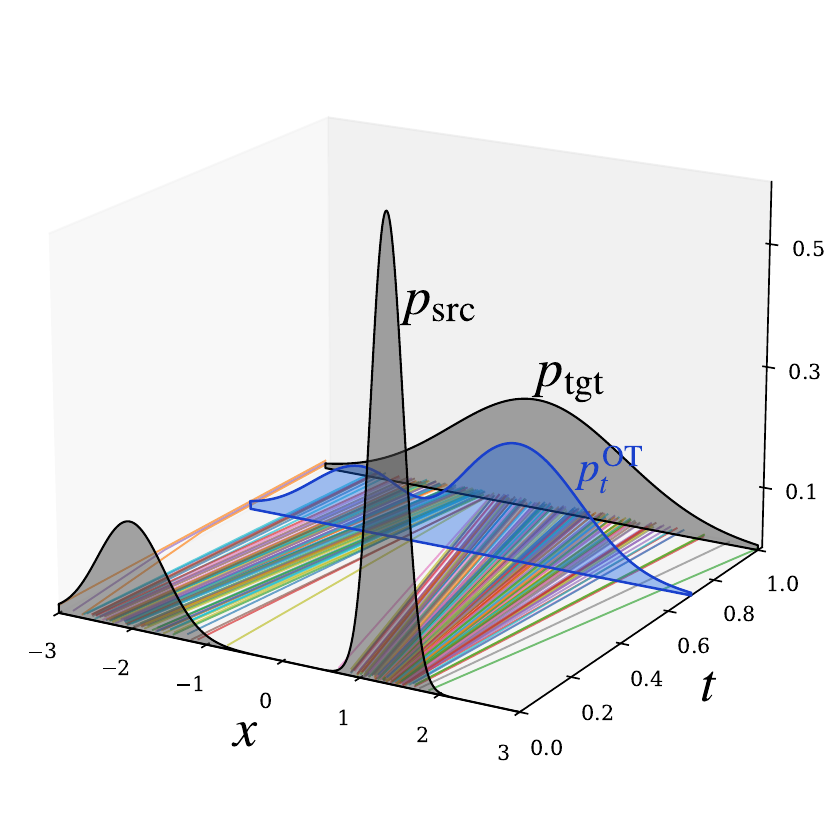}
    \caption{ \textbfs{Illustration of dynamic view of OT.} The interpolation $p_t^{\mathrm{OT}}$ evolves continuously in time, 
providing the least-cost transport plan that deterministically maps 
$p_{\mathrm{src}}$ to $p_{\mathrm{tgt}}$ (the McCann's displacement interpolation).
\figcredit{Created by the authors.}}
    \label{fig:ill-ot}
\end{figure}

Instead of mapping distributions directly in a static manner, as in the Monge--Kantorovich formulation, transport can also be modeled as a continuous-time flow:
\begin{equation*}
    p_0 := p_{\mathrm{src}} \to p_t \to p_1 := p_{\mathrm{tgt}}, \quad t \in [0,1].
\end{equation*}
This dynamic formulation of optimal transport, introduced by \citet{benamou2000computational}, seeks a smooth velocity field $ \rvv_t(\rvx) $ that describes how mass in $ p_t(\rvx) $ evolves over time.

The Benamou--Brenier formulation\footnote{Benamou–Brenier formulation describes how to compute the $ \mathcal{W}_2 $ distance by minimizing kinetic energy over continuous paths of measures and velocities.} shows that, for the quadratic cost $ c(\rvx, \rvy) = \|\rvx - \rvy\|_2^2 $ (i.e., the $ \mathcal{W}_2 $ distance), the optimal value of the static OT problem in \Cref{eq:ot} is equal to the optimal value of the kinetic energy minimization problem:
\begin{mdframed}
   \begin{equation}\label{eq:bb-ot}
    \mathcal{W}_2^2(p_{\mathrm{src}}, p_{\mathrm{tgt}}) = \min_{\substack{ (p_t, \rvv_t) \text{ s.t. }
        \partial_t p_t + \nabla \cdot (p_t \rvv_t) = 0, \\
        p_0 = p_{\mathrm{src}},\,\, p_1 = p_{\mathrm{tgt}}
    }}
    \int_0^1 \int_{\mathbb{R}^D} \|\rvv_t(\rvx)\|^2 p_t(\rvx) \diff\rvx \diff t
\end{equation} 
\end{mdframed}
where $p_t$ is a probability distribution on $\mathbb{R}^D$ for each $t \in [0,1]$. In particular, The optimal transport flow $p_t(\rvx)$ follows \textit{McCann’s displacement interpolation}:
\begin{equation*}
    \rmT_t^*(\rvx) = (1 - t)\rvx + t\rmT^*(\rvx),
\end{equation*}
where $\rmT^*(\rvx)$ is the OT map that transports $p_{\mathrm{src}}$ to $p_{\mathrm{tgt}}$. This linear interpolation moves mass along straight lines with constant velocity: $p_t = \rmT_t^*\#p_{\mathrm{src}}$ for each $t \in [0,1]$. 

The optimal transport map $\rmT^*$ satisfies the \emph{Monge–Ampère equation}:
\begin{equation}\label{eq:monge-ampere}
    p_{\mathrm{tgt}}\left(\nabla \psi(\rvx)\right) \det\left(\nabla^2 \psi(\rvx)\right) = p_{\mathrm{src}}(\rvx),
\end{equation}
where $\rmT^*(\rvx) = \nabla \psi(\rvx)$ for some convex function $\bpsi$ by Brenier’s theorem. However, this nonlinear PDE is typically intractable for explicit solutions. Note that this is precisely the change-of-variables relation used by normalizing flows (c.f., \Cref{eq:nf-change-of-var}): flows parametrize an invertible transport map with a tractable Jacobian determinant, but do not in general impose the gradient-of-potential structure $\rmT^*=\nabla\psi$; consequently, a trained flow can differ substantially from the Brenier/OT map.

\subsection{Entropy-Regularized Optimal Transport (EOT)}

To motivate EOT concretely, consider empirical distributions built from samples. 
Suppose $p_{\mathrm{src}}$ is supported on points $\{\rvx^{(i)}\}_{i=1}^n \subset \R^D$ with weights $a_i$, 
and $p_{\mathrm{tgt}}$ on $\{\rvy^{(j)}\}_{j=1}^n \subset \R^D$ with weights $b_j$. 
A coupling is then an $n\times n$ nonnegative matrix $\gamma=(\gamma_{ij})$ whose row sums match $a$ and column sums match $b$. 
Each entry $\gamma_{ij}$ represents the amount of mass transported from $\rvx^{(i)}$ to $\rvy^{(j)}$\footnote{Empirical (discrete) measures provide a principled proxy for continuous distributions. 
When the ground cost is $c(x,y)=d(x,y)^p$ (so the OT value equals $W_p^p$) and the measures have finite $p$th moments, the empirical measures converge to the population in $W_p$ with quantitative rates; see \citet{fournier2015rate} and the overview in \citet{peyre2019computational}.}.

\paragraph{Why Regularize OT?}
Classical OT in this discrete setting (obtained by taking counting measures in the continuous formulation \Cref{eq:ot}) reduces to minimizing
\[
\min_{\gamma=(\gamma_{ij})}\sum_{i,j} C_{ij}\,\gamma_{ij},
\]
over all feasible couplings $\gamma=(\gamma_{ij})$, 
where $C_{ij} = c(\rvx^{(i)},\rvy^{(j)})$ is the cost of moving one unit of mass from source point $\rvx^{(i)}$ to target point $\rvy^{(j)}$, for a prescribed ground cost 
$c:\R^D\times\R^D\to\R_{\ge 0}$ (e.g., $c(\rvx,\rvy)=\|\rvx-\rvy\|_2^2$).

Two main issues arise:
\begin{enumerate}
    \item \textbfs{Non-Uniqueness and Instability:} The minimizer $\gamma^*$ need not be unique. 
    For example, if two transport plans achieve the same minimum cost, the solver may select either one. 
    Consequently, small changes in the inputs $(a,b,C)$ (such as moving a sample, adjusting weights, or slightly perturbing costs) can cause abrupt jumps in the solution.  
    \item \textbfs{High Computational Cost:} The problem is a linear program with $n^2$ variables and $2n$ constraints. 
    Practical solvers (e.g., Hungarian algorithm~\citep{kuhn1955hungarian,munkres1957algorithms}, network simplex~\citep{peyre2019computational}) typically scale as $\mathcal{O}(n^3)$, which is infeasible for large $n$.
\end{enumerate}

To overcome these bottlenecks,  EOT objective function introduces a regularization term to the classical OT problem, controlled by a parameter $\varepsilon > 0$:
\begin{mdframed}
    \begin{align}\label{eq:eot-epsilon}
\begin{aligned}
        \text{EOT}_\varepsilon(p_{\mathrm{src}}, p_{\mathrm{tgt}}) &:= \min_{\gamma \in \Gamma(p_{\mathrm{src}}, p_{\mathrm{tgt}})} \int c(\rvx,  \rvy)  \diff\gamma(\rvx, \rvy) 
        + \varepsilon \mathcal{D}_{\text{KL}}\left(\gamma \Vert M \right).
\end{aligned}
\end{align}
\end{mdframed}
The reference measure $M$ is typically chosen as the product of the marginals, $p_{\mathrm{src}} \otimes p_{\mathrm{tgt}}$. The KL divergence term is directly related to the Shannon entropy of the transport plan $\gamma$:
\[
\mathcal{D}_{\text{KL}}(\gamma \,\Vert\, p_{\mathrm{src}} \otimes p_{\mathrm{tgt}}) = -\mathcal{H}(\gamma) + \text{Constant},
\]
where $\mathcal{H}(\gamma) := -\int \gamma(\rvx, \rvy) \log \gamma(\rvx, \rvy) \diff \rvx \diff \rvy$.

The addition of this regularization term yields several theoretical and practical advantages, which we briefly outline below:
\paragraph{Why Entropy Regularizer Helps?}
\begin{enumerate}
  \item \textbfs{Mass Spreading.}
  Since $t\mapsto t\log t$ is convex and grows rapidly for large $t$, minimizing $\int \gamma\log\gamma$
  penalizes \emph{peaky} couplings (some $\gamma(\rvx,\rvy)$ very large, most near zero).
  For fixed total mass $\int \gamma=1$, it favors plans where $\gamma(\rvx,\rvy)$ is more evenly distributed
  over $(\rvx,\rvy)\in\mathbb{R}^D\times\mathbb{R}^D$.
  Equivalently, maximizing Shannon entropy promotes higher ``uncertainty'' (diffuseness).

  \item \textbfs{Strict Convexity and Uniqueness.}
  Because $\mathcal{H}$ is strictly concave, the objective in \Cref{eq:eot-epsilon} is strictly convex in $\gamma$,
  yielding a \emph{unique} minimizer $\gamma^*_\varepsilon$ that depends continuously on $(p_{\mathrm{src}},p_{\mathrm{tgt}},c)$.

  \item \textbfs{Sinkhorn Form and Positivity.}
  Under mild conditions\footnote{We assume that $c<\infty$ holds $p_{\mathrm{src}}\otimes p_{\mathrm{tgt}}$-almost everywhere, and that the marginal kernel integrals are finite and positive. 
For simplicity, we focus on the case where $\gamma^*_\varepsilon$, $p_{\mathrm{src}}$, and $p_{\mathrm{tgt}}$ admit densities with respect to the Lebesgue measure.
},
  the optimizer has the \emph{Schrödinger/Sinkhorn form}
  \[
  \gamma^*_\varepsilon(\rvx,\rvy) = u(\rvx)\, \exp \bigl(-\tfrac{c(\rvx,\rvy)}{\varepsilon}\bigr)  \,v(\rvy)p_{\mathrm{src}}(\rvx) p_{\mathrm{tgt}}(\rvy),
  \]
  for positive scaling functions $u,v$ (unique up to a global factor). In practice, the continuous formulation is approximated with finite samples, reducing EOT to a finite (sampled) Sinkhorn iteration. 
The entropic objective is strictly convex, and the scaling (Sinkhorn/IPFP) algorithm solves it efficiently~\citep{sinkhorn1964relationship,cuturi2013sinkhorn}. 
For a dense problem with $n$ support points per marginal (an $n\times n$ kernel), each Sinkhorn iteration costs $\mathcal{O}(n^2)$ time and $\mathcal{O}(n^2)$ memory, 
making the method more scalable and practical~\citep{altschuler2017near}.

  \item \textbfs{Limits in $\varepsilon$.}
As $\varepsilon \to 0$, the optimal plan $\gamma^*_\varepsilon$ becomes increasingly concentrated, approaching a (possibly singular) classical OT coupling (we will revisit this connection in \Cref{subsec:eot-ot}). 
As $\varepsilon$ increases, $\gamma^*_\varepsilon$ gradually spreads out and approaches the independent coupling $p_{\mathrm{src}}\otimes p_{\mathrm{tgt}}$. 
\end{enumerate}

\subsection{Schrödinger Bridge (SB)}\label{subsec:sb}

\paragraph{KL Formulation of SB.}
The Schrödinger Bridge (SB) problem, introduced by Erwin Schrödinger in the 1930s,
asks the following question. Suppose particles move according to some simple
reference dynamics, such as Brownian motion. Now imagine that we observe the
particles at two times: at $t=0$ their distribution is $p_{\mathrm{src}}$, and
at $t=1$ it is $p_{\mathrm{tgt}}$. Among all possible stochastic processes
that connect these two distributions, which one deviates the least from the
reference dynamics? Here ``deviation'' is measured by the KL divergence, so the
solution to the SB problem is the most likely way to deform Brownian motion
into a process that satisfies the prescribed boundary conditions.

To make this precise, let $\rvx_{0:T}:=\{\rvx_t\}_{t\in[0,T]}$ denote a complete
trajectory of the process. We write $P$ for the \emph{law of trajectories},
that is, the probability distribution over entire sample paths. The time-$t$
marginal of $P$ is denoted by $p_t$ (or $P_t$), which describes the
distribution of the state $\rvx_t$ at a single time. Formally, for a measurable
set $A\subseteq\mathbb R^D$,
\[
p_t(A) = P(\rvx_t \in A).
\]
In other words, $p_t$ can be viewed as the empirical distribution obtained by
sampling many full trajectories from $P$ and then collecting the states at
time $t$—for instance, as a histogram if the state is one-dimensional.

Consider a reference diffusion $\{\rvx_t\}_{t\in[0,T]}$ governed by the SDE
\begin{align}\label{eq:general-sb-sde}
    \diff \rvx_t  =  \rvf(\rvx_t,t)\,\diff t  +  g(t)\,\diff \rvw_t,
\end{align}
where $\rvf\colon \R^D\times[0,T]\!\to\!\R^D$, $g\colon[0,T]\!\to\!\R$, and
$\{\rvw_t\}_{t\in[0,T]}$ is a standard Brownian motion.  
Let $R$ denote the \emph{path law} (joint distribution) of the full trajectory
$\rvx_{0:T}:=\{\rvx_t\}_{t\in[0,T]}$, and let $R_0$ denote the initial marginal of the reference process; this $R$ will serve as the \emph{reference}
trajectory distribution.

With this notation, the SB problem seeks a trajectory law $P$
that is closest to $R$ in KL divergence while matching the prescribed endpoint
marginals:
\begin{align}\label{eq:sb-epsilon-general}
    \mathrm{SB}(p_{\mathrm{src}}, p_{\mathrm{tgt}})
    := \min_{P}\, \mathcal{D}_{\mathrm{KL}}(P\|R)
    \quad \text{s.t.} \quad P_0 = p_{\mathrm{src}},\;\; P_T = p_{\mathrm{tgt}}.
\end{align}
The optimizer $P^*$ depends on the chosen reference process $R$.

\paragraph{Stochastic Control View of SB.}

\begin{figure}[th!]
    \centering
    \includegraphics[width=0.75\linewidth]{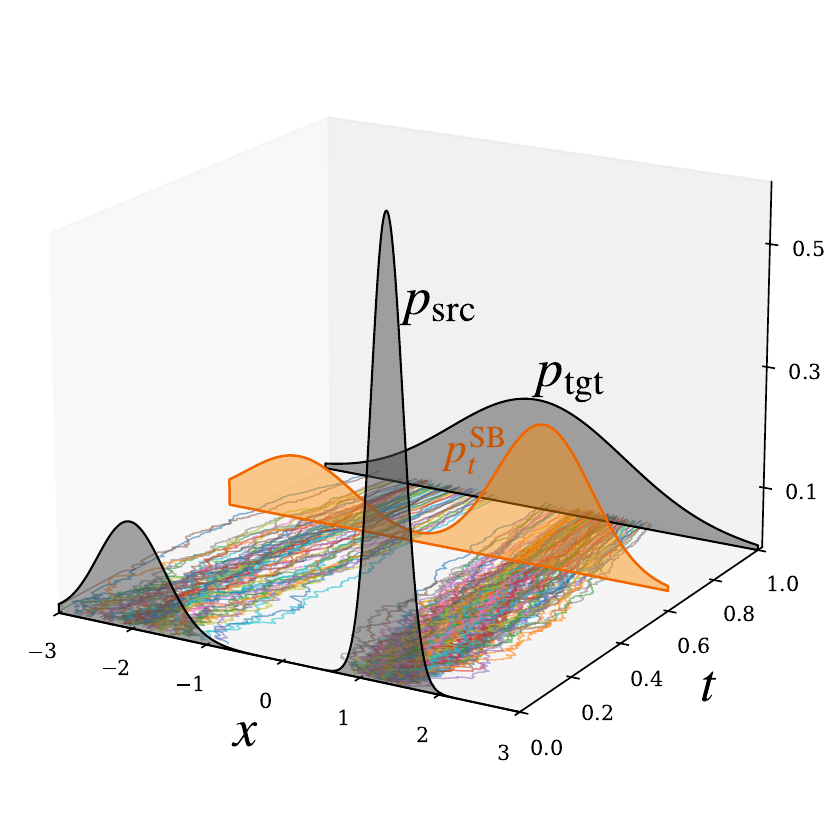}
    \caption{\textbfs{Illustration of stochastic control view of SB.} The bridge seeks the stochastic path that deviates least from the reference while connecting $p_{\mathrm{src}}$ and $p_{\mathrm{tgt}}$.
    \figcredit{Created by the authors.}}
    \label{fig:ill-sb}
\end{figure}

Rather than optimizing over arbitrary path distributions $P$ in
\Cref{eq:sb-epsilon-general}, a more tractable approach is to take the reference
dynamics as an anchor and allow it to drift. This is done by
introducing a time-dependent drift $\rvv_t(\rvx_t)$, which perturbs the
reference process and generates a family of candidate trajectory
distributions. The resulting dynamics take the form of a \emph{controlled
diffusion}:
\[
    \diff \rvx_t
    = \bigl[\rvf(\rvx_t,t) + \rvv_t(\rvx_t)\bigr]\diff t + g(t) \diff \rvw_t,
\]
where $\rvv_t:\R^D\to\R^D$ is the drift to be optimized later (\Cref{eq:sb-kinetic}).
 Under standard integrability
conditions (e.g., Novikov’s condition) and by Girsanov’s theorem
(see \Cref{subsec:girsanov}), the KL divergence between the controlled law $P$
and the reference $R$ admits the dynamic (kinetic) form
\begin{align*}
    \mathcal{D}_{\mathrm{KL}}(P\|R)
    &=
    \mathcal{D}_{\mathrm{KL}}(p_{\mathrm{src}}\|R_0)
    +
    \E_{P}\!\left[
    \frac{1}{2}\int_0^T
    \frac{\|\rvv_t(\rvx_t)\|^2}{g^2(t)}
    \diff t
    \right]
    \\
    &=
    \mathcal{D}_{\mathrm{KL}}(p_{\mathrm{src}}\|R_0)
    +
    \frac{1}{2}\int_0^T\!\int_{\R^D}
    \frac{\|\rvv_t(\rvx)\|^2}{g^2(t)}
    p_t(\rvx)\diff\rvx\diff t .
\end{align*}
where $p_t$ is the time–$t$ marginal of $\rvx_t$ under the controlled process. The
second equality follows from the law of total expectation. Since $P_0=p_{\mathrm{src}}$ is fixed in the SB problem, the first term is
constant with respect to the control and may therefore be omitted from the
optimization. In particular, it vanishes when $R_0=p_{\mathrm{src}}$.

Hence, the SB problem can be reformulated as minimizing the expected control
energy over all admissible drifts $\rvv_t$ that steer the process from
$p_{\mathrm{src}}$ at $t=0$ to $p_{\mathrm{tgt}}$ at $t=T$
\citep{dai1991stochastic,pra1990markov,pavon1991free,chen2016relation}.
Since the initial marginal is fixed to $p_{\mathrm{src}}$, we may take the
reference process to have $R_0=p_{\mathrm{src}}$ without changing the
optimizer; then
$\mathcal{D}_{\mathrm{KL}}(p_{\mathrm{src}}\|R_0)=0$.
This leads to the \emph{stochastic control formulation}:
\begin{mdframed}
\begin{align}\label{eq:sb-kinetic}
\begin{aligned}
\mathrm{SB}(p_{\mathrm{src}}, p_{\mathrm{tgt}})
=
\min_{\substack{
\mathbf{v}_t \text{ s.t. }
\diff \rvx_t =
\left[\rvf(\rvx_t,t)+\mathbf{v}_t(\rvx_t)\right]\diff t
+g(t)\diff\rvw_t,\\
\rvx_0\sim p_{\mathrm{src}},\,
\rvx_T\sim p_{\mathrm{tgt}}
}}
\frac{1}{2}
\int_0^T\int_{\mathbb{R}^D}
\frac{\norm{\rvv_t(\rvx)}^2}{g^2(t)}
p_t(\rvx)\diff\rvx\diff t .
\end{aligned}
\end{align}
\end{mdframed}
Importantly, the endpoint distributions $p_{\mathrm{src}}$ and $p_{\mathrm{tgt}}$ are arbitrary;
the control $\rvv_t$ is chosen precisely to ``bridge'' the reference dynamics between
these marginals while staying as close as possible (in KL divergence) to the reference
process $R$.

\paragraph{A Special Brownian Reference.}
\Cref{eq:sb-kinetic} resembles the Benamou--Brenier formulation of OT
in \Cref{eq:bb-ot}, especially when the reference process
$R^\varepsilon$ (with $\varepsilon>0$) is chosen to be a Brownian
motion\footnote{This resemblance can be made precise. In an equivalent
time-symmetric fluid-dynamic formulation, if $\tilde{\rvv}_t$ denotes
the current velocity satisfying the continuity equation
$\partial_t \rho_t + \nabla \cdot (\rho_t \tilde{\rvv}_t)=0$, then the
Brownian-reference SB objective (\Cref{eq:sb-epsilon-kinetic}) can be
written as a Benamou--Brenier-type kinetic term, up to a conventional
constant factor, plus a Fisher information term:
\[
\int_0^T\!\!\int
\left[
\frac{1}{2}\|\tilde{\rvv}_t(\rvx)\|^2
+
\frac{\varepsilon^2}{8}\|\nabla_{\rvx}\log \rho_t(\rvx)\|^2
\right]\rho_t(\rvx)\,\diff\rvx\,\diff t.
\]
Thus, in the Brownian-reference / quadratic-cost case, entropic OT can
be viewed as a Fisher-information-regularized version of dynamic OT; see
Problem~4.6 in~\citet{chen2021stochastic}.}:
\[
\diff \rvx_t = \sqrt{\varepsilon}\,\diff\rvw_t,
\]
so that $\rvf \equiv \mathbf{0}$ and $g(t) \equiv \sqrt{\varepsilon}$.

In this setting, the SB problem seeks a path distribution $P$ that stays closest
(in KL divergence) to the Brownian reference $R^\varepsilon$, while matching the
endpoint marginals:
\begin{align}\label{eq:sb-epsilon}
    \mathrm{SB}_\varepsilon(p_{\mathrm{src}}, p_{\mathrm{tgt}})
    := \min_{P} \mathcal{D}_{\mathrm{KL}}(P \| R^\varepsilon)
    \quad \text{s.t.} \quad P_0 = p_{\mathrm{src}}, \;\; P_T = p_{\mathrm{tgt}}.
\end{align}
The equivalent stochastic control formulation then becomes
\begin{align}\label{eq:sb-epsilon-kinetic}
\begin{aligned}
    \mathrm{SB}_\varepsilon(p_{\mathrm{src}}, p_{\mathrm{tgt}})
    = \min_{\substack{
        \rvv_t \text{ s.t. } \diff \rvx_t = \rvv_t(\rvx_t)\,\diff t + \sqrt{\varepsilon}\,\diff \rvw_t, \\
        \rvx_0 \sim p_{\mathrm{src}},\;\; \rvx_T \sim p_{\mathrm{tgt}}
    }}
    \frac{1}{2\varepsilon}\int_0^T \!\!\int_{\R^D}
        \|\rvv_t(\rvx)\|^2 \, p_t(\rvx)\,\diff \rvx\,\diff t,
\end{aligned}
\end{align}
where $p_t$ denotes the time--$t$ marginal of diffusion driven by the drift $\rvv_t$:
\[
\diff \rvx_t = \rvv_t(\rvx_t)\,\diff t + \sqrt{\varepsilon}\,\diff \rvw_t.
\]

\paragraph{Why We Need to Specify a Reference Distribution?}
Classical OT does not require a reference path distribution because the
prescribed ground cost already tells us how expensive it is to pair a source
point with a target point. The optimization therefore compares admissible
couplings directly through this cost. In the SB problem, however, we compare
entire stochastic trajectories rather than only their endpoints. We therefore
need a reference path measure $R$ that specifies what the uncontrolled
dynamics would normally look like. The quantity
$\mathcal{D}_{\mathrm{KL}}(P\|R)$ then measures how much an admissible process
$P$ must deviate from these reference dynamics in order to satisfy the
prescribed endpoint distributions.

\paragraph{Coupled PDE Characterization.}  
A convenient way to describe the SB solution is through two 
space–time potentials $\Psi(x,t)$ and $\widehat{\Psi}(x,t)$. Let
$p_t^{\mathrm{SB}}$ denote the marginal at time $t\in[0,T]$ of the optimal
trajectory law $P^*$ in \Cref{eq:sb-epsilon-general}. Then one has the
symmetric factorization~\citep{dai1991stochastic}
\begin{align}\label{eq:sb-symmetry}
p_t^{\mathrm{SB}}(x) = \Psi(x,t) \widehat{\Psi}(x,t),
\end{align}
where $\Psi$ and $\widehat{\Psi}$ solve the (linear) \emph{Schrödinger system}~\citep{caluya2021wasserstein,chen2021stochastic,chenlikelihood}:
\begin{align}
\begin{aligned}\label{eq:pde-schrodinger}
\frac{\partial \Psi}{\partial t}(\mathbf{x},t)
  &= -\nabla_{\mathbf{x}} \Psi(\mathbf{x},t) \cdot \rvf(\mathbf{x},t)
     - \frac{g^2(t)}{2}\,\Delta_\mathbf{x} \Psi(\mathbf{x},t),  \\
\frac{\partial \widehat{\Psi}}{\partial t}(\mathbf{x},t)
  &= -\nabla_{\mathbf{x}}\!\cdot\!\bigl(\widehat{\Psi}(\mathbf{x},t)\,\rvf(\mathbf{x},t)\bigr)
     + \frac{g^2(t)}{2}\,\Delta_\mathbf{x}\widehat{\Psi}(\mathbf{x},t) 
\\ \text{subject to}\quad
\Psi(\mathbf{x},0)&\,\widehat{\Psi}(\mathbf{x},0) = p_{\mathrm{src}}(\mathbf{x}),
\quad
\Psi(\mathbf{x},T)\,\widehat{\Psi}(\mathbf{x},T) = p_{\mathrm{tgt}}(\mathbf{x}). 
\end{aligned}
\end{align}

\subparagraph{Forward-Time Schrödinger Bridge SDE.}
Once $\Psi$ is known, the optimal dynamics is the reference diffusion tilted by
the space–time factor $\Psi$:
\begin{align}\label{eq:sb-forward-sde}
    \diff \rvx_t = \bigl[\rvf(\rvx_t, t) + g^2(t)\,\nabla_{\mathbf{x}} \log \Psi(\rvx_t, t)\bigr]\diff t
+ g(t)\diff \rvw_t, \quad \rvx_0 \sim p_{\mathrm{src}}.
\end{align}
Let $Q$ denote the trajectory law of \Cref{eq:sb-forward-sde}
(so $Q_0=p_{\mathrm{src}}$ and $Q_T=p_{\mathrm{tgt}}$ by
\Cref{eq:sb-symmetry,eq:pde-schrodinger}). Then
\emph{$Q=P^*$} and the minimizer $\rvv^*$ to \Cref{eq:sb-kinetic} is (see \citep{chen2021stochastic}'s Section 4.6): 
\[
\rvv_t^{*}(\rvx) = g^2(t) \nabla_\rvx \log \Psi(\rvx,t).
\]
That is,
drift correction $g^2\nabla_\rvx\log\Psi$ is precisely the
minimal KL perturbation of the reference needed to match the endpoint
marginals. 


\subparagraph{Reverse-Time Schrödinger Bridge SDE.}
The same optimal path law can be generated reverse in time. A convenient way
to see the form of the reverse-time drift is to conceptually use the standard time-reversal identity for diffusions:
\[
\rvb^{-}(\rvx,t)=\rvb^{+}(\rvx,t)-g^2(t)\nabla_\rvx\log p_t^{\mathrm{SB}}(\rvx),
\]
where $\rvb^{+}=\rvf+g^2\nabla\log\Psi$ and $p_t=\Psi\,\widehat{\Psi}$. This gives
\[
\rvb^{-}(\rvx,t)=\rvf(\rvx,t)-g^2(t) \nabla_\rvx\log \widehat{\Psi}(\rvx,t).
\]
Thus the reverse-time SDE reads
\begin{align}\label{eq:sb-backward-sde}
\diff \rvx_t
= \Big[\rvf(\rvx_t,t) - g^2(t) \nabla_{\rvx}\log \widehat{\Psi}(\rvx_t,t)\Big]\diff t
+ g(t)\,\diff \bar{\rvw}_t, \quad \rvx_T \sim p_{\mathrm{tgt}} .
\end{align}
Equivalently, reparametrizing time by $\rvy_\tau := \rvx_{T-\tau}$ so that
$\tau$ increases from $0$ to $T$. Then $\rvy_\tau$ evolves forward in $\tau$ from $\rvy_0 \sim p_{\mathrm{tgt}}$ as
\begin{align}\label{eq:sb-reverse-forward-tau}
\begin{aligned}
    \diff \rvy_\tau
= \big[-\rvf(\rvy_\tau,T-\tau) + g^2(T-\tau) \nabla_{\rvy}\log \widehat{\Psi}(\rvy_\tau,T-\tau)\big]\diff \tau
+ g(T-\tau) \diff \rvw_\tau.
\end{aligned}
\end{align}

Both the forward and reverse descriptions yield the same optimal path law $P^*$ which are linked by
\[
\nabla\log p_t^{\mathrm{SB}}=\nabla\log\Psi+\nabla\log\widehat\Psi,
\qquad
\rvb^{-}=\rvb^{+}-g^2\,\nabla\log p_t^{\mathrm{SB}},
\]
so their marginals coincide with $p_t^{\mathrm{SB}}$ at every time.  The additional drift terms $g^2\nabla\log\Psi$ (forward) and
$-\,g^2\nabla\log\widehat{\Psi}$ (reverse-time) act as control forces that steer
the reference diffusion to match the endpoint marginals while staying closest
to the reference in relative entropy.

\subparagraph{Practical Obstacles to the Coupled PDE Approach.}
To construct the generative process based on \Cref{eq:sb-backward-sde}, one must solve the coupled PDEs in \Cref{eq:pde-schrodinger} to obtain the backward Schrödinger potential $\widehat{\Psi}$. However, these PDEs are notoriously difficult to solve, even in low-dimensional settings, which makes their direct application in generative modeling challenging. To circumvent this, several works have proposed alternative strategies: leveraging Score SDE techniques to iteratively solve each half-bridge problem ($p_{\mathrm{tgt}} \leftarrow p_{\mathrm{src}}$ and $p_{\mathrm{tgt}} \rightarrow p_{\mathrm{src}}$)~\citep{de2021diffusion}; optimizing surrogate likelihood bounds~\citep{chenlikelihood, liu20232}; or designing simulation-free training based on an analytical solution of the posterior $\rvx_t \vert \rvx_0, \rvx_T$ for sample pairs $(\rvx_0, \rvx_T) \sim p_{\mathrm{src}} \otimes p_{\mathrm{tgt}}$~\citep{liu20232}. We do not delve into the technical details here but briefly discuss the connection between diffusion models and SB in \Cref{sec:dm-and-sb}.

\subsection{Global Pushforwards and Local Dynamics: An OT Analogy for DGMs}
From the optimal-transport viewpoint (in \Cref{eq:ot}), one can leverage deep generative models to learn a transport (pushforward) map from a simple prior to the data, i.e., 
$\rmG_{\bm{\phi}\#}p_{\mathrm{prior}} \approx p_{\mathrm{data}}$. 
Although $\rmG_{\bm{\phi}}$ generally does not coincide with the optimal transport map 
(except in works~\citep{genevay2018learning,onken2021ot} that impose an OT objective under suitable conditions), 
the Benamou--Brenier formulation (in \Cref{eq:bb-ot}) provides a complementary, dynamic perspective. 
Rather than directly learning a single global map, it describes transport as a continuous flow generated by a time-dependent local vector field, tracing a smooth path between $p_{\mathrm{prior}}$ and $p_{\mathrm{data}}$. 
This dynamic formulation parallels the relationship between the static Schrödinger Bridge problem (in \Cref{eq:sb-epsilon-general}) and its stochastic-control counterpart (in \Cref{eq:sb-kinetic}), 
where the optimal coupling is realized as a controlled diffusion process. 
A similar analogy emerges in generative modeling: 
standard DGMs such as GANs or VAEs learn a global pushforward map, 
whereas diffusion models learn a time-dependent local vector field that drives the generative dynamics.

\newpage
\section{Relationship of Variant Optimal Transport Formulations}\label{sec:relation-ot}

\begin{figure}[ht!]
\centering
\resizebox{\textwidth}{!}{
\begin{tikzpicture}[
    node distance=2.2cm and 1.2cm,  
    block/.style={
        rectangle, draw, thick, rounded corners,
        fill=white!10, text width=13em,   
        minimum height=5em, align=center, font=\normalsize 
    },
    arrow/.style = {ultra thick, -{Stealth[length=4mm, width=4mm]}},
    dashed_arrow/.style={-Latex, ultra thick, dashed, color=gray, -{Stealth[length=4mm, width=4mm]}},
    label/.style={midway, font=\small, align=center}
]

\node[block, text width=11em] (sde) {
    {\large \textbfs{SB}$_\varepsilon$ \\ \textbfs{(Stochastic Control)}}\\[0.6em] 
See \Cref{eq:sb-epsilon-kinetic}.
};

\node[block, text width=11em, below=of sde] (ode) {
   {\large \textbfs{OT} \\ \textbfs{(Dynamic Formulation)}
   \\ [0.5em]
See \Cref{eq:bb-ot}.
   }
};

\node[block, text width=11em, right=of sde, xshift=1.5cm] (sb) {
    {\large \textbfs{SB}$_\varepsilon$ \\ \textbfs{(Path-Space Formulation)}}\\[0.5em]
$\displaystyle
\min_{\substack{P :\, P_0 = p_{\rm src} \\ \phantom{P :\, } P_T = p_{\rm tgt}}}
  \mathcal{D}_\mathrm{KL}(P \| R^\varepsilon)
$
};

\node[block, right=of sb, xshift=1.5cm] (eot) {
    {\large \textbfs{EOT}$_\varepsilon$ }\\[0.5em]
    $\displaystyle
        \min_{\gamma \in \Gamma(p_{\rm src}, p_{\rm tgt})}
        \int \norm{\rvx-\rvy}_2^2  \diff\gamma
         + \varepsilon  \mathcal{D}_{\mathrm{KL}}(\gamma \| p_{\rm src} \otimes p_{\rm tgt})
    $

};

\node[block, below=of eot] (ot) {
 {\large \textbfs{OT \\ (Static Formulation)} }\\[0.6em]
    $\displaystyle
        \min_{\gamma \in \Gamma(p_{\rm src}, p_{\rm tgt})}
        \int \norm{\rvx-\rvy}_2^2 \diff\gamma
    $
};

\draw[arrow] (sde) -- (ode)
    node[label, left]
    {$\varepsilon \to 0$}
    node[label, right]
    {(iv)};
\draw[arrow,<->] (sde.east) -- (sb.west)
  node[pos=.5, above, text width=9em, align=center, sloped]
    {\footnotesize$p_t = P^*(\rvx_t \in \cdot)$}
  node[pos=.5, below, text width=9em, align=center, sloped]{(i)};
\draw[arrow,<->] (sb.east) -- (eot.west)
  node[pos=.5, above, text width=8em, align=center, sloped]
    {Endpoint projection\\ \footnotesize$\gamma^* = (\rvx_0,\rvx_T)_\# P^*$}
  node[pos=.5, below, text width=9em, align=center, sloped]
    {(ii)};
\draw[arrow] (eot) -- (ot)
    node[label, left]
    {$\varepsilon \to 0$}
    node[label, right]
    {(iii)};
\draw[arrow, <->] (ode) -- ([yshift=0.3cm] ot.west)
    node[label, below]
    {$p_t = \bPsi_t\#\gamma^*$ where $\bPsi_t(\rvx,\rvy) = (1-t) \rvx + t \rvy$};
    
\end{tikzpicture}
}
\caption{\textbfs{Relationship between variants of optimal transport with $c(\rvx,\rvy)=\|\rvx-\rvy\|_2^2$ and Reference $R^\varepsilon$ in SB.}
We summarize the relationships: (i) the stochastic-control and path-space
KL formulations of $\mathrm{SB}_\varepsilon$ are equivalent; (ii) for the
Brownian reference with the corresponding quadratic-cost normalization,
the endpoint projection of $\mathrm{SB}_\varepsilon$ yields
$\mathrm{EOT}_\varepsilon$; (iii) $\mathrm{EOT}_\varepsilon$ converges to
static OT as $\varepsilon\to0$; and (iv) the appropriately rescaled
Brownian-reference Schr\"odinger problem converges to dynamic OT as
$\varepsilon\to0$. \figcredit{Created by the authors.}} 
\label{fig:ot-dm}
\end{figure}

Before delving into the technical details, it is helpful to clarify how the different formulations of optimal transport and its entropic regularizations are connected. At a high level, these problems can be viewed as related (see \Cref{fig:ot-dm} for their diagram for connection):
\begin{enumerate}
    \item[(i)] \textbfs{Path-Space SB $\Longleftrightarrow$ Stochastic-Control SB.}
    The path-space KL formulation of $\mathrm{SB}_\varepsilon$ and its
    stochastic-control formulation describe the same bridge problem relative
    to the chosen reference process (see \Cref{subsec:sb}).

    \item[(ii)] \textbfs{Brownian-Reference SB $\Longleftrightarrow$ EOT$_\varepsilon$.}
    For the Brownian reference
    \[
    \diff \rvx_t=\sqrt{\varepsilon}\,\diff\rvw_t,
    \]
    with the corresponding quadratic-cost normalization, the endpoint coupling
    induced by the optimal SB path law solves the associated
    $\mathrm{EOT}_\varepsilon$ problem (see \Cref{subsec:sb-eot}).

    \item[(iii)] \textbfs{EOT$_\varepsilon$ $\longrightarrow$ Static OT.}
    As $\varepsilon\to0$, the entropy-regularized endpoint coupling approaches
    the corresponding classical OT problem, and the optimal values converge to
    the OT value under the assumptions discussed in
    \Cref{subsec:eot-ot}.

    \item[(iv)] \textbfs{SB$_\varepsilon$ $\longrightarrow$ Dynamic OT.}
    Under the corresponding small-noise scaling, the Brownian-reference
    Schr\"odinger problem converges as $\varepsilon\to0$ to the
    Benamou--Brenier dynamic formulation of OT
    (see \Cref{subsec:sb-ot}).
\end{enumerate}

Together, these non-trivial relationships provide a compact view across stochastic control, entropy-regularized, and classical OT frameworks.

\begin{figure}[th!]
    \centering
    \includegraphics[width=\linewidth]{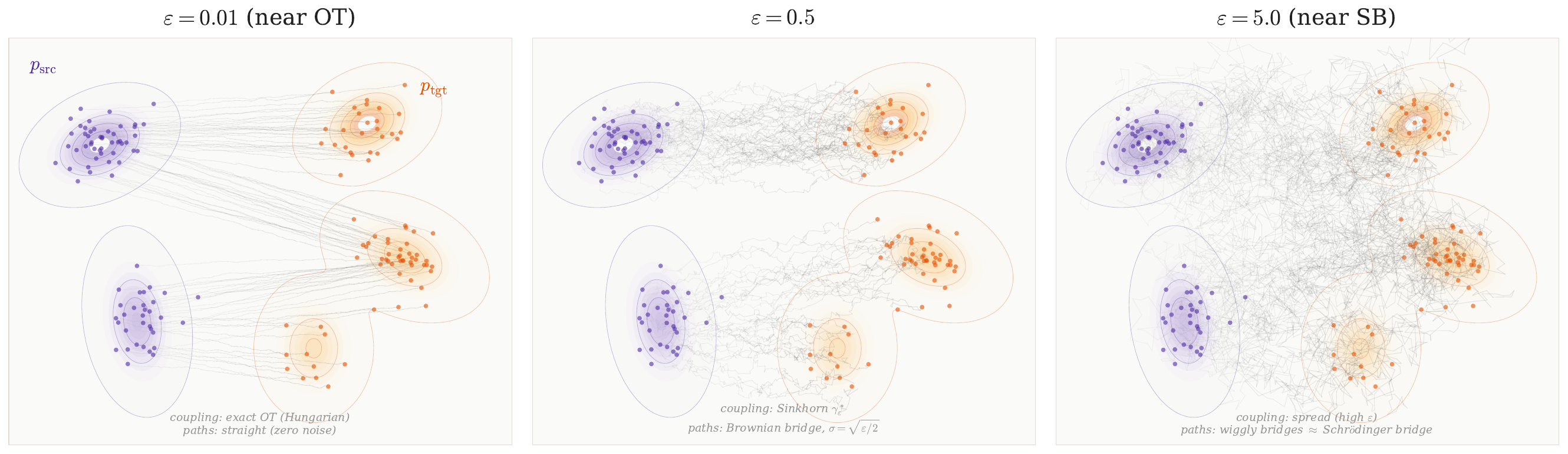}
    \caption{\textbfs{Effect of the regularization parameter $\varepsilon$ on
    entropic optimal transport (connections (iii) and (iv)).}
    As $\varepsilon$ increases, two things change simultaneously:
    the coupling $\gamma^*_\varepsilon$ spreads mass to more distant
    targets, and the transport paths become increasingly stochastic.
    Left ($\varepsilon=0.01$): paths are nearly straight and each
    source point maps to a single nearby target, recovering classical OT.
    Center ($\varepsilon=0.5$): paths acquire visible fluctuations
    and the coupling begins to spread.
    Right ($\varepsilon=2.5$): paths are highly stochastic,
    approaching the Schr\"odinger bridge regime;
    in the limit $\varepsilon\to\infty$, the coupling degenerates to
    the independent product
    $p_{\mathrm{src}}\otimes p_{\mathrm{tgt}}$.
    The couplings are computed via the Sinkhorn algorithm
    (Hungarian algorithm for $\varepsilon\to 0$);
    conditional on endpoints, the paths are Brownian bridges
    with diffusion coefficient $\sigma=\sqrt{\varepsilon/2}$
    from the Mikami duality (\Cref{subsec:sb-eot}).
    \figcredit{Created by the authors with AI-assisted coding.}}
    \label{fig:eot-epsilon}
\end{figure}

\subsection{SB and EOT are (Dual) Equivalent}\label{subsec:sb-eot}
In this section, we present two complementary perspectives showing that SB are essentially equivalent to EOT.  
Unlike classical optimal transport, which produces a single deterministic map, SB yields a \emph{stochastic} flow of particles: mass is transported probabilistically, with marginals evolving under diffusion-like dynamics.

From the static viewpoint, SB coincides with EOT, where the goal is to find a coupling between the two endpoint distributions that balances transport cost with entropy.  
From the dynamic viewpoint, SB describes a controlled diffusion process that remains as close as possible to a simple reference (such as Brownian motion) while still matching the desired endpoints.   Each perspective independently establishes the equivalence, providing two consistent ways to understand SB/EOT as canonical formulations of distribution-to-distribution transformation.

\paragraph{Static Schr\"odinger Bridge.}
Let
\[
  \widetilde R^\varepsilon(\rvx,\rvy)
  := \frac{1}{Z_\varepsilon}\,e^{-c(\rvx,\rvy)/\varepsilon}\,p_{\mathrm{src}}(\rvx)\,p_{\mathrm{tgt}}(\rvy),
\]
with normalizing constant
\[
Z_\varepsilon := \iint e^{-c(\rvx, \rvy)/\varepsilon}\, p_{\mathrm{src}}(\rvx)\, p_{\mathrm{tgt}}(\rvy)\,\mathrm d\rvx\,\mathrm d\rvy.
\]
Then, for any admissible coupling \(\gamma\in\Gamma(p_{\mathrm{src}},p_{\mathrm{tgt}})\), a direct computation gives
\[
\int c\, \mathrm d\gamma
+\varepsilon \mathcal{D}_\mathrm{KL}\!\bigl(\gamma \Vert p_{\mathrm{src}}\!\otimes\!p_{\mathrm{tgt}}\bigr)
=
\varepsilon \mathcal{D}_\mathrm{KL}\!\bigl(\gamma \Vert \widetilde R^\varepsilon\bigr)
-\varepsilon\log Z_\varepsilon.
\]
Therefore,
\begin{align}
\begin{aligned}\label{eq:eot-sb-static}
    \min_{\gamma\in\Gamma(p_{\mathrm{src}},p_{\mathrm{tgt}})}
  \Bigl\{\int c \,\mathrm d\gamma
  + \varepsilon \mathcal{D}_\mathrm{KL}\!\bigl(\gamma \Vert p_{\mathrm{src}}\!\otimes\!p_{\mathrm{tgt}}\bigr)\Bigr\}
  = \varepsilon \min_{\gamma\in\Gamma(p_{\mathrm{src}},p_{\mathrm{tgt}})}
  &\mathcal{D}_\mathrm{KL}\!\bigl(\gamma \Vert \widetilde R^\varepsilon\bigr)
  \\&-\varepsilon\log Z_\varepsilon,
\end{aligned}
\end{align}
so the entropic OT problem is equivalent, up to an additive constant, to the static Schr\"odinger Bridge problem
\[
\min_{\gamma\in\Gamma(p_{\mathrm{src}},p_{\mathrm{tgt}})}
\mathcal{D}_\mathrm{KL}(\gamma\Vert \widetilde R^\varepsilon).
\]

\paragraph{Dynamic Equivalence (Brownian Reference).} We can also understand this equivalence from a dynamical viewpoint. 
A classical result \citep{mikami2006duality} states  that entropic OT with quadratic cost
\[
c(\rvx,\rvy)=\frac{\|\rvy-\rvx\|^2}{2T}
\]
is affinely equivalent to the SB problem where the
reference path law $R^\varepsilon$ is Brownian motion on $[0,T]$,
\[
\mathrm{d}\rvx_t=\sqrt{\varepsilon}\,\mathrm{d}\rvw_t.
\]
Here, ``affinely equivalent'' means the optimal values differ by a positive scaling and an
additive constant (independent of the decision variable), so the minimizers coincide.
In particular, let $P^*$ be the optimal path distribution for SB. The key
connection to EOT is obtained by looking only at the two endpoints of this
optimal stochastic process. If
$\rvx_{[0:T]}\sim P^*$, then the endpoint pair
$(\rvx_0,\rvx_T)$ induces the coupling
\[
\gamma^*:=(\rvx_0,\rvx_T)_\#P^*,
\]
and $\gamma^*$ is an optimal transport plan for the corresponding EOT
problem.

Conversely, starting from an optimal EOT coupling $\gamma^*$, we may first
sample a pair of endpoints
$(\rvx_0,\rvx_T)\sim\gamma^*$ and then, conditioned on these endpoints,
sample a trajectory from the reference diffusion conditioned to connect
$\rvx_0$ to $\rvx_T$. Mixing these conditional diffusion paths according
to $\gamma^*$ recovers the optimal SB path distribution $P^*$.

In words, the optimal stochastic process from the dynamic SB problem
induces the optimal endpoint coupling of the static EOT problem. Conversely,
the optimal EOT coupling specifies how the endpoints should be paired,
while the reference diffusion specifies the optimal stochastic path between
each paired set of endpoints.

The key idea to derive this fact is that the KL divergence over paths can be broken down according to the endpoints, which means the Schr\"odinger bridge problem reduces to a KL divergence just over the joint distribution of $(\rvx_0,\rvx_T)$. For Brownian motion the transition density between $\rvx$ and $\rvy$ has a Gaussian form, so its negative log is quadratic:
\[
-\varepsilon\log p_T(\rvy\mid\rvx)=\frac{\|\rvy-\rvx\|^2}{2T}+\text{const}.
\]
This shows that the endpoint KL is exactly the same as the entropic OT objective with quadratic cost, up to an irrelevant constant.

\paragraph{SB with General Reference Determines the Endpoint EOT Cost.}
The connection between SB and EOT is not restricted to Brownian motion.
For a general well-posed reference process, its transition law tells us how
``natural'' it is under the reference dynamics to move from $\rvx$ at time
$0$ to $\rvy$ at time $T$. Endpoint pairs that are unlikely under the
reference should therefore carry a larger transport cost.

More precisely, let $p_T(\rvy|\rvx)$ denote the transition density of the
reference process over $[0,T]$. Up to a scaling and additive terms that
depend only on the fixed endpoint marginals, the corresponding endpoint
cost takes the form
\[
c(\rvx,\rvy)
\propto
-\log p_T(\rvy|\rvx).
\]
Thus, at the level of endpoint couplings, the reference dynamics induce the
cost used by the corresponding EOT problem: likely transitions receive
smaller cost, while unlikely transitions receive larger cost.

The full reference process contains additional information beyond this
endpoint cost. Once an endpoint pair $(\rvx,\rvy)$ is selected, the
reference dynamics also determine the conditional stochastic paths used to
connect those endpoints. In this sense, EOT determines how the endpoints
are optimally paired, while SB additionally describes the stochastic
transport between each paired set of endpoints.

\subsection{EOT$_\varepsilon$ is Reduced to OT as $\varepsilon\to 0$}\label{subsec:eot-ot}

Let $\gamma^*_\varepsilon$ denote the optimal plan for
$\mathrm{EOT}_\varepsilon$, and let $\gamma^*$ denote an optimal plan of
the unregularized OT problem in \Cref{eq:ot}. Under standard assumptions,
as $\varepsilon\to0$ the EOT optimal values converge to the OT value, and
any limiting entropic optimal plan is OT-optimal; when the OT optimizer is
unique, the optimal plans converge to $\gamma^*$.

This convergence result is both fundamental and practically important. One of the reasons is that the entropy-regularized OT problem $\mathrm{EOT}_\varepsilon$ admits efficient numerical solutions via algorithms such as Sinkhorn. Thus, the result provides theoretical justification for using $\mathrm{EOT}_\varepsilon$ with small $\varepsilon$ as a computationally tractable proxy for the classical OT problem in \Cref{eq:ot}, even when the cost function $c(\rvx, \rvy)$ is more general than the quadratic case.

\thmp{(Informal) EOT$_\varepsilon$ Converges to OT.}{eot-ot}{
Under standard assumptions ensuring the entropic OT limit, as
$\varepsilon \to 0$, the optimal values converge:
\[
\lim_{\varepsilon \to 0}
\mathrm{EOT}_\varepsilon(p_{\mathrm{src}}, p_{\mathrm{tgt}})
=
\mathrm{OT}(p_{\mathrm{src}}, p_{\mathrm{tgt}}).
\]
Moreover, any limiting optimal plan is an optimal coupling of the
unregularized OT problem. In particular, when the OT optimizer is unique,
the entropic optimal plans converge to that optimizer.
}
{For a rigorous proof, we refer to the literature~\citep{mikami2008optimal, peyre2019computational}. Below we provide a heuristic derivation of the value convergence.

Let us denote the corresponding optimal values by
\[
V_\varepsilon := \mathrm{EOT}_\varepsilon(p_{\mathrm{src}}, p_{\mathrm{tgt}}), \quad
V_0 := \mathrm{OT}(p_{\mathrm{src}}, p_{\mathrm{tgt}})
\]
for notational simplicity. 

The elementary upper-bound argument below applies when an OT-optimal
coupling has finite relative entropy with respect to
$p_{\mathrm{src}}\otimes p_{\mathrm{tgt}}$. In general, the rigorous
value convergence can be established by approximating optimal couplings
with finite-entropy couplings, or more systematically through
$\Gamma$-convergence arguments.

\proofparagraph{Upper Bound.} By optimality of $\gamma^*_\varepsilon$, its value $V_\varepsilon$ is bounded by the cost of using the plan $\gamma^*$:
\[
    V_\varepsilon \leq \int c  \diff\gamma^* + \varepsilon \mathcal{D}_{\text{KL}}(\gamma^* \| p_{\mathrm{src}} \otimes p_{\mathrm{tgt}}).
\]
Assuming the KL term is a finite constant $K$, we get $V_\varepsilon \leq V_0 + \varepsilon K$. Taking the limit superior yields $\limsup_{\varepsilon \to 0} V_\varepsilon \leq V_0$.

\proofparagraph{Lower Bound.} Since the KL-divergence is non-negative, $V_\varepsilon \geq \int c  \diff\gamma^*_\varepsilon$. By definition of $V_0$ as the minimal transport cost, any plan's cost is at least $V_0$, so $\int c  \diff\gamma^*_\varepsilon \geq V_0$. This implies $V_\varepsilon \geq V_0$ for all $\varepsilon > 0$, and thus $\liminf_{\varepsilon \to 0} V_\varepsilon \geq V_0$.

Combining the upper and lower bounds shows the convergence of the optimal value, $\lim_{\varepsilon \to 0} V_\varepsilon = V_0$. The corresponding convergence of optimal plans requires additional
compactness arguments: any limiting plan is OT-optimal, and full
convergence follows when the OT optimizer is unique.

 }

\subsection{SB$_\varepsilon$ is Reduced to OT as $\varepsilon\to 0$}\label{subsec:sb-ot} 
For direct comparison with \Cref{eq:bb-ot}, we rescale time so that
$T=1$ throughout this subsection. For each $\varepsilon > 0$, let $\mathbf{v}_t^\varepsilon$ be a minimizer of the SB problem as in \Cref{eq:sb-epsilon-kinetic}, and let $p_t^\varepsilon$ be the marginal distribution of the controlled SDE $\mathbf{x}_t$ induced by $\mathbf{v}_t^\varepsilon$. Then $p_t^\varepsilon$ satisfies the associated Fokker–Planck equation. In contrast, denote by $(p_t^0, \mathbf{v}_t^0)$ a minimizer of the Benamou–Brenier formulation of optimal transport (see \Cref{eq:bb-ot}).

The following theorem\footnote{A rigorous formulation uses $\Gamma$-convergence of the
appropriately rescaled Schr\"odinger functionals. Together with suitable
compactness assumptions, this yields convergence of optimal values and
corresponding limit points of minimizers. Since this requires more technical background, we omit the details here and state only the conceptual result.} states that as $\varepsilon \to 0$, the SB problem converges to the OT problem. This result is practically important for reasons similar to those in Theorem~\ref{thm:eot-ot}. The objective $ \mathrm{SB}_\varepsilon $ can be efficiently solved using Sinkhorn type algorithms, yielding a numerically tractable and differentiable proxy for optimal transport. This is especially valuable in high dimensional or large scale settings, where direct solvers (e.g., based on the Benamou–Brenier formulation) become computationally expensive.

\thmp{(Informal) SB$_\varepsilon$ Converges to OT.}{sb-ot}{ As $\varepsilon \to 0$, we have:
\[
\lim_{\varepsilon \to 0}
2\varepsilon\,\mathrm{SB}_\varepsilon
(p_{\mathrm{src}},p_{\mathrm{tgt}})
=
\mathcal{W}_2^2(p_{\mathrm{src}},p_{\mathrm{tgt}}),
\]
where OT is of the Benamou–Brenier formulation as in \Cref{eq:bb-ot}. Moreover, under suitable compactness and regularity assumptions, the
marginals $p_t^\varepsilon$ and the corresponding velocity fields
$\mathbf{v}_t^\varepsilon$ converge, in the appropriate weak sense, along
subsequences to a limiting pair $(p_t^0,\mathbf{v}_t^0)$ that solves the
Benamou--Brenier problem. In particular, when the Benamou--Brenier
minimizer is unique, the optimal Schr\"odinger dynamics converge to this
OT solution in the corresponding limiting sense.
}{A full rigorous proof of the convergence result is beyond our scope; we refer the reader to \citet{leonard2012schrodinger,leonard2014survey} for detailed derivations. Nevertheless, we can heuristically understand why this convergence may hold.

In the stochastic control formulation of the SB problem \Cref{eq:sb-epsilon-kinetic}, the controlled SDE is given by:
\[
\diff \rvx_t = \mathbf{v}_t^\varepsilon(\rvx_t)\,\diff t + \sqrt{\varepsilon}\,\diff \mathbf{w}_t.
\]
As $\varepsilon \to 0$, the noise term vanishes, and the SDE formally approaches a deterministic ODE:
\[
\diff \rvx_t = \mathbf{v}_t^0(\rvx_t)\,\diff t.
\]
Together with \Cref{eq:sb-epsilon-kinetic}, this suggests that the
appropriately rescaled SB objective converges to the Benamou--Brenier
action:
\[
\lim_{\varepsilon \to 0}
2\varepsilon\,\mathrm{SB}_\varepsilon
(p_{\mathrm{src}},p_{\mathrm{tgt}})
=
\mathcal{W}_2^2(p_{\mathrm{src}},p_{\mathrm{tgt}}).
\]

In parallel, the marginal density $p_t^\varepsilon$ satisfies the Fokker-Planck equation:
\[
\partial_t p_t^\varepsilon + \nabla \cdot \left(p_t^\varepsilon \mathbf{v}_t^\varepsilon\right) = \frac{\varepsilon}{2}\,\Delta p_t^\varepsilon.
\]
Again, as $\varepsilon \to 0$, the diffusion term vanishes, and the equation formally reduces to the continuity equation:
\[
\partial_t p_t^0 + \nabla \cdot \left(p_t^0 \mathbf{v}_t^0\right) = 0.
\]
}

Until now, we have presented the fundamental equivalences (under their respective assumptions) between EOT and SB, as well as their important connection to OT through a limiting process, illustrated in \Cref{fig:ot-dm}. Next, we will explore how diffusion models connect to these concepts.

\newpage

\section{Is Diffusion Model's SDE Optimal Solution to SB Problem?}\label{sec:dm-and-sb}

\subsection{Diffusion models as a Special Case of Schrödinger Bridges}

SB framework extends (score-based) diffusion models by enabling nonlinear interpolation between arbitrary source and target distributions. It achieves this by adding control drift terms derived from scalar potentials $\Psi(\mathbf{x}, t)$ and $\widehat{\Psi}(\mathbf{x}, t)$, which guide a reference diffusion process to match prescribed endpoint marginals (see \Cref{eq:pde-schrodinger}) and follow the decomposition:
\[
\nabla\log{\Psi(x,t)}+\nabla\log{\hat{\Psi}(x,t)}=\nabla \log p_t^{\mathrm{SB}}(\mathbf{x}).
\]
This generalization allows the model to move beyond standard Gaussian priors and generate samples from broader distributions.

\paragraph{Connection to Diffusion Models.}

Diffusion models arise as a special case of the SB framework. Suppose the potential is constant, $\Psi(\mathbf{x}, t) \equiv 1$. Under this assumption, the second PDE in \Cref{eq:pde-schrodinger} reduces to the standard Fokker–Planck equation, whose solution is the marginal density of the reference process:
\begin{align}\label{eq:sb-backward-potential}
    \widehat{\Psi}(\mathbf{x}, t) = p_t^{\mathrm{SB}}(\mathbf{x}).
\end{align}
The corresponding SB forward SDE thus becomes the uncontrolled reference process:
\[
    \diff\mathbf{x}_t = \mathbf{f}(\mathbf{x}_t, t)\diff t + g(t)\diff\mathbf{w}_t,
\]
and the SB backward SDE simplifies to:
\[
    \diff\mathbf{x}_t = \left[\mathbf{f}(\mathbf{x}_t, t) - g^2(t) \nabla \log p_t^{\mathrm{SB}}(\mathbf{x}_t) \right]\diff t + g(t)\diff\bar{\mathbf{w}}_t,
\]
which matches Anderson’s reverse-time SDE used in diffusion models. This correspondence shows that diffusion models can be interpreted as the zero-control limit of SB, where no additional drift is introduced by the potentials.

\paragraph{Boundary Conditions and Generality.}
The above reduction is purely formal unless the boundary constraints are compatible. For arbitrary
source/target $(p_{\mathrm{src}},p_{\mathrm{tgt}})$, the PDE boundary conditions in
\Cref{eq:pde-schrodinger} are generally not satisfied by the choice $\Psi\equiv 1$.
Full SB resolves this by learning nontrivial potentials that induce a nonlinear control
drift, bending the reference dynamics to match any prescribed endpoints. By contrast,
diffusion models fix one endpoint to a simple prior (typically Gaussian) and learn only the
reverse-time score to reach the data. With this perspective, SB is the more flexible umbrella: with nontrivial
potentials it bridges arbitrary endpoints; with $\Psi\equiv 1$ it collapses to the diffusion-model
case above. We additionally remark that in the standard linear diffusion model, $p_T \approx p_{\mathrm{prior}}$ holds only as $T \to \infty$, so the match to the prior is merely approximate.

\subsection{Diffusion Models as Schrödinger Half-Bridges}

In this section, we explain why diffusion models are not full Schr\"odinger bridges, 
but can instead be understood through the relaxed notion of \emph{Schr\"odinger half-bridges}. 
A half-bridge enforces only one endpoint constraint (either $p_{\mathrm{prior}}$ or $p_{\mathrm{data}}$) rather than both, 
making it a one-sided variant of the full bridge. 
Before formalizing this connection, we introduce the definition of Schr\"odinger half-bridges, 
building on the general formulation in \Cref{eq:sb-epsilon-general} with arbitrary $p_{\mathrm{src}}$ and $p_{\mathrm{tgt}}$. 
We will then return to diffusion models and show how the half-bridge viewpoint naturally applies when the 
endpoints are given by $p_{\mathrm{prior}}$ and $p_{\mathrm{data}}$.

\paragraph{Schrödinger Half-Bridges}

The SB problem asks for a stochastic process whose law is
closest (in KL divergence) to a simple reference process, while \emph{matching
two endpoint distributions} $p_{\mathrm{src}}$ and $p_{\mathrm{tgt}}$.
Solving the full bridge requires enforcing both boundary conditions,
which is often computationally difficult. A useful relaxation is the
\emph{half-bridge} problem: instead of matching both endpoints, we match
only one of them. 

Formally, let $R$ be the reference path distribution. 
The \emph{forward half-bridge} seeks a path distribution $P$ minimizing
\[
    \min_{P: P_0 = p_{\mathrm{src}}}
    \mathcal{D}_{\mathrm{KL}}(P \,\|\, R),
\]
subject to the single constraint $P_0 = p_{\mathrm{src}}$. 
Similarly, the \emph{backward half-bridge} constrains only the terminal
distribution,
\[
    \min_{P: P_T = p_{\mathrm{tgt}}}
    \mathcal{D}_{\mathrm{KL}}(P \,\|\, R).
\]
In words, the forward half-bridge asks: among all processes starting
from the desired initial distribution, which one looks most like the
reference? The backward half-bridge asks the same question for processes
ending at the desired terminal distribution. By combining these two
relaxations iteratively, one can approximate the full SB.

\paragraph{Diffusion Models Miss Exact Endpoint Matching.}
A key difference between diffusion models and the SB framework lies in the treatment of the terminal distribution $p_T$. In standard diffusion models, the forward SDE is typically linear in $\rvx_t$ (see \Cref{eq:forward-linear-sde}) and designed so that $p_T$ approximates the  prior only as $T \to \infty$:
\[
    p_T \approx p_{\mathrm{prior}}.
\]
At finite time, however, $p_T$ is the distribution of
$\alpha_T\rvx_0+\sigma_T\beps$ and therefore generally still depends on
$p_{\mathrm{data}}$; it need not coincide exactly with the prescribed
Gaussian prior (see \Cref{subsec:rigorous-proof-linear-sde}).

In contrast, the SB framework enforces exact marginal matching at a finite time $T$ by introducing an additional control drift of the form $g^2(t)\nabla_\mathbf{x}\log \Psi(\mathbf{x}, t)$. This ensures that the terminal distribution precisely satisfies $p_T = p_{\mathrm{prior}}$, regardless of the initial data distribution $p_0 = p_{\mathrm{data}}$. In summary:
\begin{itemize}
    \item \textbfs{Diffusion Models:} $p_T \approx p_{\mathrm{prior}}$, asymptotically as $T \to \infty$,
    \item \textbfs{Schrödinger Bridge:} $p_T = p_{\mathrm{prior}}$ exactly at finite $T$, enabled by solving for the control potentials $\Psi$ and $\widehat{\Psi}$.
\end{itemize}

\paragraph{Diffusion Schrödinger Bridge.} 
The fixed forward noising process can be viewed as a forward
Schr\"odinger half-bridge anchored at $p_{\mathrm{data}}$ relative to the
chosen reference dynamics. Its terminal marginal is the distribution
produced by that reference process and is generally only approximately
$p_{\mathrm{prior}}$; hence it is not the full bridge between the
prescribed pair $(p_{\mathrm{data}},p_{\mathrm{prior}})$.

To address this, the Diffusion Schrödinger Bridge (DSB)~\citep{de2021diffusion} alternates between matching both endpoint marginals by following the idea of the Iterative Proportional Fitting (IPF) algorithm, an alternating projection method. This extends diffusion models to solve the full SB problem as follows\footnote{Although this description uses $p_{\mathrm{data}}$ and $p_{\mathrm{prior}}$, the DSB framework applies to any pair of endpoint distributions.}:

\begin{itemize}
    \item \textbfs{Step 0: Reference Process.} Initialize with $P^{(0)} := R_{\text{fwd}}$, the reference forward SDE:
    \[
        \mathrm{d} \mathbf{x}_t = \mathbf{f}(\mathbf{x}_t, t)   \mathrm{d} t + g(t)   \mathrm{d} \mathbf{w}_t, \quad \mathbf{x}_0 \sim p_{\text{data}}.
    \]
    This ensures $P_0^{(0)} = p_{\text{data}}$, but typically $P_T^{(0)} \neq p_{\text{prior}}$.

    \item \textbfs{Step 1: Backward Pass.} Compute the process $P^{(1)}$ that matches $p_{\text{prior}}$ at time $T$ while staying close to $P^{(0)}$:
    \[
        P^{(1)} = \argmin_{P : P_T = p_{\text{prior}}} \mathcal{D}_{\text{KL}}(P  \|  P^{(0)}).
    \]
    This is achieved via approximating the oracle score function  with a neural network $\mathbf{s}_{\bm{\phi}^\times}$, which results in the reverse-time SDE:
    \[
        \mathrm{d} \mathbf{x}_t = \left[ \mathbf{f}(\mathbf{x}_t, t) - g^2(t) \mathbf{s}_{\bm{\phi}^\times}(\mathbf{x}_t, t) \right] \mathrm{d} t + g(t)  \mathrm{d} \bar{\mathbf{w}}_t,
    \]
    simulated backward from $\mathbf{x}_T \sim p_{\text{prior}}$.
    
    \item \textbfs{Iteration.} The process $P^{(1)}$ satisfies $P_T^{(1)} = p_{\text{prior}}$, but its initial marginal $P_0^{(1)}$ typically deviates from $p_{\text{data}}$. IPF addresses this by learning a forward SDE to adjust $P_0^{(1)}$ back to $p_{\text{data}}$, followed by another backward pass to enforce $p_{\text{prior}}$. This alternation continues, refining the process until convergence to the optimal bridge $P^*$, which satisfies both $P_0^* = p_{\text{data}}$ and $P_T^* = p_{\text{prior}}$. \citet{de2021diffusion} prove convergence under mild conditions.
\end{itemize}

The DSB formulation above is historically important and conceptually clear, but its practical implementation can be computationally demanding. The forward--backward projections are carried out iteratively, so that each IPF iteration requires solving a separate score-matching problem, while approximation errors may also accumulate across rounds. More broadly, a growing body of work seeks to alleviate these difficulties by developing Schr\"odinger bridge formulations with improved stability and scalability, and in some cases with reduced dependence on repeatedly fitting separate diffusion models throughout the iterative procedure. We do not discuss these developments in detail here, but mention them to orient interested readers to a broader literature on practical SB methods~\citep{shi2023dsbm,liu2024gsbm}.

\newpage
\section{Is Diffusion Model's ODE an Optimal Map to OT Problem?}\label{sec:pfode-ot}
In this section, we focus on quadratic-cost optimal transport problem.

\subsection{PF-ODE Flow Is Generally Not Optimal Transport}
A natural question is whether the deterministic flow map induced by a
diffusion process coincides with the optimal transport map under quadratic
cost. The answer is no in general, and explicit counterexamples are already
available in Gaussian settings. In particular,
\citet{tanana2017comparison} showed that, for a diffusion flow with a more
general Gaussian reference, the resulting deterministic transport map can
differ from the Brenier optimal transport map even when the endpoint
distributions are Gaussian. Since both maps are explicit in this setting,
their difference can be verified directly.

Below we present the stronger construction of
\citet{lavenant2022flow} in the standard-Gaussian
Ornstein--Uhlenbeck setting that is directly relevant to diffusion models.
Their argument establishes non-optimality for a broader class of initial
distributions and provides structural insight into why the corresponding
PF-ODE flow map need not be an optimal transport map.

\begin{figure}[th!]
    \centering
    \includegraphics[width=\linewidth]{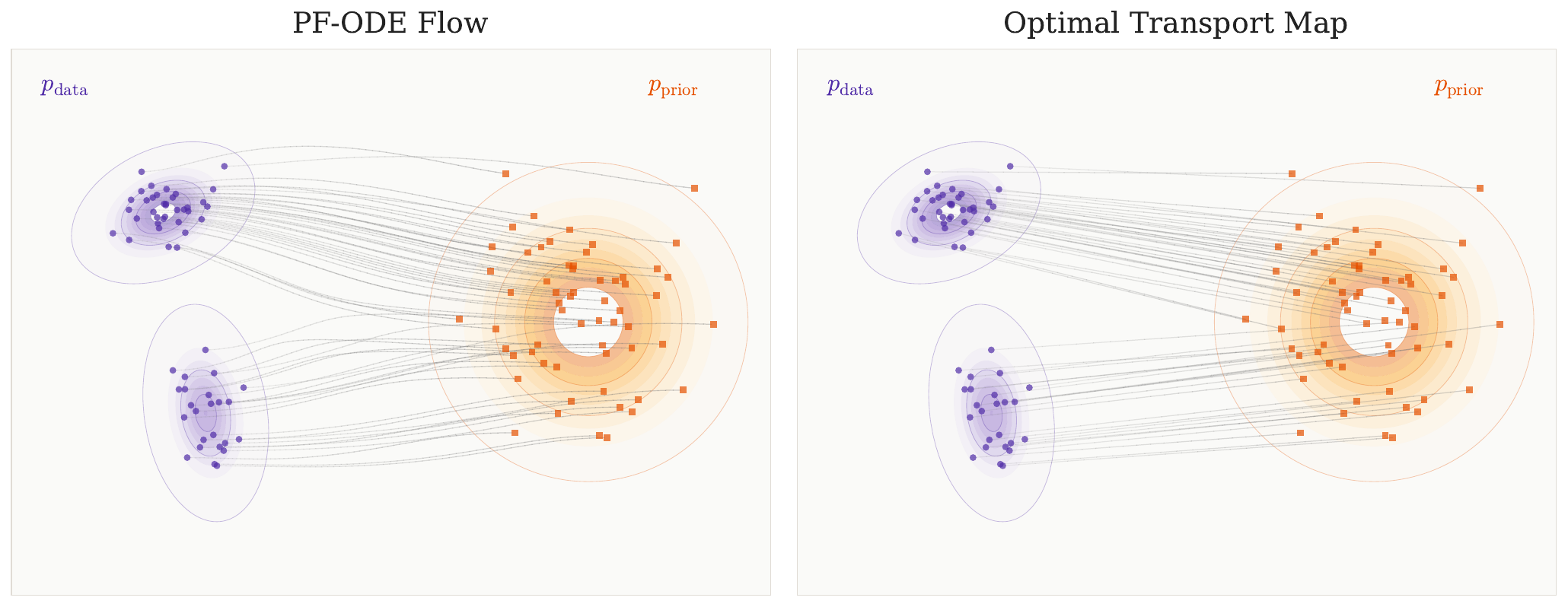}
    \caption{Comparing the PF-ODE flow map with the true OT map
    from $p_{\mathrm{data}}$ to $p_{\mathrm{prior}}$ in 2D.
    Right: the quadratic-cost OT map, computed exactly via the
    Hungarian algorithm.
    Left: the PF-ODE flow map obtained by integrating the
    PF-ODE. The two maps visibly differ,
    illustrating that the PF-ODE does not generally recover
    the optimal transport map.
    \figcredit{Created by the authors with AI-assisted coding.}}
    \label{fig:pfode-vs-ot-2d}
\end{figure}

\paragraph{Setup.} We consider a VP SDE, specifically the Ornstein–Uhlenbeck process, which evolves a smooth initial density $p_0$ toward the standard Gaussian $\mathcal{N}(\bm{0}, \rmI)$:
\[
\diff \rvx(t) = -\rvx(t) \diff t + \sqrt{2} \diff \rvw(t), \quad \rvx(0) \sim p_0.
\]
The associated PF-ODE is given by
\[
\frac{\diff \mathbf{S}_t(\rvx)}{\diff t} = -\mathbf{S}_t(\rvx) - \nabla \log p_t(\mathbf{S}_t(\rvx)), \quad \mathbf{S}_0(\rvx) = \rvx.
\]
Here, $\mathbf{S}_t$ denotes the flow map pushing forward $p_0$ to the marginal $p_t$:
\[
\left(\mathbf{S}_t\right) \# p_0 = p_t,
\quad\text{that is,}\quad
p_t(\rvy) = \int_{\mathbb{R}^D} \delta(\rvy - \mathbf{S}_t(\rvx))   p_0(\rvx) \diff \rvx.
\]
These densities $p_t$ evolve via the Fokker-Planck equation:
\[
\frac{\partial p_t}{\partial t} = \nabla \cdot (\rvx p_t) + \Delta p_t.
\]
This is equivalent to a continuity equation with velocity field:
\[
\rvv_t(\rvx) = -\rvx - \nabla \log p_t(\rvx),
\]
whose flow is given by $\mathbf{S}_t(\rvx)$. In other words, the PF-ODE can be written as: 
\[
\frac{\diff \mathbf{S}_t(\rvx)}{\diff t} = \rvv_t\left(\mathbf{S}_t(\rvx)\right).
\]

As $t \to \infty$, the map transports the initial distribution to the prior:
\[
\mathbf{S}_\infty \# p_0 = \mathcal{N}(\bm{0}, \rmI) =: p_{\text{prior}}.
\]
\paragraph{Objective of \citet{lavenant2022flow}'s Argument.}

\citet{lavenant2022flow} do not directly assess whether the terminal map $\mathbf{S}_\infty$ from $p_0$ to the Gaussian is optimal. Instead, they construct a specific initial distribution $p_0$ and examine the entire PF-ODE trajectory. Their key observation is that optimality may fail at some point along the flow.

They consider the intermediate marginal $p_t = \mathbf{S}_t \# p_0$ and define the residual transport map from $p_{t_0}$ to the Gaussian as
\[
\mathbf{T}_{t \to \infty} := \mathbf{S}_\infty \circ \mathbf{S}_{t}^{-1}, \quad \text{for all } t\geq 0.
\]
The core of their argument shows that, for a carefully chosen $p_0$, there exists a time $t_0 \geq 0$ such that $\mathbf{T}_{t_0 \to \infty}$ is not the quadratic-cost optimal transport map  from the new starting distribution $p_{t_0}$ to $\mathcal{N}(\bm{0}, \rmI)$.

This result demonstrates that PF-ODE flows do not, in general, yield optimal transport maps, and that the property of optimality can break down for certain initial distributions.

\paragraph{Some Tools.} The argument of \citet{lavenant2022flow} crucially relies on the following result, known as \emph{Brenier's theorem}:
\begin{quote}
    \begin{theorem}[Informal Brenier's Theorem]\label{thm:brenier}
Let $\nu_1, \nu_2$ be two probability distributions on $\mathbb{R}^D$ with smooth densities. A smooth map $\rmT : \mathbb{R}^D \to \mathbb{R}^D$ is the optimal transport from $\nu_1$ to $\nu_2$ (under quadratic cost) if and only if $\rmT = \nabla u$ for some convex function $u$. In this case, $\mathrm{D} \rmT$ is symmetric and positive semi-definite, and $u$ satisfies the Monge–Ampère equation:
\[
\det \mathrm{D}^2 u(\mathbf{x}) = \frac{\nu_1(\mathbf{x})}{\nu_2(\nabla u(\mathbf{x}))}.
\]
\end{theorem}
\end{quote}

The proof also implicitly uses the following fact, which we will not repeat each time: \emph{a map is the optimal transport between two distributions if and only if its inverse is the optimal transport in the reverse direction}.

\paragraph{Proof Sketch: PF-ODE Is Not an OT Map in General.} \citet{lavenant2022flow} employ a proof by contradiction: they assume that for every $t\ge0$, the map
\[
\mathbf{T}_t  =  \mathbf{S}_t \circ \mathbf{S}_\infty^{-1}
\]
is the quadratic-cost optimal transport map from $\mathcal{N}(\bm{0},\rmI)$ to $p_t$.

\subparagraph{Step 1:  Brenier's Theorem.}  By Brenier’s Theorem, the Jacobian of any optimal transport map from Gaussian must be symmetric and positive semi-definite. Thus, 
\[
\mathrm{D} \mathbf{T}_t(\rvx) = \mathrm{D} \mathbf{S}_t(\mathbf{S}_\infty^{-1}(\rvx)) \mathrm{D} (\mathbf{S}_\infty^{-1})(\rvx)
\]
must be symmetric for all $t$ and $\rvx$. Here, $\mathrm{D} \mathbf{T}_t(\rvx)$ denotes the total differentiation with respect to $\rvx$.
\subparagraph{Step 2: Time-Differentiating the Symmetry Condition.}  
Differentiating in time:
\[
\frac{\partial}{\partial t} \mathrm{D}\mathbf{T}_t(\rvx)
= \left( \frac{\partial}{\partial t} \mathrm{D} \mathbf{S}_t \right)(\mathbf{S}_\infty^{-1}(\rvx)) \mathrm{D} (\mathbf{S}_\infty^{-1})(\rvx).
\]
Given that the symmetry holds for all $t$, it follows that this derivative remains symmetric.

Using the flow ODE (differentiating in $\rvx$), we obtain:
\[
\frac{\partial (\mathrm{D}\mathbf{S}_t)}{\partial t}
= \mathrm{D}\rvv_t(\mathbf{S}_t) \cdot \mathrm{D}\mathbf{S}_t
= \left( -\rmI - \mathrm{D}^2 \log p_t(\mathbf{S}_t) \right) \cdot \mathrm{D}\mathbf{S}_t.
\]
Combining the above, we see that
\[
\left( -\rmI - \mathrm{D}^2 \log p_t(\mathbf{S}_t) \right) \cdot\mathrm{D}\mathbf{S}_t \cdot \mathrm{D} (\mathbf{S}_\infty^{-1})
\]
is symmetric for all $t\geq 0$.

At $t = 0$, we have $\mathbf{S}_0 = \rmI$ and $\mathrm{D}\mathbf{S}_0 = \rmI$, yielding:
\[
\left( -\rmI - \mathrm{D}^2 \log p_0(\mathbf{S}_\infty^{-1}(\rvx)) \right) \cdot \mathrm{D} (\mathbf{S}_\infty^{-1})(\rvx) \quad \text{is symmetric}.
\]

\subparagraph{Step 3: The Commutation Condition.}  Since $\mathbf{T}_0 = \mathbf{S}_\infty^{-1}$ is assumed to be optimal, its Jacobian $D\mathbf{T}_0 = \mathrm{D}(\mathbf{S}_\infty^{-1})$ is symmetric.  Moreover, the Hessian $\mathrm{D}^2\log p_0$ is symmetric.  Recall that two symmetric matrices multiply to a symmetric matrix if and only if they commute.  Hence, for all $\mathbf{x}\in\mathbb{R}^D$,
\[
\mathrm{D}^2\log p_0\bigl(\mathbf{S}_\infty^{-1}(\mathbf{x})\bigr)
\quad\text{must commute with}\quad
\mathrm{D}(\mathbf{S}_\infty^{-1})(\mathbf{x}) .
\]
Setting $\mathbf{y}=\mathbf{S}_\infty^{-1}(\mathbf{x})$ gives the equivalent condition: for all $\rvy\in\mathbb{R}^D$,
\[
\mathrm{D}^2\log p_0(\mathbf{y})
\quad\text{must commute with}\quad
\mathrm{D}\mathbf{S}_\infty(\mathbf{y}) .
\]

Now, we transform this condition into a more computable form.  
Since $\mathbf{S}_\infty$ is optimal between $p_0$ and $\mathcal{N}(\bm{0},\rmI)$, Brenier’s theorem guarantees that $\mathbf{S}_\infty = \nabla u$ for some convex function $u$.  From the Monge–Ampère equation, it follows that:
\[
\log p_0(\rvy) = \log \det(\mathrm{D}^2 u(\rvy)) - \frac{1}{2} \|\nabla u(\rvy)\|^2 + \text{Constant}.
\]
The condition becomes (with $\mathrm{D}\rmS_\infty = \mathrm{D}^2 u$): 
\begin{mdframed}
\begin{align}\label{eq:necessary-ot}
\mathrm{D}^2 \left(\log \det \mathrm{D}^2 u - \frac{1}{2} \|\nabla u\|^2\right)
\quad\text{must commute with}\quad
\mathrm{D}^2 u.
\end{align}
\end{mdframed}
This yields a necessary condition for $\mathbf{T}_t$ to be optimal.
\subparagraph{Step 4: Constructing the Counterexample.} Let us show how to leverage this necessary condition to derive a contradiction.

Assume we can construct a convex function $u$ such that  
\[
\mathrm{D}^2 \left( \log \det \mathrm{D}^2 u(\mathbf{x}) - \frac{1}{2} \lvert \nabla u(\mathbf{x}) \rvert^2 \right)
\]
does not commute with $\mathrm{D}^2 u(\mathbf{x})$ for some $\mathbf{x} \in \mathbb{R}^D$.  Defining $p_0 = (\nabla u)^{-1} \# \mathcal{N}(\bm{0}, \rmI)$, Brenier's theorem implies that $\nabla u$ is the optimal transport from $p_0$ to $\mathcal{N}(\bm{0}, \rmI)$. However, the condition in \Cref{eq:necessary-ot} fails, leading to a contradiction. Thus, our goal is to construct such a function. 
Consider  
\[
u(\mathbf{x}) = \frac{1}{2} \|\mathbf{x}\|^2 + \varepsilon \phi(\mathbf{x}), \quad \text{for a small } \varepsilon.
\]
Then $\mathrm{D}^2 u(\mathbf{0}) = \rmI + \varepsilon \mathrm{D}^2 \phi(\mathbf{0})$, and the commutation condition at $\mathbf{x} = \mathbf{0}$ requires $\mathrm{D}^2 \phi(\mathbf{0})$ to commute with $\mathrm{D}^2 (\Delta \phi)(\mathbf{0})$. 

For example, in $\mathbb{R}^2$, choosing  
\[
\phi(x_1, x_2) = x_1 x_2 + x_1^4
\]
provides a counterexample where the Hessian $\mathrm{D}^2 \log p_0$ and the Jacobian $\mathrm{D}^2 u$ do not commute.

This contradiction shows that $\mathbf{T}_t$ cannot be optimal for all $t \geq 0$. Therefore, there exists some $t_0 \geq 0$ such that the map $\mathbf{T}_{t_0 \to \infty}$ is not optimal.

\newpage

\subsection{Can Canonical Linear Flow and Reflow Leads to an OT Map?}\label{subsec:ot-linear-reflow} 
We have seen that the PF-ODE (especially in VP type forward kernel) is generally not an OT map. One natural question now is:
\begin{question}
Does the linear interpolation flow $(1-t)\rvx_0 + t \rvx_1$ with $\rvx_0\sim p_{\mathrm{src}}$ and $\rvx_1\sim p_{\mathrm{tgt}}$, when applied to the independent coupling $\pi(\rvx_0, \rvx_1) = p_{\mathrm{src}}(\rvx_0)p_{\mathrm{tgt}}(\rvx_1)$, recover the OT map?
\end{question}
The answer to the question is no.
 
Nevertheless, combining a linear path with a given coupling offers a practical upper bound on the true OT cost.  Among all possible paths, linear interpolation provides the tightest such upper bound, as we will see in the following discussion.

\paragraph{Canonical Linear Flow and Optimal Transport.}
Focusing on optimal transport with quadratic cost, we consider the equivalent form of \Cref{eq:ot}, the Benamou--Brenier formulation in \Cref{eq:bb-ot}:
\begin{equation*}
    \mathcal{K}\left(p_{\mathrm{src}}, p_{\mathrm{tgt}}\right) :=  \min_{\substack{ (p_t, \rvv_t) \text{ s.t. }
        \partial_t p_t + \nabla \cdot (p_t \rvv_t) = 0, \\
        p_0 = p_{\mathrm{src}},\,\, p_1 = p_{\mathrm{tgt}}
    }}
    \int_0^1 \int_{\mathbb{R}^D} \|\rvv_t(\rvx)\|^2 p_t(\rvx) \diff\rvx \diff t.
\end{equation*}
Unlike the static Monge formulation (\Cref{eq: Monge's Formulation}),
where computing the optimal map requires solving the Monge--Amp\`ere
equation (\Cref{eq:monge-ampere}), the Benamou--Brenier problem admits a
convex reformulation after introducing the momentum field
$\rvm_t := p_t \rvv_t$. Nevertheless, discretizing this convex problem
on a grid in $\mathbb{R}^D$ remains intractable in high dimensions.

While solving the Benamou-Brenier formulation is generally intractable,  \citet{liu2022rectified,lipman2024flow} reveal that its kinetic energy admits a practical upper bound. This is achieved by restricting the search to a simpler family of \emph{conditional flows}, where each path is defined by its fixed endpoints $(\rvx_0, \rvx_1)$ drawn from a coupling $\pi_{0,1}$ of the source and target distributions. Within this \emph{conditional flow} family, the canonical linear interpolation emerges as the optimal choice, as formalized below.

\proppp{An Upper Bound on OT Kinetic Energy via Conditional Flows}{ot-conditional-flow}{
Let $\pi_{0,1}$ be any coupling between $p_{\mathrm{src}}$ and $p_{\mathrm{tgt}}$.
\begin{enumerate}
    \item[(1)] The kinetic energy is bounded above by the expected path energy of any conditional flow $\bm{\Psi}_t(\rvx_0, \rvx_1)$ that connects the endpoints:
    \begin{equation*}
        \mathcal{K}\left(p_{\mathrm{src}}, p_{\mathrm{tgt}}\right) \le \mathbb{E}_{(\rvx_0, \rvx_1) \sim \pi_{0,1}} \left[ \int_0^1 \|\bm{\Psi}'_t(\rvx_0, \rvx_1)\|^2 \diff t \right].
    \end{equation*}
    \item[(2)] The unique conditional flow $\bm{\Psi}^*_t$ that minimizes the upper bound on the right-hand side is the linear interpolation path:
    \begin{equation*}
        \bm{\Psi}^*_t(\rvx_0, \rvx_1) = (1 - t) \rvx_0 + t \rvx_1.
    \end{equation*}
  Substituting this optimal path yields the tightest version of the bound:
    \begin{equation*}
        \mathcal{K}\left(p_{\mathrm{src}}, p_{\mathrm{tgt}}\right) \le \mathbb{E}_{(\rvx_0, \rvx_1) \sim \pi_{0,1}} \|\rvx_1 - \rvx_0\|^2.
    \end{equation*}
\end{enumerate}
}{The proof relies on a straightforward application of Jensen's inequality and the tower property of conditional expectations, before solving the simplified variational problem with the Euler-Lagrange equation; we refer to \citep{lipman2024flow}'s Section 4.7 for the complete argument.}

In other words, the linear interpolation $\bm{\Psi}^*_t$ (i.e., the forward kernel used by Flow Matching and Rectified Flow) minimizes an upper bound on the true kinetic energy for any chosen coupling $\pi_{0,1}$.

We emphasize that optimality within this class of conditional flows does not guarantee global optimality on the marginal distributions.

\paragraph{Reflow and Optimal Transport.} 
The most naive transport plan between two distributions is to connect their samples with straight lines using a simple independent coupling. However, this approach is demonstrably not optimal, as the failure lies not in the straight-line paths themselves, but in the inefficient initial pairing of points. 

The Reflow procedure may offer a constructive response. It is an iterative algorithm designed specifically to correct this pairing, and crucially, each step is guaranteed to be cost-non-increasing \citep{liu2022flow}. This property suggests Reflow systematically pushes the transport plan towards a more optimal configuration, which naturally motivates the central question of its convergence.
\begin{question}
    What happens if we apply the \texttt{Rectify} operator iteratively? Can the resulting sequence of transport plans converge to the optimal one, or does the fixed point of the Reflow process yield the OT map?
\end{question}
The short answer is no in general. Below, we explain what may go wrong. To recall, the Reflow procedure iteratively refines the coupling between $p_{\mathrm{src}}$ and $p_{\mathrm{tgt}}$ via the update:
\[
\pi^{(k+1)} = \texttt{Rectify}(\pi^{(k)}),
\]
initialized with the product coupling $\pi^{(0)} := p_{\mathrm{src}}(\rvx_0) p_{\mathrm{tgt}}(\rvx_1)$. More precisely, $\texttt{Rectify}$ output the updated coupling $\pi^{(k+1)}$ via the following: At each iteration $k = 0, 1, 2, \ldots$, a velocity field $\rvv_t^{(k)}$ is learned via:
\[
\rvv_t^{(k)} \in \argmin_{\rvu_t} \, \mathcal{L}(\rvu_t\big\vert\pi^{(k)}),
\]
where $\mathcal{L}(\rvu_t\big\vert\pi^{(k)})$ is the loss (e.g., RF or FM loss) defined in \Cref{eq:reflow-loss}. Here, for notational simplicity, we adopt a non-parametric formulation for the velocity field, rather than a parameterized form $\bm{\phi}$ employed in other contexts.
The updated coupling is then given by:
\[
\pi^{(k+1)}(\rvx_0, \rvx_1) := p_{\mathrm{src}}(\rvx_0)\, \delta\big(\rvx_1 - \bm{\Psi}_1^{(k)}(\rvx_0)\big),
\]
where $\bm{\Psi}_1^{(k)}$ denotes the solution map at time $t=1$ obtained by integrating $\rvv_t^{(k)}$ from initial condition $\rvx_0$.

It has been observed in \citep{liu2022flow} that for a coupling $\pi$ between $p_{\mathrm{src}}$ and $p_{\mathrm{tgt}}$, the existence of a velocity field $\rvv_t$ that minimizes the Reflow loss, that is, satisfies $\mathcal{L}(\rvv_t |\pi) = 0$, does not necessarily imply that the transport is optimal.

Motivated by the Benamou–Brenier framework, where the optimal transport velocity is known to be the gradient of a potential function, \citet{liu2022rectified} proposed an additional constraint: the velocity field $\rvv_t$ should be a potential field. Accordingly, the objective in \Cref{eq:reflow-loss} is modified to restrict $\rvv_t$ to the space of gradient vector fields, also known as potential flows:
\begin{align}\label{eq:reflow-loss-gradient}
    \rvw_t^{(k)} \in \argmin_{\substack{\rvu_t:\, \rvu_t=\nabla\varphi \\ \text{for some }\varphi\colon\mathbb{R}^D\to\mathbb{R}}} \, \mathcal{L}(\rvu_t\big\vert\pi^{(k)}),
\end{align}
with the rest of the procedure remaining the same as in \texttt{Rectify}.
We denote this associated operator as $\texttt{Rectify}_{\perp}$, emphasizing the projection onto irrotational vector fields. 

Let $\pi$ be a coupling between $p_{\mathrm{src}}$ and $p_{\mathrm{tgt}}$. \citet{liu2022flow} conjecture the following equivalence characterizing optimality:
\begin{enumerate}[label=(\textit{\roman*})]
  \item $\pi$ \textit{is an optimal transport coupling.}
  \item $\pi$ \textit{is a fixed point of the potential rectification operator:}
  \[
    \pi = \texttt{Rectify}_{\perp}(\pi).
  \]
  \item \textit{There exists a gradient velocity field $\rvv_t = \nabla \varphi_t$ such that the rectify loss vanishes:}
  \[
    \mathcal{L}(\rvv_t |\pi) = 0.
  \]
\end{enumerate}
However, \citet{hertrich2025relation} exhibit two types of counterexamples:
\begin{enumerate}
  \item When the intermediate distributions $p_t$ have disconnected support, one can find fixed points of $\texttt{Rectify}_{\perp}$ with zero Reflow loss and gradient velocity fields that nonetheless fail to produce the optimal coupling.
  \item Even when both endpoint distributions are Gaussian, there exist couplings whose loss is arbitrarily small but whose deviation from the optimal coupling is arbitrarily large.
\end{enumerate}

Therefore, while rectified flows may yield strong generative models, their reliability as optimal transport solvers remains limited. This highlights an important gap between generative modeling and principled optimal transport theory, inviting further research at their intersection.

Finally, we note that transport cost does not always correlate with downstream performance; as such, computing the exact optimal transport map may not necessarily lead to better practical outcomes. Nonetheless, variants of optimal transport remain fundamental to many problems in science and engineering. Diffusion models offer a powerful framework for exploring these challenges.

\newpage
\section{Closing Remarks}
In this chapter, we revisited the classical problem of transporting one
distribution to another through the lens of optimal transport and its
entropic variants. We reviewed optimal transport in both its static
Monge--Kantorovich form and its dynamic Benamou--Brenier formulation,
where transport is realized by a deterministic flow minimizing kinetic
cost. We then discussed entropy-regularized optimal transport and the
Schr\"odinger bridge (SB), which replace deterministic transport by the
most likely controlled diffusion relative to a reference process. From
this perspective, diffusion models lie naturally near the SB viewpoint
on the stochastic side, while their PF-ODEs define deterministic
transport maps on the ODE side. As we emphasized, however, the PF-ODE
transport is generally \emph{not} the quadratic-cost optimal transport
map: it is one deterministic coupling among many, rather than the
minimizer of the Benamou--Brenier action.

This also highlights a more basic question: before asking whether a
transport is optimal, \emph{when can two distributions be connected by a
deterministic map at all?} In many settings, existence is not the main
difficulty. Under mild regularity assumptions, one can often find a
measurable map $\rmT$ such that
$\rmT_\# p_{\mathrm{src}} = p_{\mathrm{tgt}}$. The harder issue is that,
beyond one dimension, relatively few classical constructions are both
explicit and robust in practice. In one dimension, the monotone
rearrangement gives a simple answer by matching quantiles. In higher
dimensions, several important constructions are known: Brenier's theorem
(Theorem~\ref{thm:brenier}) yields the quadratic-cost optimal map under
suitable regularity; the Knothe--Rosenblatt
rearrangement~\citep{rosenblatt1952remarks,knothe1957contributions}
builds a triangular map from iterated conditional distributions; and the
Dacorogna--Moser approach~\citep{dacorogna1990partial} constructs smooth
diffeomorphic transports through a velocity field solving a divergence
equation. Each gives a principled route to deterministic transport, but
each also relies on structural assumptions or nontrivial analytical
machinery.

From this viewpoint, diffusion-based flow maps provide an appealing and
flexible alternative. Rather than prescribing a closed-form transport
map, diffusion models and flow-matching methods learn a time-dependent
vector field whose induced flow pushes a simple source distribution
toward the target. This learned map need not coincide with the Brenier
map or any other classical canonical construction, and in general it
does not solve a prescribed optimal transport problem. Nevertheless, it
offers a practical and expressive route to map-based sampling in high
dimensions, where explicit transport-map constructions are often
unavailable or difficult to compute. In this sense, diffusion flows are
best understood not as replacements for classical transport theory, but
as modern learned transport mechanisms that sit alongside optimal
transport maps, triangular rearrangements, and PDE-based constructions
within the broader landscape of deterministic transport.
\newpage

\part{Sampling of Diffusion Models}

\vspace*{\fill}  
\begin{figure*}[th!]
\begin{center}
\resizebox{\textwidth}{!}{%
\centering
\begin{tikzpicture}[
    box/.style={rectangle, draw, thick, minimum width=3cm, minimum height=2cm, align=center},
    topbox/.style={rectangle, draw, thick, minimum width=12cm, minimum height=1.5cm, align=center},
    arrow/.style={->, thick},
    roundbox/.style={rectangle, draw, thick, rounded corners=10pt, minimum width=4cm, minimum height=3cm, align=center}
]

\node[topbox, label=above:{\parbox{5cm}{\centering  {Chapter}~\ref{ch:score-sde}}}] (top) at (0,4) {\\
\textbfs{Generation with Diffusion Model $\mathbf{v}^*(\rvx, t)$} \\[0.3cm]
$\Longleftrightarrow$ Solve the ODE backward from $T$ to $0$ with $\mathbf{x}(T)\sim p_{\mathrm{prior}}$ (more generally, from $s$ to $t$ with $s>t$): \\[0.2cm]
$\displaystyle\frac{\diff\mathbf{x}(t)}{\diff t} = \mathbf{v}^*(\mathbf{x}(t), t)$ \\[0.3cm]
$\displaystyle\Longleftrightarrow \mathbf{x}(0) = \mathbf{x}(T) + \int_0^T \mathbf{v}^*(\mathbf{x}(t), t)\diff t$
};

\node[roundbox, label=below:{\parbox{5cm}{\centering  {Chapter}~\ref{ch:guidance} 
 }}] (left) at (-6,-2) {
\\ \large\textbfs{Steering Generation} \\[0.4cm]
\parbox{5cm}{
\begin{align*}
    \displaystyle\mathbf{x}(0) &= \mathbf{x}(T) + \int_0^T [\mathbf{v}^*(\mathbf{x}(t), t) \\ 
    &\qquad \textcolor{darkgray}{+ \textbf{Guidance}}] \diff t
\end{align*}}
};

\node[roundbox, label=below:{\parbox{5cm}{\centering {Chapter}~\ref{ch:solvers} 
 }}] (center) at (0,-2) {
\\ \large\textbfs{Fast Generation} \\
\large\textbfs{with Numerical Solvers} \\[-0.6cm]
$\displaystyle\mathbf{x}(0) = \mathbf{x}(T) + \eqnmarkbox[Plum]{velocity}{\int_0^T \mathbf{v}^*(\mathbf{x}(t), t)\diff t}$ \\[-0.8cm]
Estimating the Integration
};

\node[roundbox, label=below:{\parbox{5cm}{\centering  {Chapter}~\ref{ch:distillation} and {Chapter}~\ref{ch:fast-scratch} 
 }}] (right) at (6,-2) {
\\ \large\textbfs{Learning a Fast} \\
\large\textbfs{Diffusion-Based Generator} \\[-0.6cm]
$\displaystyle\mathbf{x}(0) = \eqnmarkbox[Pink]{target}{\mathbf{x}(T) + \int_0^T \mathbf{v}^*(\mathbf{x}(t), t)\diff t}$ \\[-0.8cm]
Learning the Integration
};

\draw[ultra thick, -{Stealth[length=4mm, width=4mm]}, rounded corners=4pt] (top.south) -- ++(-6,-1) -- (left.north);
\draw[ultra thick, -{Stealth[length=4mm, width=4mm]}, rounded corners=4pt] (top.south) -- (center.north);
\draw[ultra thick, -{Stealth[length=4mm, width=4mm]}, rounded corners=4pt] (top.south) -- ++(6,-1) -- (right.north);

\end{tikzpicture}
}
\end{center}
\end{figure*}
\vspace*{\fill}  

\chapter{Guidance and Controllable Generation}\label{ch:guidance}
Diffusion models are powerful generative frameworks. In the unconditional setting, the goal is to learn $p_{\text{data}}(\rvx)$ and generate samples without external input.

Many applications, however, require \emph{conditional generation}, where outputs satisfy user-specified criteria. This can be achieved by steering an unconditional model or directly learning the conditional distribution $p_0(\rvx | \rvc)$, with condition $\rvc$ (e.g., label, text description, or sketch) guiding the process.

This chapter builds on a principled view of the conditional score, which decomposes into an \emph{unconditional direction} and a \emph{guidance direction} that nudges samples toward the condition while preserving realism. We explain why guidance is essential, show how the conditional score serves as a unifying interface for control, and survey ways to approximate or replace the guidance term. We then distinguish \emph{control} (meeting the condition) from \emph{alignment} (meeting human preference under the condition), and describe how preferences can be incorporated into the same framework. Finally, we discuss direct optimization of preference without additional reward models (i.e., a learned scorer that assigns higher values to outputs better aligned with human preference).

\chapterlearning{

  \item understand how conditioning modifies the velocity field used during diffusion sampling;

  \item derive the key ideas behind classifier guidance and classifier-free guidance, and understand how guidance strength controls conditioning;

  \item understand how training-free guidance uses task-specific losses or measurement models to steer a pre-trained diffusion model, including for inverse problems;

  \item distinguish exact conditional sampling from surrogate-guided dynamics and their fixed-time tilted-density interpretations;

  \item understand how preference-based alignment, from RLHF to DPO and Diffusion-DPO, further reshapes a conditional generative model toward preferred outputs.

}{The chapter assumes the Score SDE/PF-ODE picture from \Cref{ch:score-sde,ch:all-equivalent}. Readers interested primarily in controllable generation may treat the later preference-alignment material as a second pass.}

\clearpage
\newpage

\section{Prologue}\label{sec:prologue-guidance}

\begin{figure}[th!]
    \centering
    \includegraphics[width=\linewidth]{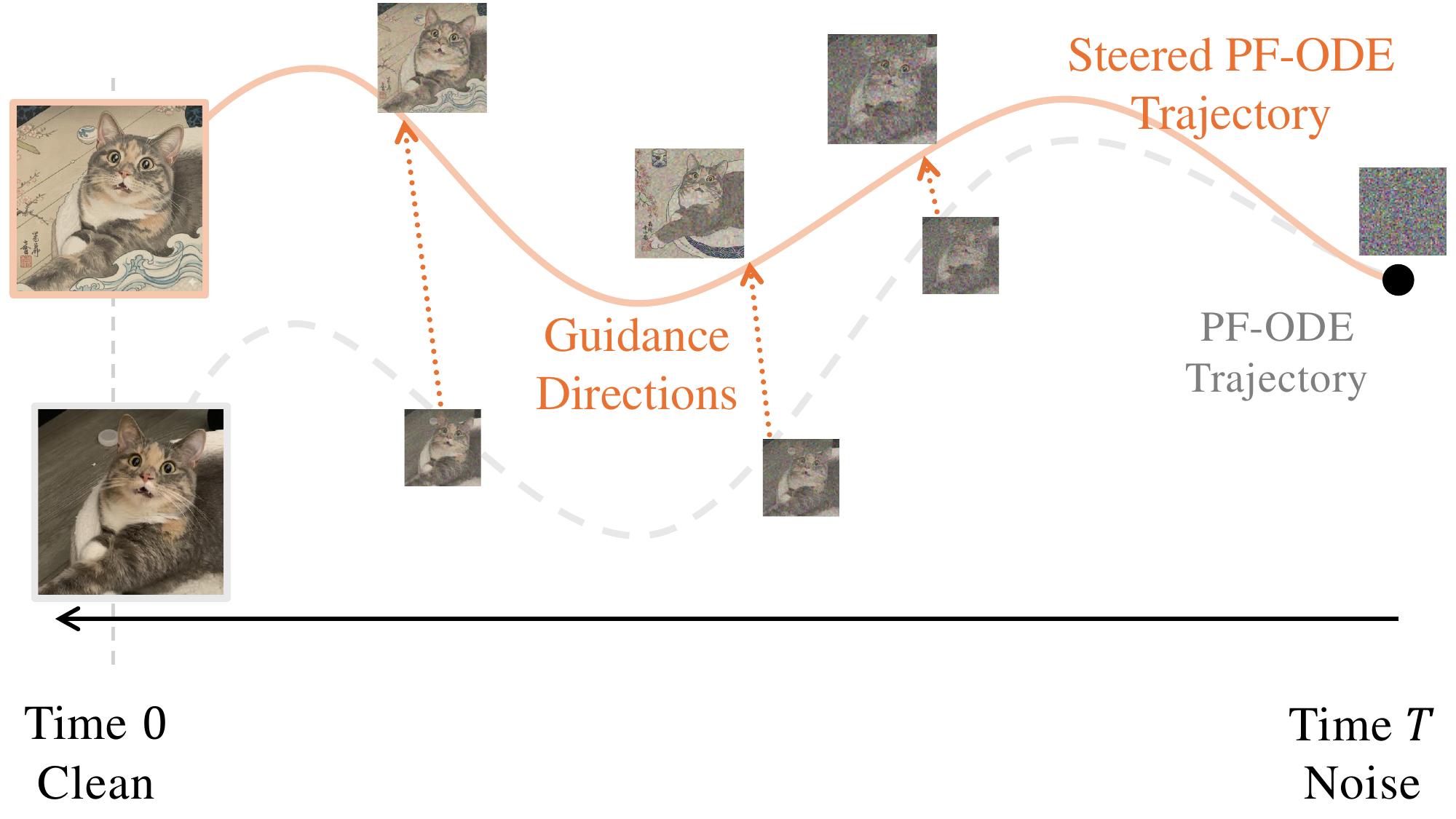}
\caption{\textbfs{Illustration of steered diffusion sampling.}
Reverse-time PF-ODE sampling begins from pure noise at the right ($t = T$) and gradually evolves toward a clean sample at the left ($t = 0$).
During this process, time-dependent guidance directions, weighted by $w_t$, modify the unconditional score field and steer the trajectory toward the desired attribute (Japanese painting style) while the sample is progressively refined from coarse to fine detail.
\figcredit{Created by the authors.}}
    \label{fig:guidance}
\end{figure}

The generation process of diffusion models proceeds in a coarse-to-fine manner, providing a flexible framework for controllable generation. At each step, a small amount of noise is removed and the sample becomes clearer, gradually revealing more structure and detail. This property enables control over the generation process: by adding a guidance term to the learned, time-dependent velocity field, we can steer the generative trajectory to reflect user intent.

A principled foundation for guidance-based sampling in diffusion models is the Bayesian decomposition of the conditional score. For each noise level $t$,
\begin{mdframed}
\begin{align}
\nabla_{\rvx_t}\log p_t(\rvx_t |\rvc)
=
\underbrace{\nabla_{\rvx_t}\log p_t(\rvx_t)}_{\text{unconditional direction}}
+
\underbrace{\nabla_{\rvx_t}\log p_t(\rvc |\rvx_t)}_{\text{guidance direction}}.
\label{eq:guidance_bayes}
\end{align}
\end{mdframed}
This identity shows that the exact conditional reverse dynamics are obtained by adding the guidance term
$\nabla_{\rvx_t}\log p_t(\rvc|\rvx_t)$ on top of the unconditional score. We will see later on that classifier guidance~\citep{dhariwal2021diffusion} estimates this term directly,
classifier-free guidance~\citep{ho2021classifier} estimates it implicitly as the difference between jointly learned conditional and unconditional scores,
and more general training-free methods may instead use a task-specific surrogate guidance direction.

With the exact conditional score, the PF-ODE becomes
\begin{align}\label{eq:condition_prob_ode}
\begin{aligned}
    \frac{\diff \mathbf{x}(t)}{\diff t} &= f(t)\mathbf{x}(t) - \frac{1}{2} g^2(t) \underbrace{\nabla_{\rvx_t} \log p_t(\rvx(t) |\rvc)}_{\text{conditional score}} \\
    &= f(t)\mathbf{x}(t) - \frac{1}{2} g^2(t) \Big[ \nabla_{\rvx_t} \log p_t(\rvx(t)) + \nabla_{\rvx_t} \log p_t(\rvc |\rvx(t)) \Big].
\end{aligned}
\end{align}
With exact scores and exact integration, these dynamics recover
$p_0(\rvx_0|\rvc)$ when initialized from the corresponding terminal marginal
$p_T(\rvx_T|\rvc)$. In standard diffusion constructions, the terminal noising
is chosen so that $p_T(\rvx_T|\rvc)$ is approximately matched by the usual
condition-independent prior.

We formulate the discussion below in score prediction for clarity. Since the
standard $\rvx$-, $\beps$-, score-, and $\rvv$-prediction parameterizations are
related by time-dependent linear transformations wherever these conversions are
well-defined, the same guidance correction can be expressed in the other
parameterizations using the relationships in \Cref{eq:predictions-equivalence}.

\paragraph{Instantiations of the Guidance Direction.}
\begin{enumerate}[leftmargin=*]
\item \textbfs{Classifier Guidance (CG).} In \Cref{sec:cg}, classifier guidance (CG) trains a time-conditional classifier
$p_{\bpsi}(\rvc | \rvx_t, t)$ on noised data $\rvx_t$ (obtained by corrupting
clean labeled samples at level $t$). At sampling time, its input gradient
provides the guidance term:
\[
\nabla_{\rvx_t}\log p_{\bpsi}(\rvc | \rvx_t, t)
 \approx  \nabla_{\rvx_t}\log p_t(\rvc | \rvx_t),
\]
which is then added to the unconditional score~\citep{dhariwal2021diffusion}.
\item \textbfs{Classifier-Free Guidance (CFG).} In \Cref{sec:cfg}, CFG directly trains a single conditional model
\[
\rvs_{\bphi}(\rvx_t,t,\rvc) \approx \nabla_{\rvx_t}\log p_t(\rvx_t|\rvc),
\]
where the unconditional model is learned jointly by randomly replacing the condition with a special null token for a fraction of the training steps.
\item \textbfs{Training-Free (Surrogate) Guidance.}
Under the standard forward noising construction, we first draw
$(\rvx_0,\rvc)\sim p_{\mathrm{data}}$ and then sample
$\rvx_t\sim p_t(\cdot|\rvx_0)$, where the forward kernel does not depend
additionally on $\rvc$. Thus
$\rvc\perp\rvx_t\mid\rvx_0$\footnote{
Indeed,
\[
p_t(\rvc|\rvx_t)
=
\int
p(\rvc,\rvx_0|\rvx_t)\,\mathrm d\rvx_0
=
\int
p(\rvc|\rvx_0,\rvx_t)\,
p(\rvx_0|\rvx_t)\,\mathrm d\rvx_0.
\]
Because the standard forward kernel satisfies
$p_t(\rvx_t|\rvx_0,\rvc)=p_t(\rvx_t|\rvx_0)$,
we have
$p(\rvc|\rvx_0,\rvx_t)=p(\rvc|\rvx_0)$,
which yields the expression below.
}.
Consequently,
\[
p_t(\rvc|\rvx_t)
=
\int
p(\rvc|\rvx_0)\,
p(\rvx_0|\rvx_t)\,
\mathrm d\rvx_0.
\]
This posterior average is generally intractable, motivating training-free
methods that instead construct a task-specific surrogate guidance direction.

In \Cref{subsec:tfg}, training-free (loss-based) guidance avoids evaluating the
conditional likelihood $p_t(\rvc|\rvx_t)$ directly. Instead, it starts from an
off-the-shelf loss $\ell(\rvx_t,\rvc;t)$ that measures how well the current
sample satisfies the desired condition, with smaller loss indicating better
agreement. From this loss, we define the positive \emph{guidance potential}
\[
\rvh_t(\rvx_t;\rvc)
:=
\exp\big(-\tau\,\ell(\rvx_t,\rvc;t)\big),
\qquad \tau>0.
\]
Thus, lower-loss samples receive larger potential values. We denote the
corresponding \emph{guidance direction} by
\[
\mathcal{G}_t(\rvx_t,\rvc)
:=
\nabla_{\rvx_t}\log \rvh_t(\rvx_t;\rvc)
=
-\tau\nabla_{\rvx_t}\ell(\rvx_t,\rvc;t).
\]

Classifier guidance fits naturally into the same picture. At the population
level, its guidance potential is
\[
\rvh_t(\rvx_t;\rvc)
=
p_t(\rvc|\rvx_t),
\]
which is approximated in practice by the learned classifier
$p_{\bpsi^\times}(\rvc|\rvx_t,t)$. Equivalently, the learned classifier
corresponds to the loss
\[
\ell(\rvx_t,\rvc;t)
=
-\log p_{\bpsi^\times}(\rvc|\rvx_t,t),
\qquad \tau=1.
\]

\end{enumerate}

At the population level, the three constructions above share the same additive
guidance template. Given a positive potential
$\rvh_t(\rvx_t;\rvc)$ and guidance strength $w_t\geq0$, define the guided score
field
\begin{align}
\rvs^{\mathrm{guide}}(\rvx_t,t,\rvc;w_t)
:=
\nabla_{\rvx_t}\log p_t(\rvx_t)
+
w_t\mathcal{G}_t(\rvx_t,\rvc).
\label{eq:generic-guided-score}
\end{align}
For loss-based guidance,
$\mathcal{G}_t(\rvx_t,\rvc)
=-\tau\nabla_{\rvx_t}\ell(\rvx_t,\rvc;t)$.
For exact classifier guidance,
$\mathcal{G}_t(\rvx_t,\rvc)
=\nabla_{\rvx_t}\log p_t(\rvc|\rvx_t)$; classifier-free guidance expresses
the same population guidance direction through the difference between
conditional and unconditional scores.

At each fixed time $t$, \Cref{eq:generic-guided-score} is, whenever
normalizable, the score of the tilted density
\[
\widetilde p_t^{\mathrm{tilt}}(\rvx_t|\rvc)
\propto
p_t(\rvx_t)\rvh_t(\rvx_t;\rvc)^{w_t}.
\]
This gives a useful energy interpretation: the base density favors samples
supported by the diffusion model, while the potential reweights them toward
the desired condition.

In practice, the corresponding learned or surrogate version of
\Cref{eq:generic-guided-score} is inserted into the reverse-time dynamics.
Importantly, the tilted density above is only a fixed-time interpretation:
the family
$\{\widetilde p_t^{\mathrm{tilt}}(\cdot|\rvc)\}_t$
need not coincide with the marginals generated by the guided sampler.
Consequently, for a generic loss or guidance schedule, integrating the guided
PF-ODE from $T$ to $0$ does not in general produce samples distributed
according to $\widetilde p_0^{\mathrm{tilt}}(\cdot|\rvc)$.
The additional cross-time consistency required for this interpretation to hold
is discussed in \Cref{subsec:guided-marginal-path}.

The effect of guidance on the sampling trajectory is illustrated in
\Cref{fig:guidance}. The same principle can also be applied on top of an
already conditional model, allowing additional control signals to modify its
sampling dynamics.

\paragraph{From Control to Better Alignment with Direct Preference Optimization.}

Strong control can be on-condition but off-preference: a sample may satisfy the
conditioning signal (e.g., the prompt) yet deviate from what humans actually
prefer. We formalize this by tilting the conditional target at $t=0$ by a
preference rating:
\[
\widetilde p_0^{\mathrm{tilt}}(\rvx_0|\rvc)
\propto
p_0(\rvx_0|\rvc)
\exp\bigl(
\lambda_{\mathrm{pref}} r(\rvx_0,\rvc)
\bigr),
\qquad
\lambda_{\mathrm{pref}}>0,
\]
where $r(\rvx_0,\rvc)$ is a scalar alignment rating for a clean sample
$\rvx_0$ and condition $\rvc$, with larger values indicating stronger
preference, and $\lambda_{\mathrm{pref}}$ controls the strength of the
preference reweighting. In the KL-regularized formulation introduced later,
the optimizer takes this form with the reference policy as the base
distribution and $\lambda_{\mathrm{pref}}=1/\beta$.

Existing methods for achieving such steerability typically collect human labels of the relative quality of model generations and fine-tune the conditional diffusion model to align with these preferences, often through reinforcement learning from human feedback (RLHF). However, RLHF is a complex and often unstable procedure: it first fits a reward model to capture human preferences, and then fine-tunes the conditional diffusion model with reinforcement learning to maximize this estimated reward while constraining policy drift from the original model.

This naturally raises the question: \emph{can we remove the reward model training stage altogether?} We address this with Diffusion-DPO~\citep{wallace2024diffusion}, an adaptation of Direct Preference Optimization~\citep{rafailov2023direct} originally developed for large language models. As described in \Cref{sec:rlhf-dpo}, Diffusion-DPO learns the preference tilt directly from pairwise choices, so the conditional diffusion model is fine-tuned to align to preferences without a separate reward model.

\clearpage
\newpage

\section{Classifier Guidance}\label{sec:cg}

\subsection{Foundation of Classifier Guidance}
Let $\rvc$ denote a conditioning variable drawn from a distribution $p(\rvc)$, such as a class label, caption, or other auxiliary information. Our goal is to draw samples from $p_0(\rvx | \rvc)$. In diffusion-based conditional generation, we realize this goal by running the reverse-time dynamics whose time marginals are $p_t(\cdot | \rvc)$. The drift of these dynamics depends on the conditional score
\[
\nabla_{\rvx_t}\log p_t(\rvx_t | \rvc), \quad t \in [0,T].
\]
Hence a standard and effective route\footnote{One could in principle obtain $p_0(\rvx | \rvc)$ from an unconditional generator via rejection or importance sampling if $p(\rvc | \rvx)$ were available and well calibrated. This is rarely practical for high-dimensional or rare conditions.} is to estimate this quantity.

A fundamental insight, based on Bayes' rule, is that the conditional score can be decomposed as:
\begin{align}
\nabla_{\rvx_t} \log p_t(\rvx_t | \rvc) 
&= \nabla_{\rvx_t} \log \left( \frac{p_t(\rvx_t)   p_t(\rvc | \rvx_t)}{p(\rvc)} \right) \nonumber \\
&= \nabla_{\rvx_t} \log p_t(\rvx_t) + \nabla_{\rvx_t} \log p_t(\rvc | \rvx_t) - \nabla_{\rvx_t} \log p(\rvc) \nonumber \\
&= \underbrace{\nabla_{\rvx_t} \log p_t(\rvx_t)}_{\text{unconditional score}} 
+ \underbrace{\nabla_{\rvx_t} \log p_t(\rvc | \rvx_t)}_{\text{classifier gradient}}, \label{eq:cg_bayes}
\end{align}
where $p_t(\rvc | \rvx_t)$ indicates a probability of $\mathbf{c}$ conditioned on $\mathbf{x}_{t}$ which predicts the condition $\rvc$ from the noisy input $\rvx_t$ at time $t$. 

This decomposition\footnote{In the last identity,
$\nabla_{\rvx_t}\log p(\rvc)=0$ because $p(\rvc)$ does not depend on
$\rvx_t$.} motivates the \emph{Classifier Guidance} (CG) approach proposed by
\citet{dhariwal2021diffusion}. To separate the population identity from the
learned classifier used in practice, first consider the oracle quantities.
For a guidance scale $\omega\geq0$, define at each fixed time $t$
\begin{align}
\widetilde p_t^{\mathrm{CG},\omega}(\rvx_t|\rvc)
\propto
p_t(\rvx_t)p_t(\rvc|\rvx_t)^\omega.
\label{eq:cg-density-omega}
\end{align}
Its score is
\begin{align}
\rvs^{\mathrm{CG}}(\rvx_t,t,\rvc;\omega)
&:=
\nabla_{\rvx_t}
\log\widetilde p_t^{\mathrm{CG},\omega}(\rvx_t|\rvc)
\nonumber\\
&=
\nabla_{\rvx_t}\log p_t(\rvx_t)
+
\omega\nabla_{\rvx_t}\log p_t(\rvc|\rvx_t).
\label{eq:cg_omega}
\end{align}
Thus $\omega$ scales the classifier-guidance direction. When $\omega=1$,
Bayes' rule gives
$\widetilde p_t^{\mathrm{CG},1}(\rvx_t|\rvc)=p_t(\rvx_t|\rvc)$, and
\Cref{eq:cg_omega} recovers the exact conditional score. For a general
$\omega$, $\widetilde p_t^{\mathrm{CG},\omega}$ provides a fixed-time tempered
interpretation of the guided field; whether it also forms the marginal path of
the guided sampler is discussed in
\Cref{subsec:guided-marginal-path}.

The scalar $\omega \ge 0$ modulates the influence of the classifier:
\begin{itemize}
    \item $\omega = 1$: recovers the true conditional score $\nabla_{\rvx_t} \log p_t(\rvx_t |\rvc)$.
    \item $\omega > 1$: amplifies the classifier signal, typically increasing conditional fidelity (often at the expense of diversity).
    \item $0 \le \omega < 1$: down-weights the classifier signal, typically increasing sample diversity while weakening conditioning.
\end{itemize}

\paragraph{Practical Approximation in CG.}
In practice, CG is a training-free method  (w.r.t.\ the diffusion model) for steering a pre-trained unconditional diffusion model,
\[
\rvs_{\bm{\phi}^\times}(\rvx_t ,t) \approx \nabla_{\rvx_t} \log p_t(\rvx_t).
\]
CG is applied only at sampling time, without modifying the diffusion model itself. To enable this, a time-dependent classifier $p_{\bpsi}(\rvc |\rvx_t,t)$ is trained separately to predict the condition $\rvc$ from noisy inputs $\rvx_t$ at different noise levels $t$. The classifier is trained in a standard way by minimizing the cross-entropy loss:
\begin{align}
\mathbb{E}_{t \sim \mathcal{U}[0,T], (\rvx, \rvc) \sim p_{\text{data}},  \bm{\epsilon} \sim \mathcal{N}(\mathbf{0}, \mathbf{I})}
\Big[ -\log p_{\bpsi}(\rvc | \rvx_t,t) \Big],
\end{align}
where $(\rvx,\rvc)\sim p_{\text{data}}$ denotes paired labeled data, and $\rvx_t = \alpha_t \rvx + \sigma_t \bm{\epsilon}$ is the noisy input at time $t$. The classifier must be trained across noise levels and typically receives $t$, or an
equivalent noise-level representation, as input so that it can approximate
$p_t(\rvc|\rvx_t)$ throughout the diffusion path.

After training, the classifier provides scores that serves as a surrogate for the true likelihood gradient:
\[
\nabla_{\rvx_t} \log p_{\bpsi^\times}(\rvc |\rvx_t,t) \approx \nabla_{\rvx_t} \log p_t(\rvc |\rvx_t).
\]

\subsection{Inference with CG}
At inference time, the learned classifier gradient
$\nabla_{\rvx_t}\log p_{\bpsi^\times}(\rvc|\rvx_t,t)$ is added to the
unconditional score estimate and scaled by $\omega$, yielding a learned
approximation to the oracle field in \Cref{eq:cg_omega}:
\begin{align*}
\rvs^{\mathrm{CG}}_{\mathrm{model}}(\rvx_t,t,\rvc;\omega)
:=
\underbrace{\rvs_{\bm{\phi}^\times}(\rvx_t,t)}_{\text{uncond.\ direction}}
+
\omega
\underbrace{\nabla_{\rvx_t}\log
p_{\bpsi^\times}(\rvc|\rvx_t,t)}_{\text{guidance direction}}
\approx
\rvs^{\mathrm{CG}}(\rvx_t,t,\rvc;\omega).
\end{align*}
The unconditional score estimate in the reverse-time SDE or PF-ODE is then
replaced by
$\rvs^{\mathrm{CG}}_{\mathrm{model}}(\rvx_t,t,\rvc;\omega)$, thereby steering the generative trajectory toward samples that align with the condition $\rvc$.

\subsection{Advantages and Limitations}

CG provides a simple and flexible mechanism for conditional generation, allowing for explicit control over the strength of conditioning via $\omega$. It can be used with any pre-trained unconditional diffusion model, requiring only an additional classifier for conditioning.

However, the approach has notable limitations:
\begin{itemize}
    \item \textbfs{Training Cost:} The classifier must be trained to operate across all noise levels, which is computationally expensive.
    \item \textbfs{Robustness:} Classifiers must generalize well to severely corrupted inputs $\rvx_t$, especially for large $t$, which can be challenging.
    \item \textbfs{Separate Training:} Since the classifier is trained independently of the diffusion model, it may not align perfectly with the learned data distribution.
\end{itemize}

\clearpage
\newpage

\section{Classifier-Free Guidance}\label{sec:cfg}
\subsection{Foundation of Classifier-Free Guidance}

\begin{figure}[th!]
    \centering
\includegraphics[width=\linewidth]{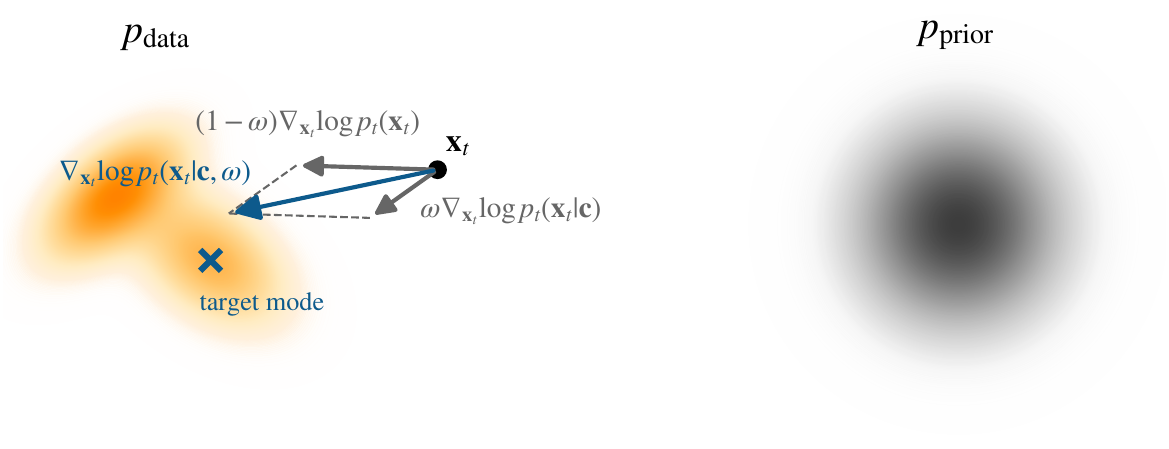}
\caption{\textbfs{Illustration of CFG.}
The guided direction is a linear combination of the conditional score
$\nabla_{\rvx_t}\log p_t(\rvx_t|\rvc)$ and the unconditional score
$\nabla_{\rvx_t}\log p_t(\rvx_t)$, controlled by $\omega$.
For $0\leq\omega\leq1$, this interpolates between the two directions; for
$\omega>1$, it extrapolates beyond the conditional score in the conditional
direction. The resulting field is used to steer the sampling trajectory toward
the conditioning signal.
\figcredit{Created by the authors.}}
    \label{fig:cfg}
\end{figure}

\emph{Classifier-free guidance} (CFG)~\citep{ho2021classifier} removes the need
for a separately trained classifier by expressing the classifier-guidance
direction through conditional and unconditional scores. Bayes' rule gives
\begin{align}
\nabla_{\rvx_t}\log p_t(\rvc|\rvx_t)
=
\nabla_{\rvx_t}\log p_t(\rvx_t|\rvc)
-
\nabla_{\rvx_t}\log p_t(\rvx_t).
\end{align}
Substituting this identity into \Cref{eq:cg_omega} gives the oracle CFG field
\begin{align}\label{eq:cfg-model}
\rvs^{\mathrm{CFG}}(\rvx_t,t,\rvc;\omega)
&:=
\nabla_{\rvx_t}\log p_t(\rvx_t)
+
\omega
\left(
\nabla_{\rvx_t}\log p_t(\rvx_t|\rvc)
-
\nabla_{\rvx_t}\log p_t(\rvx_t)
\right)
\nonumber\\
&=
\omega
\underbrace{\nabla_{\rvx_t}\log p_t(\rvx_t|\rvc)}_{\text{conditional score}}
+
(1-\omega)
\underbrace{\nabla_{\rvx_t}\log p_t(\rvx_t)}_{\text{unconditional score}}.
\end{align}
With exact scores, \Cref{eq:cfg-model} is algebraically the same guided field
as \Cref{eq:cg_omega}. At $\omega=1$, it reduces to the exact conditional
score. For other guidance strengths, its fixed-time tilted-density
interpretation should be distinguished from the marginals generated by the
guided dynamics, as discussed in
\Cref{subsec:guided-marginal-path}.

The hyperparameter $\omega$ again plays a critical role in controlling the influence of the conditioning information (we take $\omega \ge 0$):
\begin{itemize}
    \item At $\omega = 0$, the model behaves as an unconditional diffusion model, completely ignoring the conditioning.
    \item At $\omega = 1$, the model uses the conditional score without additional guidance.
    \item For $\omega > 1$, the model places more emphasis on the conditional score and less on the unconditional score, strengthening alignment with $\rvc$ but typically reducing diversity.
\end{itemize}

\subsection{Training and Sampling of CFG}

\paragraph{Joint Training of Unconditional and Conditional Diffusion Models via CFG.}
CFG requires a diffusion model trained to support both conditional and
unconditional predictions. Rather than training two separate models, CFG uses
a single model $\rvs_{\bm{\phi}}(\rvx_t,t,\rvc)$ and randomly replaces the
conditioning input by a null token $\emptyset$ during training:
\begin{itemize}
    \item with the null input, the model learns the unconditional prediction
    $\rvs_{\bm{\phi}}(\rvx_t,t,\emptyset)$;
    \item with the original condition, it learns the conditional prediction
    $\rvs_{\bm{\phi}}(\rvx_t,t,\rvc)$.
\end{itemize}
The null condition is used with probability $p_\mathrm{uncond}$, a
user-specified condition-dropout probability. The full training procedure is
shown in \Cref{alg:cfg_training}, alongside standard unconditional training in
\Cref{alg:no_cfg_training}. The guidance weight $\omega$ is introduced only
at sampling time and is not part of this training objective.

\begin{center}
\begin{minipage}[t]{0.48\textwidth}
  \begin{algorithm}[H]
  {\small
    {\small \caption{Uncond. DM \label{alg:no_cfg_training} }}
    \begin{algorithmic}[1]
      \Repeat
        \State $\rvx \sim p_{\text{data}}(\rvx)$
        \State $t \sim \mathcal{U}[0, T]$
        \State $\bm{\epsilon} \sim \mathcal{N}(\mathbf{0}, \mathbf{I})$
        \State $\rvx_t = \alpha_t \rvx + \sigma_t \bm{\epsilon}$
        \State Take gradient step on:
        \Statex \quad $\nabla_{\bm{\phi}} \left\| \rvs_{\bm{\phi}}(\rvx_t, t) - \rvs \right\|^2$
      \Until{converged}
    \end{algorithmic}
    }
  \end{algorithm}
\end{minipage}
\hfill
\begin{minipage}[t]{0.48\textwidth}
  \begin{algorithm}[H]
      {\small
    {\small \caption{CFG for Cond. DM \label{alg:cfg_training}}}
    \begin{algorithmic}[1]
      \Require \textcolor{orange}{$p_\mathrm{uncond}$: prob. of unconditional dropout}
      \Repeat
        \State \textcolor{orange}{$(\rvx, \rvc) \sim p_{\text{data}}(\rvx, \rvc)$}
        \State \textcolor{orange}{$\rvc \gets \emptyset$ with prob. $p_\mathrm{uncond}$}
        \State $t \sim \mathcal{U}[0, T]$
        \State $\bm{\epsilon} \sim \mathcal{N}(\mathbf{0}, \mathbf{I})$
        \State $\rvx_t = \alpha_t \rvx + \sigma_t \bm{\epsilon}$
        \State Take gradient step on:
        \Statex \quad \textcolor{orange}{$\nabla_{\bm{\phi}} \left\| \rvs_{\bm{\phi}}(\rvx_t, t, \rvc) - \rvs \right\|^2$}
      \Until{converged}
    \end{algorithmic}
    }
  \end{algorithm}
\end{minipage}
\end{center}

\paragraph{Conditioned Sampling with CFG.}

Once the model $\rvs_{\bm{\phi}^\times}(\rvx_t,t,\rvc)$ is trained using
Algorithm~\ref{alg:cfg_training}, CFG forms the learned guided field
\begin{align}\label{eq:cfg-model-nn}
\rvs_{\mathrm{model}}^{\mathrm{CFG}}(\rvx_t,t,\rvc;\omega)
&:=
\omega
\underbrace{\rvs_{\bm{\phi}^\times}(\rvx_t,t,\rvc)}_{\text{conditional score estimate}}
+
(1-\omega)
\underbrace{\rvs_{\bm{\phi}^\times}(\rvx_t,t,\emptyset)}_{\text{unconditional score estimate}}
\nonumber\\
&\approx
\rvs^{\mathrm{CFG}}(\rvx_t,t,\rvc;\omega).
\end{align}
During sampling, the unconditional score estimate in the reverse-time SDE or
PF-ODE is replaced by
$\rvs_{\mathrm{model}}^{\mathrm{CFG}}$.
Thus CFG defines a modified sampling field by combining the learned
conditional and unconditional score estimates.

This formulation enables controllable generation by adjusting $\omega$,
providing a simple post-training trade-off between conditioning strength and
sample diversity without requiring a separately trained classifier or
unconditional diffusion model.



\clearpage
\newpage

\section{Training-Free Guidance}
Training-free guidance modifies the sampling dynamics of a fixed pre-trained
diffusion model using an external condition, loss, or measurement objective,
without retraining the diffusion model itself. The Bayes decomposition in
\Cref{eq:guidance_bayes} provides a useful motivation in score space: guidance
adds a condition-dependent direction to the unconditional score. In practice,
a task-specific loss can be used to construct corrections either to the
predicted clean sample or to the score used during sampling.

We first introduce these two general forms in \Cref{subsec:tfg}. We then ask
what distribution such surrogate-guided dynamics actually generate in
\Cref{subsec:guided-marginal-path}, before turning to representative
training-free approaches for inverse problems in
\Cref{subsec:inverse-problem}.

\paragraph{Setup and Notations.}
Let $\rvc$ denote a conditioning variable. We assume access to a pre-trained
diffusion model $\rvs_{\bm{\phi}^\times}(\rvx_t,t)$ expressed in score
prediction\footnote{Here, we adopt the score and $\rvx$-prediction
parameterizations for simplicity; other parameterizations such as
$\beps$-prediction can be handled analogously.}. In addition, suppose we are
given a task loss
\[
\ell(\cdot,\rvc)\colon\mathbb{R}^D\to\mathbb{R}
\]
that measures disagreement with the desired condition, so that smaller
$\ell(\rvx,\rvc)$ indicates better agreement. Examples include a perceptual
distance to a reference target, or a feature-space distance or negative
similarity derived from a pre-trained representation such as
CLIP~\citep{radford2021learning}.

Consider the standard linear--Gaussian forward noising kernel
\[
p_t(\cdot|\rvx_0)
:=
\mathcal N\bigl(\cdot;\alpha_t\rvx_0,\sigma_t^2\rmI\bigr).
\]
We borrow the DDIM update introduced later in \Cref{eq:ddim-update-all} as a
concrete example:
\begin{align}\label{eq:ddim-update-clean-noise-score}
\rvx_{t\rightarrow t-1}
=
\alpha_{t-1}
\underbrace{\hat{\rvx}_0(\rvx_t)}_{\text{in data space}}
-
\sigma_{t-1}\sigma_t
\underbrace{\hat{\rvs}(\rvx_t)}_{\text{in noise space}},
\end{align}
where
$\hat{\rvx}_0(\rvx_t):=\rvx_{\bm{\phi}^\times}(\rvx_t,t)$ is the clean
$\rvx$-prediction and
$\hat{\rvs}(\rvx_t):=\rvs_{\bm{\phi}^\times}(\rvx_t,t)$ is the score
prediction at noise level $t$.

\subsection{Conceptual Framework for Training-Free Guidance}\label{subsec:tfg}

A common strategy is to use gradients of the task loss to modify either the
predicted clean sample or the score term in the sampling update. To distinguish
these two operations, define the data-space direction
\[
\mathcal G_t^{\mathrm{data}}(\rvx_t,\rvc)
:=
-
\left.
\nabla_{\rvx_0}\ell(\rvx_0,\rvc)
\right|_{\rvx_0=\hat{\rvx}_0(\rvx_t)}.
\]
For the noise-space correction, we reuse the guidance direction introduced in
the Prologue,
\[
\mathcal G_t(\rvx_t,\rvc)
:=
\nabla_{\rvx_t}\log\rvh_t(\rvx_t;\rvc),
\]
for a positive guidance potential $\rvh_t$, specified below.

Using the DDIM update in \Cref{eq:ddim-update-clean-noise-score}, the two
corrections can be written compactly as
\begin{align}\label{eq:ddim-update-clean-noise-guide}
\begin{aligned}
\rvx_{t\rightarrow t-1}
=
\alpha_{t-1}
\underbrace{\left(
\hat{\rvx}_0(\rvx_t)
+
{\color{orange}
\eta_t^{\text{data}}
\mathcal G_t^{\text{data}}(\rvx_t,\rvc)}
\right)}_{\text{A. data space}}
-
\sigma_{t-1}\sigma_t
\underbrace{\left(
\hat{\rvs}(\rvx_t)
+
{\color{orange}
\eta_t^{\text{latent}}
\mathcal G_t(\rvx_t,\rvc)}
\right)}_{\text{B. noise space}},
\end{aligned}
\end{align}
where
$\eta_t^{\text{data}},\eta_t^{\text{latent}}\geq0$
control the strengths of the data-space and noise-space corrections,
respectively.

\paragraph{A. Guidance in Data Space.}
The data-space correction directly improves the current clean estimate by
taking a local step that decreases the task loss:
\[
\hat{\rvx}_0(\rvx_t)
\;\longmapsto\;
\hat{\rvx}_0(\rvx_t)
+
\eta_t^{\text{data}}
\mathcal G_t^{\text{data}}(\rvx_t,\rvc).
\]
Thus the clean prediction is first moved toward better agreement with the
condition and then used in the subsequent sampling update. This correction
can also be applied repeatedly when multiple optimization steps are desired.

Representative examples include MGPD~\citep{he2023manifold} and
UGD~\citep{bansal2023universal}.

\paragraph{B. Guidance in Noise Space.}
A closely related strategy applies guidance directly to the score used during
sampling. Following the potential view introduced in
\Cref{sec:prologue-guidance}, a common construction evaluates the task loss on
the current denoised estimate and chooses
\[
\rvh_t(\rvx_t;\rvc)
:=
\exp\!\left(
-\tau\,\ell\bigl(\hat{\rvx}_0(\rvx_t),\rvc\bigr)
\right),
\qquad \tau>0.
\]
By the definition of $\mathcal G_t$ above, this gives
\begin{align*}
\mathcal G_t(\rvx_t,\rvc)
=
\nabla_{\rvx_t}\log\rvh_t(\rvx_t;\rvc)
=
-\tau
\nabla_{\rvx_t}
\ell\bigl(\hat{\rvx}_0(\rvx_t),\rvc\bigr).
\end{align*}
Accordingly, the score term in
\Cref{eq:ddim-update-clean-noise-guide} becomes
\[
\hat{\rvs}(\rvx_t)
+
\eta_t^{\text{latent}}
\mathcal G_t(\rvx_t,\rvc).
\]
Comparing this practical update with the population-level construction in
\Cref{eq:generic-guided-score},
$\eta_t^{\text{latent}}$ plays the role of the generic guidance strength
$w_t$, while $\hat{\rvs}(\rvx_t)$ replaces the exact unconditional score by
its learned estimate.

When $\rvh_t(\rvx_t;\rvc)$ is proportional to the true conditional likelihood
$p_t(\rvc|\rvx_t)$ up to a factor independent of $\rvx_t$,
$\mathcal G_t(\rvx_t,\rvc)$ recovers the Bayes guidance direction
$\nabla_{\rvx_t}\log p_t(\rvc|\rvx_t)$. For a generic task loss,
$\mathcal G_t$ instead defines a surrogate guidance direction.

Evaluating $\mathcal G_t$ requires differentiating through the
$\rvx$-prediction:
\[
\mathcal G_t(\rvx_t,\rvc)
=
-\tau
\left[
\nabla_{\rvx_t}\hat{\rvx}_0(\rvx_t)
\right]^\top
\left.
\nabla_{\rvx_0}\ell(\rvx_0,\rvc)
\right|_{\rvx_0=\hat{\rvx}_0(\rvx_t)},
\]
which may incur substantial computational cost in practice.

Representative examples include
\citep{yu2023freedom,chung2022diffusion,bansal2023universal}.

The constructions above specify how guidance modifies each sampling step.
They do not, by themselves, determine the marginal distributions obtained
after these modified dynamics are integrated across all noise levels. We
clarify this distinction next.

\subsection{What Distribution Does Guided Sampling Produce?}
\label{subsec:guided-marginal-path}

The potential view above has an exact fixed-time density interpretation at the
population level. In particular, replacing the learned unconditional score
estimate $\hat{\rvs}$ by the exact score, a positive potential
$\rvh_t(\rvx;\rvc)$ and guidance strength $w_t\geq0$ define the ideal guided
field
\[
\nabla_{\rvx}\log p_t(\rvx)
+
w_t\nabla_{\rvx}\log\rvh_t(\rvx;\rvc).
\]
At each fixed $t$, this is the score of
\begin{align}
\widetilde p_t^{\mathrm{tilt}}(\rvx|\rvc)
=
\frac{1}{Z_t(\rvc)}
p_t(\rvx)\rvh_t(\rvx;\rvc)^{w_t},
\qquad
Z_t(\rvc)
=
\int
p_t(\rvx)\rvh_t(\rvx;\rvc)^{w_t}\,\diff\rvx,
\label{eq:guided-fixed-time-tilt}
\end{align}
provided $0<Z_t(\rvc)<\infty$. The learned field used in practice approximates
this ideal construction through the learned score estimate
$\hat{\rvs}(\rvx_t)$.

For the loss-based construction above,
\[
\rvh_t(\rvx_t;\rvc)
=
\exp\!\left[
-\tau\ell\bigl(\hat{\rvx}_0(\rvx_t),\rvc\bigr)
\right],
\]
and $\eta_t^{\mathrm{latent}}$ plays the role of $w_t$.

This fixed-time identity does not determine the distribution generated after
the guided dynamics are integrated across noise levels. Consequently, for a
generic loss and guidance schedule, solving the guided PF-ODE from $T$ to $0$
does not in general produce samples distributed according to
$\widetilde p_0^{\mathrm{tilt}}(\cdot|\rvc)$.

The reason is cross-time consistency: densities at different noise levels
cannot be chosen independently. For
$\{\widetilde p_t^{\mathrm{tilt}}\}_t$ to form the marginal path associated
with the same forward diffusion, they must also satisfy the corresponding
Fokker-Planck equation:
\[
\partial_t \widetilde p_t^{\mathrm{tilt}}
+
\nabla_{\rvx} \cdot
\bigl(f(t)\rvx \widetilde p_t^{\mathrm{tilt}}\bigr)
-
\frac{1}{2}g^2(t)
\Delta_{\rvx}\widetilde p_t^{\mathrm{tilt}}
=
0.
\]
When this condition holds, the corresponding PF-ODE
\[
\frac{\diff\rvx(t)}{\diff t}
=
f(t)\rvx(t)
-
\frac{1}{2}g^2(t)
\nabla_{\rvx}\log
\widetilde p_t^{\mathrm{tilt}}\left(\rvx(t)|\rvc\right)
\]
has $\widetilde p_t^{\mathrm{tilt}}$ as its marginal path when initialized from
the corresponding terminal distribution
$\widetilde p_T^{\mathrm{tilt}}(\cdot|\rvc)$.

The compatibility condition can be checked directly. Substituting
\Cref{eq:guided-fixed-time-tilt} into the Fokker-Planck equation and using
the fact that $p_t$ itself satisfies the same equation gives
\begin{align*}
&
\partial_t \widetilde p_t^{\mathrm{tilt}}
+
\nabla_{\rvx} \cdot
\bigl(f(t)\rvx \widetilde p_t^{\mathrm{tilt}}\bigr)
-
\frac{1}{2}g^2(t)
\Delta_{\rvx}\widetilde p_t^{\mathrm{tilt}}
\\
&=
\widetilde p_t^{\mathrm{tilt}}
\Bigg[
\partial_t
\log \bigl(\rvh_t^{ w_t}\bigr)
+
f(t)\rvx\cdot
\nabla_{\rvx}\log \bigl(\rvh_t^{ w_t}\bigr)
-
g^2(t)
\nabla_{\rvx}\log p_t
\cdot
\nabla_{\rvx}\log \bigl(\rvh_t^{ w_t}\bigr)
\\[-2pt]
&\hspace{3.4cm}
-
\frac{1}{2}g^2(t)
\left(
\Delta_{\rvx}\log \bigl(\rvh_t^{ w_t}\bigr)
+
\left\|
\nabla_{\rvx}\log \bigl(\rvh_t^{ w_t}\bigr)
\right\|^2
\right)
-
\frac{Z_t'}{Z_t}
\Bigg],
\end{align*}
where the arguments $(\rvx;\rvc)$ are suppressed for readability.
Thus, the tilted family is a valid marginal path only when the bracketed
quantity vanishes. A generic task loss and guidance schedule need not satisfy
this condition. The mismatch can therefore remain even with exact score
evaluation and exact ODE integration: it comes from the guided dynamics
themselves rather than from numerical discretization. This is an instance of
the broader Fokker-Planck consistency requirement discussed in
\citep{lai2023fp}.

Two simple cases clarify the distinction. When $w_t\equiv0$, the tilt vanishes
and the original unconditional path $p_t$ is recovered. For exact classifier
guidance, under the standard setting where the forward diffusion corrupts
$\rvx$ while leaving $\rvc$ fixed,
\[
\rvh_t(\rvx;\rvc)
=
p_t(\rvc|\rvx),
\qquad
w_t\equiv1.
\]
Bayes' rule then gives
\[
\widetilde p_t^{\mathrm{tilt}}(\rvx|\rvc)
=
p_t(\rvx|\rvc),
\]
so the tilted family is exactly the conditional marginal path. This explains
the special role of guidance weight $1$ for the oracle classifier likelihood.

For a general guidance strength, for example
\[
p_t(\rvx) p_t(\rvc|\rvx)^\omega,
\]
the expression still defines a valid fixed-time tilted density whenever it is
normalizable, but it is not generally guaranteed to describe the marginals of
the guided sampler. Special choices may satisfy the required compatibility
condition, so $\omega\neq1$ alone does not imply inconsistency.

Correction-based approaches such as Feynman-Kac
Correctors~\citep{skreta2025feynman} explicitly account for this cross-time
mismatch when the goal is to follow a prescribed sequence of tilted
distributions more faithfully.

\subsection{Examples of Training-Free Approaches to Inverse Problems}\label{subsec:inverse-problem}
The principle introduced in \Cref{subsec:tfg} has important applications in inverse problems. We begin with an overview of the background and then provide several concrete examples illustrating how to leverage pre-trained diffusion models for inference-time inverse problem solving.

\paragraph{Background on Inverse Problems.}
Let $\mathcal A$ denote a known forward operator, possibly nonlinear, and
suppose that
\begin{align}\label{eq:inverse_problem}
\rvy
=
\mathcal A(\rvx_0)
+
\sigma_{\rvy}\rvz,
\qquad
\rvz\sim\mathcal N(\bm 0,\rmI).
\end{align}
The Bayesian objective is to characterize the posterior
$p_0(\rvx_0|\rvy)$, which combines the measurement model with prior knowledge
about plausible clean signals. Blind inverse problems extend this setting by
treating unknown parameters of $\mathcal A$ as additional quantities to infer.

Classical inverse-problem methods commonly combine a data-fidelity term with an
analytical or hand-designed prior in a variational or Bayesian formulation.
Supervised restoration methods may instead learn a direct mapping from
measurements to clean signals using paired data. Pre-trained diffusion models
provide a third route: a learned prior over clean data can be reused across
measurement operators, with measurement information introduced only at
inference time.

\paragraph{Pre-Trained Diffusion Models as Inverse Problem Solvers.}
For the measurement model above, Bayes' rule gives
\begin{align}
\nabla_{\rvx_t}\log p_t(\rvx_t|\rvy)
=
\underbrace{\nabla_{\rvx_t}\log p_t(\rvx_t)}_{\text{data score}}
+
\underbrace{\nabla_{\rvx_t}\log p_t(\rvy|\rvx_t)}_{\text{measurement guidance}}.
\label{eq:bayes_rule}
\end{align}
The data score can be supplied by a pre-trained diffusion model, whereas the
measurement term is generally unavailable in closed form. Training-free
inverse solvers therefore introduce measurement information during sampling,
but they need not all do so in the same mathematical form.

A useful distinction is between two common strategies. Some methods directly
enforce measurement consistency by modifying or projecting the intermediate
sample. Others approximate the likelihood score
$\nabla_{\rvx_t}\log p_t(\rvy|\rvx_t)$ and add it to the data score through
\Cref{eq:bayes_rule}.

\subparagraph{ILVR~\citep{choi2021ilvr}.}
ILVR provides a simple example of the first strategy. Let $\phi_N$ denote its
linear low-pass operator. At each reverse step, one first samples an
unconditional proposal
\[
\rvx_{t-1}'
\sim
p_{\bm\theta}(\rvx_{t-1}'|\rvx_t)
\]
and noises the reference image to the same level,
\[
\rvy_{t-1}
\sim
p_{t-1}(\rvy_{t-1}|\rvy).
\]
The proposal is then refined according to
\[
\rvx_{t-1}
=
\phi_N(\rvy_{t-1})
+
(\rmI-\phi_N)\rvx_{t-1}'.
\]
Thus, the prescribed low-frequency component comes from the noised reference,
while the complementary component comes from the unconditional diffusion
proposal. This is a sample-space consistency correction rather than an
explicit approximation of
$\nabla_{\rvx_t}\log p_t(\rvy|\rvx_t)$.

\subparagraph{Diffusion Posterior Sampling (DPS)~\citep{chung2022diffusion}.}
For a known nonlinear forward operator $\mathcal A$ with additive Gaussian
measurement noise $\sigma_{\rvy}>0$, \emph{Diffusion Posterior Sampling}
(DPS) approximates
\begin{align}\label{eq:dps-score}
\nabla_{\rvx_t}\log p_t(\rvy|\rvx_t)
\approx
\nabla_{\rvx_t}
\log p\bigl(\rvy|X_0=\hat\rvx_0(\rvx_t)\bigr),
\end{align}
where
$\hat\rvx_0(\rvx_t):=\E[\rvx_0|\rvx_t]$
is the posterior mean of the clean sample given $\rvx_t$, typically estimated
from the pre-trained diffusion model using Tweedie's formula
(\Cref{eq:tweedie}). 

This plug-in approximation replaces the posterior average of the measurement
likelihood by its value at the posterior mean:
\begin{align*}
  p_t(\rvy|\rvx_t)
  &= \int p_t(\rvy|\rvx_t, \rvx_0)  p(\rvx_0|\rvx_t)  \mathrm{d}\rvx_0 \\
  &= \int p_t(\rvy|\rvx_0)  p(\rvx_0|\rvx_t)  \mathrm{d}\rvx_0 
   \approx  p_t\bigl(\rvy|X_0=\hat\rvx_0(\rvx_t)\bigr),
\end{align*}
where we have used that $\rvy$ depends only on $\rvx_0$ (not on $\rvx_t$) given $\rvx_0$, and the approximation holds under the assumption that the posterior $p(\rvx_0|\rvx_t)$ is tightly peaked around its mean.

Since 
\[
  p_t\bigl(\rvy|X_0 = \hat\rvx_0(\rvx_t)\bigr)
  = \mathcal{N}\bigl(\rvy;  \mathcal{A}(\hat\rvx_0(\rvx_t)),  \sigma_{\rvy}^2 \rmI\bigr),
\]
we compute
\begin{align*}
  \nabla_{\rvx_t} \log p_t(\rvy|\rvx_t)
  &\approx \nabla_{\rvx_t} \log \mathcal{N}\bigl(\rvy;  \mathcal{A}(\hat\rvx_0),  \sigma_{\rvy}^2 \rmI\bigr) \\
  &= -\frac{1}{2\sigma_{\rvy}^2}  \nabla_{\rvx_t} \bigl\| \rvy - \mathcal{A}(\hat\rvx_0) \bigr\|^2 \quad  \\
  &= \frac{1}{\sigma_{\rvy}^2}  \left[ \mathcal{J}_{\mathcal{A}}\big(\hat\rvx_0(\rvx_t)\big) \cdot \nabla_{\rvx_t} \hat\rvx_0(\rvx_t) \right]^\top \bigl( \rvy - \mathcal{A}\big(\hat\rvx_0(\rvx_t)\big) \bigr),
\end{align*}
where $\mathcal{J}_{\mathcal{A}}\big(\hat\rvx_0(\rvx_t)\big) := \nabla_{\rvx_0} \mathcal{A}(\rvx) \big|_{\rvx = \hat\rvx_0(\rvx_t)}$ denotes the Jacobian of the forward operator with respect to its input. This formula propagates the gradient through the score approximation pipeline, reflecting how the measurement likelihood changes with respect to perturbations in the noisy sample $\rvx_t$.

For linear inverse problems, this further simplifies to:
\begin{align*}
    \nabla_{\rvx_t} \log p_t(\rvy | \rvx_t) \approx \eqnmarkbox[OliveGreen]{guidance}{\frac{1}{\sigma_{\rvy}^2}} \,\,    \eqnmarkbox[Plum]{proj}{\left[ \rmA \cdot \nabla_{\rvx_t} \hat\rvx_0(\rvx_t) \right]^\top} \left(\eqnmarkbox[NavyBlue]{merror}{\rvy - \rmA(\hat\rvx_0(\rvx_t))}\right).
\end{align*}

A large body of work explores diffusion-based inverse problem solvers by proposing various approximations for $\nabla_{\rvx_t} \log p_t(\rvy | \rvx_t)$.  For a comprehensive overview, we refer readers to the survey by~\citet{daras2024survey}.

\paragraph{A Related Supervised Iterative-Restoration Viewpoint and Its Connection to the PF-ODE.}
The discussion above solves inverse problems by combining a pre-trained diffusion prior with the Bayes-rule decomposition in \Cref{eq:bayes_rule}. More specifically, these methods use the score of the prior distribution together with the measurement model to guide the reconstruction process. At the same time, they also reflect a broader idea: instead of recovering the clean signal in one step, they gradually improve a corrupted input through many small updates. A similar high-level strategy also appears in supervised image restoration, but from a different starting point. Rather than using a pre-trained score model as a prior, these methods learn the refinement procedure directly from paired corrupted and clean examples.

A representative example is InDI~\citep{delbracio2023inversion}. Unlike the Bayes-rule-based approaches above, InDI begins with paired data $(\rvy,\rvx_0)$, where $\rvy$ is a corrupted observation and $\rvx_0$ is the corresponding clean target. Importantly, it does not assume in general that the corruption from $\rvx_0$ to $\rvy$ is known analytically or that it is Gaussian. 

The basic idea is to connect the two endpoints by the interpolation path
\[
\rvx_t=(1-t)\rvx_0+t\rvy,\qquad t\in[0,1].
\]
This path should be understood only as a convenient trajectory from the clean target to the corrupted observation. It is not a Gaussian noising process in general. Along this path, for any $0\le s\le t$, one has the exact relation
\[
\rvx_s=\Bigl(1-\frac{s}{t}\Bigr)\rvx_0+\frac{s}{t}\rvx_t.
\]
Taking conditional expectation given $\rvx_t$ yields
\[
\mathbb E[\rvx_s|\rvx_t]
=
\Bigl(1-\frac{s}{t}\Bigr)\mathbb E[\rvx_0|\rvx_t]
+
\frac{s}{t}\rvx_t.
\]
This shows that, to move from the current state $\rvx_t$ to a slightly less degraded state, it is enough to predict the clean target from $\rvx_t$.

Based on this idea, one trains a time-conditioned predictor $\rmF_{\bphi}(\rvx_t,t)$ from paired examples by minimizing the regression objective
\[
\min_{\bphi}\;
\mathbb E_{(\rvx_0,\rvy)\sim p(\rvx_0,\rvy),  t\sim p(t)}
\Bigl\|
\rmF_{\bphi}\bigl((1-t)\rvx_0+t\rvy, t\bigr)-\rvx_0
\Bigr\|_2^2,
\]
where $p(\rvx_0,\rvy)$ denotes the joint distribution of data pairs consisting of a clean target $\rvx_0$ and its corrupted observation $\rvy$, and $p(t)$ specifies how the interpolation time is sampled during training.
That is, $\rmF_{\bphi}$ is trained to predict the clean target from an intermediate degraded state along the interpolation path. For this squared-error loss, the Bayes-optimal predictor is the conditional expectation
\[
\rmF^*(\rvx_t,t)=\mathbb E[\rvx_0|\rvx_t].
\]
Therefore, if the model class is sufficiently expressive and the optimization is successful, the learned predictor approximates the average clean target associated with the intermediate state $\rvx_t$ at time $t$. In this sense, the training objective is similar in spirit to diffusion-model training: in both cases, one learns a time-conditioned regression from an intermediate corrupted state to a cleaner target. The difference is that diffusion models generate such training pairs synthetically from a prescribed forward corruption process, whereas here the endpoint pairs $(\rvy,\rvx_0)$ come directly from supervised restoration data.

At inference time, one starts from $\hat{\rvx}_1=\rvy$ and repeatedly refines the estimate by
\[
\hat{\rvx}_{t-\Delta t}
=
\frac{\Delta t}{t}\rmF_{\bphi}(\hat{\rvx}_t,t)
+
\Bigl(1-\frac{\Delta t}{t}\Bigr)\hat{\rvx}_t.
\]
Hence, the final reconstruction is produced by gradual refinement rather than by a single direct prediction. In the continuous-time limit, this leads to InDI's residual-flow ODE
\[
\frac{\diff \rvx_t}{\diff t}
=
\frac{\rvx_t-\rmF_{\bphi}(\rvx_t,t)}{t}.
\]

Now suppose we further specialize to the Gaussian denoising setting, where the corruption is assumed to take the known analytic form
\[
\rvy=\rvx_0+\sigma\beps,
\qquad \beps\sim\mathcal N(\bm 0,\rmI),
\]
with known noise level \(\sigma\). Under this specialization, InDI's residual-flow ODE reduces exactly to the probability-flow ODE. Indeed, the interpolation path becomes
\[
\rvx_t=(1-t)\rvx_0+t\rvy
=
\rvx_0+t\sigma\beps,
\]
which is precisely the standard diffusion form
\[
\rvx_t=\alpha_t\rvx_0+\sigma_t\beps,
\qquad\text{with}\qquad
\alpha_t=1,\;\sigma_t=t\sigma.
\]
Therefore, the predictor \(\rmF_{\bphi}(\rvx_t,t)\approx \mathbb E[\rvx_0\mid \rvx_t]\) is now learning to denoise a Gaussian-corrupted intermediate state \(\rvx_t\). In this case, the residual-flow ODE becomes
\[
\frac{\diff \rvx_t}{\diff t}
=
\frac{\rvx_t-\mathbb E[\rvx_0\mid \rvx_t]}{t}
=
-\dot{\sigma}_t\sigma_t\nabla_{\rvx_t}\log p_t(\rvx_t),
\]
where the last equality follows from \(\sigma_t=t\sigma\) together with Tweedie's formula,
\[
\mathbb E[\rvx_0\mid \rvx_t]
=
\rvx_t+\sigma_t^2\nabla_{\rvx_t}\log p_t(\rvx_t).
\]
This is exactly the PF-ODE for this Gaussian denoising path, written in either denoiser form or score form. Thus, under the Gaussian denoising specialization, InDI suggests an alternative interpretation of diffusion modeling: one may view it as supervised denoiser learning on Gaussian-corrupted data, with the PF-ODE arising as the continuous-time limit of gradual refinement.

\newpage
\section{From Reinforcement Learning to Direct Preference Optimization for Model Alignment}\label{sec:rlhf-dpo}

In the pursuit of aligning generative models with human intent, the prevailing paradigm has been Reinforcement Learning from Human Feedback (RLHF). While effective, RLHF is a complex, multi-stage process that can be unstable. This section introduces \emph{Direct Preference Optimization (DPO)}~\citep{rafailov2023direct}, a more streamlined and stable method that reaches the same goal without explicit reward modeling or reinforcement learning. We then outline its extension to diffusion models via \emph{Diffusion-DPO}~\citep{wallace2024diffusion}.

\subsection{The Motivation: Circumventing the Pitfalls of RLHF}

The goal of alignment is to steer a base, pre-trained model (e.g., an SFT model) toward outputs that humans prefer. RLHF proceeds in three stages. First, \emph{supervised fine-tuning (SFT)} trains a base model on prompt–response pairs. Second, \emph{reward modeling (RM)} fits a model on preference data consisting of prompts $\mathbf{c}$ and paired responses (a preferred ``winner'' $\mathbf{x}_w$ and a dispreferred ``loser'' $\mathbf{x}_l$), learning a scalar $r(\mathbf{c},\mathbf{x})$ with $r(\mathbf{c},\mathbf{x}_w)>r(\mathbf{c},\mathbf{x}_l)$. Third, \emph{RL fine-tuning} optimizes the SFT model (policy $\pi$\footnote{A policy maps a prompt/history (state) to a distribution over responses/actions.}) with an algorithm such as PPO~\citep{schulman2017proximal}, maximizing expected reward from $r$ while regularizing by a KL penalty that keeps $\pi$ close to the reference/SFT distribution.

Despite its impact, this pipeline suffers from drawbacks: the RL stage is unstable and computationally expensive because it is on-policy—each update requires freshly generated samples from the current model; it also requires training and hosting multiple large models (SFT, reward, and sometimes a value model); and it optimizes only a proxy for human preferences, so flaws in the reward model can be exploited. This motivates a central question:
\begin{question}
    Can we eliminate explicit reward modeling and the unstable RL step, directly optimizing the model on preference data?
\end{question}

Direct Preference Optimization (DPO) streamlines alignment by replacing the multi-stage RLHF pipeline with a single, supervised-style step. Instead of training a separate reward model and running unstable RL algorithms like PPO, DPO directly fits the policy to preference pairs using a simple logistic loss, while staying close to a fixed reference model. The key insight is that the KL-regularized RLHF objective can be rewritten so that the log-likelihood ratio between the policy and the reference acts as an implicit reward. This preserves the same regularization toward the reference policy but avoids costly rollouts and explicit reward modeling.

In \Cref{subsec:rlhf-bradley-terry}, we briefly review the RLHF pipeline and its reliance on large reward models and RL fine-tuning. In \Cref{subsec:dpo}, we present DPO, originally proposed for language models, which circumvents reward model training and simplifies alignment fine-tuning. Finally, in \Cref{subsec:diffusion-dpo}, we extend this idea to diffusion models, introducing Diffusion-DPO as a practical and stable alignment method in the generative modeling setting.

\subsection{RLHF: Bradley-Terry View}\label{subsec:rlhf-bradley-terry}

\paragraph{Short Introduction to RLHF.}
RLHF begins with a learned judge: a reward model $r_{\bm\psi}$ that assigns a scalar preference score to candidate responses for the same prompt $\mathbf c$. The dataset $\mathcal D$ consists of pairs $(\tilde{\mathbf x},\mathbf x)$ annotated with a label $y$ indicating whether $\tilde{\mathbf x}$ is preferred over $\mathbf x$. The label can be binary $y\in\{0,1\}$ or a soft value $y\in[0,1]$ obtained by aggregating multiple raters. The training objective is a simple logistic loss
\begin{align}\label{eq:rm-emp}
\begin{aligned}
    \mathcal L_{\mathrm{RM}}(\bm\psi)
=
- \mathbb E_{(\mathbf c,\tilde{\mathbf x},\mathbf x,y)\sim\mathcal D}
\Big[
y &\log \sigma \big(r_{\bm\psi}(\mathbf c,\tilde{\mathbf x})-r_{\bm\psi}(\mathbf c,\mathbf x)\big)
\\ &+(1-y)\log \big(1-\sigma \big(r_{\bm\psi}(\mathbf c,\tilde{\mathbf x})-r_{\bm\psi}(\mathbf c,\mathbf x)\big)\big)
\Big],
\end{aligned}
\end{align}
where $\sigma(u)=1/(1+e^{-u})$. In practice, preference pairs in $\mathcal D$ may originate from various sources: curated responses, model snapshots at different checkpoints, or generations from a pre-trained conditional diffusion model.  
A standard convention is to store them in an ordered format $(\text{winner}, \text{loser})$.  
Under this convention we simply set $y=1$, and \Cref{eq:rm-emp} reduces to the special case (with $\tilde\rvx = \rvx^w$ and $\rvx = \rvx^l$):
\begin{equation}\label{eq:rm-bt}
\mathcal L_{\mathrm{RM}}(\bm\psi)
=
- \mathbb E_{(\mathbf c,\mathbf x_w,\mathbf x_l)\sim\mathcal D}
\Big[\log \sigma \big(r_{\bm\psi}(\mathbf c,\mathbf x_w)-r_{\bm\psi}(\mathbf c,\mathbf x_l)\big)\Big].
\end{equation}

\paragraph{Bradley-Terry View and KL Connection.}
It is standard to interpret
\[
p_{r_{\bm\psi}}(\tilde{\mathbf x}\succ \mathbf x|\mathbf c)
:= \sigma \big(r_{\bm\psi}(\mathbf c,\tilde{\mathbf x})-r_{\bm\psi}(\mathbf c,\mathbf x)\big)
\]
through the Bradley-Terry (BT) model~\citep{bradley1952rank}, which converts two scalar scores into a win probability.  
This formulation highlights two key properties: (i) only the \emph{difference} of scores matters (so $r_{\bm\psi}(\mathbf c,\cdot)$ is shift-invariant), and (ii) the loss pushes the predicted winner’s score above the loser’s score. To see (ii) intuitively, consider one pair with label $y\in\{0,1\}$ and define
\[
\Delta r := r_{\bm\psi}(\mathbf c,\tilde{\mathbf x})-r_{\bm\psi}(\mathbf c,\mathbf x),
\quad
p:=\sigma(\Delta r),\quad \sigma(u)=\tfrac{1}{1+e^{-u}}.
\]
The per-example logistic loss is
\[
\ell=-\big[y\log p+(1-y)\log(1-p)\big].
\]
Then
\[
\frac{\partial \ell}{\partial \Delta r}=\sigma(\Delta r)-y.
\]
Under gradient descent with step size $\eta>0$, the score gap updates as
\[
\Delta r \leftarrow \Delta r-\eta\big(\sigma(\Delta r)-y\big).
\]
Hence, if $y=1$ (``$\tilde{\mathbf x}$ wins''), then $\sigma(\Delta r)-1\le 0$, so $\Delta r$ increases (winner up, loser down); if $y=0$, $\Delta r$ decreases.

Each per-example term in \Cref{eq:rm-emp} can be viewed as the cross-entropy between the observed Bernoulli label and the model’s predicted win probability:
\[
-\big[y\log p_{r_{\bm\psi}}+(1-y)\log(1-p_{r_{\bm\psi}})\big]
= \mathcal D_{\mathrm{KL}} \big(\mathrm{Bern}(y) \big\| \mathrm{Bern}(p_{r_{\bm\psi}})\big)
+ \mathcal H \big(\mathrm{Bern}(y)\big),
\]
where $\mathcal H$ is the entropy of the target Bernoulli distribution.
Averaging over the dataset $\mathcal D$ gives
\begin{align}\label{eq:reward_loss}
    \mathcal L_{\mathrm{RM}}(\bm\psi)
= \mathbb E_{\mathcal D} \Big[\mathcal D_{\mathrm{KL}}\left(\mathrm{Bern}(y) \big\| \mathrm{Bern}(p_{r_{\bm\psi}})\right)\Big]
+ \underbrace{\mathbb E_{\mathcal D} \big[\mathcal H(\mathrm{Bern}(y))\big]}_{\text{independent of }\bm\psi}.
\end{align}
Thus, minimizing the logistic loss is equivalent to minimizing the KL divergence between the empirical Bernoulli distribution of human labels and the model’s predicted Bernoulli distribution. In the binary case ($y\in\{0,1\}$), this equivalence is exact; for soft labels ($y\in[0,1]$), the result holds up to an entropy constant offset. Intuitively, the reward model is trained to adjust its win probabilities until they align with the empirical human win rates observed in the dataset.

From this point onward, we adopt the most common convention where $\mathcal D$ stores pairs in an ordered format: $(\rvx^w, \rvx^l, \rvc) \sim \mathcal D$. Under this convention, the label is always $y=1$, and the loss simplifies to the ordered form given in \Cref{eq:rm-bt}, which we will use in the following discussion.

\paragraph{KL Regularized Policy Optimization (with Fixed Reward).}
With the fitted reward $r:=r_{\bm\psi^\times}$ trained via \Cref{eq:rm-bt}, and a conditional pre-trained diffusion model $p_{\bm{\phi}^\times}(\mathbf x|\mathbf c)$, RLHF then adjusts a learnable policy $\pi_{\bm\theta}(\mathbf x|\mathbf c)$, usually fine-tuned on top of $p_{\bm{\phi}^\times}(\mathbf x|\mathbf c)$, toward higher-reward responses.  
At the same time, the policy is regularized to stay close to a reference model, taken as the pre-trained diffusion model $\pi_{\mathrm{ref}}(\mathbf x|\mathbf c):=p_{\bm{\phi}^\times}(\mathbf x|\mathbf c)$, using a $\mathcal{D}_{\mathrm{KL}}$ penalty:
\begin{equation}\label{eq:rlhf-policy-correct}
\max_{\bm\theta} 
\mathbb E_{\mathbf c\sim p(\mathbf c)}
\Big[
\mathbb E_{\mathbf x\sim \pi_{\bm\theta}(\cdot|\mathbf c)}\big[r_{\bm\psi}(\mathbf c,\mathbf x)\big]
-\beta \mathcal{D}_{\mathrm{KL}} \big(\pi_{\bm\theta}(\cdot|\mathbf c) \big\| \pi_{\mathrm{ref}}(\cdot|\mathbf c)\big)
\Big],
\end{equation}
which makes the two forces explicit: seek samples the judge prefers, but stay close to the pre-trained reference.

We remark that the reward objective in \Cref{eq:rm-bt} uses only labeled pairs and does not require that $\mathcal D$ be generated by the reference model (i,e., the pre-trained conditional diffusion model). While not required, collecting pairs from models close to the intended policy can reduce distribution shift and make the learned reward more reliable in the region where it will be used.

In summary, RLHF proceeds in two stages: first fit the reward $r^*$ by minimizing the loss in \Cref{eq:rm-bt} (equivalently, the expected binary $\mathcal{D}_{\mathrm{KL}}$ in \Cref{eq:reward_loss}); then optimize the policy $\pi^*$ by solving \Cref{eq:rlhf-policy-correct}.

\newpage

\subsection{DPO Framework}\label{subsec:dpo}
\paragraph{The Bridge from RLHF.}
The KL-regularized policy objective in \Cref{eq:rlhf-policy-correct} has a simple closed-form solution for each prompt $\mathbf c$, given the fitted reward $r := r_{\bm\psi^\times}$, expressed in the following energy-based form~\citep{peters2010relative}:
\begin{equation}\label{eq:rlhf-closed-form}
\pi^*(\mathbf x|\mathbf c)
=
\frac{1}{Z(\mathbf c)} \pi_{\mathrm{ref}}(\mathbf x|\mathbf c)\exp(r(\mathbf c,\mathbf x)/\beta),
\end{equation}
where $\pi_{\mathrm{ref}}(\mathbf x|\mathbf c):=p_{\bm{\phi}^\times}(\mathbf x|\mathbf c)$,  and $Z(\mathbf c)$ is the partition function ensuring $\int \pi^*(\mathbf x|\mathbf c) \diff \mathbf x=1$.

For smaller $\beta$, $\exp(r/\beta)$ becomes sharper, so $\pi^*$ concentrates on high reward regions: reward dominates, the policy moves farther from $\pi_{\mathrm{ref}}$, diversity decreases, and training may become unstable or prone to reward hacking. For larger $\beta$, $\exp(r/\beta)$ flattens, keeping $\pi^*$ closer to $\pi_{\mathrm{ref}}$: the KL term dominates, updates are conservative, diversity follows the reference, but reward gains are limited.

Since our aim is to fine-tune the policy directly (without training a separate reward model), \Cref{eq:rlhf-closed-form} lets us \emph{define} an \textit{implicit reward} from any policy. We introduce below:
\paragraph{Defining an Implicit Reward Motivated by Inverting \Cref{eq:rlhf-closed-form}.}
\Cref{eq:rlhf-closed-form} suggests an immediate inversion: for any policy $\pi$ (with support contained in $\pi_{\mathrm{ref}}$), define
\begin{equation}\label{eq:implicit-reward}
r_{\pi}(\mathbf c,\mathbf x)
 = 
\beta \log\frac{\pi(\mathbf x|\mathbf c)}{\pi_{\mathrm{ref}}(\mathbf x|\mathbf c)}
 + \beta \log Z(\mathbf c).
\end{equation}
Then \Cref{eq:rlhf-closed-form} holds with $\pi$ in place of $\pi^*$, i.e., $\pi$ would be the optimizer of \Cref{eq:rlhf-policy-correct} for the reward function $r_{\pi}$.
In this sense, $r_{\pi}$ is an \emph{implicit (policy-induced) reward}: it is identified up to the prompt-dependent constant $\beta\log Z(\mathbf c)$, which vanishes in any pairwise comparison such as in the BT model:
\[
r_{\pi}(\mathbf c,\mathbf x_w)-r_{\pi}(\mathbf c,\mathbf x_l)
=
\beta\Big(
\log\frac{\pi(\mathbf x_w|\mathbf c)}{\pi_{\mathrm{ref}}(\mathbf x_w|\mathbf c)}
-
\log\frac{\pi(\mathbf x_l|\mathbf c)}{\pi_{\mathrm{ref}}(\mathbf x_l|\mathbf c)}
\Big).
\]
This cancellation is exactly what makes the constant irrelevant for preference learning and leads directly to the DPO loss on log-probability differences.

\paragraph{DPO's Training Loss.}
Plug the implicit reward \Cref{eq:implicit-reward} into the BT model of \Cref{eq:rm-bt} for a labeled pair $(\mathbf x_w,\mathbf x_l)$ under the same prompt $\mathbf c$. 
The constants $\log Z(\mathbf c)$ cancel between winner and loser, yielding a single logistic-loss objective on log-probability differences:
\begin{align*}
\mathcal L_{\mathrm{DPO}}(\bm\theta;\pi_{\mathrm{ref}})
&=
-\mathbb E_{(\mathbf c,\mathbf x_w,\mathbf x_l)\sim\mathcal D} \left[
\log \sigma \Big(
\beta\big(
\log\frac{\pi_{\bm\theta}(\mathbf x_w|\mathbf c)}{\pi_{\mathrm{ref}}(\mathbf x_w|\mathbf c)}
-
\log\frac{\pi_{\bm\theta}(\mathbf x_l|\mathbf c)}{\pi_{\mathrm{ref}}(\mathbf x_l|\mathbf c)}
\big)
\Big)
\right].
\end{align*}
In words: DPO pushes up the (temperature-scaled) advantage of the winner over the loser, measured as the difference of log-likelihood improvements over the reference:
\[
-\log \sigma \left(
  \beta 
  \bigl[
    \text{log-ratio difference of }
    \tfrac{\pi_{\bm\theta}}{\pi_{\mathrm{ref}}}
    \text{ at } \mathbf x_w \text{ vs.\ } \mathbf x_l
  \bigr]
\right).
\]
This achieves the goal of RLHF in a single, stable classification-style
fine-tuning stage, without training an explicit reward model.

\subsection{Diffusion-DPO}\label{subsec:diffusion-dpo}
Throughout this subsection, we retain the policy notation from
\Cref{subsec:dpo}: $\pi_{\bm\theta}$ and $\pi_{\mathrm{ref}}$ denote the
learned and reference diffusion policies, including their path and transition
distributions, while $p_t(\rvx_t|\rvx_0)$ continues to denote the fixed
forward noising process.
\paragraph{Why Naive DPO Fails for Diffusion Models?}
Evaluating the sample likelihood $\pi_\btheta(\mathbf x|\mathbf c)$ in diffusion models requires the instantaneous change-of-variables formula (divergence of the drift) of ODE solving (see \Cref{eq:score-sde-likelihood})\footnote{In discrete-time diffusion models (e.g., DDPM), evaluating $\pi_{\bm\theta}(\mathbf x_0| \mathbf c)$ requires marginalizing over the latent reverse trajectory $\mathbf x_{1:T}$.
}, which is computationally intensive. Moreover, differentiating through the entire sampling trajectory can suffer from vanishing or exploding gradients. To avoid these issues, Diffusion-DPO works at the \emph{path} level. We take the discrete-time diffusion model (e.g., DDPM) as an illustrative example; the continuous-time diffusion model is analogous.

\paragraph{Defining Pathwise Implicit Rewards.}
Let a trajectory be
$\mathbf x_{0:T}:=(\mathbf x_T,\ldots,\mathbf x_0)$ under the reverse-time
Markov chain with conditionals
$\pi(\mathbf x_{t-1}|\mathbf x_t,\mathbf c)$. Here, $\mathbf x_T$ is the
highest-noise state and $\mathbf x_0$ is the clean output.

Human preferences are defined on the final output through
$r(\mathbf c,\mathbf x_0)$. Since $\mathbf x_0$ is part of every diffusion
trajectory, the reward itself does not need to change when we move to path
space. The useful step is instead to regularize the entire reverse process.
By the chain rule for KL divergence, the KL divergence between the full path
distributions is at least as large as the KL divergence between their endpoint
marginals. This motivates the pathwise objective
\[
\max_{\btheta}\mathbb E_{\mathbf c\sim p(\mathbf c)}
\Big[
\mathbb E_{\mathbf x_{0:T}\sim\pi_{\btheta}(\cdot|\mathbf c)}
\big[r(\mathbf c,\mathbf x_0)\big]
-\beta
\mathcal D_{\mathrm{KL}}
\big(
\pi_{\btheta}(\cdot|\mathbf c)
\big\|
\pi_{\mathrm{ref}}(\cdot|\mathbf c)
\big)
\Big],
\]
where $\pi_{\btheta}(\cdot|\mathbf c)$ and
$\pi_{\mathrm{ref}}(\cdot|\mathbf c)$ denote the corresponding path
distributions. The reward term is unchanged, while the KL penalty now also
controls how the latent denoising trajectories deviate from the reference.

At the distributional level, assuming the normalizer is finite, the optimizer
for each prompt $\mathbf c$ has the energy-based form
\begin{equation}\label{eq:path-closed-form}
\pi^*(\mathbf x_{0:T}|\mathbf c)
=
\frac{1}{Z(\mathbf c)}
\pi_{\mathrm{ref}}(\mathbf x_{0:T}|\mathbf c)
\exp\big(r(\mathbf c,\mathbf x_0)/\beta\big).
\end{equation}
As in ordinary DPO, we can invert this relation and define the
\emph{implicit path reward} induced by any policy $\pi$:
\[
R_{\pi}(\mathbf c,\mathbf x_{0:T})
:=
\beta\log
\frac{\pi(\mathbf x_{0:T}|\mathbf c)}
{\pi_{\mathrm{ref}}(\mathbf x_{0:T}|\mathbf c)}
+\beta\log Z(\mathbf c).
\]
At the pathwise optimum,
$R_{\pi^*}(\mathbf c,\mathbf x_{0:T})=r(\mathbf c,\mathbf x_0)$.
For a general policy, however, $R_{\pi}$ may also depend on the latent
trajectory. Since the preference data compare only the clean endpoints, we
therefore average this implicit path reward over trajectories associated with
each endpoint.

\paragraph{From Pathwise Implicit Rewards to DPO.}
For a labeled endpoint pair $(\mathbf x_0^w,\mathbf x_0^l)$ under the same
prompt $\mathbf c$, the Bradley--Terry margin is obtained by comparing the
conditional averages of their implicit path rewards:
\begin{align}\label{eq:diffusion-dpo}
\begin{aligned}
\mathcal L_{\mathrm{Diff\text{-}DPO}}(\bm\theta;\pi_{\mathrm{ref}})
&:=
- \mathbb E_{(\mathbf c,\mathbf x_0^w,\mathbf x_0^l)\sim\mathcal D}
 \left[\log \sigma \big(\Delta R(\mathbf c;\bm\theta)\big)\right], \quad \text{where}\\
\Delta R(\mathbf c;\bm\theta)
&:=
\underbrace{\mathbb E_{\mathbf x_{1:T}^w\sim \pi_{\bm\theta}(\cdot | \mathbf x_0^w,\mathbf c)}
 \Big[R_{\pi_{\bm\theta}} \big(\mathbf c,(\mathbf x_0^w,\mathbf x_{1:T}^w)\big)\Big]}_{\text{winner path expectation}}
\\ &\quad\quad\quad-
\underbrace{\mathbb E_{\mathbf x_{1:T}^l\sim \pi_{\bm\theta}(\cdot | \mathbf x_0^l,\mathbf c)}
 \Big[R_{\pi_{\bm\theta}} \big(\mathbf c,(\mathbf x_0^l,\mathbf x_{1:T}^l)\big)\Big]}_{\text{loser path expectation}}. 
\end{aligned}
\end{align}
Here, the expectation $\mathbb E_{\mathbf x_{1:T}\sim\pi_{\bm\theta}(\cdot|\mathbf x_0,\mathbf c)}[\cdot]$ means:
given a fixed endpoint $\mathbf x_0$ (e.g., the winner $\mathbf x_0^w$) from the dataset, we take an
expectation over latent denoising trajectories $\mathbf x_{1:T}$ under the model-induced conditional path
distribution (the posterior over reverse-time trajectories) that, with kernels
$\pi_{\bm\theta}(\mathbf x_{t-1}|\mathbf x_t,\mathbf c)$, could produce $\mathbf x_0$.
Since these intermediate states are unobserved, we average the path reward over all such trajectories.

However, \Cref{eq:diffusion-dpo} is impractical for three practical reasons:
\begin{enumerate}[leftmargin=*]
\item \textbfs{Endpoint Conditioning Induces an Intractable Path Posterior.}
The term $\mathbb E_{\pi_{\bm\theta}(\mathbf x_{1:T}|\mathbf x_0,\mathbf c)}[\cdot]$ averages over reverse paths constrained to hit $\mathbf x_0$, whereas the sampler runs $\mathbf x_T \to \cdots \to \mathbf x_0$ without this constraint. Conditioning on the endpoint creates a diffusion-bridge posterior with generally no closed form and costly sampling.

\item \textbfs{Nested, $\bm\theta$-Coupled Expectations.}
The loss $-\log\sigma(\Delta R(\mathbf c;\bm\theta))$ with
\[
\Delta R=\mathbb E_{\text{paths}|\mathbf x_0^w,\mathbf c}[R_{\pi_{\bm\theta}}]
-\mathbb E_{\text{paths}|\mathbf x_0^l,\mathbf c}[R_{\pi_{\bm\theta}}]
\]
has both the path joint distribution and the integrand $R_{\pi_{\bm\theta}}$ depending on $\bm\theta$. Thus $\nabla_{\bm\theta}$ must differentiate through the sampling distribution, leading to REINFORCE/pathwise couplings and high-variance gradients.

\item \textbfs{Long Chains, Large Sums, and Expensive Backpropagation.}
In $R_{\pi_{\bm\theta}}(\mathbf c,\mathbf x_{0:T})$, computing
\[
\beta \left[\log \pi_{\bm\theta}(\mathbf x_{0:T}|\mathbf c)
-\log \pi_{\mathrm{ref}}(\mathbf x_{0:T}|\mathbf c)\right]
\]
requires $\mathcal O(T)$ per-step log-densities with $T \sim 10^2$–$10^3$, for both policy $\pi_{\bm\theta}$ and reference $\pi_{\mathrm{ref}}$, and for both winner/loser paths. Backpropagating through these stochastic chains (or bridge samplers) is memory and compute heavy and can be unstable; repeating this over many samples per pair and across all triplets pushes training beyond practical budgets.
\end{enumerate}

\paragraph{Toward a Tractable Surrogate for \Cref{eq:diffusion-dpo}.}
To make this computable, we apply a key mathematical insight. By leveraging properties of diffusion models and applying Jensen's inequality, we can optimize a tractable upper bound on this loss. This transforms the problem from evaluating an entire path's likelihood to evaluating an expectation over the individual, single-step transitions within the path:

Because $-\log\sigma(\cdot)$ is convex, Jensen’s inequality yields an upper bound by moving the inner expectations outside the nonlinear function:
\begin{align*}
\mathcal L_{\mathrm{Diff\text{-}DPO}}(\bm\theta;\pi_{\mathrm{ref}})
\le  
-\mathbb E_{(\mathbf c,\mathbf x_0^w,\mathbf x_0^l)\sim\mathcal D}
  \mathbb E_{\substack{\mathbf x^w_{1:T}\sim \pi_{\bm\theta}(\cdot | \mathbf x_0^w,\mathbf c)\\[1pt]
                       \mathbf x^l_{1:T}\sim \pi_{\bm\theta}(\cdot | \mathbf x_0^l,\mathbf c)}}
\left[
\log \sigma \Big(
R_{\pi_{\bm\theta}}(\mathbf c,\mathbf x^w_{0:T})
-
R_{\pi_{\bm\theta}}(\mathbf c,\mathbf x^l_{0:T})
\Big)
\right].
\end{align*}
Using the implicit-reward identity $R_{\pi_{\bm\theta}}=\beta\log\frac{\pi_{\bm\theta}}{\pi_{\mathrm{ref}}}+\beta\log Z(\mathbf c)$ and cancellation of the constant between winner and loser, the bound becomes
\begin{align}\label{eq:dpo-path-bound}
\mathcal L_{\mathrm{Diff\text{-}DPO}}(\bm\theta;\pi_{\mathrm{ref}})
& \le
-\mathbb E
\left[
\log \sigma \Big(
\beta\big(
\log\frac{\pi_{\bm\theta}(\mathbf x^w_{0:T}| \mathbf c)}{\pi_{\mathrm{ref}}(\mathbf x^w_{0:T}| \mathbf c)}
-
\log\frac{\pi_{\bm\theta}(\mathbf x^l_{0:T}| \mathbf c)}{\pi_{\mathrm{ref}}(\mathbf x^l_{0:T}| \mathbf c)}
\big)
\Big)
\right].
\end{align}

\subparagraph{A Tractable Surrogate (Stepwise Form).} We now exploit the Markov property of the reverse process to decompose the upper bound of $\mathcal L_{\mathrm{Diff\text{-}DPO}}$. This allows us to express the path-level preference as a sum of per-step contributions, converting the intractable pathwise loss into a tractable single-step estimator. The resulting form yields a DPO-style log-sigmoid loss whose inner margin reduces to a DSM-style MSE difference. Concretely, for the reverse chain,
\begin{align*}
    \pi_{\bm\theta}(\mathbf x_{0:T}| \mathbf c)
&=\pi_{\bm\theta}(\mathbf x_T| \mathbf c) \prod_{t=1}^{T}
\pi_{\bm\theta}(\mathbf x_{t-1}| \mathbf x_t,\mathbf c),
\\
\pi_{\mathrm{ref}}(\mathbf x_{0:T}| \mathbf c)
&=\pi_{\mathrm{ref}}(\mathbf x_T| \mathbf c) \prod_{t=1}^{T}
\pi_{\mathrm{ref}}(\mathbf x_{t-1}| \mathbf x_t,\mathbf c).
\end{align*}
Hence
\[
\frac{\pi_{\bm\theta}(\mathbf x_{0:T}| \mathbf c)}{\pi_{\mathrm{ref}}(\mathbf x_{0:T}| \mathbf c)}
=
\frac{\pi_{\bm\theta}(\mathbf x_T| \mathbf c)}{\pi_{\mathrm{ref}}(\mathbf x_T| \mathbf c)}
\prod_{t=1}^{T}
\frac{\pi_{\bm\theta}(\mathbf x_{t-1}| \mathbf x_t,\mathbf c)}
     {\pi_{\mathrm{ref}}(\mathbf x_{t-1}| \mathbf x_t,\mathbf c)}.
\]
If the prior at time $T$ is the same for both models,
$\pi_{\bm\theta}(\mathbf x_T| \mathbf c)=\pi_{\mathrm{ref}}(\mathbf x_T| \mathbf c)$,
then the first factor equals $1$, and taking logs yields
\[
\log\frac{\pi_{\bm\theta}(\mathbf x_{0:T}| \mathbf c)}{\pi_{\mathrm{ref}}(\mathbf x_{0:T}| \mathbf c)}
=\sum_{t=1}^{T}
\log\frac{\pi_{\bm\theta}(\mathbf x_{t-1}| \mathbf x_t,\mathbf c)}
         {\pi_{\mathrm{ref}}(\mathbf x_{t-1}| \mathbf x_t,\mathbf c)}.
\]
It follows that the bound in \Cref{eq:dpo-path-bound} can be written as
\[
\mathcal L_{\mathrm{Diff\text{-}DPO}}(\bm\theta;\pi_{\mathrm{ref}})
  \le  - \E \left[\log\sigma \Big(\beta \sum_{t=1}^T \Delta_t\Big)\right],
\]
where each per-step contribution is
\[
\Delta_t
= \log\frac{\pi_{\bm\theta}(\mathbf x^w_{t-1}|\mathbf x^w_t,\mathbf c)}
               {\pi_{\mathrm{ref}}(\mathbf x^w_{t-1}|\mathbf x^w_t,\mathbf c)}
- \log\frac{\pi_{\bm\theta}(\mathbf x^l_{t-1}|\mathbf x^l_t,\mathbf c)}
               {\pi_{\mathrm{ref}}(\mathbf x^l_{t-1}|\mathbf x^l_t,\mathbf c)}.
\]

To reduce the sum over timesteps to a single randomly selected transition,
observe that
$\beta\sum_{s=1}^T\Delta_s
=
\E_{t\sim\mathcal U\{1,\dots,T\}}[\beta T\Delta_t]$.
Since $-\log\sigma(\cdot)$ is convex, Jensen's inequality therefore gives,
after taking the existing outer expectations,
\[
\mathcal L_{\mathrm{Diff\text{-}DPO}}(\bm\theta;\pi_{\mathrm{ref}})
\leq
-\E\left[
\log\sigma\bigl(\beta T\Delta_t\bigr)
\right],
\]
where the expectation is over
$(\mathbf c,\mathbf x_0^w,\mathbf x_0^l)\sim\mathcal D$,
$t\sim\mathcal U\{1,\dots,T\}$, and
$\mathbf x_{t-1:t}^w\sim
\pi_{\bm\theta}(\cdot|\mathbf x_0^w,\mathbf c)$ and
$\mathbf x_{t-1:t}^l\sim
\pi_{\bm\theta}(\cdot|\mathbf x_0^l,\mathbf c)$.
Although the bound now depends on only one transition, sampling these
endpoint-conditioned transition pairs remains intractable.

Following Diffusion-DPO, we approximate the intractable model-induced
conditional path using the fixed forward noising process. For each clean
endpoint $\mathbf x_0$, we first sample $\mathbf x_t
\sim
p_t(\mathbf x_t|\mathbf x_0)$.
Conditioned on $(\mathbf x_0,\mathbf x_t)$, the forward process induces the
tractable posterior $p(\mathbf x_{t-1}|\mathbf x_t,\mathbf x_0)$.
Averaging the reverse-transition log-ratio under this posterior gives
\begin{align*}
&\E_{p(\mathbf x_{t-1}|\mathbf x_t,\mathbf x_0)}
\left[
\log
\frac{
\pi_{\bm\theta}(\mathbf x_{t-1}|\mathbf x_t,\mathbf c)
}{
\pi_{\mathrm{ref}}(\mathbf x_{t-1}|\mathbf x_t,\mathbf c)
}
\right]
\nonumber\\&=
-
\mathcal D_{\mathrm{KL}}
\left(
p(\mathbf x_{t-1}|\mathbf x_t,\mathbf x_0)
\,\middle\|\,
\pi_{\bm\theta}(\mathbf x_{t-1}|\mathbf x_t,\mathbf c)
\right)+
\mathcal D_{\mathrm{KL}}
\left(
p(\mathbf x_{t-1}|\mathbf x_t,\mathbf x_0)
\,\middle\|\,
\pi_{\mathrm{ref}}(\mathbf x_{t-1}|\mathbf x_t,\mathbf c)
\right).
\end{align*}

For Gaussian reverse kernels with the same prescribed reverse variance, these
KL divergences reduce to weighted denoising regression errors. Under
$\boldsymbol\epsilon$-prediction,
\begin{align}
\E_{p(\mathbf x_{t-1}|\mathbf x_t,\mathbf x_0)}
\left[
\log
\frac{
\pi_{\bm\theta}(\mathbf x_{t-1}|\mathbf x_t,\mathbf c)
}{
\pi_{\mathrm{ref}}(\mathbf x_{t-1}|\mathbf x_t,\mathbf c)
}
\right]
=
-\lambda_t
\left[
\left\|
\hat{\boldsymbol\epsilon}_{\bm\theta}(\mathbf x_t,t,\mathbf c)
-\boldsymbol\epsilon
\right\|^2
-
\left\|
\hat{\boldsymbol\epsilon}_{\mathrm{ref}}(\mathbf x_t,t,\mathbf c)
-\boldsymbol\epsilon
\right\|^2
\right],
\end{align}
where $\mathbf x_t
=
\alpha_t\mathbf x_0
+
\sigma_t\boldsymbol\epsilon$ with $\boldsymbol\epsilon\sim\mathcal N(\mathbf 0,\mathbf I)$, and $\lambda_t>0$ collects the schedule-dependent weighting factors.

For notation simplicity, define for any noised sample $(\mathbf x_t,\boldsymbol\epsilon)$:
\[
\Delta\mathrm{MSE}(\mathbf x_t;\boldsymbol\epsilon)
:= \big\|\hat{\boldsymbol\epsilon}_{\bm\theta}(\mathbf x_t,t,\mathbf c)-\boldsymbol\epsilon\big\|^2
 - \big\|\hat{\boldsymbol\epsilon}_{\mathrm{ref}}(\mathbf x_t,t,\mathbf c)-\boldsymbol\epsilon\big\|^2.
\]
This yields the practical single-step Diffusion-DPO surrogate
 for $\mathcal L_{\mathrm{Diff\text{-}DPO}}$:
\begin{mdframed}
\begin{align*}
    &\tilde{\mathcal L}_{\mathrm{Diff\text{-}DPO}}(\bm\theta;\pi_{\mathrm{ref}})
    :=
    \\&-\mathbb E_{\substack{(\mathbf c,\mathbf x_0^w,\mathbf x_0^l)\sim\mathcal D\\
    t\sim\mathcal U\{1,\dots,T\}, \boldsymbol\epsilon^w,\boldsymbol\epsilon^l\sim\mathcal N(\mathbf 0,\mathbf I)}}
    \Big[
    \log\sigma \Big(
    \beta T\lambda_t\big(\Delta\mathrm{MSE}(\mathbf x_t^l;\boldsymbol\epsilon^l)-\Delta\mathrm{MSE}(\mathbf x_t^w;\boldsymbol\epsilon^w)\big)
    \Big)
    \Big],
\end{align*}
\end{mdframed}
where
$\mathbf x_t^w=\alpha_t\mathbf x_0^w+\sigma_t\boldsymbol\epsilon^w$
and
$\mathbf x_t^l=\alpha_t\mathbf x_0^l+\sigma_t\boldsymbol\epsilon^l$.
In practice, one may also use shared noise
$\boldsymbol\epsilon^w=\boldsymbol\epsilon^l$
as a variance-reduction variant; this does not change the overall DPO-style form
of the objective, but slightly alters the stochastic estimator.

Intuitively, minimizing $\tilde{\mathcal L}_{\mathrm{Diff\text{-}DPO}}$ increases the reference-normalized winner-over-loser margin. Equivalently, it encourages the model to reduce denoising error on the winner more than on the loser, measured relative to the frozen reference model. The reference model does not contribute a direct gradient term, but it remains inside the margin within the $\log\sigma(\cdot)$ nonlinearity, so it still shapes the DPO-style per-example weighting.

\newpage
\section{Closing Remarks}\label{sec:ch8_cr}

This chapter has shifted our focus from foundational principles to the practical challenge of controllable generation. We established a unified framework for guidance based on the Bayesian decomposition of the conditional score, which elegantly separates the generative process into an unconditional direction and a steering term.

We saw this principle manifest in several powerful techniques. We covered methods that require dedicated training, such as Classifier Guidance (CG), which uses an external classifier, and the more efficient Classifier-Free Guidance (CFG), which learns conditional and unconditional scores within a single model. We also explored flexible training-free guidance methods, which can steer a pre-trained model at inference time by defining a surrogate guidance potential from a task-specific loss, enabling applications from artistic control to solving inverse problems without any retraining.

Beyond simple conditioning, we delved into the nuanced task of aligning model outputs with human preferences. After reviewing the standard but complex RLHF pipeline, we introduced Direct Preference Optimization (DPO) and its novel adaptation, Diffusion-DPO, as a more direct and stable alternative. This approach elegantly bypasses the need for an explicit reward model and reinforcement learning by deriving a loss directly from preference data.

Through these techniques, we have assembled a powerful toolkit for steering the generative process. However, a major practical hurdle remains untouched: the significant computational cost and latency of the iterative sampling process itself. Having addressed what to generate, we now turn to the equally important question of how fast we can generate it. The next chapter will tackle this challenge directly:
\begin{enumerate}
    \item We will leverage the insight that sampling is equivalent to solving an ODE to explore sophisticated numerical solvers designed to drastically reduce the number of required steps.
    \item We will investigate a sequence of influential methods, including DDIM, DEIS, and the DPM-Solver family, which have made diffusion models far more practical by accelerating sampling speed by orders of magnitude.
\end{enumerate}

\chapter{Numerical Solvers for Fast Diffusion Sampling}\label{ch:solvers}

The generation process of a diffusion model, which maps noise to data samples, can be formulated through a reverse-time SDE or its
associated PF-ODE. This procedure is inherently slow, since it relies on numerical solvers that approximate solution trajectories with many small integration steps (see \Cref{app:de} for a brief introduction). Accelerating inference has therefore become a central research objective. Broadly, existing approaches fall into two categories:
\begin{itemize}
    \item \textbfs{Training-Free Approaches:} The focus of this chapter. These methods develop advanced numerical solvers to improve the efficiency of diffusion sampling without additional training.
    \item \textbfs{Training-Based Approaches:} Covered in \Cref{ch:distillation,ch:fast-scratch}. These techniques either distill a pre-trained diffusion model into a fast generator, or directly learn the ODE flow map (solution) so that only a few sampling steps are required.
\end{itemize}
SDE-based samplers (e.g., Euler--Maruyama) may yield more diverse samples due to stochasticity but typically require more steps~\citep{xu2023restart}. Here we focus on ODE-based generation, whose principles extend naturally to the SDE setting.

\chapterlearning{

  \item view diffusion sampling as a numerical integration problem for the PF-ODE;

  \item distinguish model approximation error from numerical discretization error;

  \item understand the main numerical ideas behind DDIM, DEIS, DPM-Solver, and DPM-Solver\texttt{++};

  \item connect modern diffusion samplers to classical numerical methods, including Euler, Runge-Kutta (RK), Adams--Bashforth (AB), and exponential integrators.

}{Readers unfamiliar with differential equations or numerical solvers may first refer to \Cref{app:de}. Throughout this chapter, keep one question in mind: what does each solver integrate exactly, and what does it approximate numerically?}

\clearpage

\section{Prologue}\label{sec:prologue-solvers}
\subsection{Fast Sampling as Numerical Integration}

The Score SDE framework~\citep{song2020score} established a key foundation by
linking discrete-time diffusion models and ELBO-based formulations
\citep{sohl2015deep,ho2020denoising} with continuous-time SDE/ODE
generative modeling. This connection gives a clean way to think about sampling:
after a diffusion model has been trained, generation can be viewed as solving a
learned differential equation from noise back to data.

The practical difficulty is speed. A naive sampler follows this reverse-time
trajectory with many small numerical steps, often requiring hundreds or
thousands of neural network evaluations. Fast sampling asks whether we can
approximate the same generation process with far fewer model calls.

This chapter focuses on \emph{training-free} acceleration: the pre-trained
diffusion model is kept fixed, and we design better numerical solvers for
sampling. Training-based acceleration methods, such as distillation or direct
few-step training, are discussed later in
\Cref{ch:distillation,ch:fast-scratch}.

The main message of this chapter is therefore simple:
\msgu{Message}{}{Once a diffusion model has been trained, training-free fast sampling is largely a
numerical integration problem.}

The literature contains many named diffusion samplers, which can make the
subject appear more fragmented than it really is. In this chapter, we first
introduce the small number of numerical ideas underlying these methods.
Only afterward do we attach the familiar method names to them. This
organization makes clear that the samplers are not unrelated tricks: they
are different ways of approximating the same PF-ODE trajectory.

\paragraph{Sampling as Solving the Empirical PF-ODE.}
Suppose we have a pre-trained diffusion model
\[
    \rvs_{\bm{\phi}^\times}(\rvx,t)
    \approx
    \nabla_{\rvx}\log p_t(\rvx),
\]
or equivalently a model written in one of the other standard parameterizations
from \Cref{sec:equivalent-parametrizations}. The empirical PF-ODE uses this
learned model in place of the true score:
\[
    \frac{\diff \rvx(t)}{\diff t}
    =
    \rvf(\rvx(t),t)
    -
    \frac{1}{2}g^2(t)\,
    \rvs_{\bm{\phi}^\times}(\rvx(t),t).
\]
Starting from $\rvx(T)\sim p_{\mathrm{prior}}$, generation integrates this
ODE backward from $t=T$ to $t=0$.

For two times $s>t$, the exact empirical update can be written in integral
form as
\begin{align}
\label{eq:solution-int-score}
\begin{aligned}
    \widetilde\bPsi_{s\to t}(\rvx_s)
    =
    \rvx_s
    +
    \int_s^t
    \left[
        \rvf(\rvx(\tau),\tau)
        -
        \frac{1}{2}g^2(\tau)
        \rvs_{\bm{\phi}^\times}(\rvx(\tau),\tau)
    \right]\diff \tau .
\end{aligned}
\end{align}
Here, $\widetilde{\bPsi}_{s\to t}$ denotes the flow map of the empirical
PF-ODE, obtained by replacing the oracle score with the learned model.
As throughout this book, a \emph{flow map} refers to the classical ODE
solution map introduced in \Cref{eq:flow-map-def}. We denote the
corresponding oracle flow map by $\bPsi_{s\to t}$. The empirical flow
map is intended to approximate the oracle map:
\[
    \widetilde{\bPsi}_{s\to t}(\rvx_s)
    \approx
    \bPsi_{s\to t}(\rvx_s).
\]
Because the integral in \Cref{eq:solution-int-score} cannot generally
be evaluated in closed form, a numerical sampler discretizes the
reverse-time interval using a decreasing grid
\begin{align}
\label{eq:time-step-def}
    T=t_0>t_1>\cdots>t_M=0,
\end{align}
starts from
\[
    \tilde{\rvx}_{t_0}=\rvx_T\sim p_{\mathrm{prior}},
\]
and produces a sequence
\[
    \tilde{\rvx}_{t_0},\tilde{\rvx}_{t_1},\ldots,\tilde{\rvx}_{t_M},
\]
where $\tilde{\rvx}_{t_i}$ approximates
$\widetilde\bPsi_{T\to t_i}(\rvx_T)$. The final iterate
$\tilde{\rvx}_{t_M}$ is used as the generated sample.

Thus, every training-free solver in this chapter answers the same question:
\begin{question}
How should we approximate the PF-ODE integral accurately using as few
model evaluations as possible?
\end{question}

\paragraph{Three Independent Choices in Solver Design.}
Once the decreasing time grid in \Cref{eq:time-step-def} has been introduced,
it is useful to distinguish three choices that are often intertwined in
diffusion sampling.

\begin{enumerate}
    \item \textbfs{Prediction Parameterization.}
    Which quantity is provided by the pre-trained diffusion model:
    score $\rvs_{\bm{\phi}^\times}$, noise
    $\beps_{\bm{\phi}^\times}$, clean data
    $\rvx_{\bm{\phi}^\times}$, or velocity
    $\rvv_{\bm{\phi}^\times}$?

    \item \textbfs{Time Parameterization.}
    Which variable is used to parameterize progress along the same continuous
    trajectory? The original diffusion time $t$ is one choice; other
    reparameterizations will be introduced later when they become useful.

    \item \textbfs{Numerical Approximation.}
    How does the solver approximate the unknown model-dependent quantity over
    a finite interval? It may use only the current evaluation, obtain
    additional evaluations within the current step, or reuse evaluations from
    previous steps.
\end{enumerate}
This separation will be important throughout the chapter. The first two
choices determine the numerical form presented to the solver, while the third
determines how the solver uses the available model evaluations.

\paragraph{How Can a Solver Use Model Evaluations?}
A useful way to organize numerical solvers is by asking where the information
for the next update comes from. Starting from the current state, a solver may
either obtain additional information \emph{within the current step} or reuse
information from \emph{previous steps}.

A \emph{single-step method} constructs the next state using only the current
state and evaluations generated within the same time interval. Euler's method
is the simplest example, using only the current velocity. Higher-order
Runge-Kutta (RK) methods may instead evaluate the velocity at one or more
intermediate states before completing the step. Thus, ``single-step'' does not
mean ``one function evaluation''; it means that no information from earlier
completed steps is required.

A \emph{multistep method} reuses information from previous completed steps.
Stored model evaluations provide information about how the dynamics have
recently changed, allowing higher-order updates without several new evaluations
within the current step. The Adams--Bashforth (AB) family is a representative
classical example.

A separate idea is to change how updates across time are scheduled. Rather
than computing all time points strictly sequentially, one may initialize states
at several times and repeatedly correct them until they become consistent with
the ODE. Such fixed-point iterations can allow model evaluations at different
time points to be carried out in parallel.

Representative classical examples, including Euler, RK,
AB, and Picard iteration, are reviewed in
\Cref{app:de-solver-examples}; their structural organization is summarized in
\Cref{eq:classic-solvers}.

\paragraph{Computational Cost and Practical Performance.}
A numerical time step need not correspond to a single neural-network
evaluation. Euler uses one evaluation per step, whereas higher-order
RK methods may require several. Multistep methods instead reuse
stored evaluations and, after initialization, can often advance using only one
new evaluation per step.

We therefore measure sampling cost by the \emph{number of function
evaluations} (NFE). If a method uses approximately $m$ model evaluations per
step over $M$ steps, then 
\[
\mathrm{NFE}\approx mM.
\]
For instance, classifier-free guidance (see \Cref{sec:cfg}) requires both
conditional and unconditional model predictions and therefore roughly doubles
the neural-network evaluation cost per sampling step.

This classical taxonomy tells us \emph{where numerical information comes from}
and how many model evaluations it costs. Diffusion PF-ODEs offer an additional
opportunity: not every part of the vector field is equally unknown. Depending
on the prediction parameterization, part of the dynamics is determined
analytically by the noise schedule, while only the remaining quantity requires
a neural-network evaluation. We next examine how exposing this separation
changes the numerical form of the PF-ODE and motivates diffusion-specific
integrators.

\subsection{Design of Diffusion-Specific Solvers}
\label{subsec:common_design_solvers}

The classical numerical methods reviewed in \Cref{app:de-solver} treat a
general ODE through evaluations of its vector field. Diffusion PF-ODEs offer
additional structure: the noise schedule is known analytically, whereas the
diffusion-model prediction is supplied by a neural network.

This creates a useful opportunity. Rather than numerically approximating
every part of the dynamics in the same way, a solver may retain the known
schedule-dependent terms analytically and approximate only the
model-dependent prediction. How clearly this structure is exposed depends on
the prediction parameterization.

\paragraph{I. Parameterizations Beyond the Score.}
Although the PF-ODE can be written in score form, modern implementations often
use noise prediction, clean-data prediction, or velocity prediction. These
parameterizations are algebraically related, but they expose different numerical
structures.

There is also an important stability reason to avoid direct score prediction in
many solvers. As $t\to 0$, the true score
$\nabla_{\rvx}\log p_t(\cdot)$ can change very rapidly, especially when the
data distribution is concentrated near a low-dimensional manifold
\citep{kim2022soft}. This makes it difficult for a neural network trained to
approximate the score directly to remain accurate near small noise levels~\footnote{To see the scaling, recall the oracle relation from
\Cref{eq:predictions-equivalence}:
\[
    \beps^*(\rvx_t,t)
    =
    -\sigma_t\nabla_{\rvx}\log p_t(\rvx_t),
\]
where $\beps^*(\rvx_t,t)
    =
    \mathbb{E}[\beps|\rvx_t]$ is the oracle noise predictor, and the perturbation kernel is $ \rvx_t|\rvx_0
    \sim
    \mathcal{N}(\rvx_t; \alpha_t\rvx_0,\sigma_t^2\rmI)$. By the orthogonality property of conditional expectation in $L^2$,
\[
    \mathbb{E}\|\beps\|_2^2
    =
    \mathbb{E}\|\beps^*\|_2^2
    +
    \mathbb{E}\|\beps-\beps^*\|_2^2,
\]
and therefore
\[
    \mathbb{E}\|\beps^*\|_2^2
    \le
    \mathbb{E}\|\beps\|_2^2
    =
    D,
\]
where $D$ is the data dimension. Thus, the oracle noise predictor has uniformly bounded expected squared norm. In contrast, the oracle score
\[
    \rvs^*(\rvx_t,t)
    :=
    \nabla_{\rvx}\log p_t(\rvx_t)
\]
satisfies
\[
    \mathbb{E}\|\rvs^*(\rvx_t,t)\|_2^2
    =
    \sigma_t^{-2}\,
    \mathbb{E}\|\beps^*(\rvx_t,t)\|_2^2
    \le
    \frac{D}{\sigma_t^2}.
\]
Thus, direct score prediction can become increasingly poorly scaled as
$\sigma_t$ becomes small, since noise prediction is converted to the score
through the factor $1/\sigma_t$. This scaling can make direct score
parameterization numerically less favorable near small noise levels.}.

For this reason, a widely used alternative is to predict the noise
$\beps_{\bm{\phi}^\times}$, or related targets such as clean-data prediction
or velocity prediction. In the noise-prediction form, the score approximation is
\[
    \rvs_{\bm{\phi}^\times}(\rvx,t)
    =
    -\frac{1}{\sigma_t}\,
    \beps_{\bm{\phi}^\times}(\rvx,t).
\]
These prediction parameterizations are algebraically related, but the
quantity supplied by the neural network is only one component of the solver
design. We next examine how the same PF-ODE appears numerically under these
different parameterizations.

\paragraph{II. One PF-ODE, Two Numerical Forms.}
Once the prediction parameterization has been chosen, the same PF-ODE can be
presented to a numerical solver in two particularly useful forms.

\subparagraph{Direct Velocity Form.}
The most direct representation is
\begin{equation}
\label{eq:direct-pfode-form}
\frac{\diff \rvx(t)}{\diff t}
=
\rvv_{\bm{\phi}^\times}(\rvx(t),t).
\end{equation}
Here, under $\rvv$-prediction, the model directly supplies the complete
empirical PF-ODE velocity $\rvv_{\bm{\phi}^\times}(\rvx(t),t)$. Numerical
integration methods therefore discretize the full right-hand side as a whole.

\subparagraph{Factored Semilinear Form.}
Score, noise, and clean-data prediction naturally expose additional
schedule-dependent structure. In each case, the PF-ODE can be written in the
generic \emph{semilinear} form
\begin{equation}
\label{eq:general-semilinear-pfode}
\frac{\diff\rvx(t)}{\diff t}
=
\underbrace{
a(t)\rvx(t)
}_{\substack{\text{known}\\\text{linear part}}}
+
\underbrace{
b(t)
}_{\substack{\text{known}\\\text{coefficient}}}
\underbrace{
\rmN_{\bm{\phi}^\times}(\rvx(t),t)
}_{\substack{\text{neural}\\\text{evaluation}}},
\quad
\rmN_{\bm{\phi}^\times}
\in
\left\{
\rvs_{\bm{\phi}^\times},
\beps_{\bm{\phi}^\times},
\rvx_{\bm{\phi}^\times}
\right\}.
\end{equation}
Here, $a(t)$ and $b(t)$ are known functions determined by the noise schedule,
whereas $\rmN_{\bm{\phi}^\times}$ denotes the neural prediction\footnote{In
standard semilinear notation, $b(t)\rmN_{\bm{\phi}^\times}$ could be absorbed
into a single nonlinear term. We keep the two factors separate because $b(t)$
is known analytically from the noise schedule, whereas
$\rmN_{\bm{\phi}^\times}$ requires a neural-network evaluation. This allows a
solver to approximate the neural prediction inside the integral while retaining
and analytically integrating the known schedule-dependent coefficient.}. The term
\emph{semilinear} refers to this decomposition: the first term is linear in the
state $\rvx(t)$, whereas the neural term may depend nonlinearly on $\rvx(t)$.
This separation is not unique, but it is particularly useful numerically because
the known linear part can often be handled analytically, leaving only the neural
contribution to be approximated numerically.

For the three standard semilinear parameterizations\footnote{In the following
sections, we switch between the SDE coefficients $(f(t),g(t))$ and the Gaussian
perturbation kernel $\rvx_t | \rvx_0
    \sim
    \mathcal{N}
    \bigl(
        \rvx_t;
        \alpha_t\rvx_0,
        \sigma_t^2\rmI
    \bigr)$.
They are related by
\[
    f(t)
    =
    \frac{\alpha_t'}{\alpha_t},
    \qquad
    g^2(t)
    =
    \frac{\diff}{\diff t}\bigl(\sigma_t^2\bigr)
    -
    2\frac{\alpha_t'}{\alpha_t}\sigma_t^2
    =
    2\sigma_t\sigma_t'
    -
    2\frac{\alpha_t'}{\alpha_t}\sigma_t^2.
\]
}, the corresponding choices of $a(t)$, $b(t)$, and
$\rmN_{\bm{\phi}^\times}$ are listed in \Cref{tb:pf-ode-parameterizations}. For example, under noise prediction,
\begin{equation}
\label{eq:noise_pf_ode}
\frac{\diff\rvx(t)}{\diff t}
=
\underbrace{
f(t)\rvx(t)
}_{\text{linear part}}
+
\underbrace{
\frac{1}{2}\frac{g^2(t)}{\sigma_t}
\beps_{\bm{\phi}^\times}(\rvx(t),t)
}_{\text{nonlinear neural part}}.
\end{equation}
The coefficient multiplying the neural prediction is known entirely from the
noise schedule; the only learned quantity in the second term is
$\beps_{\bm{\phi}^\times}(\rvx(t),t)$.

\begin{table}[th!]
\caption{Factored PF-ODE forms for score, noise, and clean-data prediction.}
\small
\centering
\renewcommand{\arraystretch}{1.35}
\begin{tabularx}{0.88\linewidth}{
    @{}
    >{\raggedright\arraybackslash}p{0.20\linewidth}
    >{\centering\arraybackslash}p{0.19\linewidth}
    >{\centering\arraybackslash}X
    >{\centering\arraybackslash}p{0.22\linewidth}
    @{}
}
\toprule
\textbfs{Prediction} &
$\boldsymbol{a(t)}$ &
$\boldsymbol{b(t)}$ &
$\boldsymbol{\rmN_{\bm{\phi}^\times}}$ \\
\midrule

$\rvs$-prediction &
$f(t)$ &
$-\frac{1}{2}g^2(t)$ &
$\rvs_{\bm{\phi}^\times}$ \\

\graymidrule

$\beps$-prediction &
$f(t)$ &
$\displaystyle \tfrac{1}{2}\tfrac{g^2(t)}{\sigma_t}$ &
$\beps_{\bm{\phi}^\times}$ \\

\graymidrule

$\rvx$-prediction &
$\displaystyle \tfrac{\sigma_t'}{\sigma_t}$ &
$\displaystyle
\alpha_t
\left(
\tfrac{\alpha_t'}{\alpha_t}
-
\tfrac{\sigma_t'}{\sigma_t}
\right)$ &
$\rvx_{\bm{\phi}^\times}$ \\

\bottomrule
\end{tabularx}
\label{tb:pf-ode-parameterizations}
\end{table}

At the oracle level, score, noise, clean-data, and velocity prediction are
exactly equivalent representations of the same PF-ODE. For an empirical model,
the same equivalence holds when the different prediction parameterizations are
obtained through exact algebraic conversion of the same model output (see
\Cref{eq:predictions-equivalence}). In this case, the same empirical PF-ODE can
be expressed through score, noise, clean-data, or velocity prediction. The
resulting formulas look different because they expose different numerical
structures, but they describe the same learned continuous dynamics. However, if separate networks are trained independently for different
prediction targets, their empirical PF-ODEs need not coincide exactly. In that
case, differences between samplers may reflect both model approximation and
numerical discretization.
\paragraph{III. Exponential Integrators for Semilinear PF-ODEs.}
The factored semilinear form in \Cref{eq:general-semilinear-pfode} separates
the analytically known linear dynamics from the model-dependent prediction.
For $s>t$, define the integrating factor
\[
\mathcal{E}(s\shortrightarrow t)
:=
\exp\left(
\int_s^t a(u)\diff u
\right).
\]
The variation-of-constants formula then gives the exact empirical PF-ODE flow
\begin{mdframed}
\begin{equation}
\label{eq:factored-variation-of-constants}
\widetilde\bPsi_{s\to t}(\rvx_s)
=
\mathcal{E}(s\shortrightarrow t)\rvx_s
+
\int_s^t
\mathcal{E}(\tau\shortrightarrow t)
b(\tau)
\rmN_{\bm{\phi}^\times}(\rvx_\tau,\tau)
\diff\tau .
\end{equation}
\end{mdframed}
We refer the reader to \Cref{app:exp-int-factor} for the general derivation.

For example, under noise prediction,
$a(t)=f(t)$,
$b(t)=\tfrac{1}{2}g^2(t)/\sigma_t$, and
$\rmN_{\bm{\phi}^\times}=\beps_{\bm{\phi}^\times}$, so
\begin{equation}
\label{eq:variation_of_constants}
\widetilde\bPsi_{s\to t}(\rvx_s)
=
\underbrace{
\mathcal{E}(s\shortrightarrow t)\rvx_s
}_{\text{exact linear evolution}}
+
\frac{1}{2}
\int_s^t
\frac{g^2(\tau)}{\sigma_\tau}
\mathcal{E}(\tau\shortrightarrow t)
\beps_{\bm{\phi}^\times}(\rvx_\tau,\tau)
\diff\tau,
\end{equation}
where
$\mathcal{E}(s\shortrightarrow t)
=
\exp\!\left(\int_s^t f(u)\diff u\right)$.

The numerical opportunity is now clear. The integrating factor and the
coefficient $b(\tau)$ are determined by the diffusion schedule; the only
unknown quantity inside the integral is the neural prediction evaluated along
the trajectory. Keeping these known factors explicit therefore allows a
first-order approximation to act only on the neural prediction, instead of
absorbing the coefficient-weighted product
$b(\tau)\rmN_{\bm{\phi}^\times}(\rvx_\tau,\tau)$ into a single unknown
residual.

\paragraph{Why the Factored Exponential-Integrator Form Can Help.}
To see the benefit of this separation, compare it with plain Euler over one
backward step from an exact empirical PF-ODE state $\rvx_s$ at time $s$ to
$t<s$. Define $\Delta s:=s-t>0$ and
$\rmN_s:=\rmN_{\bm{\phi}^\times}(\rvx_s,s)$.

Plain Euler treats the complete PF-ODE velocity as locally constant,
$a(\tau)\rvx_\tau+
b(\tau)\rmN_{\bm{\phi}^\times}(\rvx_\tau,\tau)
\approx a(s)\rvx_s+b(s)\rmN_s$ for $\tau\in[t,s]$, giving
\begin{align}
\label{eq:euler-step}
\tilde\rvx_t^{\mathrm{Euler}}
=
\rvx_s
-
\Delta s
\left[
a(s)\rvx_s+b(s)\rmN_s
\right].
\end{align}
Instead, we may hold only the neural prediction fixed,
$\rmN_{\bm{\phi}^\times}(\rvx_\tau,\tau)\approx\rmN_s$, while retaining the
known schedule dependence in
\Cref{eq:factored-variation-of-constants}. This gives the
\emph{first-order factored exponential-integrator update}\footnote{The update
in \Cref{eq:factored-ei-step} is not plain Euler in the original state and time
variables. After an integrating-factor transformation of the state and, on intervals
where the resulting time coordinate is monotone, a corresponding time
reparameterization, it can be written as an ordinary explicit Euler step. We make this connection concrete for diffusion
sampling in the DDIM section.}
\begin{align}
\label{eq:factored-ei-step}
\tilde\rvx_t^{\mathrm{EI}}
=
\mathcal E(s\shortrightarrow t)\rvx_s
+
\left[
\int_s^t
\mathcal E(\tau\shortrightarrow t)b(\tau)\diff\tau
\right]\rmN_s.
\end{align}
Both methods use one neural-network evaluation per step. Plain Euler freezes
the complete PF-ODE velocity, while the factored update keeps the known
schedule dependence and freezes only the neural prediction. This factorization
matters: if
$b(\tau)\rmN_{\bm{\phi}^\times}(\rvx_\tau,\tau)$
is instead treated as a single residual and held fixed, the result is
conventional exponential Euler. Thus, the factored update retains one more
known quantity, $b(\tau)$, inside the integral; the two updates coincide when
$b(\tau)$ is constant over the step.

A simple frozen-prediction example makes this difference concrete. Suppose
$a(t)\equiv a$, $b(t)\equiv b$, and
$\rmN_{\bm{\phi}^\times}(\rvx_\tau,\tau)\equiv\rmN_s$ over the step. For
$a\neq0$,
\begin{align*}
\tilde\rvx_t^{\mathrm{EI}}
&=
e^{-a\Delta s}\rvx_s
+
\frac{b}{a}
\left(
e^{-a\Delta s}-1
\right)\rmN_s,
\\
\tilde\rvx_t^{\mathrm{Euler}}
&=
(1-a\Delta s)\rvx_s
-
b\Delta s\,\rmN_s.
\end{align*}
Expanding the factored update gives
\begin{align*}
\tilde\rvx_t^{\mathrm{EI}}
=
\tilde\rvx_t^{\mathrm{Euler}}
+
\frac{(\Delta s)^2}{2}
\left[
a^2\rvx_s
+
ab\,\rmN_s
\right]
+
\mathcal O((\Delta s)^3).
\end{align*}
Thus, the two updates agree to first order for small steps, while the factored
update retains the higher-order effects of the known dynamics. The difference
can become important for larger steps. For example, when $a>0$, the exact
multiplier $e^{-a\Delta s}$ remains positive and contractive for every
$\Delta s>0$, whereas the Euler approximation $1-a\Delta s$ becomes negative
when $a\Delta s>1$ and has magnitude greater than one when $a\Delta s>2$.
Plain Euler can therefore distort dynamics that the factored form already
handles analytically.

In the full PF-ODE, the neural prediction generally varies along the
trajectory. Subtracting \Cref{eq:factored-ei-step} from the exact
variation-of-constants formula gives
\begin{align}
\label{eq:ei-local-defect}
\widetilde\bPsi_{s\to t}(\rvx_s)
-
\tilde\rvx_t^{\mathrm{EI}}
=
\int_s^t
\mathcal E(\tau\shortrightarrow t)b(\tau)
\Big[
\rmN_{\bm{\phi}^\times}(\rvx_\tau,\tau)
-
\rmN_s
\Big]\diff\tau.
\end{align}
Hence, once the known schedule-dependent dynamics are handled analytically,
the remaining discretization error is governed by the variation of the neural
prediction along the trajectory. This separation is the basic numerical
principle that several diffusion-specific solvers will exploit in different
ways.

\subsection{Roadmap from Numerical Ideas to Diffusion Samplers}
The remainder of this chapter shows how these numerical ideas appear in modern
diffusion samplers. We begin with DDIM in \Cref{sec:revisit-ddim}, which
constructs a first-order update from the current model prediction. We then study
DEIS in \Cref{sec:exponential}, which uses predictions from previous steps to
build higher-order approximations. DPM-Solver in \Cref{sec:dpm} develops
higher-order updates using additional model evaluations within the current
step. DPM-Solver\texttt{++} in \Cref{sec:dpm++} develops two corresponding
variants based on clean-data prediction: a single-step variant,
DPM-Solver\texttt{++}(2S), which uses additional evaluations within the current
step, and a multistep variant, DPM-Solver\texttt{++}(2M), which reuses
predictions from previous steps. Finally, \Cref{sec:picard} introduces
ParaDiGMS, which updates several time points together through fixed-point
iteration to enable parallel computation across time.

The table below provides a roadmap connecting these diffusion samplers with
the classical numerical ideas reviewed in \Cref{eq:classic-solvers}.

\begin{center}
\small
\renewcommand{\arraystretch}{1.25}
\begin{tabularx}{0.98\linewidth}{
    @{}
    >{\raggedright\arraybackslash}p{0.21\linewidth}
    >{\raggedright\arraybackslash}p{0.32\linewidth}
    >{\raggedright\arraybackslash}X
    >{\raggedright\arraybackslash}p{0.13\linewidth}
    @{}
}
\toprule
\textbfs{Classical Idea} &
\textbfs{How the Update Is Constructed} &
\textbfs{Diffusion Examples} &
\textbfs{Where} \\
\midrule

Euler-type first-order update &
Use the model prediction at the beginning of the step to advance to the next
state. &
DDIM; DPM-Solver-1 &
\Cref{sec:revisit-ddim,sec:dpm} \\

\graymidrule

RK-type staged update &
Evaluate the model at additional intermediate states before completing the
current step. &
Higher-order DPM-Solver; DPM-Solver\texttt{++}(2S) &
\Cref{sec:dpm,sec:dpm++} \\

\graymidrule

AB-type multistep update &
Reuse model predictions from previous steps to construct the next update. &
DEIS; DPM-Solver\texttt{++}(2M) &
\Cref{sec:exponential,sec:dpm++} \\

\graymidrule

Picard-type fixed-point iteration &
Update several time points together and repeatedly refine their states. &
ParaDiGMS &
\Cref{sec:picard} \\

\bottomrule
\end{tabularx}
\end{center}

These connections provide a high-level numerical roadmap for the methods
developed below. Diffusion solvers further benefit from structure specific to
the PF-ODE, including different prediction parameterizations, analytically known
schedule-dependent coefficients, and alternative time parameterizations.

To keep the presentation systematic, each section follows the same broad
structure: we first derive the diffusion solver and explain the numerical idea
it exploits, then connect it to the corresponding classical solver principle,
and finally relate it to other diffusion solvers developed in this chapter.
This organization makes clear both the common numerical backbone and the
diffusion-specific ingredients that distinguish the resulting methods.

\clearpage
\newpage

\section{\texorpdfstring{DDIM}{DDIM}}\label{sec:revisit-ddim}

\emph{Denoising Diffusion Implicit Models} (DDIM)~\citep{song2020denoising}
is one of the earliest and most widely used methods for accelerating diffusion
sampling. Although DDIM was originally introduced from a probabilistic
perspective, its deterministic update also admits a particularly clean
interpretation through the PF-ODE. We therefore introduce DDIM here under the
common numerical ODE-solving umbrella of this chapter, and defer its original
variational perspective to \Cref{subsec:ddim}.

The Prologue in \Cref{sec:prologue-solvers} leaves us with a concrete numerical question: once the known
schedule-dependent factors have been retained analytically, how should we
approximate the neural prediction inside the remaining integral? DDIM takes
the simplest possible answer: \emph{use its value at the beginning of the step}.
We first derive the resulting update, and then clarify its relation to
classical Euler integration and to the higher-order diffusion solvers
introduced later.

\subsection{DDIM as a First-Order Diffusion Solver}\label{subsec:ddim-different-para}

Consider one sampling step from starting point\footnote{Throughout this chapter, $\tilde{\rvx}_s$ denotes the state at time level $s$ used to initialize the current solver step, typically obtained from a previous numerical step. The tilde distinguishes it from $\rvx_s$, which denotes an oracle sample from the exact diffusion distribution, either $\rvx_s \sim p_s$ or, conditioned on $\rvx_0$, $\rvx_s \sim p_s(\,\cdot \mid \rvx_0)$. Hence, $\tilde{\rvx}_s$ need not be exactly distributed according to $p_s$.} $\tilde\rvx_s$ at time $s$ to $t<s$. Under noise prediction,
\Cref{eq:variation_of_constants} gives the exact empirical PF-ODE flow
\begin{align*}
\widetilde\bPsi_{s\to t}(\tilde\rvx_s)
=
\mathcal E(s\shortrightarrow t)\tilde\rvx_s
+
\frac{1}{2}
\int_s^t
\frac{g^2(\tau)}{\sigma_\tau}
\mathcal E(\tau\shortrightarrow t)
\beps_{\bm{\phi}^\times}(\rvx_\tau,\tau)
\diff\tau .
\end{align*}
The schedule-dependent factors are known analytically; the only unknown
quantity inside the integral is the neural prediction
$\beps_{\bm{\phi}^\times}(\rvx_\tau,\tau)$ evaluated along the unknown
trajectory.

From the PF-ODE viewpoint, deterministic DDIM uses the simplest first-order
approximation: it holds the noise prediction fixed at its value at the
beginning of the step,
\[
\beps_{\bm{\phi}^\times}(\rvx_\tau,\tau)
\approx
\beps_{\bm{\phi}^\times}(\tilde\rvx_s,s),\quad\text{for}\quad \tau\in[t,s],
\]
 while retaining the known schedule-dependent factors
inside the integral. This gives
\begin{align*}
\tilde\rvx_t
=
\mathcal E(s\shortrightarrow t)\tilde\rvx_s
+
\left[
\frac{1}{2}
\int_s^t
\frac{g^2(\tau)}{\sigma_\tau}
\mathcal E(\tau\shortrightarrow t)
\diff\tau
\right]
\beps_{\bm{\phi}^\times}(\tilde\rvx_s,s).
\end{align*}

Using $f(t)=\alpha_t'/\alpha_t$, the integrating factor simplifies to
$\mathcal E(s\shortrightarrow t)=\alpha_t/\alpha_s$. The remaining
schedule-dependent coefficient also has a convenient closed form:
$\tfrac{1}{2}\tfrac{g^2(\tau)}{\sigma_\tau}
\mathcal E(\tau\shortrightarrow t)
=
\alpha_t
\tfrac{\diff}{\diff\tau}
\bigl(\tfrac{\sigma_\tau}{\alpha_\tau}\bigr)$.
Therefore, once the neural prediction is held fixed over the step, the
remaining integral can be evaluated exactly from the noise schedule.

\proppp{DDIM as a First-Order Factored EI
Update}{ddim-euler}{
For $s>t$, deterministic DDIM gives
\begin{align}
\label{eq:ddim-euler}
\tilde\rvx_t
=
\frac{\alpha_t}{\alpha_s}\tilde\rvx_s
+
\alpha_t
\left(
\frac{\sigma_t}{\alpha_t}
-
\frac{\sigma_s}{\alpha_s}
\right)
\beps_{\bm{\phi}^\times}(\tilde\rvx_s,s).
\end{align}
}{
The result follows by integrating
$\frac{\diff}{\diff\tau}(\sigma_\tau/\alpha_\tau)$
from $s$ to $t$.
}

The approximation made by DDIM is therefore very specific: the known
schedule evolution is retained over the full interval, while only the
change of the neural prediction is neglected.

There is an analogous derivation under clean-data prediction. The exact
empirical PF-ODE flow can be written as
\begin{align*}
\widetilde\bPsi_{s\to t}(\tilde\rvx_s)
=
\frac{\sigma_t}{\sigma_s}\tilde\rvx_s
+
\sigma_t
\int_s^t
\frac{\diff}{\diff\tau}
\left(
\frac{\alpha_\tau}{\sigma_\tau}
\right)
\rvx_{\bm{\phi}^\times}(\rvx_\tau,\tau)
\diff\tau.
\end{align*}
Using the same first-order approximation as above, we hold
$\rvx_{\bm{\phi}^\times}(\rvx_\tau,\tau)$ fixed at its value at time $s$.
The remaining schedule-dependent integral is then evaluated exactly, giving
\[
\tilde\rvx_t
=
\frac{\sigma_t}{\sigma_s}\tilde\rvx_s
+
\alpha_s
\left(
\frac{\alpha_t}{\alpha_s}
-
\frac{\sigma_t}{\sigma_s}
\right)
\rvx_{\bm{\phi}^\times}(\tilde\rvx_s,s).
\]
Thus, freezing either the noise prediction in the noise-prediction PF-ODE or
the clean-data prediction in the clean-data PF-ODE produces the same DDIM
update. This correspondence will reappear later in DPM-Solver and
DPM-Solver\texttt{++}, which build higher-order approximations from the noise-
and clean-data-prediction formulations, respectively.

\paragraph{Equivalent Forms and Interpretation.}
At the oracle level, the standard prediction targets are pointwise
algebraically equivalent through \Cref{eq:predictions-equivalence}.
Therefore, the deterministic DDIM update derived above from noise prediction
can be rewritten at the starting point $s$ in terms of clean-data or score
prediction without changing the finite-step update. The following corollary
collects these equivalent expressions.

\cornp{Equivalent Forms of the Deterministic DDIM Update}{
Let $s>t$, and consider a state $\tilde\rvx_s$ at time $s$. The oracle
deterministic DDIM update admits the equivalent expressions
\begin{align}
\begin{aligned}
\label{eq:ddim-update-all}
\tilde\rvx_t
&=
\frac{\alpha_t}{\alpha_s}\tilde\rvx_s
+
\alpha_t
\left(
\frac{\sigma_t}{\alpha_t}
-
\frac{\sigma_s}{\alpha_s}
\right)
\beps^*(\tilde\rvx_s,s)
\\
&=
\frac{\sigma_t}{\sigma_s}\tilde\rvx_s
+
\alpha_s
\left(
\frac{\alpha_t}{\alpha_s}
-
\frac{\sigma_t}{\sigma_s}
\right)
\rvx^*(\tilde\rvx_s,s)
\\
&=
\frac{\alpha_t}{\alpha_s}\tilde\rvx_s
+
\sigma_s^2
\left(
\frac{\alpha_t}{\alpha_s}
-
\frac{\sigma_t}{\sigma_s}
\right)
\rvs^*(\tilde\rvx_s,s)
\\
&=
\alpha_t
\underset{\substack{
\approx\rvx_{\bm{\phi}^\times}\\
\text{estimated clean}
}}{
\underbrace{\rvx^*(\tilde\rvx_s,s)}
}
+
\sigma_t
\underset{\substack{
\approx\beps_{\bm{\phi}^\times}\\
\text{estimated noise}
}}{
\underbrace{\beps^*(\tilde\rvx_s,s)}
}.
\end{aligned}
\end{align}
}

The last identity in \Cref{eq:ddim-update-all} gives a particularly intuitive
view of DDIM. Starting from $\tilde\rvx_s$, DDIM keeps the clean and noise
estimates inferred at time $s$ fixed and recombines them using the target
schedule coefficients:
\[
\tilde\rvx_t
=
\alpha_t\rvx^*(\tilde\rvx_s,s)
+
\sigma_t\beps^*(\tilde\rvx_s,s).
\]
This mirrors the shared-noise coupling of the forward Gaussian path,
$\rvx_\tau=\alpha_\tau\rvx_0+\sigma_\tau\beps$: conditioned on a fixed pair
$(\rvx_0,\beps)$, moving from one time to another changes only the coefficients
$(\alpha_\tau,\sigma_\tau)$. DDIM applies the same idea using the clean and
noise estimates available at the beginning of the step.

At the conditional level, where $\rvx_0$ and $\beps$ are fixed, this
recombination follows the corresponding shared-noise path exactly. At the
marginal level, however, $\rvx^*(\tilde\rvx_s,s)$ and
$\beps^*(\tilde\rvx_s,s)$ are posterior estimates inferred from
$\tilde\rvx_s$. Consequently, a finite DDIM step is generally an approximation
to the marginal PF-ODE flow, rather than an exact draw from $p_t$.

\begin{figure}[th!]
    \centering
    \includegraphics[width=0.86\linewidth]{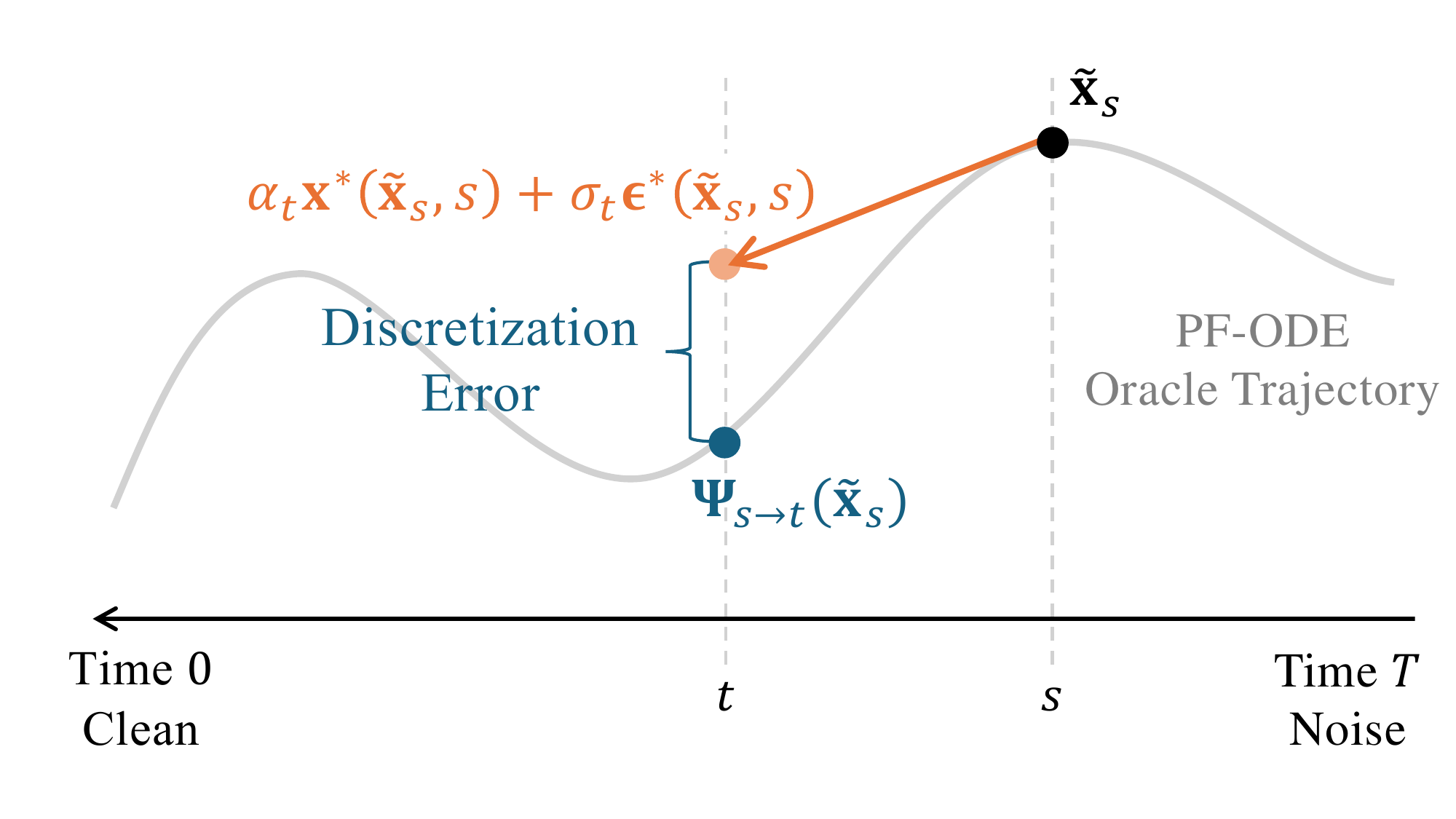}
    \caption{\textbfs{Finite-step DDIM versus the empirical PF-ODE flow.}
    Starting from $\tilde\rvx_s$, the empirical PF-ODE evolves continuously
    to $\widetilde\bPsi_{s\to t}(\tilde\rvx_s)$, whereas DDIM retains the
    start-time neural prediction over the whole interval while evolving the
    known schedule analytically. Their discrepancy is the one-step
    discretization error in \Cref{eq:ddim-local-defect}.
    \figcredit{Created by the authors.}}
    \label{fig:euler}
\end{figure}

\paragraph{Why Does the Prediction Held Fixed Matter?}
At any fixed state and time, the standard prediction parameterizations are
algebraically equivalent through \Cref{eq:predictions-equivalence}. A
finite-step solver, however, introduces an additional approximation by holding
one chosen prediction fixed over an interval. Such a freezing approximation is
preserved under a change of parameterization only when the converted prediction
also remains constant along the resulting approximate trajectory.

DDIM provides a particularly clean example. Let
\[
\widehat\rvx_0
:=
\rvx_{\bm{\phi}^\times}(\tilde\rvx_s,s),
\qquad
\widehat\beps
:=
\beps_{\bm{\phi}^\times}(\tilde\rvx_s,s),
\]
where, if only one quantity is directly predicted by the model, the other is
defined by the algebraic conversion in \Cref{eq:predictions-equivalence}.
Hence, by construction,
$\tilde\rvx_s=\alpha_s\widehat\rvx_0+\sigma_s\widehat\beps$.

Holding this start-time clean/noise pair fixed defines the DDIM interpolation
over the entire step:
\[
\tilde\rvx_\tau^{\mathrm{DDIM}}
=
\alpha_\tau\widehat\rvx_0
+
\sigma_\tau\widehat\beps,
\qquad
\tau\in[t,s].
\]
This requires no additional model evaluations: once
$(\widehat\rvx_0,\widehat\beps)$ is fixed at time $s$, the schedule
coefficients $(\alpha_\tau,\sigma_\tau)$ are known for every intermediate
$\tau$. At $\tau=s$ the expression recovers the starting state, while at
$\tau=t$ it gives the DDIM update. The intermediate values simply describe
the finite-step trajectory induced by the DDIM approximation.

The special role of noise and clean-data prediction is now immediate. Along
this trajectory,
\[
\frac{
\tilde\rvx_\tau^{\mathrm{DDIM}}
-
\sigma_\tau\widehat\beps
}{
\alpha_\tau
}
=
\widehat\rvx_0,
\qquad
\frac{
\tilde\rvx_\tau^{\mathrm{DDIM}}
-
\alpha_\tau\widehat\rvx_0
}{
\sigma_\tau
}
=
\widehat\beps.
\]
Thus, the algebraically converted clean and noise predictions remain constant
together along the DDIM trajectory. Consequently, holding
$\beps_{\bm{\phi}^\times}(\rvx_\tau,\tau)
\approx\beps_{\bm{\phi}^\times}(\tilde\rvx_s,s)$ or holding
$\rvx_{\bm{\phi}^\times}(\rvx_\tau,\tau)
\approx\rvx_{\bm{\phi}^\times}(\tilde\rvx_s,s)$ leads to the same DDIM
update. In this sense, $\widehat\rvx_0$ and $\widehat\beps$ are the two fixed
components of the DDIM shared-noise path, while
$(\alpha_\tau,\sigma_\tau)$ provide the known time-dependent weights.

However, score and velocity behave differently. DDIM keeps the inferred
clean/noise pair $(\widehat\rvx_0,\widehat\beps)$ fixed, but the score and
velocity obtained from this pair generally change with time.

For example, converting the fixed (at time $s$) noise estimate to score prediction at an
intermediate time $\tau$ gives
\[
\rvs(\tilde\rvx_\tau^{\mathrm{DDIM}},\tau)
=
-\frac{\widehat\beps}{\sigma_\tau}
=
\frac{\sigma_s}{\sigma_\tau}
\rvs_{\bm{\phi}^\times}(\tilde\rvx_s,s).
\]
Thus, even though $\widehat\beps$ is fixed, its corresponding score generally
varies with $\tau$ through the factor $1/\sigma_\tau$. Directly imposing
\[
\rvs_{\bm{\phi}^\times}(\rvx_\tau,\tau)
\approx
\rvs_{\bm{\phi}^\times}(\tilde\rvx_s,s),
\qquad
\tau\in[t,s],
\]
therefore makes a different finite-step approximation and generally yields a
different first-order factored solver. Similarly, the fixed clean/noise pair induces the velocity
\[
\frac{\diff}{\diff\tau}
\tilde\rvx_\tau^{\mathrm{DDIM}}
=
\alpha_\tau'\widehat\rvx_0
+
\sigma_\tau'\widehat\beps.
\]
This velocity generally changes with $\tau$ because the schedule derivatives
$\alpha_\tau'$ and $\sigma_\tau'$ change. Holding the raw velocity prediction
fixed instead,
$\rvv_{\bm{\phi}^\times}(\rvx_\tau,\tau)
\approx
\rvv_{\bm{\phi}^\times}(\tilde\rvx_s,s)$,
gives plain Euler in the original diffusion time; its relation to DDIM is
discussed in \Cref{subsec:ddim-classical-euler}.

\rmkb{
Prediction parameterizations are pointwise algebraically equivalent, but
finite-step freezing need not commute with a time-dependent conversion. DDIM
keeps a consistent start-time clean/noise pair fixed; the corresponding score
and complete velocity generally vary over the step. Hence directly freezing
the raw score or velocity defines a different first-order method. The velocity
case is quantified in \Cref{subsec:ddim-classical-euler}.
}

\subsection{Relation to Euler-Type Methods}
\label{subsec:ddim-classical-euler}

The discussion above shows that DDIM is a first-order method, but its relation
to the classical Euler method requires some care. Once the known diffusion
schedule is separated from the learned prediction, different first-order
schemes arise depending on \emph{what is held fixed over a finite step}.

We compare three one-evaluation constructions:
\begin{enumerate}
    \item \emph{plain Euler}, which holds the complete PF-ODE velocity fixed;
    \item \emph{exponential Euler}, which treats the known linear propagation
    analytically but holds the remaining coefficient-weighted residual fixed;
    \item \emph{DDIM}, which keeps the start-time clean/noise estimates fixed
    while retaining their finite-step evolution under the known diffusion
    schedule.
\end{enumerate}
These distinctions are easiest to see from two equivalent views of the same
DDIM update. The clean/noise form makes its difference from plain Euler
transparent, whereas the factored semilinear form clarifies its relation to
exponential Euler. We then identify the remaining discretization error and
show the precise sense in which DDIM itself can be viewed as an Euler method
after a schedule-aligned change of coordinates.

\paragraph{DDIM versus Plain Euler.}
DDIM and plain Euler use the same start-time model information, but propagate
it differently over a finite step. Plain Euler locally approximates the
complete PF-ODE velocity, whereas DDIM retains the finite-step evolution of
the known diffusion schedule while holding the model prediction fixed.
Consequently, for smooth schedules, the two updates agree to first order and
generally differ by terms of order $\mathcal O((\Delta s)^2)$. We make this
difference explicit below.

Consider one backward step from time $s$ to $t<s$, with
$\Delta s:=s-t>0$, and suppose the start-time clean and noise predictions are
algebraically consistent:
\[
\tilde\rvx_s
=
\alpha_s
\rvx_{\bm{\phi}^\times}(\tilde\rvx_s,s)
+
\sigma_s
\beps_{\bm{\phi}^\times}(\tilde\rvx_s,s).
\]
The corresponding empirical PF-ODE velocity at time $s$ is
\[
\rvv_{\bm{\phi}^\times}(\tilde\rvx_s,s)
=
\alpha_s'
\rvx_{\bm{\phi}^\times}(\tilde\rvx_s,s)
+
\sigma_s'
\beps_{\bm{\phi}^\times}(\tilde\rvx_s,s).
\]
Plain Euler holds this complete velocity fixed over the step, $\tau\in[t,s]$, and gives:
\[
\tilde\rvx_t^{\mathrm{Euler}}
=
\tilde\rvx_s
-
\Delta s\,
\rvv_{\bm{\phi}^\times}(\tilde\rvx_s,s).
\]
By contrast, DDIM keeps the clean/noise-prediction pair at the starting time fixed, while evolving the known schedule coefficients from $(\alpha_s,\sigma_s)$ to $(\alpha_t,\sigma_t)$:
\[
\tilde\rvx_t^{\mathrm{DDIM}}
=
\tilde\rvx_s
+
(\alpha_t-\alpha_s)
\rvx_{\bm{\phi}^\times}(\tilde\rvx_s,s)
+
(\sigma_t-\sigma_s)
\beps_{\bm{\phi}^\times}(\tilde\rvx_s,s).
\]
Their exact finite-step difference is therefore
\begin{align*}
\tilde\rvx_t^{\mathrm{DDIM}}
-
\tilde\rvx_t^{\mathrm{Euler}}
={}&
\left[
\alpha_t-\alpha_s+\Delta s\,\alpha_s'
\right]
\rvx_{\bm{\phi}^\times}(\tilde\rvx_s,s)
+
\left[
\sigma_t-\sigma_s+\Delta s\,\sigma_s'
\right]
\beps_{\bm{\phi}^\times}(\tilde\rvx_s,s).
\end{align*}
For smooth schedules,
\[
\alpha_t-\alpha_s+\Delta s\,\alpha_s'
=
\tfrac{(\Delta s)^2}{2}\alpha_s''
+
\mathcal O((\Delta s)^3), \quad
\sigma_t-\sigma_s+\Delta s\,\sigma_s'
=
\tfrac{(\Delta s)^2}{2}\sigma_s''
+
\mathcal O((\Delta s)^3).
\]
Hence,
\begin{align*}
\tilde\rvx_t^{\mathrm{DDIM}}
-
\tilde\rvx_t^{\mathrm{Euler}}
=
\tfrac{(\Delta s)^2}{2}
\left[
\alpha_s''
\rvx_{\bm{\phi}^\times}(\tilde\rvx_s,s)
+
\sigma_s''
\beps_{\bm{\phi}^\times}(\tilde\rvx_s,s)
\right]
+
\mathcal O((\Delta s)^3).
\end{align*}
Thus, DDIM and plain Euler agree to first order but generally differ over a
finite step. Plain Euler replaces the finite schedule changes
$(\alpha_t-\alpha_s,\sigma_t-\sigma_s)$ by their local linearizations,
whereas DDIM retains the finite schedule evolution analytically.

A useful special case follows immediately. If both $\alpha_\tau$ and
$\sigma_\tau$ are \emph{affine in $\tau$ over $[t,s]$}, then
\[
\alpha_t-\alpha_s=-\Delta s\,\alpha_s',
\qquad
\sigma_t-\sigma_s=-\Delta s\,\sigma_s',
\]
so DDIM and plain Euler coincide exactly. This includes, for example, the
schedules (see also discussion in \Cref{subsec:concept-why-v-affine}) 
\[(\alpha_t,\sigma_t)=(1,t)\quad \text{and}\quad (\alpha_t,\sigma_t)=(1-t,t) .
\]
Thus, the finite-step distinction between the
two methods is driven by the nonlinearity of the known schedule in the
original diffusion time.

\paragraph{DDIM versus Exponential Euler.}
The comparison above used the clean/noise representation to show how DDIM
differs from plain Euler in the original diffusion time. We now return to the
semilinear representation from the prologue to distinguish DDIM from
conventional exponential Euler. The difference is simple: both methods retain
the known linear evolution analytically, but they hold different quantities
fixed inside the remaining integral.

Under noise prediction, the factored PF-ODE in
\Cref{eq:general-semilinear-pfode} has
\[
a(\tau)=f(\tau),
\qquad
b(\tau)=\frac{g^2(\tau)}{2\sigma_\tau},
\qquad
\rmN_{\bm{\phi}^\times}
=
\beps_{\bm{\phi}^\times}.
\]
Its exact variation-of-constants formula contains the model-dependent term
\[
b(\tau)\beps_{\bm{\phi}^\times}(\rvx_\tau,\tau).
\]

Conventional exponential Euler treats this coefficient-weighted quantity as a
single residual and holds the whole residual fixed at the beginning of the
step:
\[
b(\tau)\beps_{\bm{\phi}^\times}(\rvx_\tau,\tau)
\approx
b(s)\beps_{\bm{\phi}^\times}(\tilde\rvx_s,s), \quad\tau\in[t,s]
\]
It therefore gives
\[
\tilde\rvx_t^{\mathrm{ExpEuler}}
=
\mathcal E(s\shortrightarrow t)\tilde\rvx_s
+
b(s)
\left[
\int_s^t
\mathcal E(\tau\shortrightarrow t)
\diff\tau
\right]
\beps_{\bm{\phi}^\times}(\tilde\rvx_s,s).
\]

DDIM uses the finer factorization emphasized in the prologue. Since
$b(\tau)$ is known from the diffusion schedule, it remains inside the
integral; only the neural prediction is held fixed:
\[
\beps_{\bm{\phi}^\times}(\rvx_\tau,\tau)
\approx
\beps_{\bm{\phi}^\times}(\tilde\rvx_s,s).
\]
Hence,
\[
\tilde\rvx_t^{\mathrm{DDIM}}
=
\mathcal E(s\shortrightarrow t)\tilde\rvx_s
+
\left[
\int_s^t
\mathcal E(\tau\shortrightarrow t)b(\tau)
\diff\tau
\right]
\beps_{\bm{\phi}^\times}(\tilde\rvx_s,s).
\]

The finite-step difference between the two updates therefore comes entirely
from the known coefficient $b(\tau)$:
\[
\tilde\rvx_t^{\mathrm{DDIM}}
-
\tilde\rvx_t^{\mathrm{ExpEuler}}
=
\left[
\int_s^t
\mathcal E(\tau\shortrightarrow t)
\bigl(b(\tau)-b(s)\bigr)
\diff\tau
\right]
\beps_{\bm{\phi}^\times}(\tilde\rvx_s,s).
\]
For smooth $b$,
\[
\tilde\rvx_t^{\mathrm{DDIM}}
-
\tilde\rvx_t^{\mathrm{ExpEuler}}
=
\frac{(\Delta s)^2}{2}
b'(s)
\beps_{\bm{\phi}^\times}(\tilde\rvx_s,s)
+
\mathcal O((\Delta s)^3).
\]
Thus, DDIM and conventional exponential Euler agree to first order. The only
difference is that exponential Euler freezes $b(\tau)$ at $b(s)$ together
with the neural prediction, while DDIM keeps the known variation of
$b(\tau)$. Therefore, the two updates coincide exactly whenever $b(\tau)$ is constant
over the step, including the constant-$b$ setting used in the Prologue
example.

Together with the preceding plain-Euler comparison, the three first-order
rules differ only in how much known structure is retained before making the
constant approximation on $\tau\in[t,s]$:
\[
\begin{array}{rcl}
\text{Plain Euler}
&:&
f(\tau)\rvx_\tau+b(\tau)\beps_\tau
\;\approx\;
f(s)\tilde\rvx_s+b(s)\widehat\beps,
\\[4pt]
\text{Exponential Euler}
&:&
b(\tau)\beps_\tau
\;\approx\;
b(s)\widehat\beps,
\\[4pt]
\text{DDIM}
&:&
\beps_\tau
\;\approx\;
\widehat\beps,
\end{array}
\]
where
$\widehat\beps
:=
\beps_{\bm{\phi}^\times}(\tilde\rvx_s,s)$.
All three use the same single neural-network evaluation at the beginning of
the step. Plain Euler approximates the complete velocity; exponential Euler
first retains the known linear propagation; DDIM additionally retains the
known coefficient $b(\tau)$ and approximates only the neural prediction.

This hierarchy concerns which analytically known terms are preserved, rather
than a universal ordering of one-step errors. Its main numerical consequence
is that the remaining approximation in DDIM can be isolated entirely to the
variation of the neural prediction, as we show next.

\paragraph{What Error Does Each Method Still Make?}
The comparisons above describe how the three numerical updates differ from one
another. We now compare each update with the exact empirical PF-ODE flow.
For this local-error analysis, suppose the step starts from an exact empirical
state $\rvx_s$ at time $s$.

For plain Euler, the exact empirical flow can be written in the direct velocity
form as
\[
\widetilde\bPsi_{s\to t}(\rvx_s)
=
\rvx_s
+
\int_s^t
\rvv_{\bm{\phi}^\times}(\rvx_\tau,\tau)
\diff\tau.
\]
Since plain Euler holds the complete velocity fixed at its start-time value,
its exact one-step defect is
\begin{align*}
\widetilde\bPsi_{s\to t}(\rvx_s)
-
\tilde\rvx_t^{\mathrm{Euler}}
=
\int_s^t
\Big[
\rvv_{\bm{\phi}^\times}(\rvx_\tau,\tau)
-
\rvv_{\bm{\phi}^\times}(\rvx_s,s)
\Big]
\diff\tau.
\end{align*}

For exponential Euler, the known linear propagation is already treated
analytically, but the coefficient-weighted residual
$b(\tau)\beps_{\bm{\phi}^\times}(\rvx_\tau,\tau)$ is held fixed. Hence
\begin{align*}
\widetilde\bPsi_{s\to t}(\rvx_s)
-
\tilde\rvx_t^{\mathrm{ExpEuler}}
=
\int_s^t
\mathcal E(\tau\shortrightarrow t)
\Big[
b(\tau)
\beps_{\bm{\phi}^\times}(\rvx_\tau,\tau)
-
b(s)
\beps_{\bm{\phi}^\times}(\rvx_s,s)
\Big]
\diff\tau.
\end{align*}

Finally, DDIM retains both the known linear propagation and the known
coefficient $b(\tau)$, and holds only the neural prediction fixed:
\begin{align}
\label{eq:ddim-local-defect}
\widetilde\bPsi_{s\to t}(\rvx_s)
-
\tilde\rvx_t^{\mathrm{DDIM}}
=
\int_s^t
\mathcal E(\tau\shortrightarrow t)b(\tau)
\Big[
\beps_{\bm{\phi}^\times}(\rvx_\tau,\tau)
-
\beps_{\bm{\phi}^\times}(\rvx_s,s)
\Big]
\diff\tau.
\end{align}

These identities reveal the relevant notion of smoothness for each first-order
scheme. If the corresponding quantity varies slowly along the exact empirical
trajectory, then the one-step approximation is accurate. More precisely,
with $\Delta s:=s-t$, standard smoothness gives
\begin{align*}
\left\|
\widetilde\bPsi_{s\to t}(\rvx_s)
-
\tilde\rvx_t^{\mathrm{Euler}}
\right\|
&\le
\frac{(\Delta s)^2}{2}
\sup_{\tau\in[t,s]}
\left\|
\tfrac{\diff}{\diff\tau}
\rvv_{\bm{\phi}^\times}(\rvx_\tau,\tau)
\right\|,
\\
\left\|
\widetilde\bPsi_{s\to t}(\rvx_s)
-
\tilde\rvx_t^{\mathrm{ExpEuler}}
\right\|
&\le
\frac{(\Delta s)^2}{2}
\sup_{\tau\in[t,s]}
\left|
\mathcal E(\tau\shortrightarrow t)
\right|\cdot
\sup_{\tau\in[t,s]}
\left\|
\tfrac{\diff}{\diff\tau}
\left[
b(\tau)
\beps_{\bm{\phi}^\times}(\rvx_\tau,\tau)
\right]
\right\|,
\\
\left\|
\widetilde\bPsi_{s\to t}(\rvx_s)
-
\tilde\rvx_t^{\mathrm{DDIM}}
\right\|
&\le
\frac{(\Delta s)^2}{2}
\sup_{\tau\in[t,s]}
\left|
\mathcal E(\tau\shortrightarrow t)b(\tau)
\right| \cdot
\sup_{\tau\in[t,s]}
\left\|
\tfrac{\diff}{\diff\tau}
\beps_{\bm{\phi}^\times}(\rvx_\tau,\tau)
\right\|.
\end{align*}

Under the usual smoothness and stability assumptions, all three methods are
\emph{globally first order} and require only one neural-network evaluation per
step. Moreover, each method is exact on a step when the quantity it holds
fixed is in fact constant along the exact empirical trajectory: the complete
PF-ODE velocity for plain Euler, the coefficient-weighted residual
$b(\tau)\beps_{\bm{\phi}^\times}(\rvx_\tau,\tau)$ for exponential Euler, and
the noise prediction
$\beps_{\bm{\phi}^\times}(\rvx_\tau,\tau)$ for DDIM.

Accordingly, there is no universal error ordering among the three methods:
each is most accurate when the quantity it holds fixed varies slowly over the
step. DDIM is particularly well suited to regimes where the neural prediction
changes slowly but the known diffusion schedule changes appreciably, since it
retains the latter analytically. Exploiting more known structure can therefore
reduce discretization error, but does not guarantee a smaller error on every
individual step.

The errors discussed here are numerical discretization errors relative to the
empirical PF-ODE; model mismatch between the empirical and oracle PF-ODEs is
a separate source of error.

\paragraph{A Precise Euler Interpretation of DDIM.}
Although DDIM and plain Euler agree to first order, they generally give
different finite-step updates in the original state $\rvx$ and diffusion time
$t$. There is, however, an exact Euler interpretation after a
schedule-aligned change of both the state and the time coordinate.

Under noise prediction, define
\begin{equation*}
\rvy_t
:=
\frac{\rvx_t}{\alpha_t},
\qquad
\rho_t
:=
\frac{\sigma_t}{\alpha_t}.
\end{equation*}
The two transformations play complementary roles. The state normalization
$\rvx_t\mapsto\rvx_t/\alpha_t$ factors out the known signal scaling
$\alpha_t$, while the new clock
$\rho_t=\sigma_t/\alpha_t$ measures progress by the noise-to-signal ratio
rather than by the original diffusion time.

To see this first at the conditional level, consider a fixed pair
$(\rvx_0,\beps)$ in the Gaussian perturbation path
\[
\rvx_t
=
\alpha_t\rvx_0+\sigma_t\beps,
\qquad
\beps\sim\mathcal N(\mathbf 0,\mathbf I).
\]
Here, $\beps$ is a fixed latent noise realization, not a neural prediction.
The transformed state satisfies
\[
\rvy_t
=
\frac{\rvx_t}{\alpha_t}
=
\rvx_0+\rho_t\beps.
\]
Hence, along this particular conditional path,
\[
\frac{\diff\rvy}{\diff\rho}
=
\beps.
\]
Thus, for fixed $(\rvx_0,\beps)$, the known schedule dependence becomes affine
in the transformed coordinate $\rho$.

The corresponding empirical PF-ODE has the same coordinate structure, but
the fixed latent noise $\beps$ is replaced by the learned, state-dependent
noise prediction. Using \Cref{eq:noise_pf_ode},
\[
\frac{\diff\rvx_t}{\diff t}
=
\frac{\alpha_t'}{\alpha_t}\rvx_t
+
\left(
\sigma_t'
-
\frac{\alpha_t'}{\alpha_t}\sigma_t
\right)
\beps_{\bm{\phi}^\times}(\rvx_t,t),
\]
we obtain
\begin{align}
\frac{\diff\rvy_t}{\diff t}
=
\frac{1}{\alpha_t}
\left(
\frac{\diff\rvx_t}{\diff t}
-
\frac{\alpha_t'}{\alpha_t}\rvx_t
\right)=
\frac{\diff\rho_t}{\diff t}\,
\beps_{\bm{\phi}^\times}(\rvx_t,t).
\label{eq:ddim-coordinate-ode-t}
\end{align}
Hence, on any interval where $\alpha_t>0$ and $\rho_t$ is strictly monotone,
we may use $\rho$ as the time coordinate. Writing $t=t(\rho)$ and defining
the transformed neural field
\[
\bar{\beps}_{\bm{\phi}^\times}(\rvy,\rho)
:=
\beps_{\bm{\phi}^\times}
\bigl(
\alpha_{t(\rho)}\rvy,
t(\rho)
\bigr),
\]
the empirical PF-ODE becomes
\begin{equation}
\label{eq:ddim-coordinate-ode-rho}
\frac{\diff\rvy}{\diff\rho}
=
\bar{\beps}_{\bm{\phi}^\times}(\rvy,\rho).
\end{equation}
The spatial and temporal dependence of the model therefore remains present;
it is simply expressed in the transformed variables $(\rvy,\rho)$.

Explicit Euler applied to \Cref{eq:ddim-coordinate-ode-rho} from
$\rho_s$ to $\rho_t$ gives
\[
\tilde\rvy_t
=
\tilde\rvy_s
+
(\rho_t-\rho_s)
\bar{\beps}_{\bm{\phi}^\times}(\tilde\rvy_s,\rho_s).
\]
Since
\[
\bar{\beps}_{\bm{\phi}^\times}(\tilde\rvy_s,\rho_s)
=
\beps_{\bm{\phi}^\times}(\tilde\rvx_s,s),
\]
substituting
$\rvy=\rvx/\alpha$ and $\rho=\sigma/\alpha$ yields
\[
\tilde\rvx_t
=
\frac{\alpha_t}{\alpha_s}\tilde\rvx_s
+
\alpha_t
\left(
\frac{\sigma_t}{\alpha_t}
-
\frac{\sigma_s}{\alpha_s}
\right)
\beps_{\bm{\phi}^\times}(\tilde\rvx_s,s),
\]
which is exactly the deterministic DDIM update.

\msgu{Coordinate Choice Is Part of the Numerical Method}{}{
An invertible change of state and time coordinates gives an equivalent
representation of the same continuous ODE solution. A finite-step
discretization, however, is generally not invariant to that representation:
``change coordinates and then apply Euler'' need not equal
``apply Euler and then change coordinates''.

The numerical significance is that a suitable coordinate system can absorb
analytically known parts of the dynamics before discretization. In DDIM,
$\rvx/\alpha$ removes the known signal scaling and
$\rho=\sigma/\alpha$ absorbs the corresponding schedule-dependent time
scaling, leaving the transformed neural field
$\bar{\beps}_{\bm{\phi}^\times}(\rvy,\rho)$ as the quantity approximated by
Euler. Thus DDIM generally differs from plain Euler in $(\rvx,t)$, while it is
exactly explicit Euler in the schedule-aligned coordinates
$(\rvy,\rho)$.
}

A dual construction holds under clean-data prediction. Normalizing the state by $\sigma_t$ and using $\alpha_t/\sigma_t$ as the time coordinate gives 
\[
\frac{\diff}{\diff t} \left( \frac{\rvx_t}{\sigma_t} \right) = \frac{\diff}{\diff t} \left( \frac{\alpha_t}{\sigma_t} \right) \rvx_{\bm{\phi}^\times}(\rvx_t,t), 
\] 
so the clean-data form of DDIM is likewise explicit Euler in its corresponding schedule-aligned coordinates.

This \emph{state--time change-of-coordinates viewpoint} also previews the role
of coordinate choice in higher-order diffusion solvers. DPM-Solver, developed
later in \Cref{sec:dpm}, uses the half-log SNR
\[
\lambda_t
:=
\log\frac{\alpha_t}{\sigma_t},
\]
which is directly related to DDIM's Euler clock:
\[
\rho_t
=
\frac{\sigma_t}{\alpha_t}
=
e^{-\lambda_t},
\qquad
\diff\rho
=
-e^{-\lambda}\diff\lambda.
\]
Therefore, the exact transformed equation
\[
\frac{\diff\rvy}{\diff\rho}
=
\bar{\beps}_{\bm{\phi}^\times}(\rvy,\rho)
\]
can equivalently be integrated as
\[
\rvy_t
=
\rvy_s
-
\int_{\lambda_s}^{\lambda_t}
e^{-\lambda}
\hat{\beps}_{\bm{\phi}^\times}
(\hat\rvx_\lambda,\lambda)
\diff\lambda,
\]
with the usual $\lambda$-parameterized model notation introduced later in
\Cref{sec:dpm}.

At first order, holding the neural prediction fixed while retaining the known
change-of-variable weight gives the same finite-step update, so
DPM-Solver-1 is exactly DDIM. Beyond first order, however, the coordinate in
which the state--time-dependent neural prediction is approximated becomes part
of the numerical method. DPM-Solver uses $\lambda$ because it yields an
exponentially weighted integral with convenient higher-order expansions.

\subsection{(Optional) The Original Variational Perspective on DDIM}\label{subsec:ddim}

The PF-ODE perspective above explains deterministic DDIM as a numerical
solver. Historically, however, DDIM was introduced from the variational
viewpoint of DDPM. Its starting question is different:
\emph{does the denoising objective uniquely determine how noisy states at
different times must be coupled?}

\paragraph{Revisiting DDPM's Variational View.}
Recall that DDPM introduces a fixed Markovian inference process whose
one-time perturbation marginals take the Gaussian form
\[
p_t(\rvx_t|\rvx_0)
=
\mathcal N
\bigl(
\rvx_t;
\alpha_t\rvx_0,
\sigma_t^2\rmI
\bigr).
\]
The corresponding variational derivation involves tractable conditionals
such as
$p(\rvx_t|\rvx_s,\rvx_0)$ for $t<s$, while the learned reverse model used
for generation cannot condition directly on the unknown clean sample
$\rvx_0$ and instead relies on the diffusion-model prediction.

The key observation behind DDIM concerns the denoising objective. We recall that a noise-prediction DDPM loss can be written as
\[
\E_t
\E_{\rvx_0,\beps}
\left[
\left\|
\beps_{\bm{\phi}}
\bigl(
\alpha_t\rvx_0+\sigma_t\beps,t
\bigr)
-
\beps
\right\|_2^2
\right],
\]
which depends on the inference process through the one-time marginals
$p_t(\rvx_t|\rvx_0)$, rather than through the full joint coupling of
$\rvx_1,\ldots,\rvx_T$ conditioned on $\rvx_0$.

\paragraph{Original DDIM Motivation.}
The central observation behind DDIM is that the denoising training objective
does not uniquely determine how noisy states at different times are coupled.
For a fixed clean sample $\rvx_0$, the usual denoising surrogate is determined
by the one-time perturbation marginals
$p_t(\rvx_t|\rvx_0)$, whereas many different joint processes across time can
share exactly the same marginals. DDIM exploits this freedom: it preserves
the perturbation marginals used during training, while changing the coupling
between different noise levels.

Concretely, for any $t<s$, consider a conditional transition
${\color{orange}\pi(\rvx_t|\rvx_s,\rvx_0)}$. We require it to be
\emph{one-step marginally consistent}:
\begin{mdframed}
\begin{align}
\label{eq:ddim-reverse-kernel-def}
\int
{\color{orange}\pi(\rvx_t|\rvx_s,\rvx_0)}
\,p_s(\rvx_s|\rvx_0)
\diff\rvx_s
=
p_t(\rvx_t|\rvx_0).
\end{align}
\end{mdframed}
In words, if
$\rvx_s\sim p_s(\cdot|\rvx_0)$, then applying the chosen transition once
produces
$\rvx_t\sim p_t(\cdot|\rvx_0)$. Hence, the marginal distribution at each
noise level is preserved even though the dependence between
$\rvx_s$ and $\rvx_t$ may be changed.

The Bayesian reverse posterior induced by the original Markovian DDPM
forward process is one valid choice of such a transition,\footnote{For the
Markov forward process,
$\rvx_0\to\rvx_t\to\rvx_s$ with $t<s$, the conditional posterior
$p(\rvx_t|\rvx_s,\rvx_0)$ satisfies
\[
p_t(\rvx_t|\rvx_0)
=
\int
p(\rvx_t|\rvx_s,\rvx_0)
\,p_s(\rvx_s|\rvx_0)
\diff\rvx_s
\]
by the law of total probability.}
but it is not uniquely determined by the marginals. DDIM instead allows
other marginal-consistent couplings, including transitions directly between
widely separated noise levels. This is the probabilistic reason that the
same trained denoiser can be reused on a coarser sampling grid with skipped
intermediate times.

\paragraph{Derivation of Discrete-Time DDIM.}
We now make the coupling freedom above explicit for the Gaussian perturbation
model
\[
p_t(\rvx_t|\rvx_0)
:=
\mathcal{N}
\bigl(
\rvx_t;
\alpha_t\rvx_0,
\sigma_t^2\mathbf{I}
\bigr),
\qquad
\rvx_0\sim p_{\mathrm{data}}.
\]
The goal is to construct, for any $t<s$, a conditional transition
${\color{orange}\pi(\rvx_t|\rvx_s,\rvx_0)}$ that satisfies the
marginal-consistency condition in \Cref{eq:ddim-reverse-kernel-def}, without
requiring it to coincide with the Bayesian posterior associated with the
original Markov forward process.

A natural choice is the Gaussian family
\begin{align}
\label{eq:ddim-reverse-kernel-param}
{\color{orange}\pi(\rvx_t|\rvx_s,\rvx_0)}
=
\mathcal{N}
\bigl(
\rvx_t;\,
a_{t,s}\rvx_0+b_{t,s}\rvx_s,\,
c_{t,s}^2\mathbf{I}
\bigr),
\end{align}
where $(a_{t,s},b_{t,s},c_{t,s})$ determine how the state at time $s$,
the clean sample, and newly injected noise are combined.

To impose marginal consistency, first draw
\[
\rvx_s
=
\alpha_s\rvx_0+\sigma_s\beps',
\qquad
\beps'\sim\mathcal{N}(\mathbf{0},\mathbf{I}),
\]
and then draw $\rvx_t$ from
\Cref{eq:ddim-reverse-kernel-param}. Equivalently,
\begin{align}
\label{eq:ddim-derivation}
\begin{aligned}
\rvx_t
&=
a_{t,s}\rvx_0
+
b_{t,s}\rvx_s
+
c_{t,s}\beps
\\
&=
a_{t,s}\rvx_0
+
b_{t,s}
\bigl(
\alpha_s\rvx_0+\sigma_s\beps'
\bigr)
+
c_{t,s}\beps
\\
&=
\bigl(
a_{t,s}+b_{t,s}\alpha_s
\bigr)\rvx_0
+
\sqrt{
b_{t,s}^2\sigma_s^2+c_{t,s}^2
}\,
\beps'',
\end{aligned}
\end{align}
where $\beps$ and $\beps'$ are independent standard Gaussian variables, and
$\beps''$ is a standard Gaussian representing their normalized linear
combination.

For the resulting marginal to equal the prescribed perturbation marginal $\rvx_t
\sim
p_t(\rvx_t|\rvx_0)
=
\mathcal{N}
\bigl(
\rvx_t;
\alpha_t\rvx_0,
\sigma_t^2\mathbf{I}
\bigr)$, the means and variances must agree. Therefore,
\begin{align*}
\alpha_t
&=
a_{t,s}+b_{t,s}\alpha_s,
\\
\sigma_t^2
&=
b_{t,s}^2\sigma_s^2+c_{t,s}^2.
\end{align*}

The important point is that these are only two constraints on the three
coefficients $(a_{t,s},b_{t,s},c_{t,s})$. Hence, the prescribed marginals do
\emph{not} determine a unique transition between $s$ and $t$. We may choose
$c_{t,s}$ as a free parameter, with
$0\leq c_{t,s}\leq\sigma_t$, and solve for the remaining coefficients.
Choosing the nonnegative branch gives
\begin{align}
\label{eq:ddim-coefficients}
b_{t,s}
&=
\frac{
\sqrt{\sigma_t^2-c_{t,s}^2}
}{
\sigma_s
},
\qquad
a_{t,s}
=
\alpha_t-\alpha_s b_{t,s}.
\end{align}
The nonnegative choice of $b_{t,s}$ gives the standard DDIM coupling.

Substituting \Cref{eq:ddim-coefficients} into
\Cref{eq:ddim-reverse-kernel-param} yields
\begin{align}
\label{eq:ddim-reverse-kernel-result}
{\color{orange}\pi(\rvx_t|\rvx_s,\rvx_0)}
=
\mathcal{N}
\left(
\rvx_t;\,
\alpha_t\rvx_0
+
\frac{
\sqrt{\sigma_t^2-c_{t,s}^2}
}{
\sigma_s
}
\bigl(
\rvx_s-\alpha_s\rvx_0
\bigr),
\,
c_{t,s}^2\mathbf{I}
\right).
\end{align}

This expression makes the coupling freedom explicit. The term
$\rvx_s-\alpha_s\rvx_0$ is the noise component already present in
$\rvx_s$, while $c_{t,s}$ controls how much \emph{new} Gaussian noise is
introduced in the transition from $s$ to $t$. Varying $c_{t,s}$ therefore
changes the coupling between the two noise levels while preserving the same
target marginal $p_t(\rvx_t|\rvx_0)$.

\rmkb{
The chosen transition
${\color{orange}\pi(\rvx_t|\rvx_s,\rvx_0)}$
is generally different from the Bayesian posterior associated with the
original Markov forward process. That posterior is recovered by one
particular choice of $c_{t,s}$, discussed below. Other choices define
different marginal-consistent couplings, including the deterministic DDIM
coupling obtained when $c_{t,s}=0$.
}

\paragraph{DDIM Sampler (Step $s\to t$).}
The marginal-consistent kernel
${\color{orange}\pi(\rvx_t|\rvx_s,\rvx_0)}$ in
\Cref{eq:ddim-reverse-kernel-result} still conditions on the unknown clean
sample $\rvx_0$. During generation, DDIM replaces $\rvx_0$ by the clean-data
estimate obtained from the pre-trained diffusion model. Under
$\beps$-prediction, we define
\[
\rvx_{\bm{\phi}^\times}(\rvx_s,s)
:=
\frac{
\rvx_s
-
\sigma_s\beps_{\bm{\phi}^\times}(\rvx_s,s)
}{
\alpha_s
},
\]
and use the plug-in transition
\[
p_{\bm{\phi}^\times}(\rvx_t|\rvx_s)
:=
{\color{orange}\pi
\bigl(
\rvx_t
\,\big|\,
\rvx_s,
\rvx_{\bm{\phi}^\times}(\rvx_s,s)
\bigr)}.
\]
Substituting this estimate into
\Cref{eq:ddim-reverse-kernel-result} yields
\begin{mdframed}
\[
\rvx_t
=
\frac{\alpha_t}{\alpha_s}\rvx_s
+
\left(
\sqrt{\sigma_t^2-c_{t,s}^2}
-
\frac{\alpha_t}{\alpha_s}\sigma_s
\right)
\beps_{\bm{\phi}^\times}(\rvx_s,s)
+
c_{t,s}\beps_t,
\qquad
\beps_t\sim\mathcal{N}(\mathbf{0},\mathbf{I}).
\]
\end{mdframed}
Thus, the model prediction determines the estimated clean/noise content
carried from $s$ to $t$, while the free coefficient
$c_{t,s}\in[0,\sigma_t]$ controls how much \emph{new} Gaussian noise is
injected during the step.

To interpret this degree of freedom, it is useful to compare with the
original Markov forward process. For $t<s$, define
\[
\alpha_{s|t}
:=
\frac{\alpha_s}{\alpha_t},
\qquad
\sigma_{s|t}^2
:=
\sigma_s^2-\alpha_{s|t}^2\sigma_t^2,
\]
so that
\[
p(\rvx_s|\rvx_t)
=
\mathcal{N}
\bigl(
\rvx_s;
\alpha_{s|t}\rvx_t,
\sigma_{s|t}^2\mathbf I
\bigr).
\]
Gaussian conditioning gives the corresponding posterior variance
\[
\operatorname{Var}
\bigl(
\rvx_t|\rvx_s,\rvx_0
\bigr)
=
\frac{\sigma_t^2}{\sigma_s^2}
\sigma_{s|t}^2\mathbf I.
\]
This identifies the DDPM posterior as one particular member of the
marginal-consistent family above and makes the role of $c_{t,s}$ explicit:

\begin{itemize}
\item \textbfs{DDPM Step (Posterior Variance).}
Choosing
\[
c_{t,s}
=
\frac{\sigma_t}{\sigma_s}\sigma_{s|t}
\]
matches the posterior variance of the original Markov forward process.
Together with the nonnegative branch in
\Cref{eq:ddim-coefficients}, the conditional kernel
${\color{orange}\pi(\rvx_t|\rvx_s,\rvx_0)}$ then coincides with the
Bayesian posterior $p(\rvx_t|\rvx_s,\rvx_0)$ of that process.

\item \textbfs{Deterministic DDIM ($\eta=0$).}
At the opposite extreme, setting $c_{t,s}=0$ removes all newly injected
noise. The practical update becomes
\[
\rvx_t
=
\frac{\alpha_t}{\alpha_s}\rvx_s
+
\left(
\sigma_t
-
\frac{\alpha_t}{\alpha_s}\sigma_s
\right)
\beps_{\bm{\phi}^\times}(\rvx_s,s),
\]
which is exactly the deterministic DDIM update derived earlier from the
PF-ODE.

At the conditional level, its shared-noise interpretation is especially
simple: if
$\rvx_s=\alpha_s\rvx_0+\sigma_s\beps$, then the deterministic transition
gives
\[
\rvx_t
=
\alpha_t\rvx_0+\sigma_t\beps.
\]
Thus, no new noise is introduced; the same pair $(\rvx_0,\beps)$ is carried
across the step while only the known schedule coefficients change.

\item \textbfs{Interpolation.}
More generally, define
\begin{align}
\label{eq:eta-ddim}
c_{t,s}
=
\eta\,
\frac{\sigma_t}{\sigma_s}\,
\sigma_{s|t},
\qquad
\eta\in[0,1].
\end{align}
The parameter $\eta$ continuously controls the stochasticity of the
transition: $\eta=0$ gives deterministic DDIM, whereas $\eta=1$ matches the
posterior variance of the corresponding DDPM Markov transition.
\end{itemize}

Hence, the DDIM family separates two ingredients that are fixed differently:
the perturbation marginals are kept unchanged, while the coupling between
successive noise levels can be varied through $c_{t,s}$. Deterministic DDIM
corresponds to preserving the existing shared noise across the step, whereas
the DDPM posterior choice injects the amount of additional randomness
prescribed by the original Markov forward process.

\paragraph{Connection to Conditional Flow Matching.}
The deterministic DDIM construction above also provides a simple connection
to conditional flow matching (CFM) in \Cref{sec:flow-matching-framework}. The key is to distinguish two objects
associated with the same conditional Gaussian path: its \emph{finite-time
map} and its \emph{instantaneous velocity}. DDIM is related to the former,
whereas CFM is built from the latter.

Condition on a clean sample $\rvx_0$ and consider the shared-noise path
\[
\rvx_\tau
=
\alpha_\tau\rvx_0
+
\sigma_\tau\beps.
\]
For fixed $(\rvx_0,\beps)$, this is an exact conditional trajectory through
the Gaussian perturbation marginals
$p_\tau(\rvx_\tau|\rvx_0)$.

\textbfs{Finite-Time View: the Conditional DDIM Map.}
Suppose the trajectory is at $\rvx_s$ at time $s$. Since the same
$\beps$ is retained, eliminating it from
$\rvx_s=\alpha_s\rvx_0+\sigma_s\beps$ gives the exact conditional map
\[
\bPsi_{s\to t}(\rvx_s|\rvx_0)
=
\frac{\sigma_t}{\sigma_s}\rvx_s
+
\left(
\alpha_t
-
\alpha_s\frac{\sigma_t}{\sigma_s}
\right)\rvx_0.
\]
This map is exact \emph{conditioned on $\rvx_0$}. It is not yet the practical
DDIM sampler, because $\rvx_0$ is unknown during generation. Replacing
$\rvx_0$ by its start-time model prediction gives precisely the deterministic
DDIM update derived above.

\textbfs{Infinitesimal-Time View: the CFM Conditional Velocity.}
Instead of asking where the conditional path moves over a finite interval,
we may ask for its instantaneous direction. Differentiating
$\rvx_t=\alpha_t\rvx_0+\sigma_t\beps$ gives
\[
\frac{\diff\rvx_t}{\diff t}
=
\alpha_t'\rvx_0+\sigma_t'\beps.
\]
Using
\[
\beps
=
\frac{\rvx-\alpha_t\rvx_0}{\sigma_t},
\]
the same velocity can be written as a function of the current state:
\[
\rvv_t^*(\rvx|\rvx_0)
=
\frac{\sigma_t'}{\sigma_t}\rvx
+
\left(
\alpha_t'
-
\alpha_t\frac{\sigma_t'}{\sigma_t}
\right)\rvx_0.
\]
This is the conditional velocity associated with the Gaussian probability
path in conditional flow matching.

CFM then connects this conditional velocity to the marginal dynamics by
averaging over the posterior uncertainty about $\rvx_0$ at the current
state:
\begin{align*}
\rvv^*(\rvx,t)
&:=
\E_{\rvx_0\sim p(\rvx_0|\rvx_t=\rvx)}
\left[
\rvv_t^*(\rvx|\rvx_0)
\right]
\\
&=
\frac{\sigma_t'}{\sigma_t}\rvx
+
\left(
\alpha_t'
-
\alpha_t\frac{\sigma_t'}{\sigma_t}
\right)
\rvx^*(\rvx,t)
\\
&=
\alpha_t'\rvx^*(\rvx,t)
+
\sigma_t'\beps^*(\rvx,t),
\end{align*}
where we used
\[
\rvx^*(\rvx,t)
=
\E[\rvx_0|\rvx_t=\rvx],
\qquad
\rvx
=
\alpha_t\rvx^*(\rvx,t)
+
\sigma_t\beps^*(\rvx,t).
\]
The resulting $\rvv^*(\rvx,t)$ is exactly the oracle marginal PF-ODE
velocity.

\msgu{DDIM and CFM}{}{
The same conditional Gaussian path gives two views:
DDIM uses its finite-time map after replacing the unknown $\rvx_0$ by a
model prediction, while CFM uses its instantaneous conditional velocity.
Posterior averaging the latter gives the marginal PF-ODE velocity.
}

This connection is exact at the level of the conditional path and its
instantaneous velocity. It does \emph{not} imply that a finite DDIM step is
the exact marginal PF-ODE flow: DDIM keeps the prediction inferred at the
beginning of the step fixed, whereas the exact marginal PF-ODE velocity
changes continuously with the evolving state.

\clearpage
\newpage

\section{\texorpdfstring{DEIS}{DEIS}}\label{sec:exponential}

In the factored exponential-integrator formula
(\Cref{eq:variation_of_constants}),
\[
\int_s^t
\frac{g^2(\tau)}{2\sigma_\tau}\,
\mathcal E(\tau\shortrightarrow t)\,
\beps_{\bm{\phi}^\times}(\rvx_\tau,\tau)
\diff\tau,
\]
all schedule-dependent factors are known; the only unknown quantity along the
trajectory is the neural prediction
$\beps_{\bm{\phi}^\times}(\rvx_\tau,\tau)$. DDIM makes the simplest
first-order approximation by holding this prediction fixed at the beginning of
the step,
\[
\beps_{\bm{\phi}^\times}(\rvx_\tau,\tau)
\approx
\beps_{\bm{\phi}^\times}(\rvx_s,s),
\qquad
\tau\in[t,s].
\]
This approximation can become inaccurate over larger steps when the neural
prediction varies appreciably along the trajectory.

DEIS~\citep{zhang2022fast} improves on this idea without introducing
additional model evaluations within the current step. Inspired by classical
multistep methods such as Adams--Bashforth
(see \Cref{app:de-solver-examples}), it reuses model predictions from previous
completed steps as \emph{anchors} and fits a polynomial that extrapolates
their recent variation into the next step. The polynomial then replaces the
unknown neural prediction inside the integral, while the known
schedule-dependent weight
$\tfrac{g^2(\tau)}{2\sigma_\tau}
\mathcal E(\tau\shortrightarrow t)$
is retained analytically. Thus, DEIS extends the constant approximation of
DDIM to a higher-order multistep polynomial approximation.

From the classical numerical viewpoint, DEIS is therefore an
Adams--Bashforth-type multistep method within the factored
exponential-integrator framework: previous neural evaluations provide the
higher-order information, while the known diffusion dynamics are retained in
the weighted integral.

We first review the polynomial extrapolation used to construct this
approximation in \Cref{subsec:poly-extrapolation}, then derive the resulting
DEIS update in \Cref{subsec:deis-lagrange}. Finally,
\Cref{sec:ddim-deis} shows that the constant, degree-zero case recovers DDIM.

\subsection{Background: Polynomial Extrapolation}
\label{subsec:poly-extrapolation}

Before deriving DEIS, we briefly review the polynomial construction it uses
to reuse previous model evaluations. The idea is simple: first fit a
polynomial through several known values, and then evaluate that polynomial
beyond the most recent anchor to predict how the quantity may continue to
change. Thus, the polynomial is an \emph{interpolant} at the known anchors
and acts as an \emph{extrapolant} over the next solver step.

\paragraph{From Anchors to a Polynomial Extrapolant.}
Suppose a time-varying quantity is known at $n{+}1$ distinct times,
\[
(\tau_0,\rmY_0),\;
(\tau_1,\rmY_1),\;
\ldots,\;
(\tau_n,\rmY_n),
\qquad
\tau_0<\tau_1<\cdots<\tau_n,
\]
where each $\rmY_j$ may be vector-valued. We would like to construct the
simplest smooth curve that exactly matches these observed values and can then
be continued beyond the most recent anchor. The unique polynomial of degree at
most $n$ passing through all anchors is the \emph{Lagrange interpolant}
\[
P_n(\tau)
=
\sum_{j=0}^{n}
\ell_j(\tau)\,\rmY_j,
\]
with basis functions
\[
\ell_j(\tau)
:=
\prod_{\substack{k=0\\k\neq j}}^{n}
\frac{\tau-\tau_k}{\tau_j-\tau_k}.
\]
The construction of $\ell_j$ is simple: every factor makes it vanish at one
of the other anchors, while the denominator normalizes its value to $1$ at
$\tau_j$. Hence,
\[
\ell_j(\tau_k)=\delta_{jk},
\]
and therefore
\[
P_n(\tau_j)=\rmY_j,
\qquad
j=0,\ldots,n.
\]
Moreover,
\[
\sum_{j=0}^{n}\ell_j(\tau)=1.
\]
Thus, $P_n(\tau)$ is a time-dependent linear combination of the observed
anchors: at each anchor time it reproduces exactly the corresponding value,
while between or beyond the anchors the polynomial combines them to describe
the local trend suggested by their variation. In DEIS, we will use this same
curve beyond the most recent anchor, turning the interpolant into an
\emph{extrapolant} for the next solver step.

The lowest-order cases illustrate the construction:

\exm{}{
\noindent\textbfs{$n=0$ (Constant).}
With one anchor, the polynomial simply keeps the most recent value fixed:
\[
P_0(\tau)=\rmY_0.
\]

\medskip
\noindent\textbfs{$n=1$ (Linear).}
With two anchors, the polynomial is the straight line joining them:
\[
P_1(\tau)
=
\frac{\tau-\tau_1}{\tau_0-\tau_1}\rmY_0
+
\frac{\tau-\tau_0}{\tau_1-\tau_0}\rmY_1.
\]

\medskip
\noindent\textbfs{$n=2$ (Quadratic).}
With three anchors, the interpolant becomes
\[
P_2(\tau)
=
\ell_0(\tau)\rmY_0
+
\ell_1(\tau)\rmY_1
+
\ell_2(\tau)\rmY_2,
\]
which can additionally capture local curvature in the observed values.
}

In DEIS, the anchors will be neural predictions from previous completed
sampling steps. The polynomial matches those stored predictions exactly and
is then extrapolated beyond the most recent anchor over the next reverse-time
interval.

\subsection{DEIS: Lagrange Polynomial Approximation of the PF-ODE Integral}
\label{subsec:deis-lagrange}

Let $n \ge 0$ be the chosen polynomial degree. At step $i$, DEIS approximates
the unknown map
$\tau \mapsto \beps_{\bphi^\times}(\rvx_\tau,\tau)$ over the next
reverse-time interval $[t_i,t_{i-1}]$ using a degree-$n$ Lagrange polynomial
constructed from past model outputs. This polynomial is then substituted into
the exponential--integrator update (\Cref{eq:variation_of_constants}) to
obtain $\tilde{\rvx}_{t_i}$. The intuition is that, instead of holding the
neural prediction constant as in DDIM, DEIS uses its recent history to capture
how it is changing over time.

\begin{figure}[th!]
    \centering
    \includegraphics[width=0.85\linewidth]{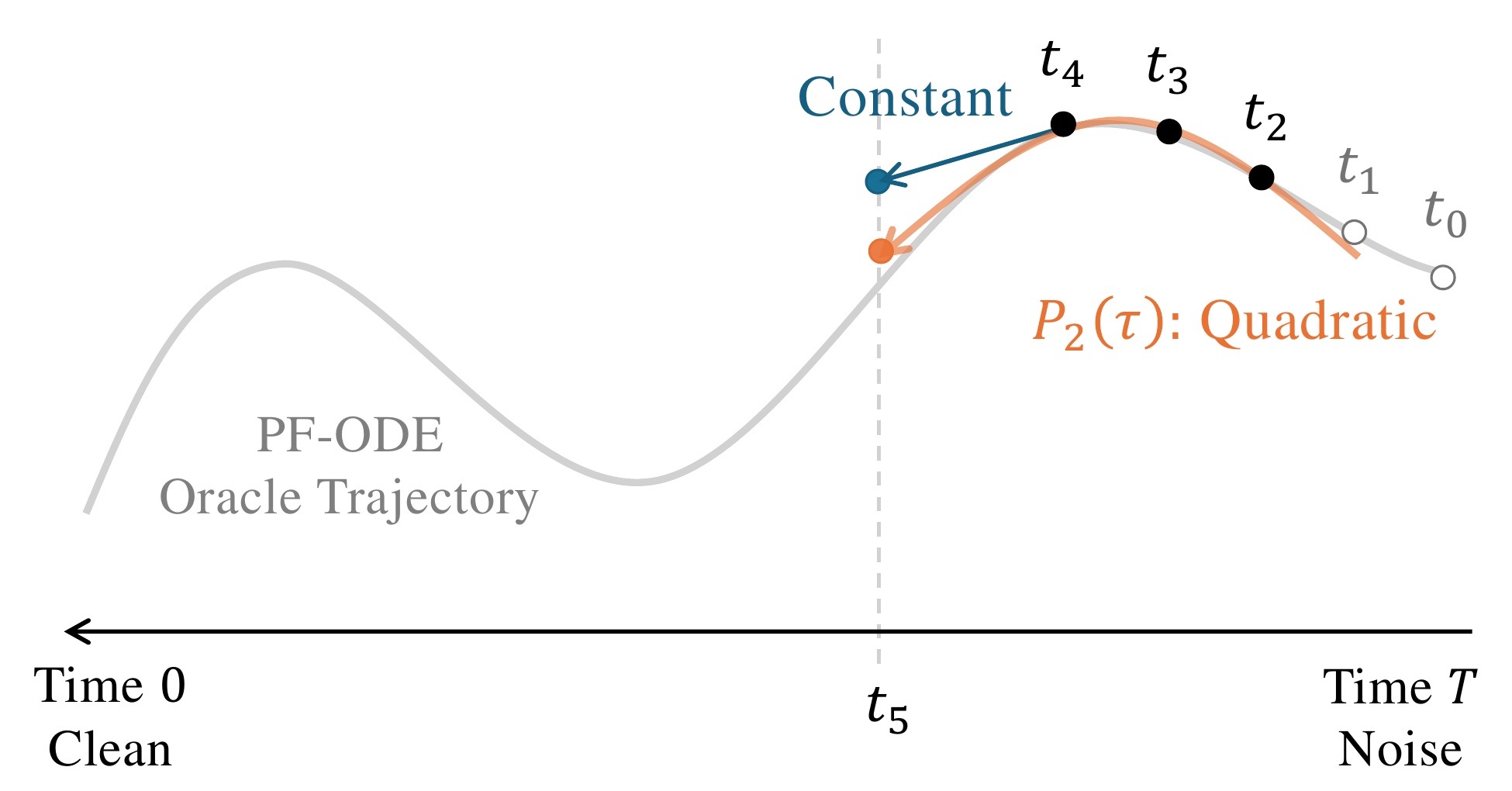}
    \caption{\textbfs{Illustration of DEIS as a multistep method.} With three past anchors at
$t_2,t_3,t_4$, DEIS builds a quadratic curve through the model outputs
and extrapolates it over the step from
$t_4$ to $t_5$. The extrapolating polynomial is then integrated against the
known diffusion-dependent weight, capturing variation of the neural prediction
that is neglected by the constant approximation used by DDIM.
\figcredit{Created by the authors.}}
    \label{fig:deis_lagrange}
\end{figure}

A degree-$n$ update needs $n{+}1$ anchors. When they are available
(\emph{sufficient history}, $i \ge n{+}1$), we use the full degree-$n$
scheme. During the initial steps (\emph{insufficient history}, $i \le n$),
we use the highest feasible degree, $i{-}1$, and increase it as more anchors
become available. We treat these two cases below.

\paragraph{Case I: $i = n+1,\dots,M$ (Sufficient History).}
With sufficient history, DEIS reuses the last $n{+}1$ model evaluations as
anchors,
\[
(\tau_j,\rmY_j) := \bigl(t_{i-1-j}, \bm{\epsilon}_{\bm{\phi}^\times}(\tilde{\rvx}_{t_{i-1-j}},t_{i-1-j})\bigr),
\quad j=0,\dots,n.
\]
Viewing
$\tau\mapsto\bm{\epsilon}_{\bm{\phi}^\times}(\rvx_\tau,\tau)$
as a smooth function along the trajectory, we construct the degree-$n$
Lagrange polynomial
\[
P_n(\tau)
= 
\sum_{j=0}^{n}
\underbrace{\Bigl[\prod_{\substack{k=0\\k\neq j}}^{n}
\frac{\tau - t_{i-1-k}}{\,t_{i-1-j} - t_{i-1-k}\,}\Bigr]}_{=:~\ell_j^{(i)}(\tau)}
\bm{\epsilon}_{\bm{\phi}^\times}\bigl(\tilde{\mathbf x}_{t_{i-1-j}}, t_{i-1-j}\bigr)
\]
which by construction satisfies $P_n(\tau_j)=\rmY_j$ for each anchor:
\[
P_n\bigl(\tau_j\bigr)
= \rmY_j =
\bm{\epsilon}_{\bm{\phi}^\times}\bigl(\tilde{\mathbf x}_{t_{i-1-j}}, t_{i-1-j}\bigr),
\quad
 j=0,\ldots,n.
\]
Each $\ell_j^{(i)}$ satisfies
\[
\ell_j^{(i)}\bigl(t_{i-1-m}\bigr)
=
\begin{cases}
1,&m=j,\\
0,&m\neq j.
\end{cases}
\]

The polynomial is then extrapolated beyond the most recent anchor
$t_{i-1}$ over the new step:
\[
\bm{\epsilon}_{\bm{\phi}^\times}(\rvx_\tau,\tau) \approx
P_n(\tau)=\sum_{j=0}^{n}\ell_j^{(i)}(\tau) 
\bm{\epsilon}_{\bm{\phi}^\times}\bigl(\tilde{\rvx}_{t_{i-1-j}}, t_{i-1-j}\bigr), 
\quad \tau \in [t_i,t_{i-1}].
\]
We then substitute $P_n(\tau)$ for
$\beps_{\bm{\phi}^\times}(\rvx_\tau,\tau)$ in the
exponential--integrator formula (\Cref{eq:variation_of_constants}):
\begin{align*}
\int_{t_{i-1}}^{t_i}
\frac{g^2(\tau)}{2\sigma_\tau}
\mathcal{E}(\tau\to t_i)
\beps_{\bm{\phi}^\times}(\rvx_\tau,\tau)\diff\tau
&\approx
\sum_{j=0}^{n}
\underbrace{
\int_{t_{i-1}}^{t_i}
\frac{g^2(\tau)}{2\sigma_\tau}
\mathcal{E}(\tau\to t_i)
\ell_j^{(i)}(\tau)\diff\tau
}_{=:~C_{i,j}}
\\
&\qquad\cdot
\beps_{\bm{\phi}^\times}
(\tilde\rvx_{t_{i-1-j}},t_{i-1-j}).
\end{align*}
The coefficients
\[
C_{i,j}
:=
\frac{1}{2}
\int_{t_{i-1}}^{t_i}
\frac{g^2(\tau)}{\sigma_\tau}
\mathcal{E}(\tau\to t_i)
\ell_j^{(i)}(\tau)\diff\tau
\]
depend only on the noise schedule and the chosen time grid and can therefore
be precomputed. For schedules admitting suitable antiderivatives, these
coefficients are available in closed form; otherwise they can be evaluated
numerically to high accuracy.

Integrating the linear part exactly with
$\mathcal{E}(t_{i-1}\to t_i)$ gives the \emph{AB-DEIS-$n$} update
\[
\tilde\rvx_{t_i}
=
\mathcal{E}(t_{i-1}\to t_i)\tilde\rvx_{t_{i-1}}
+
\sum_{j=0}^{n}
C_{i,j}
\beps_{\bm{\phi}^\times}
(\tilde\rvx_{t_{i-1-j}},t_{i-1-j}).
\]
For the local truncation statements below, let $h$ denote a characteristic
local step size and assume neighboring step ratios remain bounded. Under
standard smoothness assumptions on the model prediction along the empirical
trajectory and regularity of the schedule-dependent weight, the full-history
degree-$n$ construction has formal order $n+1$, with one-step local defect
$\mathcal O(h^{n+2})$.

\paragraph{Case II: $i=1,\dots,n$ (Insufficient History).}
During the initial steps, only $i$ past model evaluations are available.
We therefore use all available anchors and set the polynomial degree to
$i{-}1$:
\[
P_{i-1}(\tau)
=
\sum_{j=0}^{i-1}
\ell_j^{(i)}(\tau) 
\bm{\epsilon}_{\bm{\phi}^\times}\bigl(\tilde{\mathbf x}_{t_{i-1-j}},  t_{i-1-j}\bigr),
\]
where $\ell_j^{(i)}$ is the Lagrange basis of degree $i{-}1$ built on the nodes at time
$\{t_{i-1}, t_{i-2},\ldots,t_{0}\}$. This matches all available anchors and
transitions into the full-history formula once $i\ge n{+}1$.

This is the standard warm-start situation for a multistep method. The first
step uses a constant approximation, the second uses a linear extrapolation,
the third uses a quadratic extrapolation, and so on, until the target degree
$n$ becomes available.

\exm{Special Cases of Lagrange Polynomials}{
\noindent \textbfs{Degree $0$ (one anchor):}
\[
P_0(\tau) = \bm{\epsilon}_{\bm{\phi}^\times}(\tilde\rvx_{t_{i-1}}, t_{i-1}).
\]
This holds the neural prediction fixed at its start-of-step value, exactly as
in the first-order DDIM approximation.

\medskip
\noindent \textbfs{Degree $1$ (two anchors):} the Lagrange polynomial is a linear map passing through the two pre-specified anchors.
\[
P_1(\tau)
= 
\underbrace{\frac{\tau - t_{i-2}}{t_{i-1} - t_{i-2}}}_{\displaystyle \ell_{i-1}(\tau)}
\,\bm{\epsilon}_{\bm{\phi}^\times}(\tilde\rvx_{t_{i-1}}, t_{i-1})
 + 
\underbrace{\frac{\tau - t_{i-1}}{t_{i-2} - t_{i-1}}}_{\displaystyle \ell_{i-2}(\tau)}
\,\bm{\epsilon}_{\bm{\phi}^\times}(\tilde\rvx_{t_{i-2}}, t_{i-2}).
\]
Here $\ell_{i-1}(\tau)$ and $\ell_{i-2}(\tau)$ are the Lagrange basis weights.
They satisfy the interpolation conditions
$P_1(t_{i-1})=\bm{\epsilon}_{\bm{\phi}^\times}(\tilde\rvx_{t_{i-1}}, t_{i-1})$ and
$P_1(t_{i-2})=\bm{\epsilon}_{\bm{\phi}^\times}(\tilde\rvx_{t_{i-2}}, t_{i-2})$,
with
$\ell_{i-1}(\tau)+\ell_{i-2}(\tau)=1$. 
}

\paragraph{Summary of AB-DEIS-$n$ Update.}
Combining the sufficient-history and warm-start cases gives the
\emph{AB-DEIS-$n$} update\footnote{``AB'' refers to the Adams--Bashforth
multistep principle, here applied within the exponential-integrator
formulation~\citep{hochbruck2010exponential}.} Here, $n$ is the target
polynomial degree, using up to $n{+}1$ past evaluations:
\begin{mdframed}
\begin{align*}
\tilde\rvx_{t_i}
= \mathcal{E}(t_{i-1} \to t_i)\,\tilde\rvx_{t_{i-1}}
+ \sum_{j=0}^{\min\{n,\, i-1\}} C_{i,j}\,\beps_{\bm{\phi}^\times}(\tilde\rvx_{t_{i-1-j}},\,t_{i-1-j}),
\end{align*}
with coefficients
\begin{align*}
C_{i,j}
:= \frac{1}{2} \int_{t_{i-1}}^{t_i}
\frac{g^2(\tau)}{\sigma_\tau}\,\mathcal{E}(\tau \to t_i)\,
\Bigg[\prod_{\substack{k=0\\k\neq j}}^{\min\{n,\, i-1\}}
\frac{\tau - t_{i-1-k}}{\,t_{i-1-j} - t_{i-1-k}\,}\Bigg]\,d\tau.
\end{align*}
\end{mdframed}

When $i\ge n{+}1$ (sufficient history), $\min\{n,\,i-1\}=n$, and the
full-history degree-$n$ step has one-step local truncation error
$\mathcal O(h^{n+2})$ under the assumptions above, corresponding
to a method of order $n+1$.

During warm start ($i\le n$), the degree is $i-1$, so the
one-step local truncation error is $\mathcal O(h^{i+1})$, and the corresponding
step is formally of order $i$. Thus, the order increases from $1$ at the first
step and reaches the full order $n+1$ once $n+1$ anchors are available.
As with standard multistep methods, lower-order warm-start steps do not by
themselves establish a global order-$(n+1)$ guarantee; such a guarantee also
requires stability and starting values of sufficient accuracy.

Very large $n$ can degrade numerical behavior because high-degree
extrapolation becomes increasingly sensitive to the spacing of the anchors,
errors in the stored model predictions, and stability considerations.
Relatively small polynomial degrees are therefore preferred in practice.

As we will see in the following subsection, the special case $n{=}0$ reduces
exactly to deterministic DDIM, i.e., the first-order factored
exponential-integrator update.
\subsection{DDIM = AB-DEIS-0}
\label{sec:ddim-deis}

The connection to DDIM becomes immediate at degree $n=0$. In this case, the
Lagrange polynomial is constant, so DEIS uses only the most recent model
prediction throughout the next step. The corresponding coefficient simplifies
to
\[
C_{i0} = \frac{1}{2} \int_{t_{i-1}}^{t_{i}} \frac{g^2(\tau)}{\sigma_\tau} \mathcal{E}(\tau \shortrightarrow t_i)   \mathrm{d}\tau.
\]
Substituting into the update formula yields the zeroth-order AB-DEIS scheme:
\begin{align}\label{eq:deis-0}
  \tilde\rvx_{t_{i}} &=  \mathcal{E}(t_{i-1} \shortrightarrow t_{i})\tilde\rvx_{t_{i-1}} + C_{i0} \bm{\epsilon}_{\bm{\phi}^\times}(\tilde\rvx_{t_{i-1}}, t_{i-1})\nonumber\\
  &=e^{\int_{t_{i-1}}^{t_{i}}f(u)\diff u} \tilde\rvx_{t_{i-1}} + \Big( \int_{t_{i-1}}^{t_{i}} \frac{g^2(\tau)}{2\sigma_\tau} e^{\int_{\tau}^{t_{i}}f(u)\diff u}   \mathrm{d}\tau
\Big) \bm{\epsilon}_{\bm{\phi}^\times}(\tilde\rvx_{t_{i-1}}, t_{i-1}).  
\end{align}
This is exactly the same first-order factored approximation used by
deterministic DDIM: the neural prediction is held fixed at
$t_{i-1}$, while the known integrating factor and schedule-dependent
coefficient remain inside the integral and are propagated over the full step.
Hence, AB-DEIS-0 and DDIM are not merely first-order equivalent; they give
the same finite-step update.

\proppnp{DDIM = AB-DEIS-0}{ddim-deis}{\Cref{eq:deis-0} is identical to the DDIM update in \Cref{eq:ddim-euler}.}

This also clarifies the role of DEIS in the solver hierarchy. We remark that the index
``$0$'' refers to the \emph{polynomial degree}, not the numerical order:
AB-DEIS-0 is still a first-order method. Higher-degree DEIS keeps the same
analytic treatment of the known diffusion dynamics but replaces this constant
prediction by polynomial extrapolation from additional past model evaluations.
Thus, DDIM is the degree-zero base case from which the DEIS multistep
construction naturally extends.

Later, in \Cref{subsec:dpm-solver++-multi}, we will encounter another route to
a multistep diffusion solver; that is, DPM-Solver\texttt{++}(2M). It starts from a local
Taylor expansion of the data prediction in log-SNR time and recycles previous
model evaluations through a backward finite difference to estimate the required
derivative. Although its representation and derivation differ from those of
DEIS, both methods use past model evaluations to capture the variation of the
neural prediction without introducing additional within-step stages. We return
to this connection in \Cref{subsec:dpm-deis}, where we compare the two
constructions more explicitly.
    
\clearpage
\newpage

\section{\texorpdfstring{DPM-Solver}{DPM-Solver}}\label{sec:dpm}

DPM-Solver~\citep{lu2022dpm} develops higher-order solvers for the empirical
PF-ODE by exploiting the same diffusion-specific structure that appeared in
our discussion of DDIM. The central goal is to obtain accurate samples using
far fewer neural-network evaluations than a first-order sampler. The original
DPM-Solver focuses on noise prediction and constructs first-, second-, and
third-order single-step solvers; later extensions, including
DPM-Solver\texttt{++}~\citep{lu2022dpm2} and
DPM-Solver-v3~\citep{zheng2023dpm}, further develop this family.

\paragraph{High-Level Idea of DPM-Solver.}
The DDIM discussion in \Cref{sec:revisit-ddim} showed that, under
$\beps$-prediction, a first-order update can be obtained by retaining the
known diffusion schedule analytically while holding the neural prediction
fixed over each step,
\[
\beps_{\bm{\phi}^\times}(\rvx_\tau,\tau)
\approx
\beps_{\bm{\phi}^\times}(\tilde\rvx_s,s).
\]
DPM-Solver~\citep{lu2022dpm} builds naturally on this idea: rather than
treating the neural prediction as constant, it seeks higher-order
approximations of how the prediction varies along the trajectory. Like DEIS,
it starts from the semilinear $\beps$-prediction form of the PF-ODE,
\begin{align}
\label{eq:noise-alpha-sigma}
\frac{\diff \rvx_t}{\diff t}
=
\frac{\alpha_t'}{\alpha_t}\rvx_t
-
\sigma_t
\left(
\frac{\alpha_t'}{\alpha_t}
-
\frac{\sigma_t'}{\sigma_t}
\right)
\beps_{\bphi^\times}(\rvx_t,t),
\end{align}
and its variation-of-constants representation introduced in the Prologue.
Its key additional ingredient is to reparameterize time by the
\emph{half-log signal-to-noise ratio}, following the log-SNR
parameterization of VDM~\citep{kingma2021variational},
\begin{align}
\label{eq:lambda}
\lambda_t
:=
\log\frac{\alpha_t}{\sigma_t}.
\end{align}
This coordinate is directly related to DDIM's schedule-aligned clock
$\rho_t=\sigma_t/\alpha_t$ (discussed around \Cref{eq:ddim-coordinate-ode-t}) through $\rho_t=e^{-\lambda_t}$, and turns the
remaining model-dependent term into a particularly convenient exponentially
weighted integral. Local Taylor approximations of the neural prediction in
$\lambda$ can then be integrated analytically, providing a systematic route
from the first-order DDIM update to higher-order diffusion solvers.

\subsection{DPM-Solver's Insight: Time Reparameterization via Log-SNR}
\label{subsec:dpm-method}   

We now make the log-SNR reparameterization in \Cref{eq:lambda} precise and
derive the corresponding PF-ODE representation.

\paragraph{Change-of-Variable to Log-SNR in PF-ODE.}
Suppose that $\alpha_t>0$, $\sigma_t>0$, and the log-SNR coordinate
$\lambda_t$ in \Cref{eq:lambda} is strictly monotone on the interval of
interest. It then admits an inverse function $t_\lambda(\cdot)$ satisfying
\[
t=t_\lambda(\lambda(t)).
\]
For schedules with $\sigma_0=0$, one has $\lambda_t\to+\infty$ as
$t\to0$, so formulas involving the endpoint $t=0$ are understood through
the corresponding limit or through a small positive terminal time in a
numerical implementation.

We then express the state, schedule, and neural prediction in the
$\lambda$-coordinate. A hat ($\hat{\cdot}$) together with $\lambda$ indicates that the corresponding quantity is
viewed in this reparameterized time coordinate. More precisely, define
\begin{align}
\begin{aligned}
\label{eq:change-of-var-x-epsilon}
\hat{\rvx}_{\lambda}
&:=
\rvx_{t_\lambda(\lambda)},
\\
\alpha_\lambda
&:=
\alpha_{t_\lambda(\lambda)},
\qquad
\sigma_\lambda
:=
\sigma_{t_\lambda(\lambda)},
\\
\hat{\bm{\epsilon}}_{\bm{\phi}^\times}(\rvx,\lambda)
&:=
\bm{\epsilon}_{\bm{\phi}^\times}
\bigl(
\rvx,
t_\lambda(\lambda)
\bigr).
\end{aligned}
\end{align}
Hence, along the empirical PF-ODE trajectory,
\[
\hat{\bm{\epsilon}}_{\bm{\phi}^\times}
(\hat\rvx_\lambda,\lambda)
=
\bm{\epsilon}_{\bm{\phi}^\times}
\bigl(
\rvx_{t_\lambda(\lambda)},
t_\lambda(\lambda)
\bigr).
\]
Since
\[
\frac{\diff\lambda_t}{\diff t}
=
\frac{\alpha_t'}{\alpha_t}
-
\frac{\sigma_t'}{\sigma_t},
\]
the PF-ODE in \Cref{eq:noise-alpha-sigma} can be written as
\[
\frac{\diff\rvx_t}{\diff t}
=
\frac{\diff\log\alpha_t}{\diff t}\rvx_t
-
\sigma_t
\frac{\diff\lambda_t}{\diff t}
\bm{\epsilon}_{\bm{\phi}^\times}(\rvx_t,t).
\]
Applying the chain rule,
\[
\frac{\diff\rvx_t}{\diff t}
=
\frac{\diff\hat\rvx_\lambda}{\diff\lambda}
\frac{\diff\lambda_t}{\diff t},
\qquad
\frac{\diff\log\alpha_t}{\diff t}
=
\frac{\diff\log\alpha_\lambda}{\diff\lambda}
\frac{\diff\lambda_t}{\diff t},
\]
gives the PF-ODE in log-SNR time:
\begin{mdframed}
\begin{align}
\label{eq:diffusion_ode_lambda_eps}
\frac{\diff\hat{\rvx}_\lambda}{\diff\lambda}
=
\frac{\diff\log\alpha_\lambda}{\diff\lambda}
\hat{\rvx}_\lambda
-
\sigma_\lambda
\hat{\bm{\epsilon}}_{\bm{\phi}^\times}
(\hat{\rvx}_\lambda,\lambda).
\end{align}
\end{mdframed}

Applying the same exponential-integrator (variation-of-constants) technique
to \Cref{eq:diffusion_ode_lambda_eps} gives the following exact solution.

\proppp{Exponentially Weighted Exact Solution}{exp-wgt-exact-sol}{
Given an initial value $\rvx_s$ at time $s>0$, the exact empirical PF-ODE
solution $\widetilde\bPsi_{s\to t}(\rvx_s)$ at time $t\in(0,s]$ can be
re-expressed as
\begin{equation}
\label{eq:analytic_solution}
\widetilde\bPsi_{s\to t}(\rvx_s)
=
\frac{\alpha_t}{\alpha_s}\rvx_s
-
\alpha_t
\int_{\lambda_s}^{\lambda_t}
e^{-\lambda}
\hat{\bm{\epsilon}}_{\bm{\phi}^\times}
(\hat\rvx_\lambda,\lambda)
\diff\lambda.
\end{equation}
When the endpoint satisfies $\sigma_0=0$, the case $t=0$ is understood
through the corresponding limit.
}{
From \Cref{eq:diffusion_ode_lambda_eps}, the integrating factor associated
with the linear term satisfies
\[
\exp\left(
\int_{\lambda_s}^{\lambda_t}
\frac{\diff\log\alpha_\lambda}{\diff\lambda}
\diff\lambda
\right)
=
\frac{\alpha_t}{\alpha_s}.
\]
Therefore, the variation-of-constants formula gives
\begin{align*}
\widetilde\bPsi_{s\to t}(\rvx_s)
&=
\frac{\alpha_t}{\alpha_s}\rvx_s
-
\int_{\lambda_s}^{\lambda_t}
\frac{\alpha_t}{\alpha_\lambda}
\sigma_\lambda
\hat{\bm{\epsilon}}_{\bm{\phi}^\times}
(\hat\rvx_\lambda,\lambda)
\diff\lambda
\\
&=
\frac{\alpha_t}{\alpha_s}\rvx_s
-
\alpha_t
\int_{\lambda_s}^{\lambda_t}
e^{-\lambda}
\hat{\bm{\epsilon}}_{\bm{\phi}^\times}
(\hat\rvx_\lambda,\lambda)
\diff\lambda,
\end{align*}
where the last equality follows from
$\sigma_\lambda/\alpha_\lambda=e^{-\lambda}$.
}

The key numerical consequence is now explicit. In $\lambda$-time, the only
model-dependent quantity in the exact solution appears inside
\[
\int_{\lambda_s}^{\lambda_t}
e^{-\lambda}
\hat{\bm{\epsilon}}_{\bm{\phi}^\times}
(\hat\rvx_\lambda,\lambda)\diff\lambda.
\]
When the model output is locally approximated by a Taylor polynomial in
$\lambda$, the resulting exponentially weighted polynomial terms can be
integrated in closed form, yielding explicit higher-order update
coefficients. Thus, as in the factored viewpoint of the Prologue, the known
schedule dependence is retained analytically while the approximation is
concentrated on the neural prediction.

\paragraph{Intuition of Why Reparameterize Time?}
For strictly monotone $\lambda(t)$, a local first-order change of variables gives
\[
|\Delta t|
\approx
\frac{|\Delta\lambda|}{|\lambda'(t)|}.
\]
Thus, if the numerical grid uses a fixed step size $|\Delta\lambda|$, the
induced step in $t$ is smaller where $|\lambda'(t)|$ is larger
(i.e., where $\lambda$ changes rapidly with $t$), and larger where
$|\lambda'(t)|$ is smaller.

The reparameterization itself does not alter the PF-ODE solution path; it only
changes how the same path is parameterized:
\[
\frac{\diff \hat\rvx_\lambda}{\diff \lambda}
=
\frac{1}{\lambda'(t)}
\frac{\diff \rvx_t}{\diff t}.
\]
Hence, where $|\lambda'(t)|$ is large, the same change of the trajectory with
respect to $t$ is represented over a larger range of $\lambda$.

More importantly, $\lambda(t)$ measures progress through the diffusion process
directly in terms of the relative balance between signal and noise. A fixed
$\Delta\lambda$ corresponds to a fixed multiplicative change in
$\alpha_t/\sigma_t$, or equivalently to a fixed additive change in log-SNR,
whereas the same $\Delta t$ may correspond to very different changes in
corruption level at different times. Therefore, choosing numerical nodes
uniformly in $\lambda$ provides a natural way to redistribute resolution along
the diffusion trajectory. The precise induced spacing in the original time
variable $t$ depends on the chosen noise schedule. Two simple examples
illustrate this effect:

\begin{itemize}
\item \textbfs{$(\alpha_t,\sigma_t)=(1-t,t)$.}
This corresponds to the linear interpolation commonly used in flow matching.
Then
\[
\lambda(t)
=
\log\frac{1-t}{t},
\qquad
\lambda'(t)
=
-\frac{1}{t(1-t)},
\]
so locally
\[
|\Delta t|
\approx
|\Delta\lambda|\,t(1-t).
\]
Hence, a uniform grid in $\lambda$ induces very small steps in $t$ near both
endpoints $t\to0$ and $t\to1$, and larger steps around the middle of the
trajectory.

\item \textbfs{$(\alpha_t,\sigma_t)=(1,t)$.}
This corresponds to the EDM parameterization~\citep{karras2022elucidating}
when the independent variable $t$ is taken directly as the noise level
$\sigma_t$. Then
\[
\lambda(t)
=
\log\frac{1}{t},
\qquad
\lambda'(t)
=
-\frac{1}{t},
\]
and therefore
\[
|\Delta t|
\approx
|\Delta\lambda|\,t.
\]
In fact, uniform spacing in $\lambda$ implies
\[
t=e^{-\lambda},
\]
so the corresponding noise levels are geometrically spaced. Thus, the grid
contains finer absolute spacing at small noise levels (high SNR) and coarser
absolute spacing at large noise levels (low SNR).
\end{itemize}

\subsection{Estimating the Integral with Taylor Expansion}\label{subsec:dpm-method-2} 
DEIS and DPM-Solver exploit the same factored principle but obtain
higher-order information in different ways. DEIS reuses predictions from
previous completed steps to extrapolate the neural prediction over the next
interval. DPM-Solver instead works in the log-SNR coordinate $\lambda$ and
uses a local Taylor expansion of the neural prediction, with the required
higher-order information obtained from additional evaluations within the
current step. We now derive this single-step construction.

From \Cref{eq:analytic_solution}, the exact empirical PF-ODE flow initialized
at the current numerical state $\tilde\rvx_s$ is
\begin{equation}
\label{eqn:analytic_solution_each_step}
\widetilde\bPsi_{s\to t}(\tilde\rvx_s)
=
\frac{\alpha_t}{\alpha_s}\tilde\rvx_s
-
\alpha_t
\int_{\lambda_s}^{\lambda_t}
e^{-\lambda}
\hat{\bm{\epsilon}}_{\bm{\phi}^\times}
(\hat\rvx_\lambda,\lambda)
\diff\lambda,
\end{equation}
where
$\hat\rvx_{\lambda_s}=\tilde\rvx_s$ and
$\hat\rvx_\lambda$ denotes the corresponding exact empirical PF-ODE
trajectory over the current step. Therefore, after the known
schedule-dependent factors have been handled analytically, the remaining
numerical problem is to approximate
\[
\lambda
\longmapsto
\hat{\bm{\epsilon}}_{\bm{\phi}^\times}
(\hat\rvx_\lambda,\lambda)
\]
inside the weighted integral.

For $n\geq1$, we expand this neural prediction about $\lambda_s$:
\begin{equation*}
\hat{\bm{\epsilon}}_{\bm{\phi}^\times}
(\hat\rvx_\lambda,\lambda)
=
\sum_{k=0}^{n-1}
\frac{(\lambda-\lambda_s)^k}{k!}
\hat{\bm{\epsilon}}_{\bm{\phi}^\times}^{(k)}
(\hat\rvx_{\lambda_s},\lambda_s)
+
\mathcal O\bigl((\lambda-\lambda_s)^n\bigr),
\end{equation*}
where
\[
\hat{\bm{\epsilon}}_{\bm{\phi}^\times}^{(k)}
(\hat\rvx_\lambda,\lambda)
:=
\frac{\diff^k}{\diff\lambda^k}
\hat{\bm{\epsilon}}_{\bm{\phi}^\times}
(\hat\rvx_\lambda,\lambda)
\]
denotes the $k$-th \emph{total derivative along the empirical PF-ODE
trajectory}. In particular,
\[
\frac{\diff}{\diff\lambda}
\hat{\bm{\epsilon}}_{\bm{\phi}^\times}
(\hat\rvx_\lambda,\lambda)
=
\frac{\partial}{\partial\lambda}
\hat{\bm{\epsilon}}_{\bm{\phi}^\times}
(\hat\rvx_\lambda,\lambda)
+
J_{\rvx}
\hat{\bm{\epsilon}}_{\bm{\phi}^\times}
(\hat\rvx_\lambda,\lambda)
\frac{\diff\hat\rvx_\lambda}{\diff\lambda},
\]
so the derivative includes both the explicit dependence on $\lambda$ and the
dependence through the evolving state.

Substituting the Taylor expansion into
\Cref{eqn:analytic_solution_each_step} gives
\begin{mdframed}
\begin{equation}
\label{eq:k-th-expansion}
\widetilde\bPsi_{s\to t}(\tilde\rvx_s)
=
\frac{\alpha_t}{\alpha_s}\tilde\rvx_s
-
\alpha_t
\sum_{k=0}^{n-1}
{\color{cyan}
\hat{\bm{\epsilon}}_{\bm{\phi}^\times}^{(k)}
(\hat\rvx_{\lambda_s},\lambda_s)}
\,{\color{orange}C_k}
+
\mathcal O(h^{n+1}),
\end{equation}
where
\[
h:=\lambda_t-\lambda_s,
\qquad
{\color{orange}C_k}
:=
\int_{\lambda_s}^{\lambda_t}
e^{-\lambda}
\frac{(\lambda-\lambda_s)^k}{k!}
\diff\lambda.
\]
The coefficients ${\color{orange}C_k}$ depend only on the log-SNR interval
and can be evaluated analytically.
\end{mdframed}

If the required trajectory derivatives were available, dropping the
$\mathcal O(h^{n+1})$ remainder would give the corresponding formal
$n$-th order update. DPM-Solver avoids explicitly differentiating the neural
network by approximating these derivatives from additional model evaluations,
as described in the next subsection.

\exm{DPM-Solver-1}{
For $n=1$, the exact empirical flow satisfies
\[
\widetilde\bPsi_{s\to t}(\tilde\rvx_s)
=
\frac{\alpha_t}{\alpha_s}\tilde\rvx_s
-
\sigma_t
\left(e^h-1\right)
\bm{\epsilon}_{\bm{\phi}^\times}(\tilde\rvx_s,s)
+
\mathcal O(h^2).
\]
Dropping the $\mathcal O(h^2)$ remainder defines
DPM-Solver-1:
\begin{align}
\label{eq:dpm-1}
\tilde\rvx_t^{\mathrm{DPM\text{-}Solver\text{-}1}}
:=
\frac{\alpha_t}{\alpha_s}\tilde\rvx_s
-
\sigma_t
\left(e^h-1\right)
\bm{\epsilon}_{\bm{\phi}^\times}(\tilde\rvx_s,s).
\end{align}
This finite-step update is exactly the deterministic DDIM update, as shown
in \Cref{subsec:ddim=dpm-solver-1}.
}

DPM-Solver-$n$ with $n \geq 2$ requires evaluating the $k^{\text{th}}$-derivative {\color{cyan}$\hat{\bm{\epsilon}}_{\bm{\phi}^\times}^{(k)}(\hat\rvx_\lambda, \lambda)$} for $k \leq n-1$. However, directly computing higher-order derivatives is computationally expensive in practice. \citet{lu2022dpm} also propose efficient approximation methods for these derivatives, which will be detailed in the next subsection.

\subsection{Implementation of DPM-Solver-$n$}\label{subsec:dpm-n-method} 

\paragraph{DPM-Solver-$n$ with $n \geq 2$.}
In practice, implementing a higher-order DPM-Solver-$n$ entails the following:
\begin{itemize}
    \item Precomputing the coefficients ${\color{orange}C_k}$;
    \item Approximating the $k^{\text{th}}$ derivative {\color{cyan}$\hat{\bm{\epsilon}}_{\bm{\phi}^\times}^{(k)}(\hat\rvx_\lambda,\lambda)$} for $k \leq n-1$ to circumvent the costly computation of exact higher-order derivatives—a challenge well-studied in the ODE literature~\citep{hochbruck2005explicit,luan2021efficient}. One common strategy is finite difference approximation.
\end{itemize}

We now elaborate on the first two points.

\subparagraph{Precomputing ${\color{orange}C_k}$.}
Let $s$ and $t$ denote the start and end times, respectively, and define
$h:=\lambda_t-\lambda_s$. The Taylor expansion of the exact empirical
PF-ODE flow takes the form
\begin{equation}
\label{eq:analytic_expansion}
\widetilde\bPsi_{s\to t}(\rvx_s)
=
\frac{\alpha_t}{\alpha_s}\rvx_s
-
\sigma_t
\sum_{k=0}^{n-1}
h^{k+1}
\varphi_{k+1}(h)
\hat{\bm{\epsilon}}_{\bm{\phi}^\times}^{(k)}
(\hat\rvx_{\lambda_s},\lambda_s)
+
\mathcal O(h^{n+1}),
\end{equation}
where, for $m\geq1$,
\[
\varphi_m(h)
:=
\frac{
e^h-\sum_{j=0}^{m-1}\frac{h^j}{j!}
}{
h^m
},
\qquad
\varphi_m(0):=\frac{1}{m!}.
\]
For the first three cases,
\begin{align*}
\varphi_1(h)
&=
\frac{e^h-1}{h},
&
\varphi_2(h)
&=
\frac{e^h-h-1}{h^2},
&
\varphi_3(h)
&=
\frac{e^h-\frac{h^2}{2}-h-1}{h^3}.
\end{align*}

\exm{Third-Order Expansion with Exact Derivatives}{
For $n=3$, \Cref{eq:analytic_expansion} becomes
\begin{equation}
\label{eq:example-dpm-solver-3-exact}
\begin{aligned}
\widetilde\bPsi_{s\to t}(\rvx_s)
=
\frac{\alpha_t}{\alpha_s}\rvx_s
&-
\sigma_t
\left(e^h-1\right)
\hat{\bm{\epsilon}}_{\bm{\phi}^\times}
(\hat\rvx_{\lambda_s},\lambda_s)
\\
&-
\sigma_t
\left(e^h-h-1\right)
\hat{\bm{\epsilon}}_{\bm{\phi}^\times}^{(1)}
(\hat\rvx_{\lambda_s},\lambda_s)
\\
&-
\sigma_t
\left(
e^h-\frac{h^2}{2}-h-1
\right)
\hat{\bm{\epsilon}}_{\bm{\phi}^\times}^{(2)}
(\hat\rvx_{\lambda_s},\lambda_s)
+
\mathcal O(h^4).
\end{aligned}
\end{equation}
}

\subparagraph{Approximating {\color{cyan}$\hat{\bm{\epsilon}}_{\bm{\phi}^\times}^{(k)}(\hat\rvx_\lambda,\lambda)$} for $k \leq n-1$.}
For $n \geq 2$, following the standard approach of single-step ODE
solvers~\citep{atkinson2009numerical}, \citet{lu2022dpm} use additional model
evaluations within the current step to approximate the higher-order
derivatives. We illustrate this idea for $n=2$.

Let $\gamma\in(0,1]$ specify a stage point in the log-SNR interval
$[\lambda_s,\lambda_t]$. Given an estimate $\tilde\rvx_s$ at $s$, define
\[
s^{\mathrm{mid}}
=
t_\lambda\left(\lambda_s+\gamma h\right),
\qquad
h:=\lambda_t-\lambda_s.
\]
For $\gamma\in(0,1)$ this is an interior point, while $\gamma=1$ corresponds
to the endpoint $t$. We first use a DPM-Solver-1 step to predict the state at
this stage:
\[
\rvx^{\mathrm{mid}}
=
\frac{\alpha_{s^{\mathrm{mid}}}}{\alpha_s}\tilde\rvx_s
-
\sigma_{s^{\mathrm{mid}}}
\left(e^{\gamma h}-1\right)
\bm{\epsilon}_{\bm{\phi}^\times}(\tilde\rvx_s,s).
\]
The difference between the model predictions at the starting and stage points
then provides a finite-difference approximation of the first total derivative
along the trajectory:
\[
\hat{\bm{\epsilon}}_{\bm{\phi}^\times}^{(1)}
(\hat\rvx_{\lambda_s},\lambda_s)
\approx
\frac{
\bm{\epsilon}_{\bm{\phi}^\times}
(\rvx^{\mathrm{mid}},s^{\mathrm{mid}})
-
\bm{\epsilon}_{\bm{\phi}^\times}
(\tilde\rvx_s,s)
}{
\gamma h
}.
\]
Substituting this approximation into the second-order expansion gives
\begin{align}
\label{eq:dpm-solver-2-app}
\begin{aligned}
\widetilde\bPsi_{s\to t}(\tilde\rvx_s)
=
\frac{\alpha_t}{\alpha_s}\tilde\rvx_s
&-
\sigma_t
\left(e^h-1\right)
\bm{\epsilon}_{\bm{\phi}^\times}(\tilde\rvx_s,s)
\\
&-
\frac{\sigma_t}{\gamma h}
\left(e^h-h-1\right)
\left[
\bm{\epsilon}_{\bm{\phi}^\times}
(\rvx^{\mathrm{mid}},s^{\mathrm{mid}})
-
\bm{\epsilon}_{\bm{\phi}^\times}
(\tilde\rvx_s,s)
\right]
+
\mathcal O(h^3).
\end{aligned}
\end{align}
Thus, dropping the $\mathcal O(h^3)$ remainder yields a second-order
single-step update.

For the midpoint choice $\gamma=\tfrac12$, DPM-Solver-2 admits the simpler
two-stage form in \Cref{alg:dpm-solver-2}. This update differs from
\Cref{eq:dpm-solver-2-app} only by $\mathcal O(h^3)$ and therefore has the
same second-order accuracy.
\begin{algorithm}[H]
    \centering
    \caption{DPM-Solver-2 (with $\gamma=\tfrac12$).}
    \label{alg:dpm-solver-2}
    \begin{algorithmic}[1]
    \Require initial value $\rvx_T$, time steps $\{t_i\}_{i=0}^M$, model ${\bm{\epsilon}}_{\bm{\phi}^\times}$
        \State $\tilde\rvx_{t_0}\gets\rvx_T$
        \For{$i\gets 1$ to $M$}
        \State $h_i \gets \lambda_{t_i}-\lambda_{t_{i-1}}$
        \State $s_i^{\mathrm{mid}}
        \gets
        t_\lambda\left(
        \tfrac{\lambda_{t_{i-1}}+\lambda_{t_i}}{2}
        \right)$
        \State
        $\rvx_i^{\mathrm{mid}}
        \gets
        \tfrac{\alpha_{s_i^{\mathrm{mid}}}}{\alpha_{t_{i-1}}}
        \tilde\rvx_{t_{i-1}}
        -
        \sigma_{s_i^{\mathrm{mid}}}
        \left(e^{h_i/2}-1\right)
        {\bm{\epsilon}}_{\bm{\phi}^\times}
        (\tilde\rvx_{t_{i-1}},t_{i-1})$
        \State
        $\tilde\rvx_{t_i}
        \gets
        \tfrac{\alpha_{t_i}}{\alpha_{t_{i-1}}}
        \tilde\rvx_{t_{i-1}}
        -
        \sigma_{t_i}
        \left(e^{h_i}-1\right)
        {\bm{\epsilon}}_{\bm{\phi}^\times}
        (\rvx_i^{\mathrm{mid}},s_i^{\mathrm{mid}})$
        \EndFor
        \State \Return $\tilde\rvx_{t_M}$
    \end{algorithmic}
\end{algorithm}

\rmkb{
The finite difference above approximates the \emph{total} $\lambda$-derivative
of the neural prediction along the empirical PF-ODE trajectory. Under standard
smoothness assumptions,
\[
\frac{
\bm{\epsilon}_{\bm{\phi}^\times}
(\rvx^{\mathrm{mid}},s^{\mathrm{mid}})
-
\bm{\epsilon}_{\bm{\phi}^\times}
(\tilde\rvx_s,s)
}{
\gamma h
}
=
\hat{\bm{\epsilon}}_{\bm{\phi}^\times}^{(1)}
(\hat\rvx_{\lambda_s},\lambda_s)
+
\mathcal O(h).
\]
Here, the intermediate state is itself obtained by a first-order
DPM-Solver-1 step, whose stage error is $\mathcal O(h^2)$; under smoothness
of the model, this does not change the $\mathcal O(h)$ accuracy of the
difference quotient.

In \Cref{eq:dpm-solver-2-app}, this derivative approximation is multiplied by
\[
e^h-h-1=\mathcal O(h^2),
\]
so its induced error is $\mathcal O(h^3)$. Hence, the resulting method is
second-order accurate for any fixed $\gamma\in(0,1]$.

This single-step construction uses two model evaluations: one at
$(\tilde\rvx_s,s)$ and one at the predicted stage
$(\rvx^{\mathrm{mid}},s^{\mathrm{mid}})$. The choice of $\gamma$ does not
change the formal second-order accuracy, but it changes the stage location
and the error constant. In particular, $\gamma=\tfrac12$ places the second
evaluation at the midpoint of the log-SNR interval, giving the commonly used
midpoint version of DPM-Solver-2.
}

\paragraph{Implementation Detail: Selection of Sampling Timesteps.}
Since DPM-Solver constructs its local high-order approximation in log-SNR time
$\lambda$, a natural choice is to place the numerical nodes uniformly in
$\lambda$. For a finite terminal time $t_M>0$, this gives
\[
\lambda_{t_i}
=
\lambda_{t_0}
+
\frac{i}{M}
\left(
\lambda_{t_M}-\lambda_{t_0}
\right),
\qquad
i=0,\ldots,M.
\]
If $\sigma_0=0$, then $\lambda_0=+\infty$, so in practice one instead uses a
finite terminal log-SNR corresponding to a small positive time, with the
endpoint handled by the corresponding limiting or final denoising step.

Although the derivation is carried out in $\lambda$-space, the pre-trained
model and schedule $(\alpha_t,\sigma_t)$ are usually parameterized by the
original diffusion time $t$. During sampling, each selected log-SNR node is
therefore mapped back through
\[
t_i=t_\lambda(\lambda_{t_i})
\]
whenever the model is evaluated or the schedule coefficients are required;
see, for example, \Cref{alg:dpm-solver-2}. Thus, $\lambda$ serves as the
numerical integration coordinate, while the sampler continues to evaluate
the original pre-trained diffusion model at its corresponding time labels.

\subsection{DDIM $=$ DPM-Solver-1}
\label{subsec:ddim=dpm-solver-1}

The first-order DPM-Solver recovers DDIM exactly. Both methods start from an
exact integral representation of the same noise-prediction PF-ODE and make
the same approximation over one step:
\[
\beps_{\bm{\phi}^\times}(\rvx_\tau,\tau)
\approx
\beps_{\bm{\phi}^\times}(\tilde\rvx_s,s).
\]
The remaining schedule-dependent terms are then integrated analytically.
DDIM applies this construction directly in the schedule-aligned form derived
earlier, whereas DPM-Solver first rewrites the integral using the log-SNR
coordinate
\[
\lambda_t
=
\log\frac{\alpha_t}{\sigma_t}.
\]
At first order, this change of numerical clock does not alter the constant
approximation of the model prediction. Consequently, the two derivations
produce exactly the same finite-step update.

\proppp{DDIM is DPM-Solver-1}{ddim-dpm}{
For a fixed diffusion schedule $(\alpha_t,\sigma_t)$, the deterministic DDIM
update in \Cref{eq:ddim-euler} is identical to the DPM-Solver-1 update in
\Cref{eq:dpm-1}.
}{
By the definition of $\lambda$,
\[
\frac{\sigma_s}{\alpha_s}
=
e^{-\lambda_s},
\qquad
\frac{\sigma_t}{\alpha_t}
=
e^{-\lambda_t}.
\]
Let
\[
h
=
\lambda_t-\lambda_s.
\]
Then
\[
e^h
=
\frac{\alpha_t/\sigma_t}{\alpha_s/\sigma_s}
=
\frac{\alpha_t\sigma_s}{\sigma_t\alpha_s},
\]
and therefore
\[
-\sigma_t(e^h-1)
=
\sigma_t
-
\frac{\alpha_t}{\alpha_s}\sigma_s
=
\alpha_t
\left(
\frac{\sigma_t}{\alpha_t}
-
\frac{\sigma_s}{\alpha_s}
\right).
\]
Substituting this identity into the DPM-Solver-1 update gives
\[
\tilde\rvx_t
=
\frac{\alpha_t}{\alpha_s}\tilde\rvx_s
+
\alpha_t
\left(
\frac{\sigma_t}{\alpha_t}
-
\frac{\sigma_s}{\alpha_s}
\right)
\beps_{\bm{\phi}^\times}(\tilde\rvx_s,s),
\]
which is exactly the deterministic DDIM update in
\Cref{eq:ddim-euler}.
}

Thus, DPM-Solver-1 and DDIM are two derivations of the same first-order
finite-step update. The log-SNR coordinate becomes consequential beyond first
order: higher-order DPM-Solvers use it to approximate how the noise prediction
changes along the trajectory, rather than treating that prediction as constant
over the whole step.

\subsection{Relation of DPM-Solver to Runge-Kutta-Type Solvers}
\label{subsec:dpm-rk}

DPM-Solver-1 is exactly the DDIM update
(\Cref{subsec:ddim=dpm-solver-1}), so its first-order behavior is already
understood. The higher-order question is how DPM-Solver improves this
one-evaluation update.

The connection is to Runge-Kutta (RK) methods. Euler uses only the vector
field at the beginning of a step, whereas RK methods obtain additional
information through \emph{stages}, i.e., evaluations at states and times
generated within the current step; see \Cref{app:de-solver-examples}.
Higher-order DPM-Solvers follow the same principle: instead of holding one
neural prediction fixed over the whole step, they introduce additional
within-step model evaluations to capture its variation. The diffusion-specific
difference is that the known schedule-dependent factors are retained
analytically, while the staged approximation is applied only to the learned
prediction.

\paragraph{DPM-Solver-2 versus Plain Midpoint RK2.}
We first compare DPM-Solver-2 with classical midpoint RK2 applied directly to
the complete noise-prediction PF-ODE in log-SNR time from
\Cref{eq:diffusion_ode_lambda_eps}. For brevity, denote its complete
right-hand side by $\rmF(\hat\rvx_\lambda,\lambda)$.

For the midpoint choice $\gamma=\tfrac12$, let
\[
\lambda_{\mathrm{mid}}
:=
\lambda_{s^{\mathrm{mid}}}
=
\lambda_s+\frac{h}{2},
\qquad
\alpha_{\mathrm{mid}}
:=
\alpha_{s^{\mathrm{mid}}},
\qquad
\sigma_{\mathrm{mid}}
:=
\sigma_{s^{\mathrm{mid}}},
\]
and abbreviate
\[
\hat\beps_s
:=
\hat\beps_{\bm{\phi}^\times}(\tilde\rvx_s,\lambda_s)
=
\beps_{\bm{\phi}^\times}(\tilde\rvx_s,s).
\]

Classical midpoint RK2 gives
\[
\rvx_{\mathrm{mid}}^{\mathrm{RK}}
=
\tilde\rvx_s
+
\frac{h}{2}
\rmF(\tilde\rvx_s,\lambda_s),
\]
followed by
\[
\tilde\rvx_t^{\mathrm{RK2}}
=
\tilde\rvx_s
+
h\,
\rmF
\left(
\rvx_{\mathrm{mid}}^{\mathrm{RK}},
\lambda_{\mathrm{mid}}
\right).
\]
Thus, plain RK2 applies its two-stage construction to the complete PF-ODE
velocity, including both the known schedule-dependent terms and the neural
prediction.

By contrast, midpoint DPM-Solver-2 in \Cref{alg:dpm-solver-2} first constructs
the stage
\[
\rvx_{\mathrm{mid}}^{\mathrm{DPM2}}
=
\frac{\alpha_{\mathrm{mid}}}{\alpha_s}\tilde\rvx_s
-
\sigma_{\mathrm{mid}}
\left(
e^{h/2}-1
\right)
\hat\beps_s.
\]
Denoting the corresponding midpoint prediction by
\[
\hat\beps_{\mathrm{mid}}^{\mathrm{DPM2}}
:=
\hat\beps_{\bm{\phi}^\times}
\left(
\rvx_{\mathrm{mid}}^{\mathrm{DPM2}},
\lambda_{\mathrm{mid}}
\right),
\]
the full update is
\[
\tilde\rvx_t^{\mathrm{DPM2}}
=
\frac{\alpha_t}{\alpha_s}\tilde\rvx_s
-
\sigma_t
\left(
e^h-1
\right)
\hat\beps_{\mathrm{mid}}^{\mathrm{DPM2}}.
\]

The structural difference is immediate. Plain midpoint RK2 discretizes the
complete PF-ODE velocity, whereas DPM-Solver-2 retains the known
diffusion-dependent evolution analytically and uses its two model evaluations
only to approximate the variation of the noise prediction.

Under standard smoothness assumptions, both are second-order methods for the
same empirical PF-ODE in $\lambda$-time. Their intermediate states generally
differ by $\mathcal O(h^2)$, and their complete one-step updates therefore
satisfy
\[
\tilde\rvx_t^{\mathrm{DPM2}}
-
\tilde\rvx_t^{\mathrm{RK2}}
=
\mathcal O(h^3).
\]
Thus, DPM-Solver-2 follows the same two-stage midpoint principle as classical
RK2, while exploiting the known diffusion structure before applying the
numerical approximation.

\paragraph{DPM-Solver-2 versus Exponential Midpoint RK2.}
A closer comparison keeps the semilinear decomposition in
\Cref{eq:diffusion_ode_lambda_eps} explicit and treats its known linear part
analytically. The corresponding \emph{exponential midpoint RK2} construction
first takes an exponential-Euler stage to the midpoint:
\[
\rvx_{\mathrm{mid}}^{\mathrm{ExpRK}}
=
\frac{\alpha_{\mathrm{mid}}}{\alpha_s}\tilde\rvx_s
-
\alpha_{\mathrm{mid}}\sigma_s
\left[
\int_{\lambda_s}^{\lambda_{\mathrm{mid}}}
\frac{1}{\alpha_\lambda}\diff\lambda
\right]
\hat\beps_s.
\]
After evaluating
\[
\hat\beps_{\mathrm{mid}}^{\mathrm{ExpRK}}
:=
\hat\beps_{\bm{\phi}^\times}
\left(
\rvx_{\mathrm{mid}}^{\mathrm{ExpRK}},
\lambda_{\mathrm{mid}}
\right),
\]
the full exponential-midpoint update is
\[
\tilde\rvx_t^{\mathrm{ExpRK2}}
=
\frac{\alpha_t}{\alpha_s}\tilde\rvx_s
-
\alpha_t\sigma_{\mathrm{mid}}
\left[
\int_{\lambda_s}^{\lambda_t}
\frac{1}{\alpha_\lambda}\diff\lambda
\right]
\hat\beps_{\mathrm{mid}}^{\mathrm{ExpRK}}.
\]
When the linear coefficient is constant, these integral coefficients reduce
to the usual $\varphi_1$ coefficients of explicit exponential midpoint RK2.

DPM-Solver-2 uses the same midpoint staging principle, but exploits the finer
noise-prediction factorization
\[
-\sigma_\lambda
\hat\beps_{\bm{\phi}^\times}(\hat\rvx_\lambda,\lambda).
\]
Since $\sigma_\lambda$ is also known from the diffusion schedule,
DPM-Solver-2 retains it inside the integral and stages only the neural
prediction. Its midpoint and full-step coefficients are therefore
\[
\alpha_{\mathrm{mid}}
\int_{\lambda_s}^{\lambda_{\mathrm{mid}}}
\frac{\sigma_\lambda}{\alpha_\lambda}\diff\lambda
=
\sigma_{\mathrm{mid}}
\left(e^{h/2}-1\right),
\quad \text{and}\quad
\alpha_t
\int_{\lambda_s}^{\lambda_t}
\frac{\sigma_\lambda}{\alpha_\lambda}\diff\lambda
=
\sigma_t
\left(e^h-1\right),
\]
respectively, recovering the update in \Cref{alg:dpm-solver-2}.

Hence, the three constructions differ in what is approximated by the RK
stages:
\[
\begin{array}{rl}
\text{\textbfs{Plain RK2:} }
&
\text{stage the complete PF-ODE velocity},
\\[4pt]

\textbf{\textbfs{ExpRK2:} }
&
\begin{aligned}[t]
&\text{propagate the linear part exactly;}\\
&\text{stage the coefficient-weighted residual }
\sigma_\lambda\hat\beps_{\bm{\phi}^\times},
\end{aligned}
\\[6pt]

\text{\textbfs{DPM-Solver-2:} }
&
\text{also retain }\sigma_\lambda\text{ analytically and stage only }
\hat\beps_{\bm{\phi}^\times}.
\end{array}
\]

The exponential-RK and DPM midpoint stages differ by
\[
\rvx_{\mathrm{mid}}^{\mathrm{DPM2}}
-
\rvx_{\mathrm{mid}}^{\mathrm{ExpRK}}
=
-
\alpha_{\mathrm{mid}}
\left[
\int_{\lambda_s}^{\lambda_{\mathrm{mid}}}
\frac{\sigma_\lambda-\sigma_s}{\alpha_\lambda}
\diff\lambda
\right]
\hat\beps_s
=
\mathcal O(h^2).
\]
For the complete step, a Taylor expansion about
$\lambda_{\mathrm{mid}}$ gives
\[
\int_{\lambda_s}^{\lambda_t}
\frac{\sigma_\lambda-\sigma_{\mathrm{mid}}}{\alpha_\lambda}
\diff\lambda
=
\mathcal O(h^3),
\]
while smoothness of the neural prediction gives
\[
\hat\beps_{\mathrm{mid}}^{\mathrm{DPM2}}
-
\hat\beps_{\mathrm{mid}}^{\mathrm{ExpRK}}
=
\mathcal O(h^2).
\]
Consequently,
\[
\tilde\rvx_t^{\mathrm{DPM2}}
-
\tilde\rvx_t^{\mathrm{ExpRK2}}
=
\mathcal O(h^3).
\]
Thus, DPM-Solver-2 and exponential midpoint RK2 agree through second order,
but DPM-Solver-2 retains one additional analytically known quantity,
$\sigma_\lambda$, inside the weighted integral.

\rmkb{
The discussion above concerns the noise-prediction formulation in which
DPM-Solver is derived. If a pre-trained model uses another prediction
parameterization, its output may be converted pointwise to noise prediction
through \Cref{eq:predictions-equivalence} at each stage. Just like the DDIM-case in \Cref{subsec:ddim-classical-euler}, this should not be
confused with directly applying a classical RK method to the raw prediction
field: pointwise conversion and finite-step discretization are separate
operations.
}

\msgu{DPM-Solver and Runge--Kutta Methods}{}{
DPM-Solver combines the within-step staging principle of Runge--Kutta
methods with the analytic treatment of known diffusion dynamics.
DPM-Solver-1 is exactly DDIM, while higher-order DPM-Solvers use additional
within-step evaluations to capture the variation of the neural prediction.
The connection to classical RK methods is structural rather than an equality
of finite-step formulas: DPM-Solver retains the known diffusion-dependent
weights analytically before approximating the learned prediction.
}

\clearpage
\newpage

\section{\texorpdfstring{DPM-Solver\texttt{++}}{DPM-Solver\texttt{++}}}\label{sec:dpm++}

\subsection{From DPM-Solver to DPM-Solver\texttt{++} for Guided Sampling}

DPM-Solver\texttt{++}~\citep{lu2022dpm2} was developed to improve
high-order sampling under strong guidance. In this regime, two practical
difficulties arise. First, large guidance scales can amplify the model output
and its derivatives, narrowing the convergence region of high-order solvers
and requiring smaller numerical steps. Second, the guided prediction may
drive the final solution outside the range represented by the training data,
leading to the train--test mismatch discussed in
\Cref{ch:guidance}.

DPM-Solver\texttt{++} addresses these issues in two complementary ways. It
reformulates the diffusion ODE using data prediction, for which data-space
corrections such as dynamic thresholding can be applied, and it develops a
multistep variant that reuses previous model evaluations to obtain smaller
effective step sizes under a fixed NFE budget. We first derive the
data-prediction formulation and then introduce its single-step and multistep
second-order solvers.

\subsection{DPM-Solver\texttt{++}'s Methodology}

DPM-Solver\texttt{++} keeps the log-SNR coordinate $\lambda$ introduced for
DPM-Solver, but expresses the empirical PF-ODE through data prediction rather
than noise prediction. Recall from \Cref{eq:predictions-equivalence} that
\[
\beps_{\bm{\phi}^\times}(\rvx_t,t)
=
\frac{
\rvx_t-\alpha_t\rvx_{\bm{\phi}^\times}(\rvx_t,t)
}{
\sigma_t
}.
\]
Substituting this relation into the noise-prediction exact solution in
\Cref{eq:analytic_solution} gives the equivalent data-prediction form
\begin{equation}
\label{eq:dpmpp-data-exact}
\widetilde\bPsi_{s\to t}(\rvx_s)
=
\frac{\sigma_t}{\sigma_s}\rvx_s
+
\sigma_t
\int_{\lambda_s}^{\lambda_t}
e^\lambda
\hat{\rvx}_{\bm{\phi}^\times}
(\hat\rvx_\lambda,\lambda)
\diff\lambda,
\end{equation}
where, following \Cref{eq:change-of-var-x-epsilon}, we define the
$\lambda$-parameterized data-prediction field by
\[
\hat{\rvx}_{\bm{\phi}^\times}
(\rvx,\lambda)
:=
\rvx_{\bm{\phi}^\times}
\bigl(
\rvx,
t_\lambda(\lambda)
\bigr).
\]
Hence, along the empirical PF-ODE trajectory,
\[
\hat{\rvx}_{\bm{\phi}^\times}
(\hat\rvx_\lambda,\lambda)
=
\rvx_{\bm{\phi}^\times}
\bigl(
\rvx_{t_\lambda(\lambda)},
t_\lambda(\lambda)
\bigr).
\]

Thus, the structure is exactly dual to DPM-Solver: the known
schedule-dependent evolution is handled analytically, while the only
model-dependent quantity inside the remaining integral is now the data
prediction. DPM-Solver\texttt{++} develops two second-order realizations of
this exact form:

\begin{itemize}
    \item \textbfs{Single-Step Solver, DPM-Solver\texttt{++}(2S).}
    Use an additional model evaluation at an intermediate stage within the
    current step, analogous to the staged construction of DPM-Solver-2.

    \item \textbfs{Multistep Solver, DPM-Solver\texttt{++}(2M).}
    Reuse data predictions from the previous two completed steps. After the
    warm start, only one new model evaluation is required per step.
\end{itemize}

\rmkb{
Without thresholding, converting a model output to data prediction is a
pointwise algebraic re-expression of the same empirical PF-ODE. The numerical
method can nevertheless change because the finite-step approximation is now
applied to the data prediction rather than to the noise prediction. If
thresholding is used, it is an additional modification of the model output
and should be viewed separately from the numerical-order analysis below.
}

\subsection{DPM-Solver\texttt{++} Single-Step by Taylor Expansion}
\label{subsec:dpm-solver++-single}

Following the DPM-Solver construction in \Cref{subsec:dpm-method-2},
DPM-Solver\texttt{++} approximates the data prediction along the empirical
trajectory by a local Taylor expansion in log-SNR time. For $k\geq0$, denote
its $k$-th total derivative along the empirical PF-ODE trajectory by
\[
\hat{\rvx}_{\bm{\phi}^\times}^{(k)}
(\hat\rvx_\lambda,\lambda)
:=
\frac{\diff^k}{\diff\lambda^k}
\hat{\rvx}_{\bm{\phi}^\times}
(\hat\rvx_\lambda,\lambda).
\]

For one step from $t_{i-1}$ to $t_i$, let $h_i
:=
\lambda_{t_i}-\lambda_{t_{i-1}}
>0$. A Taylor expansion about $\lambda_{t_{i-1}}$ gives
\[
\hat{\rvx}_{\bm{\phi}^\times}
(\hat\rvx_\lambda,\lambda)
=
\sum_{k=0}^{n-1}
\frac{
(\lambda-\lambda_{t_{i-1}})^k
}{
k!
}
\hat{\rvx}_{\bm{\phi}^\times}^{(k)}
\bigl(
\hat\rvx_{\lambda_{t_{i-1}}},
\lambda_{t_{i-1}}
\bigr)
+
\mathcal O
\bigl(
(\lambda-\lambda_{t_{i-1}})^n
\bigr).
\]
Substituting this expansion into \Cref{eq:dpmpp-data-exact} yields
\begin{align}
\label{eq:dpmpp-taylor-expansion}
\widetilde\bPsi_{t_{i-1}\to t_i}
(\tilde\rvx_{t_{i-1}})
=
\frac{\sigma_{t_i}}{\sigma_{t_{i-1}}}
\tilde\rvx_{t_{i-1}}
+
\alpha_{t_i}
\sum_{k=0}^{n-1}
h_i^{k+1}
\varphi_{k+1}(-h_i)
\hat{\rvx}_{\bm{\phi}^\times}^{(k)}
\bigl(
\hat\rvx_{\lambda_{t_{i-1}}},
\lambda_{t_{i-1}}
\bigr)
+
\mathcal O(h_i^{n+1}),
\end{align}
where $\varphi_m$ is defined in \Cref{eq:analytic_expansion}. Here we used
\[
\int_{\lambda_{t_{i-1}}}^{\lambda_{t_i}}
e^\lambda
\frac{
(\lambda-\lambda_{t_{i-1}})^k
}{
k!
}
\diff\lambda
=
e^{\lambda_{t_i}}
h_i^{k+1}
\varphi_{k+1}(-h_i),
\]
together with
$\sigma_{t_i}e^{\lambda_{t_i}}=\alpha_{t_i}$.

For $n=1$, dropping the $\mathcal O(h_i^2)$ remainder gives
\[
\tilde\rvx_{t_i}
=
\frac{\sigma_{t_i}}{\sigma_{t_{i-1}}}
\tilde\rvx_{t_{i-1}}
+
\alpha_{t_i}
\left(
1-e^{-h_i}
\right)
\rmD_{i-1},
\qquad
\rmD_{i-1}
:=
\rvx_{\bm{\phi}^\times}
(\tilde\rvx_{t_{i-1}},t_{i-1}),
\]
which is exactly the deterministic DDIM update in data-prediction form.

For $n=2$, the first total derivative is approximated using an additional
within-step stage. For the midpoint choice, define
\[
s_i^{\mathrm{mid}}
:=
t_\lambda
\left(
\lambda_{t_{i-1}}+\frac{h_i}{2}
\right).
\]
A first-order DPM-Solver\texttt{++} step predicts the midpoint state
\[
\rvu_i
=
\frac{
\sigma_{s_i^{\mathrm{mid}}}
}{
\sigma_{t_{i-1}}
}
\tilde\rvx_{t_{i-1}}
+
\alpha_{s_i^{\mathrm{mid}}}
\left(
1-e^{-h_i/2}
\right)
\rmD_{i-1},
\]
at which we evaluate
\[
\rmD_i^{\mathrm{mid}}
:=
\rvx_{\bm{\phi}^\times}
(\rvu_i,s_i^{\mathrm{mid}}).
\]
The midpoint DPM-Solver\texttt{++}(2S) update then takes the particularly
simple form
\[
\tilde\rvx_{t_i}
=
\frac{\sigma_{t_i}}{\sigma_{t_{i-1}}}
\tilde\rvx_{t_{i-1}}
+
\alpha_{t_i}
\left(
1-e^{-h_i}
\right)
\rmD_i^{\mathrm{mid}}.
\]
As with midpoint DPM-Solver-2, this simplified two-stage update differs from
the direct second-order Taylor/finite-difference construction only by
$\mathcal O(h_i^3)$ and therefore retains second-order accuracy.

This yields the single-step DPM-Solver\texttt{++} construction. When $n=1$,
it reduces exactly to the deterministic DDIM update in data-prediction form.
When $n=2$, approximating the first total $\lambda$-derivative using an
additional within-step model evaluation gives
\emph{DPM-Solver\texttt{++}(2S)}, the data-prediction counterpart of
DPM-Solver-2 in \Cref{alg:dpm-solver-2}. Its midpoint implementation is
summarized in \Cref{alg:dpmpp-2s-mid}.
\begin{algorithm}[H]
    \centering
    \caption{DPM-Solver\texttt{++}(2S): a midpoint special case.}
    \label{alg:dpmpp-2s-mid}
    \begin{algorithmic}[1]
    \Require initial value $\rvx_T$, time steps $\{t_i\}_{i=0}^M$,
    data-prediction model $\rvx_{\bphi^\times}$

    \State $\tilde\rvx_{t_0}\gets\rvx_T$
    \State $\lambda_{t_i}\gets\log(\alpha_{t_i}/\sigma_{t_i})$
    for $i=0,\ldots,M$
    \State
    $\rmD_0
    \gets
    \rvx_{\bphi^\times}(\tilde\rvx_{t_0},t_0)$
    \Comment{cache start prediction}

    \For{$i\gets1$ \textbf{to} $M$}
        \State
        $h_i
        \gets
        \lambda_{t_i}-\lambda_{t_{i-1}}$

        \State
        $s_i^{\mathrm{mid}}
        \gets
        t_\lambda
        \left(
        \tfrac{\lambda_{t_{i-1}}+\lambda_{t_i}}{2}
        \right)$

        \State
        $\rvu_i
        \gets
        \tfrac{\sigma_{s_i^{\mathrm{mid}}}}{\sigma_{t_{i-1}}}
        \tilde\rvx_{t_{i-1}}
        +
        \alpha_{s_i^{\mathrm{mid}}}
        \left(
        1-e^{-h_i/2}
        \right)
        \rmD_{i-1}$
        \Comment{predict midpoint stage}

        \State
        $\rmD_i^{\mathrm{mid}}
        \gets
        \rvx_{\bphi^\times}
        (\rvu_i,s_i^{\mathrm{mid}})$
        \Comment{evaluate at midpoint}

        \State
        $\tilde\rvx_{t_i}
        \gets
        \tfrac{\sigma_{t_i}}{\sigma_{t_{i-1}}}
        \tilde\rvx_{t_{i-1}}
        +
        \alpha_{t_i}
        \left(
        1-e^{-h_i}
        \right)
        \rmD_i^{\mathrm{mid}}$

        \If{$i<M$}
            \State
            $\rmD_i
            \gets
            \rvx_{\bphi^\times}
            (\tilde\rvx_{t_i},t_i)$
            \Comment{cache for next step}
        \EndIf
    \EndFor

    \State \Return $\tilde\rvx_{t_M}$
    \end{algorithmic}
\end{algorithm}

\subsection{DPM-Solver\texttt{++} Multistep by Recycling History}
\label{subsec:dpm-solver++-multi}

DPM-Solver\texttt{++}(2S) obtains second-order information by introducing an
additional model evaluation \emph{within the current step}.
DPM-Solver\texttt{++}(2M) takes a different route: it reuses model
predictions from previous completed steps to estimate the same local
variation. After initialization, this requires only one new model evaluation
per step, allowing more, and hence smaller, numerical steps under the same
NFE budget. This is particularly useful under strong guidance, where large
steps can reduce numerical stability.

The idea is simple. Recall that the second-order expansion in
\Cref{eq:dpmpp-taylor-expansion} requires both the current data prediction
and its first total derivative with respect to $\lambda$. The single-step
method estimates this derivative using a new within-step stage; the multistep
method instead estimates it from the recent history of model predictions.

For the first step, no such history is available, so we initialize with the
first-order data-prediction update
\[
\tilde\rvx_{t_1}
=
\frac{\sigma_{t_1}}{\sigma_{t_0}}
\tilde\rvx_{t_0}
+
\alpha_{t_1}
\left(
1-e^{-h_1}
\right)
\rmD_0,
\qquad
\rmD_0
:=
\rvx_{\bm{\phi}^\times}
(\tilde\rvx_{t_0},t_0).
\]
This is exactly the deterministic DDIM update in data-prediction form.

Once two completed-step predictions are available, for $i\geq2$ define
\[
\rmD_{i-1}
:=
\rvx_{\bm{\phi}^\times}
(\tilde\rvx_{t_{i-1}},t_{i-1}),
\qquad
\rmD_{i-2}
:=
\rvx_{\bm{\phi}^\times}
(\tilde\rvx_{t_{i-2}},t_{i-2}).
\]
Keeping the first two terms of
\Cref{eq:dpmpp-taylor-expansion} gives
\begin{align*}
\widetilde\bPsi_{t_{i-1}\to t_i}
(\tilde\rvx_{t_{i-1}})
=
\frac{\sigma_{t_i}}{\sigma_{t_{i-1}}}
\tilde\rvx_{t_{i-1}}
&+
\alpha_{t_i}
\left(
1-e^{-h_i}
\right)
\rmD_{i-1}
\\
&+
\alpha_{t_i}
\left(
h_i-1+e^{-h_i}
\right)
\hat{\rvx}_{\bm{\phi}^\times}^{(1)}
\bigl(
\hat\rvx_{\lambda_{t_{i-1}}},
\lambda_{t_{i-1}}
\bigr)
+
\mathcal O(h_i^3).
\end{align*}
Thus, the only additional quantity needed for a second-order update is the
first total $\lambda$-derivative of the data prediction.

Instead of evaluating the model at a new intermediate stage,
DPM-Solver\texttt{++}(2M) estimates this derivative from the previous
completed step:
\[
\hat{\rvx}_{\bm{\phi}^\times}^{(1)}
\bigl(
\hat\rvx_{\lambda_{t_{i-1}}},
\lambda_{t_{i-1}}
\bigr)
\approx
\frac{
\rmD_{i-1}-\rmD_{i-2}
}{
h_{i-1}
}.
\]
Under standard smoothness assumptions, this backward difference has error
$\mathcal O(h_{i-1})$. Substituting it into the second-order expansion and
collecting the two prediction terms gives
\[
\tilde\rvx_{t_i}^{\mathrm{poly}}
=
\frac{\sigma_{t_i}}{\sigma_{t_{i-1}}}
\tilde\rvx_{t_{i-1}}
+
\alpha_{t_i}
\left(
1-e^{-h_i}
\right)
\rmD_i^{\mathrm{poly}},
\]
where
\[
\rmD_i^{\mathrm{poly}}
:=
\rmD_{i-1}
+
\frac{\beta(h_i)}{h_{i-1}}
\left(
\rmD_{i-1}-\rmD_{i-2}
\right),
\qquad
\beta(h)
:=
\frac{
h-1+e^{-h}
}{
1-e^{-h}
}.
\]
When consecutive step ratios remain bounded, this weighted multistep update
has one-step local defect $\mathcal O(h_i^3)$.

\paragraph{The Practical DPM-Solver\texttt{++}(2M) Update.}
The expression above is already a valid second-order multistep construction.
DPM-Solver\texttt{++}(2M) uses a slightly simpler coefficient that preserves
the same order. Since
\[
\beta(h)
=
\frac{h}{2}
+
\frac{h^2}{12}
+
\mathcal O(h^3),
\]
define the step ratio
\[
r_i
:=
\frac{h_i}{h_{i-1}}
\]
and replace $\rmD_i^{\mathrm{poly}}$ by
\[
\rmD_i^{\mathrm{sim}}
:=
\rmD_{i-1}
+
\frac{r_i}{2}
\left(
\rmD_{i-1}-\rmD_{i-2}
\right).
\]
Equivalently,
\[
\rmD_i^{\mathrm{sim}}
=
\left(
1+\frac{r_i}{2}
\right)\rmD_{i-1}
-
\frac{r_i}{2}\rmD_{i-2}.
\]
The interpretation is straightforward: the two most recent model predictions
tell us the local direction in which the data prediction is changing, and
$\rmD_i^{\mathrm{sim}}$ extrapolates that trend into the current step.

Under the same smoothness and bounded-step-ratio assumptions,
\[
\rmD_i^{\mathrm{poly}}
-
\rmD_i^{\mathrm{sim}}
=
\mathcal O(h_i^2).
\]
Since
\[
\alpha_{t_i}
\left(
1-e^{-h_i}
\right)
=
\mathcal O(h_i),
\]
this simplification changes the state update only by
$\mathcal O(h_i^3)$. Hence, it preserves second-order accuracy, giving the
practical DPM-Solver\texttt{++}(2M) update
\begin{mdframed}
\[
\tilde\rvx_{t_i}
=
\frac{\sigma_{t_i}}{\sigma_{t_{i-1}}}
\tilde\rvx_{t_{i-1}}
+
\alpha_{t_i}
\left(
1-e^{-h_i}
\right)
\rmD_i^{\mathrm{sim}}.
\]
\end{mdframed}

After computing $\tilde\rvx_{t_i}$, the model is evaluated once at the new
state,
\[
\rmD_i
=
\rvx_{\bm{\phi}^\times}
(\tilde\rvx_{t_i},t_i),
\]
and this prediction is cached for the next step. Thus, after the warm start,
each DPM-Solver\texttt{++}(2M) step requires only one new model evaluation.

\rmkb{
For uniform log-SNR steps, $r_i=1$, so
\[
\rmD_i^{\mathrm{sim}}
=
\frac{3}{2}\rmD_{i-1}
-
\frac{1}{2}\rmD_{i-2}.
\]
The coefficients $(\frac32,-\frac12)$ are the familiar AB2 extrapolation
weights for the \emph{model prediction}. The complete
DPM-Solver\texttt{++}(2M) update is nevertheless not plain AB2: the known
diffusion-dependent factor
$\alpha_{t_i}(1-e^{-h_i})$ is retained analytically.
}

The complete warm-start and two-step procedure is summarized in
\Cref{alg:dpmpp-2m}. The key distinction from
DPM-Solver\texttt{++}(2S) is where the second-order information comes from:
2S obtains it from an additional stage within the current step, whereas 2M
recycles model evaluations from previous completed steps.

\begin{algorithm}[H]
    \centering
    \caption{DPM-Solver\texttt{++}(2M).}
    \label{alg:dpmpp-2m}
    \begin{algorithmic}[1]
    \Require initial value $\rvx_T$, time steps $\{t_i\}_{i=0}^M$,
    data-prediction model $\rvx_{\bphi^\times}$

    \State $\tilde\rvx_{t_0}\gets\rvx_T$
    \State $\lambda_{t_i}\gets\log(\alpha_{t_i}/\sigma_{t_i})$
    for $i=0,\ldots,M$
    \State
    $\rmD_0
    \gets
    \rvx_{\bphi^\times}
    (\tilde\rvx_{t_0},t_0)$
    \Comment{cache start prediction}

    \Statex
    \noindent\textbfs{Case I. Warm start ($i=1$)}

    \State
    $h_1
    \gets
    \lambda_{t_1}-\lambda_{t_0}$

    \State
    $\tilde\rvx_{t_1}
    \gets
    \tfrac{\sigma_{t_1}}{\sigma_{t_0}}
    \tilde\rvx_{t_0}
    +
    \alpha_{t_1}
    \left(
    1-e^{-h_1}
    \right)
    \rmD_0$
    \Comment{DDIM in data prediction}

    \If{$M>1$}
        \State
        $\rmD_1
        \gets
        \rvx_{\bphi^\times}
        (\tilde\rvx_{t_1},t_1)$
        \Comment{cache for multistep update}
    \EndIf

    \Statex
    \noindent\textbfs{Case II. Two-step multistep update ($i\geq2$)}

    \For{$i\gets2$ \textbf{to} $M$}
        \State
        $h_i
        \gets
        \lambda_{t_i}-\lambda_{t_{i-1}}$

        \State
        $r_i
        \gets
        h_i/h_{i-1}$
        \Comment{step ratio}

        \State
        $\rmD_i^{\mathrm{sim}}
        \gets
        \left(
        1+\tfrac12 r_i
        \right)
        \rmD_{i-1}
        -
        \tfrac12 r_i
        \rmD_{i-2}$

        \State
        $\tilde\rvx_{t_i}
        \gets
        \tfrac{\sigma_{t_i}}{\sigma_{t_{i-1}}}
        \tilde\rvx_{t_{i-1}}
        +
        \alpha_{t_i}
        \left(
        1-e^{-h_i}
        \right)
        \rmD_i^{\mathrm{sim}}$

        \If{$i<M$}
            \State
            $\rmD_i
            \gets
            \rvx_{\bphi^\times}
            (\tilde\rvx_{t_i},t_i)$
            \Comment{one new model call; cache for next step}
        \EndIf
    \EndFor

    \State \Return $\tilde\rvx_{t_M}$
    \end{algorithmic}
\end{algorithm}

\subsection{Relation of DPM-Solver\texttt{++}(2M) to DEIS}
\label{subsec:dpm-deis}

DPM-Solver\texttt{++}(2M) and DEIS use the same basic multistep idea:
reuse model predictions from previous completed steps to approximate how the
neural prediction changes over the next interval, while keeping the known
diffusion-dependent weights analytic. The difference is the representation in
which this idea is applied. DEIS was formulated earlier using noise prediction
in the original diffusion time, whereas DPM-Solver\texttt{++}(2M) uses data
prediction in log-SNR time.

To expose the connection directly, we can apply the DEIS polynomial
extrapolation principle to the exact data-prediction integral underlying
DPM-Solver\texttt{++}. Recall that
\[
\widetilde\bPsi_{t_{i-1}\to t_i}
(\tilde\rvx_{t_{i-1}})
=
\frac{\sigma_{t_i}}{\sigma_{t_{i-1}}}
\tilde\rvx_{t_{i-1}}
+
\sigma_{t_i}
\int_{\lambda_{t_{i-1}}}^{\lambda_{t_i}}
e^\lambda
\hat{\rvx}_{\bm{\phi}^\times}
(\hat\rvx_\lambda,\lambda)
\diff\lambda.
\]
The known weight $e^\lambda$ is already handled analytically. The only
quantity that must be approximated over the step is therefore the data
prediction
\[
\lambda
\longmapsto
\hat{\rvx}_{\bm{\phi}^\times}
(\hat\rvx_\lambda,\lambda).
\]

Using the two most recent cached predictions
$\rmD_{i-2}$ and $\rmD_{i-1}$, the simplest nonconstant approximation is the
unique affine polynomial through these two anchors. In Lagrange form,
\[
P_1(\lambda)
=
\frac{
\lambda-\lambda_{t_{i-2}}
}{
h_{i-1}
}
\rmD_{i-1}
-
\frac{
\lambda-\lambda_{t_{i-1}}
}{
h_{i-1}
}
\rmD_{i-2}.
\]
The same polynomial can be written more transparently around the most recent
anchor as
\[
P_1(\lambda)
=
\rmD_{i-1}
+
\frac{
\lambda-\lambda_{t_{i-1}}
}{
h_{i-1}
}
\left(
\rmD_{i-1}-\rmD_{i-2}
\right).
\]
The first expression is the Lagrange form, while the second is the
backward-Newton form of the \emph{same} affine polynomial. Thus, Lagrange
interpolation and backward differences are not different numerical
approximations here; they are simply two representations of the same
two-anchor extrapolant.

Applying this extrapolant inside the DPM-Solver\texttt{++} integral gives
\[
\tilde\rvx_{t_i}^{\mathrm{poly}}
=
\frac{\sigma_{t_i}}{\sigma_{t_{i-1}}}
\tilde\rvx_{t_{i-1}}
+
\sigma_{t_i}
\int_{\lambda_{t_{i-1}}}^{\lambda_{t_i}}
e^\lambda
P_1(\lambda)
\diff\lambda.
\]
Because the weight $e^\lambda$ is known, the two required moments can be
evaluated exactly:
\[
\sigma_{t_i}
\int_{\lambda_{t_{i-1}}}^{\lambda_{t_i}}
e^\lambda\diff\lambda
=
\alpha_{t_i}
\left(
1-e^{-h_i}
\right),
\]
and
\[
\sigma_{t_i}
\int_{\lambda_{t_{i-1}}}^{\lambda_{t_i}}
e^\lambda
\left(
\lambda-\lambda_{t_{i-1}}
\right)
\diff\lambda
=
\alpha_{t_i}
\left(
h_i-1+e^{-h_i}
\right).
\]
Substituting these coefficients yields
\begin{align*}
\tilde\rvx_{t_i}^{\mathrm{poly}}
=
\frac{\sigma_{t_i}}{\sigma_{t_{i-1}}}
\tilde\rvx_{t_{i-1}}
&+
\alpha_{t_i}
\left(
1-e^{-h_i}
\right)
\rmD_{i-1}
\\
&+
\alpha_{t_i}
\left(
h_i-1+e^{-h_i}
\right)
\frac{
\rmD_{i-1}-\rmD_{i-2}
}{
h_{i-1}
}.
\end{align*}
Equivalently,
\[
\tilde\rvx_{t_i}^{\mathrm{poly}}
=
\frac{\sigma_{t_i}}{\sigma_{t_{i-1}}}
\tilde\rvx_{t_{i-1}}
+
\alpha_{t_i}
\left(
1-e^{-h_i}
\right)
\rmD_i^{\mathrm{poly}},
\]
where $\rmD_i^{\mathrm{poly}}$ is defined in
\Cref{subsec:dpm-solver++-multi}.

This shows the precise connection to DEIS. Before the practical coefficient
simplification, the DPM-Solver\texttt{++} second-order multistep update is
exactly what one obtains by applying the degree-one DEIS extrapolation
principle to the data-prediction/log-SNR integral. In this sense, it is the
data-prediction/log-SNR analogue of AB-DEIS-1.

The practical DPM-Solver\texttt{++}(2M) update further simplifies this
coefficient using the leading-order expansion
\[
\beta(h_i)
=
\frac{h_i}{2}
+
\mathcal O(h_i^2),
\]
which gives
\[
\frac{\beta(h_i)}{h_{i-1}}
=
\frac{h_i}{2h_{i-1}}
+
\mathcal O\left(
\frac{h_i^2}{h_{i-1}}
\right).
\]
Accordingly, DPM-Solver\texttt{++}(2M) replaces
$\beta(h_i)/h_{i-1}$ by $h_i/(2h_{i-1})$ in the effective prediction.
Since
\[
\rmD_{i-1}-\rmD_{i-2}
=
\mathcal O(h_{i-1}),
\]
this changes the effective prediction by
\[
\mathcal O\left(
\frac{h_i^2}{h_{i-1}}
\right)
\mathcal O(h_{i-1})
=
\mathcal O(h_i^2).
\]
The coefficient multiplying the effective prediction in the state update is
$\alpha_{t_i}(1-e^{-h_i})=\mathcal O(h_i)$, so the resulting change in the
state update is only $\mathcal O(h_i^3)$. Therefore,
DPM-Solver\texttt{++}(2M) and the degree-one weighted-polynomial update are
generally different finite-step formulas, but they agree through second
order.

This also highlights why the connection to DEIS is specific to
DPM-Solver\texttt{++}(2M): 2M obtains its additional second-order information
by extrapolating previous completed-step predictions, whereas
DPM-Solver\texttt{++}(2S) obtains it from a new intermediate stage within the
current step.

The complete DPM-Solver\texttt{++}(2M) procedure, including its first-order
warm start and the caching of one new data prediction after each step, is
summarized in \Cref{alg:dpmpp-2m}.

\subsection{Closing Remark}
Although our main discussion has focused on DPM-Solver and
DPM-Solver\texttt{++}, it is useful to briefly place them in a broader
context. DPM-Solver is derived in the noise-prediction formulation, whereas
DPM-Solver\texttt{++} reformulates the solver using data prediction, which
can behave more favorably under strong CFG. DPM-Solver-v3~\citep{zheng2023dpm}
takes this idea further by introducing a more general model representation
whose coefficients are chosen using precomputed model statistics to reduce
local discretization error. In this sense, it makes the representation itself
part of the numerical solver design.

Another related direction is the predictor--corrector framework. In the Score
SDE framework (see \Cref{ch:score-sde}), a \emph{predictor} advances the
sample using a numerical step of the reverse-time dynamics, while a
\emph{corrector}, typically a Langevin-type update at the same noise level,
further refines the sample using the score (see \Cref{sec:SMLD}). This
predict--then--refine pattern also has a classical numerical analogue. For
example, Heun's method first predicts an endpoint with Euler and then improves
the numerical update using an averaged slope. Methods such as
UniPC~\citep{zhao2023unipc} develop predictor--corrector constructions
specifically for diffusion sampling. We do not pursue these extensions here,
but they illustrate the same broader theme of this chapter: effective
diffusion solvers combine classical numerical ideas with structure specific
to the diffusion PF-ODE.

\clearpage
\newpage

\section{\texorpdfstring{(Optional) ParaDiGMs}{(Optional) ParaDiGMs}}\label{sec:picard}

\begin{figure}[th!]
  \centering
  \tikzset{
    >={Stealth},
    state/.style={
      rectangle,
      rounded corners=3pt,
      draw=black,
      fill=gray!10,
      minimum size=9mm,       
      inner sep=0.3pt,
      outer sep=0pt,
      align=center,
      text height=1.6ex,      
      text depth=.25ex
    }
  }

  \subcaptionbox{Sequential sampling by time-stepping estimation
      in generation process.\label{fig:chaingraph}}%
      [.485\textwidth][c]{%
        \begin{adjustbox}{max width=\linewidth,center}
          \begin{tikzpicture}[->,thick,font=\small, x=1.4cm, y=1.4cm]
            \useasboundingbox (-0.35,-0.9) rectangle (3.55,0.9);
            \node[state] (x0) at (0,0)   {$\rvx_T$};
            \node[state] (x1) at (1,0)   {$\rvx_{T-1}$};
            \node[state] (x2) at (2,0)   {$\rvx_{T-2}$};
            \node[state] (xT) at (3.2,0) {$\rvx_0$};
            \node at ($(x2)!0.5!(xT)$) {$\dots$};
            \draw (x0) -- (x1);
            \draw (x1) -- (x2);
          \end{tikzpicture}
        \end{adjustbox}
        \vspace{0.36cm}
      }
  \hfill
  \subcaptionbox{Picard iterations with skip dependencies.\label{fig:connectedgraph}}%
      [.485\textwidth][c]{%
        \begin{adjustbox}{max width=\linewidth,center}
          \begin{tikzpicture}[->,thick,font=\small, x=1.4cm, y=1.4cm]
            \useasboundingbox (-0.35,-2.3) rectangle (3.55,0.9);
            \node[state] (x0k)  at (0,0)   {$\rvx_T^{k}$};
            \node[state] (x1k)  at (1,0)   {$\rvx_{T-1}^{k}$};
            \node[state] (x2k)  at (2,0)   {$\rvx_{T-2}^{k}$};
            \node[state] (xTk)  at (3.2,0) {$\rvx_0^{k}$};
            \node[state] (x0kp) at (0,-1.6)   {$\rvx_T^{k+1}$};
            \node[state] (x1kp) at (1,-1.6)   {$\rvx_{T-1}^{k+1}$};
            \node[state] (x2kp) at (2,-1.6)   {$\rvx_{T-2}^{k+1}$};
            \node[state] (xTkp) at (3.2,-1.6) {$\rvx_0^{k+1}$};
            \node at ($(x2k)!0.5!(xTk)$) {$\dots$};
            \node at ($(x2kp)!0.5!(xTkp)$) {$\dots$};
            \draw (x0k) -- (x1kp);
            \draw (x0k) -- (x2kp);
            \draw (x0k) -- (xTkp);
            \draw (x1k) -- (x2kp);
            \draw (x1k) -- (xTkp);
            \draw (x2k) -- (xTkp);
          \end{tikzpicture}
        \end{adjustbox}
      }

  \vspace{-0.4em}
  \caption{\textbfs{Comparisons of two computation graphs.}
Left: conventional time-stepping ODE solving, where the solution is propagated sequentially across time. 
Right: Picard iteration, which enables parallel computation by updating all time nodes simultaneously using the results from the previous iteration, thereby avoiding the strictly sequential nature of time-stepping.
\figcredit{Adapted from \citet{shih2023parallel}.}
}
\end{figure}

\subsection{From Time-Stepping to Time-Parallel Solver}

In the previous sections, we focused on the time–stepping approach, which estimates the trajectory by evolving from the prior time $T$ toward an arbitrary $t\in[0,T]$. 
Let 
\[
\rvv_{\bphi^\times}(\rvx,t):=\rvf(\rvx, t) - \frac{1}{2}g^2(t) \rvs_{\bphi^\times}(\rvx,t)
\]
denote the empirical PF–ODE drift from a pre-trained diffusion model. The exact evolution from $T$ to any intermediate time $t$ is:
\begin{equation}\label{eq:picard_T_to_t}
    \widetilde\bPsi_{T\to t}\big(\rvx(T)\big)
    =
    \rvx(T)
    + \int_{T}^{t}\rvv_{\bphi^\times}\big(\rvx(\tau),\tau\big)\diff\tau,
    \quad \rvx(T)\sim p_{\mathrm{prior}}.
\end{equation}
Time–stepping schemes approximate this integral using discrete updates based on past timesteps.  

In this section, we turn to the \emph{time–parallel} approach, exemplified by \emph{ParaDiGMS}, which builds on classical Picard iteration to enable parallel integration across time. The key idea behind ParaDiGMS is to \emph{trade computational resources for faster simulation}.

\subsection{Methodology of ParaDiGMS}

\paragraph{From Trajectories to Picard Iteration as a Fixed-Point Update.}
Fix the initial state of reverse-time sampling,
\[
\rvx_T\sim p_{\mathrm{prior}}.
\]
The integral expression in \Cref{eq:picard_T_to_t} can be viewed as a map
that takes an entire candidate trajectory and produces an updated one.
Formally, given any path $\{\rvy(\tau)\}_{\tau\in[0,T]}$, define
\[
(\mathcal{L}[\rvy(\cdot)])(t)
=
\rvx_T
+
\int_T^t
\rvv_{\bphi^\times}\big(\rvy(\tau),\tau\big)\diff\tau,
\qquad
t\in[0,T].
\]
That is, $\mathcal{L}$ keeps the initial state $\rvx_T$ fixed and updates
the rest of the trajectory using the velocity evaluated along the candidate
path.

A solution trajectory $\rvx^*(\cdot)$ of the empirical PF-ODE with
$\rvx^*(T)=\rvx_T$ is precisely a \emph{fixed point} of this map:
\[
\rvx^*(t)
=
\mathcal{L}[\rvx^*(\cdot)](t)
=
\rvx_T
+
\int_T^t
\rvv_{\bphi^\times}\big(\rvx^*(\tau),\tau\big)\diff\tau.
\]
Thus, instead of constructing the trajectory strictly one time step after
another, we may search for an entire trajectory that is consistent with this
integral equation.

Starting from an initial guess $\rvx^{(0)}(\cdot)$ satisfying
$\rvx^{(0)}(T)=\rvx_T$, for example the constant path
$\rvx^{(0)}(t)\equiv\rvx_T$, Picard iteration repeatedly applies this map:
\begin{align}\label{eq:picard_iter_T_to_t}
\rvx^{(k+1)}(t)
=
\mathcal{L}[\rvx^{(k)}(\cdot)](t)
=
\rvx_T
+
\int_T^t
\rvv_{\bphi^\times}
\big(\rvx^{(k)}(\tau),\tau\big)\diff\tau,
\end{align}
with $k=0,1,2,\ldots$. Each iteration evaluates the velocity along the previous trajectory and uses
these evaluations to update the trajectory at all time points.

\paragraph{Discrete Picard on a $T$ to $0$ Grid.}
To turn \Cref{eq:picard_iter_T_to_t} into a practical algorithm, we place a
uniform, decreasing grid on $[0,T]$ by choosing a step count $M$, setting
$\Delta t := T/M$, and defining
\[
t_j := T - j \Delta t,
\quad j=0,1,\ldots,M,
\]
so $t_0=T$ and $t_M=0$. Denote sampled iterates by
\[
\rvx^{(k)}_j := \rvx^{(k)}(t_j).
\]
Because the grid runs reversely in time, the
integral from $T$ to $t_j$ has negative orientation. Approximating it by
left endpoints on the partition $\{[t_{i+1},t_i]\}_{i=0}^{j-1}$ gives
\[
\int_{T}^{t_j}\rvv_{\bphi^\times}(\rvx^{(k)}(\tau), \tau) \diff\tau
 \approx 
- \Delta t \sum_{i=0}^{j-1} \rvv_{\bphi^\times}\big(\rvx^{(k)}_{i}, t_i\big),
\]
since each small integral over $[t_{i+1},t_i]$ equals
$- \int_{t_i}^{t_{i+1}} \cdot\diff\tau$. Substituting this approximation
into \Cref{eq:picard_iter_T_to_t} yields the discrete Picard update
\begin{equation}\label{eq:discrete_picard_decreasing}
\rvx^{(k+1)}_j
=
\rvx_T
-
\Delta t
\underbrace{
\sum_{i=0}^{j-1}
\rvv_{\bphi^\times}
\big(\rvx^{(k)}_i,t_i\big)
}_{\substack{\text{cumulative sum}\\\text{of drifts}}},
\qquad
j=1,\ldots,M.
\end{equation}
This scheme is simple and parallel-friendly: each drift evaluation
$\rvv_{\bphi^\times} \big(\rvx^{(k)}_i,t_i\big)$ depends only on the previous
iterate at the same time node $t_i$, so all $i=0,\dots,j - 1$
evaluations can be computed independently across the grid. The integral is then
recovered by a cumulative sum, performed either serially or via a parallel
\emph{prefix-sum (scan/sliding windows)}.

\paragraph{Sliding Windows and Parallel Evaluation.}
\begin{figure}[th!]
    \centering
    \begin{minipage}[t]{0.30\textwidth}
        \centering
        \includegraphics[width=\linewidth]{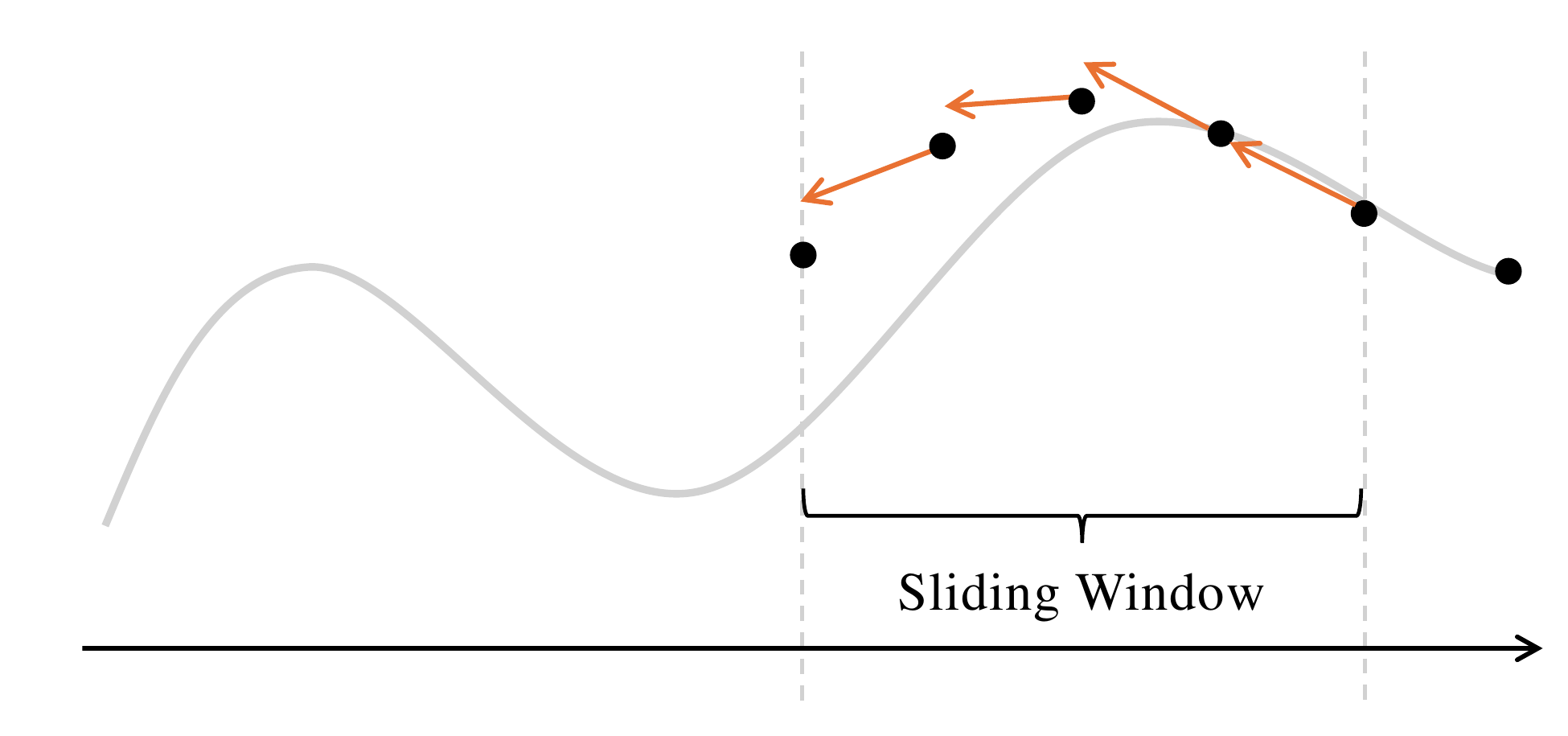}
        \caption{Compute the drift of $\rvx^{(k)}_{\ell:\ell+p}$ on a batch window of size $p=4$, in parallel}
        \label{fig:picard_1}
    \end{minipage}%
    \hspace{0.03\textwidth} 
    \begin{minipage}[t]{0.30\textwidth}
        \centering
        \includegraphics[width=\linewidth]{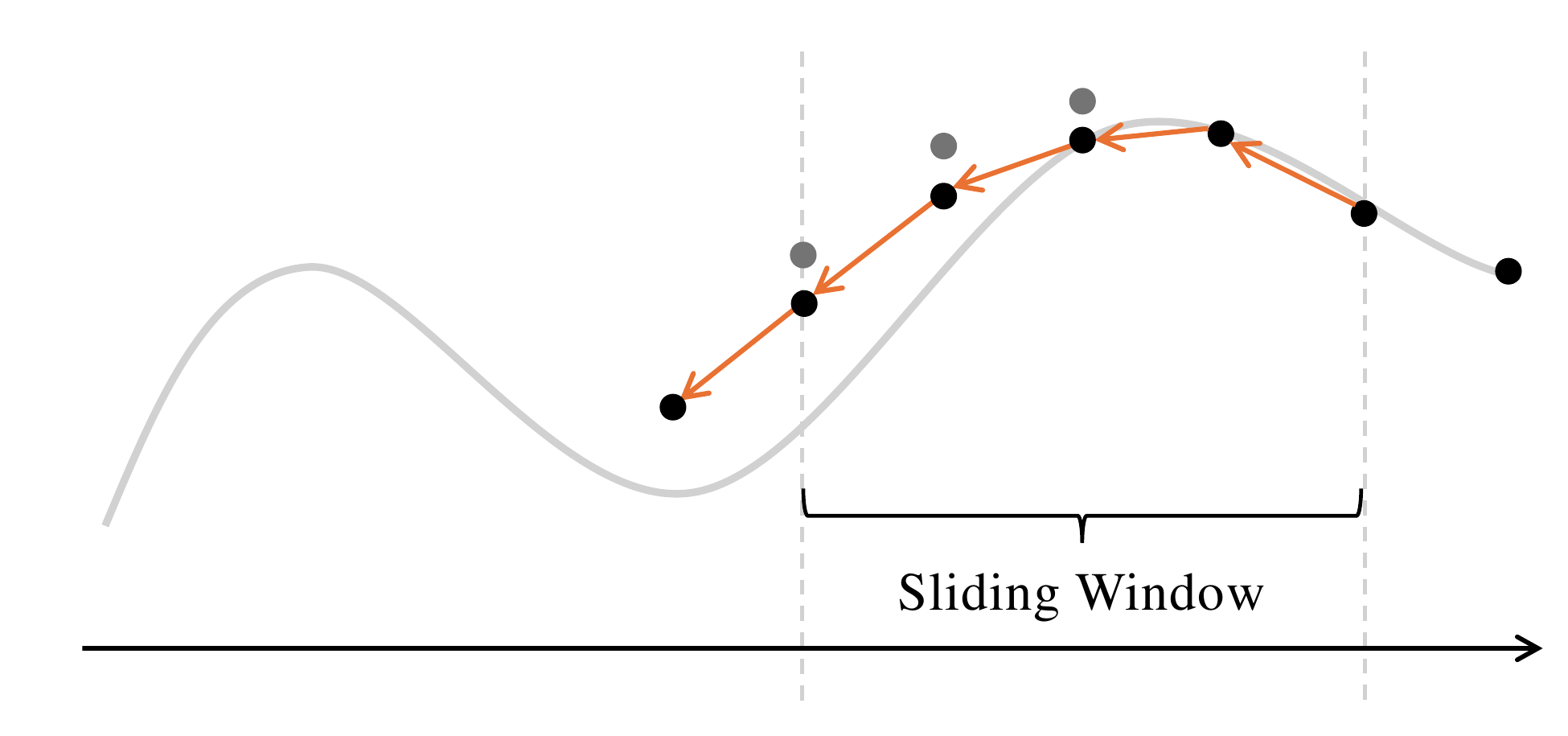}
        \caption{Update the values to $\rvx^{(k+1)}_{\ell:\ell+p}$ using the cumulative drift of points in the window}
        \label{fig:picard_2}
    \end{minipage}%
    \hspace{0.03\textwidth} 
    \begin{minipage}[t]{0.30\textwidth}
        \centering
        \includegraphics[width=\linewidth]{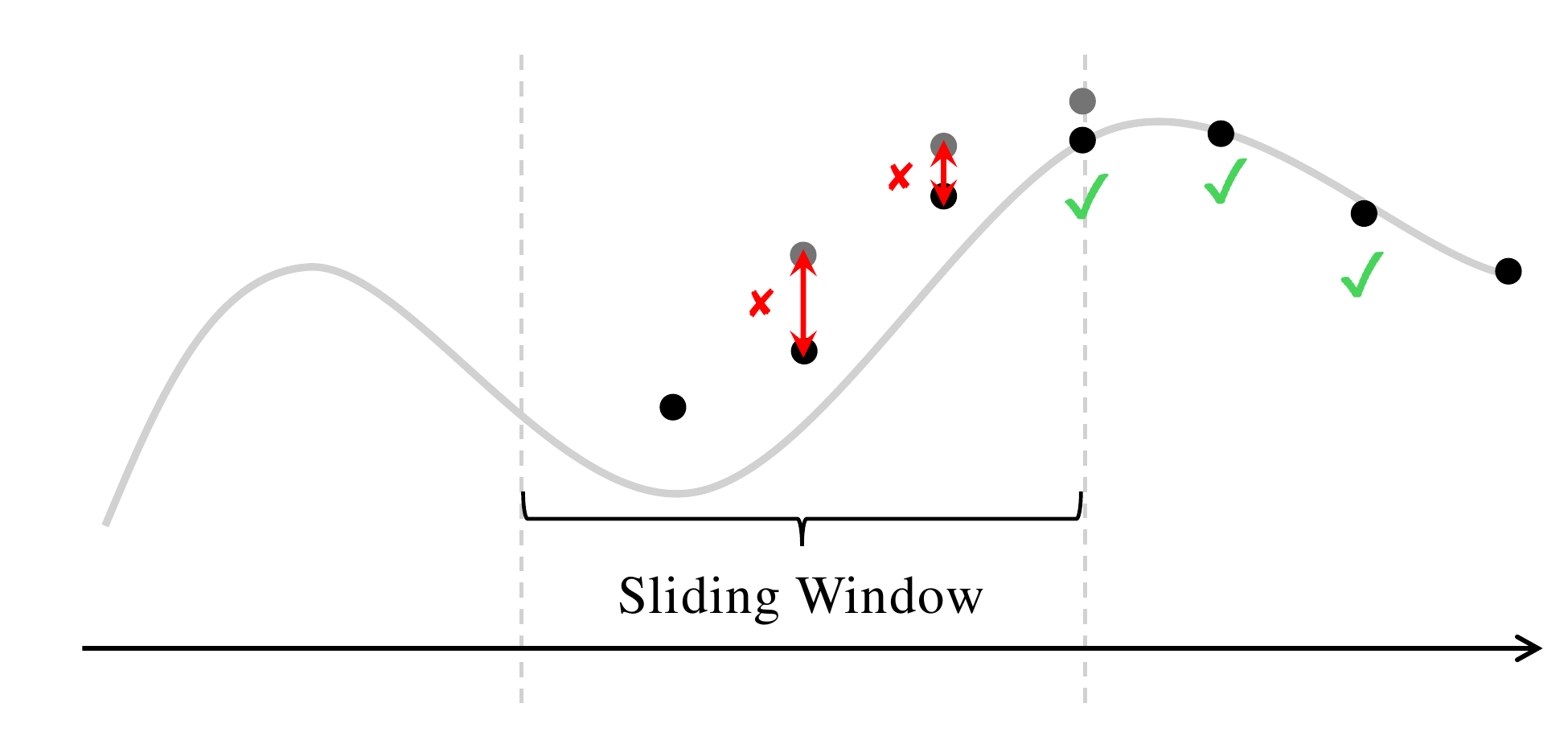}
        \caption{Determine how far to slide the window forward, based on the error $\lVert \rvx^{(k+1)}_i - \rvx^{(k)}_i \rVert^2$.}
        \label{fig:picard_3}
    \end{minipage}
    \figcredit{Adapted from \citet{shih2023parallel}.}
    \label{fig:picard}
\end{figure}

The discrete Picard update \Cref{eq:discrete_picard_decreasing} expresses
each $\rvx^{(k+1)}_j$ as the fixed initial value $\rvx_T$ minus a cumulative
sum of drifts. To limit memory and exploit parallel hardware, it is convenient
to apply the same idea locally on short, sliding blocks of indices.

Fix a window length $p$ and a left index $\ell$; the window then covers
$j=\ell,\ldots,\ell+p$ with $t_\ell>t_{\ell+1}>\cdots>t_{\ell+p}$. During
iteration $k$:

\subparagraph{Step 1. Parallel Drift Evaluation on the Window.}
Compute, in parallel and using only the previous iterate,
\[
\rvv_{\bphi^\times} \big(\rvx^{(k)}_{\ell+i}, t_{\ell+i}\big),
\quad i=0,1,\ldots,p-1.
\]
These are the $p$ local increments needed to advance from the left edge
$t_\ell$ across the window.

\subparagraph{Step 2. Left–Anchored Cumulative Updates.}
Form the windowed updates by anchoring at $j=\ell$ and accumulating the drift across subintervals:
\begin{equation}\label{eq:window_update}
\rvx^{(k+1)}_{\ell+j+1}
 = 
\rvx^{(k)}_{\ell}
 - \Delta t \sum_{i=0}^{j}
\rvv_{\bphi^\times} \big(\rvx^{(k)}_{\ell+i}, t_{\ell+i}\big),
\quad j=0,1,\ldots,p-1.
\end{equation}
This is the local, windowed counterpart of
\Cref{eq:discrete_picard_decreasing}, with the current left-boundary value
$\rvx^{(k)}_\ell$ serving as the anchor for the window. The minus sign
reflects the decreasing time direction. The inner sum is a prefix sum over
the windowed drifts, so all partial sums can be produced efficiently on
parallel hardware.

\subparagraph{Step 3. Progress Control and Window Advance.}
Having formed the left–anchored cumulative updates on the current window (Step~2),
we now decide how far to slide that window. We measure local convergence
by the pointwise Picard change
\[
\mathtt{error}_j
:=
\big\|\rvx^{(k+1)}_{\ell+j}-\rvx^{(k)}_{\ell+j}\big\|^2,
\quad j=1,\ldots,p-1,
\]
and compare it against prescribed tolerances $\mathtt{tol}_{\ell+j}$. That is, $\mathtt{error}_j$ measures how much the iterate at node $\ell+j$ changed during the last Picard
update. If this number is small, it indicates local agreement between the two
successive approximations and hence local convergence of the fixed-point
iteration at that node. If it is large, that node has not settled yet and needs
more Picard smoothing.

The \emph{stride} is chosen as the first index in the window that fails this
test (or the full window length $p$ if none fail):
\[
\mathtt{stride}
:=
\min\Big(\,\{\,j\ge 1:\ \mathtt{error}_j>\mathtt{tol}_{\ell+j}\,\}\ \cup\ \{p\}\Big).
\]
We then slide the window by setting $\ell \leftarrow \ell+\mathtt{stride}$.  In words: we accept all nodes from the left edge up to (but not including) the
first one that has not converged; if all nodes have converged, we accept the
entire window. We then slide the window by that many accepted nodes,
$\ell \leftarrow \ell+\mathtt{stride}$, and continue. This advances by at
most the window length $p$, never skipping any node that has not met its
tolerance. If sliding would overrun the grid end $M$, we truncate the window to
$p \leftarrow \min\{p,\,M-\ell\}$ and proceed.

When the window moves forward it uncovers new time nodes that
have no values yet. To start Picard iteration there, we simply copy the value
from the left boundary of the window and use it as an initial guess. This
``constant extrapolation'' is cheap and stable, and will be corrected by later
updates. If desired, one can replace it by more accurate guesses, such as linear
or polynomial extrapolations from past points.

This completes the procedure of ParaDiGMS. We summarize the algorithm in \Cref{alg:paradigms}.

\begin{algorithm}[t]
  \caption{ParaDiGMS with Sliding Windows}
  \label{alg:paradigms}
  \begin{algorithmic}[1]
    \Require Drift $\rvv_{\bphi^\times}(\rvx,t)$; $\{t_j\}_{j=0}^{M}$; window length $p$; $\{\mathtt{tol}_j\}_{j=1}^{M}$
    \Ensure Approximate trajectory $\{\rvx^{(k)}_j\}_{j=0}^{M}$ with $\rvx^{(k)}_M$ at $t=0$
    \State $k \gets 0$, $\ell \gets 0$ 
    \State Sample $\rvx^{(0)}_0 \sim p_{\mathrm{prior}}$; set $\rvx^{(0)}_j \gets \rvx^{(0)}_0$ for $j=1,\ldots,\min(p,M)$  \Statex \Comment{constant extrapolation}
    \While{$\ell < M$}
      \State $J \gets \min(p,  M-\ell)$   \Comment{current window length}
      \State \textbfs{Step 1: Parallel} 
      \State For $i=0,\ldots,J-1$: $g_i \gets \rvv_{\bphi^\times} \big(\rvx^{(k)}_{\ell+i},\, t_{\ell+i}\big)$ 
      \Statex \Comment{drifts from previous iterate (Picard freezing)}
      \State Compute prefix sums $S_j \gets \sum_{i=0}^{j} g_i$ for $j=0,\ldots,J-1$ 
      \Statex \Comment{scan over windowed drifts}
      \State \textbfs{Step 2: Cumulative Updates} 
      \State For $j=0,\ldots,J-1$: $\displaystyle \rvx^{(k+1)}_{\ell+j+1} \gets \rvx^{(k)}_{\ell} - \Delta t\, S_j$ 
      \Statex \Comment{left-anchored update; cf.\ \Cref{eq:window_update}}
      \State \textbfs{Step 3: Progress Control and Window Advance} 
      \State For $j=1,\ldots,J-1$: $\mathtt{error}_j \gets \big\|\rvx^{(k+1)}_{\ell+j}-\rvx^{(k)}_{\ell+j}\big\|^2$ \Statex \Comment{pointwise Picard change}
      \State $\displaystyle \mathtt{stride} \gets \min \Big(\,\{\,j\in\{1,\ldots,J-1\}:~\mathtt{error}_j>\mathtt{tol}_{\ell+j}\,\}\ \cup\ \{J\}\Big)$
      \State \textbfs{Initialize New Nodes} 
        \State For $r=1,\ldots,\min\{\mathtt{stride},\,M-\ell-J\}$:
        $\rvx^{(k+1)}_{\ell+J+r}
        \gets
        \rvx^{(k+1)}_{\ell+J}$ 
      \Statex \Comment{constant extrapolation into newly exposed indices}
      \State $\ell \gets \ell + \mathtt{stride}$;\quad $k \gets k+1$
    \EndWhile
    \State \Return $\{\rvx^{(k)}_j\}_{j=0}^{M}$
  \end{algorithmic}
\end{algorithm}

\subsection{Relation to Time-Stepping Solvers}

\paragraph{Selection of Sliding Window Size.}
To understand the role of the window size, first consider the smallest
choice $p=1$. In this case, there is no time parallelism, and the method
reduces to its sequential base update. For the left-endpoint discretization
in \Cref{eq:window_update}, this is simply a plain Euler step.

More generally, ParaDiGMS specifies how computations across different time
points can be parallelized, while the numerical update used at each step can
be chosen separately. For example, the Euler update can be replaced by DDIM
or DPM-Solver, giving ParaDDIM or ParaDPMSolver. Increasing $p$ then exposes
more time points to parallel computation without changing the underlying time
grid or the total number of steps $M$. Picard iteration repeatedly corrects
these states toward consistency with the chosen base solver, with convergence
controlled by the prescribed local tolerances.

\paragraph{Compatibility with Higher--Order Solvers (e.g., DPM).}
The Picard framework determines how states at different time points are updated
and evaluated in parallel, while the numerical rule used to approximate each
step can be chosen separately.

For example, instead of the left-endpoint rule in
\Cref{eq:window_update}, we may use the trapezoidal rule:
\begin{align*}
\rvx^{(k+1)}_{\ell+j+1}
=
\rvx^{(k)}_{\ell}
-\Delta t\Bigg[
\tfrac{1}{2}
\rvv_{\bphi^\times}
\big(\rvx^{(k)}_{\ell},t_{\ell}\big)
+
\sum_{i=1}^{j}
\rvv_{\bphi^\times}
\big(\rvx^{(k)}_{\ell+i},t_{\ell+i}\big)
+
\tfrac{1}{2}
\rvv_{\bphi^\times}
\big(\rvx^{(k)}_{\ell+j+1},t_{\ell+j+1}\big)
\Bigg],
\end{align*}
where $j=0,\ldots,p-1$. For $j=0$, this is the familiar one-step trapezoidal update using the two
endpoint drifts. For larger $j$, the same rule accumulates the trapezoidal
contributions across the window. Since all drifts are evaluated using the
previous Picard iterate, they can still be computed in parallel, while the
interior contributions can be accumulated with a prefix sum.

More generally, ParaDiGMS can use an existing sequential sampler such as
DDIM or DPM-Solver as its base discretization. In that case, the Picard
update is constructed from the corresponding base-solver increment rather
than from the trapezoidal formula above. The precise form depends on the
chosen solver, since higher-order methods may use intermediate stages or
previous model evaluations.

For a ParaDiGMS construction based on a given sequential solver, exact
fixed-point convergence recovers the corresponding sequential discrete
trajectory. With a finite stopping tolerance, the final result also contains
a remaining fixed-point iteration error in addition to the discretization
error of the base solver.

\newpage
\section{Closing Remarks}\label{sec:ch9_cr}

This chapter has confronted one of the most significant practical limitations
of diffusion models: their slow, iterative sampling process. We have explored
training-free approaches that accelerate generation by designing better
numerical solvers for the empirical PF-ODE:
\begin{enumerate}
    \item We began with DDIM, a first-order factored exponential-integrator
    update that retains the known diffusion schedule analytically while
    holding the neural prediction fixed.
    \item We then studied DEIS, which extends this idea to higher-order
    multistep updates by reusing past model evaluations to extrapolate how the
    neural prediction changes.
    \item Finally, we examined the DPM-Solver family, which introduces the
    log-SNR time coordinate and constructs higher-order updates from
    within-step model evaluations. DPM-Solver\texttt{++} further develops
    single-step and multistep variants in the data-prediction formulation.
\end{enumerate}

Together, these methods illustrate a common principle: exploit the analytically
known diffusion structure and concentrate numerical approximation on the
learned model prediction. They differ in the prediction and time
parameterizations, and in whether higher-order information is obtained from
within-step stages or previous-step history. ParaDiGMS provides a complementary
direction by reorganizing computation across time through fixed-point
iteration and parallel evaluation. Such solver developments can reduce the
sampling cost from hundreds or thousands of model evaluations to only tens in
many practical settings.

However, these training-free methods remain fundamentally iterative: they
numerically approximate a continuous generative trajectory, even when parts of
the computation can be parallelized. This raises a natural question:
\emph{can we achieve high-quality generation in just one or a very few steps?}

The final Part~D of this monograph explores this question through
training-based acceleration. In \Cref{ch:distillation}, we study
distillation-based methods, where a fast \textit{student} generator learns to
reproduce the behavior of a slow pre-trained \textit{teacher}. In
\Cref{ch:fast-scratch}, we turn to methods that directly learn few-step
generators from scratch defined by an ODE flow.

This shift from designing a better numerical solver to directly learning a
fast approximation of the generative solution map aims to combine the quality
of diffusion models with the speed of one- or few-step generation.

\clearpage
\newpage

\part{Toward Learning Fast Diffusion-Based Generators}

\chapter{Distillation-Based Methods for Fast Sampling}\label{ch:distillation}

This chapter introduces training-based approaches that accelerate diffusion model sampling by teaching new generators to produce samples in only one or a few steps. The central idea, called \emph{distillation}, is to let a fast student model learn from a slow, pre-trained diffusion model (teacher) sampler. While the teacher may require hundreds of steps, the student can achieve comparable quality in only a few steps\footnote{Here, distillation means reducing sampling steps rather than model size.}. Unlike solver-based acceleration, which improves the numerical integration scheme, distillation directly trains a generator to take efficient shortcuts. We highlight two main paradigms: \emph{distribution level distillation}, which skips simulating the full trajectory and instead aligns the student’s output distribution with the teacher’s, and \emph{flow map level distillation}, which trains the student to reproduce the teacher’s sampling path in a faster and more compact way.

\chapterlearning{

  \item distinguish the two main distillation targets: the teacher's distribution and its state-to-state flow map;

  \item understand distribution-level distillation, where the student matches the teacher's noisy marginals through score-based supervision;

  \item understand flow-map distillation, where the student learns teacher-generated transitions, with Progressive Distillation progressively replacing two teacher steps by one student step.

}{With a pre-trained diffusion model as the teacher, the key distinction in this chapter is what the student is trained to reproduce: the teacher's distribution or its flow map. The next chapter asks how such fast maps can be learned without a teacher.}


\clearpage
\newpage

\section{Prologue}\label{sec:distillation-prologue}
A central bottleneck of diffusion models is their slow sampling speed.  

As shown through Tweedie's formula (\Cref{subsec:four-equiv-para}), a diffusion model can be interpreted as an ``$\rvx$-prediction'' model, $\rvx_{\bm{\phi}^\times}(\rvx_t, t)$, trained to recover the expected clean data from a noisy input $\rvx_t$ at noise level $t$:
\[
    \rvx_{\bm{\phi}^\times}(\rvx_t, t) \approx \mathbb{E}[\rvx_0 |\rvx_t],
\]
where the expectation is taken with respect to $p(\rvx_0 |\rvx_t)$, representing all plausible clean data corresponding to $\rvx_t$.  
A natural idea is to use $\rvx_{\bm{\phi}^\times}(\rvx_t, t)$ for one step generation. Yet because this denoiser averages over many plausible outcomes, the prediction can become overly smooth, and generation with only a few denoising steps leads to blurry, low quality samples.  

On the other hand, as discussed in \Cref{subsec:sde-sample}, diffusion sampling follows an ODE or SDE trajectory through a long sequence of iterative steps. This produces high fidelity samples, but repeated neural-network evaluations make sampling expensive. For a deterministic ODE solver of order $n$, under standard regularity assumptions, the local truncation error is $\mathcal{O}(h^{n+1})$ and the global discretization error over a fixed time interval is $\mathcal{O}(h^n)$, where $h=\max_i |t_i-t_{i-1}|$ is the step size. Thus, using a coarser discretization generally increases numerical error for a fixed learned vector field, although sample fidelity need not decrease monotonically with the number of solver steps. This creates a practical trade off between sampling accuracy and computational cost. 

To overcome this bottleneck, a major line of research is \emph{distillation}, which assumes access to a well-trained diffusion model (the \emph{teacher}) and trains a generator (the \emph{student}) to reproduce its behavior through a single feed forward or few step computation. 
This compresses the teacher’s many sampling steps into a fast process, effectively bypassing slow iterative solvers while maintaining high sample fidelity.

Below, we introduce two perspectives on distillation: \emph{distribution level distillation} and \emph{flow map level distillation}\footnote{Chronologically, flow map level distillation appeared earlier, with \emph{Knowledge Distillation} (KD)~\citep{luhman2021knowledge} in 2021 and \emph{Progressive Distillation} (PD)~\citep{salimans2021progressive} in 2022, preceding the family of distribution level distillation approaches that emerged around 2023. For smoother exposition and connection to the next chapter, however, we present distribution level distillation first.}.

\subsection{Distribution Level Distillation}  
The goal of distribution-based distillation is to train a one-step generator $\rmG_{\bm{\theta}}(\rvz)$ that maps noise $\rvz \sim p_{\mathrm{prior}}$ to a sample $\hat\rvx = \rmG_{\bm{\theta}}(\rvz)$, inducing a distribution $p_{\bm{\theta}}(\hat\rvx)$ that approximates the target data distribution $p_{\mathrm{data}}(\rvx)$. This is typically achieved by minimizing a statistical divergence
\[
\min_{\bm{\theta}}  \mathcal{D}\!\left(p_{\bm{\theta}}(\hat\rvx),  p_{\mathrm{data}}(\hat\rvx)\right),
\]
where $\mathcal{D}$ denotes a suitable divergence measurement such as KL.  

In practice, distribution based methods align the generator’s distribution with the empirical distribution $p_{\bm{\phi}^\times}(\rvx)$ produced by a pre-trained diffusion model:
\[
\min_{\bm{\theta}}  \mathcal{D}\!\left(p_{\bm{\theta}}(\hat\rvx),  p_{\bm{\phi}^\times}(\hat\rvx)\right),
\]
where $p_{\bm{\phi}^\times}$ serves as a surrogate for $p_{\mathrm{data}}$. Rather than evaluating this divergence explicitly, these methods construct tractable estimators of its gradient using the pre-trained teacher together with quantities estimated from the student's current distribution. For KL-based distribution matching, this gradient takes the form of a difference between the student and target score functions.

This formulation distills multi-step generative processes of diffusion models into a single step model through distributional alignment. We detail this approach in \Cref{sec:VSD}.

\subsection{Flow Map Level Distillation} We consider the PF-ODE, which can be expressed for any prediction model (see \Cref{eq:predictions-equivalence}):
\begin{align}\label{eq:pf-ode-ct}
    \frac{\diff \rvx(\tau)}{\diff \tau} 
    = f(\tau) \rvx(\tau) - \tfrac{1}{2}g^2(\tau) \nabla_\rvx \log p_\tau(\rvx(\tau)) 
    =: \rvv^*(\rvx(\tau), \tau).
\end{align}

Its solution map (flow map), starting from $\rvx_s$ at time $s$ and evolving reversely to time $t \leq s$, is denoted by $\bPsi_{s\to t}(\rvx_s)$; that is,
\begin{align}\label{eq:int_target_gt}
\begin{aligned}
    \bPsi_{s\to t}(\rvx_s) 
    &\coloneqq \rvx_s + \int_s^t \rvv^*(\rvx(\tau), \tau) \diff \tau,
\end{aligned}
\end{align}
where the integral solves the PF-ODE. Intuitively, $\bPsi_{s\to t}$ transports, $\rvx_s$, noise at time $s$ to less noisy states at time $t$ (ultimately data at $t=0$).

Sampling from a diffusion model corresponds to evaluating $\bPsi_{T\to 0}(\rvx_T)$ for $\rvx_T \sim p_{\mathrm{prior}}$. Typically, this integral is approximated by iterative numerical solvers leveraging the velocity field $\rvv$ (see \Cref{ch:solvers}), but requires many steps (e.g., at least $10$-$20$ NEFs even in DPM-Solver), making sampling slower than classic one-step generative models such as GAN. This motivates a natural question:
\begin{question}
    Can we learn the solution map $\bPsi_{s\to t}(\rvx_s)$ directly? 
\end{question}
In particular, learning a map $\bPsi_{T\to 0}(\rvx_T)$ with $\rvx_T \sim p_{\mathrm{prior}}$ enables one-step generation.

\paragraph{Trajectory Distillation.}

Trajectory distillation seeks to train a neural generator that approximates the solution map at the instance level. Since the PF-ODE integral rarely admits a closed form, it must be approximated numerically during training. To formalize, we introduce the general solver notation
\begin{align}\label{eq:solver}
    \mathtt{Solver}_{s\to t}(\rvx_s;\,\bm{\phi}^\times)
    \quad \text{or simply} \quad
    \mathtt{Solver}_{s\to t}(\rvx_s),
\end{align}
denoting numerical integration of the empirical PF-ODE from $s$ to $t$ starting
at $\rvx_s$, with teacher parameters $\bm{\phi}^\times$ (omitted when clear from
context).

\paragraph{An Early Approach: Direct Knowledge Distillation.}
To enable few step or even one step generation, a direct approach is to train a generator $\rmG_\btheta(\rvx_T, T,0)$ to imitate the output of a numerical solver evaluated along the full trajectory:
\[
\rmG_\btheta(\rvx_T, T,0)\approx \mathtt{Solver}_{T\to 0}(\rvx_T), 
\qquad \rvx_T \sim p_{\mathrm{prior}}.
\]
This idea underlies one of the earliest trajectory distillation methods, \emph{Knowledge Distillation}~\citep{luhman2021knowledge}, which uses the regression loss
\[
\mathcal{L}_{\mathrm{KD}}(\btheta)
:= 
\E_{\rvx_T \sim p_{\mathrm{prior}}}
\left\|
\rmG_\btheta(\rvx_T, T,0)
-
\mathtt{Solver}_{T\to 0}(\rvx_T)
\right\|_2^2.
\]
While this approach provides direct supervision from the pre-trained teacher,
it cannot leverage the strong supervision available in the original training data.
In addition, it is computationally expensive if ODE integration is invoked within
the training loop, since each parameter update requires solving the ODE to form
targets. Finally, because the generator learns only a global mapping from $T$ to $0$, it may lose controllability for steering the generation process from intermediate states. Consequently, most controllable generation techniques introduced in \Cref{ch:guidance} cannot be directly applied.

\paragraph{Preface to Progressive Distillation.}
\emph{Progressive Distillation} (PD)~\citep{salimans2021progressive} trains a time–conditional $ \mathtt{Student} $ using \emph{local} supervision from $\mathtt{Teacher}$ fragments. 
Let $t_0=T>t_1>\cdots>t_N=0$ be a fixed time grid. 
The $ \mathtt{Teacher} $ provides time–stepping maps $\mathtt{Teacher}_{t_k\to t_{k+1}}$ for $k=0,\ldots,N-1$.

Rather than supervising only the one-jump $T\to 0$, PD trains the $ \mathtt{Student}  $ two–step skip map to match two consecutive $ \mathtt{Teacher}$ steps:
\[
\mathtt{Student}_{t_k\to t_{k+2}}
\;\approx\;
\mathtt{Teacher}_{t_{k+1}\to t_{k+2}}
\circ
\mathtt{Teacher}_{t_{k}\to t_{k+1}},
\]
for $k = 0, 2, 4, \dots$.
The matching is performed using a simple regression loss (e.g., mean squared error).

After training on locally paired fragments, the $ \mathtt{Student} $ no longer follows every time interval of the original grid. 
Instead, it advances on every other time point,
\[
t_0 \to t_2 \to t_4 \to \cdots \to t_N,
\]
which means that each $ \mathtt{Student} $ step effectively covers two consecutive $ \mathtt{Teacher}$  steps. 
Consequently, the $ \mathtt{Student} $ completes the same overall time span $[0,T]$ using only $N/2$ transitions. 

After this stage, the trained $ \mathtt{Student} $ replaces the $ \mathtt{Teacher}$  to serve as the new reference model. 
The entire procedure is then repeated on the coarser grid, where the time step doubles 
$(N \to N/2 \to N/4 \to \cdots)$,
progressively distilling the trajectory into fewer and fewer steps until the desired number of inference steps is reached. 
This iterative halving preserves the global time horizon while continually compressing the temporal resolution of the generative process.

\paragraph{A Unified Perspective of Flow Map Learning.}
Various methods, including KD and PD, can be expressed within a unified loss framework:
\begin{align}\label{eq:general-cm-loss}
  \mathcal{L}_{\mathrm{oracle}}(\btheta)
  := \E_{s,t} \E_{\rvx_s \sim p_s} 
  \left[  w(s,t)  d \big(\rmG_\btheta(\rvx_s, s,t),  \bPsi_{s\to t}(\rvx_s)\big)\right],
\end{align}
where $\bPsi_{s\to t}$ is the oracle flow map, $w(s,t) \geq 0$ specifies how different time pairs $(s,t)$ are weighted,
$d(\cdot,\cdot)$ is a discrepancy measure such as
$d(\rvx,\rvy)=\|\rvx-\rvy\|_2^2$ or $d(\rvx,\rvy)=\|\rvx-\rvy\|_1$,  and $p_s$ denotes the
forward noised marginal at time $s$. Because $\bPsi_{s\to t}$ is not available
in closed form, one must rely on approximations, typically through a pre-trained
diffusion model (teacher) or another tractable surrogate.

KD appears as a simple instance of \Cref{eq:general-cm-loss} by considering only the endpoint pair $(s,t)=(T,0)$. Equivalently, one may take the degenerate weighting
$w(s,t)=\delta(s{-}T)\delta(t)$.
Assuming $p_T=p_{\mathrm{prior}}$\footnote{This equality holds when the noise schedule reaches the prior exactly at $T$; otherwise, one typically has $p_T\approx p_{\mathrm{prior}}$ for sufficiently large $T$.}, the oracle objective becomes
\[
\E_{\rvx_T\sim p_T}
\big\|\rmG_\btheta(\rvx_T,T,0)-\bPsi_{T\to 0}(\rvx_T)\big\|_2^2
\approx \mathcal{L}_{\mathrm{KD}}(\btheta),
\]
where the practical KD target replaces the oracle flow map by
$\mathtt{Solver}_{T\to 0}\approx\bPsi_{T\to 0}$.
An alternative perspective on this formulation is presented in
\Cref{app-sec:flow-map}.

PD also fits this template, but instead of supervising only with the single extreme pair $(T,0)$, it uses many \emph{nearby} time pairs and enforces a simple \emph{local consistency} rule:
a short step followed by another short step should match the direct two–step move. We return to this in \Cref{eq:pd-local}.

In practice, the main challenge is that the oracle flow map $\bPsi_{s\to t}$ generally has no closed-form expression, making direct supervision infeasible.  A range of methods have been developed to approximate this
target efficiently, but their success often hinges on the quality of the teacher
model. We will return to \Cref{eq:general-cm-loss} in \Cref{ch:fast-scratch}, presenting a principled framework for training-from-scratch methods that eliminate the teacher from the learning loop.

\clearpage
\newpage

\section{\texorpdfstring{Distribution-Based Distillation}{Distribution-Based Distillation}}\label{sec:VSD}

Several works have pursued this distribution-based distillation concurrently under different names,
including Distributional Matching Distillation (DMD)~\citep{yin2024one,yin2024improved},
Variational Score Distillation (VSD)~\citep{pooledreamfusion,wang2023prolificdreamer,luo2023diff,lu2024simplifying},
and Score Identity Distillation (SiD)~\citep{zhou2024score}.
Despite technical differences, they share the same principle:
train a generator whose forward-noised marginals match those of the teacher.
We focus on VSD as a representative formulation, since the others follow similar principles.\footnote{
Historically, VSD~\citep{wang2023prolificdreamer} was introduced for text-to-3D generation, where a pre-trained 2D diffusion model provides distributional supervision through a differentiable renderer. Its single-particle (Dirac) specialization recovers the basic SDS~\citep{pooledreamfusion} update, up to the corresponding time-weighting and prediction-parameterization conventions.
}

\subsection{Formulation of VSD as a Representative Approach}
\paragraph{Forward Process.}
Let $\{p_t\}_{t \in [0,T]}$ denote the marginal densities of a forward diffusion process induced by
\[
\rvx_t = \alpha_t \rvx_0 + \sigma_t \bm{\epsilon}, \quad \bm{\epsilon} \sim \mathcal{N}(\bm{0}, \rmI),
\]
with initial distribution $p_0 = p_{\mathrm{data}}$.  
In contrast, let $p_0^{\bm{\theta}}$ denote the distribution of synthetic samples generated by a deterministic one-step generator $\rmG_{\bm{\theta}}(\rvz)$ from latent variables $\rvz \sim p_{\mathrm{prior}}(\rvz)$. Define $\{p_t^{\bm{\theta}}\}_{t \in [0,T]}$ as the marginal densities obtained by applying the same forward diffusion process to $p_0^{\bm{\theta}}$, that is,
\begin{align}\label{eq:vsd-forward}
    \rvx_t^{\bm{\theta}} := \alpha_t \rmG_{\bm{\theta}}(\rvz) + \sigma_t \bm{\epsilon}, 
\end{align}
where $\rvz \sim p_{\mathrm{prior}}$ and $ \bm{\epsilon} \sim \mathcal{N}(\bm{0}, \rmI)$. Thus, both $p_t$ and $p_t^{\bm{\theta}}$ share the same Gaussian diffusion kernel 
$p_t(\rvx_t| \rvx_0)$ but differ in their starting distributions 
($p_{\mathrm{data}}$ vs.\ $p_0^{\bm{\theta}}$ of one-step synthetic samples).

\paragraph{Training Objective and Gradient.}
The literature typically adopts the KL divergence to match the distributions $p_t$ and $p_t^{\bm{\theta}}$, commonly by minimizing
\begin{align*}
    \mathcal{L}_{\text{VSD}}(\bm{\theta}) :=\mathbb{E}_t \left[ \omega(t)  \mathcal{D}_{\text{KL}}(p_t^{\bm{\theta}} \, \| \, p_t) \right]
= \mathbb{E}_{t, \rvz, \bm{\epsilon}} \left[ \omega(t) \left( \log p_t^{\bm{\theta}}(\rvx_t^{\bm{\theta}}) - \log p_t(\rvx_t^{\bm{\theta}}) \right) \right],
\end{align*}
where $\omega(t)\geq 0$ is a time-dependent weighting function. We will discuss in \Cref{subsec:vsd-discussion} why the KL divergence plays a
special role in distribution-level distillation.

As shown in \citep{wang2023prolificdreamer}, at the population level, the global optimum is achieved when
$p_0^{\bm{\theta}^*}=p_{\mathrm{data}}$, indicating that the generator's distribution matches the data distribution. Thus, the VSD objective provides a principled distribution-matching loss for learning $p_{\mathrm{data}}$.

However, the density-based formulation of the objective lacks an efficient training mechanism. Fortunately, by taking the gradient with respect to $\bm{\theta}$, we arrive at the expression in \Cref{eq:vsd-grad}, which is summarized in the following proposition. For notational simplicity, we denote $\hat\rvx_t := \rvx_t^{\bm{\theta}}$ as defined in \Cref{eq:vsd-forward}.

\proppp{$\bm{\theta}$-Gradient of $\mathcal{L}_{\text{VSD}}$}{vsd-grad}{
We have
\begin{align}\label{eq:vsd-grad}
\begin{aligned}
        \nabla_{\bm{\theta}}\mathcal{L}_{\text{VSD}}(\bm{\theta}) = \mathbb{E}_{t, \rvz, \bm{\epsilon}} \left[ \omega(t)\alpha_t  \left( 
\nabla_{\rvx} \log p_t^{\bm{\theta}}(\hat\rvx_t) - \nabla_{\rvx} \log p_t(\hat\rvx_t) 
\right) \cdot \partial_{\bm{\theta}} \rmG_{\bm{\theta}}(\rvz) \right].
\end{aligned}
\end{align}
}{
The derivation applies the chain rule:
\begin{align*}
&\nabla_{\btheta}\mathcal{L}_{\text{VSD}}(\btheta)
=
\nabla_{\btheta}\E_t
\left[
\omega(t)\mathcal{D}_{\mathrm{KL}}(p_t^{\btheta}\|p_t)
\right]
\\
&= \E_{t,\rvz,\beps}\!\left[
\omega(t)\,
\partial_\btheta
\big(\log p_t^{\btheta}(\hat\rvx_t)-\log p_t(\hat\rvx_t)\big)
\right] \\
&= \E_{t,\rvz,\beps} \left[
\omega(t)\left(
\underbrace{\partial_{\btheta}\log p_t^{\btheta}(\hat\rvx_t)}_{\text{first}}
+ (\nabla_{\rvx}\log p_t^{\btheta}(\hat\rvx_t))^\top \partial_{\btheta}\hat\rvx_t
- (\nabla_{\rvx}\log p_t(\hat\rvx_t))^\top \partial_{\btheta}\hat\rvx_t
\right)\right].
\end{align*}
The first term vanishes by the score-function identity:
\[
\E_{\hat\rvx_t\sim p_t^{\btheta}}\!\left[\partial_{\btheta}\log p_t^{\btheta}(\hat\rvx_t)\right]
= \int \partial_{\btheta} p_t^{\btheta}(\rvx)  \diff\rvx
= \partial_{\btheta}  \int p_t^{\btheta}(\rvx)  \diff\rvx
= \partial_{\btheta}(1)=0.
\]
Using the reparameterization $\hat\rvx_t=\alpha_t\rmG_{\btheta}(\rvz)+\sigma_t\beps$ gives
$\partial_{\btheta}\hat\rvx_t=\alpha_t\partial_{\btheta}\rmG_{\btheta}(\rvz)$, hence
\[
\nabla_{\btheta}\mathcal{L}_{\text{VSD}}(\btheta)
= \E_{t,\rvz,\beps} \left[\omega(t) \alpha_t 
\big(\nabla_{\rvx}\log p_t^{\btheta}(\hat\rvx_t)-\nabla_{\rvx}\log p_t(\hat\rvx_t)\big)^\top
\partial_{\btheta}\rmG_{\btheta}(\rvz)\right].
\]
This proves \Cref{eq:vsd-grad}.
See \Cref{app-sec:flow-map} for details.
}

We observe that the score functions naturally emerge when taking the gradient
with respect to $\bm{\theta}$. Consequently, we require approximations of
the score $\nabla_{\rvx}\log p_t^{\bm{\theta}}(\hat\rvx_t)$ for the
one-step generator and $\nabla_{\rvx}\log p_t(\hat\rvx_t)$ for the data
distribution, as will be detailed in the following subsection.

\subsection{Training Pipeline of VSD}\label{subsec:vsd-training}
Existing works~\citep{yin2024one,yin2024improved,pooledreamfusion,wang2023prolificdreamer,luo2023diff,lu2024simplifying} typically address this via a bi-level optimization approach: training a new diffusion model on samples from $\rmG_{\bm{\theta}}(\rvz)$ to approximate $\nabla_{\rvx} \log p_t^{\bm{\theta}}(\hat\rvx_t)$, and employing a pre-trained diffusion model to provide a proxy for the intractable oracle score function $\nabla_{\rvx} \log p_t(\hat\rvx_t)$ on synthetic samples $\hat\rvx_t$. More precisely, training proceeds by alternating between two phases:
\begin{itemize}
    \item \textbfs{Score Estimation Phase.} Fix $\bm{\theta}$. Let $\hat{\rvx}_0=\rmG_{\bm{\theta}}(\rvz)$ and
    $\hat{\rvx}_t=\alpha_t \hat{\rvx}_0+\sigma_t\beps$ with $\rvz\sim p_{\mathrm{prior}}$, $\beps\sim\mathcal{N}(\bm{0},\rmI)$.
    Train $s_{\bm{\zeta}}$ by DSM using the known Gaussian diffusion kernel $p_t(\rvx_t|\rvx_0)$:
    \[
    \mathcal{L}_{\text{DSM}}(\bm{\zeta};\bm{\theta})
    =\E_{t,\rvz,\beps}\Big\|\,\rvs_{\bm{\zeta}}(\hat{\rvx}_t,t)
    - \nabla_{\rvx_t}\log p_t(\hat{\rvx}_t|\hat{\rvx}_0)\,\Big\|^2,
    \]
    which yields $\rvs_{\bm{\zeta}}(\cdot,t)\approx \nabla_{\rvx}\log p_t^{\bm{\theta}}(\cdot)$ at optimum (for fixed $\bm{\theta}$).

  \item \textbfs{Generator Update Phase.} With $s_{\bm{\zeta}}$ frozen (stop-grad),  $\bm{\theta}$ is updated by using the gradient in \Cref{eq:vsd-grad}, replacing the individual score terms by their respective proxies:
  \[
    \rvs_{\bm{\zeta}}(\hat\rvx_t,t)\approx \nabla_{\rvx}\log p_t^{\bm{\theta}}(\hat\rvx_t), \,\, \text{and} \,\, \rvs_{\bm{\phi}^\times}(\hat{\rvx}_t,t)\approx \nabla_{\rvx}\log p_t(\hat\rvx_t)\ \ \text{(teacher)}.
  \]
  \Cref{eq:vsd-grad} then approximately becomes:
  \[
    \nabla_{\bm{\theta}}\mathcal{L}_{\text{VSD}}(\bm{\theta})
    \approx \mathbb{E}_{t,\rvz,\bm{\epsilon}}\Big[\omega(t)\alpha_t
      \big(\rvs_{\bm{\zeta}}(\hat{\rvx}_t,t) - \rvs_{\bm{\phi}^\times}(\hat{\rvx}_t,t)\big)^{\top}
      \partial_{\bm{\theta}}\rmG_{\bm{\theta}}(\rvz)
    \Big].
  \]
\end{itemize}
These two phases are alternated throughout training. Ideally, the auxiliary
model tracks the current student score
$\rvs_{\bm{\zeta}}(\cdot,t)\approx\nabla_{\rvx}\log p_t^{\bm{\theta}}(\cdot)$,
while the pre-trained teacher provides an approximation to the target score
$\rvs_{\bm{\phi}^\times}(\cdot,t)\approx\nabla_{\rvx}\log p_t(\cdot)$.
At the population level, if these two score functions agree at a noised level
with $\sigma_t>0$ and $\alpha_t\neq 0$, then their corresponding normalized
densities agree, $p_t^{\bm{\theta}}=p_t$. Since Gaussian noising is injective
when $\alpha_t\neq 0$, this further implies
$p_0^{\bm{\theta}}=p_{\mathrm{data}}$.

In practice, both scores are represented by finite learned networks, so these
equalities characterize the ideal population solution rather than a guarantee
of alternating optimization.

\subsection{Beyond KL: General Divergences and Why KL Is Special}
\label{subsec:vsd-discussion}

The KL divergence is not essential to the distribution-matching principle of VSD.
In principle, one may use a general $f$-divergence (see \Cref{eq:f-div}):
\[
\mathcal{D}_f(p_t^{\btheta}\|p_t)
=
\int p_t(\rvx)\,
f\!\left(
\frac{p_t^{\btheta}(\rvx)}{p_t(\rvx)}
\right)\diff \rvx.
\]
What makes the reverse KL particularly convenient is its gradient structure.
Define the density ratio
\[
r_t(\rvx)
:=
\frac{p_t^{\btheta}(\rvx)}{p_t(\rvx)}.
\]
For a twice-differentiable convex $f$ and positive densities, the pathwise
gradient can be written as
\[
\nabla_{\btheta}\mathcal{D}_f(p_t^{\btheta}\|p_t)
=
\E_{\hat\rvx_t\sim p_t^{\btheta}}
\left[
r_t(\hat\rvx_t)f''\!\big(r_t(\hat\rvx_t)\big)
\big(
\nabla_{\rvx}\log p_t^{\btheta}(\hat\rvx_t)
-
\nabla_{\rvx}\log p_t(\hat\rvx_t)
\big)^\top
\partial_{\btheta}\hat\rvx_t
\right].
\]
Conceptually, the gradient has three components:
\[
\begin{aligned}
\nabla_{\btheta}\mathcal{D}_f
=
\E\Big[
\underbrace{r_t f''(r_t)}_{\text{density-ratio weighting}}
\times
\underbrace{
\big(
\nabla_{\rvx}\log p_t^{\btheta}
-
\nabla_{\rvx}\log p_t
\big)^\top
}_{\text{score mismatch}}
\times
\underbrace{
\partial_{\btheta}\hat\rvx_t
}_{\text{parameter sensitivity}}
\Big].
\end{aligned}
\]
The first term determines \emph{where and how strongly} to update, the second
gives the score-mismatch direction, and the last maps this sample-space
signal back to the generator parameters.

This expression reveals a useful common structure: all such $f$-divergences
use the same \emph{score-mismatch direction},
\[
\nabla_{\rvx}\log p_t^{\btheta}(\rvx)
-
\nabla_{\rvx}\log p_t(\rvx)
=
\nabla_{\rvx}\log r_t(\rvx),
\]
while the choice of divergence changes how strongly different regions are
weighted through $r_t f''(r_t)$. Thus, for the usual strictly convex choices of $f$, the population
distribution-matching condition remains $p_t^{\btheta}=p_t$, while the choice
of divergence changes how different density-ratio regimes are weighted during
optimization through $r_t f''(r_t)$.

For the reverse KL used in VSD,
$f(r)=r\log r$, and therefore
\[
r f''(r)=1.
\]
The density-ratio weighting disappears, recovering the score-difference
gradient in \Cref{eq:vsd-grad}. Consequently, once the student and target
scores are available, reverse-KL training does not require an additional
pointwise density-ratio estimator. By contrast, a general $f$-divergence
requires the factor $r_t f''(r_t)$; although the score difference provides
$\nabla_{\rvx}\log r_t$, the ratio $r_t$ itself is not directly available
for an implicit generator and generally requires additional estimation.

For example, the forward KL corresponds to $f(r)=-\log r$, for which
$r f''(r)=1/r$. It therefore places greater weight on regions where the
model density is small relative to the target density. This illustrates how
different divergences can share the same underlying score-mismatch direction
while emphasizing different density-ratio regimes.

Finally, this score-mismatch view connects directly to the Wasserstein
gradient-flow perspective in \Cref{sec:wgf-distribution-level-training}.
At the distribution level, reverse-KL gradient flow moves particles according
to the target score minus the model score. The VSD parameter gradient contains
the same score mismatch, pulled back through the generator Jacobian
$\partial_{\btheta}\hat\rvx_t$. Thus, the Wasserstein view describes the ideal
distribution-space motion, while VSD realizes this motion through updates of a
finite-dimensional shared generator.

\clearpage
\newpage

\section{\texorpdfstring{Progressive Distillation}{Progressive Distillation}}\label{sec:PD}

Progressive Distillation (PD)~\citep{salimans2021progressive} consists of two procedures that together enable a diffusion model to learn the PF-ODE trajectory more efficiently. 
The key idea is to progressively reduce the number of integration steps required for high-quality sampling while retaining fidelity to the teacher trajectory.
\begin{itemize}
    \item \textbfs{Distillation Operation:} Distills a deterministic sampler (e.g., DDIM) based on a pre-trained teacher model (initially a diffusion model) into a student model that reproduces the same trajectory using only half as many sampling steps.
    \item \textbfs{Progressive Operation:} Repeats this distillation process iteratively, each time halving the number of steps, until the student can generate high-quality samples within a small fixed budget (typically $1$–$4$ steps).
\end{itemize}

\begin{figure}[th!]
    \centering
    \includegraphics[width=0.93\linewidth]{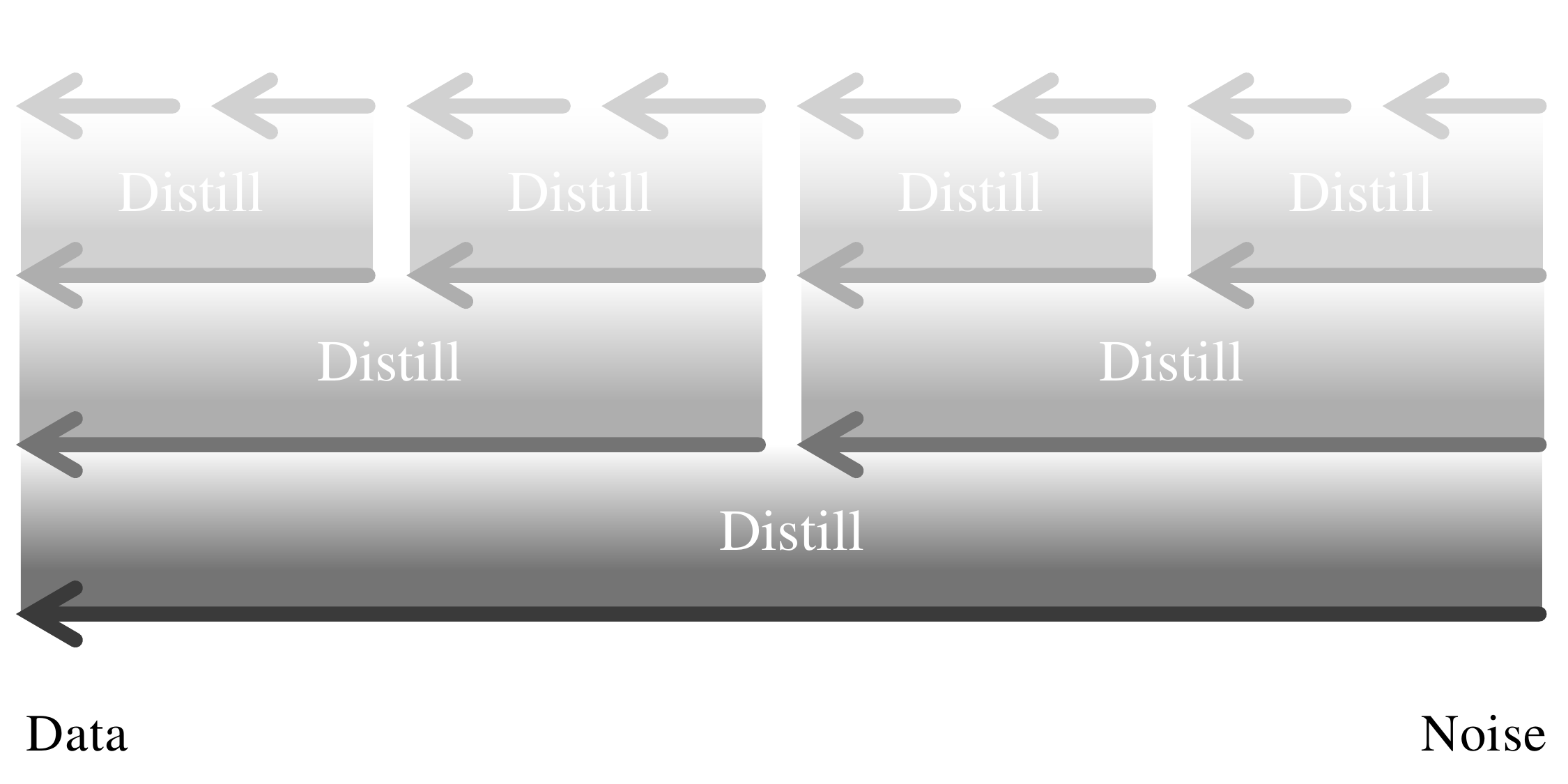} 
    \caption{\textbfs{Illustration of Progressive Distillation (PD).} 
At each round, the student model is trained so that a single step reproduces the effect of two adjacent teacher steps. 
This process distills $N$ teacher steps into $N/2$ student steps, and repeating the procedure progressively halves the trajectory length until the desired step count is reached. 
The arrows indicate how multi-step teacher transitions are compressed into fewer student steps, moving from noise to data.
\figcredit{Created by the authors.}
}
    \label{fig:pd-illustration}
\end{figure}

We first introduce the distillation operation of PD in \Cref{subsec:pd-distillation}, and then summarize the entire training pipeline in \Cref{subsec:pd-training}. \Cref{subsec:pd-guidance} presents an extension for CFG guidance.

\subsection{Distillation Operation in PD}\label{subsec:pd-distillation}

In this section, we fix DDIM in the $\rvx$-prediction parameterization as the time–stepping rule and still write $\mathtt{Solver}_{s \to t}$ for the deterministic map obtained by plugging the current teacher’s $\rvx$-denoiser into DDIM.

In the first PD round (teacher = pre–trained diffusion model), this coincides with integrating the diffusion PF–ODE via DDIM; in later rounds (teacher = previous student), $\mathtt{Solver}_{s \to t}$ is simply the DDIM transition induced by the current teacher, not the original diffusion PF–ODE.

The distillation step is as follows: starting from a noisy input $\rvx_s$ (a perturbed version of clean data, $\rvx_s = \alpha_s \rvx_0 + \sigma_s \bm{\epsilon}$), the student is trained to predict a target ${\color{orange}\tilde\rvx}$ so that a single student step $s\!\to\! t$ reproduces the teacher’s two consecutive steps $s\!\to\! u\!\to\! t$.
Let $\rvx_{\bm{\phi}^\times}(\rvx,\tau)$ denote the teacher’s $\rvx$-prediction denoiser in this round.
Applying the teacher–induced DDIM transition twice gives
\[
\tilde\rvx_u := \mathtt{Solver}_{s\to u}\!\left(\rvx_s;\,\rvx_{\bm{\phi}^\times}\right),
\qquad
\tilde\rvx_t := \mathtt{Solver}_{u\to t}\!\left(\tilde\rvx_u;\,\rvx_{\bm{\phi}^\times}\right).
\]
Here, we use the notation of \Cref{eq:solver} to denote the deterministic transition map from $s$ to $t$ (starting at $\rvx_s$) induced by plugging $\rvx_{\bm{\phi}^\times}$ into DDIM.

\begin{question}
What is the pseudo-clean ${\color{orange}\tilde\rvx}$ at time $s$ such that the solver produces the same output $\tilde\rvx_t$ when stepping directly $s \to t$ as it does via $s \to u \to t$? Specifically, determine ${\color{orange}\tilde\rvx}$ satisfying:
\[
\tilde\rvx_t = \mathtt{Solver}_{s\to t}\left(\rvx_s; {\color{orange}\tilde\rvx}\right).
\]
\end{question}

Once a closed-form expression for ${\color{orange}\tilde\rvx}$ is obtained, we train a student model $\rvf_{\bm{\theta}}(\rvx_s, s)$ (also an $\rvx$-prediction model here) to approximate the ``two-steps-in-one'' target ${\color{orange}\tilde\rvx}$ by minimizing
\begin{align}\label{eq:pd-optimization}
\min_{\bm{\theta}} \E_s
\mathbb{E}_{\rvx_s\sim p_s}\!\left[w(\lambda_s)\,\big\|\rvf_{\bm{\theta}}(\rvx_s, s) - {\color{orange}\tilde\rvx}\big\|_2^2\right].
\end{align}

In the following, we show that the DDIM rule yields ${\color{orange}\tilde\rvx}$ in closed form through elementary algebra (note that the result holds for both discrete and continuous time):
\lemp{Two-Steps-in-One Target ${\color{orange}\tilde\rvx}$ of DDIM}{pd-ddim}{Starting from an initial condition $\rvx_s$, if the solver is taken as DDIM, then the ``two-step-in-one'' target ${\color{orange}\tilde\rvx}$ can be computed as
\begin{align*}
    {\color{orange}\tilde\rvx} 
    &=\frac{\sigma_s}{\alpha_t\sigma_s - \alpha_s\sigma_t} \tilde\rvx_t - \frac{\sigma_t}{\alpha_t\sigma_s - \alpha_s\sigma_t} \rvx_s.
\end{align*}
Here, $\tilde\rvx_t$ is obtained by applying DDIM (in \Cref{eq:ddim-update-all}) twice, from $s \to u \to t$:
\begin{align*}
    s \to u: & \quad    && \tilde\rvx_u = \frac{\sigma_u}{\sigma_s}\rvx_s + \alpha_s \left(\frac{\alpha_u}{\alpha_s} - \frac{\sigma_u}{\sigma_s} \right)\rvx_{\bm{\phi}^\times}(\rvx_s, s)  \\
    u \to t: & \quad    && \tilde\rvx_t = \frac{\sigma_t}{\sigma_u}\tilde\rvx_u + \alpha_u \left(\frac{\alpha_t}{\alpha_u} - \frac{\sigma_t}{\sigma_u} \right)\rvx_{\bm{\phi}^\times}(\tilde\rvx_u, u).
\end{align*}
}{$\tilde\rvx_t$ must be matched with the one-step DDIM from $s$ to $t$, $\tilde\rvx_t'$, expressed as:
\[
s \to t: \quad \tilde\rvx_t' = \frac{\sigma_t}{\sigma_s}\rvx_s + \alpha_s \left(\frac{\alpha_t}{\alpha_s} - \frac{\sigma_t}{\sigma_s} \right) {\color{orange}\tilde\rvx}.
\]
By equating $\tilde\rvx_t'$ and $\tilde\rvx_t$, we can solve for ${\color{orange}\tilde\rvx}$ in terms of $\tilde\rvx_t$, $s$, and $t$:
\begin{align}
    &\tilde\rvx_t = \tilde\rvx_t' \nonumber \\
    \Longleftrightarrow\,\, & \tilde\rvx_t = \frac{\sigma_t}{\sigma_s}\rvx_s + \alpha_s \left(\frac{\alpha_t}{\alpha_s} - \frac{\sigma_t}{\sigma_s} \right) {\color{orange}\tilde\rvx} \nonumber \\
    \Longleftrightarrow\,\, & {\color{orange}\tilde\rvx} =  \frac{\tilde\rvx_t - \frac{\sigma_t}{\sigma_s}\rvx_s}{\alpha_s \left(\frac{\alpha_t}{\alpha_s} - \frac{\sigma_t}{\sigma_s} \right)} \label{eq:x-pd-paper} \\
    \Longleftrightarrow\,\, & {\color{orange}\tilde\rvx} = \frac{\sigma_s}{\alpha_t\sigma_s - \alpha_s\sigma_t} \tilde\rvx_t - \frac{\sigma_t}{\alpha_t\sigma_s - \alpha_s\sigma_t} \rvx_s. \nonumber
\end{align}
}
With this formula, PD computes the pseudo-clean target at time $s$ whose single DDIM step $s\!\to\!t$ lands exactly at the two-step output $\tilde\rvx_t$.

\paragraph{Practical Discrete Time Grids and Loss.}
In practice, we fix a decreasing grid $t_0=T>t_1>\cdots>t_N=0$ and, for brevity, write $s:=t_k$, $u:=t_{k+1}$, $t:=t_{k+2}$.
The teacher provides one step maps $\mathtt{Teacher}_{t_k\to t_{k+1}}$, and the student learns a two step skip map that matches the teacher composition:
\[
\mathtt{Student}_{t_k\to t_{k+2}}
 \approx 
\mathtt{Teacher}_{t_{k+1}\to t_{k+2}}
\circ
\mathtt{Teacher}_{t_{k}\to t_{k+1}}.
\]

For the standard halving schedule, assume $N$ is even and sample
triplets $(s,u,t)=(t_k,t_{k+1},t_{k+2})$ with
$k\in\{0,2,\ldots,N-2\}$.
The objective \Cref{eq:pd-optimization} becomes
\[
\min_{\bm{\theta}} 
\mathbb{E}_{\,k \sim \mathcal{U}\{0,2,\ldots,N{-}2\}}
\mathbb{E}_{\,\rvx_{t_k}\sim p_{t_k}}\!
\Big[
  w(\lambda_{t_k})\,
  \big\|
  \rvf_{\bm{\theta}}(\rvx_{t_k},t_k) - {\color{orange}\tilde\rvx}^{(k)}
  \big\|_2^2
\Big],
\]
where the teacher two-step target ${\color{orange}\tilde\rvx}^{(k)}$ is computed via Lemma~\ref{pd-ddim}.  If the grid is uniform, one may write $t_k = T(1 - k/N)$ so that
\[
s = T\Big(1 - \frac{k}{N}\Big),\quad
u = T\Big(1 - \frac{k+1}{N}\Big),\quad
t = T\Big(1 - \frac{k+2}{N}\Big),
\]
corresponding to evenly spaced time steps of size $\Delta s =  T/N$.

\subsection{Entire Training Pipeline of PD and Its Sampling}\label{subsec:pd-training}
After training on locally paired fragments via \Cref{eq:pd-optimization}, the $\mathtt{Student}$ no longer follows every interval of the original grid. Instead, each learned step covers two consecutive $\mathtt{Teacher}$ steps, so the $\mathtt{Student}$ advances on every other time point,
\[
t_0 \to t_2 \to t_4 \to \cdots \to t_N,
\]
and thus traverses the same horizon $[0,T]$ using only $N/2$ transitions. After this stage, the trained $\mathtt{Student}$ replaces the $\mathtt{Teacher}$ as the new denoiser model. The procedure is then repeated on the coarser grid (the time step doubles), yielding the progression
\[
N \;\to\; N/2 \;\to\; N/4 \;\to\; \cdots,
\]
until the desired number of inference steps is reached. At each iteration, the new $\mathtt{Student}$ is initialized from the updated $\mathtt{Teacher}$. This iterative halving preserves the global time horizon while progressively compressing the temporal resolution of the generative process.

\paragraph{Sampling.}
At inference time, using the (DDIM) solver with the current $\mathtt{Student}$ as the denoiser, the sampler advances on the coarser grid induced by training. After the first round it takes ``skip-2'' jumps ($t_0 \to t_2 \to \cdots \to t_N$), after the next round ``skip-4'' ($t_0 \to t_4 \to \cdots \to t_N$), and so on, halving the number of sampling steps at each iteration while keeping the same start and end times.

\subsection{Additional Discussion: Local Semigroup Matching and the Possibility of Generalized Solvers}

\paragraph{Progressive Distillation as Local Semigroup Matching.}
Within the unified objective \Cref{eq:general-cm-loss}, the intractable oracle
target $\bPsi_{s\to 0}$ is replaced by a teacher–induced surrogate that uses the
\emph{semigroup property} of the ODE flow (see more details later in \Cref{eq:semi-group}): evolving from $s$ to $t$ should be
equivalent to going from $s$ to any intermediate $u$ and then from $u$ to $t$,
\[
\bPsi_{s\to t} = \bPsi_{u\to t}\circ\bPsi_{s\to u}.
\]
PD enforces this locally by training the student’s one–step map to match the
teacher’s composition of two adjacent one–step fragments:
\begin{align}\label{eq:pd-local}
  \E_{s}\E_{\rvx_s\sim p_s}
  \big\|
    \underbrace{\rmG_{\btheta}(\rvx_s, s, s-2\Delta s)}_{\text{student one-step}}
    -
    \underbrace{\mathtt{Solver}_{s-\Delta s\to s-2\Delta s}\big(\mathtt{Solver}_{s\to s-\Delta s}(\rvx_s)\big)}_{\text{teacher two-step composition}}
  \big\|_2^2.
\end{align}
When the teacher solver accurately approximates the underlying ODE flow,
the composition in \Cref{eq:pd-local} approximates the corresponding semigroup
relation on a short decreasing grid (take $s>u>t$ with $u=s-\Delta s$ and $t=s-2\Delta s$):
\begin{align*}
  \bPsi_{s\to s-2\Delta s}
  &= \bPsi_{s-\Delta s\to s-2\Delta s}\circ\bPsi_{s\to s-\Delta s}
  \\&\approx
  \mathtt{Solver}_{s-\Delta s\to s-2\Delta s}\circ \mathtt{Solver}_{s\to s-\Delta s},
\end{align*}
so training only requires short teacher fragments, rather than a full rollout
from time $s$ all the way to $0$.

To connect back to the few–step denoiser view in \Cref{eq:general-cm-loss}, define
the student’s few–step map as a composition of learned jumps:
\[
\underbrace{\rmG_{\btheta}(\rvx_s, s, 0)}_{\text{few-step denoiser}}
=
\big(\rmG_{\btheta}(\,\cdot\,, 2\Delta s, 0)\circ\cdots\circ \rmG_{\btheta}(\,\cdot\,, s, s-2\Delta s)\big)(\rvx_s).
\]
Conceptually, \Cref{eq:pd-local} provides an efficient \emph{local surrogate} for the
global regression
\[
\E_{s,\rvx_s}\Big\|\rmG_{\btheta}(\rvx_s,s,0)\;-\;(\mathtt{Solver})^{\circ}_{s\to 0}(\rvx_s)\Big\|_2^2,
\]
where $(\mathtt{Solver})^{\circ}_{s\to 0}$ denotes the teacher’s full
composition from $s$ to $0$ on a grid with step size $\Delta s$, serving as a
proxy for $\bPsi_{s\to 0}$.

\paragraph{Can we Use Other Solvers?}
Above, we used DDIM in the $\rvx$-prediction parameterization as a concrete
PF--ODE sampler. The basic distillation principle is more general: several
fine teacher steps produce a target endpoint, and the student is trained so
that one coarser step reaches the same endpoint.

The closed-form pseudo-target derived above is available whenever the coarse
solver update can be inverted with respect to the student's network prediction.
For DDIM, the update is affine in a single $\rvx$-prediction, so this inversion
is analytic. The same idea applies to other solvers whose one-step update admits
a comparable inversion.

For more complicated solvers, such a single-output inversion may not be
available. Multi-stage methods use several network evaluations, multi-step
methods depend on previous solver states, and implicit methods require solving
coupled equations. In such cases, one can instead compute the endpoint of the
coarse student step and directly match it to the endpoint produced by the fine
teacher steps, as in \Cref{eq:pd-local}, backpropagating through the student
solver when feasible. Since a coarse step may itself require multiple neural
function evaluations, fewer time steps do not necessarily imply the same
reduction in NFE.

For a stochastic sampler, the same pathwise logic applies after fixing or
coupling the random draws used by the teacher and student. Without such
coupling, individual trajectories need not share the same endpoint, and the
appropriate target is generally distributional rather than pathwise matching.

\subsection{PD with Guidance}\label{subsec:pd-guidance}
\citet{meng2023distillation} proposed a two-stage pipeline for distilling
classifier-free guided (CFG) diffusion models: (1) \emph{distill the guidance}
into a single network that takes the guidance weight as input, and (2) apply
\emph{progressive distillation (PD)} to reduce the sampling steps. They
demonstrated this both in pixel space and in latent space (e.g., Stable Diffusion).

\paragraph{Stage-One Distillation: Distilling Guidance.}
Let $\rvx_{\bm{\phi}^\times}(\rvx_s, s, \rvc)$ denote the (pre-trained) conditional
diffusion model output in the ``$\rvx$-prediction'' parameterization (i.e., a
clean estimate) at time $s$ and condition $\rvc$; the condition can also
be null, $\rvc=\emptyset$ (unconditional branch). The $\omega$-weighted CFG
combination in \Cref{eq:cfg-model-nn} can be written as
\begin{equation}\label{eq:cfg-model-clean}
   \rvx_{\bm{\phi}^\times}^{\,\omega}(\rvx_s, s, \rvc)
   := (1+\omega)\,\rvx_{\bm{\phi}^\times}(\rvx_s, s, \rvc)
      - \omega\,\rvx_{\bm{\phi}^\times}(\rvx_s, s, \emptyset),
\end{equation}
where $\omega \sim p_\omega(\omega)$ for some CFG weighting distribution $p_\omega$, typically $p_\omega(\omega)=\mathcal{U}[\omega_{\min},\omega_{\max}]$.

Stage-one introduces a new model $\rvx_{\bm{\theta}_1}(\rvx_s, s, \rvc, \omega)$
that directly takes $\omega$ as input and learns to reproduce the CFG
output $\rvx_{\bm{\phi}^\times}^{\,\omega}(\rvx_s, s, \rvc)$ by supervised
regression:
\[
\min_{\bm{\theta}_1}\;
\mathbb{E}_{\omega\sim p_\omega, s, \rvx\sim p_{\mathrm{data}}, \rvx_s \sim p(\rvx_s\mid \rvx)}
\lambda(s)
\big\|\rvx_{\bm{\theta}_1}(\rvx_s, s, \rvc, \omega)-\rvx_{\bm{\phi}^\times}^{\omega}(\rvx_s, s, \rvc)\big\|_2^2.
\]
Here $\lambda(s)$ is a standard schedule-dependent weighting; sampling $\omega$
each iteration teaches a single network to emulate CFG at arbitrary
guidance strengths.

\paragraph{Stage-Two Distillation: PD.}
The stage-one model $\rvx_{\bm{\theta}_1}(\rvx_s, s, \rvc, \omega)$ serves as the
teacher in PD and is progressively distilled into a student
$\rvx_{\bm{\theta}_2}(\rvx_s, s, \rvc, \omega)$ with fewer sampling steps,
following \Cref{subsec:pd-training}. At each iteration, the number of steps is
halved (e.g., $N \to N/2 \to N/4 \to \cdots$).

\newpage
\section{Closing Remarks}\label{sec:ch10_cr}

This chapter has introduced our first major paradigm for training-based acceleration. Having exhausted training-free improvements via numerical solvers, we shifted our focus to a new strategy: training a fast \textit{student} generator that learns to replicate the behavior of a slow, pre-trained \textit{teacher} diffusion model.

We explored two primary distillation philosophies. First, in distribution-based distillation, represented by methods like Variational Score Distillation (VSD), the student's output distribution is trained to match the teacher's. This is achieved by aligning their respective score functions across different noise levels, providing a stable, non-adversarial objective. Second, in flow map distillation, we saw how methods like Progressive Distillation (PD) train the student to directly approximate the teacher's solution trajectory. PD's iterative approach, where each round halves the number of sampling steps, proved to be a powerful and practical method for compressing a long iterative process into just a few steps.

These distillation techniques successfully bridge the gap between the high sample quality of iterative diffusion models and the inference speed of one-step generators, offering a compelling pathway to efficient, high-fidelity synthesis.

However, the reliance on a pre-trained teacher model introduces a two-stage pipeline: first train a slow but powerful teacher, then distill it into a fast student. This raises a fundamental question at the forefront of generative modeling research: Can we bypass the teacher entirely?

Is it possible to design a standalone training principle that learns these fast, few-step generators directly from data? The final chapter of this monograph will address this question.
\begin{enumerate}
    \item We will explore pioneering methods such as Consistency Models that learn the mapping from any point on an ODE trajectory to its destination point.
    \item We will delve into generalized concepts of Consistency Models which learn to map any point on an ODE trajectory to another point in a single step.
\end{enumerate}

This shift from improving the solver or distilling a solution to learning the solution map itself represents a significant step toward a new class of generative models that are both principled and highly efficient by design. 

\clearpage
\newpage

\chapter{Learning Fast Generators from Scratch}\label{ch:fast-scratch}

\epigraph{
    \textit{Truth is ever to be found in simplicity, and not in the multiplicity and confusion of things.
}}{Isaac Newton}

In \Cref{ch:distillation}, we saw that slow iterative samplers in diffusion models can be compressed into few-step generators through distillation. From an engineering perspective, two-stage pipelines are practical because they divide a complex generative training task into clear, independent objectives. The first stage learns the data distribution, while the second accelerates sampling or enhances quality. This separation allows each stage to be optimized independently, making the overall system easier to manage, more stable, and more reliable.

In this chapter, however, the focus shifts to a central question driving the progress of deep generative modeling:
\begin{question}
    Can we design a standalone generative principle that trains in a stable and efficient way, fast sampling, and allows users to easily guide or control what is produced?
\end{question}
In this chapter we pursue this direction and discuss an alternative approach:
training few-step diffusion-based generators \emph{without} relying on a pre-trained model.
Our focus is the \emph{flow-map model}. Here, a \emph{flow map} is the
classical solution map of an ODE, a notion rooted in differential
equations and dynamical systems and closely connected, in continuum and
fluid mechanics, to following the motion of material particles.
Specifically, $\bPsi_{s\to t}$ takes a state at time $s$ to the state
reached at time $t$ under the prescribed dynamics. A flow-map model
learns to approximate this map directly, allowing a long trajectory to
be traversed in one or a few network evaluations%
\footnote{See \Cref{eq:flow-map-def} and the surrounding discussion for
the mathematical and historical background of flow maps.}.
This formulation provides a principled way to transport probability mass from the prior distribution $p_{\mathrm{prior}}$ to the data distribution $p_{\mathrm{data}}$,
while preserving the marginal distributions $p_t$ specified by the forward diffusion process at each intermediate time.

\chapterlearning{

  \item distinguish an instantaneous velocity field from its ODE solution map (flow map);

  \item understand the semigroup property of exact ODE flows and how it motivates consistency-based learning objectives;

  \item understand Consistency Model, Consistency Trajectory Model, and Mean Flow as representative ways to learn special or general flow maps;

  \item distinguish oracle flow-map identities from their practical approximations, and understand how suitable parametrizations enable few-step generation while retaining access to local velocity information.

}{The central object in this chapter is the ODE flow map. The individual methods should be viewed as different parametrizations and training strategies for learning this same underlying object.}

\newpage
\section{Prologue}

\begin{figure}[th]
  \centering
  \resizebox{0.95\linewidth}{!}{%
    \begin{tikzpicture}[x=2.6cm,y=1cm,line cap=butt]
      \tikzset{>={Stealth[length=12pt,width=16pt]}}
      \def\DotR{3pt}
      \def\LabelAngle{45}
      \def\DateAbove{4pt}
      \def\TextBelow{16pt}
      \newcommand{\nl}{\\}

      \draw[ultra thick,-Stealth] (0,0) -- (6.5,0);

      \foreach \x/\date/\desc in {
        0.5/{2021/01}/{{\color{PersA}{KD}}\nl{\color{PersA}{Section~\ref{sec:distillation-prologue}}}},
        1.6/{2022/02}/{{\color{PersA}{PD}}\nl{\color{PersA}{Section~\ref{sec:PD}}}},
        2.7/{2023/03}/{{\color{PersA}{CM}}\nl{\color{PersA}{Section~\ref{sec:discrete-cm}}}},
        3.8/{2023/10}/{{\color{PersB}{CTM}}\nl{\color{PersB}{Section~\ref{sec:CTM}}}},
        4.9/{2024/10}/{{\color{PersA}{sCM}}\nl{\color{PersA}{Section~\ref{sec:conti-cm}}}},
        6.0/{2025/05}/{{\color{PersB}{MF}}\nl{\color{PersB}{Section~\ref{sec:MF}}}}
      }{
        \fill[black] (\x,0) circle[radius=\DotR];
        \node[above=\DateAbove,align=center] at (\x,0) { \date};
        \node[below=\TextBelow,anchor=north,rotate=\LabelAngle,inner sep=1pt] at (\x,0)
          {\shortstack[c]{\strut \desc}};
      }
    \end{tikzpicture}%
  }
\caption{\textbfs{Timeline of diffusion-based methods for learning flow
maps.}\;
The underlying flow map is the classical ODE solution map; the dates
shown here refer only to modern machine-learning methods that distill,
approximate, or learn this map. We use {\color{PersA}blue} for the
special case {\color{PersA}$\bPsi_{s\to 0}$} and
{\color{PersB}orange} for the general map
{\color{PersB}$\bPsi_{s\to t}$}.
\figcredit{Created by the authors.}}
  \label{fig:timeline}
\end{figure}

\paragraph{Motivation of Flow Map Models.}
In \Cref{ch:distillation} we showed how the inaccessible regression target in the oracle flow-map loss
$\mathcal{L}_{\mathrm{oracle}}(\btheta)$ (see \Cref{eq:general-cm-loss:rep})
can be estimated by distilling knowledge from a pre-trained diffusion model to obtain few-step generators.
This route is effective and practical: a two-stage pipeline can be engineered for robustness
and often remains competitive in both data and compute efficiency.

In this chapter, we shift focus to a broader challenge at the core of deep generative modeling:
\emph{Can we establish a standalone generative principle that enables stable, scalable, and efficient training,
fast sampling, and generation that can be easily steered by user intentions, without relying on a pre-trained model?}
Designing such standalone principles lies at the center of generative modeling.

Diffusion models offer a useful design principle: start with a continuous-time forward process that gradually transforms data into a simple prior (noise) as a reference, and frame the modeling task as learning the reverse-time transport that restores this process to match the desired marginal distributions.  
This time-dependent formulation also makes it easier to steer the generation process at intermediate steps, compared to one-shot generative maps.  
Specializing to diffusion-motivated methods, this leads to the question:
\begin{question}
Can we learn the flow map $\bPsi_{s\to t}(\cdot)$ with a network
$\rmG_\btheta(\cdot,s,t)$ (a \emph{flow map model}) without access to pre-trained models,
while maintaining high-fidelity generation?
\end{question}
\begin{figure}[th!]
    \centering
    \includegraphics[width=0.9\linewidth]{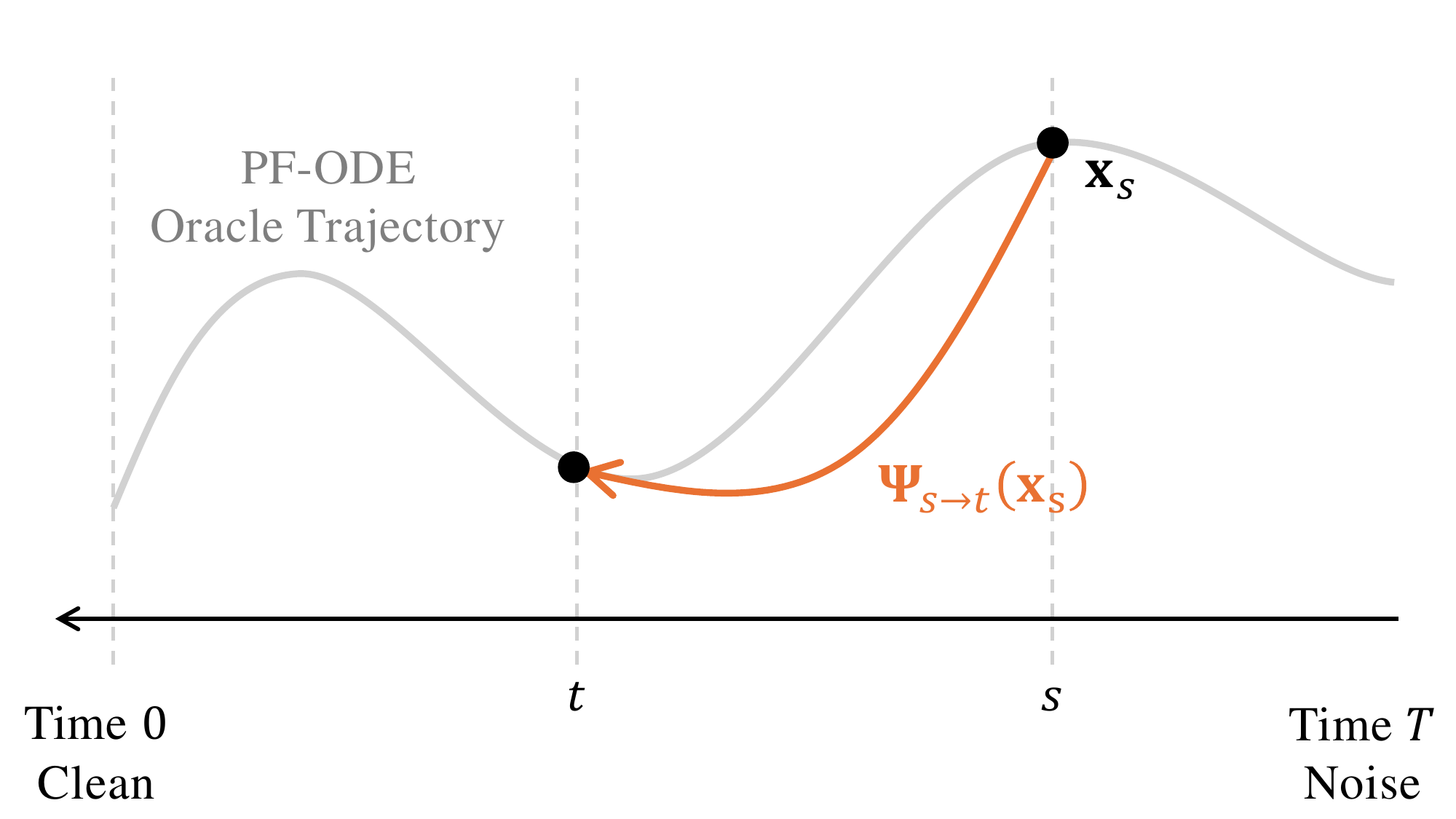}
    \caption{\textbfs{Illustration of the flow map.} Starting from any state $\rvx_s$ at time $s$, 
the flow map $\bPsi_{s \to t}$ transports it to the corresponding point on the same PF-ODE oracle trajectory at time $t$. \figcredit{Created by the authors.}}
    \label{fig:flow-map-illustration}
\end{figure}

This chapter develops methods toward this goal, organized around a single objective
that also underlies distillation and provides a unified view of flow-map formulations~\citep{boffi2024flow,hu2025cmt}:
\begin{mdframed}
\begin{equation}\tag{\ref*{eq:general-cm-loss}}\label{eq:general-cm-loss:rep}
  \mathcal{L}_{\mathrm{oracle}}(\btheta)
  := \mathbb{E}_{s,t}\,\mathbb{E}_{\rvx_s \sim p_s}\Big[
     w(s,t)  d\big(\rmG_\btheta(\rvx_s,s,t),  \bPsi_{s\to t}(\rvx_s)\big)
  \Big].
\end{equation}
\end{mdframed}

\addtocounter{equation}{-1}
Here $s,t$ are sampled from some time distribution (e.g., uniform),
$w(s,t) \ge 0$ assigns weights to the time pairs $(s,t)$, and $d(\cdot,\cdot)$
is a discrepancy measure such as the squared $\ell_2$ norm. The oracle flow map
$\bPsi_{s\to t}$ represents the ideal transformation that takes a state
$\rvx_s$ at time $s$ and transports it directly to time $t$:
\[
\bPsi_{s\to t}(\rvx_s) = \rvx_s + \int_s^t \rvv^*(\rvx_u,u)\,\mathrm{d}u,
\]
where the oracle drift is given as
\[
\rvv^*(\rvx_u,u) = \E \left[\alpha_u'\rvx_0+ \sigma_u'\beps | \rvx_u\right],
\]
while equivalent parametrizations are also possible  (see \Cref{ch:all-equivalent}), with common choices including the $\rvx$-prediction and $\rvv$-prediction forms. 

At the optimum of the oracle loss, the learned model recovers the true flow map
exactly:
\[
\rmG^*(\rvx_s, s, t) = \bPsi_{s\to t}(\rvx_s), \quad \text{for all }\, s,t, \text{ and }\, \rvx_s \sim p_s.
\]

Because the flow map $\bPsi_{s\to t}$ cannot be expressed in closed form, it must be approximated. 
One option, discussed in \Cref{ch:distillation}, is to rely on a pre-trained diffusion model. 
Alternatively, as we will see in this chapter, new and more tractable surrogates can be introduced. 
For clarity, existing approaches can be broadly categorized according to whether the training procedure queries a teacher during the loop: 
\emph{distillation}, which explicitly calls a teacher model, and 
\emph{training from scratch}, which avoids teacher calls by constructing self-contained surrogates.

Building on this principled
objective, we now turn to systematic approaches for learning flow map models,
with the aim of developing methods that are practical while also producing
generations that more accurately reflect the true data distribution and are
computationally efficient. We begin with a high-level introduction to this paradigm.
\paragraph{Special Flow Map: Consistency Functions.}
\emph{Consistency Models}~\citep{song2023consistency} represent one of the earliest pioneering approaches to flow-map learning. 
They learn a few-step denoiser $\rvf_\btheta(\cdot,s)$ that approximates the special case of the flow map to the origin:
\[
\bPsi_{s\to 0}(\cdot), \qquad s \in (0,T].
\]
The key idea is that every state $\rvx_s$ is mapped to the time-$0$ endpoint
$\bPsi_{s\to 0}(\rvx_s)$ of the PF-ODE trajectory passing through it. Formally, the oracle training objective for the CM family~\citep{song2023consistency,songimproved,geng2024consistency,lu2024simplifying} is
\begin{align}\label{eq:oracle-cm}
    \mathcal{L}_{\methodl{oracle}{CM}}(\btheta) 
    := \E_{s} \E_{\rvx_s\sim p_s} \left[ w(s)  d \left(\rvf_\btheta(\rvx_s,s), \bPsi_{s\to 0}(\rvx_s)\right)\right].
\end{align}

In practice, however, the oracle $\bPsi_{s\to 0}(\rvx_s)$ is unavailable. It is therefore replaced by a \emph{stop-gradient} target, denoted as $\rvf_{\btheta^-}$, taken from a slightly earlier step $\bPsi_{s\to s-\Delta s}(\rvx_s)$ on the same trajectory:
\[
\bPsi_{s\to 0}(\rvx_s) \approx \rvf_{\btheta^-} \left(\bPsi_{s\to s-\Delta s}(\rvx_s),\,s-\Delta s\right), \quad \Delta s > 0,
\]
where $\bPsi_{s\to s-\Delta s}(\rvx_s)$ itself must also be approximated. Two
practical strategies are available: (i) \emph{distillation}, which relies on a
pre-trained diffusion model, and (ii) \emph{training from scratch}, which uses a
one-point estimate without teacher guidance.


\paragraph{General Flow Map.} Two representative approaches are the \emph{Consistency Trajectory Model} (CTM) and \emph{Mean Flow} (MF).
\subparagraph{Consistency Trajectory Models.} \emph{Consistency Trajectory Model} (CTM)\\~\citep{kim2023consistency} is the first work to learn the general flow map $\bPsi_{s\to t}$ for arbitrary start and end times, and can be viewed as a concrete instance under the unified objective of \Cref{eq:general-cm-loss:rep}. CTM adopts an Euler-inspired parametrization by expressing the oracle flow map as
\begin{align*}
\bPsi_{s\to t}(\rvx_s)
:= \rvx_s + \int_s^t \rvv^*(\rvx_u,u)\diff u 
= \frac{t}{s} \rvx_s
+ \frac{s-t}{s}
\underbrace{\Big[\rvx_s + \frac{s}{s-t} \int_s^t \rvv^*(\rvx_u,u) \diff u\Big]}_{\approx~\rvg_{\btheta}},
\end{align*}
which motivates the neural parameterization
\[
\rmG_{\btheta}(\rvx_s,s,t)
:= \frac{t}{s}\,\rvx_s + \frac{s-t}{s}\,\rvg_\btheta(\rvx_s,s,t),
\]
where $\rvg_\btheta$ is a neural network trained so that $\bPsi_{s\to t}(\rvx_s)\approx \rmG_{\btheta}(\rvx_s,s,t)$.

Since the oracle $\bPsi_{s\to t}(\rvx_s)$ is inaccessible, CTM trains against a \emph{stop-gradient} target evaluated at an intermediate time $u$:
\[
\bPsi_{s\to t}(\rvx_s)
 \approx 
\rmG_{\btheta^-} \big(\bPsi_{s\to u}(\rvx_s),\,u,\,t\big),
\qquad u\in[t,s],
\]
where the intermediate state $\bPsi_{s\to u}(\rvx_s)$ is approximated in one of
two ways: (i) \emph{distillation}, which uses a few-step solver applied to a
pre-trained diffusion teacher, or (ii) \emph{training from scratch}, which
constructs a self-induced teacher directly through the $\rmG_\btheta$
parametrization.

\subparagraph{Mean Flow.}
Mean Flow (MF)~\citep{geng2025mean} builds on flow matching by modeling the \emph{average drift} over an interval $[t,s]$ (with $t\le s$):
\[
\rvh_{\btheta}(\rvx_s,s,t)
 \approx 
\rvh^*(\rvx_s,s,t)
:= \frac{1}{\,t-s\,}\int_s^t \rvv^*(\rvx_u,u)\,\diff u,
\]
also aligning with \Cref{eq:general-cm-loss:rep}.
Differentiating the identity
\[
(t-s)\,\rvh^*(\rvx_s,s,t)  =  \int_s^t \rvv^*(\rvx_u,u)\,\diff u
\]
with respect to $s$ yields a self-referential relation that motivates the MF objective
\[
\mathcal{L}_{\mathrm{MF}}(\btheta)
:= \E_s\,\E_{\rvx_s\sim p_s} \Big[w(s)\,\big\|
\rvh_{\btheta}(\rvx_s,s,t)-\rvh_{\btheta^-}^{\mathrm{tgt}}(\rvx_s,s,t)
\big\|_2^2\Big],
\]
with stop-gradient target
\[
\rvh_{\btheta^-}^{\mathrm{tgt}}(\rvx_s,s,t)
:= \rvv^*(\rvx_s,s)  -  (s-t) \left((\partial_\rvx \rvh_{\btheta^-})\,\rvv^*(\rvx_s,s) + \partial_s \rvh_{\btheta^-}\right).
\]
In practice, the oracle velocity $\rvv^*(\rvx_s,s)$ must also be approximated.
Two common strategies are: (i) \emph{distillation}, which leverages a
pre-trained diffusion model trained with flow matching, or (ii) \emph{training
from scratch}, which uses the one-point conditional velocity
$\alpha'_s\,\rvx_0+\sigma'_s\,\beps$ derived from the forward corruption process
$\rvx_s = \alpha_s \rvx_0+\sigma_s \beps$.

\subparagraph{Relationship Between CTM and MF.}
CTM and MF approximate the same path integral but parameterize different surrogates of it:
\begin{align*}
\bPsi_{s\to t}(\rvx_s)
&:= \rvx_s  +  \int_s^t \rvv^*(\rvx_u,u)\,\diff u
\\[2pt]
&= \frac{t}{s}\,\rvx_s
 +  \frac{s-t}{s}\,
\underbrace{\Big[\rvx_s + \frac{s}{\,s-t\,} \int_s^t \rvv^*(\rvx_u,u)\,\diff u\Big]}_{\approx~\rvg_{\btheta} }
\\[4pt]
&= \rvx_s  +  (t-s)\,
\underbrace{\Big[\frac{1}{\,t-s\,} \int_s^t \rvv^*(\rvx_u,u)\,\diff u\Big]}_{\approx~\rvh_{\btheta} }.
\end{align*}
In words, CTM learns an \emph{slope displacement} through $\rvg_\btheta$, while MF learns the \emph{average drift} $\rvh_\btheta$; both are consistent ways to approximate the same integral that defines $\bPsi_{s\to t}$.

\paragraph{What Happens Next?}
We begin with the CM family, which focuses on the specific flow map
$\bPsi_{s\to 0}$. This part covers both its discrete time origin in
\Cref{sec:discrete-cm} and its continuous time extension in
\Cref{sec:conti-cm}. We then move on to the general flow map and provide a
detailed discussion of two key representatives, CTM and MF. Their
parameterizations, training strategies, and practical approximations are
presented in \Cref{sec:CTM} and \Cref{sec:MF}, respectively.

We remark that the \emph{Elucidating Diffusion Model} (EDM) introduced in
\Cref{sec:edm} offers systematic guidelines for designing the network
parameterization of the $\rvx$-prediction model and has demonstrated strong
empirical performance. Although this section can be considered optional, the
EDM formulation serves as a valuable foundation for CM-style models.

For clarity of exposition later on, we do not strictly follow the chronological order in which these approaches appeared. 
Instead, we organize the discussion by conceptual relationships. 
Nevertheless, to acknowledge originality and respect chronology, we provide the historical timeline in \Cref{fig:timeline}.


\clearpage
\newpage

\section{Special Flow Map: Consistency Model in Discrete Time}\label{sec:discrete-cm}

\paragraph{An Important Principle of Flow Maps: The Semigroup Property.}
\begin{figure}[th!]
    \centering
    \includegraphics[width=0.9\linewidth]{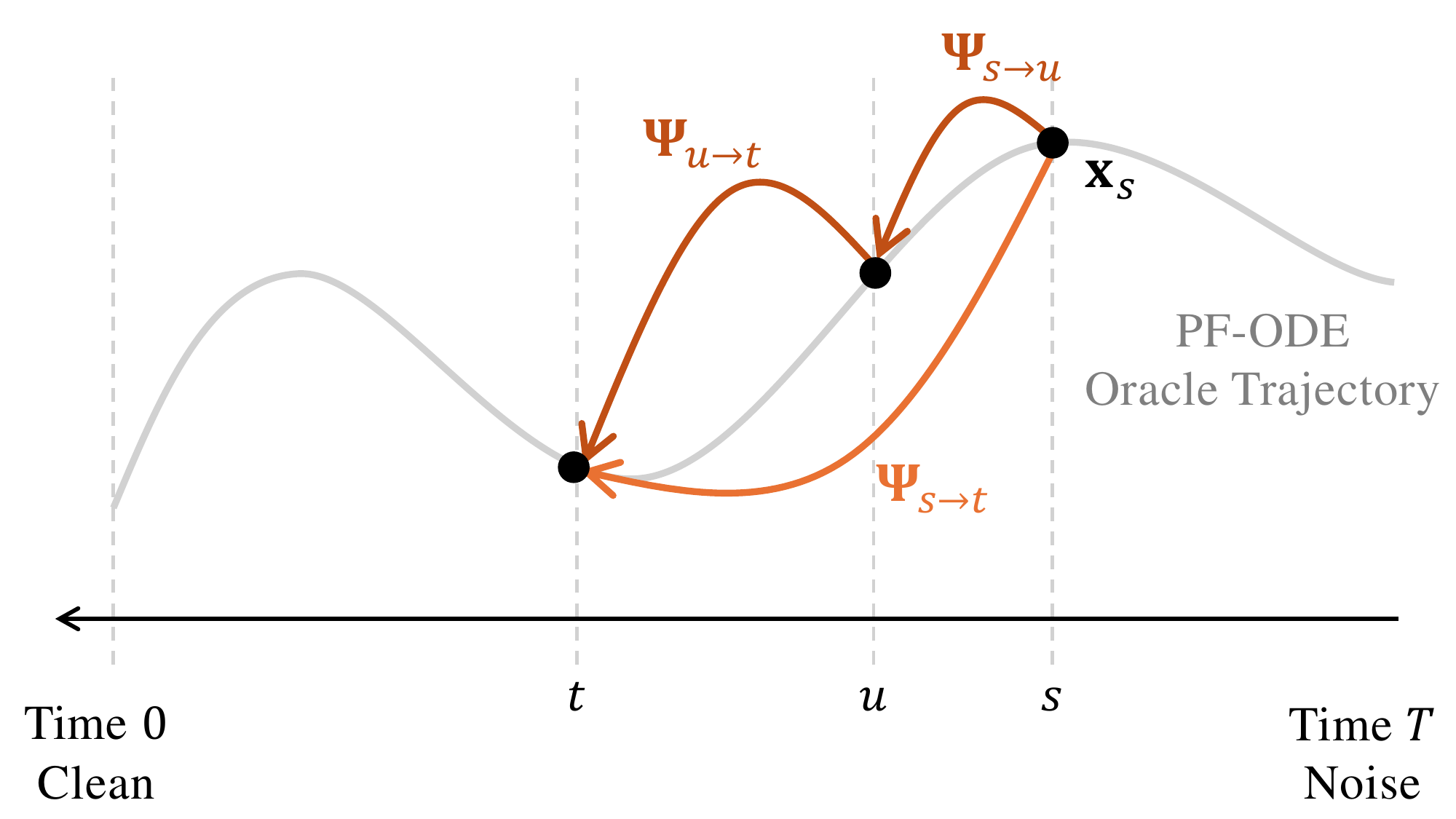}
    \caption{\textbfs{Illustration of the flow map semigroup property.} 
This property states that transitioning from $s$ to $u$ and then from $u$ to $t$ 
is equivalent to transitioning directly from $s$ to $t$.
\figcredit{Created by the authors.}}
    \label{fig:semigroup}
\end{figure}

Consistency Models (introduced in \Cref{sec:discrete-cm,sec:conti-cm}) and their generalization, the Consistency Trajectory Model (\Cref{sec:CTM}), define their regression targets by exploiting a key mathematical structure of flow maps. 
This structure is the fundamental \emph{semigroup property}:
\begin{mdframed}
   \begin{align}\label{eq:semi-group}
       \bPsi_{u\to t}\circ \bPsi_{s\to u}=\bPsi_{s\to t},
       \,\,\,
       \bPsi_{s\to s}=\rmI, \quad \text{for all }\, s, u, t \in [0,T].
   \end{align} 
\end{mdframed}
Intuitively, this means that if we first evolve a state from $s$ to $u$ (through $\bPsi_{s\rightarrow u}$)  and then from $u$ to $t$ (through $\bPsi_{u\rightarrow t}$), we end up at exactly the same point as if we had evolved directly from $s$ to $t$. 
This is nothing more than the basic principle of ODE solving\footnote{The semigroup property follows from the uniqueness theorem for ODE initial value problems (see \Cref{app:de}).}: once the starting point of a flow is specified, its future evolution is completely determined, and it follows a single well-defined path. Whether we follow this path in one long step or divide it into smaller intervals, we still move along the same trajectory and arrive at the same final state.

To build further intuition for the semigroup property, consider the solution trajectory $\{\rvx(s)\}_{s \in [0,T]}$ of the PF-ODE
\begin{align*}
    \frac{\diff \rvx(\tau)}{\diff \tau} = \rvv^*(\rvx(\tau), \tau),
\end{align*}
with a fixed initial condition $\rvx(T)$ at time $T$, solved backward in time.  
If we fix the terminal time at $t=0$, the corresponding flow map can be written more simply as
\[
\rvf^*(\cdot, s) := \bPsi_{s\to 0}(\cdot),
\]
which is referred to as the \emph{consistency function}.  
By construction, this function inherits several fundamental properties directly from the semigroup identity of \Cref{eq:semi-group} with $t=0$:

\begin{enumerate}
    \item[(i)] \textbfs{Global Consistency:} every point along the trajectory maps to the same clean endpoint,
    \[
       \rvf^*(\rvx(s), s) = \rvx(0), \quad \text{for all }\, s \in [0,T].
    \]
    This is because 
\begin{align*}
\rvf^*(\rvx(s),s)
&= \bPsi_{s\to 0}\big(\bPsi_{0\to s}(\rvx(0))\big)
= \big(\bPsi_{s\to 0}\circ \bPsi_{0\to s}\big)(\rvx(0))
\\&= \bPsi_{0\to 0}(\rvx(0))
= \rvx(0).
\end{align*}
    \item[(ii)] \textbfs{Self Consistency:} any two points along the same trajectory must give identical outputs,
    \begin{align}\label{eq:cm-self-consistency}
       \rvf^*(\rvx(s), s) = \rvf^*(\rvx(u), u), \quad \text{for all }\, s, u \in [0,T].
    \end{align}
    This is a direct re-interpretation of the semigroup identity: $\Psi_{s\rightarrow 0}\circ\Psi_{0\rightarrow s}=\Psi_{u\rightarrow 0}\circ\Psi_{0\rightarrow u}$.
    \item[(iii)] \textbfs{Local Consistency:} the consistency function is invariant with respect to $s$ when evaluated along the trajectory,
    \begin{align}\label{eq:cm-diff}
        \frac{\diff}{\diff s}\,\rvf^*(\rvx(s),s) = 0,
        \qquad \rvf^*(\rvx(0),0) = \rvx(0).
    \end{align}
This follows from global consistency, which states that $\rvf^*(\rvx(s), s)$ does
not change with $s$ along the trajectory.
\end{enumerate}
The three properties are all equivalent. Each states that along any solution trajectory $s\mapsto \rvx(s)$,
the flow-to-origin/consistency map $\rvf^*(\rvx(s),s)=\bPsi_{s\to 0}(\rvx(s))$
yields the same terminal point $\rvx(0)$, independent of the starting time.

\paragraph{Goal of Consistency Models.}
A CM aims to train a neural network 
$\rvf_{\bm{\theta}} \colon \mathbb{R}^D \times [0,T] \to \mathbb{R}^D$ 
to approximate the special flow map $\bPsi_{s\to 0}$, i.e., consistency function\footnote{The concept of a consistency function for an ODE generalizes to a function
$\rvf(\rvx,t)$ for an SDE such that $\rvf(\rvx_t,t)$ is a (local) martingale with
respect to the SDE’s natural filtration, i.e.
\[
\E[\rvf(\rvx_t,t)|\rvx_s]=\rvf(\rvx_s,s),\quad\text{for all }\quad t\ge s.
\]
This generalization was proposed/observed in \citep{daras2023consistent,lai2023fp},
and the theoretical connections are summarized in \citep{lai2023equivalence}.}. 
The key idea is to enforce consistency along PF-ODE trajectories: states lying
on the same ODE trajectory should map to the same time-$0$ endpoint
(more precisely, this corresponds to the special case $t=0$ and
$u=s-\Delta s$ in \Cref{eq:semi-group}).

There are, however, multiple ways to realize this goal. 
The choice depends on whether a pre-trained diffusion model is available and whether training is carried out in a discrete-time or continuous-time regime. 
We begin by summarizing these variants in \Cref{tb:consistency-methods} and illustrating their objectives in \Cref{fig:cd-ct-loss-distill}. 
The subsequent sections (\Cref{sec:discrete-cm,sec:conti-cm}) then gradually develop the details of each approach.

\begin{table}[th]
  \caption{Training Objectives of Consistency Models}
  \small
  \centering
  \resizebox{0.76\textwidth}{!}{
  \begin{tabular}{ccc}
     \toprule
     & \textbfs{Distillation} & \textbfs{From Scratch} \\
     \midrule
     \textbfs{Discrete-time} & \Cref{eq:cd-discrete-approx-N} & \Cref{eq:ct-discrete-approx-N} \\
     \textbfs{Continuous-time} & \Cref{eq:cd-continuous-time}  & \Cref{eq:ct-continuous-time} \\
     \bottomrule
  \end{tabular}
  }
  \label{tb:consistency-methods}
\end{table}

\subsection{Discrete-Time Approximations for Learning a Consistency Function}\label{subsec:approx-cm}

In principle, a consistency function can be learned by minimizing the oracle loss \Cref{eq:oracle-cm}:
\[
    \mathcal{L}_{\methodl{oracle}{CM}}(\btheta) 
    := \E_{s} \E_{\rvx_s\sim p_s} \left[ w(s)  d \left(\rvf_\btheta(\rvx_s,s), \bPsi_{s\to 0}(\rvx_s)\right)\right].
\]
This objective enforces that every noisy sample $\rvx_s$ is mapped back to its clean endpoint $\bPsi_{s\to 0}(\rvx_s)$.  

The challenge is that the oracle map $\bPsi_{s\to 0}(\rvx_s)$ is not available in practice.  
To overcome this, \citet{song2023consistency} exploit the \emph{semigroup property}: any noisy state and its consecutive step along the same PF-ODE trajectory must map to the same clean endpoint.  
Concretely, the oracle target is replaced by a \emph{stop-gradient} target taken from a slightly earlier point on the trajectory:
\begin{align*}
    \bPsi_{s\to 0}(\rvx_s) 
    &= \bPsi_{s-\Delta s \to 0} \left(\bPsi_{s\to s-\Delta s}(\rvx_s)\right) \\
    &\approx \rvf_{\btheta^-} \left(\bPsi_{s\to s-\Delta s}(\rvx_s),\,s-\Delta s\right), 
    \qquad \Delta s > 0,
\end{align*}
where $\bm{\theta}^-$ are parameters under the stop-gradient operator. 
A further difficulty is that the intermediate state $\bPsi_{s\to s-\Delta s}(\rvx_s)$ has no closed form either and must itself be approximated.  
Two practical regimes have been proposed:

\paragraph{With Pre-trained Diffusion Model (Consistency Distillation).}
Suppose we have access to a pre-trained diffusion model. 
Consistency Distillation (CD) leverages the teacher model to approximate the intermediate state $\bPsi_{s\to s-\Delta s}(\rvx_s)$ by simulating only a single backward ODE step:
\[
 \bPsi_{s\to s-\Delta s}(\rvx_s)  \approx  \mathtt{Solver}_{s \to s-\Delta s}(\rvx_s).
\]

More concretely, a pre-trained diffusion model provides an estimate of the score function 
$\rvs_{\bm{\phi}^\times}(\rvx_s, s) \approx \nabla_{\rvx_s}\log p_s(\rvx_s)$. 
Using this, one can perform a one-step DDIM update from $\rvx_s$ to obtain an approximation of the state at $s' = s-\Delta s$:
\begin{align*}
    \bPsi_{s\to s-\Delta s}(\rvx_s) 
    &\approx \frac{\alpha_{s'}}{\alpha_s}\,\rvx_s 
    + \sigma_s^2 \left( \frac{\alpha_{s'}}{\alpha_s} - \frac{\sigma_{s'}}{\sigma_s} \right)\nabla_{\rvx_s}\log p_s(\rvx_s) \\
    &\approx \frac{\alpha_{s'}}{\alpha_s}\,\rvx_s 
    + \sigma_s^2 \left( \frac{\alpha_{s'}}{\alpha_s} - \frac{\sigma_{s'}}{\sigma_s} \right)\rvs_{\bm{\phi}^\times}(\rvx_s, s) \\
    &:= \tilde{\rvx}_{s'}^{\bm{\phi}^\times}.
\end{align*}

Combining this construction with the stop-gradient target yields a practical \emph{discrete-time} proxy for the oracle loss $\mathcal{L}_{\methodl{oracle}{CM}}(\btheta)$. 
Formally, over a partition $0 = s_1 < s_2 < \cdots < s_N = T$, the CD training objective is given by
\begin{mdframed}
\begin{align}\label{eq:cd-discrete-approx-N}
    \mathcal{L}_{\text{CD}}^N(\bm{\theta}, \bm{\theta}^-; \bm{\phi}^\times) 
    := \E_{\rvx_0, \beps, i} \Big[\, \omega(s_i)\,
    d\big(\rvf_{\bm{\theta}}(\rvx_{s_{i+1}}, s_{i+1}),\, 
          \rvf_{\bm{\theta}^-}(\tilde{\rvx}_{s_i}^{\bm{\phi}^\times}, s_i)\big)\Big].
\end{align}
\end{mdframed}
Here, $\omega(\cdot)$ is a time-dependent weight, $d(\cdot,\cdot)$ is a distance measurement, and $\bm{\theta}^-$ indicates stop-gradient parameters, which prevent collapse to trivial solutions (e.g., constant predictions).

\paragraph{Without Pre-Trained Diffusion Model (Consistency Training).}
When no pre-trained diffusion model is available, the oracle score $\nabla_{\rvx}\log p_s(\rvx_s)$ can still be estimated directly using a simple one-point approximation (albeit with high variance). 
Recall that it admits the conditional expectation form:
\begin{align*}
    \nabla_{\rvx_s} \log p_s(\rvx_s)
    &= \E_{\rvx_0 \sim p(\rvx_0|\rvx_s)} \left[ \nabla_{\rvx_s}\log p(\rvx_s|\rvx_0) \right] \\
    &= \E_{\rvx_0 \sim p(\rvx_0|\rvx_s)} \left[ -\frac{\rvx_s - \alpha_s \rvx_0}{\sigma_s^2} \right].
\end{align*}
The identity above suggests a simple one-sample estimator. 
If $\rvx_s$ is obtained from a paired sample $(\rvx_0,\beps)$ via $\rvx_s = \alpha_s \rvx_0 + \sigma_s \beps$, then 
\[
\widehat{\nabla_{\rvx}\log p_s}(\rvx_s)
:=-\frac{\boldsymbol\epsilon}{\sigma_s}
=-\frac{\rvx_s-\alpha_s \rvx_0}{\sigma_s^2}
\]
serves as an \emph{unbiased} estimator of the score at $\rvx_s$ (conditionally unbiased with respect to $p(\rvx_0|\rvx_s)$). It corresponds exactly to the \emph{conditional score} used as the regression target in denoising score matching.

Plugging this estimate into the DDIM one-step update from $s$ to $s' = s-\Delta s$ (see \Cref{eq:ddim-update-all}) yields
\begin{align}\label{eq:one-pt-score}
\begin{aligned}
    \bPsi_{s\to s'}(\rvx_s)
&\approx
\frac{\alpha_{s'}}{\alpha_s}\,\rvx_s 
+ \sigma_s^2 \left(\frac{\alpha_{s'}}{\alpha_s}-\frac{\sigma_{s'}}{\sigma_s}\right)\nabla_{\rvx_s}\log p_s(\rvx_s) && \text{(DDIM)}
\\
&\approx
\frac{\alpha_{s'}}{\alpha_s}\,\rvx_s 
+ \sigma_s^2 \left(\frac{\alpha_{s'}}{\alpha_s}-\frac{\sigma_{s'}}{\sigma_s}\right)\widehat{\nabla_{\rvx_s}\log p_s}(\rvx_s) && \text{(1-pt score)}
\\
&=
\frac{\alpha_{s'}}{\alpha_s}\,\rvx_s 
-\left(\frac{\alpha_{s'}}{\alpha_s}-\frac{\sigma_{s'}}{\sigma_s}\right)(\rvx_s-\alpha_s\rvx_0)
\\
&= \alpha_{s'}\rvx_0+\sigma_{s'}\beps, 
\end{aligned}
\end{align}
where\footnote{The last identity follows directly from the forward corruption process $\rvx_s=\alpha_s \rvx_0+\sigma_s \beps$ by elementary algebra.}
$\rvx_0$ is the same data sample and $\beps$ is the same Gaussian noise used to
construct $\rvx_s$.

This leads to a teacher-free discrete-time surrogate of the oracle objective $\mathcal{L}_{\methodl{oracle}{CM}}$, written as
\begin{mdframed}
\begin{align}\label{eq:ct-discrete-approx-N}
    \mathcal{L}_{\text{CT}}^N(\bm{\theta}, \bm{\theta}^-) 
    := \E_{\rvx_0, \beps, i} \left[ 
       \omega(s_i)\, d \left(\rvf_{\bm{\theta}}(\rvx_{s_{i+1}}, s_{i+1}),\, 
       \rvf_{\bm{\theta}^-}(\rvx_{s_i}, s_i)\right)
    \right],
\end{align}
\end{mdframed}
with $\rvx_{s_i} = \alpha_{s_i}\rvx_0 + \sigma_{s_i}\beps$ and $\rvx_{s_{i+1}} = \alpha_{s_{i+1}}\rvx_0 + \sigma_{s_{i+1}}\beps$.

Using $\alpha_{s'}\rvx_0+\sigma_{s'}\beps$ directly as an approximation of $\bPsi_{s\to s'}(\rvx_s)$ without expectation introduces high variance\footnote{The one-point (conditional) score estimate $\widehat{\nabla_{\rvx}\log}p_s(\rvx_s)$ can be viewed as a one-sample Monte Carlo estimator, which is \emph{conditionally unbiased} given $\rvx_s$: averaging this estimator over the (generally intractable) clean posterior $p(\cdot|\rvx_s)$ recovers the true score as
\[
\nabla_{\rvx_s}\log p_s(\rvx_s)
= \E_{\rvx_0\sim p(\cdot|\rvx_s)}
 \left[\widehat{\nabla_{\rvx}\log}p_s(\rvx_s)\right].
\]}. 
Recall, however, the analogous conditional-expectation argument in denoising
score matching (see \Cref{sec:conditional-trick}). Here, the one-point state
estimate is unbiased before it is passed through the nonlinear target network
and discrepancy, so the resulting losses need not agree exactly. Conditioning
on $\rvx_s$, the first-order error cancels in expectation; under suitable
smoothness and moment assumptions, the remaining discrepancy is
$\mathcal{O}(\Delta s^2)$, as formalized below.

\thmp{CM-CT Equivalence up to Error $\mathcal{O}(\Delta s^2)$}{ct-cm}{
Let $s' := s - \Delta s$, and define
\begin{align*}
\mathcal{L}_{\mathrm{CM}}(\btheta,\btheta^-)
&:= \E_{s,\rvx_0,\beps} \big[w(s)\,d\big(\rvf_{\btheta}(\rvx_s,s), \rvf_{\btheta^-}(\rvx_{s'}^{\mathrm{DDIM}},s')\big)\big],\\
\mathcal{L}_{\mathrm{CT}}(\btheta,\btheta^-)
&:= \E_{s,\rvx_0,\beps} \big[w(s) d\big(\rvf_{\btheta}(\rvx_s,s), \rvf_{\btheta^-}(\rvx_{s'},s')\big)\big],
\end{align*}
where
\[
\rvx_{s'}^{\mathrm{DDIM}}
:= \frac{\alpha_{s'}}{\alpha_s}\rvx_s
+ \sigma_s^2 \left(\frac{\alpha_{s'}}{\alpha_s}-\frac{\sigma_{s'}}{\sigma_s}\right)\nabla_{\rvx_s}\log p_s(\rvx_s)
\]
is the oracle DDIM update.  Both  $\rvx_s = \alpha_s\rvx_0 + \sigma_s\beps$  and $\rvx_{s'} = \alpha_{s'}\rvx_0 + \sigma_{s'}\beps$
share the same pair $(\rvx_0,\beps)$ with 
$\rvx_0 \sim p_{\mathrm{data}}$ and 
$\beps \sim \mathcal{N}(\bm{0},\rmI)$.
Then,
\[
\mathcal{L}_{\mathrm{CM}}(\btheta,\btheta^-)
= \mathcal{L}_{\mathrm{CT}}(\btheta,\btheta^-) + \mathcal{O}(\Delta s^2).
\]
}{First, note that the DDIM update with the oracle score equals the conditional mean,
\[
\rvx_{s'}^{\mathrm{DDIM}} = \E[\rvx_{s'} | \rvx_s],
\]
which can also be verified from \Cref{eq:one-pt-score} by taking the expectation over $p(\cdot|\rvx_s)$.
Next, perform a Taylor expansion of
\[
d(\rvf_{\btheta}(\rvx_s,s), \rvf_{\btheta^-}(\cdot,s'))
\]
around
$\rvx_{s'}^{\mathrm{DDIM}}=\E[\rvx_{s'}|\rvx_s]$.
The one-point update recovers
$\rvx_{s'}=\alpha_{s'}\rvx_0+\sigma_{s'}\beps$ pathwise for the same
$(\rvx_0,\beps)$. Since
\[
\E[\rvx_{s'}-\rvx_{s'}^{\mathrm{DDIM}}\mid\rvx_s]=0,
\]
the linear Taylor term vanishes after conditioning on $\rvx_s$.
Under the smoothness and moment assumptions described above, the remaining
expected Taylor remainder is $\mathcal{O}(\Delta s^2)$. Hence,
$\mathcal{L}_{\mathrm{CT}}
=\mathcal{L}_{\mathrm{CM}}+\mathcal{O}(\Delta s^2)$.
A detailed derivation is provided in \Cref{app-sec:flow-map}.
}

In summary, CD leverages a teacher model for initialization and supervision, which often stabilizes optimization and reduces variance.  In contrast, Consistency Training (CT) requires no pre-trained model and can therefore be trained entirely from scratch.
Despite this difference, CT serves as a fully standalone generative model.
\paragraph{Practical Considerations.}
In practice, \citet{song2023consistency} adopt the EDM formulation
of \citet{karras2022elucidating} (see \Cref{sec:edm}) with the forward corruption
kernel $    \rvx_s = \rvx_0 + s \bm{\epsilon}$, and use the neural network parameterization proposed therein
(cf.\ \Cref{eq:D-parametrization}):
\[
    \rvf_{\bm{\theta}}(\rvx, s)
    = c_{\mathrm{skip}}(s) \rvx
      + c_{\mathrm{out}}(s)\,
        \rmF_{\bm{\theta}} \left(c_{\mathrm{in}}(s)\,\rvx,\, c_{\mathrm{noise}}(s)\right),
\]
where $\rmF_{\bm{\theta}}$ is a neural network and the coefficients follow
\Cref{eq:edm-simple-coefficients}. This parameterization has the important
boundary property
\[
    \rvf_{\bm{\theta}}(\rvx, 0) = \rvx,
\]
which enforces consistency at time zero and ensures the network output matches
its input when no noise is present.


\subsection{Sampling with Consistency Model}\label{subsec:cm-sampling}
Once a consistency model $\rvf_{\bm{\theta}^\times}$ is trained, either in continuous or discrete time, it can be used to generate samples in a single step or a few steps. The algorithm is summarized in \Cref{alg:cm_sampling}.

\paragraph{One-Step Generation.} Given an initial latent $\hat{\rvx}_T$ sampled from the prior distribution (in practice,  $\mathcal{N}(\bm{0}, T^2\rmI)$), a clean sample can be generated via a single function evaluation: $\rvf_{\bm{\theta}^\times}(\hat{\rvx}_T, T)$.

\paragraph{Multi-Step Generation.} With pre-selected timesteps 
$T>\tau_1 > \tau_2 > \cdots > \tau_{M-1}=0$, start from initial noise $\hat\rvx_T$ and alternate between noise injection and large clean jumps via the consistency model at earlier time points, gradually refining the sample:
\begin{align*}
    \hat\rvx_T \xrightarrow[\text{get a clean}]{\text{long jump}} \rvf_{\bm{\theta}^\times}(\hat\rvx_T, T)
    \xrightarrow[\text{to level $\tau_1$}]{\text{add noise}} \hat\rvx_{\tau_1}
    \xrightarrow[\text{get a clean}]{\text{long jump}} \rvf_{\bm{\theta}^\times}(\hat\rvx_{\tau_1}, \tau_1)
    \xrightarrow[\text{to level $\tau_2$}]{\text{add noise}} \cdots.
\end{align*}

\begin{algorithm}[thb!]
    {\small
        \caption{CM's Sampling with One-Step or Multi-Step Generation \label{alg:cm_sampling}}
    \begin{algorithmic}[1]
        \Require Consistency model $\rvf_{\bm{\theta}^\times}(\cdot, \cdot)$, sequence of time points $T>\tau_1 > \tau_2 > \cdots > \tau_{M-1}=0$, initial noise $\hat{\rvx}_T$
        \If{one-step}
            \State $\rvx \leftarrow \rvf_{\bm{\theta}^\times}(\hat{\rvx}_T, T)$
        \Else
            \State $\rvx \leftarrow \rvf_{\bm{\theta}^\times}(\hat{\rvx}_T, T)$
            \For{$m=1$ \textbf{to} $M-1$}
                \State Sample $\bm{\epsilon} \sim \mathcal{N}(\bm{0}, \bfI)$
                \State $\hat{\rvx}_{\tau_m} \leftarrow \alpha_{\tau_m}\rvx + \sigma_{\tau_m} \bm{\epsilon}$
                \State $\rvx \leftarrow \rvf_{\bm{\theta}^\times}(\hat{\rvx}_{\tau_m}, \tau_m)$
            \EndFor
        \EndIf
        \Ensure $\rvx$
    \end{algorithmic}
    }
\end{algorithm}

\clearpage
\newpage

\section{Special Flow Map: Consistency Model in Continuous Time}\label{sec:conti-cm}

We now move beyond the discrete-time setting of consistency models and consider
a continuous-time perspective. Unlike the discrete approach, which fixes a time
grid and trains only on those sampled points, the continuous formulation treats
the flow map as defined for all times. This shift eliminates the dependence on
an arbitrary discretization and provides a more principled alignment with the
underlying dynamics. It also helps reduce the approximation errors that
naturally arise from discretized integration, and ensures consistency is
enforced globally rather than only at selected steps.

\subsection{Continuous-Time Consistency Model}\label{subsec:conti-ct}
To motivate the continuous time formulation, we first revisit
\Cref{eq:cm-diff}, which describes the condition under which time derivatives
can be taken. Using the chain rule, we arrive at
\begin{align}\label{eq:conti-ct-motivation}
\begin{aligned}
        &\frac{\diff }{\diff s} \rvf^*(\rvx(s), s) = 0 \\
    \Longleftrightarrow\,\, 
    &\left(\nabla_{\rvx} \rvf^*\right)(\rvx(s), s) \cdot 
    \underbrace{\frac{\diff }{\diff s} \rvx(s)}_{\text{ODE velocity}} + 
    \left(\frac{\partial }{\partial s} \rvf^*\right)(\rvx(s), s) = 0,
\end{aligned}
\end{align}
where the trajectory $\rvx(s)$ follows the PF-ODE
\begin{align*}
    \frac{\diff }{\diff s}\rvx(s) =  \rvv^*(\rvx(s), s).
\end{align*}

This relationship shows that the consistency function $\rvf^*$ remains constant
along any solution trajectory of the ODE. The velocity field $\rvv^*$ can be
estimated in practice either from a pre-trained diffusion model (when such a
model is available) or from a direct one-point approximation, such as
$\alpha_s' \rvx_0 + \sigma_s' \beps$, as explained in
\Cref{sec:discrete-cm}.  

\Cref{eq:conti-ct-motivation} suggests a natural continuous-time characterization of consistency along PF-ODE trajectories. One possible approach is to enforce this differential condition directly by minimizing the residual, in a manner similar to physics-informed neural networks (PINNs)~\citep{raissi2018deep,boffi2024flow}:
\begin{align*}
\min_{\bm{\theta}}
\mathbb{E}_{s,\rvx_s\sim p_s}
\left[
\left\|
\partial_s\rvf_{\bm{\theta}}(\rvx_s,s)
+
\big(\partial_{\rvx}\rvf_{\bm{\theta}}(\rvx_s,s)\big)
\rvv^*(\rvx_s,s)
\right\|_2^2
\right].
\end{align*}

In practice, however, \citet{song2023consistency,lu2024simplifying} employ teacher-based consistency objectives rather than directly optimizing this residual. This motivates asking whether the discrete consistency loss admits a meaningful infinitesimal counterpart as $\Delta s \to 0$. The proposition below shows that, at the comparison point where the stop-gradient target coincides with the online network, the scaled discrete-time gradient converges to the gradient of a continuous-time objective:
\begin{align}\label{eq:cm-discrete}
\mathcal{L}_{\mathrm{CM}}^{\Delta s}(\btheta,\btheta^-)
:= \E \left[\omega(s) \big\|
\rvf_{\btheta}(\rvx_s,s) -
\rvf_{\btheta^-} \big(\bPsi_{s\to s-\Delta s}(\rvx_s), s-\Delta s\big)
\big\|_2^2\right].
\end{align}
For a uniform partition underlying
\Cref{eq:cd-discrete-approx-N,eq:ct-discrete-approx-N}, let
\[
\Delta s_N:=\frac{T}{N-1}.
\]
Then $\Delta s_N\to0$ as $N\to\infty$, so the infinitesimal limit in
\Cref{eq:cm-discrete} corresponds to refining these discrete objectives.

We summarize this key idea in the following proposition.

\proppp{Continuous-Time Consistency Training}{continuous-time-ct}{The following convergence result holds at $\bm{\theta}=\bm{\theta}^-$:
\begin{align*}
    \left.\lim_{\Delta s\to 0}\frac{1}{\Delta s}\nabla_{\btheta}\mathcal{L}_{\mathrm{CM}}^{\Delta s}(\bm{\theta}, \bm{\theta}^-)\right|_{\bm{\theta}=\bm{\theta}^-}
    =
    \left.\nabla_{\bm{\theta}} \mathcal{L}^\infty_{\mathrm{CM}}(\bm{\theta}, \bm{\theta}^-)\right|_{\bm{\theta}=\bm{\theta}^-}.
\end{align*}
Here,
\begin{align*}
    \mathcal{L}^\infty_{\mathrm{CM}}(\bm{\theta},\bm{\theta}^-)
    :=
    \mathbb{E}_{s,\rvx_0,\bm{\epsilon}}\left[
    2\omega(s)\,
    \rvf_{\bm{\theta}}^\top(\rvx_s, s)\cdot
    \frac{\diff }{\diff s}\rvf_{\bm{\theta}^-}(\rvx_s, s)
    \right],
\end{align*}
and the total differentiation identity is
\begin{align}\label{eq:total-diff}
    \frac{\diff}{\diff s}\rvf_{\btheta^-}(\rvx_s,s)
    =
    \partial_s \rvf_{\btheta^-}(\rvx_s,s)
    + \big(\partial_{\rvx}\rvf_{\btheta^-}(\rvx_s,s)\big)\rvv^*(\rvx_s,s).
\end{align}
}{A first-order Taylor expansion of the stop-gradient target around $(\rvx_s, s)$ shows that the loss $\mathcal{L}_{\mathrm{CM}}^{\Delta s}$ behaves, up to $\mathcal{O}(\Delta s^2)$, like an inner product between the student update $\nabla_{\btheta}\rvf_{\btheta}(\rvx_s,s)$ and the tangent change $\tfrac{\diff}{\diff s}\rvf_{\btheta^-}(\rvx_s,s)$. Evaluating at $\bm{\theta}=\bm{\theta}^-$ removes the zeroth-order discrepancy term, so the scaled gradient satisfies
\[
\left.\lim_{\Delta s\to 0}\frac{1}{\Delta s}\nabla_{\btheta}\mathcal{L}_{\mathrm{CM}}^{\Delta s}\right|_{\bm{\theta}=\bm{\theta}^-}
=
\left.\nabla_{\btheta}\E\left[
2\omega(s)\,
\rvf_{\btheta}^\top(\rvx_s,s)\cdot
\frac{\diff}{\diff s}\rvf_{\btheta^-}(\rvx_s,s)
\right]\right|_{\bm{\theta}=\bm{\theta}^-},
\]
which is the claimed identity. We defer the proof to \Cref{app-sec:flow-map}.
}

The result above is written under the gradient operator
$\nabla_{\bm{\theta}}$ so that terms involving $\bm{\theta}^-$ vanish, since
$\bm{\theta}^-$ is treated as constant under stop-gradient.  Note
that $\tfrac{\diff}{\diff s}\rvf_{\btheta^-}(\rvx_s,s)$ denotes the total
derivative along the oracle trajectory, rather than a simple partial time
derivative.  

In summary, the continuous-time consistency model can be trained using the
gradient of the following pseudo-objective (ignoring the factor $2$):
\begin{mdframed}
    \begin{align}\label{eq:conti-ct-loss}
    \mathcal{L}^\infty_{\mathrm{CM}}(\bm{\theta},\bm{\theta}^-) := \mathbb{E}_{s,\rvx_0, \bm{\epsilon}}\left[\omega(s)\rvf_{\bm{\theta}}^\top(\rvx_s, s) \cdot \frac{\diff }{\diff s} \rvf_{\bm{\theta}^-}(\rvx_s, s) \right]. 
\end{align}
\end{mdframed}

\subsection{Training Continuous-Time Consistency Model}
Similar to the discrete-time case discussed in \Cref{subsec:approx-cm}, 
we now clarify the practical approximation of the tangent term in 
\Cref{eq:conti-ct-loss}, which involves the inaccessible oracle velocity $\rvv^*$:
\begin{align*}
    \frac{\diff}{\diff s}\rvf_{\btheta^-}(\rvx_s,s)
    = \partial_s \rvf_{\btheta^-}(\rvx_s,s)
    + \big(\partial_{\rvx}\rvf_{\btheta^-}(\rvx_s,s)\big)\,\rvv^*(\rvx_s,s).
\end{align*}
After training a continuous-time CM, sampling follows the same procedure as in
the discrete time case (\Cref{subsec:cm-sampling}).

\paragraph{Continuous-Time Consistency Distillation.}
If a pre-trained diffusion model is available such that
$\rvv_{\bm{\phi}^\times} \approx \rvv^*$, then the tangent vector
$\tfrac{\diff}{\diff s}\rvf_{\btheta^-}(\rvx_s,s)$ in
\Cref{eq:total-diff} can be approximated by the surrogate
\begin{align}\label{eq:cd-continuous-time}
    \frac{\diff}{\diff s}\rvf_{\btheta^-}(\rvx_s,s) 
     \approx 
    \partial_s \rvf_{\btheta^-}(\rvx_s,s)
    + \big(\partial_{\rvx}\rvf_{\btheta^-}(\rvx_s,s)\big)\,
      \rvv_{\bm{\phi}^\times}(\rvx_s,s).
\end{align}
We denote the resulting objective as
$\mathcal{L}^\infty_{\mathrm{CM}}(\bm{\theta},\bm{\theta}^-;\bm{\phi}^\times)$.
Accordingly, Proposition~\ref{continuous-time-ct} can be restated as
\[
\lim_{N\to\infty} 
\frac{1}{\Delta s_N}\,\nabla_{\bm{\theta}}\,
\mathcal{L}_{\mathrm{CD}}^{N}(\bm{\theta},\bm{\theta}^-;\bm{\phi}^\times)
= 
\nabla_{\bm{\theta}}\,
\mathcal{L}_{\mathrm{CD}}^{\infty}(\bm{\theta},\bm{\theta}^-;\bm{\phi}^\times).
\]

\paragraph{Continuous-Time Consistency Training (from Scratch).} On the other hand, if a pre-trained diffusion model is not available, the oracle
velocity $\rvv^*$ can be approximated using the one point conditional estimate
$\alpha_s'\rvx_0 + \sigma_s'\beps$. In this case, the tangent vector
$\tfrac{\diff}{\diff s}\rvf_{\btheta^-}(\rvx_s,s)$ in
\Cref{eq:total-diff} is replaced by the surrogate
\begin{align}\label{eq:ct-continuous-time}
    \frac{\diff}{\diff s}\rvf_{\btheta^-}(\rvx_s,s) 
     \approx 
    \partial_s \rvf_{\btheta^-}(\rvx_s,s)
    + \big(\partial_{\rvx}\rvf_{\btheta^-}(\rvx_s,s)\big) 
      \left(\alpha_s'\rvx_0 + \sigma_s'\beps\right).
\end{align}
We denote the resulting objective as
$\mathcal{L}^\infty_{\mathrm{CT}}(\bm{\theta},\bm{\theta}^-)$, which corresponds
to the training from scratch setting. Accordingly,
Proposition~\ref{continuous-time-ct} can be restated as
\[
\lim_{N\to\infty} 
\frac{1}{\Delta s_N}\,\nabla_{\bm{\theta}} 
\mathcal{L}_{\mathrm{CT}}^{N}(\bm{\theta},\bm{\theta}^-)
=
\nabla_{\bm{\theta}} 
\mathcal{L}_{\mathrm{CT}}^{\infty}(\bm{\theta},\bm{\theta}^-).
\]

So far, we have introduced all the fundamental approaches listed in 
\Cref{tb:consistency-methods} to realize the learning of the consistency 
function $\bPsi_{s\to 0}$. 
To provide a clearer overview, \Cref{fig:cd-ct-loss-distill} summarizes the 
relationships among the different loss functions for training consistency 
functions. 
The figure also indicates whether each method relies on a pre-trained diffusion 
model and distinguishes between continuous time and discrete time objectives.

\begin{figure}[ht]
\centering
\scalebox{0.9}{ 
\begin{tikzpicture}[
  box/.style={draw, rounded corners, thick, minimum width=3.5cm, minimum height=1cm, align=center},
  arrow/.style={thick, -{Latex[length=3mm]}},
  xshift=0cm, yshift=0cm
]

\node[box] (discCD) {\small{Discrete-Time CD}};
\node[box, right=4.5cm of discCD] (discCT) {\small{Discrete-Time CT}};
\node[box, below=2cm of discCD] (contCD) {\small{Continuous-Time CD}};
\node[box, below=2cm of discCT] (contCT) {\small{Continuous-Time CT}};

\draw[arrow] (discCD) -- node[midway, above] {\footnotesize 
\shortstack{
$\mathcal{L}_{\text{CD}}(\bm{\theta},\bm{\theta}^-)  =\mathcal{L}_{\text{CT}}(\bm{\theta},\bm{\theta}^-) + \mathcal{O}(\Delta s^2)$
\\
\scriptsize (Theorem~\ref{thm:ct-cm})
}
} (discCT);
\draw[arrow] (discCD) -- node[midway, left] {
  \footnotesize
  \shortstack{
    $\displaystyle\lim_{N\rightarrow\infty} \frac{1}{\Delta s_N}\nabla_{\bm{\theta}}\mathcal{L}^N_{\text{CD}}$ \\ $=\nabla_{\bm{\theta}}\mathcal{L}^\infty_{\text{CD}}  $ \\
\scriptsize (Theorem 5)   
  }
} (contCD);
\draw[arrow] (discCT) -- node[midway, right] {
  \footnotesize
  \shortstack{
    $\displaystyle\lim_{N\rightarrow\infty} \frac{1}{\Delta s_N}\nabla_{\bm{\theta}}\mathcal{L}^N_{\text{CT}}  $\\
    $= \nabla_{\bm{\theta}}\mathcal{L}^\infty_{\text{CT}}$ \\
\scriptsize (Theorem 6) 
  }
} (contCT);
\draw[arrow] (discCD) to[out=-30,in=150] node[midway, above, yshift=-8.5mm, text width=5.2cm, align=center] {
  \footnotesize
  \shortstack{
    $\displaystyle\lim_{N\rightarrow\infty} \frac{1}{\Delta s_N}\nabla_{\bm{\theta}}\mathcal{L}^N_{\text{CD}}(\bm{\theta},\bm{\theta}^-;\bm{\phi}^\times)  $\\
    $= \nabla_{\bm{\theta}}\mathcal{L}^\infty_{\text{CT}}(\bm{\theta},\bm{\theta}^-)$ \\
\scriptsize (Theorem 6) 
  }
} (contCT);
\draw[arrow] (contCD) -- node[midway, below] {
  \footnotesize
  \shortstack{
    $\mathcal{L}^\infty_{\text{CD}}(\bm{\theta},\bm{\theta}^-;\bm{\phi}^\times)= \mathcal{L}^\infty_{\text{CT}}(\bm{\theta},\bm{\theta}^-)$ 
  }
} (contCT);
\end{tikzpicture}
}
\caption{\textbfs{Diagram showing relationships between discrete/continuous-time CD and CT under the $\ell_2$ distance metric: $d(\rvx,\rvy)=\|\rvx-\rvy\|_2^2$.} The marked theorems follow the labeling in \citep{song2023consistency}. Whenever the theorems involve CT, we assume a perfect score: $\rvs_{\bm{\phi}^\times}(\rvx, t) \equiv \nabla_\rvx \log p_t(\rvx)$. $\mathcal{L}^\infty_{\text{CT}}$ is defined in \Cref{eq:conti-ct-loss}, while $\mathcal{L}^\infty_{\text{CD}}$ is defined in \Cref{eq:cd-continuous-time}.
\figcredit{Created by the authors.}}
\label{fig:cd-ct-loss-distill}
\end{figure}

However, the tangent vector 
$\tfrac{\diff }{\diff s}\rvf_{\bm{\theta}^-}$ 
often causes instability during training. 
In the following optional section, we present techniques from 
\emph{Simplifying, Stabilizing and Scaling Continuous Time Consistency Models} 
(sCM)~\citep{lu2024simplifying} that mitigate these issues.

\subsection{(Optional) Practical Considerations of Continuous-Time CT} 
Our interest lies in the training from scratch scenario, since it yields a
standalone generative model that does not rely on external pre-trained diffusion
models. Hence, we focus our discussion on the continuous time case.

In practice, however, training directly with \Cref{eq:conti-ct-loss} is often unstable, 
as the term $\tfrac{\diff}{\diff s}\rvf_{\bm{\theta}^-}$ can exhibit large or unbounded time derivatives, 
leading to exploding gradients during optimization. To overcome this, suitable
parameterizations and stabilization strategies are typically required
\citep{geng2025consistency,lu2024simplifying}. As summarized in
\Cref{subsec:summary-disentagle}, the main factors that influence stable
training include the \emph{diffusion process}, \emph{parameterization choices},
\emph{time weighting function}, and \emph{time sampling distribution}, all of
which should be carefully designed and disentangled also in continuous-time CM.

\paragraph{Diffusion Process.}  Instead of using the standard diffusion parameterization 
$\rvx_s = \alpha_s \rvx_0 + \sigma_s \bm{\epsilon}$ with 
$\bm{\epsilon} \sim \mathcal{N}(\bm{0}, \rmI)$, 
\citet{lu2024simplifying} adopt a trigonometric schedule. 
This schedule, although mathematically equivalent to the original form 
(as shown in \Cref{eq:trig-para}), provides a cleaner structure and a better 
separation in the training objective, which contributes to improved stability 
during training
\footnote{Intuitively, both the trigonometric functions and their derivatives are bounded, which helps prevent scale explosion in terms like $\frac{\diff}{\diff s} \rvf_{\bm{\theta}^-}$. A detailed discussion is provided later.
}. In addition, they incorporate the standard deviation $\sigma_{\mathrm{d}}$ of the data distribution $p_{\mathrm{data}}$, in line with EDM’s design in \Cref{subsec:D-design-edm}:
\begin{align}\label{eq:trig-scm}
    \rvx_s := \cos(s) \rvx_0 + \sin(s) \rvz, \quad \text{where } \rvz \sim \mathcal{N}(\bm{0}, \sigma_{\mathrm{d}}^2 \rmI).
\end{align}

This formulation is fully general. For any diffusion process of the form $\rvx_s = \alpha_s \rvx_0 + \sigma_s \bm{\epsilon}$ with $\bm{\epsilon} \sim \mathcal{N}(\bm{0}, \rmI)$, we can equivalently write:
\[
\rvx_s = \alpha_s \rvx_0 + \frac{\sigma_s}{\sigma_\mathrm{d}} \cdot (\sigma_{\mathrm{d}} \bm{\epsilon}),
\]
by defining $\rvz := \sigma_{\mathrm{d}} \bm{\epsilon}$, $\alpha'_s := \alpha_s$, and $\sigma'_s := \frac{\sigma_s}{\sigma_{\mathrm{d}}}$. The transformed pair $(\alpha'_s, \sigma'_s)$ can then be mapped to the trigonometric form $(\cos(s), \sin(s))$ using the normalization described in \Cref{eq:any-to-trig}.

\paragraph{Parametrizations.}
By considering the analogous principles of EDM in \Cref{subsec:D-design-edm}, \citet{lu2024simplifying} propose the following parametrization for the neural network similar to \Cref{eq:D-parametrization}:
\begin{align*}
    \rvf_{\bm{\theta}}(\rvx, s) := c_{\mathrm{skip}}(s)\rvx + c_{\mathrm{out}}(s)\rmF_{\bm{\theta}}\left(c_{\mathrm{in}}(s)\rvx, c_{\mathrm{noise}}(s)\right).
\end{align*}
Here, $c_{\mathrm{skip}}(s)$, $c_{\mathrm{out}}(s)$, and $c_{\mathrm{in}}(s)$ can be derived using the same criteria presented in \Cref{subsec:D-design-edm} (see Appendix B of \citet{lu2024simplifying} for detailed derivations), and are given by
\[
c_{\mathrm{skip}}(s) = \cos(s), \quad c_{\mathrm{out}}(s) = -\sigma_{\mathrm{d}} \sin(s), \quad c_{\mathrm{in}}(s) \equiv \frac{1}{\sigma_{\mathrm{d}}}.
\]
This is considered along with the default choice $c_{\mathrm{noise}}(s) = s$, where $\partial_s c_{\mathrm{noise}}(s)$ is bounded to ensure training stability, as will be discussed around \Cref{eq:ds-decomposition}. This leads to the following parametrization under the trigonometric schedule:
\begin{align}\label{eq:trig-nn}
    \rvf_{\bm{\theta}}(\rvx, s) = \cos(s)\rvx - \sin(s)\sigma_{\mathrm{d}}\rmF_{\bm{\theta}}\left(\frac{\rvx}{\sigma_{\mathrm{d}}}, c_{\mathrm{noise}}(s)\right).
\end{align}
We note that this parametrization also ensures that the neural network always satisfies the boundary condition $\rvf_{\bm{\theta}}(\rvx, 0) \equiv \rvx$ for all $\rvx$, which is an essential property of a consistency function.

\subparagraph{Techniques for Stabilizing Tangent Training.}
Under the trigonometric schedule and the network parametrization described in \Cref{eq:trig-nn}, the gradient of the loss in \Cref{eq:conti-ct-loss} becomes
\begin{align}\label{eq:conti-ct-loss-trig}
    \nabla_{\bm{\theta}}\mathcal{L}^\infty_{\text{CT}}(\bm{\theta},\bm{\theta}^-)  =\nabla_{\bm{\theta}} \mathbb{E}_{s, \rvx_0, \bm{\epsilon}} \left[ -\omega(s) \sigma_{\mathrm{d}} \sin(s) \rmF_{\bm{\theta}}^\top\left( \frac{\rvx_s}{\sigma_{\mathrm{d}}}, s \right) \cdot \frac{\diff \rvf_{\bm{\theta}^-}}{\diff s} (\rvx_s, s) \right].
\end{align}
 In theory, training with the gradient update in \Cref{eq:conti-ct-loss-trig} may be sufficient to learn a consistency function. However, \citet{lu2024simplifying} empirically observed that the training process can become unstable in practice due to the behavior of the tangent function, given by
\begin{align*}
&\underbrace{\frac{\diff \rvf_{\bm{\theta}^-}(\rvx_s, s)}{\diff s}}_{\text{\textbfs{A.}}}
= \\&-\cos(s) \left( \sigma_{\mathrm{d}} \rmF_{\bm{\theta}^-}\left( \frac{\rvx_s}{\sigma_{\mathrm{d}}}, s \right) - \frac{\diff \rvx_s}{\diff s} \right) - \sin(s) \left( \rvx_s + \sigma_{\mathrm{d}} \frac{\diff \rmF_{\bm{\theta}^-}}{\diff s}\left( \frac{\rvx_s}{\sigma_{\mathrm{d}}}, c_{\mathrm{noise}}(s) \right) \right).
\end{align*}
In particular, instability was observed in the term
\begin{align*}
\underbrace{\sin(s) \frac{\diff \rmF_{\bm{\theta}^-}}{\diff s}\left( \frac{\rvx_s}{\sigma_{\mathrm{d}}}, c_{\mathrm{noise}}(s) \right)}_{\text{\textbfs{B.}}}
= \sin(s) \nabla_{\rvx_s} \rmF_{\bm{\theta}^-} \frac{\diff \rvx_s}{\diff s}
+ \sin(s) \partial_s \rmF_{\bm{\theta}^-}.
\end{align*}
More specifically, the instability arises from the component
\begin{align}\label{eq:ds-decomposition}
        \sin(s) \partial_s \rmF_{\bm{\theta}^-} = \sin(s) \underbrace{\frac{\partial c_{\mathrm{noise}}(s)}{\partial s} \cdot \frac{\partial \text{emb}(c_{\mathrm{noise}})}{\partial c_{\mathrm{noise}}}}_{\text{\textbfs{C.}}} \cdot \frac{\partial \rmF_{\bm{\theta}^-}}{\partial \text{emb}(c_{\mathrm{noise}})}.
\end{align}
Here, we follow a common practice in the DM and CM literature by applying a positional or Fourier embedding, denoted by $\text{emb}(\cdot)$, to the time variable $c_{\mathrm{noise}}(s)$:
\[
s \mapsto c_{\mathrm{noise}}(s) \mapsto \text{emb}(c_{\mathrm{noise}}(s)) \mapsto \rmF_{\bm{\theta}^-}\left( \frac{\rvx_s}{\sigma_{\mathrm{d}}}, \text{emb}(c_{\mathrm{noise}}(s)) \right).
\]

Therefore, some additional empirical techniques are introduced to mitigate the instability:
\begin{itemize}
    \item \textbfs{A. Tangent Normalization.} Explicitly normalize the tangent function by replacing $\frac{\diff }{\diff s}\rvf_{\theta^-}$ with
    $    \frac{\frac{\diff }{\diff s} \rvf_{\theta^-}}{\left\| \frac{\diff }{\diff s} \rvf_{\theta^-} \right\|_2 + c}$,
    where $c>0$ is a constant set empirically. Alternatively, clipping the tangent within $[-1, 1]$ can also effectively cap its variance.
    \item \textbfs{B. Tangent Warm-Up.} Since the term $\sin(s)(\rvx_s + \sigma_{\mathrm{d}} \frac{\diff }{\diff s}\rmF_{\bm{\theta}^-})$ may induce instability, an optional technique can be applied by replacing the coefficient $\sin(s)$ with $r \cdot \sin(s)$, where $r$ linearly increases from 0 to 1 over the first few training iterations.
    \item \textbfs{C. Time Embedding.} In light of the derivative chain in \Cref{eq:ds-decomposition}, \citet{lu2024simplifying} opted for a smaller magnitude parameter to control the derivative $\frac{\partial \text{emb}(c_{\mathrm{noise}})}{\partial c_{\mathrm{noise}}}$. For a similar reason, $c_{\mathrm{noise}}(s) = s$ is chosen, where $\partial_s c_{\mathrm{noise}}(s)=1$, a bounded constant.
\end{itemize}
On top of these, architectural changes for improved normalization (for stability) and efficient JVP-based computation of $\frac{\diff}{\diff s}\rvf_{\theta^-}$ are often necessary, but beyond our scope.

\paragraph{Time-Weighting Function.}
Manual design of the time-weighting function $\omega(s)$ may lead to suboptimal performance. To address this, following a similar approach to EDM-2~\citep{karras2024analyzing}, \citet{lu2024simplifying}  learn an adaptive weighting function $\omega_{\bm{\varphi}}(s)$ to balance the training loss variance across different times $s$ (see \Cref{eq:effect-adaptive-wgt} for the desired outcome). 

To elaborate further, we observe that the objective function in \Cref{eq:conti-ct-loss-trig} takes the form 
\[
\mathbb{E}_{s, \rvx_0, \bm{\epsilon}} \left[\rmF_{\bm{\theta}}^\top \rvy \right],\quad\text{with}\quad \rvy=-\omega(s) \sigma_{\mathrm{d}} \sin(s)\frac{\diff \rvf_{\bm{\theta}^-}}{\diff s}.
\]
Since $\rvy$ is a vector independent of $\bm{\theta}$, \Cref{eq:conti-ct-loss-trig} is equivalent to
\begin{align*}
    \nabla_{\bm{\theta}} \mathbb{E}_{s, \rvx_0, \bm{\epsilon}} \left[\rmF_{\bm{\theta}}^\top \rvy \right]
    = \frac{1}{2} \nabla_{\bm{\theta}} \mathbb{E}_{s, \rvx_0, \bm{\epsilon}} \left\| \rmF_{\bm{\theta}} - \rmF_{\bm{\theta}^-} + \rvy \right\|_2^2.
\end{align*}
Based on this observation, \citet{lu2024simplifying} propose additionally training an adaptive weighting network $\omega_{\bm{\varphi}}(s)$ to estimate the loss norm, formulated as the following minimization problem:
\begin{align*}
    \min_{\bm{\varphi}} \mathbb{E}_{s, \rvx_0, \bm{\epsilon}} \left[ \frac{e^{\omega_{\bm{\varphi}}(s)}}{D} \left\| \rmF_{\bm{\theta}} - \rmF_{\bm{\theta}^-} + \rvy \right\|_2^2 - \omega_{\bm{\varphi}}(s) \right].
\end{align*}
To understand the effect of the adaptive weighting, observe that the optimal solution $\omega^*(s)$ (obtained by taking the partial derivative of the above objective with respect to $\omega_{\bm{\varphi}}$) satisfies
\begin{align}\label{eq:effect-adaptive-wgt}
    \mathbb{E}_{\rvx_0,\bm{\epsilon}}
    \left[
    \frac{e^{\omega^*(s)}}{D}
    \left\|
    \rmF_{\bm{\theta}}
    - \rmF_{\bm{\theta}^-}
    + \rvy
    \right\|_2^2
    \right]
    = 1,
    \qquad \text{for almost every } s.
\end{align}
That is, for an unconstrained pointwise weighting function, the expected
rescaled loss is normalized separately at each time $s$. As a result, the adaptive weighting effectively reduces the variance of the training loss across different time steps, leading to more balanced and stable training.

\paragraph{Time Sampling Distribution.} \citet{lu2024simplifying} opt to sample $\tan(s)$ from a log-normal proposal distribution~\citep{karras2022elucidating}, that is,
\begin{align}\label{eq:s-dist}
    \log\big(\sigma_{\mathrm{d}} \tan(s)\big) \sim \mathcal{N}(\cdot;P_{\text{mean}}, P_{\text{std}}^2).
\end{align}
Here, $P_{\text{mean}}$ and $P_{\text{std}}$ are two hyper-parameters. Following \citet{lu2024simplifying}, we can use the prior weighting
\[
\omega(s)=\frac{1}{\sigma_{\mathrm{d}}\tan(s)},
\]
so that
\[
-\omega(s)\sigma_{\mathrm{d}}\sin(s)
\frac{\diff\rvf_{\bm{\theta}^-}}{\diff s}
=
-\cos(s)\frac{\diff\rvf_{\bm{\theta}^-}}{\diff s}.
\]

\paragraph{Summary of Training Objective.} In summary of the aforementioned discussion, the final training loss is expressed as:
\begin{align*}
    &\mathcal{L}_{\text{sCM}}(\bm{\theta}, \bm{\varphi})
    := \\&\mathbb{E}_{s, \rvx_0, \bm{\epsilon}} \left[
    \frac{e^{\omega_{\bm{\varphi}}(s)}}{D}
    \left\|
    \rmF_{\bm{\theta}}\left(\frac{\rvx_s}{\sigma_{\mathrm{d}}}, s\right)
    - \rmF_{\bm{\theta}^-}\left(\frac{\rvx_s}{\sigma_{\mathrm{d}}}, s\right)
    - \cos(s) \frac{\diff \rvf_{\bm{\theta}^-}}{\diff s}\left(\rvx_s, s\right)
    \right\|_2^2
    - \omega_{\bm{\varphi}}(s)
    \right].
\end{align*}
Here, $s$ is sampled according to \Cref{eq:s-dist}, and $\rvx_s$ is computed via \Cref{eq:trig-scm}. The model trained with this loss is referred to as \emph{sCM}, and its training procedure is summarized in \Cref{alg:scm}.

\begin{algorithm}[tb]
{\small
\caption{Training of Continuous-time Consistency Models (sCM) \label{alg:scm}}
\begin{algorithmic}[1]
\Require dataset $\mathcal{D}$ with std. $\sigma_{\mathrm{d}}$, pre-trained DM $\mathbf{F}_{\text{pretrain}}$ with parameter $\bm{\theta}_{\text{pretrain}}$, model $\mathbf{F}_{\bm{\theta}}$, weighting $\omega_{\boldsymbol{\varphi}}$, learning rate $\eta$, proposal $(P_{\text{mean}}, P_{\text{std}})$, constant $c$, warmup iteration $H$
\State \textbfs{Init:} $\bm{\theta} \gets \bm{\theta}_{\text{pretrain}}$, Iters $\gets 0$
\Repeat
    \State $\rvx_0 \sim \mathcal{D}$, $\mathbf{z} \sim \mathcal{N}(\bm{0}, \sigma_{\mathrm{d}}^2 \mathbf{I})$, $\tau \sim \mathcal{N}(P_{\text{mean}}, P_{\text{std}}^2)$, $s \gets \arctan\left(\frac{e^\tau}{\sigma_{\mathrm{d}}}\right)$
    \State $\rvx_s \gets \cos(s) \rvx_0 + \sin(s) \mathbf{z}$
    \If{consistency training}
        \State $\frac{\diff\rvx_s}{\diff s} \gets \cos(s) \mathbf{z} - \sin(s) \rvx_0$
    \Else
        \State $\frac{\diff\rvx_s}{\diff s} \gets \sigma_{\mathrm{d}} \mathbf{F}_{\text{pretrain}}\left(\frac{\rvx_s}{\sigma_{\mathrm{d}}}, s\right)$
    \EndIf
    \State $r \gets \min\left(1, \frac{\text{Iters}}{H}\right)$ \hfill \Comment{ Tangent warmup}
    \State $\rvw \gets - \cos^2(s) (\sigma_{\mathrm{d}} \mathbf{F}_{\bm{\theta}}^{-} - \frac{\diff\rvx_s}{\diff s}) - r  \cos(s) \sin(s) \left(\rvx_s + \sigma_{\mathrm{d}} \frac{\diff\mathbf{F}_{\bm{\theta}^-}}{\diff s}\right)$
    \State $\rvw \gets \frac{\rvw}{\|\rvw\| + c}$ \hfill \Comment{ Tangent normalization}
    \State $\mathcal{L}_{\text{sCM}}(\bm{\theta}, \boldsymbol{\varphi}) \gets\frac{e^{\omega_{\boldsymbol{\varphi}}(s)}}{D} \norm{ \mathbf{F}_{\bm{\theta}}\left( \frac{\rvx_s}{\sigma_{\mathrm{d}}}, s\right) - \mathbf{F}_{\bm{\theta}^{-}}\left( \frac{\rvx_s}{\sigma_{\mathrm{d}}}, s\right) - \rvw }^2_2 - \omega_{\boldsymbol{\varphi}}(s)$
    \Statex  \hfill \Comment{ Adaptive weighting}
    \State $(\bm{\theta}, \boldsymbol{\varphi}) \gets (\bm{\theta}, \boldsymbol{\varphi}) - \eta \nabla_{\bm{\theta}, \boldsymbol{\varphi}} \mathcal{L}_{\text{sCM}}(\bm{\theta}, \boldsymbol{\varphi})$
    \State Iters $\gets$ Iters + 1
\Until{convergence}
\end{algorithmic}
}
\end{algorithm}

\clearpage
\newpage

\section{\texorpdfstring{General Flow Map: Consistency Trajectory Model}{General Flow Map: Consistency Trajectory Model}}\label{sec:CTM}

Consistency Trajectory Model (CTM)~\citep{kim2023consistency} is among the first methods to learn a \emph{general} flow map $\bPsi_{s\to t}$.

\paragraph{Setup of CTM in Practice.}
Similar to the CM family, CTM originally follows the formulation of EDM~\citep{karras2022elucidating} (\Cref{sec:edm}), 
using the PF-ODE in $\rvx$-prediction form with the noise schedule 
$\alpha_t = 1$ and $\sigma_t = t$. 
Under this setup, the PF-ODE becomes
\begin{align*}
    \frac{\diff \rvx(\tau)}{\diff \tau} 
    = \frac{\rvx(\tau) - \E[\rvx|\rvx(\tau)]}{\tau}.
\end{align*}
Starting from $\rvx_s$ at time $s$ and evolving backward to an ending
time $t\le s$,
the corresponding flow map (solution) can be written equivalently as
\begin{align*}
\rvx_s + \int_s^t 
    \frac{\rvx_\tau - \E[\rvx |\rvx_\tau]}{\tau}\, \diff \tau.
\end{align*}

CTM adopts an Euler-inspired parameterization: applying a single-step Euler
solver to the PF-ODE\footnote{For affine-in-time schedules, such as
$(\alpha_t,\sigma_t)=(1,t)$ or $(1-t,t)$, the plain Euler step in diffusion
time coincides exactly with the deterministic DDIM update. For general
schedules, the two agree only to first order. See discussion in \Cref{subsec:ddim-classical-euler}.} yields
\[
\rvx_{s\to t}^{\mathrm{Euler}}
= \rvx_s - (s-t)\,\frac{\rvx_s - \E[\rvx|\rvx_s]}{s}
= \frac{t}{s}\,\rvx_s + \Big(1 - \frac{t}{s}\Big)\E[\rvx|\rvx_s],
\]
where $\rvx_{s\to t}^{\mathrm{Euler}}$ approximates the solution at time $t$ given the state $\rvx_s$ at time $s$.

While the EDM setup provides a simple illustrative case, 
CTM allows broader noise schedules defined by an arbitrary linear Gaussian forward kernel $(\alpha_t, \sigma_t)$ 
and expresses the PF-ODE in $\rvv$-prediction form:
\[
\bPsi_{s\to t}(\rvx_s)
= \rvx_s + \int_s^t \rvv^*(\rvx_u,u) \diff u.
\]
In the discussion that follows, we focus on this general formulation.

\subsection{CTM Parametrization for Flexible Transition Learning}
Following the single-step Euler solver of the PF-ODE above, 
CTM rewrites the oracle flow map $\bPsi_{s\to t}$ 
as a convex combination of the input $\rvx_s$ and a residual function $\rvg^*$:
\begin{align*}
\bPsi_{s\to t}(\rvx_s)
:= \rvx_s + \int_s^t \rvv^*(\rvx_u,u) \diff u 
= \frac{t}{s}\,\rvx_s
+ \frac{s-t}{s}
\underbrace{\Big[\rvx_s + \frac{s}{s-t} \int_s^t \rvv^*(\rvx_u,u) \diff u\Big]}_{=:~\rvg^*}.
\end{align*}
where the residual term $\mathbf{g}^*$ is defined as
\begin{align}\label{eq:ctm-g-oracle}
\mathbf{g}^*(\mathbf{x}_{s}, s, t) &:= \rvx_s + \frac{s}{s-t} \int_s^t \rvv^*(\rvx_u,u)\diff u.
\end{align}
This motivates the neural parameterization
\begin{align}\label{eq:ctm_para_1}
    \rmG_{\btheta}(\rvx_s,s,t)
:= \frac{t}{s}\,\rvx_s + \frac{s-t}{s}\,\rvg_{\btheta}(\rvx_s,s,t),
\end{align}
where $\rvg_{\btheta}$ is a neural network that aims at $\rvg_{\btheta} \approx \rvg^*$, 
hence $\rmG_{\btheta}(\rvx_s,s,t)$ is trained to approximate the oracle flow map,
\[
\rmG_{\btheta}(\rvx_s,s,t) \approx \bPsi_{s\to t}(\rvx_s).
\]
Therefore, CTM naturally fits within the general consistency-mapping framework of  
\Cref{eq:general-cm-loss:rep}, which aligns the learned mapping with the oracle flow map.

Moreover, this formulation inherently satisfies the initial condition
\[
\rmG_{\btheta}(\rvx_s,s,s)=\rvx_s,
\]
without requiring any explicit enforcement during training.

\paragraph{Advantages of CTM's Parametrizations.}
A crucial characteristic of $\rvg^*$ becomes evident when taking the limit as $t$ approaches $s$ (i.e., the same ending time as the starting time):
\proppp{Properties of $\rvg^*$}{ctm-g-para}{\begin{enumerate}
    \item[(i)] \textbfs{Recovering Diffusion Model:} 
    \[
    \rvg^*(\rvx_s, s, s)=\lim_{t\rightarrow s}\mathbf{g}^*(\mathbf{x}_{s},s,t)=\rvx_s - s\rvv^*(\rvx_s,s).
    \]
    \item[(ii)] \textbfs{Integration Representation:} 
    \[
    \mathbf{g}^*(\mathbf{x}_{s},s,t)=\rvx_s - s\rvv^*(\rvx_s,s)+\mathcal{O}(\vert t-s\vert).
    \]
\end{enumerate}}{From the definition of $\rvg^*$, we obtain
\begin{align*}
    \lim_{t\rightarrow s}\mathbf{g}^*(\mathbf{x}_{s},s,t)=\mathbf{x}_{s}-s\lim_{t\rightarrow s}\frac{1}{t-s}\int_{s}^{t}\rvv^*(\mathbf{x}_{\tau}, \tau)\diff \tau=\rvx_s - s\rvv^*(\rvx_s,s).
\end{align*}
 This proved the first identity. For the second claim, from the Taylor expansion, we have 
\begin{align*}
    \mathbf{g}^*(\mathbf{x}_{s},s,t)&=\mathbf{x}_{s}-\frac{s}{s-t}\int_{t}^{s}\rvv^*(\rvx_\tau, \tau)\diff \tau\\
    &=\mathbf{x}_{s}-\frac{s}{s-t}\bigg[ (s-t)\rvv^*(\rvx_s, s) + \mathcal{O}((t-s)^{2}) \bigg]\\
    &=\rvx_s - s\rvv^*(\rvx_s,s)+\mathcal{O}(\vert t-s\vert).
\end{align*}}
From this proposition, we can conclude that
\begin{enumerate}
    \item Estimating $\rvg^*$ enables approximating not only the finite $s$-to-$t$ transition (for $s \ge t$) but also the \textit{infinitesimal} $s$-to-$s$ transition characterized by the instantaneous velocity $\rvv^*$.
    \item $\mathbf{g}^*(\mathbf{x}_{s},s,t)$  is interpreted as a denoiser-like quantity added with a residual term of the Taylor expansion.
\end{enumerate}
Therefore, by leveraging CTM's parameterization in
\Cref{eq:ctm_para_1}, learning
$\mathbf{G}_{\bm{\theta}} \approx \bPsi_{s\to t}$ for $t<s$
corresponds, up to the nonzero factor $(s-t)/s$, to learning
$\mathbf{g}_{\bm{\theta}} \approx \mathbf{g}^*$, enabling long-jump
capability through $\mathbf{G}_{\bm{\theta}}$. Moreover, the diagonal
value of $\mathbf{g}^*$ encodes the diffusion model's instantaneous
velocity through
\[
\rvv^*(\rvx_s,s)
=
\frac{\rvx_s-\rvg^*(\rvx_s,s,s)}{s},
\]
which can be directly supervised by the diffusion-model loss introduced
below. Thus, CTM's parameterization provides a single framework for
learning both finite-time transitions and local diffusion dynamics,
combining the long-jump capability of flow-map models with access to the
instantaneous dynamics of diffusion models.

In the next two sections, we first present CTM's consistency loss
(\Cref{subsec:ctm-consistency}), which supports both distillation and
training-from-scratch, and enforces the semigroup property to achieve
$\mathbf{G}_{\bm{\theta}}(\cdot, s, t) \approx \bPsi_{s\to t}(\cdot, s, t)$.
We then describe auxiliary losses (\Cref{subsec:ctm-auxiliary}) that arise
naturally from the parametrization in \Cref{eq:ctm_para_1}, including diffusion model loss and
GAN loss, which further improve CTM's performance significantly.

\subsection{Consistency Loss in CTM}\label{subsec:ctm-consistency}
CTM aims to approximate the oracle solution map
\[
\mathbf{G}_{\bm{\theta}}(\cdot, s, t) \approx \bPsi_{s\to t}(\cdot, s, t),
\]
for any $s \ge t$. Since the oracle $\bPsi_{s\to t}$ is usually not available in
closed form, CTM builds a feasible regression target by enforcing the
\emph{semigroup property} (\Cref{eq:semi-group}): for any $s \ge u \ge t$,
\[
\bPsi_{u\to t}\circ \bPsi_{s\to u} = \bPsi_{s\to t}.
\]
Depending on whether a pre-trained diffusion model is available,
the flow map $\bPsi_{s\to t}$ can be approximated in different ways.
Throughout, we assume $s \ge u \ge t \in [0,T]$.

\paragraph{Training via Distillation.}
Assume access to a pre-trained diffusion model producing
$\rvv_{\bm{\phi}^\times}(\rvx_s, s) \approx \rvv^*(\rvx_s, s)$. Then the
PF-ODE is approximated by the empirical dynamics
\begin{align}
\frac{\diff \rvx(\tau)}{\diff \tau}
= \rvv_{\bm{\phi}^\times}(\rvx_\tau, \tau).
\end{align}
CTM trains $\mathbf{G}_{\bm{\theta}}$ to match a numerical solver
$\mathtt{Solver}_{s\to t}(\rvx_s;\bm{\phi}^\times)$ applied to this empirical
ODE, which serves as a computable proxy for the oracle:
\[
\mathbf{G}_{\bm{\theta}}(\rvx_s, s, t)
\approx \mathtt{Solver}_{s\to t}(\rvx_s;\bm{\phi}^\times)
\approx \bPsi_{s\to t}(\rvx_s, s, t).
\]
With a strong teacher, the solver can recover $\bPsi_{s\to t}$ up to
discretization error, so the optimal student closely matches the ground truth
(see \citep{kim2023consistency}, Propositions 3 and 4).

\begin{figure}
    \centering
    \includegraphics[width=0.9\linewidth]{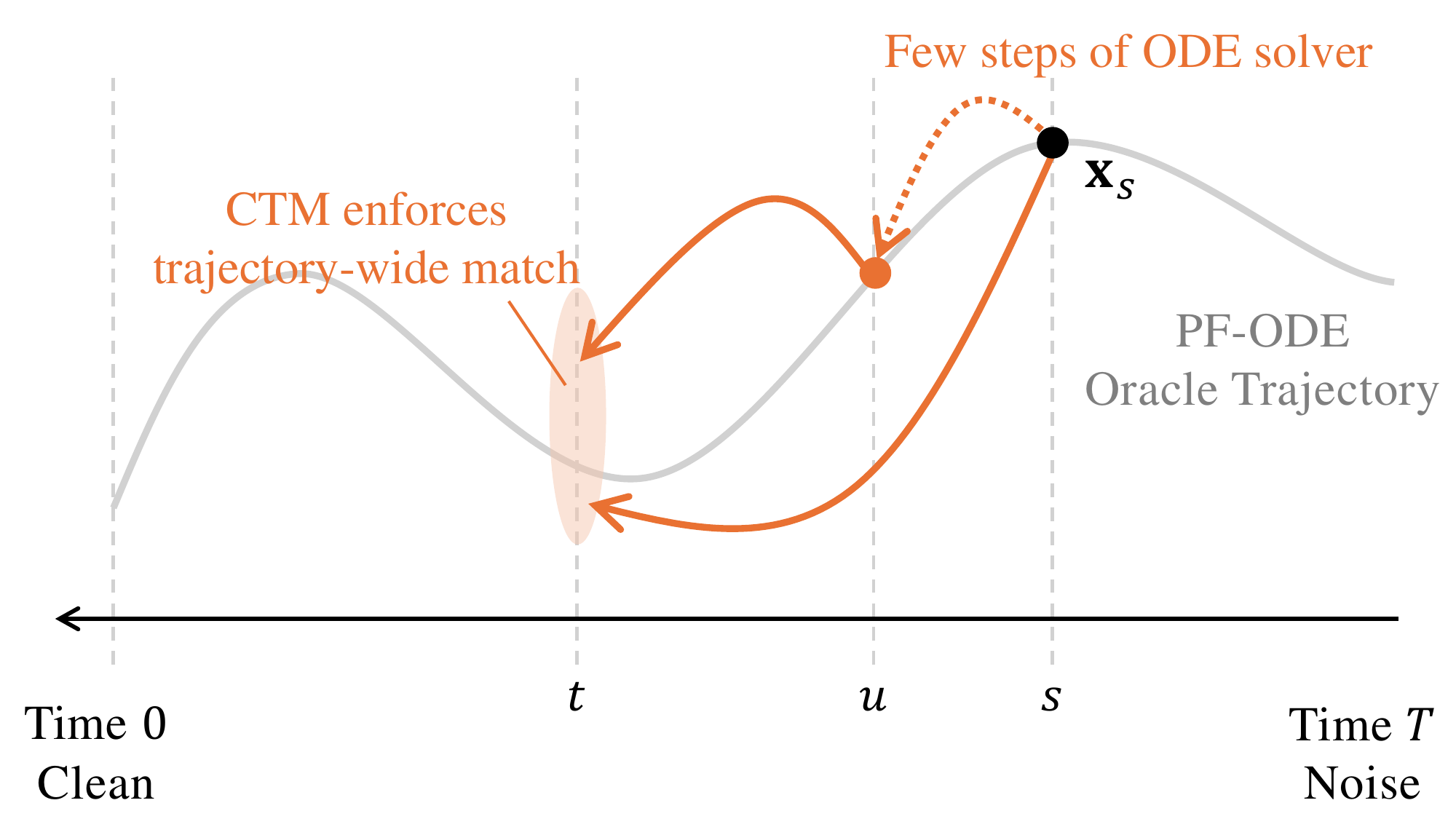}
    \caption{\textbfs{Illustration of CTM's semigroup property.} For any $s \ge u \ge t$, CTM enforces
    $\rmG_{\btheta}(\rvx_s, s,  t)
\approx
\rmG_{\btheta^-} \big(\mathtt{Solver}_{s\to u}(\rvx_s), u ,  t\big)$,
i.e., a short solver segment $s \to u$ followed by a CTM ``jump'' to $t$ matches the direct CTM map $s\to t$. The solver may be a pre-trained diffusion or a CTM's self-induced teacher. \figcredit{Adapted from \citet{kim2023consistency}.}}
    \label{fig:ill_models}
\end{figure}

However, solving across the full interval $[t,s]$ during training loop can be costly when $s$ and $t$ are far
apart. To improve efficiency and provide a smoother signal, CTM introduces
\emph{soft consistency matching}, which operationalizes the semigroup property.
As illustrated in \Cref{fig:ill_models}, CTM compares two predictions at time
$t$: the direct student output $\mathbf{G}_{\bm{\theta}}(\rvx_s, s, t)$, and a
mixed teacher–student path that first advances the teacher from $s$ to a random
$u \sim \mathcal{U}[t,s)$, then lets the student jump from $u$ to $t$:
\[
\mathbf{G}_{\bm{\theta}^-} \big(\mathtt{Solver}_{s\to u}(\rvx_s;\bm{\phi}^\times),\, u,\, t\big).
\]
The student is trained to match this composite prediction:
\begin{align}\label{eq:soft}
\underbrace{\mathbf{G}_{\bm{\theta}}(\rvx_s, s, t)}_{\approx\, \bPsi_{s\to t}(\rvx_s)}
 \approx 
\underbrace{\mathbf{G}_{\bm{\theta}^-} \big(\mathtt{Solver}_{s\to u}(\rvx_s;\bm{\phi}^\times), u, t\big)}_{\approx\, \bPsi_{u\to t}(\bPsi_{s\to u}(\rvx_s))},
\end{align}
where $\bm{\theta}^-$ is a stop gradient copy of $\mathbf{G}_{\bm{\theta}}$.

By varying $u$, CTM interpolates between global and local supervision:
\begin{itemize}
    \item \textbfs{Global Consistency} ($u=t$): the teacher is solved over the
    full interval $s\to t$, so the target reduces to the full teacher
    transition at time $t$.

    \item \textbfs{Local Consistency} ($u=s-\Delta s$): the student learns
    from a short teacher step near $s$; when $t=0$, this reduces to consistency
    distillation.
\end{itemize}

To reinforce sample quality while aligning trajectories, both predictions are
mapped to time $0$ by the stop gradient student and compared in a feature
space metric $d$:
\begin{align*}
\rvx_{\mathrm{est}}(\rvx_s,s,t) &:= \mathbf{G}_{\bm{\theta}^-} \big(\mathbf{G}_{\bm{\theta}}(\rvx_s,s,t),\, t,\, 0\big),\\
\rvx_{\mathrm{target}}(\rvx_s,s,u,t) &:= \mathbf{G}_{\bm{\theta}^-} \big(
\mathbf{G}_{\bm{\theta}^-}(\mathtt{Solver}_{s\to u}(\rvx_s;\bm{\phi}^\times), u, t),\, t,\, 0\big).
\end{align*}
The CTM consistency loss is
\begin{mdframed}
    \begin{align}\label{eq:ctm-consist-distill}
\begin{aligned}
\mathcal{L}_{\mathrm{consist}}(\bm{\theta}; \bm{\phi}^\times)
:=
\mathbb{E}_{s \in [0,T]}\mathbb{E}_{t \in [0,s]}\mathbb{E}_{u \in [t,s)}
\mathbb{E}_{\rvx_0}\mathbb{E}_{\rvx_s |\rvx_0}
\Big[ d\big(\rvx_{\mathrm{est}}, \rvx_{\mathrm{target}}\big) \Big],
\end{aligned}
\end{align}
\end{mdframed}
which encourages the student to match the empirical PF-ODE solution while
preserving generation quality.

\paragraph{Training from Scratch.}
Leveraging CTM's special parameterization (Proposition~\ref{ctm-g-para}(i)), 
\[
\rvg^*(\rvx_\tau, \tau, \tau)
= \rvx_\tau - \tau\,\rvv^*(\rvx_\tau,\tau)
 \implies 
\rvv^*(\rvx_\tau,\tau)
= \frac{\rvx_\tau - \rvg^*(\rvx_\tau, \tau, \tau)}{\tau}.
\]
We can therefore replace the oracle residual function
$\rvg^*(\cdot,\tau,\tau)$ with CTM's own estimate
$\rvg_{\bm{\theta}^-}(\cdot,\tau,\tau)$ for $\tau \in (0,T]$,
which yields a self-induced empirical PF-ODE:
\begin{align}\label{eq:emp-pf-ode-ctm}
\frac{\diff \rvx(\tau)}{\diff \tau}
= \frac{\rvx(\tau) - \rvg_{\bm{\theta}^-} \left(\rvx(\tau),\tau,\tau\right)}{\tau}.
\end{align}

We then solve this self-induced ODE to construct a tractable transition target:
\[
\mathbf{G}_{\bm{\theta}}(\rvx_s, s, t)
\approx \mathtt{Solver}_{s\to t}(\rvx_s;\bm{\theta}^-).
\]
This target approaches the oracle flow map only insofar as the diagonal estimate
$\rvg_{\bm{\theta}^-}(\cdot,\tau,\tau)$ accurately recovers the target local
dynamics. The diffusion-model loss introduced below provides direct supervision
for this local quantity.

As in the distillation case \Cref{eq:soft}, full integration over $[t,s]$ can be
costly when $s$ and $t$ are far apart. CTM therefore enforces the semigroup
property to obtain a shorter supervision path:
\[
\underbrace{\mathbf{G}_{\bm{\theta}}(\rvx_s, s, t)}_{\approx\, \bPsi_{s\to t}(\rvx_s)}
 \approx 
\underbrace{\mathbf{G}_{\bm{\theta}^-} \big(\mathtt{Solver}_{s\to u}(\rvx_s;\bm{\theta}^-),\, u,\, t\big)}_{\approx\, \bPsi_{u\to t}(\bPsi_{s\to u}(\rvx_s))},
\]
where $u \sim \mathcal{U}[t,s)$ and $\bm{\theta}^-$ is a stop gradient copy of
the student. The only change from distillation is that the external teacher
$\rvv_{\bm{\phi}^\times}$ is replaced by the self-induced teacher
$\rvg_{\bm{\theta}^-}$.

To couple trajectory matching with sample quality, both predictions are mapped
to time $0$ using the stop gradient student and compared in feature space. The
target without any pre-trained model is
\[
\hat{\rvx}_{\mathrm{target}}
:= \mathbf{G}_{\bm{\theta}^-} \big(
\mathbf{G}_{\bm{\theta}^-}(\mathtt{Solver}_{s\to u}(\rvx_s;\bm{\theta}^-),\, u,\, t),\, t,\, 0\big),
\]
which replaces $\rvx_{\mathrm{target}}$ in \Cref{eq:ctm-consist-distill}, and leads to:
\begin{mdframed}
    \begin{align}\label{eq:ctm-consist-scratch}
\begin{aligned}
\mathcal{L}_{\mathrm{consist}}(\bm{\theta}; \btheta^-)
:=
\mathbb{E}_{s \in [0,T]}\mathbb{E}_{t \in [0,s]}\mathbb{E}_{u \in [t,s)}
\mathbb{E}_{\rvx_0}\mathbb{E}_{\rvx_s |\rvx_0}
\Big[ d\big(\rvx_{\mathrm{est}}, \hat\rvx_{\mathrm{target}}\big) \Big],
\end{aligned}
\end{align}
\end{mdframed}
Conceptually, this is
self-distillation within CTM: the model supplies its own short horizon teacher
signals while the student learns the full transition.

\subsection{Auxiliary Losses in CTM}\label{subsec:ctm-auxiliary}

(Self-)distillation can underperform the teacher because it optimizes only teacher generated targets, lacking direct supervision from real data. By contrast, CTM can naturally incorporate data driven regularizers, for example by augmenting its objective with denoising score matching and an adversarial (GAN) term~\citep{goodfellow2014generative}, to better learn the flow map. 

\paragraph{Natural Integration of Diffusion Loss.}
The diffusion–model loss (more precisely, the conditional flow matching loss; see \Cref{eq:fm-cfm}) integrates naturally into CTM and provides a fixed regression target that facilitates the learning of the flow map model. To see this, note that we have
\[
\rvv^*(\rvx_s, s)
= \frac{\rvx_s - \rvg^*(\rvx_s, s, s)}{s},
\qquad 
\rvg^*(\rvx_s, s, s) \approx \rvg_{\btheta}(\rvx_s, s, s).
\]
This naturally induces a velocity parametrization through $\rvg_{\btheta}$:
\[
\rvv_{\btheta}(\rvx_s, s)
:= \frac{1}{s}\bigl(\rvx_s - \rvg_{\btheta}(\rvx_s, s, s)\bigr).
\]
Using the linear Gaussian path
\[
\rvx_s = \alpha_s \rvx_0 + \sigma_s \beps,
\qquad
\rvx_0 \sim p_{\mathrm{data}}, 
\beps \sim \mathcal N(0, \mathbf I),
\]
the diffusion model loss can be written as
\begin{align}\label{eq:ctm-dsm}
\mathcal L_{\mathrm{DM}}(\btheta)
&:= 
\E_{\rvx_0, \beps, s}
\Bigl[
w(s)
\bigl\|
 \rvv_{\btheta}(\rvx_s, s) - \left(\alpha_s' \rvx_0 + \sigma_s' \beps\right)
\bigr\|_2^2
\Bigr].
\end{align}

$\mathcal L_{\mathrm{DM}}$ improves accuracy when $t$ is close to $s$ by explicitly supervising small
jumps along the trajectory. In this regime, the factor $1-\tfrac{t}{s}$ in
\Cref{eq:ctm_para_1} approaches zero, which can weaken gradients and slow
learning; $\mathcal L_{\mathrm{DM}}$ supplies a stronger local signal and stabilizes training.  

Conceptually, \Cref{eq:ctm-consist-distill,eq:ctm-consist-scratch} enforce
trajectory matching (zeroth order), while \Cref{eq:ctm-dsm} enforces slope
matching (first order).

\paragraph{(Optional) GAN Loss.}
While consistency and diffusion model loss provide strong regression signals, they can yield
overly smooth outputs. CTM therefore optionally adds an adversarial term to
encourage sharper, more realistic samples by aligning the generator
distribution with the data distribution. With a discriminator
$D_{\bm{\zeta}}$ that distinguishes real $\rvx_{0}\sim p_{\mathrm{data}}$ from
generated $\rvx_{\mathrm{est}}(\rvx_{s},s,t)$, the objective is
\begin{align*} 
&\mathcal{L}_{\mathrm{GAN}}(\bm{\theta}, \bm{\zeta}) := \\ &\mathbb{E}_{\mathbf{x}_0} \big[\log D_{\bm{\zeta}}(\mathbf{x}_0)\big] + \mathbb{E}_{s \in [0, T]} \mathbb{E}_{t \in [0, s]} \mathbb{E}_{\mathbf{x}_0} \mathbb{E}_{\mathbf{x}_s \vert \mathbf{x}_0} \big[\log (1 - D_{\bm{\zeta}}( \mathbf{x}_{\mathrm{est}}(\mathbf{x}_s, s, t)))\big], \end{align*}
where $D_{\bm{\zeta}}$ is maximized and $\rmG_{\bm{\theta}}$ is minimized.
Intuitively, the discriminator acts as an adaptive perceptual distance that
encourages realistic detail. Theoretically, the GAN term drives distributional
matching (Jensen–Shannon divergence) between $p_{\mathrm{data}}$ and the model
distribution induced by $\rmG_{\bm{\theta}}$~\citep{goodfellow2014generative},
which can raise fidelity beyond the teacher.

\paragraph{Overall CTM Objective.}
In summary, CTM unifies (self-)distillation, diffusion, and GAN losses into a single training framework:
\begin{mdframed}
    \begin{align*}
\mathcal{L}_{\mathrm{CTM}}(\bm{\theta},\bm{\zeta})
:=  \mathcal{L}_{\mathrm{consist}}(\bm{\theta};  \bm{\phi}^\times/\bm{\theta}^-)
+ \lambda_{\mathrm{DM}} \mathcal{L}_{\mathrm{DM}}(\bm{\theta})
+ \lambda_{\mathrm{GAN}} \mathcal{L}_{\mathrm{GAN}}(\bm{\theta},\bm{\zeta}),
\end{align*}
\end{mdframed}
where the teacher is either an external pre-trained model $\bm{\phi}^\times$ or
the self-induced teacher $\bm{\theta}^-$. The regression style components
$\mathcal{L}_{\mathrm{consist}}$ and $\mathcal{L}_{\mathrm{DM}}$ act as strong
regularizers, while the optional GAN term improves fine scale detail without
sacrificing stability~\citep{kim2024pagoda}.

\subsection{Flexible Sampling with CTM}
CTM learns the general flow map $\bPsi_{s\to t}$ for any $s>t$, which means it supports
anytime to anytime transitions. This property enables flexible sampling strategies. 
For example, CTM proposes \emph{$\gamma$ sampling}, where the hyperparameter $\gamma$
controls the stochasticity during generation.
In addition, CTM can reuse standard inference techniques developed for diffusion models,
such as ODE based solvers and exact likelihood computation.

In what follows, we fix a discrete time grid for sampling $T=\tau_0 > \tau_1 > \tau_2 > \cdots > \tau_{M}=0$.

\begin{algorithm}[H]
    \centering
    \caption{CTM's $\gamma$-sampling}\label{alg:gamma}
    \begin{algorithmic}[1]
    \Require Trained CTM $\mathbf{G}_{\bm{\theta}^\times}$, $\gamma\in[0,1]$, $T=\tau_0>\tau_1 > \tau_2 > \cdots > \tau_{M}=0$.
    \State Start from $\mathbf{x}_{\tau_{0}}\sim p_{\mathrm{prior}}=\mathcal{N}(\bm{0}, T^2\rmI)$
        \For{$n=0$ to $M-1$}
        \State $\tilde{\tau}_{n+1}\leftarrow \sqrt{1-\gamma^{2}}\tau_{n+1}$
        \State Denoise $\mathbf{x}_{\tilde{\tau}_{n+1}}\leftarrow \mathbf{G}_{\bm{\theta}^\times}(\mathbf{x}_{\tau_{n}},\tau_{n},\tilde{\tau}_{n+1})$
        \State Diffuse $\mathbf{x}_{\tau_{n+1}}\leftarrow\mathbf{x}_{\tilde{\tau}_{n+1}}+\gamma \tau_{n+1}\bm{\epsilon}$, where $\bm{\epsilon}\sim\mathcal{N}(\bm{0}, \rmI)$
        \EndFor
        \Ensure  $\mathbf{x}_{\tau_{M}}$
    \end{algorithmic}
\end{algorithm}

\paragraph{Methodology of $\gamma$-Sampling.}
CTM's $\gamma$-sampling introduces a unified family of samplers that arises naturally from learning a general flow map model.  
It encompasses prior approaches, such as CM's multistep sampling (see \Cref{alg:cm_sampling}) 
and time-stepping-style sampling, which is conceptually similar to ODE solvers.  
The parameter $\gamma$ directly controls the degree of semantic change during generation, 
making $\gamma$ sampling a flexible and task aware strategy for diverse downstream applications.

\begin{figure}[t]
    \centering
    \includegraphics[width=\linewidth]{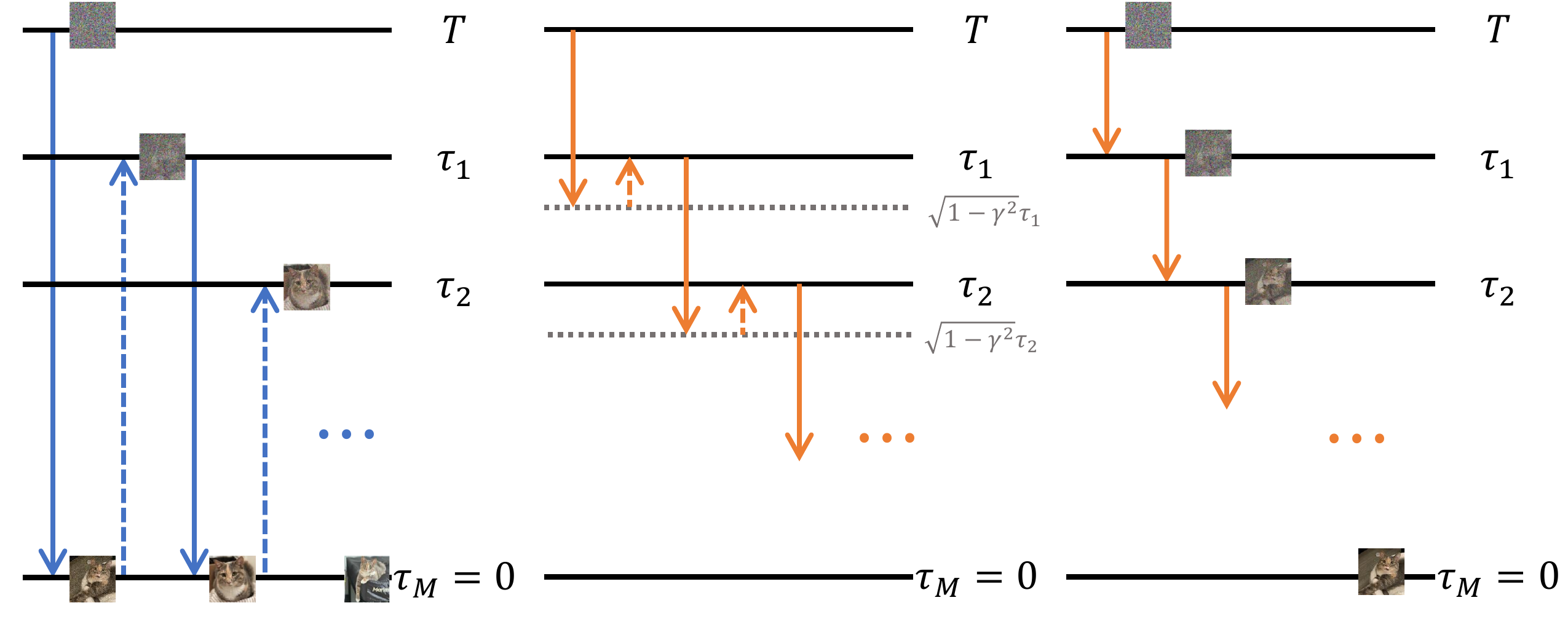}
    \caption{\textbfs{Illustration of $\gamma$-sampling with varying $\gamma$ value.} The procedure alternates between denoising with a network evaluation and adding noise in reverse,   $(\tau_{n}\xrightarrow{\text{Denoise}} \sqrt{1-\gamma^{2}}\tau_{n+1}\xrightarrow{\text{Noisify}} \tau_{n+1})_{n=0}^{M-1}$.
The leftmost panel illustrates $\gamma=1$, corresponding to the fully stochastic case.  
The rightmost panel shows $\gamma=0$, corresponding to the fully deterministic case.  
The middle panel depicts intermediate values $\gamma\in(0,1)$, which interpolate between these two extremes. 
\figcredit{Adapted from \citet{kim2023consistency}.}}
    \label{fig:gamma_sampling}
\end{figure}

\begin{itemize}
\item \Cref{fig:gamma_sampling}-(Left): When $\gamma=1$, it coincides with the multistep sampling introduced in CM (i.e., a special flow map $\bPsi_{s\to 0}$), which is fully stochastic and results in semantic variation when the number of steps changes.

\item \Cref{fig:gamma_sampling}-(Right): When $\gamma=0$, it reduces to fully deterministic time-stepping, which estimates the solution trajectory of the PF-ODE. A key distinction between $\gamma$ sampling with $\gamma=0$ and conventional time-stepping ODE-based sampling is that CTM avoids the discretization errors of numerical solvers.

\item \Cref{fig:gamma_sampling}-(Middle): When $0 < \gamma < 1$, $\gamma$-sampling interpolates between the two extremes by allowing a controlled amount of stochasticity to be injected during sampling. 
\end{itemize}
We highlight that values $\gamma\in[0,1)$ require the general flow map
$\bPsi_{s\to t}$, whereas the special case $\gamma=1$ reduces to
CM-style multistep sampling and requires only $\bPsi_{s\to0}$.

\paragraph{Analysis of $\gamma$-Sampling.} CTM empirically observed that CM's multistep sampling degrades in quality once the number of steps $M \geq 4$.  
To explain this phenomenon, CTM analyzed the underlying cause: when $\gamma \neq 0$, each neural ``jump'' introduces a small mismatch, and these mismatches accumulate as the model iteratively maps states toward time zero. This error accumulation explains why long multi-step runs can perform poorly. We formalize this idea in the following proposition.

\proppp{(Informal) 2-step $\gamma$-sampling}{ctm-gamma}{
\label{th:sampling_agg_error_2_steps}
Let $\tau\in(0,T)$ and $\gamma \in [0,1]$. Let $p_{\bm{\theta}^{*},2}$ denote as the density obtained from the $\gamma$-sampler with the optimal CTM, following the transition sequence $
    T\rightarrow\sqrt{1-\gamma^2}\tau\rightarrow \tau \rightarrow 0
$, starting from $p_T$. Then
 \begin{align*}
    \mathcal D_{\mathrm{TV}}\big( p_{\mathrm{data}}, p_{\bm{\theta}^{*},2} \big)         = \mathcal{O}\left(\sqrt{T-\sqrt{1-\gamma^2}\tau+ \tau}  \right).
 \end{align*}
 Here, $\mathcal{D}_{\mathrm{TV}}$ denotes the total variation  between distributions (see \Cref{eq:f-div}).
}{
We refer the reader to Theorem 8 of~\citet{kim2023consistency} for the general case when the number of sampling steps is $M$. }
The insights from the above theorem can be summarized as follows:
\begin{itemize}
    \item \textbfs{When} $\gamma = 1$ (corresponding to CM's multistep sampling)\textbfs{:} The method performs iterative long-range transitions from $\tau_n$ to $0$ at each step $n$. This leads to error accumulation on the order of
    \begin{align*}
        \mathcal{O}\left(\sqrt{T + \tau_1 + \cdots + \tau_M}\right).
    \end{align*}
    
    \item \textbfs{When} $\gamma = 0$ (corresponding to CTM's deterministic multistep sampling)\textbfs{:} Such temporal overlap between transitions is eliminated. This avoids error accumulation and yields a tighter bound of $\mathcal{O}(\sqrt{T})$. Empirically, CTM with $\gamma = 0$ provides a favorable trade-off between sampling speed and sample quality: increasing the number of sampling steps improves generation quality without introducing instability.
\end{itemize}

\paragraph{CTM Supports Diffusion Inference.} Since CTM learns the score function (or denoiser) directly through $\rvg_{\bm{\theta}}$, 
thanks to its parametrization in \Cref{eq:ctm_para_1}, it is compatible with inference techniques 
originally developed for diffusion models. For instance, one can compute exact likelihoods 
(\Cref{subsec:sde-sample}) or apply advanced samplers such as DDIM or DPM (\Cref{ch:solvers}) 
for generation, by using $\rvg_{\bm{\theta}}(\cdot, s, s)$.

\newpage

\section{\texorpdfstring{General Flow Map: Mean Flow}{General Flow Map: Mean Flow}}\label{sec:MF}

Just as diffusion models admit many equivalent parameterizations and training objectives, a general flow map $\bPsi_{s\to t}$ can also be learned in multiple plausible ways.  
In this section, we introduce \emph{Mean Flow} (MF)~\citep{geng2025mean}, a later representative of the general flow map family $\bPsi_{s\to t}$ that illustrates an alternative yet principled perspective on how such maps can be effectively learned.

\subsection{Modeling and Training of Mean Flow}

In contrast to CM and CTM, which build on the EDM framework, MF is based on the 
flow matching formulation ($\alpha_t = 1-t$ and $\sigma_t  = t$ for $t\in[0,1]$). 
Rather than directly parameterizing the flow map, MF learns the \emph{average drift}
over an interval $[t,s]$ (with $t<s$):
\[
\rvh_{\btheta}(\rvx_s, s, t)  \approx 
\rvh^*(\rvx_s, s, t) := \frac{1}{t-s}\int_s^t \rvv^*(\rvx_u,u) \diff u.
\]
The corresponding oracle loss is
\begin{align}\label{eq:mf-oracle}
\mathbb{E}_{t<s} \mathbb{E}_{\rvx_s\sim p_s}\Big[w(s) 
\|\rvh_{\btheta}(\rvx_s,s,t) - \rvh^*(\rvx_s,s,t)\|_2^2\Big].
\end{align}
In particular, when $s \to t$, the loss function reduces to the flow matching loss:
\begin{align}\label{eq:mf-to-fm}
    \mathbb{E}_{t} \mathbb{E}_{\rvx_t\sim p_t}\Big[w(t) 
\|\rvh_{\btheta}(\rvx_t,t,t) - \rvv^*(\rvx_t,t)\|_2^2\Big],
\end{align}
learning the instantaneous velocity.
We will see later in \Cref{subsec:equivalence-ctm-mf} that MF remains consistent with the 
general objective in \Cref{eq:general-cm-loss:rep}, 
but approaches it from a different (while equivalent) perspective.
Since the oracle regression target $\rvh^*(\rvx_s,s,t)$ does not admit a closed form in general, MF constructs a surrogate by exploiting an identity obtained from differentiating
\[
(t-s)\,\rvh^*(\rvx_s,s,t) = \int_s^t \rvv^*(\rvx_u,u)\,\diff u
\]
with respect to $s$. This yields the \emph{MF identity}:
\begin{align}\label{eq:mf-identity}
\begin{aligned}
    \rvh^*(\rvx_s,s,t)
&= \rvv^*(\rvx_s,s) - (s-t) \tfrac{\diff}{\diff s}\rvh^*(\rvx_s,s,t) \\
&= \rvv^*(\rvx_s,s) - (s-t)\Big[(\partial_{\rvx}\rvh^*)(\rvx_s,s,t) \rvv^*(\rvx_s,s) 
+ \partial_s \rvh^*(\rvx_s,s,t)\Big],
\end{aligned}
\end{align}
where the second line applies the chain rule together with 
\[
\frac{\diff}{\diff s}\rvx_s = \rvv^*(\rvx_s,s).
\]

Motivated by this identity, MF replaces the intractable oracle with a stop-gradient surrogate, leading to the practical training objective
\begin{mdframed}
\begin{align}\label{eq:mf-practice}
    \mathcal{L}_{\mathrm{MF}}(\btheta)
:= \mathbb{E}_{t<s} \mathbb{E}_{\rvx_s\sim p_s}\Big[w(s)\,
\|\rvh_{\btheta}(\rvx_s,s,t) - \rvh_{\btheta^-}^{\mathrm{tgt}}(\rvx_s,s,t)\|_2^2\Big],
\end{align}
\end{mdframed}
where the regression target is defined as
\begin{align*}
    \rvh_{\btheta^-}^{\mathrm{tgt}}&(\rvx_s,s,t)
:= \\ &\rvv^*(\rvx_s,s) - (s-t)\Big[\underbrace{(\partial_{\rvx}\rvh_{\btheta^-})(\rvx_s,s,t) \rvv^*(\rvx_s,s)
+ \partial_s \rvh_{\btheta^-}(\rvx_s,s,t)}_{\mathrm{JVP}}\Big].
\end{align*}

In practice, the oracle velocity $\rvv^*$ cannot be computed in closed form and
must instead be approximated. Two common strategies are available: relying on a
pre-trained diffusion model (\emph{distillation}) or constructing a direct
estimator from data (\emph{training from scratch}). Regardless of the choice, one ultimately needs to compute a
\emph{Jacobian--vector product}  (JVP) of the target network
$\rvh_{\btheta^-}$:
\[ 
\left[\partial_\rvx \rvh_{\btheta^-}, \partial_s \rvh_{\btheta^-}, \partial_t \rvh_{\btheta^-}\right]^\top \cdot \left[\rvv^*, 1, 0\right]
\]

\paragraph{Distillation.}
Use a pre-trained diffusion model with a flow matching backbone,
$\rvv_{\bphi^\times} \approx \rvv^*$.

\paragraph{Training from scratch.}
Use the one point conditional velocity $\alpha_s'\rvx_0 + \sigma_s'\beps$,
obtained from the forward noise injection $\rvx_s=\alpha_s\rvx_0+\sigma_s\beps$ with
$\beps\sim\mathcal N(\bm{0},\rmI)$. This gives an unbiased single sample estimate of the
instantaneous drift at level $s$ when evaluated at paired $(\rvx_0,\beps)$.


\subsection{Sampling of Mean Flow}
Once a MF $\rvh_{\btheta^\times}$ is trained, it naturally recovers a proxy of the flow
map. For any starting point $\rvx_s$, the map from $s$ to $t$ is (approximately) given by
\[
\bPsi_{s\to t}(\rvx_s) = \rvx_s + (t-s)\,\rvh^*(\rvx_s,s,t)
  \approx   \rvx_s + (t-s)\,\rvh_{\btheta^\times}(\rvx_s,s,t).
\]

This enables both one-step and multi-step sampling. For example, drawing
$\rvx_T\sim p_{\mathrm{prior}}$, the one-step generation of a clean sample is
\[
\rvx_0  \leftarrow  \rvx_T - T\,\rvh_{\btheta^\times}(\rvx_T,T,0).
\]

Alternatively, multi-step generation can be performed by preparing a time grid
and applying the map sequentially, in the same time-stepping manner used in
CTM. Since MF learns a general flow map, it also supports $\gamma$-sampling
as in CTM, where a controllable hyperparameter $\gamma$ injects stochasticity
into the sampling process.

\subsection{Relationship between CTM and MF}\label{subsec:equivalence-ctm-mf}
At first sight CTM and MF may appear unrelated.  
In fact, both provide different parameterizations of the same oracle flow
map $\bPsi_{s\to t}$, and their corresponding squared oracle regression
errors are related by a time-dependent weighting~\citep{hu2025cmt}.

\paragraph{Relationship of Parameterizations.}
Both methods operate under the same general framework but represent the learned function in distinct ways.  
The flow map can be written equivalently as
\begin{align*}
\bPsi_{s\to t}(\rvx_s)
& = \rvx_s + \int_s^t \rvv^*(\rvx_u, u)\diff u \\
&= \frac{t}{s}\rvx_s
   + \frac{s-t}{s}\underbrace{\Bigg[\rvx_s + \frac{s}{s-t}\int_s^t \rvv^*(\rvx_u,u)\,\diff u\Bigg]}_{\approx\,\rvg_\btheta} \\
&= \rvx_s + (t-s)\underbrace{\Bigg[\frac{1}{t-s}\int_s^t \rvv^*(\rvx_u,u)\,\diff u\Bigg]}_{\approx\,\rvh_\btheta}.
\end{align*}
Here, the first is the definition of the flow map, the second form highlights the CTM parametrization through $\rvg_\btheta$ (see \Cref{eq:ctm-g-oracle,eq:ctm_para_1}), while the last highlights the MF parametrization through $\rvh_\btheta$.

\paragraph{Relationship of Training Loss.} Given the above reinterpretation of the oracle flow map $\bPsi_{s\to t}$ in terms of the CTM parametrization 
\[
\rvg_\btheta(\rvx_s,s,t) \approx \rvg^*(\rvx_s,s,t):=\rvx_s + \frac{s}{s-t}\int_s^t \rvv^*(\rvx_u,u) \diff u
\]
 and the MF parametrization 
\[
\rvh_\btheta(\rvx_s,s,t) \approx \rvh^*(\rvx_s,s,t):=\frac{1}{t-s}\int_s^t \rvv^*(\rvx_u,u) \diff u,
\] 
we now show that the training losses of CTM and MF are in fact equivalent.

Consider the relation
\[
\rvg_\btheta(\rvx_s,s,t) := \rvx_s - s \rvh_\btheta(\rvx_s,s,t),
\]
and take $d(\rvx,\rvy):=\|\rvx-\rvy\|^2$ as an example. Substituting into \Cref{eq:general-cm-loss:rep} and viewing $\rmG_\btheta$ as CTM’s flow-map parameterization (\Cref{eq:ctm_para_1}) gives
\begin{align} &~d\big(\rmG_\btheta(\rvx_s,s,t), \bPsi_{s\to t}(\rvx_s)\big) \notag \\ = &~\norm{\rmG_\btheta(\rvx_s,s,t) - \bPsi_{s\to t}(\rvx_s)}^2 \notag \\ = &~\norm{\left(\frac{t}{s}\rvx_s + \frac{s-t}{s} \rvg_\btheta(\rvx_s,s,t)\right)-\left(\frac{t}{s}\rvx_s + \frac{s-t}{s}\Big[\rvx_s + \frac{s}{s-t}\int_s^t \rvv^*(\rvx_u,u) \diff u\Big]\right)}^2 \notag \\ = &~\left(\frac{s-t}{s}\right)^2\norm{\rvg_\btheta(\rvx_s,s,t)-\left(\rvx_s + \frac{s}{s-t}\int_s^t \rvv^*(\rvx_u,u) \diff u\right)}^2 \label{eq:g-para} \\ = &~\left(\frac{s-t}{s}\right)^2\norm{\left(\rvx_s - s \rvh_\btheta(\rvx_s,s,t)\right)-\left(\rvx_s + \frac{s}{s-t}\int_s^t \rvv^*(\rvx_u,u) \diff u\right)}^2 \notag \\ = &~\left(\frac{s-t}{s}\right)^2\norm{\left(\rvx_s - s \rvh_\btheta(\rvx_s,s,t)\right)-\left(\rvx_s + \frac{s}{s-t}\int_s^t \rvv^*(\rvx_u,u) \diff u\right)}^2 \notag \\ = &~\left(s-t\right)^2\norm{ \rvh_\btheta(\rvx_s,s,t)- \left(\frac{1}{t-s}\int_s^t \rvv^*(\rvx_u,u)\diff u \right)}^2 \label{eq:h-para} \end{align}
Hence,
\begin{align*} 
\frac{1}{s^2} \Bigg\|\rvg_\btheta(\rvx_s,s,t)-\rvg^*(\rvx_s,s,t) \Bigg\|^2 = \Bigg\|\rvh_\btheta(\rvx_s,s,t)-\rvh^*(\rvx_s,s,t) \Bigg\|^2. 
\end{align*}
Putting aside the specific algorithm used to learn the flow map, the CTM and MF losses share the same mathematical form, differing only by a weighting function. Moreover, setting $t=0$ in either case recovers the CM setting ($\bPsi_{s\to 0}$), where
each state maps directly to the clean data.

\paragraph{Auxiliary Loss in Practice.}
In CTM, training is performed with the consistency loss in \Cref{eq:ctm-consist-scratch} jointly with its self-defined diffusion model loss in \Cref{eq:ctm-dsm}. A similar strategy is adopted in MF. As shown in \Cref{eq:mf-to-fm}, when $s \to t$, the MF loss reduces to the standard flow matching objective. In practice, MF controls the ratio between pairs with $s \neq t$ and those with $s = t$; consequently, the overall optimization becomes a mixture of the MF objective in \Cref{eq:mf-practice} and the flow matching objective in \Cref{eq:mf-to-fm}. 

Both parametrizations are able to provide a smooth transition from diffusion-model training, which learns instantaneous velocity with a fixed regression target, to flow-map learning, which employs a stop-gradient pseudo-regression target.

\paragraph{Both CTM and MF Parameterizations Enable Flexible Inference.}
Both CTM ($\rmG_\btheta(\rvx_s,s,t)$) and MF ($\rvh_\btheta(\rvx_s,s,t)$) aim to approximate the underlying flow map $\bPsi_{s\to t}$:
\[
\rmG_\btheta(\rvx_s,s,t)\approx\bPsi_{s\to t},\quad  \text{and}
\quad 
\rvx_s+(t-s)\rvh_\btheta(\rvx_s,s,t)\approx\bPsi_{s\to t}.
\]
Since both models learn an explicit mapping between any two time steps, they naturally support CTM’s $\gamma$-sampling and remain compatible with inference techniques originally developed for diffusion models, such as guidance (\Cref{ch:guidance}), exact likelihood computation (\Cref{eq:score-sde-likelihood}), and accelerated sampling with higher-order solvers (\Cref{ch:solvers}). 
This compatibility arises because their parameterizations recover the instantaneous diffusion drift in the infinitesimal limit $t\to s$:
\[
\rvg^*(\rvx_s,s,s)=\rvx_s-s\,\rvv^*(\rvx_s,s),
\quad  \text{and}
\quad 
\rvh^*(\rvx_s,s,s)=\rvv^*(\rvx_s,s).
\]
This property is not shared by specialized flow map formulations $\bPsi_{s\to0}$, such as those in the CM family. Thus, both CTM and MF can be regarded as flexible and general flow map formulations that generalize diffusion-based inference to direct time-to-time mappings.

\paragraph{Conclusion.}

The relationship between CTM and MF is analogous to the relationship between
different parametrizations of diffusion models
(\Cref{sec:equivalent-parametrizations}): at the oracle level, they encode the
same underlying flow map. Under squared-error regression and the parametrization
relation above, their oracle regression errors differ only by a known
time-dependent scaling. Their practical training objectives, however, remain
different because they use different surrogate targets and optimization
procedures.

Although CTM and MF are related in this way, they differ conceptually and algorithmically in how the target is learned. The CTM target is defined in the ``integral'' form, as it acts directly on the flow map. The MF target, on the other hand, is defined in the ``differential'' form, as it relies on finite differences of the flow map. Because a target in the integral form is not directly tractable, CTM must solve an ODE during training. In contrast, MF leverages the MF identity (see \Cref{eq:mf-identity}) to construct a tractable target in the differential form. This relationship is analogous to that between energy-based and score-based methods: the former learns a probability density itself (in the integral form), while the latter matches the gradient of that density (in the differential form). In practice, CTM and MF may behave differently due to factors such as loss weighting, network architecture, and optimization dynamics, which can cause one method to perform better than the other under certain conditions.

This perspective suggests that CTM and MF are not the only viable formulations. Other parameterizations of the flow map could also enable efficient and stable training, potentially leading to new standalone generative models. Exploring these alternatives may further enrich the landscape of diffusion models and their flow map extensions, pushing the boundaries of few-step generation.

\clearpage
\newpage

\section{Closing Remarks}\label{sec:ch11_cr}

This final chapter has brought our exploration full circle: from slow iterative
diffusion samplers to fast few-step generators (learned from scratch). The common object behind the methods in this chapter is the oracle flow map
introduced in \Cref{eq:general-cm-loss:rep}:
\[
\bPsi_{s\to t}(\rvx_s)
=
\rvx_s+\int_s^t \rvv^*(\rvx_\tau,\tau)\,\diff \tau.
\]
Learning this map replaces many small numerical updates by one or a few learned
jumps.

At this point, it is useful to step back from the individual algorithms.
Knowledge Distillation (\Cref{sec:distillation-prologue}), Progressive
Distillation (\Cref{sec:PD}), CM (\Cref{sec:discrete-cm}), CTM
(\Cref{sec:CTM}), and MF (\Cref{sec:MF}) were introduced through
different training constructions, but they all exploit, explicitly or
implicitly, the same ODE solution-map structure.

The flow map itself is a classical object from differential equations and
dynamical systems, with closely related particle-based and field-based
descriptions in continuum and fluid mechanics; see
\Cref{eq:flow-map-def} and the surrounding historical discussion.
Here, we use three classical descriptions of the same ODE flow, namely
the \emph{semigroup}, \emph{Lagrangian}, and \emph{Eulerian} views, as an
organizing lens for the modern learning methods developed above%
\footnote{In modern generative modeling, \citet{boffi2024flow} bring together
structural ideas that already appear across PD, CM, CTM, and MF, and use them to
describe the learning of diffusion PF-ODE flow maps in a common language. The
main ingredients therefore largely come from these existing lines of work. Our
presentation and illustrations in this part are also informed by the exposition
of \citet{dieleman2026flowmaps}.}

\paragraph{Semigroup View: Split the Trip.}
As introduced in \Cref{eq:semi-group}, the finite-step identity is
\[
\bPsi_{u\to t}\big(\bPsi_{s\to u}(\rvx_s)\big)
=
\bPsi_{s\to t}(\rvx_s).
\]
As illustrated in \Cref{fig:semigroup}, the intuition is the same as traveling along a fixed route: going from $s$ to
$u$ and then from $u$ to $t$ lands at the same point as going directly from
$s$ to $t$.  Within this view, KD learns the full transition $\bPsi_{T\to0}$, PD
compresses two adjacent teacher steps into one student step, CM learns the
special anytime-to-clean map $\bPsi_{s\to0}$, and CTM extends the same
composition principle to the general map $\bPsi_{s\to t}$.

\begin{figure}[t]
    \centering

    \begin{subfigure}[t]{0.485\textwidth}
        \centering
        \includegraphics[width=\linewidth]{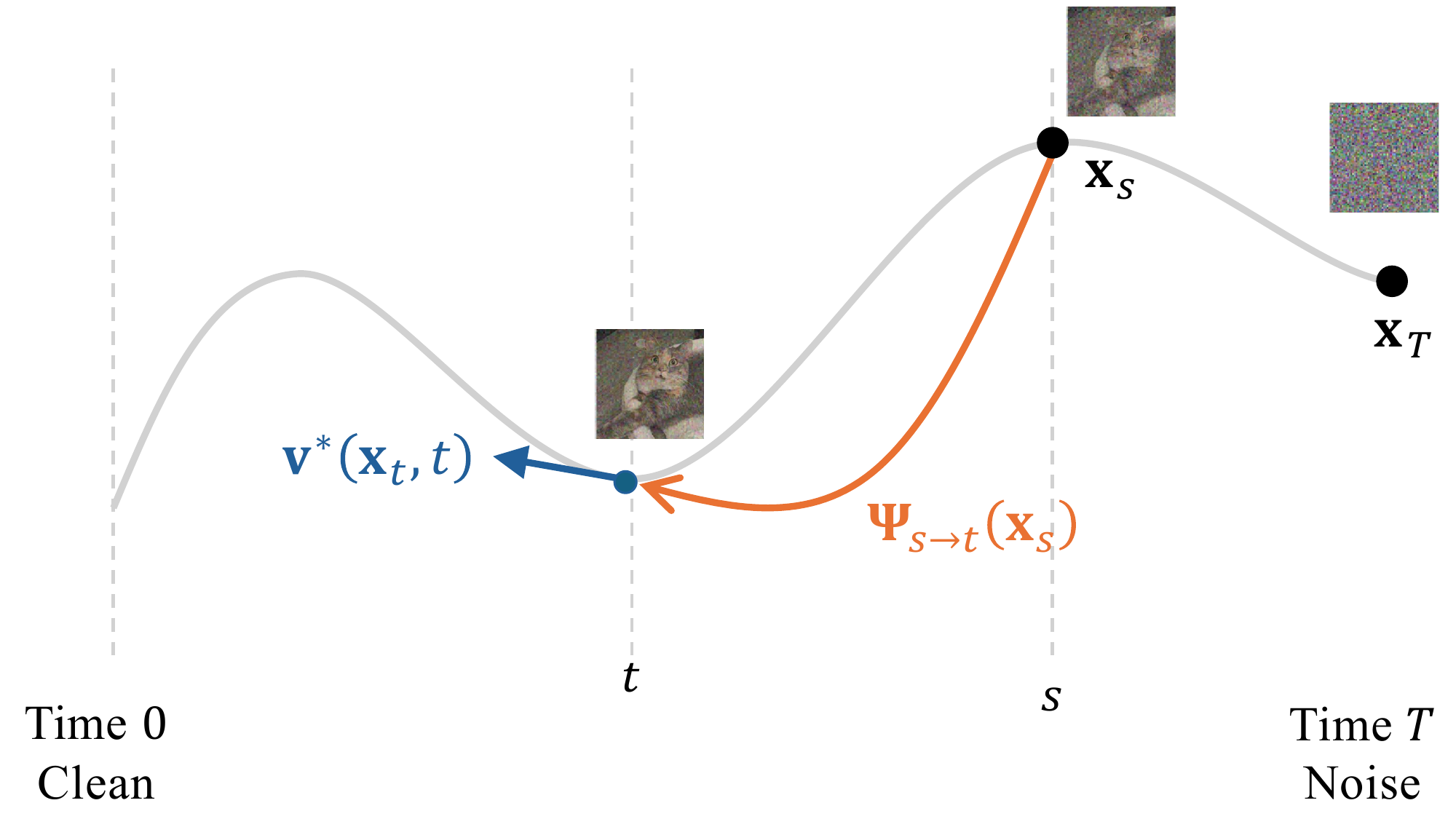}
        \caption{Base ending time $t$.}
        \label{fig:lagrangian-base}
    \end{subfigure}
    \hfill
    \begin{subfigure}[t]{0.485\textwidth}
        \centering
        \includegraphics[width=\linewidth]{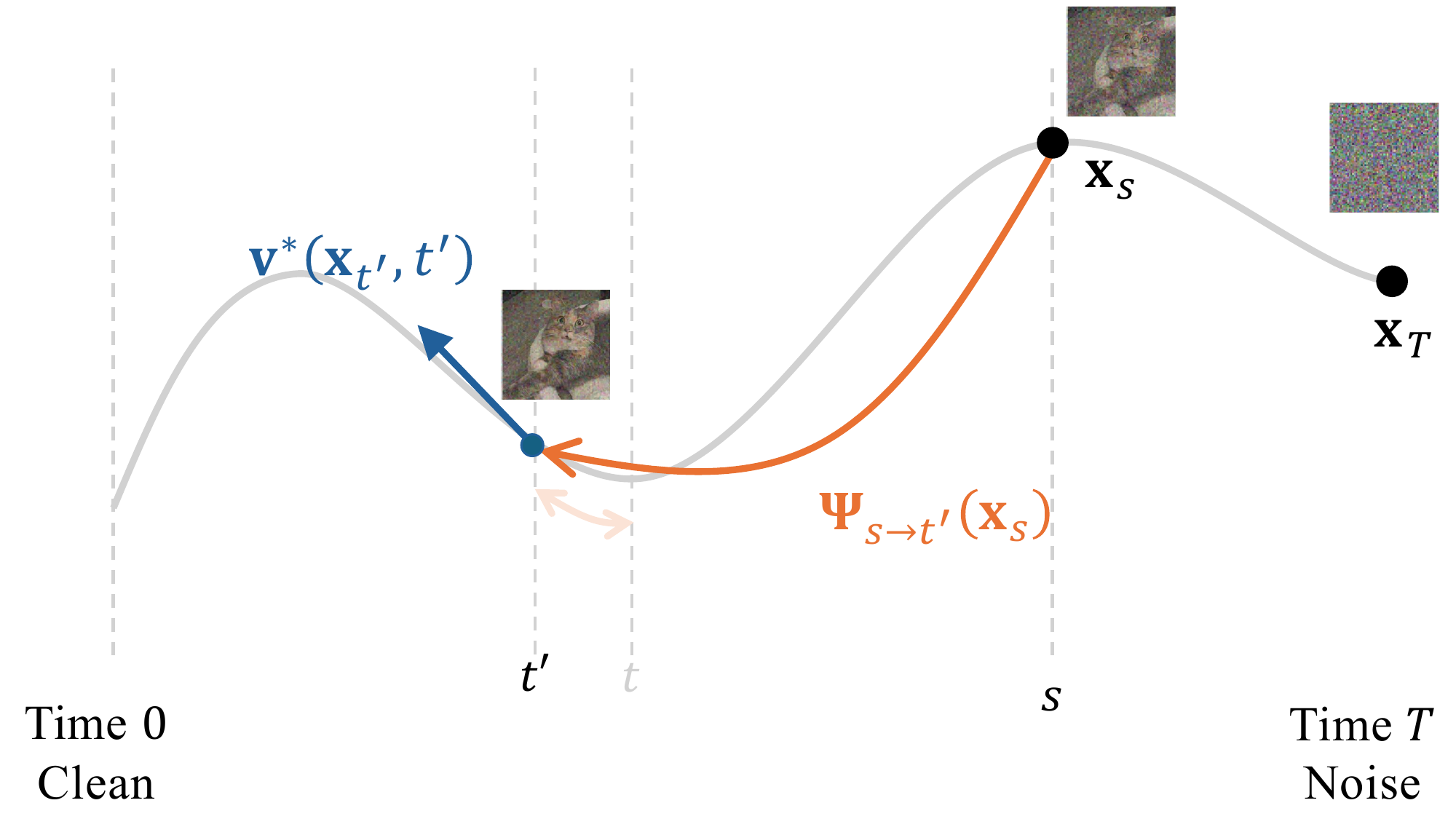}
        \caption{Moved ending time $t'$.}
        \label{fig:lagrangian-moved}
    \end{subfigure}

    \caption{\textbfs{Lagrangian consistency of flow maps.}
    The starting state--time pair $(\rvx_s,s)$ is fixed, and the ending time is moved from $t$ to
    $t'$. The output of the flow map moves along the same PF-ODE trajectory, and
    its infinitesimal change equals the ODE velocity:
    $\partial_t \bPsi_{s\to t}(\rvx_s)
    = \rvv^*(\bPsi_{s\to t}(\rvx_s),t)$.}
    \figcredit{Adapted from \citet{dieleman2026flowmaps}.}
    \label{fig:lagrangian-consistency}
\end{figure}

\paragraph{Lagrangian View: Move the Ending Point.}
Fix the starting state--time pair $(\rvx_s,s)$ and vary the ending time
$t$. The ending point then moves along the same trajectory:
\[
\partial_t \bPsi_{s\to t}(\rvx_s)
=
\rvv^*(\bPsi_{s\to t}(\rvx_s),t).
\]
The name ``Lagrangian'' follows the classical particle viewpoint: we
track the ending point along the trajectory as the ending time changes.
Infinitesimally, the ending point moves with the local ODE velocity
evaluated at that point.

\paragraph{Eulerian View: Move the Starting Point.}
Fix the ending time $t$ and slide the starting point along the same trajectory.
The ending point at time $t$ remains fixed:
\[
\frac{\diff}{\diff s}\bPsi_{s\to t}(\rvx_s)=0.
\]
Since $\frac{\diff}{\diff s}\rvx_s=\rvv^*(\rvx_s,s)$, the chain rule gives
\[
\partial_s\bPsi_{s\to t}(\rvx_s)
+
\big(\partial_{\rvx}\bPsi_{s\to t}\big)(\rvx_s)\rvv^*(\rvx_s,s)
=
0.
\]
The term ``Eulerian'' invokes the classical field-based viewpoint: Here, the ending time $t$
is fixed, while the starting state--time pair $(\rvx_s,s)$ moves along the
same ODE trajectory; the ending point at time $t$ therefore remains unchanged. For CM, the ending time is fixed at $t=0$, so the ending point is the clean sample:
\[
    \rvf^*(\rvx_s,s)=\bPsi_{s\to0}(\rvx_s).
\]
Thus, \Cref{eq:conti-ct-motivation} is the starting-time consistency condition
for the special flow map studied in \Cref{sec:discrete-cm,sec:conti-cm}.

This view also recaps the relationship between CTM and MF from
\Cref{subsec:equivalence-ctm-mf}.  Both approximate the same path integral
\[
\bPsi_{s\to t}(\rvx_s)
=
\rvx_s+\int_s^t\rvv^*(\rvx_\tau,\tau)\,\diff\tau,
\]
but expose different trainable surrogates.  CTM represents the ending point through a displacement-style
parameterization, while MF represents the same ending point through the
average drift
\[
\rvh^*(\rvx_s,s,t)
=
\frac{1}{t-s}\int_s^t\rvv^*(\rvx_\tau,\tau)\,\diff\tau.
\]
Substituting
\[
\bPsi_{s\to t}(\rvx_s)=\rvx_s+(t-s)\rvh^*(\rvx_s,s,t)
\]
into the Eulerian starting-time identity yields
\[
\rvh^*(\rvx_s,s,t)
=
\rvv^*(\rvx_s,s)
-
(s-t)
\left[
\partial_s\rvh^*(\rvx_s,s,t)
+
(\partial_{\rvx}\rvh^*)(\rvx_s,s,t)\rvv^*(\rvx_s,s)
\right].
\]
This is the basic Mean Flow identity: the average drift over a long interval is
the instantaneous drift at the starting point, corrected by how the average drift
changes as the starting point slides along the same trajectory.

\begin{figure}[t]
    \centering

    \begin{subfigure}[t]{0.485\textwidth}
        \centering
        \includegraphics[width=\linewidth]{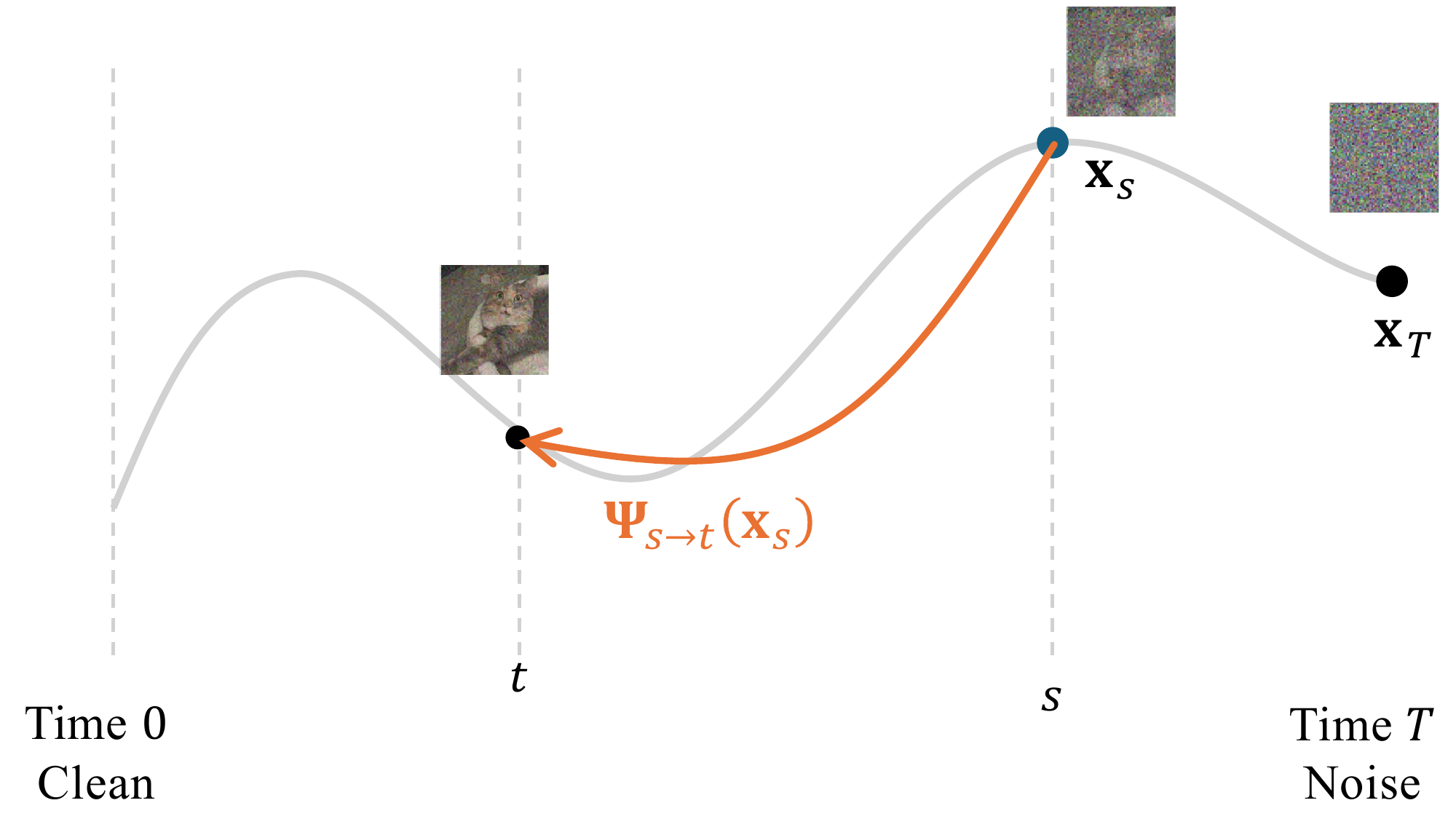}
        \caption{Base starting time $s$.}
        \label{fig:eulerian-base}
    \end{subfigure}
    \hfill
    \begin{subfigure}[t]{0.485\textwidth}
        \centering
        \includegraphics[width=\linewidth]{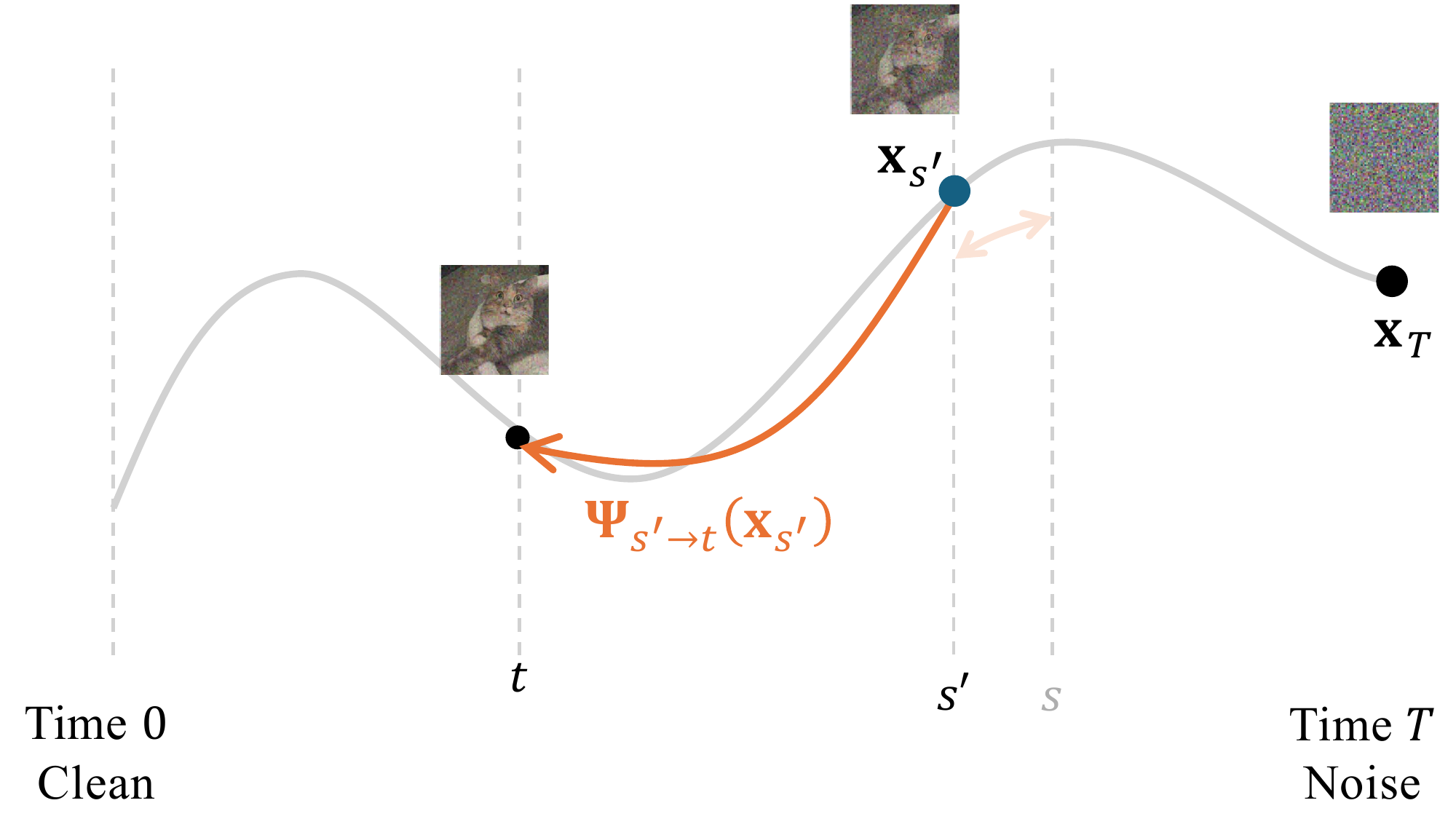}
        \caption{Moved starting time $s'$.}
        \label{fig:eulerian-moved}
    \end{subfigure}

    \caption{\textbfs{Eulerian consistency of flow maps.}
    The ending time $t$ is fixed, while the starting time is moved from $s$ to
    $s'$. Since the corresponding starting points lie on the same PF-ODE
trajectory, the ending point remains unchanged:
    $\bPsi_{s\to t}(\rvx_s)=\bPsi_{s'\to t}(\rvx_{s'})$.
    Infinitesimally, this gives
    $\frac{\diff}{\diff s}\bPsi_{s\to t}(\rvx_s)=0$, or equivalently
    $\partial_s\bPsi_{s\to t}(\rvx_s)
    +(\partial_{\rvx}\bPsi_{s\to t})(\rvx_s)\rvv^*(\rvx_s,s)=0$.}
        \figcredit{Adapted from \citet{dieleman2026flowmaps}.}
    \label{fig:eulerian-consistency}
\end{figure}

\paragraph{Mathematical Equivalence of the Three Perspectives.}
For an exact ODE flow, the semigroup, Lagrangian, and Eulerian identities
all express the same simple principle: under standard existence,
uniqueness, and smoothness assumptions, an ODE determines a unique
trajectory through each state--time pair. That is, once a state at a
given time is fixed, there is only one path to follow. The semigroup view
splits this path into consecutive pieces, the Lagrangian view follows its
ending point as the ending time changes, and the Eulerian view moves the
starting point along the same path while keeping the ending point fixed.
Thus, the three identities are equivalent descriptions of the same ODE
flow~\citep{arnold1988geometrical,chorin1993mathematical}. Their
connection becomes especially clear by viewing the semigroup identity as
the finite-interval statement, and the Lagrangian and Eulerian identities
as its two infinitesimal forms, obtained by shrinking a short interval near the ending time or near the
starting time, respectively.

The semigroup identity gives the finite-interval form of this principle:
\[
    \bPsi_{u\to t}\big(\bPsi_{s\to u}(\rvx_s)\big)
    =
    \bPsi_{s\to t}(\rvx_s),
    \qquad
    \bPsi_{s\to s}=\rmI .
\]
The two differential views are obtained by shrinking one of the two intervals.

\subparagraph{Ending-Time Limit: the Lagrangian Identity.}
Fix the starting state--time pair $(\rvx_s,s)$ and shrink the final
segment of the trip. For $\Delta t>0$,
\[
    \bPsi_{t\to t-\Delta t}
    \big(\bPsi_{s\to t}(\rvx_s)\big)
    =
    \bPsi_{s\to t-\Delta t}(\rvx_s).
\]
Let $\rvx_t:=\bPsi_{s\to t}(\rvx_s)$.  A first-order expansion of the short
flow gives
\[
    \bPsi_{t\to t-\Delta t}(\rvx_t)
    =
    \rvx_t-\Delta t\,\rvv^*(\rvx_t,t)
    +\mathcal{O}\big((\Delta t)^2\big).
\]
Therefore,
\[
    \frac{
    \bPsi_{s\to t-\Delta t}(\rvx_s)
    -
    \bPsi_{s\to t}(\rvx_s)}
    {-\Delta t}
    =
    \rvv^*(\rvx_t,t)
    +\mathcal{O}(\Delta t).
\]
Taking $\Delta t\to0$ yields the Lagrangian identity
\[
    \partial_t\bPsi_{s\to t}(\rvx_s)
    =
    \rvv^*\big(\bPsi_{s\to t}(\rvx_s),t\big).
\]
Thus, when the starting state--time pair $(\rvx_s,s)$ is fixed and the
ending time changes, the ending point moves with the ODE velocity
evaluated at that point.

\subparagraph{Starting-Time Limit: the Eulerian Identity.} Now fix the ending time $t$ and move the starting state--time pair
backward by a short interval $\Delta s>0$ along the same trajectory:
\[
    \rvx_{s-\Delta s}
    :=
    \bPsi_{s\to s-\Delta s}(\rvx_s)
    =
    \rvx_s-\Delta s\,\rvv^*(\rvx_s,s)
    +\mathcal{O}\big((\Delta s)^2\big).
\]
The semigroup identity gives
\[
    \bPsi_{s-\Delta s\to t}(\rvx_{s-\Delta s})
    =
    \bPsi_{s\to t}(\rvx_s).
\]
Expanding the left-hand side in both the starting state and the starting time gives
\[
    \bPsi_{s-\Delta s\to t}(\rvx_{s-\Delta s})
    =
    \bPsi_{s\to t}(\rvx_s)
    -
    \Delta s\,
    \big(\partial_{\rvx}\bPsi_{s\to t}\big)(\rvx_s)\rvv^*(\rvx_s,s)
    -
    \Delta s\,
    \partial_s\bPsi_{s\to t}(\rvx_s)
    +
    \mathcal{O}\big((\Delta s)^2\big).
\]
Comparing with the semigroup identity and dividing by $-\Delta s$ gives
\[
    \partial_s\bPsi_{s\to t}(\rvx_s)
    +
    \big(\partial_{\rvx}\bPsi_{s\to t}\big)(\rvx_s)\rvv^*(\rvx_s,s)
    =
    \mathcal{O}(\Delta s).
\]
Taking $\Delta s\to0$ yields the Eulerian identity
\[
    \partial_s\bPsi_{s\to t}(\rvx_s)
    +
    \big(\partial_{\rvx}\bPsi_{s\to t}\big)(\rvx_s)\rvv^*(\rvx_s,s)
    =
    0.
\]
Here, the ending time $t$ is fixed, while the starting state--time pair
moves along the same trajectory. The ending point therefore remains
unchanged.

\subparagraph{Recovering the Flow Map from Either Differential Identity.}
Conversely, either differential identity
together with the boundary condition $\bPsi_{s\to s}=\rmI$ recovers the same
flow map.  The Lagrangian identity evolves the ending point from
$\bPsi_{s\to s}(\rvx_s)=\rvx_s$.  The Eulerian identity says that
$\bPsi_{\tau\to t}(\rvx_\tau)$ is constant along the characteristic
$\rvx_\tau=\bPsi_{s\to \tau}(\rvx_s)$, hence
\[
    \bPsi_{s\to t}(\rvx_s)
    =
    \bPsi_{t\to t}(\rvx_t)
    =
    \rvx_t.
\]
Although these perspectives are equivalent at the oracle level, they lead to
different practical algorithms once approximations are introduced: one must
choose which identity to enforce, where to place stop-gradient targets, how to
sample time pairs, and how to parameterize the map.

\paragraph{Closing.}
The central shift explored in this chapter is from learning a local velocity
field to learning finite pieces of its ODE solution map. A conventional
diffusion sampler repeatedly evaluates the local field and numerically
integrates it. A flow-map model instead amortizes part of this integration
during training, so that a long interval can be traversed with one or a few
network evaluations.

This creates a direct route to fast generation, but moves difficulty from
sampling into learning. The model must approximate a more global object, and
its success depends on the parameterization of the map, the construction of
tractable targets, the sampling and weighting of time pairs, and the
stabilization of optimization.

Viewed in this way, flow-map learning is not a new dynamical concept but a
modern learning strategy built around a classical one. It preserves the
trajectory-based foundation of diffusion models while enlarging the design
space for efficient generation. The objective is no longer only to integrate
the generative ODE more efficiently at inference time, but to learn useful
finite-time transitions of its solution map directly.

\clearpage
\newpage

\part*{Epilogue and Outlook}
\addcontentsline{toc}{part}{Epilogue and Outlook}
\chapter{A Unifying Principle and the Road Ahead}\label{ch:epilogue}

\epigraph{\textit{The purpose of abstraction is not to be vague, but to create a new semantic level in which one can be absolutely precise.}}{Edsger W.\ Dijkstra}

Every chapter of this book has been about the same thing: how probability mass moves under a prescribed transformation, and how to reverse, approximate, or exploit that motion to generate data. In this final chapter, we step back to make this structural unity explicit. We first revisit the change-of-variable principle that underpins everything we have covered (\Cref{sec:epilogue-backbone}), then explain how the same probability-transport viewpoint extends to discrete state spaces through Markov kernels and continuous-time Markov chains (\Cref{sec:epilogue-discrete}), and close with reflections on the broader landscape (\Cref{sec:epilogue-closing}).

\chapterlearning{

  \item explain probability transport as the common principle underlying the
  generative modeling frameworks developed throughout this book;

  \item connect probability evolution in continuous spaces, through the
  continuity and Fokker--Planck equations, with its finite-state counterpart,
  the master equation;

  \item explain how reverse modeling changes in discrete state spaces, where
  transition kernels, probability ratios, and jump rates replace ordinary
  Euclidean score and velocity fields;

  \item compare the variational, score/ratio, and flow perspectives across
  continuous and discrete state spaces, identifying what remains common and
  what changes.

}{This chapter is an epilogue rather than a new method chapter. Read it as a
synthesis: focus on which principles remain unchanged, and which mathematical
objects must change, when moving from continuous to discrete state spaces.}

\clearpage
\newpage

\section{The Density-Transport Backbone}\label{sec:epilogue-backbone}

Throughout this book, we have followed two closely related lines of development.
The first is \emph{diffusion models themselves}, including variational, score-based, and flow-based formulations, which offer different mathematical descriptions of how probability laws evolve and how such evolution may be reversed for generation.
The second is \emph{diffusion-motivated fast generators}, most notably flow map models, which are built on the same transport viewpoint but aim to learn more direct and efficient generators.
Despite their differences in form, all of these methods are rooted in a single organizing idea: generative modeling is fundamentally a problem of \emph{transporting probability mass}. Starting from a simple reference distribution, such as Gaussian noise, we seek to construct a process that gradually connects it to a complex data distribution.

This viewpoint is useful because it shifts attention away from individual model families and toward a more fundamental question:
\emph{how does a probability law evolve when the underlying samples are moved by a prescribed dynamics?}
The mathematical language for answering this question is the change-of-variable formula and its continuous-time extensions. In this sense, the change-of-variable principle is not merely a technical tool; it is the conceptual backbone underlying both diffusion models and the fast generators they inspire.

\paragraph{A Common Three-Step Structure.}
At a high level, the methods studied in this book all follow the same simple pattern:
\begin{enumerate}[leftmargin=2em]
    \item Define a forward process that gradually moves data toward a simple reference distribution.
    \item Understand how the probability law evolves under that process through the appropriate law-evolution equation.
    \item Learn a generative mechanism that reverses, approximates, or otherwise exploits this evolution in order to produce samples.
\end{enumerate}
What varies across formulations is the choice of forward process, the mathematical form of the law-evolution equation, and the way the generative mechanism is parameterized and trained.

\paragraph{The Change-of-Variable Hierarchy.}
In \Cref{app:continuity}, we developed a systematic progression of change-of-variable formulas, from deterministic bijections to stochastic differential equations. That progression is worth recalling here, because it reveals a single transport principle appearing in several forms.

\FloatBarrier 

\begin{figure}[H]
  \centering

  \begin{subfigure}{\textwidth}
    \centering
    \includegraphics[width=\textwidth]{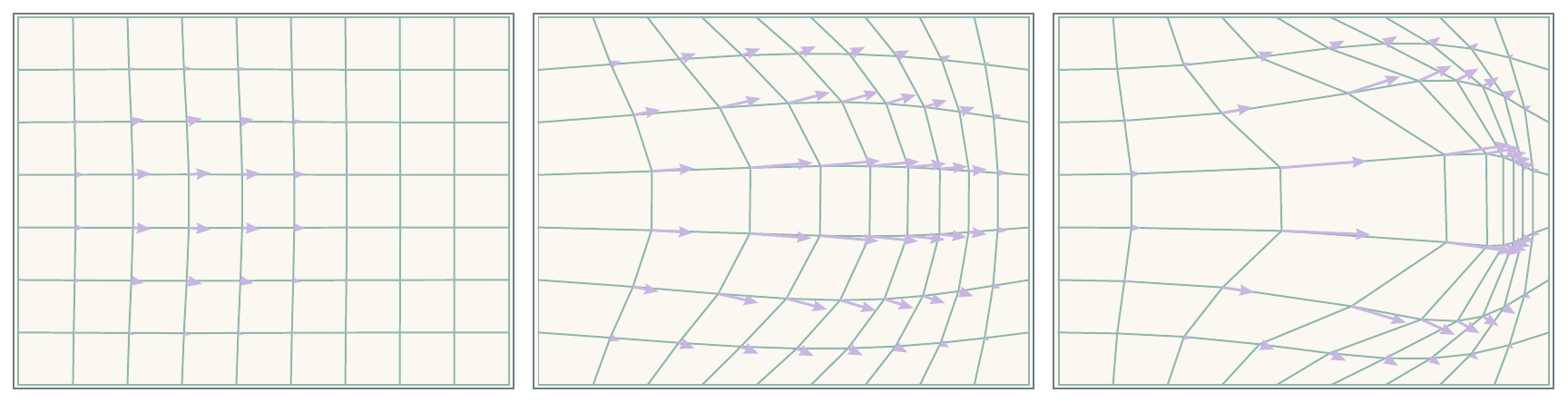}
        \caption{Vector field illustrations. The arrows represent velocities that move particles through space, deforming the underlying grid accordingly.}
  \end{subfigure}

  \vspace{0.5em} 

  \begin{subfigure}{\textwidth}
    \centering
    \includegraphics[width=\textwidth]{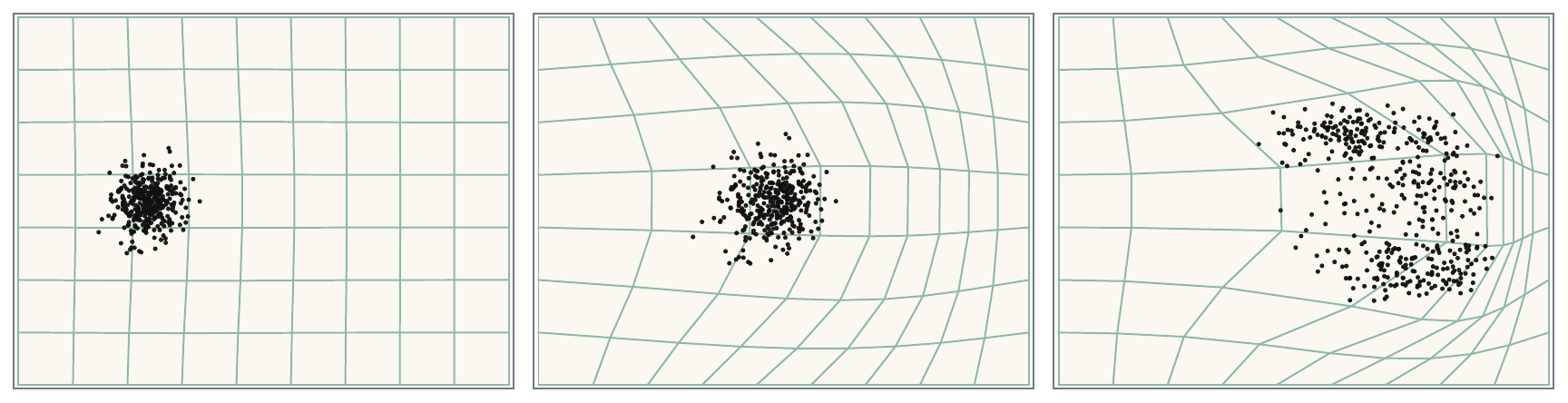}
    \caption{Particle-cloud dynamics. A predefined vector field generates a flow that transports particles from their initial state.}
  \end{subfigure}

  \vspace{0.5em}

  \begin{subfigure}{\textwidth}
    \centering
    \includegraphics[width=\textwidth]{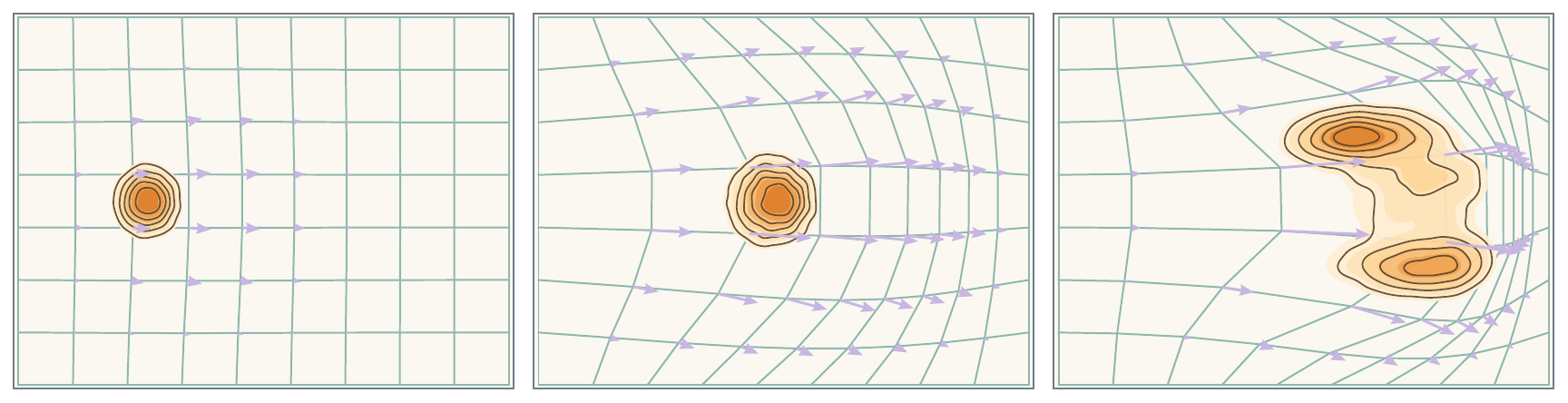}
    \caption{Density evolution. As particles are advected by the vector field, the density contours deform accordingly, reflecting how the flow reshapes the underlying distribution.}
  \end{subfigure}

  \caption{\textbfs{Illustrations of particle and density dynamics under a vector field.}
Each column shows successive time snapshots from left to right. \figcredit{Adapted from \citet{lipman2024flow}.}
}
  \label{fig:flow-3x3rows}
\end{figure}

\begin{enumerate}[leftmargin=2em]
    \item \textbfs{Single Bijection.}
    A smooth invertible map $\bPsi:\R^D\to\R^D$ transforms a density $p_0$ into $p_1$ via
    \[
    p_1(\rvx_1)
    =
    p_0\bigl(\bPsi^{-1}(\rvx_1)\bigr)
    \left|\det \partial_{\rvx_1}\bPsi^{-1}(\rvx_1)\right|.
    \]
    The Jacobian determinant records the local change of volume: when the map stretches space, the density decreases; when it compresses space, the density increases.

    \item \textbfs{Composed Bijections.}
    Chaining many bijections, with $\rvx_k=\bPsi_k(\rvx_{k-1})$, yields
    \[
    \log p_L(\rvx_L)
    =
    \log p_0(\rvx_0)
    -
    \sum_{k=1}^L
    \log\left|\det \partial_{\rvx_{k-1}}\bPsi_k(\rvx_{k-1})\right|.
    \]
    This is the basic principle behind normalizing flows. More importantly, it already illustrates the central message: once particle motion is specified, the evolution of the density follows accordingly.

    \item \textbfs{Continuous-Time Deterministic Flow (ODE).}
    Taking the infinitesimal-step limit leads to the continuity equation,
    \[
    \frac{\partial p_t(\rvx)}{\partial t}
    +
    \nabla\cdot\bigl(p_t(\rvx)\,\rvv(\rvx,t)\bigr)
    =
    0,
    \]
    where particles evolve according to
    \[
    \frac{\diff \rvx(t)}{\diff t} = \rvv(\rvx(t),t).
    \]
    This is the continuous-time expression of mass conservation. Probability mass is not created or destroyed; it simply flows through space under the velocity field $\rvv$.

    \item \textbfs{Continuous-Time Stochastic Flow (SDE).}
    Adding Brownian noise yields the Fokker--Planck equation,
    \[
    \frac{\partial p_t(\rvx)}{\partial t}
    =
    -\nabla\cdot\bigl(\rvv(\rvx,t)\,p_t(\rvx)\bigr)
    +
    \frac{1}{2}g^2(t)\,\Delta p_t(\rvx),
    \]
    which corresponds to the stochastic dynamics
    \[
    \diff\rvx(t) = \rvv(\rvx(t),t)\diff t + g(t)\diff\rvw(t).
    \]
    The first term describes directed transport by the drift, while the second captures the random spreading induced by noise.
\end{enumerate}

These are not unrelated formulas.
The multi-layer change-of-variable rule is the discrete-time version of transport under repeated deterministic transformations.
Its infinitesimal deterministic limit gives the continuity equation.
Adding stochastic perturbation leads to the Fokker--Planck equation.
Each level extends the previous one, while preserving the same underlying principle: probability mass is conserved, and the probability law adjusts to reflect how the underlying dynamics moves samples through space.

From this perspective, the main formulations in this book are easier to place in context.
Variational methods describe generation through denoising transitions on a time grid. Score-based methods work with stochastic dynamics and learn the score field governing the reverse-time SDE.
Flow-based methods work with deterministic transport and learn a velocity field whose induced marginal distributions follow a prescribed family of intermediate laws.
Fast generators, in turn, build on this same backbone, but seek more direct ways to exploit the learned transport for efficient sampling.

The central takeaway is therefore simple.
Diffusion modeling is not best understood as a single algorithm tied to one architecture or one noise schedule. It is a \emph{design principle}: specify a forward transport process, understand the induced evolution of probability laws, and then construct a generator on top of that structure. Once this viewpoint is in place, many seemingly different methods become natural variations of the same idea.

A natural question then arises: does this principle extend beyond continuous state spaces? In the next section, we show that it does.

\clearpage
\newpage

\section{Beyond Continuous States: Probability Evolution on Discrete Spaces}
\label{sec:epilogue-discrete}

So far, every model in this book has operated under a common structural
assumption: the state on which the generative dynamics acts lives in a
continuous Euclidean space $\R^D$. Whether the entries of $\rvx$ represent
pixel intensities, audio samples, or continuous latent variables, the
mathematics is the same: particles move along continuous trajectories governed
by an ODE or SDE, and the probability density evolves through the continuity or
Fokker--Planck equation. The framework rests on the differential structure of $\R^D$: gradients and
divergences are defined through derivatives, and deterministic dynamics can be
described by differentiable trajectories generated by vector fields.

This distinction is about the modeling state, not about the raw modality.
Discrete observations can also be handled by first choosing a continuous code
and a readout. For a raw symbol $a$, let $\rmE(a)\in\R^d$ be its code,
and let $\rmD$ map codes back to symbols, or more generally to a
categorical distribution:
\[
\text{raw discrete symbol } a
\;\xrightarrow{\;\rmE\;}\;
\text{continuous code } \rmE(a)\in\R^d
\;\xrightarrow{\;\rmD\;}\;
\text{discrete readout}.
\]
The representation may be fixed, such as a one-hot, categorical-vector, or
simplex-valued code~\citep{richemond2022categorical};
on an injective codebook, the readout can be deterministic and may satisfy
$\rmD(\rmE(a))=a$. Alternatively, the representation may be learned,
such as a token embedding or an autoencoder latent
\citep{li2022diffusionlm,dieleman2022continuous}; then $\rmD$ is a learned
readout and $\rmD\circ\rmE$ need not be exactly invertible. Once the
representation is fixed, the diffusion is continuous-state: Gaussian noising in
an ambient code space, or a geometry-adapted continuous process for a constrained
code space, is handled by the continuous machinery developed earlier in the
book. The finite-state route studied below is different: the evolving state
itself remains discrete.

The present section studies the complementary case, where the state itself
remains finite and discrete state throughout the forward and reverse processes. Many data types naturally fit this view: text is a sequence of
vocabulary tokens, molecular graphs use finitely many atom and bond types, and
protein sequences are strings over a finite alphabet. In such a finite state
space there is no canonical infinitesimal displacement from one state to another,
no smooth trajectory through token values, and no ordinary gradient with respect
to the state itself. This raises a natural question:
\begin{question}
    If diffusion is really about transporting probability mass, what does that idea become when the state space is finite?
\end{question}

The structural backbone remains the same, but the mathematical objects change.
In continuous space, probability laws evolve under ODE/SDE dynamics through the
continuity or Fokker--Planck equation. In a finite state space, probability mass
moves between states according to Markov transition kernels or jump rates, and
the law evolves by a finite-dimensional conservation law: the \emph{master
equation}, also called the Kolmogorov forward equation. At an abstract level,
both settings follow the same recipe:

\msg{Observation}{}{Choose a forward corruption process, understand its induced probability path, and learn a reverse transport mechanism.}

In continuous Gaussian diffusion, this reverse mechanism appears as Gaussian
reverse conditionals, scores, or velocity fields. In finite state spaces, it
appears instead as reverse transition probabilities, probability ratios, or jump
rates. These are different mathematical objects, but they play the same
conceptual role.

The goal of this chapter is not to present one discrete diffusion algorithm as
definitive. Instead, we develop a roadmap that should remain stable as the
literature changes. The roadmap is organized by the object used to describe
reverse transport: transition kernels, probability ratios, or jump rates. We
proceed in four steps: discrete-time transitions (\Cref{subsec:epilogue-dtmc}),
their continuous-time limit (\Cref{subsec:epilogue-ctmc}), three perspectives on
reverse modeling that mirror the variational, score/ratio, and flow viewpoints
of the continuous case (\Cref{subsec:epilogue-three-views}), and a comparison
highlighting what is genuinely different from the continuous setting
(\Cref{subsec:epilogue-genuine-diff}).

\subsection{Discrete-Time Transitions}
\label{subsec:epilogue-dtmc}

Consider a forward process over $K$ discrete categories, where $K$
includes any special corruption tokens, such as a mask token, required
by the chosen forward process.  Each state is represented as a
\emph{one-hot} column vector $\rvx\in\{0,1\}^{K}$ with exactly one
entry equal to~$1$; the set of all such vectors
is\footnote{An equivalent formulation, standard in the CTMC and
stochastic-processes
literature~\citep{norris1998markov,kelly1979reversibility}, represents
each state as a scalar index $x\in\{1,\dots,K\}$, writes transition
rates as $[\rmQ_t]_{ij}$, and the marginal as $p_t(x)$.  The one-hot
encoding adopted here is notationally equivalent but aligns more closely
with the discrete diffusion
literature~\citep{austin2021structured,sahoo2024simple,lou2024discrete},
where inner products such as $\langle\rvx_{\bphi},\rvx\rangle$ and
matrix--vector products such as $\rmP^{\top}\rvx$ appear naturally in
training objectives.}
\[
\mathcal{V}
= \bigl\{\rvx\in\{0,1\}^{K}:\textstyle\sum_{i=1}^{K}x_i=1\bigr\}
= \{\rve_1,\ldots,\rve_K\}.
\]
We write $\rve_i$ for the $i$-th standard basis vector, namely the one-hot
encoding of category~$i$.  For example, each $\rve_i$ can represent
a token, a discrete latent code, an atom type, a bond type, or an amino acid.

For sequence data such as text or proteins, a sample consists of $N$
tokens, each taking values in a vocabulary of size~$K$. The full state space
is then $\mathcal{V}^N$, whose size grows exponentially with sequence length.
In many discrete diffusion models, the \emph{forward} corruption acts
independently across positions, so the single-token Markov process is the
basic local building block. We therefore focus first on one token, because it
exposes the core probability-transport mechanism without notational overload.
For full sequences, the same formulas apply to full configurations. If the
forward corruption acts independently across positions, the conditional
forward kernel factorizes positionwise.\footnote{The resulting marginal
distribution over corrupted sequences generally need not factorize, because
the clean-data distribution can contain dependencies across positions.}
The learned reverse model, however, need not be independent across positions:
a Transformer or graph neural network can observe the entire corrupted object
and use global context to predict local or global reverse transitions.

\begin{figure}[th!]
    \centering
    \includegraphics[width=\linewidth]{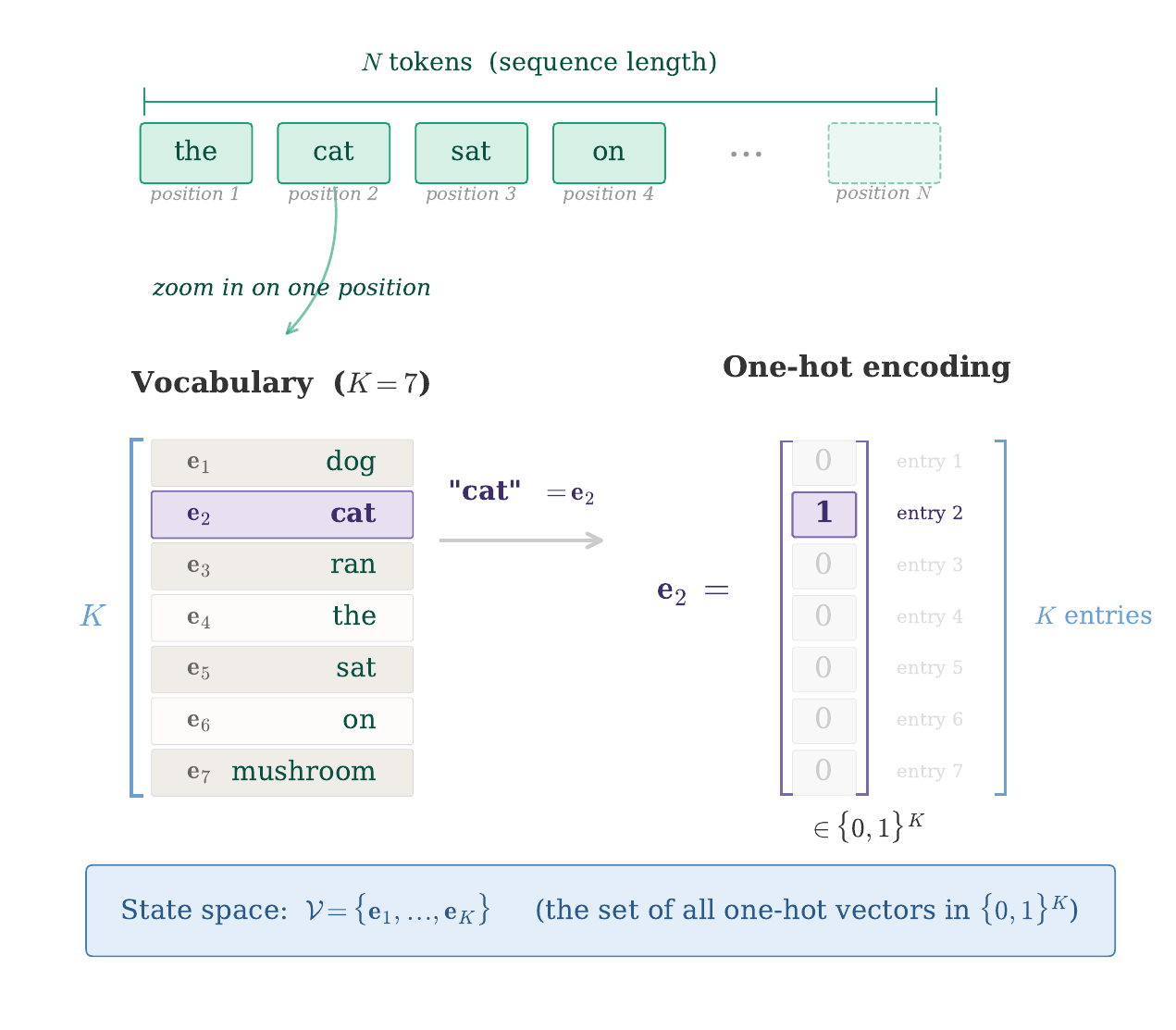}
    \caption{\textbfs{From sequences to one-hot states.}
    A data sample such as a text or protein sequence consists of $N$
    tokens, each taking values in a finite alphabet or vocabulary of size~$K$.  This chapter focuses on
    the evolution of a single token among $K$ states.  Zooming in on one
    position, here ``cat'', the token is represented by the basis vector
    $\rve_2$ and encoded as a one-hot vector in $\{0,1\}^K$, namely a
    $K$-dimensional vector whose second entry is~$1$ and all other
    entries are~$0$.  The state space
    $\mathcal{V}=\{\rvx\in\{0,1\}^K:\sum_{i=1}^K x_i=1\}
    =\{\rve_1,\ldots,\rve_K\}$ is the set of all such one-hot vectors.
    The ordering of the vocabulary entries is arbitrary and unrelated to
    the position of a token in the sequence. \figcredit{Created by the authors with AI-assisted coding.}}
    \label{fig:notation-overview}
\end{figure}

At step~$k$, the system occupies one of these $K$ states.  We denote the
scalar probability of being in state~$\rve_i$ by
\[
p_k(i) := \Pr(\rvx_k=\rve_i),
\qquad i=1,\ldots,K.
\]
Collecting these scalar probabilities gives the marginal probability vector
\[
\rvp_k
:=
\bigl(p_k(1),\;\dots,\;p_k(K)\bigr)^\top
=
\E[\rvx_k]
\in \Delta^{K-1}\subset\R^K,
\qquad
\sum_{i=1}^K p_k(i)=1,
\]
where $\Delta^{K-1}:=\bigl\{\rvp\in\R^K:p_i\geq0,\;
\sum_{i=1}^K p_i=1\bigr\}$ denotes the probability simplex.
Equivalently, $\Pr(\rvx_k=\rve_i)=p_k(i)$, or briefly
$\rvx_k\sim\mathrm{Cat}(\rvp_k)$.

Thus $\rvp_k$ is the full law of the discrete random state $\rvx_k$, while
$p_k(i)$ is its $i$-th scalar entry.  At continuous time we use the same
convention:
\[
p_t(i):=\Pr(\rvx_t=\rve_i),
\qquad
\rvp_t:=\bigl(p_t(1),\ldots,p_t(K)\bigr)^\top.
\]
The vector $\rvp_t$ is the discrete analogue of a density: it tells us how
the total probability mass is distributed across the $K$ states.

Following the convention used throughout this book, bold lowercase letters
($\rvp$, $\rvx$) denote vectors, bold uppercase letters ($\rmP$, $\rmQ$)
denote matrices, and unbolded indexed quantities such as $p_k(i)$,
$[\rmP_k]_{ij}$, and $[\rmQ_t]_{ij}$ denote scalar entries.  Unbolded
$p$ with random variables as arguments, such as
$p(\rvx_{k-1}|\rvx_k)$, denotes a probability mass function or transition
kernel under the fixed forward process; $p_{\bphi}$ denotes the learned
reverse model.

How does the system move between states? A single transition step is
specified by a \emph{transition matrix} $\rmP_k \in \R^{K \times K}$.
Its entry $[\rmP_k]_{ij}$ gives the probability that the system,
currently in state~$\rve_i$, moves to state~$\rve_j$ in one step:
\[
[\rmP_k]_{ij} \;=\; \Pr(\rvx_{k+1} = \rve_j | \rvx_k = \rve_i).
\]
Thus the full matrix has the following structure:
\[
\rmP_k \;=\;
\underbrace{
\begin{pmatrix}
[\rmP_k]_{11} & [\rmP_k]_{12} & \cdots & [\rmP_k]_{1K} \\
[\rmP_k]_{21} & [\rmP_k]_{22} & \cdots & [\rmP_k]_{2K} \\
\vdots & \vdots & \ddots & \vdots \\
[\rmP_k]_{K1} & [\rmP_k]_{K2} & \cdots & [\rmP_k]_{KK}
\end{pmatrix}
}_{\substack{\text{columns: next state } j}}
\quad
\left.
\vphantom{
\begin{pmatrix}
a \\ a \\ \vdots \\ a
\end{pmatrix}
}
\right\}
\text{\footnotesize rows: current state } i.
\]
Reading across row $i$ answers: ``given that the system is currently
in state $\rve_i$, what is the probability of landing in each possible
next state?'' Each row is therefore a probability distribution over destinations:
\[
[\rmP_k]_{ij} \geq 0
\quad \text{for all } i,j,
\qquad\qquad
\sum_{j=1}^{K} [\rmP_k]_{ij} = 1
\quad \text{for each } i.
\]
The non-negativity condition says every transition probability is valid,
and the row-sum constraint says all mass leaving state~$\rve_i$ is accounted
for: the system always ends up somewhere.

\paragraph{How the Distribution Evolves.}
With the transition matrix in hand, we now ask a basic question: if the
system starts with distribution $\rvp_k$, what is the distribution after
one step? The answer follows from simple probability accounting. The
probability of being in state~$\rve_j$ after the transition is the total
mass arriving at~$\rve_j$ from all possible source states:
\begin{equation}
p_{k+1}(j) = \sum_{i=1}^K p_k(i)\,[\rmP_k]_{ij}.
\label{eq:discrete-transport-one-step}
\end{equation}
Each term in the sum represents the mass initially at state~$\rve_i$,
weighted by the probability of jumping from~$\rve_i$ to~$\rve_j$.
In vector form, this becomes
\[
\rvp_{k+1} = \rmP_k^\top \rvp_k.
\]
Equivalently, the conditional distribution of the next state given the
current one-hot state~$\rvx_k$ is
\[
\Pr(\rvx_{k+1} | \rvx_k)
= \mathrm{Cat}(\rvx_{k+1};\,\rmP_k^\top\,\rvx_k),
\]
since $\rmP_k^\top\,\rve_i$ extracts the $i$-th row of~$\rmP_k$, which
contains the transition probabilities from state~$\rve_i$.

A system evolving by this rule is called a \emph{discrete-time Markov
chain}: conditional on the current state $\rvx_k$, the distribution of
$\rvx_{k+1}$ does not depend on the earlier states. This plays the finite-state
role of the law-evolution formulas in continuous space. There, a bijection $\bPsi$ reshapes a density, and the
Jacobian determinant compensates for local volume change. In a finite
state space, there is no volume element to track. Instead, the transition
matrix $\rmP_k$ directly specifies how much probability mass leaves each
source state and arrives at each destination. The row-sum constraint
guarantees that total probability is conserved, just as the divergence
structure does in the continuity equation.

Repeating this over multiple steps, with transition matrices
$\rmP_0, \dots, \rmP_{L-1}$, yields
\begin{equation*}
\rvp_L = \rmP_{L-1}^\top \cdots \rmP_1^\top \rmP_0^\top\, \rvp_0.
\end{equation*}
This mirrors the multi-layer change-of-variable formula in
\Cref{eq:cov-multi-layer-nolog}. In the continuous deterministic setting, density
evolution is accumulated through Jacobian determinants; here, the
probability vector is propagated through successive Markov operators.
The underlying idea is the same: probability mass is tracked through a
sequence of prescribed transformations.

\paragraph{Inverting a Single Step.}
Given the forward machinery, it is natural to ask how a single
transition can be reversed for generation.  Suppose the system is observed in
state~$\rve_j$ at step~$k{+}1$.  What is the probability that it was in
state~$\rve_i$ at the previous step~$k$?  This is a direct application
of Bayes' rule:
\begin{align}
\Pr(\rvx_k = \rve_i | \rvx_{k+1} = \rve_j)
= \frac{\Pr(\rvx_{k+1} = \rve_j | \rvx_k = \rve_i)\;\Pr(\rvx_k = \rve_i)}
        {\Pr(\rvx_{k+1} = \rve_j)}  
= [\rmP_k]_{ij}\frac{p_k(i)}{p_{k+1}(j)},
\label{eq:discrete-reverse-kernel}
\end{align}
whenever $p_{k+1}(j)>0$. This is the discrete-time reverse kernel associated with the forward
transition~$\rmP_k$ together with the current marginal law. Its form is
intuitive: among all the ways that mass could have arrived at
state~$\rve_j$, each source~$\rve_i$ contributes in proportion to how
much mass was there, through $p_k(i)$, and how likely the jump was,
through $[\rmP_k]_{ij}$. The numerator and denominator in \Cref{eq:discrete-reverse-kernel} each have a clear interpretation:
\begin{itemize}[leftmargin=2em]
    \item \textbfs{Numerator:}
    The term $[\rmP_k]_{ij}\,p_k(i)$ is the joint probability of being in
    state~$\rve_i$ at step~$k$ and then jumping to~$\rve_j$ at
    step~$k{+}1$.

    \item \textbfs{Denominator:}
    The term
    \[
    p_{k+1}(j) = \sum_{i'} p_k(i')\,[\rmP_k]_{i'j}
    \]
    is the total probability of arriving at state~$\rve_j$ from any
    source. It serves as the normalizing constant, ensuring that the
    reverse probabilities sum to one over~$i$.
\end{itemize}

A crucial point is that the reverse kernel is \emph{not} the matrix inverse of $\rmP_k$. A Markov transition may merge probability mass from many states into one state, and it need not be invertible as a linear map. The probabilistic reverse is instead a Bayesian inverse, and it depends not only on the forward transition matrix but also on the marginal distribution $\rvp_k$. This is exactly the difficulty faced in generative modeling: $\rvp_k$ is obtained by pushing the unknown data distribution through the forward process, so it is usually not available in closed form.  The reverse-modeling perspectives below can be understood as different ways of representing or learning the missing information needed to implement this Bayesian inverse.

\subsection{Continuous-Time Limit: The Master Equation}
\label{subsec:epilogue-ctmc}

In the continuous-space story, shrinking the step size of composed
bijections to zero yielded the continuity equation
(\Cref{subsec:cov-continuity-eq}). The same limiting idea works here. When the
time steps become infinitesimally small, a discrete-time Markov chain
becomes a continuous-time Markov chain (CTMC), and the distribution
evolves according to the master equation.

\begin{figure}[th!]
    \centering
    \includegraphics[width=0.8\linewidth]{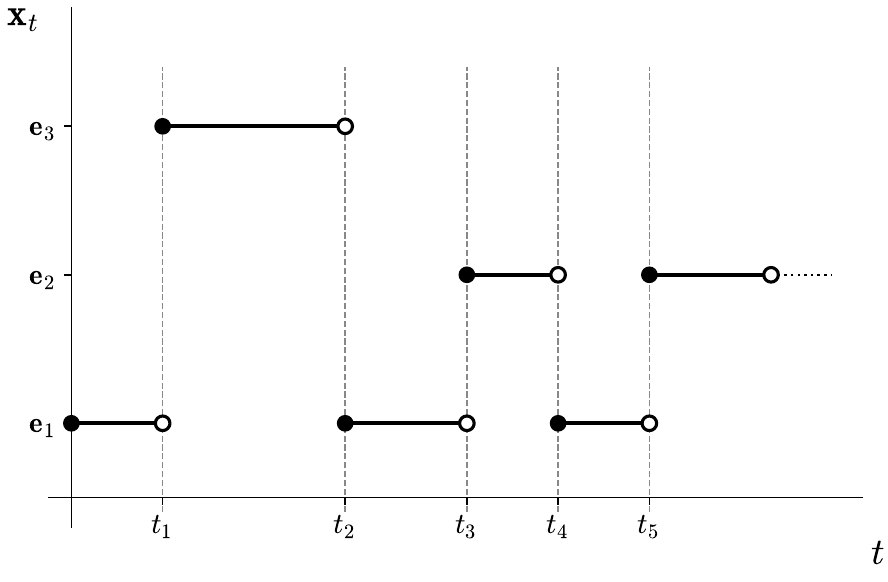}
    \caption{\textbfs{Illustration of a CTMC sample path.}
Let the state space be $\mathcal{V}=\{\rve_1,\rve_2,\rve_3\}$, labeled as $S_1,S_2,S_3$ on the vertical axis.
The trajectory is piecewise constant: the process remains at $\rvx_t=\rve_i$ for a random holding time, then jumps instantaneously to another state $\rve_j$ at times $t_1,t_2,\ldots$.
Filled circles mark the state occupied immediately after each jump, and open circles the state just before the jump, reflecting the right-continuous convention.
The dotted segment indicates continuation beyond the displayed window.
These jump times and destinations are governed by the rate matrix $\rmQ_t$: from state $\rve_i$, the process jumps to $\rve_j$ ($j\neq i$) at instantaneous rate $[\rmQ_t]_{ij}$.
\figcredit{Adapted from \citet{campbell2022continuous}.}}
    \label{fig:ctmc-trajectory}
\end{figure}

\paragraph{Infinitesimal Transitions.}
In the previous subsection, each transition matrix $\rmP_k$ described a
full step from time $k$ to time $k{+}1$. Now imagine making these steps
finer and finer. Let $\rmP_{t,\Delta t}$ denote the transition matrix
that moves the system forward by a small duration $\Delta t$ starting at
time $t$: its entry $[\rmP_{t,\Delta t}]_{ij}$ is the probability of
moving from state~$\rve_i$ to state~$\rve_j$ during the interval
$[t,\,t{+}\Delta t]$.

When $\Delta t$ is small, most mass stays put and only a small fraction
jumps. This means $\rmP_{t,\Delta t}$ is close to the identity, and we
can expand it as
\begin{equation}
\rmP_{t,\Delta t}
\;=\; \rmI + \rmQ_t\,\Delta t + \mathcal{O}(\Delta t^2),
\label{eq:generator-expansion}
\end{equation}
where $\rmQ_t \in \R^{K \times K}$ is the \emph{rate matrix}, also
called the generator, at time $t$. If $\rmQ_t$ is sufficiently regular in time, the remainder can be written as $\mathcal{O}(\Delta t^2)$ locally. The entries of $\rmQ_t$ are instantaneous
\emph{rates} rather than probabilities. Specifically:
\begin{itemize}[leftmargin=2em]
    \item $[\rmQ_t]_{ij} \geq 0$ for $i \neq j$: this is the rate at which
      the process jumps from state~$\rve_i$ to state~$\rve_j$.
      Multiplying by $\Delta t$ gives the leading-order jump probability
      over a short interval.
    \item $[\rmQ_t]_{ii} = -\sum_{j \neq i} [\rmQ_t]_{ij}$: the
      diagonal entry is negative and records the total rate at which
      probability \emph{leaves} state~$\rve_i$. This ensures that each row
      of $\rmQ_t$ sums to zero, which is the rate-level expression of
      probability conservation.\footnote{This follows directly from the
      row-sum constraint on $\rmP_{t,\Delta t}$. Since each row of a
      transition matrix sums to $1$, substituting the expansion
      $\rmP_{t,\Delta t} = \rmI + \rmQ_t\,\Delta t
      + o(\Delta t)$ gives
      $1 + \Delta t \sum_{j} [\rmQ_t]_{ij}
      + o(\Delta t) = 1$,
      which forces $\sum_{j} [\rmQ_t]_{ij} = 0$. Separating the
      diagonal term yields
      $[\rmQ_t]_{ii} = -\sum_{j \neq i} [\rmQ_t]_{ij}$.}
\end{itemize}
Reading off the expansion, the probability of jumping from~$\rve_i$ to
$\rve_j$ with $j\neq i$ during a short interval is approximately
$[\rmQ_t]_{ij}\,\Delta t$, while the probability of staying
at~$\rve_i$ is approximately
$1 - \sum_{j \neq i} [\rmQ_t]_{ij}\,\Delta t$.

\paragraph{Master Equation.}
With infinitesimal transitions in hand, the continuous-time limit follows
immediately. Substituting the expansion in \Cref{eq:generator-expansion}
into $\rvp_{t+\Delta t} = \rmP_{t,\Delta t}^\top\,\rvp_t$ and taking
$\Delta t \to 0$ gives
\begin{mdframed}
\begin{equation}
\frac{\diff}{\diff t}\,\rvp_t = \rmQ_t^\top\,\rvp_t.
\label{eq:master-equation-vector}
\end{equation}
\end{mdframed}
This is the \emph{master equation}, also known as the \emph{Kolmogorov
forward equation} for CTMCs~\citep{norris1998markov}. Recall that $\rvp_t$ is the probability
vector whose $j$-th entry $p_t(j)$ gives the probability of being in
state~$\rve_j$ at time~$t$. Writing out the $j$-th component of
\Cref{eq:master-equation-vector} makes the meaning concrete:
\begin{equation}
\frac{\diff}{\diff t}\,p_t(j)
= \underbrace{\sum_{i \neq j} p_t(i)\,[\rmQ_t]_{ij}}_{\text{inflow: mass arriving from other states}}
\;-\;
\underbrace{p_t(j)\sum_{\ell \neq j} [\rmQ_t]_{j\ell}}_{\text{outflow: mass departing to other states}}.
\label{eq:master-equation-component}
\end{equation}
The rate of change of $p_t(j)$ is simply inflow minus outflow. This is
probability conservation expressed as a differential equation.

\paragraph{Comparison with the Continuity Equation.}
In continuous space, the continuity equation
\[
\partial_t\,p_t(\rvx) + \nabla \cdot
\big(p_t(\rvx)\,\rvv(\rvx,t)\big) = 0
\]
says that density changes are driven by the net flux of probability mass
under the velocity field $\rvv$. The master equation says the same
thing in finite-state language: $p_t(j)$ changes because of net probability flux governed by the
rate matrix $\rmQ_t$. The velocity field and the rate matrix play
analogous structural roles: both prescribe how probability mass is
redistributed, and both enforce conservation. The main difference is geometric: in $\R^D$, mass flows through neighboring spatial locations; in a finite state space, mass jumps along allowed edges of a transition graph.

\begin{figure}[th!]
    \centering
    \includegraphics[width=\linewidth]{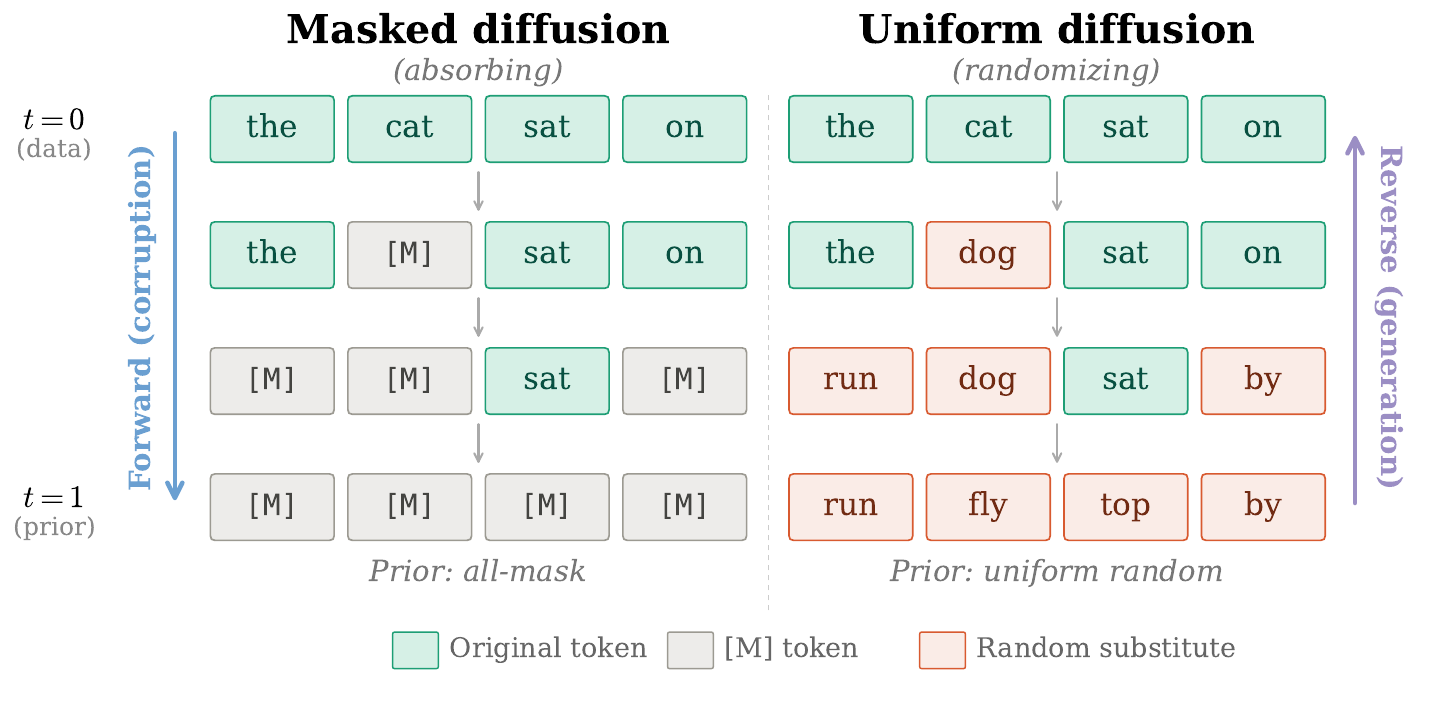}
\caption{\textbfs{Illustration of masked and uniform diffusion forward corruption and reverse generation.}
The two examples shown here admit an interpolating marginal of the form
$\Pr(\rvx_t|\rvx_0)=\mathrm{Cat}(\rvx_t;\alpha_t\rvx_0+(1-\alpha_t)\bm{\pi})$,
where the limiting prior $\bm{\pi}$ and decay factor $\alpha_t$ are determined
by the chosen forward rate matrix.  In masked diffusion, tokens are
absorbed into the mask state, with limiting prior $\bm{\pi}=\rvm=\rve_K$.  In
uniform diffusion, tokens are repeatedly randomized, with limiting prior
$\bm{\pi}_{\mathrm{unif}}=\frac{1}{K}\mathbf{1}$.  Generation reverses the
chosen corruption process, starting from the corresponding simple reference distribution, or from a close finite-time approximation to it.
\figcredit{Created by the authors with AI-assisted coding.}}
    \label{fig:mask-uniform}
\end{figure}

\paragraph{Forward Corruption.}
Just as in the continuous case, we need a forward process that gradually
destroys the structure of the data distribution and drives it toward a simple
reference distribution that is easy to sample from.  This is the discrete analogue of choosing a
Gaussian noising process in continuous diffusion.  The design goal is the same:
the corruption should be progressive, analytically or computationally tractable, and strong enough
that, by terminal time $T$, the data distribution has been largely washed out.

The rate matrix can be chosen in many ways. It may be uniform, absorbing, nearest-neighbor, graph-based, chemically structured, grammar-aware, or otherwise adapted to the data domain. The following two examples are therefore not meant to exhaust the taxonomy; they are canonical cases that make the mechanics transparent.
\subparagraph{Masked Diffusion (Absorbing).} Reserve the last
      category as a mask token, written $\rvm := \rve_K$.  Clean data
      $\rvx_0$ occupies one of the first $K{-}1$ categories, so
      $\langle\rvx_0,\rvm\rangle = 0$.  The rate matrix is defined so
      that every ordinary state transitions to the mask state at
      rate~$\beta(t)$, while the mask state never transitions out:
\[
[\rmQ_t]_{iK}=\beta(t),\qquad
[\rmQ_t]_{ii}=-\beta(t),\qquad
[\rmQ_t]_{ij}=0
\quad
\text{for } i<K,\; j\notin\{i,K\},
\]
and
\[
[\rmQ_t]_{Kj}=0
\quad
\text{for all } j.
\]
      The idea is simple: each token is independently replaced by the
      mask at rate~$\beta(t)$, and once masked it stays masked.

      The forward marginal conditioned on a clean state~$\rvx_0$ is
      \[
      \Pr(\rvx_t | \rvx_0)
      = \mathrm{Cat}\bigl(\rvx_t;\;
          \alpha_t\,\rvx_0 + (1-\alpha_t)\,\rvm\bigr),
      \]
      where
      \[
      \alpha_t
      =
      \exp\left(-\int_0^t\beta(s)\,\diff s\right).
      \]
      At time~$t$, the token equals its clean value~$\rvx_0$ with
      probability~$\alpha_t$ and is masked with probability~$1-\alpha_t$.
      As time increases, more and more probability mass accumulates on the
      mask state.  If $\int_0^T\beta(s)\diff s$ is large, then $\alpha_T$ is small and the terminal distribution is close to the all-mask reference distribution.

\subparagraph{Uniform Diffusion.} Instead of erasing a token, this
      process gradually randomizes it.  One convenient continuous-time normalization is to let a token jump to each of the other $K-1$ states at the same rate:
      \[
      [\rmQ_t]_{ij} = \frac{\beta(t)}{K - 1}
      \quad\text{for } i \neq j,
      \qquad
      [\rmQ_t]_{ii} = -\beta(t).
      \]
      The first formula says that from state~$\rve_i$, the token jumps
      to each alternative state at the same instantaneous rate.  The second says that
      the total rate of leaving state~$\rve_i$ is~$\beta(t)$.

      The forward marginal again takes the interpolating form
      \[
      \Pr(\rvx_t|\rvx_0)
      = \mathrm{Cat} \bigl(\rvx_t;\;
          \alpha_t\,\rvx_0 + (1-\alpha_t)\,\bm{\pi}_{\mathrm{unif}}
        \bigr),
      \]
      where
      $\bm{\pi}_{\mathrm{unif}}=\tfrac{1}{K}\mathbf{1}\in\R^K$ is the
      uniform probability vector.  Under the rate normalization above,
      the decay factor is
      \[
      \alpha_t
      =
      \exp\left(
      -\frac{K}{K-1}\int_0^t \beta(s)\,\diff s
      \right).
      \]
      As time passes, repeated randomization washes out the structure of
      the data distribution and pushes it toward the uniform distribution
      on~$\mathcal{V}$.

More generally, the forward process need not have the simple mixture form
\[
\Pr(\rvx_t|\rvx_0)
= \mathrm{Cat}\bigl(\rvx_t;\;\alpha_t\rvx_0+(1-\alpha_t)\bm{\pi}\bigr).
\]
That form is useful for intuition and appears in common masked and uniform
constructions, but the Markov law-evolution viewpoint is more general:
transition matrices govern discrete-time evolution, and rate matrices govern
continuous-time evolution through the master equation. Structured discrete diffusion models can use transition rules that respect edit distance, graph adjacency, molecular validity, vocabulary structure, or other domain knowledge. In all cases, the same principle remains: the forward Markov process defines a probability path, and generation learns to move along that path in reverse.

\subsection{Three Perspectives on Reverse Modeling}
\label{subsec:epilogue-three-views}

Once the forward process and its law evolution are in place, the remaining task is the same as in the continuous case: learn a reverse process whose marginal distributions retrace the forward evolution in reverse time, turning simple reference samples into data. In Part~B,
we organized continuous diffusion models into three perspectives:
variational (\Cref{ch:variational}), score-based
(\Cref{ch:score-based,ch:score-sde}), and flow-based
(\Cref{ch:flow-based}). The same three-way viewpoint carries over to the discrete setting, with one important caveat: these are not mutually exclusive schools or fixed paper categories. They are three ways of identifying what unknown object must be learned in order to reverse the probability path.

\paragraph{Variational Perspective: Learn Reverse Transition Kernels.}
In the continuous case (\Cref{sec:ddpm}), the variational viewpoint
treats diffusion as a hierarchical latent-variable model. The forward
process defines a chain of increasingly noisy latents, and the goal is
to learn reverse transitions that undo the corruption one step at a
time. The true reverse kernel $\Pr(\rvx_{k-1}|\rvx_k)$ is intractable
because it depends on the marginal law of $\rvx_k$, which involves the
unknown data distribution. The key insight of DDPM is to condition on
the clean sample $\rvx_0$: the forward posterior
$\Pr(\rvx_{k-1}|\rvx_k, \rvx_0)$ is Gaussian in closed form, and
Theorem~\ref{thm:equiv-marginal-kl} shows that training against this
conditional posterior is equivalent, up to terms independent of the learned reverse model, to training
against the intractable marginal reverse kernel. The ELBO then
decomposes into a sum of KL terms (\Cref{eq:ddpm-diffusion}), one per
noise level, each asking the learned model to match a tractable
forward posterior.

The same logic applies in the discrete setting. Write $p$ for the joint law
obtained by drawing $\rvx_0$ from the data distribution and then applying the
fixed forward chain, and write $p_{\bphi}$ for the learned reverse model. Consider a chain of
corrupted one-hot states
\[
\rvx_0 \;\xrightarrow{\;\rmP_0\;}\; \rvx_1 \;\xrightarrow{\;\rmP_1\;}\;
\cdots \;\xrightarrow{\;\rmP_{L-1}\;}\; \rvx_L,
\]
where $\rvx_0$ is drawn from the data distribution and $\rvx_L$ is close to
a simple reference distribution, such as all-mask or uniform.  The transitions
are now categorical rather than Gaussian, but the variational structure is the
same:
\begin{itemize}[leftmargin=2em]
    \item The true reverse $p(\rvx_{k-1}|\rvx_k)$ is intractable,
      because it depends on the marginal at step~$k$.
    \item Conditioning on $\rvx_0$ makes the forward posterior tractable
      via Bayes' rule:
      \[
      p(\rvx_{k-1}|\rvx_k, \rvx_0)
      =
      \frac{p(\rvx_k|\rvx_{k-1})\;p(\rvx_{k-1}|\rvx_0)}
             {p(\rvx_k|\rvx_0)},
      \]
      where every factor on the right is determined by the forward
      process.
    \item Training against these conditional posteriors has the same optimal
      reverse kernel as training against the intractable marginal reverse
      kernels; the difference consists of terms independent of $p_{\bphi}$.
\end{itemize}

Schematically, the discrete diffusion ELBO contains denoising terms of the form
\[
\E_{p(\rvx_{0:L})}
 \Big[
\mathcal{D}_{\mathrm{KL}} \big(
  p(\rvx_{k-1}|\rvx_k, \rvx_0)
  \;\|\;
  p_{\bphi}(\rvx_{k-1}|\rvx_k)
\big)
\Big],
\qquad k=1,\ldots,L.
\]
Depending on the exact model, the full ELBO also includes boundary terms such
as the terminal prior term and the reconstruction term near $k=0$.  The key
point is that each denoising term asks the learned reverse kernel to match a
tractable Bayesian posterior determined by the forward corruption.

Here $p_{\bphi}(\rvx_{k-1}=\rve_i|\rvx_k=\rve_j)$ denotes the neural-network
parameterized reverse transition probability: given that the system is in
state~$\rve_j$ at step~$k$, it gives the probability of having come from
state~$\rve_i$ at step~$k{-}1$.  For each fixed~$\rve_j$, these values form a
categorical distribution over previous states.

The main formal difference from continuous DDPM is that these KL divergences
are finite sums over discrete states rather than closed-form Gaussian
expressions. This reflects a special property of Gaussian distributions: the product of two Gaussian densities in the same variable is proportional to another Gaussian. In continuous DDPM, the
forward posterior is computed by multiplying Gaussian factors, so the result is itself Gaussian and
the KL divergence reduces to a simple algebraic expression involving
only means and covariances (\Cref{sec:ddpm}). In the discrete setting,
Bayes' rule yields a categorical distribution instead, and each KL term is a sum over states. This sum is exact and conceptually simple, but for large vocabularies or full sequence spaces it may require structure, factorization, subsampling, or specialized parametrization.

A practical point about parametrization is also worth noting. The ELBO asks the
learned reverse kernel $p_{\bphi}(\rvx_{k-1}|\rvx_k)$ to match a Bayesian
posterior, but it does not force one unique neural parametrization of that
kernel. One common choice is a clean-state, or $x_0$-prediction,
parametrization. The exact marginal reverse kernel satisfies the identity
\[
p(\rvx_{k-1}|\rvx_k)
=
\sum_{a}
p(\rvx_{k-1}|\rvx_k,\rvx_0=\rve_a)\,
p(\rvx_0=\rve_a|\rvx_k),
\]
where the sum is over valid clean categories. This is simply marginalization
over the unknown clean state. Since the posterior $p(\rvx_0|\rvx_k)$ depends on
the data distribution, it is not available in closed form. A model may therefore
predict a clean-state distribution $\widehat p_{\bphi}(\rvx_0|\rvx_k)$ and use
it to define
\[
p_{\bphi}(\rvx_{k-1}|\rvx_k)
:=
\sum_{a}
p(\rvx_{k-1}|\rvx_k,\rvx_0=\rve_a)\,
\widehat p_{\bphi}(\rvx_0=\rve_a|\rvx_k).
\]
If $\widehat p_{\bphi}(\rvx_0|\rvx_k)$ were equal to the true posterior
$p(\rvx_0|\rvx_k)$, this mixture would recover the exact reverse kernel. In
practice, it is a useful way to tie the learned reverse transition to the known
forward posterior.

This clean-state parametrization is common in multinomial/D3PM-style
discrete-time diffusion models~\citep{hoogeboom2021argmax,austin2021structured},
but it is not the definition of discrete diffusion. D3PM itself notes that one
could instead predict the reverse logits directly, and later models use other
structured parametrizations tailored to the forward process
\citep{sahoo2024simple}. Thus clean-state prediction should be viewed as one
important parametrization of the variational perspective, not as the only way to
model the reverse process.

\paragraph{Score/Ratio Perspective: Learn Local Probability Comparisons.}
In the continuous case (\Cref{ch:score-based,ch:score-sde}), the
score-based viewpoint starts from a simple question: given a corrupted
sample $\rvx_t \sim p_t$, what local information about the current density is needed to reverse the diffusion? The answer is the score function
$\nabla_{\rvx}\log p_t(\rvx)$, which appears in the reverse-time
SDE and the probability-flow ODE\@. Learning the score with a neural
network therefore suffices for generation.

In a finite state space, there is no ambient Euclidean geometry, no
ordinary notion of infinitesimal displacement, and no gradient with respect to the state. A sample does not move
by taking a small Euclidean step; it changes by jumping from one state to another.
So the score $\nabla_{\rvx}\log p_t(\rvx)$ has no direct literal analogue. But
the reverse process must still answer an analogous question: \emph{if the
system is currently in state~$\rve_j$, how likely is each candidate source state~$\rve_i$ relative to it?}

\subparagraph{Deriving the Reverse Jump Rate.}
To identify the needed quantity, consider a forward CTMC $\{\rvx_t\}_{0\leq t\leq T}$ with rate
matrix~$\rmQ_t$ and marginal $\rvp_t$.  The entry $[\rmQ_t]_{ij}$ is the jump rate from
state~$\rve_i$ to state~$\rve_j$.  Over a short interval~$\Delta t$, the
probability of the infinitesimal forward event
$\rvx_t=\rve_i$ and $\rvx_{t+\Delta t}=\rve_j$, for $i\neq j$, is approximately
\[
p_t(i)\,[\rmQ_t]_{ij}\,\Delta t.
\]

Now define the reverse-time process $\bar{\rvx}_s := \rvx_{T-s}$ for $0\leq s\leq T$. Its marginal at reverse time $s$ is $\bar{\rvp}_s=\rvp_{T-s}$. If the reverse process has rate matrix $\bar{\rmQ}_s$, then the corresponding reverse jump from $\rve_j$ to $\rve_i$ must reproduce the same infinitesimal path probability in the reversed temporal order. Taking the limit $\Delta t\to 0$ gives the off-diagonal time-reversal formula
\begin{mdframed}
\begin{equation}
[\bar{\rmQ}_s]_{ji}
= [\rmQ_{T-s}]_{ij}\;\frac{p_{T-s}(i)}{p_{T-s}(j)},
\qquad i \neq j,
\label{eq:reverse-rate-discrete-reverse-time}
\end{equation}
\end{mdframed}
whenever $p_{T-s}(j)>0$. Equivalently, if $t$ denotes the corresponding forward time and $\widetilde{\rmQ}_t$ denotes the reverse generator indexed by that forward time, then
\begin{equation}
[\widetilde{\rmQ}_t]_{ji}
= [\rmQ_t]_{ij}\;\frac{p_t(i)}{p_t(j)},
\qquad i \neq j.
\label{eq:reverse-rate-discrete}
\end{equation}
The diagonal entries are fixed by the row-sum-to-zero constraint,
\[
[\widetilde{\rmQ}_t]_{jj}
=
-\sum_{i\neq j}[\widetilde{\rmQ}_t]_{ji}.
\]
This is the standard time-reversal formula for CTMCs
\citep{kelly1979reversibility,anderson2012continuous}, applied to
discrete diffusion by \citet{campbell2022continuous}. The formula is understood only on states with positive marginal probability. Moreover, the
reverse process only reverses probability flux along transitions allowed by the
forward process: if $[\rmQ_t]_{ij}=0$, then the corresponding reverse rate from
$\rve_j$ to $\rve_i$ is also zero. The statement is not an equilibrium detailed-balance assumption; it is a local time-reversal identity for infinitesimal path probabilities.

\subparagraph{The Ratio as a Discrete Analogue of the Score.}
Now inspect \Cref{eq:reverse-rate-discrete}.  The forward rate
$[\rmQ_t]_{ij}$ is known, since the forward process is fixed by design.
The unknown quantity is the ratio $p_t(i)/p_t(j)$,
which compares how likely state~$\rve_i$ is relative to
state~$\rve_j$ under the current marginal.  In inner-product notation,
$p_t(i) = \langle\rvp_t,\rve_i\rangle$, so the ratio can be written as
$\langle\rvp_t,\rve_i\rangle / \langle\rvp_t,\rve_j\rangle$.

Why is this a discrete analogue of the score?  In continuous space, the
score $\nabla_{\rvx}\log p_t(\rvx)$ measures how the log-density changes
under an infinitesimal displacement.  In a finite state space, the
natural local comparison is instead across an allowed jump from one state
to another.  Along a jump from~$\rve_j$ to~$\rve_i$, the change in
log-probability is
\[
\log p_t(i) - \log p_t(j)
= \log \frac{p_t(i)}{p_t(j)},
\]
a finite difference of log-probabilities, playing the role that a
directional derivative of $\log p_t$ plays in~$\R^D$. This comparison is naturally \emph{edge-local}: the ratio is only needed for
pairs of states connected by the forward jump structure.  Thus, the discrete
score is not a Euclidean vector field over $\mathbb{R}^D$; it is a collection
of local probability comparisons along allowed transitions. For full sequences, the same statement applies with $i$ and $j$ replaced by full configurations, often neighboring configurations under a Hamming-type or graph-based transition structure.

\subparagraph{What the Network Learns.}
In continuous score-based models, a neural network
$\rvs_{\bm{\phi}}(\rvx, t)$ is trained to approximate the score
$\nabla_{\rvx}\log p_t(\rvx)$, and the learned score is plugged into
the reverse-time SDE to generate samples. The discrete case follows the
same logic at the level of information: the network must provide the time-dependent ratio information
needed to determine reverse jump rates via
\Cref{eq:reverse-rate-discrete}. Generation then proceeds by simulating
the reverse-time Markov chain or CTMC using the learned transition probabilities or rates.

Different formulations represent this information in different ways. Ratio-based models such as SEDD~\citep{lou2024discrete}
estimate ratios $p_t(i)/p_t(j)$ or their log-forms using losses inspired
by score matching. Other continuous-time formulations
\citep{campbell2022continuous,sun2023scorebased} learn equivalent
reverse quantities through objectives such as matching conditional transition probabilities. The idea of treating
local probability comparisons as a discrete analogue of the score is
formalized by the \emph{concrete score} of \citet{meng2022concrete}.
Despite these differences in parametrization, the underlying principle
is shared: learn the local probability comparisons required by time reversal, construct the
reverse transitions or rates, and simulate backward.

\paragraph{Flow Perspective: Learn a Discrete Transport Field.}
In the continuous case (\Cref{ch:flow-based}), the flow-based viewpoint
takes a different route: instead of learning the score and reversing an
SDE, one directly learns a velocity field $\rvv_t(\rvx)$ whose ODE flow
transports a reference distribution to the data distribution. The
training strategy is flow matching
(\Cref{sec:flow-matching-framework}): the marginal velocity field is
intractable, but a \emph{conditional} velocity field
$\rvv_t(\rvx | \rvx_0)$, conditioned on a data or endpoint sample, is available in closed form. The marginal and conditional
targets are related by
\[
\rvv_t(\rvx)
= \E \big[\rvv_t(\rvx | \rvx_0) | \rvx_t = \rvx\big],
\]
an average over all possible conditioning variables given the current state. Computing this average explicitly would require knowing the continuous marginal
density $p_t(\rvx)$.  The key training trick (\Cref{subsec:fm-instantiations}) is that,
under squared-error regression, the conditional target has the same population
minimizer as the intractable marginal target. Training therefore avoids explicit marginal computation by using samples from tractable conditional paths.

The discrete case follows the same idea, with rate matrices replacing velocity fields. Since a discrete state cannot
move along a continuous trajectory through token space, the role of the velocity field is played by a rate matrix, also called a
Markov generator, $\rmQ_t$. Recall that the entry $[\rmQ_t]_{ij}$ specifies
the jump rate from state~$\rve_i$ to state~$\rve_j$; the actual
probability mass sent from~$\rve_i$ to~$\rve_j$ per unit time is the flux
\[
J_t(i,j)=p_t(i)\,[\rmQ_t]_{ij}.
\]
In this sense, a rate matrix is a discrete transport field: it determines how probability mass is redistributed over time.

There is one important subtlety. A probability path $\{\rvp_t\}_{t\in[0,T]}$ does not uniquely determine a rate matrix. Many different generators can produce the same derivative $\diff\rvp_t/\diff t$ in the master equation, just as different transport fields can sometimes realize the same marginal evolution under different geometric constraints. A discrete flow-matching method therefore specifies not only a probability path, but also a rule for assigning conditional jump rates or probability currents along that path.

As in continuous flow matching, the marginal rate matrix is generally
intractable, but conditional versions can often be made explicit.  If we
condition on an endpoint, for example a clean data state and/or a reference state, one can prescribe a conditional probability
path and derive tractable conditional jump rates along that path. A neural
network is then trained, under an appropriate proper loss, to predict the corresponding transition or rate information from corrupted
samples. The marginal generator is recovered through the same conditional-expectation principle: average the conditional transport object over the unknown endpoint distribution given the current state, but learn this average by supervised regression or classification against conditional targets rather than computing it explicitly.

The parallel with continuous flow matching is therefore:
\begin{itemize}[leftmargin=2em]
    \item In continuous flow matching, the network is trained against a
      conditional velocity target; the marginal velocity emerges as the
      optimal predictor under the flow-matching loss.
    \item In discrete flow matching, the network is trained against conditional
      transition or jump-rate targets; the marginal discrete transport field
      emerges by the analogous conditional-expectation mechanism.
\end{itemize}
In both cases, training avoids the intractable marginal law by working
with tractable conditional paths and sampled intermediate states.

This perspective is developed by \citet{campbell2024generative} and
\citet{gat2024discrete} under the heading of discrete flow models or discrete flow matching. More broadly, the flow perspective should be read as a statement about \emph{what object is learned}: a transport field on a discrete state space, represented by transitions, rates, or probability currents.

\subsection{What Is Similar and What Is Genuinely Different?}
\label{subsec:epilogue-genuine-diff}

The following table summarizes the parallel between continuous and
discrete diffusion models. Reading across each row shows the same
conceptual ingredient instantiated in two different settings:

\begin{table}[!t]
\caption{\textbfs{Continuous and discrete views of probability transport.}
The same generative modeling ingredients appear in both continuous and finite
discrete state spaces, but the mathematical objects change: densities become
probability vectors, ODE/SDE dynamics become Markov chains or CTMCs, and
reverse modeling is expressed through transition kernels, score/ratio
information, or discrete transport fields.}
\label{tb:continuous-discrete-diffusion}
\centering
\begingroup
\scriptsize
\setlength{\tabcolsep}{4pt}
\renewcommand{\arraystretch}{1.08}
\begin{tabular}{p{0.15\linewidth} p{0.40\linewidth} p{0.40\linewidth}}
\toprule
\textbfs{State}
& \textbfs{Continuous} $(\rvx \in \R^D)$
& \textbfs{Discrete} $(\rvx \in \mathcal{V} \subset \{0,1\}^K)$ \\
\midrule

\textbfs{Law}
& Density $p_t(\rvx)$
& Probability vector $\rvp_t \in \Delta^{K-1}$ \\
\graymidrule

\textbfs{Sample \newline Dynamics}
& ODE or SDE trajectories
& Markov-chain or CTMC jumps \\
\graymidrule

\textbfs{Law \newline Evolution}
& Continuity or Fokker--Planck equation
& Transition-matrix propagation or master equation \\
\graymidrule

\textbfs{Transport \newline Object}
& Velocity, drift, diffusion, or score field
& Transition kernel, rate matrix, ratio, or probability current \\
\graymidrule

\textbfs{Reference \newline Law}
& Simple distribution, e.g.\ Gaussian
& Simple distribution, e.g.\ uniform or all-mask \\

\midrule
\multicolumn{3}{c}{\textbfs{Three Perspectives on Reverse Modeling}} \\
\midrule

\textbfs{Variational}
& ELBO with Gaussian or continuous transitions
& ELBO with categorical transitions \\
\graymidrule

\textbfs{Score/Ratio}
& Learn score $\nabla_{\rvx}\log p_t(\rvx)$
& Learn ratios or log-ratios, e.g.\ $p_t(i)/p_t(j)$ \\
\graymidrule

\textbfs{Flow}
& Learn velocity field $\rvv_t(\rvx)$
& Learn transition, rate, or current field \\

\bottomrule
\end{tabular}
\endgroup
\end{table}
\noindent
The parallel is clean, but the discrete setting
is not simply the continuous theory with gradients replaced by ratios. Several deeper differences are worth highlighting.

\paragraph{No Literal Inverse Map.}
In a deterministic continuous normalizing flow, the reverse map is literally the inverse of the forward map. In a discrete Markov process, the forward kernel may merge mass from many states and need not be invertible. The reverse process is therefore not obtained by matrix inversion. It is a Bayesian or time-reversal construction that depends on the current marginal law. This is why reverse transition probabilities involve factors such as $p_k(i)/p_{k+1}(j)$, and reverse jump rates involve ratios such as $p_t(i)/p_t(j)$.

\paragraph{Geometry and Sample Paths.}
In continuous Euclidean space, individual samples evolve along
continuous-time trajectories: ODE solutions are smooth, and SDE sample
paths are continuous even though they are typically nowhere
differentiable. This differential structure makes objects such as
velocity fields, score functions, and divergences meaningful. In a
finite state space, by contrast, an individual trajectory is a
\emph{pure-jump path}: it stays at one state for a random holding time
and then jumps abruptly to another. What evolves smoothly is not the
sample path itself, but the probability vector $\rvp_t$, which evolves
continuously in time on the probability simplex according to the master
equation. Discrete diffusion therefore still admits a unifying
description at the level of probability evolution, but its sample-level
dynamics are governed by jump processes rather than by continuous ODE/SDE
paths in the data space.

\paragraph{Forward Corruption Structure and Reverse Parametrization.}
In discrete diffusion, the forward corruption process is often a substantive modeling choice. In continuous Gaussian diffusion, one may vary the schedules $\alpha_t$ and $\sigma_t$, but the forward kernel often remains Gaussian, so the main effect is to control the rate and scale of corruption. In discrete diffusion, by contrast, the transition rule itself can change: one may use uniform corruption, masking, nearest-neighbor transitions, graph-structured transitions, domain-specific transitions, or more general probability paths. As emphasized by D3PM~\citep{austin2021structured}, the forward process can therefore incorporate domain-specific inductive bias directly, and this in turn shapes the reverse generation problem.

This difference also changes the status of reverse parameterizations. In the Gaussian setting, common targets such as noise, denoised data, score, and velocity are linked by simple time-dependent affine transformations. In the discrete setting, reverse transition probabilities, clean-state predictors, probability ratios, and jump rates are connected through normalization, Bayes' rule, and the chosen transition graph. They are related, but not merely interchangeable Gaussian-style reparameterizations. As a result, the variational, score/ratio, and flow descriptions are genuinely different views of the reverse problem, not just different names for the same algebraic target.

\paragraph{Sampling Mechanism.}
The sampling procedure also differs.  In continuous diffusion, generation is
usually implemented by numerically integrating an ODE or SDE over a chosen time
grid.  In discrete diffusion, generation means simulating a reverse Markov
chain or CTMC\@.  In discrete time, this amounts to sampling categorical reverse
transitions.  In continuous time, it may involve simulating jumps from learned
rates, for example by event-based simulation or time discretization.  Thus,
the role played by numerical ODE/SDE solvers in continuous diffusion is replaced
by categorical transition sampling or jump-process simulation in the discrete
setting.

\paragraph{Why Stop Here?}
Our goal in this section is not to develop the full machinery of discrete diffusion models, but to identify the principle that remains invariant across state spaces. The key point is structural: in continuous spaces, distributional evolution is governed by change-of-variable principles and their differential forms, including the continuity equation and the Fokker--Planck equation. In discrete spaces, the corresponding role is played by Markov operators and the master equation. In both settings, the modeling logic is the same: define a forward corruption process, understand how it transports the distribution, and then learn a reverse transport mechanism for generation. This logic does not depend on whether the underlying state space is continuous or discrete.

What changes are the mathematical objects used to realize this program, not the program itself. In continuous spaces, one works with densities, gradients, velocity fields, and SDE or ODE dynamics. In discrete spaces, one works with probability vectors, transition kernels, jump rates, ratios, and probability currents. These differences matter technically, and they lead to distinct questions of parameterization, training, and sampling. But they do not alter the deeper conclusion: the probability-transport viewpoint is not tied to a particular state space, architecture, or paper lineage. It is a general organizing principle broad enough to encompass both continuous and discrete diffusion within a single conceptual framework.

\clearpage
\newpage

\section{Closing Remarks of the Book}\label{sec:epilogue-closing}
Diffusion models may at first appear as a collection of acronyms, parametrizations, and algorithmic variants. Yet the central idea is simple. They provide a way to construct generators by prescribing how probability laws evolve over time. From this viewpoint, generation is not a mysterious black box, but the controlled transport of a cloud of particles under a forward process and its learned reverse.

Throughout this book, this viewpoint has appeared in several forms. In continuous state spaces, variational, score-based, and flow-based formulations provide different ways to describe and learn the reverse of a prescribed law evolution. Diffusion-motivated fast generators, such as flow-map models, belong to the same story as well: they do not abandon the transport viewpoint, but instead ask whether the underlying generative dynamics can be represented more directly and executed more efficiently. In discrete state spaces, the same structural idea reappears in the language of Markov kernels, jump processes, ratios, and master equations.

The main message of this book is therefore not that one particular formulation should replace all others. Rather, diffusion is best understood as a general principle for constructing generators from a prescribed forward process~\citep{lai2026unified}. Under this viewpoint, different state spaces naturally lead to different reverse representations. In continuous settings, one may work with conditional Gaussians, score functions, velocity fields, or more direct flow maps. In discrete settings, one may work with reverse transition probabilities, clean-state predictors, probability ratios, jump rates, or probability currents associated with the underlying Markov process. What matters is not the label of the parametrization, but whether the forward process, the induced evolution of probability laws, and the learned generative mechanism fit together coherently.

We hope this book has made that structure clear: a forward process induces an evolution of probability laws, and generative modeling amounts to learning how to reverse, approximate, or exploit that evolution. Specific methods will continue to change, but this viewpoint gives a stable way to understand why they work, how they relate, and where new developments are likely to come from.

\epigraph{\textit{The important thing is not to stop questioning. Curiosity has its own reason for existence.}}{Albert Einstein}



\appendix

\chapter{Crash Course on Differential Equations}\label{app:de}

Differential equations (DEs) are fundamental tools for modeling dynamic systems and can be broadly categorized into \emph{ordinary differential equations} (ODEs), \emph{stochastic differential equations} (SDEs), and \emph{partial differential equations} (PDEs). 

ODEs describe how a system’s state changes over time according to a precise rule, so that knowing the starting point determines the future path exactly. SDEs add randomness to this evolution, modeling how noise or uncertainty influences the system’s behavior, making the outcome probabilistic rather than fixed. PDEs explain how functions depending on several variables, such as time and space, evolve together, capturing phenomena like heat spreading, waves moving, or \emph{the time evolution of probability densities in stochastic systems} (Spoiler: Fokker-Planck equation). These types of differential equations form a fundamental language for understanding how systems evolve over time and space under both deterministic and random influences.

In this chapter, we provide essential prerequisites on differential equations.

\chapterlearning{
  \item interpret an ODE through its velocity field, solution trajectory, and the role of existence and uniqueness;

  \item use integrating factors to expose the linear structure of semilinear ODEs and understand why this structure is useful for diffusion ODE solvers;

  \item distinguish single-step, multistep, and parallel-in-time numerical viewpoints, and recognize Euler, Runge--Kutta, Adams--Bashforth, and Picard iteration as representative examples;

  \item understand how an SDE extends deterministic dynamics with Brownian noise, how its differential notation relates to the It\^o integral, and how Euler--Maruyama discretizes the resulting stochastic dynamics.
}{
This appendix provides the differential-equation background used throughout
the book. Focus on the connection between dynamics, trajectories, and
numerical approximation; the later diffusion chapters will specialize these
ideas to probability-flow ODEs and stochastic diffusion processes.
}
\clearpage
\newpage

\section{Foundation of Ordinary Differential Equations}

This section introduces the fundamental theory of ODEs, emphasizing the uniqueness of solutions given an initial condition. It also covers practical methods for solving ODEs using numerical solvers.

\subsection{Intuition of Ordinary Differential Equation}

The deterministic process is called an \emph{ordinary differential equation} (ODE). In the multivariate case, we consider systems of the form:
\begin{equation}\label{eq:general_ode}
    \frac{\diff \mathbf{x}(t)}{\diff t} = \mathbf{v}(\mathbf{x}(t), t),
\end{equation}
where $\mathbf{x}(t) \in \mathbb{R}^D$ is a vector-valued function representing the state of the system at time $t$, and $\mathbf{v}: \mathbb{R}^D \times \mathbb{R} \to \mathbb{R}^D$ is a vector field specifying the direction and magnitude of change at each point in space and time.

\paragraph{High-Level Intuition for Solving ODEs.}
\begin{figure}[th]
    \centering
    \includegraphics[width=\linewidth]{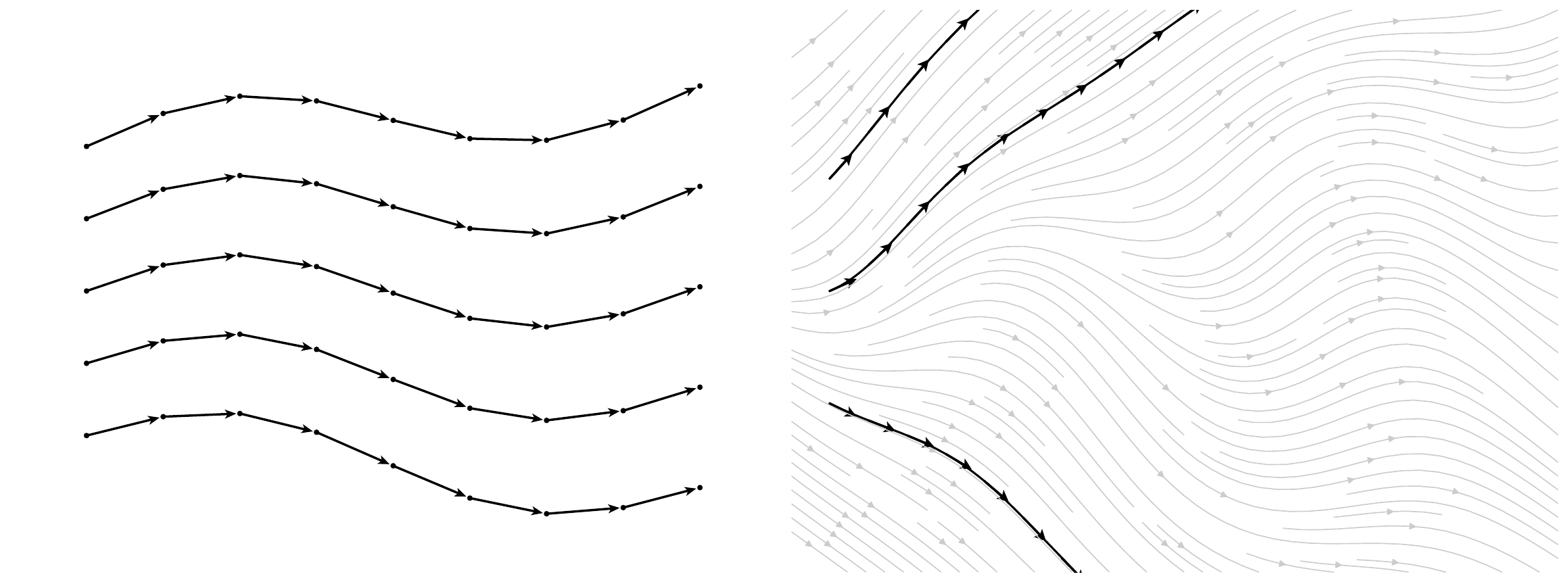}
    \caption{\textbfs{ODE illustration.}  
A velocity field $\rvv(\rvx,t)$ assigns a drift vector at every point.  
A solution trajectory $\rvx(t)$ is a path whose tangent always matches the local drift.  
The left panel shows step-by-step solver updates (dots and arrows) approximating the path,  
while the right panel shows the exact trajectories (black) flowing consistently with the velocity field. Without specifying an initial state $\rvx(0)$, there are infinitely many trajectories whose instantaneous changes match the same velocity field.  
Once $\rvx(0)$ is fixed, however, the ODE determines a unique path $\rvx(t)$ that flows according to the drift.  
\figcredit{Created by the authors.}}
    \label{fig:ode-illustration}
\end{figure}

To build intuition, imagine the vector field $\mathbf{v}(\mathbf{x}, t)$ as a dynamic landscape of arrows that tells you how a point $\mathbf{x}$ should move at any given time $t$. Solving the differential equation means tracing out a curve $\mathbf{x}(t)$ through this field such that the tangent (i.e., the instantaneous velocity) of the curve at any point aligns with the vector given by $\mathbf{v}(\mathbf{x}(t), t)$.
\begin{itemize}
    \item \textbfs{Vector Field Perspective:} The function $\mathbf{v}(\mathbf{x}, t)$ defines how things should move: it gives the local ``instructions'' for motion or change.
    \item \textbfs{Trajectory Perspective:} The solution $\mathbf{x}(t)$ is a path that a particle would follow if it obeys the rule set by the vector field $\mathbf{v}$ at every instant.
\end{itemize}
Thus, solving an ODE is like placing a particle in a flow field and observing where it goes over time.

\subsection{Existence and Uniqueness of Ordinary Differential Equations}
So far, we have seen that solving an ODE means finding a path that follows the directions given by the vector field at every point. Intuitively, this is like tracing the trajectory of a particle as it moves along the flow defined by the velocities.

But this picture leads to an important question: 
\begin{question}
    If we pick a starting point, can we be sure there really is a path that follows these directions? And if there is, is that path unique, or could the particle suddenly jump onto a different trajectory?
\end{question}

Answering these questions is essential because it tells us whether the system’s behavior can be reliably predicted from its starting position.  
The \emph{Existence and Uniqueness Theorem} provides conditions on the vector field that guarantee exactly one path starting from any given initial point.  
This ensures the solution behaves consistently and forms a cornerstone of the theory of ODEs.

\paragraph{Local (in Time) Existence and Uniqueness Theorem.}

Below, we state a \emph{local} version of the theorem, which asserts existence and uniqueness of a solution in a neighborhood of the initial time for a given initial condition.

\thmp{Local Existence and Uniqueness}{ode-local}{
Let $\mathbf{v}(\mathbf{x}, t)$ be a continuous function with respect to $\mathbf{x}$ and $t$ in a domain $D \subseteq \mathbb{R}^D \times \mathbb{R}$. If $\mathbf{v}$ satisfies the Lipschitz condition with respect to $\mathbf{x}$:
\[
\|\mathbf{v}(\mathbf{x}_1, t) - \mathbf{v}(\mathbf{x}_2, t)\| \leq L \|\mathbf{x}_1 - \mathbf{x}_2\| \quad \forall (\mathbf{x}_1, t), (\mathbf{x}_2, t) \in D,
\]
where $L > 0$ is a constant, then for every initial condition $\mathbf{x}(t_0) = \mathbf{x}_0$, there exists a unique solution $\mathbf{x}(t)$ to \Cref{eq:general_ode} defined on some interval $[t_0 - \delta, t_0 + \delta]$.}{(Proof Outline) The Existence and Uniqueness Theorem can be demonstrated constructively using the Picard-Lindel\"of iteration method. The method generates a sequence of functions $\{\mathbf{x}_n(t)\}$ that converges to the solution $\mathbf{x}(t)$. The iteration is defined as:
\[
\mathbf{x}_{n+1}(t) = \mathbf{x}_0 + \int_{t_0}^t \mathbf{v}(\mathbf{x}_n(s), s) \diff s.
\]
\begin{itemize}
    \item Start with an initial guess $\mathbf{x}_0(t) = \mathbf{x}_0$.
    \item Iteratively refine $\mathbf{x}_n(t)$ using the integral form.
    \item Convergence is guaranteed under the Lipschitz condition by applying Contraction Mapping Theorem.
\end{itemize}
}
The essence of the proof is rooted in the Picard–Lindelöf iteration method, whose core idea is also leveraged in \Cref{sec:picard} to accelerate the sampling process of diffusion models.

\paragraph{Global (in Time) Existence and Uniqueness Theorem.}
While the Local Existence and Uniqueness Theorem guarantees the existence of solutions on a small time interval, the ``global (in time) existence and uniqueness theorem'' extends this result to the entire interval $[t_0, T]$ under additional regularity conditions. A well-known result in this category is the \emph{Carathéodory theorem}, which ensures the global existence and uniqueness of solutions to ODEs under two key assumptions: local Lipschitz continuity in the state variable and a linear growth bound.

\begin{enumerate}[label=(\roman*)]
    \item \textbfs{Local Lipschitz condition in $\mathbf{x}$:} There exists a function $\mathrm{Lip}(t)$, integrable on $[0, T]$, such that for all $\mathbf{x}_1, \mathbf{x}_2 \in \mathbb{R}^D$,
    \[
    \|\rvv(\mathbf{x}_1, t) - \rvv(\mathbf{x}_2, t)\| \leq \mathrm{Lip}(t) \|\mathbf{x}_1 - \mathbf{x}_2\|.
    \]
    
    \item \textbfs{Linear growth condition:} There exists a function $M(t)$, integrable on $[0, T]$, such that for all $\mathbf{x} \in \mathbb{R}^D$,
    \[
    \|\rvv(\mathbf{x}, t)\| \leq M(t)(1 + \|\mathbf{x}\|).
    \]
\end{enumerate}
We refer the reader to \citep{reid1971ordinary} for a comprehensive discussion of the assumptions, formal statement, and detailed proof of the theorem.

\rmkb{To apply these theorems to the probability flow ODE in diffusion models (see \Cref{eq:prob_ode_gt}), it may be necessary to impose additional assumptions, such as conditions (i) and (ii), on the score function $\nabla_\rvx \log p_t(\rvx)$. These assumptions can be reasonably accepted without further justification by readers not focused on technical details.
}

In summary, when an initial condition is given to an ODE defined by a time-dependent velocity field, the trajectory of the particle flow is uniquely determined.

\paragraph{Uniqueness Implies Non-Intersection of Solutions} The uniqueness of solutions in ODEs, as guaranteed by the Local Existence and Uniqueness Theorem, implies a fundamental property: two different solution trajectories, starting from different initial conditions, cannot cross each other. This reflects the deterministic nature of ODEs, ensuring that each state evolves along a unique path. The following corollary formalizes this result.

\corp{Non-Intersection of Solutions}{
Consider two solutions $\rvx_1(t)$ and $\rvx_2(t)$ to the ODE
\[
\frac{\diff \rvx(t)}{\diff t} = \rvv\left(\rvx(t),t\right), \quad t \in [0,T].
\]
Suppose they have distinct initial values $\rvx_1(0) \neq \rvx_2(0)$. Then, these solutions do not intersect on $[0,T]$, i.e.,
\[
\rvx_1(t) \neq \rvx_2(t) \quad \text{for all } t \in [0,T].
\]
}{Assume, for the sake of contradiction, that there exists some $t^* \in (0,T]$ such that
\[
\mathbf{x}_1(t^*) = \mathbf{x}_2(t^*).
\]
Define the first time at which the two solutions meet as
\[
t_0 := \inf \{ t \in [0,T]|\mathbf{x}_1(t) = \mathbf{x}_2(t) \}.
\]
Since $\mathbf{x}_1(0) \neq \mathbf{x}_2(0)$ and $t^*$ is contained in this set, it follows that $t_0 > 0$. By continuity of $\mathbf{x}_1$ and $\mathbf{x}_2$, we have
\[
\mathbf{x}_1(t_0) = \mathbf{x}_2(t_0).
\]

Consider the initial value problem
\[
\frac{\mathrm{d}\mathbf{x}(t)}{\mathrm{d}t} = \mathbf{v}(\mathbf{x}(t), t), \quad \mathbf{x}(t_0) = \mathbf{x}_1(t_0).
\]
By the uniqueness theorem for ODEs, both $\mathbf{x}_1$ and $\mathbf{x}_2$ must coincide on the interval $[t_0, T]$. Applying uniqueness backward in time similarly implies that the two solutions coincide on $[0, t_0]$.
Therefore, the solutions satisfy
\[
\mathbf{x}_1(t) = \mathbf{x}_2(t) \quad \text{for all } t \in [0, T],
\]
which contradicts the assumption that $\mathbf{x}_1(0) \neq \mathbf{x}_2(0)$. Hence, we conclude that
\[
\mathbf{x}_1(t) \neq \mathbf{x}_2(t) \quad \text{for all } t \in [0, T].
\]

}

By guaranteeing non-intersecting solution paths, this theorem offers hidden yet crucial support for the flow map model (see \Cref{ch:distillation,ch:fast-scratch}).

\subsection{Exponential Integration Factor}\label{app:exp-int-factor}
Even an ODE determined by a general time-varying velocity $\rvv$ does not admit a closed-form solution, in some special cases, we can solve them analytically or by reducing its formulation to a better structural one.  
\paragraph{An Illustrative Example.}
Consider the following linear scalar ODE:
\[
\frac{\diff \rvx(t)}{\diff t} = L(t)\rvx(t),
\]
where $L(t) \in \mathbb{R}$ is a continuous function. This equation is solvable in closed form, and its solution is well known (for any $s$ and $t$):
\[
\rvx(t) = \rvx(s) \cdot \exp\left( \int_s^t L(\tau) \,\diff \tau \right).
\]
This formula demonstrates how the solution evolves according to an exponential factor that accumulates the effect of the time-dependent coefficient $L(t)$. This motivates the use of \emph{exponential integration factors}:
\begin{align}\label{eq:exp-int-factor}
    \mathcal{E}(s \shortrightarrow t) := \exp\left( \int_s^t L(\tau)\diff \tau \right),
\end{align}
especially in more general settings where the dynamics include both linear and nonlinear components.

\paragraph{Semilinear ODEs and Exponential Integration Factors.}
We now consider a broader class of ODEs known as \emph{semilinear} ODEs. These equations separate the dynamics into a linear part (in the state variable) and a nonlinear remainder:
\begin{align}\label{eq:semilinear-ode}
    \frac{\diff \rvx(t)}{\diff t} = L(t)\rvx(t) + \rmN(\rvx(t), t),
\end{align}
where $\rvx(t) \in \mathbb{R}^D$ is the state vector, $L(t)$ is a scalar-valued continuous function, and $\rmN: \mathbb{R}^D \times [0,T] \to \mathbb{R}^D$ is a nonlinear vector field. This semilinear structure arises naturally in many physical and engineering systems. In particular, it also appears in the probability flow ODE formulation of diffusion models (see \Cref{eq:prob_ode_gt}). Recognizing this structure enables integrating-factor representations that
handle the known linear dynamics analytically and leave the remaining term to
be approximated numerically. This principle plays a central role in the
diffusion ODE solvers studied in \Cref{ch:solvers}.

\subparagraph{Step 1: Isolate the Non-Linear Term via an Integration Factor.}
Observing that we can isolate the nonlinear part by subtracting the linear drift from the semiliner ODE in \Cref{eq:semilinear-ode}:
\[
\frac{\diff \rvx(t)}{\diff t} - L(t)\rvx(t) = \rmN(\rvx(t), t).
\]
To absorb the linear term, we multiply both sides by the inverse integration factor:
\[
\mathcal{E}^{-1}(s \shortrightarrow t) = \exp\left(-\int_s^t L(\tau)\diff \tau \right).
\]
Now apply the product rule to the left-hand side:
\begin{align*}
\mathcal{E}^{-1}(s \shortrightarrow t) \left( \frac{\diff \rvx(t)}{\diff t} - L(t)\rvx(t) \right)
&= \frac{\diff}{\diff t} \left[ \mathcal{E}^{-1}(s \shortrightarrow t) \rvx(t) \right].
\end{align*}
Hence, the equation becomes:
\[
\frac{\diff}{\diff t} \left[ \mathcal{E}^{-1}(s \shortrightarrow t) \rvx(t) \right]
= \mathcal{E}^{-1}(s \shortrightarrow t) \rmN(\rvx(t), t).
\]

This transformation simplifies the original equation by isolating the nonlinear component, allowing us to focus entirely on the nonlinear dynamics in a transformed coordinate system.

\subparagraph{Step 2: Integrate Over Time.}
We now integrate both sides from $s$ to $t$:
\[
\int_s^t \frac{\diff}{\diff \tau} \left[ \mathcal{E}^{-1}(s \shortrightarrow \tau) \rvx(\tau) \right] \diff \tau
= \int_s^t \mathcal{E}^{-1}(s \shortrightarrow \tau) \rmN(\rvx(\tau), \tau)\diff \tau.
\]
The left-hand side is simply the difference of the transformed variable evaluated at $t$ and $s$:
\[
\mathcal{E}^{-1}(s \shortrightarrow t) \rvx(t) - \rvx(s).
\]
Hence, we obtain:
\[
\mathcal{E}^{-1}(s \shortrightarrow t) \rvx(t)
= \rvx(s) + \int_s^t \mathcal{E}^{-1}(s \shortrightarrow \tau) \rmN(\rvx(\tau), \tau)\diff \tau.
\]

\subparagraph{Step 3: Solve for $\rvx(t)$.}
Multiplying both sides by the exponential flow $\mathcal{E}(s \shortrightarrow t)$ gives the solution:
\begin{align}\label{eq:exp-int-factor-integration}
    \rvx(t) = \underbrace{\mathcal{E}(s \shortrightarrow t)  \rvx(s)}_{\text{linear part}} + \underbrace{\int_s^t \mathcal{E}(\tau \shortrightarrow t) \rmN(\rvx(\tau), \tau)\diff \tau}_{\text{nonlinear part}} .
\end{align}

The solution naturally separates into a linear and a nonlinear component. Exponential integrators exploit this structure by solving the linear part in exactly closed form and discretizing only the nonlinear residual. This ensures that the step size is dictated by the nonlinear dynamics rather than by the potentially large linear coefficient (see, for instance, the comparison between the exponential Euler update and the vanilla Euler update in \Cref{sec:prologue-solvers}).

\subsection{Numerical Solvers of Ordinary Differential Equations}
\label{app:de-solver}

We consider the ODE in \Cref{eq:general_ode},
\[
\frac{\diff\mathbf{x}(t)}{\diff t}
=
\mathbf{v}(\mathbf{x}(t),t),
\]
together with an initial condition. Solving the ODE means finding the
continuous trajectory $\mathbf{x}(t)$ whose instantaneous change agrees with
the velocity field $\mathbf{v}$ at every time.

Equivalently, the solution satisfies the integral equation
\begin{equation}
\label{eq:integral_form}
\mathbf{x}(t)
=
\mathbf{x}(0)
+
\int_0^t
\mathbf{v}(\mathbf{x}(\tau),\tau)\diff\tau.
\end{equation}
If this integral cannot be evaluated in closed form, we approximate the
trajectory numerically by evaluating the velocity field at finitely many
points.

\paragraph{From a Continuous Trajectory to Numerical Updates.}
A numerical solver first chooses a discrete time grid
\[
t_0,t_1,\ldots,t_M
\]
and produces corresponding approximations
\[
\mathbf{x}_0,\mathbf{x}_1,\ldots,\mathbf{x}_M
\]
to the continuous trajectory.

The gap
\[
h_i:=t_{i+1}-t_i
\]
is the \emph{step size}. Smaller steps generally provide more local
information about the trajectory but require more function evaluations.
The purpose of a numerical solver is therefore to use the available
evaluations of $\mathbf{v}$ efficiently so that accurate trajectories can be
obtained with relatively few steps.

Different solvers mainly differ in \emph{where the information used for the
next update comes from}. This leads to the following useful organization.

\paragraph{A Structural View of Numerical Solvers.}
For the methods relevant to this book, it is useful to distinguish two broad
numerical viewpoints.

\begin{itemize}
    \item \textbfs{Time-Stepping Methods:}
    Starting from the current state, the solver advances the trajectory from
    one time point to the next. Euler, Runge--Kutta, and Adams--Bashforth
    methods are representative examples.

    Within time-stepping methods, we further distinguish:
    \begin{itemize}
        \item a \emph{single-step method}, which constructs the next state
        from the current state and any additional evaluations generated
        within the current interval; and

        \item a \emph{multistep method}, which additionally reuses
        information from previous completed steps.
    \end{itemize}

    \item \textbfs{Parallel-in-Time Viewpoint:}
    Instead of completing all time points strictly one after another, one may
    formulate the trajectory computation so that tentative states at several
    times can be updated together. Fixed-point constructions such as Picard
    iteration provide a basic mechanism for this idea.
\end{itemize}

A single-step method should not be confused with a single
\emph{function evaluation}. For example, Euler uses one velocity evaluation
per step, whereas a Runge--Kutta method may evaluate the velocity several
times at intermediate \emph{stages} while still being a single-step method.
The defining distinction is whether information from earlier completed steps
is required.

\msgu{Mental Picture}{}{
The numerical ODE solver ideas used later for diffusion sampling can be organized as follows:
\begin{align}\label{eq:classic-solvers}
    \begin{array}{c}
\begin{array}{rcl}
\text{\textbfs{Time-stepping}}
&\rightarrow&
\begin{cases}
\text{single-step: }
    & \text{Euler, Runge--Kutta},\\
\text{multistep}
    & \text{Adams--Bashforth},
\end{cases}
\\[12pt]
\text{\textbfs{Parallel-in-time}}
&\rightarrow&
\text{Picard fixed-point iteration}.
\end{array}
\\[6pt]
\downarrow
\\[-2pt]
\text{The same numerical ideas can be applied forward or backward in time.}
\end{array}
\end{align}
}

The distinction is especially useful later in diffusion sampling:
Runge--Kutta-type constructions obtain richer information by looking
\emph{within} the current step, whereas Adams--Bashforth-type constructions
obtain it by looking \emph{backward} at previously computed values.

\paragraph{Solving ODEs in Forward and Reverse Time.}
The taxonomy above does not depend on the direction in which the ODE is
integrated. In ordinary forward integration,
\[
t_{n+1}>t_n,
\qquad
h_n=t_{n+1}-t_n>0.
\]
Diffusion sampling instead typically starts from a noisy terminal state and
uses a decreasing time grid,
\[
t_{n+1}<t_n,
\qquad
h_n=t_{n+1}-t_n<0.
\]

The same numerical formulas apply in either direction when the step size is
treated as a signed quantity. Thus, reverse-time integration of an ODE does
not require a new family of numerical solvers; Euler, Runge--Kutta,
multistep, and related methods can all be applied on a decreasing time grid.

This simple reversal should be distinguished from reversing an SDE, where
the stochastic dynamics require additional analysis. We discuss that issue
in the following section.

\subsection{Representative Numerical Solvers}
\label{app:de-solver-examples}

The taxonomy above tells us where a solver obtains information for its next
update. We now make these ideas concrete through a few representative
examples.

We deliberately keep the list short. Euler illustrates the simplest
single-step update; Runge--Kutta methods illustrate the use of additional
stages within the current step; Adams--Bashforth illustrates reuse of past
history; and Picard iteration illustrates the fixed-point viewpoint behind
parallel-in-time methods.

Throughout the examples below, let
\[
t_{n+1}=t_n+h.
\]
The step size $h$ is signed: $h>0$ for forward integration and $h<0$ for
reverse integration.

\paragraph{Euler's Method.}
Euler's method is the simplest single-step method. It uses only the velocity
at the beginning of the current interval:
\[
\mathbf{x}_{n+1}
=
\mathbf{x}_n
+
h\,\mathbf{v}(\mathbf{x}_n,t_n).
\]
Geometrically, Euler looks at the current direction of motion and follows
that direction for one complete step.

Under standard smoothness assumptions, Euler has local truncation error
$\mathcal O(h^2)$ and global error $\mathcal O(h)$. Its simplicity makes it
a useful first-order baseline, but its accuracy deteriorates when the
velocity changes substantially over one step.

\paragraph{Runge--Kutta Methods.}
Euler uses only the velocity at the beginning of the interval. Runge--Kutta
(RK) methods obtain more information by evaluating the velocity at additional
\emph{stages} generated within the same step.

They are therefore \emph{single-step, multi-stage} methods: all information
used to construct $\mathbf{x}_{n+1}$ is generated from the current starting
point $\mathbf{x}_n$, but one step may contain several function evaluations.

We illustrate the basic RK principle using two second-order methods.

\subparagraph{Explicit Midpoint Method.}
The explicit midpoint method first uses the current velocity to predict the
state halfway through the interval:
\[
\mathbf{x}_{\mathrm{mid}}
=
\mathbf{x}_n
+
\frac{h}{2}
\mathbf{v}(\mathbf{x}_n,t_n).
\]
It then evaluates the velocity at this predicted midpoint and uses that
direction for the complete update:
\[
\mathbf{x}_{n+1}
=
\mathbf{x}_n
+
h\,
\mathbf{v}
\left(
\mathbf{x}_{\mathrm{mid}},
t_n+\frac{h}{2}
\right).
\]

Thus, instead of assuming that the starting velocity remains representative
over the entire interval, the midpoint method ``looks inside'' the step
before deciding how to complete it.

\subparagraph{Heun's Method (Improved Euler).}
Heun's method is another two-stage, second-order Runge--Kutta method.
Instead of looking at the midpoint, it first uses Euler's method to predict
the endpoint:
\[
\mathbf{x}_{\mathrm{pred}}
=
\mathbf{x}_n
+
h\,\mathbf{v}(\mathbf{x}_n,t_n).
\]
It then evaluates the velocity at this predicted endpoint and averages the
starting and predicted ending velocities:
\[
\mathbf{x}_{n+1}
=
\mathbf{x}_n
+
\frac{h}{2}
\left[
\mathbf{v}(\mathbf{x}_n,t_n)
+
\mathbf{v}(\mathbf{x}_{\mathrm{pred}},t_n+h)
\right].
\]

The explicit midpoint and Heun methods both use two stages and, under
standard smoothness assumptions, have local truncation error
$\mathcal O(h^3)$ and global error $\mathcal O(h^2)$. They differ only in
how the additional within-step information is collected and combined.

Higher-order Runge--Kutta methods extend the same principle by using more
stages; their standard formulas are omitted here. In diffusion sampling,
\citet{karras2022elucidating} use Heun's method directly as an ODE solver.

\paragraph{Adams--Bashforth Method.}
Runge--Kutta methods obtain additional information by looking
\emph{within} the current step. A multistep method instead looks
\emph{backward} and reuses information from previous completed steps.

For example, with a uniform step size $h$, the second-order
Adams--Bashforth method (AB2) is
\[
\mathbf{x}_{n+1}
=
\mathbf{x}_n
+
h
\left[
\frac{3}{2}
\mathbf{v}(\mathbf{x}_n,t_n)
-
\frac{1}{2}
\mathbf{v}(\mathbf{x}_{n-1},t_{n-1})
\right].
\]
The second piece of information is therefore a stored velocity from the
previous step rather than a new evaluation inside the current interval.

Under standard smoothness assumptions and sufficiently accurate
initialization, AB2 has local truncation error $\mathcal O(h^3)$ and global
error $\mathcal O(h^2)$. Because the first step has no previous history, a
multistep method must first be initialized using another solver.

The formula above assumes a uniform grid. On nonuniform grids, the
multistep coefficients depend on the neighboring step sizes. This
history-reuse principle will later reappear in the diffusion-specific
multistep solvers in
\Cref{sec:exponential,subsec:dpm-solver++-multi}.

\paragraph{Picard Iteration.}
The methods above approximate the trajectory by constructing local
time-stepping updates. Picard iteration follows a different idea: it
repeatedly refines an entire candidate trajectory through the integral form
of the ODE.

Starting from an initial guess
$\mathbf{x}^{(0)}(t)$ satisfying
$\mathbf{x}^{(0)}(0)=\mathbf{x}(0)$, Picard iteration defines
\[
\mathbf{x}^{(k+1)}(t)
=
\mathbf{x}(0)
+
\int_0^t
\mathbf{v}
\bigl(
\mathbf{x}^{(k)}(s),s
\bigr)
\diff s.
\]
Each iteration therefore evaluates the velocity along the previous candidate
trajectory and uses those evaluations to construct a new one.

A fixed point of this iteration satisfies the original ODE integral equation.
Under the standard conditions of the Picard--Lindel\"of theorem, the
iteration converges locally to the ODE solution.

For our later purposes, the important observation is computational:
within one Picard iteration, the velocity at each time depends on the
\emph{previous} candidate trajectory. Evaluations at different time points
can therefore be carried out independently, which opens the possibility of
parallel-in-time computation. This viewpoint will be revisited in
\Cref{sec:picard}.

\newpage
\clearpage

\section{Foundation of Stochastic Differential Equations}\label{app:sde-intro}

Stochastic Differential Equations (SDEs) are an extension of ordinary differential equations (ODEs) that incorporate randomness, providing a mathematical framework for modeling systems affected by uncertainty. This chapter introduces SDEs, beginning with the discretization of ODEs, extending to the discretization of SDEs, and culminating in a discussion of general SDEs, including Ito's calculus and Ito's formula.

\subsection{From ODEs to SDEs: An Intuitive Introduction}
Let us begin with a ODE describing the deterministic evolution of a state variable $ \rvx(t) \in \mathbb{R}^D $:
\begin{equation}
    \frac{\diff\rvx(t)}{\diff t} = \rvf(\rvx(t), t), \quad \rvx(0) = \rvx_0.
    \label{eq:ode}
\end{equation}
Here, $ \rvf: \mathbb{R}^D \times [0,T] \rightarrow \mathbb{R}^D $ is a time-dependent velocity field that governs the dynamics of $ \rvx(t) $. The solution to this ODE is a smooth trajectory $ t \mapsto \rvx(t) $, fully determined by the initial condition $ \rvx_0 $.

\paragraph{Discretization Perspective.} To build intuition, consider an Euler discretization of \Cref{eq:ode} over small time steps $ \Delta t $:
\begin{equation*}
    \rvx_{t+\Delta t} = \rvx_t + \rvf(\rvx_t, t) \Delta t.
\end{equation*}
This approximation becomes more accurate as $ \Delta t \to 0 $, converging (under standard regularity conditions on $ \rvf $) to the exact solution of the ODE.

\paragraph{Introducing Randomness: From ODE to SDE.} In many real-world systems, perfect knowledge of the dynamics is unrealistic. Noise, uncertainty, or unmodeled interactions may affect the evolution. To incorporate such randomness, we augment the ODE with a stochastic term:
\begin{equation}
    \rvx_{t+\Delta t} = \rvx_t + \rvf(\rvx_t, t)\Delta t + g(t)\sqrt{\Delta t} \cdot \bm{\epsilon}_t,
    \label{eq:euler-sde}
\end{equation}
where
\begin{itemize}
    \item $g: [0,T] \to \mathbb{R}$ is a diffusion coefficient (possibly dependent on both state and time, though here assumed time-dependent only),
    \item $\bm{\epsilon}_t \sim \mathcal{N}(\bm{0}, \rmI_D)$ are i.i.d. standard Gaussian vectors.
\end{itemize}

This modified update rule reflects not just deterministic drift, but also random perturbations scaled by $ \sqrt{\Delta t} $. The scaling ensures that the stochastic perturbation remains finite in the limit $ \Delta t \to 0 $. Importantly, this formulation gives rise to a \emph{continuous-time stochastic process} as $ \Delta t \to 0 $, which leads us to the framework of SDE.

\paragraph{Stochastic Differential Equations.}
Formally, the limit of the discrete update~\Cref{eq:euler-sde} as $ \Delta t \to 0 $ defines the SDE:
\begin{equation}
    \diff \rvx(t) = \rvf(\rvx(t), t)\diff t + g(t) \diff \rvw(t).
    \label{eq:sde}
\end{equation}
Here, $ \rvw(t) \in \mathbb{R}^D $ is a \emph{Wiener process} (standard Brownian motion), a continuous-time stochastic process characterized by:
\begin{itemize}
    \item \textbfs{Initial State:} $ \rvw(0) = 0 $ almost surely;
    \item \textbfs{Independent Increments:}  for $ 0 \leq s < t $, the increment $ \rvw(t) - \rvw(s) $ is independent of the past;
    \item \textbfs{Gaussian Increments:} 
    \begin{align}\label{eq:wiener-gaussian}
        \rvw(t) - \rvw(s) \sim \mathcal{N}\bigl(\bm{0}, (t - s)\rmI_D \bigr)
    \end{align}
    \item \textbfs{Continuity:} Sample paths $ t \mapsto \rvw(t) $ are almost surely continuous but nowhere differentiable.
\end{itemize}
In addition, the notation
\[
\diff \rvw(t) := \rvw(t + \diff t) - \rvw(t)
\]
is often used to denote the infinitesimal increment of the Wiener process. 

While suggestive, this notation is heuristic and should not be interpreted as a classical differential (e.g., in the Riemann or Lebesgue sense), since Brownian paths are almost surely nowhere differentiable. Instead, it serves as a formal shorthand to express the Gaussian increments property:
\[
\diff \rvw(t) \sim \mathcal{N}(0, \diff t\, \mathrm{I}_D),
\]
meaning that over an infinitesimal time interval of length $ \diff t $, the increment of the Wiener process behaves like a Gaussian random variable with zero mean and covariance $ \diff t\, \mathrm{I}_D $.

\subsection{Further Explanation of \Cref{eq:sde}}
The SDE in \Cref{eq:sde} should be understood in its \emph{integral form}:
\begin{align}\label{eq:sde-integral-form}
    \rvx(t) = \rvx(0) + \int_0^t \rvf(\rvx(s), s) \diff s + \int_0^t g(s) \diff \rvw(s),
\end{align}
interpreted in the \emph{Itô sense}.
Here, the first term is a classical (Riemann or Lebesgue) integral representing the accumulated deterministic drift, while the second term is an \emph{Itô stochastic integral}, which integrates with respect to the Wiener process $ \rvw(t) $. We do not provide a full rigorous construction of the Itô integral, but offer the following intuition.

\paragraph{Intuition for It\^o Integration.}
The It\^o integral can be understood as the limit of discrete sums of
the form\footnote{For square-integrable integrands, this convergence holds in the $L^2$ sense (mean-square), which in turn implies convergence in probability. The $L^2$ viewpoint is natural because it connects directly to It\^o's isometry, the key identity underlying the standard and rigorous construction of the It\^o integral. This isometry links stochastic integrals to standard integrals in expectation ($\E\bigl[\bigl\|\int_0^T \bm{\psi}(t)\,\mathrm{d}\rvw_t \bigr\|^2\bigr] = \E\bigl[\int_0^T \|\bm{\psi}(t)\|_F^2\,\mathrm{d}t\bigr]$ with Frobenius norm
$\|\bm{\psi}(t)\|_F^2$); we state and use it in \Cref{subsec:rigorous-proof-linear-sde}.}
\[
\sum_i g(t_i)\bigl(w(t_{i+1})-w(t_i)\bigr),
\]
where the integrand $g(t)$ is evaluated at the \emph{left} endpoint
$t_i$ of each subinterval. As the partition becomes finer, these sums
converge to the It\^o integral. The left-point evaluation rule is what
distinguishes It\^o integration from classical integrals, which may use
midpoints or other evaluation rules. Because Brownian paths are
continuous yet almost surely nowhere differentiable, classical
integration does not directly apply. The It\^o integral handles this
irregularity and captures the cumulative effect of stochastic
fluctuations over time.

\paragraph{Use of Differential Notation.}
Expressions such as $ \diff \rvx(t) $, $ \diff t $, and $ \diff \rvw(t) $ are not classical differentials. Instead, they are formal notations representing infinitesimal increments of the respective processes. While heuristic, they are widely used for their convenience in expressing SDEs analogously to ODEs and facilitate formal manipulations within Itô calculus.

How Itô calculus is applied in diffusion models will be explained in \Cref{app:Ito}.

\paragraph{Comparison with ODEs.}
In ODEs, e.g.,
\[
\frac{\diff \rvx(t)}{\diff t} = \rvf(\rvx(t), t),
\]
the integral form
\[
\rvx(t) = \rvx(0) + \int_0^t \rvf(\rvx(\tau), \tau) \diff \tau
\]
is justified by the \emph{Fundamental Theorem of Calculus}, which ensures that differentiable functions can be recovered from their derivatives.

By contrast, in SDEs such as \Cref{eq:sde}, there is no direct analog of this theorem because Brownian motion lacks differentiability, and stochastic integrals do not follow the classical chain rule. Instead, Itô calculus introduces alternative tools (e.g., Itô's lemma) to analyze and manipulate stochastic dynamics.

Thus, while the differential notation for SDEs is compact and intuitive, a rigorous understanding depends on interpreting them via their integral formulation using Itô integrals.

\subsection{A Numerical Solver for SDE.} Like ODEs, the SDE in \Cref{eq:sde} admits a unique solution\footnote{The solution is in the \emph{strong} sense, meaning that $\mathbf{x}(t)$ satisfies the SDE in its integral form (see \Cref{eq:sde-integral-form}) with respect to the given Brownian motion $\mathbf{w}(t)$ on a fixed probability space. We omit the detailed technical definitions here.} if $\mathbf{f}(\cdot, t)$ and $g(\cdot)$ satisfy some smoothness conditions: $\mathbf{f}(\cdot, t)$ is Lipschitz and of linear growth in $\mathbf{x}$, and $g(\cdot)$ is square integrable.

For general SDEs as in \Cref{eq:sde}, closed-form solutions are generally unavailable, so numerical methods are necessary. A common approach is the \emph{Euler–Maruyama method}, which generalizes Euler’s method for ODEs and, indeed, we have already seen it in \Cref{eq:euler-sde}. It approximates the drift term $\rvf(\rvx(t), t)$ over a time step $\Delta t$ and simulates the stochastic noise $g(t) \diff \rvw(t)$ using Gaussian increments $\sqrt{\Delta t} \, \boldsymbol{\epsilon}_t$ with $\boldsymbol{\epsilon}_t \sim \mathcal{N}(\mathbf{0}, \mathbf{I})$. 

Later, in \Cref{subsec:rigorous-proof-linear-sde}, we will see that a linear SDE admits a closed-form solution.

\chapter{Density Evolution: From Change of Variable to Fokker–Planck}\label{app:continuity}

Understanding how probability densities evolve under transformations is fundamental in both probability theory and generative modeling. In particular, diffusion models aim to construct generative processes whose induced density paths reverse a pre-defined forward process. This evolution is governed by the continuity equation or, in the stochastic case, the Fokker-Planck equation.

Although these names may sound unfamiliar or intimidating, they are in fact continuous-time analogues of the change-of-variable formula from basic calculus. In \Cref{sec:cov}, it builds up to them by presenting a progression of change-of-variable formulas, starting from deterministic bijections, and culminating in stochastic differential equations. This progression naturally bridges discrete mappings and continuous-time flow dynamics. See \Cref{fig:unified_framework_CoV} for an overview of this unified framework.

In \Cref{sec:intuition-continuity}, we provide a physical and intuitive interpretation of the continuity equation, emphasizing its connection to the conservation of density in dynamical systems.

\chapterlearning{
  \item connect the change-of-variable formula to the continuity equation for ODEs and the Fokker--Planck equation for SDEs;

  \item understand the continuity equation as conservation of probability mass, and distinguish particle motion from density evolution;

  \item distinguish the Lagrangian and Eulerian views of transport, including why a density path alone need not determine a unique velocity field;

  \item understand why an SDE and its probability-flow ODE can share the same marginal densities while having different sample paths.
}{
The key principle is that the change-of-variable formula describes how
moving samples changes their probability distribution. This appendix
develops that idea from deterministic maps to the continuity and
Fokker--Planck equations, with Wasserstein gradient flow as an optional
extension to distribution-level training.
}

\begin{figure}[ht!]
\centering
\scalebox{0.9}{ 
\begin{tikzpicture}[
    mainbox/.style={draw, thick, rectangle, rounded corners=0pt, minimum width=13cm, minimum height=2.5cm, inner sep=5mm},
    equationbox/.style={draw, thick, rectangle, rounded corners=5pt, fill=white, inner sep=3mm, text width=5.2cm, align=center},
    smalltransformbox/.style={draw, thick, rectangle, rounded corners=5pt, fill=white, inner sep=2mm, text width=3.8cm, align=center},
    boxtitle/.style={font=\bfseries\sffamily\large},
    myarrow/.style={-Latex, very thick, black},
    arrowlabel/.style={font=\normalsize, align=left, anchor=west},
    eqfont/.style={font=\normalsize}
]

\node (transform_title) at (-0.5,-0.5) [boxtitle] {Transform};
\node (density_title) at (6cm,-0.5) [boxtitle] {Density};

\node (box1) at (3cm,-2.2cm) [mainbox] {};
\node (transform1) at ($(box1.west) + (3.2cm, 0)$) [smalltransformbox, eqfont] 
    {$ \rvx_0\xrightarrow{\,\bPhi\,} \rvx_1$};
\node (density1) at ($(box1.east) - (3.5cm, 0)$) [equationbox, eqfont]
    {$p_0(\rvx_0) = p_1(\rvx_1) \left| \det \frac{\partial \bPhi(\rvx_0)}{\partial \rvx_0} \right|$};

\node (box2) at (3cm,-6.2cm) [mainbox] {};
\node (transform2) at ($(box2.west) + (3.2cm, 0)$) [equationbox, eqfont, text width=5cm]
    {$\rvx_0\xrightarrow{\,\bPhi_1\,} \rvx_1\xrightarrow{\,\bPhi_2\,} \cdots\xrightarrow{\,\bPhi_L\,} \rvx_L$};
\node (density2) at ($(box2.east) - (3.5cm, 0)$) [equationbox, eqfont]
    {$
      \begin{array}{c}
        \textstyle \log p_0(\rvx_0) = \log p_L(\rvx_L) +\\ \sum_{k=0}^{L-1} \log \left| \det \frac{\partial \bPhi_{k+1}(\rvx)}{\partial \rvx_{k}} \right|
      \end{array}
    $
    };

\draw [myarrow] ($(box1.south) + (0,0)$) -- ($(box2.north) + (0,0)$);
\node [arrowlabel, right=0.2cm of $(box1.south)!0.5!(box2.north)$] {Multiple Bijections};

\node (box3) at (3cm,-10.2cm) [mainbox] {};
\node (transform3) at ($(box3.west) + (3.2cm, 0)$) [equationbox, eqfont]
    {$\frac{\diff\rvx(t)}{\diff t} = \rvf(\rvx(t), t)$, \\defining the flow map $\bPhi_{0\to t}$};
\node (density3) at ($(box3.east) - (3.5cm, 0)$) [equationbox, eqfont]
    {$\partial_t p_t(\rvx) = -\nabla \cdot (\rvf(\rvx, t) p_t(\rvx))$};

\draw [myarrow] ($(box2.south) + (0,0)$) -- ($(box3.north) + (0,0)$);
\node [arrowlabel, right=0.2cm of $(box2.south)!0.5!(box3.north)$] {Continuous-Time Limit};

\node (box4) at (3cm,-14.2cm) [mainbox] {};
\node (transform4) at ($(box4.west) + (3.2cm, 0)$) [equationbox, eqfont]
    {$\diff\rvx(t) = \rvf(\rvx(t), t)\diff t + g(t)\diff\rvw(t)$};
\node (density4) at ($(box4.east) - (3.5cm, 0)$) [equationbox, eqfont]
    {$
      \begin{array}{c}
        \textstyle \partial_t p_t(\rvx) = -\nabla \cdot (\rvf(\rvx, t) p_t(\rvx)) \\
        \textstyle + \frac{1}{2} g^2(t)\Delta p_t(\rvx)
      \end{array}
    $};

\draw [myarrow] ($(box3.south) + (0,0)$) -- ($(box4.north) + (0,0)$);
\node [arrowlabel, right=0.2cm of $(box3.south)!0.5!(box4.north)$] {With Gaussian Noise};

\end{tikzpicture}
}
\caption{A unified \emph{change-of-variables formula}. From top to bottom: (1) a single bijection; (2) composition of multiple bijections; (3) continuous-time deterministic flow governed by an ODE and the associated continuity equation; (4) stochastic flow modeled by an SDE and the corresponding Fokker–Planck equation. \figcredit{Created by the authors.}}
\label{fig:unified_framework_CoV}
\end{figure}

\newpage

\section{Change-of-Variable Formula: \\From Deterministic Maps to Stochastic Flows}\label{sec:cov}

In this section, we aim to demystify the continuity equation and the Fokker-Planck equation by drawing analogies to the classic change-of-variable formula from calculus. We begin with the familiar single-variable case, extend it to the multivariate setting and to probability densities (\Cref{subsec:cov-det}), then generalize to compositions of bijective maps whose continuous-time limit leads to the continuity equation (\Cref{subsec:cov-continuity-eq}). Finally, we incorporate stochasticity by introducing random noise, which naturally extends the continuity equation to the Fokker-Planck equation (\Cref{subsec:cov-fpe}).

\subsection{Change-of-Variable Formula for Deterministic Maps}\label{subsec:cov-det}
We move particles according to a deterministic map and study how their \emph{law} (density) evolves. The key principle is conservation of probability mass, grounded in a fundamental result from calculus and probability: the change-of-variable formula. This formula describes how integrals, and therefore probability densities, transform under smooth bijective mappings. To build intuition, we first consider a single update step, and then extend the discussion to sequential transformations.

\paragraph{Single Update.}
Think of a single update rule induced by applying a vector field (analogous to a force) 
$\bm{\Psi}:\R^D \to \R^D$ for one unit of time. 
Starting from an initial particle state $\rvx_0$, its next state is given by
\[
\rvx_1 = \bm{\Psi}(\rvx_0).
\]

\paragraph{Underlying Pattern (a Density) and How it Moves.}
If the initial states follow an underlying ``law/pattern'' described by a density $p_0$ (i.e., $\rvx_0\sim p_0$),
then applying $\bm{\Psi}$ produces a new density $p_1$ for $\rvx_1$ (i.e., $\rvx_1\sim p_1$).
Assuming $\bm{\Psi}$ is a smooth bijection, $p_1$ is obtained from $p_0$ via the standard \emph{change-of-variables formula}:
\begin{mdframed}
\begin{equation}\label{eq:change-of-variable-density}
p_{1}(\rvx_1) = p_{0}(\bm{\Psi}^{-1}(\rvx_1)) \cdot \left| \det \left( \frac{\partial \bm{\Psi}^{-1}}{\partial \rvx_1} \right) \right|.
\end{equation}
\end{mdframed}
Here $\tfrac{\partial \bm{\Psi}}{\partial \rvx}$ is the Jacobian matrix of $\bm{\Psi}$, denoted $\partial_\rvx \bm{\Psi}$.  
Equivalently, in the original coordinates,
\[
p_0(\rvx_0) \;=\; p_1\!\big(\bm{\Psi}(\rvx_0)\big)\;\big|\det \partial_\rvx \bm{\Psi}(\rvx_0)\big|.
\]
In words, $\bm{\Psi}$ reshapes the density $p_0$ into $p_1$. The factor $\big|\det \partial_\rvx \bm{\Psi}\big|$ represents the local change in volume; since probability mass is conserved, the density compensates by its inverse.  

As a simple case, if $\bm{\Psi}$ is linear with an invertible matrix $\rmA$ (i.e., $\rvx_1=\rmA\rvx_0$), then
\[
p_1(\rvx_1)\;=\;p_0(\rmA^{-1}\rvx_1)\,\big|\det \rmA^{-1}\big|.
\]

Schematically, we can read it as:
\[
\begin{array}{lccc}
   \text{\textbfs{Sample:}} & \rvx_0 & \xrightarrow{\hspace{0.2cm} \bm{\Psi} \hspace{0.2cm}} & \rvx_1  \\
   \text{\textbfs{Density:}} & p_{\rvx_0}(\rvx_0) & \xrightarrow{\hspace{0.2cm} \bm{\Psi} \hspace{0.2cm}} & p_{\rvx_1}(\rvx_1) 
\end{array}
\]

\paragraph{Why is \Cref{eq:change-of-variable-density} the Change-of-Variables Formula?}
This comes directly from the familiar rule in calculus.

\subparagraph{Single-Variable Case.}
Let $y = \Psi(x)$ be smooth and invertible. Rewriting an integral over $y$ in terms of $x$ gives
\[
\int g(y)\,\diff y = \int g(\Psi(x)) \cdot |\Psi'(x)|\,\diff x,
\]
where $|\Psi'(x)|$ compensates for interval stretching or compression, ensuring area preservation.

\subparagraph{Multivariate Case.}
For $\bm{\Psi}: \R^D \to \R^D$ with $\rvy=\bm{\Psi}(\rvx)$,
\[
\int g(\rvy)\,\diff\rvy
= \int g(\bm{\Psi}(\rvx)) \,\big|\det(\partial_\rvx \bm{\Psi})\big| \,\diff\rvx,
\]
so infinitesimal volumes transform as
\[
\diff \rvy = \big|\det(\partial_\rvx \bm{\Psi})\big|\,\diff \rvx.
\]

From this, the density formula in \Cref{eq:change-of-variable-density} follows:
\begin{align*}
p_{\rvy}(\rvy)
= \int_{\R^D} \delta(\rvy - \bm{\Psi}(\rvx))\, p_{\rvx}(\rvx)\diff \rvx = p_{\rvx}\!\big(\bm{\Psi}^{-1}(\rvy)\big)\,\Big|\det\!\left(\frac{\partial \bm{\Psi}^{-1}}{\partial \rvy}\right)\Big|.
\end{align*}

\paragraph{Composing Multiple Bijections.} We now apply several updates in sequence.  
Let $\rvx_k = \bm{\Psi}_k(\rvx_{k-1})$ for $k=1,\dots,L$; that is,
\[
\rvx_0 \xrightarrow{\ \bm{\Psi}_1\ } \rvx_1 \xrightarrow{\ \bm{\Psi}_2\ } \cdots
\xrightarrow{\ \bm{\Psi}_L\ } \rvx_L,
\]
where each $\bm{\Psi}_k:\R^D \to \R^D$ is a smooth bijection.  
If the initial state follows density $p_0$ (i.e., $\rvx_0 \sim p_0$), then the sequence of updates induces densities $p_1,\dots,p_L$ for $\rvx_1,\dots,\rvx_L$.  

Because probability mass is conserved at each step, the densities evolve according to
\[
p_{k}(\rvx_k) \;=\; p_{k-1}(\rvx_{k-1}) \,
   \Big|\det \partial_{\rvx_{k-1}} \bm{\Psi}_k(\rvx_{k-1})\Big|^{-1},
   \quad k=1,\dots,L.
\]
By recursion, the final density at $\rvx_L$ is
\begin{mdframed}
\begin{align}\label{eq:cov-multi-layer-nolog}
    p_{\rvx_L}(\rvx_L) 
    = p_{\rvx_0}(\rvx_0) 
    \cdot \prod_{k=1}^L \left| \det \left( \frac{\partial \bm{\Psi}_k}{\partial \rvx_{k-1}} \right) \right|^{-1}.
\end{align}
\end{mdframed}
Equivalently, in log-density form:
\begin{align*}
    \log p_{\rvx_L}(\rvx_L) 
    = \log p_{\rvx_0}(\rvx_0) 
    - \sum_{k=1}^L \log \left| \det \left( \frac{\partial \bm{\Psi}_k}{\partial \rvx_{k-1}} \right) \right|.
\end{align*}
This expression reflects how each transformation $\bm{\Psi}_k$ stretches or contracts volume, as captured by the Jacobian determinant. The accumulation of these local volume changes along the transformation path determines the final probability density under the composed map.

\Cref{eq:cov-multi-layer-nolog} serves as the core principle underlying Normalizing Flows (see \Cref{subsec:NODE}).

\clearpage
\newpage

\subsection{Continuous-Time Limit: Continuity Equation}\label{subsec:cov-continuity-eq}

We now pass from discrete updates to a continuous description.  
Suppose the particle motion is driven by a time-varying velocity field 
$\rvf:\R^D \times [0,T]\to \R^D$.  
Imagine evolving a particle $\rvx_0 \sim p_0$ through infinitely many small bijective updates.  
At each step $t$ of length $\Delta t>0$, the update is
\[
\rvx_{t+\Delta t} = \bm{\Psi}(\rvx_t) 
:= \rvx_t + \Delta t\, \rvf(\rvx_t, t).
\]
As $\Delta t \to 0$, the composition of these updates converges to a continuous flow governed by a velocity field $\rvf:\R^D \times [0,T]\to \R^D$:
\begin{align}\label{eq:ode-f-continuity}
    \frac{\diff \rvx(t)}{\diff t} = \rvf(\rvx(t), t),
    \qquad \rvx(0) = \rvx_0 \sim p_0.
\end{align}
Under suitable smoothness assumptions (see \Cref{app:de}), this ODE admits a unique solution for each initial condition, which defines a deterministic flow map 
$\bm{\Psi}_{0\to t}:\R^D \to \R^D$.  
In other words, $\bm{\Psi}_{0\to t}$ brings the initial state $\rvx_0$ to the solution of \Cref{eq:ode-f-continuity} at time $t$:
\[
\bm{\Psi}_{0\to t}(\rvx_0) 
= \rvx_0 + \int_{0}^{t} \rvf(\rvx(\tau), \tau)\,\diff\tau.
\]
As a result, the whole distribution also moves: the initial density $p_0$ is transported into the new density $p_t$, the law of $\rvx(t)$.  
Formally, this is written as a \emph{pushforward}:
\[
p_t = \big(\bm{\Psi}_{0\to t}\big)_{\#} p_0.
\]
When $\bm{\Psi}_{0\to t}$ is smooth and invertible, this reduces to the familiar change-of-variables rule:
\[
p_t(\rvx) 
= p_0\!\big(\bm{\Psi}_{t\to 0}(\rvx)\big)\,
   \big|\det \partial_\rvx\bm{\Psi}_{t\to 0}(\rvx)\big|
= \int \delta\!\left(\rvx - \bm{\Psi}_{0\to t}(\rvx_0)\right) 
   p_0(\rvx_0)\,\diff\rvx_0. 
\]

\paragraph{Continuity Equation: How the Density Moves in Time.}
Rather than writing a separate formula for the density at each time, we can describe how it moves continuously using a differential equation in space $\rvx$ and time $t$.
The idea is simple: probability mass is conserved, and the velocity field $\rvf$ only redistributes it in space.  
This gives the \emph{continuity equation}:
\begin{mdframed}
\begin{align}\label{eq:continuity-cov}
    \frac{\partial}{\partial t} p_t(\rvx) 
    + \nabla \cdot \big( p_t(\rvx)\,\rvf(\rvx, t) \big) = 0.
\end{align}
\end{mdframed}
Here the divergence term $\nabla \cdot (p_t \rvf)$ measures how the flow locally expands or compresses the density, ensuring total probability remains $1$.

This partial differential equation (PDE) ensures that probability mass is conserved as the flow moves particles.  
In fact, it can be viewed as the continuous-time analogue of the change-of-variables formula.

\paragraph{Derivation of Continuity Equation via  Change-of-Variables Formula.} Conceptually, the continuity equation can also be obtained by taking the continuous-time limit of \Cref{eq:cov-multi-layer-nolog}. Here, however, we adopt a more direct derivation based on \Cref{eq:change-of-variable-density}.
\subparagraph{Discretization and Change-of-Variable Formula.}
Consider 
\[
\mathbf{x}_{t+\Delta t} := \bm{\Psi}(\mathbf{x}_t) = \mathbf{x}_t + \Delta t\, \rvf(\mathbf{x}_t, t),
\]
which is actually the forward Euler discretization of the ODE in \Cref{eq:ode-f-continuity} over a small time interval $\Delta t > 0$.
The Jacobian of the map $\bm{\Psi}$ with respect to $\mathbf{x}_t$ expands as
\[
\frac{\partial \bm{\Psi}}{\partial \mathbf{x}_t} = \rmI + \Delta t \nabla_{\mathbf{x}} \rvf(\mathbf{x}_t, t) + \mathcal{O}(\Delta t^2),
\]
so its determinant satisfies
\[
\det\left( \frac{\partial \bm{\Psi}}{\partial \mathbf{x}_t} \right) = 1 + \Delta t\, \nabla \cdot \rvf(\mathbf{x}_t, t) + \mathcal{O}(\Delta t^2).
\]
This uses the standard expansion $\det(\rmI + \Delta t\, \mathbf{A}) = 1 + \Delta t\, \Tr(\mathbf{A}) + \mathcal{O}(\Delta t^2)$ as $\Delta t \to 0$, along with $\nabla \cdot \rvf = \Tr(\nabla_{\mathbf{x}} \rvf)$.

Applying the change-of-variables formula, the log-density evolves as
\begin{align*}
\log p_{t+\Delta t}(\mathbf{x}_{t+\Delta t}) = \log p_t(\mathbf{x}_t) - \Delta t\, \nabla \cdot \rvf(\mathbf{x}_t, t) + \mathcal{O}(\Delta t^2).
\end{align*}
That is,
\begin{align}
    \label{eq:log-density-diff}
\log p_{t+\Delta t}(\mathbf{x}_{t+\Delta t}) - \log p_t(\mathbf{x}_t) = - \Delta t\, \nabla \cdot \rvf(\mathbf{x}_t, t) + \mathcal{O}(\Delta t^2).
\end{align}

\subparagraph{Using Taylor Expansion.}
Now, we expand the left-hand side via multivariate Taylor expansion:
\begin{align*}
    &\log p_{t+\Delta t}(\rvx_{t+\Delta t}) 
    - \log p_t(\rvx_t) 
    \\= &\Delta t\, \partial_t \log p_t(\rvx_t) 
    + (\rvx_{t+\Delta t} - \rvx_t)^\top \nabla_{\rvx_t} \log p_t(\rvx_t) 
    + \mathcal{O}(\Delta t^2).
\end{align*}
Substituting $\rvx_{t+\Delta t} - \rvx_t = \rvf(\rvx_t, t)\Delta t$ yields:
\begin{align*}
    &\log p_{t+\Delta t}(\rvx_{t+\Delta t}) 
    - \log p_t(\rvx_t) 
    \\= & \Delta t\, \partial_t \log p_t(\rvx_t) 
    + \Delta t\, \rvf(\rvx_t, t)^\top \nabla_{\rvx_t} \log p_t(\rvx_t) 
    + \mathcal{O}(\Delta t^2).
\end{align*}
Matching terms with \Cref{eq:log-density-diff} and letting $\Delta t \to 0$, we conclude that
\begin{equation*}
    \partial_t \log p_t(\rvx_t) 
    = - \nabla_{\rvx_t} \cdot \rvf(\rvx_t, t) 
    - \rvf(\rvx_t, t)^\top \nabla_{\rvx_t} \log p_t(\rvx_t).
\end{equation*}
Exponentiating and using the product rule yields the continuity equation.

\paragraph{Lagrangian and Eulerian Views of Deterministic Transport.}
The same deterministic transport admits two complementary descriptions.
The \emph{Lagrangian} view follows individual particles through the flow
map $\bm{\Psi}_{0\to t}$, whereas the \emph{Eulerian} view describes the
velocity $\rvf(\rvx,t)$ and density $p_t(\rvx)$ at fixed spatial
locations. Under suitable regularity, these descriptions are consistent:
a prescribed velocity field generates a particle flow, whose pushforward
density satisfies the continuity equation.

There is nevertheless an important asymmetry. Once the velocity field and
initial density are fixed, the resulting flow and density evolution are
determined. A prescribed density path alone, however, generally does not
determine a unique velocity field or particle flow.

\subparagraph{From Velocity to Flow and Density.}
So far, we have assumed that the velocity field $\rvf$ is given.  
The particle ODE
\[
\frac{\diff \rvx(t)}{\diff t} = \rvf(\rvx(t), t)
\]
describes how each particle moves, while the density PDE
\[
\partial_t p_t + \nabla \!\cdot \!\big(p_t \rvf_t\big) = 0
\]
describes how the entire distribution of particles evolves.  
These two views are connected: moving particles according to the ODE automatically produces a density that satisfies the PDE.  
In this case, the particle flow $\bm{\Psi}_{0\to t}$ is uniquely determined: starting from $\rvx(0)\sim p_0$, each trajectory $\rvx(t)$ is fixed, and the resulting density $p_t$ follows the continuity equation.  
Here, particle dynamics and density dynamics are fully consistent.

\subparagraph{From Density Path to Velocity: Non-Uniqueness.}
If instead we begin only with the density path $t \mapsto p_t$ (e.g., \Cref{subsec:fm-instantiations} in flow matching), the velocity field is no longer uniquely determined.  
For example, if a vector field $\mathbf w_t$ satisfies
\[
\nabla_{\rvx}\cdot\!\big(p_t(\rvx)\,\mathbf w_t(\rvx)\big)=0
\qquad\text{(no net flux w.r.t.\ $p_t$),}
\]
then both $\rvf_t$ and $\rvf_t+\mathbf w_t$ give rise to the same density evolution.  
Thus a single density path may correspond to many different flows, and choosing one particular particle flow $\bm{\Psi}_{0\to t}$ amounts to picking a specific velocity field among these possibilities.

Not every given path $p_t$ can actually arise from particles moving under some velocity field.  
The continuity equation (\Cref{eq:continuity-cov}) provides the consistency check for whether a density path can be ``generated by a flow''.  
We say that $p_t$ is \emph{realizable} (or \emph{generated by} $\rvf$) if there exists a velocity field $\rvf$ such that particles following
\[
\dfrac{\diff \rvx(t)}{\diff t} = \rvf(\rvx(t), t)
\]
produce exactly the densities $p_t$ through the flow map $\bm{\Psi}_{0\to t}$.  That is, realizability holds when $p_t$ and $\rvf$ together satisfy \Cref{eq:continuity-cov}.

Intuitively, realizability means that the snapshots of $p_t$ over time can be explained by particles moving under some velocity field, rather than being an arbitrary sequence of distributions.

When this condition holds, the density $p_t$ is nothing more than the pushforward of the initial density $p_0$ along the flow map $\bm{\Psi}_{0\to t}$.  
In this case, the familiar change-of-variables formula applies:
\begin{align*}
    p_t = \big(\bm{\Psi}_{0\to t}\big)_{\#} p_0 
= p_0\!\big(\bm{\Psi}_{t\to 0}(\rvx)\big)\,
   \big|\det \partial_\rvx\bm{\Psi}_{t\to 0}(\rvx)\big|
= \int \delta \left(\rvx - \bm{\Psi}_{0\to t}(\rvx_0)\right) 
   p_0(\rvx_0)\,\diff\rvx_0.
\end{align*}

\paragraph{(Optional) Conditioning.}
If an additional conditioning variable $\rvz \sim \pi(\rvz)$ is introduced, the same reasoning applies for each fixed $\rvz$:
\[
\frac{\diff \rvx(t)}{\diff t}=\rvv_t(\rvx(t)|\rvz)
\]
with pushforward $p_t(\cdot|\rvz)=(\bPsi_{0\to t}(\cdot;\rvz))_{\#}p_0$, and continuity equation
\[
\partial_t p_t(\rvx|\rvz)+\nabla\!\cdot\!\big(p_t(\rvx|\rvz)\,\rvv_t(\rvx|\rvz)\big)=0
\]
The marginal density is then 
\[
p_t(\rvx)=\int p_t(\rvx|\rvz)\pi(\rvz)\diff\rvz.
\]



\subsection{Stochastic Processes: Fokker-Planck Equation}\label{subsec:cov-fpe}

When noise is added, the dynamics follow the SDE as in \Cref{eq:sde}:
\[
\diff \rvx(t) = \rvf(\rvx(t), t)\diff t + g(t)\diff \rvw(t).
\]
Then, the density $ p_t(\rvx) $ satisfies the Fokker-Planck equation:
\begin{mdframed}
    \begin{align*}
    \frac{\partial p_t(\rvx)}{\partial t}
&= - \nabla \cdot \left( \rvf(\rvx, t)\, p_t(\rvx) \right)
+ \frac{1}{2} g^2(t)\, \Delta p_t(\rvx)
\\& = - \nabla \cdot \left(  \left({\rvf(\rvx, t)}-{\frac{1}{2} g^2(t) \nabla_\rvx \log p_t(\rvx)}\right) p_t(\rvx)\right).
\end{align*}
\end{mdframed}
$ \Delta p_t = \nabla \cdot \nabla_\rvx p_t $ is the Laplacian operator. In the first equality, the first term describes transport of probability mass by the deterministic drift $\rvf$, while the second term models the spreading (diffusion) of the density due to stochastic noise with variance proportional to $\tfrac{1}{2} g^2(t) $. The second equality rewrites the same density evolution as a continuity
equation with velocity
\[
\rvf(\rvx,t)
-
\frac{1}{2}g^2(t)\nabla_{\rvx}\log p_t(\rvx).
\]
Therefore, the deterministic ODE driven by this
velocity therefore evolves through the same one-time marginal densities
$p_t$ as the SDE. This is the probability-flow ODE used throughout
the book.

\rmkb{
The key point is that the SDE and PF-ODE agree on the \emph{density path},
but generally not on how samples at different times are coupled. At any
fixed time $t$, both produce the same distribution $p_t$; equivalently,
their particle ``snapshots'' have the same histogram.

Across time, however, their sample paths are different. For the PF-ODE,
once the state $\rvx_s$ is fixed, the later state is determined by
$\bPsi_{s\to t}(\rvx_s)$. For the SDE, fresh Brownian noise is injected after time $s$, so the later
state remains random even after conditioning on $\rvx_s$.

Consequently, the two processes may have different transition
distributions, temporal correlations, and pathwise events, even though
their marginal density at every individual time is identical. Thus, the
PF-ODE reproduces the SDE's \emph{marginal density evolution}, rather than
its full stochastic path law. See \Cref{fig:score-sde-three-dynamics} for illustration.
}

The derivation of the Fokker-Planck equation is more involved; we refer the reader to \Cref{subsec:rigorous-proof-fpe}.

\newpage
\clearpage

\section{Intuition of the Continuity Equation}\label{sec:intuition-continuity}
In this section, we give a physical interpretation of the continuity equation, highlighting its role as a conservation law for probability density in a dynamical system.

\subsection{Physical Interpretation of the Continuity Equation}

Consider a small fixed control volume (a rectangular box) in 3D space with one corner at $\mathbf{x} = (x, y, z)$ and side lengths $\Delta x$, $\Delta y$, and $\Delta z$. Let $p(\mathbf{x}, t)$ denote the density of a conserved quantity (e.g., mass or probability) at position $\mathbf{x}$ and time $t$. For a sufficiently small box, the total amount of the quantity inside the box is approximately:
\[
\text{Total quantity in box} \approx p(\mathbf{x}, t) \Delta x \Delta y \Delta z.
\]

\paragraph{How Does the Total Change?}
Changes in the total quantity can only arise from flux across the box’s boundary. Let $\mathbf{j}(\mathbf{x}, t)$ denote the flux vector, representing the amount of quantity flowing per unit area per unit time.

\paragraph{Flux in the $x$-Direction.}
The inflow through the left face (at $x$) is approximately:
\[
j_x(x, y, z, t) \Delta y \Delta z,
\]
and the outflow through the right face (at $x + \Delta x$) is:
\[
j_x(x + \Delta x, y, z, t) \Delta y \Delta z.
\]
Thus, the net \emph{influx} in the $x$-direction is:
\[
\left[ j_x(x, y, z, t) - j_x(x + \Delta x, y, z, t) \right] \Delta y \Delta z.
\]

\paragraph{Net Flux in All Directions.}
Analogous terms arise in the $y$- and $z$-directions:
\begin{align*}
\left[ j_y(x, y, z, t) - j_y(x, y + \Delta y, z, t) \right] \Delta x \Delta z, \\
\left[ j_z(x, y, z, t) - j_z(x, y, z + \Delta z, t) \right] \Delta x \Delta y.
\end{align*}
Summing all contributions, the total net \emph{influx} into the box is:
\[
- \nabla \cdot \mathbf{j}(\mathbf{x}, t) \Delta x \Delta y \Delta z.
\]
Equivalently, the total net \emph{outflux} from the box is:
\[
\nabla \cdot \mathbf{j}(\mathbf{x}, t) \Delta x \Delta y \Delta z.
\]

\paragraph{Rate of Change Inside the Box.}
The rate of change of the total quantity within the box is:
\[
\frac{\partial p}{\partial t}(\mathbf{x}, t) \Delta x \Delta y \Delta z.
\]

\paragraph{Conservation Principle.}
Assuming the quantity is conserved (e.g., total mass or probability is constant in time), the rate of change equals the negative of the net outflux:
\[
\frac{\partial p}{\partial t}(\mathbf{x}, t) \Delta x \Delta y \Delta z = - \nabla \cdot \mathbf{j}(\mathbf{x}, t) \Delta x \Delta y \Delta z.
\]

Canceling the common volume factor and taking the small-box limit, we obtain the local form of the continuity equation:
\[
\frac{\partial p}{\partial t} + \nabla \cdot \mathbf{j} = 0.
\]
This equation has a simple interpretation: if $\nabla\cdot\mathbf{j}(\mathbf{x},t)>0$, more quantity flows out of the small box than into it, so the density inside decreases. If $\nabla\cdot\mathbf{j}(\mathbf{x},t)<0$, more quantity flows in than out, so the density inside increases.

\subsection{Derivation of the Continuity Equation from Conservation Laws}

The small-box argument above gives the local intuition. We now write the same conservation principle in a coordinate-free form over an arbitrary control volume.

The continuity equation formalizes the conservation of a physical quantity, such as mass or charge, in a dynamical system. Let $p(\mathbf{x}, t)$ denote the density of the conserved quantity at position $\mathbf{x} \in \mathbb{R}^D$ and time $t \in [0, T]$, and let $\mathbf{v}(\mathbf{x}, t)$ denote the velocity field. When the quantity is transported by particles moving with velocity $\mathbf{v}(\mathbf{x},t)$, the flux is
\[
\mathbf{j}(\mathbf{x},t)=p(\mathbf{x},t)\mathbf{v}(\mathbf{x},t).
\]

\paragraph{Step 1: Rate of Change within a Control Volume.}  
Consider an arbitrary control volume $V \subset \mathbb{R}^D$ with boundary $\partial V$. The total amount of the conserved quantity in $V$ is
\[
\int_V p(\mathbf{x}, t) \diff V,
\]
whose time derivative gives the rate of accumulation:
\[
\frac{\partial}{\partial t} \int_V p(\mathbf{x}, t) \diff V.
\]

\paragraph{Step 2: Net Flux Through the Boundary.}  
The quantity exits $V$ through $\partial V$ with outward normal vector $\mathbf{n}$. The net outward flux is
\[
\int_{\partial V} p(\mathbf{x}, t) \mathbf{v}(\mathbf{x}, t) \cdot \mathbf{n} \diff S.
\]

\paragraph{Step 3: Conservation Principle.}  
Conservation implies that the rate of accumulation within $V$ equals the negative of the net outward flux:
\[
\frac{\partial}{\partial t} \int_V p \diff V + \int_{\partial V} p \mathbf{v} \cdot \mathbf{n} \diff S = 0.
\]

\paragraph{Step 4: Divergence Theorem.}  
Applying the divergence theorem to convert the surface integral to a volume integral:
\[
\int_{\partial V} p \mathbf{v} \cdot \mathbf{n} \diff S = \int_V \nabla \cdot (p \mathbf{v}) \diff V.
\]
Hence,
\[
\frac{\partial}{\partial t} \int_V p \diff V + \int_V \nabla \cdot (p \mathbf{v}) \diff V = 0.
\]
Equivalently,
\[
\int_V \left(\frac{\partial p}{\partial t}+\nabla\cdot(p\mathbf{v})\right)\diff V=0.
\]
Since the control volume $V$ is arbitrary, the integrand must vanish pointwise. This yields the continuity equation:
\[
\frac{\partial p}{\partial t}+\nabla\cdot(p\mathbf{v})=0.
\]
Using $\mathbf{j}=p\mathbf{v}$, this is the same as
\[
\frac{\partial p}{\partial t}+\nabla\cdot\mathbf{j}=0.
\]

\paragraph{Particle-Level Intuition.}
Expanding the divergence term gives
\[
\frac{\partial p}{\partial t}
+
\mathbf{v}\cdot\nabla p
+
p\,\nabla\cdot\mathbf{v}
=
0.
\]
The first two terms describe the total change of the density along a moving particle:
\[
\frac{\diff}{\diff t}p(\mathbf{x}_t,t)
=
\frac{\partial p}{\partial t}(\mathbf{x}_t,t)
+
\mathbf{v}(\mathbf{x}_t,t)\cdot\nabla p(\mathbf{x}_t,t).
\]
Therefore,
\[
\frac{\diff}{\diff t}p(\mathbf{x}_t,t)
=
-
p(\mathbf{x}_t,t)\nabla\cdot\mathbf{v}(\mathbf{x}_t,t).
\]
This provides another useful interpretation: density decreases when nearby particles spread apart, and density increases when nearby particles contract. Thus, the velocity field tells us how individual particles move, while the continuity equation tells us how the entire density evolves.

\clearpage
\newpage

\section{(Optional) Wasserstein Gradient Flows as Distribution-Level Training}
\label{sec:wgf-distribution-level-training}

We now step back from diffusion-specific dynamics and take a broader
distribution-level view of deep generative modeling.  We have discussed that the continuity equation describes how a probability
distribution moves when its particles move.   The same language can also be used
to understand the training of deep generative models: a training algorithm updates
the model parameters, the updated generator moves generated particles, and the
movement of these particles changes the induced model distribution
$p_{\bphi}$.

From this viewpoint, different generative modeling methods can be organized
according to how the model distribution $p_{\bphi}$ is driven toward the data
distribution $p_{\mathrm{data}}$.  Wasserstein gradient flow gives one
particularly clean way to describe this movement: it treats training as an
idealized flow of probability mass in distribution space, which is then
approximated by finite-dimensional neural network updates.

A neural generator $\rmG_{\bphi}$ defines both a function and a probability
distribution.  If
\[
    \rvz\sim p_{\mathrm{prior}},
    \qquad
    \rvx=\rmG_{\bphi}(\rvz),
\]
then the generated distribution is the pushforward
\[
    p_{\bphi}:=(\rmG_{\bphi})_\# p_{\mathrm{prior}}.
\]
From the generative modeling viewpoint, the object we ultimately care about is
how $p_{\bphi}$ approaches the data distribution $p_{\mathrm{data}}$.

This motivates a distribution-level question:
\begin{question}
Can we first describe the ideal way to move the model distribution
$p_{\bphi}$ toward $p_{\mathrm{data}}$, and then update the neural network
parameters so that the generator approximately realizes this movement?
\end{question}

Throughout this optional section, $\tau$ denotes \emph{training time}: an
idealized continuous-time version of the discrete training iteration index.  It
should not be confused with diffusion/noising time.  We write $\bphi_\tau$ for
the parameter at training time $\tau$, and $p_{\bphi_\tau}$ for the
corresponding model distribution.

\paragraph{Moving Model Distributions by Moving Particles.}
We first describe an ideal distribution-level evolution, before accounting
for the finite-dimensional parameterization of the neural generator.
Suppose particles from the current model distribution move according to a
training-time velocity field $\rvw_\tau$:
\[
    \frac{\diff \rvx_\tau}{\diff \tau}
    =
    \rvw_\tau(\rvx_\tau),
    \qquad
    \rvx_\tau\sim p_{\bphi_\tau}.
\]
Then the model distribution evolves according to the continuity equation
\[
    \partial_\tau p_{\bphi_\tau}(\rvx)
    +
    \nabla_{\rvx}\cdot
    \big(
        p_{\bphi_\tau}(\rvx)\rvw_\tau(\rvx)
    \big)
    =
    0.
\]
Thus, $\rvw_\tau$ describes how each generated particle moves, while the
continuity equation describes how the entire generated distribution moves\footnote{This velocity should not be confused with the diffusion-time PF-ODE velocity
$\rvv^*(\rvx,t)$.  The variable $t$ or $s$ describes noise/denoising time,
whereas $\tau$ describes training-time evolution of the model distribution.}.

\paragraph{Choosing the Velocity: Gradient Descent in Distribution Space.}
Now we need to choose a velocity field $\rvw_\tau$ that moves
$p_{\bphi_\tau}$ toward $p_{\mathrm{data}}$.  Let
\[
    \mathcal{D}(p_{\bphi_\tau},p_{\mathrm{data}})
\]
be a discrepancy between the current model distribution and the data
distribution.  \emph{Wasserstein gradient flow}~\citep{jordan1998variational,ambrosio2005gradient} chooses the particle velocity that
decreases this discrepancy most directly under the geometry of moving
probability mass.

To choose the velocity field, we first ask how the discrepancy changes when the
current distribution is infinitesimally perturbed.  Think of
$\mathcal{D}(p,p_{\mathrm{data}})$ as a global energy of the distribution
$p$.  Its first variation gives a pointwise energy potential: it tells us
which regions of space are costly for the current distribution and which regions
are favorable.

Concretely, for a small perturbation $p+\epsilon h$, where
$\epsilon>0$ is small and $\int h(\rvx)\diff\rvx=0$, the first variation is
defined by
\[
    \mathcal{D}(p+\epsilon h,p_{\mathrm{data}})
    =
    \mathcal{D}(p,p_{\mathrm{data}})
    +
    \epsilon
    \int
    \frac{\delta\mathcal{D}}{\delta p}
    (p,p_{\mathrm{data}})(\rvx)\,
    h(\rvx)\diff\rvx
    +
    \mathop{\mathrm{o}}(\epsilon).
\]
Here $p$ is a dummy distribution variable.  After taking the variation, we
evaluate it at the current model distribution $p=p_{\bphi_\tau}$.

We define the resulting pointwise energy potential by
\[
    E_\tau(\rvx)
    :=
    \frac{\delta\mathcal{D}}{\delta p}
    (p_{\bphi_\tau},p_{\mathrm{data}})(\rvx).
\]
Intuitively, if $E_\tau(\rvx)$ is large, then placing more probability mass
near $\rvx$ would increase the discrepancy.  If $E_\tau(\rvx)$ is small,
then moving mass toward $\rvx$ is favorable.  Therefore, the Wasserstein
gradient-flow velocity moves particles downhill in this potential:
\[
    \rvw_\tau(\rvx)
    =
    -
    \nabla_{\rvx}E_\tau(\rvx).
\]
With this choice,
\[
    \frac{\diff}{\diff \tau}
    \mathcal{D}(p_{\bphi_\tau},p_{\mathrm{data}})
    =
    -
    \int
    p_{\bphi_\tau}(\rvx)
    \left\|
        \nabla_{\rvx}E_\tau(\rvx)
    \right\|_2^2
    \diff\rvx
    \le 0.
\]
Therefore, Wasserstein gradient flow is gradient descent in the
$2$-Wasserstein geometry of probability distributions\footnote{We remark that this ideal
distribution-space dynamics should not be confused with ordinary gradient
descent on the neural parameters $\bphi$: the latter is constrained by the
finite-dimensional generator family and generally realizes only an
approximation to the desired particle motion.}.  The global discrepancy $\mathcal{D}$ defines the
energy of the whole distribution, the first variation $E_\tau$ defines the
local energy landscape seen by particles, and
$-\nabla_{\rvx}E_\tau$ gives the ideal particle velocity for decreasing the
discrepancy.

\paragraph{Realizing the Distributional Velocity with a Neural Network.}
The velocity above is a distribution-level object: it says how generated
particles should move.  A neural generator, however, cannot move every particle
independently.  It can only move particles through a shared parameter update of
$\bphi$.

At training time $\tau$, the current generator produces particles
\[
    \rvx_i
    =
    \rmG_{\bphi_\tau}(\rvz_i),
    \qquad
    \rvz_i\sim p_{\mathrm{prior}},
    \qquad
    i=1,\ldots,N.
\]
A small Wasserstein step moves these particles to
\[
    \widetilde{\rvx}_i
    =
    \rvx_i+\eta\rvw_\tau(\rvx_i),
\]
where $\eta>0$ is a small step size.  The neural network is then updated so
that its new outputs match these moved particles:
\[
    \bphi_{\tau+\eta}
    \approx
    \arg\min_{\bphi}
    \frac{1}{N}
    \sum_{i=1}^N
    \left\|
        \rmG_{\bphi}(\rvz_i)
        -
        \operatorname{sg}
        \big(
            \rvx_i+\eta\rvw_\tau(\rvx_i)
        \big)
    \right\|_2^2.
\]
Here $\operatorname{sg}(\cdot)$ denotes stop-gradient, so the moved particles
are treated as fixed targets.

For a small parameter update $\Delta\bphi$, we have the linearization
\[
    \rmG_{\bphi_\tau+\Delta\bphi}(\rvz_i)
    \approx
    \rmG_{\bphi_\tau}(\rvz_i)
    +
    \big(\partial_{\bphi}\rmG_{\bphi_\tau}(\rvz_i)\big)\Delta\bphi.
\]
Therefore, the parameter update approximately solves
\[
    \min_{\Delta\bphi}
    \frac{1}{N}
    \sum_{i=1}^N
    \left\|
        \big(\partial_{\bphi}\rmG_{\bphi_\tau}(\rvz_i)\big)\Delta\bphi
        -
        \eta\rvw_\tau(\rvx_i)
    \right\|_2^2.
\]
In words, Wasserstein gradient flow specifies the desired distribution-level
particle motion, while neural network training projects this motion onto the
set of movements that can be realized by changing the finite-dimensional
parameter $\bphi$.

\paragraph{Forward-KL Training: Data-Side Likelihood Learning.}
A classical distribution-level objective is the data-side, or forward, KL
divergence:
\[
    \mathcal{D}_{\mathrm{FKL}}(p_{\bphi},p_{\mathrm{data}})
    :=
    \mathcal D_{\mathrm{KL}}(p_{\mathrm{data}}\|p_{\bphi})
    =
    \int
    p_{\mathrm{data}}(\rvx)
    \log
    \frac{p_{\mathrm{data}}(\rvx)}{p_{\bphi}(\rvx)}
    \diff\rvx.
\]
Since $p_{\mathrm{data}}$ is fixed, minimizing this objective is equivalent to
maximum likelihood (see \Cref{eq:MLE}):
\[
    \min_{\bphi}
    \mathcal D_{\mathrm{KL}}(p_{\mathrm{data}}\|p_{\bphi})
    \quad
    \Longleftrightarrow
    \quad
    \max_{\bphi}
    \E_{\rvx\sim p_{\mathrm{data}}}
    \big[
        \log p_{\bphi}(\rvx)
    \big].
\]
This is the distribution-level view behind many likelihood-based models.
Autoregressive models and normalizing flows optimize exact likelihoods;
likelihood-based VAEs optimize variational lower bounds; energy-based models
optimize likelihood through positive and negative phases; and diffusion models
can be related to likelihood or score-matching surrogates through their
variational and denoising objectives.

From the Wasserstein gradient-flow viewpoint, the forward KL has first
variation
\[
    \frac{\delta}{\delta p}
    \mathcal D_{\mathrm{KL}}(p_{\mathrm{data}}\|p)
    =
    -
    \frac{p_{\mathrm{data}}}{p}.
\]
Evaluating at $p=p_{\bphi_\tau}$ gives the velocity
\[
    \rvw_\tau^{\mathrm{FKL}}(\rvx)
    =
    \nabla_{\rvx}
    \left(
        \frac{p_{\mathrm{data}}(\rvx)}
             {p_{\bphi_\tau}(\rvx)}
    \right).
\]
This expression involves the density ratio
$p_{\mathrm{data}}/p_{\bphi_\tau}$, which is usually difficult to estimate
for implicit generators. In practice, forward-KL or maximum-likelihood
training is most direct when likelihoods, variational bounds, or suitable
approximations to the likelihood gradient are available. Score matching
provides a different route to distribution learning by matching score
functions rather than directly realizing the forward-KL Wasserstein flow.

The main strength of forward-KL training is that it is data-covering and
statistically well grounded when a likelihood or a variational surrogate is
available.  Its limitation is that it often requires tractable densities,
variational bounds, score-matching surrogates, or expensive negative sampling,
which makes it less directly compatible with arbitrary implicit generators.

\paragraph{Reverse-KL Training: Model-Side Score-Mismatch Dynamics.}
The model-side, or reverse, KL divergence is
\[
    \mathcal{D}_{\mathrm{RKL}}(p_{\bphi},p_{\mathrm{data}})
    :=
    \mathcal D_{\mathrm{KL}}(p_{\bphi}\|p_{\mathrm{data}})
    =
    \int
    p_{\bphi}(\rvx)
    \log
    \frac{p_{\bphi}(\rvx)}
         {p_{\mathrm{data}}(\rvx)}
    \diff\rvx.
\]
Its first variation is
\[
    \frac{\delta}{\delta p}
    \mathcal D_{\mathrm{KL}}(p\|p_{\mathrm{data}})
    =
    \log p-\log p_{\mathrm{data}}+1.
\]
Evaluating at $p=p_{\bphi_\tau}$ gives the Wasserstein velocity
\[
    \rvw_\tau^{\mathrm{RKL}}(\rvx)
    :=
    \nabla_{\rvx}\log p_{\mathrm{data}}(\rvx)
    -
    \nabla_{\rvx}\log p_{\bphi_\tau}(\rvx).
\]
Thus, reverse-KL training induces a score-mismatch force:
\[
    \text{target score}
    -
    \text{model score}.
\]
The target score attracts generated particles toward high-density regions of the
data distribution, while the model score repels particles from regions where the
current model distribution is already too concentrated.

In diffusion distillation and distribution matching (see \Cref{sec:VSD}), this idea is usually
applied at a noise level $s$.  Let $p_s$ denote the teacher or noised-data
distribution, and let $p_{\bphi_\tau,s}$ denote the noised distribution of
current generator samples.  The reverse-KL objective
\[
    \E_s
    \left[
        w(s)
        \mathcal D_{\mathrm{KL}}(p_{\bphi_\tau,s}\|p_s)
    \right]
\]
induces the score-mismatch velocity
\[
    \rvw_{\tau,s}^{\mathrm{RKL}}(\rvx)
    :=
    \nabla_{\rvx}\log p_s(\rvx)
    -
    \nabla_{\rvx}\log p_{\bphi_\tau,s}(\rvx).
\]
At the parameter level, for generated noisy samples
$\hat\rvx_s\sim p_{\bphi_\tau,s}$, one obtains gradients of the form
\[
    \nabla_{\bphi}
    \mathcal D_{\mathrm{KL}}(p_{\bphi,s}\|p_s)
    \big|_{\bphi=\bphi_\tau}
    =
    \E_{\hat\rvx_s\sim p_{\bphi_\tau,s}}
    \left[
        \left(
            \nabla_{\rvx}\log p_{\bphi_\tau,s}(\hat\rvx_s)
            -
            \nabla_{\rvx}\log p_s(\hat\rvx_s)
        \right)^\top
        \partial_{\bphi}\hat\rvx_s
    \right].
\]
Thus, the score difference
\[
\nabla_{\rvx}\log p_s(\hat\rvx_s)
-
\nabla_{\rvx}\log p_{\bphi_\tau,s}(\hat\rvx_s)
\]
provides the distribution-level descent direction that drives the parameter
gradient. The actual movement of individual generated samples is mediated by
the shared neural parameterization, as discussed above. This score-difference gradient is the basic distribution-level mechanism
behind reverse-KL-based methods such as DMD and VSD: estimate the target and
model scores on generated samples and use their difference to update the
generator.

The main strength of reverse-KL training is that it is naturally compatible with
implicit generators once the relevant scores can be estimated.  It is also
mode-seeking, which can improve visual fidelity in one-step or few-step
generation.  Its limitation is that it can drop modes, depends strongly on the
quality of score estimates, and may require an additional model-score network or
a strong teacher.

The same attraction-repulsion interpretation also appears in drifting-style
particle updates~\citep{weber2023the,deng2026generative}.  In the terminology
of \citet{lai2026unified}, DMD/VSD/SiD provide parametric score-based
instantiations of this score-mismatch dynamics, while drifting-style
approaches~\citep{weber2023the,deng2026generative} provide a
nonparametric kernel-based instantiation.

\paragraph{Closing Remarks.}
In summary, the continuity equation provides the common language behind both
diffusion dynamics and distribution-level training.  In diffusion models, it
describes how the intermediate distributions $p_t$ evolve along a prescribed
noise or probability-flow dynamics.  In Wasserstein gradient flow, the same
equation describes how the model distribution $p_{\bphi_\tau}$ evolves during
training as generated particles move toward the data distribution
$p_{\mathrm{data}}$.

This viewpoint therefore broadens the role of the continuity equation beyond
diffusion models.  It lets us interpret generative modeling as the design of
distributional dynamics: different methods choose different discrepancies,
different particle velocities, and different parameter updates to drive
$p_{\bphi}$ toward $p_{\mathrm{data}}$.  From this perspective, diffusion
models, likelihood-based models, distribution-matching methods, and
score-mismatch-based approaches can be viewed under a shared distribution-level
taxonomy.

\clearpage
\newpage

\chapter{Behind the Scenes of Diffusion Models:\\ It\^o’s Calculus and Girsanov's Theorem}\label{app:Ito}

(Score-based) Diffusion models are built on SDEs: a drift that pushes states and a Brownian term that jitters them. Unlike ODE paths, Brownian paths are nowhere differentiable, so the ordinary chain rule fails. In this section, we introduce two fundamental tools that make the math precise:
\begin{itemize}
\item \textbfs{It\^o’s Formula} is the correct chain rule for stochastic trajectories. It tells us how a function $\rvh(\rvx_t,t)$ evolves when $\rvx_t$ follows an SDE. It enables derivations of the Fokker–Planck equation, moment dynamics, the It\^o product rule, and the identities used in score-based training.
\item \textbfs{Girsanov’s Theorem} is a change-of-measure result on \emph{path probabilities}. It quantifies how likelihoods change when the noise is fixed but the drift is altered. This links likelihood-weighted score matching to path-space KL divergence
and, through a variational bound, to data likelihood.
\end{itemize}

With these tools, the standard diffusion model derivations (Fokker–Planck, reverse time SDE, training objectives, and likelihood relations) follow cleanly and without hand waving.

\chapterlearning{
  \item understand why ordinary calculus is not enough for SDEs, and how It\^o's formula gives the correct chain rule for stochastic trajectories;

  \item see how It\^o's formula works behind the scenes in important diffusion-model derivations, including the Fokker--Planck equation and linear SDEs;

  \item understand Girsanov's theorem as a way to compare the probabilities of stochastic paths when their drifts are different but their diffusion is the same;

  \item distinguish matching distributions at each time from matching the full stochastic path, and see why likelihood-weighted score matching is related to path-space KL divergence and data likelihood.
}{
This appendix develops some of the mathematical tools that operate behind
the scenes of diffusion models. It\^o calculus makes stochastic
differentiation precise, while Girsanov's theorem explains how changing
the drift changes the probability of entire stochastic paths. These tools
underlie several standard diffusion-model derivations that are often used
without being shown explicitly.
}

\newpage

\section{It\^o's Formula: The Chain Rule for Random Processes}
Standard calculus does not directly apply to stochastic processes because Wiener processes are not differentiable in the classical sense. Instead, we use It\^o’s calculus, which provides rules for working with stochastic integrals. 

\subsection{Motivation: Why Do We Need a Special Chain Rule?}
Consider a deterministic time-varying function $\rvy_t$ that evolves smoothly with time $t$ (e.g., an ODE). If we have a function $\rvh(\rvy_t, t)$, the usual chain rule tells us:
\[
\frac{\diff \rvh}{\diff t} = \frac{\partial \rvh}{\partial t}  + \nabla_{\rvy}\rvh  \frac{\diff \rvy_t}{\diff t} .
\]
Here, $\nabla_{\rvy}\rvh$ is the Jacobian of $\rvh$. This works perfectly for deterministic paths $\rvy_t$.

\begin{question}
    But what happens if $\rvx_t$ is a stochastic process, say, driven by an SDE 
\[
\diff \rvx_t = \rvf(\rvx_t, t)\diff t + g(t) \diff \rvw_t
\]
as in \Cref{eq:sde}? What SDE does the process $\rvh(\rvx_t, t)$ satisfy?
\end{question}

\paragraph{Why the Ordinary Chain Rule Fails?}
Naïvely applying the classical chain rule yields
\[
\diff \rvh = \frac{\partial \rvh}{\partial t} \diff t + \nabla_{\rvy}\rvh  \cdot \diff \rvx_t.
\]
However, Brownian increments scale as
$\mathcal{O}(\sqrt{\diff t})$, while their quadratic variation is encoded
in It\^o calculus by
\[
(\diff \rvw_t)^2 = \diff t.
\]
Thus, second-order terms in $\diff \rvw_t$ \textit{do not} vanish in stochastic calculus, unlike classical calculus where $(\diff t)^2$ terms are negligible.

\exm{Simple Example--$h(x_t) = x_t^2$}{
To see the intuition, let us consider the simple real-valued function $h(x_t) = x_t^2 \in \mathbb{R}$ where the random variable $x_t \in \mathbb{R}$ satisfies
\[
\diff x_t = \sigma \diff w_t, \quad
x_0=0,
\]
with a constant $\sigma>0$. 
If we try the classical chain rule,
\[
\diff h = 2 x_t \diff x_t = 2 x_t \sigma \diff w_t.
\]
If this were true, the expectation of $h(x_t)$ would be constant in time because 
\[
\mathbb{E}[\diff h ] =  2 \sigma \mathbb{E}[x_t  \diff w_t]= 2 \sigma \mathbb{E}[x_t]  \underbrace{\mathbb{E}[\diff w_t]}_{=0}=0.
\]
But we know from classical Brownian motion properties (see \Cref{eq:wiener-gaussian}) that
\[
\mathbb{E}[x_t^2] = \sigma^2 t,
\]
which grows linearly in time. So the ordinary chain rule misses an important term.
}

\subsection{Deriving 1D It\^o's Formula from Taylor Expansion}\label{subsec:1D-Ito}

\paragraph{Deterministic Chain Rule via Taylor Expansion.}
To understand why the classical chain rule fails for stochastic processes defined by SDEs, we first revisit it in the deterministic setting using Taylor expansion. We consider the scalar case: $y_t \in \mathbb{R}$ and $h(\cdot,\cdot)\in \mathbb{R}$. Formally treating $\diff y_t = y_{t+\diff t} - y_t$, with $\diff t \approx 0$, we expand:
\begin{align*}
&h(y_{t+\diff t}, t+\diff t) 
- h(y_t, t) \\
= ~&\frac{\partial h}{\partial t} \diff t
+ \frac{\partial h}{\partial y} \diff y_t  + \frac{1}{2} \left(\textcolor{gray}{\frac{\partial^2 h}{\partial y^2}  (\diff y_t)^2}
+ 2 \textcolor{gray}{\frac{\partial^2 h}{\partial t \partial y} \diff t \diff y_t}
+  \textcolor{gray}{\frac{\partial^2 h}{\partial t^2} (\diff t)^2}
\right)
+ \mathcal{O}(\diff t^3),
\end{align*}
Here, $\diff t \diff y_t = \left(\frac{\diff y_t}{\diff t}\right)(\diff t)^2 = \mathcal{O}(\diff t^2)$, and similarly $(\diff t)^2 = \mathcal{O}(\diff t^2)$. Therefore, all the gray parts are ignorable, and the full differential is:
\[
\diff h
= \frac{\partial h}{\partial t} \diff t
+ \frac{\partial h}{\partial y} \diff y_t
+ \mathcal{O}(\diff t^2).
\]

\paragraph{It\^o's Formula via Stochastic Taylor Expansion.}
Now consider a stochastic process $x_t \in \mathbb{R}$ governed by the SDE:
\[
\diff x_t = f(x_t, t) \diff t + g(t) \diff w_t,
\]
where $w_t$ is standard Brownian motion. We aim to compute the differential of a scalar-valued function $h(x_t, t)$.

Using the stochastic Taylor expansion~\citep{kloeden1992stochastic}, which retains second-order terms in $\diff x_t$, we have:
\begin{align*}
&h(x_{t+\diff t}, t+\diff t) 
- h(x_t, t) \\
=~&\frac{\partial h}{\partial t} \diff t 
+ \frac{\partial h}{\partial x} \diff x_t 
+ \frac{1}{2} \left( \frac{\partial^2 h}{\partial x^2} (\diff x_t)^2 
+ 2\textcolor{gray}{{\frac{\partial^2 h}{\partial t \partial x} \diff t \diff x_t}} 
+ \textcolor{gray}{{ \frac{\partial^2 h}{\partial t^2} (\diff t)^2}} \right)
+ \cdots
\end{align*}
\subparagraph{Negligible Cross Terms.}
By the scaling property of Brownian motion (\Cref{eq:wiener-gaussian}), 
\[
\diff w_t = \mathcal{O}(\sqrt{\diff t}) \quad \Rightarrow \quad \diff t \cdot \diff w_t = \mathcal{O}((\diff t)^{3/2}).
\]
Therefore,
\begin{align*}
\diff t \cdot \diff x_t 
&= \diff t \left(f \diff t + g \diff w_t\right) 
= f (\diff t)^2 + g \cdot \diff t \cdot \diff w_t 
= \mathcal{O}((\diff t)^{3/2}).
\end{align*}
So, the gray terms are negligible: $ \mathcal{O}((\diff t)^{3/2}) $ or smaller.
\subparagraph{Second-Order Term $(\diff x_t)^2$.}
Expanding using the SDE:
\begin{align*}
(\diff x_t)^2 
&= \left( f \diff t + g \diff w_t \right)^2 \\
&= f^2 (\diff t)^2 + 2fg \diff t \diff w_t + g^2 (\diff w_t)^2 \\
&= g^2(t) \diff t,
\end{align*}
where the last equality uses the It\^o differential rules
$(\diff t)^2=0$, $\diff t\,\diff w_t=0$, and
$(\diff w_t)^2=\diff t$, with the last identity understood in the
quadratic-variation sense.

Combining terms, we obtain the differential:
\[
\diff h(x_t, t)
= \frac{\partial h}{\partial t} \diff t
+ \frac{\partial h}{\partial x} \diff x_t
+ \frac{1}{2} \frac{\partial^2 h}{\partial x^2} g^2(t) \diff t.
\]
Substituting $ \diff x_t = f(x_t, t) \diff t + g(t) \diff w_t $ yields:
\[
\diff h(x_t, t)
= \left( \frac{\partial h}{\partial t} + f \frac{\partial h}{\partial x}
+ \frac{1}{2} g^2 \frac{\partial^2 h}{\partial x^2} \right) \diff t
+ g \frac{\partial h}{\partial x} \diff w_t.
\]
This is the 1D version of \emph{It\^o's formula}.

\exm{Simple Example--$h(x_t) = x_t^2$}{
We revisit the simple example: $ h(x_t) = x_t^2 $, where the stochastic process $ x_t \in \mathbb{R} $ satisfies
\[
\diff x_t = \sigma \diff w_t,
\]
with a constant $ \sigma > 0 $.  
 Applying It\^o’s formula correctly to $ h(x_t) = x_t^2 $, we obtain:
\[
\diff h(x_t) = \diff (x_t^2) = 2 x_t \diff x_t + \sigma^2 \diff t.
\]
Substituting $ \diff x_t = \sigma \diff w_t $, this becomes:
\[
\diff(x_t^2) = 2 x_t \sigma \diff w_t + \sigma^2 \diff t.
\]
}

\subsection{It\^o's Formula: The Chain Rule for SDEs}
\label{subsec:ito-summary}
We summarize the one-dimensional It\^o's formula derived above. Using similar arguments, the result extends naturally to the multi-dimensional setting. While we omit the detailed derivation, we state the general formula for completeness.

Finally, we illustrate an application of It\^o's formula by deriving the \emph{It\^o product rule}, which enables computation of $\diff(\rvx_t^\top \rvy_t)$ for stochastic processes $\rvx_t$ and $\rvy_t$.

\paragraph{$1$D It\^o's Formula.}
Let $x_t \in \mathbb{R}$ be a stochastic process satisfying the SDE:
\[
\diff x_t = f(x_t, t) \diff t + g(t) \diff w_t.
\]
For a scalar function $h \colon \mathbb{R} \times [0, T] \to \mathbb{R}$, the process $h(x_t, t)$ satisfies:
\begin{mdframed}
\[
\diff h(x_t, t)
= \left( \frac{\partial h}{\partial t} + f \frac{\partial h}{\partial x}
+ \frac{1}{2} g^2 \frac{\partial^2 h}{\partial x^2} \right) \diff t
+ g \frac{\partial h}{\partial x} \diff w_t.
\]
\end{mdframed}

\paragraph{Multidimensional It\^o's Formula with Scalar Output.}
Let $\rvx_t \in \mathbb{R}^D$ satisfy the SDE:
\[
\diff \rvx_t = \rvf(\rvx_t, t) \diff t + g(t) \diff \rvw_t,
\]
where $\rvf \colon \mathbb{R}^D \times [0, T] \to \mathbb{R}^D$, $g \colon [0, T] \to \mathbb{R}$, and $\rvw_t \in \mathbb{R}^D$ is a $D$-dimensional Brownian motion. Let $h \colon \mathbb{R}^D \times [0, T] \to \mathbb{R}$ be a scalar-valued function. Then $h(\rvx_t, t)$ satisfies:
\begin{mdframed}
\begin{align}\label{eq:Ito-scalar}
    \diff h(\rvx_t, t)
= \left( \frac{\partial h}{\partial t}
+ \nabla_{\rvx} h^\top \rvf
+ \frac{1}{2} g^2(t) \operatorname{Tr} \left( \nabla^2_{\rvx} h \right) \right) \diff t
+ g(t) \nabla_{\rvx} h^\top \diff \rvw_t,
\end{align}
\end{mdframed}
where $\nabla_{\rvx} h \in \mathbb{R}^D$ is the gradient and $\nabla^2_{\rvx} h \in \mathbb{R}^{D \times D}$ is the Hessian matrix of $h$ with respect to $\rvx$.

\exm{It\^o's Product Rule}{
Let $\rvx_t, \rvy_t \in \mathbb{R}^D$ be vector-valued stochastic processes governed by the SDEs:
\[
\begin{aligned}
\diff \rvx_t &= \rva(\rvx_t, t) \diff t + b(t) \diff \rvw_t, \\
\diff \rvy_t &= \rvc(\rvy_t, t) \diff t + d(t) \diff \rvw_t,
\end{aligned}
\]
where $\rva, \rvc: \mathbb{R}^D \times [0, T] \to \mathbb{R}^D$ are vector fields, and $b(t), d(t) \in \mathbb{R}$ are scalar-valued functions. Here, $\rvw_t \in \mathbb{R}^D$ denotes a standard $D$-dimensional Brownian motion.

We aim to derive the SDE for the scalar-valued process
\[
z(t) := \rvx_t^\top \rvy_t.
\]

Applying the multivariate It\^o formula to the bilinear function $h(\rvx, \rvy) := \rvx^\top \rvy$, we obtain:
\[
\diff(\rvx^\top \rvy) = (\diff \rvx)^\top \rvy + \rvx^\top \diff \rvy + \Tr\left[ \diff \rvx \cdot (\diff \rvy)^\top \right].
\]

The It\^o correction term is computed as:
\begin{align*}
    \diff \rvx \cdot (\diff \rvy)^\top &= b(t) \diff \rvw_t \cdot \left[ d(t) \diff \rvw_t \right]^\top \\&= b(t) d(t) \diff \rvw_t \cdot \diff \rvw_t^\top \\&= b(t)d(t) \diff t \cdot \rmI_D.
\end{align*}
Thus,
\[
\Tr\left[ \diff \rvx \cdot (\diff \rvy)^\top \right] = b(t)d(t) \Tr(\rmI_D) \diff t = D  b(t)d(t) \diff t.
\]

Putting everything together, the resulting SDE is:
\begin{align}\label{eq:ito-product}
\begin{aligned}
        \diff(\rvx^\top \rvy) &= (\diff \rvx)^\top \rvy + \rvx^\top \diff \rvy + D  b(t)d(t) \diff t
\end{aligned}
\end{align}
}



\subsection{It\^o's Formula's Application: Derivation of Fokker-Planck Equation}\label{subsec:rigorous-proof-fpe}
In this section, we apply It\^o's formula from \Cref{eq:Ito-scalar} to derive the Fokker–Planck equation, a PDE that characterizes the time evolution of the probability density $ p_t(\rvx) $ associated with the $ D $-dimensional diffusion process defined by the SDE in \Cref{eq:sde}.

\paragraph{Step 1: Apply It\^o's Formula.}
Let $\phi(\rvx, t)$ be a smooth test function $\phi \colon \mathbb{R}^D \times [0, T] \to \mathbb{R}$. By It\^o’s Formula:
\[
\diff \phi(\rvx_t, t) = \left( \frac{\partial \phi}{\partial t} + \nabla_{\rvx} \phi^\top \rvf(\rvx_t, t) + \frac{1}{2} g^2(t) \Tr[\nabla_{\rvx}^2 \phi] \right) \diff t + g(t) \nabla_{\rvx} \phi^\top \diff \rvw_t.
\]
\paragraph{Step 2: Take Expectation.}
Taking expectation and using that the It\^o integral has zero expectation
under the standard integrability conditions:
\[
\mathbb{E}[\diff \phi(\rvx_t, t)] = \mathbb{E} \left[ \left( \frac{\partial \phi}{\partial t} + \nabla_{\rvx} \phi^\top \rvf(\rvx_t, t) + \frac{1}{2} g^2(t) \Tr[\nabla_{\rvx}^2 \phi] \right) \diff t \right].
\]
\paragraph{Step 3: Express Expectation via Density.}
This expectation can be written as:
\[
\mathbb{E}[\diff \phi(\rvx_t, t)] = \int \left( \frac{\partial \phi}{\partial t} + \nabla_{\rvx} \phi^\top \rvf(\rvx, t) + \frac{1}{2} g^2(t) \Tr[\nabla_{\rvx}^2 \phi] \right) p_t(\rvx) \diff \rvx \diff t.
\]
\paragraph{Step 4: Integrate by Parts.}
Use integration by parts (divergence theorem) in $\mathbb{R}^D$:
\begin{align*}
    \int \nabla_{\rvx} \phi^\top \rvf p_t \diff \rvx &= -\int \phi \nabla_{\rvx} \cdot (\rvf p_t) \diff \rvx, \\
    \int \Tr[\nabla_{\rvx}^2 \phi] p_t  \diff \rvx &= \int \phi \Delta p_t \diff \rvx.
\end{align*}
\paragraph{Step 5: Substitute and Rearrange.}
Note that
\[
\mathbb{E}[\phi(\rvx_t,t)] = \int \phi(\rvx,t) p_t(\rvx) \diff\rvx.
\]
Differentiating in time, we have
\[
\frac{\diff}{\diff t}\mathbb{E}[\phi(\rvx_t,t)]
= \frac{\diff}{\diff t}\int \phi(\rvx,t) p_t(\rvx) \diff\rvx
= \int\Bigl(\partial_t\phi(\rvx,t) p_t(\rvx) + \phi(\rvx,t) \partial_t p_t(\rvx)\Bigr)\diff\rvx.
\]
Therefore,
\[
\mathbb{E}[\diff \phi(\rvx_t,t)]
= \frac{\diff}{\diff t}\mathbb{E}[\phi(\rvx_t,t)] \diff t
= \int\Bigl(\partial_t\phi(\rvx,t) p_t(\rvx) + \phi(\rvx,t) \partial_t p_t(\rvx)\Bigr)\diff\rvx \diff t.
\]
Equating this expression with Step~3, the terms
$\int \partial_t\phi\,p_t\,\diff\rvx\,\diff t$ cancel. Applying Step~4 to
the remaining spatial terms gives
\[
\int
\phi(\rvx,t)
\left[
\frac{\partial p_t}{\partial t}
+
\nabla_{\rvx}\cdot\bigl(\rvf(\rvx,t)p_t(\rvx)\bigr)
-
\frac{1}{2}g^2(t)\Delta p_t(\rvx)
\right]
\diff\rvx
=
0.
\]
Since this holds for every smooth test function $\phi$, we obtain the
Fokker--Planck equation
\[
\frac{\partial p_t(\rvx)}{\partial t}
=
-\nabla_{\rvx}\cdot\bigl(\rvf(\rvx,t)p_t(\rvx)\bigr)
+
\frac{1}{2}g^2(t)\Delta p_t(\rvx).
\]

\subsection{It\^o's Formula Application: Closed-Form Solution of a Linear SDE}\label{subsec:rigorous-proof-linear-sde}
This subsection demonstrates how to obtain a closed-form solution for a linear SDE by using an integration factor (similar to the ODE case) and It\^o's formula. The approach mirrors classical techniques for solving linear ODEs, but adapted to the stochastic setting.

We consider a linear SDE of the form
\begin{align}\label{eq:app-linear-sde}
    \mathrm{d}\rvx_t = f(t)\rvx_t \mathrm{d}t + g(t) \mathrm{d}\rvw_t,
\end{align}
where $f(t)$ and $g(t)$ are deterministic functions and $\rvw_t$ is a standard Wiener process. 
\paragraph{Closed-Form Solution of a Linear SDE.} We derive the explicit solution to the linear SDE in \Cref{eq:app-linear-sde} using the method of integrating factors. This type of forward SDE commonly arises in diffusion models (see \Cref{sec:dm-today}).

\subparagraph{Step 1: Define an Integration Factor.} Let
\[
\Psi(t) := \exp\left( -\int_0^t f(s) \mathrm{d}s \right),
\quad \text{and define} \quad
\rvy_t := \Psi(t)\rvx_t.
\]
\subparagraph{Step 2: Apply It\^o's Formula.}  
We apply It\^o's formula to the function $\rvh(\rvx, t) := \Psi(t)\rvx$. This is actually a special case of the It\^o product rule in \Cref{eq:ito-product}. Since $\Psi(t)$ is deterministic, there is no cross-variation term, and the formula simplifies to:
\begin{align*}
    \mathrm{d}\rvy_t 
    &= \mathrm{d}[\Psi(t)\rvx_t] \\
    &= \Psi'(t)\rvx_t \mathrm{d}t + \Psi(t) \mathrm{d}\rvx_t \\
    &= -f(t)\Psi(t)\rvx_t \mathrm{d}t + \Psi(t)\left[f(t)\rvx_t \mathrm{d}t + g(t) \mathrm{d}\rvw_t\right] \\
    &= \Psi(t)g(t) \mathrm{d}\rvw_t.
\end{align*}
Hence,
\[
\rvy_t = \rvy_0 + \int_0^t \Psi(s)g(s) \mathrm{d}\rvw(s) = \rvx_0 + \int_0^t \Psi(s)g(s) \mathrm{d}\rvw(s),
\]
since $\Psi(0) = 1$.
\subparagraph{Step 3: Solve for $\rvx_t$.}  
Using $\rvx_t = \Psi(t)^{-1}\rvy_t$, we obtain
\begin{align}\label{eq:lienar-sde-solution}
    \rvx_t &=  e^{ \int_0^t f(s)\mathrm{d}s}
    \left[ \rvx_0 + \int_0^t e^{ -\int_0^s f(r) \mathrm{d}r } g(s) \mathrm{d}\rvw(s) \right].
\end{align}
This provides an explicit solution to the vector-valued SDE.

Below, we demonstrate two alternative approaches to compute the analytical form of $p_t(\rvx_t|\rvx_0)$.

\paragraph{Analysis of the Closed-Form Solution.} \Cref{eq:lienar-sde-solution} reconfirms that $p_t(\rvx_t|\rvx_0)$ is Gaussian. To see this, define
\[
\phi(s) := e^{-\int_0^s f(u) \diff u} g(s),
\]
which is a deterministic matrix-valued function of time (assuming $f(u)$ and $g(s)$ are deterministic). The It\^o integral $\int_0^t \phi(s) \diff \rvw_s$ is then a zero-mean Gaussian random variable, as it is the stochastic integral of a deterministic function with respect to Brownian motion. Therefore, $\rvx_t|\rvx_0$ is an affine transformation of a Gaussian random variable and hence itself Gaussian. Its distribution is fully characterized by its conditional mean and covariance.

We define the conditional mean and covariance (given initial condition $\rvx_0$) as
\[
\rvm(t) := \mathbb{E}[\rvx_t | \rvx_0], \quad \rmP(t) := \mathbb{E}[(\rvx_t - \rvm(t))(\rvx_t - \rvm(t))^\top|\rvx_0].
\]

\subparagraph{Mean.}
Using linearity of expectation and the fact that the It\^o integral of a deterministic function has zero mean:
\begin{align*}
    \rvm(t) = \mathbb{E}\left[ e^{\int_0^t f(s) \diff s} \left( \rvx_0 + \int_0^t \phi(s) \diff \rvw_s \right) \bigg| \rvx_0 \right] = e^{\int_0^t f(s) \diff s} \rvx_0.
\end{align*}

\subparagraph{Covariance.}
Let  $\rvz_t := \int_0^t \phi(s) \diff \rvw_s$. Then $\rvx_t - \rvm(t) = A(t) \rvz_t$, so
\[
\rmP(t) = e^{2\int_0^t f(s) \diff s} \mathbb{E}[\rvz_t \rvz_t^\top].
\]
By It\^o isometry\footnote{It\^o's isometry links stochastic integrals to standard integrals in expectation; we omit the proof as it requires the full machinery of It\^o calculus. For a process $\bm{\psi} : [0,T] \to \mathbb{R}^{D \times D}$, It\^o isometry states
\[
\mathbb{E}\left[\left\| \int_0^T \bm{\psi}(t)   \mathrm{d}\rvw_t \right\|^2 \right]
= \mathbb{E}\left[\int_0^T \|\bm{\psi}(t)\|_F^2   \mathrm{d}t \right],
\]
where $\|\bm{\psi}(t)\|_F^2 = \sum_{i,j=1}^D |\psi_{ij}(t)|^2$ is the Frobenius norm. For $\bm{\psi}(t) \in \mathbb{R}^D$ (a vector), the integral is scalar and the isometry simplifies to
\[
\mathbb{E}\left[\left(\int_0^T \bm{\psi}(t)   \mathrm{d}\rvw_t\right)^2\right] = \mathbb{E}\left[\int_0^T \|\bm{\psi}(t)\|^2   \mathrm{d}t \right].
\]
},
\[
\mathbb{E}[\rvz_t \rvz_t^\top] = \left( \int_0^t \phi^2(s) \diff s \right) \rmI_D,
\]
hence,
\[
\rmP(t) = e^{2\int_0^t f(s) \diff s} \left( \int_0^t \left( e^{-\int_0^s f(u) \diff u} g(s) \right)^2 \diff s \right) \rmI_D.
\]
This shows the conditional covariance is isotropic.

\paragraph{Derivation of Mean and Variance ODEs in \Cref{eq:mean-var-evolution}.}
Alternatively, we can derive the moment evolution equations directly from the linear SDE \Cref{eq:app-linear-sde}.

\subparagraph{Mean Evolution.}
Taking the conditional expectation of both sides of the SDE and using linearity:
\[
\frac{\diff \rvm(t)}{\diff t} = \mathbb{E}[f(t) \rvx_t|\rvx_0] = f(t) \mathbb{E}[\rvx_t|\rvx_0] = f(t) \rvm(t).
\]

\vspace{0.3cm}

\subparagraph{Covariance Evolution.}
Define the centered process $ \tilde{\rvx}_t := \rvx_t - \rvm(t) $. Applying It\^o's product rule (see \Cref{eq:ito-product}):
\[
\diff \left( \tilde{\rvx}_t \tilde{\rvx}_t^\top \right)
= \diff \tilde{\rvx}_t \cdot \tilde{\rvx}_t^\top 
+ \tilde{\rvx}_t \cdot \diff \tilde{\rvx}_t^\top 
+ \diff \tilde{\rvx}_t \cdot \diff \tilde{\rvx}_t^\top.
\]

From the SDE, we compute:
\[
\diff \tilde{\rvx}_t = \diff \rvx_t - \diff \rvm(t) = f(t) \tilde{\rvx}_t  \diff t + g(t)  \diff \rvw_t.
\]

Substituting into the product rule and taking expectation:
\begin{align*}
    \frac{\diff \rmP(t)}{\diff t}
    &= \mathbb{E}[f(t) \tilde{\rvx}_t \tilde{\rvx}_t^\top + \tilde{\rvx}_t \tilde{\rvx}_t^\top f(t) + g^2(t) \rmI_D] \\
    &= 2 f(t) \rmP(t) + g^2(t) \rmI_D.
\end{align*}
Thus, we recover the moment evolution equations in \Cref{eq:mean-var-evolution}.

\newpage

\section{Change-of-Variable For Measures: Girsanov's Theorem in Diffusion Models}
Diffusion models harness SDEs to transform simple noise into rich data distributions. At the heart of this transformation lies a profound idea: we can reinterpret randomness by modifying only the deterministic part of an SDE (the drift) while preserving its underlying stochasticity. This is precisely where Girsanov’s theorem enters the picture.

\paragraph{The Core Idea.}
Consider an observed continuous trajectory that describes the data's evolution from time $t=0$ to $t=T$, denoted as $\rvx_{0:T}:=\{ \rvx_t|t \in [0,T] \}$. Girsanov's theorem addresses a fundamental question:
\begin{question}
    Given this single observed path, what is its likelihood if we assume it was generated by one SDE, versus if we assume it was generated by a different SDE?
\end{question}
We compare two hypothetical models for generating the same trajectory. They
have the same initial distribution and diffusion coefficient, but differ in
their deterministic ``push'' or drift function. We assume $\rvx_0$ has the same initial distribution for both assumed generating processes.

To build intuition, imagine $\rvx_{0:T}$ as a wiggly line drawn on paper.
One hypothesis assigns trajectory probabilities according to drift $\rvf$,
while another uses a different drift $\tilde{\rvf}$ with the same diffusion
coefficient. Girsanov's theorem compares the corresponding path measures and
quantifies how changing the drift changes the relative weighting assigned to
the same trajectories.

\paragraph{The Setup.}
Let $P_{\rvf}$ and $P_{\tilde{\rvf}}$ denote the path measures induced by
two SDEs with the same initial distribution and the same diffusion
coefficient, but different drifts:
\begin{align*}
    \diff\rvx_t
    &= \rvf(\rvx_t,t)\diff t + g(t)\diff\rvw_t,
    \\
    \diff\rvx_t
    &= \tilde{\rvf}(\rvx_t,t)\diff t + g(t)\diff\rvw_t.
\end{align*}
For simplicity, assume $g(t)>0$ on the interval of interest and the
standard integrability conditions required by Girsanov's theorem. Take
$P_{\tilde{\rvf}}$ as the reference path measure, and let $\rvw_t$ denote
the Brownian motion under this reference measure. Define
\[
\bm{\delta}_t
:=
\rvf(\rvx_t,t)-\tilde{\rvf}(\rvx_t,t).
\]

\paragraph{Girsanov's Likelihood Ratio.}
Girsanov's theorem gives the Radon--Nikodym derivative between the two path
measures:
\begin{mdframed}
\[
\frac{\diff P_{\rvf}}{\diff P_{\tilde{\rvf}}}(\rvx_{0:T})
=
\exp\left(
\int_0^T
\bm{\delta}_t^\top g(t)^{-1}\diff\rvw_t
-
\frac{1}{2}
\int_0^T
\big\|g(t)^{-1}\bm{\delta}_t\big\|^2
\diff t
\right).
\]
\end{mdframed}
This compact formula is an exponential of two integrals. The first is an It\^o integral, while the second is a standard Riemann integral. This ratio is crucial in diffusion models, allowing us to compare
different stochastic dynamics and to connect path-space discrepancies
with likelihood-based training objectives.

Girsanov's theorem is best understood as a change-of-variable formula for \emph{measures}. Just as a change of variables in calculus transforms an integral between coordinate systems via the Jacobian determinant, Girsanov's theorem provides the corresponding factor (the Radon–Nikodym derivative) to transform probabilities or expectations between two stochastic processes, when the drift changes but the diffusion remains the same.

\subsection{Girsanov’s Theorem as a Bridge Between Likelihood Training and Score Matching}\label{subsec:girsanov}

After understanding how Girsanov's theorem relates the likelihoods of a single
path under different drift assumptions, we now delve into its implications for
diffusion models.

Recall the forward SDE in diffusion models:
\[
\diff \rvx_t = \rvf(\rvx_t, t) \diff t + g(t) \diff \rvw_t,
\]
which induces a \emph{path distribution} $P$ over full trajectories
$\rvx_{0:T} := \{\rvx_t\}_{t=0}^T$ (that is, the joint law of the process over
the entire time interval).
The reverse-time SDE, parameterized by a learnable score function
$\rvs_{\bm{\phi}}(\rvx_t, t)$, is given by
\[
\diff \rvx_t = \big[\rvf(\rvx_t, t) - g^2(t) \rvs_{\bm{\phi}}(\rvx_t, t)\big] \diff t
+ g(t) \diff \bar{\rvw}_t,
\]
which in turn defines another path distribution
$P_{\bm{\phi}}$ over trajectories.

\paragraph{Two Notions in Diffusion Models.}
It is useful to distinguish two levels at which two stochastic processes can
agree:
\begin{itemize}
    \item \textbfs{Concept 1. Marginal Distribution Matching:}
    The processes have the same distribution $p_t(\rvx_t)$ at every fixed
    time $t$. Thus, their time-indexed ``snapshots'' agree.

    \item \textbfs{Concept 2. Joint Path Distribution Matching:}
    The processes have the same probability law over entire trajectories.
    This stronger notion also matches how states at different times are
    coupled, including their transition laws and temporal dependencies.
\end{itemize}

Matching the full path law automatically implies marginal matching, since
each $p_t$ is obtained by projecting the path law onto the state at time
$t$. The converse is generally false: two processes can have identical
marginals at every time while inducing different couplings between those
marginals. The SDE and its associated PF-ODE provide an important example:
they share the same marginal density path, but generally have different
sample trajectories and path laws.

\paragraph{Girsanov Connects Score Error to Path-Law Mismatch.}
The exact reverse-time SDE satisfies the stronger notion above. Under
standard regularity conditions, when initialized from the exact terminal
marginal $p_T$ and equipped with the true score
$\nabla_{\rvx}\log p_t(\rvx)$, it is the time reversal of the forward
diffusion and therefore reproduces its full time-reversed path law.

Let $P^{\mathrm{rev}}$ denote this exact reverse-time path law, and let
$P_{\bm{\phi}}$ denote the learned reverse-time path law obtained by
replacing the true score with $\rvs_{\bm{\phi}}$ and initializing from
$p_{\mathrm{prior}}$. Decomposing the path-space KL into the mismatch between the initial
distributions and the conditional path laws, and applying Girsanov's
theorem to the latter after expressing both processes using the same
increasing reverse-time clock, gives, in the original diffusion-time
label $t$,
\begin{align}\label{eq:girsanov-KL}
\begin{aligned}
\mathcal{D}_{\mathrm{KL}}
\bigl(P^{\mathrm{rev}}\|P_{\bm{\phi}}\bigr)
=
&\;
\mathcal{D}_{\mathrm{KL}}
\bigl(p_T\|p_{\mathrm{prior}}\bigr)
\\
&+
\frac{1}{2}
\int_0^T
g^2(t)
\E_{\rvx_t\sim p_t}
\left[
\left\|
\rvs_{\bm{\phi}}(\rvx_t,t)
-
\nabla_{\rvx}\log p_t(\rvx_t)
\right\|^2
\right]
\diff t .
\end{aligned}
\end{align}
The first term measures the mismatch between the true terminal distribution
and the prior used to initialize generation; it vanishes when
$p_T=p_{\mathrm{prior}}$. The second term measures the accumulated mismatch
between the learned and exact reverse drifts.

Thus, score matching with the particular likelihood weighting $g^2(t)$
has a direct path-space interpretation: minimizing the weighted score error
minimizes the KL divergence between the exact and learned reverse path laws,
up to the fixed terminal-distribution term. When denoising score matching is
used instead of marginal score matching, the corresponding objectives differ
by a parameter-independent constant. A different time weighting generally
does not correspond to this same path-space KL.

\paragraph{Connection to Data Likelihood.}
The path-space KL also controls the discrepancy between the generated and
data marginals. Since taking the endpoint of a trajectory is a measurable
projection and KL divergence cannot increase under marginalization,
\begin{align}\label{eq:girsanov-likelihood}
\mathcal{D}_{\mathrm{KL}}
\bigl(p_{\mathrm{data}}\|p_{\bm{\phi}}\bigr)
\leq
\mathcal{D}_{\mathrm{KL}}
\bigl(P^{\mathrm{rev}}\|P_{\bm{\phi}}\bigr),
\end{align}
where $p_{\bm{\phi}}$ denotes the generated marginal at data time.

Combining this with \Cref{eq:girsanov-KL} shows that likelihood-weighted
score matching, together with the terminal-distribution mismatch, provides
an upper bound on the data-side KL divergence and therefore on the expected
negative log-likelihood up to a constant independent of $\bm{\phi}$
\citep{song2021maximum}.
\chapter{Supplementary Materials and Proofs}\label{app:proof}
\section{Variational Perspective}\label{app-sec:var-proof}

\subsection{Theorem~\ref{thm:equiv-marginal-kl}: Equivalence Between Marginal and Conditional KL Minimization}\label{subsec:proof-equiv-marginal-kl}
\paragraph{\textit{Proof.}} \textbfs{Derivation of \Cref{eq:kl-matching}.}

We start by expanding the right-hand side expectation:
\begin{align*}
    &\mathbb{E}_{p(\rvx_0, \rvx_i)}\left[
    \mathcal{D}_{\text{KL}} \big(p(\rvx_{i-1}|\rvx_i, \rvx_0) \| p_\phi(\rvx_{i-1}|\rvx_i)\big)
    \right]
    \\
    =& \int \int p(\rvx_0, \rvx_i)   
    \mathcal{D}_{\text{KL}} \big(p(\rvx_{i-1}|\rvx_i, \rvx_0) \| p_\phi(\rvx_{i-1}|\rvx_i)\big) 
    \diff\rvx_0 \diff\rvx_i.
\end{align*}
By the definition of KL divergence,
\begin{align*}
    &\mathcal{D}_{\text{KL}} \big(p(\rvx_{i-1}|\rvx_i, \rvx_0) \| p_\phi(\rvx_{i-1}|\rvx_i)\big)
    = \int p(\rvx_{i-1}|\rvx_i, \rvx_0) 
    \log \frac{p(\rvx_{i-1}|\rvx_i, \rvx_0)}{p_\phi(\rvx_{i-1}|\rvx_i)} 
    \diff\rvx_{i-1}.
\end{align*}
Substituting this into the expectation, we have
\begin{align*}
    &\int \int \int p(\rvx_0, \rvx_i)  p(\rvx_{i-1}|\rvx_i, \rvx_0) 
    \log \frac{p(\rvx_{i-1}|\rvx_i, \rvx_0)}{p_\phi(\rvx_{i-1}|\rvx_i)} 
    \diff\rvx_{i-1} \diff\rvx_0 \diff\rvx_i.
\end{align*}
Using the chain rule of probability,
\begin{align*}
    p(\rvx_0, \rvx_i) 
    &= p(\rvx_i)   p(\rvx_0|\rvx_i),
\end{align*}
we rewrite the integral as
\begin{align*}
    &\int p(\rvx_i) \int p(\rvx_0|\rvx_i) \int p(\rvx_{i-1}|\rvx_i, \rvx_0)
    \log \frac{p(\rvx_{i-1}|\rvx_i, \rvx_0)}{p_\phi(\rvx_{i-1}|\rvx_i)} 
    \diff\rvx_{i-1} \diff\rvx_0 \diff\rvx_i.
\end{align*}
This allows us to express the expectation in nested form:
\begin{align*}
    &\mathbb{E}_{p(\rvx_i)} \left[
    \mathbb{E}_{p(\rvx_0|\rvx_i)} \left[
    \mathbb{E}_{p(\rvx_{i-1}|\rvx_i, \rvx_0)} \left[
    \log \frac{p(\rvx_{i-1}|\rvx_i, \rvx_0)}{p_\phi(\rvx_{i-1}|\rvx_i)}
    \right]
    \right]
    \right].
\end{align*}

Next, we apply the decomposition of the logarithm:
\begin{align*}
    \log \frac{p(\rvx_{i-1}|\rvx_i, \rvx_0)}{p_\phi(\rvx_{i-1}|\rvx_i)}
    =\log \frac{p(\rvx_{i-1}|\rvx_i, \rvx_0)}{p(\rvx_{i-1}|\rvx_i)} 
    + \log \frac{p(\rvx_{i-1}|\rvx_i)}{p_\phi(\rvx_{i-1}|\rvx_i)}.
\end{align*}
Substituting this back into the expectation gives two terms:
\begin{align*}
    &\mathbb{E}_{p(\rvx_i)} \left[
    \mathbb{E}_{p(\rvx_0|\rvx_i)} \left[
    \mathbb{E}_{p(\rvx_{i-1}|\rvx_i, \rvx_0)} \left[
    \log \frac{p(\rvx_{i-1}|\rvx_i, \rvx_0)}{p(\rvx_{i-1}|\rvx_i)}
    \right]
    \right]
    \right]
    \\
    &+
    \mathbb{E}_{p(\rvx_i)} \left[
    \mathbb{E}_{p(\rvx_0|\rvx_i)} \left[
    \mathbb{E}_{p(\rvx_{i-1}|\rvx_i, \rvx_0)} \left[
    \log \frac{p(\rvx_{i-1}|\rvx_i)}{p_\phi(\rvx_{i-1}|\rvx_i)}
    \right]
    \right]
    \right].
\end{align*}
Since the second logarithmic term does not depend on $\rvx_0$, by the law of total probability
\begin{align*}
    \mathbb{E}_{p(\rvx_0|\rvx_i)} \left[
    \mathbb{E}_{p(\rvx_{i-1}|\rvx_i, \rvx_0)} \left[
    \log \frac{p(\rvx_{i-1}|\rvx_i)}{p_\phi(\rvx_{i-1}|\rvx_i)}
    \right]
    \right]
    = \mathbb{E}_{p(\rvx_{i-1}|\rvx_i)} \left[
    \log \frac{p(\rvx_{i-1}|\rvx_i)}{p_\phi(\rvx_{i-1}|\rvx_i)}
    \right].
\end{align*}
Similarly, the first term is the KL divergence
\begin{align*}
    \mathbb{E}_{p(\rvx_0|\rvx_i)} \left[
    \mathcal{D}_{\text{KL}} \big(p(\rvx_{i-1}|\rvx_i, \rvx_0)  \|  p(\rvx_{i-1}|\rvx_i)\big)
    \right].
\end{align*}
Putting it all together, we obtain the decomposition:
\begin{align*}
    &\mathbb{E}_{p(\rvx_0, \rvx_i)} \left[
    \mathcal{D}_{\text{KL}} \big(p(\rvx_{i-1}|\rvx_i, \rvx_0)  \|  p_\phi(\rvx_{i-1}|\rvx_i)\big)
    \right]
    \\
    =& \mathbb{E}_{p(\rvx_i)} \left[
    \mathbb{E}_{p(\rvx_0|\rvx_i)} \left[
    \mathcal{D}_{\text{KL}} \big(p(\rvx_{i-1}|\rvx_i, \rvx_0)  \|  p(\rvx_{i-1}|\rvx_i)\big)
    \right]
    \right]
    \\
    &+
    \mathbb{E}_{p(\rvx_i)} \left[
    \mathcal{D}_{\text{KL}} \big(p(\rvx_{i-1}|\rvx_i)  \|  p_\phi(\rvx_{i-1}|\rvx_i)\big)
    \right].
\end{align*}

\textbfs{Proof of Optimality.} To prove:
\[
p^*(\rvx_{i-1}|\rvx_i)
=
p(\rvx_{i-1}|\rvx_i)
=
\mathbb{E}_{p(\rvx_0|\rvx_i)}
\left[
p(\rvx_{i-1}|\rvx_i,\rvx_0)
\right],
\qquad
\rvx_i \sim p_i.
\]
The first identity follows from the fact that the KL divergence $\mathcal{D}_{\mathrm{KL}}(p \| p_{\bm{\phi}})$ is minimized when $p^* = p$, assuming the parameterization is sufficiently expressive. The second identity follows directly from the law of total probability.
\hfill$\blacksquare$

\subsection{Theorem~\ref{thm:ddpm-elbo}: ELBO of Diffusion Model}\label{app:vlb-proof}

\paragraph{\textit{Proof.}} For notational simplicity, we denote
\[
\rvx_{1:L}:=(\rvx_1,\ldots,\rvx_L).
\]

\textbfs{Step 1: Apply Jensen's Inequality.}
The marginal log-likelihood is
\[
\log p_{\bm{\phi}}(\rvx_0)
=
\log \int p_{\bm{\phi}}(\rvx_0,\rvx_{1:L})\,\diff \rvx_{1:L},
\]
where the joint generative distribution is
\[
p_{\bm{\phi}}(\rvx_0,\rvx_{1:L})
=
p_{\text{prior}}(\rvx_L)\prod_{i=1}^L p_{\bm{\phi}}(\rvx_{i-1}| \rvx_i).
\]

We introduce the forward noising distribution
\[
p(\rvx_{1:L}| \rvx_0)
=
\prod_{i=1}^L p(\rvx_i| \rvx_{i-1}),
\]
and rewrite
\[
\log p_{\bm{\phi}}(\rvx_0)
=
\log \int
p(\rvx_{1:L}| \rvx_0)
\frac{p_{\bm{\phi}}(\rvx_0,\rvx_{1:L})}{p(\rvx_{1:L}| \rvx_0)}
\,\diff \rvx_{1:L}.
\]
Applying Jensen's inequality yields
\[
\log p_{\bm{\phi}}(\rvx_0)
\ge
\mathbb{E}_{p(\rvx_{1:L}| \rvx_0)}
\left[
\log
\frac{p_{\bm{\phi}}(\rvx_0,\rvx_{1:L})}{p(\rvx_{1:L}| \rvx_0)}
\right]
=: \mathcal{L}_{\text{ELBO}}(\rvx_0;\bm{\phi}),
\]
and hence
\[
-\log p_{\bm{\phi}}(\rvx_0)
\le
-\mathcal{L}_{\text{ELBO}}(\rvx_0;\bm{\phi}).
\]

\textbfs{Step 2: Expand the ELBO.}
Substituting the factorizations into the ELBO gives
\begin{align*}
\mathcal{L}_{\text{ELBO}}(\rvx_0;\bm{\phi})
&=
\mathbb{E}_{p(\rvx_{1:L}| \rvx_0)}
\Bigg[
\log p_{\text{prior}}(\rvx_L)
+
\sum_{i=1}^L \log p_{\bm{\phi}}(\rvx_{i-1}| \rvx_i)
-
\sum_{i=1}^L \log p(\rvx_i| \rvx_{i-1})
\Bigg].
\end{align*}
Therefore,
\begin{align*}
-\mathcal{L}_{\text{ELBO}}(\rvx_0;\bm{\phi})
&=
\mathbb{E}_{p(\rvx_{1:L}| \rvx_0)}
\Bigg[
\log \frac{p(\rvx_L| \rvx_0)}{p_{\text{prior}}(\rvx_L)}
+
\sum_{i=2}^L
\log \frac{p(\rvx_{i-1}| \rvx_i,\rvx_0)}{p_{\bm{\phi}}(\rvx_{i-1}| \rvx_i)}
-
\log p_{\bm{\phi}}(\rvx_0| \rvx_1)
\Bigg].
\end{align*}
Here we used the identity
\[
p(\rvx_{1:L}| \rvx_0)
=
p(\rvx_L| \rvx_0)\prod_{i=2}^L p(\rvx_{i-1}| \rvx_i,\rvx_0),
\]
which follows from the Markov structure of the forward process.

Now group the terms according to their dependence:
\begin{align*}
-\mathcal{L}_{\text{ELBO}}(\rvx_0;\bm{\phi})
&=
\underbrace{
\mathbb{E}_{p(\rvx_L| \rvx_0)}
\left[
\log \frac{p(\rvx_L| \rvx_0)}{p_{\text{prior}}(\rvx_L)}
\right]
}_{\mathcal{L}_{\text{prior}}(\rvx_0)}
+
\underbrace{
\mathbb{E}_{p(\rvx_1| \rvx_0)}
\left[
-\log p_{\bm{\phi}}(\rvx_0| \rvx_1)
\right]
}_{\mathcal{L}_{\text{recon.}}(\rvx_0;\bm{\phi})}
\\
&\quad+
\underbrace{
\sum_{i=2}^L
\mathbb{E}_{p(\rvx_i| \rvx_0)}
\left[
\mathcal{D}_{\mathrm{KL}}
\big(
p(\rvx_{i-1}| \rvx_i,\rvx_0)
\|
p_{\bm{\phi}}(\rvx_{i-1}| \rvx_i)
\big)
\right]
}_{\mathcal{L}_{\text{diffusion}}(\rvx_0;\bm{\phi})}.
\end{align*}
This proves the claimed decomposition.
\hfill$\blacksquare$

\clearpage
\newpage

\section{Score-Based Perspective}\label{app-sec:score-proof}

\subsection{Proposition~\ref{sm-trace}: Tractable Score Matching via Integration by Parts}\label{app-sec:sm-trace}
\paragraph{\textit{Proof.}}

 \textbfs{Expanding $\mathcal{L}_{\text{SM}}(\bm{\phi})$. }
        Let us expand the squared difference inside the expectation:
        \begin{align*}
        \mathcal{L}_{\text{SM}}(\bm{\phi}) &= \frac{1}{2}\mathbb{E}_{\rvx \sim p_{\text{data}}(\rvx)} \left[ \norm{\rvs_{\bm{\phi}}(\rvx)}_2^2 - 2 \langle \rvs_{\bm{\phi}}(\rvx), \rvs(\rvx) \rangle + \norm{\rvs(\rvx)}_2^2 \right] \\
        &= \frac{1}{2}\mathbb{E}_{\rvx \sim p_{\text{data}}(\rvx)} \left[ \norm{\rvs_{\bm{\phi}}(\rvx)}_2^2 \right] - \mathbb{E}_{\rvx \sim p_{\text{data}}(\rvx)} \left[ \langle \rvs_{\bm{\phi}}(\rvx), \rvs(\rvx) \rangle \right] \\
        &\quad + \frac{1}{2}\mathbb{E}_{\rvx \sim p_{\text{data}}(\rvx)} \left[ \norm{\rvs(\rvx)}_2^2 \right].
        \end{align*}
        We now focus on the cross-product term:
        \[
        \mathbb{E}_{\rvx \sim p_{\text{data}}(\rvx)} \left[ \langle \rvs_{\bm{\phi}}(\rvx), \rvs(\rvx) \rangle \right].
        \]
        Using the fact that
        \[
        \nabla_{\rvx} \log p_{\text{data}}(\rvx) = \frac{\nabla_{\rvx}p_{\text{data}}(\rvx)}{p_{\text{data}}(\rvx)},
        \]
        and assuming $p_{\text{data}}(\rvx)$ is not zero (e.g., on its support), the cross-product term becomes:
        \begin{align*}
            \mathbb{E}_{\rvx \sim p_{\text{data}}(\rvx)} \left[ \langle \rvs_{\bm{\phi}}(\rvx), \rvs(\rvx) \rangle \right] 
             &= \int \rvs_{\bm{\phi}}(\rvx)^\top  \nabla_{\rvx} \log p_{\text{data}}(\rvx) p_{\text{data}}(\rvx) \diff \rvx \\
             &= \int \rvs_{\bm{\phi}}(\rvx)^\top  \nabla_{\rvx}p_{\text{data}}(\rvx) \diff \rvx \\
             &= \sum_{i=1}^{D} \int  s_{\bm{\phi}}^{(i)}(\rvx) \partial_{x_i}p_{\text{data}}(\rvx) \diff \rvx,
        \end{align*}
        where $s_{\bm{\phi}}^{(i)}(\rvx)$ is the $i$-th component of the score function 
        \[
        \rvs_{\bm{\phi}} = \left( s_{\bm{\phi}}^{(1)}, s_{\bm{\phi}}^{(2)}, \dots, s_{\bm{\phi}}^{(D)} \right).
        \]
        \proofparagraph{Integration by Parts.}
        We use the following integration-by-parts formula~\citep{evans10}, which is derived from standard calculus:
        \begin{lemma*}
            Let $u, v$ be differentiable real-valued functions on a ball $\mathbb{B}(\bm{0}, R) \subset \mathbb{R}^D$ of radius $R>0$. Then for $i = 1, \ldots, D$, the formula holds:
        \[
        \int_{\mathbb{B}(\bm{0}, R)} u   \partial_{x_i} v   \diff \rvx = -\int_{\mathbb{B}(\bm{0}, R)} v   \partial_{x_i} u   \diff \rvx + \int_{\partial \mathbb{B}(\bm{0}, R)} u v \nu_i   \diff S,
        \]
        where $\nu = (\nu_1, \ldots, \nu_D)$ is the outward unit normal to the boundary $\partial \mathbb{B}(\bm{0}, R)$---a sphere with radius $R>0$, and $\diff S$ is the surface measure on $\partial \mathbb{B}(\bm{0}, R)$.
        \end{lemma*}
        We apply this formula to $u(\rvx) := s_{\bm{\phi}}^{(i)}(\rvx)$ and $v(\rvx) = p_{\text{data}}(\rvx)$ for all $i = 1, \dots, D$, assuming that
        \[
        |u(\rvx) v(\rvx)| \to 0 \quad \text{as } R \to \infty.
        \]
        Summing the results over all $i = 1, \dots, D$, we get:
        \begin{align*}
            \mathbb{E}_{\rvx \sim p_{\text{data}}(\rvx)} \left[ \langle \rvs_{\bm{\phi}}(\rvx), \rvs(\rvx) \rangle \right] 
             &= - \sum_{i=1}^{D} \int  \partial_{x_i}s_{\bm{\phi}}^{(i)}(\rvx) p_{\text{data}}(\rvx) \diff \rvx \\
             &= - \mathbb{E}_{\rvx \sim p_{\text{data}}(\rvx)} \left[\Tr\left(\nabla_{\rvx}\rvs_{\bm{\phi}}(\rvx)\right) \right].
        \end{align*}
        Combining all results, we have:
        \begin{align*}
        \mathcal{L}_{\text{SM}}(\bm{\phi}) &= \underbrace{\mathbb{E}_{\rvx \sim p_{\text{data}}(\rvx)} \left[\Tr\left(\nabla_{\rvx}\rvs_{\bm{\phi}}(\rvx)\right) + \frac{1}{2}  \norm{\rvs_{\bm{\phi}}(\rvx)}_2^2 \right]}_{\widetilde{\mathcal{L}}_{\text{SM}}(\bm{\phi})} \\
        &\quad + \underbrace{\frac{1}{2}\mathbb{E}_{\rvx \sim p_{\text{data}}(\rvx)} \left[ \norm{\rvs(\rvx)}_2^2 \right]}_{=:C},
        \end{align*}
        where $C$ depends only on the distribution $p_{\text{data}}$, which concludes the proof.
\hfill$\blacksquare$

\subsection{Theorem~\ref{thm:sm-dsm}: Equivalence Between SM and DSM Minimization}\label{app-sec:sm-dsm}
\paragraph{\textit{Proof.}} Expanding both $\mathcal{L}_{\text{SM}}(\bm{\phi}; \sigma)$ and $\mathcal{L}_{\text{DSM}}(\bm{\phi}; \sigma)$, we have:
\begin{align*}
 &\mathcal{L}_{\text{SM}}(\bm{\phi}; \sigma) = \frac{1}{2} \mathbb{E}_{\tilde{\rvx} \sim p_\sigma(\tilde{\rvx})} 
 \begin{aligned}[t]
         \Big[ 
    \norm{\rvs_{\bm{\phi}}(\tilde{\rvx}; \sigma)}_2^2 
    &- 2 \rvs_{\bm{\phi}}(\tilde{\rvx}; \sigma)^\top  \nabla_{\tilde{\rvx}} \log p_\sigma(\tilde{\rvx}) \\
    &+ \norm{\nabla_{\tilde{\rvx}} \log p_\sigma(\tilde{\rvx})}_2^2 
    \Big], 
 \end{aligned}
\\
&\mathcal{L}_{\text{DSM}}(\bm{\phi}; \sigma) = \frac{1}{2} \mathbb{E}_{ p_{\text{data}}(\rvx)p_{\sigma}(\tilde{\rvx} | \rvx)}
 \begin{aligned}[t]
         \Big[
    \norm{\rvs_{\bm{\phi}}(\tilde{\rvx}; \sigma)}_2^2 
    &- 2 \rvs_{\bm{\phi}}(\tilde{\rvx}; \sigma)^\top \nabla_{\tilde{\rvx}} \log p_{\sigma}(\tilde{\rvx} | \rvx) \\
    &+ \norm{\nabla_{\tilde{\rvx}} \log p_{\sigma}(\tilde{\rvx} | \rvx)}_2^2 
    \Big].
 \end{aligned}
\end{align*}
Subtracting the two equations yields:
\begin{align*}
 &\mathcal{L}_{\text{SM}}(\bm{\phi}; \sigma) -\mathcal{L}_{\text{DSM}}(\bm{\phi}; \sigma) 
 \\= \frac{1}{2} &\Bigg( \mathbb{E}_{\tilde{\rvx} \sim p_\sigma(\tilde{\rvx})} \norm{\rvs_{\bm{\phi}}(\tilde{\rvx}; \sigma)}_2^2 - \mathbb{E}_{ p_{\text{data}}(\rvx)p_{\sigma}(\tilde{\rvx} | \rvx)} \norm{\rvs_{\bm{\phi}}(\tilde{\rvx}; \sigma)}_2^2\Bigg)
 \\ \quad - &  
 \begin{aligned}[t]
 \Bigg(
      \mathbb{E}_{\tilde{\rvx} \sim p_\sigma(\tilde{\rvx})}  &\left[ \rvs_{\bm{\phi}}(\tilde{\rvx}; \sigma)^\top  \nabla_{\tilde{\rvx}} \log p_\sigma(\tilde{\rvx}) \right] \\ 
      &- \mathbb{E}_{ p_{\text{data}}(\rvx)p_{\sigma}(\tilde{\rvx} | \rvx)}\left[ \rvs_{\bm{\phi}}(\tilde{\rvx}; \sigma)^\top  \nabla_{\tilde{\rvx}} \log p_\sigma(\tilde{\rvx}| \rvx)\right]
      \Bigg)
 \end{aligned}
 \\ \quad +  \frac{1}{2}&\Bigg(\mathbb{E}_{\tilde{\rvx} \sim p_\sigma(\tilde{\rvx})}  \norm{\nabla_{\tilde{\rvx}} \log p_\sigma(\tilde{\rvx})}_2^2  -  \mathbb{E}_{ p_{\text{data}}(\rvx)p_{\sigma}(\tilde{\rvx} | \rvx)} \norm{\nabla_{\tilde{\rvx}} \log p_{\sigma}(\tilde{\rvx} | \rvx)}_2^2   \Bigg).
\end{align*}
Next, we address the three terms one at a time. For the first term, since $p_\sigma(\tilde{\rvx}) = \int p_{\sigma}(\tilde{\rvx} | \rvx) p_{\text{data}}(\rvx) \diff\rvx$, we can rewrite it as:
\begin{align*}
    \mathbb{E}_{\tilde{\rvx} \sim p_\sigma(\tilde{\rvx})} \norm{\rvs_{\bm{\phi}}(\tilde{\rvx}; \sigma)}_2^2 
    &= \int \Big( \int p_{\sigma}(\tilde{\rvx} | \rvx) p_{\text{data}}(\rvx)  \diff\rvx \Big) \norm{\rvs_{\bm{\phi}}(\tilde{\rvx}; \sigma)}_2^2 \diff \tilde{\rvx} 
    \\ &= \int p_{\text{data}}(\rvx) \int p_{\sigma}(\tilde{\rvx} | \rvx) \norm{\rvs_{\bm{\phi}}(\tilde{\rvx}; \sigma)}_2^2 \diff \tilde{\rvx} \diff\rvx
    \\ &= \mathbb{E}_{ p_{\text{data}}(\rvx)p_{\sigma}(\tilde{\rvx} | \rvx)} \norm{\rvs_{\bm{\phi}}(\tilde{\rvx}; \sigma)}_2^2.
\end{align*}
Thus, the first term is zero. For the second term:
\begin{align}\label{eq:dsm-sec-term}
\begin{aligned}
          &\mathbb{E}_{\tilde{\rvx} \sim p_\sigma(\tilde{\rvx})}\left[ \rvs_{\bm{\phi}}(\tilde{\rvx}; \sigma)^\top  \nabla_{\tilde{\rvx}} \log p_\sigma(\tilde{\rvx}) \right] 
      \\=&\int  p_\sigma(\tilde{\rvx}) \rvs_{\bm{\phi}}(\tilde{\rvx}; \sigma)^\top  \frac{\nabla_{\tilde{\rvx}} p_\sigma(\tilde{\rvx})}{p_\sigma(\tilde{\rvx})}  \diff\tilde\rvx
        \\=&\int   \rvs_{\bm{\phi}}(\tilde{\rvx}; \sigma)^\top  \nabla_{\tilde{\rvx}} \int p_{\sigma}(\tilde{\rvx} | \rvx) p_{\text{data}}(\rvx) \diff\rvx \diff\tilde\rvx
    \\=&\int \int  \rvs_{\bm{\phi}}(\tilde{\rvx}; \sigma)^\top    \nabla_{\tilde{\rvx}}p_{\sigma}(\tilde{\rvx} | \rvx) p_{\text{data}}(\rvx) \diff\tilde\rvx \diff\rvx 
      \\=  &\mathbb{E}_{ p_{\text{data}}(\rvx)p_{\sigma}(\tilde{\rvx} | \rvx)}\left[ \rvs_{\bm{\phi}}(\tilde{\rvx}; \sigma)^\top  \nabla_{\tilde{\rvx}} \log p_\sigma(\tilde{\rvx}| \rvx)\right].
\end{aligned}
\end{align}
Thus, it is also zero. For the third term, note that:
\[
C:=\frac{1}{2}\Bigg(\mathbb{E}_{\tilde{\rvx} \sim p_\sigma(\tilde{\rvx})}  \norm{\nabla_{\tilde{\rvx}} \log p_\sigma(\tilde{\rvx})}_2^2  -  \mathbb{E}_{ p_{\text{data}}(\rvx)p_{\sigma}(\tilde{\rvx} | \rvx)} \norm{\nabla_{\tilde{\rvx}} \log p_{\sigma}(\tilde{\rvx} | \rvx)}_2^2   \Bigg)
\]
depends only on $p_{\text{data}}(\rvx)$ and $p_{\sigma}(\tilde{\rvx}|\rvx)$, and hence it is constant with respect to $\bm{\phi}$.
\hfill$\blacksquare$

\subsection{Lemma~\ref{tweedie}: Tweedie's Formula}\label{app-sec:tweedie}
We first state a more general form of Tweedie’s formula, which considers time-dependent Gaussian perturbations, and we provide its proof below.

\paragraph{Tweedie’s Identity with Time-Dependent Parameters.}
Let
$\rvx_t|\rvx_0 \sim
\mathcal{N}\big(\cdot; \alpha_t\rvx_0, \sigma_t^2\rmI\big)$. Then Tweedie's formula says
\begin{align*}
\alpha_t \mathbb{E}_{\mathbf{x}_0 \sim p(\mathbf{x}_0|\rvx_t)}[\mathbf{x}_0|\rvx_t] = \rvx_t + \sigma_t^2 \nabla_{\rvx_t} \log p_t(\rvx_t),
\end{align*}
where the expectation is taken over the posterior distribution $p(\mathbf{x}_0|\rvx_t)$ of $\mathbf{x}_0$ given the observed $\rvx_t$, and $p_t(\rvx_t)$ is the marginal density of $\rvx_t$.

\paragraph{\textit{Proof.}}

\paragraph{Marginal Density and Its Score.}
We recall that the marginal density of $\rvx_t$ is given by
\begin{align*}
p_t(\rvx_t) = \int p_t(\rvx_t|\rvx_0)  p_0(\rvx_0)  \diff\rvx_0.
\end{align*}
We now compute the score function:
\begin{align*}
\nabla_{\rvx_t} \log p_t(\rvx_t) 
= \frac{\nabla_{\rvx_t} p_t(\rvx_t)}{p_t(\rvx_t)} = \frac{1}{p_t(\rvx_t)} \int \nabla_{\rvx_t} p_t(\rvx_t|\rvx_0)  p_0(\rvx_0)  \diff\rvx_0.
\end{align*}
We therefore need to compute the gradient of the conditional density.

\paragraph{Gradient of the Conditional and Rearrangement.}
The gradient of the conditional Gaussian density is:
\begin{align*}
\nabla_{\rvx_t} p_t(\rvx_t|\rvx_0) 
= -p_t(\rvx_t|\rvx_0) \cdot \sigma_t^{-2} (\rvx_t - \alpha_t \rvx_0).
\end{align*}
Substituting this into the previous expression, we have:
\begin{align*}
\nabla_{\rvx_t} p_t(\rvx_t) 
&= \int \nabla_{\rvx_t} p_t(\rvx_t|\rvx_0)  p_0(\rvx_0)  \diff\rvx_0 \\
&= -\sigma_t^{-2} \int (\rvx_t - \alpha_t \rvx_0)  p_t(\rvx_t|\rvx_0)  p_0(\rvx_0)  \diff\rvx_0 \\
&= -\sigma_t^{-2} \int (\rvx_t - \alpha_t \rvx_0)  p(\rvx_0|\rvx_t)  p_t(\rvx_t)  \diff\rvx_0 \\
&= -p_t(\rvx_t)  \sigma_t^{-2} \left(\rvx_t - \alpha_t  \mathbb{E}_{p(\rvx_0|\rvx_t)}[\rvx_0\vert \rvx_t] \right).
\end{align*}
Dividing both sides by $p_t(\rvx_t)$, we obtain:
\begin{align*}
\nabla_{\rvx_t} \log p_t(\rvx_t) 
= -\sigma_t^{-2} \left(\rvx_t - \alpha_t  \mathbb{E}_{p(\rvx_0|\rvx_t)}[\rvx_0\vert \rvx_t] \right).
\end{align*}
Rearranging yields:
\begin{align*}
\rvx_t + \sigma_t^{2} \nabla_{\rvx_t} \log p_t(\rvx_t) 
= \alpha_t  \mathbb{E}_{p(\rvx_0|\rvx_t)}[\rvx_0\vert \rvx_t].
\end{align*}
This completes the derivation.

\subsection{Stein’s Identity and Surrogate SURE Objective}\label{app-sec:stein-identity}

\paragraph{Stein's Identity.} Stein’s identity is the integration-by-parts technique that turns expectations under an unknown density  into expectations of observable functions and their derivatives, which cancels the partition function and enables unbiased, tractable objectives and tests without ever evaluating the unknown density or the partition function. We begin with the simplest one-dimensional case and then extend it to the form needed to prove the surrogate loss for SURE.
\subparagraph{1D, Standard Normal Case.}
If $z \sim \mathcal{N}(0,1)$ and $f$ has suitable decay, then Stein's identity states:
\[
\E[f'(z)]  =  \E[z f(z)] .
\]
Denote $\phi(z) := \tfrac{1}{\sqrt{2\pi}}  e^{-z^2/2}$, the one-dimensional standard normal density.
The proof follows by integration by parts, using $\phi'(z) = -z\phi(z)$, together with the vanishing boundary term. 
To see this precisely, we compute
\[
\E[f'(z)] 
= \int_{-\infty}^{\infty} f'(z)  \phi(z)  \diff z.
\]
By integration by parts, with $u=\phi(z)$ and $\diff v=f'(z)  \diff z$, we obtain
\[
\int f'(z)  \phi(z)  \diff z
= \Big[ f(z)  \phi(z) \Big]_{-\infty}^{\infty}
- \int f(z)  \phi'(z)  \diff z.
\]
Since $\phi'(z) = -z\phi(z)$ and $f(z)\phi(z)\to 0$ as $|z|\to\infty$ (decay condition), the boundary term vanishes and we have
\[
\E[f'(z)]  =  \int f(z)  z  \phi(z)  \diff z  =  \E[z f(z)].
\]
This completes the derivation and proves the one-dimensional case of Stein's identity.

\subparagraph{Multivariate, Standard Normal Case.}
If $\rvz \sim \mathcal{N}(\mathbf{0},\rmI_D)$ and $g:\R^D \to \R$, then Stein's identity is
\[
\E[\nabla g(\rvz)]  =  \E[\rvz g(\rvz)].
\]
Equivalently, for $\mathbf{u}:\R^D \to \R^D$,
\begin{align}\label{eq:stein-multi}
    \E[ \nabla_{\rvz} \cdot\mathbf{u}(\rvz)]  =  \E[\rvz^\top \mathbf{u}(\rvz)] .
\end{align}

\subparagraph{Identity Needed for SURE.}
With $\tilde{\rvx}=\rvx+\sigma\rvz$, where $\rvz \sim \mathcal{N}(\mathbf{0},\rmI_D)$, and any vector function $\mathbf{a}$ of suitable regularity,
\begin{align}\label{eq:stein-for-sure}
    \E\big[(\tilde{\rvx}-\rvx)^\top \mathbf{a}(\tilde{\rvx})  \big|  \rvx\big]
 =  \sigma^2 \E\big[ \nabla_{\tilde{\rvx}} \cdot\mathbf{a}(\tilde{\rvx})  \big|  \rvx\big]. 
\end{align}
This is obtained by applying \Cref{eq:stein-multi} and using the chain rule.

\paragraph{Deriving SURE from the Conditional MSE.}
Let $\rmD:\R^D \to \R^D$ be a denoiser and define
\[
R(\rmD;\rvx):=\E \left[\|\rmD(\tilde{\rvx})-\rvx\|_2^2| \rvx\right].
\]
Expand around $\tilde{\rvx}$:
\[
\begin{aligned}
&~R(\rmD;\rvx) \\
=&~\E \left[\|\rmD(\tilde{\rvx})-\tilde{\rvx}\|^2  \big|  \rvx\right]
+2 \E \left[(\rmD(\tilde{\rvx})-\tilde{\rvx})^\top(\tilde{\rvx}-\rvx)  \big|  \rvx\right]
+\E \left[\|\tilde{\rvx}-\rvx\|^2  \big|  \rvx\right] \\
=&~\E \left[\|\rmD(\tilde{\rvx})-\tilde{\rvx}\|^2  \big|  \rvx\right]
+2\Big(\underbrace{\E[(\tilde{\rvx}-\rvx)^\top \rmD(\tilde{\rvx})|\rvx]}_{\substack{\sigma^2 \E[ \nabla_{\tilde{\rvx}} \cdot\rmD(\tilde{\rvx})|\rvx] \\  \text{ by \Cref{eq:stein-for-sure}}
}}
-\underbrace{\E[(\tilde{\rvx}-\rvx)^\top \tilde{\rvx}|\rvx]}_{ \sigma^2 D}\Big)
\\&\qquad\qquad\qquad+\underbrace{\E[\|\tilde{\rvx}-\rvx\|^2|\rvx]}_{ \sigma^2 D} \\
=&~\E \left[\|\rmD(\tilde{\rvx})-\tilde{\rvx}\|^2
+2\sigma^2  \nabla_{\tilde{\rvx}} \cdot\rmD(\tilde{\rvx})
-D\sigma^2 | \rvx\right].
\end{aligned}
\]
Therefore the \emph{observable} surrogate
\[ 
\mathrm{SURE}(\rmD;\tilde{\rvx})
:= \|\rmD(\tilde{\rvx})-\tilde{\rvx}\|_2^2
+2\sigma^2  \nabla_{\tilde{\rvx}} \cdot\rmD(\tilde{\rvx})
-D\sigma^2 
\]
satisfies $\E \left[\mathrm{SURE}(\rmD;\tilde{\rvx}) \big| \rvx\right]=R(\rmD;\rvx)$.
Minimizing SURE (in expectation or empirically) is thus equivalent to minimizing the true conditional MSE while using only noisy observations.

\subsection{Theorem~\ref{thm:fpe}: Marginal Alignment via Fokker–Planck}\label{app-sec:scoresde-fpe-proof}

\paragraph{\textit{Proof.}}
\paragraph{Part 1: Fokker-Planck Equation for the Forward SDE.}

Consider the forward SDE:
\[
\diff \rvx(t) = \rvf(\rvx(t), t)  \diff t + g(t)  \diff \rvw(t).
\]
The diffusion matrix is $\sigma(t) = g(t) I_D$, so $\sigma(t) \sigma(t)^T = g^2(t) I_D$. The Fokker-Planck equation for the marginal density $p_t(\rvx)$ of $\rvx(t)$ is:
\[
\partial_t p_t(\rvx) = -\nabla_{\rvx} \cdot \bigl[ \rvf(\rvx, t) p_t(\rvx) \bigr] + \frac{1}{2} \sum_{i,j=1}^D \frac{\partial^2}{\partial x_i \partial x_j} \bigl[ (g^2(t) \delta_{ij}) p_t(\rvx) \bigr].
\]
Compute the diffusion term:
\[
\sum_{i,j=1}^D \frac{\partial^2}{\partial x_i \partial x_j} \bigl[ g^2(t) \delta_{ij} p_t(\rvx) \bigr] = \sum_{i=1}^D \frac{\partial^2}{\partial x_i^2} \bigl[ g^2(t) p_t(\rvx) \bigr] = g^2(t) \Delta_{\rvx} p_t(\rvx).
\]
Thus:
\[
\partial_t p_t(\rvx) = -\nabla_{\rvx} \cdot \bigl[ \rvf(\rvx, t) p_t(\rvx) \bigr] + \frac{1}{2} g^2(t) \Delta_{\rvx} p_t(\rvx).
\]
Now, rewrite using:
\[
\rvv(\rvx, t) = \rvf(\rvx, t) - \frac{1}{2} g^2(t) \nabla_{\rvx} \log p_t(\rvx).
\]
Since $\nabla_{\rvx} \log p_t(\rvx) = \frac{\nabla_{\rvx} p_t(\rvx)}{p_t(\rvx)}$, compute:
\[
\nabla_{\rvx} \cdot \bigl[ \rvv(\rvx, t) p_t(\rvx) \bigr] = \nabla_{\rvx} \cdot \biggl[ \rvf(\rvx, t) p_t(\rvx) - \frac{1}{2} g^2(t) \frac{\nabla_{\rvx} p_t(\rvx)}{p_t(\rvx)} p_t(\rvx) \biggr].
\]
The second term is:
\[
\nabla_{\rvx} \cdot \biggl[ -\frac{1}{2} g^2(t) \nabla_{\rvx} p_t(\rvx) \biggr] = -\frac{1}{2} g^2(t) \Delta_{\rvx} p_t(\rvx).
\]
Thus:
\[
\nabla_{\rvx} \cdot \bigl[ \rvv(\rvx, t) p_t(\rvx) \bigr] = \nabla_{\rvx} \cdot \bigl[ \rvf(\rvx, t) p_t(\rvx) \bigr] - \frac{1}{2} g^2(t) \Delta_{\rvx} p_t(\rvx).
\]
Therefore:
\[
\partial_t p_t(\rvx) = -\nabla_{\rvx} \cdot \bigl[ \rvv(\rvx, t) p_t(\rvx) \bigr],
\]
verifying the Fokker-Planck equation in both forms.

\paragraph{Part 2: PF-ODE Marginal Densities.}

Consider the PF-ODE:
\[
\frac{\diff \tilde\rvx(t)}{\diff t} = \rvv(\tilde\rvx(t), t), \quad \rvv(\rvx, t) = \rvf(\rvx, t) - \frac{1}{2} g^2(t) \nabla_{\rvx} \log p_t(\rvx).
\]

\textbfs{Forward Direction: $\tilde\rvx(0) \sim p_0$. } Let $\tilde\rvx(t)$ follow the PF-ODE with $\tilde\rvx(0) \sim p_0$. The flow map $\bm{\Psi}_t: \mathbb{R}^D \to \mathbb{R}^D$ is defined by:
\[
\frac{\diff}{\diff t} \bm{\Psi}_t(\mathbf{x}_0) = \rvv(\bm{\Psi}_t(\mathbf{x}_0), t), \quad \bm{\Psi}_0(\mathbf{x}_0) = \mathbf{x}_0.
\]
Since $\tilde\rvx(t) = \bm{\Psi}_t(\tilde\rvx(0))$, the density $\tilde p_t(\rvx)$ of $\tilde\rvx(t)$ satisfies the continuity equation:
\[
\partial_t \tilde p_t(\rvx) = -\nabla_{\rvx} \cdot \bigl[ \rvv(\rvx, t) \tilde p_t(\rvx) \bigr].
\]
Since $\tilde\rvx(0) \sim p_0$, we have $\tilde p_0(\rvx) = p_0(\rvx)$. From Part 1, $p_t(\rvx)$ satisfies:
\[
\partial_t p_t(\rvx) = -\nabla_{\rvx} \cdot \bigl[ \rvv(\rvx, t) p_t(\rvx) \bigr].
\]
Both $\tilde p_t$ and $p_t$ satisfy the same continuity equation with the same initial condition $p_0$. Assuming sufficient smoothness (e.g., $\rvv \in C^1$), the solution is unique in some appropriate function space, so $\tilde p_t = p_t$. Thus, $\tilde\rvx(t) \sim p_t$ for all $t \in [0, T]$.

\vspace{0.3cm}
\textbfs{Backward Direction: $\tilde\rvx(T) \sim p_T$. } Now, let $\tilde\rvx(t)$ follow the PF-ODE backward from $t = T$ to $t = 0$, with $\tilde\rvx(T) \sim p_T$. The ODE is:
\[
\frac{\diff}{\diff t} \tilde\rvx(t) = \rvv(\tilde\rvx(t), t).
\]
Let $s = T - t$, so the backward ODE becomes:
\[
\frac{\diff}{\diff s} \tilde\rvx(T - s) = -\rvv(\tilde\rvx(T - s), T - s).
\]
The density $\tilde p_{T-s}(\rvx)$ of $\tilde\rvx(T - s)$ satisfies:
\[
\partial_s \tilde p_{T-s}(\rvx) = \nabla_{\rvx} \cdot \bigl[ \rvv(\rvx, T - s) \tilde p_{T-s}(\rvx) \bigr].
\]
Since $\tilde\rvx(T) \sim p_T$, we have $\tilde p_T = p_T$. The Fokker-Planck equation for $p_t$ at $t = T - s$ is:
\[
\partial_t p_{T-s}(\rvx) = -\nabla_{\rvx} \cdot \bigl[ \rvv(\rvx, T - s) p_{T-s}(\rvx) \bigr].
\]
Since $\partial_t = -\partial_s$, we get:
\[
\partial_s p_{T-s}(\rvx) = \nabla_{\rvx} \cdot \bigl[ \rvv(\rvx, T - s) p_{T-s}(\rvx) \bigr].
\]
Both $\tilde p_{T-s}$ and $p_{T-s}$ satisfy the same PDE with the same initial condition at $s = 0$ ($\tilde p_T = p_T$). Uniqueness implies $\tilde p_{T-s} = p_{T-s}$, so $\tilde\rvx(t) = \tilde\rvx(T - s) \sim p_{T-s} = p_t$, for all $t \in [0, T]$.

\paragraph{Part 3: Reverse-Time SDE Marginal Densities.}
We verify the marginal distributions using the forward-running reverse clock
\[
s:=T-t.
\]
Define
\[
\rvy(s):=\bar\rvx(T-s),
\qquad s\in[0,T].
\]
Since the reverse process starts from $\bar\rvx(T)\sim p_T$, we have
\[
\rvy(0)\sim p_T.
\]

Under this change of clock, the reverse-time SDE becomes
\[
\diff\rvy(s)
=
\Bigl[
-\rvf(\rvy(s),T-s)
+
g^2(T-s)\nabla_{\rvx}\log p_{T-s}(\rvy(s))
\Bigr]\diff s
+
g(T-s)\diff\rvw^{\mathrm{rev}}(s),
\]
where $\rvw^{\mathrm{rev}}$ is a standard Wiener process under the
forward-running clock $s$.

Let $q_s$ denote the density of $\rvy(s)$. Its Fokker--Planck equation is
\begin{align*}
\partial_s q_s(\rvx)
&=
-\nabla_{\rvx}\cdot
\Bigl(
\bigl[
-\rvf(\rvx,T-s)
+
g^2(T-s)\nabla_{\rvx}\log p_{T-s}(\rvx)
\bigr]
q_s(\rvx)
\Bigr)
\\
&\qquad
+
\tfrac12 g^2(T-s)\Delta_{\rvx}q_s(\rvx).
\end{align*}

We now check that
\[
q_s(\rvx)=p_{T-s}(\rvx)
\]
satisfies this equation. Using
\[
\nabla_{\rvx}\log p_{T-s}(\rvx)\,p_{T-s}(\rvx)
=
\nabla_{\rvx}p_{T-s}(\rvx),
\]
the right-hand side becomes
\begin{align*}
&\nabla_{\rvx}\cdot
\bigl[
\rvf(\rvx,T-s)p_{T-s}(\rvx)
\bigr]
-
g^2(T-s)\Delta_{\rvx}p_{T-s}(\rvx)
\\
&\qquad
+
\tfrac12 g^2(T-s)\Delta_{\rvx}p_{T-s}(\rvx)
\\
&=
\nabla_{\rvx}\cdot
\bigl[
\rvf(\rvx,T-s)p_{T-s}(\rvx)
\bigr]
-
\tfrac12 g^2(T-s)\Delta_{\rvx}p_{T-s}(\rvx).
\end{align*}

On the other hand, the forward Fokker--Planck equation gives
\[
\partial_t p_t(\rvx)
=
-\nabla_{\rvx}\cdot
\bigl[\rvf(\rvx,t)p_t(\rvx)\bigr]
+
\tfrac12g^2(t)\Delta_{\rvx}p_t(\rvx).
\]
Since $t=T-s$, the chain rule gives
\begin{align*}
\partial_s p_{T-s}(\rvx)
&=
-\partial_t p_t(\rvx)\big|_{t=T-s}
\\
&=
\nabla_{\rvx}\cdot
\bigl[
\rvf(\rvx,T-s)p_{T-s}(\rvx)
\bigr]
-
\tfrac12g^2(T-s)\Delta_{\rvx}p_{T-s}(\rvx).
\end{align*}
This is exactly the Fokker--Planck equation of the reversed process.

Finally, $q_0=p_T$, which matches the initial distribution
$\rvy(0)\sim p_T$. Under the regularity assumptions ensuring uniqueness of the
Fokker--Planck solution,
\[
q_s=p_{T-s},
\qquad s\in[0,T].
\]
Returning to the original diffusion clock $t=T-s$, we obtain
\[
\bar\rvx(t)\sim p_t.
\]
Thus, when the reverse-time SDE is solved backward from $t=T$ to $t=0$, it
follows the same marginal distributions as the forward SDE, but in the reverse
order.
\hfill$\blacksquare$

\subsection{Proposition~\ref{dsm-minimizer}: Minimizer of SM and DSM}\label{app-sec:min-dsm}

\paragraph{\textit{Proof.}} To find the minimizer $\mathbf{s}^*$, we first consider a fixed time $t$ and analyze the inner expectation in the objective function:
\begin{align*}
\mathcal  J(t, \bm{\phi}) 
&:= \mathbb{E}_{\rvx_0 \sim p_{\text{data}}} 
\mathbb{E}_{\rvx_t \sim p_t(\cdot|\rvx_0)} 
\left[\left\| \mathbf{s}_{\bm{\phi}}(\rvx_t, t) - \nabla_{\rvx_t} \log p_t(\rvx_t|\rvx_0) \right\|_2^2 \right].
\end{align*}
For this expectation to be minimized, we need to find 
$\mathbf{s}_{\bm{\phi}}(\rvx_t, t)$ 
that minimizes the expected squared error for each $\rvx_t$.
We can rewrite this expectation using the joint distribution of $X_0$ and $X_t$:
\begin{align*}
\mathcal J(t, \bm{\phi}) 
= \iint p_{\text{data}}(\rvx_0) p_t(\rvx_t |\rvx_0) 
\left\| \mathbf{s}_{\bm{\phi}}(\rvx_t, t) - \nabla_{\rvx_t} \log p(\rvx_t |  \rvx_0) \right\|_2^2 
 \diff\rvx_0  \diff\rvx_t.
\end{align*}
For each fixed $\rvx_t$, we need to minimize:
\begin{align*}
\int p(\rvx_0 | X_t = \rvx_t) p_t( \rvx_t) 
\left\| \mathbf{s}_{\bm{\phi}}(\rvx_t, t) - \nabla_{\rvx_t} \log p(\rvx_t |  \rvx_0) \right\|_2^2 
 \diff\rvx_0.
\end{align*}
Since $p_t(\rvx_t)$ is constant with respect to 
$\mathbf{s}_{\bm{\phi}}(\rvx_t, t)$, this is equivalent to minimizing:
\begin{align*}
\int p( \rvx_0 | X_t = \rvx_t) 
\left\| \mathbf{s}_{\bm{\phi}}(\rvx_t, t) - \nabla_{\rvx_t} \log p(\rvx_t |  \rvx_0) \right\|_2^2 
  \diff\rvx_0
\end{align*}
This is minimized when $\mathbf{s}_{\bm{\phi}}(\rvx_t, t)$ equals the conditional expectation:
\begin{align*}
\mathbf{s}^*(\rvx_t, t) 
= \mathbb{E}_{X_0 \sim p(X_0 | X_t = \rvx_t)} 
\left[ \nabla_{\rvx_t} \log p(\rvx_t | X_0) \right].
\end{align*}
Now we need to prove that this equals 
$\nabla_{\rvx_t} \log p_t(\rvx_t)$. By Bayes' rule and the definition of marginal probability:
\begin{align*}
p_t(\rvx_t) 
= \int p_t(\rvx_t | \rvx_0) 
p_{\text{data}}(\rvx_0)   \diff\rvx_0.
\end{align*}
Taking the logarithm and then the gradient with respect to $\rvx_t$:
\begin{align*}
\nabla_{\rvx_t} \log p_t(\rvx_t) 
= \frac{\nabla_{\rvx_t} p_t(\rvx_t)}{p_t(\rvx_t)}
= \frac{\nabla_{\rvx_t} \int p_t(\rvx_t | \rvx_0) 
p_{\text{data}}(\rvx_0)   \diff\rvx_0}
{\int p_t(\rvx_t | \rvx_0) 
p_{\text{data}}(\rvx_0)   \diff\rvx_0}.
\end{align*}
Under suitable regularity conditions, we can exchange the gradient and integral:
\begin{align*}
\nabla_{\rvx_t} \log p_t(\rvx_t) 
= \frac{\int \nabla_{\rvx_t} p_t(\rvx_t | \rvx_0) 
p_{\text{data}}(\rvx_0)   \diff\rvx_0}
{\int p_t(\rvx_t | \rvx_0) 
p_{\text{data}}(\rvx_0)   \diff\rvx_0}.
\end{align*}
Using
\[
\nabla_{\rvx_t}p_t(\rvx_t|\rvx_0)
=
p_t(\rvx_t|\rvx_0)
\nabla_{\rvx_t}\log p_t(\rvx_t|\rvx_0),
\]
and Bayes' rule, we obtain
\begin{align*}
\nabla_{\rvx_t}\log p_t(\rvx_t)
&=
\int
p(\rvx_0|\rvx_t)
\nabla_{\rvx_t}\log p_t(\rvx_t|\rvx_0)
\diff\rvx_0 \\
&=
\mathbb{E}_{p(\rvx_0|\rvx_t)}
\left[
\nabla_{\rvx_t}\log p_t(\rvx_t|\rvx_0)
\right]
=
\mathbf{s}^*(\rvx_t,t),
\end{align*}
which completes the proof.
\hfill$\blacksquare$

\subsection{Closed-Form Score Function of a Gaussian}

For future reference, we summarize the formula for the score of a general multivariate normal distribution as the following lemma:

\lemp{Score of Gaussian}{score-general-gaussian}{Let $\rvx \in \mathbb{R}^D$ and consider the multivariate normal distribution
\[
p(\tilde{\rvx} | \rvx) := \mathcal{N}(\tilde{\rvx}; \bm{\mu}, \bm{\Sigma}),
\]
where $\bm{\mu} \in \mathbb{R}^D$ is the mean and $\bm{\Sigma} \in \mathbb{R}^{D \times D}$ is an invertible covariance matrix. Its score function is
\begin{align}\label{eq:score-gaussian-formula}
    \nabla_{\tilde{\rvx}} \log p(\tilde{\rvx} | \rvx) = -\bm{\Sigma}^{-1} (\tilde{\rvx} - \bm{\mu}).
\end{align}
}{The density function of $p(\tilde{\rvx} | \rvx)$ is given by:
\[
p(\tilde{\rvx} | \rvx) = \frac{1}{(2\pi)^{D/2} |\bm{\Sigma}|^{1/2}} \exp\left( -\frac{1}{2} (\tilde{\rvx} - \bm{\mu})^\top \bm{\Sigma}^{-1} (\tilde{\rvx} - \bm{\mu}) \right).
\]
To compute the score function, we first take the log of $ p(\tilde{\rvx} | \rvx) $:
\[
\log p(\tilde{\rvx} | \rvx) = -\frac{D}{2} \log(2\pi) - \frac{1}{2} \log |\bm{\Sigma}| - \frac{1}{2} (\tilde{\rvx} - \bm{\mu})^\top \bm{\Sigma}^{-1} (\tilde{\rvx} - \bm{\mu}).
\]
Now, we compute the gradient of $ \log p(\tilde{\rvx} | \rvx) $ with respect to $ \tilde{\rvx} $:
\[
\nabla_{\tilde{\rvx}} \log p(\tilde{\rvx} | \rvx) = - \frac{1}{2} \nabla_{\tilde{\rvx}} \left( (\tilde{\rvx} - \bm{\mu})^\top \bm{\Sigma}^{-1} (\tilde{\rvx} - \bm{\mu}) \right).
\]
Using the chain rule, we get:
\[
\nabla_{\tilde{\rvx}} \left( (\tilde{\rvx} - \bm{\mu})^\top \bm{\Sigma}^{-1} (\tilde{\rvx} - \bm{\mu}) \right) = 2 \bm{\Sigma}^{-1} (\tilde{\rvx} - \bm{\mu}).
\]
Thus, the score function is:
\begin{align}\label{eq:gaussian-score}
    \nabla_{\tilde{\rvx}} \log p(\tilde{\rvx} | \rvx) = - \bm{\Sigma}^{-1} (\tilde{\rvx} - \bm{\mu}).
\end{align}
}

\clearpage
\newpage

\section{Flow-Based Perspective}\label{app-sec:flow-proof}

\subsection{Lemma~\ref{instant-change-of-var}: Instantaneous Change of Variables}
\paragraph{\textit{Proof.}}

\textbfs{Approach 1: Change-of-Variables Formula. } We denote $p(\rvx(t), t)$ by $p_t(\rvx_t)$. Starting from the ODE discretization
\[
\rvz_{t+\Delta t} = \rvz_t + \Delta t  \rmF(\rvz_t, t),
\]
the change-of-variables formula for normalizing flows (\Cref{eq:nf-change-of-var-log}) gives
\begin{align*}
    \log p_{t+\Delta t}(\rvz_{t+\Delta t}) 
    &= \log p_t(\rvz_t) - \log \Big| \det\big(\rmI + \Delta t \nabla_{\rvz} \rmF(\rvz_t, t)\big) \Big| \\
    &= \log p_t(\rvz_t) - \Tr\Big(\log(\rmI + \Delta t \nabla_{\rvz} \rmF(\rvz_t, t)) \Big) \\
    &= \log p_t(\rvz_t) - \Delta t  \Tr\big(\nabla_{\rvz} \rmF(\rvz_t, t)\big) + \mathcal{O}(\Delta t^2),
\end{align*}
where we used $\log \det = \Tr \log$ and the expansion for small $\Delta t$.
Taking the limit $\Delta t \to 0$ yields the continuous-time differential equation for the log-density. Indeed, the same trick is applied in \Cref{eq:log-det-tr}.

\vspace{0.3cm}
\textbfs{Approach 2: Continuity Equation. } We can also leverage the continuity equation, which essentially serves as the change-of-variables formula:
\[
\partial_t p(\rvz,t) = - \nabla_{\rvz} \cdot \big(\rmF(\rvz,t) p(\rvz,t)\big).
\]
Expanding the divergence,
\[
\partial_t p = - \big( (\nabla_{\rvz} \cdot \rmF) p + \rmF \cdot \nabla_{\rvz} p \big).
\]
Along trajectories $\rvz(t)$ satisfying $\frac{\diff \rvz}{\diff t} = \rmF(\rvz(t), t)$, the total time derivative is
\begin{align*}
\frac{\diff}{\diff t} p(\rvz(t), t) 
&= \nabla_{\rvz} p \cdot \frac{\diff \rvz}{\diff t} + \partial_t p \\
&= \nabla_{\rvz} p \cdot \rmF - \big( (\nabla_{\rvz} \cdot \rmF) p + \rmF \cdot \nabla_{\rvz} p \big) \\
&= - (\nabla_{\rvz} \cdot \rmF) p.
\end{align*}
Dividing by $p(\rvz(t), t)$, we conclude
\[
\frac{\diff}{\diff t} \log p(\rvz(t), t) = - \nabla_{\rvz} \cdot \rmF(\rvz(t), t).
\]
\hfill$\blacksquare$

\subsection{Theorem~\ref{thm:continuity-mass}: Mass Conservation Criterion}\label{app-sec:mass-conservation}

\paragraph{Some Prerequisites: Flow Map and Flow-Induced Density.} For any initial position $\mathbf{x}_0\in\mathbb{R}^D$, the flow map 
$\bm{\Psi}_t:\mathbb{R}^D\to\mathbb{R}^D$ is the unique solution of the ODE
\[
\frac{\diff}{\diff t} \bm{\Psi}_t(\mathbf{x}_0)
= \mathbf{v}_t\bigl(\bm{\Psi}_t(\mathbf{x}_0)\bigr), 
\quad \bm{\Psi}_0(\mathbf{x}_0)=\mathbf{x}_0.
\]
Under our regularity assumptions, $\bm{\Psi}_t$ is continuously differentiable in both $t$ and $\mathbf{x}_0$.

The flow‑induced density $p_t^{\mathrm{fwd}}$ is the pushforward of the initial density $p_0$ by $\bm{\Psi}_t$:
\[
p_t^{\mathrm{fwd}}(\mathbf{x})
= p_0\bigl(\bm{\Psi}_t^{-1}(\mathbf{x})\bigr)
  \Bigl|\det \bigl(\nabla\bm{\Psi}_t^{-1}(\mathbf{x})\bigr)\Bigr|.
\]
This gives the density at $\mathbf{x}$ and time $t$ of particles that started from $p_0=p_{\mathrm{data}}$ and evolved under the velocity field $\mathbf{v}_t$.

\paragraph{\textit{Informal Proof:}} \textbfs{Sufficient Condition: $p_t^{\mathrm{fwd}} = p_t$ $\Rightarrow$ Continuity Equation. }
In \Cref{subsec:cov-continuity-eq} we obtained a \emph{strong solution} of the continuity equation by taking the continuous time limit of the discrete change of variable formula. In that approach, the density $p_t$ is assumed smooth enough that all derivatives exist classically and the PDE holds pointwise. Here, we offer a complementary derivation in the \emph{weak sense}: the continuity equation is imposed only after integrating against arbitrary smooth test functions, which relaxes the regularity requirements on both $p_t$ and the velocity field $\mathbf{v}_t$. This weak formulation is not only more rigorous since it accommodates less regular solutions but is also the standard framework in PDE theory and numerical analysis.

For any smooth test function $ \varphi(\mathbf{x}) $ with compact support, the pushforward property gives:
\begin{align*}
    \int p_t^{\mathrm{fwd}}(\mathbf{x}) \varphi(\mathbf{x})  \diff\mathbf{x} &= \int p_0(\bm{\Psi}_t^{-1}(\mathbf{x})) \left| \det\left( \nabla \bm{\Psi}_t^{-1}(\mathbf{x}) \right) \right| \varphi(\mathbf{x}) \diff\mathbf{x}\\ &=  \int p_0(\mathbf{y}) \varphi(\bm{\Psi}_t(\mathbf{y})) \diff\mathbf{y},
\end{align*}
where the second equality follows from the change of variables $ \mathbf{x} = \bm{\Psi}_t(\mathbf{y}) $, with $ \diff\mathbf{y} = \left| \det\left( \nabla \bm{\Psi}_t^{-1}(\mathbf{x}) \right) \right| \diff\mathbf{x} $.

Differentiate both sides with respect to $ t $:
\[
\frac{\diff}{\diff t} \int p_t^{\mathrm{fwd}}(\mathbf{x}) \varphi(\mathbf{x}) \diff\mathbf{x} = \frac{\diff}{\diff t} \int p_0(\mathbf{y}) \varphi(\bm{\Psi}_t(\mathbf{y})) \diff\mathbf{y}.
\]

The left-hand side is:
\[
\int \frac{\partial p_t^{\mathrm{fwd}}}{\partial t}(\mathbf{x}) \varphi(\mathbf{x}) \diff\mathbf{x}.
\]

On the right-hand side:
\[
\int p_0(\mathbf{y}) \nabla \varphi(\bm{\Psi}_t(\mathbf{y})) \cdot \mathbf{v}_t(\bm{\Psi}_t(\mathbf{y})) \diff\mathbf{y},
\]
since $ \frac{\partial \bm{\Psi}_t}{\partial t}(\mathbf{y}) = \mathbf{v}_t(\bm{\Psi}_t(\mathbf{y})) $.
Change variables to $ \mathbf{x} = \bm{\Psi}_t(\mathbf{y}) $, so 
\[
\diff\mathbf{y} = \left| \det\left( \nabla \bm{\Psi}_t^{-1}(\mathbf{x}) \right) \right| \diff\mathbf{x} ,
\]
 and
\[
p_0(\mathbf{y}) = p_t^{\mathrm{fwd}}(\mathbf{x}) \left| \det\left( \nabla \bm{\Psi}_t(\mathbf{y}) \right) \right| = \frac{p_t^{\mathrm{fwd}}(\mathbf{x})}{\left| \det\left( \nabla \bm{\Psi}_t^{-1}(\mathbf{x}) \right) \right|}.
\]
Thus, the right-hand side becomes:
\[
\int p_t^{\mathrm{fwd}}(\mathbf{x}) \nabla \varphi(\mathbf{x}) \cdot \mathbf{v}_t(\mathbf{x}) \diff\mathbf{x}.
\]

Apply integration by parts, using the compact support of $ \varphi $:
\[
\int p_t^{\mathrm{fwd}}(\mathbf{x}) \nabla \varphi(\mathbf{x}) \cdot \mathbf{v}_t(\mathbf{x}) \diff\mathbf{x} = -\int \varphi(\mathbf{x}) \nabla \cdot (p_t^{\mathrm{fwd}}(\mathbf{x}) \mathbf{v}_t(\mathbf{x})) \diff\mathbf{x}.
\]
Equating both sides:
\[
\int \left[ \frac{\partial p_t^{\mathrm{fwd}}}{\partial t}(\mathbf{x}) + \nabla \cdot (p_t^{\mathrm{fwd}}(\mathbf{x}) \mathbf{v}_t(\mathbf{x})) \right] \varphi(\mathbf{x}) \diff\mathbf{x} = 0.
\]
Since $ \varphi $ is arbitrary, we conclude:
\[
\frac{\partial p_t^{\mathrm{fwd}}}{\partial t} + \nabla \cdot (p_t^{\mathrm{fwd}} \mathbf{v}_t) = 0.
\]
Given $ p_t^{\mathrm{fwd}} = p_t $, this implies:
\[
\frac{\partial p_t}{\partial t} + \nabla \cdot (p_t \mathbf{v}_t) = 0.
\]

\textbfs{Necessary Condition: Continuity Equation $ \Rightarrow p_t^{\mathrm{fwd}} = p_t $.} Suppose $ p_t $ satisfies the continuity equation:
\[
\frac{\partial p_t}{\partial t} + \nabla \cdot (p_t \mathbf{v}_t) = 0,
\]
with the initial condition $ p_0(\mathbf{x}) = p_{\mathrm{data}}(\mathbf{x}) $. We know $ p_t^{\mathrm{fwd}} $ satisfies the same continuity equation, as shown above, and:
\[
p_0^{\mathrm{fwd}}(\mathbf{x}) = p_0(\bm{\Psi}_0^{-1}(\mathbf{x})) \left| \det\left( \nabla \bm{\Psi}_0^{-1}(\mathbf{x}) \right) \right| = p_0(\mathbf{x}),
\]
since $ \bm{\Psi}_0(\mathbf{x}) = \mathbf{x} $. Thus, both densities share the same initial condition $ p_0 = p_{\mathrm{data}} $.

The continuity equation can be rewritten as:
\[
\frac{\partial p}{\partial t} + \mathbf{v}_t \cdot \nabla p + p \nabla \cdot \mathbf{v}_t = 0.
\]
This is a first-order linear PDE. Assuming $ \mathbf{v}_t $ is continuously differentiable and globally Lipschitz, and $ p_t $ is sufficiently smooth, the method of characteristics guarantees a unique solution in the space of smooth functions. Since $ p_t $ and $ p_t^{\mathrm{fwd}} $ satisfy the same PDE and initial condition, we conclude:
\[
p_t(\mathbf{x}) = p_t^{\mathrm{fwd}}(\mathbf{x})
\]
for all $t\in[0,1]$ and $\mathbf{x} \in \mathbb{R}^D$. 

This completes the proof of the equivalence.
\hfill$\blacksquare$

\clearpage
\newpage

\section{Theoretical Supplement: A Unified and Systematic View on Diffusion Models}\label{app-sec:equiv-para}

\subsection{Proposition~\ref{equi-para}: Equivalence of Parametrizations}

\paragraph{\textit{Proof:}} 
As in Theorem~\ref{dsm-minimizer} on the DSM loss, the global optimum of the matching loss
\[
\mathbb{E}_t\left[\omega(t)\,\mathbb{E}_{\rvx_0,\bm{\epsilon}}\left[\|\cdot\|_2^2\right]\right]
\]
is attained when the inner expectation
\[
\mathbb{E}_{\rvx_0,\bm{\epsilon}}\left[\|\cdot\|_2^2\right]
\]
is minimized for each fixed $t$. Since this is a standard mean squared error problem, the minimizer is unique.
From denoising score matching~\citep{vincent2011connection}, Theorem~\ref{dsm-minimizer} shows the optimal score function satisfies
\begin{align*}
    \rvs^*(\rvx_t,t) = \mathbb{E}_{p(\rvx_0|\rvx_t)}\left[\nabla_{\rvx}\log p_t(\rvx_t|\rvx_0)\right] = \nabla_{\rvx_t} \log p_t(\rvx_t).
\end{align*}
Using the identity $\nabla_{\rvx}\log p_t(\rvx_t|\rvx_0) = -\frac{1}{\sigma_t} \bm{\epsilon}$ for $\bm{\epsilon} \sim \mathcal{N}(\bm{0},\rmI)$, we obtain
\begin{align*}
    \rvs^*(\rvx_t,t) = -\frac{1}{\sigma_t}\, \bm{\epsilon}^*(\rvx_t,t),
\end{align*}
where $\bm{\epsilon}^*(\rvx_t,t) = \mathbb{E}_{p(\rvx_0|\rvx_t)}[\bm{\epsilon}|\rvx_t]$ is the optimal $\beps$-prediction for $\mathcal{L}_{\text{noise}}({\bm{\phi}})$.
For the $\rvx$-prediction loss $\mathcal{L}_{\text{clean}}$, the optimal estimator is
\[
\rvx^*(\rvx_t,t) = \mathbb{E}_{p(\rvx_0|\rvx_t)}[\rvx_0|\rvx_t],
\]
which, by Tweedie's formula, relates to the score via
\[
\alpha_t\, \rvx^*(\rvx_t,t) = \rvx_t + \sigma_t^2\, \rvs^*(\rvx_t,t).
\]
For the $\rvv$-prediction loss $\mathcal{L}_{\text{velocity}}$, the optimal estimator is
\begin{align*}
    \rvv^*(\rvx_t,t) = \mathbb{E}_{p(\rvx_0|\rvx_t)}[\alpha_t' \rvx_0 + \sigma_t' \bm{\epsilon}|\rvx_t] 
    = \alpha_t' \rvx^* + \sigma_t' \bm{\epsilon}^*.
\end{align*}
Substituting the expressions for $\rvx^*$ and $\bm{\epsilon}^*$ in terms of $\rvs^*$ yields
\begin{align*}
    \rvv^*(\rvx_t,t)
    = \frac{\alpha_t'}{\alpha_t} \rvx_t + \left( \frac{\alpha_t'}{\alpha_t} \sigma_t^2 - \sigma_t \sigma_t' \right) \rvs^*(\rvx_t,t) = f(t) \rvx_t - \frac{1}{2}g^2(t) \rvs^*(\rvx_t,t),
\end{align*}
which completes the derivation.

\hfill$\blacksquare$

\clearpage
\newpage

\section{Theoretical Supplement: Learning Fast Diffusion-Based Generators}\label{app-sec:flow-map}

\subsection{Knowledge Distillation Loss as an Instance of the General Framework \Cref{eq:general-cm-loss}}

We begin with the oracle KD loss
\[
\mathcal{L}_{\mathrm{KD}}^{\text{oracle}}(\btheta)
=\E_{\rvx_T\sim p_T}\,
\big\|\rmG_\btheta(\rvx_T,T,0)-\bPsi_{T\to 0}(\rvx_T)\big\|_2^2,
\]
where $p_T = p_{\mathrm{prior}}$. For the deterministic ODE flow map $\bPsi$ (which satisfies the semigroup property and is bijective along trajectories), the marginals satisfy the pushforward identities
\[
p_t
= \bPsi_{0\to t}\,\sharp\,p_{\mathrm{data}}
= \bPsi_{T\to t}\,\sharp\,p_{\mathrm{prior}}.
\]
Hence,
\[
\bPsi_{s\to T}\,\sharp\,p_s=p_T
\quad\text{and}\quad
\bPsi_{T\to 0}\circ\bPsi_{s\to T}=\bPsi_{s\to 0}.
\]
Changing variables $\rvx_T=\bPsi_{s\to T}(\rvx_s)$ with $\rvx_s\sim p_s$ gives
\[
\mathcal{L}_{\mathrm{KD}}^{\text{oracle}}(\btheta)
= \E_{\rvx_s\sim p_s} 
\Big\|\rmG_\btheta \big(\bPsi_{s\to T}(\rvx_s),T,0\big)
- \bPsi_{s\to 0}(\rvx_s)\Big\|_2^2.
\]
Define the pulled-back student $\widetilde{\rmG}_\btheta(\rvx_s,s,0):=
\rmG_\btheta(\bPsi_{s\to T}(\rvx_s),T,0)$ to express the same loss in the unified flow-map form (at $t=0$):
\begin{align*}
\mathcal{L}_{\mathrm{KD}}^{\text{oracle}}(\btheta)
= \E_{\rvx_s\sim p_s} 
\big\|\widetilde{\rmG}_\btheta(\rvx_s,s,0)-\bPsi_{s\to 0}(\rvx_s)\big\|_2^2.
\end{align*}
This derivation relies on change of variables through the oracle flow and the semigroup property.

\hfill$\blacksquare$

\paragraph{Transport Interpretation.}
Proposition~\ref{vsd-grad} also admits a distributional interpretation.
For
\[
\hat\rvx_t
=
\alpha_t\rmG_\btheta(\rvz)+\sigma_t\beps,
\]
an infinitesimal parameter perturbation induces the data-space displacement
$(\partial_\btheta\hat\rvx_t)\,\diff\btheta$. Accordingly, the VSD gradient
can be written as
\[
\nabla_{\btheta}\mathcal{L}_{\mathrm{VSD}}
=
\E\!\left[
\omega(t)
\big(\partial_\btheta\hat\rvx_t\big)^\top
\left(
\nabla_{\rvx}\log p_t^{\btheta}(\hat\rvx_t)
-
\nabla_{\rvx}\log p_t(\hat\rvx_t)
\right)
\right].
\]
Thus, the score mismatch provides the data-space gradient signal, while the
generator Jacobian maps this signal back to parameter space; gradient descent
then updates the parameters in the opposite direction. This is the parametric counterpart of the
distribution-level transport view in
\Cref{sec:wgf-distribution-level-training}.

\subsection{Theorem~\ref{thm:ct-cm}: CM Equals CT up to $\mathcal{O}(\Delta s^2)$}

\paragraph{\textit{Proof:}} 

\textbfs{Step 1: DDIM Update (with Oracle Score) Is the Conditional Mean.}
\begin{align*}
\hat\rvx_{s'}
&:= \frac{\alpha_{s'}}{\alpha_s} \rvx_s 
+ \sigma_s^2 \left(\frac{\alpha_{s'}}{\alpha_s}-\frac{\sigma_{s'}}{\sigma_s}\right)\nabla_{\rvx_s}\log p_s(\rvx_s)
\\
&= \frac{\alpha_{s'}}{\alpha_s}\bigl(\rvx_s+\sigma_s^2\nabla_{\rvx_s}\log p_s(\rvx_s)\bigr)
- \sigma_{s'}\sigma_s \nabla_{\rvx_s}\log p_s(\rvx_s)
\\
&= \alpha_{s'} \E[\rvx_0|\rvx_s]
+ \sigma_{s'}\Bigl(\frac{\rvx_s-\alpha_s \E[\rvx_0|\rvx_s]}{\sigma_s}\Bigr)
\\
&= \alpha_{s'} \E[\rvx_0|\rvx_s] + \sigma_{s'} \E[\beps|\rvx_s]
\\
&= \E[\alpha_{s'}\rvx_0+\sigma_{s'}\beps|\rvx_s]
\\
&= \E[\rvx_{s'}|\rvx_s].
\end{align*}
Here we use Tweedie's formula
\[
\E[\rvx_0 |\rvx_s]=\frac{\rvx_s+\sigma_s^2\nabla_{\rvx_s}\log p_s(\rvx_s)}{\alpha_s},
\]
together with the shared-noise coupling
\[
\rvx_s=\alpha_s\rvx_0+\sigma_s\beps,
\qquad
\rvx_{s'}=\alpha_{s'}\rvx_0+\sigma_{s'}\beps
\quad (s'=s-\Delta s),
\]
in the third and fourth equalities. We also note that, more precisely, $\hat\rvx_{s'}=\E[\rvx_{s'}|\rvx_s]$ should be read as
$\hat\rvx_{s'}=\E[\rvx_{s'}|\rvx_s, s]$, since $s'=s-\Delta s$ is deterministically fixed once $s$ is sampled.

\textbfs{Step 2: Expand CT Around the Conditional Mean $\hat\rvx_{s'}$.}

Recall that $s$ is sampled first and we set $s':=s-\Delta s$ deterministically.
\begin{align*}
\mathcal{L}_{\mathrm{CT}}
&= \E_{s,\rvx_s}\E_{\rvx_{s'}|\rvx_s} \Big[w(s) d\big(\rvf_{\btheta}(\rvx_s,s), \rvf_{\btheta^-}(\rvx_{s'},s')\big)\Big]
\\
&\overset{\text{(1)}}{=} \E_{s,\rvx_s}\E_{\rvx_{s'}|\rvx_s} \Big[
w(s) d\big(\rvf_{\btheta}(\rvx_s,s), \rvf_{\btheta^-}(\hat\rvx_{s'},s')\big)
\\[-2pt]
&\qquad\qquad\qquad\qquad
+ w(s) \partial_2 d\big(\rvf_{\btheta}(\rvx_s,s),\rvf_{\btheta^-}(\hat\rvx_{s'},s')\big)\big[\partial_1\rvf_{\btheta^-}(\hat\rvx_{s'},s')(\rvx_{s'}-\hat\rvx_{s'})\big]
\\[-2pt]
&\qquad\qquad\qquad\qquad
+ w(s) \mathcal O \big(\|\rvx_{s'}-\hat\rvx_{s'}\|^2\big)\Big]
\\
&\overset{\text{(2)}}{=} \E_{s,\rvx_s} \Big[w(s) d\big(\rvf_{\btheta}(\rvx_s,s), \rvf_{\btheta^-}(\hat\rvx_{s'},s')\big)\Big]
\\
&\quad+ \E_{s,\rvx_s} \Big[w(s) \partial_2 d\big(\rvf_{\btheta}(\rvx_s,s),\rvf_{\btheta^-}(\hat\rvx_{s'},s')\big)\big[\partial_1\rvf_{\btheta^-}(\hat\rvx_{s'},s') \E_{\rvx_{s'}|\rvx_s}(\rvx_{s'}-\hat\rvx_{s'})\big]\Big]
\\
&\quad+ \E_{s,\rvx_s}\E_{\rvx_{s'}|\rvx_s} \Big[w(s) \mathcal{O} \big(\|\rvx_{s'}-\hat\rvx_{s'}\|^2\big)\Big]
\\
&\overset{\text{(3)}}{=} \E_{s,\rvx_s} \Big[w(s) d\big(\rvf_{\btheta}(\rvx_s,s), \rvf_{\btheta^-}(\hat\rvx_{s'},s')\big)\Big]
+ \E_{s,\rvx_s}\E_{\rvx_{s'}|\rvx_s} \Big[w(s) \mathcal{O} \big(\|\rvx_{s'}-\hat\rvx_{s'}\|^2\big)\Big]
\\
&= \mathcal{L}_{\mathrm{CM}} 
+ \mathcal{O} \big(\E\|\rvx_{s'}-\hat\rvx_{s'}\|^2\big)
\\
&\overset{\text{(4)}}{=} \mathcal{L}_{\mathrm{CM}} + \mathcal{O}(\Delta s^2).
\end{align*}
Here, (1) applies a second-order Taylor expansion of 
\[
h(\rvx') := d\big(\rvf_{\btheta}(\rvx_s,s), \rvf_{\btheta^-}(\rvx',s')\big)
\]
around $\hat\rvx_{s'}$ in its second argument. (2) uses the \textit{tower property} (a.k.a.\ the law of total expectation):
\[
\E_{s,\rvx_s}\E_{\rvx_{s'}|\rvx_s,s}[\cdot]
=\E_{s,\rvx_s}[\cdot],
\]
and notes that inside $\E_{\rvx_{s'}|\rvx_s,s}$, the terms
\[
w(s),\ \rvf_{\btheta}(\rvx_s,s),\ \hat\rvx_{s'},\ s'
\]
do not depend on the random variable $\rvx_{s'}$ (since $s' = s-\Delta s$ and $\hat\rvx_{s'}=\E[\rvx_{s'}|\rvx_s,s]$ are fixed once $(s,\rvx_s)$ is fixed). Hence the only $\rvx_{s'}$-dependence is through $(\rvx_{s'}-\hat\rvx_{s'})$, and the inner expectations reduce to
$\E_{\rvx_{s'}|\rvx_s,s}(\rvx_{s'}-\hat\rvx_{s'})$ and
$\E_{\rvx_{s'}|\rvx_s,s}\|\rvx_{s'}-\hat\rvx_{s'}\|^2$. (3) uses
\[
\E[\rvx_{s'}-\hat\rvx_{s'}|\rvx_s,s]=\bm 0
\quad\text{because}\quad
\hat\rvx_{s'}=\E[\rvx_{s'}|\rvx_s,s].
\]
For (4), note that
\[
\E\|\rvx_{s'}-\hat\rvx_{s'}\|^2
=\E\,\Var(\rvx_{s'}|\rvx_s,s)
\le \E\|\rvx_{s'}-\rvx_s\|^2.
\]
Moreover,
\[
\rvx_{s'}-\rvx_s=(\alpha_{s'}-\alpha_s)\rvx_0+(\sigma_{s'}-\sigma_s)\beps,
\]
and for a smooth scheduler, $\alpha_{s'}-\alpha_s=\mathcal{O}(\Delta s)$ and
$\sigma_{s'}-\sigma_s=\mathcal{O}(\Delta s)$, hence
\[
\E\|\rvx_{s'}-\rvx_s\|^2=\mathcal{O}(\Delta s^2),
\]
which yields
\[
\E\|\rvx_{s'}-\hat\rvx_{s'}\|^2=\mathcal{O}(\Delta s^2).
\]

\hfill$\blacksquare$

\subsection{Proposition~\ref{continuous-time-ct}: Continuous-Time Limit of the CT Gradient}

\paragraph{\textit{Proof:}}

We simplify the notation by working with \Cref{eq:cm-discrete} in the form
\begin{align*}
\mathcal{L}_{\mathrm{CM}}^{\Delta s}(\btheta,\btheta^-)
:=
\E \left[\omega(s)\big\|
\rvf_{\btheta}(\rvx_s,s) -
\rvf_{\btheta^-}\big(\bPsi_{s\to s-\Delta s}(\rvx_s),\, s-\Delta s\big)
\big\|_2^2\right].
\end{align*}
For notational simplicity, write
\[
\tilde\rvx_{s-\Delta s}:=\bPsi_{s\to s-\Delta s}(\rvx_s).
\]

Define
\[
\delta \rvf
:= \rvf_{\btheta}(\rvx_s,s)-\rvf_{\btheta^-}(\rvx_s,s).
\]
Then
\begin{align*}
\big\|\rvf_{\btheta}(\rvx_s, s)- \rvf_{\btheta^-}(\tilde\rvx_{s-\Delta s}, s-\Delta s)\big\|_2^2
=
\Big\|\delta \rvf
+
\Big(\rvf_{\btheta^-}(\rvx_s, s)-\rvf_{\btheta^-}(\tilde\rvx_{s-\Delta s}, s-\Delta s)\Big)\Big\|_2^2.
\end{align*}

Applying a Taylor expansion at $(\rvx_s,s)$ gives
\begin{align*}
\rvf_{\btheta^-}(\rvx_s,s) -
\rvf_{\btheta^-}(\tilde\rvx_{s-\Delta s}, s-\Delta s)
=
\frac{\diff}{\diff s}\rvf_{\btheta^-}(\rvx_s,s)\,\Delta s
+ \mathcal{O}(|\Delta s|^2),
\end{align*}
where we use the first-order expansion of the oracle flow map,
\[
\rvx_s - \bPsi_{s\to s-\Delta s}(\rvx_s)
= \rvv^*(\rvx_s,s)\,\Delta s + \mathcal{O}(|\Delta s|^2),
\]
together with the total differentiation identity
\[
\frac{\diff}{\diff s}\rvf_{\btheta^-}(\rvx_s,s)
=
\partial_s \rvf_{\btheta^-}(\rvx_s,s)
+
\big(\partial_{\rvx}\rvf_{\btheta^-}(\rvx_s,s)\big)\,\rvv^*(\rvx_s,s).
\]

Therefore,
\begin{align*}
\big\|\rvf_{\btheta}(\rvx_s, s)- \rvf_{\btheta^-}(\tilde\rvx_{s-\Delta s}, s-\Delta s)\big\|_2^2
=
\|\delta \rvf\|_2^2
+
2\,\delta \rvf^\top \frac{\diff}{\diff s}\rvf_{\btheta^-}(\rvx_s,s)\,\Delta s
+
\mathcal{O}(|\Delta s|^2).
\end{align*}

Now differentiate with respect to $\btheta$. Since $\btheta^-$ is treated as fixed with respect to $\nabla_{\btheta}$ (i.e., no gradient is taken through the stop-gradient target), there is no gradient through $\rvf_{\btheta^-}$. Hence
\begin{align*}
\nabla_{\btheta}\mathcal{L}_{\mathrm{CM}}^{\Delta s}(\btheta,\btheta^-)
&=
\nabla_{\btheta}\E\!\left[\omega(s)\|\delta \rvf\|_2^2\right]\\
&\qquad\qquad
+
2\,\E\!\left[
\omega(s)\,
\nabla_{\btheta}\rvf_{\btheta}(\rvx_s,s)^\top
\frac{\diff}{\diff s}\rvf_{\btheta^-}(\rvx_s,s)
\right]\Delta s
+
\mathcal{O}(|\Delta s|^2).
\end{align*}

At the comparison point where the target network is the stop-gradient copy of the online network, i.e., $\btheta=\btheta^-$, we have
\[
\delta \rvf
=
\rvf_{\btheta}(\rvx_s,s)-\rvf_{\btheta^-}(\rvx_s,s)
=
0,
\]
and therefore
\[
\nabla_{\btheta}\E\!\left[\omega(s)\|\delta \rvf\|_2^2\right]=0.
\]
Thus,
\begin{align*}
\left.
\nabla_{\btheta}\mathcal{L}_{\mathrm{CM}}^{\Delta s}(\btheta,\btheta^-)
\right|_{\btheta=\btheta^-}
&=
\left.
2\,\E\!\left[
\omega(s)\,
\nabla_{\btheta}\rvf_{\btheta}(\rvx_s,s)^\top
\frac{\diff}{\diff s}\rvf_{\btheta^-}(\rvx_s,s)
\right]
\right|_{\btheta=\btheta^-}\Delta s
+
\mathcal{O}(|\Delta s|^2).
\end{align*}

Dividing by $\Delta s$ and taking the limit yields
\begin{align*}
\left.
\lim_{\Delta s\to 0}\frac{1}{\Delta s}\nabla_{\btheta}\mathcal{L}_{\mathrm{CM}}^{\Delta s}(\btheta,\btheta^-)
\right|_{\btheta=\btheta^-}
&=
\left.
\nabla_{\btheta}\E\!\left[
2\,\omega(s)\,
\rvf_{\btheta}^\top(\rvx_s,s)\,
\frac{\diff}{\diff s}\rvf_{\btheta^-}(\rvx_s,s)
\right]
\right|_{\btheta=\btheta^-}.
\end{align*}
This proves the identity at $\btheta=\btheta^-$.

\hfill$\blacksquare$

\newpage
\section{(Optional) Elucidating Diffusion Model (EDM)}\label{sec:edm}

We introduce specific criteria for neural network parameterization design in
the $\rvx$-prediction model, as proposed in Elucidating Diffusion Models
(EDM)~\citep{karras2022elucidating}. EDM provides a simple 
recipe that makes the training process easier to optimize and more reliable.
The $\rvx$-prediction model is expressed as a time-dependent skip connection combined with a
scaled residual (\Cref{eq:D-parametrization}). The central idea is to normalize both the inputs and the regression targets to
unit variance at all times, and to adjust the skip path so that residual errors
are not amplified as time evolves.
 This recipe
has become a widely adopted default in diffusion model implementations and
extends naturally to flow map learning, especially the family of Consistency
Models.

\subsection{Criteria for Neural Network $\rvx_{\bm{\phi}}$ Design}\label{subsec:D-design-edm}

EDM considers a parametrization of the $\rvx$-prediction model, denoted with a slight abuse of notation as $\rvx_{\bm{\phi}}(\rvx, t)$\footnote{As discussed, all four prediction types are equivalent and can be reduced to the $\rvx$-prediction case. EDM adopts this formulation, which is both well-studied and naturally aligned with the goal of generating clean samples of flow map models.}, in the following form: 
\begin{align}\label{eq:D-parametrization}
    \rvx_{\bm{\phi}}(\rvx, t) := c_{\mathrm{skip}}(t) \rvx + c_{\mathrm{out}}(t) \rmF_{\bm{\phi}}\left(c_{\mathrm{in}}(t) \rvx, c_{\mathrm{noise}}(t)\right).
\end{align}
Here, $c_{\mathrm{skip}}(t)$, $c_{\mathrm{out}}(t)$, $c_{\mathrm{in}}(t)$, and $c_{\mathrm{noise}}(t)$ are time-dependent functions.  
They are chosen to enhance stability and performance during training, based on practical considerations that will be introduced shortly.

Plugging this in \Cref{eq:general-dm-loss-D}, then the objective function becomes after straightforward algebraic manipulation\footnote{We start from the $\rvx$-prediction diffusion loss by substituting the parameterization given in \Cref{eq:D-parametrization}:
\begin{align*}
    \begin{aligned}
      &~\mathbb{E}_{\rvx_0, \bm{\epsilon}, t} \left[\omega(t)\norm{ \rvx_{\bm{\phi}}(\rvx_t, t) - \rvx_0}_2^2\right] 
      \\= &~\mathbb{E}_{\rvx_0, \bm{\epsilon}, t} \left[\omega(t)\Big\Vert c_{\mathrm{skip}}(t) \underbrace{(\alpha_t \rvx_0+\sigma_t\beps)}_{\rvx_t} + c_{\mathrm{out}}(t) \rmF_{\bm{\phi}}\left(c_{\mathrm{in}}(t) \rvx_t, c_{\mathrm{noise}}(t)\right) - \rvx_0\Big\Vert_2^2\right]
        \\= &~\mathbb{E}_{\rvx_0, \bm{\epsilon}, t} \left[\omega(t)c_{\mathrm{out}}^2(t)\Big\Vert  \rmF_{\bm{\phi}}\left(c_{\mathrm{in}}(t) \rvx, c_{\mathrm{noise}}(t)\right) - \underbrace{\left(\frac{\left(1-c_{\mathrm{skip}}(t)\alpha_t\right)\rvx_0-c_{\mathrm{skip}}(t)\sigma_t \bm{\epsilon}}{c_{\mathrm{out}}(t)}\right)}_{\rvx_{\mathrm{tgt}}(t)}\Big\Vert_2^2\right]
        \\ = &~\mathrm{\Cref{eq:loss-F}}.
    \end{aligned}
\end{align*}

}:
\begin{align}\label{eq:loss-F}
   \mathbb{E}_{\rvx_0, \bm{\epsilon}, t} \left[\omega(t)c_{\mathrm{out}}^2(t)\norm{ \rmF_{\bm{\phi}}\left(c_{\mathrm{in}}(t)\rvx_t, c_{\mathrm{noise}}(t)\right) - \rvx_{\mathrm{tgt}}(t)}_2^2\right].
\end{align}
Here, the regression target $\rvx_{\mathrm{tgt}}(t)$ is obtained as:
\[
\rvx_{\mathrm{tgt}}(t) = \frac{\left(1-c_{\mathrm{skip}}(t)\alpha_t\right)\rvx_0-c_{\mathrm{skip}}(t)\sigma_t \bm{\epsilon}}{c_{\mathrm{out}}(t)}.
\]

By incorporating the standard deviation of $p_{\text{data}}$, denoted as $\sigma_{\mathrm{d}}$, EDM proposes the following design criterion for network parameterization, which may be described as the \emph{unit variance criterion}.

\paragraph{Unit Variance of Input.} 
\begin{align*}
    &\text{Var}_{\rvx_0,\bm{\epsilon}}\left[ c_{\mathrm{in}}(t)\rvx_t\right] = 1 \nonumber
    \\ \Longleftrightarrow &~c_{\mathrm{in}}^2(t) \text{Var}_{\rvx_0,\bm{\epsilon}}\left[ \alpha_t \rvx_0 +\sigma_t \bm{\epsilon} \right] = 1 \nonumber
    \\ \Longleftrightarrow &~c_{\mathrm{in}}(t)  = \frac{1}{\sqrt{\sigma_{\mathrm{d}}^2 \alpha_t^2 + \sigma_t^2}},
\end{align*}
taking the positive root. 
\paragraph{Unit Variance of Training Target.}
\begin{align}\label{eq:c-out}
    &\text{Var}_{\rvx_0,\bm{\epsilon}}\left[ \rvx_{\mathrm{tgt}}(t) \right] = 1 \nonumber
    \\ \Longleftrightarrow &~c_{\mathrm{out}}^2(t)  = \left(1-c_{\mathrm{skip}}(t)\alpha_t\right)^2 \sigma_{\mathrm{d}}^2 + c_{\mathrm{skip}}^2(t)\sigma_t^2, 
\end{align}
with centered data ($\E[\rvx_0]=\bm{0}$).
\paragraph{Minimize the Error Amplification from $\rmF_{\bm{\phi}}$ to $\rvx_{\bm{\phi}}$.} EDM aims to mitigate the amplification of the network's learning error from $\rmF_{\bm{\phi}}$ to $\rvx_{\bm{\phi}}$. This is achieved by selecting $c_{\mathrm{skip}}$ to minimize $c_{\mathrm{out}}$:
\[
c_{\mathrm{skip}}^* \in \argmin_{c_{\mathrm{skip}}}  c_{\mathrm{out}}^2.
\]
Using the standard approach of solving $\frac{\partial c_{\mathrm{out}}}{\partial c_{\mathrm{skip}}} = 0$ for the critical point $c_{\mathrm{skip}}^*$, we obtain
\begin{align*}
    c_{\mathrm{skip}}^*(t) = \frac{\alpha_t \sigma_{\mathrm{d}}^2 }{\alpha_t^2 \sigma_{\mathrm{d}}^2  + \sigma_t^2}. 
\end{align*}
Substituting this into \Cref{eq:c-out}, the optimal value is given by
\begin{align*}
    c_{\mathrm{out}}^*(t) = \pm \frac{\sigma_t \sigma_{\mathrm{d}} }{\sqrt{\alpha_t^2 \sigma_{\mathrm{d}}^2  + \sigma_t^2}}. 
\end{align*}
By convention, we use the nonnegative branch for the output scale:
\[
c_{\mathrm{out}}^*(t) = \frac{\sigma_t \sigma_{\text d}}
{\sqrt{\alpha_t^2 \sigma_{\text d}^2 + \sigma_t^2}}
\;\;\;(\ge 0),
\]
which ensures $c_{\mathrm{out}}(0)=0$ and $c_{\mathrm{out}}(t)\to\sigma_{\text d}$ as $\sigma_t$ is sufficiently large,
yielding the intuitive limits $\rvx_{\bm{\phi}}(\rvx_t,0)\approx \rvx_t$ and
$\rvx_{\bm{\phi}}(\rvx_t,t)\approx \sigma_{\text d}\,\rmF_{\bm{\phi}}(\cdot)$.

We summarize these coefficients as follows:
\begin{mdframed}
Denoting $R_t := \alpha_t^2 \sigma_{\mathrm{d}}^2+ \sigma_t^2$, we have the following selections:
    \begin{align}\label{eq:edm-coeff}
        \begin{aligned}
            c_{\mathrm{in}}(t)  = \frac{1}{\sqrt{R_t}}, \quad
            c_{\mathrm{skip}}(t) = \frac{\alpha_t \sigma_{\mathrm{d}}^2 }{R_t}, \quad
            c_{\mathrm{out}}(t) =  \frac{\sigma_t \sigma_{\mathrm{d}} }{\sqrt{R_t}}.
        \end{aligned}
    \end{align}
\end{mdframed}

With \Cref{eq:edm-coeff}, the regression target $\rvx_{\mathrm{tgt}}(t)$ simplifies to
\[
\rvx_{\mathrm{tgt}}(t) = \frac{1}{\sigma_{\mathrm{d}}}\frac{\sigma_t \rvx_0 - \alpha_t \sigma_{\mathrm{d}}^2\bm{\epsilon} }{\sqrt{R_t}}.
\] 
Additionally, substituting these expressions into \Cref{eq:D-parametrization} and \Cref{eq:loss-F} allows us to simplify both the parametrization and the loss function, yielding:
\begin{align*}
    \rvx_{\bm{\phi}}(\rvx, t)=\frac{\alpha_t \sigma_{\mathrm{d}}^2 }{R_t} \rvx + \frac{\sigma_t \sigma_{\mathrm{d}} }{\sqrt{R_t}}\rmF_{\bm{\phi}}\left(\frac{1}{\sqrt{R_t}}\rvx, c_{\mathrm{noise}}(t)\right),
\end{align*}
and
\begin{align*}
   \mathbb{E}_{\rvx_0, \bm{\epsilon}, t} \left[\omega(t)\frac{\sigma_t^2  }{R_t}\norm{ \sigma_{\mathrm{d}} \rmF_{\bm{\phi}}\left(\frac{1}{\sqrt{R_t}}\rvx_t, c_{\mathrm{noise}}(t)\right) - \left(\frac{\sigma_t \rvx_0 - \alpha_t \sigma_{\mathrm{d}}^2\bm{\epsilon} }{\sqrt{R_t}}\right)}_2^2\right].
\end{align*}

With this parameterization and the conditions $\alpha_0 \approx 1$ and $\sigma_0 \approx 0$, we observe that  
\[
c_{\mathrm{skip}}(0) \approx 1 \quad \text{and} \quad c_{\mathrm{out}}(0) \approx 0.
\]

\paragraph{Selection of $c_{\mathrm{noise}}(t)$.}
It provides a noise-level embedding to $\rmF_{\bm{\phi}}$; any
one-to-one mapping of the noise level $\sigma_t$ (e.g., $c_{\mathrm{noise}}(t)=\log \sigma_t$) is suitable.

\subsection{A Common EDM Special Case: $\alpha_t = 1$, $\sigma_t = t$}

We consider the simplified noise schedule used in EDM, where
$\alpha_t = 1$ and $\sigma_t = t$, which also appears in the discussion of
CTM in \Cref{sec:CTM}. Under this setting, the forward process becomes
\[
\rvx_t = \rvx_0 + t\bm{\epsilon},
\quad
\text{with }
\rvx_0 \sim p_{\text{data}},
\,\,
\bm{\epsilon} \sim \mathcal{N}(\bm{0}, \rmI),
\]
corresponding to the perturbation kernel
\[
p_t(\rvx_t|\rvx_0)
=
\mathcal{N}(\rvx_t; \rvx_0, t^2\rmI).
\]

The resulting marginal density is
\[
p_t(\rvx)
=
\int
\mathcal{N}(\rvx;\rvx_0,t^2\rmI)
p_{\mathrm{data}}(\rvx_0)
\diff\rvx_0
=
\bigl[
p_{\mathrm{data}}
*
\mathcal{N}(\bm{0},t^2\rmI)
\bigr](\rvx).
\]
Thus, the exact terminal marginal is
$p_T=p_{\mathrm{data}}*\mathcal{N}(\bm{0},T^2\rmI)$.
When $T$ is sufficiently large relative to the scale of the data, this
marginal is commonly approximated by
$\mathcal{N}(\bm{0},T^2\rmI)$, which is used as the sampling prior:
\[
p_{\mathrm{prior}}(\rvx_T)
:=
\mathcal{N}(\rvx_T;\bm{0},T^2\rmI).
\]

Under this setup, the PF-ODE based on $\rvx$-prediction $\rvx_{\bm{\phi}^\times}$ (see \Cref{eq:clean_pf_ode}) simplifies to
\begin{align*}
    \frac{\diff \rvx(t)}{\diff t} = \frac{\rvx(t) - \rvx_{\bm{\phi}^\times}(\rvx(t), t)}{t}.
\end{align*}

Substituting this formulation into \Cref{eq:edm-coeff}, the neural network parameterization coefficients become
\begin{align}\label{eq:edm-simple-coefficients}
    c_{\mathrm{in}}(t) = \frac{1}{\sqrt{\sigma_{\mathrm{d}}^2 + t^2}}, \quad
    c_{\mathrm{skip}}(t) = \frac{\sigma_{\mathrm{d}}^2}{\sigma_{\mathrm{d}}^2 + t^2}, \quad
    c_{\mathrm{out}}(t) = \pm \frac{t \sigma_{\mathrm{d}}}{\sqrt{\sigma_{\mathrm{d}}^2 + t^2}}.
\end{align}
From these expressions, we observe:
\begin{itemize}
    \item When $t \approx 0$, the noise level is negligible, so
    $c_{\mathrm{skip}} \approx 1$ and $c_{\mathrm{out}} \approx 0$. In this limit,
    the skip path dominates and the network essentially passes through its
    input,
    \[
    \rvx_{\bm{\phi}}(\rvx, t) \approx \rvx.
    \]
    \item When $t \gg 0$, the input is heavily corrupted by noise, so
    $c_{\mathrm{skip}} \approx 0$ and $c_{\mathrm{out}} \approx \sigma_{\mathrm{d}}$.
    In this regime, the skip path vanishes and the model output is determined
    entirely by the learned residual,
    \[
    \rvx_{\bm{\phi}}(\rvx, t) \approx \sigma_{\mathrm{d}} \rmF_{\bm{\phi}}\left(c_{\mathrm{in}}(t)\rvx, c_{\mathrm{noise}}(t)\right),
    \]
    meaning that the network $\rmF_{\bm{\phi}}$ predicts a scaled proxy of the clean
signal from a normalized noisy input; at high noise levels the model output is
therefore determined entirely by the learned denoising function.
\end{itemize}
In short, the parameterization smoothly interpolates from an identity map at small $t$
to a scaled residual predictor on standardized inputs at large $t$.

\backmatter  

\printbibliography

\end{document}